# Artificial Intelligence Index Report 2025

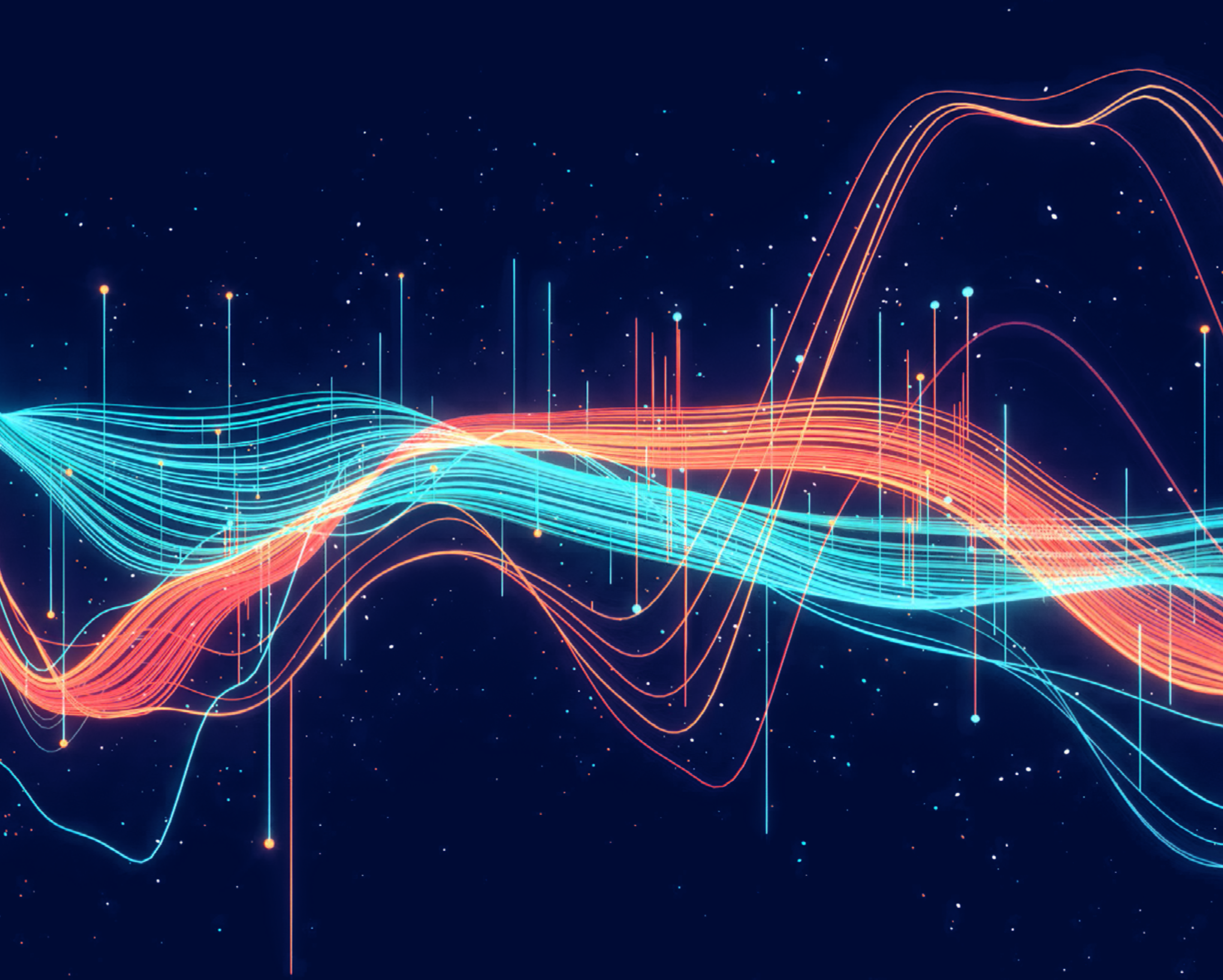

**Stanford University**
Human-Centered
Artificial Intelligence

# Introduction to the AI Index Report 2025

Welcome to the eighth edition of the AI Index report. The 2025 Index is our most comprehensive to date and arrives at an important moment, as AI's influence across society, the economy, and global governance continues to intensify. New in this year's report are in-depth analyses of the evolving landscape of AI hardware, novel estimates of inference costs, and new analyses of AI publication and patenting trends. We also introduce fresh data on corporate adoption of responsible AI practices, along with expanded coverage of AI's growing role in science and medicine.

Since its founding in 2017 as an offshoot of the One Hundred Year Study of Artificial Intelligence, the AI Index has been committed to equipping policymakers, journalists, executives, researchers, and the public with accurate, rigorously validated, and globally sourced data. Our mission has always been to help these stakeholders make better-informed decisions about the development and deployment of AI. In a world where AI is discussed everywhere—from boardrooms to kitchen tables—this mission has never been more essential.

The AI Index continues to lead in tracking and interpreting the most critical trends shaping the field—from the shifting geopolitical landscape and the rapid evolution of underlying technologies, to AI's expanding role in business, policymaking, and public life. Longitudinal tracking remains at the heart of our mission. In a domain advancing at breakneck speed, the Index provides essential context—helping us understand where AI stands today, how it got here, and where it may be headed next.

Recognized globally as one of the most authoritative resources on artificial intelligence, the AI Index has been cited in major media outlets such as The New York Times, Bloomberg, and The Guardian; referenced in hundreds of academic papers; and used by policymakers and government agencies around the world. We have briefed companies like Accenture, IBM, Wells Fargo, and Fidelity on the state of AI, and we continue to serve as an independent source of insights for the global AI ecosystem.

# Message From the Co-directors

As AI continues to reshape our lives, the corporate world, and public discourse, the AI Index continues to track its progress—offering an independent, data-driven perspective on AI's development, adoption, and impact, across time and geography.

What a year 2024 has been for AI. The recognition of AI's role in advancing humanity's knowledge is reflected in Nobel prizes in physics and chemistry, and the Turing award for foundational work in reinforcement learning. The once-formidable Turing Test is no longer considered an ambitious goal, having been surpassed by today's sophisticated systems. Meanwhile, AI adoption has accelerated at an unprecedented rate, as millions of people are now using AI on a regular basis both for their professional work and leisure activities. As high-performing, low-cost, and openly available models proliferate, AI's accessibility and impact are set to expand even further.

After a brief slowdown, corporate investment in AI rebounded. The number of newly funded generative AI startups nearly tripled, and after years of sluggish uptake, business adoption accelerated significantly in 2024. AI has moved from the margins to become a central driver of business value.

Governments, too, are ramping up their involvement. Policymakers are no longer just debating AI—they're investing in it. Several countries launched billion-dollar national AI infrastructure initiatives, including major efforts to expand energy capacity to support AI development. Global coordination is increasing, even as local initiatives take shape.

Yet trust remains a major challenge. Fewer people believe AI companies will safeguard their data, and concerns about fairness and bias persist. Misinformation continues to pose risks, particularly in elections and the proliferation of deepfakes. In response, governments are advancing new regulatory frameworks aimed at promoting transparency, accountability, and fairness. Public attitudes are also shifting. While skepticism remains, a global survey in 2024 showed a notable rise in optimism about AI's potential to deliver broad societal benefits.

AI is no longer just a story of what's possible—it's a story of what's happening now and how we are collectively shaping the future of humanity. Explore this year's AI Index report and see for yourself.

**Yolanda Gil and Raymond Perrault**
Co-directors, AI Index Report

# Top Takeaways

**1. AI performance on demanding benchmarks continues to improve.** In 2023, researchers introduced new benchmarks—MMMU, GPQA, and SWE-bench—to test the limits of advanced AI systems. Just a year later, performance sharply increased: scores rose by 18.8, 48.9, and 67.3 percentage points on MMMU, GPQA, and SWE-bench, respectively. Beyond benchmarks, AI systems made major strides in generating high-quality video, and in some settings, language model agents even outperformed humans in programming tasks with limited time budgets.

---

**2. AI is increasingly embedded in everyday life.** From healthcare to transportation, AI is rapidly moving from the lab to daily life. In 2023, the FDA approved 223 AI-enabled medical devices, up from just six in 2015. On the roads, self-driving cars are no longer experimental: Waymo, one of the largest U.S. operators, provides over 150,000 autonomous rides each week, while Baidu's affordable Apollo Go robotaxi fleet now serves numerous cities across China.

---

**3. Business is all in on AI, fueling record investment and usage, as research continues to show strong productivity impacts.** In 2024, U.S. private AI investment grew to $109.1 billion—nearly 12 times China's $9.3 billion and 24 times the U.K.'s $4.5 billion. Generative AI saw particularly strong momentum, attracting $33.9 billion globally in private investment—an 18.7% increase from 2023. AI business usage is also accelerating: 78% of organizations reported using AI in 2024, up from 55% the year before. Meanwhile, a growing body of research confirms that AI boosts productivity and, in most cases, helps narrow skill gaps across the workforce.

---

**4. The U.S. still leads in producing top AI models—but China is closing the performance gap.** In 2024, U.S.-based institutions produced 40 notable AI models, compared to China's 15 and Europe's three. While the U.S. maintains its lead in quantity, Chinese models have rapidly closed the quality gap: performance differences on major benchmarks such as MMLU and HumanEval shrank from double digits in 2023 to near parity in 2024. China continues to lead in AI publications and patents. Model development is increasingly global, with notable launches from the Middle East, Latin America, and Southeast Asia.

---

**5. The responsible AI ecosystem evolves—unevenly.** AI-related incidents are rising sharply, yet standardized RAI evaluations remain rare among major industrial model developers. However, new benchmarks like HELM Safety, AIR-Bench, and FACTS offer promising tools for assessing factuality and safety. Among companies, a gap persists between recognizing RAI risks and taking meaningful action. In contrast, governments are showing increased urgency: In 2024, global cooperation on AI governance intensified, with organizations including the OECD, EU, U.N., and African Union releasing frameworks focused on transparency, trustworthiness, and other core responsible AI principles.

---

# Top Takeaways (cont'd)

**6. Global AI optimism is rising—but deep regional divides remain.** In countries like China (83%), Indonesia (80%), and Thailand (77%), strong majorities see AI products and services as more beneficial than harmful. In contrast, optimism remains far lower in places like Canada (40%), the United States (39%), and the Netherlands (36%). Still, sentiment is shifting: Since 2022, optimism has grown significantly in several previously skeptical countries, including Germany (+10%), France (+10%), Canada (+8%), Great Britain (+8%), and the United States (+4%).

---

**7. AI becomes more efficient, affordable, and accessible.** Driven by increasingly capable small models, the inference cost for a system performing at the level of GPT-3.5 dropped over 280-fold between November 2022 and October 2024. At the hardware level, costs have declined by 30% annually, while energy efficiency has improved by 40% each year. Open-weight models are closing the gap with closed models, reducing the performance difference from 8% to just 1.7% on some benchmarks in a single year. Together, these trends are rapidly lowering the barriers to advanced AI.

---

**8. Governments are stepping up on AI—with regulation and investment.** In 2024, U.S. federal agencies introduced 59 AI-related regulations—more than double the number in 2023—and issued by twice as many agencies. Globally, legislative mentions of AI rose 21.3% across 75 countries since 2023, marking a ninefold increase since 2016. Alongside growing attention, governments are investing at scale: Canada pledged $2.4 billion, China launched a $47.5 billion semiconductor fund, France committed €109 billion, India pledged $1.25 billion, and Saudi Arabia's Project Transcendence represents a $100 billion initiative.

---

**9. AI and computer science education is expanding—but gaps in access and readiness persist.** Two-thirds of countries now offer or plan to offer K–12 CS education—twice as many as in 2019—with Africa and Latin America making the most progress. In the U.S., the number of graduates with bachelor's degrees in computing has increased 22% over the last 10 years. Yet access remains limited in many African countries due to basic infrastructure gaps like electricity. In the U.S., 81% of K–12 CS teachers say AI should be part of foundational CS education, but less than half feel equipped to teach it.

---

**10. Industry is racing ahead in AI—but the frontier is tightening.** Nearly 90% of notable AI models in 2024 came from industry, up from 60% in 2023, while academia remains the top source of highly cited research. Model scale continues to grow rapidly—training compute doubles every five months, datasets every eight, and power use annually. Yet performance gaps are shrinking: the Elo skill score difference between the top and 10th-ranked models fell from 11.9% to 5.4% in a year, and the top two are now separated by just 0.7%. The frontier is increasingly competitive—and increasingly crowded.

---

# Top Takeaways (cont'd)

**11. AI earns top honors for its impact on science.** AI's growing importance is reflected in major scientific awards: Two Nobel Prizes recognized work that led to deep learning (physics) and to its application to protein folding (chemistry), while the Turing Award honored groundbreaking contributions to reinforcement learning.

---

**12. Complex reasoning remains a challenge.** AI models excel at tasks like International Mathematical Olympiad problems but still struggle with complex reasoning benchmarks like PlanBench. They often fail to reliably solve logic tasks even when provably correct solutions exist, limiting their effectiveness in high-stakes settings where precision is critical.

# Steering Committee

**Chair**
Raymond Perrault
SRI International

**Chair-elect**
Yolanda Gil
University of Southern California, Information Sciences Institute

**Members**

Erik Brynjolfsson
Stanford University

Jack Clark
Anthropic, OECD

John Etchemendy
Stanford University

Katrina Ligett
Hebrew University

Terah Lyons
JPMorgan Chase & Co.

James Manyika
Google, University of Oxford

Juan Carlos Niebles
Stanford University, Salesforce

Vanessa Parli
Stanford University

Yoav Shoham
Stanford University, AI21 Labs

Russell Wald
Stanford University

Toby Walsh
UNSW Sydney

# Staff and Researchers

**Research Manager and Editor-in-Chief**
Nestor Maslej, Stanford University

**Research Associate**
Loredana Fattorini, Stanford University

**Affiliated Researchers**
Elif Kiesow Cortez, Stanford Law School Research Fellow
Julia Betts Lotufo, Researcher
Anka Reuel, Stanford University
Alexandra Rome, Researcher
Angelo Salatino, Knowledge Media Institute, The Open University
Lapo Santarlasci, IMT School for Advanced Studies Lucca

**Graduate Researchers**
Emily Capstick, Stanford University
Malou van Draanen Glismann, Stanford University
Njenga Kariuki, Stanford University

**Undergraduate Researchers**
Armin Hamrah, Claremont McKenna College
Sukrut Oak, Stanford University
Ngorli Fiifi Paintsil, Stanford University
Andrew Shi, Stanford University

# How to Cite This Report

Nestor Maslej, Loredana Fattorini, Raymond Perrault, Yolanda Gil, Vanessa Parli, Njenga Kariuki, Emily Capstick, Anka Reuel, Erik Brynjolfsson, John Etchemendy, Katrina Ligett, Terah Lyons, James Manyika, Juan Carlos Niebles, Yoav Shoham, Russell Wald, Toby Walsh, Armin Hamrah, Lapo Santarlasci, Julia Betts Lotufo, Alexandra Rome, Andrew Shi, Sukrut Oak. "The AI Index 2025 Annual Report," AI Index Steering Committee, Institute for Human-Centered AI, Stanford University, Stanford, CA, April 2025. https://doi.org/10.48550/arXiv.2504.07139

# Public Data and Tools

The AI Index 2025 Report is supplemented by raw data and an interactive tool. We invite each reader to use the data and the tool in a way most relevant to their work and interests.

- Raw data and charts: The public data and high-resolution images of all the charts in the report are available on Google Drive.
- Global AI Vibrancy Tool: Compare the AI ecosystems of over 30 countries. The Global AI Vibrancy tool will be updated in the summer of 2025.

# AI Index and Stanford HAI

The AI Index is an independent initiative at the Stanford Institute for Human-Centered Artificial Intelligence (HAI).

Artificial Intelligence Index

Stanford University Human-Centered Artificial Intelligence

The AI Index was conceived within the One Hundred Year Study on Artificial Intelligence (AI100).

The AI Index welcomes feedback and new ideas for next year. Contact us at nmaslej@stanford.edu.

The AI Index acknowledges that while authored by a team of human researchers, its writing process was aided by AI tools. Specifically, the authors used ChatGPT and Claude to help tighten and copy edit initial drafts. The workflow involved authors writing the original copy and utilizing AI tools as part of the editing process.

## Supporting Partners

## Analytics and Research Partners

# Contributors

The AI Index would like to acknowledge the following individuals by chapter and section for their contributions of data, analysis, advice, and expert commentary included in the AI Index Report 2025:

**Introduction**

Loredana Fattorini, Yolanda Gil, Nestor Maslej, Vanessa Parli, Ray Perrault

**Chapter 1: Research and Development**

Nancy Amato, Andrea Brown, Ben Cottier, Lucía Ronchi Darré, Virginia Dignum, Meredith Ellison, Robin Evans, Loredana Fattorini, Yolanda Gil, Armin Hamrah, Katrina Ligett, Nestor Maslej, Maurice Pagnucco, Ngorli Fiifi Paintsil, Vanessa Parli, Ray Perrault, Robi Rahman, Christine Raval, Vesna Sabljakovic-Fritz, Angelo Salatino, Lapo Santarlasci, Andrew Shi, Nathan Sturtevant, Daniel Weld, Kevin Xu, Meg Young

**Chapter 2: Technical Performance**

Rishi Bommasani, Erik Brynjolfsson, Loredana Fattorini, Tobi Gertsenberg, Yolanda Gil, Noah Goodman, Nicholas Haber, Armin Hamrah, Sanmi Koyejo, Percy Liang, Katrina Ligett, Nestor Maslej, Juan Carlos Niebles, Sukrut Oak, Vanessa Parli, Marco Pavone, Ray Perrault, Anka Reuel, Andrew Shi, Yoav Shoham, Toby Walsh

**Chapter 3: Responsible AI**

Medha Bankhwal, Emily Capstick, Dmytro Chumachenko, Patrick Connolly, Natalia Dorogi, Loredana Fattorini, Ann Fitz-Gerald, Yolanda Gil, Armin Hamrah, Ariel Lee, Katrina Ligett, Shayne Longpre, Natasha Maniar, Nestor Maslej, Katherine Ottenbreit, Halyna Padalko, Vanessa Parli, Ray Perrault, Brittany Presten, Anka Reuel, Roger Roberts, Andrew Shi, Georgio Stoev, Shekhar Tewari, Dikshita Venkatesh, Cayla Volandes, Jakub Wiatrak

**Chapter 4: Economy**

Medha Bankhwal, Erik Brynjolfsson, Mar Carpanelli, Cara Christopher, Michael Chui, Natalia Dorogi, Heather English, Murat Erer, Loredana Fattorini, Yolanda Gil, Heather Hanselman, Rosie Hood, Vishy Kamalapuram, Kory Kantenga, Njenga Kariuki, Akash Kaura, Elena Magrini, Nestor Maslej, Katherine Ottenbreit, Vanessa Parli, Ray Perrault, Brittany Presten, Roger Roberts, Cayla Volandes, Casey Weston, Hansen Yang

**Chapter 5: Science and Medicine**

Russ Altman, Kameron Black, Jonathan Chen, Jean-Benoit Delbrouck, Joshua Edrich, Loredana Fattorini, Alejandro Lozano, Yolanda Gil, Ethan Goh, Armin Hamrah, Fateme Nateghi Haredasht, Tina Hernandez-Boussard, Yeon Mi Hwang, Rohan Koodli, Arman Koul, Curt Langlotz, Ashley Lewis, Chase Ludwig, Stephen P. Ma, Abdoul Jalil Djiberou Mahamadou, David Magnus, James Manyika, Nestor Maslej, Gowri Nayar, Madelena Ng, Sophie Ostmeier, Vanessa Parli, Ray Perrault, Malkiva Pillai, Ossian Karl-Johan Ferdinand Rabow, Sean Riordan, Brennan Geti Simon, Kotoha Togami, Artem Trotsyuk, Maya Varma, Quinn Waeiss, Betty Xiong

**Chapter 6: Policy**

Elif Kiesow Cortez, Loredana Fattorini, Yolanda Gil, Julia Betts Lotufo, Vanessa Parli, Ray Perrault, Alexandra Rome, Lapo Santarlasci, Georgio Stoev, Russell Wald, Daniel Zhang

# Contributors (cont'd)

**Chapter 7: Education**

John Etchemendy, Loredana Fattorini, Lili Gangas, Yolanda Gil, Rachel Goins, Laura Hinton, Sonia Koshy, Kirsten Lundgren, Nestor Maslej, Lisa Cruz Novohatski, Vanessa Parli, Ray Perrault, Allison Scott, Andreen Soley, Bryan Twarek, Laurens Vehmeijer

**Chapter 8: Public Opinion**

Emily Capstick, John Etchemendy, Loredana Fattorini, Yolanda Gil, Njenga Kariuki, Nestor Maslej, Vanessa Parli, Ray Perrault

---

The AI Index would like to acknowledge the following individuals by chapter and section for their contributions of data, analysis, advice, and expert commentary included in the AI Index Report 2025:

**Organizations**

**Accenture**
Arnab Chakraborty, Patrick Connolly, Shekhar Tewari, Dikshita Venkatesh, Jakub Wiatrak

**Epoch AI**
Ben Cottier, Robi Rahman

**GitHub**
Lucía Ronchi Darré, Kevin Xu

**Lightcast**
Cara Christopher, Elena Magrini

**LinkedIn**
Mar Carpanelli, Akash Kaura, Kory Kantenga, Rosie Hood, Casey Weston

**McKinsey & Company**
Medha Bankhwal, Natalia Dorogi, Natasha Maniar, Katherine Ottenbreit, Brittany Presten, Roger Roberts, Cayla Volandes

**Quid**
Heather English, Hansen Yang

The AI Index also thanks Jeanina Matias, Nancy King, Carolyn Lehman, Shana Lynch, Jonathan Mindes, and Michi Turner for their help in preparing this report; Christopher Ellis for his help in maintaining the AI Index website; and Annie Benisch, Stacey Sickels Boyce, Marc Gough, Caroline Meinhardt, Drew Spence, Casey Weston, Madeleine Wright, and Daniel Zhang for their work in helping promote the report.

# Table of Contents

ACCESS THE PUBLIC DATA

# Report Highlights

## CHAPTER 1:
## Research and Development

**1. Industry continues to make significant investments in AI and leads in notable AI model development, while academia leads in highly cited research.** Industry's lead in notable model development, highlighted in the two previous AI Index reports, has only grown more pronounced, with nearly 90% of notable models in 2024 (compared to 60% in 2023) originating from industry. Academia has remained the single leading institutional producer of highly cited (top 100) publications over the past three years.

---

**2. China leads in AI research publication totals, while the United States leads in highly influential research.** In 2023, China produced more AI publications (23.2%) and citations (22.6%) than any other country. Over the past three years, U.S. institutions have contributed the most top-100-cited AI publications.

---

**3. AI publication totals continue to grow and increasingly dominate computer science.** Between 2013 and 2023, the total number of AI publications in venues related to computer science and other scientific disciplines nearly tripled, increasing from approximately 102,000 to over 242,000. Proportionally, AI's share of computer science publications has risen from 21.6% in 2013 to 41.8% in 2023.

---

**4. The United States continues to be the leading source of notable AI models.** In 2024, U.S.-based institutions produced 40 notable AI models, significantly surpassing China's 15 and Europe's combined total of three. In the past decade, more notable machine learning models have originated from the United States than any other country.

---

**5. AI models get increasingly bigger, more computationally demanding, and more energy intensive.** New research finds that the training compute for notable AI models doubles approximately every five months, dataset sizes for training LLMs every eight months, and the power required for training annually. Large-scale industry investment continues to drive model scaling and performance gains.

---

**6. AI models become increasingly cheaper to use.** The cost of querying an AI model that scores the equivalent of GPT-3.5 (64.8) on MMLU, a popular benchmark for assessing language model performance, dropped from $20.00 per million tokens in November 2022 to just $0.07 per million tokens by October 2024 (Gemini-1.5-Flash-8B)—a more than 280-fold reduction in approximately 18 months. Depending on the task, LLM inference prices have fallen anywhere from 9 to 900 times per year.

---

# Report Highlights

## CHAPTER 1:
Research and Development (cont'd)

**7. AI patenting is on the rise.** Between 2010 and 2023, the number of AI patents has grown steadily and significantly, ballooning from 3,833 to 122,511. In just the last year, the number of AI patents has risen 29.6%. As of 2023, China leads in total AI patents, accounting for 69.7% of all grants, while South Korea and Luxembourg stand out as top AI patent producers on a per capita basis.

---

**8. AI hardware gets faster, cheaper, and more energy efficient.** New research suggests that machine learning hardware performance, measured in 16-bit floating-point operations, has grown 43% annually, doubling every 1.9 years. Price performance has improved, with costs dropping 30% per year, while energy efficiency has increased by 40% annually.

---

**9. Carbon emissions from AI training are steadily increasing.** Training early AI models, such as AlexNet (2012), had modest amounts of carbon emissions at 0.01 tons. More recent models have significantly higher emissions for training: GPT-3 (2020) at 588 tons, GPT-4 (2023) at 5,184 tons, and Llama 3.1 405B (2024) at 8,930 tons. For perspective, the average American emits 18 tons of carbon per year.

## CHAPTER 2:
Technical Performance

**1. AI masters new benchmarks faster than ever.** In 2023, AI researchers introduced several challenging new benchmarks, including MMMU, GPQA, and SWE-bench, aimed at testing the limits of increasingly capable AI systems. By 2024, AI performance on these benchmarks saw remarkable improvements, with gains of 18.8 and 48.9 percentage points on MMMU and GPQA, respectively. On SWE-bench, AI systems could solve just 4.4% of coding problems in 2023—a figure that jumped to 71.7% in 2024.

---

**2. Open-weight models catch up.** Last year's AI Index revealed that leading open-weight models lagged significantly behind their closed-weight counterparts. By 2024, this gap had nearly disappeared. In early January 2024, the leading closed-weight model outperformed the top open-weight model by 8.0% on the Chatbot Arena Leaderboard. By February 2025, this gap had narrowed to 1.7%.

---

# Report Highlights

## CHAPTER 2:
Technical Performance (cont'd)

**3. The gap closes between Chinese and U.S. models.** In 2023, leading American models significantly outperformed their Chinese counterparts—a trend that no longer holds. At the end of 2023, performance gaps on benchmarks such as MMLU, MMMU, MATH, and HumanEval were 17.5, 13.5, 24.3, and 31.6 percentage points, respectively. By the end of 2024, these margins had narrowed substantially to 0.3, 8.1, 1.6, and 3.7 percentage points.

**4. AI model performance converges at the frontier.** According to last year's AI Index, the Elo score difference between the top and 10th-ranked model on the Chatbot Arena Leaderboard was 11.9%. By early 2025, this gap had narrowed to 5.4%. Likewise, the difference between the top two models shrank from 4.9% in 2023 to just 0.7% in 2024. The AI landscape is becoming increasingly competitive, with high-quality models now available from a growing number of developers.

**5. New reasoning paradigms like test-time compute improve model performance.** In 2024, OpenAI introduced models like o1 and o3 that are designed to iteratively reason through their outputs. This test-time compute approach dramatically improved performance, with o1 scoring 74.4% on an International Mathematical Olympiad qualifying exam, compared to GPT-4o's 9.3%. However, this enhanced reasoning comes at a cost: o1 is nearly six times more expensive and 30 times slower than GPT-4o.

**6. More challenging benchmarks are continually being proposed.** The saturation of traditional AI benchmarks like MMLU, GSM8K, and HumanEval, coupled with improved performance on newer, more challenging benchmarks such as MMMU and GPQA, has pushed researchers to explore additional evaluation methods for leading AI systems. Notable among these are Humanity's Last Exam, a rigorous academic test where the top system scores just 8.80%; FrontierMath, a complex mathematics benchmark where AI systems solve only 2% of problems; and BigCodeBench, a coding benchmark where AI systems achieve a 35.5% success rate—well below the human standard of 97%.

**7. High-quality AI video generators demonstrate significant improvement.** In 2024, several advanced AI models capable of generating high-quality videos from text inputs were launched. Notable releases include OpenAI's SORA, Stable Video Diffusion 3D and 4D, Meta's Movie Gen, and Google DeepMind's Veo 2. These models produce videos of significantly higher quality compared to those from 2023.

# Report Highlights

## CHAPTER 2:
Technical Performance (cont'd)

**8. Smaller models drive stronger performance.** In 2022, the smallest model registering a score higher than 60% on MMLU was PaLM, with 540 billion parameters. By 2024, Microsoft's Phi-3-mini, with just 3.8 billion parameters, achieved the same threshold—the equivalent of a 142-fold reduction in two years.

**9. Complex reasoning remains a problem.** Even though the addition of mechanisms such as chain-of-thought reasoning has significantly improved the performance of LLMs, these systems still cannot reliably solve problems for which provably correct solutions can be found using logical reasoning, such as arithmetic and planning, especially on instances larger than those they were trained on. This has a significant impact on the trustworthiness of these systems and their suitability in high-risk applications.

**10. AI agents show early promise.** The launch of RE-Bench in 2024 introduced a rigorous benchmark for evaluating complex tasks for AI agents. In short time-horizon settings (two-hour budget), top AI systems score four times higher than human experts, but as the time budget increases, human performance surpasses AI—outscoring it two to one at 32 hours. AI agents already match human expertise in select tasks, such as writing Triton kernels, while delivering results faster and at lower costs.

## CHAPTER 3:
Responsible AI

**1. Evaluating AI systems with responsible AI (RAI) criteria is still uncommon, but new benchmarks are beginning to emerge.** Last year's AI Index highlighted the lack of standardized RAI benchmarks for LLMs. While this issue persists, new benchmarks such as HELM Safety and AIR-Bench help to fill this gap.

**2. The number of AI incident reports continues to increase.** According to the AI Incidents Database, the number of reported AI-related incidents rose to 233 in 2024—a record high and a 56.4% increase over 2023.

# Report Highlights

## CHAPTER 3:
Responsible AI (cont'd)

**3. Organizations acknowledge RAI risks, but mitigation efforts lag.** A McKinsey survey on organizations' RAI engagement shows that while many identify key RAI risks, not all are taking active steps to address them. Risks including inaccuracy, regulatory compliance, and cybersecurity were top of mind for leaders with only 64%, 63%, and 60% of respondents, respectively, citing them as concerns.

---

**4. Across the globe, policymakers demonstrate a significant interest in RAI.** In 2024, global cooperation on AI governance intensified, with a focus on articulating agreed-upon principles for responsible AI. Several major organizations—including the OECD, European Union, United Nations, and African Union—published frameworks to articulate key RAI concerns such as transparency and explainability, and trustworthiness.

---

**5. The data commons is rapidly shrinking.** AI models rely on massive amounts of publicly available web data for training. A recent study found that data use restrictions increased significantly from 2023 to 2024, as many websites implemented new protocols to curb data scraping for AI training. In actively maintained domains in the C4 common crawl dataset, the proportion of restricted tokens jumped from 5–7% to 20–33%. This decline has consequences for data diversity, model alignment, and scalability, and may also lead to new approaches to learning with data constraints.

---

**6. Foundation model research transparency improves, yet more work remains.** The updated Foundation Model Transparency Index—a project tracking transparency in the foundation model ecosystem—revealed that the average transparency score among major model developers increased from 37% in October 2023 to 58% in May 2024. While these gains are promising, there is still considerable room for improvement.

---

**7. Better benchmarks for factuality and truthfulness.** Earlier benchmarks like HaluEval and TruthfulQA, aimed at evaluating the factuality and truthfulness of AI models, have failed to gain widespread adoption within the AI community. In response, newer and more comprehensive evaluations have emerged, such as the updated Hughes Hallucination Evaluation Model leaderboard, FACTS, and SimpleQA.

---

**8. AI-related election misinformation spread globally, but its impact remains unclear.** In 2024, numerous examples of AI-related election misinformation emerged in more than a dozen countries and across over 10 social media platforms, including during the U.S. presidential election. However, questions remain about the measurable impacts of this problem, with many expecting misinformation campaigns to have affected elections more profoundly than they did.

---

# Report Highlights

## CHAPTER 3:
Responsible AI (cont'd)

**9. LLMs trained to be explicitly unbiased continue to demonstrate implicit bias.** Many advanced LLMs—including GPT-4 and Claude 3 Sonnet—were designed with measures to curb explicit biases, but they continue to exhibit implicit ones. The models disproportionately associate negative terms with Black individuals, more often associate women with humanities instead of STEM fields, and favor men for leadership roles, reinforcing racial and gender biases in decision making. Although bias metrics have improved on standard benchmarks, AI model bias remains a pervasive issue.

---

**10. RAI gains attention from academic researchers.** The number of RAI papers accepted at leading AI conferences increased by 28.8%, from 992 in 2023 to 1,278 in 2024, continuing a steady annual rise since 2019. This upward trend highlights the growing importance of RAI within the AI research community.

## CHAPTER 4:
Economy

**1. Global private AI investment hits record high with 26% growth.** Corporate AI investment reached $252.3 billion in 2024, with private investment climbing 44.5% and mergers and acquisitions up 12.1% from the previous year. The sector has experienced dramatic expansion over the past decade, with total investment growing more than thirteenfold since 2014.

---

**2. Generative AI funding soars.** Private investment in generative AI reached $33.9 billion in 2024, up 18.7% from 2023 and over 8.5 times higher than 2022 levels. The sector now represents more than 20% of all AI-related private investment.

---

**3. The U.S. widens its lead in global AI private investment.** U.S. private AI investment hit $109.1 billion in 2024, nearly 12 times higher than China's $9.3 billion and 24 times the U.K.'s $4.5 billion. The gap is even more pronounced in generative AI, where U.S. investment exceeded the combined total of China and the European Union plus the U.K. by $25.4 billion, expanding on its $21.8 billion gap in 2023.

---

**4. Use of AI climbs to unprecedented levels.** In 2024, the proportion of survey respondents reporting AI use by their organizations jumped to 78% from 55% in 2023. Similarly, the number of respondents who reported using generative AI in at least one business function more than doubled—from 33% in 2023 to 71% last year.

---

# Report Highlights

## CHAPTER 4:
Economy (cont'd)

**5. AI is beginning to deliver financial impact across business functions, but most companies are early in their journeys.** Most companies that report financial impacts from using AI within a business function estimate the benefits as being at low levels. 49% of respondents whose organizations use AI in service operations report cost savings, followed by supply chain management (43%) and software engineering (41%), but most of them report cost savings of less than 10%. With regard to revenue, 71% of respondents using AI in marketing and sales report revenue gains, 63% in supply chain management, and 57% in service operations, but the most common level of revenue increases is less than 5%.

---

**6. Use of AI shows dramatic shifts by region, with Greater China gaining ground.** While North America maintains its leadership in organizations' use of AI, Greater China demonstrated one of the most significant year-over-year growth rates, with a 27 percentage point increase in organizational AI use. Europe followed with a 23 percentage point increase, suggesting a rapidly evolving global AI landscape and intensifying international competition in AI implementation.

---

**7. China's dominance in industrial robotics continues despite slight moderation.** In 2023, China installed 276,300 industrial robots, six times more than Japan and 7.3 times more than the United States. Since surpassing Japan in 2013, when China accounted for 20.8% of global installations, its share has risen to 51.1%. While China continues to install more robots than the rest of the world combined, this margin narrowed slightly in 2023, marking a modest moderation in its dramatic expansion.

---

**8. Collaborative and interactive robot installations become more common.** In 2017, collaborative robots represented a mere 2.8% of all new industrial robot installations, a figure that climbed to 10.5% by 2023. Similarly, 2023 saw a rise in service robot installations across all application categories except medical robotics. This trend indicates not just an overall increase in robot installations but also a growing emphasis on deploying robots for human-facing roles.

---

**9. AI is driving significant shifts in energy sources, attracting interest in nuclear energy.** Microsoft announced a $1.6 billion deal to revive the Three Mile Island nuclear reactor to power AI, while Google and Amazon have also secured nuclear energy agreements to support AI operations.

---

**10. AI boosts productivity and bridges skill gaps.** Last year's AI Index was among the first reports to highlight research showing AI's positive impact on productivity. This year, additional studies reinforced those findings, confirming that AI boosts productivity and, in most cases, helps narrow the gap between low- and high-skilled workers.

# Report Highlights

## CHAPTER 5:
Science and Medicine

**1. Bigger and better protein sequencing models emerge.** In 2024, several large-scale, high-performance protein sequencing models, including ESM3 and AlphaFold 3, were launched. Over time, these models have grown significantly in size, leading to continuous improvements in protein prediction accuracy.

**2. AI continues to drive rapid advances in scientific discovery.** AI's role in scientific progress continues to expand. While 2022 and 2023 marked the early stages of AI-driven breakthroughs, 2024 brought even greater advancements, including Aviary, which trains LLM agents for biological tasks, and FireSat, which significantly enhances wildfire prediction.

**3. The clinical knowledge of leading LLMs continues to improve.** OpenAI's recently released o1 set a new state-of-the-art 96.0% on the MedQA benchmark—a 5.8 percentage point gain over the best score posted in 2023. Since late 2022, performance has improved 28.4 percentage points. MedQA, a key benchmark for assessing clinical knowledge, may be approaching saturation, signaling the need for more challenging evaluations.

**4. AI outperforms doctors on key clinical tasks.** A new study found that GPT-4 alone outperformed doctors—both with and without AI—in diagnosing complex clinical cases. Other recent studies show AI surpassing doctors in cancer detection and identifying high-mortality-risk patients. However, some early research suggests that AI-doctor collaboration yields the best results, making it a fruitful area of further research.

**5. The number of FDA-approved, AI-enabled medical devices skyrockets.** The FDA authorized its first AI-enabled medical device in 1995. By 2015, only six such devices had been approved, but the number spiked to 223 by 2023.

**6. Synthetic data shows significant promise in medicine.** Studies released in 2024 suggest that AI-generated synthetic data can help models better identify social determinants of health, enhance privacy-preserving clinical risk prediction, and facilitate the discovery of new drug compounds.

**7. Medical AI ethics publications are increasing year over year.** The number of publications on ethics in medical AI nearly quadrupled from 2020 to 2024, rising from 288 in 2020 to 1,031 in 2024.

# Report Highlights

## CHAPTER 5:
Science and Medicine (cont'd)

**8. Foundation models come to medicine.** In 2024, a wave of large-scale medical foundation models were released, ranging from general-purpose multimodal models like Med-Gemini to specialized models such as EchoCLIP for echocardiology, VisionFM for ophthalmology, and ChexAgent for radiology.

**9. Publicly available protein databases grow in size.** Since 2021, the number of entries in major public protein science databases has grown significantly, including UniProt (31%), PDB (23%), and AlphaFold (585%). This expansion has important implications for scientific discovery.

**10. AI research recognized by two Nobel Prizes.** In 2024, AI-driven research received top honors, with two Nobel Prizes awarded for AI-related breakthroughs. Google DeepMind's Demis Hassabis and John Jumper won the Nobel Prize in Chemistry for their pioneering work on protein folding with AlphaFold. Meanwhile, John Hopfield and Geoffrey Hinton received the Nobel Prize in Physics for their foundational contributions to neural networks.

## CHAPTER 6:
Policy and Governance

**1. U.S. states are leading the way on AI legislation amid slow progress at the federal level.** In 2016, only one state-level AI-related law was passed, increasing to 49 by 2023. In the past year alone, that number more than doubled to 131. While proposed AI bills at the federal level have also increased, the number passed remains low.

**2. Governments across the world invest in AI infrastructure.** Canada announced a $2.4 billion AI infrastructure package, while China launched a $47.5 billion fund to boost semiconductor production. France committed $117 billion to AI infrastructure, India pledged $1.25 billion, and Saudi Arabia's Project Transcendence includes a $100 billion investment in AI.

**3. Across the world, mentions of AI in legislative proceedings keep rising.** Across 75 countries, AI mentions in legislative proceedings increased by 21.3% in 2024, rising to 1,889 from 1,557 in 2023. Since 2016, the total number of AI mentions has grown more than ninefold.

# Report Highlights

## CHAPTER 6:
Policy and Governance (cont'd)

**4. AI safety institutes expand and coordinate across the globe.** In 2024, countries worldwide launched international AI safety institutes. The first emerged in November 2023 in the U.S. and the U.K. following the inaugural AI Safety Summit. At the AI Seoul Summit in May 2024, additional institutes were pledged in Japan, France, Germany, Italy, Singapore, South Korea, Australia, Canada, and the European Union.

---

**5. The number of U.S. AI-related federal regulations skyrockets.** In 2024, 59 AI-related regulations were introduced—more than double the 25 recorded in 2023. These regulations came from 42 unique agencies, twice the 21 agencies that issued them in 2023.

---

**6. U.S. states expand deepfake regulations.** Before 2024, only five states—California, Michigan, Washington, Texas, and Minnesota—had enacted laws regulating deepfakes in elections. In 2024, 15 more states, including Oregon, New Mexico, and New York, introduced similar measures. Additionally, by 2024, 24 states had passed regulations targeting deepfakes.

## CHAPTER 7:
Education

**1. Access to and enrollment in high school computer science (CS) courses in the U.S. has increased slightly from the previous school year, but gaps remain.** Student participation varies by state, race and ethnicity, school size, geography, income, gender, and disability.

---

**2. CS teachers in the U.S. want to teach AI but do not feel equipped to do so.** Despite the 81% of CS teachers who agree that using AI and learning about AI should be included in a foundational CS learning experience, fewer than half of high school CS teachers feel equipped to teach AI.

---

**3. Two-thirds of countries worldwide offer or plan to offer K–12 CS education.** This fraction has doubled since 2019, with African and Latin American countries progressing the most. However, students in African countries have the least amount of access to CS education due to schools' lack of electricity.

---

# Report Highlights

## CHAPTER 7:
Education (cont'd)

**4. Graduates who earned their master's degree in AI in the U.S. nearly doubled between 2022 and 2023.** While increased attention on AI will be slower to emerge in the number of bachelor's and PhD degrees, the surge in master's degrees could indicate a developing trend for all degree levels.

---

**5. The U.S. continues to be a global leader in producing information, technology, and communications (ICT) graduates at all levels.** Spain, Brazil, and the United Kingdom follow the U.S. as top producers at various levels, while Turkey boasts the best gender parity.

## CHAPTER 8:
Public Opinion

**1. The world grows cautiously optimistic about AI products and services.** Among the 26 nations surveyed by Ipsos in both 2022 and 2024, 18 saw an increase in the proportion of people who believe AI products and services offer more benefits than drawbacks. Globally, the share of individuals who see AI products and services as more beneficial than harmful has risen from 52% in 2022 to 55% in 2024.

---

**2. The expectation and acknowledgment of AI's impact on daily life is rising.** Around the world, two thirds of people now believe that AI-powered products and services will significantly impact daily life within the next three to five years—an increase of 6 percentage points since 2022. Every country except Malaysia, Poland, and India saw an increase in this perception since 2022, with the largest jumps in Canada (17%) and Germany (15%).

---

**3. Skepticism about the ethical conduct of AI companies is growing, while trust in the fairness of AI is declining.** Globally, confidence that AI companies protect personal data fell from 50% in 2023 to 47% in 2024. Likewise, fewer people today believe that AI systems are unbiased and free from discrimination compared to last year.

---

**4. Regional differences persist regarding AI optimism.** First reported in the 2023 AI Index, significant regional differences in AI optimism endure. A large majority of people believe AI-powered products and services offer more benefits than drawbacks in countries like China (83%), Indonesia (80%), and Thailand (77%), while only a minority share this view in Canada (40%), the United States (39%), and the Netherlands (36%).

---

# Report Highlights

## CHAPTER 8:
Public Opinion (cont'd)

**5. People in the United States remain distrustful of self-driving cars.** A recent American Automobile Association survey found that 61% of people in the U.S. fear self-driving cars, and only 13% trust them. Although the percentage who expressed fear has declined from its 2023 peak of 68%, it remains higher than in 2021 (54%).

---

**6. There is broad support for AI regulation among local U.S. policymakers.** In 2023, 73.7% of local U.S. policymakers—spanning township, municipal, and county levels—agreed that AI should be regulated, up significantly from 55.7% in 2022. Support was stronger among Democrats (79.2%) than Republicans (55.5%), though both registered notable increases over 2022.

---

**7. AI optimism registers sharp increase among countries that previously showed the most skepticism.** Globally, optimism about AI products and services has increased, with the sharpest gains in countries that were previously the most skeptical. In 2022, Great Britain (38%), Germany (37%), the United States (35%), Canada (32%), and France (31%) were among the least likely to view AI as having more benefits than drawbacks. Since then, optimism has grown in these countries by 8%, 10%, 4%, 8%, and 10%, respectively.

---

**8. Workers expect AI to reshape jobs, but fear of replacement remains lower.** Globally, 60% of respondents agree that AI will change how individuals do their job in the next five years. However, a smaller subset of respondents, 36%, believe that AI will replace their jobs in the next five years.

---

**9. Sharp divides exist among local U.S. policymakers on AI policy priorities.** While local U.S. policymakers broadly support AI regulation, their priorities vary. The strongest backing is for stricter data privacy rules (80.4%), retraining for the unemployed (76.2%), and AI deployment regulations (72.5%). However, support drops significantly for a law enforcement facial recognition ban (34.2%), wage subsidies for wage declines (32.9%), and universal basic income (24.6%).

---

**10. AI is seen as a time saver and entertainment booster, but doubts remain on its economic impact.** Global perspectives on AI's impact vary. While 55% believe it will save time, and 51% expect it will offer better entertainment options, fewer are confident in its health or economic benefits. Only 38% think AI will improve health, whilst 36% think AI will improve the national economy, 31% see a positive impact on the job market, and 37% believe it will enhance their own jobs.

# CHAPTER 1:
Research and Development

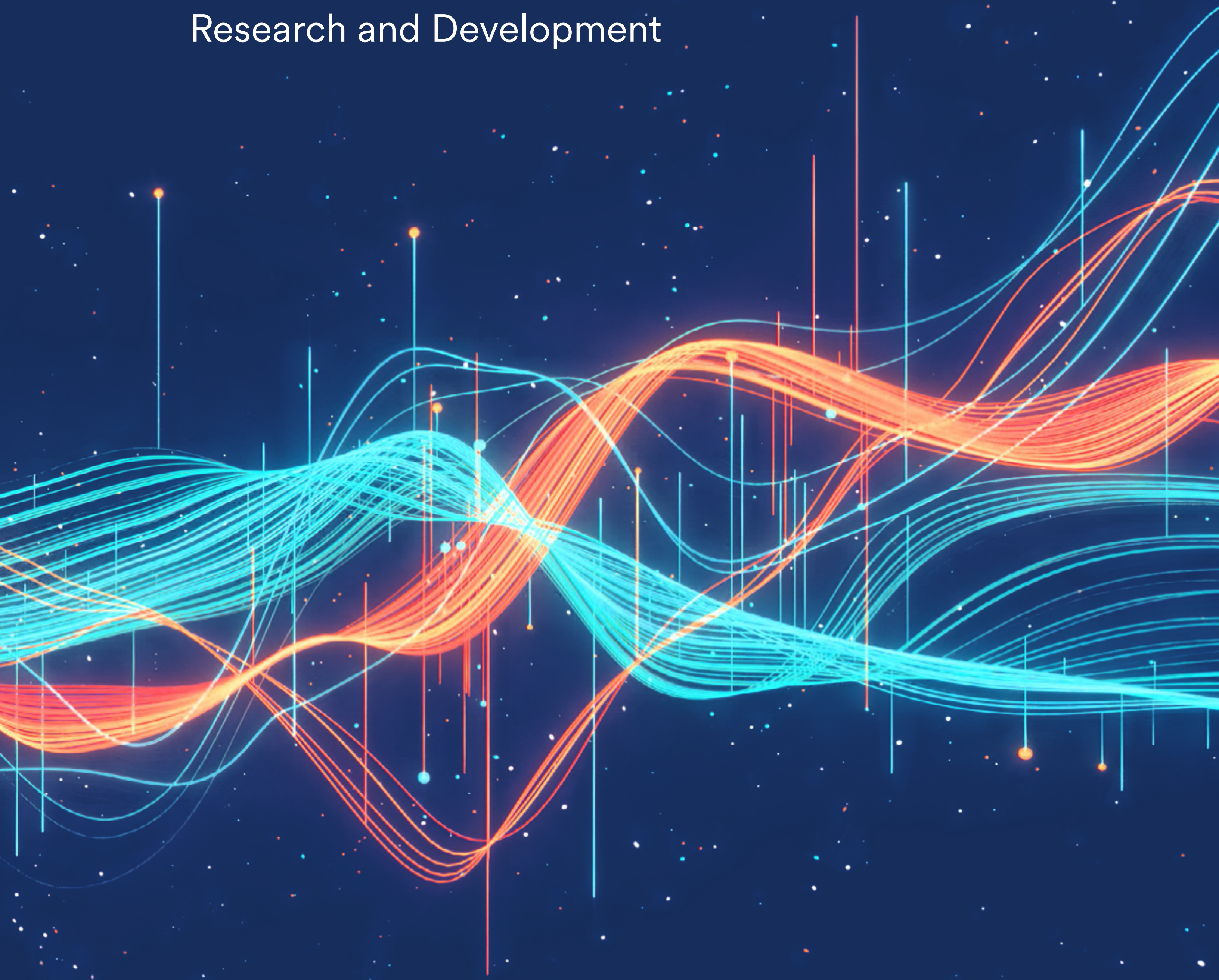

# Chapter 1: Research and Development

ACCESS THE PUBLIC DATA

**CHAPTER 1:**
Research and Development

# Overview

This chapter explores trends in AI research and development, beginning with an analysis of AI publications, patents, and notable AI systems. These topics are examined through the lens of the countries, organizations, and sectors producing them. The chapter also covers AI model training costs, AI conference attendance, and open-source AI software. New additions this year include profiles of the evolving AI hardware ecosystem, an assessment of AI training's energy requirements and environmental impact, and a temporal analysis of model inference costs.

**CHAPTER 1:**
Research and Development

# Chapter Highlights

**1. Industry continues to make significant investments in AI and leads in notable AI model development, while academia leads in highly cited research.** Industry's lead in notable model development, highlighted in the two previous AI Index reports, has only grown more pronounced, with nearly 90% of notable models in 2024 (compared to 60% in 2023) originating from industry. Academia has remained the single leading institutional producer of highly cited (top 100) publications over the past three years.

---

**2. China leads in AI research publication totals, while the United States leads in highly influential research.** In 2023, China produced more AI publications (23.2%) and citations (22.6%) than any other country. Over the past three years, U.S. institutions have contributed the most top-100-cited AI publications.

---

**3. AI publication totals continue to grow and increasingly dominate computer science.** Between 2013 and 2023, the total number of AI publications in venues related to computer science and other scientific disciplines nearly tripled, increasing from approximately 102,000 to over 242,000. Proportionally, AI's share of computer science publications has risen from 21.6% in 2013 to 41.8% in 2023.

---

**4. The United States continues to be the leading source of notable AI models.** In 2024, U.S.-based institutions produced 40 notable AI models, significantly surpassing China's 15 and Europe's combined total of three. In the past decade, more notable machine learning models have originated from the United States than any other country.

---

**5. AI models get increasingly bigger, more computationally demanding, and more energy intensive.** New research finds that the training compute for notable AI models doubles approximately every five months, dataset sizes for training LLMs every eight months, and the power required for training annually. Large-scale industry investment continues to drive model scaling and performance gains.

---

# Chapter Highlights (cont'd)

**6. AI models become increasingly affordable to use.** The cost of querying an AI model that scores the equivalent of GPT-3.5 (64.8) on MMLU, a popular benchmark for assessing language model performance, dropped from $20.00 per million tokens in November 2022 to just $0.07 per million tokens by October 2024 (Gemini-1.5-Flash-8B)—a more than 280-fold reduction in approximately 18 months. Depending on the task, LLM inference prices have fallen anywhere from 9 to 900 times per year.

---

**7. AI patenting is on the rise.** Between 2010 and 2023, the number of AI patents has grown steadily and significantly, ballooning from 3,833 to 122,511. In just the last year, the number of AI patents has risen 29.6%. As of 2023, China leads in total AI patents, accounting for 69.7% of all grants, while South Korea and Luxembourg stand out as top AI patent producers on a per capita basis.

---

**8. AI hardware gets faster, cheaper, and more energy efficient.** New research suggests that machine learning hardware performance, measured in 16-bit floating-point operations, has grown 43% annually, doubling every 1.9 years. Price performance has improved, with costs dropping 30% per year, while energy efficiency has increased by 40% annually.

---

**9. Carbon emissions from AI training are steadily increasing.** Training early AI models, such as AlexNet (2012), had modest amounts of carbon emissions at 0.01 tons. More recent models have significantly higher emissions for training: GPT-3 (2020) at 588 tons, GPT-4 (2023) at 5,184 tons, and Llama 3.1 405B (2024) at 8,930 tons. For perspective, the average American emits 18 tons of carbon per year.

# 1.1 Publications

The figures below show the global count of English-language AI publications from 2010 to 2023, categorized by affiliation type, publication type, and region. New to this year's report, the AI Index includes a section analyzing trends among the 100 most-cited AI publications, which can offer insights into particularly high-impact research. This year, the AI Index analyzed AI publication trends using the OpenAlex database. As a result, the numbers in this year's report differ slightly from those in previous editions.[1] Given that there is a significant lag in the collection of publication metadata, and that in some cases it takes until the middle of any given year to fully capture the previous year's publications, in this year's report, the AI Index team elected to examine publication trends only through 2023.

## Overview

The following section reports on trends in the total number of English-language AI publications.

### Total Number of AI Publications

Figure 1.1.1 displays the global count of AI publications. These are the publications with a computer science (CS) label in the OpenAlex catalog that were classified by the AI Index as being related to AI.[2] Between 2013 and 2023, the total number of AI

**Number of AI publications in CS worldwide, 2013–23**
Source: AI Index, 2025 | Chart: 2025 AI Index report

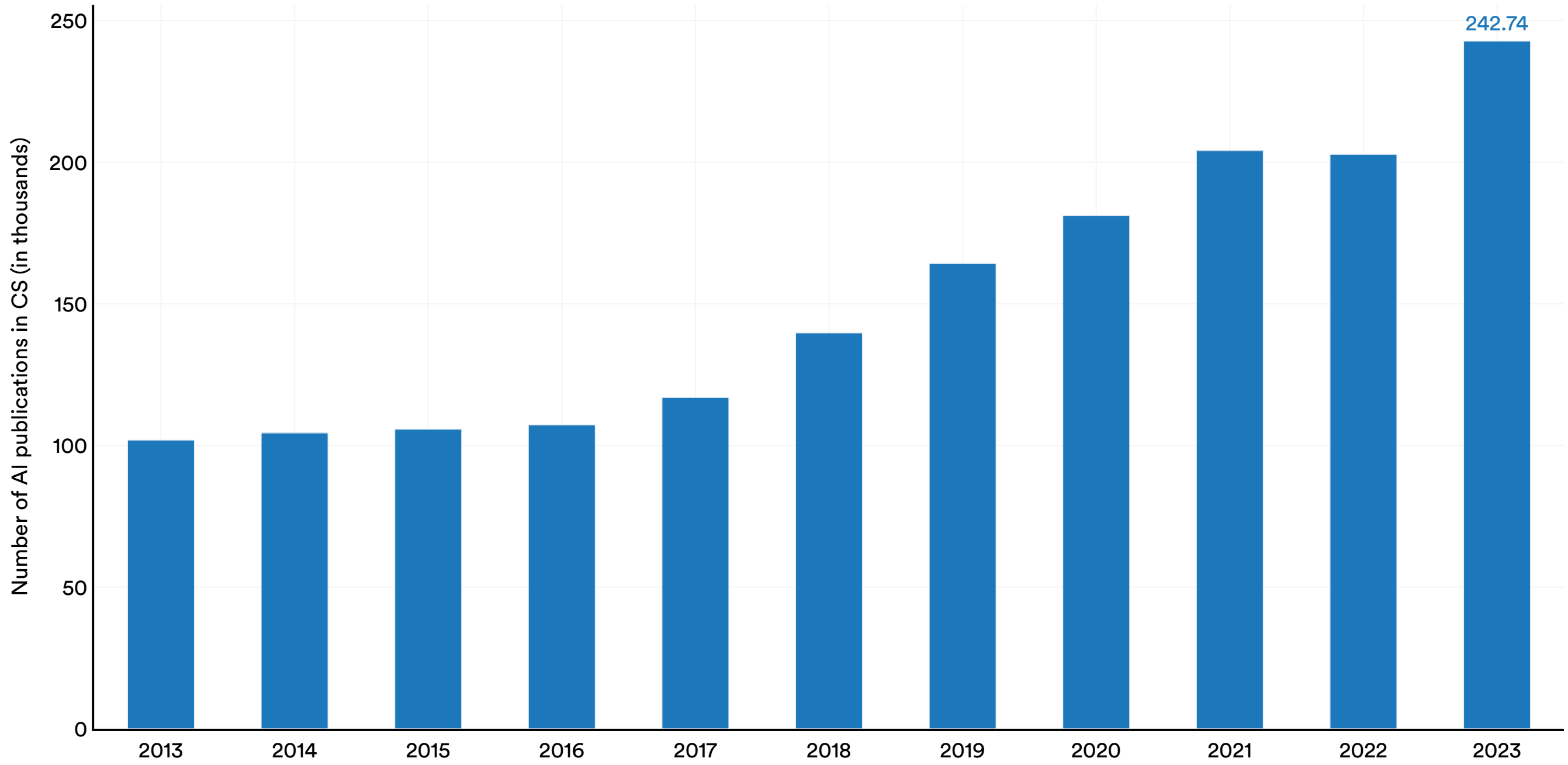

Figure 1.1.1

1 OpenAlex is a fully open catalog of scholarly metadata, including scientific papers, authors, institutions, and more. The AI Index used OpenAlex as a bibliographic database and automatically classified AI-related research using the latest version of the CSO Classifier. In previous years, the Index relied on third-party providers with different underlying data sources and classification methods. As a result, this year's findings differ slightly from those included in previous reports. Additionally, the AI Index applied the classifier only to papers that OpenAlex categorized under the broad field of computer science. This approach may have led to an undercount of AI-related publications by excluding research from fields like social sciences that employ AI methodologies but fall outside the computer science–designated classification.

2 The CSO Classifier (v3.3) is an automated text classification system designed to categorize research papers in computer science using a comprehensive ontology of 15,000 topics and 166,000 relationships, including emerging fields like GenAI, LLMs, and prompt engineering. It processes metadata (such as title and abstract) through three modules: a syntactic module for exact topic matches, a semantic module leveraging word embeddings to infer related topics, and a post-processing module that refines results by filtering outliers and adding relevant higher-level areas.

publications more than doubled, rising from approximately 102,000 in 2013 to more than 242,000 in 2023. The increase over the last year was a meaningful 19.7%. Many fields within computer science, from hardware and software engineering to human-computer interaction, are now contributing to AI. As a result, the observed growth reflects a broader and increased interest in AI across the discipline.

**AI publications in CS (% of total) worldwide, 2013–23**
Source: AI Index, 2025 | Chart: 2025 AI Index report

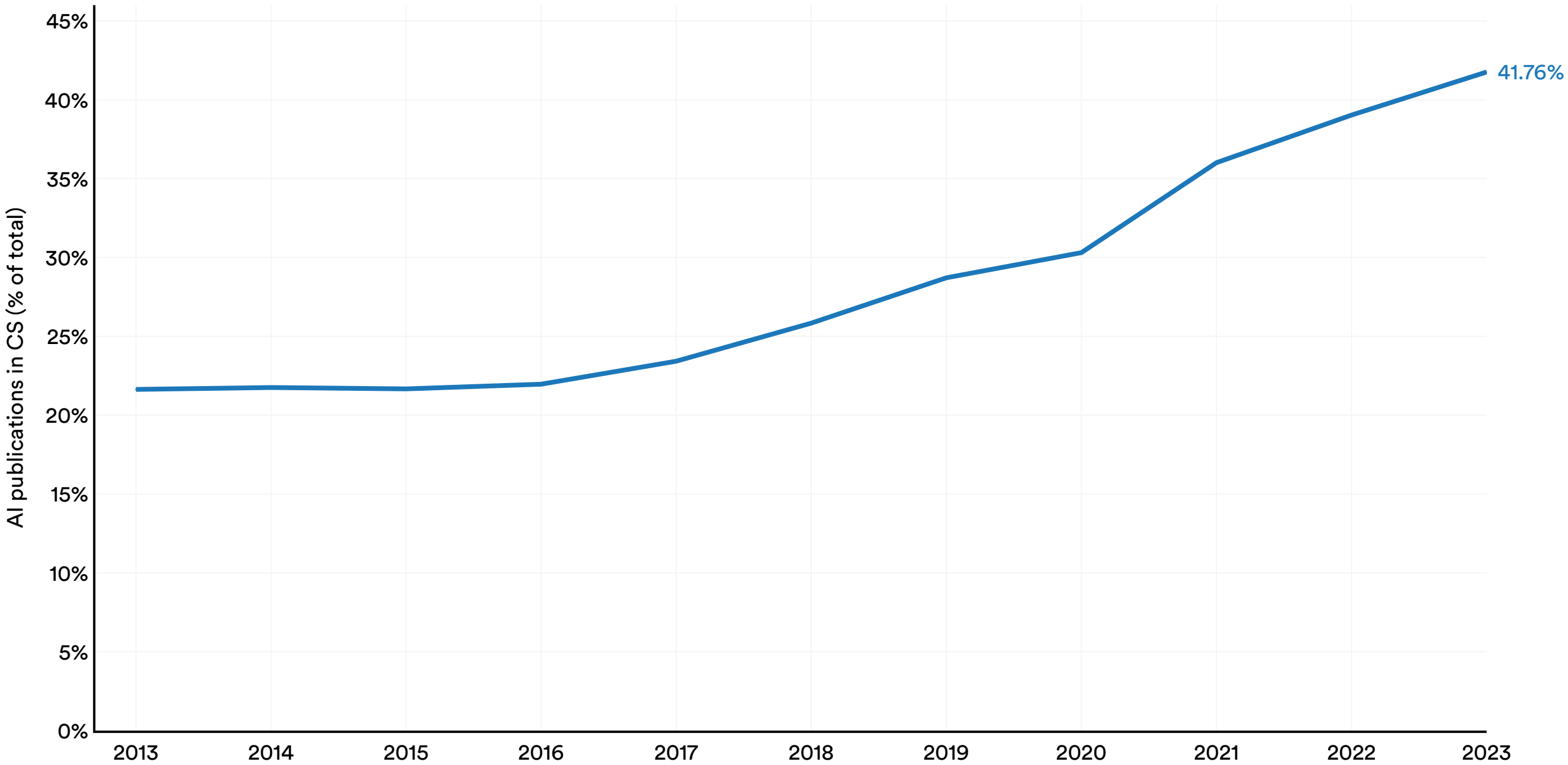

Figure 1.1.2

Figure 1.1.2 shows the proportion of computer science publications in the OpenAlex corpus classified as AI-related. Figure 1.1.2 features the same data included in Figure 1.1.1 but in a proportional form. The share of AI publications has grown significantly, almost doubling from 2013 to 2023.

## By Venue

AI researchers publish their work across various venues. Figure 1.1.3 visualizes the total number of AI publications by venue type. In 2023, journals accounted for the largest share of AI publications (41.8%), followed by conferences (34.3%). Even though the total number of journal and conference publications has increased since 2013, the share of AI publications in journals and conferences has steadily declined, from 52.6% and 36.4% in 2013 to 41.8% and 34.3%, respectively, in 2023. Conversely, AI publications in repositories like arXiv have seen a growing share.

### Number of AI publications in CS by venue type, 2013–23

Source: AI Index, 2025 | Chart: 2025 AI Index report

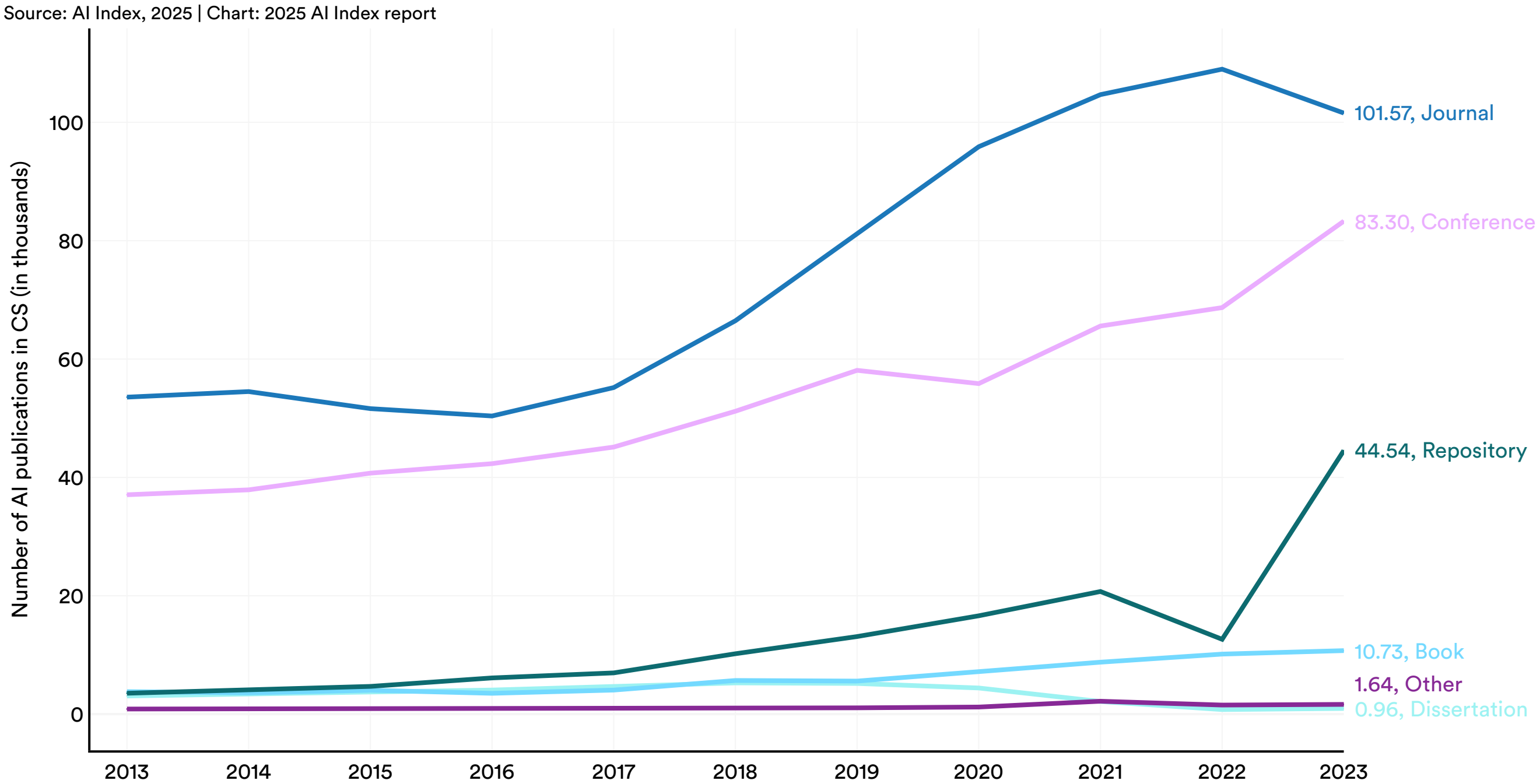

Figure 1.1.3

### By National Affiliation

Figure 1.1.4 visualizes AI publications over time by region.[3] In 2023, East Asia and the Pacific led AI research output, accounting for 34.5% of all AI publications, followed by Europe and Central Asia (18.2%) and North America (10.3%).[4]

While Figure 1.1.4 examines the geographic distribution of AI publications, identifying which regions produce the most research, Figure 1.1.5 focuses on citations, measuring the share of total AI publication citations attributed to work originating from each region. As of 2023, AI publications from East Asia and the Pacific accounted for the largest share of AI article citations at 37.1% (Figure 1.1.5). In 2017, citation shares from East Asia and the Pacific and North America were roughly equal, but since then, North American and European citation shares have declined, while East Asia and the Pacific's share has risen sharply.

**AI publications in CS (% of total) by region, 2013–23**
Source: AI Index, 2025 | Chart: 2025 AI Index report

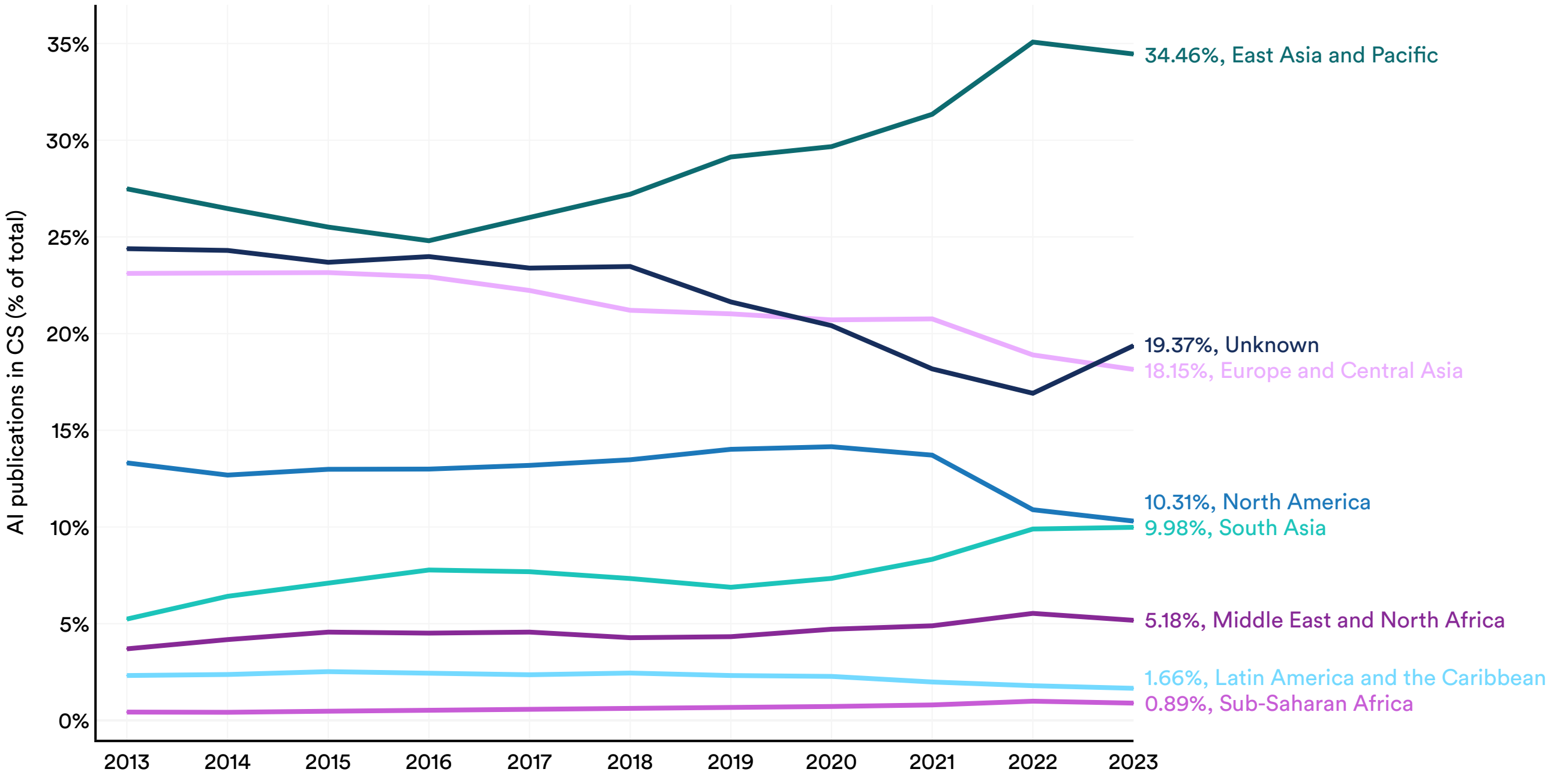

Figure 1.1.4

3 Regions in this chapter are classified according to the World Bank analytical grouping. The AI Index determines the country affiliation of authors using the "countries" field from the authorship data. This field lists all the countries an author is affiliated with, as retrieved from OpenAlex based on institutional affiliations. These affiliations can be explicitly stated in the paper or inferred from the author's most recent publications. When counting publications by country, the AI Index assigns one count to each country linked to the publication. For example, if a paper has three authors, two affiliated with institutions in the U.S. and one in China, the publication is counted once for the U.S. and once for China.

4 A publication may have an "unknown" country affiliation when the author's institutional affiliation is missing or incomplete. This issue arises due to various factors, including unstructured or omitted institution names, platform functional deficiencies, group authorship practices, unstandardized affiliation labeling, document type inconsistencies, or the author's limited publication record. The problem as it relates to OpenAlex is addressed in this paper; however, the issue of missing institutions pertains to other bibliographic databases as well.

**AI publication citations in CS (% of total) by region, 2013–23**
Source: AI Index, 2025 | Chart: 2025 AI Index report

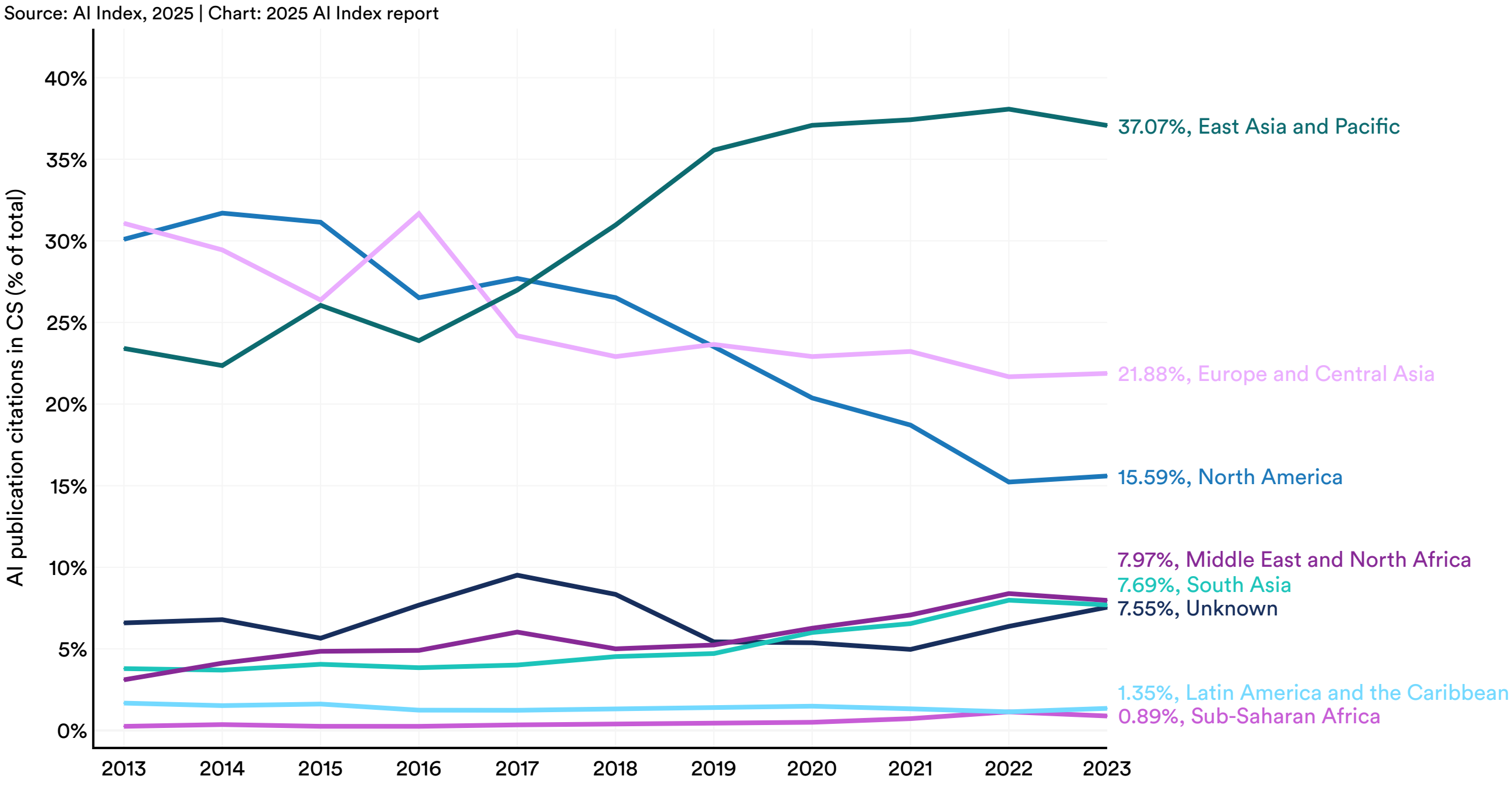

Figure 1.1.5

In 2023, China was the global leader in AI article publications, accounting for 23.2% of the total, compared to 15.2% from Europe and 9.2% from India (Figure 1.1.6).[5] Since 2016, China's share has steadily increased, while the proportion attributed to Europe has declined. AI publications attributed to the United States remained relatively stable until 2021 but have shown a slight decline since then.

**AI publications in CS (% of total) by select geographic areas, 2013–23**
Source: AI Index, 2025 | Chart: 2025 AI Index report

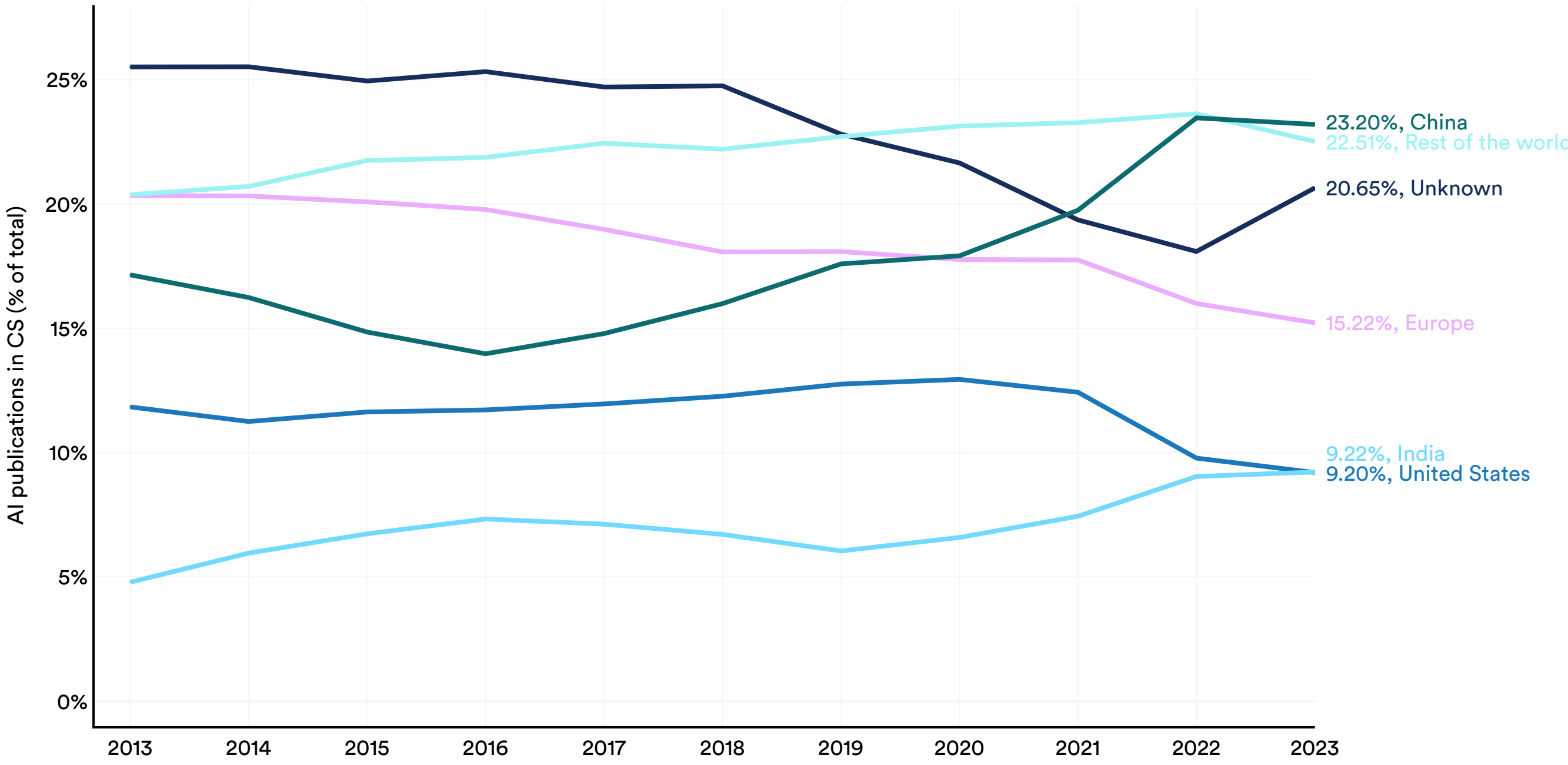

Figure 1.1.6[6]

5 For the "Europe" designation in this and other chapters of the report, the AI Index follows the list of countries defined by the United Nations Statistics Division.

6 To maintain concision, the AI Index visualized results for a select group of countries. However, full results for all countries will be available on the AI Index's Global Vibrancy Tool, which is set to be updated in summer 2025. For immediate access to country-specific research and development data, please contact the AI Index team.

In 2023, Chinese AI publications accounted for 22.6% of all AI citations, followed by Europe at 20.9% and the United States at 13.0% (Figure 1.1.7). As with total AI publications, the late 2010s marked a turning point when China surpassed Europe and the U.S. as the leading source of AI publication citations.

**AI publication citations in CS (% of total) by select geographic areas, 2013–23**

Source: AI Index, 2025 | Chart: 2025 AI Index report

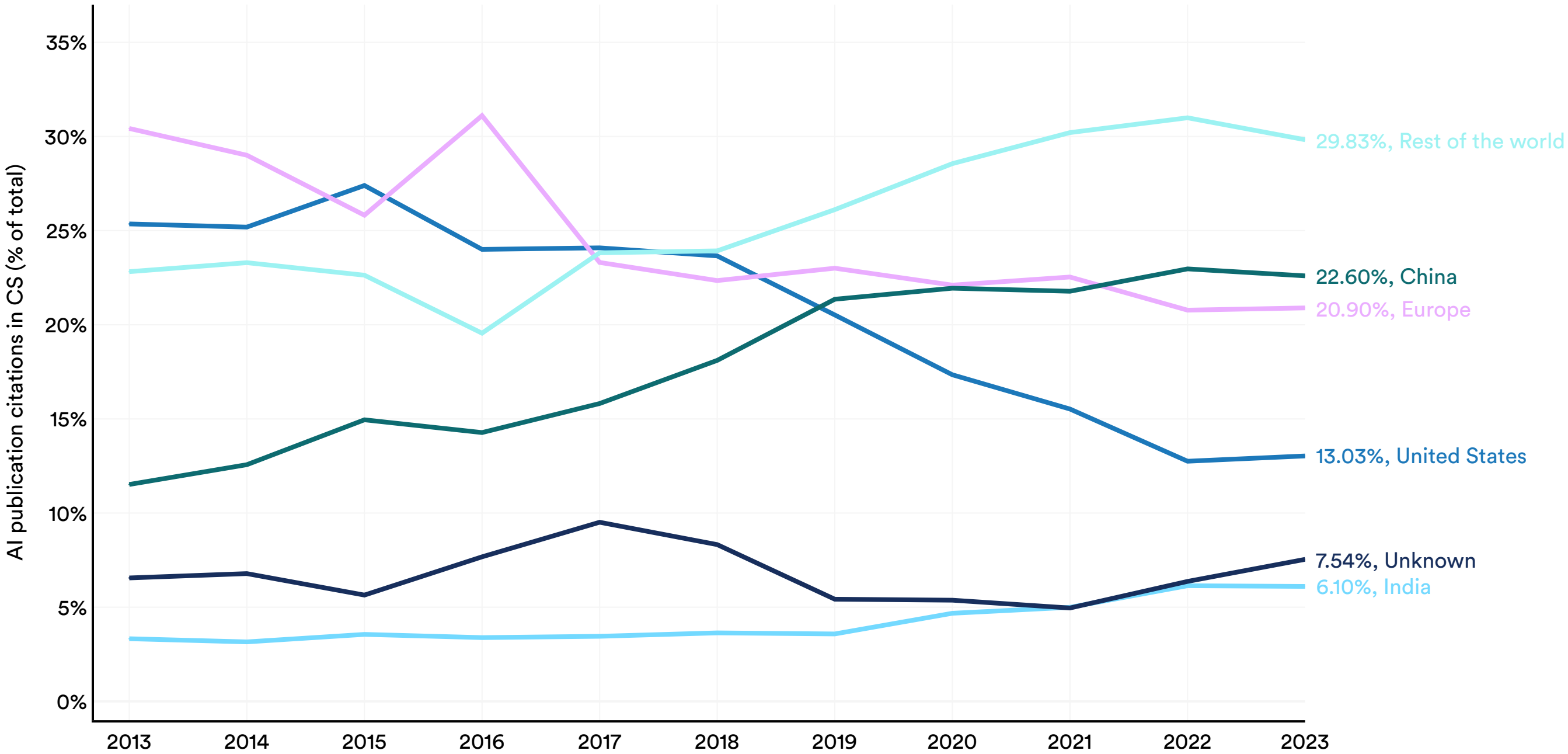

Figure 1.1.7

## By Sector

Academic institutions remain the primary source of AI publications worldwide (Figure 1.1.8). In 2013, they accounted for 85.9% of all AI publications, a figure that remained high, at 84.9%, in 2023. Industry contributed 7.1% of AI publications in 2023, followed by government institutions at 4.9% and nonprofit organizations at 1.7%.

### AI publications in CS (% of total) by sector, 2013–23

Source: AI Index, 2025 | Chart: 2025 AI Index report

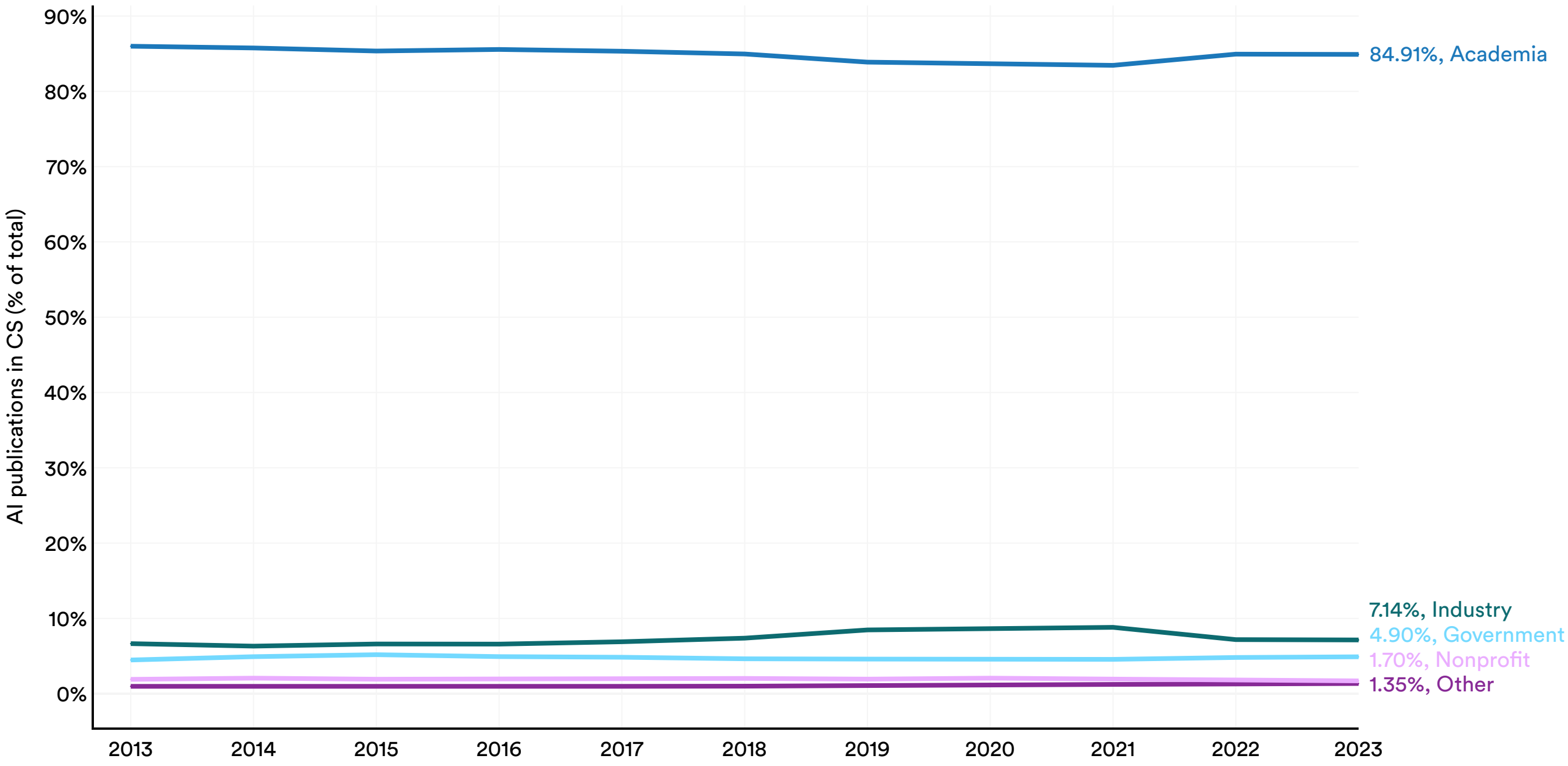

Figure 1.1.8[7]

7 For Figures 1.1.8 and 1.1.9, publications with unknown affiliations were excluded from the final visualization.

AI publications emerge from various sectors in differing proportions across geographic regions. In the United States, a higher share of AI publications (16.5%) comes from industry compared to China (8.0%) (Figure 1.1.9). Among major geographic areas, China has the highest percentage of AI publications originating from the education sector (84.5%).

**AI publications in CS (% of total) by sector and select geographic areas, 2023**
Source: AI Index, 2025 | Chart: 2025 AI Index report

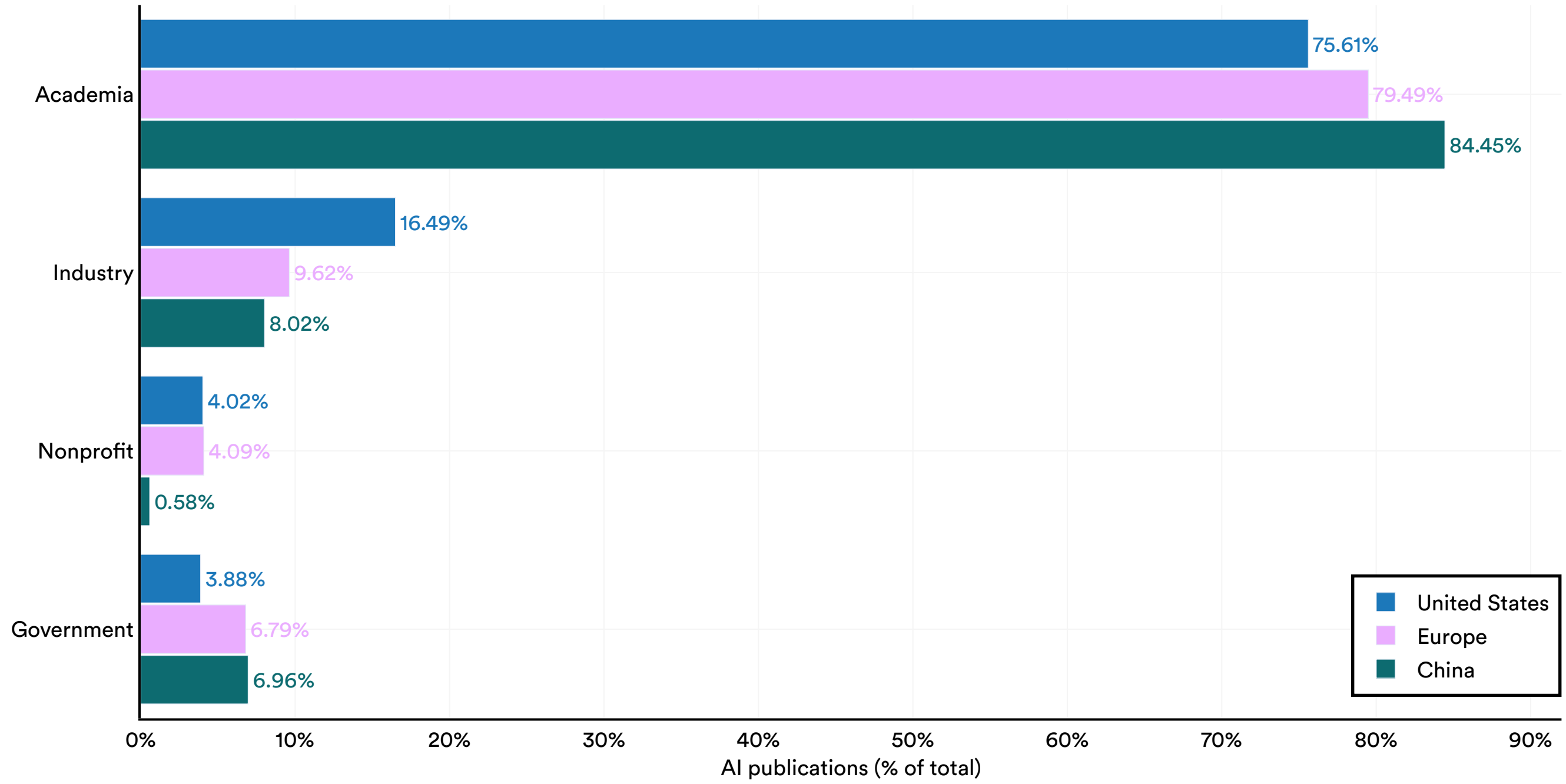

Figure 1.1.9

## By Topic

Machine learning was the most prevalent research topic in AI publications in 2023, comprising 75.7% of publications, followed by computer vision (47.2%), pattern recognition (25.9%) and natural language processing (17.1%) (Figure 1.1.10). Over the past year, there has been a sharp increase in publications on generative AI.

### Number of AI publications by select top topics, 2013–23

Source: AI Index, 2025 | Chart: 2025 AI Index report

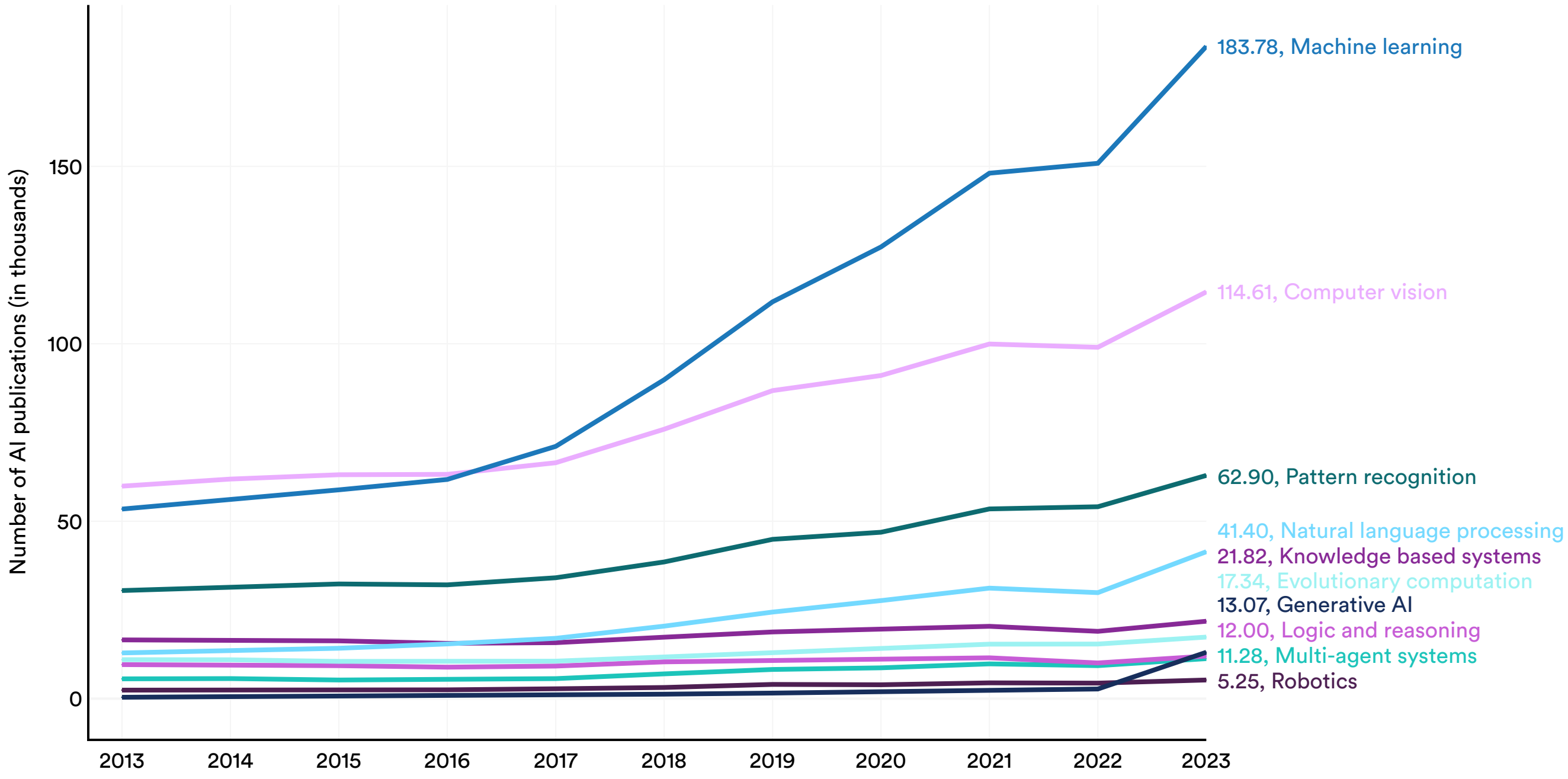

Figure 1.1.10[8]

8 The AI Index categorized papers using its own topic classifier. It is possible for a single publication to be assigned multiple topic labels.

## Top 100 Publications

While tracking total AI publications provides a broad view of research activity, focusing on the most-cited papers offers a perspective of the field's most influential work. This analysis sheds light on where some of the most groundbreaking and influential AI research is emerging. This year, the AI Index identified the 100 most-cited AI publications in 2021, 2022, and 2023, using citation data from OpenAlex. This analysis was further supplemented with insights from Google Scholar and Semantic Scholar.[9] Some of the most highly cited AI publications in 2023 included OpenAI's GPT-4 technical report, Meta's Llama 2 technical report, and Google's PaLM-E technical report. It is important to note that due to citation lag, the most-cited papers in this year's report may change in future editions.

### By National Affiliation

Figure 1.1.11 illustrates the geographic distribution of the top 100 most-cited AI publications by year. From 2021 to 2023, the U.S. consistently had the highest number of top-cited publications, with 64 in 2021, 59 in 2022, and 50 in 2023.[10] In each of these years, China ranked second. Since 2021, the U.S. share of top AI publications has gradually declined.

**Number of highly cited publications in top 100 by select geographic areas, 2021–23**
Source: AI Index, 2025 | Chart: 2025 AI Index report

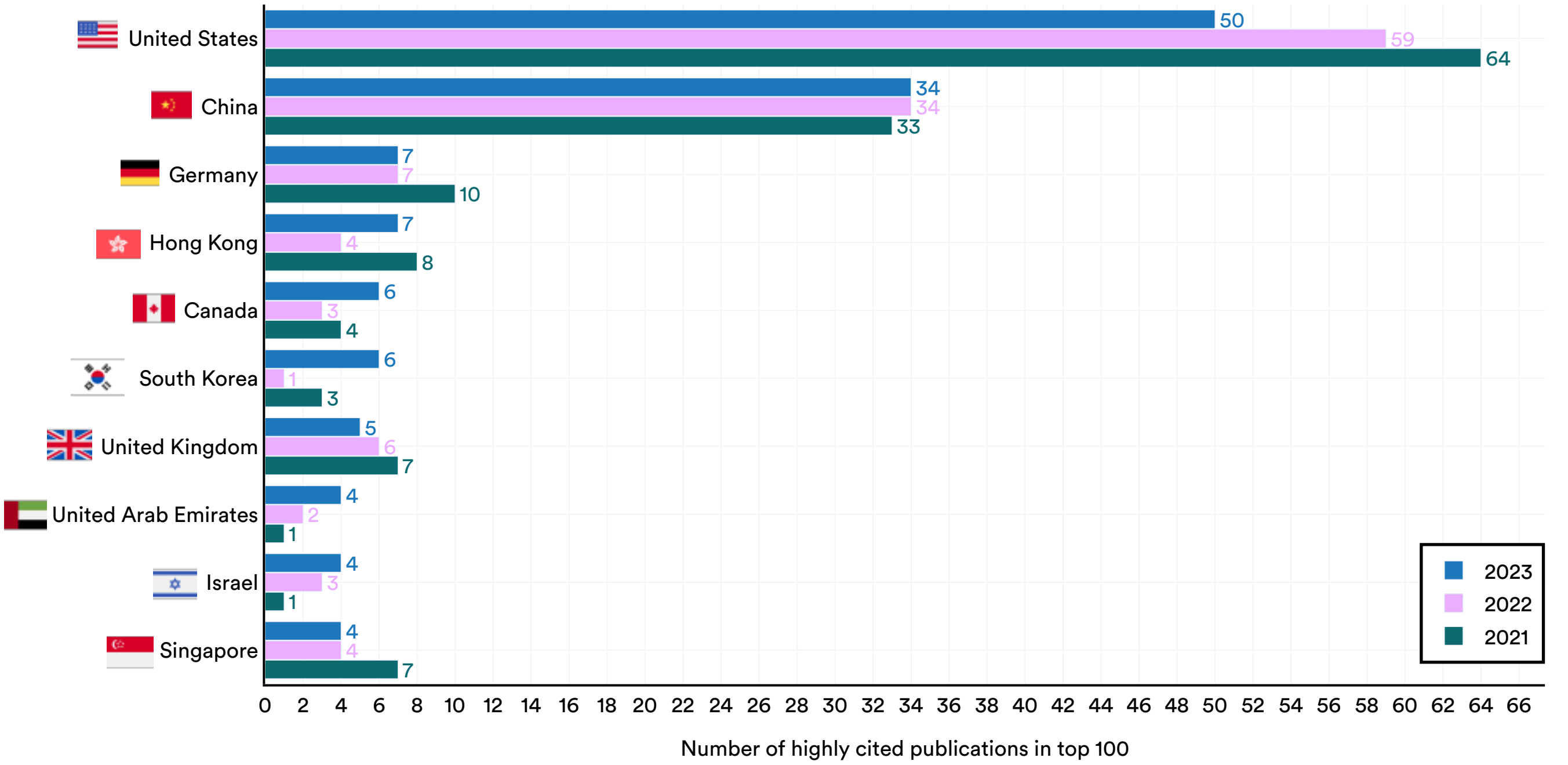

Figure 1.1.11

9 The full methodological guide can be accessed in the Appendix, along with the list of the top 100 articles.

10 A publication can have multiple authors from different countries or organizations. For example, if a paper includes authors from multiple countries, each country is credited once. As a result, the totals in this section's figures exceed 100.

### By Sector

Academia consistently produces the most top-cited AI publications, with 42 in 2023, 27 in 2022, and 34 in 2021 (Figure 1.1.12). Notably, there was a sharp decline in industry contributions, with the number of top 100 publications dropping from 17 in 2021 and 19 in 2022 to just 7 in 2023. As AI research grows more competitive, many industrial AI labs are publishing less frequently or disclosing fewer details about their research in their publications.

**Number of highly cited publications in top 100 by sector, 2021–23**

Source: AI Index, 2025 | Chart: 2025 AI Index report

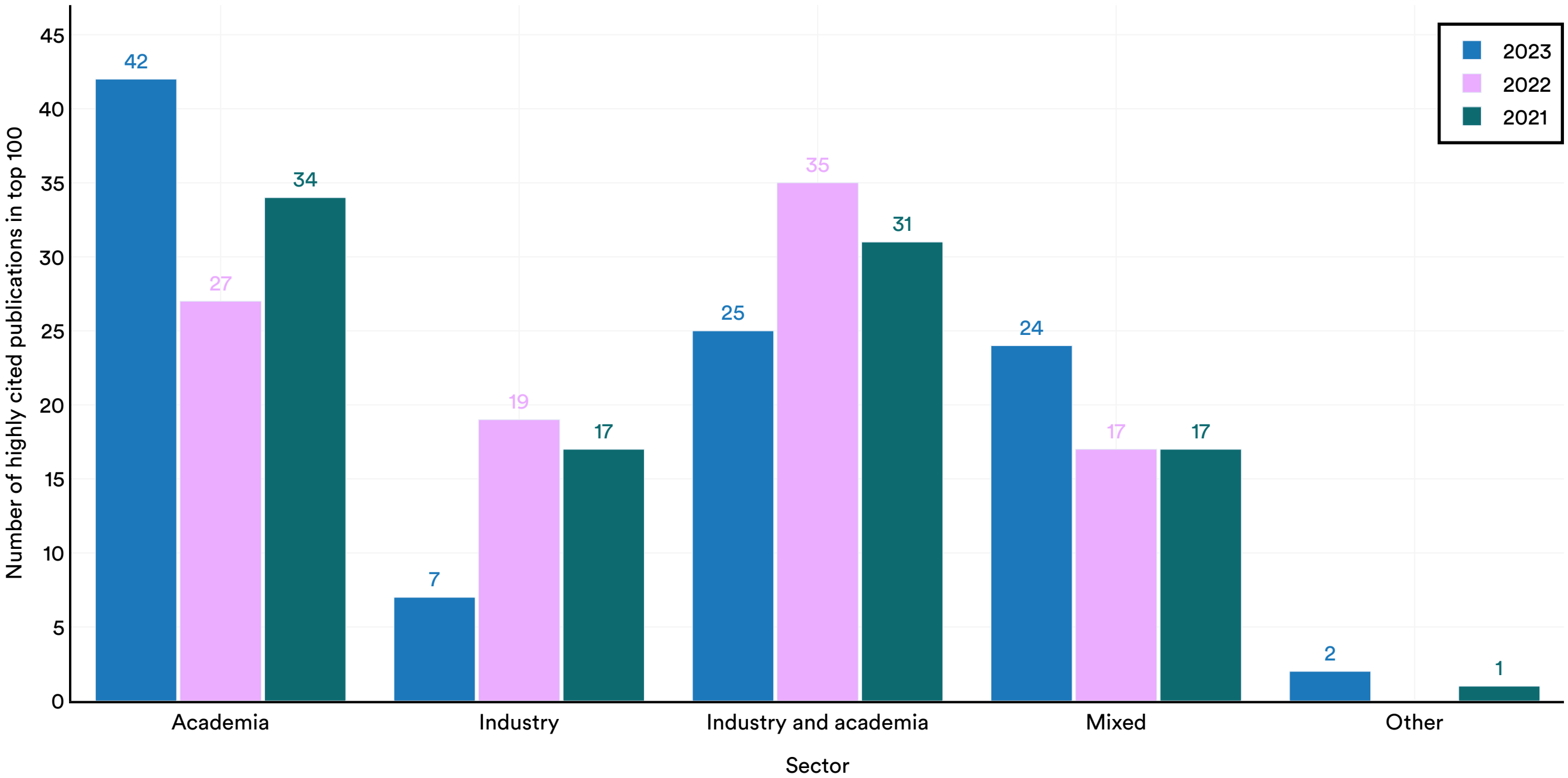

Figure 1.1.12[11]

11 The "mixed" designation includes all intersector collaborations that are not industry and academia (e.g., industry and government, academia and nonprofit). Some institutions lack data for 2021 because they did not have papers included in the top 100 that year. Since papers can have multiple authors from different institutions, the total institutional tags in Figure 1.1.12 may exceed 100. Also, because two of the papers had authors with an unknown sectoral affiliation, the total sum of publications in Figure 1.1.12 is 98.

## By Organization

Figure 1.1.13 highlights the organizations that produced the top 100 most-cited AI publications from 2021 to 2023. Some organizations may have empty bars on the chart if they lacked a top 100 publication in a given year. Additionally, Figure 1.1.13 highlights only the top 10 institutions, though many others contribute significant research.

Google led each year, but it tied with Tsinghua University in 2023, when both contributed eight publications to the top 100. In 2023, Carnegie Mellon University was the highest-ranked U.S. academic institution.

**Number of highly cited publications in top 100 by organization, 2021–23**
Source: AI Index, 2025 | Chart: 2025 AI Index report

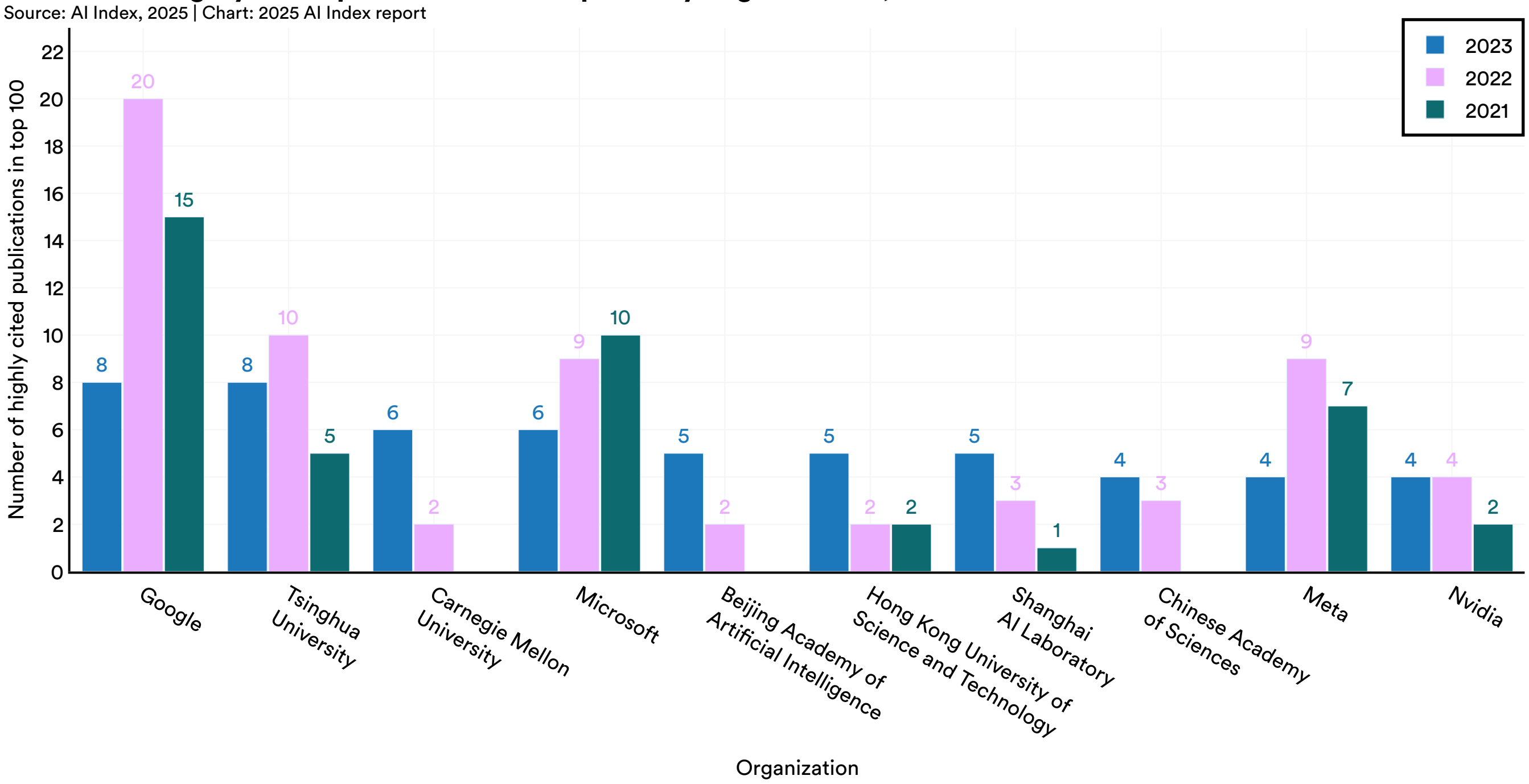

Figure 1.1.13

This section examines trends over time in global AI patents, which can reveal important insights into the evolution of innovation, research, and development within AI. Additionally, analyzing AI patents can reveal how these advances are distributed globally. Similar to the publications data, there is a noticeable delay in AI patent data availability, with 2023 being the most recent year for which data is accessible. The data in this section is sourced from patent-level bibliographic records in PATSTAT Global, a comprehensive database provided by the European Patent Office (EPO).[12]

# 1.2 Patents

## Overview

Figure 1.2.1 examines the global growth in granted AI patents from 2010 to 2023. Over the past dozen years, the number of AI patents has grown steadily and significantly, increasing from 3,833 in 2010 to 122,511 in 2023. In the last year, the number of AI patents has risen 29.6%.

**Number of AI patents granted worldwide, 2010–23**
Source: AI Index, 2025 | Chart: 2025 AI Index report

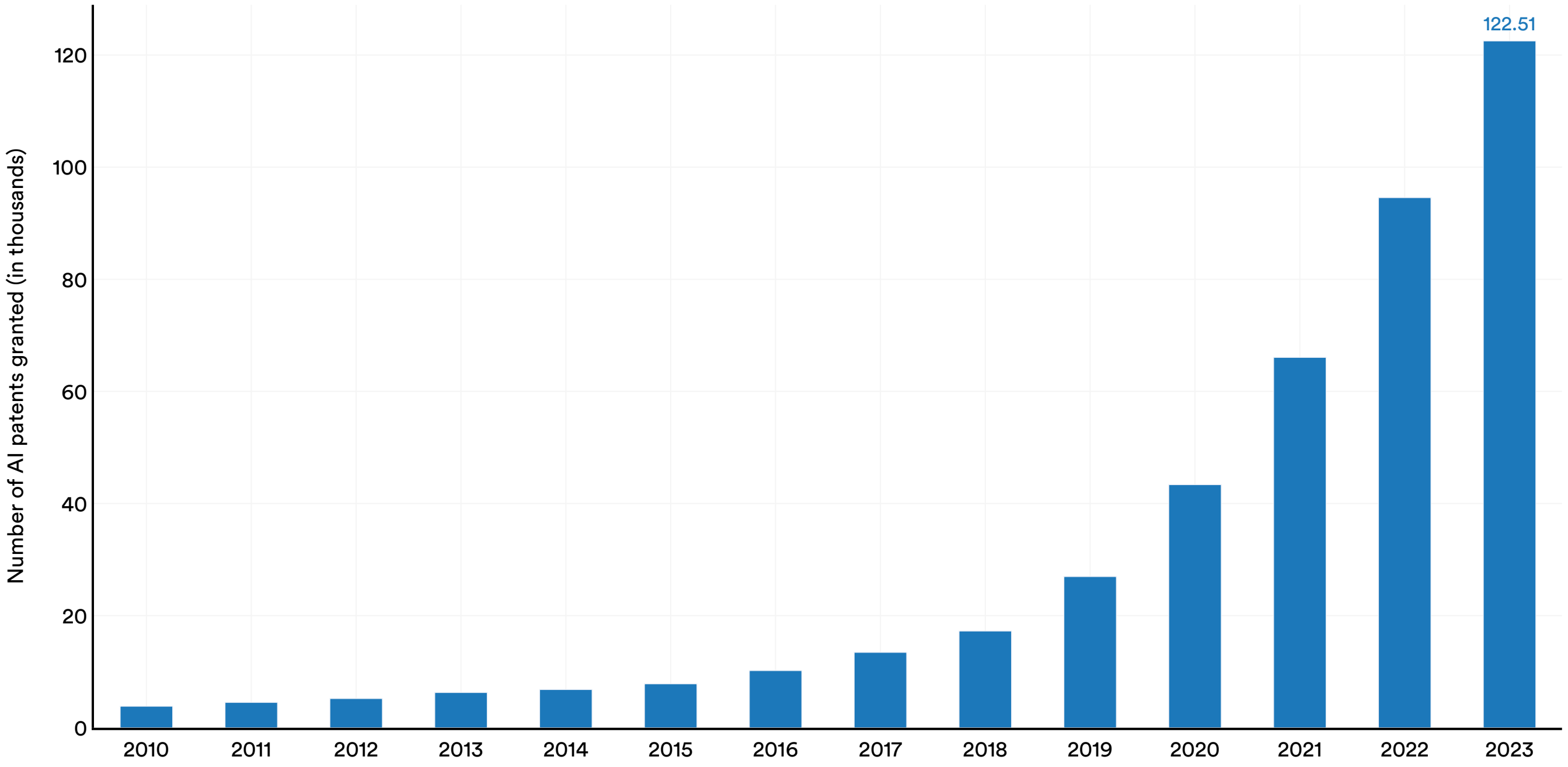

Figure 1.2.1

12 More details on the methodology behind the patent analysis in this section can be found in the Appendix.

## By National Affiliation

Figure 1.2.2 showcases the regional breakdown of granted AI patents, as in the number of patents filed in different regions across the world. As of 2023, the bulk of the world's granted AI patents (82.4%) originated from East Asia and the Pacific, with North America being the next largest contributor at 14.2%. Since 2010, the gap in AI patent grants between East Asia and the Pacific and North America has steadily widened.

**Granted AI patents (% of world total) by region, 2010–23**

Source: AI Index, 2025 | Chart: 2025 AI Index report

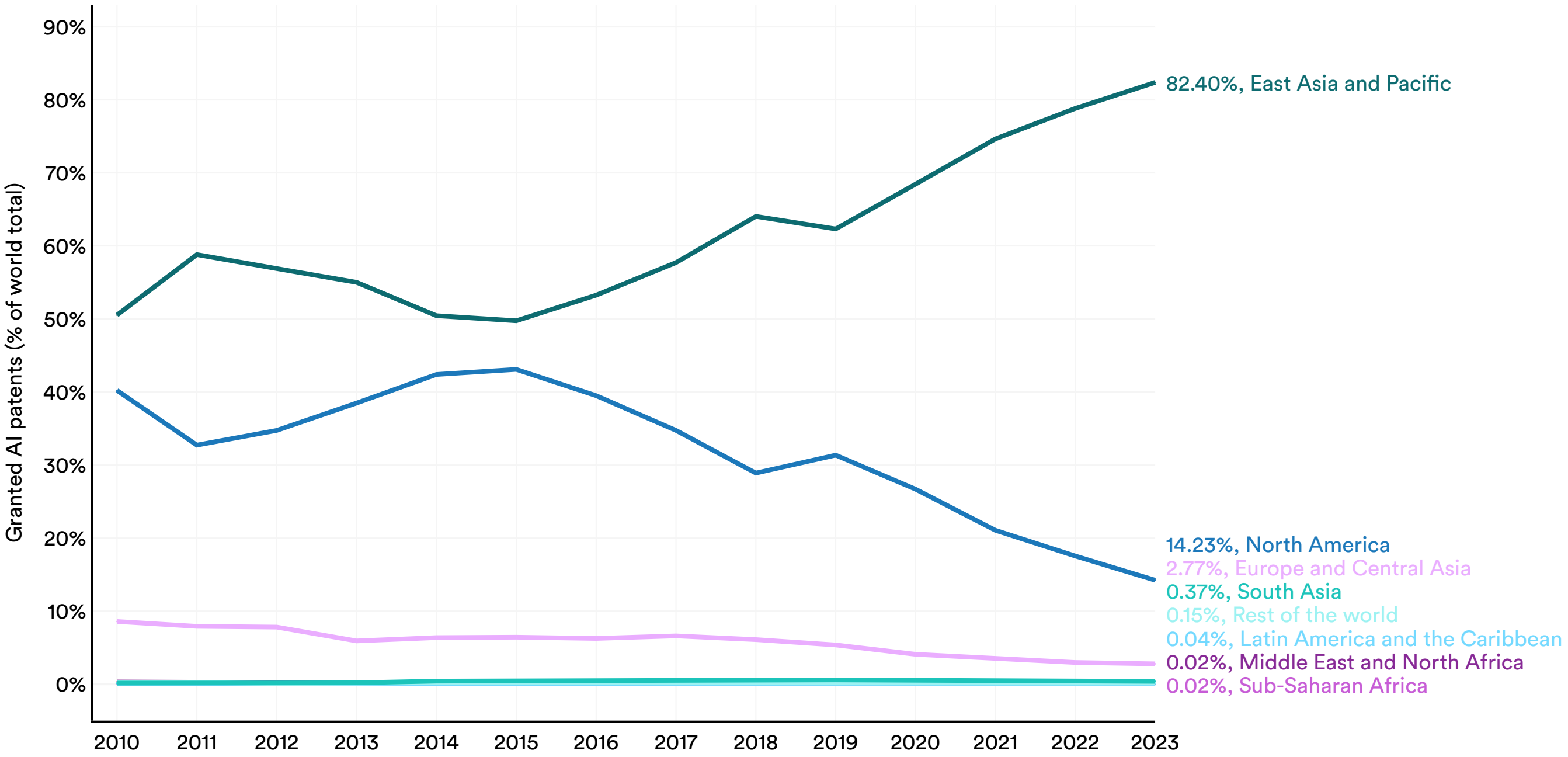

Figure 1.2.2[13]

13 Patent standards and laws vary across countries and regions, so these charts should be interpreted with caution. More detailed country-level patent information will be released in a subsequent edition of the AI Index's Global Vibrancy Tool.

Disaggregated by geographic area, the majority of the world's granted AI patents are from China (69.7%) and the United States (14.2%) (Figure 1.2.3). The share of AI patents originating from the United States has declined from a peak of 42.8% in 2015.

Figure 1.2.3 and Figure 1.2.4 document which countries lead in AI patents per capita. In 2023, the country with the most granted AI patents per 100,000 inhabitants was South Korea (17.3), followed by Luxembourg (15.3) and China (6.1) (Figure 1.2.3). Figure 1.2.5 highlights the change in granted AI patents per capita from 2013 to 2023. Luxembourg, China and Sweden experienced the greatest increase in AI patenting per capita during that time period.

**Granted AI patents (% of world total) by select geographic areas, 2010–23**
Source: AI Index, 2025 | Chart: 2025 AI Index report

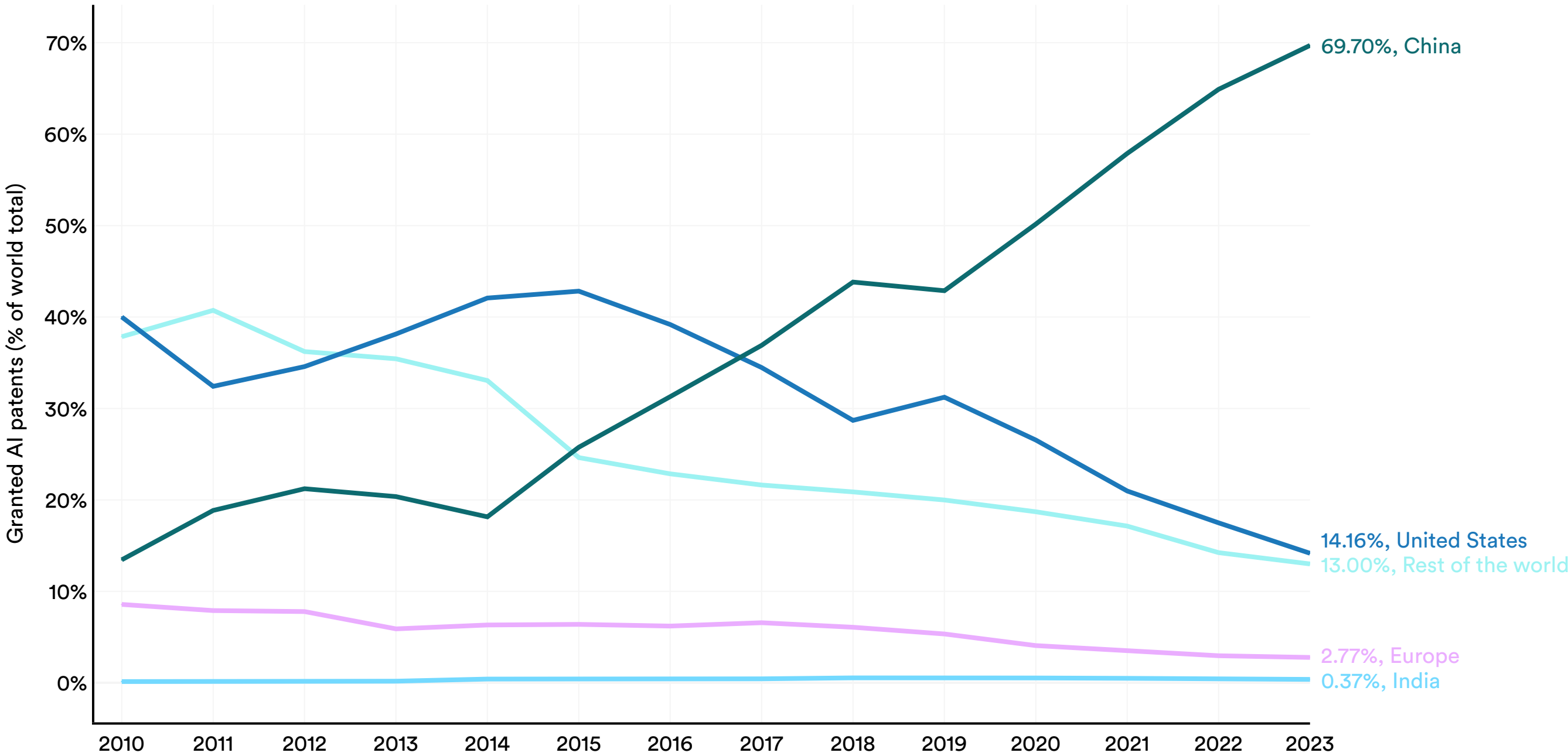

Figure 1.2.3

**Granted AI patents per 100,000 inhabitants by country, 2023**
Source: AI Index, 2025 | Chart: 2025 AI Index report

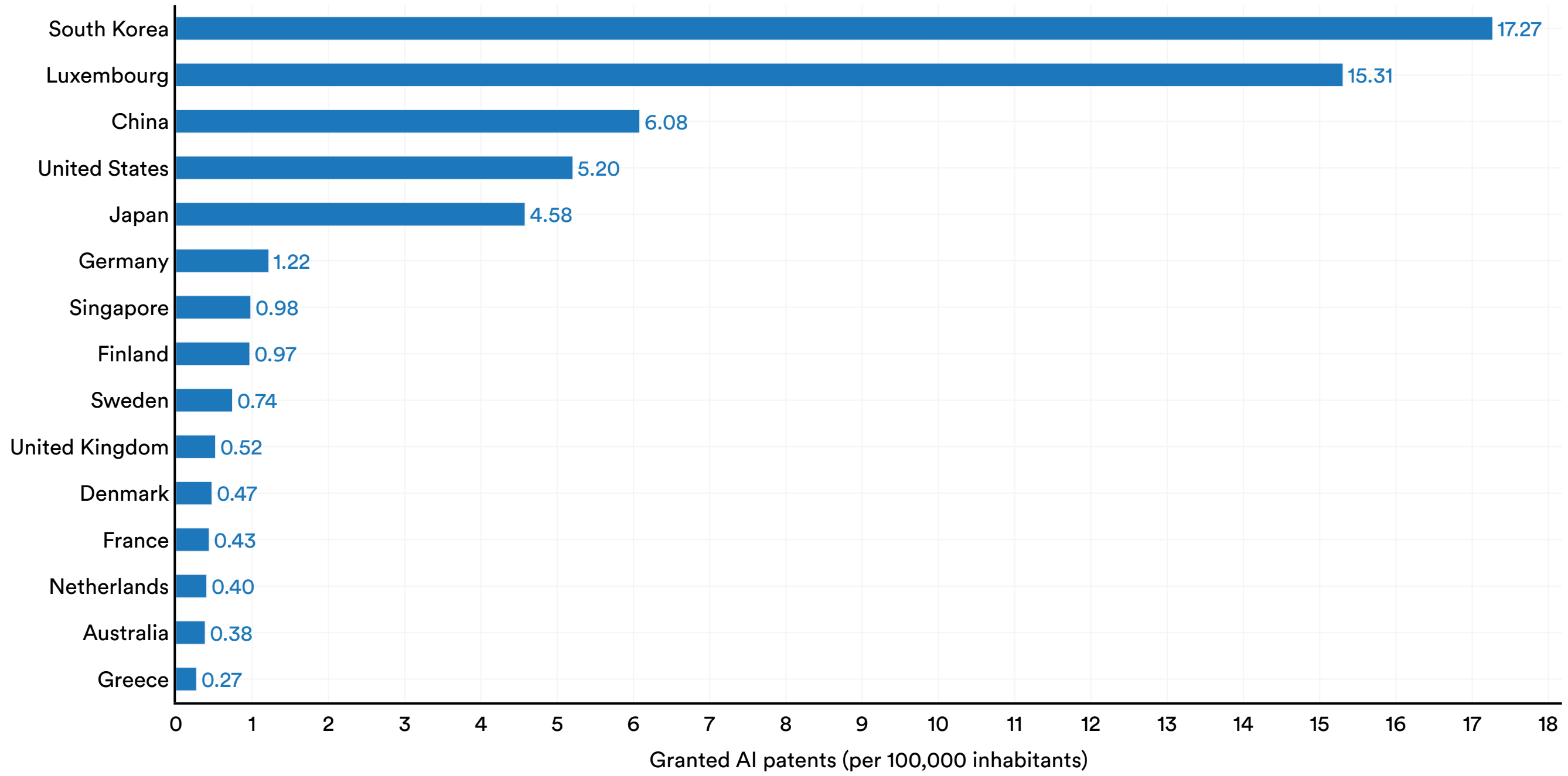

Figure 1.2.4

**Percentage change of granted AI patents per 100,000 inhabitants by country, 2013 vs. 2023**
Source: AI Index, 2025 | Chart: 2025 AI Index report

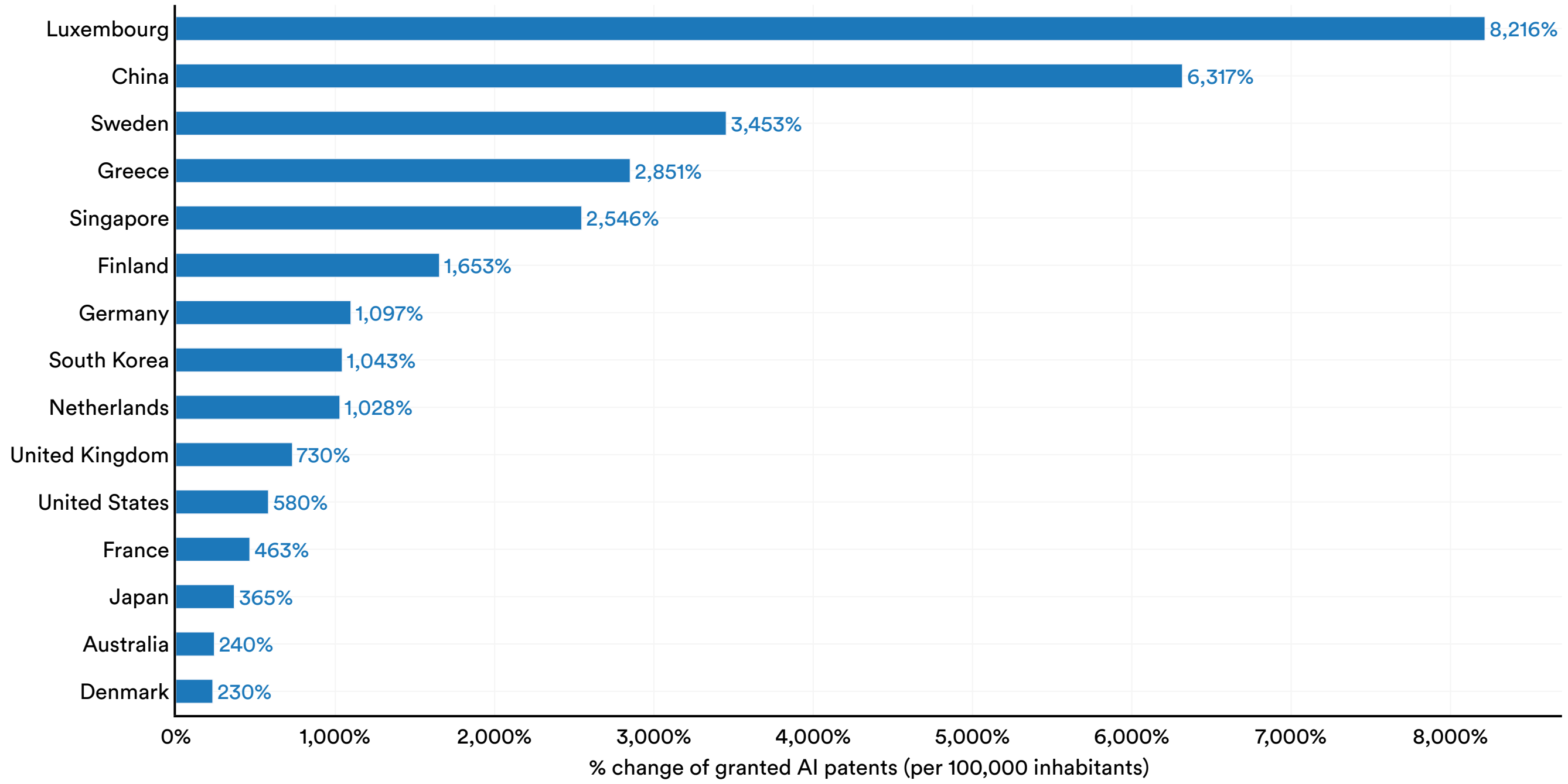

Figure 1.2.5

This section explores notable AI models.[14] Epoch AI, an AI Index data provider, uses the term "notable machine learning models" to designate particularly influential models within the AI/machine learning ecosystem. Epoch maintains a database of 900 AI models released since the 1950s, selecting entries based on criteria such as state-of-the-art advancements, historical significance, or high citation rates. Since Epoch manually curates the data, some models considered notable by some may not be included. Analyzing these models provides a comprehensive overview of the machine learning landscape's evolution, both in recent years and over the past few decades.[15] Some models may be missing from the dataset; however, the dataset can reveal trends in relative terms. Examples of notable AI models include GPT-4o, Claude 3.5, and AlphaGeometry.

Within this section, the AI Index explores trends in notable models from various perspectives, including country of origin, originating organization, gradient of model release, parameter count, and compute usage. The analysis concludes with an examination of machine learning training as well as inference costs.

# 1.3 Notable AI Models

## By National Affiliation

To illustrate the evolving geopolitical landscape of AI, the AI Index shows the country of origin of notable models. Figure 1.3.1 displays the total number of notable AI models attributed to the location of researchers' affiliated institutions.[16] In 2024, the United States led with 40 notable AI models, followed by China with 15 and France with three. All major geographic groups, including the United States, China, and Europe, reported releasing fewer notable models in 2024 than in the previous year (Figure 1.3.2). Since 2003, the United States has produced more models than other major countries such as the United Kingdom, China, and Canada (Figure 1.3.3).

It is difficult to pinpoint the exact cause of the decline in total model releases, but it may stem from a combination of factors: increasingly large training runs, the growing complexity of AI technology, and the heightened challenge of developing new modeling approaches. Epoch AI's curation of

**Number of notable AI models by select geographic areas, 2024**
Source: Epoch AI, 2025 | Chart: 2025 AI Index report

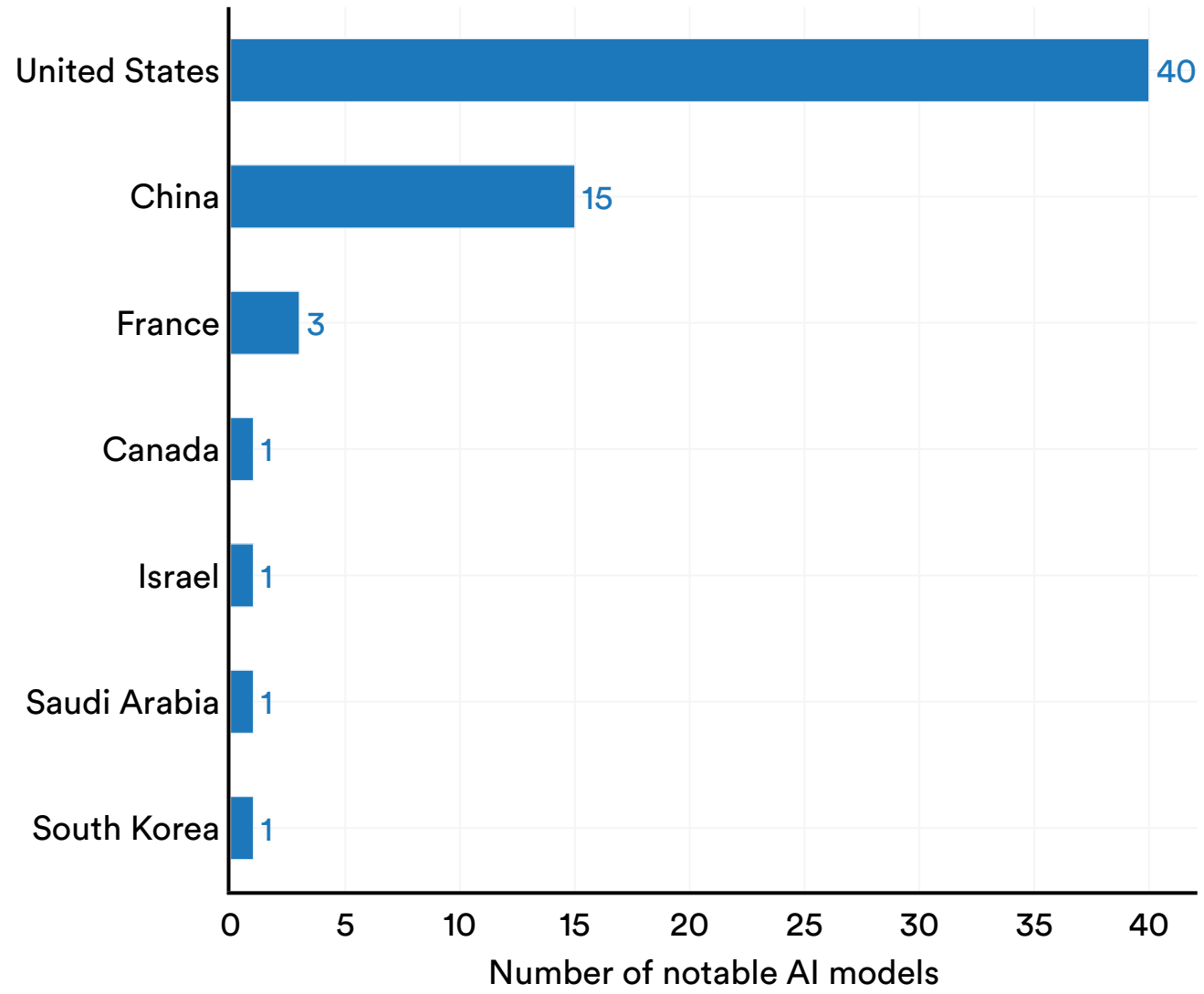

Figure 1.3.1[17]

**Number of notable AI models by select geographic areas, 2003–24**
Source: Epoch AI, 2025 | Chart: 2025 AI Index report

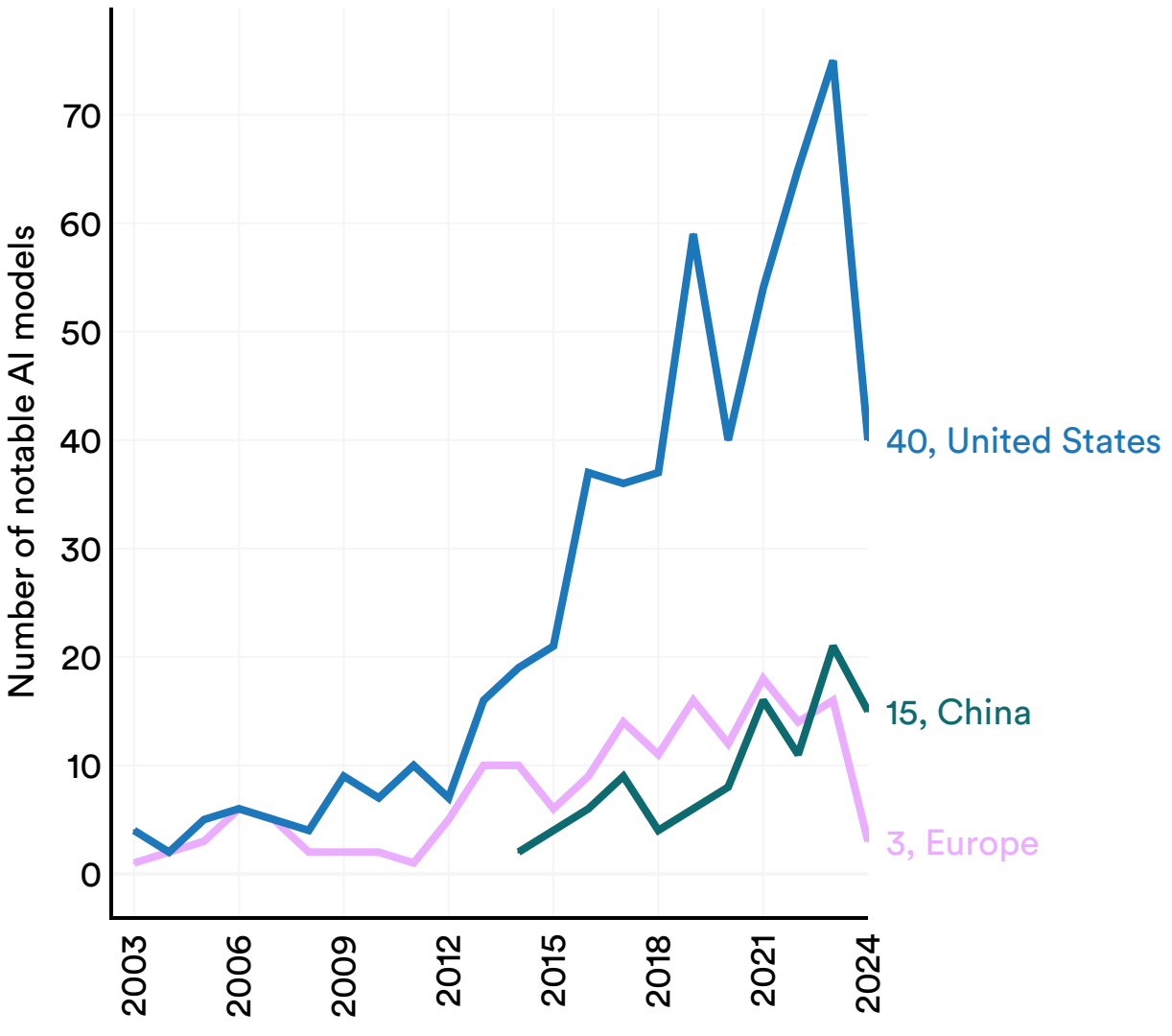

Figure 1.3.2

14 "AI system" refers to a computer program or product based on AI, such as ChatGPT. "AI model" includes a collection of parameters whose values are learned during training, such as GPT-4.

15 New and historic models are continually added to the Epoch AI database, so the total year-by-year counts of models included in this year's AI Index might not exactly match those published in last year's report. The data is from a snapshot taken on March 17, 2025.

16 A machine learning model is associated with a specific country if at least one author of the paper introducing it has an affiliation with an institution based in that country. In cases where a model's authors come from several countries, double-counting can occur.

17 This chart highlights model releases from a select group of geographic areas. More comprehensive data on model releases by country will be available in the upcoming AI Index Global Vibrancy Tool release.

notable models may overlook releases from certain countries that receive less coverage. The AI Index, in cooperation with Epoch, is committed to improving global representation in the AI model ecosystem. If readers believe that models from specific countries are missing, they are encouraged to contact the AI Index team, which will work to address the issue.

**Number of notable AI models by geographic area, 2003–24 (sum)**
Source: Epoch AI, 2025 | Chart: 2025 AI Index report

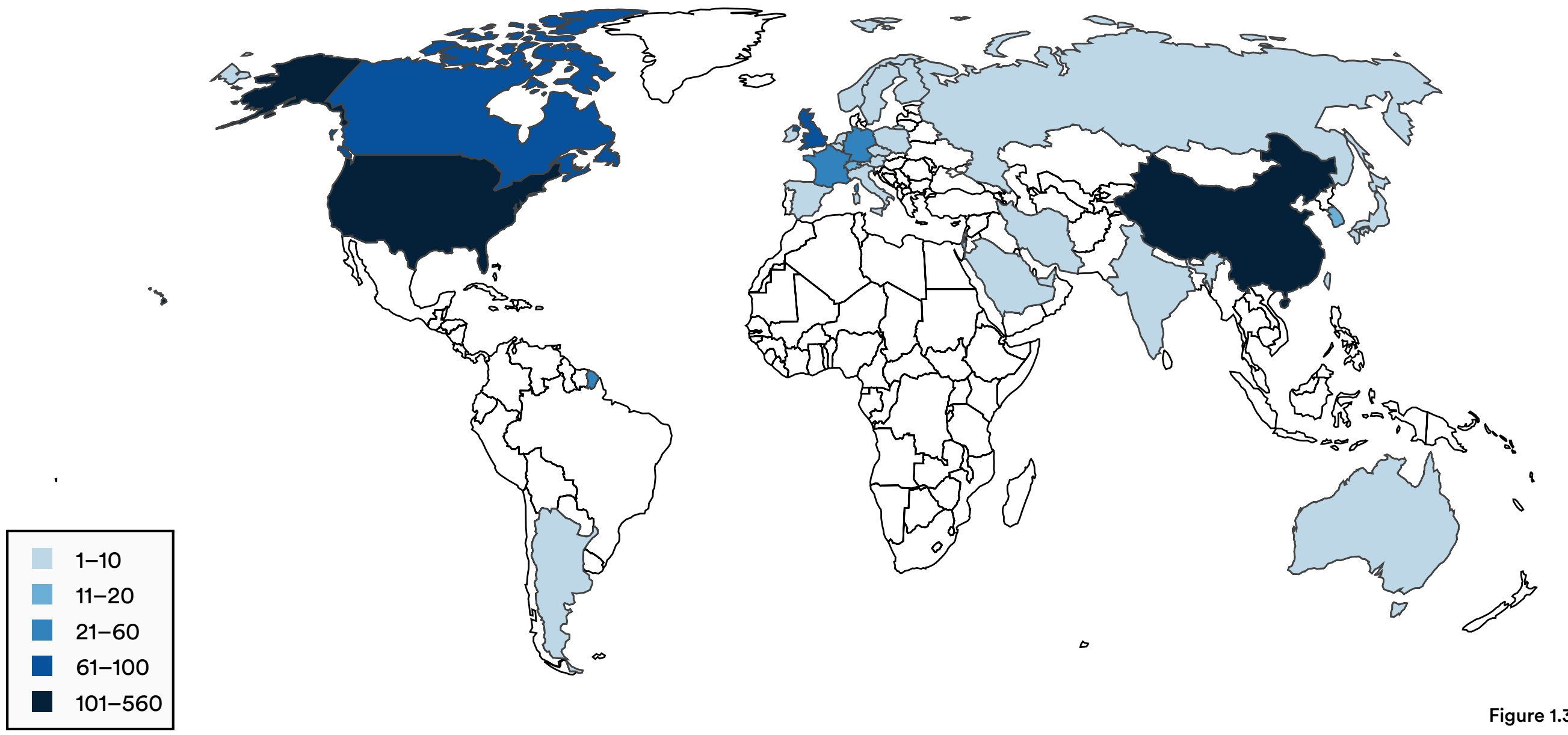

Figure 1.3.3

## By Sector

Figure 1.3.4 illustrates the sectoral origin of notable AI releases by the year the models were released. Epoch categorizes models based on their source: Industry includes companies such as Google, Meta, and OpenAI; academia covers universities like Tsinghua, MIT, and Oxford; government refers to state-affiliated research institutes like the UK's Alan Turing Institute for AI and Abu Dhabi's Technology Innovation Institute; and research collectives encompass nonprofit AI research organizations such as the Allen Institute for AI and the Fraunhofer Institute.

Until 2014, academia led in terms of releasing machine learning models. Since then, industry has taken the lead. According to Epoch AI, in 2024, industry produced 55 notable AI models. That same year, Epoch AI identified no notable AI models originating from academia (Figure 1.3.5).[18] Over time, industry-academia collaborations have contributed to a growing number of models. The proportion of notable AI models originating from industry has steadily increased over the past decade, growing to 90.2% in 2024.

18 This figure should be interpreted with caution. A count of zero academic models does not mean that no notable models were produced by academic institutions in 2023, but rather that Epoch AI has not identified any as notable. Additionally, academic publications often take longer to gain recognition, as highly cited papers introducing significant architectures may take years to achieve prominence.

## Number of notable AI models by sector, 2003–24

Source: Epoch AI, 2025 | Chart: 2025 AI Index report

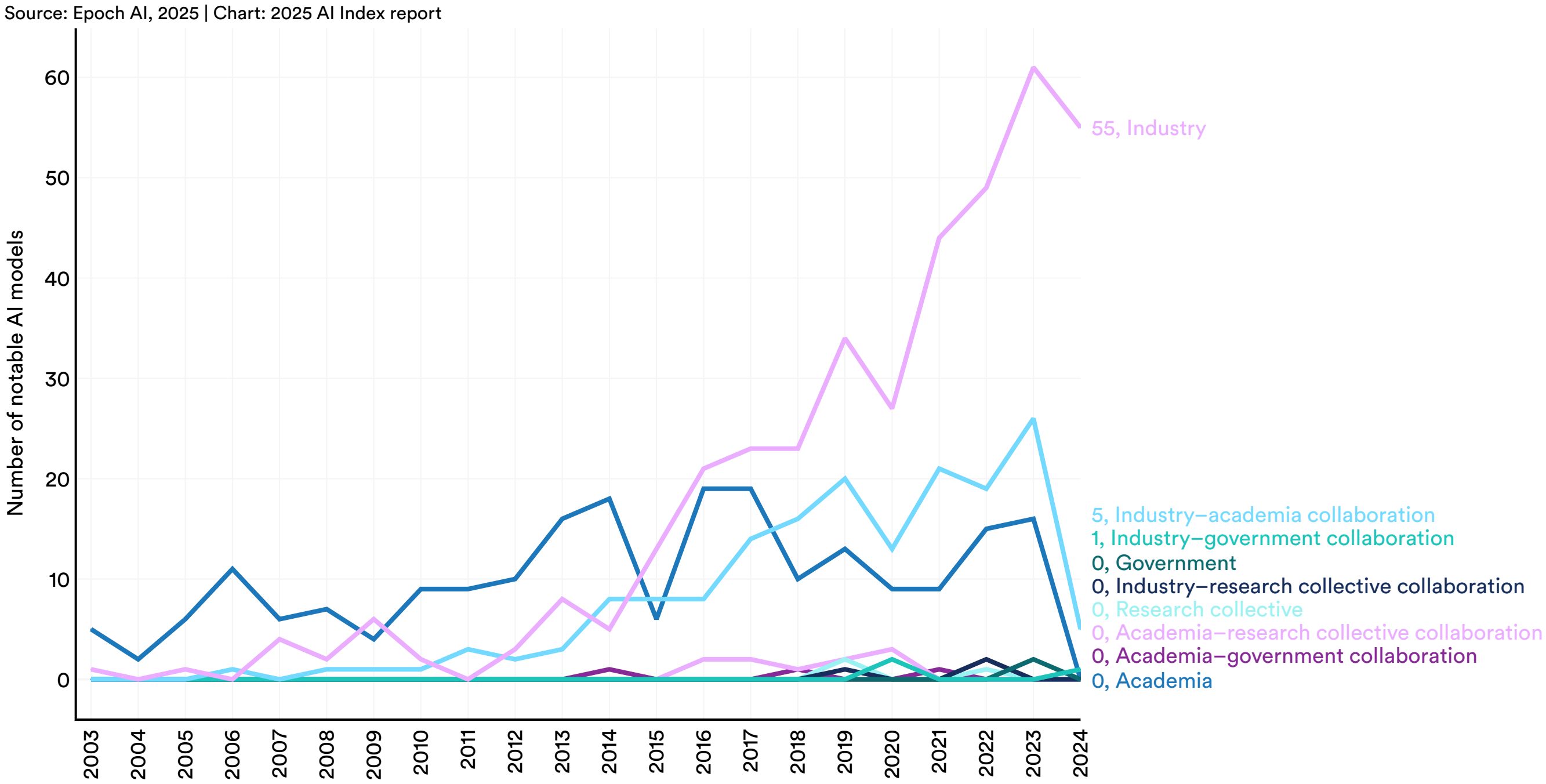

Figure 1.3.4

## Notable AI models (% of total) by sector, 2003–24

Source: Epoch AI, 2025 | Chart: 2025 AI Index report

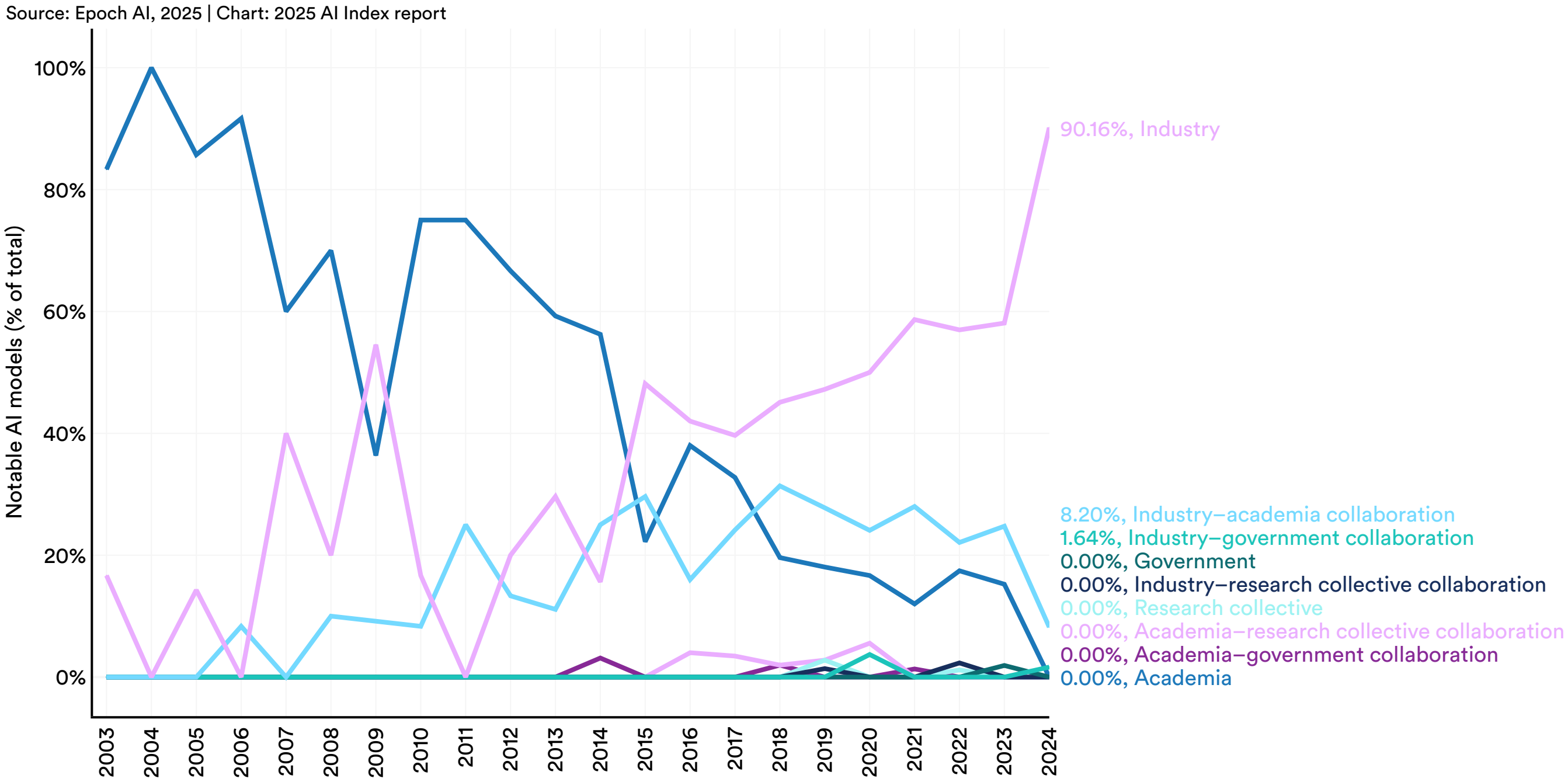

Figure 1.3.5

## By Organization

Figure 1.3.6 and Figure 1.3.7 highlight the organizations leading in the production of notable machine learning models in 2024 and over the past decade. In 2024, the top contributors were Google (7), OpenAI (7 models), and Alibaba (6). Since 2014, Google has led with 187 notable models, followed by Meta (82) and Microsoft (39). Among academic institutions, Carnegie Mellon University (25), Stanford University (25), and Tsinghua University (22) have been the most prolific since 2014.

**Number of notable AI models by organization, 2024**
Source: Epoch AI, 2025 | Chart: 2025 AI Index report

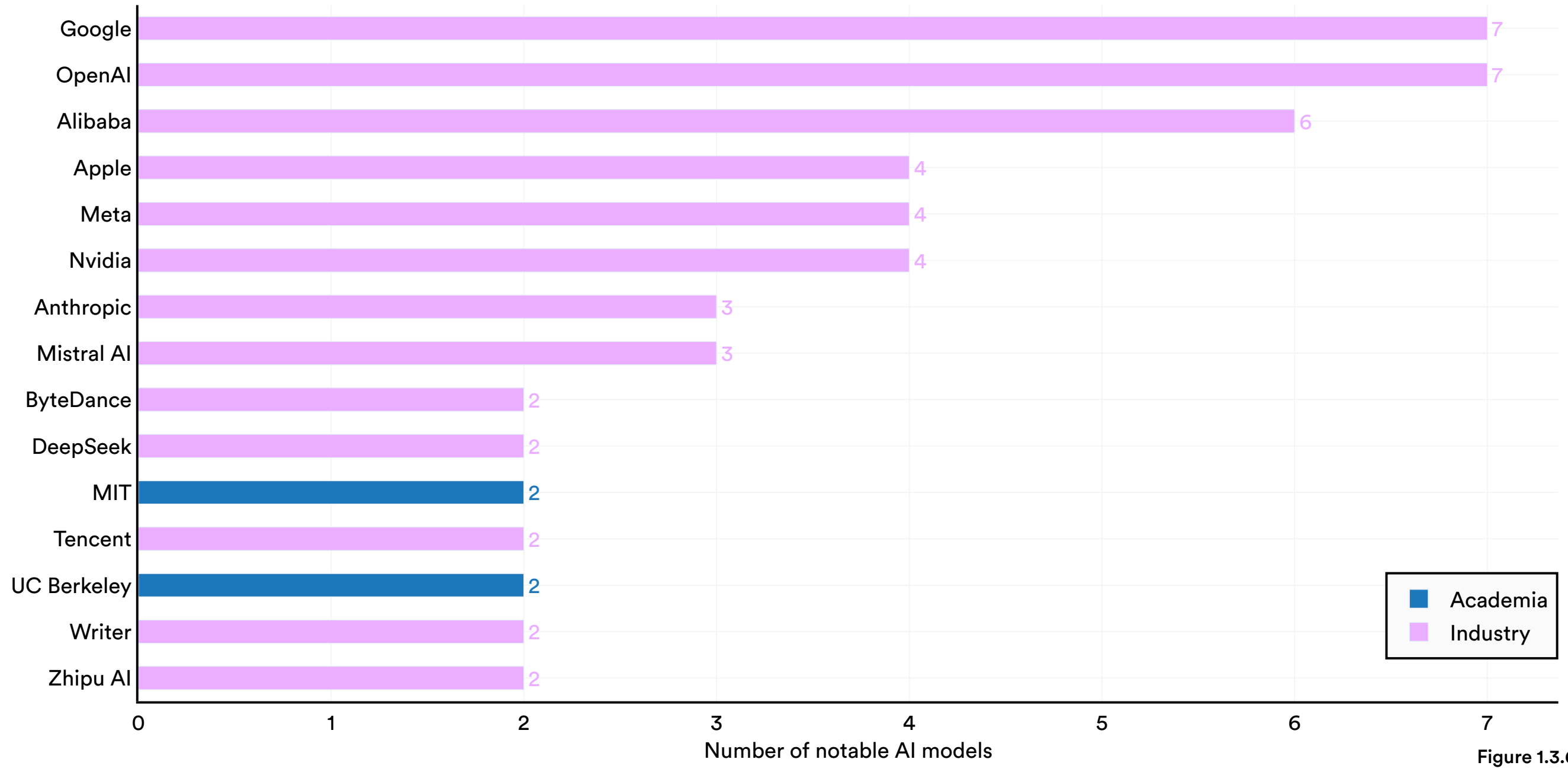

Figure 1.3.6[19]

**Number of notable AI models by organization, 2014–24 (sum)**
Source: Epoch AI, 2025 | Chart: 2025 AI Index report

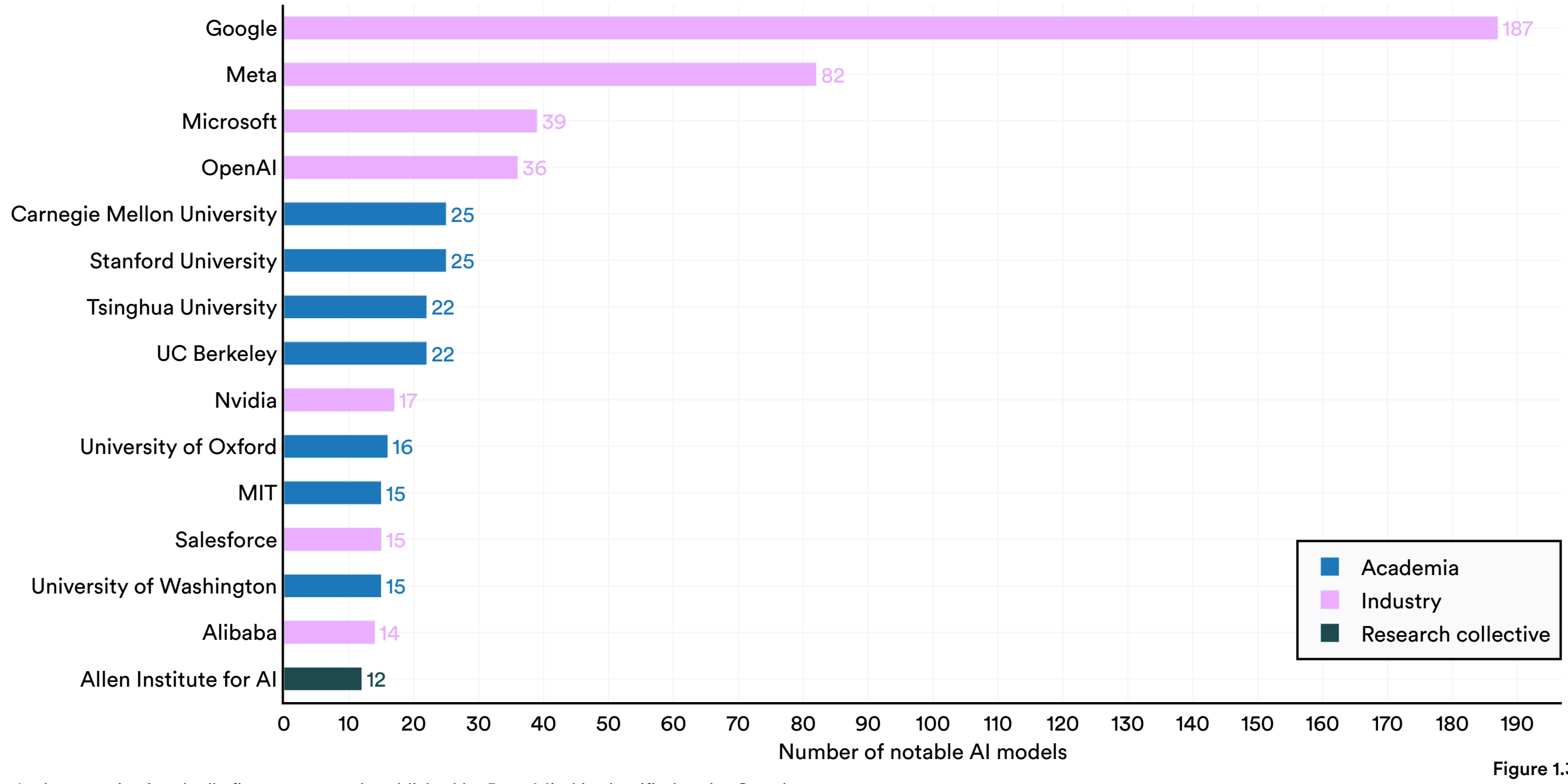

Figure 1.3.7

19 In the organizational tally figures, research published by DeepMind is classified under Google.

## Model Release

Machine learning models are released under various access types, each with varying levels of openness and usability. API access models, like OpenAI's o1, allow users to interact with models via queries without direct access to their underlying weights. Open weights (restricted use) models, like DeepSeek's-V3, provide access to their weights but impose limitations, such as prohibiting commercial use or redistribution. Hosted access (no API) models, like Gemini 2.0 Pro, refer to models available through a platform interface but without programmatic access. Open weights (unrestricted) models, like AlphaGeometry, are fully open, allowing free use, modification, and redistribution. Open weights (noncommercial) models, like Mistral Large 2, share their weights but restrict use to research or noncommercial purposes. Lastly, unreleased models, like ESM3 98B, remain proprietary, accessible only to their developers or select partners. The unknown designation refers to models that have unclear or undisclosed access types.

Figure 1.3.8 illustrates the different access types under which models have been released.[20] In 2024, API access was the most common release type, with 20 of 61 models made available this way, followed by open weights with restricted use and unreleased models.

Figure 1.3.9 visualizes machine learning model access types over time from a proportional perspective. In 2024, most AI models were released via API access (32.8%), which has seen a steady rise since 2020.

### Number of notable AI models by access type, 2014–24

Source: Epoch AI, 2025 | Chart: 2025 AI Index report

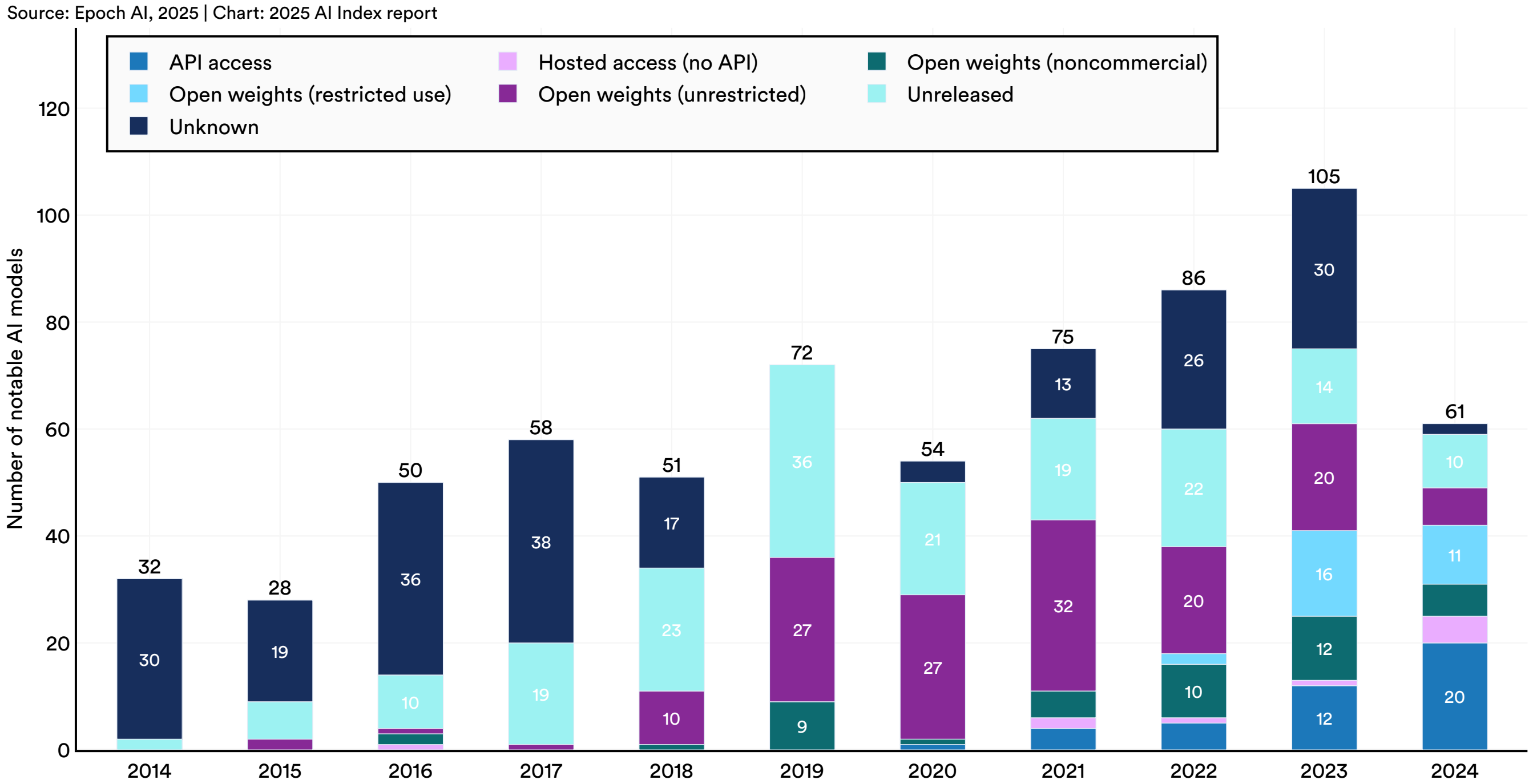

Figure 1.3.8[21]

20 Hosted access refers to using computing resources or services (such as software, hardware, or storage) provided remotely by a third party, rather than personally owning or managing them. Instead of running software or infrastructure locally, hosted access involves accessing these resources via the cloud or another remote service, typically over the internet. For example, using GPUs through platforms like AWS, Google Cloud, or Microsoft Azure—rather than running them on one's own hardware—is considered hosted access.

21 Not all models in the Epoch database are categorized by access type, so the totals in Figures 1.3.8 through 1.3.10 may not fully align with those reported elsewhere in the chapter.

## Notable AI models (% of total) by access type, 2014–24

Source: Epoch AI, 2025 | Chart: 2025 AI Index report

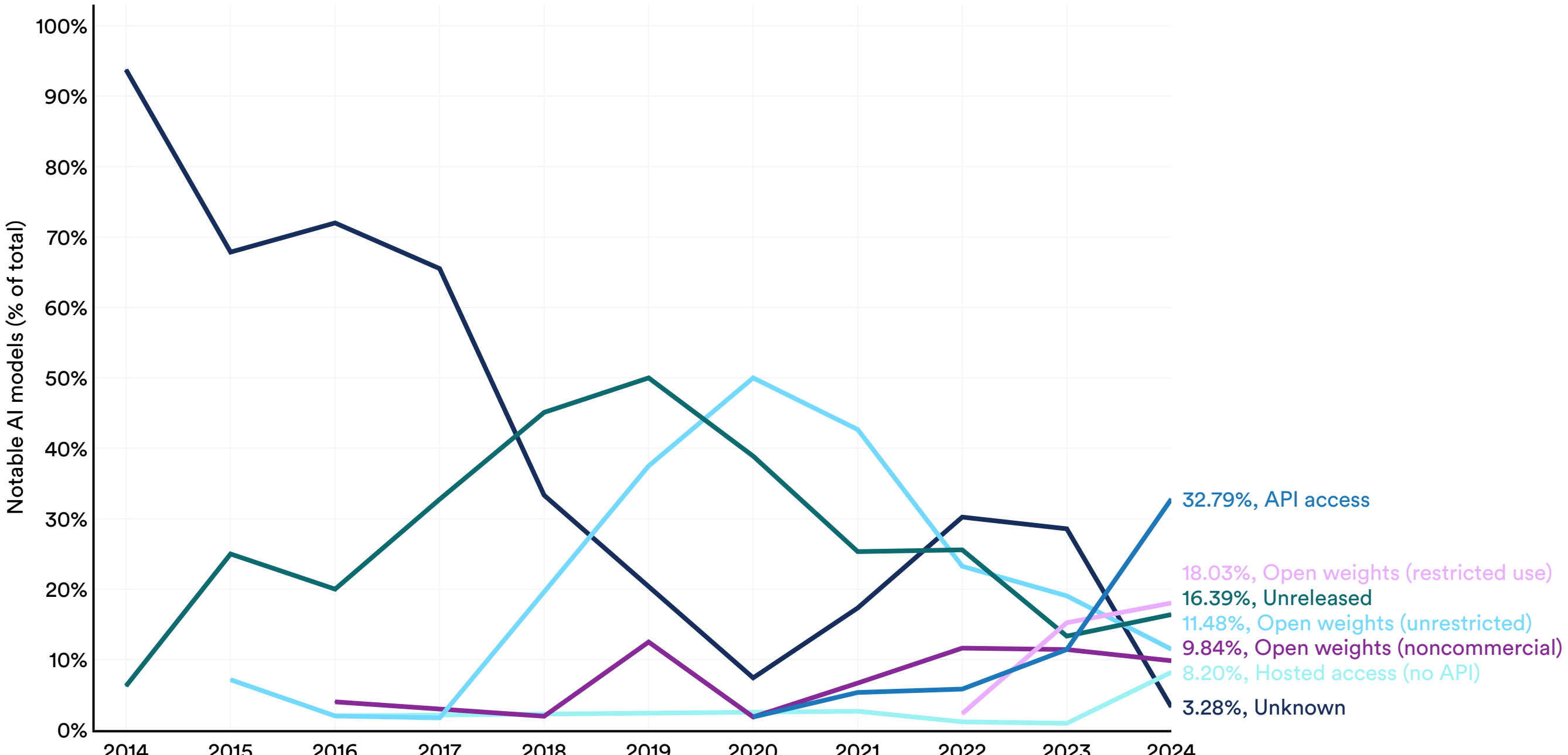

Figure 1.3.9

In traditional open-source software releases, all components, including the training code, are typically made available. However, this is often not the case with AI technologies, where even developers who release model weights may withhold the training code. Figure 1.3.10 categorizes notable AI models by the openness of their code release. In 2024, the majority—60.7%—were launched without corresponding training code.

## Number of notable AI models by training code access type, 2014–24

Source: Epoch AI, 2025 | Chart: 2025 AI Index report

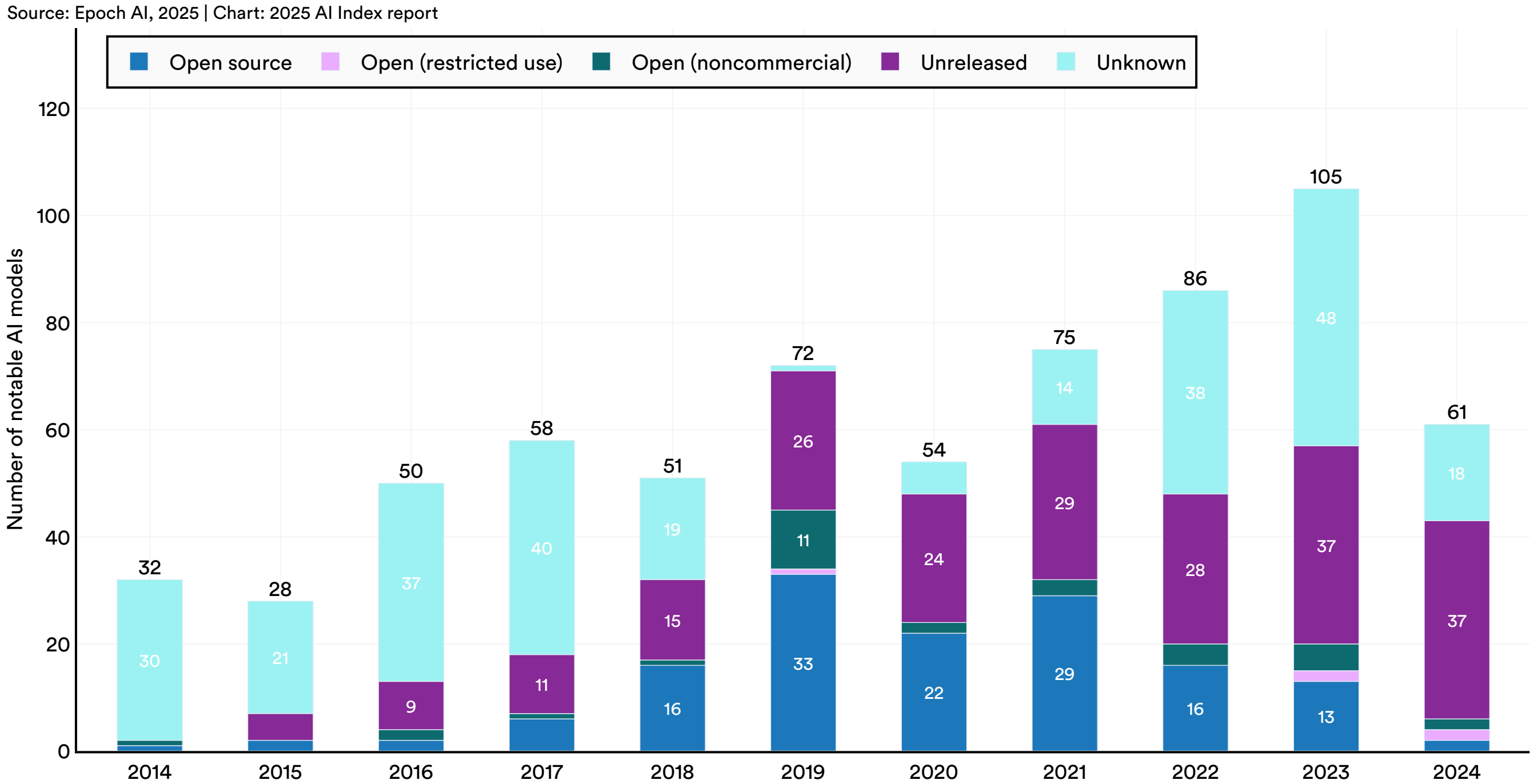

Figure 1.3.10

## Parameter Trends

Parameters in machine learning models are numerical values learned during training that determine how a model interprets input data and makes predictions. Models with more parameters require more data to be trained, but they can take on more tasks and typically outperform models with fewer parameters.

Figure 1.3.11 demonstrates the parameter count of machine learning models in the Epoch dataset, categorized by the sector from which the models originate. Figure 1.3.12 visualizes the same data, but for a smaller selection of notable models. Parameter counts have risen sharply since the early 2010s, reflecting the growing complexity of their architecture, greater availability of data, improvements in hardware, and proven efficacy of larger models. High-parameter models are particularly notable in the industry sector, underscoring the substantial financial resources available to industry to cover the computational costs of training on vast volumes of data. Several of the figures below use a log scale to reflect the exponential growth in AI model parameters and compute in recent years.

**Number of parameters of notable AI models by sector, 2003–24**
Source: Epoch AI, 2025 | Chart: 2025 AI Index report

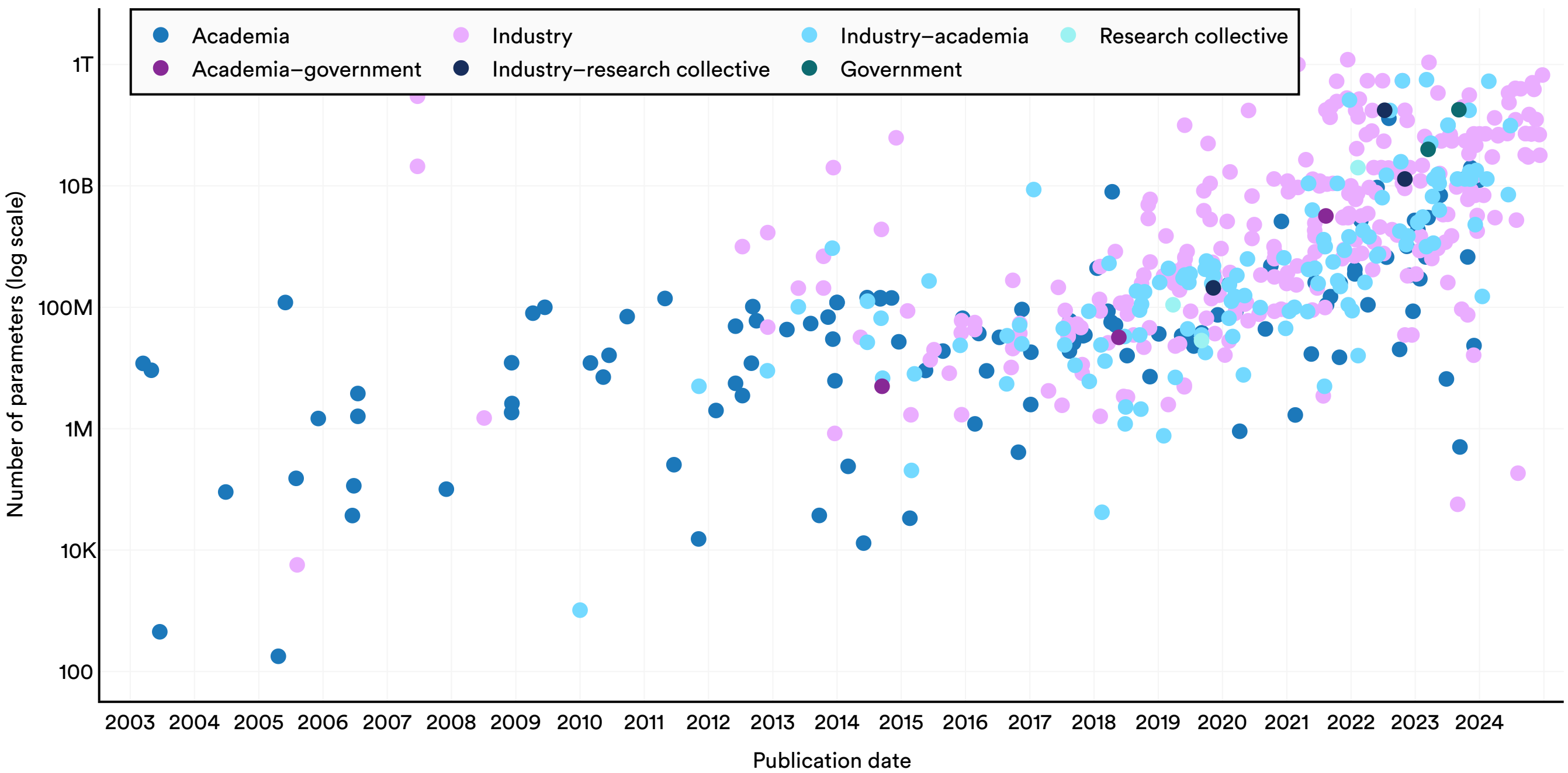

Figure 1.3.11

## Number of parameters of select notable AI models by sector, 2012–24

Source: Epoch AI, 2025 | Chart: 2025 AI Index report

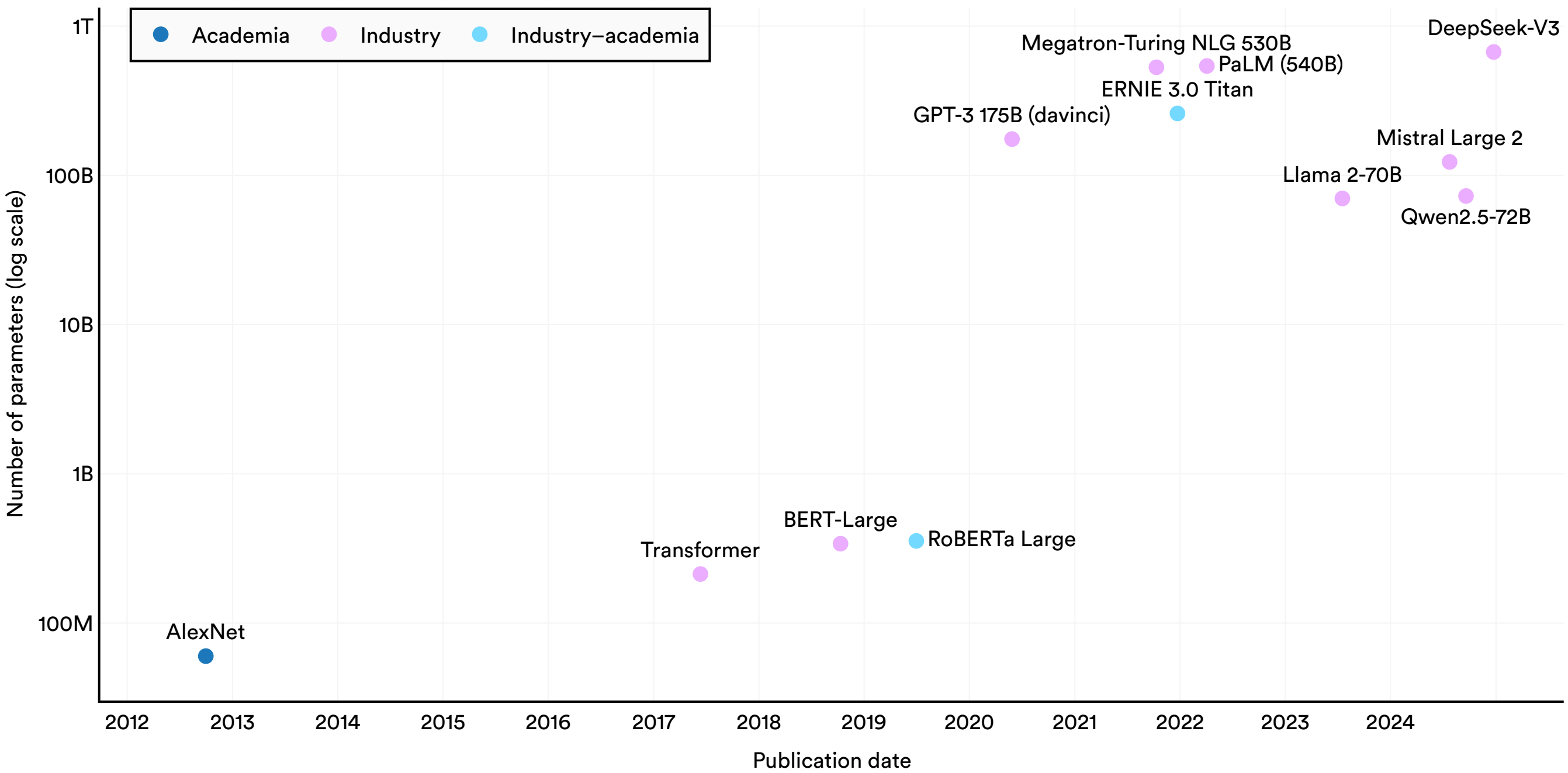

Figure 1.3.12

As model parameter counts have increased, so has the volume of data used to train AI systems. Figure 1.3.13 illustrates the growth in dataset sizes used to train notable machine learning models. The Transformer model, released in 2017 and widely credited with sparking the large language model revolution, was trained on approximately 2 billion tokens. By 2020, GPT-3 175B—one of the models underpinning the original ChatGPT—was trained on an estimated 374 billion tokens. In contrast, Meta's flagship LLM, Llama 3.3, released in the summer of 2024, was trained on roughly 15 trillion tokens. According to Epoch AI, LLM training datasets double in size approximately every eight months.

**Training dataset size of notable AI models, 2010–24**
Source: Epoch AI, 2025 | Chart: 2025 AI Index report

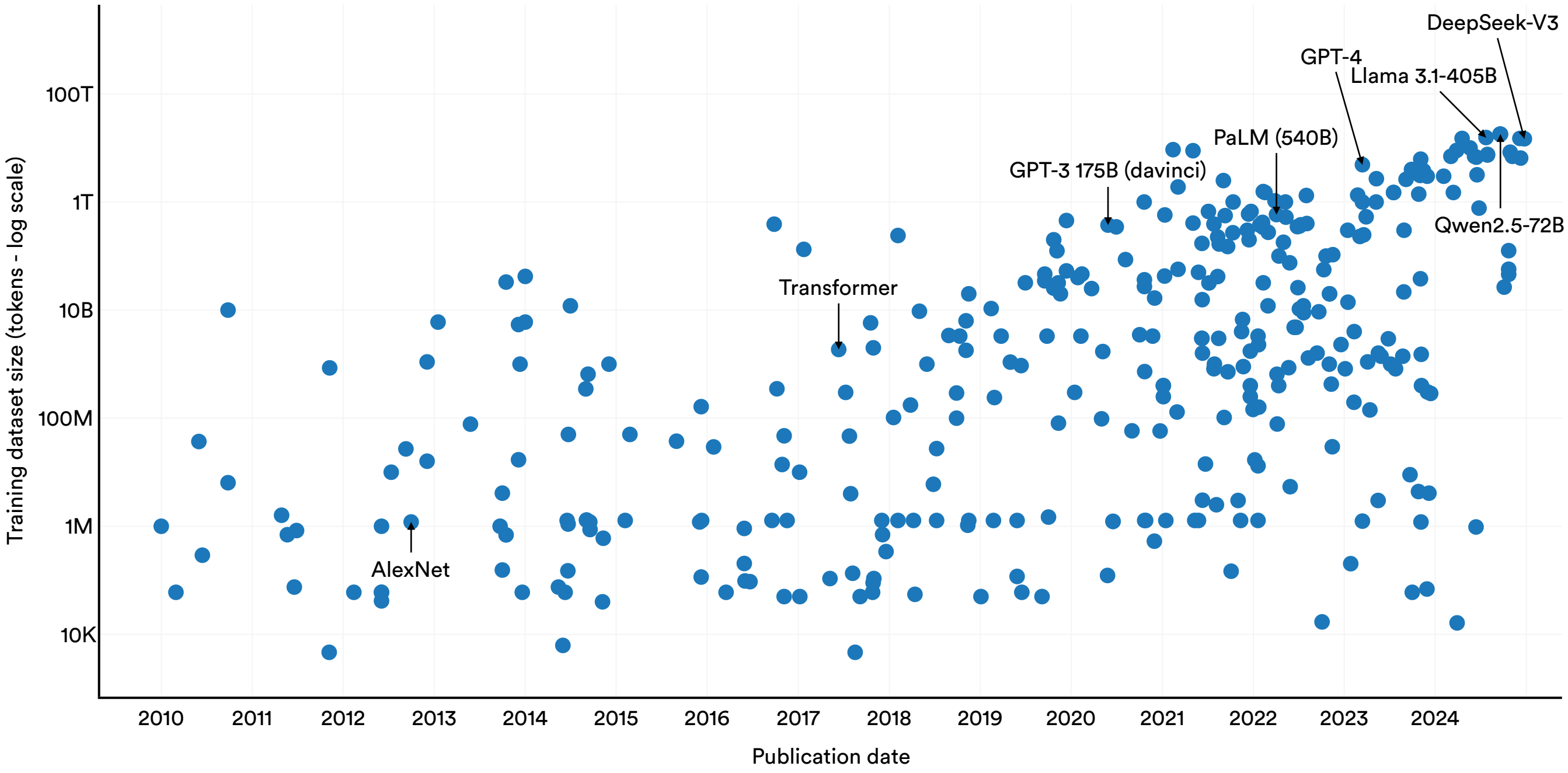

Figure 1.3.13

Training models on increasingly large datasets has led to significantly longer training times (Figure 1.3.14). Some state-of-the-art models, such as Llama 3.1-405B, required approximately 90 days to train—a typical window by today's standards. Google's Gemini 1.0 Ultra, released in late 2023, took around 100 days. This stands in stark contrast to AlexNet, one of the first models to leverage GPUs for enhanced performance, which trained in just five to six days in 2012. Notably, AlexNet was trained on far less advanced hardware.

**Training length of notable AI models, 2010–24**

Source: Epoch AI, 2025 | Chart: 2025 AI Index report

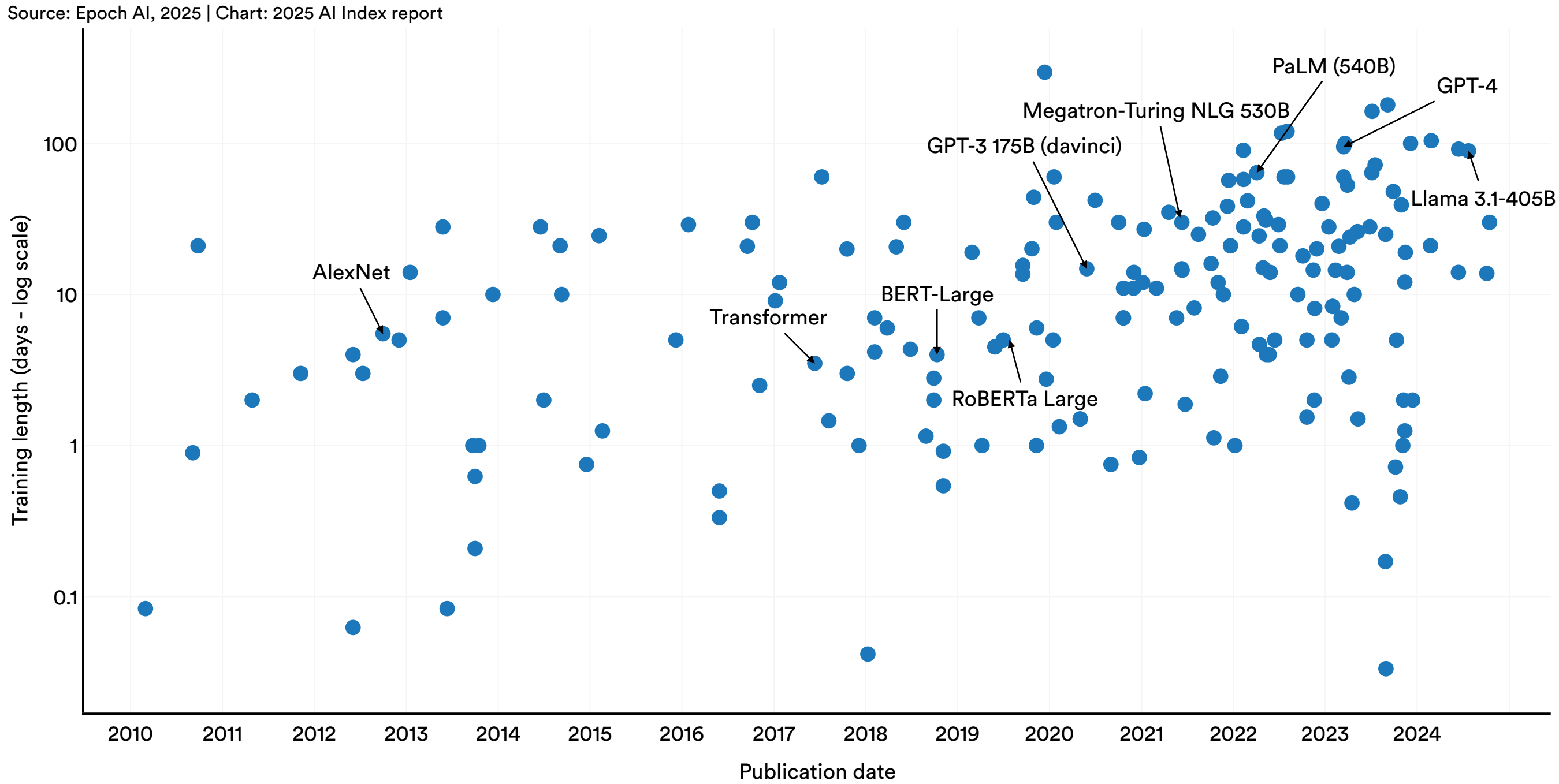

Figure 1.3.14

## Compute Trends

The term "compute" in AI models denotes the computational resources required to train and operate a machine learning model. Generally, the complexity of the model and the size of the training dataset directly influence the amount of compute needed. The more complex a model is, and the larger the underlying training data, the greater the amount of compute required for training. Before the final training run, researchers conduct numerous test runs throughout the R&D phase. While training a single model is relatively inexpensive, the cumulative cost of multiple R&D runs and the necessary datasets quickly becomes significant. These figures reflect only the final training run, not the entire R&D process.

Figure 1.3.15 visualizes the training compute required for notable machine learning models over the past 22 years. Recently, the compute usage of notable AI models has increased exponentially.[22] Epoch estimates that the training compute of notable AI models doubles roughly every five months. This trend has been especially pronounced in the last five years. This rapid rise in compute demand has important implications. For instance, models requiring more computation often have larger environmental footprints, and companies typically have more access to computational resources than academic institutions. For reference, Chapter 2 of the AI Index analyzes the relationship between improvements in computational resources and model performance.

**Training compute of notable AI models by sector, 2003–24**
Source: Epoch AI, 2025 | Chart: 2025 AI Index report

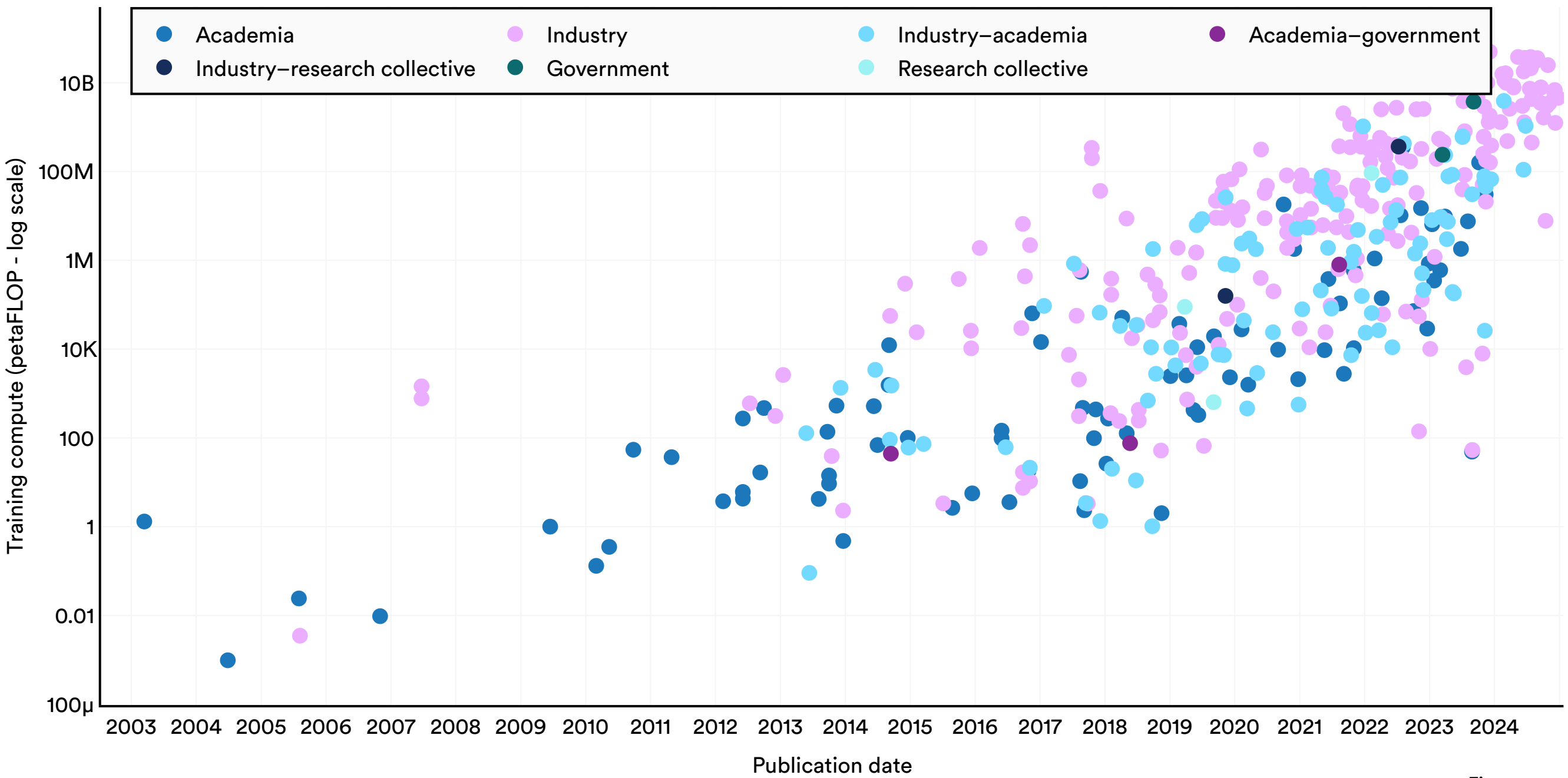

Figure 1.3.15[23]

22 FLOP stands for "floating-point operation." A floating-point operation is a single arithmetic operation involving floating-point numbers, such as addition, subtraction, multiplication, or division. The number of FLOP a processor or computer can perform per second is an indicator of its computational power. The higher the FLOP rate, the more powerful the computer. The number of floating-point operations used to train an AI model reflects its requirement for computational resources during development.

23 Estimating training compute is an important aspect of AI model analysis, yet it often requires indirect measurement. When direct reporting is unavailable, Epoch estimates compute by using hardware specifications and usage patterns or by counting arithmetic operations based on model architecture and training data. In cases where neither approach is feasible, benchmark performance can serve as a proxy to infer training compute by comparing models with known compute values. Full details of Epoch's methodology can be found in the documentation section of their website.

Figure 1.3.16 highlights the training compute of notable machine learning models since 2012. For example, AlexNet, one of the models that popularized the now standard practice of using GPUs to improve AI models, required an estimated 470 petaFLOP for training.[24] The original Transformer, released in 2017, required around 7,400 petaFLOP. OpenAI's GPT-4o, one of the current state-of-the-art foundation models, required 38 billion petaFLOP. Creating cutting-edge AI models now demands a colossal amount of data, computing power, and financial resources that are not available to academia. Most leading AI models are coming from industry, a trend that was first highlighted in last year's AI Index. Although the gap has slightly narrowed this year, the trend persists.

**Training compute of notable AI models by domain, 2012–24**

Source: Epoch AI, 2025 | Chart: 2025 AI Index report

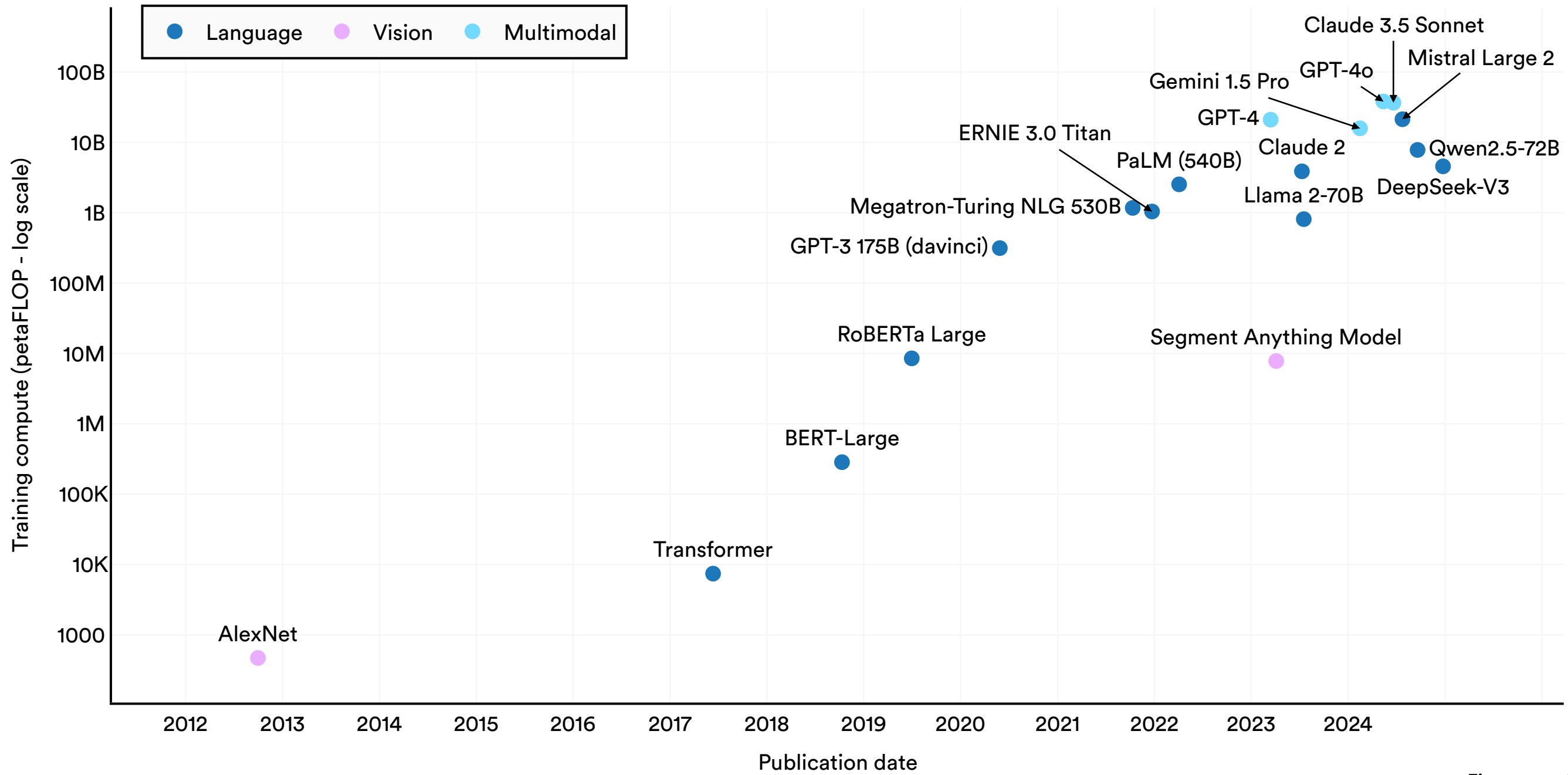

Figure 1.3.16

24 A petaFLOP (PFLOP) is a unit of computing power equal to one quadrillion ($10^{15}$) floating-point operations per second.

The launch of DeepSeek's V3 model in December 2024 garnered significant attention, particularly because it achieved exceptionally high performance while requiring far fewer computational resources than many leading LLMs. Figure 1.3.17 compares the training compute of notable machine learning models from the United States and China, highlighting a key trend: Top-tier AI models from the U.S. have generally been far more computationally intensive than Chinese models. According to Epoch AI, the top 10 Chinese language models by training compute have scaled at a rate of about three times per year since late 2021—considerably slower than the five times per year trend observed in the rest of the world since 2018.

**Training compute of select notable AI models in the United States and China, 2018–24**

Source: Epoch AI, 2025 | Chart: 2025 AI Index report

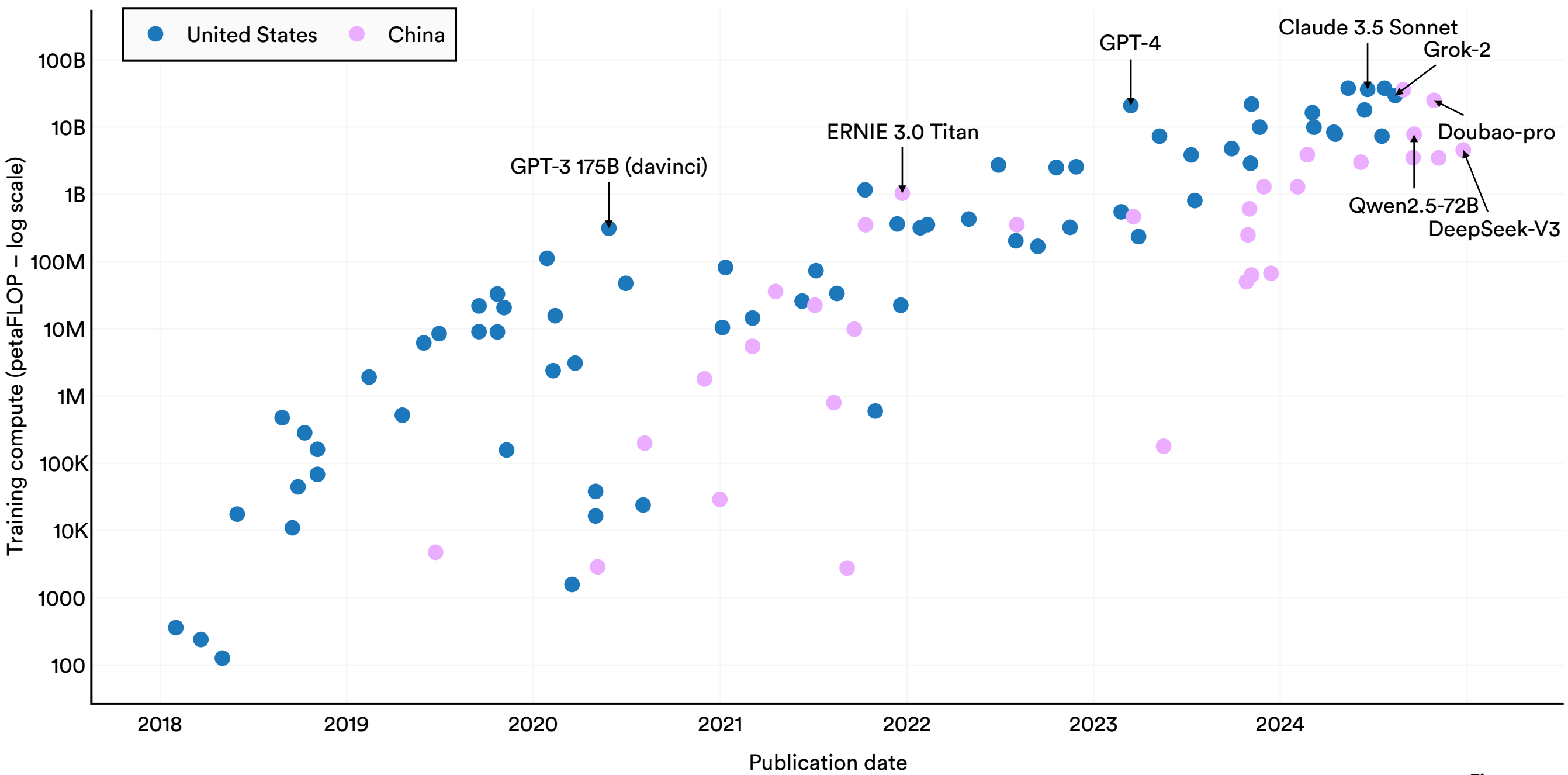

Figure 1.3.17

**Highlight:**

## Will Models Run Out of Data?

One of the key drivers of substantive algorithmic improvements in AI systems has been the scaling of models and their training on ever-larger datasets. However, as the supply of internet training data becomes increasingly depleted, concerns have grown about the sustainability of this scaling approach and the potential for a data bottleneck, where returns to scale diminish. Last year's AI Index explored various factors in this debate, including the availability of existing internet data and the potential for training models on synthetic data. New research this year suggests that the current stock of data may last longer than previously expected.

Epoch AI has updated its previous estimates for when AI researchers might run out of data. In its latest research, the team estimated the total effective stock of data available for training models according to token count (Figure 1.3.18). Common Crawl, an open repository of web crawl data frequently used in AI training, is estimated to contain a median of 130 trillion tokens. The indexed web holds approximately 510 trillion tokens, while the entire web contains around 3,100 trillion. Additionally, the total stock of images is estimated at 300 trillion, and video at 1,350 trillion.

**Estimated median data stocks**

Source: Epoch AI, 2025 | Chart: 2025 AI Index report

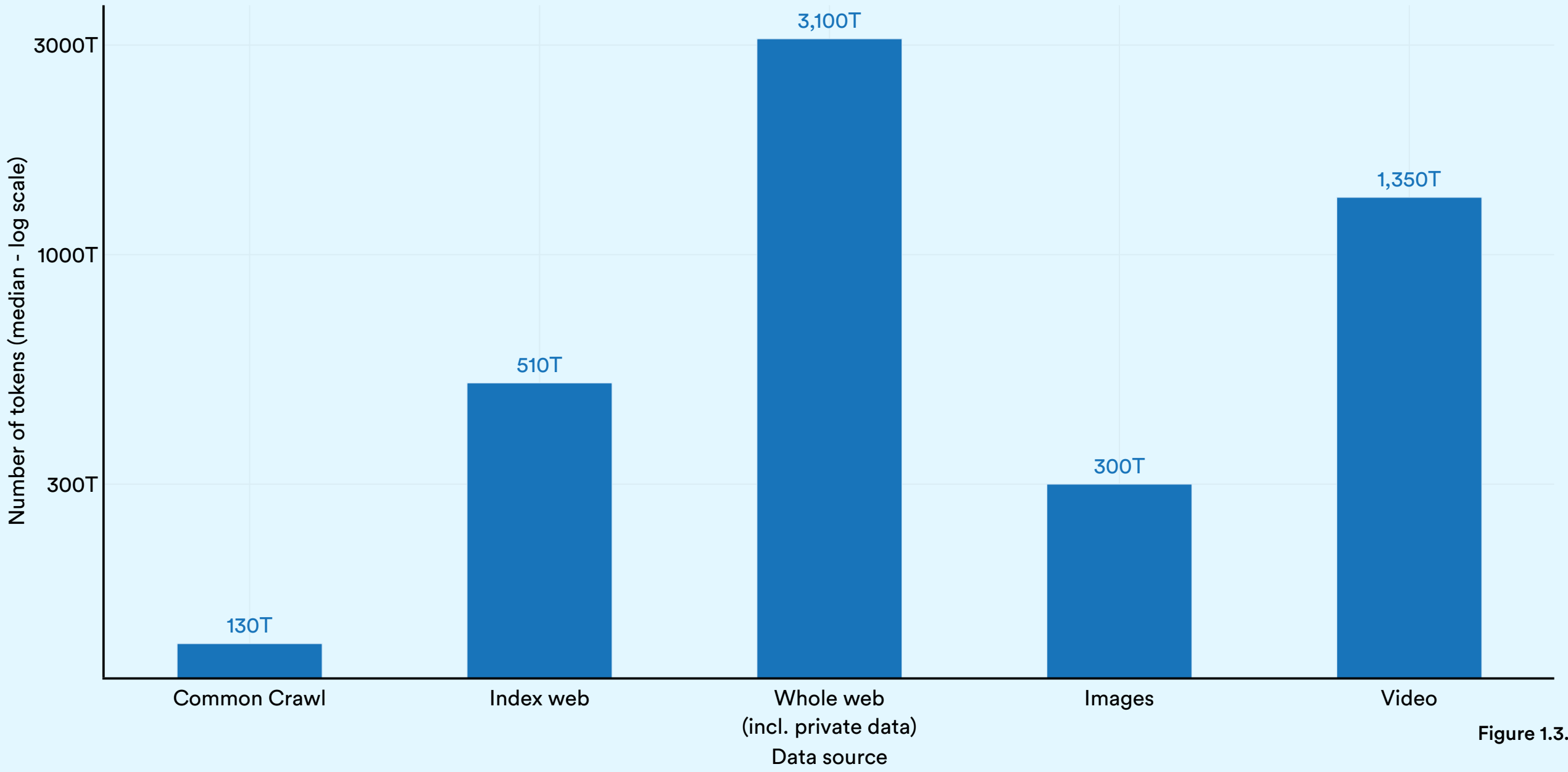

Figure 1.3.18

**Highlight:**

# Will Models Run Out of Data? (cont'd)

The Epoch AI research team projects, with an 80% confidence interval, that the current stock of training data will be fully utilized between 2026 and 2032 (Figure 1.3.19). Several factors influence the point in time when data is likely to run out. One key factor is the historical growth of dataset sizes, which depends on how many people generate and contribute content to the internet. Another important factor is computer usage. If models are trained in a compute-optimal manner, the available data stock can last longer. However, if models are overtrained to achieve more compute-efficient inference performance, the stock is likely to be depleted sooner. When AI models are overtrained, meaning they are trained for an extended period beyond the typical point of diminishing returns, they may achieve more compute-efficient inference—that is, they can process prompts (make predictions, generate text, etc.) using less computational power. However, this comes at a cost: The stock (i.e., data available to train the model) may be depleted more quickly.

**Projections of the stock of public text and data usage**
Source: Epoch AI, 2025 | Chart: 2025 AI Index report

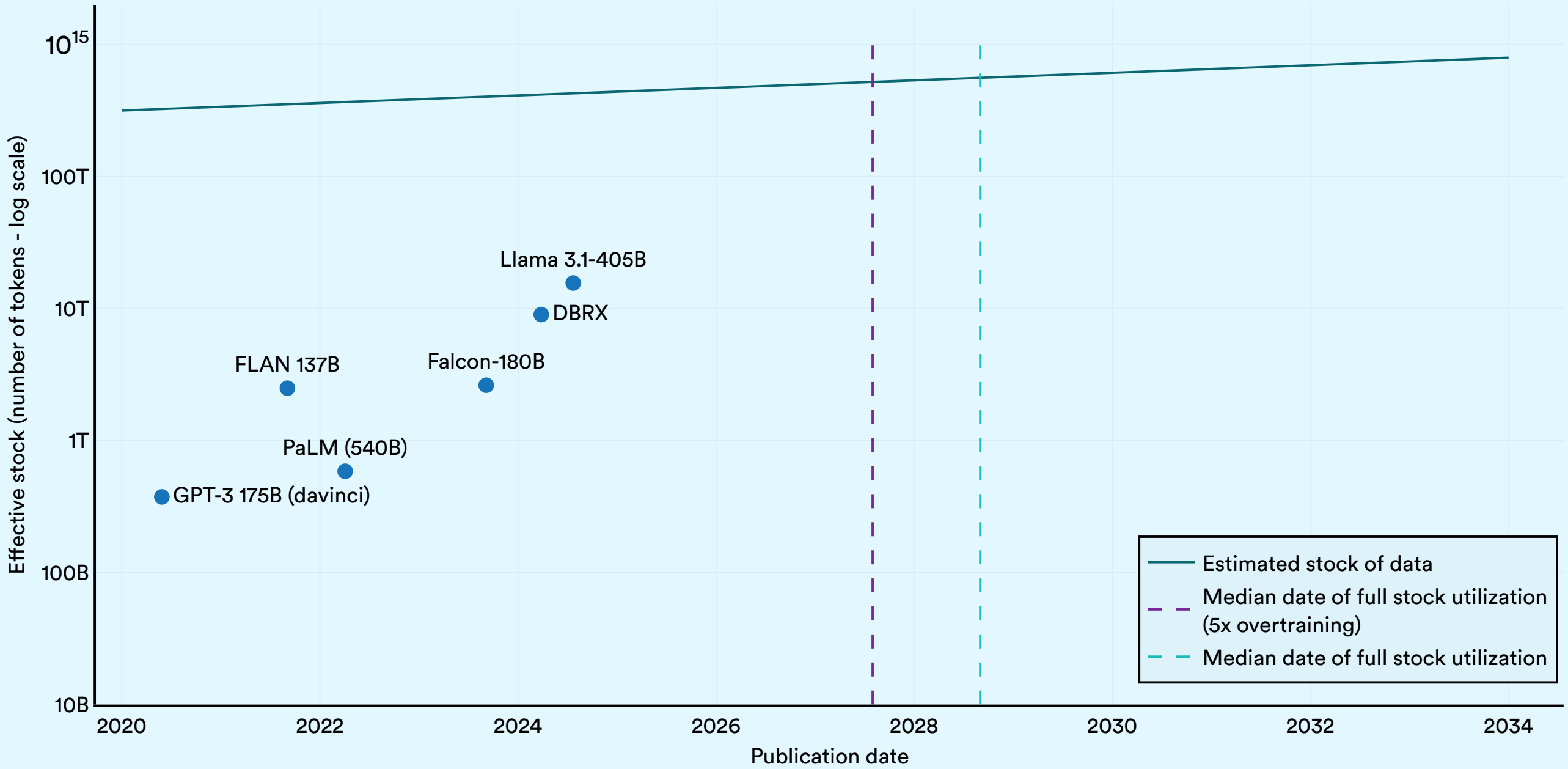

Figure 1.3.19

**Highlight:**

## Will Models Run Out of Data? (cont'd)

These projections differ slightly from Epoch's earlier estimates, which predicted that high-quality text data would be depleted by 2024. The revised projections reflect an updated methodology that incorporates new research showing that web data performs better than curated corpora and that models can be trained on the same datasets multiple times. The realization that carefully filtered web data is effective and that repeated training on the same dataset is viable has expanded estimates of the available data stock. As a result, the Epoch researchers pushed back their forecasts of when data depletion might occur.

Using synthetic data—data generated by AI models themselves—to train models has also been suggested as a solution to potential data shortages. The 2024 AI Index suggests there are limitations associated with this approach, namely that models trained this way are likely to lose representation of the tails of distributions when performing repeated training cycles on synthetic data. This leads to degraded model output quality. This phenomenon was observed across different model architectures, including variational autoencoders (VAEs), Gaussian mixture models (GMMs), and LLMs. However, newer research suggests that when synthetic data is layered on top of real data, rather than replacing it, the model collapse phenomenon does not occur. While this accumulation does not necessarily improve performance or reduce test loss (lower test loss indicates better model performance), it also does not result in the same degree of degradation as outright data replacement (Figure 1.3.20).

**Effect of data accumulation on language models pretrained on TinyStories**
Source: Gerstgrasser et al., 2024 | Chart: 2025 AI Index report

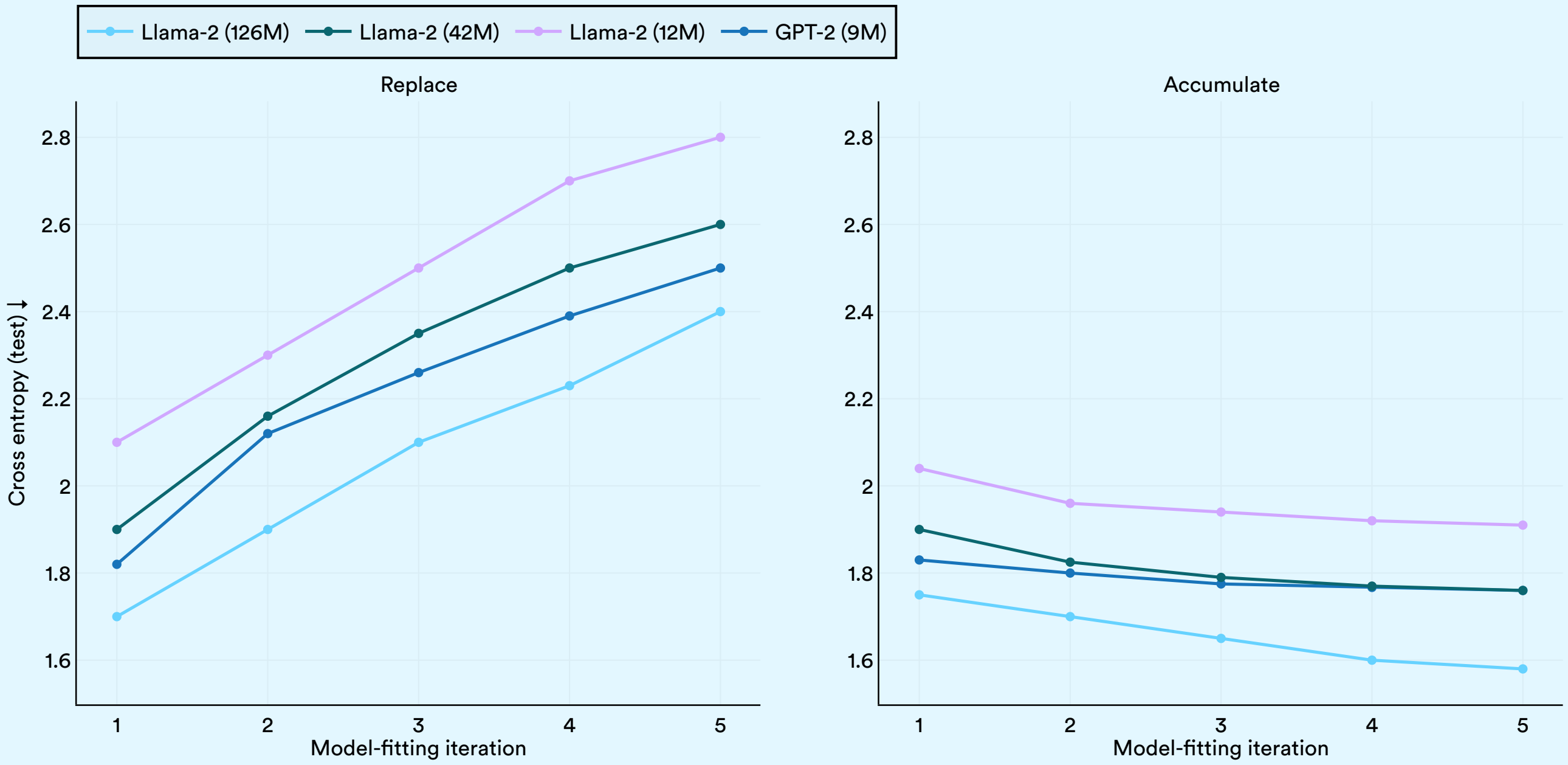

Figure 1.3.20

**Highlight:**

## Will Models Run Out of Data? (cont'd)

This year, there have been advances in generating high-fidelity synthetic data. However, synthetic data is still generally distinguishable from real data, and there is no existing scalable method to achieve the same performance training LLMs on synthetic data compared to real data. A team of Slovenian researchers compared the performance of models trained on synthetic and real data across multiple architectures and datasets. They evaluated how well synthetic relational data preserves key characteristics of the original data ("fidelity") and remains useful for downstream tasks ("utility"). They found that most methods are systematically detectable as synthetic, especially once relational information is considered. Furthermore, performance typically deteriorates compared to real data–trained models, but some methods still yield moderately good predictive scores. In a few experiments, synthetic data outperformed real data such as using Synthetic Data Vault (SDV) vs. Walmart data to train an XGBoost classifier. The researchers showed that training on the synthetic dataset achieves a lower mean squared error (MSE). There is also evidence that synthetic data shows promise in the healthcare domain. More specifically, some model architectures lead to enhanced performance on classification and prediction tasks by training on synthetically augmented datasets, increasing F1 scores or AUROC by 5%–10% on minority classes.[25]

There are concerns around the quality and fidelity of synthetically generated data, as LLMs are known to hallucinate and provide factually incorrect outputs. When training on hallucinated content in datasets, models can experience compounded degradation in output quality. New techniques have been developed to combat this issue. For example, researchers from Stanford and the University of North Carolina at Chapel Hill have used automated fact-checking and confidence scores to rank factuality scores of model response pairs. The FactTune-FS methods introduced by these researchers have tended to outperform other RLHF and decoding-based methods for factuality improvement (Figure 1.3.21). Human-in-the-loop approaches to label preferred responses have also been used to align language models. While promising, the human-in-the-loop approaches tend to be more expensive. Finally, post hoc filtering and debiasing methods can be used to remove anomalies in synthetic data before the training stage.

25 AUROC (area under the receiver operating characteristic) curve is a widely used metric for evaluating AI model performance, particularly in classification tasks.

**Highlight:**

# Will Models Run Out of Data? (cont'd)

**Factual accuracy: percentage of correct answers in biographies**

Source: Tian et al., 2023 | Chart: 2025 AI Index report

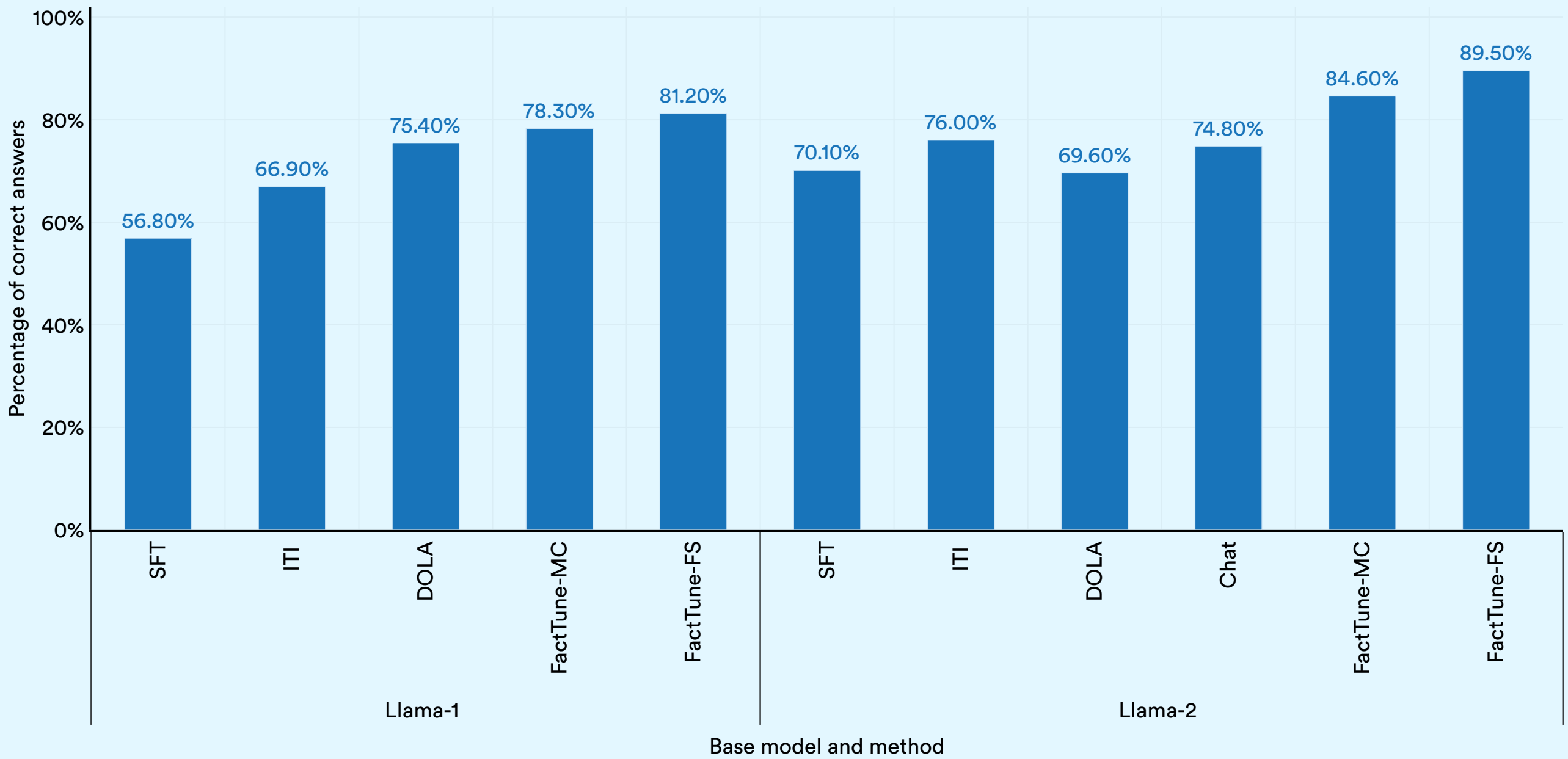

Figure 1.3.21

As the prevalence of synthetic data grows, particularly with an increasing share of web content being AI-generated, future models will inevitably be trained on non-human-generated material. While synthetic data offers the advantage of a near-infinite supply, effectively leveraging it for model training requires a deeper understanding of its impact on learning dynamics and performance. One approach to expanding datasets is data augmentation, which modifies real data—such as tilting or image mixing—to create new variations while preserving essential characteristics. Both synthetic data generation and data augmentation present opportunities to enhance AI models, but their effective use demands further research.

## Inference Cost

Last year's AI Index highlighted the rapidly rising training costs of frontier LLM systems. This year, in addition to updating its analysis on training costs, the Index examines how inference costs for frontier systems have evolved over time. Inference costs refer to the expense of querying a trained model, and they are typically measured in USD per million tokens. Data on AI token pricing comes from both Artificial Analysis and Epoch AI's proprietary database on API pricing. The reported price is a 3:1 weighted average of input and output token prices.

To analyze inference costs, the AI Index worked with Epoch to measure how costs have decreased for a fixed AI performance threshold. This standardized approach facilitates a more accurate comparison. While newer models may cost more, they also tend to perform significantly better—so comparing them directly to older, less capable models can obscure the real trend: AI performance per dollar has improved substantially. For instance, the inference cost for an AI model scoring the equivalent of GPT-3.5 (64.8) on MMLU, a popular benchmark for assessing language model performance, dropped from $20 per million tokens in November 2022 to just $0.07 per million tokens by October 2024 (Gemini-1.5-Flash-8B)—a more than 280-fold reduction in approximately 1.5 years. A similar trend is evident in the cost of models scoring above 50% on GPQA, a substantially more challenging benchmark than MMLU. There, inference costs declined from $15 per million tokens in May 2024 to $0.12 per million tokens by December 2024 (Phi 4). Epoch AI estimates that, depending on the task, LLM inference costs have been falling anywhere from nine to 900 times per year.

**Inference price across select benchmarks, 2022–24**
Source: Epoch AI, 2025; Artificial Analysis, 2025 | Chart: 2025 AI Index report

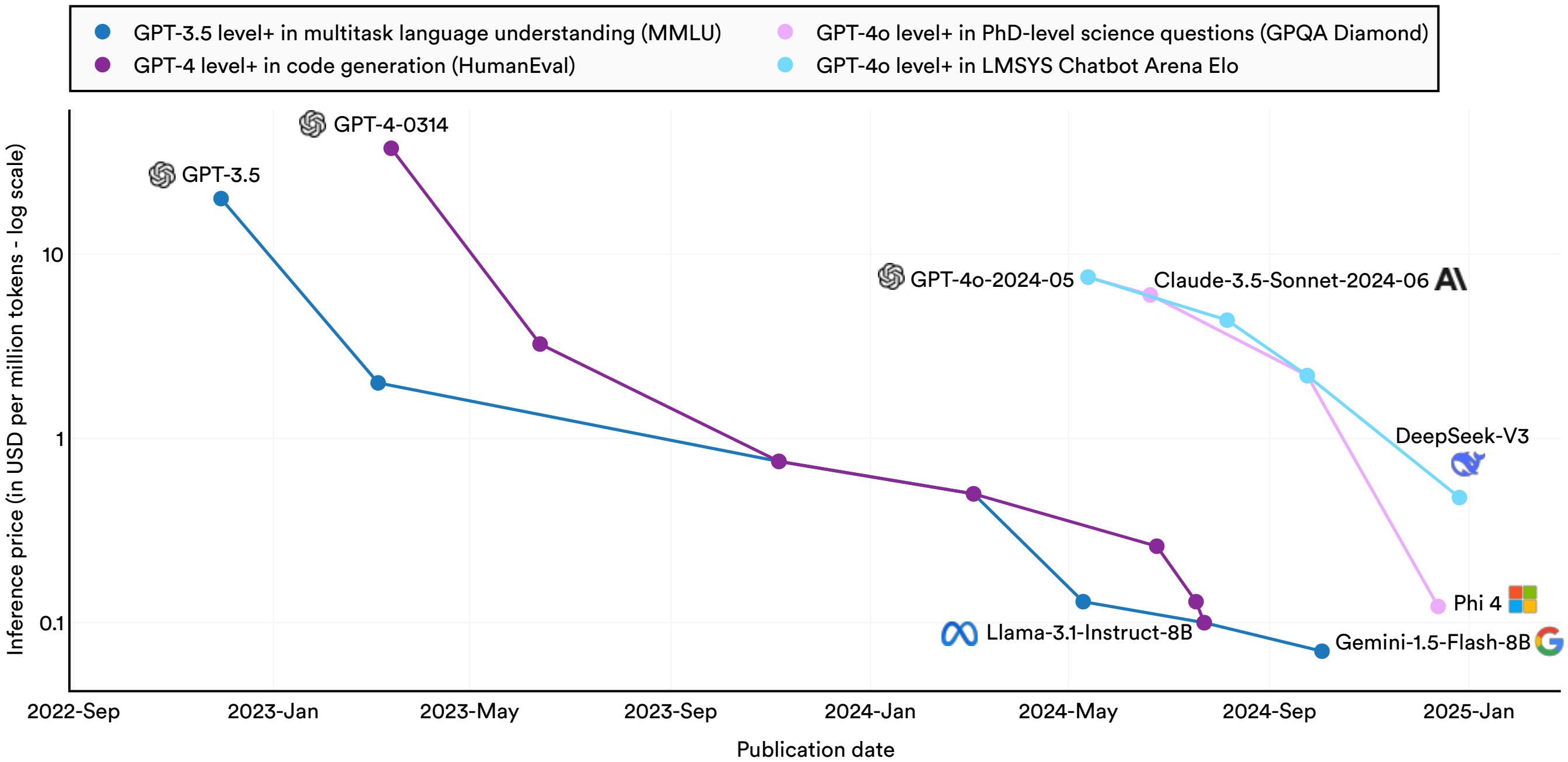

Figure 1.3.22

The inference cost to achieve a given level of performance has declined notably over time. However, state-of-the-art models remain more expensive than some of the previously mentioned alternatives. Figure 1.3.23 illustrates the cost per million tokens for leading models from developers such as OpenAI, Meta, and Anthropic.[26] These top-tier models are generally priced higher than smaller models from the same companies, reflecting the premium required for cutting-edge performance.

**Output price per million tokens for select models**
Source: Artificial Analysis, 2025 | Chart: 2025 AI Index report

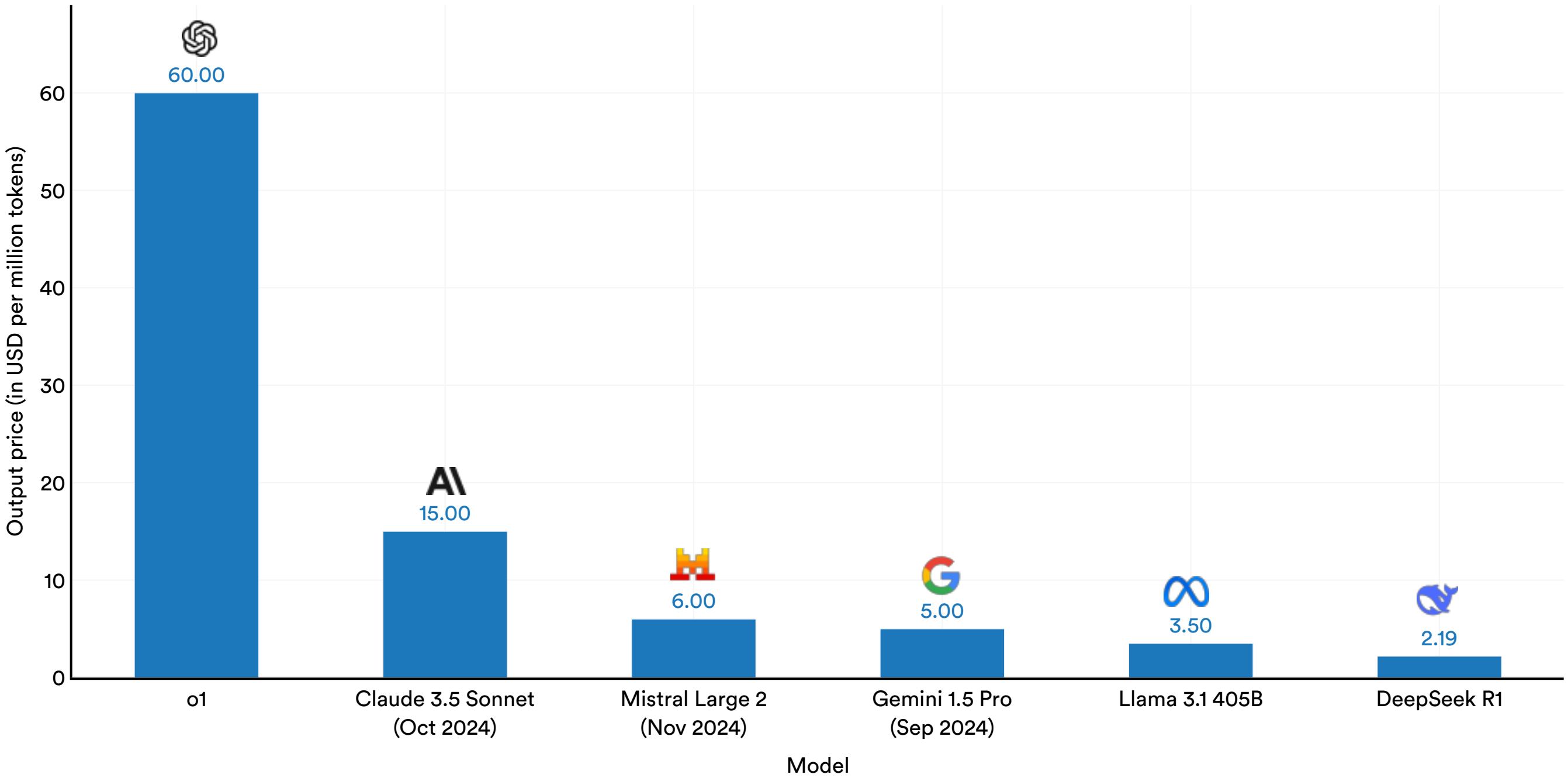

Figure 1.3.23

## Training Cost

A frequent discussion around foundation models pertains to their high training costs. While AI companies rarely disclose exact figures, costs are widely estimated to reach into the millions of dollars—and continue to rise. OpenAI CEO Sam Altman, for instance, indicated that training GPT-4 exceeded $100 million. In July 2024, Anthropic CEO Dario Amodei noted that model training runs costing around $1 billion were already underway. Even more recent models, such as DeepSeek-V3, reportedly cost less—about $6 million—but overall, training remains extremely expensive.[27]

Understanding the costs associated with training AI models remains important, yet detailed cost information remains scarce. Last year, the AI Index published initial estimates on the costs of training foundation models. This year, the AI Index once again partnered with Epoch AI to update and refine these estimates. To calculate costs for cutting-edge models, the Epoch team analyzed factors such as training duration, hardware type, quantity, and utilization rates, relying on information from academic publications, press releases, and technical reports.[28]

26 The Index visualizes a selection of state-of-the-art models with publicly available pricing as of February 2025. Since publication, newer models may have been released and pricing may have changed.

27 Some reports have disputed the stated cost of DeepSeek-V3, arguing that when factoring in employee salaries, capital expenditures, and research expenses, the actual development costs were significantly higher.

28 A detailed report on Epoch's research methodology is available in this paper.

Figure 1.3.24 visualizes the estimated training cost associated with select AI models, based on cloud compute rental prices. Figure 1.3.25 visualizes the training cost of all AI models for which the AI Index has estimates.

AI Index estimates validate suspicions that in recent years model training costs have significantly increased. For example, in 2017, the original Transformer model, which introduced the architecture that underpins virtually every modern LLM, cost around \$670 to train. RoBERTa Large, released in 2019, which achieved state-of-the-art results on many canonical comprehension benchmarks like SQuAD and GLUE, cost around \$160,000 to train. Fast-forward to 2023, and training costs for OpenAI's GPT-4 were estimated around \$79 million.

One of the few 2024 models for which Epoch could estimate training costs was Llama 3.1-405B, with an estimated cost of \$170 million. As the AI landscape grows more competitive, companies are disclosing less about their training processes, making it increasingly difficult to estimate computational costs.

As established in previous AI Index reports, there is a direct correlation between the training costs of AI models and their computational requirements. As illustrated in Figure 1.3.26, models with greater computational training needs cost substantially more to train.

**Estimated training cost of select AI models, 2019–24**
Source: Epoch AI, 2024 | Chart: 2025 AI Index report

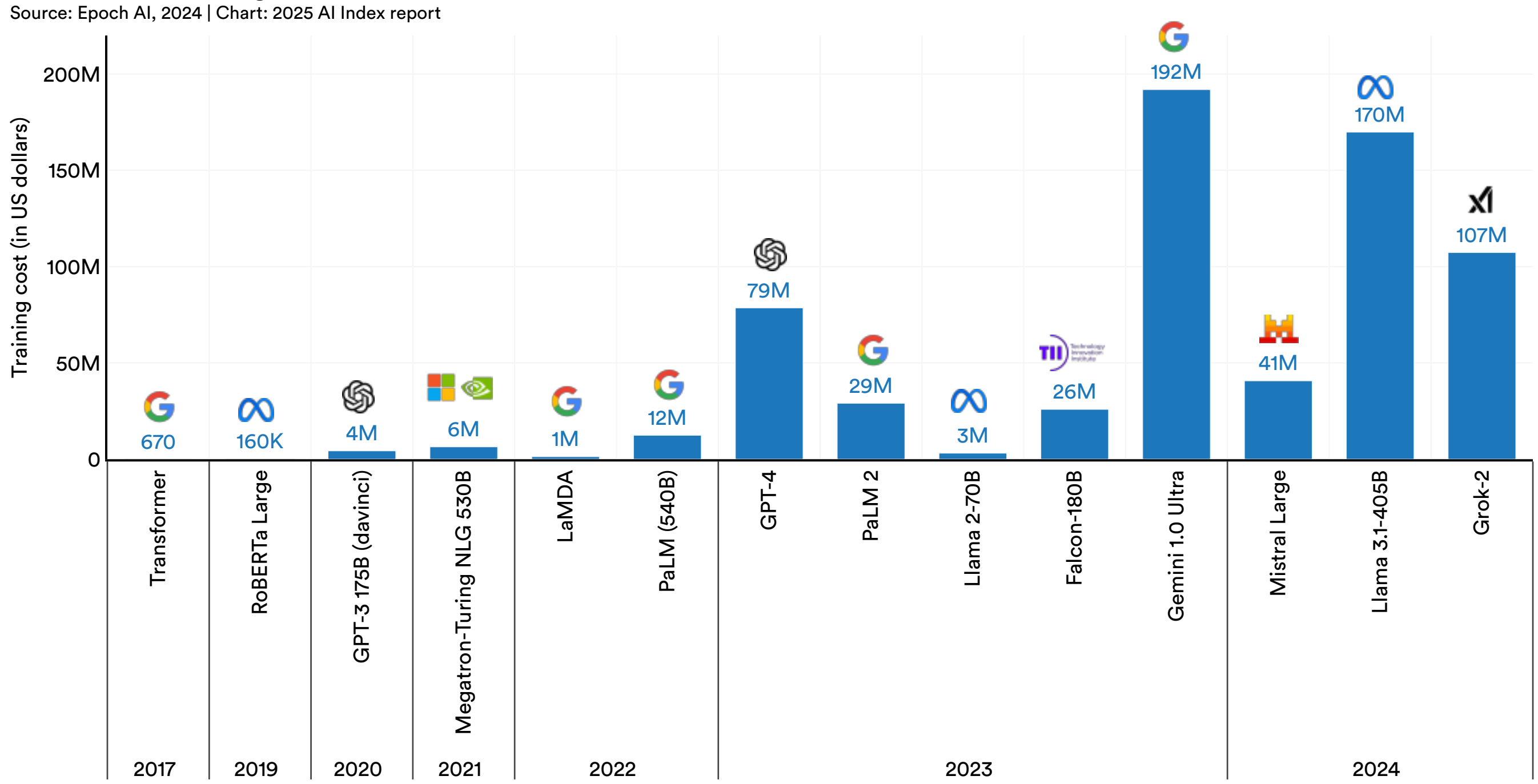

Figure 1.3.24

29 The cost figures reported in this section are inflation-adjusted.

### Estimated training cost of select AI models, 2016–24
Source: Epoch AI, 2024 | Chart: 2025 AI Index report

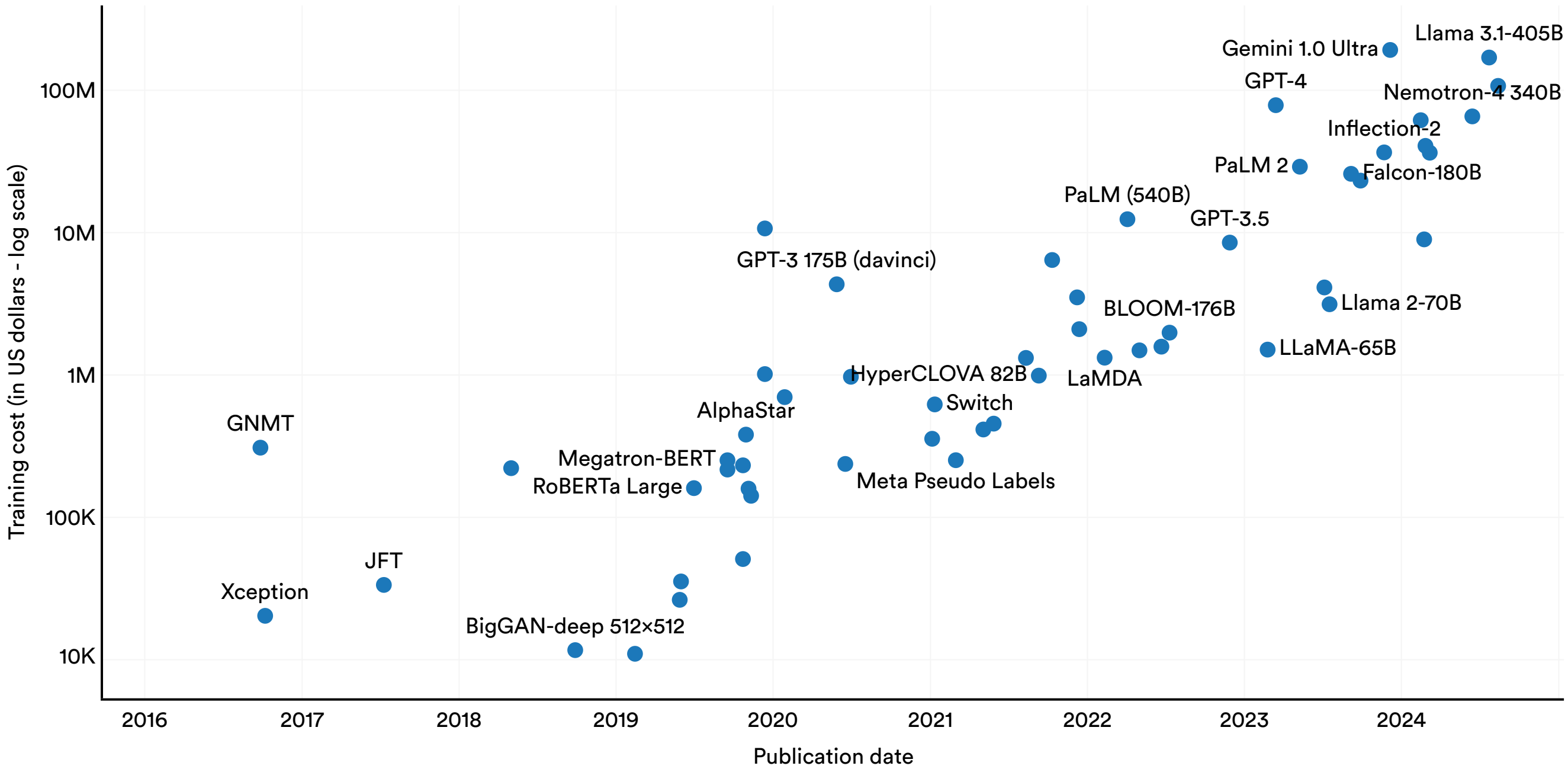

Figure 1.3.25

### Estimated training cost and compute of select AI models
Source: Epoch AI, 2024 | Chart: 2025 AI Index report

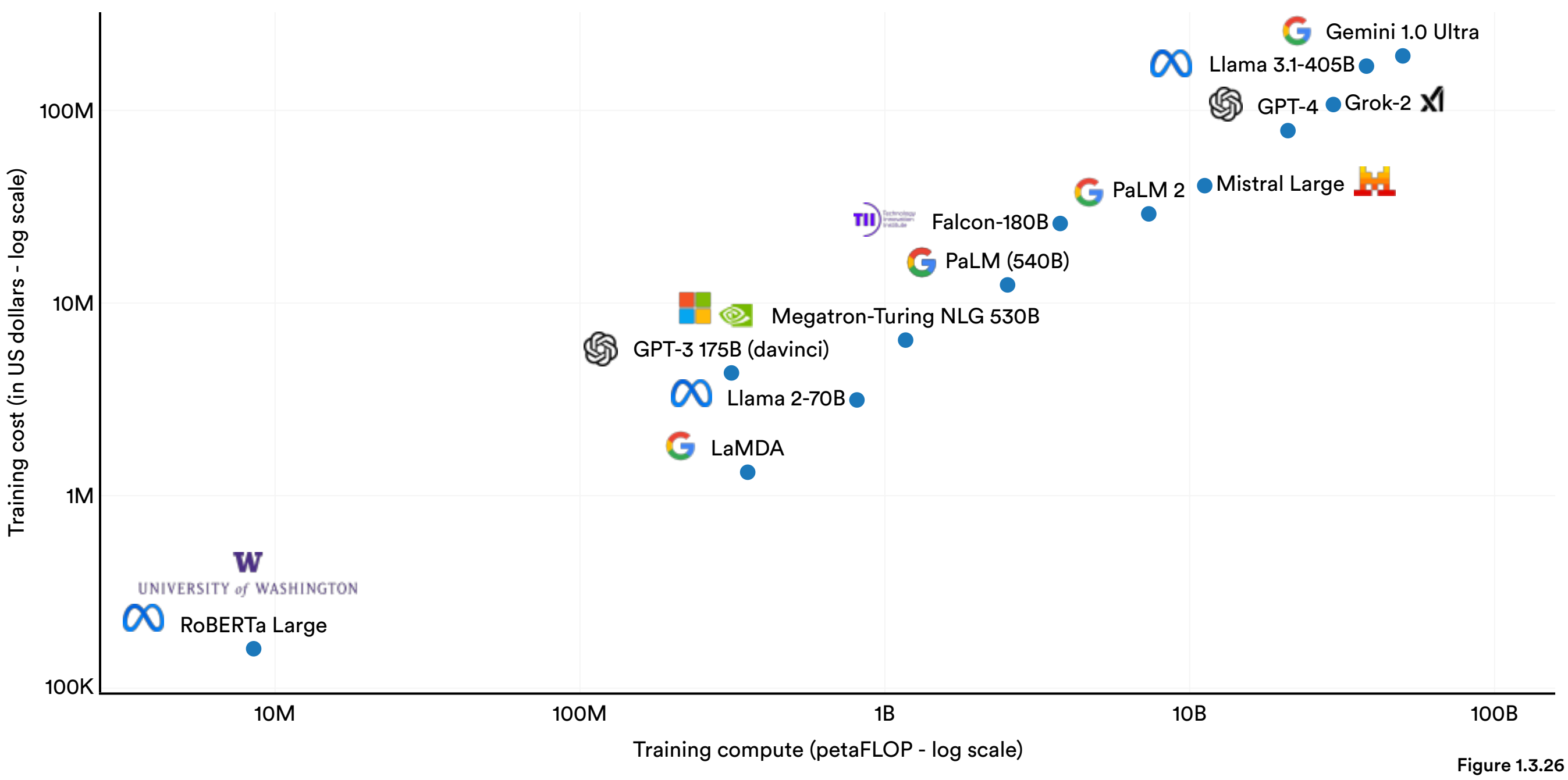

Figure 1.3.26

Hardware advancements play a critical role in driving AI progress. While scaling models and training on larger datasets have led to significant performance improvements, these advances have largely been enabled by improvements in hardware—particularly the development of more powerful and efficient GPUs (graphics processing units). GPUs accelerate complex computations, allowing models to process vast amounts of data in parallel and significantly reducing training time. This section of the Index leverages data from Epoch AI to analyze key trends in machine learning hardware and its impact on AI development.

While this section currently emphasizes compute performance (FLOP/s), network bandwidth—the speed at which GPUs communicate—is equally critical. Although data on network bandwidth of data centers is limited, future editions of the AI Index will aim to include this information.

# 1.4 Hardware

## Overview

Figure 1.4.1 illustrates the peak computational performance of ML hardware across different precision types, where precision refers to the number of bits used to represent numerical values, particularly floating-point numbers, in computations. The choice of precision depends on the specific goal. For instance, lower-precision hardware, which requires fewer bits and has lower memory bandwidth, is ideal for optimizing computation speed and energy efficiency. This is particularly beneficial for AI models running on edge or mobile devices or in scenarios where inference speed is a priority. On the other hand, higher-precision hardware preserves greater numerical accuracy, making it essential for scientific computing and applications sensitive to precision errors. Of the precisions visualized in the figures below, FP32 has the highest precision, TF32 offers medium-high precision, and Tensor-FP16/BF16 and FP16 are lower-precision formats optimized for speed and efficiency.

Measured in 16-bit floating-point operations, Epoch estimates that machine learning hardware performance has grown over the period 2008–2024 at an annual rate of approximately 43%, doubling every 1.9 years. According to Epoch, this progress has been driven by increased transistor counts, advancements in semiconductor manufacturing, and the development of specialized hardware for AI workloads.

**Peak computational performance of ML hardware for different precisions, 2008–24**
Source: Epoch AI, 2025 | Chart: 2025 AI Index report

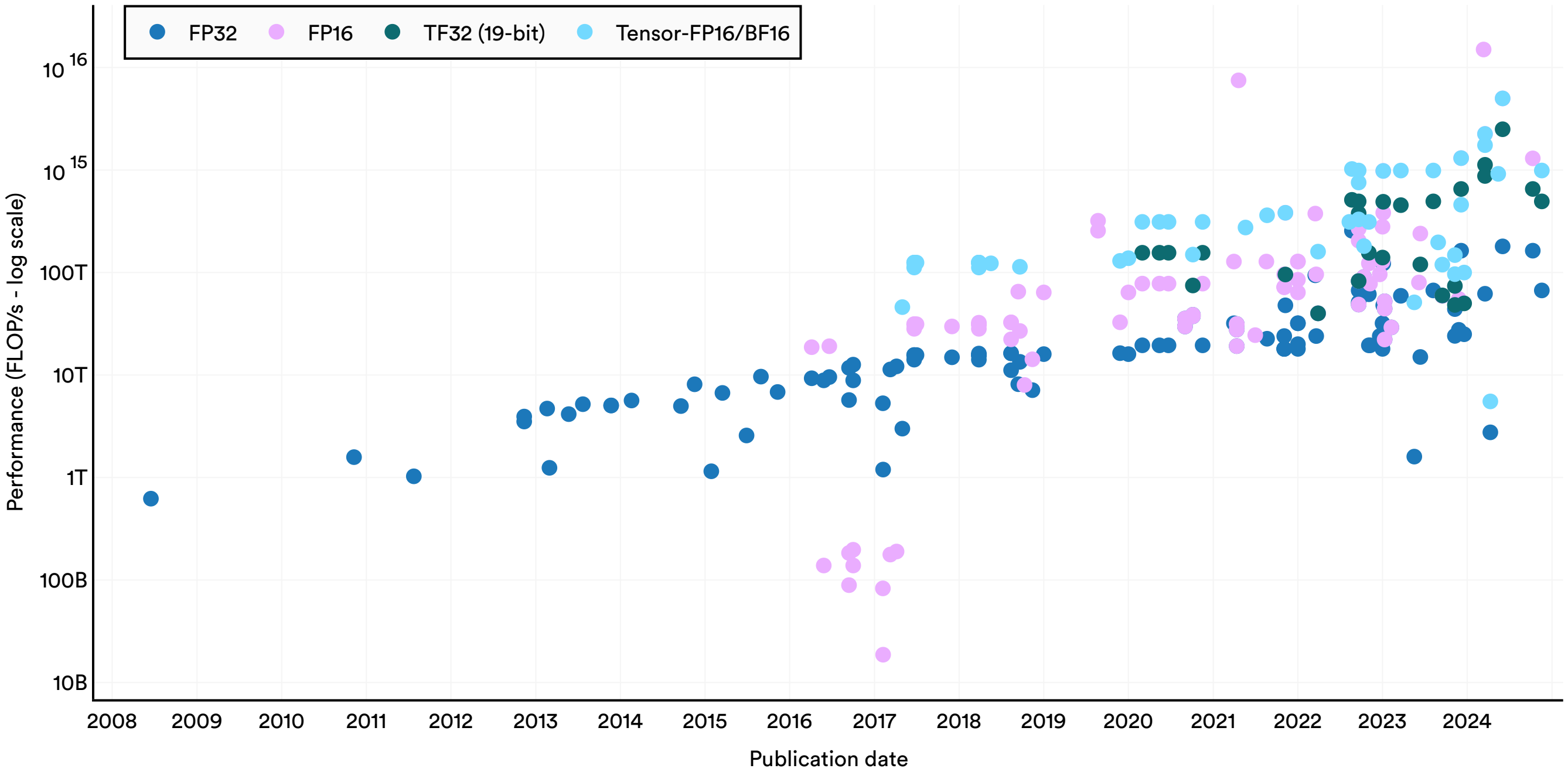

Figure 1.4.1

The price-performance of leading machine learning hardware has steadily improved. Figure 1.4.2 illustrates the performance of selected Nvidia data center GPUs—among the most commonly used for AI training—in FLOP per second. Figure 1.4.3 visualizes the price-performance of those same GPUs, measured in FLOP per second per dollar. For example, the H100 GPU, announced in March 2022, achieves 22 billion FLOP per second per dollar, which is approximately 1.7 times the price-performance of the A100 (launched in June 2020) and 16.9 times that of the P100 (released in April 2016). Epoch estimates that hardware with a fixed performance level decreases in cost by 30% annually, making AI training increasingly affordable, scalable, and conducive to model improvements.

**Performance of leading Nvidia data center GPUs for machine learning**

Source: Epoch AI, 2025 | Chart: 2025 AI Index report

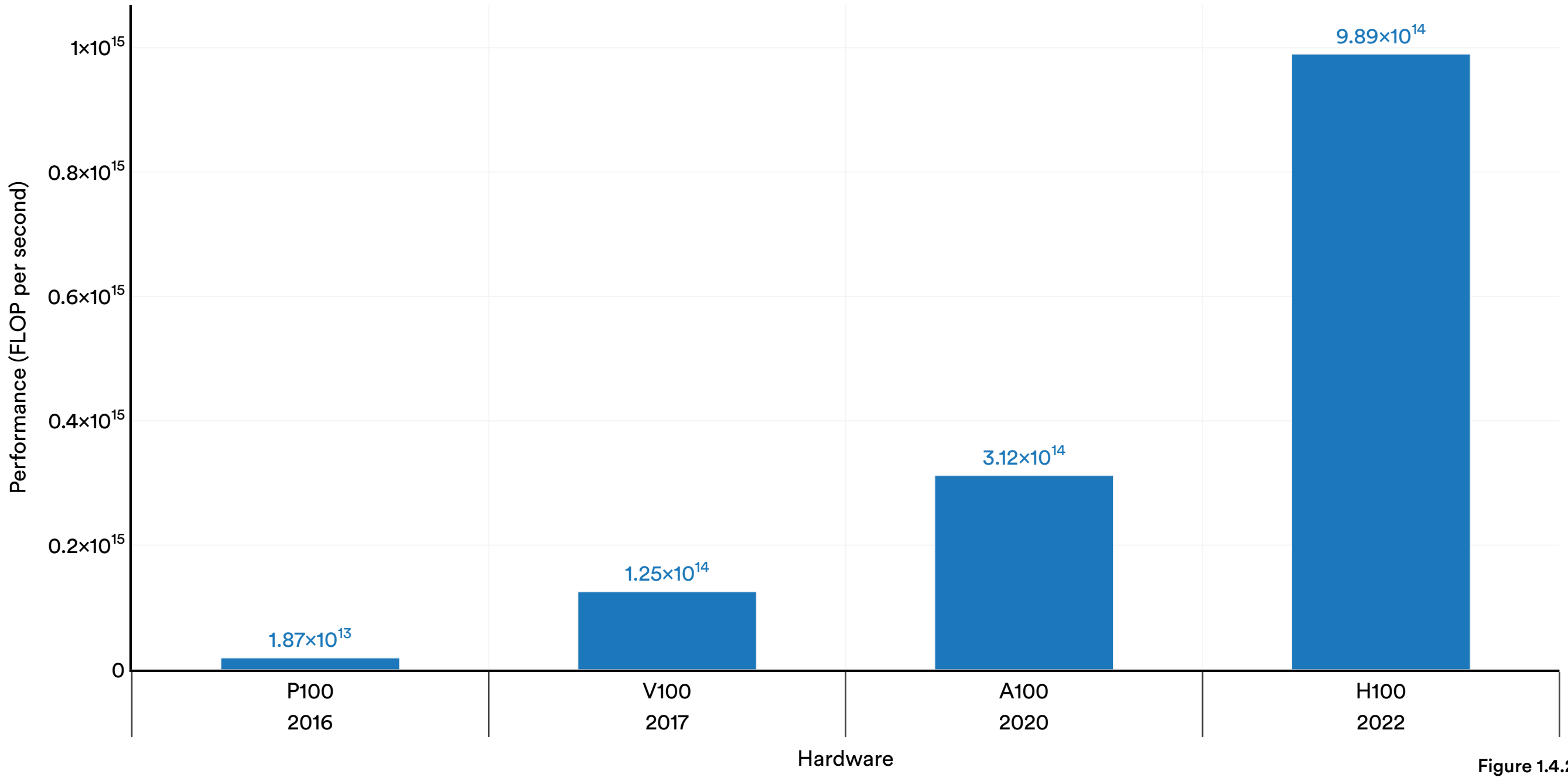

Figure 1.4.2

Figure 1.4.4, based on the Epoch AI notable machine learning models dataset, examines the hardware used to train notable machine learning models. As of 2024, the most commonly reported hardware was the A100, used by 64 models, followed by the V100. An increasing number of models are now being trained on the H100, with 15 reported by the end of 2024.

### Price-performance of leading Nvidia data center GPUs for machine learning

Source: Epoch AI, 2025 | Chart: 2025 AI Index report

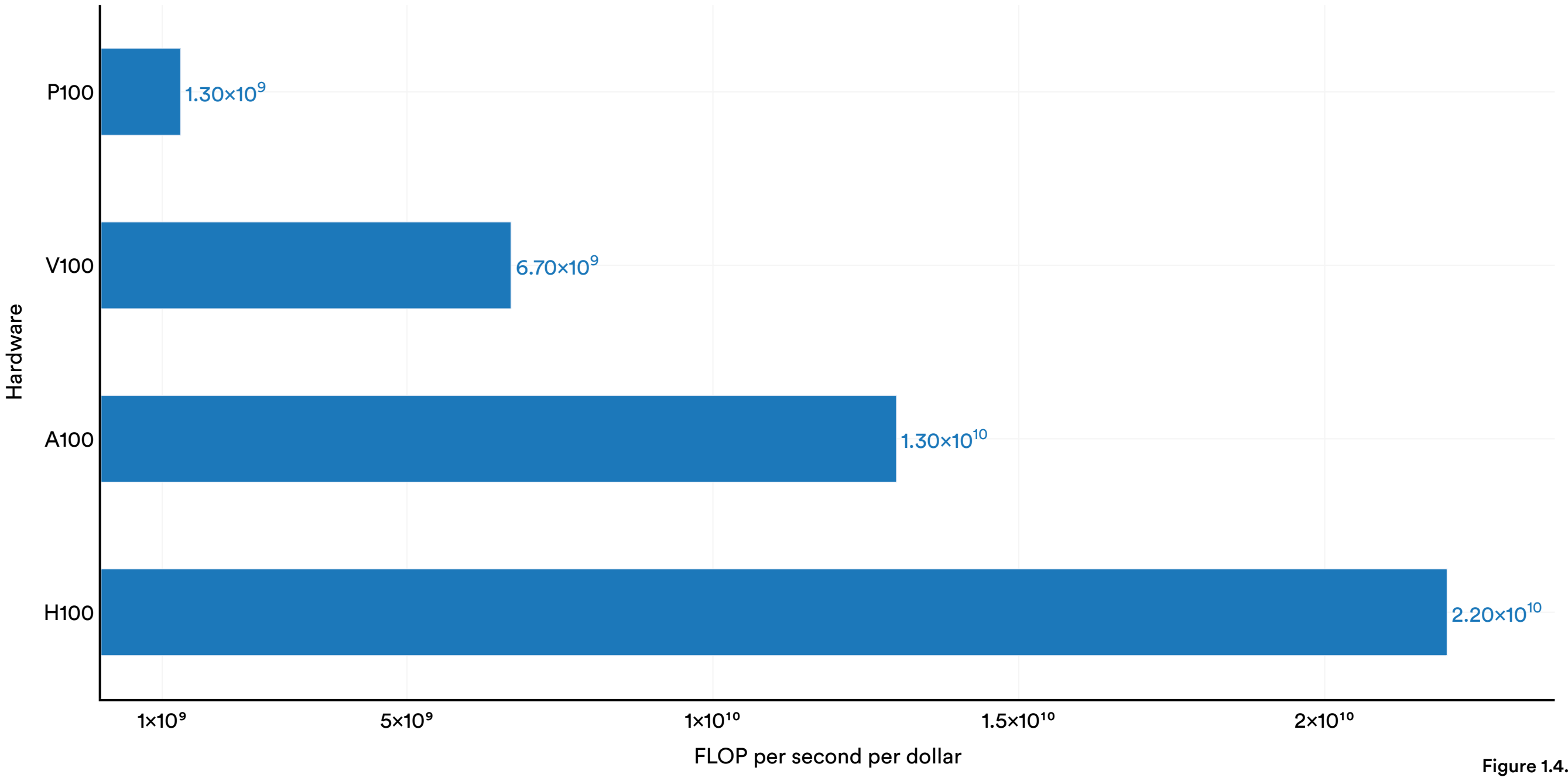

Figure 1.4.3

### Cumulative number of notable AI models trained by accelerator, 2017–24

Source: Epoch AI, 2025 | Chart: 2025 AI Index report

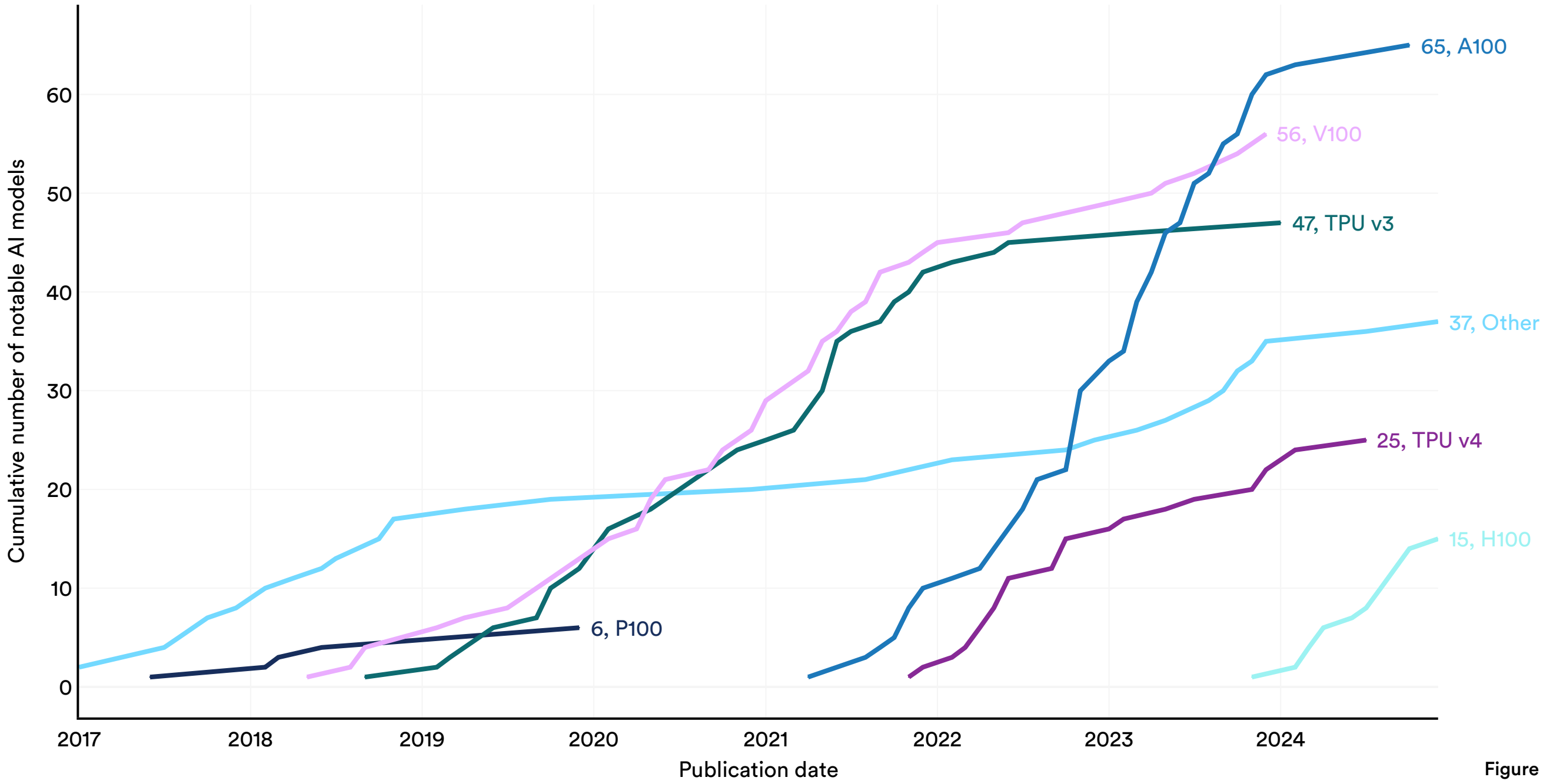

Figure 1.4.4

**Highlight:**

## Energy Efficiency and Environmental Impact

Training AI systems requires substantial energy, making the energy efficiency of machine learning hardware a critical factor. Epoch AI reports that ML hardware has become increasingly energy efficient over time, improving by approximately 40% per year. Figure 1.4.5 illustrates the energy efficiency of Tensor-FP16 precision hardware, measured in FLOP/s per watt. For instance, the Nvidia B100, released in March 2024, achieved an energy efficiency of 2.5 trillion FLOP/s per watt, compared to the Nvidia P100, released in April 2016, which reported 74 billion FLOP/s per watt. This means the B100 is 33.8 times more energy efficient than the P100.

**Energy efficiency of leading machine learning hardware, 2016–24**

Source: Epoch AI, 2025 | Chart: 2025 AI Index report

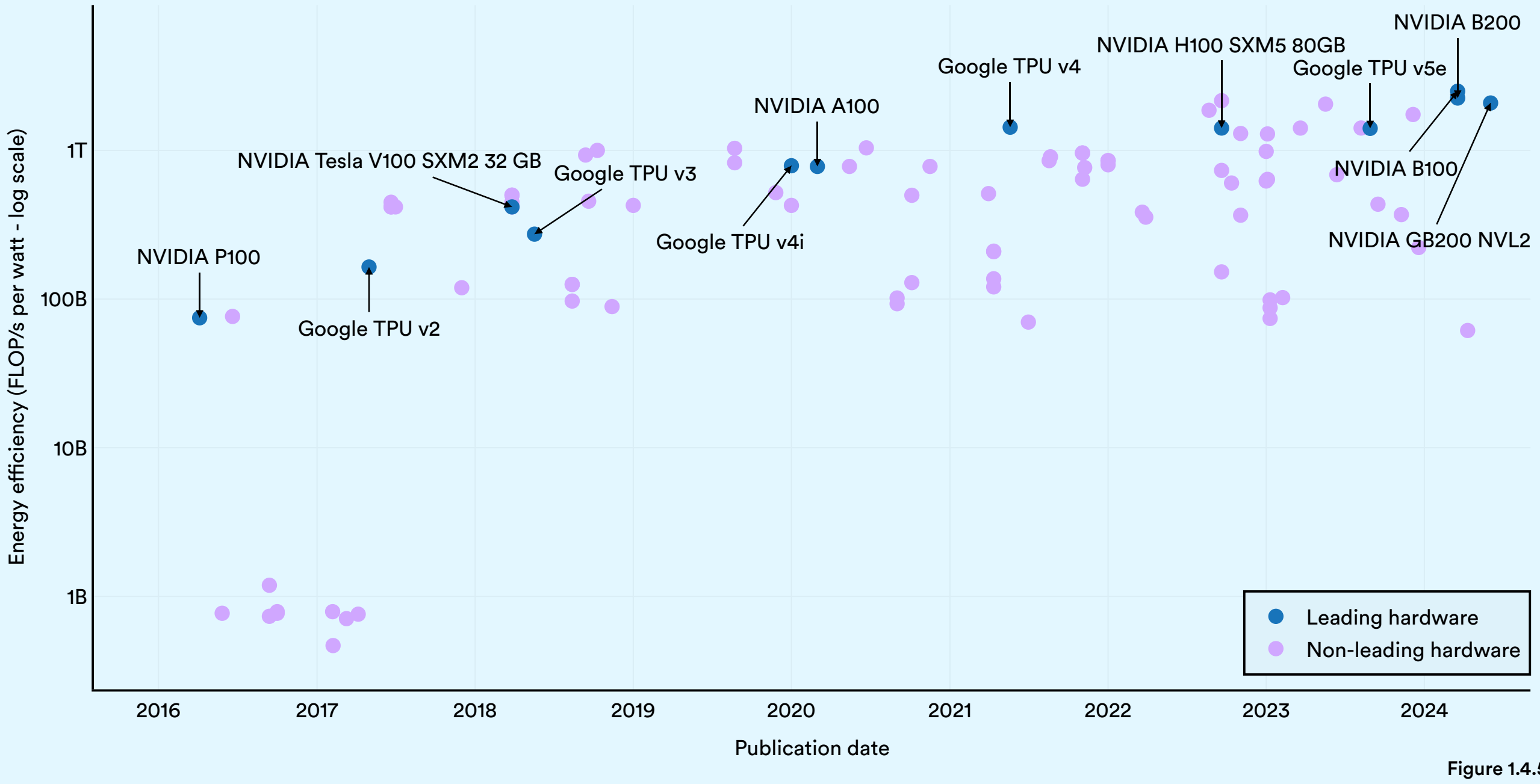

Figure 1.4.5

Highlight:

## Energy Efficiency and Environmental Impact (cont'd)

Despite significant improvements in the energy efficiency of AI hardware, the overall power consumption required to train AI systems continues to rise rapidly. Figure 1.4.6 illustrates the total power draw, measured in watts, for training various state-of-the-art AI models. For example, the original Transformer, introduced in 2017, consumed an estimated 4,500 watts. In contrast, PaLM, one of Google's first flagship LLMs, had a power draw of 2.6 million watts—almost 600 times that of the Transformer. Llama 3.1-405B, released in the summer of 2024, required 25.3 million watts, consuming over 5,000 times more power than the original Transformer. According to Epoch AI, the power required to train frontier AI models is doubling annually. The rising power consumption of AI models reflects the trend of training on increasingly larger datasets.

Unsurprisingly, given that the total amount of power used to train AI systems has increased over time, so has the amount of carbon emitted by the models. Many factors determine the amount of carbon emitted by AI systems, including the number of parameters in a model, the power usage effectiveness of a data center, and the grid carbon intensity.[30]

**Total power draw required to train frontier models, 2011–24**

Source: Epoch AI, 2025 | Chart: 2025 AI Index report

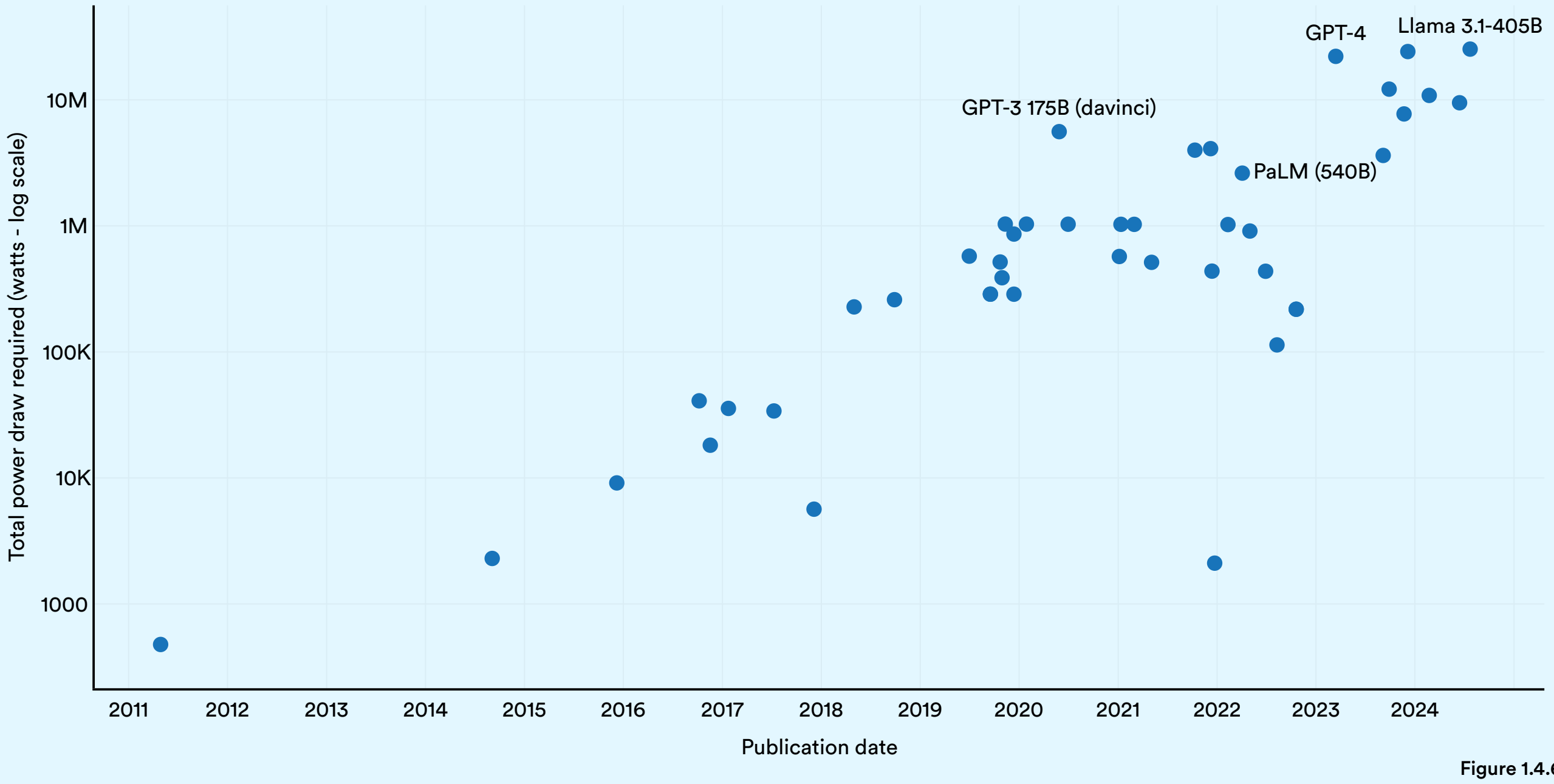

Figure 1.4.6

30 Power usage effectiveness (PUE) is a metric used to evaluate the energy efficiency of data centers. It is the ratio of the total amount of energy used by a computer data center facility, including air conditioning, to the energy delivered to computing equipment. The higher the PUE, the less efficient the data center.

**Highlight:**

## Energy Efficiency and Environmental Impact (cont'd)

Figure 1.4.7 illustrates the carbon emissions of selected AI models, sorted by their release year. To estimate these emissions, the AI Index used carbon data published by model developers and supplemented it with calculations from a widely used online AI training emissions calculator. This step was necessary as many developers do not disclose their models' carbon footprints. The calculator estimates emissions based on the type of hardware used for training, total training hours, cloud provider, and training region.[31]

The carbon emissions from training frontier AI models have steadily increased over time. While AlexNet's emissions were negligible, GPT-3 (released in 2020) reportedly emitted around 588 tons of carbon during training, GPT-4 (2023) emitted 5,184 tons, and Llama 3.1 405B (2024) emitted 8,930 tons. DeepSeek V3, released in 2024, and whose performance is comparable to OpenAI's o1, is estimated to have emissions comparable to the GPT-3, released five years ago. For context, on average, Americans emit 18.08 tons of carbon per capita per year.

**Estimated carbon emissions from training select AI models and real-life activities, 2012–24**

Source: AI Index, 2025; Strubell et al., 2019 | Chart: 2025 AI Index report

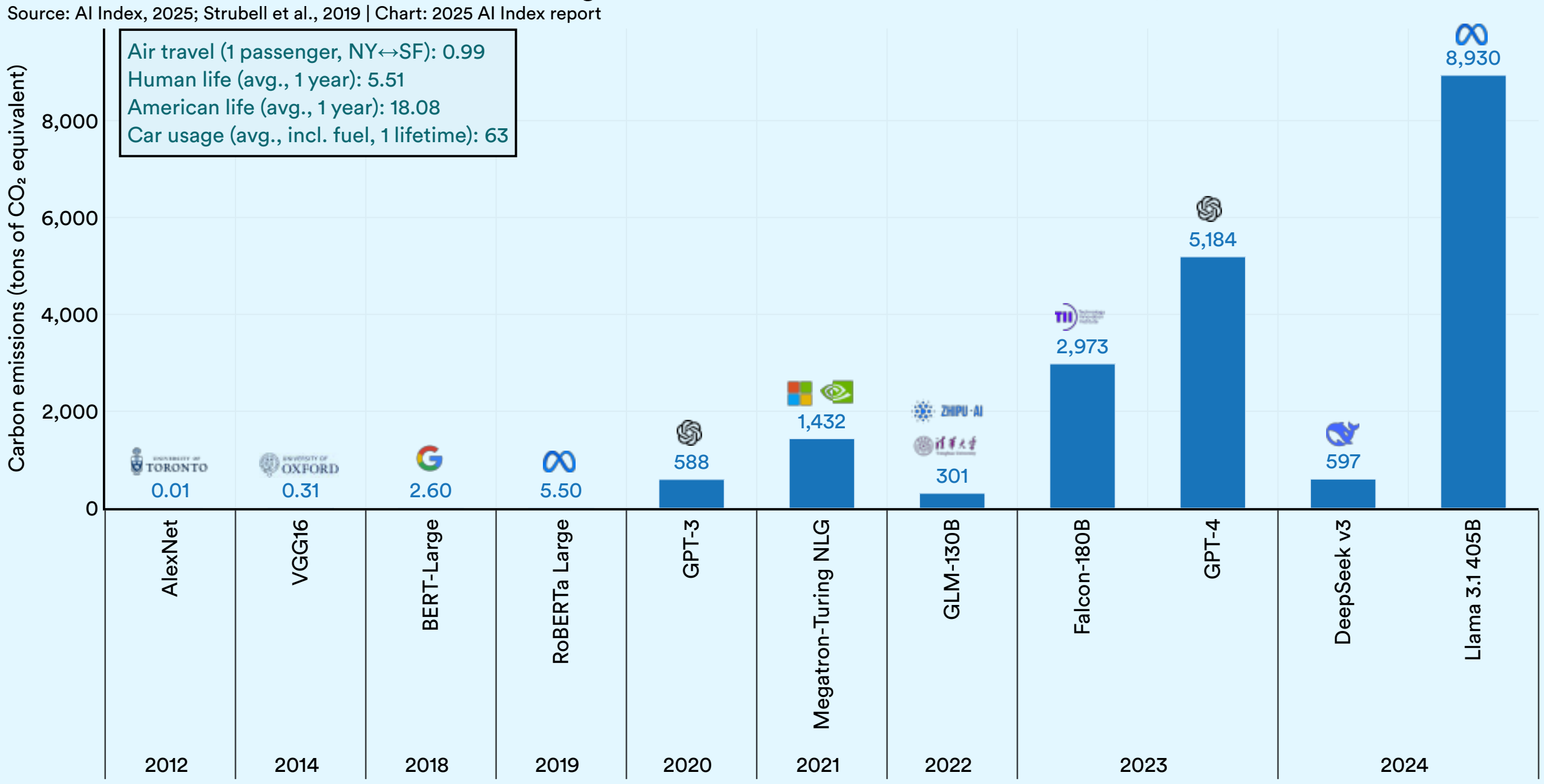

Figure 1.4.7

31 The AI Index sourced input data—such as training hardware and duration—for the emissions calculator from various online sources. To validate the accuracy of the calculator, the Index compared the calculator's estimates with actual emissions reported by developers and found that the results were largely consistent. The full estimation methodology is detailed in the Appendix.

**Highlight:**

## Energy Efficiency and Environmental Impact (cont'd)

**Estimated carbon emissions and number of parameters by select AI models**
Source: AI Index, 2025 | Chart: 2025 AI Index report

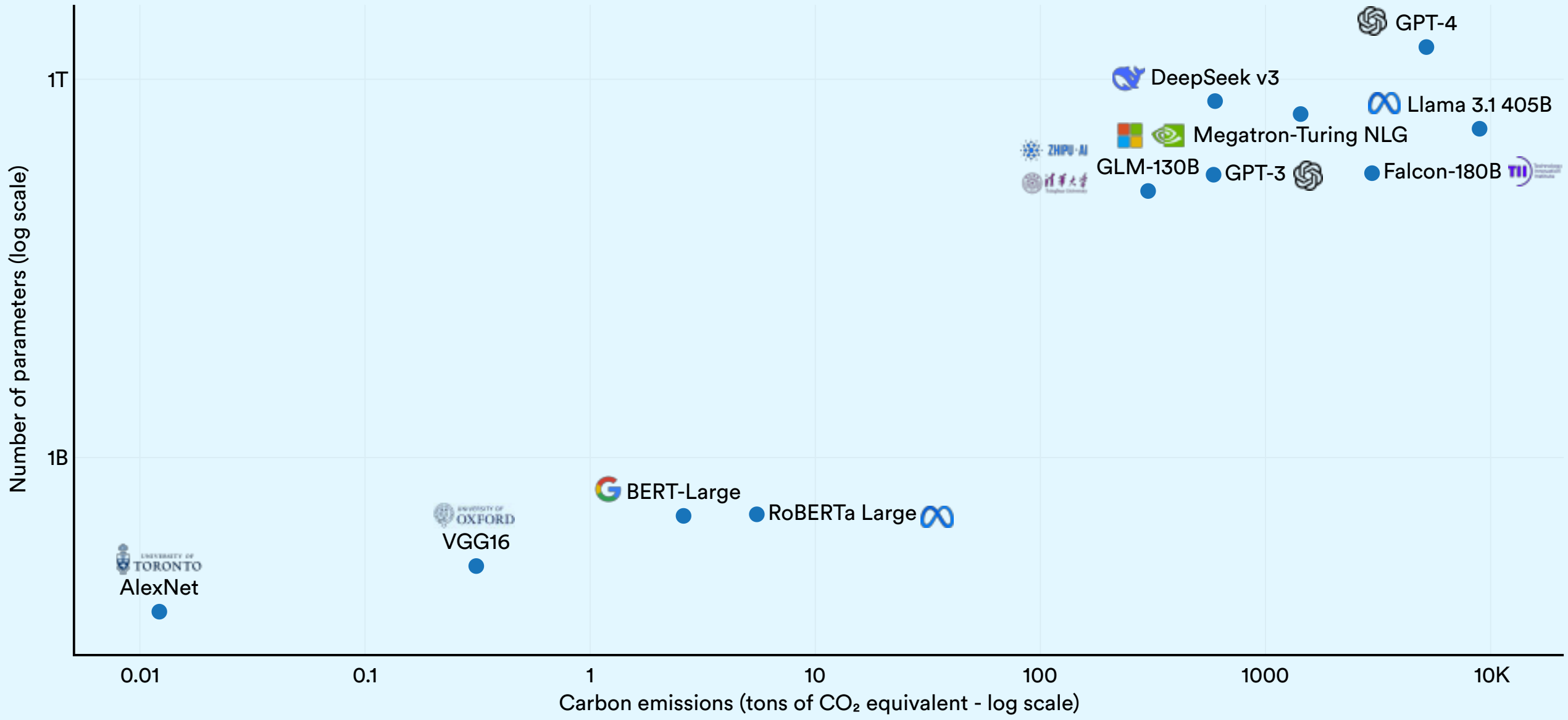

Figure 1.4.8

AI conferences serve as essential platforms for researchers to present their findings and network with peers and collaborators. Over the past two decades, these conferences have expanded in scale, quantity, and prestige. This section explores trends in attendance at major AI conferences.

# 1.5 AI Conferences

## Conference Attendance

Figure 1.5.1 graphs attendance at a selection of AI conferences since 2010. In 2020 the pandemic forced conferences to be held fully online, increasing attendance significantly. This was followed by a decline in attendance, likely due to the shift back to in-person formats, returning attendance in 2022 to prepandemic levels. Since then, there has been a steady growth in conference attendance, increasing almost 21.7% from 2023 to 2024.[32] Since 2014, the annual number of attendees has risen by more than 60,000, reflecting not just a growing interest in AI research but also the emergence of new AI conferences.

Neural Information Processing Systems (NeurIPS) remains the most attended AI conference, attracting almost 20,000 participants in 2024 (Figure 1.5.2 and Figure 1.5.3). Among the major AI conferences, NeurIPS, CVPR, ICML, ICRA, ICLR, IROS and AAAI experienced increases in attendance over the last year.

**Attendance at select AI conferences, 2010–24**
Source: AI Index, 2024 | Chart: 2025 AI Index report

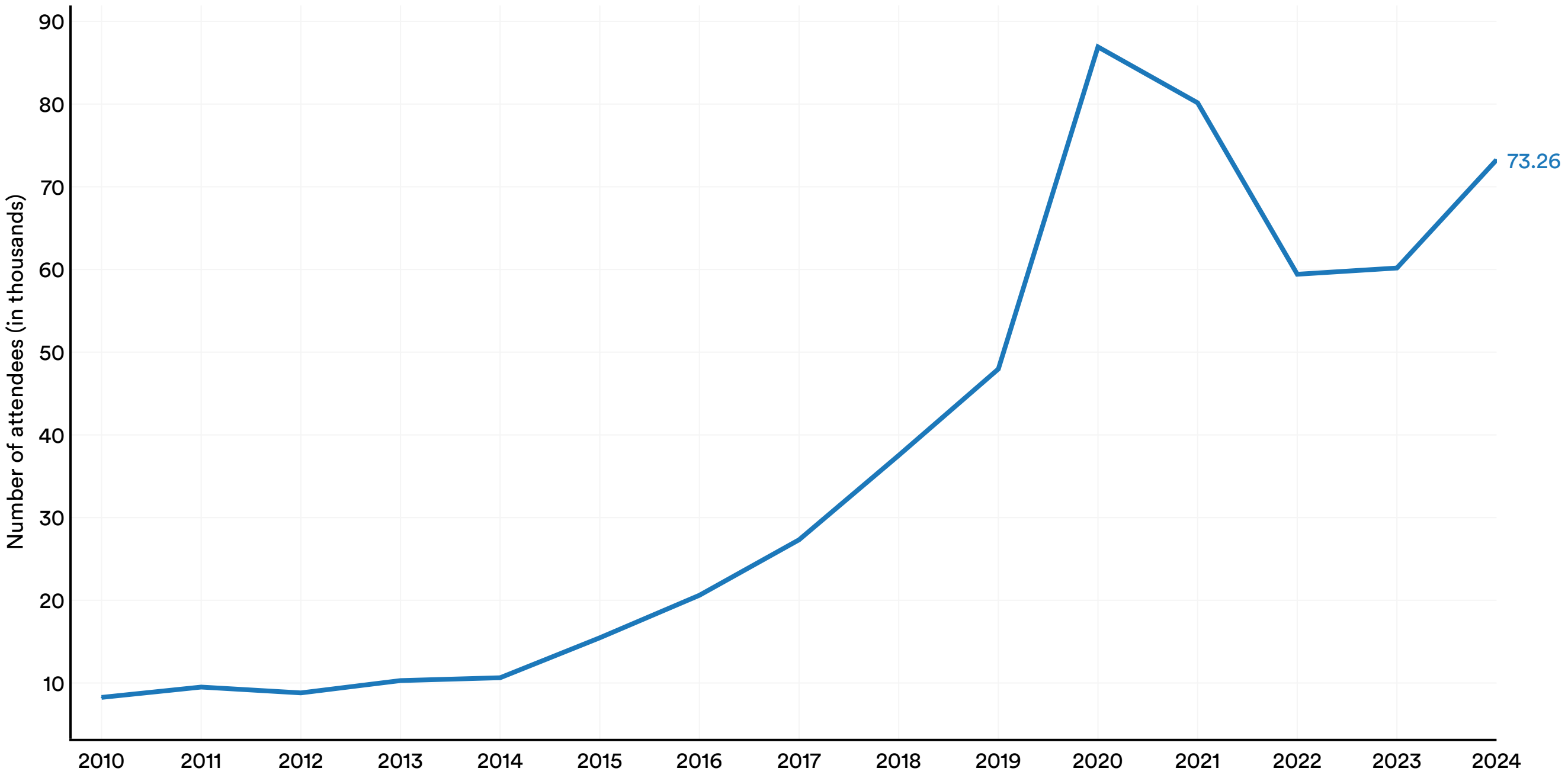

Figure 1.5.1

32 This data should be interpreted with caution given that many conferences in the last few years have had virtual or hybrid formats. Conference organizers report that measuring the exact attendance numbers at virtual conferences is difficult, as virtual conferences allow for higher attendance of researchers from around the world. The AI Index reports total attendance figures, encompassing virtual, hybrid, and in-person participation. The conferences for which the AI Index tracked data include AAAI, AAMAS, CVPR, EMNLP, FAccT, ICAPS, ICCV, ICLR, ICML, ICRA, IJCAI, IROS, KR, NeurIPS, and UAI.

## Attendance at large conferences, 2010–24

Source: AI Index, 2024 | Chart: 2025 AI Index report

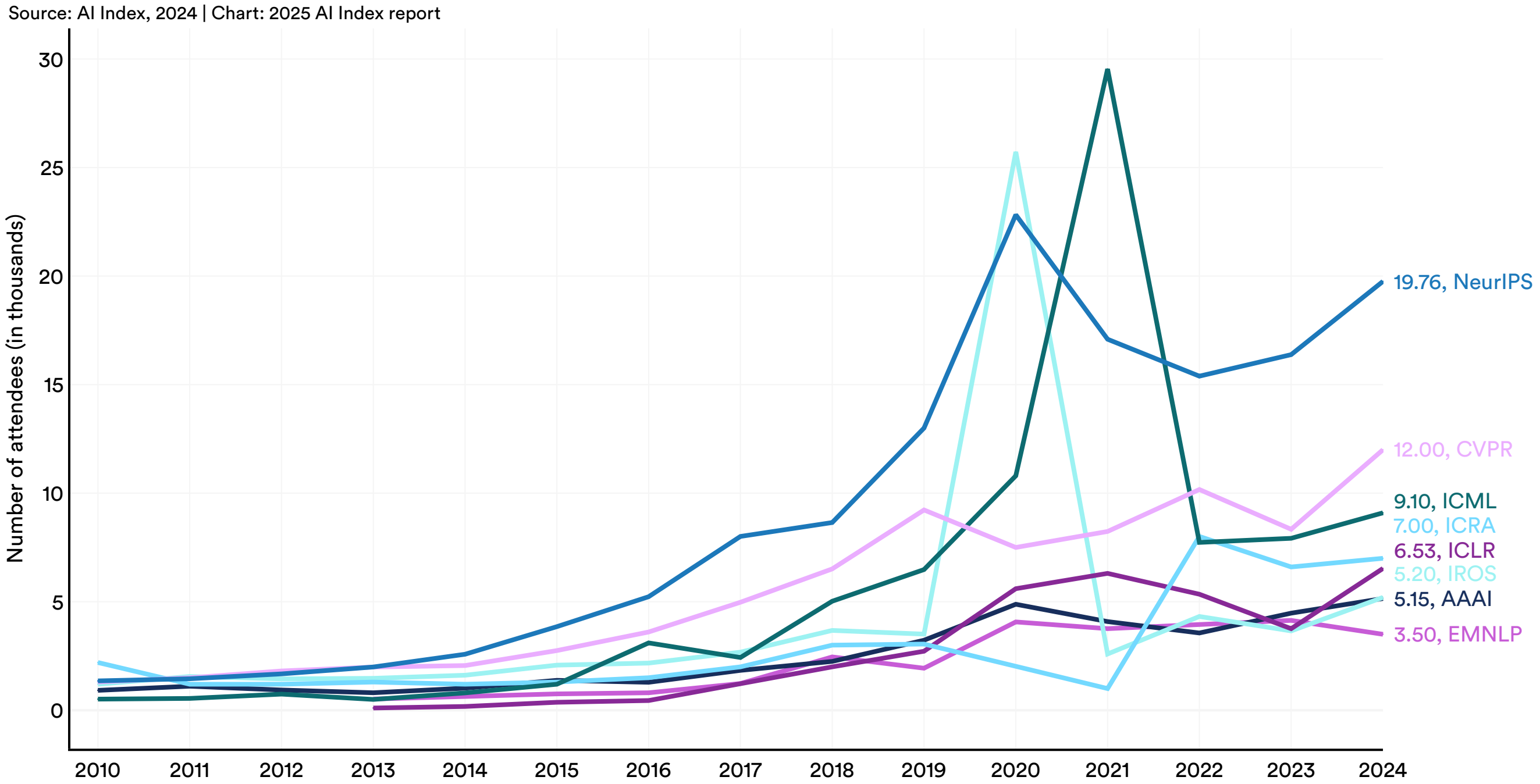

Figure 1.5.2[33]

## Attendance at small conferences, 2010–24

Source: AI Index, 2024 | Chart: 2025 AI Index report

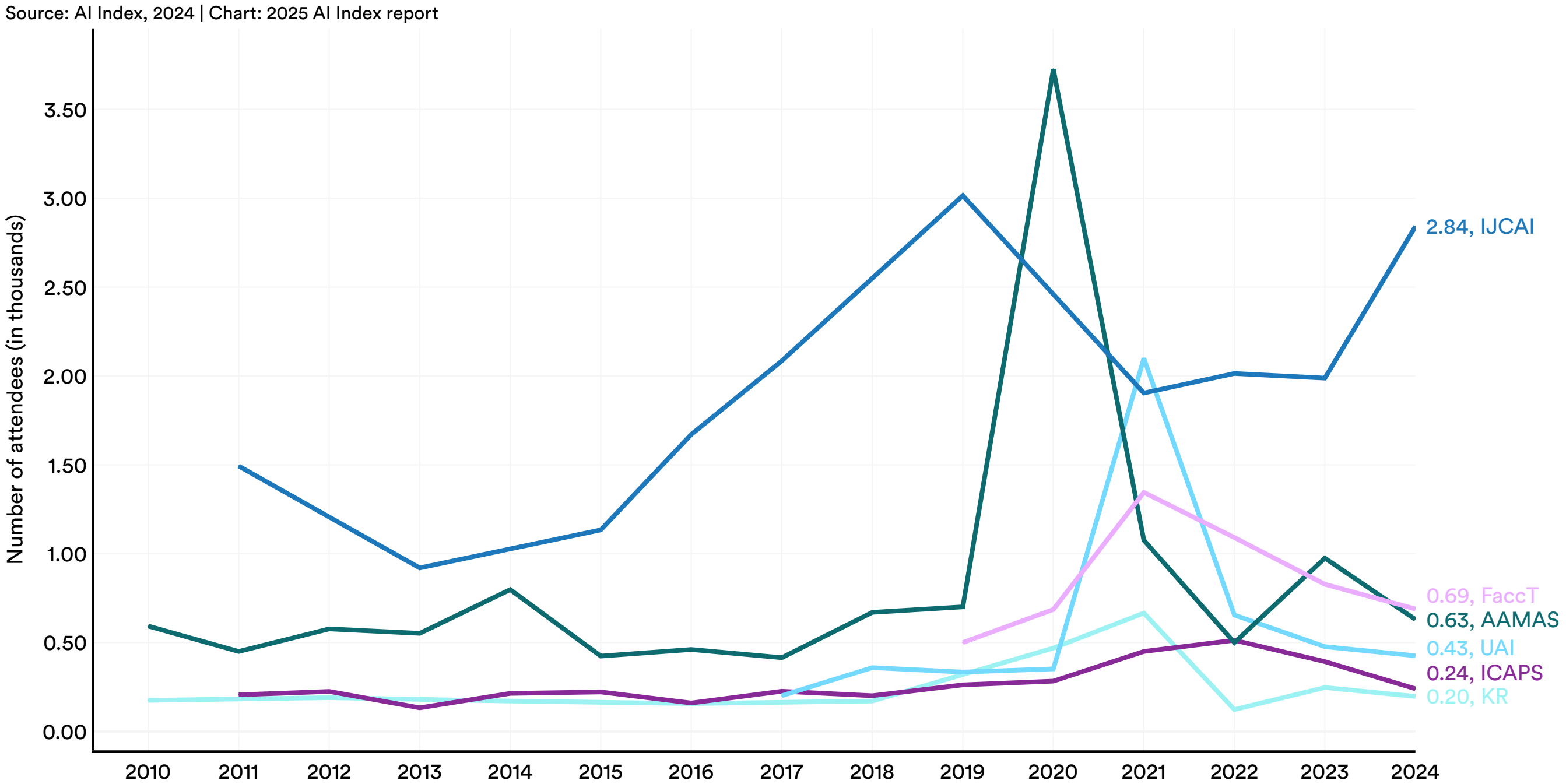

Figure 1.5.3

33 The significant spike in ICML attendance in 2021 was likely due to the conference being held virtually that year.

GitHub is a web-based platform that enables individuals and teams to host, review, and collaborate on code repositories. Widely used by software developers, GitHub facilitates code management, project collaboration, and open-source software support. This section draws on data from GitHub that provides insights into broader trends in open-source AI software development not reflected in academic publication data.[34]

# 1.6 Open-Source AI Software

## Projects

A GitHub project comprises a collection of files, including source code, documentation, configuration files, and images, that together make up a software project. Figure 1.6.1 looks at the total number of GitHub AI projects over time.[35] Since 2011, the number of AI-related GitHub projects has consistently increased, growing from 1,549 in 2011 to approximately 4.3 million in 2024. Notably, there was a sharp 40.3% rise in the total number of GitHub AI projects in the last year alone.

**Number of GitHub AI projects, 2011–24**

Source: GitHub, 2024 | Chart: 2025 AI Index report

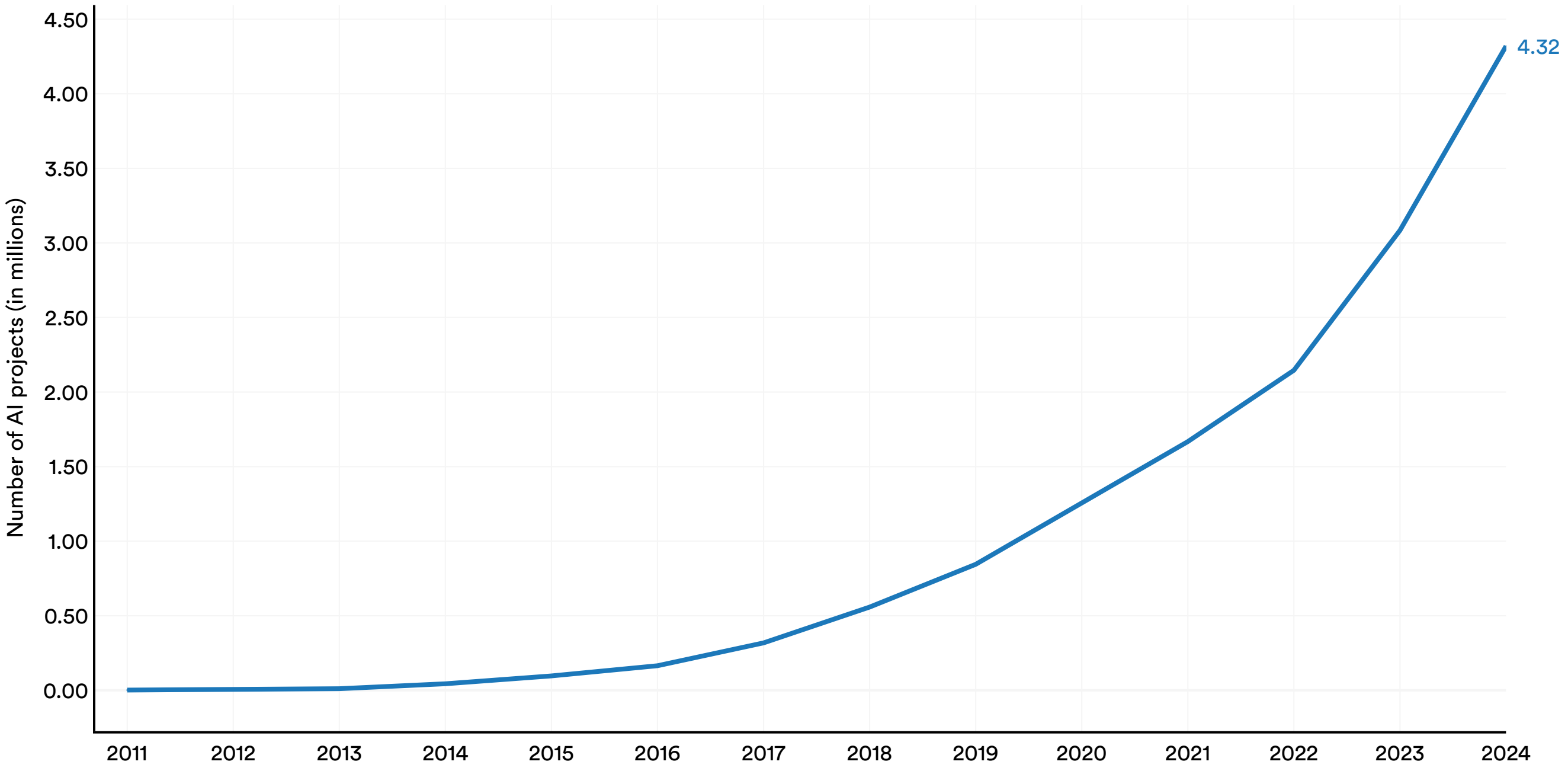

Figure 1.6.1

34 This year, GitHub updated its methodology to capture a broader range of AI-related topics, including more recent developments. As a result, the figures in this year's AI Index may not align with those from previous editions. Chinese researchers often use alternative sites to GitHub for code sharing, such as Gitee and GitCode, but the data from those sites is not included in this report. A full methodological description is available in the Appendix.

35 GitHub used AI-topic classification methods to identify AI-related repositories. Details on the methodology are available in the Appendix.

Figure 1.6.2 reports GitHub AI projects by geographic area since 2011. As of 2024, a significant share of GitHub AI projects were located in the United States, accounting for 23.4% of contributions. India was the second largest contributor with 19.9%, followed closely by Europe, which accounted for 19.5%. Notably, the share of open-source AI projects on GitHub from U.S.-based developers has declined since 2016 and appears to have stabilized in recent years.

**GitHub AI projects (% of total) by geographic area, 2011–24**
Source: GitHub, 2024 | Chart: 2025 AI Index report

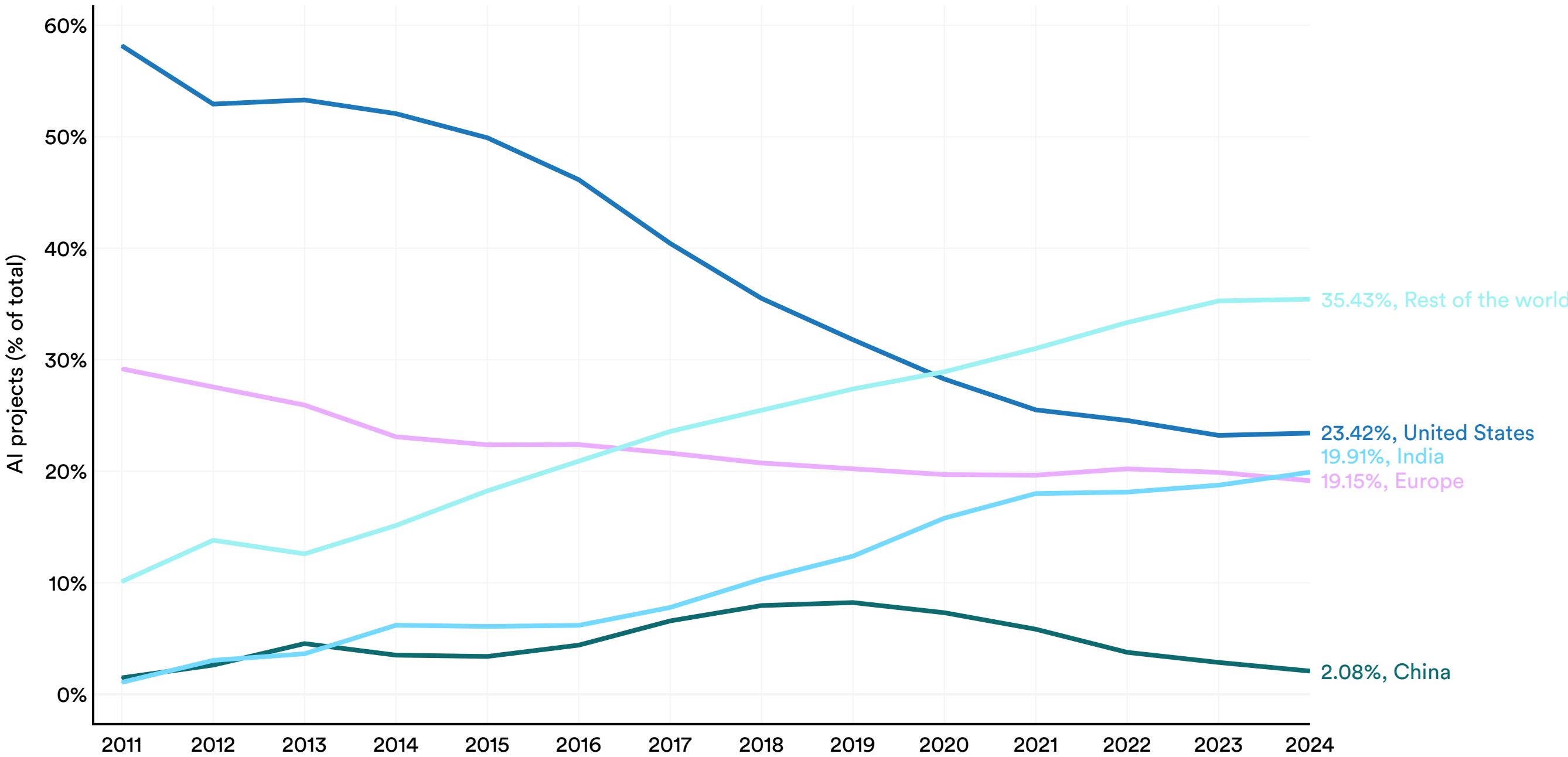

Figure 1.6.2

## Stars

GitHub users can show their interest in a repository by "starring" it, a feature similar to liking a post on social media, which signifies support for an open-source project. Among the most starred repositories are libraries such as TensorFlow, OpenCV, Keras, and PyTorch, which enjoy widespread popularity among software developers in the broader developer community beyond AI. TensorFlow, Keras, and PyTorch are popular libraries for building and deploying machine learning models, while OpenCV offers a variety of tools for computer vision, such as object detection and feature extraction.

The total number of stars for AI-related projects on GitHub continued to rise last year, increasing from 14.0 million in 2023 to 17.7 million in 2024 (Figure 1.6.3).[36] This follows a particularly sharp rise from 2022 to 2023, when the total more than doubled.

**Number of GitHub stars in AI projects, 2011–24**
Source: GitHub, 2024 | Chart: 2025 AI Index report

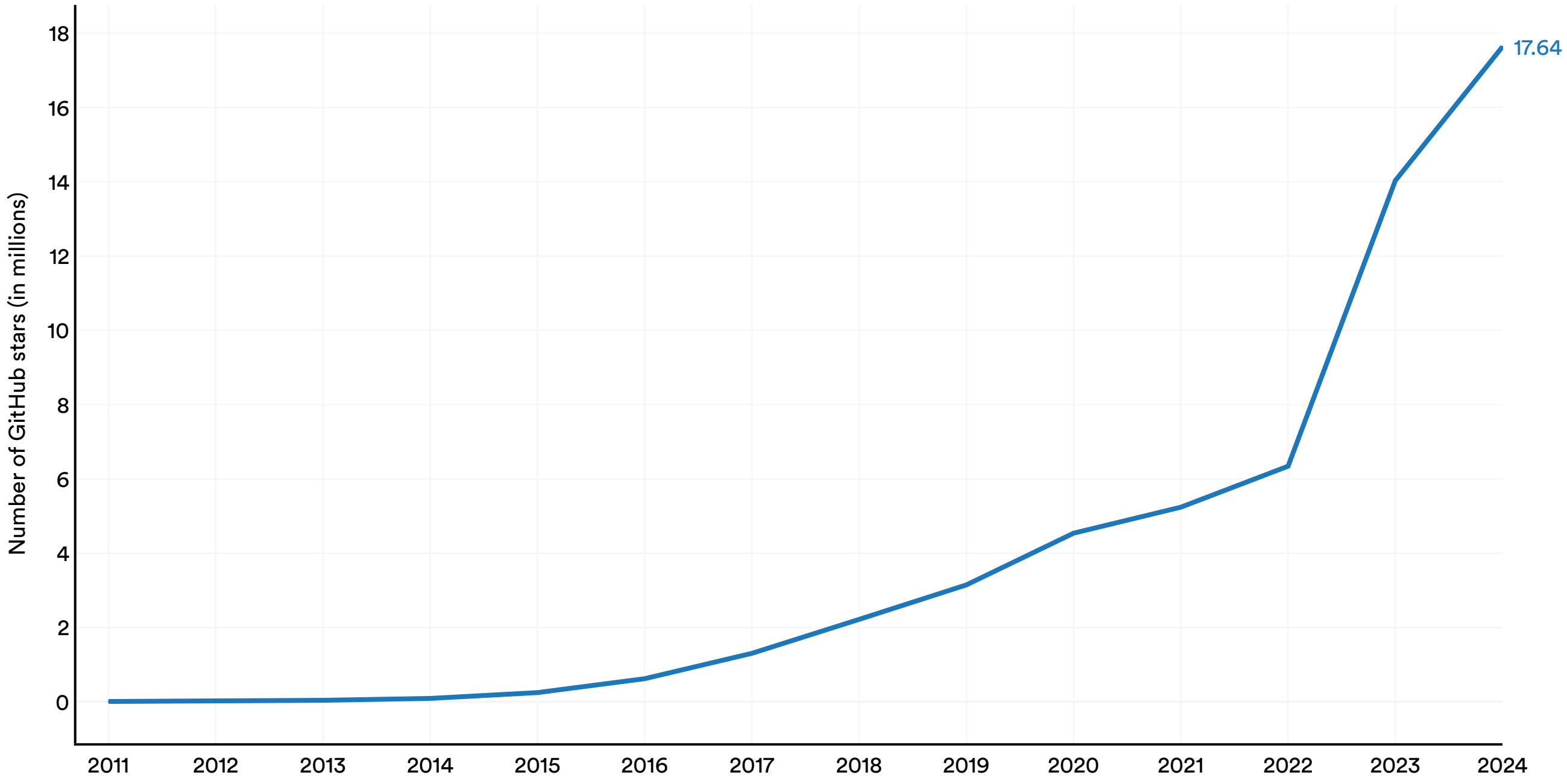

Figure 1.6.3

36 Figure 1.6.3 shows new stars given to GitHub projects within a year, not the total accumulated over time.

In 2024, the United States led in receiving the highest number of GitHub stars, totaling 21.1 million (Figure 1.6.4). All major geographic regions sampled, including Europe, China, and India, saw a year-over-year increase in the total number of GitHub stars awarded to projects located in their countries.

**Number of GitHub stars by geographic area, 2011–24**
Source: GitHub, 2024 | Chart: 2025 AI Index report

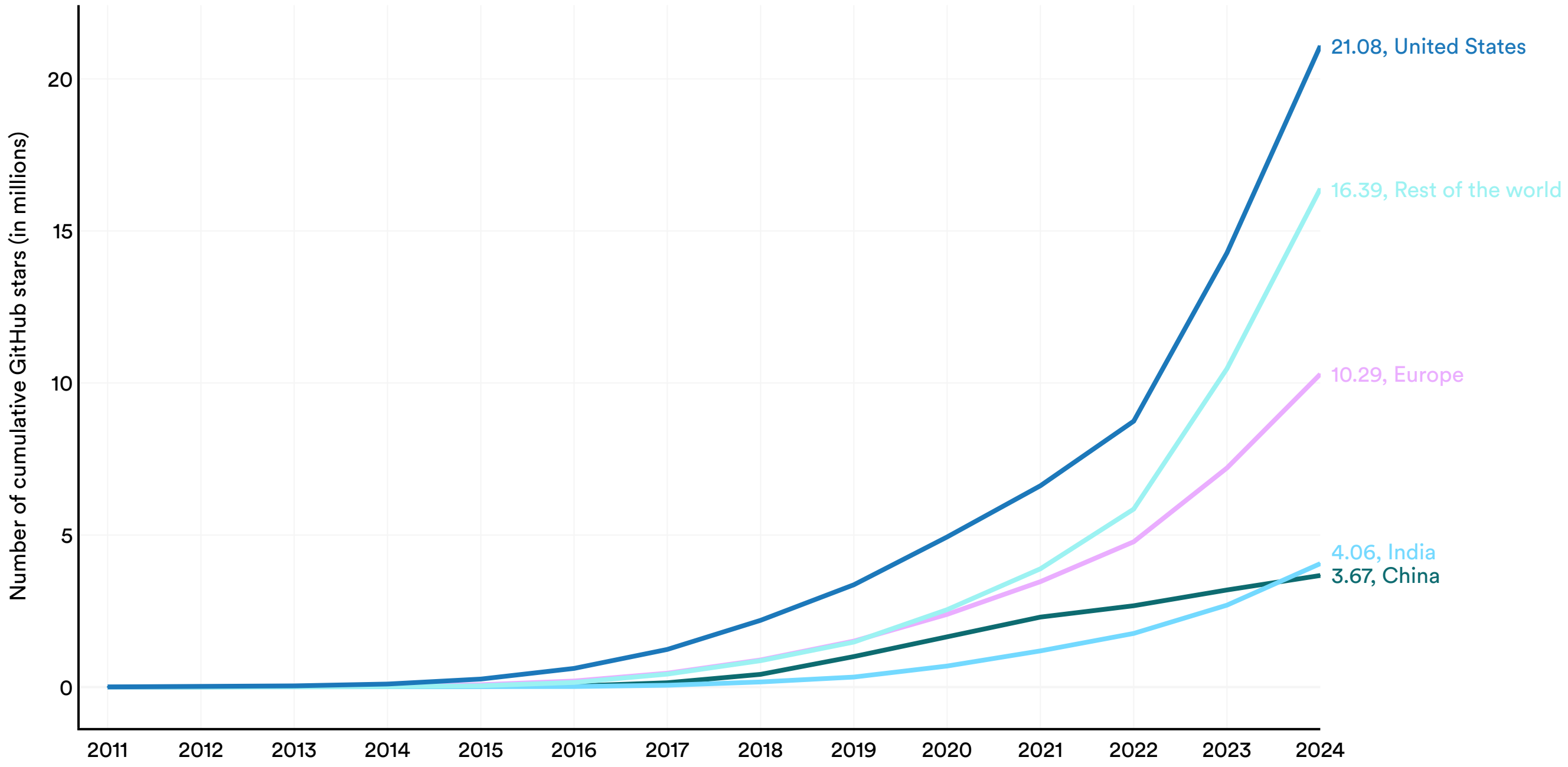

Figure 1.6.4

# CHAPTER 2:
Technical Performance

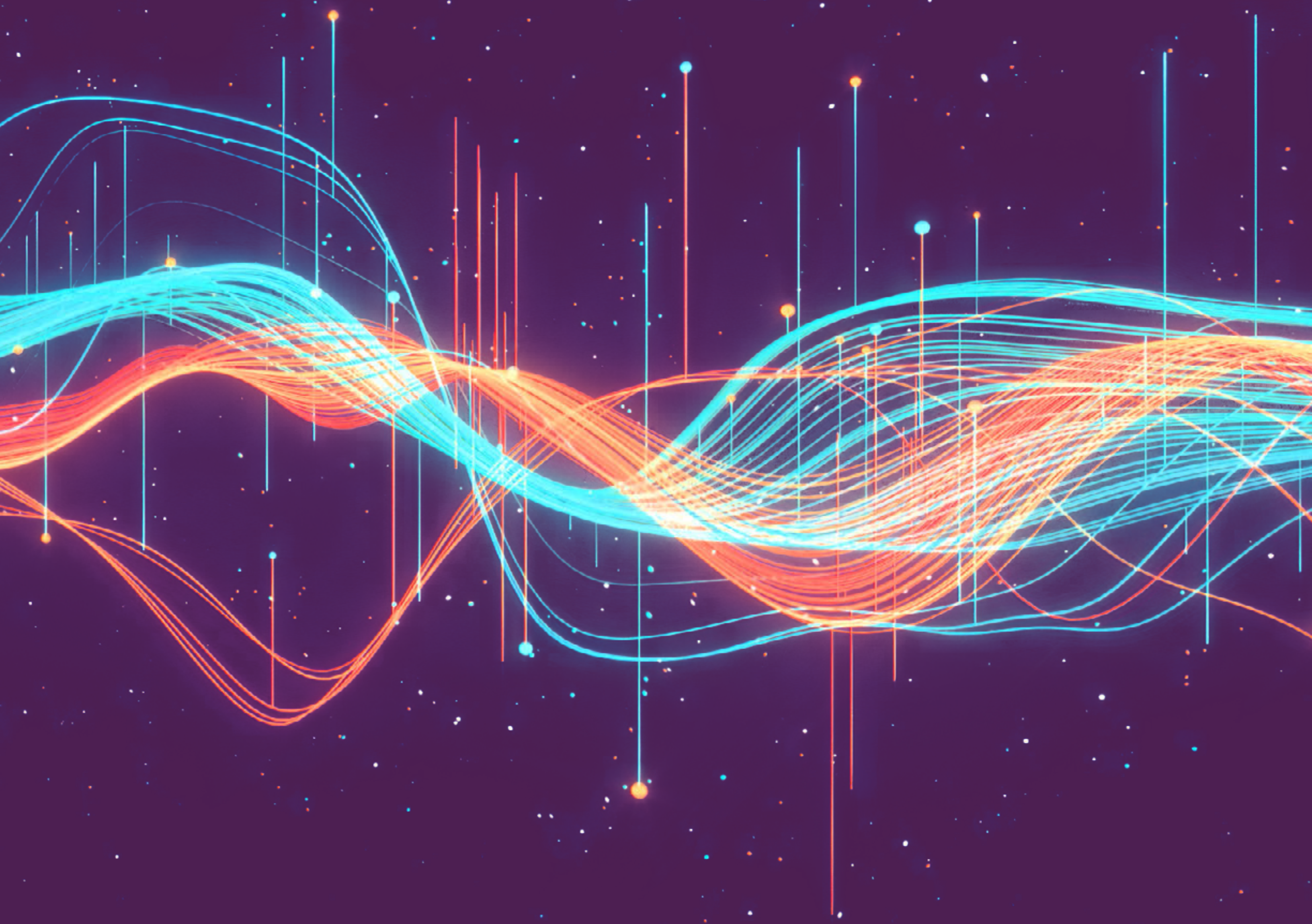

# Chapter 2: Technical Performance

# Chapter 2: Technical Performance (cont'd)

ACCESS THE PUBLIC DATA

**CHAPTER 2:**
Technical Performance

# Overview

The Technical Performance section of this year's AI Index provides a comprehensive overview of AI advancements in 2024. It begins with a high-level summary of AI technical progress, covering major AI-related launches, the state of AI capabilities, and key trends—such as the rising performance of open-weight models, the convergence of frontier model performance, and the improving quality of Chinese LLMs. The chapter then examines the current state of various AI capabilities, including language understanding and generation, retrieval-augmented generation, coding, mathematics, reasoning, computer vision, speech, and agentic AI. New this year are significantly expanded analyses of performance trends in robotics and self-driving cars.

# Chapter Highlights

**1. AI masters new benchmarks faster than ever.** In 2023, AI researchers introduced several challenging new benchmarks, including MMMU, GPQA, and SWE-bench, aimed at testing the limits of increasingly capable AI systems. By 2024, AI performance on these benchmarks saw remarkable improvements, with gains of 18.8 and 48.9 percentage points on MMMU and GPQA, respectively. On SWE-bench, AI systems could solve just 4.4% of coding problems in 2023—a figure that jumped to 71.7% in 2024.

---

**2. Open-weight models catch up.** Last year's AI Index revealed that leading open-weight models lagged significantly behind their closed-weight counterparts. By 2024, this gap had nearly disappeared. In early January 2024, the leading closed-weight model outperformed the top open-weight model by 8.04% on the Chatbot Arena Leaderboard. By February 2025, this gap had narrowed to 1.70%.

---

**3. The gap between Chinese and US models closes.** In 2023, leading American models significantly outperformed their Chinese counterparts—a trend that no longer holds. At the end of 2023, performance gaps on benchmarks such as MMLU, MMMU, MATH, and HumanEval were 17.5, 13.5, 24.3, and 31.6 percentage points, respectively. By the end of 2024, these differences had narrowed substantially to just 0.3, 8.1, 1.6, and 3.7 percentage points.

---

**4. AI model performance converges at the frontier.** According to last year's AI Index, the Elo score difference between the top and 10th-ranked model on the Chatbot Arena Leaderboard was 11.9%. By early 2025, this gap had narrowed to just 5.4%. Likewise, the difference between the top two models shrank from 4.9% in 2023 to just 0.7% in 2024. The AI landscape is becoming increasingly competitive, with high-quality models now available from a growing number of developers.

---

**5. New reasoning paradigms like test-time compute improve model performance.** In 2024, OpenAI introduced models like o1 and o3 that are designed to iteratively reason through their outputs. This test-time compute approach dramatically improved performance, with o1 scoring 74.4% on an International Mathematical Olympiad qualifying exam, compared to GPT-4o's 9.3%. However, this enhanced reasoning comes at a cost: o1 is nearly six times more expensive and 30 times slower than GPT-4o.

---

**CHAPTER 2:**
Technical Performance

# Chapter Highlights (cont'd)

**6. More challenging benchmarks are continually proposed.** The saturation of traditional AI benchmarks like MMLU, GSM8K, and HumanEval, coupled with improved performance on newer, more challenging benchmarks such as MMMU and GPQA, has pushed researchers to explore additional evaluation methods for leading AI systems. Notable among these are Humanity's Last Exam, a rigorous academic test where the top system scores just 8.80%; FrontierMath, a complex mathematics benchmark where AI systems solve only 2% of problems; and BigCodeBench, a coding benchmark where AI systems achieve a 35.5% success rate—well below the human standard of 97%.

---

**7. High-quality AI video generators demonstrate significant improvement.** In 2024, several advanced AI models capable of generating high-quality videos from text inputs were launched. Notable releases include OpenAI's SORA, Stable Video 3D and 4D, Meta's Movie Gen, and Google DeepMind's Veo 2. These models produce videos of significantly higher quality compared to those from 2023.

---

**8. Smaller models drive stronger performance.** In 2022, the smallest model registering a score higher than 60% on MMLU was PaLM, with 540 billion parameters. By 2024, Microsoft's Phi-3-mini, with just 3.8 billion parameters, achieved the same threshold. This represents a 142-fold reduction in over two years.

---

**9. Complex reasoning remains a problem.** Even though the addition of mechanisms such as chain-of-thought reasoning has significantly improved the performance of LLMs, these systems still cannot reliably solve problems for which provably correct solutions can be found using logical reasoning, such as arithmetic and planning, especially on instances larger than those they were trained on. This has a significant impact on the trustworthiness of these systems and their suitability in high-risk applications.

---

**10. AI agents show early promise.** The launch of RE-Bench in 2024 introduced a rigorous benchmark for evaluating complex tasks for AI agents. In short time-horizon settings (two-hour budget), top AI systems score four times higher than human experts, but as the time budget increases, human performance surpasses AI—outscoring it two to one at 32 hours. AI agents already match human expertise in select tasks, such as writing Triton kernels, while delivering results faster and at lower costs.

The Technical Performance chapter begins with a high-level overview of significant model releases in 2024 and reviews the current state of AI technical performance.

# 2.1 Overview of AI in 2024

## Timeline: Significant Model and Dataset Releases

As chosen by the AI Index Steering Committee, here are some of the most notable model and dataset releases of 2024.

| Date | Name | Category | Creator(s) | Significance | Image |
|---|---|---|---|---|---|
| Jan 19, 2024 | Stable LM 2 | LLM | Stability AI | Stability's latest language model builds on the original Stable LM, offering enhanced performance. With only 1.6 billion parameters, it is designed to run efficiently on portable devices such as laptops and smartphones. | Figure 2.1.1<br>Source: Wikipedia, 2025 |
| Feb 8, 2024 | Aya Dataset | Dataset | Cohere for AI, Beijing Academy of AI, Cohere, Binghamton University | A collection of 513 million prompt-completion pairs spanning 114 languages, released as part of Cohere's Aya initiative. This paper and its accompanying dataset represent significant milestones in multilingual instruction tuning. | Figure 2.1.2<br>Source: Cohere, 2025 |
| Feb 15, 2024 | Gemini 1.5 Pro | LLM | Google DeepMind | Google's Gemini model set a new benchmark with its 1M token context window, far exceeding GPT-4 Turbo's 128K token limit. |  <br>Figure 2.1.3<br>Source: Google, 2024 |
| Feb 20, 2024 | SDXL-Lightning | Text-to-image | ByteDance | Developed by ByteDance, the creators of TikTok, this model was among the fastest text-to-image systems at its release, generating high-quality synthetic images in under a second. Its speed was achieved through progressive adversarial distillation, unlike other models that rely on diffusion-based techniques. | 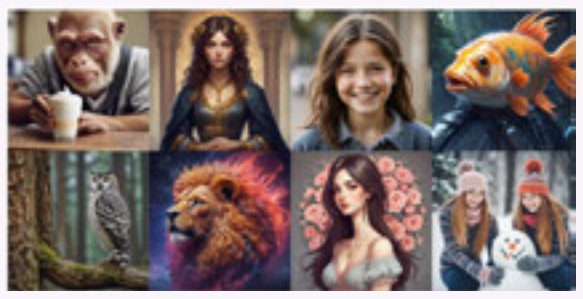<br>Figure 2.1.4<br>Source: Hugging Face, 2025 |
| Mar 4, 2024 | Claude 3 | LLM | Anthropic | Anthropic's latest LLM outperforms GPT-4 and Gemini on nearly all industry benchmarks, reduces incorrect prompt refusals, and delivers significantly higher accuracy. | 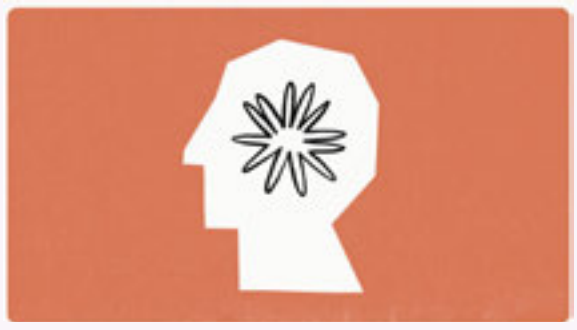<br>Figure 2.1.5<br>Source: Anthropic, 2025 |
| Mar 7, 2024 | Inflection-2.5 | LLM | Inflection AI | Inflection's flagship product, "Pi," featured an exceptional model with GPT-4–level performance while using only 40% of its computing resources. Just two weeks after the model's release, Microsoft acquired Inflection for $650 million. | 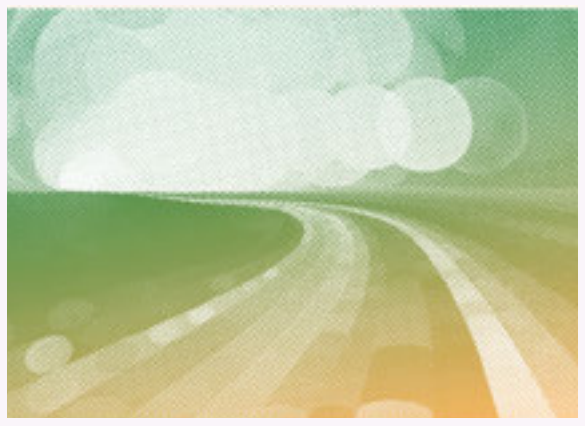 Figure 2.1.6 Source: Inflection, 2025 |
| Mar 19, 2024 | Moirai and LOTSA | Model/ dataset | Salesforce | Salesforce unveils Moirai, a foundation model for universal forecasting, alongside LOTSA—a diverse, large-scale time series dataset with 27 billion observations spanning nine domains. | 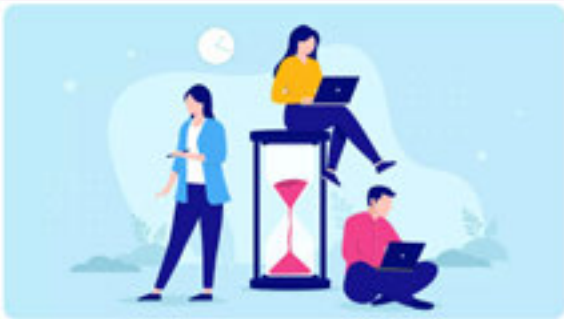 Figure 2.1.7 Source: Salesforce, 2025 |
| Mar 27, 2024 | DBRX | LLM | Databricks | Databricks' open-source mixture-of-experts (MoE) LLM is a fine-grained model, surpassing similar small MoE models like Mixtral and Grok. This transformer decoder-only model features 132B parameters (36B active per input) and was trained on 12 trillion tokens. | 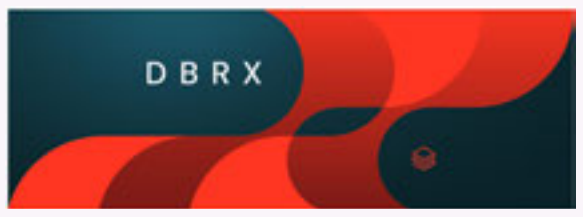  Figure 2.1.8 Source: Databricks, 2025 |
| Apr 2, 2024 | Stable Audio 2 | Text-to-song and song-to-song | Stability AI | The latest version of Stable Audio, Stability's AI-powered song generator, now supports audio-to-audio functionality. Users can upload songs and manipulate them using natural language prompts for seamless customization. | 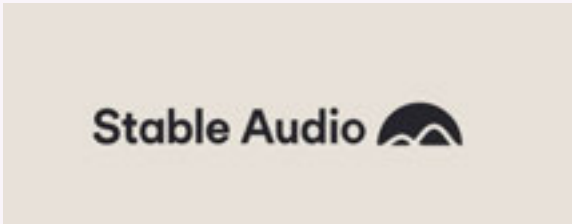  Figure 2.1.9 Source: Stability AI, 2025 |
| Apr 17, 2024 | Llama 3 | LLM | Meta | The Llama 3 series debuts with 8B and 70B parameter text-based models, ranking among the highest performing models of their size to date. | 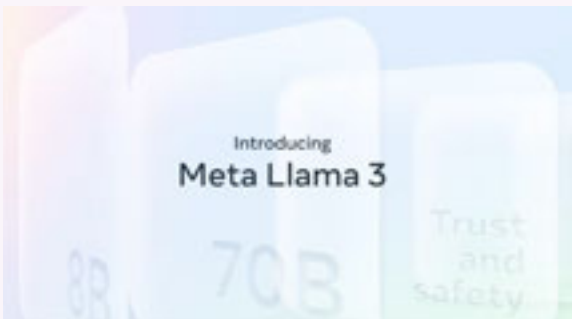  Figure 2.1.10 Source: Meta, 2025 |
| May 13, 2024 | GPT-4o | Multimodal | OpenAI | GPT-4o is a new multimodal model capable of processing inputs in any combination of text, audio, images, and video, and generating outputs in the same formats. It responds to audio in as little as 320 milliseconds, matching human response times. | 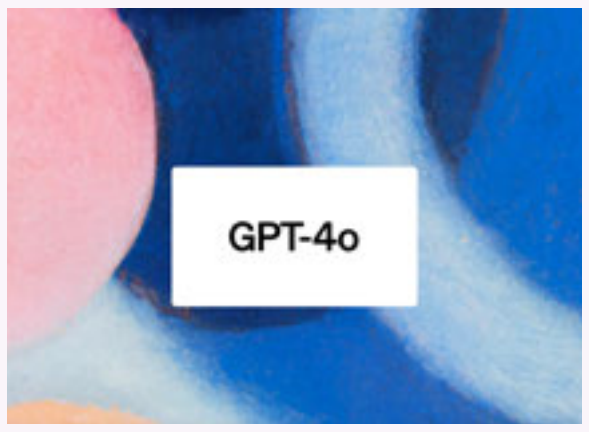  Figure 2.1.11 Source: OpenAI, 2024 |
| Jun 7, 2024 | Qwen2 | LLM | Alibaba | Qwen2, developed by China's Alibaba, is a series of advanced base and instruction-tuned models. These models rival competitors like Llama 3-70B and Mixtral-8x22B in performance across numerous benchmarks. | 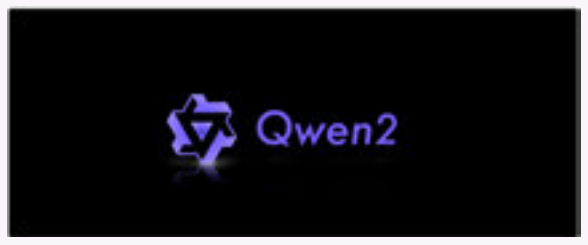  Figure 2.1.12<br>Source: Qwen, 2024 |
| Jun 17, 2024 | Runway Gen-3 | Text-to-video and image-to-video | Runway | Runway's upgraded video generation model sets a new standard for the field, particularly excelling in creating photorealistic humans with vivid and expressive emotionality. | 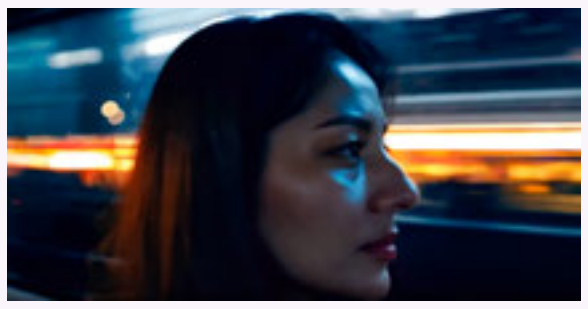 Figure 2.1.13<br>Source: Runway, 2024 |
| Jul 23, 2024 | Llama 3.1 405B | LLM | Meta | Meta has released its largest model to date, the final in the Llama 3.1 family, featuring 405B parameters. Upon its release, it became the most capable openly available foundation model, rivaling many closed models across a variety of benchmarks. | 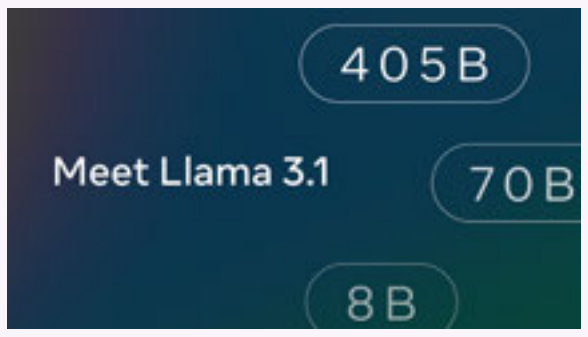  Figure 2.1.14<br>Source: Meta, 2024 |
| Aug 12, 2024 | Falcon Mamba | LLM | Technology Innovation Institute in Abu Dhabi | A powerful new 7B parameter model, built on the Mamba State Space Language Model (SSLM) architecture, enables Falcon—one of the few government-created AI models—to dynamically adjust parameters and filter out irrelevant inputs, making it more efficient than transformer-based models. | 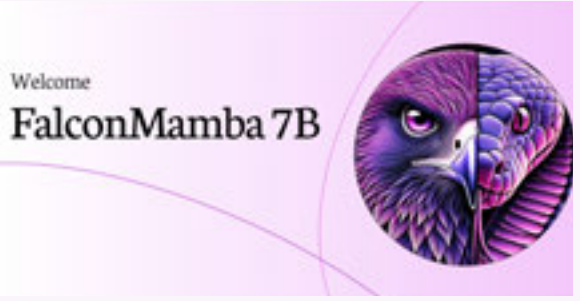  Figure 2.1.15<br>Source: Hugging Face, 2025 |
| Aug 13, 2024 | Grok-2 | Text-to-text and text-to-image | xAI | Developed by xAI, Grok is an advanced text- and image-generation model that excels in image creation, advanced reasoning, and problem-solving. Its launch was particularly notable, as it quickly rivaled the performance of leading models despite xAI being founded only in March 2023. | 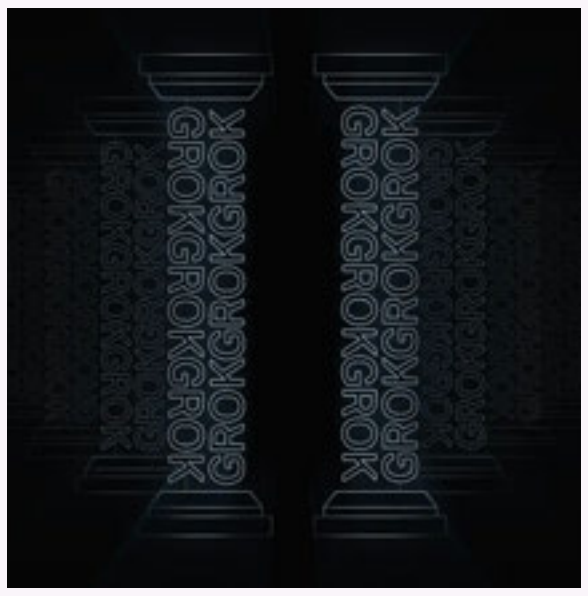 Figure 2.1.16<br>Source: xAI, 2025 |
| Aug 15, 2024 | Imagen 3 | Text-to-image | Google Labs | Google's updated AI image generator achieves the highest Elo score on the GenAI-Bench image benchmark, setting a new standard for quality in AI-generated visuals. | 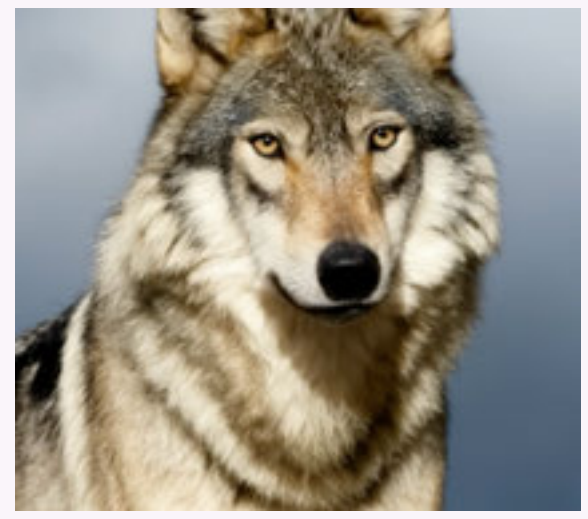 Figure 2.1.17<br>Source: Google, 2025 |
| Aug 22, 2024 | Jamba 1.5 | LLM | AI21 Labs | The first LLM to combine state-space models with transformers, delivering high-quality results for text-based applications. This hybrid approach significantly enhances speed while preserving the quality of outputs. | 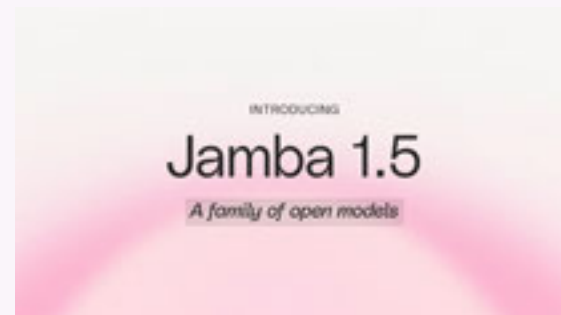  Figure 2.1.18<br>Source: AI21, 2025 |
| Aug 29, 2024 | SynthID v2 | Tool | Google | SynthID v2 is the updated version of SynthID, Google's watermarking and identification software. It now supports AI-generated content across images, video, audio, and text, and offers enhanced tracking and verification capabilities. | 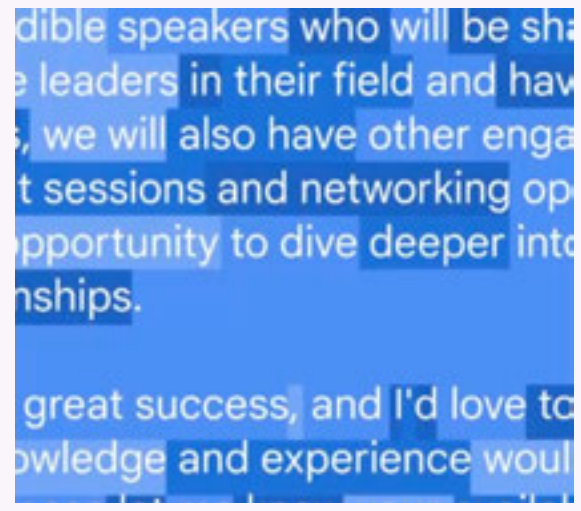 Figure 2.1.19<br>Source: Google, 2025 |
| Sep 11, 2024 | NotebookLM Podcast Tool | Text-to-podcast | Google Labs | The second end-to-end AI podcast generator to hit the market, following Synthpod, went viral. It gained popularity among students leveraging NotebookLM for studying and tech employees using it to listen to AI-generated summaries. | 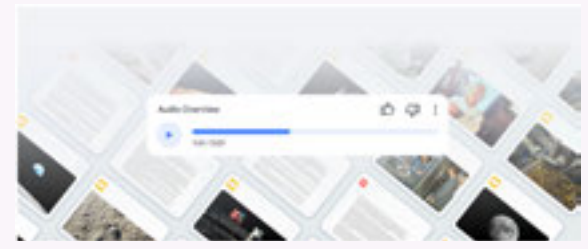 Figure 2.1.20<br>Source: Google, 2025 |
| Sep 12, 2024 | o1-preview | Language, math, biology | OpenAI | OpenAI's first model in the "o series" is designed for advanced reasoning and tackling complex tasks. It is significantly more powerful than GPT, particularly in math, science, and coding. | 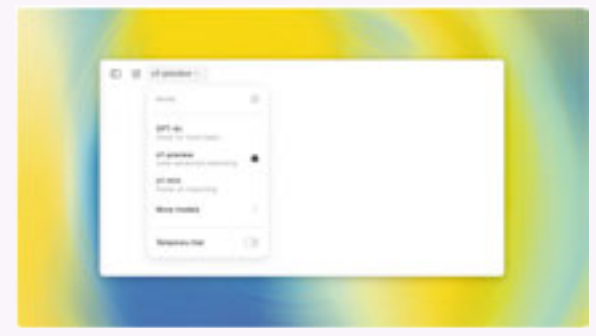 Figure 2.1.21<br>Source: OpenAI, 2025 |
| Sep 17, 2024 | NVLM (D, H, X) | Vision, language | Nvidia | Nvidia released three open-access models for vision-language tasks, achieving top scores on OCRBench (for optical character recognition) and VQAv2 (for natural language understanding). | 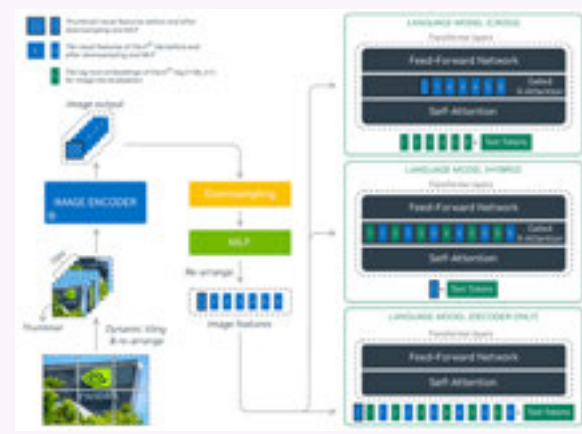 Figure 2.1.22<br>Source: Dai et al., 2024 |
| Sep 19, 2024 | Qwen2.5 | LLM | Alibaba | Qwen2.5, the latest series of foundation models from Chinese e-commerce giant Alibaba, includes a range of efficient smaller models and specialized coding and math models designed for targeted functionality. | 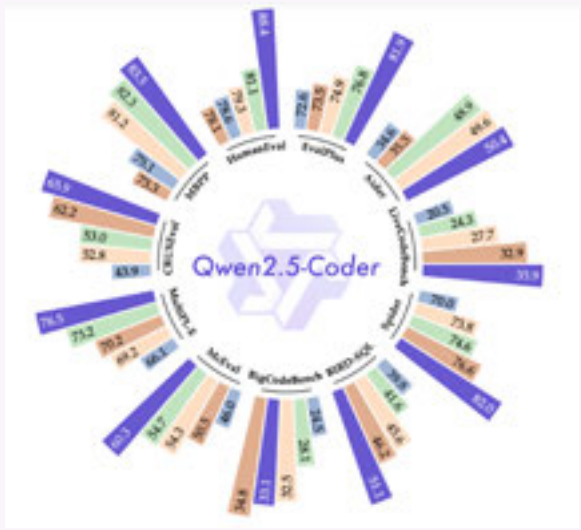<br>Figure 2.1.23<br>Source: Qwen, 2025 |
| Oct 16, 2024 | Ministral | LLM | Mistral | Ministral is a pair of compact models (3B and 8B parameters) that outperformed Gemma and Llama models of similar size across all major industry-recognized benchmarks. | 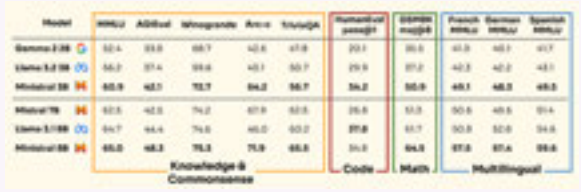<br>Figure 2.1.24<br>Source: Mistral, 2025 |
| Oct 22, 2024 | Anthropic Computer Use | Agentic Capability | Anthropic | Anthropic Computer Use is a groundbreaking computer control feature for Claude 3.5 Sonnet users, allowing Claude to move the cursor, type, and autonomously complete tasks on the user's computer in real time. | 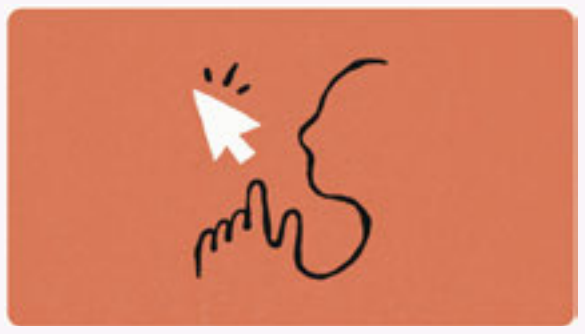<br>Figure 2.1.25<br>Source: Anthropic, 2025 |
| Oct 28, 2024 | Apple Intelligence | iPhone feature | Apple | Apple's suite of AI-powered features includes Image Playground (for image creation), Genmoji (for custom emoji creation), Siri integration with ChatGPT, and more. | 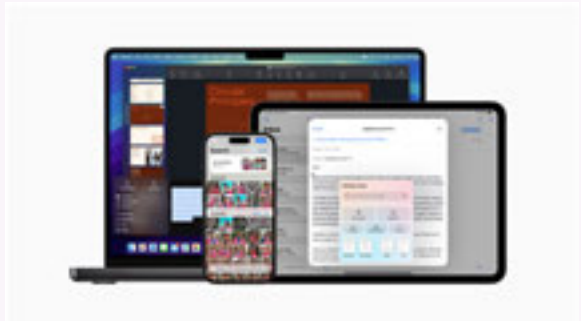<br>Figure 2.1.26<br>Source: Apple, 2025 |
| Dec 3, 2024 | Nova Pro | Multimodal | Amazon | Nova Pro is the most powerful model in Amazon Web Services' Nova family, capable of processing both visual and textual information. It especially excels at analyzing financial documents. | 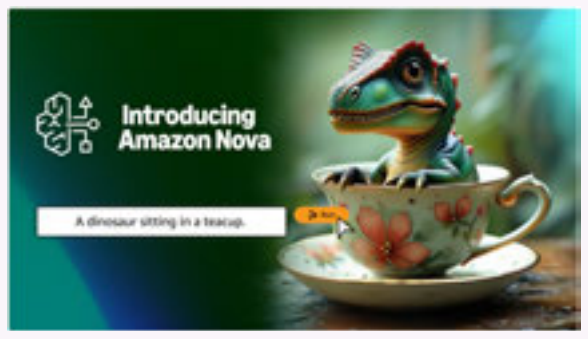<br>Figure 2.1.27<br>Source: Amazon, 2025 |
| Dec 11, 2024 | Gemini 2 | LLM | Google DeepMind | The improved version of Gemini, Google's LLM, now includes computer control along with image and audio generation capabilities. It is twice as fast as Gemini 1.5 Pro and offers significantly enhanced performance in coding and image analysis. | 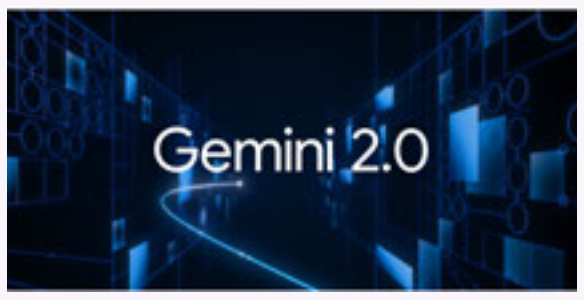<br>Figure 2.1.28<br>Source: Google, 2025 |
| Dec 12, 2024 | Sora | Text-to-video | OpenAI | OpenAI's highly anticipated video generation model can create videos up to 20 seconds long at 1080p resolution for ChatGPT Pro users (and five seconds at 720p for ChatGPT Plus users). Sora demos had been circulating at tech meetups since early 2024, but OpenAI delayed the official release to improve model safety. | 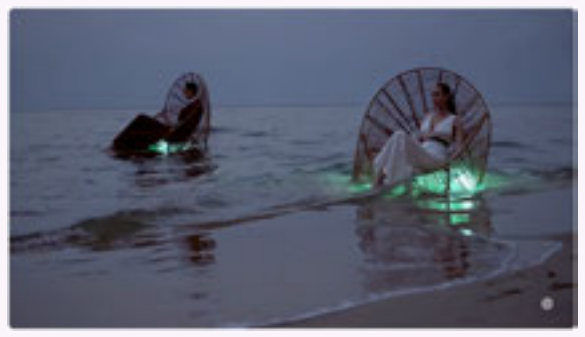 Figure 2.1.29<br>Source: OpenAI, 2025 |
| Dec 13, 2024 | Global MMLU | Dataset | Cohere | A multilingual evaluation set featuring professionally translated MMLU questions across 42 languages, designed to serve as a more global AI benchmark. It evaluates AI performance in diverse languages while addressing Western biases in the original MMLU dataset, where an estimated 28% of questions rely on Western cultural knowledge. | 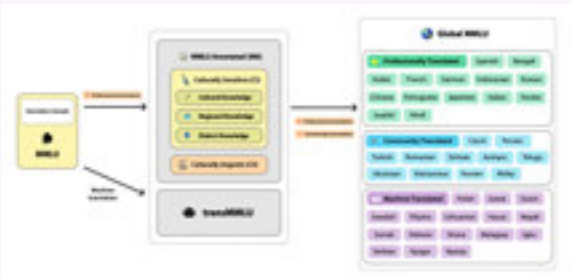 Figure 2.1.30<br>Source: Singh et al., 2025 |
| Dec 20, 2024 | o3 (beta) | Multimodal | OpenAI | OpenAI's newest frontier model, released for safety testing by AI researchers, outperforms all previous models in SWE, competition code, competition math, PhD-level science, and research math benchmarks. It also set a new record on the ARC-AGI benchmark, achieving 87.5% on the ARC Prize team's private holdout set. | 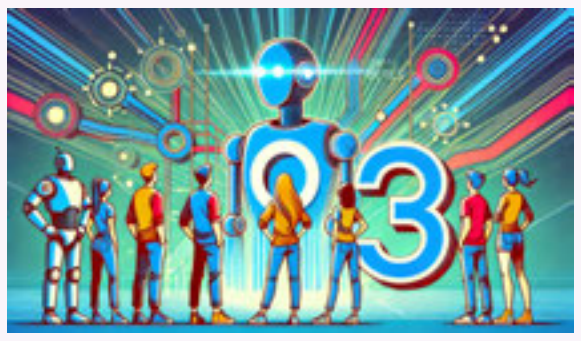 Figure 2.1.31<br>Source: VentureBeat, 2025 |
| Dec 27, 2024 | DeepSeek-V3 | LLM | DeepSeek | DeepSeek V3, an open-source model developed with significantly fewer computing resources than state-of-the-art models, outperforms leading models on benchmarks like MMLU and GPQA. | deepseek<br>Figure 2.1.32<br>Source: Dirox, 2025 |

# State of AI Performance

In this section, the AI Index offers a high-level view into major AI trends that occurred in 2024.

## Overall Review

Last year's AI Index highlighted that AI had already surpassed human performance across many tasks, with only a few exceptions, such as competition-level mathematics and visual commonsense reasoning. Over the past year, AI systems have continued to improve, exceeding human performance on several of these previously challenging benchmarks.

Figure 2.1.33 illustrates the progress of AI systems relative to human baselines for eight AI benchmarks corresponding to 11 tasks (e.g., image classification or basic-level reading comprehension).[1] The AI Index team selected one benchmark to represent each task. This year, the AI Index team added newly released benchmarks, such as GPQA Diamond and MMMU, to showcase the progress of AI systems in tackling extremely challenging cognitive tasks.

**Select AI Index technical performance benchmarks vs. human performance**
Source: AI Index, 2025 | Chart: 2025 AI Index report

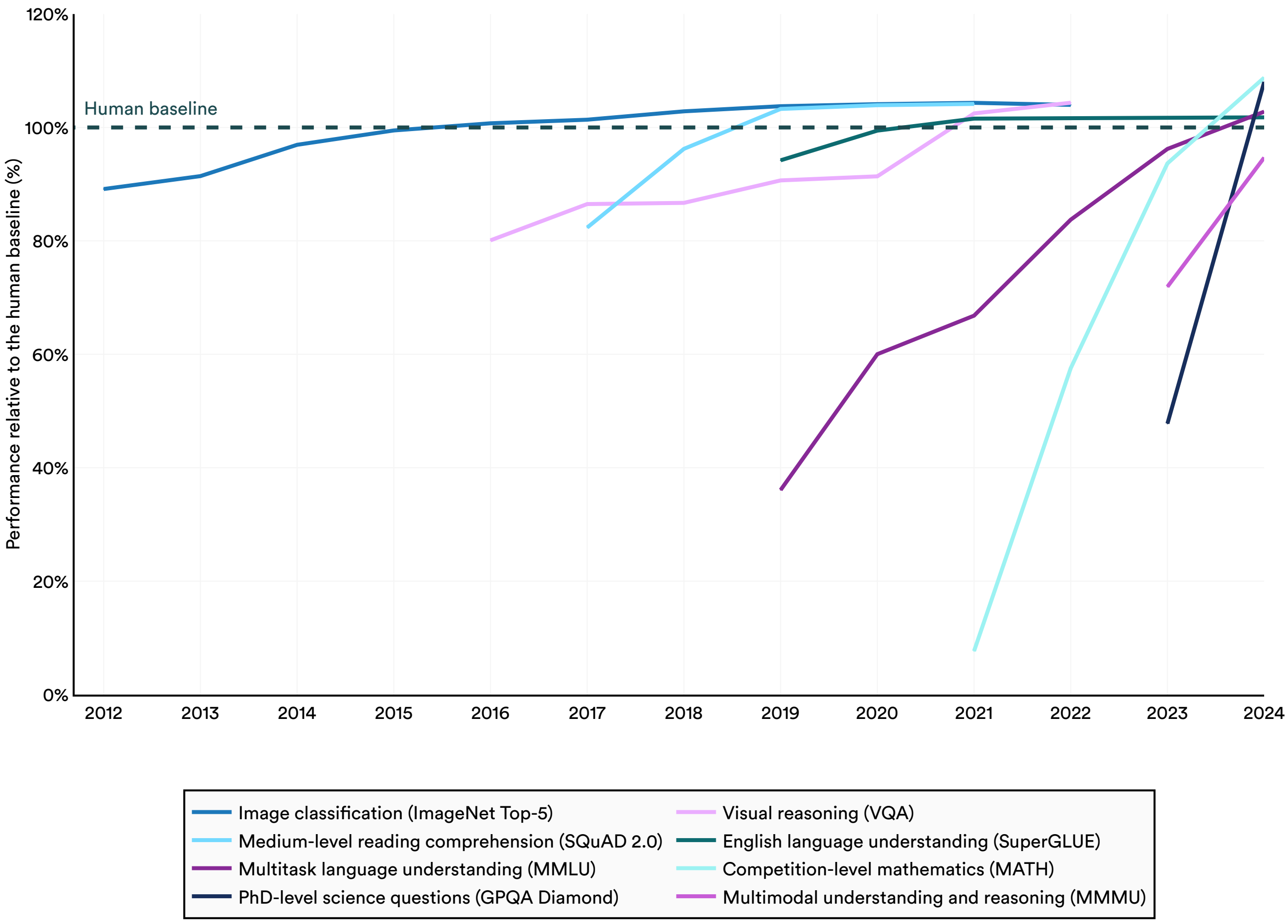

Figure 2.1.33[2]

1 An AI benchmark is a standardized test used to evaluate the performance and capabilities of AI systems on specific tasks. For example, ImageNet is a canonical AI benchmark that features a large collection of labeled images, and AI systems are tasked with classifying these images accurately. Tracking progress on benchmarks has been a standard way for the AI community to monitor the advancement of AI systems.

2 In Figure 2.1.33, the values are scaled to establish a standard metric for comparing different benchmarks. The scaling function is calibrated such that the performance of the best model for each year is measured as a percentage of the human baseline for a given task. A value of 105% indicates, for example, that a model performs 5% better than the human baseline

As of 2024, there are very few task categories where human ability surpasses AI. Even in these areas, the performance gap between AI and humans is shrinking rapidly. For example, on MATH, a benchmark for competition-level mathematics, state-of-the-art AI systems are now 7.9 percentage points ahead of human performance, a significant improvement from the 0.3-point gap in 2024.[3] Similarly, on MMMU, a benchmark for complex, multidisciplinary, expert-level questions, the best 2024 model, o1, scored 78.2%, only 4.4 points below the human benchmark of 82.6%. Conversely, at the end of 2023, Google Gemini scored 59.4%, further illustrating the rapid advancements in AI performance on cognitively demanding tasks.

### Closed vs. Open-Weight Models

AI models can be released with different levels of openness. Certain models, like Google's Med-Gemini, remain entirely closed, accessible only to their developers. Meanwhile, models such as OpenAI's GPT-4o and Anthropic's Claude 3.5 provide limited public access through APIs. However, weights for these models are not released, preventing independent modification or thorough public scrutiny. In contrast, weights for Meta's Llama 3.3 and Stable Video 4D are fully available, allowing anyone to modify and use them freely.[4]

Perspectives on open versus closed-weight AI models are sharply divided. Advocates of open-weight models highlight their potential to reduce market monopolies, spur innovation, improve security and robustness, and enhance transparency within the AI ecosystem. For example, Meta's Llama models have been leveraged to create tools like Meditron, power military applications, and drive the development of numerous open-weight models worldwide. However, critics warn that open-weight models pose significant security risks, including the spread of disinformation and the creation of bioweapons, arguing for a more cautious and controlled approach.

Last year's AI Index highlighted a notable performance gap between closed and open-weight LLM models. Figure 2.1.34 illustrates the performance trends of the top closed-weight and open-weight LLMs on the Chatbot Arena Leaderboard, a public platform for benchmarking LLM performance. In early January 2024, the leading closed-weight model outperformed the top open-weight model by 8.0%. By February 2025, this gap had narrowed to 1.7%.

The same trend is evident across other question-answering benchmarks. In 2023, closed-weight models consistently outperformed open-weight counterparts on nearly every major benchmark—MMLU, HumanEval, MMMU, and MATH. However, by 2024, the gap had narrowed significantly (Figure 2.1.35). For instance, in late 2023, closed-weight models led open models on MMLU by 15.9 points, but by the end of 2024, that difference had shrunk to just 0.1 percentage point. This rapid improvement was largely driven by Meta's summer release of Llama 3.1, followed by the launch of other high-performing open-weight models, such as DeepSeek's V3.

3 The benchmark data in this figure, along with those in other sections of this chapter, was collected in early January 2025. Since the publication of the AI Index, individual benchmark scores may have improved.

4 In the software community, "open source" refers to software released under a license that grants users the right to use, study, modify, and distribute both the software and its source code freely. Open-weight models, though more accessible than closed-weight models, are not necessarily fully open source, as the underlying code or training data is often withheld.

**Performance of top closed vs. open models on LMSYS Chatbot Arena**
Source: LMSYS, 2025 | Chart: 2025 AI Index report

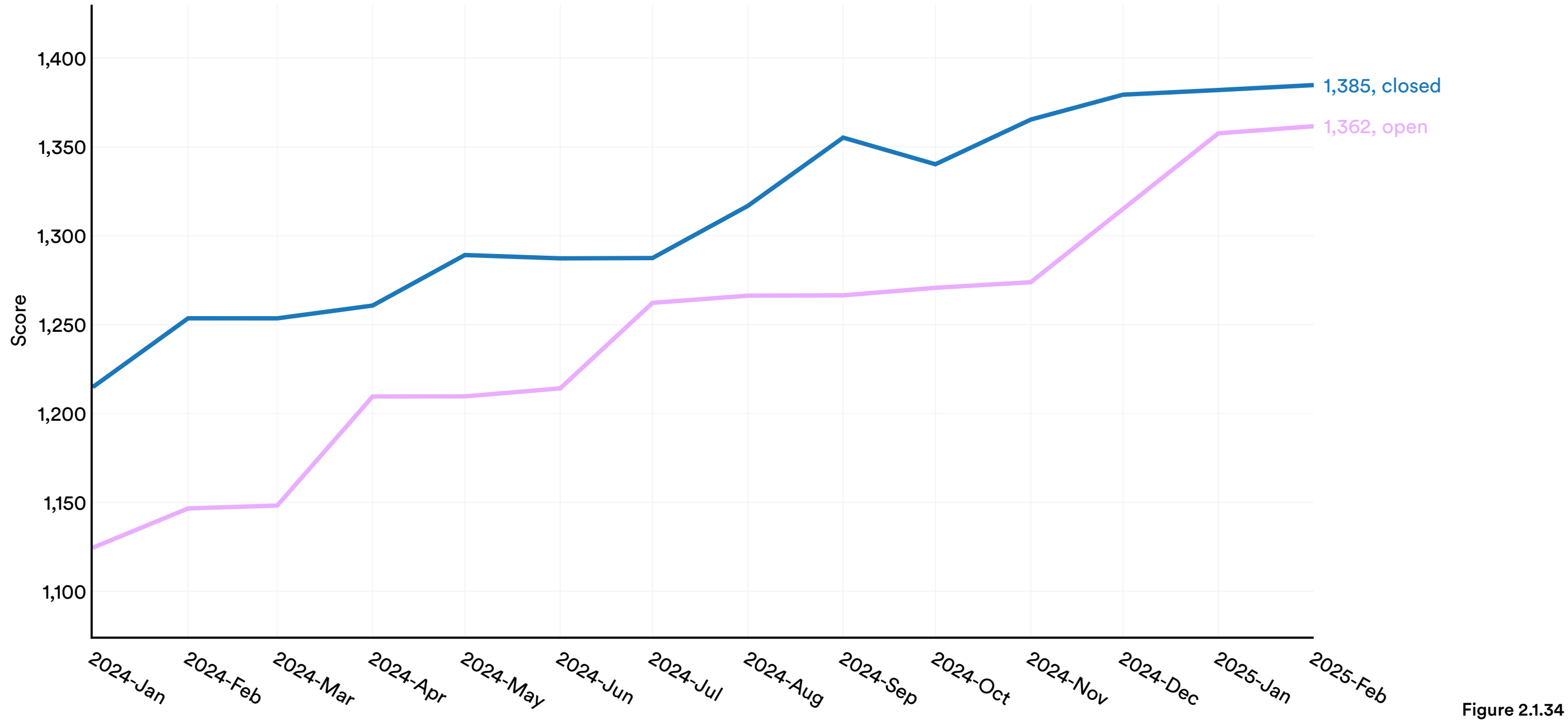

Figure 2.1.34

**Performance of top closed vs. open models on select benchmarks**
Source: AI Index, 2025 | Chart: 2025 AI Index report

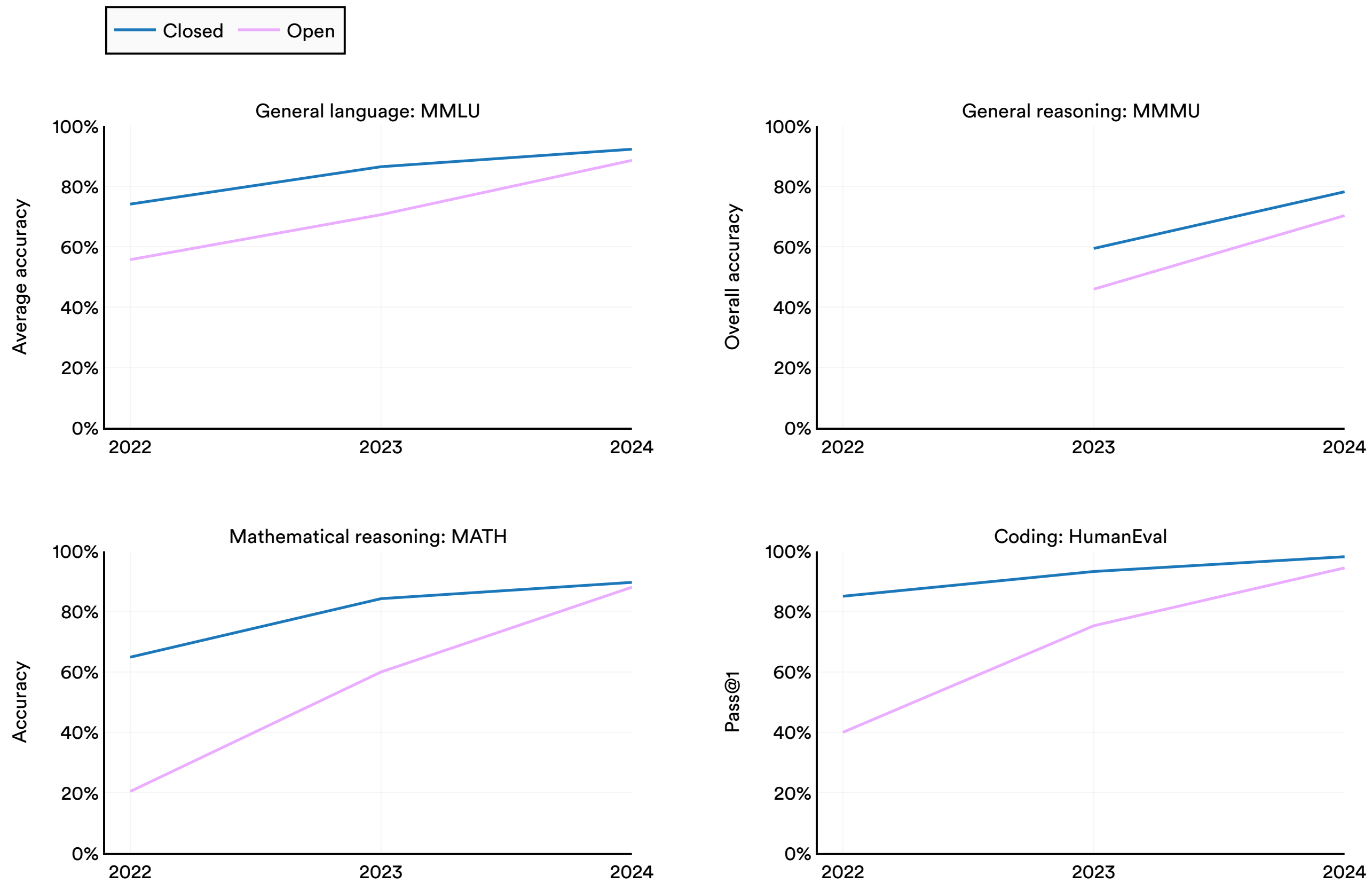

Figure 2.1.35

## US vs. China Technical Performance

The United States has historically dominated AI research and model development, with China consistently ranking second. Recent evidence, however, suggests the landscape is rapidly changing and that China-based models are catching up to their U.S. counterparts.

In 2023, leading American models significantly outperformed their Chinese counterparts. On the LMSYS Chatbot Arena, the top U.S. model outperformed the best Chinese model by 9.3% in January 2024. By February 2025, this gap had narrowed to just 1.7% (Figure 2.1.36). At the end of 2023, on benchmarks such as MMLU, MMMU, MATH, and HumanEval, the performance gaps were 17.5, 13.5, 24.3, and 31.6 percentage points, respectively (Figure 2.1.37). By the end of 2024, these differences had narrowed significantly to just 0.3, 8.1, 1.6, and 3.7 percentage points. The launch of DeepSeek-R1 garnered attention for another reason: The company reported achieving its results using only a fraction of the hardware resources typically required to train such a model. Beyond impacting U.S. stock markets, DeepSeek's R1 launch raised doubts about the effectiveness of U.S. semiconductor export controls.

**Performance of top United States vs. Chinese models on LMSYS Chatbot Arena**
Source: LMSYS, 2025 | Chart: 2025 AI Index report

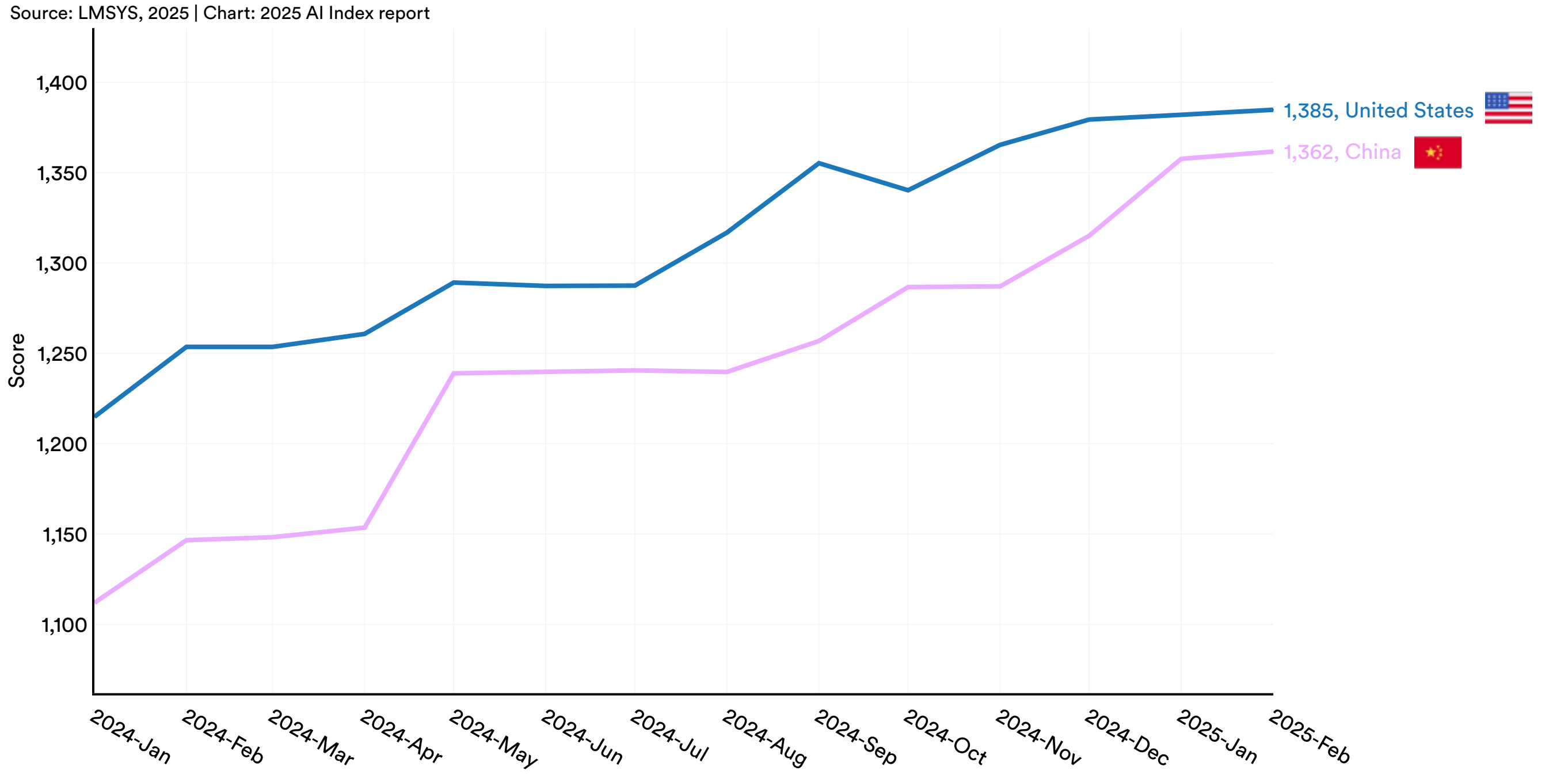

Figure 2.1.36

### Performance of top United States vs. Chinese models on select benchmarks

Source: AI Index, 2025 | Chart: 2025 AI Index report

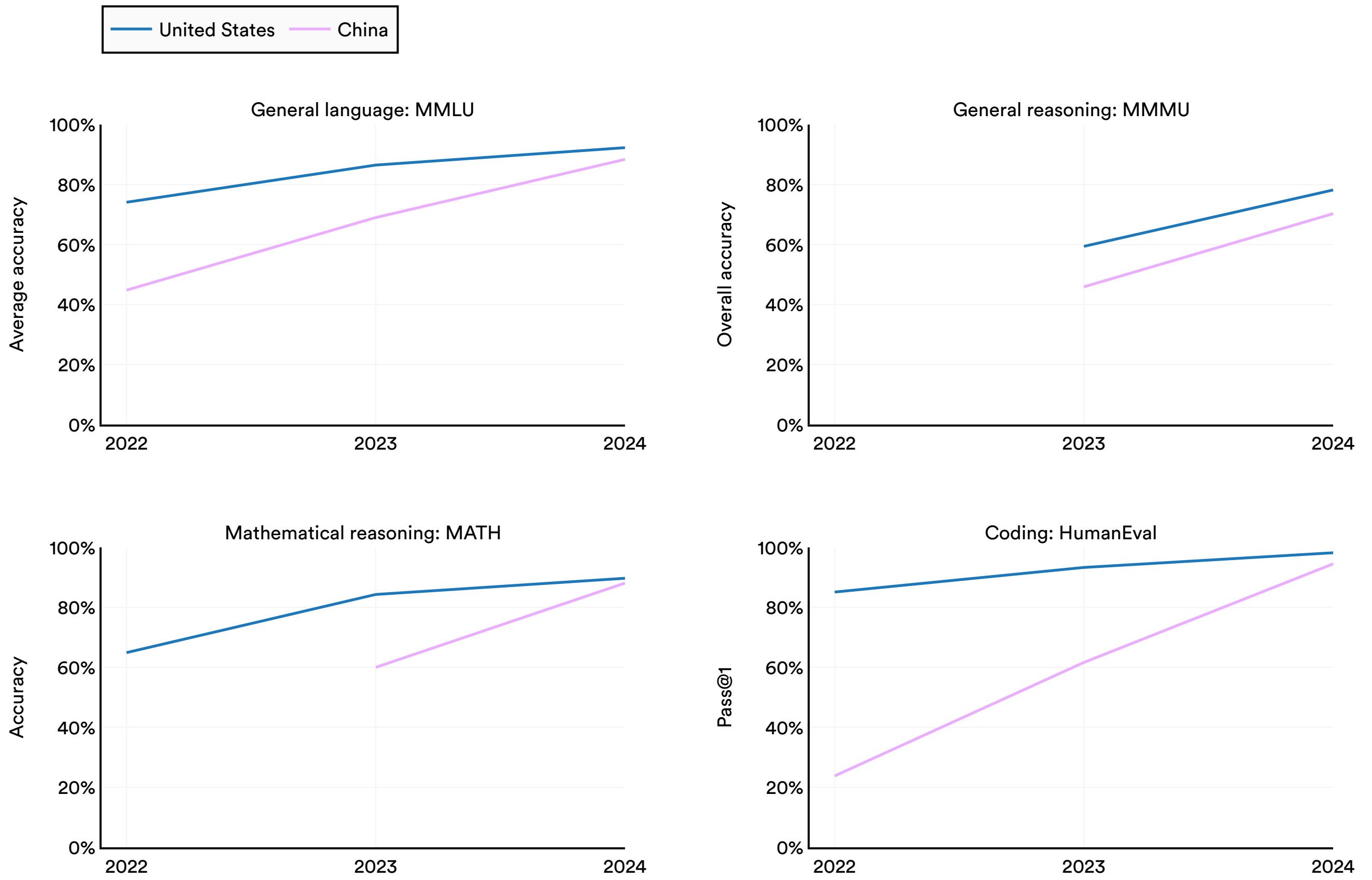

Figure 2.1.37

## Improved Performance From Smaller Models

Recent AI progress has been driven by scaling—the idea that increasing model size and training data improves performance. While scaling has significantly boosted AI capabilities, a notable recent trend is the emergence of smaller high-performing models. Figure 2.1.38 illustrates the reduction in size of the smallest model that scores above 60% on MMLU, a widely used language model benchmark. For context, early models powering ChatGPT, such as GPT-3.5 Turbo, scored around 70% on MMLU. In 2022, the smallest model surpassing 60% on MMLU was PaLM, with 540 billion parameters. By 2024, Microsoft's Phi-3 Mini, with just 3.8 billion parameters, achieved the same threshold, marking a 142-fold reduction in model size over two years.

2024 was a breakthrough year for smaller AI models. Nearly every major AI developer released compact, high-performing models, including GPT-4o mini, o1-mini, Gemini 2.0 Flash, Llama 3.1 8B, and Mistral Small 3.[5] The rise of small models is significant for several reasons. It demonstrates increasing algorithmic efficiency, allowing developers to achieve more with less data and at lower training cost. These efficiency gains, combined with growing datasets, could lead to even higher-performing models. Additionally, inference on smaller models is typically faster and less expensive. Their emergence also lowers the barrier to entry for AI developers and businesses looking to integrate AI into their operations.

**Smallest AI models scoring above 60% on MMLU, 2022–24**
Source: Abdin et al., 2024 | Chart: 2025 AI Index report

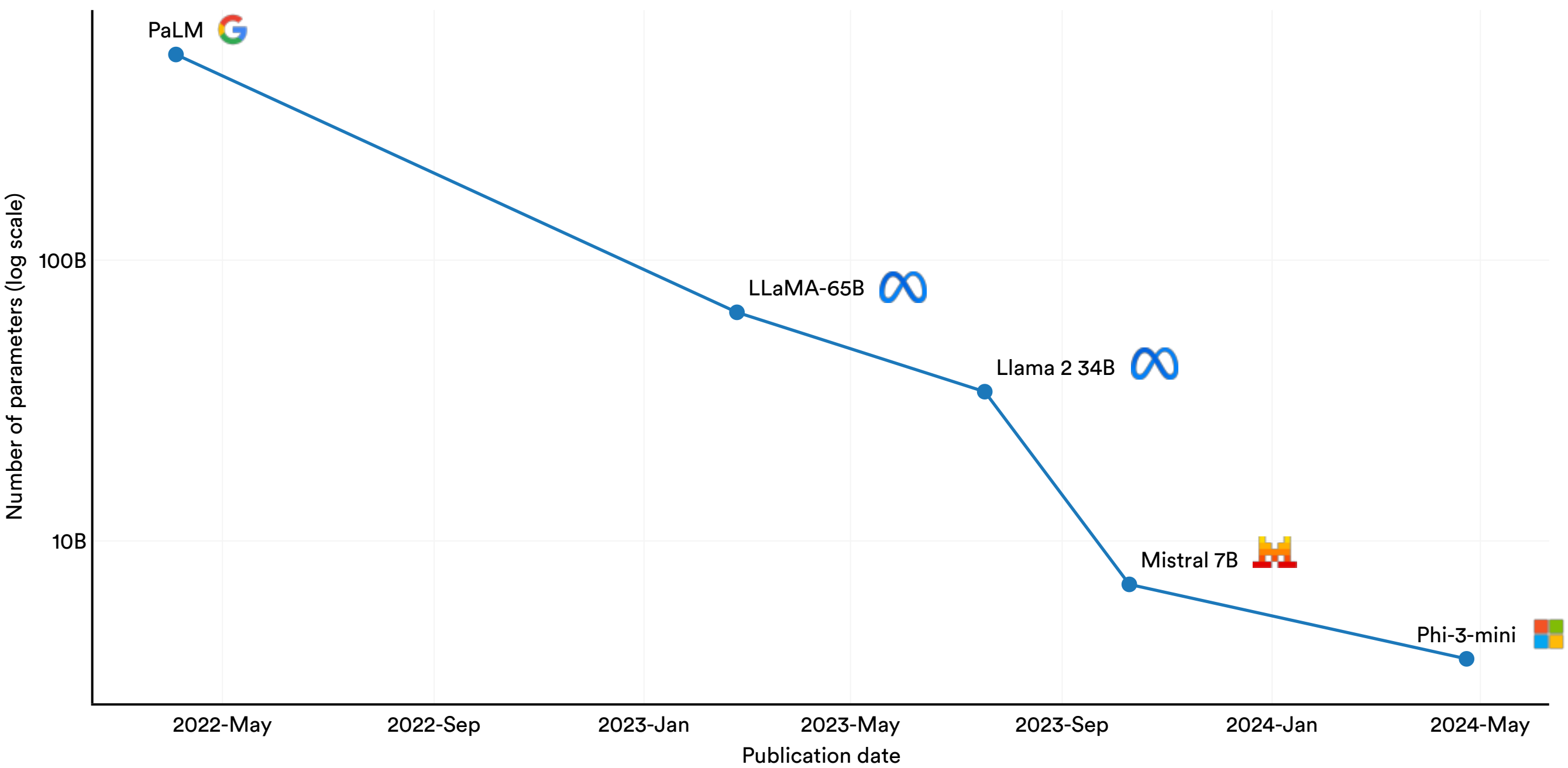

Figure 2.1.38

5 These are just a few of the small models launched in 2024.

## Model Performance Converges at the Frontier

In recent years, AI model performance at the frontier has converged, with multiple providers now offering highly capable models. This marks a shift from late 2022, when ChatGPT's launch—widely seen as AI's breakthrough into public consciousness—coincided with a landscape dominated by just two major players: OpenAI and Google. OpenAI, founded in 2015, released GPT-3 in 2020, while Google introduced models like PaLM and Chinchilla in 2022.

Since then, new players have entered the scene, including Meta with its Llama models, Anthropic with Claude, High-Flyer's DeepSeek, Mistral's Le Chat, and xAI with Grok. As competition has intensified, model performance has increasingly converged (Figure 2.1.39). According to last year's AI Index, the performance gap between the highest- and 10th-ranked models on the Chatbot Arena Leaderboard—a widely used AI ranking platform—was 11.9%. By early 2025, it had narrowed to 5.4%. Similarly, the difference between the top two models fell from 4.9% in 2023 to just 0.7% in 2024. The AI landscape is becoming more competitive, validating 2023 predictions that AI companies lack a technological moat to shield them from rivals.

Performance of top models on LMSYS Chatbot Arena by select providers
Source: LMSYS, 2025 | Chart: 2025 AI Index report

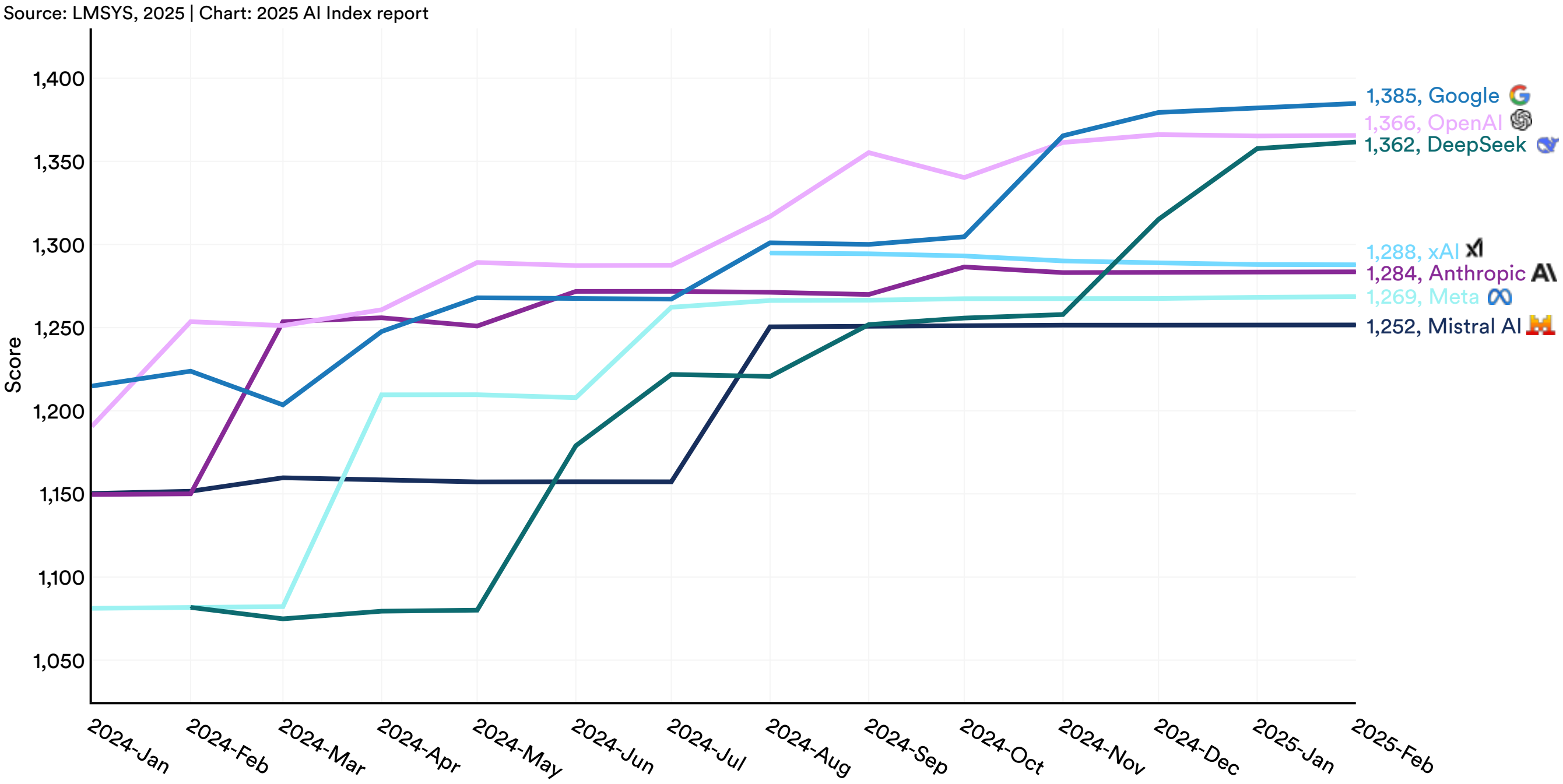

Figure 2.1.39

## Benchmarking AI

For years, the AI Index has used benchmarks to monitor the technical progress of AI systems over time. While benchmarks remain a key tool in this effort, it is important to acknowledge their limitations and guide the community toward more effective benchmarking practices.

As noted in last year's AI Index, many prominent AI benchmarks are reaching saturation. With AI systems advancing rapidly, even newly designed, more challenging tests often remain relevant for only a few years. Some experts suggest that the era of new academic benchmarks may be coming to an end. To truly assess the capabilities of AI systems, more rigorous and comprehensive evaluations are needed.

Additionally, when model developers release new models, they typically report benchmark scores, which are often accepted at face value by the broader community. However, this approach has flaws. In some cases, companies use nonstandard prompting techniques, making model-to-model comparisons unreliable. For example, when Google launched Gemini Ultra, it reported an MMLU benchmark score using a chain-of-thought prompting technique that other developers did not use. Additionally, third-party researchers have documented cases where models perform worse in independent testing compared with the results first reported by their developers.

There are critical aspects of intelligence that do not easily lend themselves to benchmarking. Benchmarks are effective for evaluating certain intelligent capabilities, such as vision and language, where tasks are discrete—e.g., classifying an image correctly or answering a multiple-choice question. However, developing benchmarks is more challenging in areas of AI such as multi-agent systems and human-AI interaction because of factors including the variability in human behaviors and the sheer diversity of correct answers.

In addition, AI advances have traditionally been evaluated in competitions designed to measure human performance, such as games and other open challenges posed to humans or machines. Games such as chess and poker involve significant intelligence, and AI systems have improved over the decades to the point of defeating the best humans at increasingly complex games. Games with a physical component or team capabilities are also a good measure of progress for AI, and the robotics community has embarked on challenging game competitions such as RoboCup for soccer-playing robots. Another area of AI where competitions are used involves coordination and teamwork where multi-agent systems demonstrate advances in distributed reasoning.

Benchmarks have been developed by the AI community for a very long time. Significant advances in AI have been possible because different approaches and methods could be evaluated against the same gold standard represented by a benchmark. In machine learning, benchmarks with different kinds of data in diverse domains have enabled significant advances. Many of these benchmarks are evaluated automatically by a third party without releasing the test data to the AI developers, which makes the evaluations more trustworthy. One interesting recent trend is that various benchmark tasks are addressed by the same model. For example, natural language was addressed for many years as a collection of separate tasks (e.g., understanding, generation, question answering), each with its own models and each with its own benchmarks. Similarly, speech tasks were benchmarked separately from language understanding or generation tasks. Today, the same model can address all language tasks, and, in some cases, a single model can address language, images, and multimodal tasks. This is a very important AI advance concerning the integration of otherwise separate intelligent tasks and capabilities.

The rapid progress of AI systems, evidenced by their consistent outperformance on benchmarks, is perhaps best illustrated by the diminishing relevance of the well-known and long-standing challenge for AI: the Turing test. Originally proposed in Alan Turing's 1950 paper "Computing Machinery and Intelligence," the test evaluates a machine's ability to exhibit humanlike intelligence. In it, a human judge engages in a text-based conversation with both a machine and a human; if the judge cannot reliably distinguish between them, the machine is said to have passed the Turing test. Recent evidence suggests that LLMs have advanced so significantly that people struggle to differentiate the best-performing language models from a human, signaling that modern AI models can pass the Turing test. While the merits and shortfalls of this test have long been debated, it remains an important historical and cultural benchmark for machine intelligence. The questioning of its relevance highlights the remarkable progress of LLMs in recent years and the evolving perception of effective computer science benchmarks and AI measurement.

In robotics, many models have emerged that address interacting with the physical world and reasoning about natural laws. A number of robotics benchmarks, such as ARMBench, focus on perception tasks. However, other benchmarks, such as VIMA-Bench, assess robot performance in simulated environments where they simultaneously incorporate perception, communication, and deep learning.

Benchmarks can also suffer from contamination, where LLMs encounter test questions that were present in their training data. A recent study by Scale found significant contamination in the performance of many LLMs on GSM8K, a widely used mathematics benchmark. Some researchers have sought to combat these contamination issues by introducing benchmarks like LiveBench, which are periodically updated with new questions from unfamiliar sources that LLMs are unlikely to have seen in their training data.

Lastly, research has shown that many benchmarks are poorly constructed. In BetterBench, researchers systematically analyzed 24 prominent benchmarks and identified systemic deficiencies: 14 failed to report statistical significance, 17 lacked scripts for result replication, and most suffered from inadequate documentation, limiting their reproducibility and effectiveness in evaluating models. Despite widespread use, benchmarks like MMLU demonstrated poor adherence to quality standards, while others, such as GPQA, performed significantly better. To address these issues, the paper proposed a 46-criteria framework covering all phases of benchmark development—design, implementation, documentation, and maintenance (Figure 2.1.40). It also introduced a publicly accessible repository to enable continuous updates and improve benchmark comparability. Figure 2.1.41, from BetterBench, assesses many prominent benchmarks on their usability and design. These findings underscore the need for standardized benchmarking to ensure reliable AI evaluation and to prevent misleading conclusions about model performance. Benchmarks have the potential to shape policy decisions and influence procurement decisions within organizations highlighting the importance of consistency and rigor in evaluation.

**Five stages of the benchmark lifecycle**
Source: Reuel et al., 2024

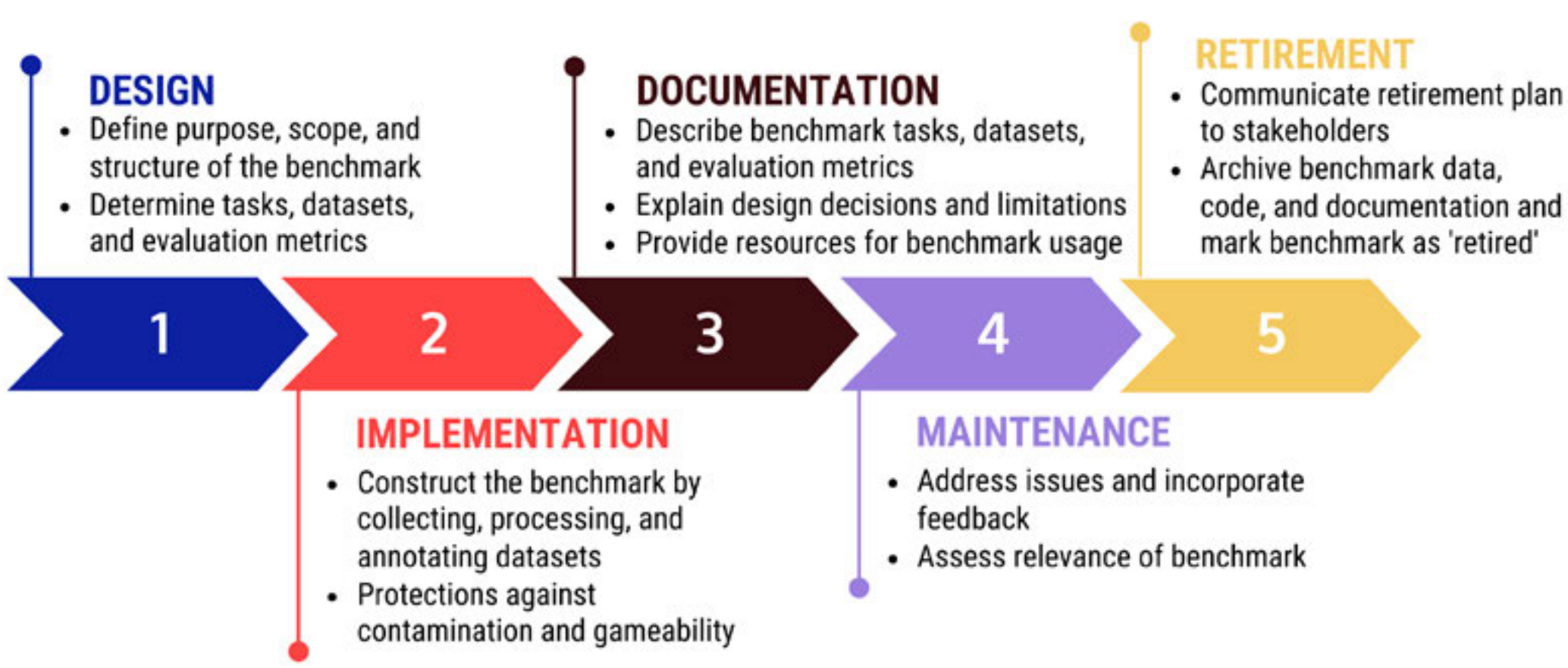

Figure 2.1.40

### Design vs. usability scores across select benchmarks

Source: Reuel et al., 2024 | Chart: 2025 AI Index report

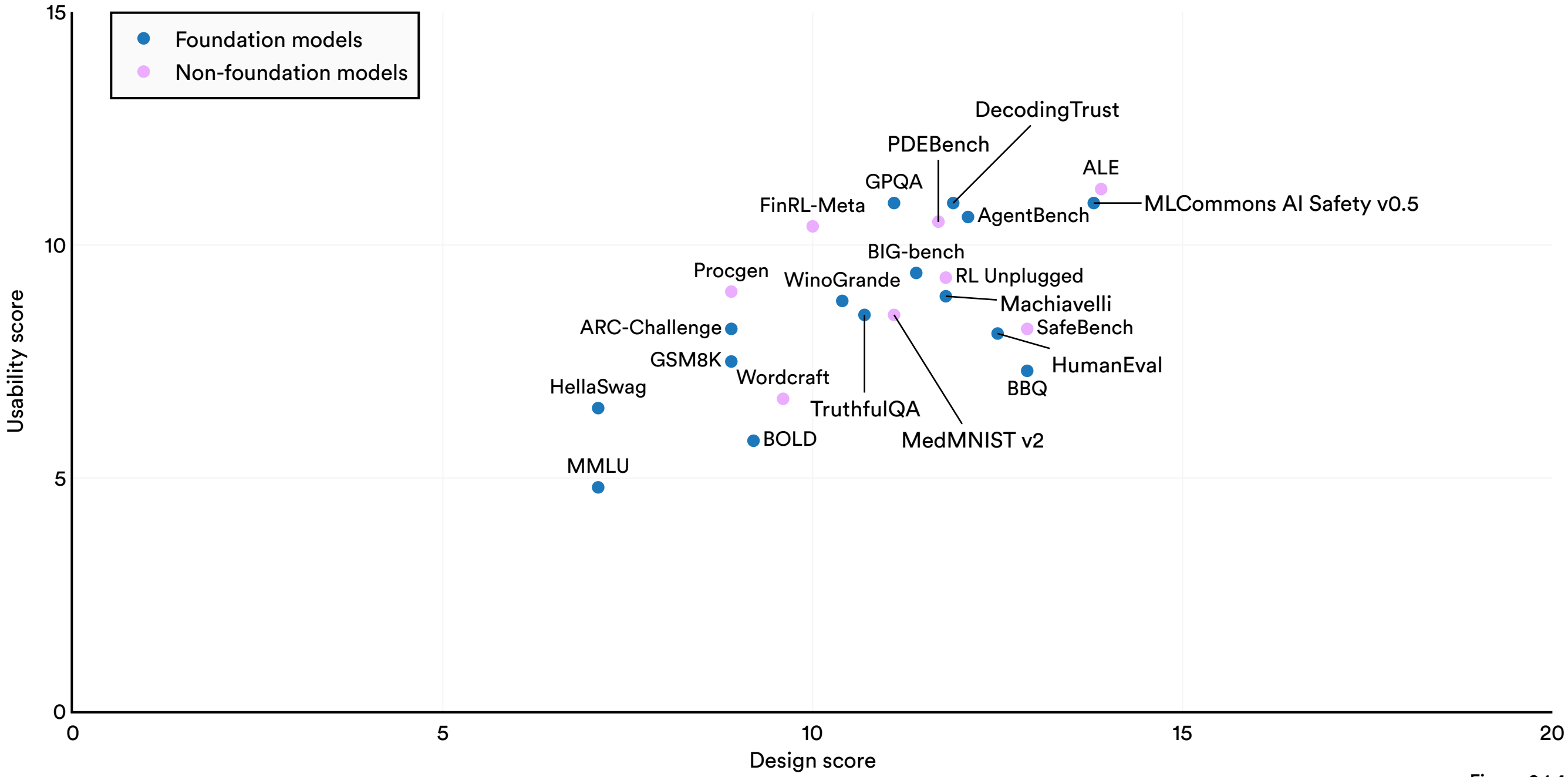

Figure 2.1.41

In this chapter, the AI Index continues to report on benchmarks, recognizing their importance in tracking AI's technical progress. As a standard practice, the Index sources benchmark scores from leaderboards, public repositories such as Papers With Code and RankedAGI, as well as company papers, blog posts, and product releases. The Index operates under the assumption that the scores reported by companies are accurate and factual. The benchmark scores in this section are current as of mid-February 2025. However, since the publication of the AI Index, newer models may have been released that surpass current state-of-the-art scores.

# 2.2 Language

Natural language processing (NLP) enables computers to understand, interpret, generate, and transform text. Current state-of-the-art models, such as OpenAI's GPT-4o, Anthropic's Claude 3.5, and Google's Gemini, are able to generate fluent and coherent prose and display high levels of language understanding ability (Figure 2.2.1). Unlike earlier versions, which were restricted to text input and output, newer language models can now reason across a growing range of input and output modalities, including audio, images, and goal-oriented tasks (Figure 2.2.2).

### A sample output from GPT-4o

**Source:** AI Index, 2025

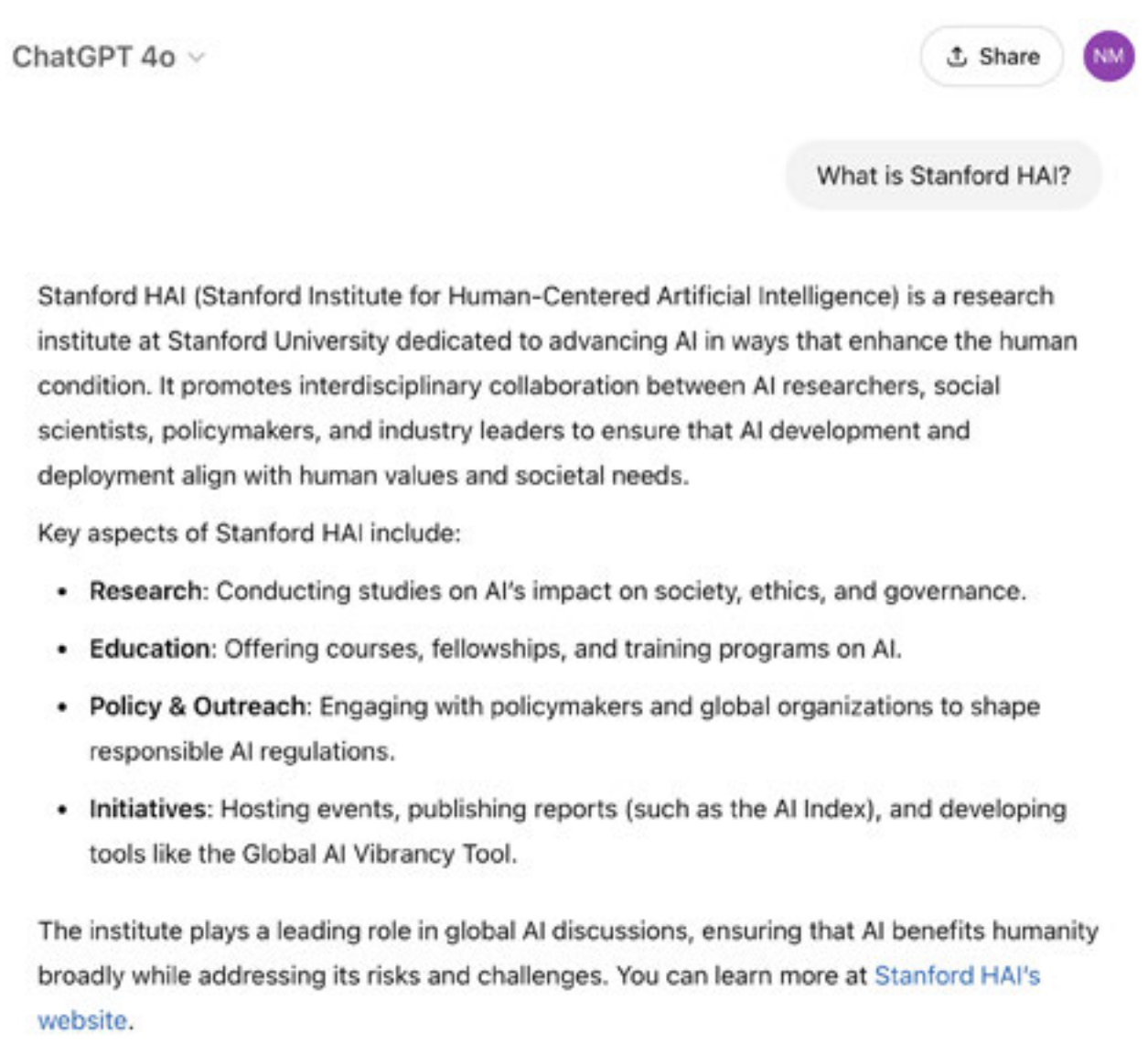

Figure 2.2.1

### Gemini 2.0 in an agentic workflow

**Source:** AI Index, 2025

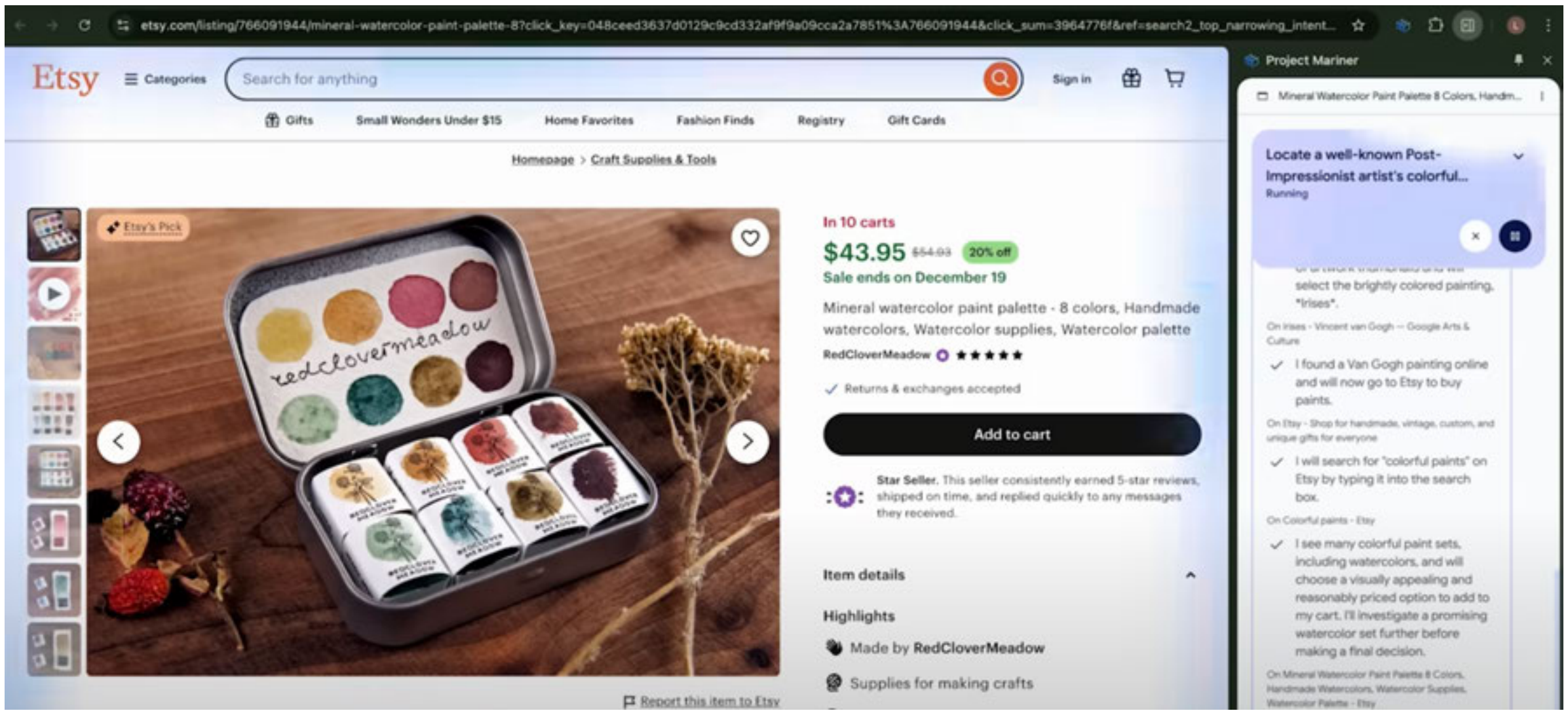

Figure 2.2.2

# Understanding

English language understanding challenges AI systems to understand the English language in various ways, such as reading comprehension and logical reasoning.

### MMLU: Massive Multitask Language Understanding

The Massive Multitask Language Understanding (MMLU) benchmark assesses model performance in zero-shot or few-shot scenarios across 57 subjects, including the humanities, STEM, and the social sciences (Figure 2.2.3). MMLU has emerged as a premier benchmark for assessing LLM capabilities: Many state-of-the-art models like GPT-4o, Claude 3.5, and Gemini 2.0 have been evaluated against MMLU.

The MMLU benchmark was created in 2020 by a team of researchers from UC Berkeley, Columbia University, University of Chicago, and University of Illinois Urbana-Champaign.

The highest recorded score on MMLU, 92.3%, was achieved by OpenAI's o1-preview model in September 2024. For comparison, GPT-4, launched in March 2023, scored 86.4% on the benchmark. Notably, one of the earliest models tested on MMLU, RoBERTa, achieved just 27.9% in 2019 (Figure 2.2.4). This latest state-of-the-art result represents a remarkable 64.4 percentage point increase over five years.

**A sample question from MMLU**
Source: Hendrycks et al., 2021

Microeconomics
One of the reasons that the government discourages and regulates monopolies is that
(A) producer surplus is lost and consumer surplus is gained.
(B) monopoly prices ensure productive efficiency but cost society allocative efficiency.
(C) monopoly firms do not engage in significant research and development.
(D) consumer surplus is lost with higher prices and lower levels of output.

Figure 2.2.3

**MMLU: average accuracy**
Source: Papers With Code, 2025 | Chart: 2025 AI Index report

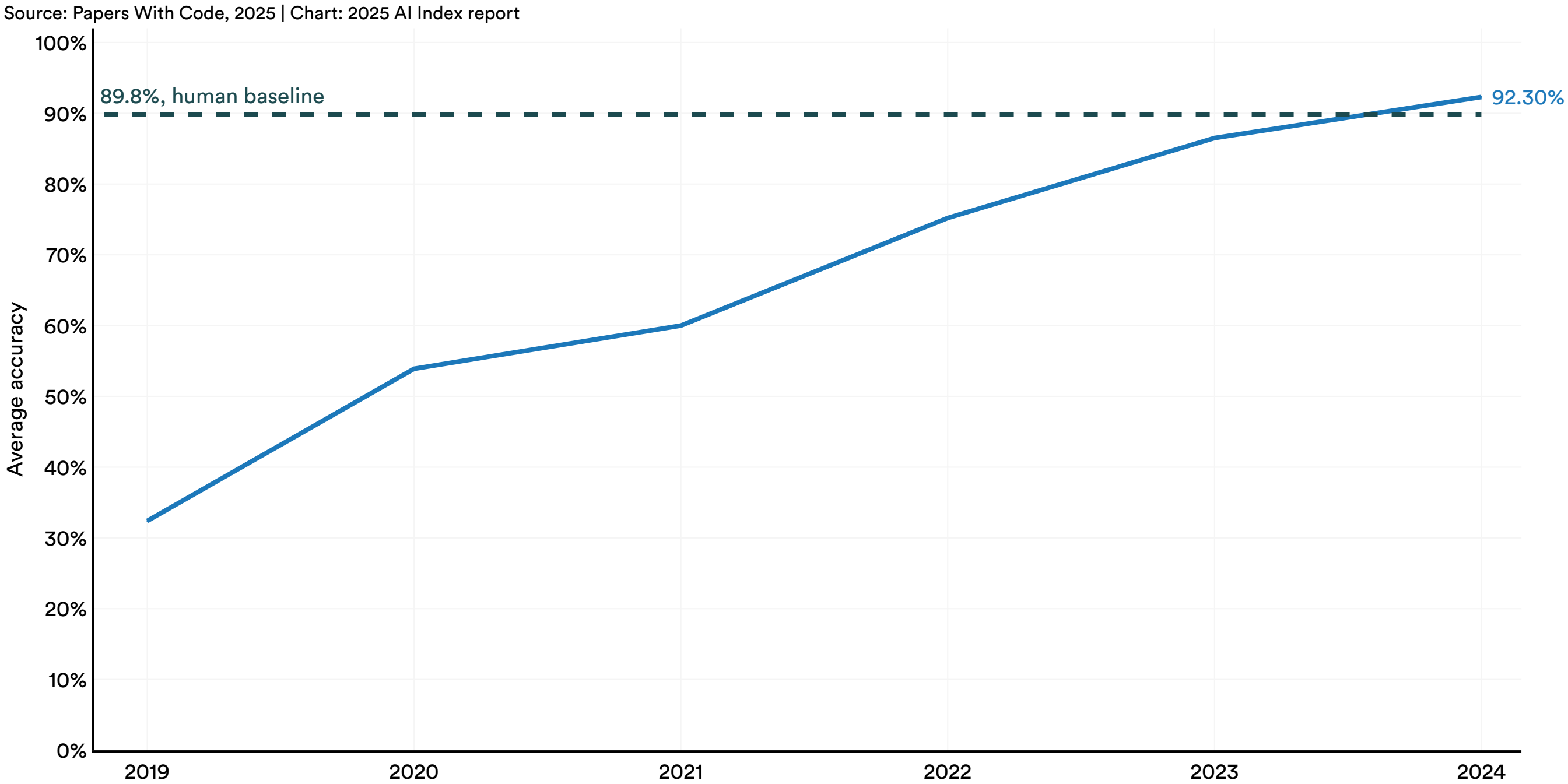

Figure 2.2.4

Despite its prominence, MMLU has faced notable criticisms. These include claims that the benchmark contains erroneous or overly simplistic questions, which may not challenge increasingly advanced systems. In 2024, a team of researchers from the University of Toronto, University of Waterloo, and Carnegie Mellon introduced MMLU-Pro, a more challenging variant of MMLU. This version eliminates noisy and trivial questions, expands complex ones, and increases the number of answer choices available to models. Figure 2.2.5 highlights performance trends on MMLU-Pro, with DeepSeek-R1 posting the highest score to date (84.0%).

Additionally, concerns have been raised about the testing landscape. Developers sometimes report MMLU scores using nonstandard prompting techniques that boost performance but can lead to misleading comparisons. Furthermore, evidence suggests that publicly reported scores by developers can differ—sometimes by as much as five percentage points—from those later evaluated by academic researchers. As such, MMLU performance results should be interpreted with caution.

# Generation

In generation tasks, AI models are tested on their ability to produce fluent and practical language responses.

### Chatbot Arena Leaderboard

The rise of capable LLMs has made it increasingly important to understand which models are preferred by the general public. Launched in 2023, the Chatbot Arena Leaderboard from LMSYS is one of the first comprehensive evaluations of public LLM preference. The leaderboard allows users to query two anonymous models and vote for the preferred generations (Figure 2.2.6). By early 2025, the platform had accumulated over 1 million votes, with users ranking one of Google's Gemini models as the community's most preferred choice.

**MMLU-Pro: overall accuracy**

Source: MMLU-Pro Leaderboard, 2025 | Chart: 2025 AI Index report

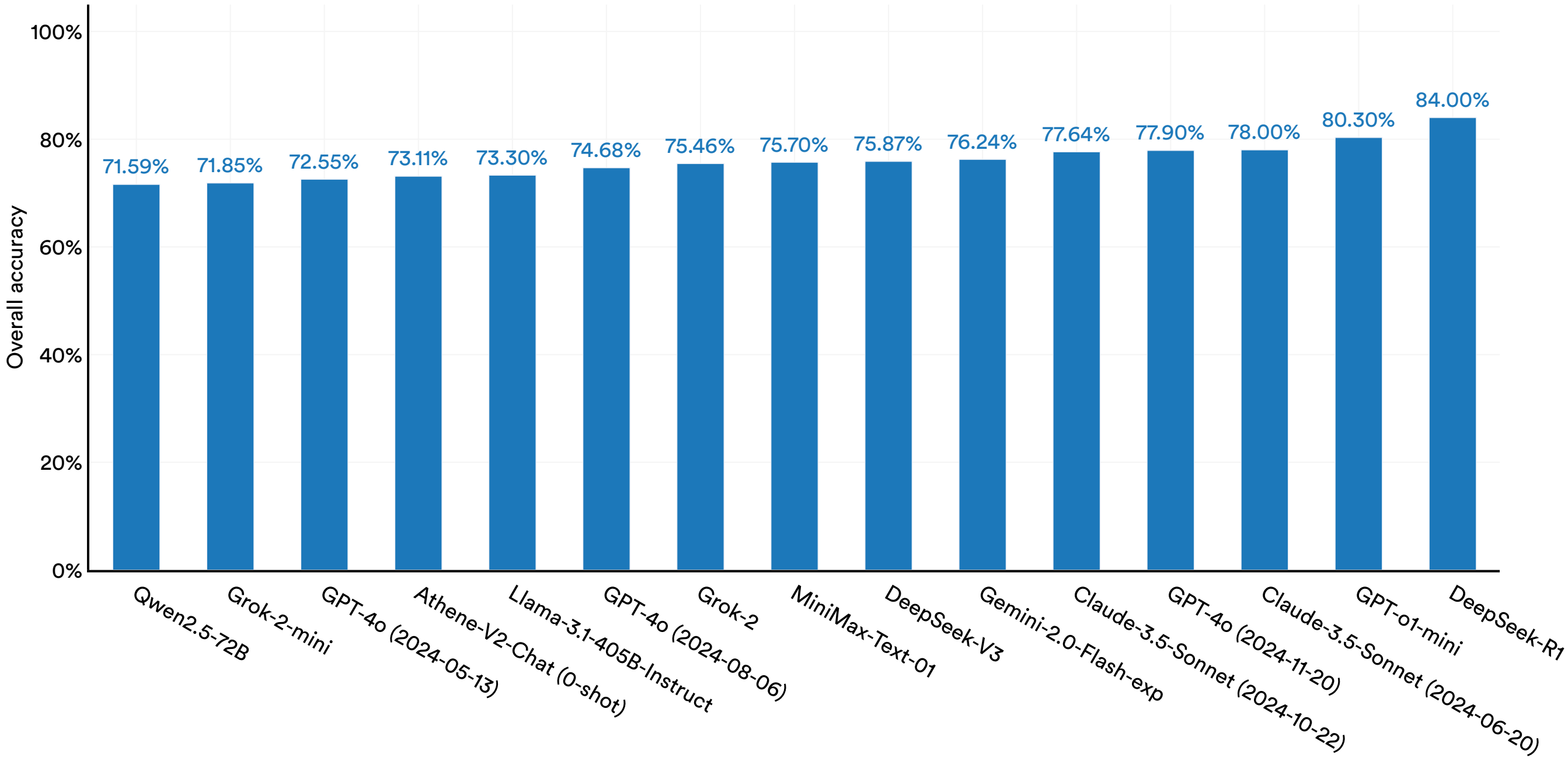

Figure 2.2.5

### A sample model response on the Chatbot Arena Leaderboard

Source: Chatbot Arena Leaderboard, 2024

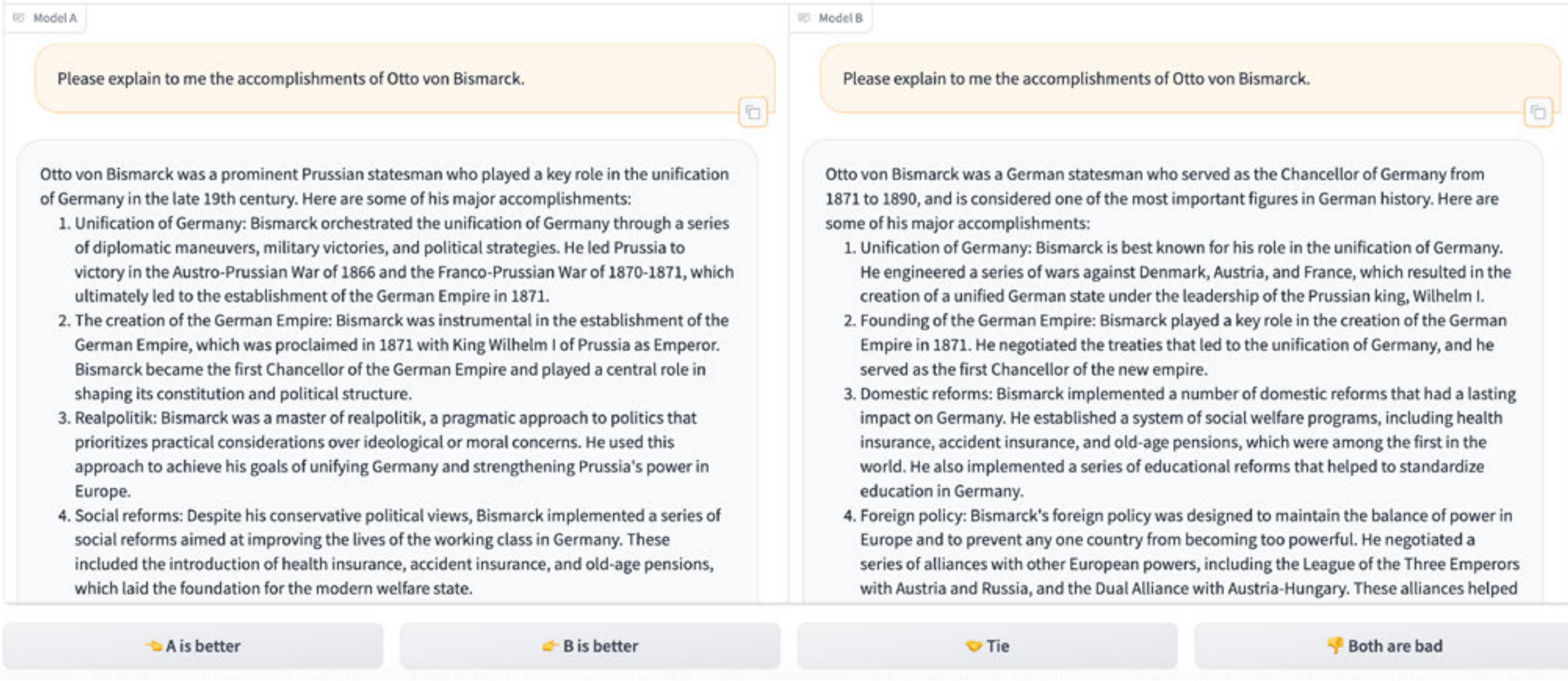

Figure 2.2.6

Figure 2.2.7 provides a snapshot of the top 10 models on the Chatbot Arena Leaderboard as of January 2025. Interestingly, the performance gap between top leaderboard models has narrowed over time. In 2023, according to data from the 2024 AI Index, the difference in Arena scores between the top model and the 10th-ranked model was 11.9%.[6] By 2025, this gap had decreased to just 5.4%. This convergence highlights a growing parity in the quality of recent LLMs.

### LMSYS Chatbot Arena for LLMs: Elo rating (overall)

Source: LMSYS, 2025 | Chart: 2025 AI Index report

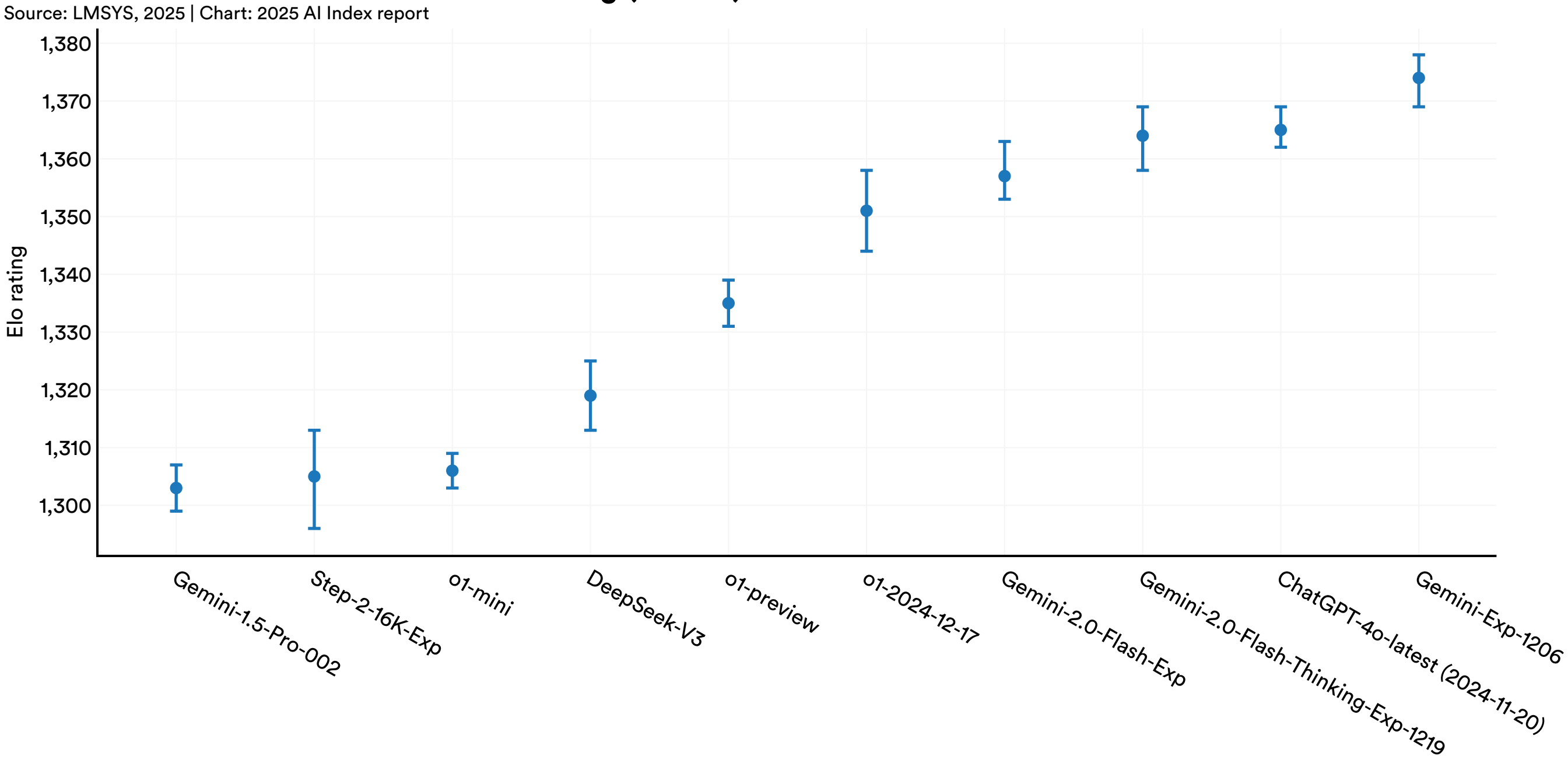

Figure 2.2.7

6 The Arena score is a relative ranking system used by the Arena Leaderboard to compare model performance. For more details on the scoring methodology, refer to the paper introducing the Chatbot Arena Leaderboard.

### Arena-Hard-Auto

One of the challenges in developing new benchmarks to keep pace with rapidly improving AI capabilities is that creating high-quality, human-curated benchmarks is often expensive and time-consuming. In response, this year saw the launch of BenchBuilder. Created by a team of UC Berkeley researchers, BenchBuilder leverages LLMs to create an automated pipeline for curating high-quality, open-ended prompts from large, crowdsourced datasets. BenchBuilder can be used to update or create new benchmarks without significant human involvement. This tool was used by the LMSYS team to develop Arena-Hard-Auto, a benchmark designed to evaluate instruction-tuned LLMs (Figure 2.2.8). Arena-Hard-Auto includes 500 challenging user queries sourced from Chatbot Arena. In this benchmark, GPT-4 Turbo serves as the judge that compares model responses against a baseline model (GPT-4-0314).

**Arena-Hard-Auto vs. other benchmarks**
Source: Li et al., 2024

| | Evaluation | Open-Ended | Prompt Curation | Prompt Source |
|---|---|---|---|---|
| Arena-Hard-Auto | Automatic | Yes | Automatic | Configurable |
| MMLU, MATH, GPQA | Automatic | No | Manual | Fixed |
| MT-Bench, AlpacaEval | Automatic | Yes | Manual | Fixed |
| Live Bench, Live Code Bench | Automatic | No | Manual | Fixed |
| Chatbot Arena | Human | Yes | Crowd-source | Crowd |

Figure 2.2.8

As of November 2024, the top-scoring models on the Arena-Hard-Auto leaderboard were o1-mini (92.0), o1-preview (90.4), and Claude-3.5-Sonnet (85.2) (Figure 2.2.9). Arena-Hard-Auto also features a style control leaderboard, which accounts for how the style of an LLM's responses might inadvertently influence user preferences. The top model on the style leaderboard is the November variant of Anthropic's Claude Sonnet 3.5 (Figure 2.2.10). Automated benchmarks like Arena-Hard-Auto have faced criticism for uneven question distribution, which limits their ability to provide a comprehensive assessment of LLM capabilities. For instance, over 50% of Arena-Hard-Auto questions focus solely on coding and debugging.

**Arena-Hard-Auto with no modification**
Source: LMSYS, 2025 | Chart: 2025 AI Index report

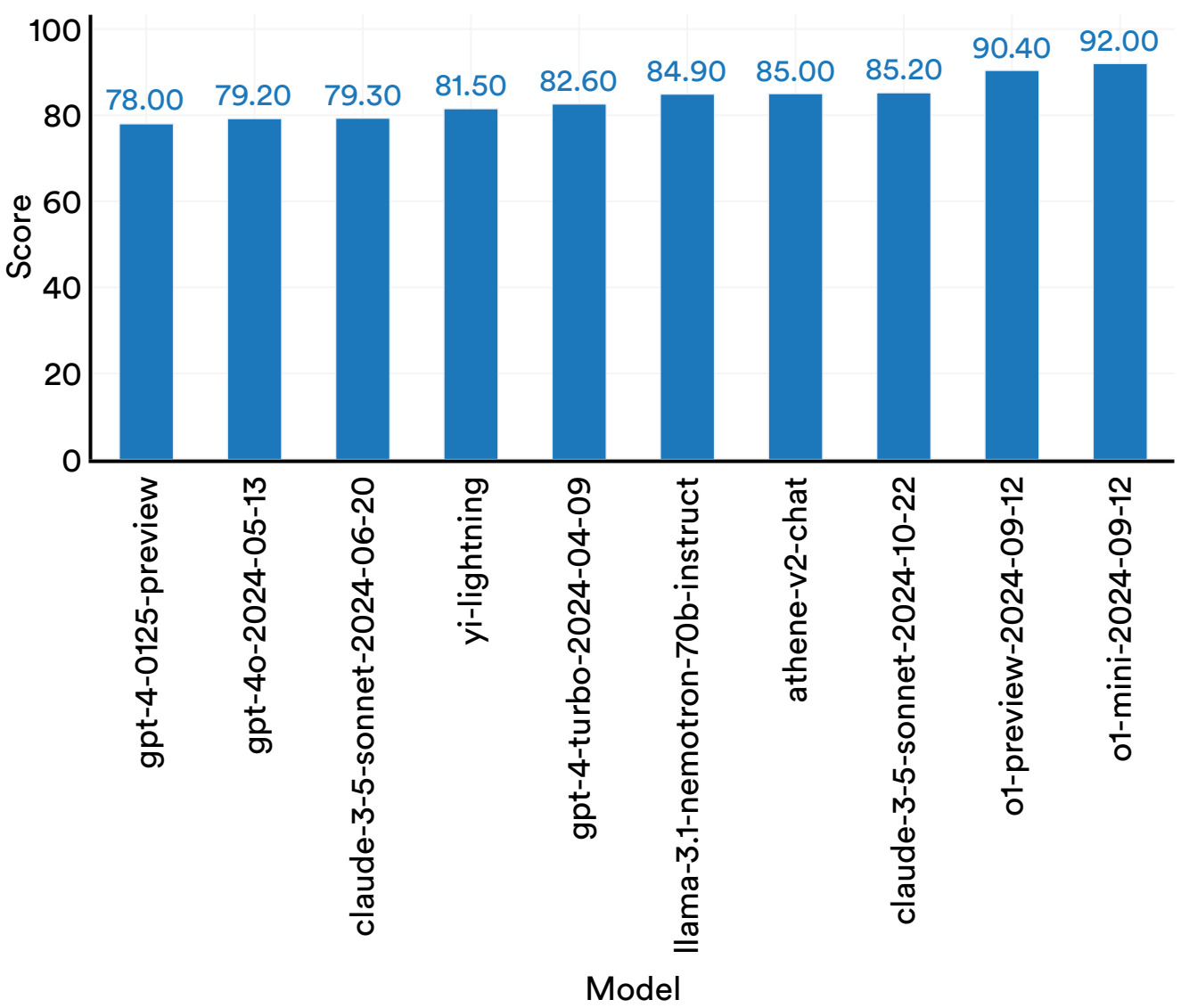

Figure 2.2.9

**Arena-Hard-Auto with style control**
Source: LMSYS, 2025 | Chart: 2025 AI Index report

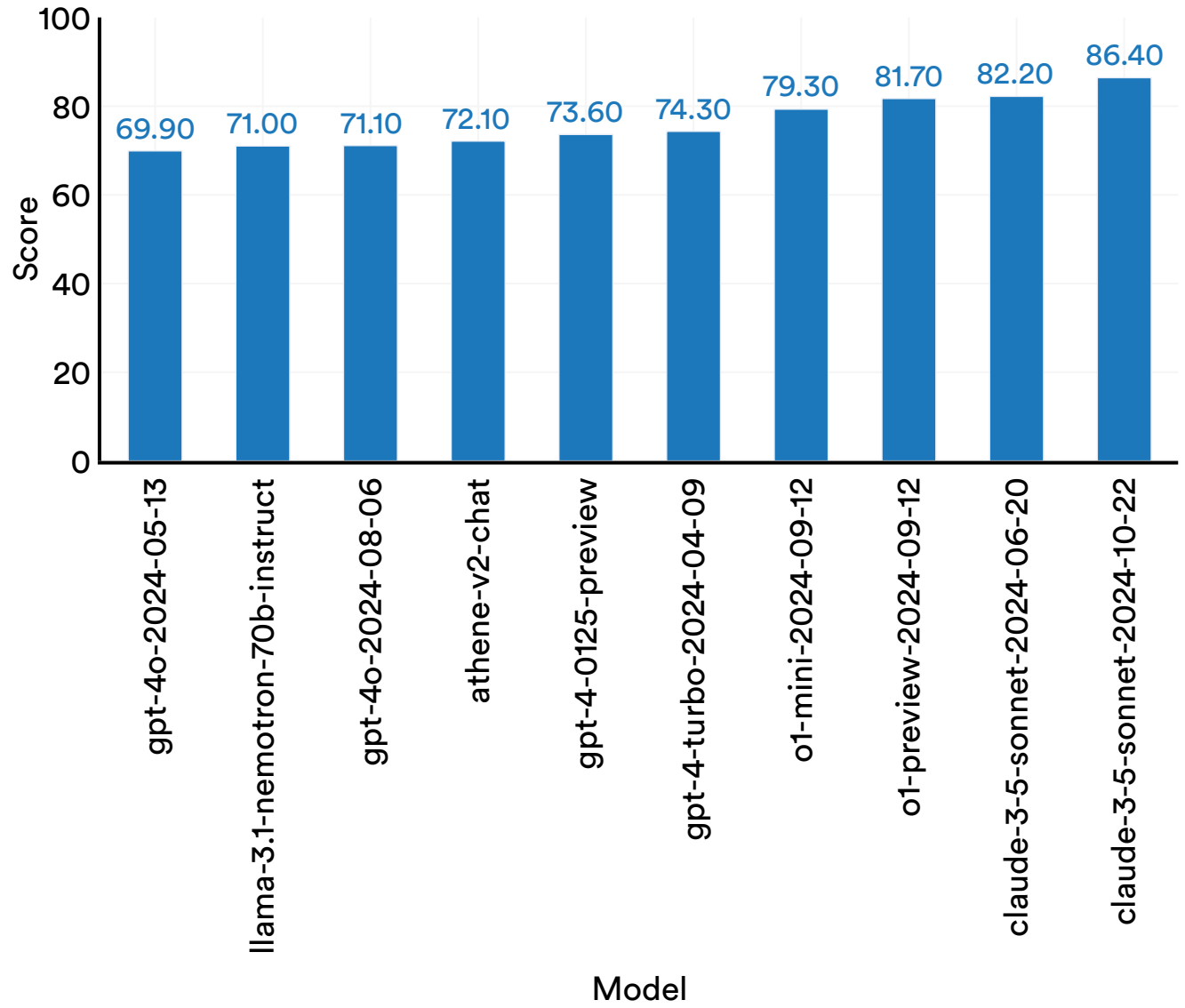

Figure 2.2.10

## WildBench

WildBench, developed by researchers from the Allen Institute for AI and the University of Washington, is a benchmark launched in 2024 to evaluate LLMs on challenging real-world queries. The creators highlight several limitations of existing LLM evaluations. For example, MMLU focuses on academic questions and does not assess open-ended, real-world problems. Similarly, benchmarks like LMSYS, which address real-world challenges, rely heavily on human oversight and lack consistency in evaluating all models with the same dataset.

### Evaluation framework for WildBench

Source: Lin et al., 2024

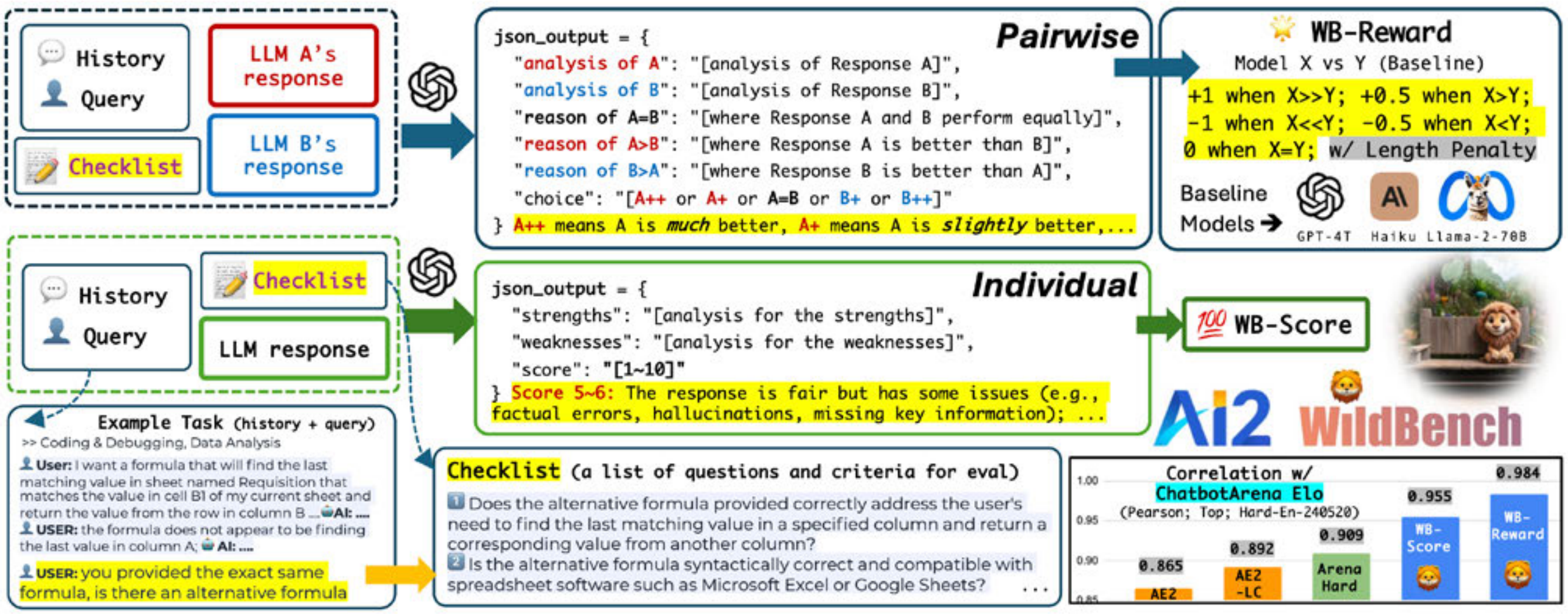

Figure 2.2.11

WildBench addresses many shortcomings of existing benchmarks by providing an automated evaluation framework for LLMs, incorporating a diverse set of real-world ("in the wild") questions that language models are likely to encounter (Figure 2.2.11). The questions in WildBench are meticulously selected from over 1 million human-chatbot interactions and are periodically updated to ensure relevance. The creators also maintain a live leaderboard to track model performance over time. Currently, the top-performing model on WildBench is GPT-4o, with an Elo score of 1227.1, narrowly surpassing the second-place model, Claude 3.5 Sonnet, which scored 1215.4 (Figure 2.2.12).

**WildBench: WB-Elo (length controlled)**
Source: WildBench Leaderboard, 2025 | Chart: 2025 AI Index report

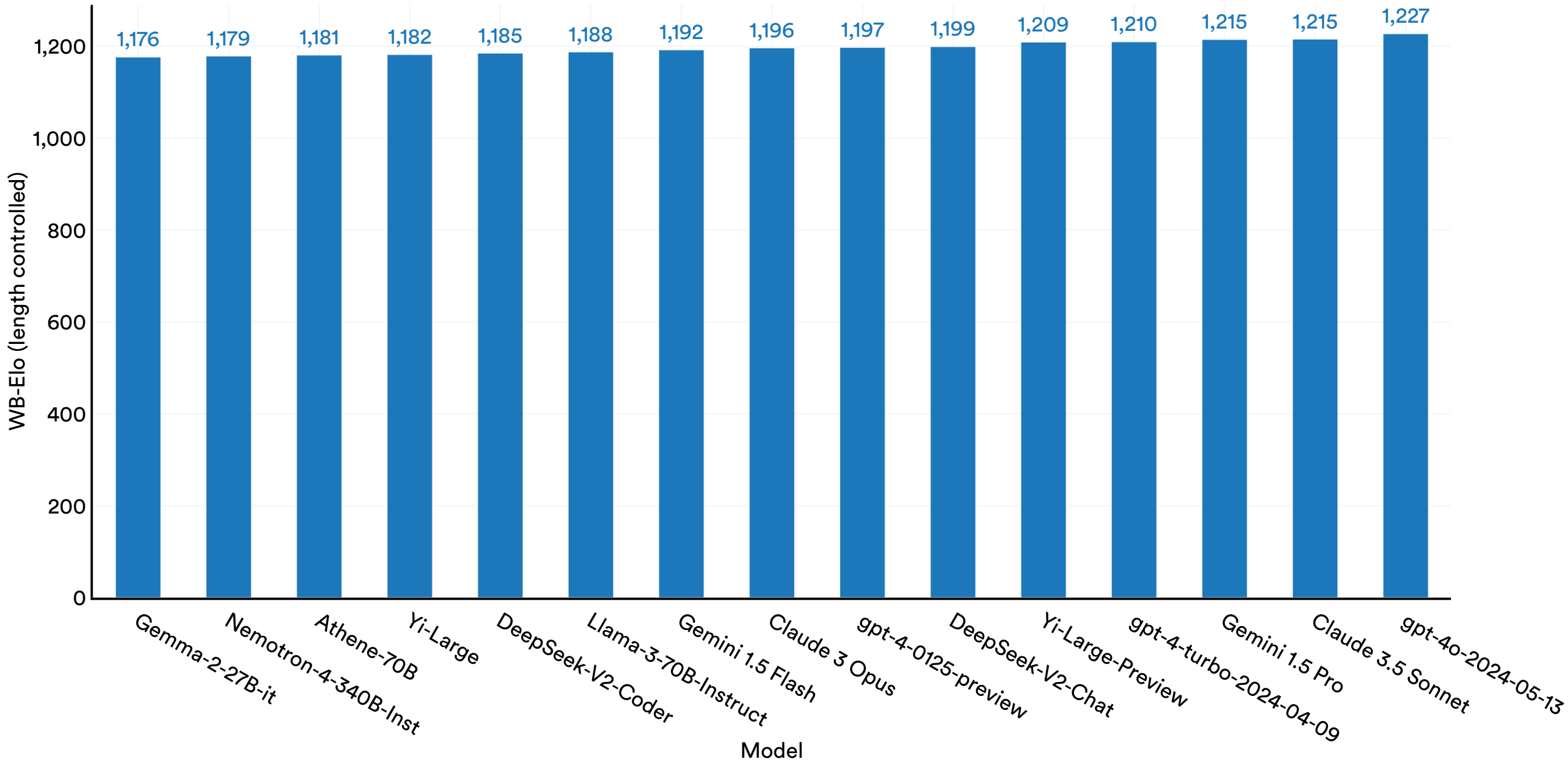

Figure 2.2.12

**Highlight:**

## o1, o3, and Inference-Time Compute

OpenAI's latest two models, o1 and o3, mark a paradigm shift in AI models' ability to "think" and exhibit signs of advanced reasoning. o1 and o3 have shown impressive results across a variety of tasks, including programming, quantum physics, and logic. The models' advanced reasoning capabilities are attributed to their chain-of-thought process and ability to iteratively check answers. This means that the models break complex problems into smaller, more manageable steps before executing them, enhancing the resulting output quality. For example, when asked to decipher scrambled text, o1 will specify its thought and reasoning process more thoroughly than GPT-4 (Figure 2.2.13). This process, through which AI systems iterate as they answer, has been referred to as inference or test-time computation.

### Chain-of-thought thinking in o1

Source: OpenAI, 2024

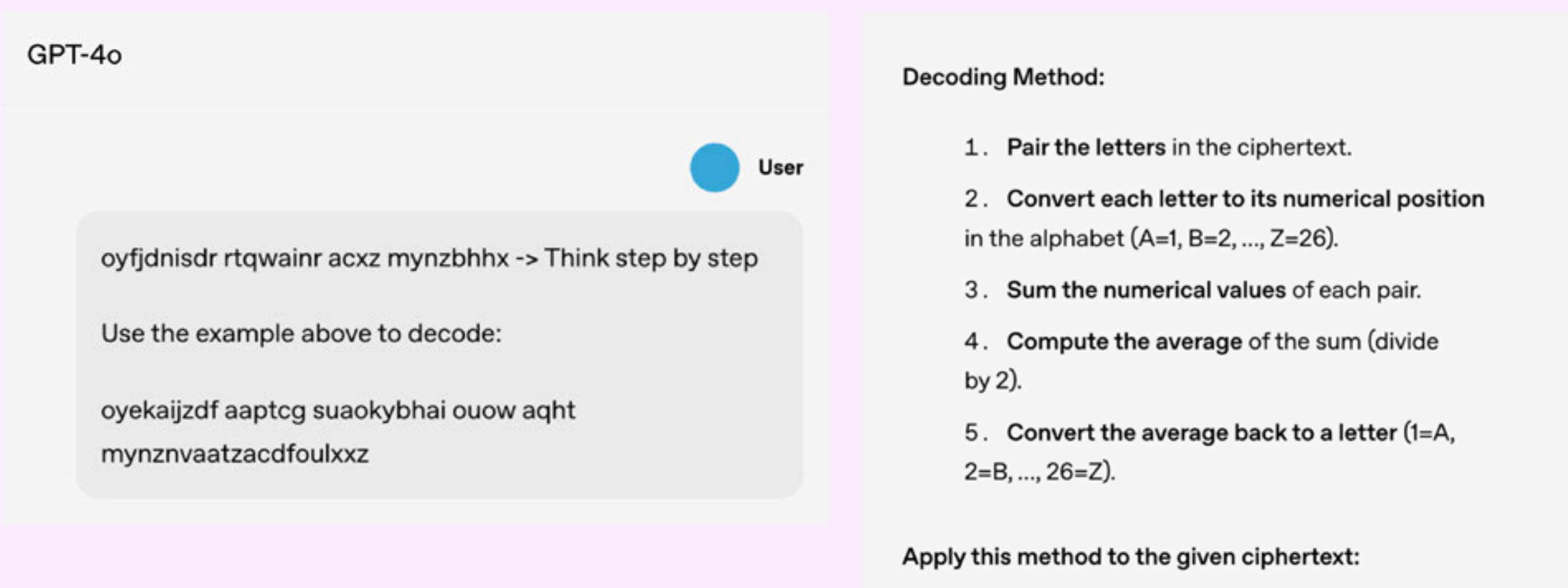

Figure 2.2.13

**Highlight:**

## o1, o3, and Inference-Time Compute (cont'd)

Figure 2.2.14 juxtaposes the scores of GPT-4o, OpenAI's previous state-of-the-art model, with o1 and o1-preview on a variety of benchmarks.[7] For example, o1 outperforms GPT-4o with a 2.8-point gain on MMLU, 34.5 points on MATH, 26.7 points on GPQA Diamond, and 65.1 points on AIME 2024, a notoriously difficult mathematics competition. Finally, o3 demonstrates more complex reasoning than any other AI model known today, posting an 87.5% accuracy rate on the ARC-AGI machine intelligence benchmark and passing the previous record of 55.5%.

**GPT-4o vs. o1-preview vs. o1 on select benchmarks**
Source: OpenAI, 2024 | Chart: 2025 AI Index report

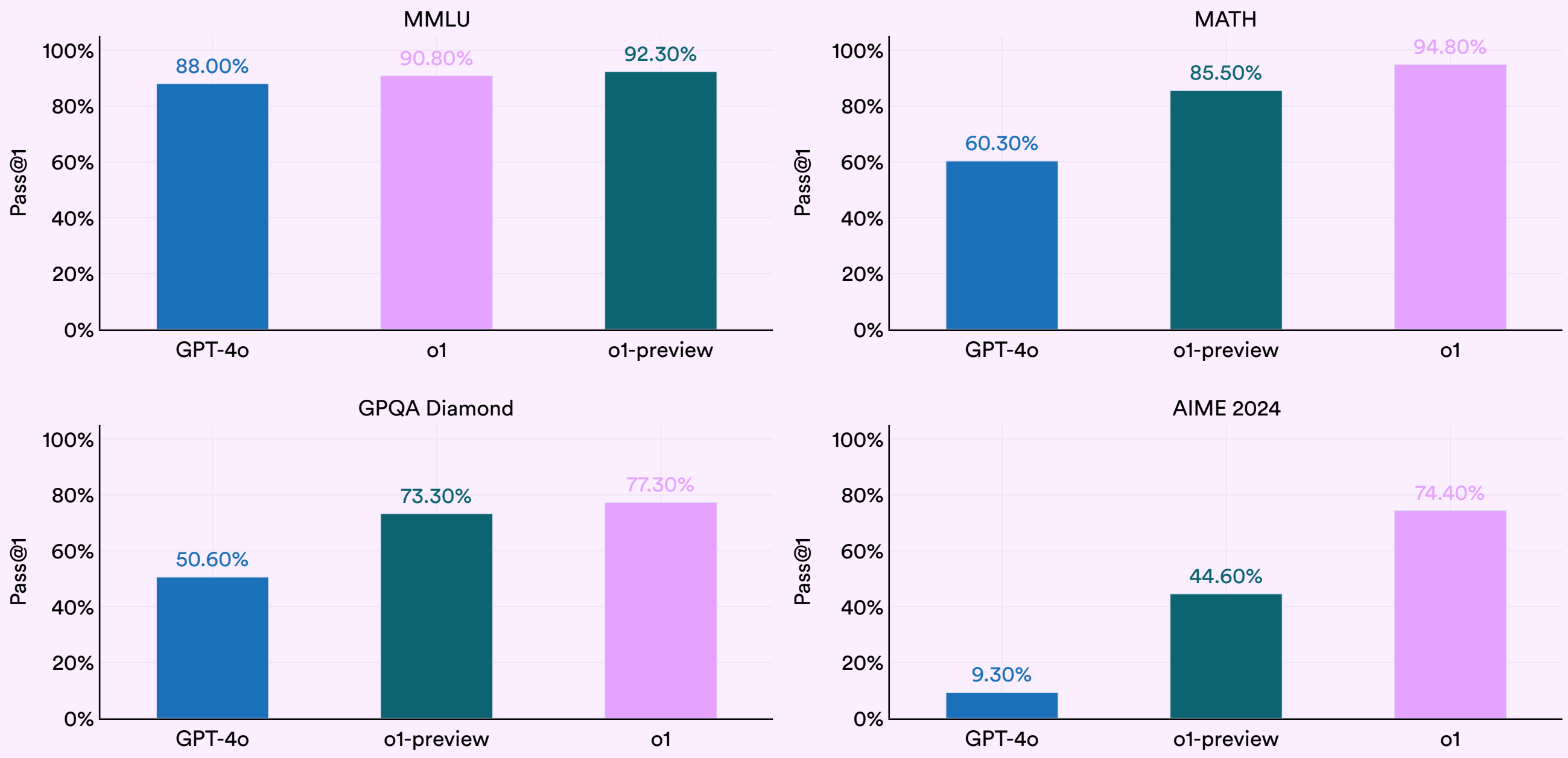

Figure 2.2.14

While these models enhance reasoning capabilities, this comes at a price—both a financial and latency cost. For example, GPT-4o costs $2.50 per 1 million input tokens and $10 per 1 million output tokens. Conversely, o1 costs $15 per 1 million input tokens and $60 per 1 million output tokens.[8] Moreover, o1 is approximately 40 times slower than GPT-4o, with 29.7 seconds to first token as opposed to GPT-4o's 0.72. The latency of o3, while not publicly available, is presumably even higher. o1 and o3's strong capabilities are likely to continue fueling powerful AI systems and agents.

OpenAI first released o1-preview to ChatGPT Plus and Teams users on Sept. 12, 2024, and released the full version of o1 (as well as access to ChatGPT Pro, a $200 monthly subscription enabling access to o1) on Dec. 5, 2024.

7 The o1-preview model is OpenAI's early release of o1, made available before its broader public launch.

8 o3 is currently only available to select researchers and developers via OpenAI's safety testing program.

## MixEval

MixEval, launched by researchers at the National University of Singapore, Carnegie Mellon University, and the Allen Institute for AI, is another newly released benchmark designed to address some of the aforementioned limitations in the current field of LLM evaluation. MixEval combines comprehensive, well-distributed, real-world user queries, similar to those found in Chatbot Arena, with ground-truth-based questions, like those featured in MMLU (Figure 2.2.15). MixEval includes various evaluation suites, with MixEval-Hard representing the more challenging version of the benchmark. This suite focuses on substantially harder queries, making it one of the most effective tools for assessing how models handle complex questions.

### Evaluation framework for MixEval

Source: Ni et al., 2024

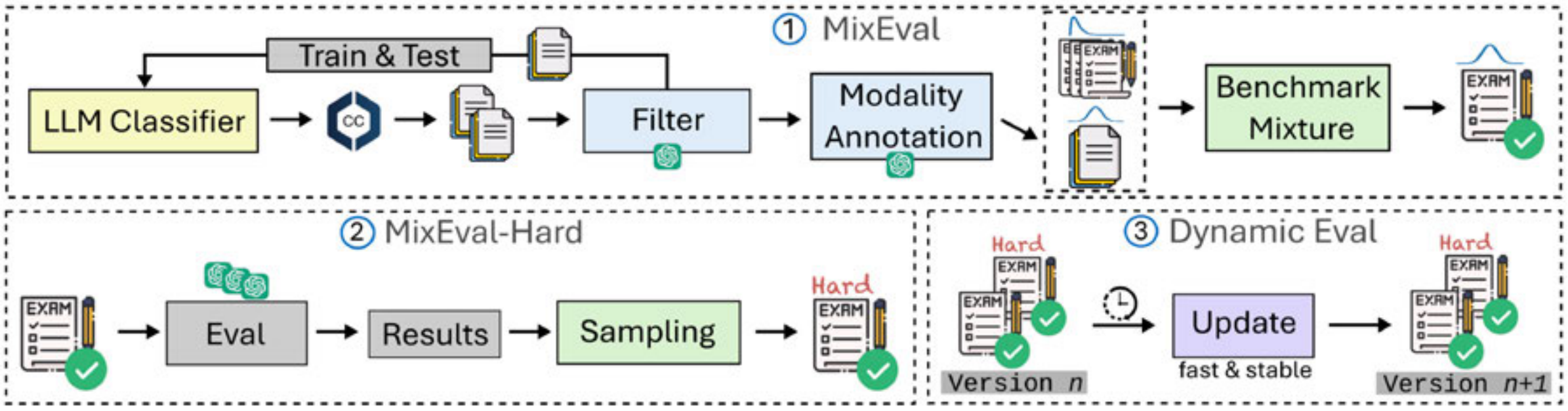

Figure 2.2.15

The highest-scoring model on the MixEval-Hard benchmark is OpenAI's o1-preview, with a score of 72.0. In second place is the Claude 3.5 Sonnet-0620 model, followed by the Llama-3 1-405B-Instruct model, which scored 66.2 (Figure 2.2.16). All three models were released in 2024.

### MixEval-Hard on chat models: score

Source: MixEval Leaderboard, 2025 | Chart: 2025 AI Index report

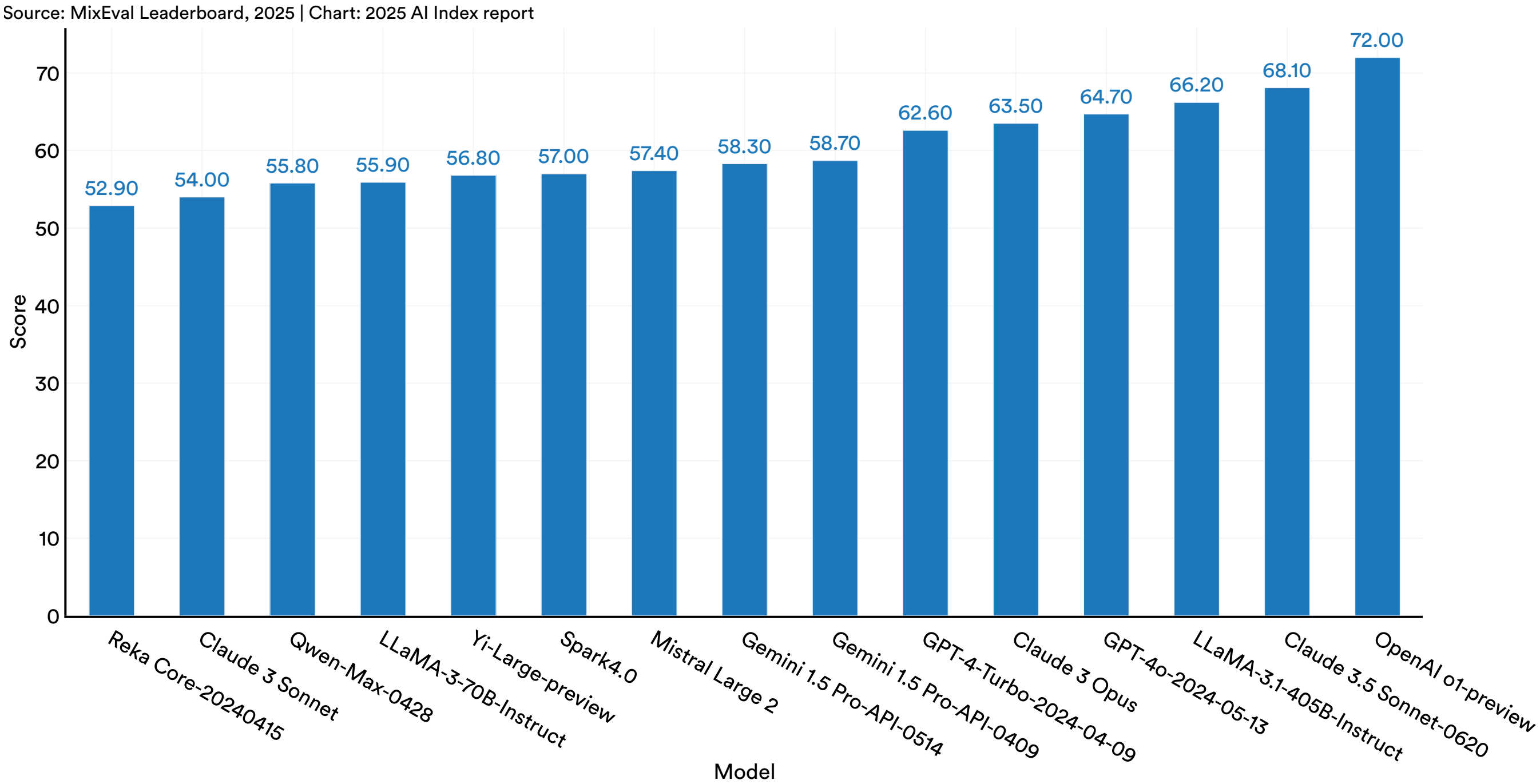

Figure 2.2.16

# RAG: Retrieval Augment Generation (RAG)

An increasingly common capability being tested in LLMs is retrieval-augmented generation (RAG). This approach integrates LLMs with retrieval mechanisms to enhance their response generation. The model first retrieves relevant information from files or documents and then generates a response tailored to the user's query based on the retrieved content. RAG has diverse use cases, including answering precise questions from large databases and addressing customer queries using information from company documents.

In recent years, RAG has received increasing attention from researchers and companies. For example, in September 2024, Anthropic introduced Contextual Retrieval, a method that significantly enhances the retrieval capabilities of RAG models. 2024 also saw the release of numerous benchmarks for evaluating RAG systems, including Ragnarok (a RAG arena battleground) and CRAG (Comprehensive RAG benchmark). Additionally, specialized RAG benchmarks, such as FinanceBench for financial question answering, have been developed to address specific use cases.

## Berkeley Function Calling Leaderboard

The Berkeley Function Calling Leaderboard evaluates the ability of LLMs to accurately call functions or tools. The evaluation suite includes over 2,000 question-function-answer pairs across multiple programming languages (such as Python, Java, JavaScript, and REST API) and spans a variety of testing domains (Figure 2.2.17).

**Data composition on the Berkeley Function Calling Leaderboard**
Source: Yan et al., 2024

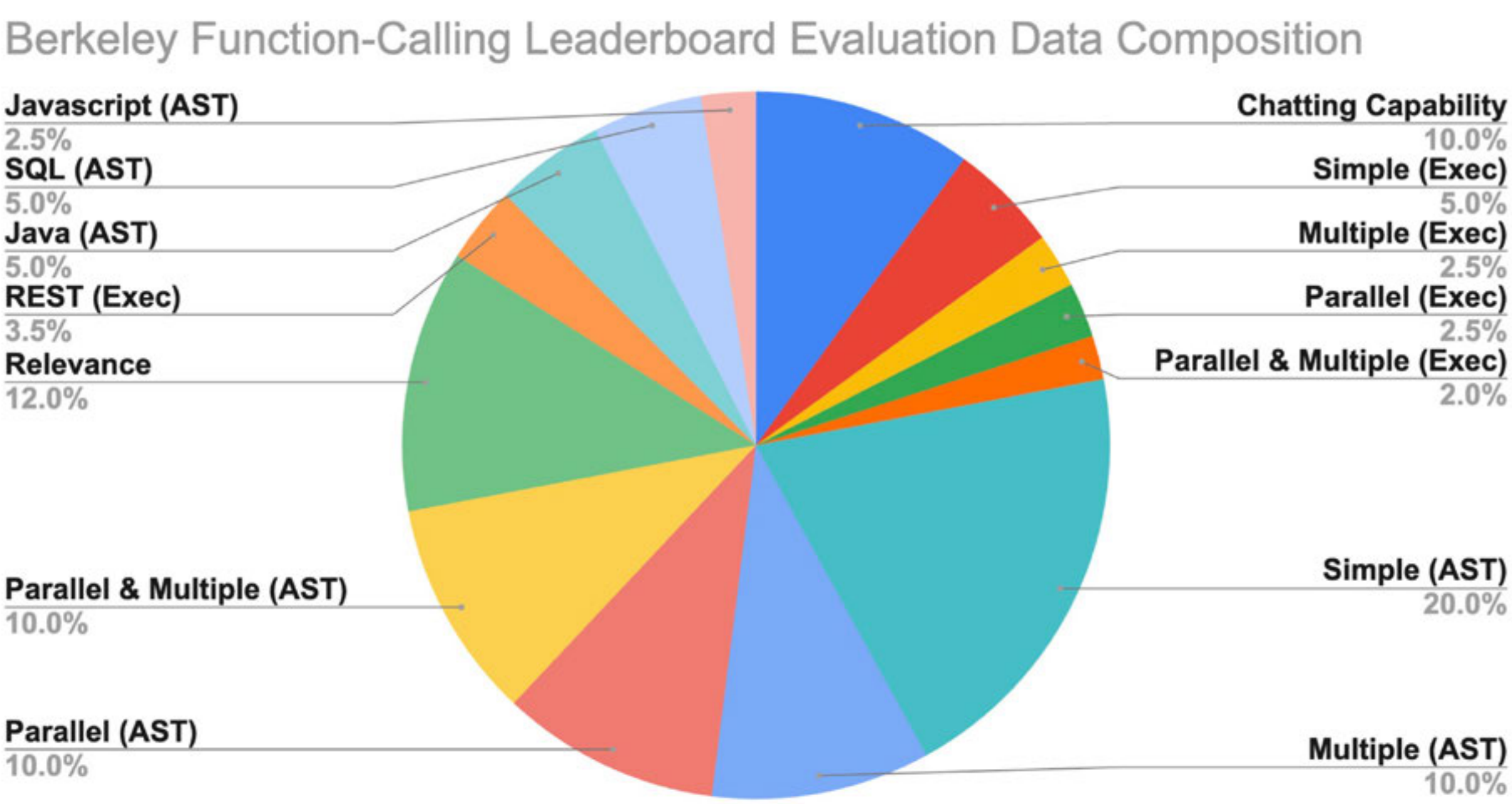

Figure 2.2.17[9]

9 In this context: AST (abstract syntax tree) refers to tasks that involve analyzing or manipulating code at the structural level, using its parsed representation as a tree of syntactic elements. Evaluations labeled with "AST" likely test an AI model's ability to understand, generate, or manipulate code in a structured manner. Exec (execution-based) indicates tasks that require actual execution of function calls to verify correctness. Evaluations labeled with "Exec" likely assess whether the AI model can correctly call and execute functions, ensuring the expected outputs are produced.

The top model on the Berkeley Function Calling Leaderboard is watt-tool-70b, a fine-tuned variant of Llama-3.3-70B-Instruct designed specifically for function calling. It achieved an overall accuracy of 74.31 (Figure 2.2.18). The next-highest-scoring model was a November variant of GPT-4o, with a score of 72.08. Performance on this benchmark has improved significantly over the course of 2024, with top models at the end of the year achieving accuracies up to 50 points higher than those recorded early in the year.

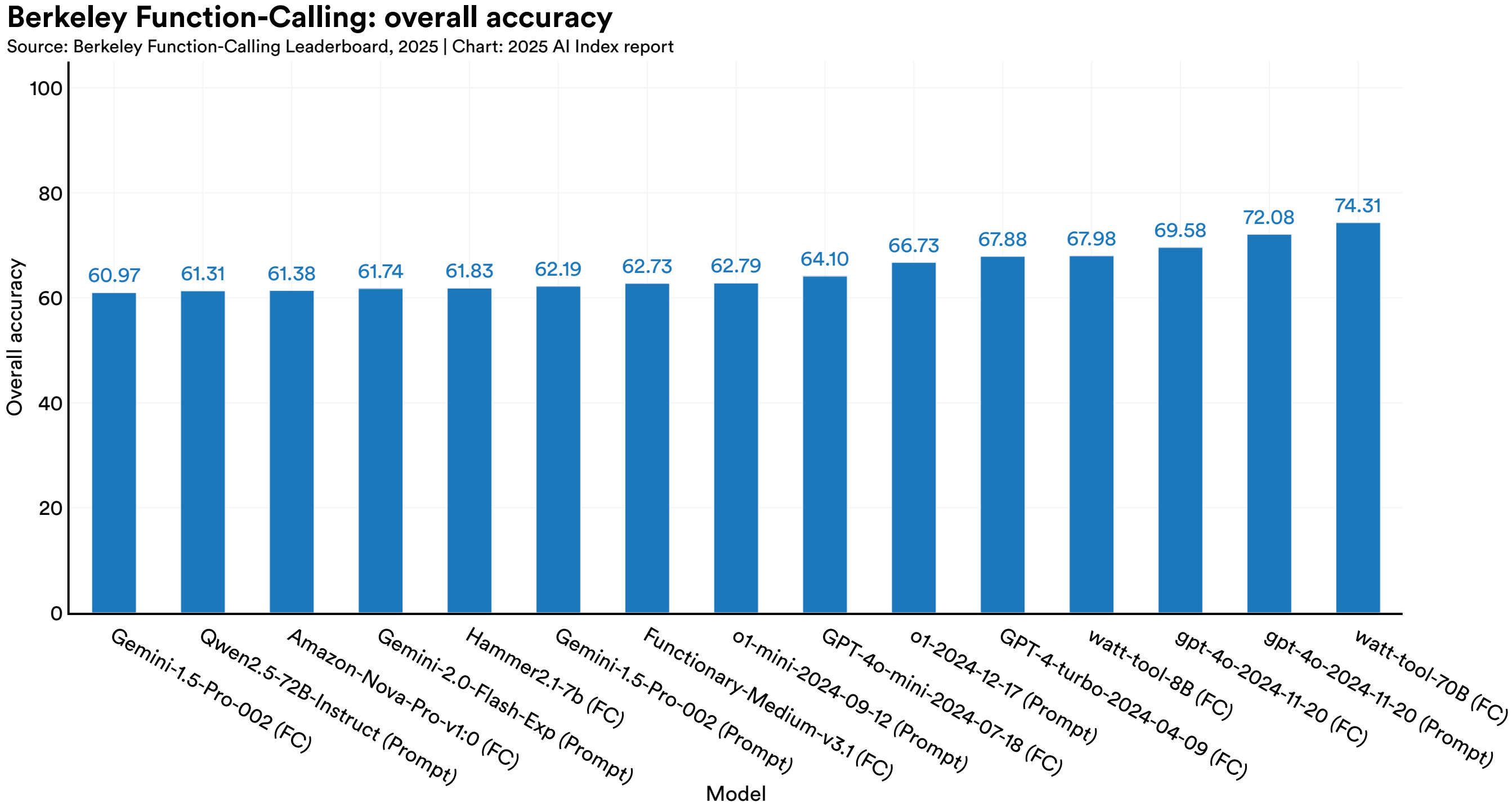

Figure 2.2.18

### MTEB: Massive Text Embedding Benchmark

The Massive Text Embedding Benchmark (MTEB), created by a team at Hugging Face and Cohere, was introduced in late 2022 to comprehensively evaluate how models perform on various embedding tasks. Embedding involves converting data, such as words, texts, or documents, into numerical vectors that capture rough semantic meanings and distance between vectors. Embedding is an essential component of RAG. During a RAG task, when users input a query, the model transforms it into an embedding vector. This transformation enables the model to then search for relevant information. MTEB includes 58 datasets spanning 112 languages and eight embedding tasks (Figure 2.2.19).[10] For example, in the bitext mining task, there are two sets of sentences from two different languages, and for every sentence in the first set, the model is tasked to find the best match in the second set.

**Tasks in the MTEB benchmark**
Source: Muennighoff et al., 2023

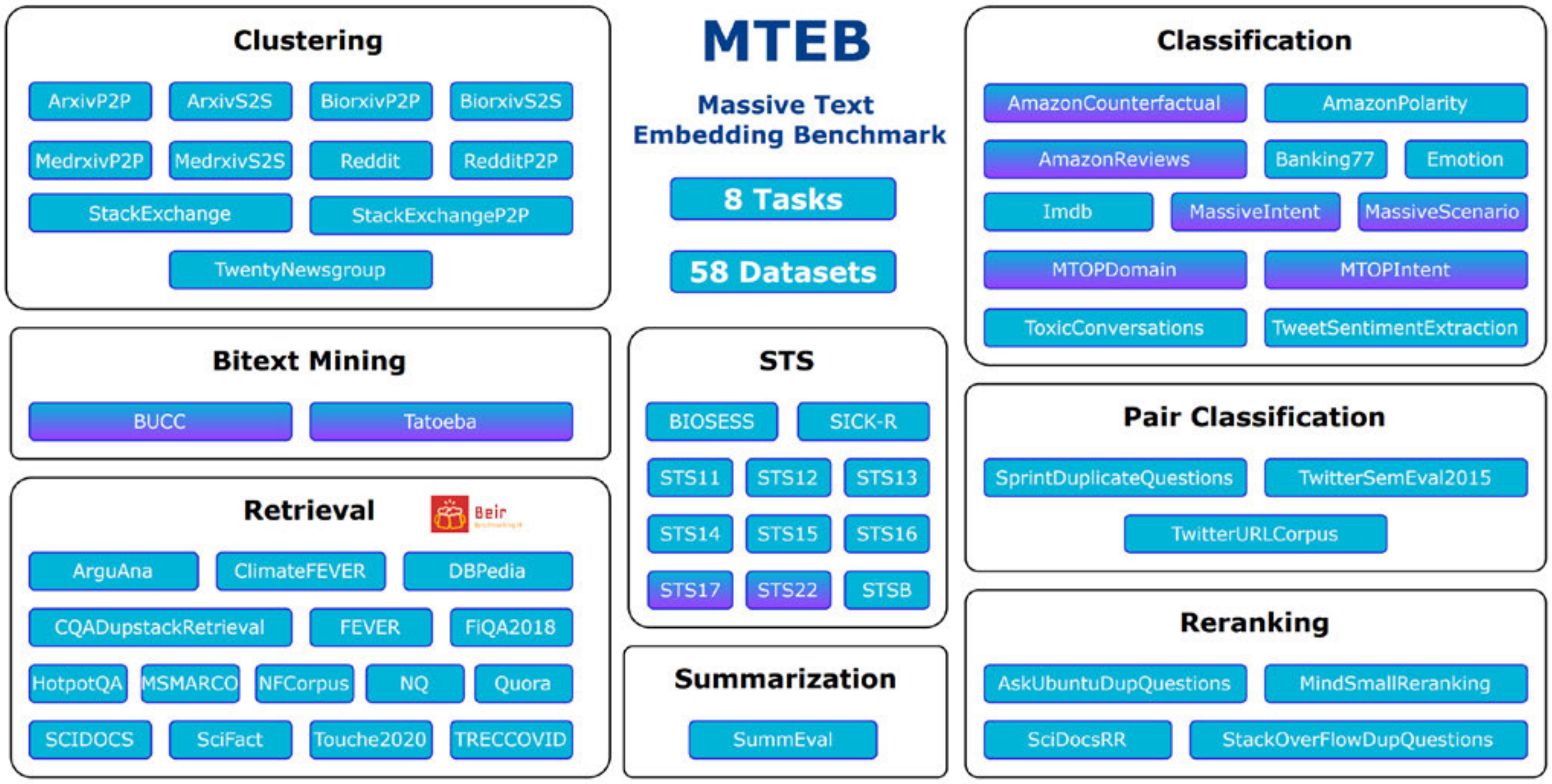

Figure 2.2.19

10 The benchmark covers the following eight tasks: bitext mining, classification, clustering, pair classification, reranking, retrieval, semantic textual similarity, and summarization. For details on each task, refer to the MTEB paper.

As of early 2025, the top-performing embedding model on the MTEB benchmark was Voyage AI's voyage-3-m-exp, with a score of 74.03. Voyage AI is focused on creating high-quality AI embedding models. The voyage-3-m-exp model is a variant of the voyage-3-large, a large foundation model specifically designed for embedding tasks, and it uses strategies like Matryoshka Representation Learning and quantization-aware training to improve its performance. The voyage-3-m-exp model narrowly outperformed NV-Embed-v2 (72.31), which held the top spot for most of 2024 (Figure 2.2.20). When the MTEB benchmark was first introduced in late 2022, the leading model achieved an average score of 59.5. Over the past two years, therefore, performance on the benchmark has meaningfully improved.

MTEB on English subsets across 56 datasets: average score
Source: MTEB Leaderboard, 2025 | Chart: 2025 AI Index report

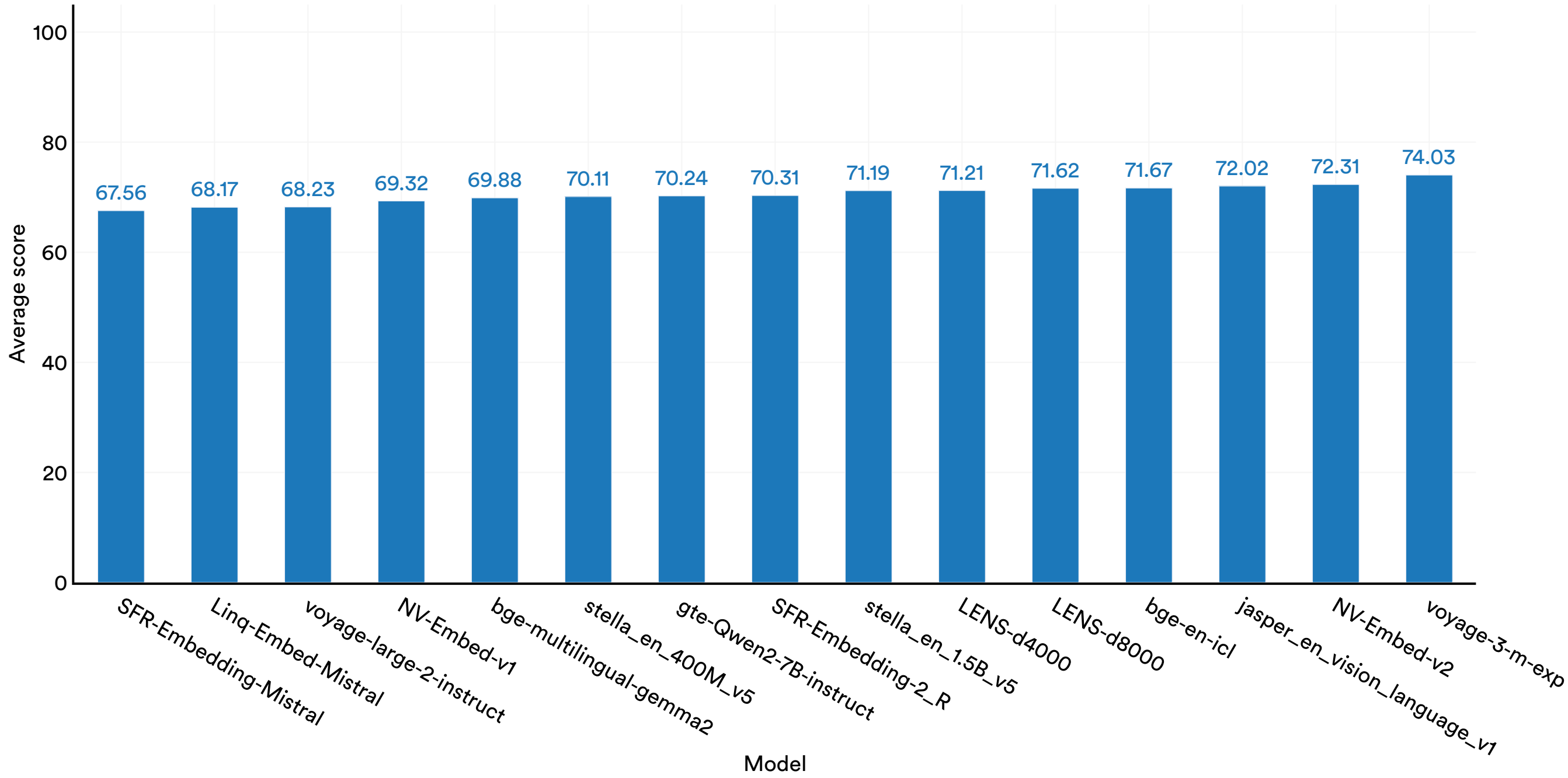

Figure 2.2.20

Highlight:

# Evaluating Retrieval Across Long Contexts

As AI models have advanced, their ability to handle longer contexts has significantly improved. For example, models like GPT-4 and Llama 2, released in 2023 by OpenAI and Meta, featured context windows of 8,000 and 4,000 tokens, respectively. In contrast, more recent models such as GPT-4o (May 2024) and Gemini 2.0 Pro Experimental (February 2025) boast context windows ranging from 128 thousand to 2 million. These extended context windows allow users to input and process increasingly large amounts of data, enabling more complex and detailed interactions.

As the context windows of LLMs have expanded, evaluating their performance in long-context settings has become increasingly important. However, existing long-context evaluation methods have been relatively limited. Typically, these evaluations focus on "needle-in-the-haystack" scenarios, where models are tasked with retrieving specific pieces of information from lengthy texts. While useful, such evaluations provide only a baseline assessment of a model's ability to function effectively in long-context environments.

In 2024, several new evaluation suites were introduced to address the limitations of long-context model assessments and improve their evaluation. One such benchmark is Nvidia's RULER, which assesses long-context performance by examining retrieval performance and multihop reasoning, aggregation, and question answering. Among the models evaluated on RULER, Gemini-1.5-Pro achieved the highest weighted performance average (95.5), followed by GPT-4 (89.0) and GLM4(88.0) (Figure 2.2.21). The researchers behind RULER also revealed that many models suffer performance issues in longer context settings. In fact, the RULER team demonstrated that while most popular LLMs claim context sizes of 32K tokens or greater, only half of them can maintain satisfactory performance at the length of 32K. This means that their actual operational context windows are shorter than those claimed by their developers (Figure 2.2.22).

RULER: weighted average score (increasing)

Source: Hsieh et al., 2024 | Chart: 2025 AI Index report

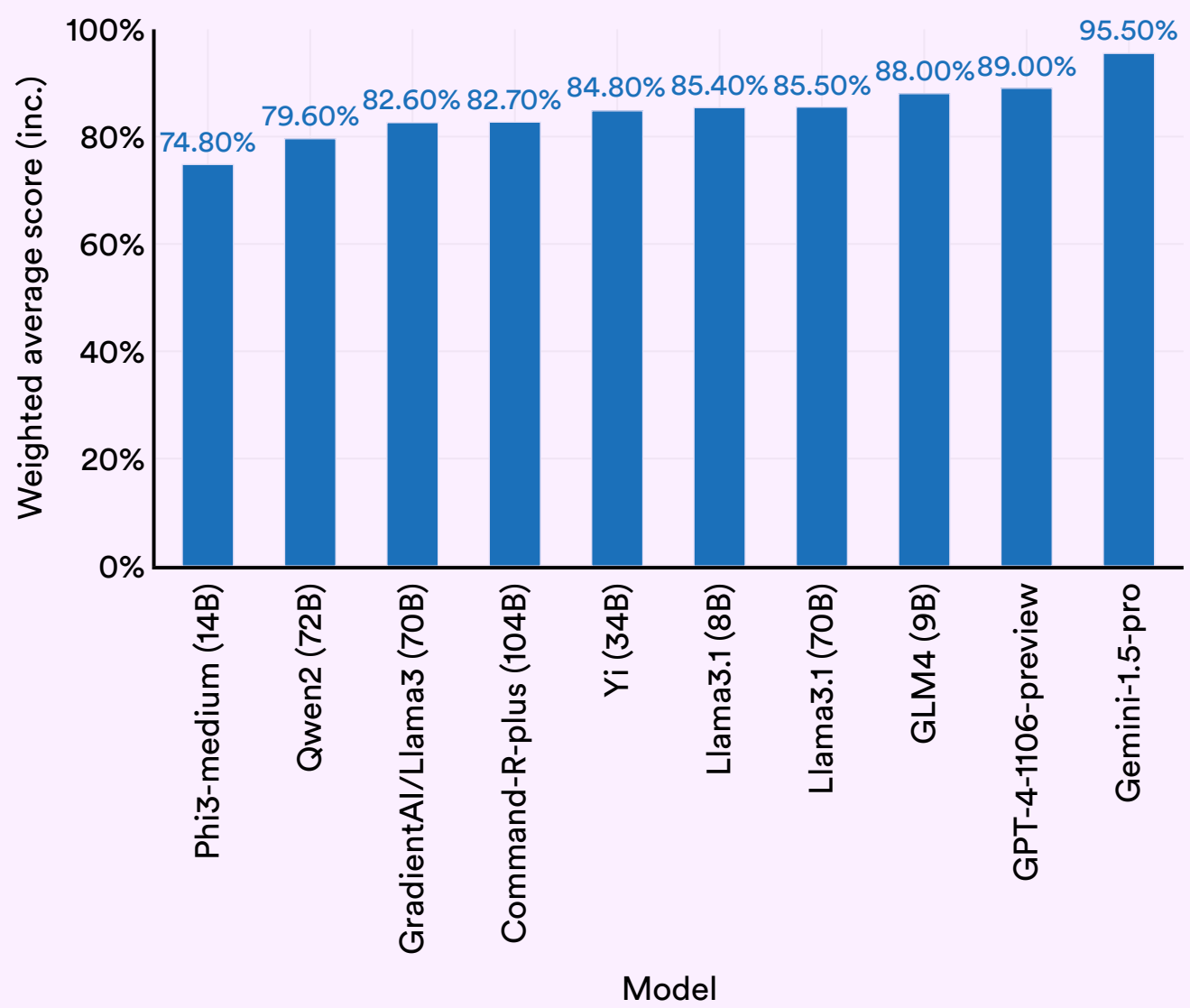

Figure 2.2.21

RULER: claimed vs. effective context length

Source: Hsieh et al., 2024 | Chart: 2025 AI Index report

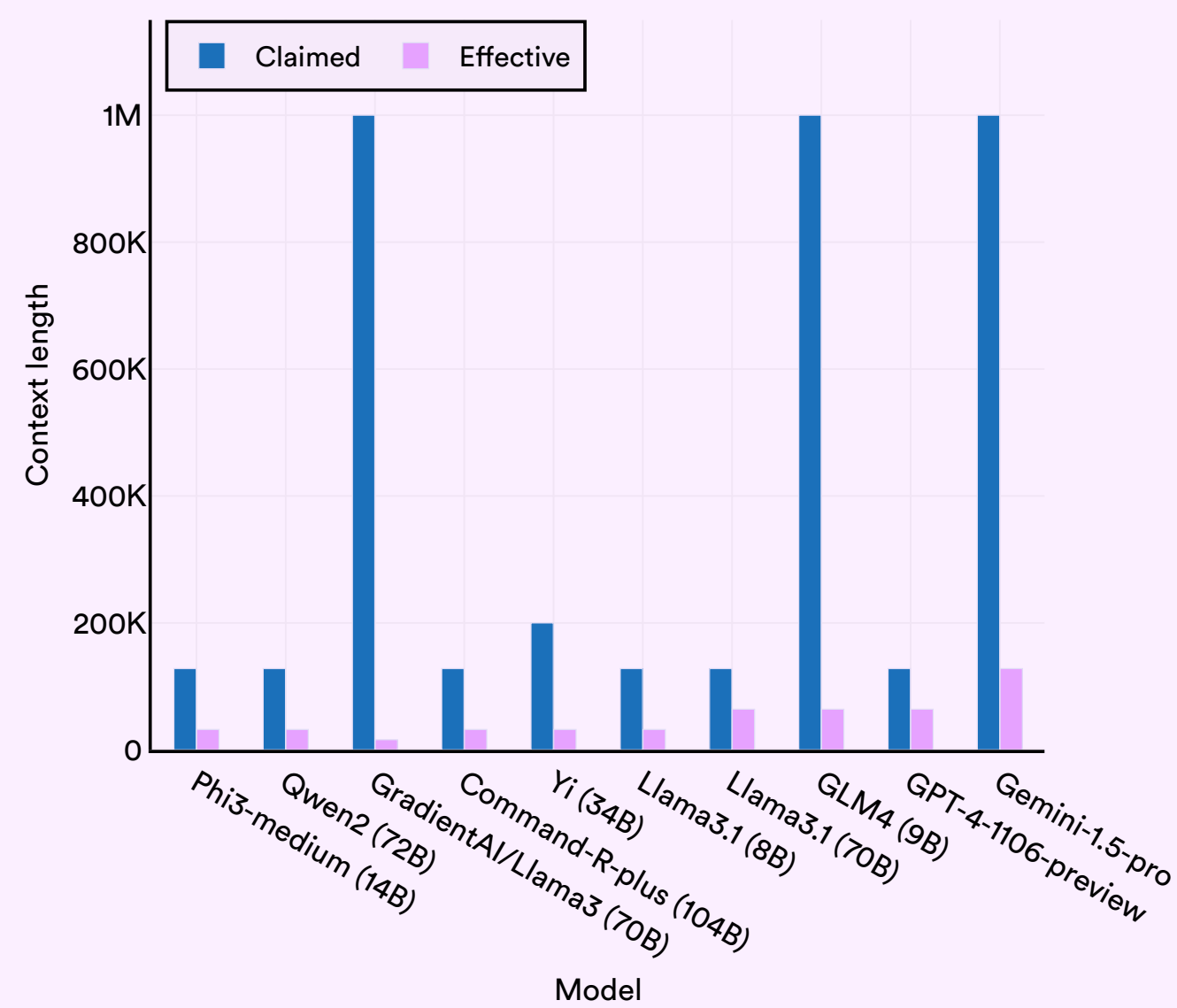

Figure 2.2.22

**Highlight:**

# Evaluating Retrieval Across Long Contexts (cont’d)

HELMET (How to Evaluate Long-Context Models Effectively and Thoroughly), an Intel and Princeton collaboration, is another long-context evaluation benchmark introduced in 2024. The researchers behind HELMET were motivated by the inadequacies of existing benchmarks, which suffered from insufficient coverage of downstream tasks, context lengths too short to test evolving long-context capabilities, and unreliable metrics (Figure 2.2.23). Even more comprehensive than RULER, HELMET features seven long-context evaluation categories, including synthetic recall, passage re-ranking, and generation with citations. Figure 2.2.24 illustrates the average performance of several notable models on the HELMET benchmark across 8K, 32K, and 128K context settings. While models like GPT-4, Claude 3.5 Sonnet, and Llama 3.1-70B struggle with performance degradation in longer context settings, others, such as Gemini 1.5 Pro and the August variant of GPT-4, maintain their effectiveness. The introduction of benchmarks like RULER and HELMET highlights how the rapid evolution of LLMs is compelling researchers to rethink and refine evaluation methodologies.

**Comparing long-context benchmarks**
Source: Yen et al., 2024
Figure 2.2.23

| | Type of tasks | | | | | | | Benchmark features | | |
|---|---|---|---|---|---|---|---|---|---|---|
| | Cite | RAG | Re-rank | Long-QA | Summ | ICL | Synthetic Recall | Robust Eval. | $L \geq 128k$ | Controllable $L$ |
| ZeroSCROLLS | ✗ | ✗ | ✗ | ✓ | ✓ | ✗ | ✗ | ✗ | ✗† | ✗ |
| LongBench | ✗ | ✓ | ✗ | ✓ | ✓ | ✓ | ✓ | ✗ | ✗† | ✗ |
| L-Eval | ✗ | ✓ | ✗ | ✓ | ✓ | ✗ | ✗ | ✓‡ | ✗† | ✗ |
| RULER | ✗ | ✗ | ✗ | ✗ | ✗ | ✗ | ✓ | ✓ | ✓ | ✓ |
| ∞Bench | ✗ | ✗ | ✗ | ✓ | ✓ | ✗ | ✓ | ✗ | ✓ | ✓ |
| HELMET (Ours) | ✓ | ✓ | ✓ | ✓ | ✓ | ✓ | ✓ | ✓ | ✓ | ✓ |

**HELMET: average score**
Source: Yen et al., 2024 | Chart: 2025 AI Index report

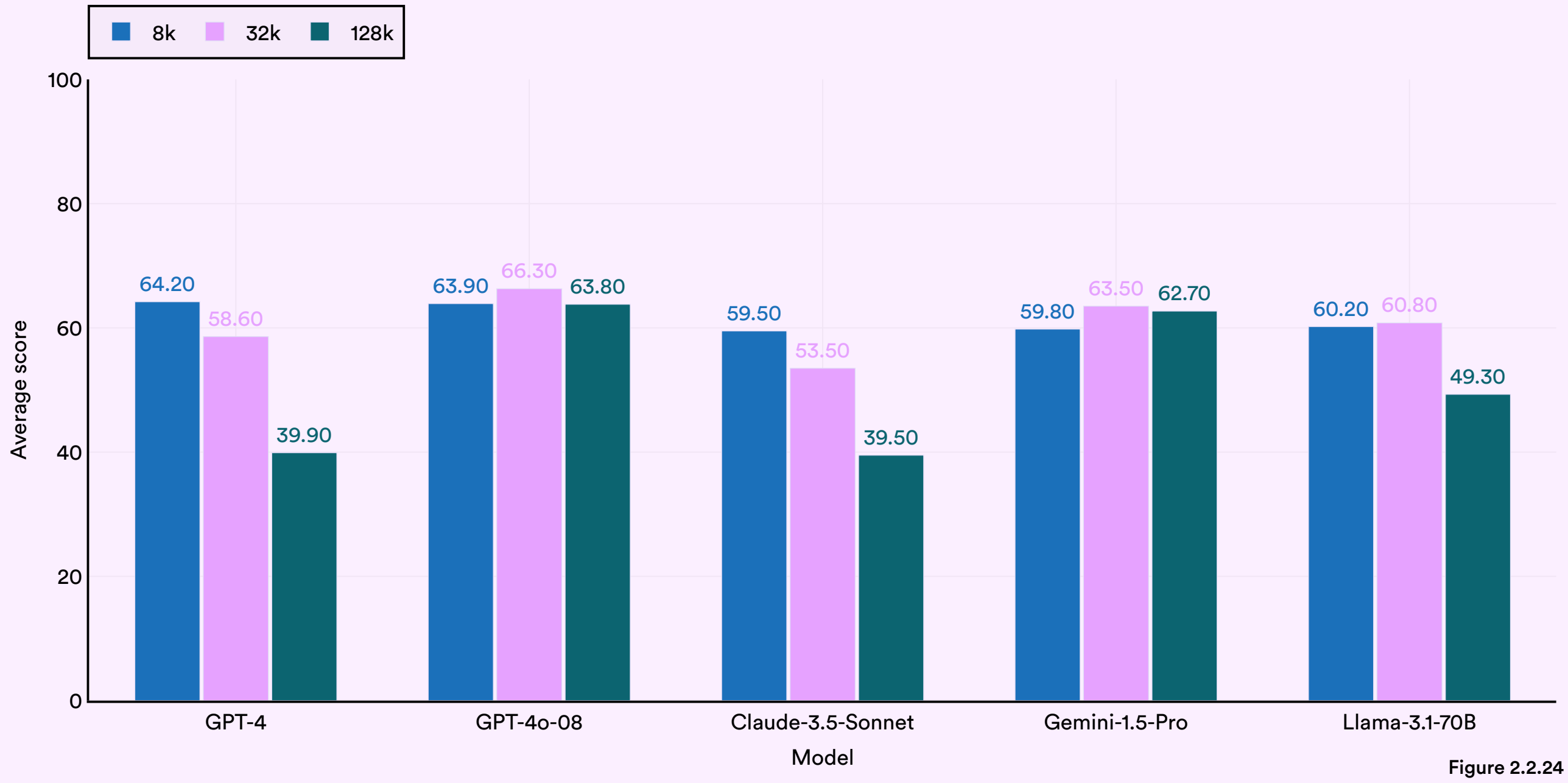

Figure 2.2.24

Computer vision allows machines to understand images and videos and to create realistic visuals from textual prompts or other inputs. This technology is widely used in fields such as autonomous driving, medical imaging, and video game development.

# 2.3 Image and Video

## Understanding

Vision models are evaluated on their ability to understand and reason about the content of images and videos. Vision understanding was one of the first AI capabilities widely tested during the deep learning era. ImageNet, created by Fei-Fei Li and extensively covered in past editions of the AI Index, served as a foundational benchmark for image understanding. As AI systems have advanced, researchers have shifted toward evaluating image models on more complex and comprehensive understanding tasks, such as those involving video or commonsense reasoning in images.

In the ImageNet era, vision algorithms were tasked with more straightforward tasks (e.g., classifying images into predefined categories). However, modern computer vision benchmarks like VCR and MVBench introduce more open-ended challenges, where no fixed categories or classes exist. In these cases, algorithms process natural language questions, identify objects from an open set of images, and generate answers based on image content or prior knowledge.

### VCR: Visual Commonsense Reasoning

Introduced in 2019 by researchers from the University of Washington and the Allen Institute for AI, the Visual Commonsense Reasoning (VCR) challenge tests the commonsense visual reasoning abilities of AI systems. In this challenge, AI systems not only answer questions based on images but also reason about the logic behind their answers (Figure 2.3.1). Performance in VCR is measured using the Q->AR score, which evaluates the machine's ability to both select the correct answer to a question (Q->A) and choose the appropriate rationale behind that answer (Q->R).

**Sample question from Visual Commonsense Reasoning (VCR) challenge**
Source: Zellers et al., 2018

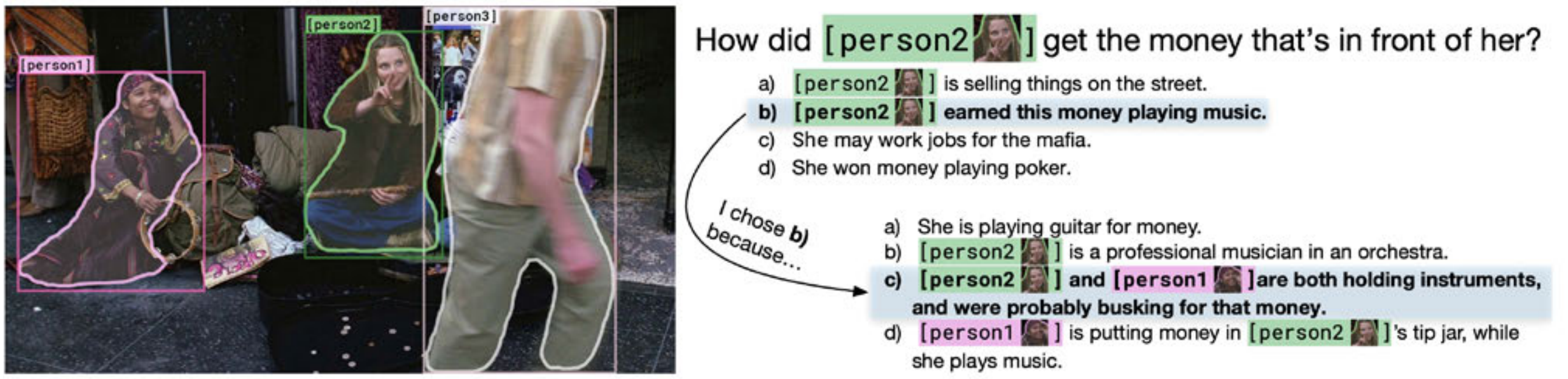

Figure 2.3.1

The VCR benchmark was one of the few benchmarks routinely featured in the AI Index where AI systems consistently fell short of the human baseline. However, 2024 marked a turning point, with AI systems finally reaching this baseline. A model posted to the leaderboard in July 2024 achieved a score of 85.0, matching the human benchmark (Figure 2.3.2). This milestone represented a significant 4.2% improvement on the benchmark since 2023. Even previously challenging benchmarks are now being surpassed.

**Visual Commonsense Reasoning (VCR) task: Q->AR score**
Source: VCR Leaderboard, 2025 | Chart: 2025 AI Index report

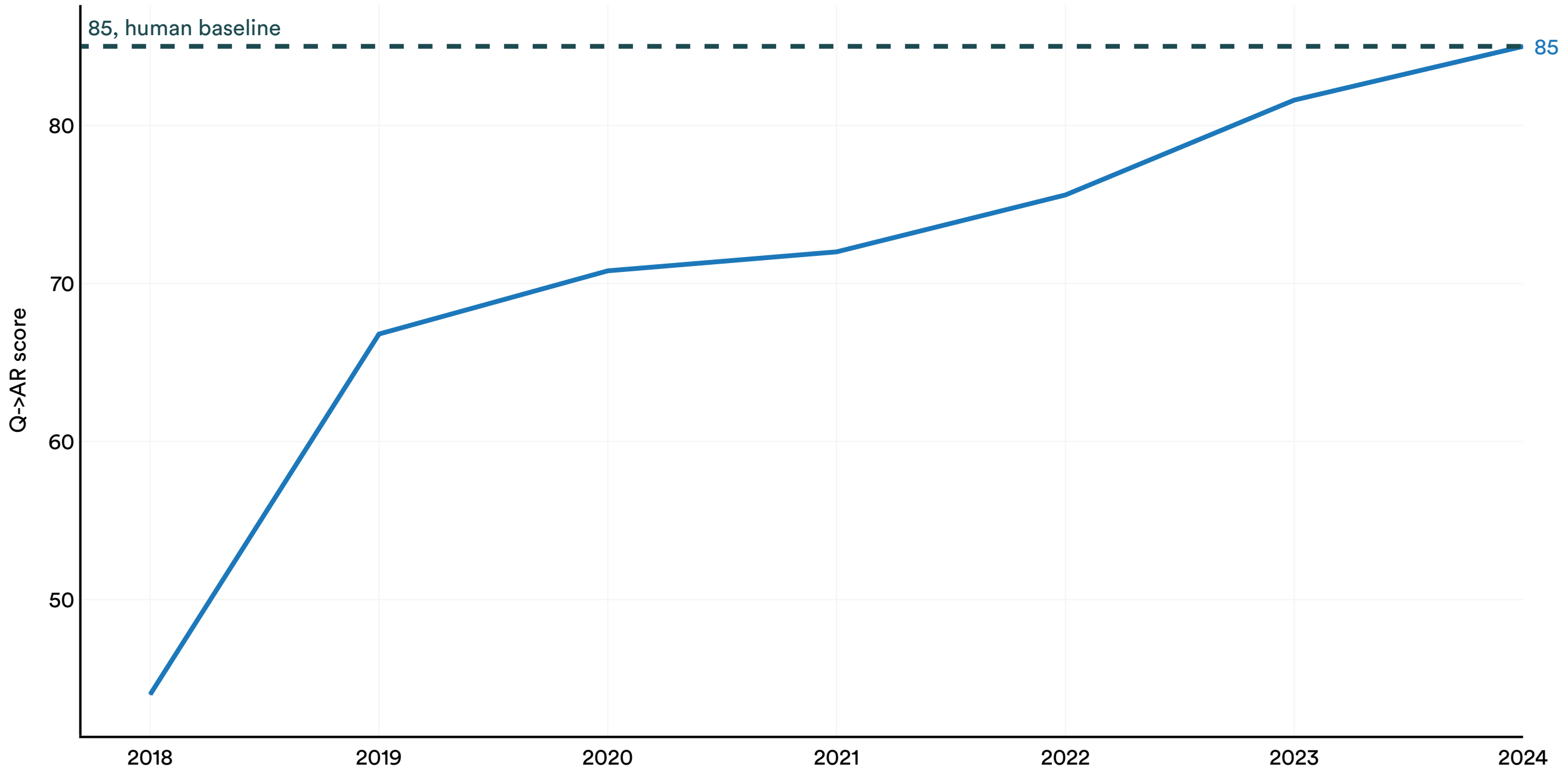

Figure 2.3.2

## MVBench

MVBench, introduced by a team of researchers from Hong Kong and China in 2023, is a challenging, multimodal, video-understanding benchmark.[11] Unlike earlier video benchmarks that primarily tested spatial understanding through static image tasks, MVBench incorporates more complex video tasks requiring temporal reasoning across multiple frames (Figure 2.3.3).

**Sample tasks on MVBench**
Source: Li et al., 2023
Figure 2.3.3

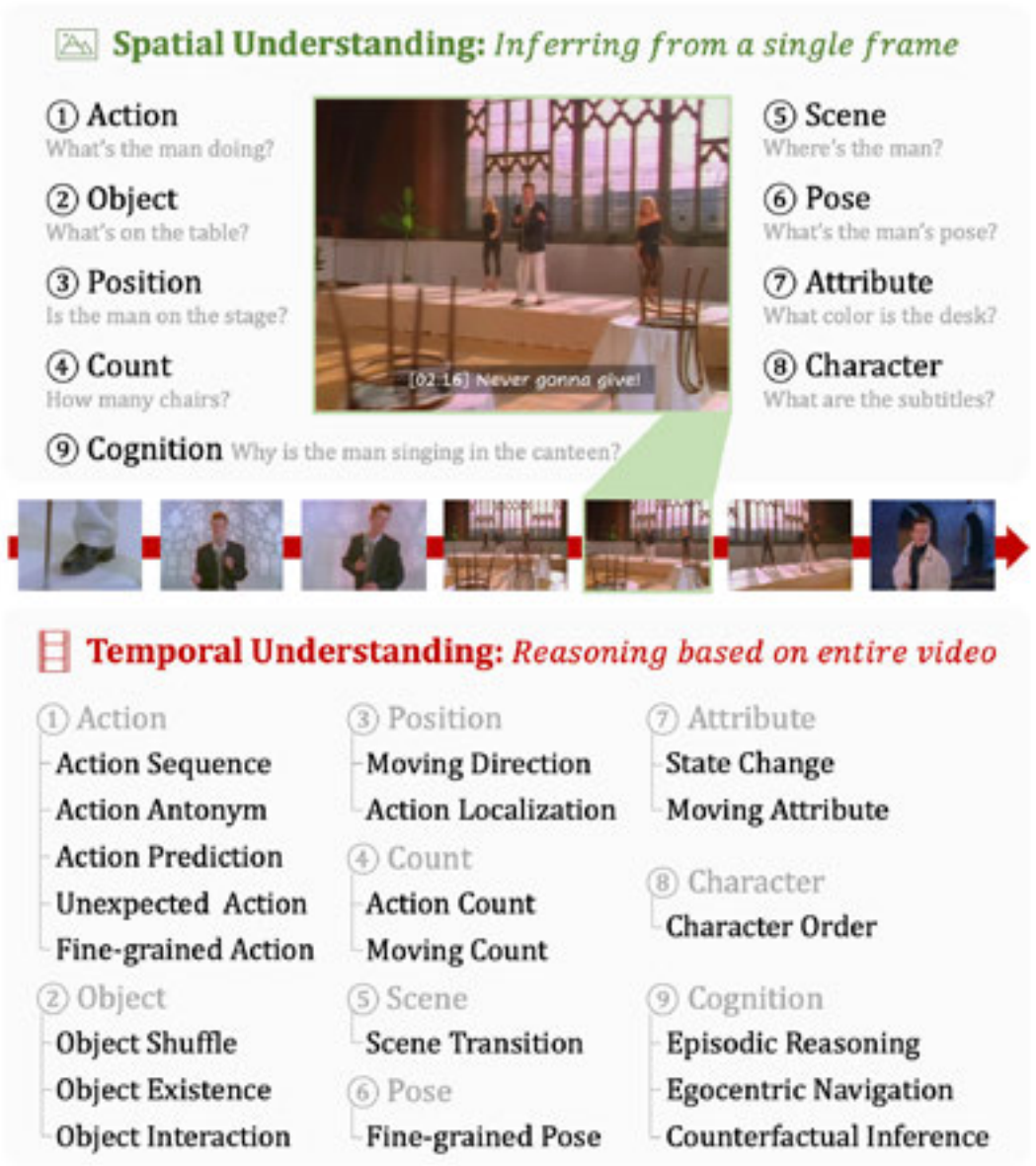

11 The researchers were affiliated with the Chinese Academy of Sciences, University of Chinese Academy of Sciences, Shanghai AI Laboratory, the University of Hong Kong, Fudan University, and Nanjing University.

As of 2024, the top model on the MVBench leaderboard is Video-CCAM-7B-v1.2, built on the Queen 2.5-7B-Instruct language model. Its score of 69.23 marks a significant 14.6% improvement on the benchmark since its introduction in late 2023 (Figure 2.3.4). These results highlight the gradual but steady progress in the dynamic video understanding capabilities of AI models.

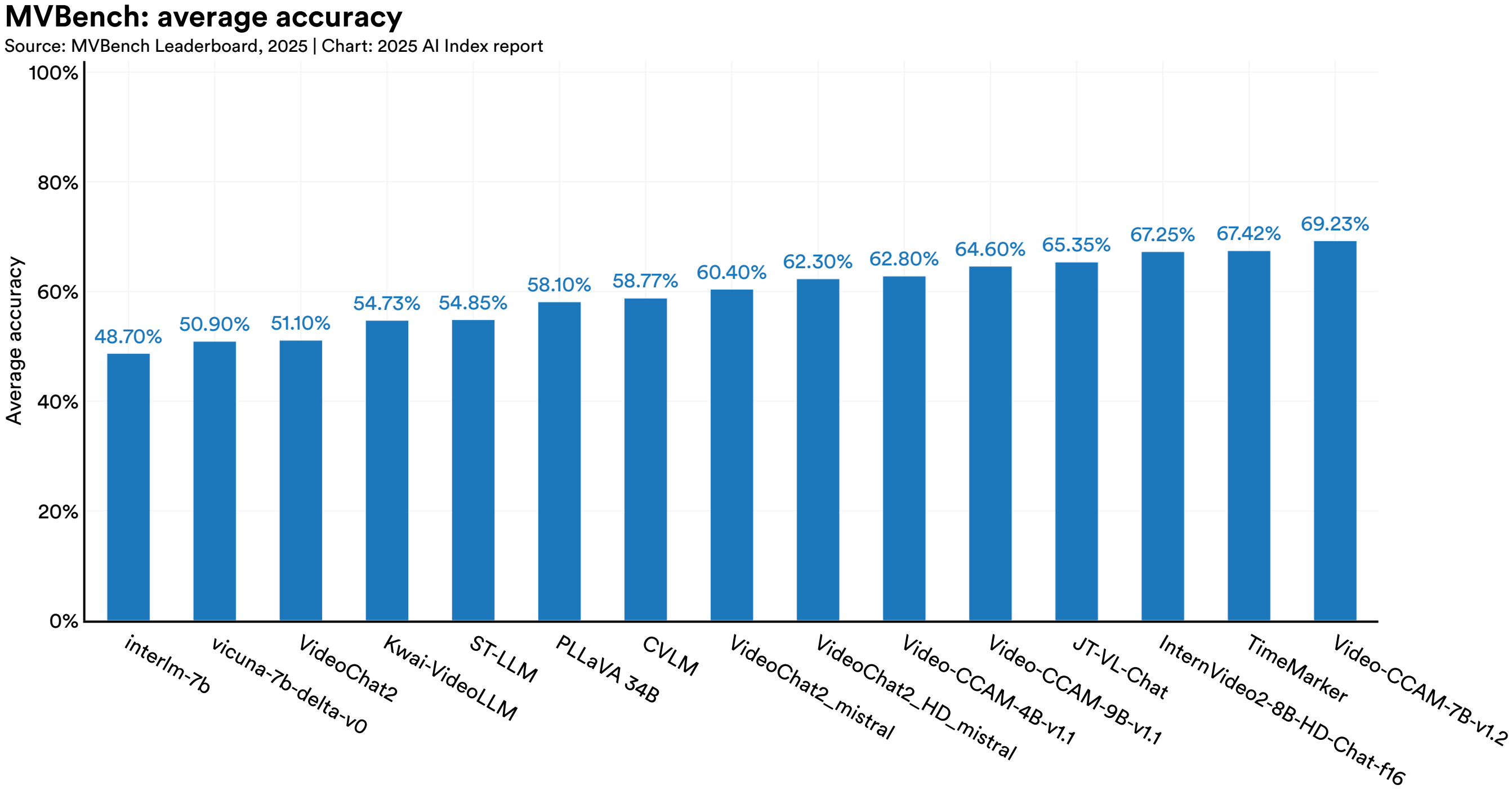

Figure 2.3.4

## Generation

Image generation is the task of generating images that are indistinguishable from real ones. As noted in last year's AI Index, today's image generators are so advanced that most people struggle to differentiate between AI-generated images and actual images of human faces (Figure 2.3.5). Figure 2.3.6 highlights several generations from various Midjourney model variants from 2022 to 2025 for the prompt "a hyper-realistic image of Harry Potter." The progression demonstrates the significant improvement in Midjourney's ability to generate hyper-realistic images over a two-year period. In 2022, the model produced cartoonish and inaccurate renderings of Harry Potter, but by 2025, it could create startlingly realistic depictions.

**Which face is real?**
Source: Which Face Is Real, 2024

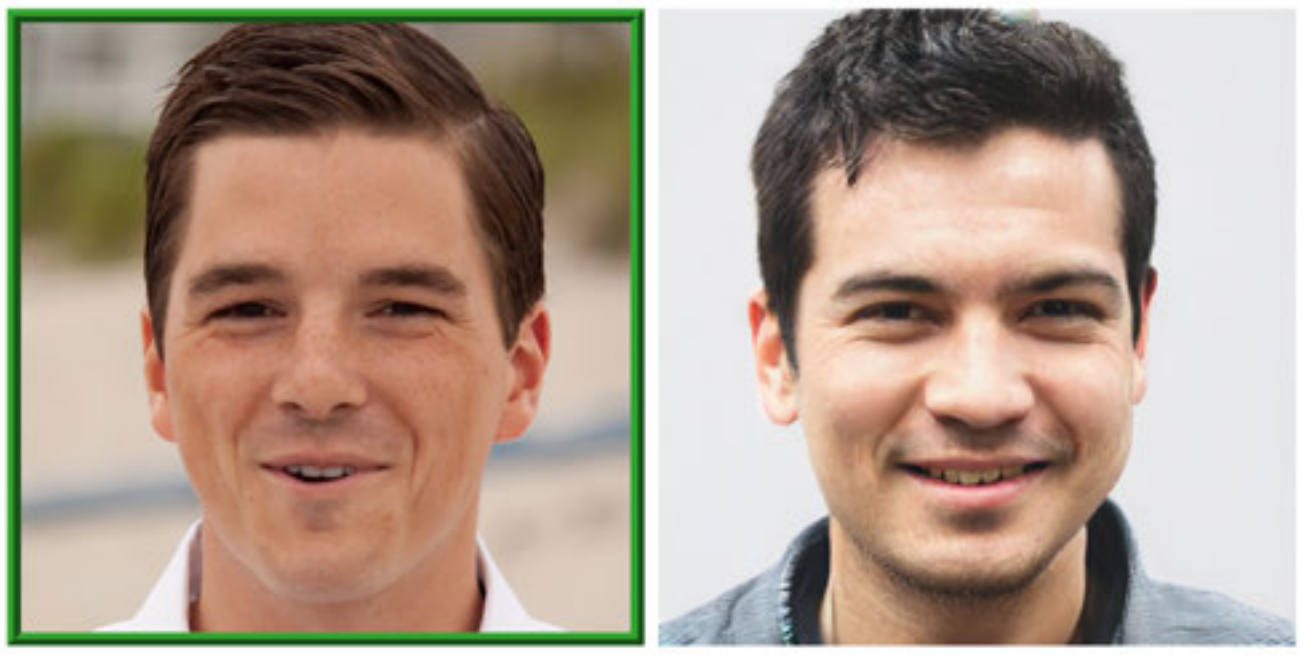

Figure 2.3.5

**Midjourney generations over time: "a hyper-realistic image of Harry Potter"**
Source: Midjourney, 2024

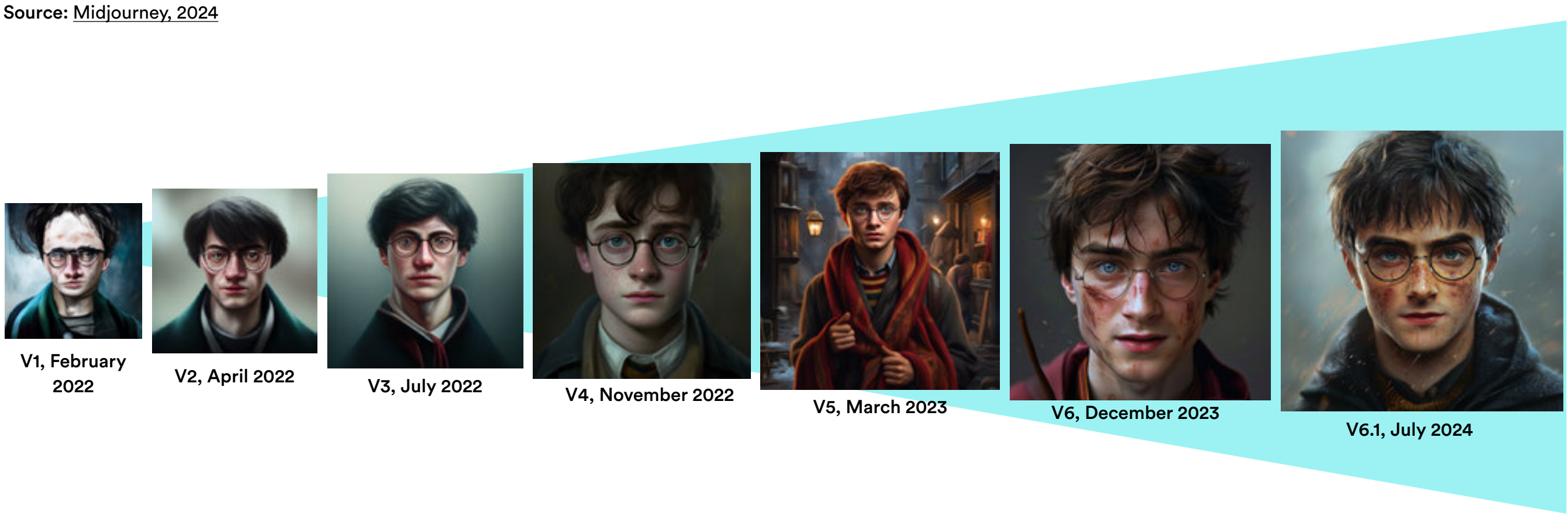

Figure 2.3.6

## Chatbot Arena: Vision

The AI community has increasingly embraced public evaluation platforms, such as the Chatbot Arena Leaderboard, to assess the capabilities of leading AI systems, including top AI image generators. This leaderboard also features a Vision Arena, which ranks the performance of over 50 vision models. Users can submit text-to-image prompts, such as “Batman drinking a coffee,” and vote for their preferred generation (Figure 2.3.7). To date, the Vision Arena has garnered more than 150,000 votes.

As of early 2025, the top-ranked vision model on the leaderboard is Google’s Gemini-2.0-Flash-Thinking-Exp-1219 (Figure 2.3.8). Similar to other Chatbot Arena categories—such as general, coding, and math—the leading models are closely clustered in performance. For example, the gap between the top model and the fourth-ranked model, ChatGPT-4o-latest (2024-11-20), is just 3.4%.

**Sample from the Chatbot Vision Arena**
Source: Chatbot Arena Leaderboard, 2025

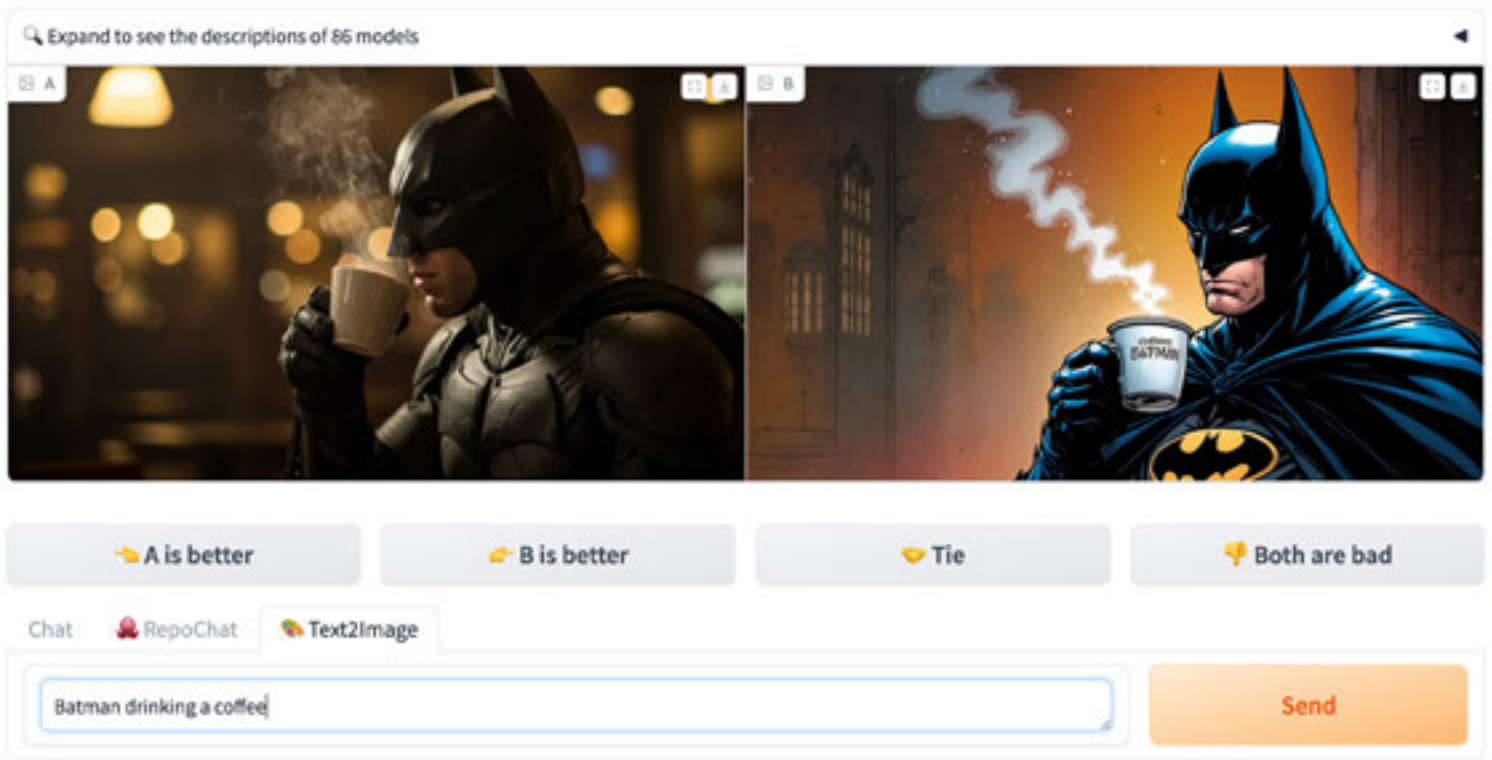

Figure 2.3.7

**LMSYS Chatbot Arena for LLMs: Elo rating (vision)**
Source: LMSYS, 2025 | Chart: 2025 AI Index report

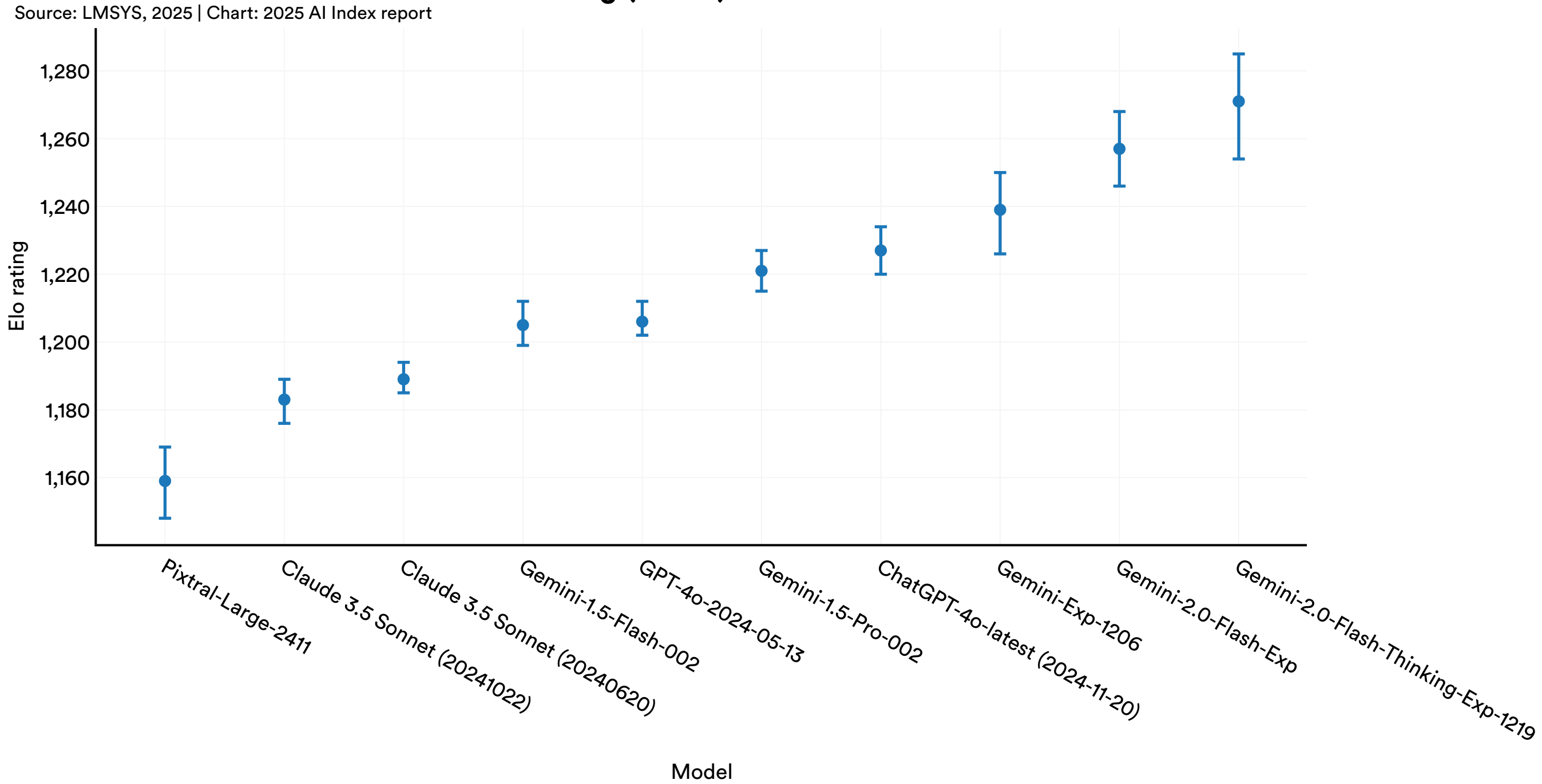

Figure 2.3.8

Highlight:

## The Rise of Video Generation

As highlighted in last year's AI Index, recent years have witnessed the rise of video generation models capable of creating videos from text prompts. While earlier models demonstrated some promise, they were plagued by significant limitations, such as producing low-quality videos, omitting sound, or generating only very short clips. However, 2024 marked a significant leap forward in AI video generation, with several major industry players unveiling advanced video generation systems.

In November 2023, Stability AI launched its Stable Video Diffusion model, their first foundation model capable of generating high-quality videos (Figure 2.3.9). The model follows a three-step process: text-to-image pretraining, video pretraining, and high-quality video fine-tuning. Shortly after, in March, Stability AI introduced Stable Video 3D, a model designed to generate multiple 3D views and videos of an object from a single image. In February 2024, OpenAI responded with a preview of Sora, its own video generation model, which moved out of research mode and became publicly accessible in December 2024. Sora can generate 20-second videos at resolutions up to 1080p (Figure 2.3.10). As a diffusion model, it creates a base video and progressively refines it by removing noise over multiple steps to enhance quality.

**Still generations from Stable Video Diffusion**
Source: Stability AI, 2025

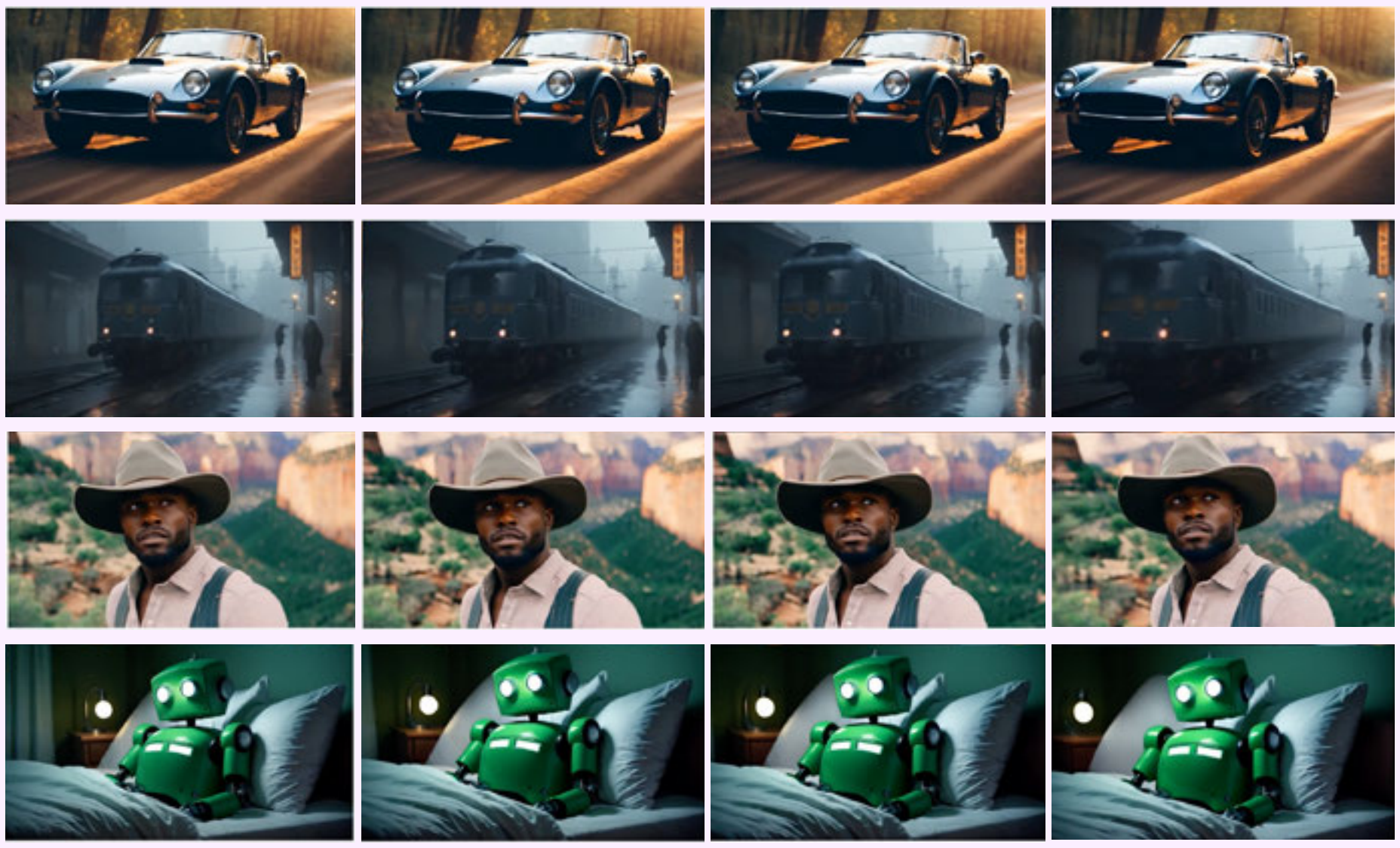

Figure 2.3.9

**Still generation from Sora**
Source: OpenAI, 2024

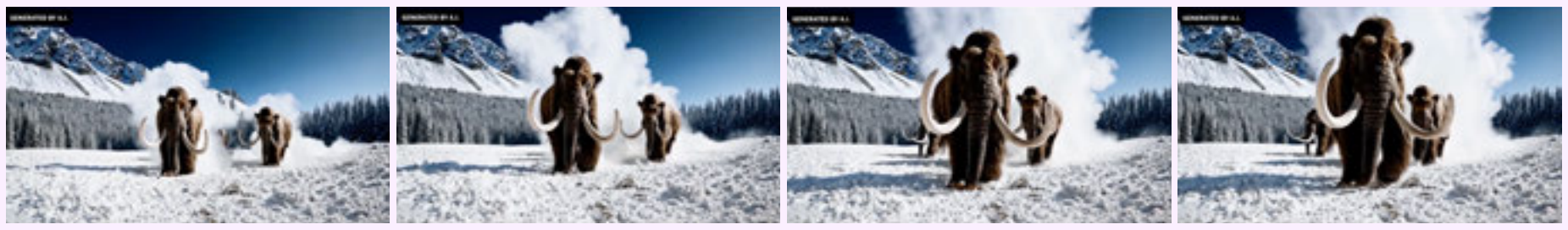

Figure 2.3.10

**Highlight:**

# The Rise of Video Generation (cont'd)

Other major tech players have entered the video generation space. In October 2024, Meta unveiled the latest version of its Movie Gen model. Unlike earlier iterations, the new Movie Gen includes advanced instruction-based video editing features, personalized video generation from images, and the ability to incorporate sound into videos. Meta's most advanced Movie Gen model can create 16-second videos at 16 frames per second, with a resolution of 1080p. Google also made significant strides in 2024, launching two major video generation models: Veo in May and Veo 2 in December. Internal benchmarking by Google revealed that Veo 2 outperformed other leading video generators, such as Meta's Movie Gen, Kling v1.5, and Sora Turbo. In user comparisons, videos generated by Veo 2 were consistently favored over those produced by competing models (Figure 2.3.11).

**Veo 2: overall preference**

Source: DeepMind, 2024 | Chart: 2025 AI Index report

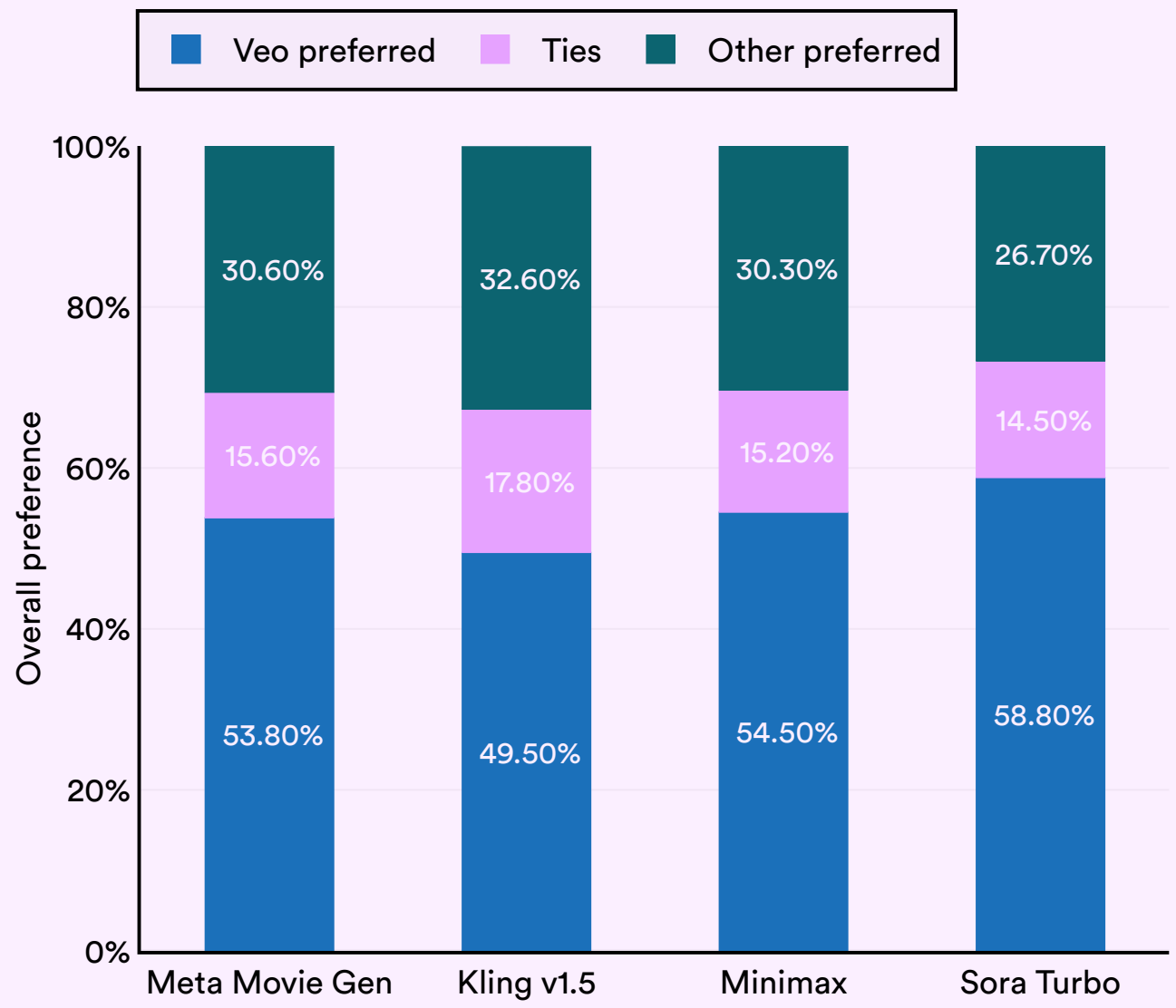

Figure 2.3.11

Smaller players have also made notable contributions to video generation, with models such as Runway's Gen-3 Alpha, Luma's Dream Machine, and Kuaishou's Kling 1.5. The remarkable progress in this field is evident when comparing videos generated in 2023 to those produced in 2024. A popular prompt on the internet, "Will Smith eating spaghetti," demonstrates this advancement, with videos generated in 2025 from one popular video generator Pika showcasing a dramatic improvement in quality compared to their 2023 counterparts (Figure 2.3.12).

**Will Smith eating spaghetti, 2023 vs. 2025**

Source: Pika, 2025

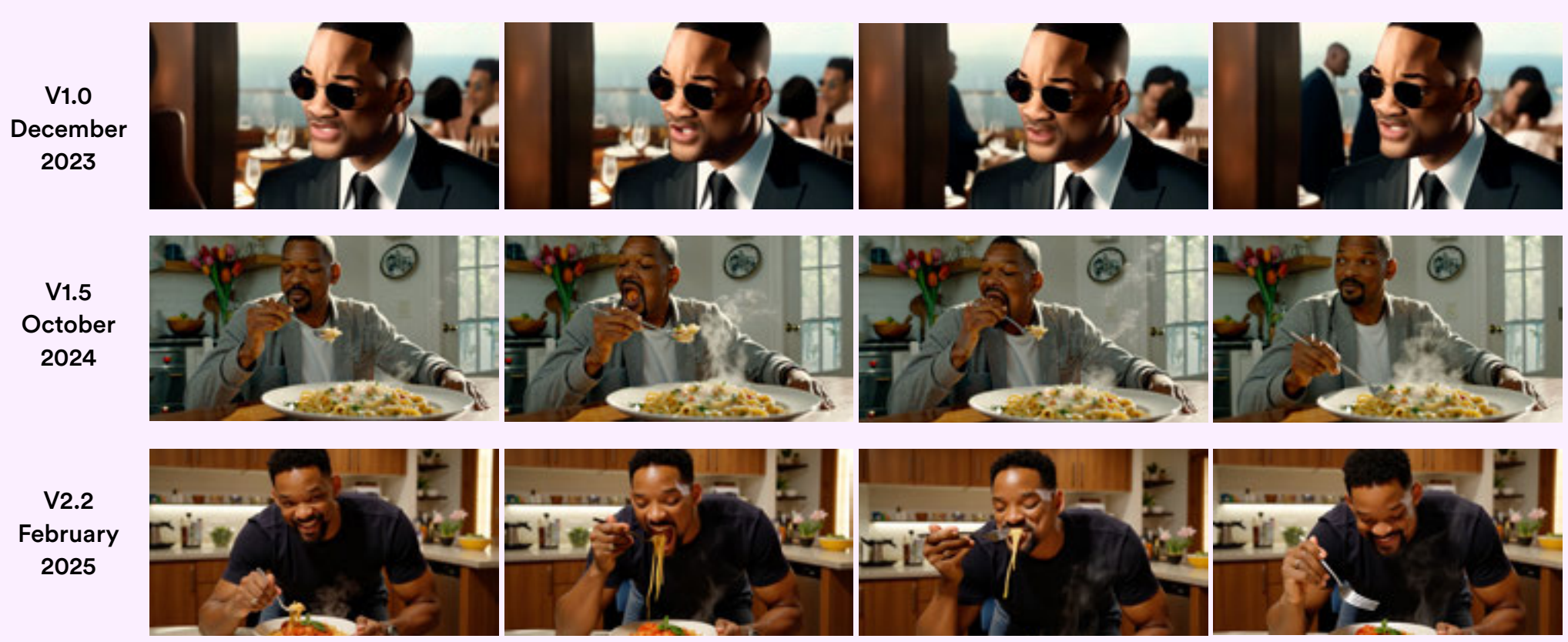

Figure 2.3.12

AI systems are adept at processing human speech, with audio capabilities that include transcribing spoken words to text and recognizing individual speakers. More recently, AI has advanced in generating synthetic audio content.

# 2.4 Speech

## Speech Recognition

Speech recognition is the ability of AI systems to identify spoken words and convert them into text. Speech recognition has progressed so much that today many computer programs and texting apps are equipped with dictation devices that can reliably transcribe speech into writing.

### LSR2: Lip Reading Sentences 2

The Oxford-BBC Lip Reading Sentences 2 (LRS2) dataset, introduced in 2017, is one of the most comprehensive public datasets for lipreading in authentic, in-the-wild scenarios (Figure 2.4.1). The dataset consists of audio-visual clips from a variety of talk shows and news programs. On automatic speech recognition (ASR) tasks, systems' ability to transcribe speech are evaluated on word error rate (WER), with lower scores indicating more precise transcription.

**Still images from the BBC lip reading sentences 2 dataset**
Source: Chung et al., 2024

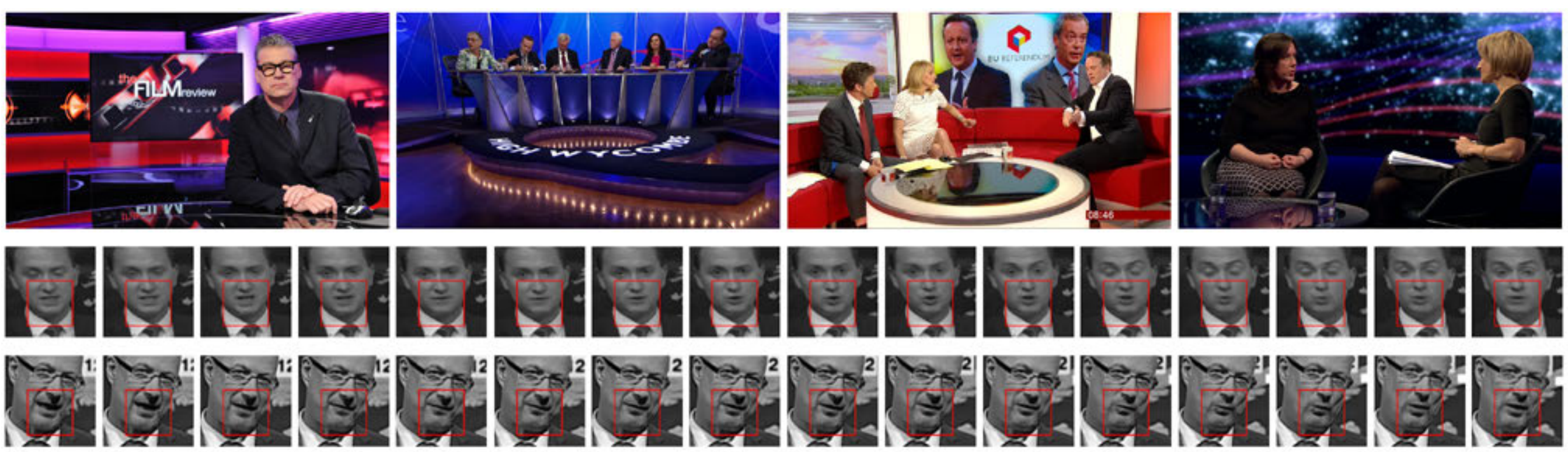

Figure 2.4.1

This year, the model Whisper-Flamingo set a new standard on the LRS2 benchmark, achieving a word error rate of 1.3 percent, surpassing the previous state-of-the-art score of 1.5 set in 2023 (Figure 2.4.2). However, given the already low WER, significant further improvements appear unlikely, suggesting that the benchmark may be nearing saturation.

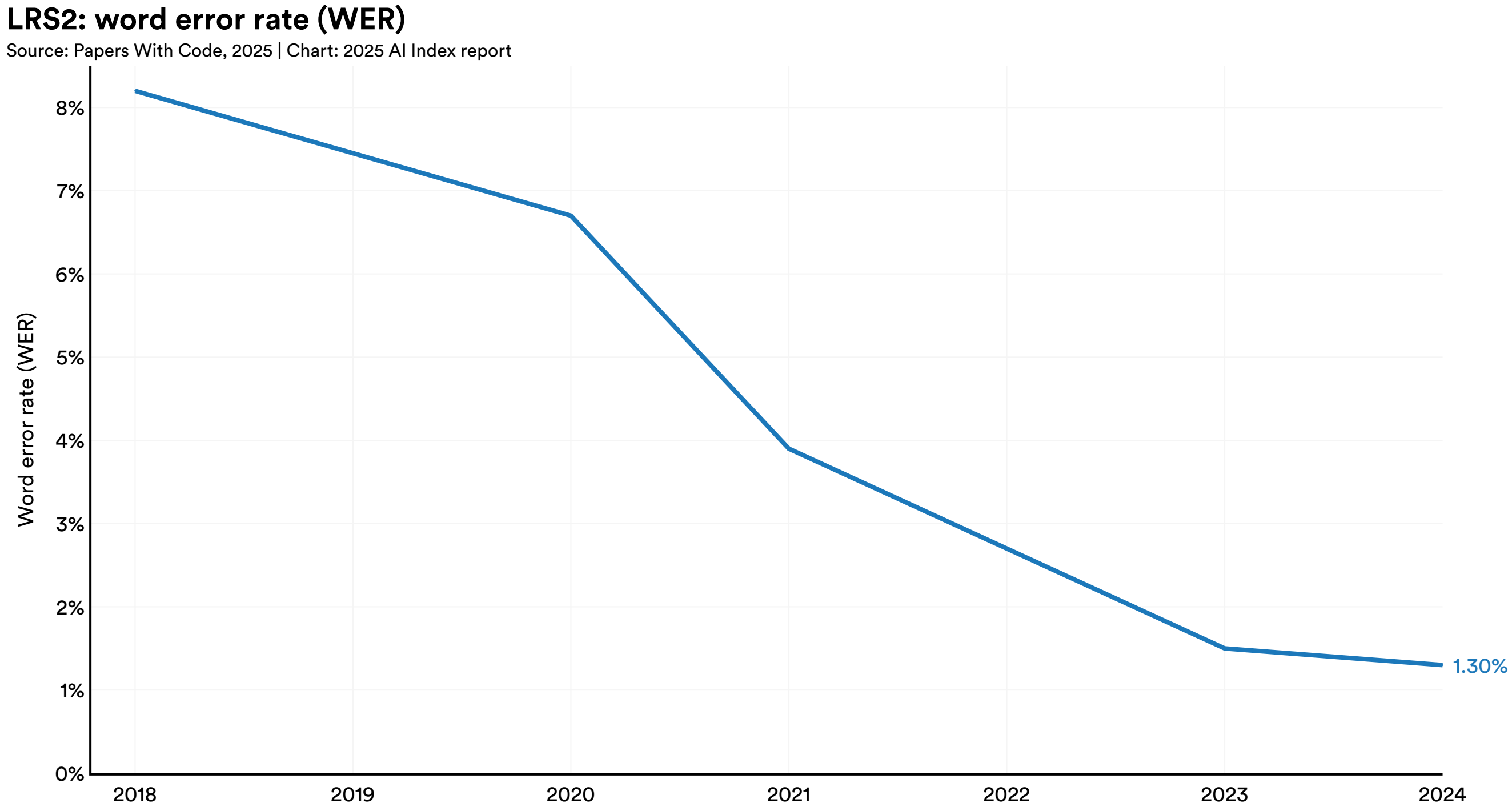

Figure 2.4.2

Coding involves the generation of instructions that computers can follow to perform tasks. Recently, LLMs have become proficient coders, serving as valuable assistants to computer scientists. There is also increasing evidence that many coders find AI coding assistants highly useful. As highlighted in last year's AI Index, LLMs have become increasingly proficient coders, to the extent that many foundational coding benchmarks, such as HumanEval, are slowly becoming saturated. In response, researchers have shifted their focus toward testing LLMs on more complex coding challenges.

# 2.5 Coding

## HumanEval

HumanEval, a benchmark introduced by OpenAI researchers in 2021, evaluates the coding abilities of AI systems through 164 challenging, handwritten programming problems (Figure 2.5.1). The current leader in HumanEval performance is Claude 3.5 Sonnet (HPT), which achieved a score of 100% (Figure 2.5.2).

**Sample HumanEval problem**
Source: Chen et al., 2023

```
def incr_list(l: list):
    """Return list with elements incremented by 1.
    >>> incr_list([1, 2, 3])
    [2, 3, 4]
    >>> incr_list([5, 3, 5, 2, 3, 3, 9, 0, 123])
    [6, 4, 6, 3, 4, 4, 10, 1, 124]
    """
    return [i + 1 for i in l]
```

Figure 2.5.1

**HumanEval: Pass@1**
Source: Papers With Code, 2025 | Chart: 2025 AI Index report

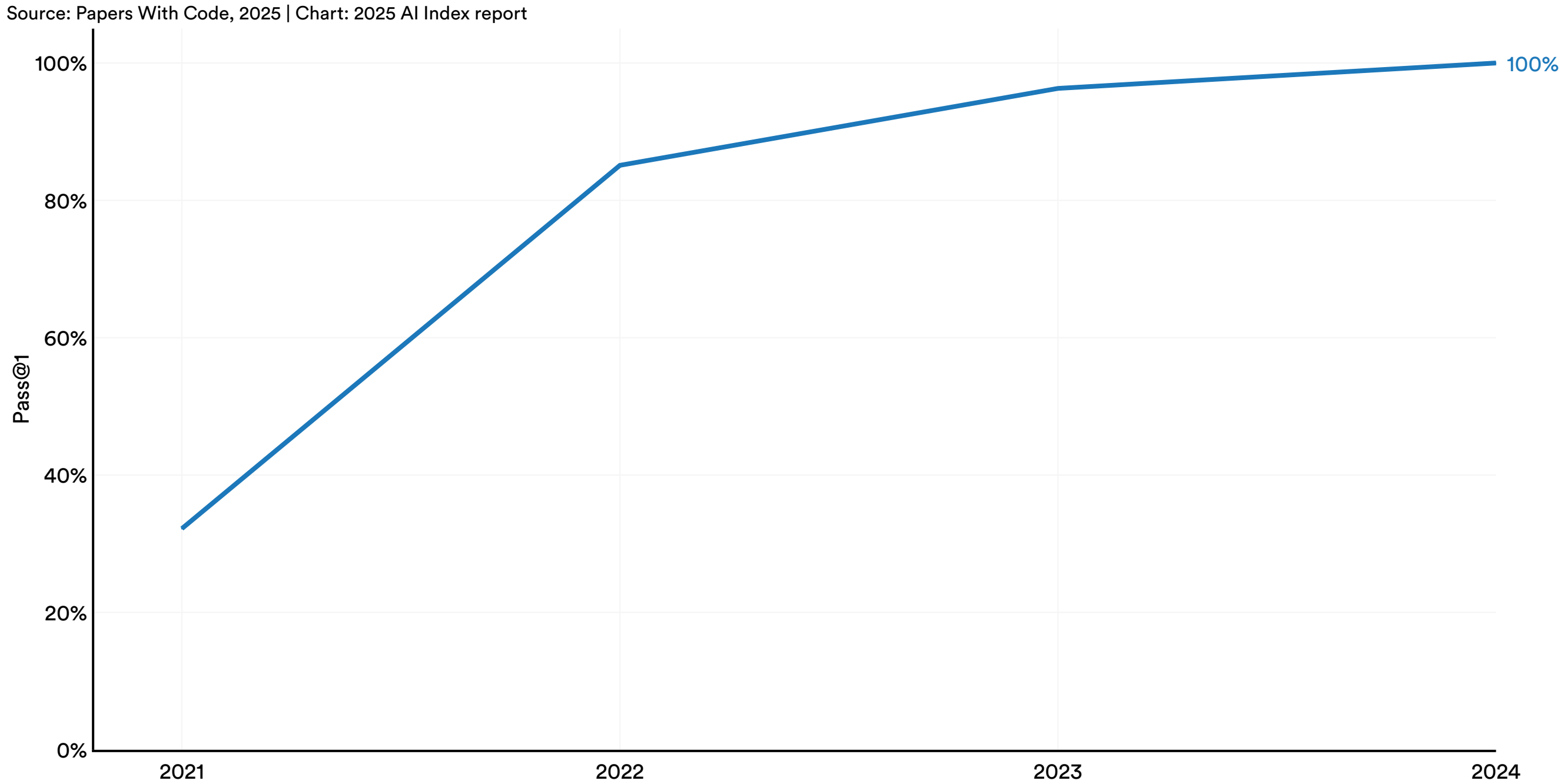

Figure 2.5.2

## SWE-bench

In October 2023, researchers from Princeton and the University of Chicago introduced SWE-bench, a dataset comprising 2,294 software engineering problems sourced from real GitHub issues and popular Python repositories (Figure 2.5.3). SWE-bench presents a tougher test for AI coding proficiency, demanding that systems coordinate changes across multiple functions, interact with various execution environments, and perform complex reasoning. SWE-bench features a Lite subset that is curated to make evaluation more accessible and a Verified subset that is filtered by a human annotator. The charts below report on the Verified score.

SWE-bench highlights the rapid improvement of LLMs on tasks that were once considered extremely demanding. At the end of 2023, the best performing model on SWE-bench achieved a score of just 4.4%. By early 2025, the top model, OpenAI's o3 model, is reported to have successfully solved 71.7% of the problems on the Verified benchmark set (Figure 2.5.4). This significant performance increase suggests that AI researchers may soon need to develop more challenging coding benchmarks to effectively test LLMs.

### A sample model input from SWE-bench

Source: Jimenez et al., 2023

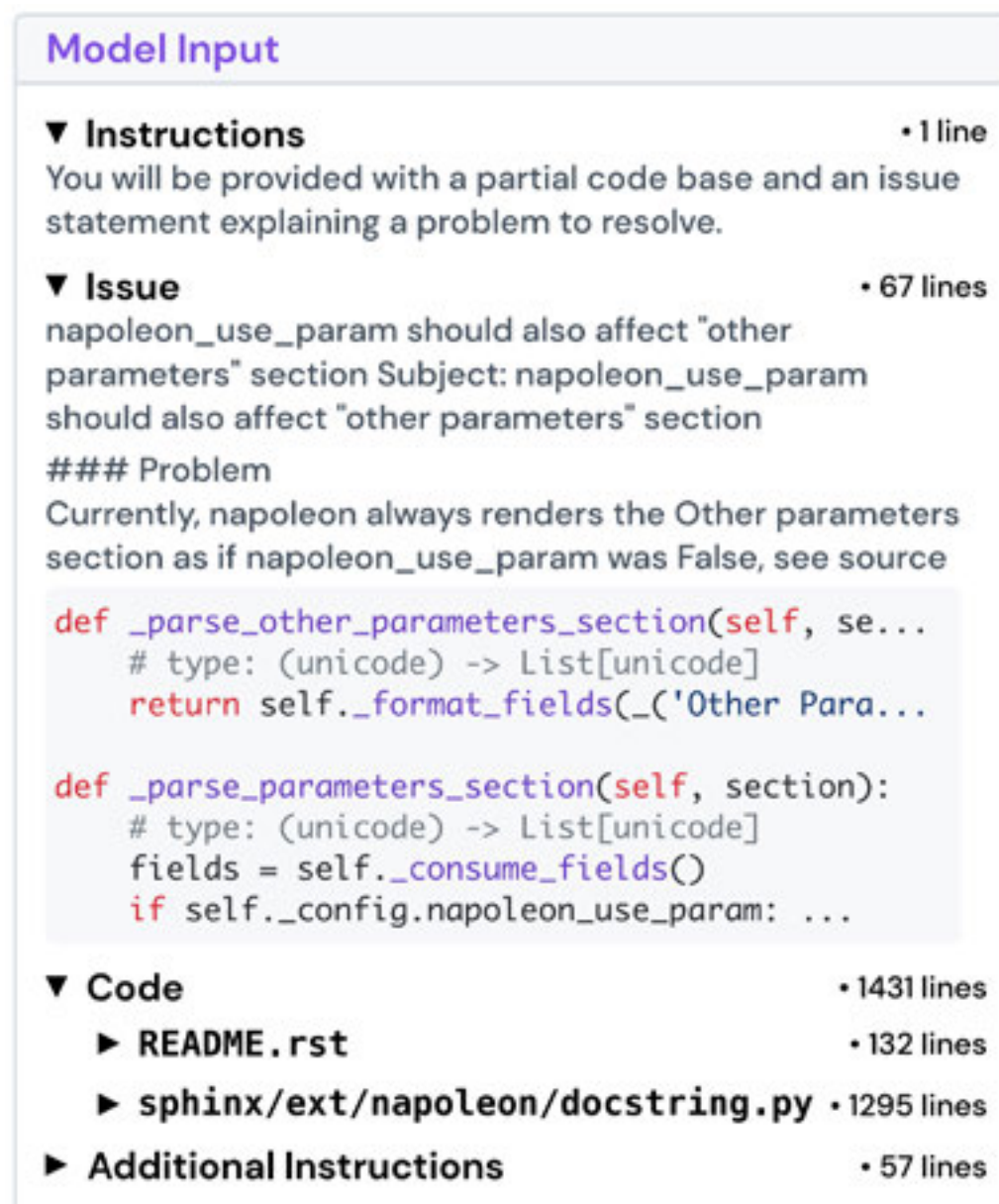

Figure 2.5.3

### SWE-bench: percent solved

Source: SWE-bench Leaderboard, 2025; OpenAI, 2024 | Chart: 2025 AI Index report

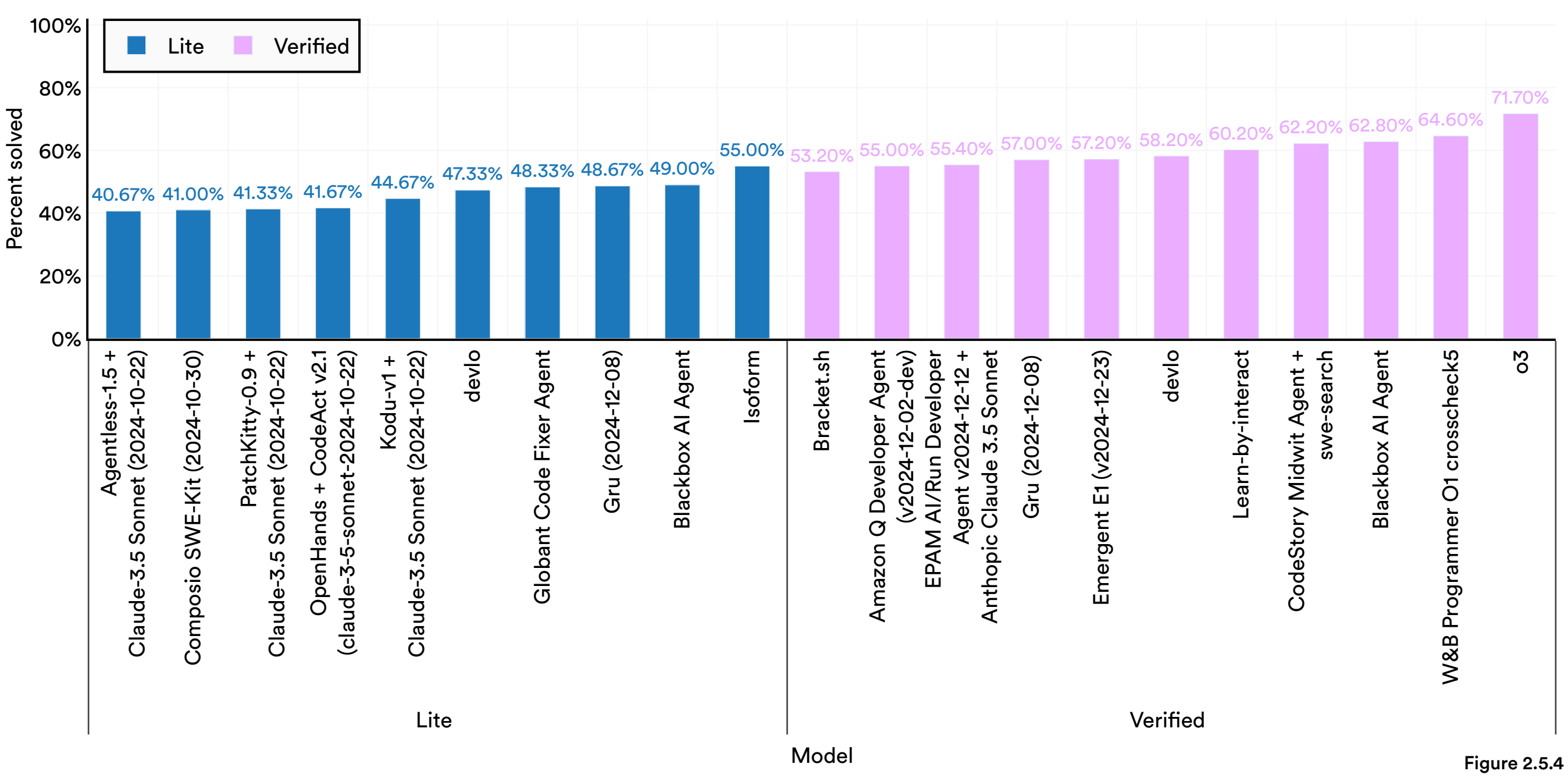

Figure 2.5.4

## BigCodeBench

One limitation of existing coding benchmarks is that many are restricted to short, self-contained algorithmic tasks or standalone function calls. However, solving complex and practical tasks often requires the ability to invoke diverse functions, such as tools for data analysis or web development. Effective coding also requires the ability to follow coding instructions expressed in language, a task not tested by many current coding benchmarks.

To address the limitations of existing coding benchmarks, an international team in 2024 released BigCodeBench, a comprehensive, diverse, and challenging benchmark for coding evaluation (Figure 2.5.5). BigCodeBench requires LLMs to invoke multiple function calls across 139 libraries and seven domains, encompassing 1,140 fine-grained tasks. Current AI systems struggle on BigCodeBench. For example, on both the "complete" (code completion based on structured docstrings) and "instruct" (code completion based on natural-language instructions) tasks on the hard subset of the benchmark, the current best model, OpenAI's o1, achieves an average score of just 35.5 (Figure 2.5.6). Models perform slightly better on the full set of the benchmark (Figure 2.5.7). BigCodeBench highlights the gap that persists for AI systems to achieve human-level coding proficiency.

### Programming tasks in BigCodeBench

Source: Zhuo et al., 2024

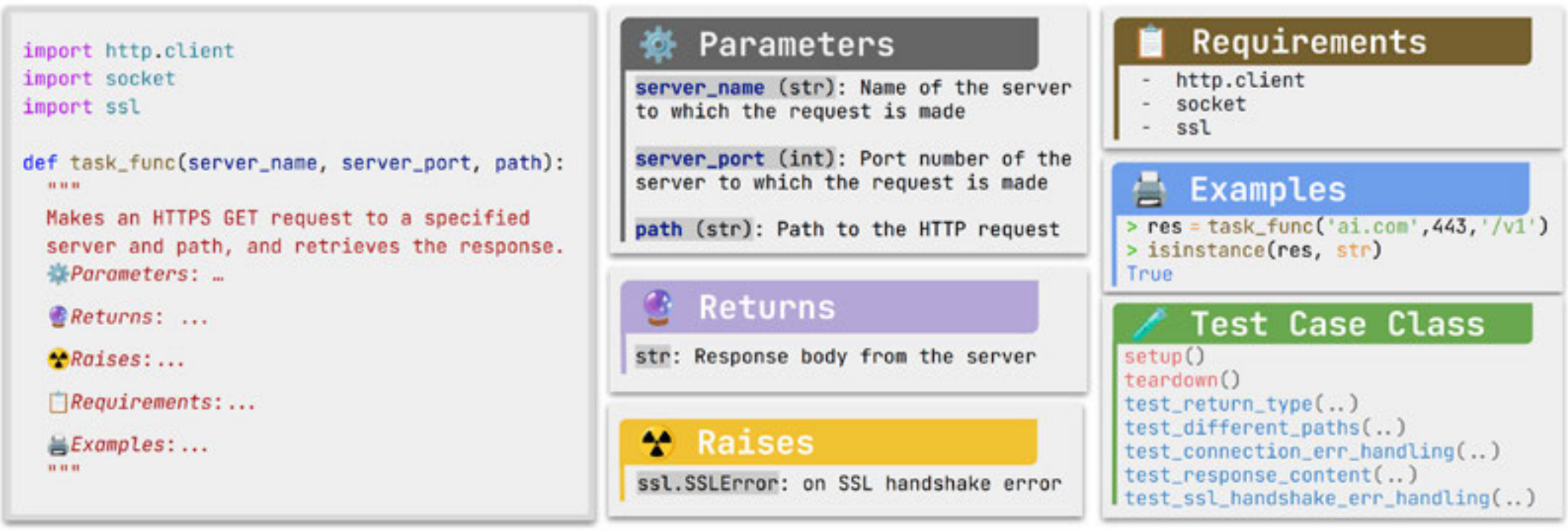

Figure 2.5.5

### BigCodeBench on the hard set: Pass@1 (average)

Source: Hugging Face, 2025 | Chart: 2025 AI Index report

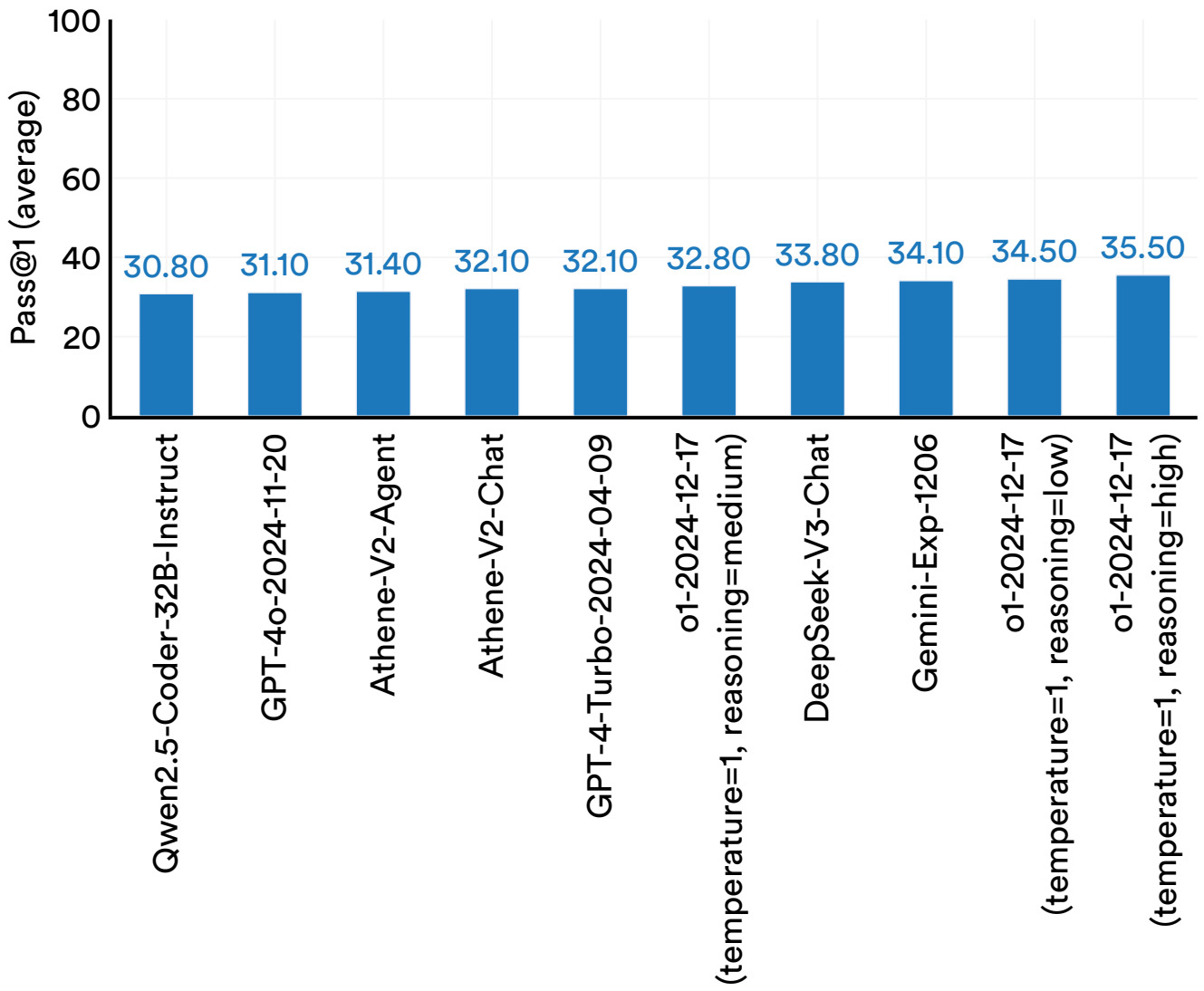

Figure 2.5.6

### BigCodeBench on the full set: Pass@1 (average)

Source: Hugging Face, 2025 | Chart: 2025 AI Index report

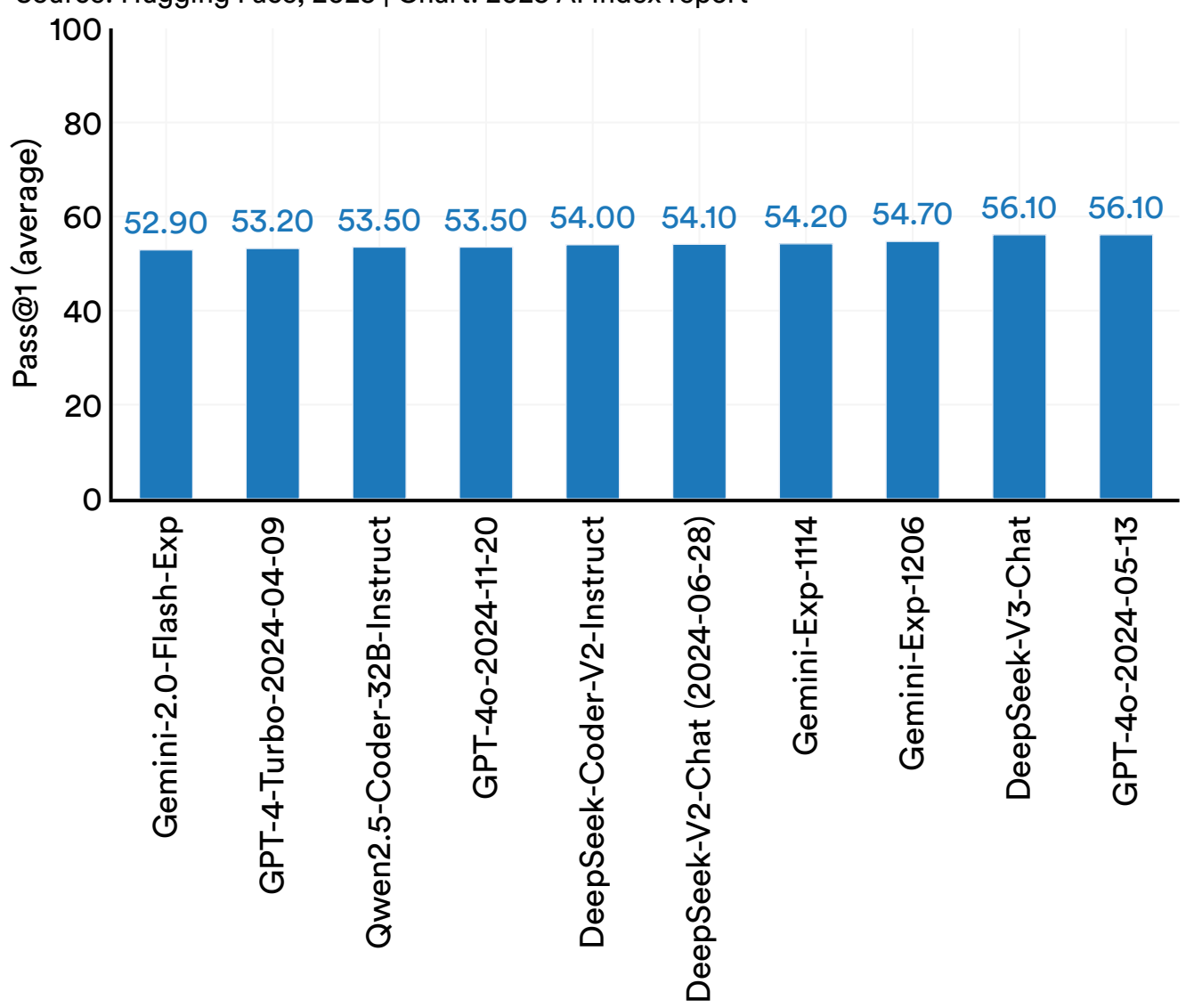

Figure 2.5.7

## Chatbot Arena: Coding

The Chatbot Arena LLM leaderboard now features a coding filter, offering valuable insights into how coders and the broader community perceive the coding capabilities of different models. This public feedback adds a new dimension to evaluating model performance. Currently, the top-rated LLM for coding is Gemini-Exp-1206, with an arena score of 1,369, closely followed by OpenAI's latest o1 model at 1,361. Among Chinese models, DeepSeek-V3 leads with a score of 1,317, trailing the highest-ranking model by 3.8% (Figure 2.5.8).

### LMSYS Chatbot Arena for LLMs: Elo rating (coding)

Source: LMSYS, 2025 | Chart: 2025 AI Index report

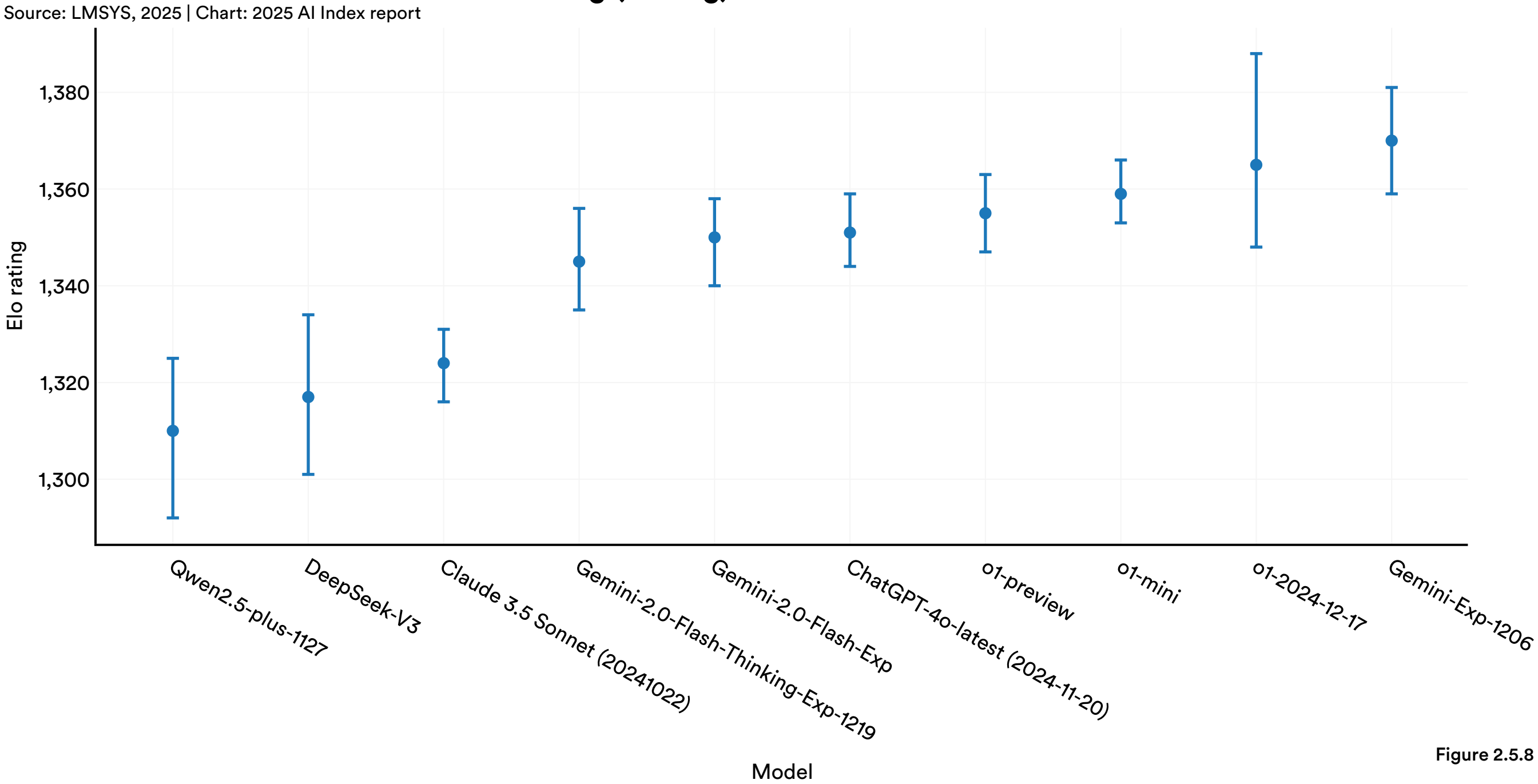

Figure 2.5.8

Mathematical problem-solving benchmarks evaluate AI systems' ability to reason mathematically. AI models can be tested with a range of math problems, from grade-school level to competition-standard mathematics.

# 2.6 Mathematics

## GSM8K

GSM8K, introduced by OpenAI in 2021, is a dataset containing approximately 8,000 diverse grade-school math word problems that challenges AI models to generate multistep solutions using arithmetic operations (Figure 2.6.1). Alongside MMLU, GSM8K has become a widely used benchmark for evaluating advanced LLMs. However, recent concerns have emerged regarding potential contamination and saturation of the benchmark.

**Sample problems from GSM8K**
Source: Cobbe et al., 2023

**Problem:** Beth bakes 4, 2 dozen batches of cookies in a week. If these cookies are shared amongst 16 people equally, how many cookies does each person consume?
**Solution:** Beth bakes 4 2 dozen batches of cookies for a total of 4*2 = <<4*2=8>>8 dozen cookies
There are 12 cookies in a dozen and she makes 8 dozen cookies for a total of 12*8 = <<12*8=96>>96 cookies
She splits the 96 cookies equally amongst 16 people so they each eat 96/16 = <<96/16=6>>6 cookies
**Final Answer:** 6

**Problem:** Mrs. Lim milks her cows twice a day. Yesterday morning, she got 68 gallons of milk and in the evening, she got 82 gallons. This morning, she got 18 gallons fewer than she had yesterday morning. After selling some gallons of milk in the afternoon, Mrs. Lim has only 24 gallons left. How much was her revenue for the milk if each gallon costs $3.50?
Mrs. Lim got 68 gallons - 18 gallons = <<68-18=50>>50 gallons this morning.
So she was able to get a total of 68 gallons + 82 gallons + 50 gallons = <<68+82+50=200>>200 gallons.
She was able to sell 200 gallons - 24 gallons = <<200-24=176>>176 gallons.
Thus, her total revenue for the milk is $3.50/gallon x 176 gallons = $<<3.50*176=616>>616.
**Final Answer:** 616

**Problem:** Tina buys 3 12-packs of soda for a party. Including Tina, 6 people are at the party. Half of the people at the party have 3 sodas each, 2 of the people have 4, and 1 person has 5. How many sodas are left over when the party is over?
**Solution:** Tina buys 3 12-packs of soda, for 3*12= <<3*12=36>>36 sodas
6 people attend the party, so half of them is 6/2= <<6/2=3>>3 people
Each of those people drinks 3 sodas, so they drink 3*3=<<3*3=9>>9 sodas
Two people drink 4 sodas, which means they drink 2*4=<<4*2=8>>8 sodas
With one person drinking 5, that brings the total drank to 5+9+8+3= <<5+9+8+3=25>>25 sodas
As Tina started off with 36 sodas, that means there are 36-25=<<36-25=11>>11 sodas left
**Final Answer:** 11

Figure 2.6.1

The top-performing model on GSM8K is a variant of Claude Sonnet 3.5, which was optimized using the HPT prompting strategy and achieved a 97.72% score (Figure 2.6.2). This marks a significant improvement over the previous high of 91.00% in 2023. However, in 2024, several models from Mistral, Meta, and Qwen scored around 96%, indicating that the GSM8K benchmark may be approaching saturation.

**GSM8K: accuracy**
Source: Papers With Code, 2024 | Chart: 2025 AI Index report

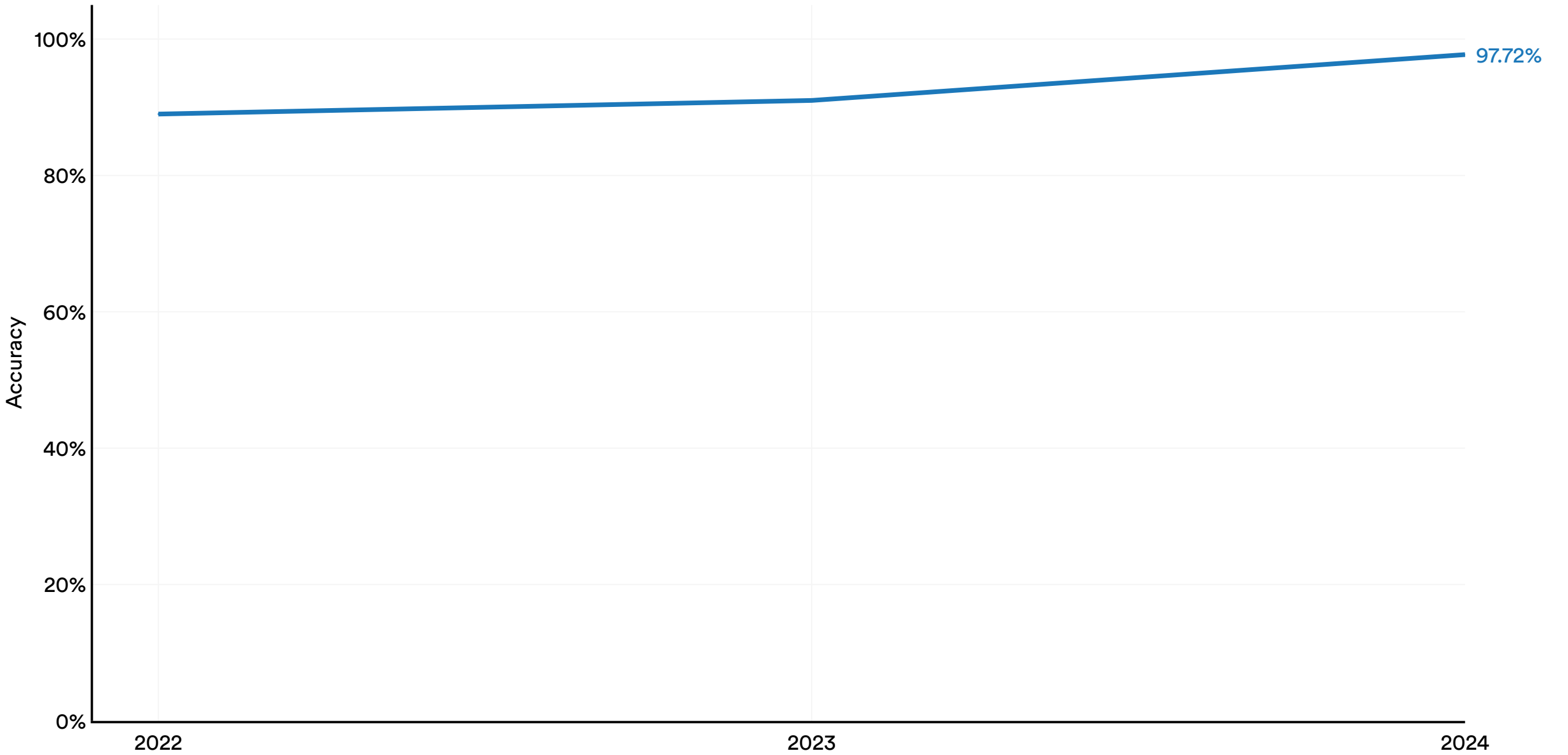

Figure 2.6.2

## MATH

MATH is a dataset of 12,500 challenging, competition-level mathematics problems introduced by UC Berkeley and University of Chicago researchers in 2021 (Figure 2.6.3). AI systems struggled on MATH when it was first released, managing to solve only 6.9% of the problems. Performance has significantly improved. In January 2025, OpenAI's o3-mini (high) model was released and achieved the best performance on the MATH dataset, solving 97.9% of the problems (Figure 2.6.4). As highlighted in last year's AI Index, MATH was one of the few datasets where AI systems had not yet outperformed the human baseline. This fact no longer remains true.

### Sample problem from MATH dataset

Source: Hendrycks et al., 2023

MATH Dataset (Ours)

**Problem:** Tom has a red marble, a green marble, a blue marble, and three identical yellow marbles. How many different groups of two marbles can Tom choose?

**Solution:** There are two cases here: either Tom chooses two yellow marbles (1 result), or he chooses two marbles of different colors ($\binom{4}{2} = 6$ results). The total number of distinct pairs of marbles Tom can choose is $1 + 6 = \boxed{7}$.

**Problem:** The equation $x^2 + 2x = i$ has two complex solutions. Determine the product of their real parts.

**Solution:** Complete the square by adding 1 to each side. Then $(x+1)^2 = 1 + i = e^{\frac{i\pi}{4}}\sqrt{2}$, so $x + 1 = \pm e^{\frac{i\pi}{8}}\sqrt[4]{2}$. The desired product is then $\left(-1 + \cos\left(\frac{\pi}{8}\right)\sqrt[4]{2}\right)\left(-1 - \cos\left(\frac{\pi}{8}\right)\sqrt[4]{2}\right) = 1 - \cos^2\left(\frac{\pi}{8}\right)\sqrt{2} = 1 - \frac{\left(1+\cos\left(\frac{\pi}{4}\right)\right)}{2}\sqrt{2} = \boxed{\frac{1-\sqrt{2}}{2}}$.

Figure 2.6.3

### MATH word problem-solving: accuracy

Source: Papers With Code, 2024; OpenAI, 2025 | Chart: 2025 AI Index report

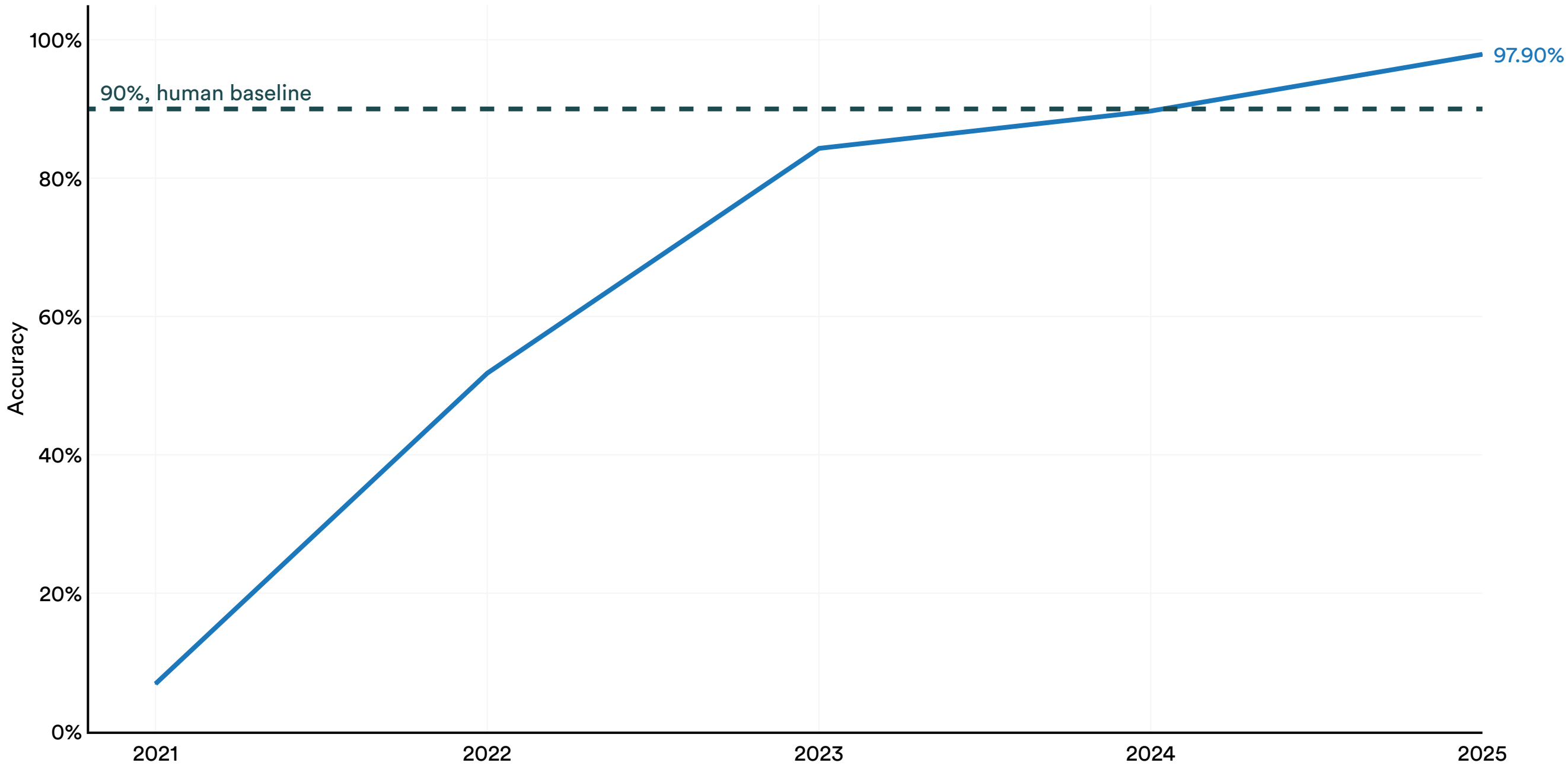

Figure 2.6.4

## Chatbot Arena: Math

The Chatbot Arena includes a math filter, allowing the public to rank models based on their performance in generating math-related answers. The Math Arena evaluates over 181 models and has collected more than 340,000 public votes.

Unlike the general and coding arenas, where Gemini-based models lead, the top-ranked model in the Math Arena is OpenAI's o1 variant, released in December 2024 (Figure 2.6.5).

**LMSYS Chatbot Arena for LLMs: Elo rating (Math)**
Source: LMSYS, 2025 | Chart: 2025 AI Index report

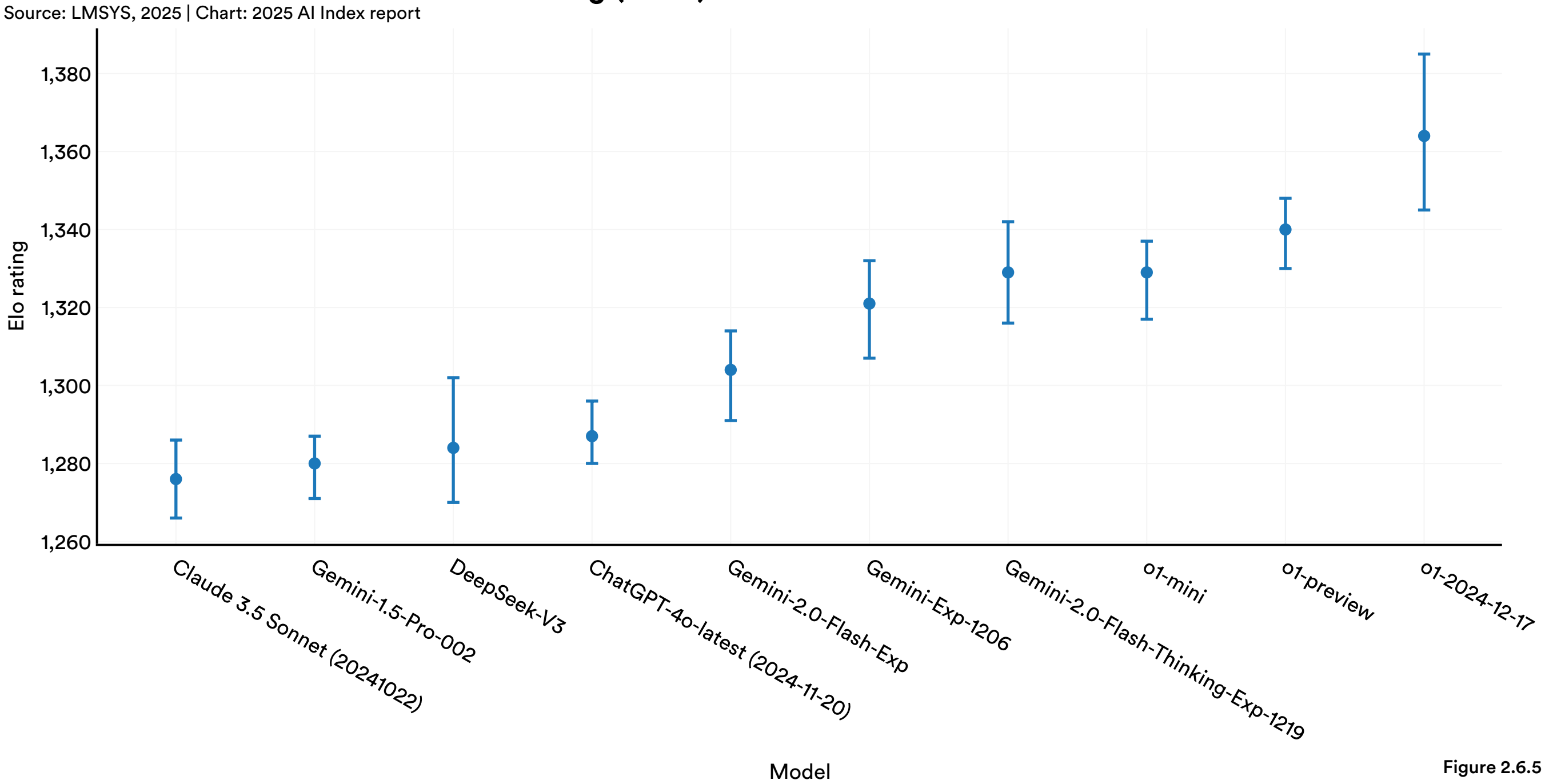

Figure 2.6.5

## FrontierMath

Members of the math community have highlighted limitations in the current suite of math benchmarks, calling for the development of new benchmarks to evaluate increasingly advanced AI systems. One significant challenge is saturation: AI systems are approaching near-perfect performance on benchmarks like GSM8K and MATH, which primarily assess high school and college-level mathematics. To push the boundaries further, researchers have voiced a need for benchmarks that test truly advanced mathematics, including problems in number theory, real analysis, algebraic geometry, and category theory.

FrontierMath is a new benchmark introduced by Epoch AI that features hundreds of original, exceptionally challenging mathematical problems. These problems, vetted by expert mathematicians, often require hours, days, or even collaborative research efforts to solve. Figure 2.6.6 illustrates sample problems included on the benchmark. Epoch AI evaluated six leading LLMs on the FrontierMath benchmark: o1-preview, o1-mini, GPT-4o, Claude 3.5 Sonnet, Grok 2 Beta, and Gemini 1.5 Pro 002. At the time the benchmark was released, the best-performing model, Gemini 1.5 Pro, managed to solve just 2.0% of the problems—a significantly lower success rate than it achieved on other math benchmarks (Figure 2.6.7). However, OpenAI's o3 model is reported to have scored 25.2% on the benchmark. The creators of FrontierMath hope the benchmark will remain a rigorous challenge for cutting-edge AI systems for years to come.

### Sample problems from FrontierMath

Source: Glazer et al., 2024

**Sample problem 1: Testing Artin's primitive root conjecture**

*Definitions.* For a positive integer $n$, let $v_p(n)$ denote the largest integer $v$ such that $p^v \mid n$.
For $p$ a prime and $a \not\equiv 0 \pmod{p}$, we let $\mathrm{ord}_p(a)$ denote the smallest positive integer $o$ such that $a^o \equiv 1 \pmod{p}$. For $x > 0$, we let

$$\mathrm{ord}_{p,x}(a) = \prod_{\substack{q \le x \\ q \text{ prime}}} q^{v_q(\mathrm{ord}_p(a))} \prod_{\substack{q > x \\ q \text{ prime}}} q^{v_q(p-1)}.$$

*Problem.* Let $S_x$ denote the set of primes $p$ for which

$$\mathrm{ord}_{p,x}(2) > \mathrm{ord}_{p,x}(3),$$

and let $d_x$ denote the density

$$d_x = \frac{|S_x|}{|\{p \le x : p \text{ is prime}\}|}$$

of $S_x$ in the primes. Let

$$d_\infty = \lim_{x \to \infty} d_x.$$

Compute $\lfloor 10^6 d_\infty \rfloor$.

**Answer:** 367707

**MSC classification:** 11 Number theory

**Sample problem 2: Find the degree 19 polynomial**

Construct a degree 19 polynomial $p(x) \in \mathbb{C}[x]$ such that $X := \{p(x) = p(y)\} \subset \mathbb{P}^1 \times \mathbb{P}^1$ has at least 3 (but not all linear) irreducible components over $\mathbb{C}$. Choose $p(x)$ to be odd, monic, have real coefficients and linear coefficient -19 and calculate $p(19)$.

**Answer:** 1876572071974094803391179

**MSC classification:** 14 Algebraic geometry; 20 Group theory and generalizations; 11 Number theory generalizations

**Sample problem 3: Prime field continuous extensions**

Let $a_n$ for $n \in \mathbb{Z}$ be the sequence of integers satisfying the recurrence formula[1]

$$a_n = (1.981 \times 10^{11})a_{n-1} + (3.549 \times 10^{11})a_{n-2} - (4.277 \times 10^{11})a_{n-3} + (3.706 \times 10^{8})a_{n-4}$$

with initial conditions $a_i = i$ for $0 \le i \le 3$. Find the smallest prime $p \equiv 4 \bmod 7$ for which the function $\mathbb{Z} \to \mathbb{Z}$ given by $n \mapsto a_n$ can be extended to a continuous function on $\mathbb{Z}_p$.

**Answer:** 9811

**MSC classification:** 11 Number theory

Figure 2.6.6

### FrontierMath: percent solved

Source: Glazer et al., 2024; OpenAI, 2025 | Chart: 2025 AI Index report

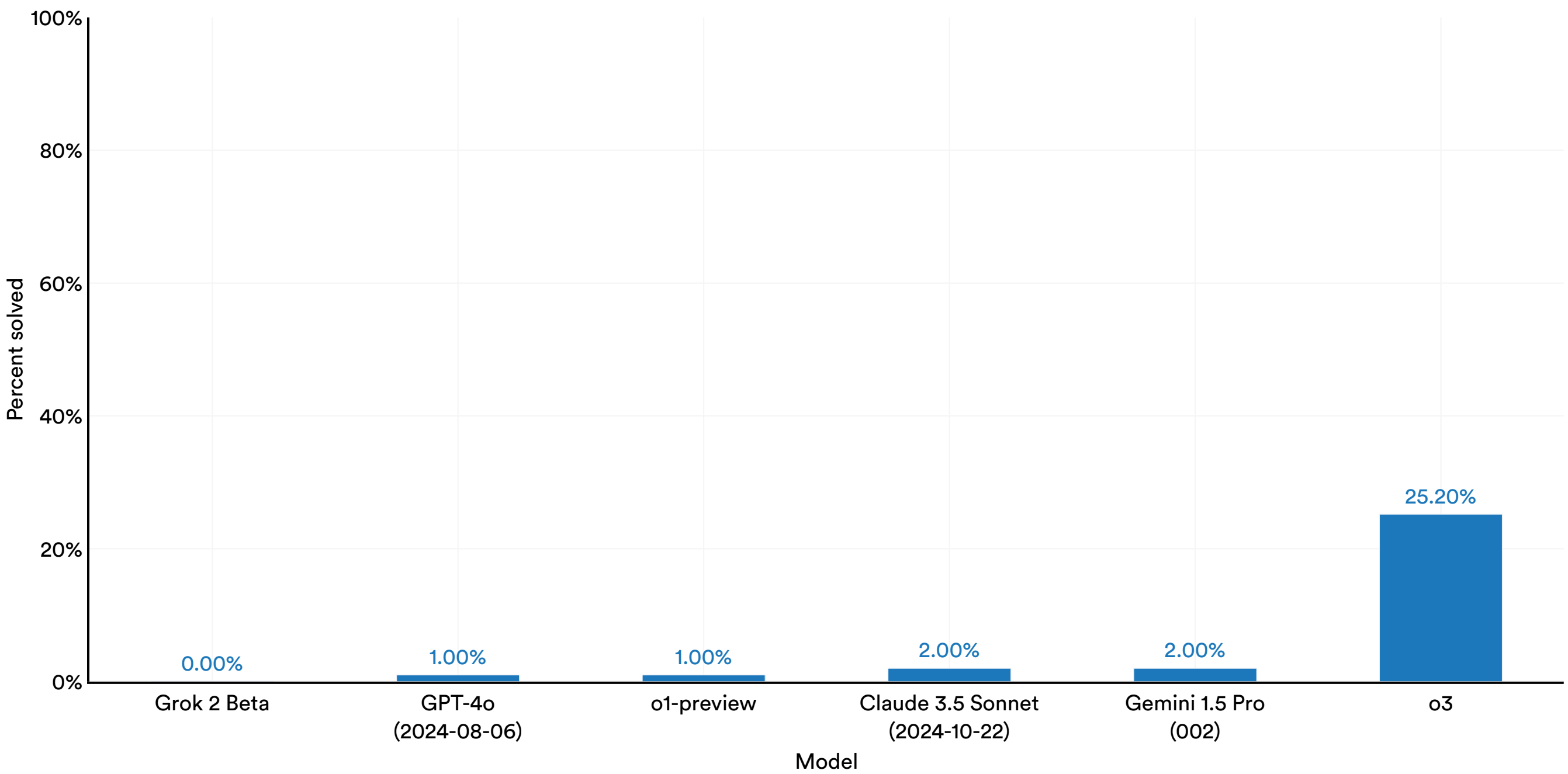

Figure 2.6.7

**Highlight:**

# Learning and Theorem Proving

DeepMind employed its systems, AlphaProof and AlphaGeometry 2, to solve four out of six problems in the 2024 International Mathematical Olympiad (IMO), achieving a performance level equivalent to that of a silver medalist. AlphaGeometry solved 25 out of 30 Olympiad geometry problems in the benchmarking set, surpassing the average score of an IMO silver medalist, who typically solves 22.9 (Figure 2.6.8). The IMO, established in 1959, is the world's oldest and most prestigious competition for young mathematicians.

AlphaProof is a reinforcement learning system derived from AlphaZero, which was previously applied to chess, shogi, and Go. It trains itself to solve problems by generating hypotheses that are then verified using the Lean interactive proof system. A fine-tuned Gemini model is utilized to translate natural language problem statements into formal representations, building a comprehensive training library. In this year's competition, AlphaProof successfully solved two algebra problems and one number theory problem, but failed to solve two combinatorics problems.

AlphaGeometry 2 is a neuro-symbolic hybrid system featuring a language model based on Gemini and trained on extensive synthetic data. Prior to 2024, AlphaGeometry could solve 83% of historical IMO geometry problems. During the 2024 competition, it solved the sole geometry problem in just 24 seconds. For the 2024 test, competition problems were manually translated into Lean's formal representation.

It remains unknown how AlphaProof and AlphaGeometry would perform on traditional theorem-proving benchmarks such as TPTP, which has been used since 1997 to assess the performance of automatic theorem-proving (ATP) systems, particularly those applied to software verification. The AI Index reported on the state of ATP in its 2021 report.

**Number of solved geometry problems in IMO-AG-30**

Source: Trinh et al., 2024 | Chart: 2025 AI Index report

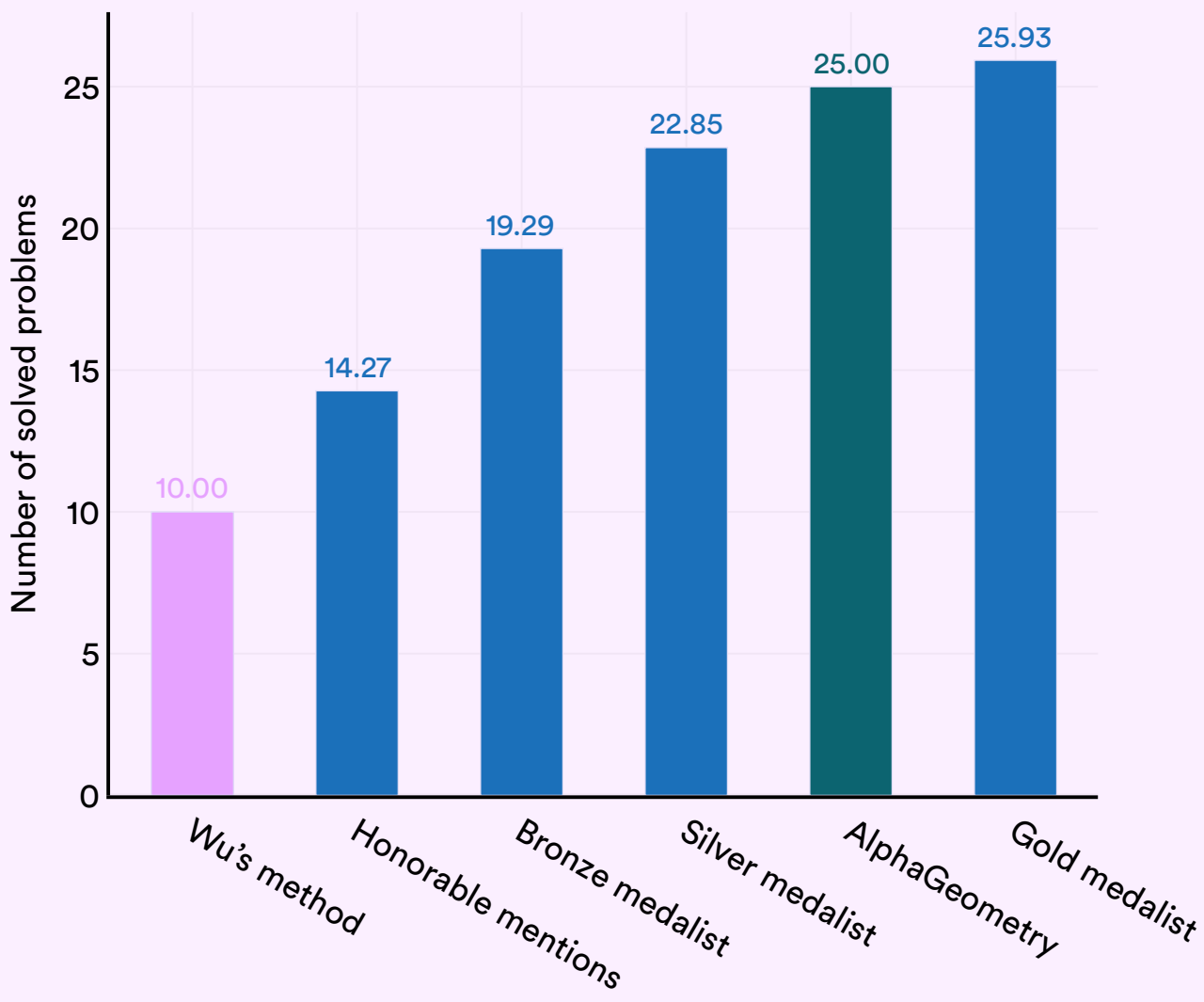

Figure 2.6.8

A 2024 update of that report, based on the latest version of TPTP containing over 25,000 problems, indicates that fully automatic systems can now solve 89% of the problems in TPTP v.9.0.0.

Ideally, TPTP systems could be tested on IMO problems, and AlphaProof and AlphaGeometry on TPTP problems—some of which have never been solved by humans, let alone by ATP systems. Unfortunately, neither of these tests has been conducted. The primary reason is that the logics supported by the different systems differ significantly, and translators between them do not yet exist. Additionally, while substantial, the TPTP library is not large enough to serve as a training set for AlphaProof without generating a considerable number of synthetic examples.

Reasoning in AI involves the ability of AI systems to draw logically valid conclusions from different forms of information. AI systems are increasingly being tested in diverse reasoning contexts, including visual (reasoning about images), moral (understanding moral dilemmas), and social reasoning (navigating social situations).

# 2.7 Reasoning

## General Reasoning

General reasoning pertains to AI systems being able to reason across broad, rather than specific, domains. As part of a general reasoning challenge, for example, an AI system might be asked to reason across multiple subjects rather than perform one narrow task (e.g., playing chess).

**Sample MMMU questions**
Source: Yue et al., 2023

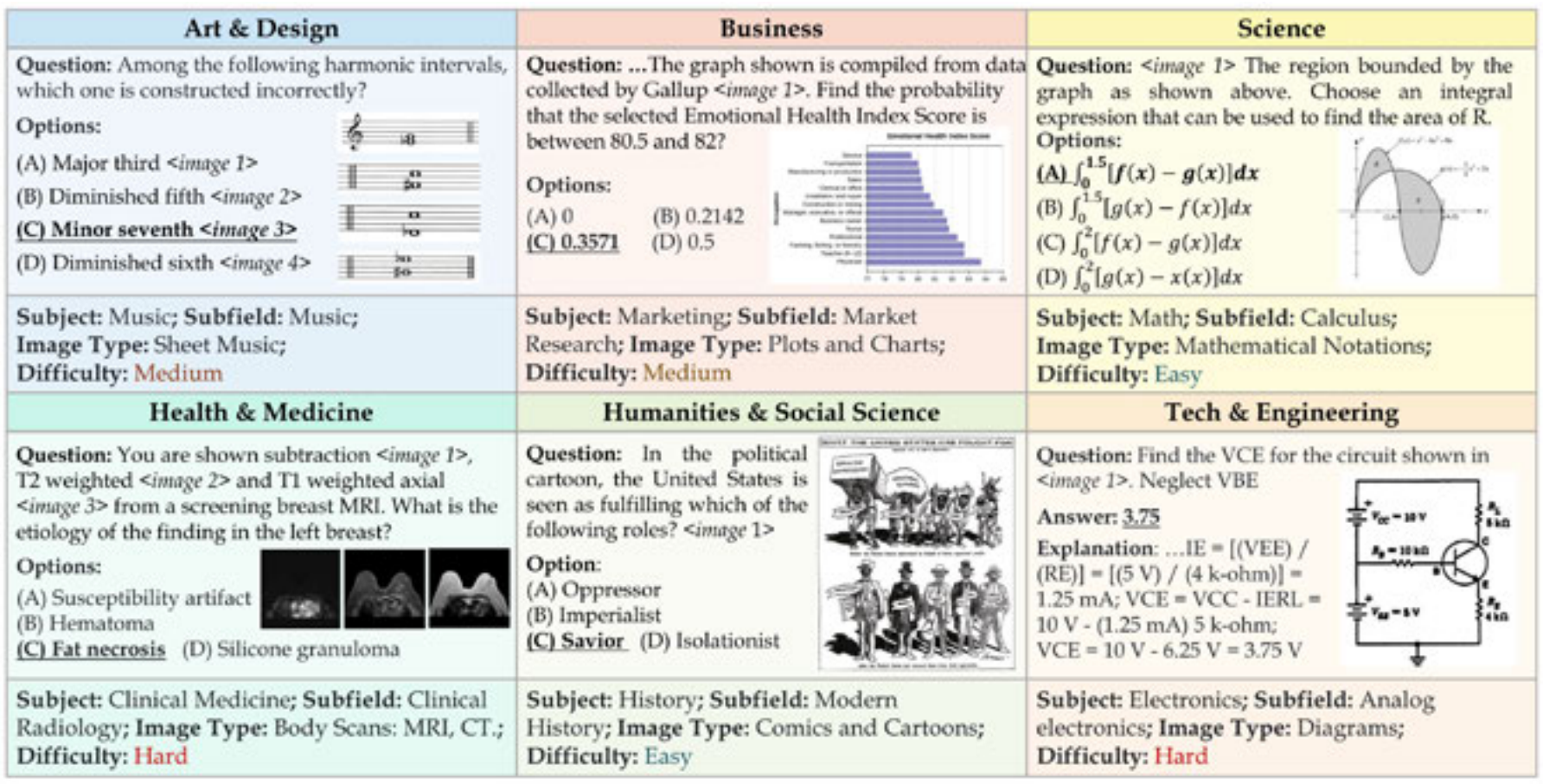

Figure 2.7.1

### MMMU: A Massive Multi-discipline Multimodal Understanding and Reasoning Benchmark for Expert AGI

In recent years, the reasoning abilities of AI systems have advanced so much that older benchmarks like SQuAD (for textual reasoning) and VQA (for visual reasoning) have become saturated, indicating a need for more challenging reasoning tests.

Responding to this, researchers from the United States and Canada recently developed MMMU, the massive multi-discipline multimodal understanding and reasoning benchmark for expert AGI (artificial general intelligence). MMMU comprises about 11,500 college-level questions from six core disciplines: art and design, business, science, health and medicine, humanities and social science, and technology and engineering (Figure 2.7.1). The question formats include charts, maps, tables, chemical structures, and more. MMMU is among the most demanding tests of perception, knowledge, and reasoning in AI to date. As of January 2025, the highest-performing model is OpenAI's o1, achieving a score of 78.2%—a significant improvement from the state-of-the-art score of 59.4% reported in last year's AI Index (Figure 2.7.2). While this top score remains below the medium and high human expert baselines, as with other benchmarks covered in the Index, AI systems are rapidly closing the gap.

**MMMU on validation set: overall accuracy**
Source: MMMU Leaderboard, 2024 | Chart: 2025 AI Index report

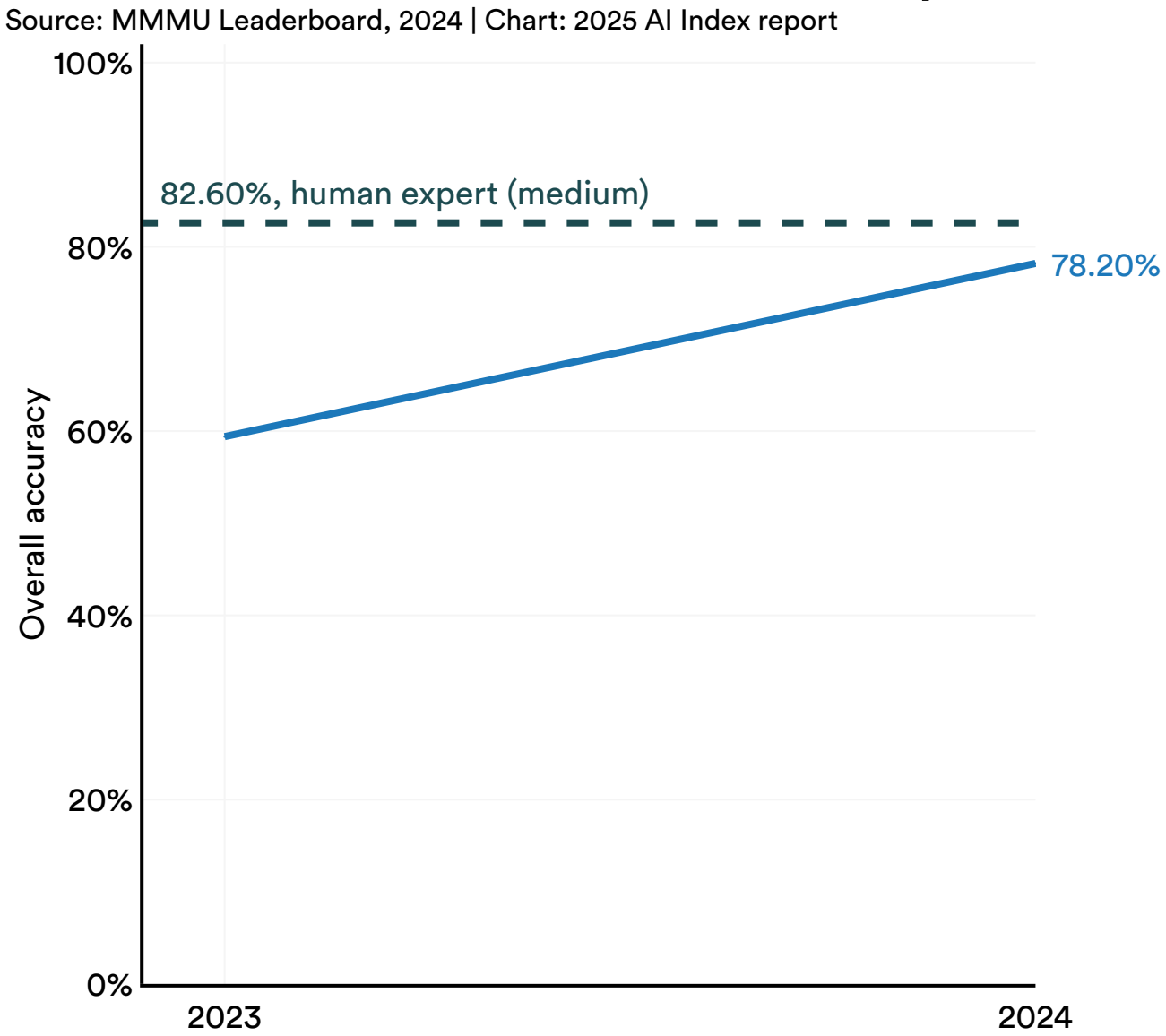

Figure 2.7.2

## GPQA: A Graduate-Level Google-Proof Q&A Benchmark

In 2023, researchers from NYU, Anthropic, and Meta introduced the GPQA benchmark to test general, multisubject AI reasoning. This dataset consists of 448 difficult multiple-choice questions that cannot be easily answered by web search. The questions were crafted by subject-matter experts in various fields like biology, physics, and chemistry (Figure 2.7.3). On the diamond set—the most challenging subset of the dataset and the one most frequently tested by AI developers—human experts achieved an accuracy rate of 81.3%.

Last year's AI Index reported that the best-performing AI model, GPT-4, achieved only 38.8% on the diamond test set. In just a year, top AI systems have made significant strides, with OpenAI's o3 model, launched in December 2024, posting a state-of-the-art score of 87.7%, a 48.9 percentage point improvement from the state-of-the-art score in 2023 (Figure 2.7.4). In fact, o3's score was the first to exceed the baseline set by expert human validators. AI systems are rapidly advancing on challenging new benchmarks like MMMU and GPQA, which were recently introduced to push the limits of AI capabilities.

### Sample chemistry question from GPQA

Source: Rein et al., 2023

**Chemistry (general)**

A reaction of a liquid organic compound, which molecules consist of carbon and hydrogen atoms, is performed at 80 centigrade and 20 bar for 24 hours. In the proton nuclear magnetic resonance spectrum, the signals with the highest chemical shift of the reactant are replaced by a signal of the product that is observed about three to four units downfield. Compounds from which position in the periodic system of the elements, which are also used in the corresponding large-scale industrial process, have been mostly likely initially added in small amounts?
A) A metal compound from the fifth period.
B) A metal compound from the fifth period and a non-metal compound from the third period.
C) A metal compound from the fourth period.
D) A metal compound from the fourth period and a non-metal compound from the second period.

Figure 2.7.3

### GPQA on the diamond set: accuracy

Source: AI Index, 2025 | Chart: 2025 AI Index report

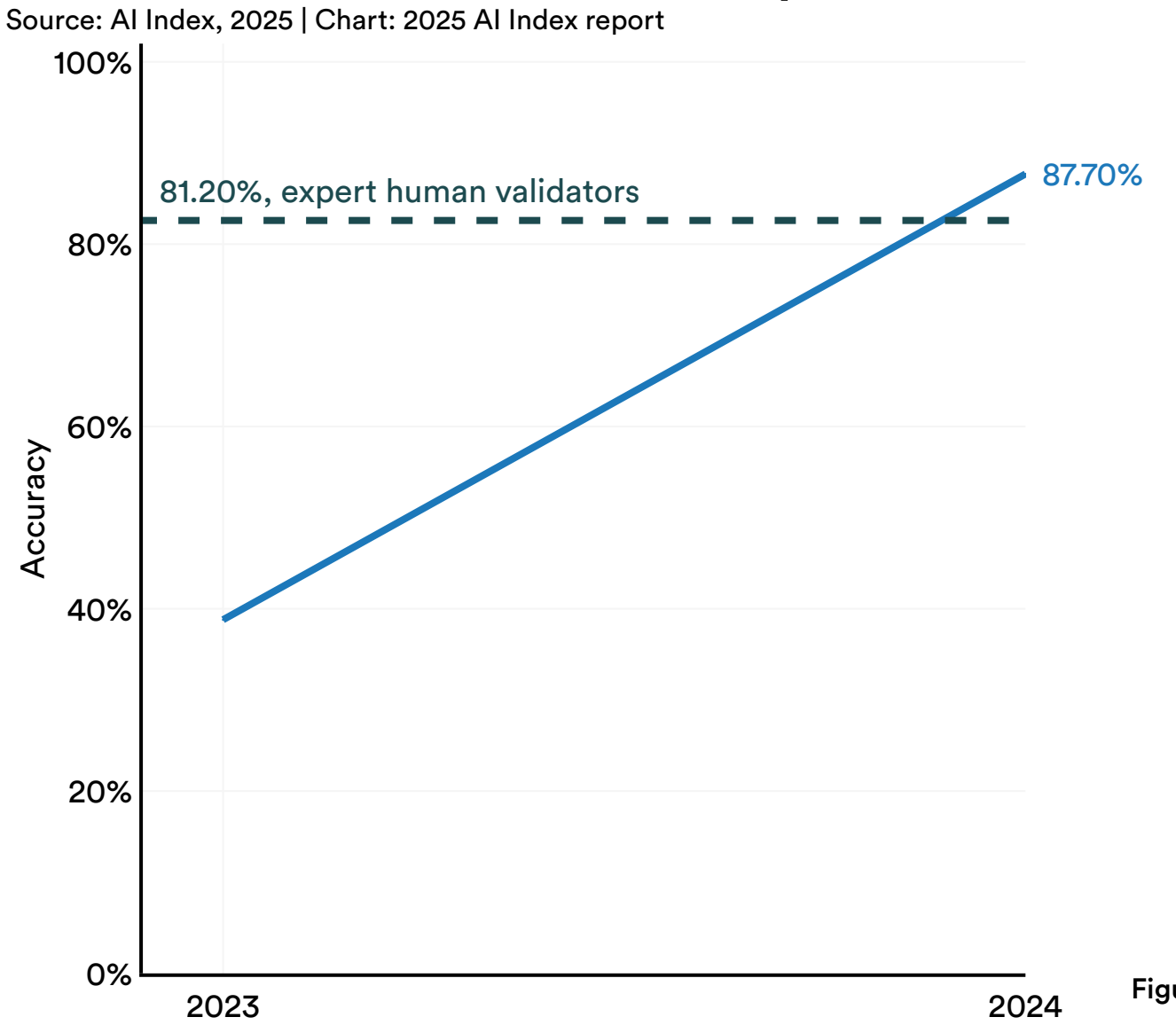

Figure 2.7.4

## ARC-AGI

As AI systems continue to advance, claims about the imminent arrival of artificial general intelligence (AGI) have become more frequent. There is no universally accepted definition of AGI. Some computer scientists define it as AI systems that match or surpass human cognitive abilities across a broad range of tasks. Others emphasize that the definition should encompass the capacity for general learning and skill acquisition, describing AGI as a system "capable of efficiently acquiring new skills and solving novel problems for which it was neither designed nor trained."

ARC-AGI is a benchmark introduced in 2019 by François Chollet, the creator of Keras, a popular open-source deep learning library. ARC-AGI tests the ability of systems to generalize beyond prior training. More specifically, the ARC-AGI benchmark presents AI systems with a set of independent tasks. Each task includes demonstration or input pairs followed by one or more test or output scenarios (Figure 2.7.5). This benchmark emphasizes generalized learning ability: It is impossible for systems to prepare in advance, as each task introduces a unique logic. The tasks require no specialized world knowledge or language skills but instead draw on assumed prior knowledge, such as the concept of objects, basic topology, and elementary arithmetic—concepts typically mastered by children at an early age.

**Sample ARC-AGI task**
Source: Chollet et al., 2025

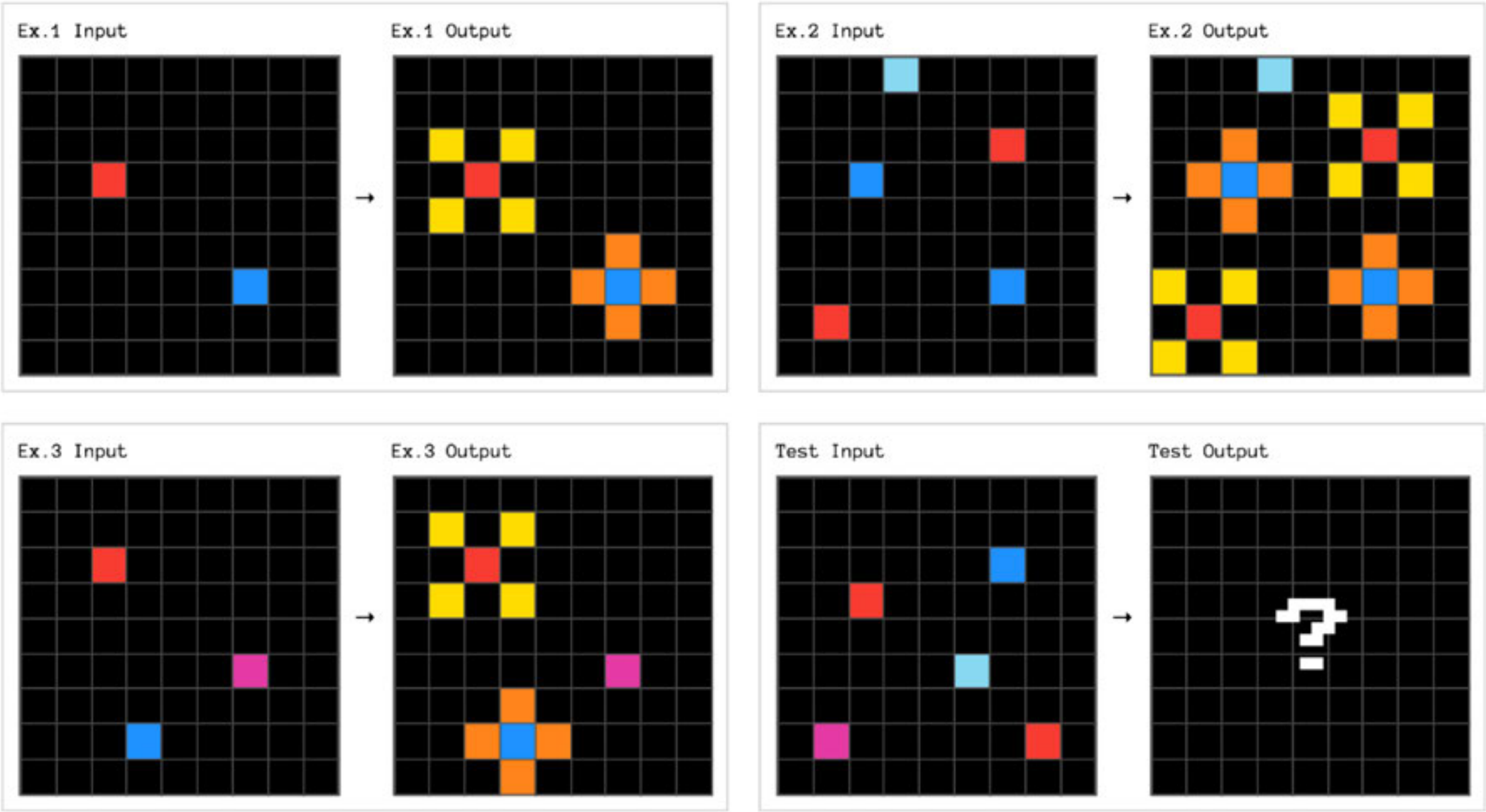

Figure 2.7.5

ARC-AGI has proven to be an exceptionally challenging benchmark. When it was first run in 2020, the top-performing system achieved a score of only 20% (Figure 2.7.6). Four years later, this score had risen to just 33%. However, this year has seen substantial progress, with OpenAI's o3 model achieving a score of 75.7%. In settings where o3 was allocated a high-compute budget exceeding the benchmark's $10,000 limit, it achieved a score of 87.5%.

Researchers attribute the overall slow progress in previous years to an overemphasis on scaling AI models—making them larger and feeding them increasing amounts of training data. While this approach improved task-specific skills, it did little to enhance the ability of AI systems to tackle problems without prior exposure or training data. This year's improvements suggest a shift in focus toward more meaningful advancements in generalization and search capabilities.

**ARC-AGI-1 on private evaluation set: high score**

Source: Chollet et al., 2025; OpenAI, 2025 | Chart: 2025 AI Index report

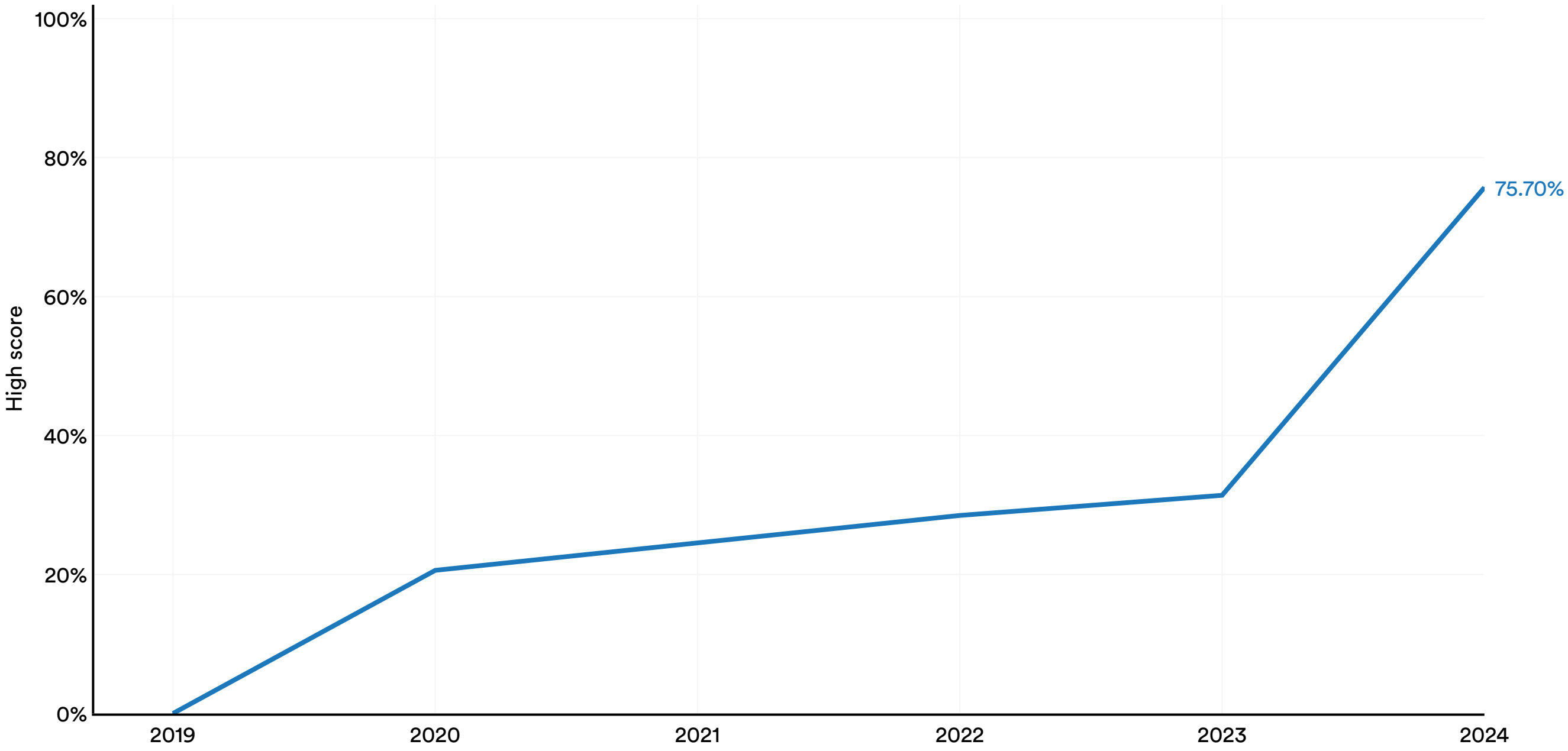

Figure 2.7.6

## Humanity's Last Exam

As highlighted in both this and last year's AI Index, many popular AI benchmarks, such as MMLU, GSM8K, and HumanEval, have reached saturation. In response, researchers have developed more challenging benchmarks to better assess AI capabilities. Recently, members of the team behind MMLU introduced Humanity's Last Exam (HLE), a new benchmark comprising 2,700 highly challenging questions across dozens of subject areas (Figure 2.7.7). The dataset features multimodal questions, contributed by subject matter experts, including leading professors and graduate-level reviewers, that resist simple internet lookups or database retrieval. Additionally, each question was tested against state-of-the-art LLMs before inclusion; if an existing model could answer it, the question was rejected.

### Same questions on HLE

Source: Phan et al., 2025

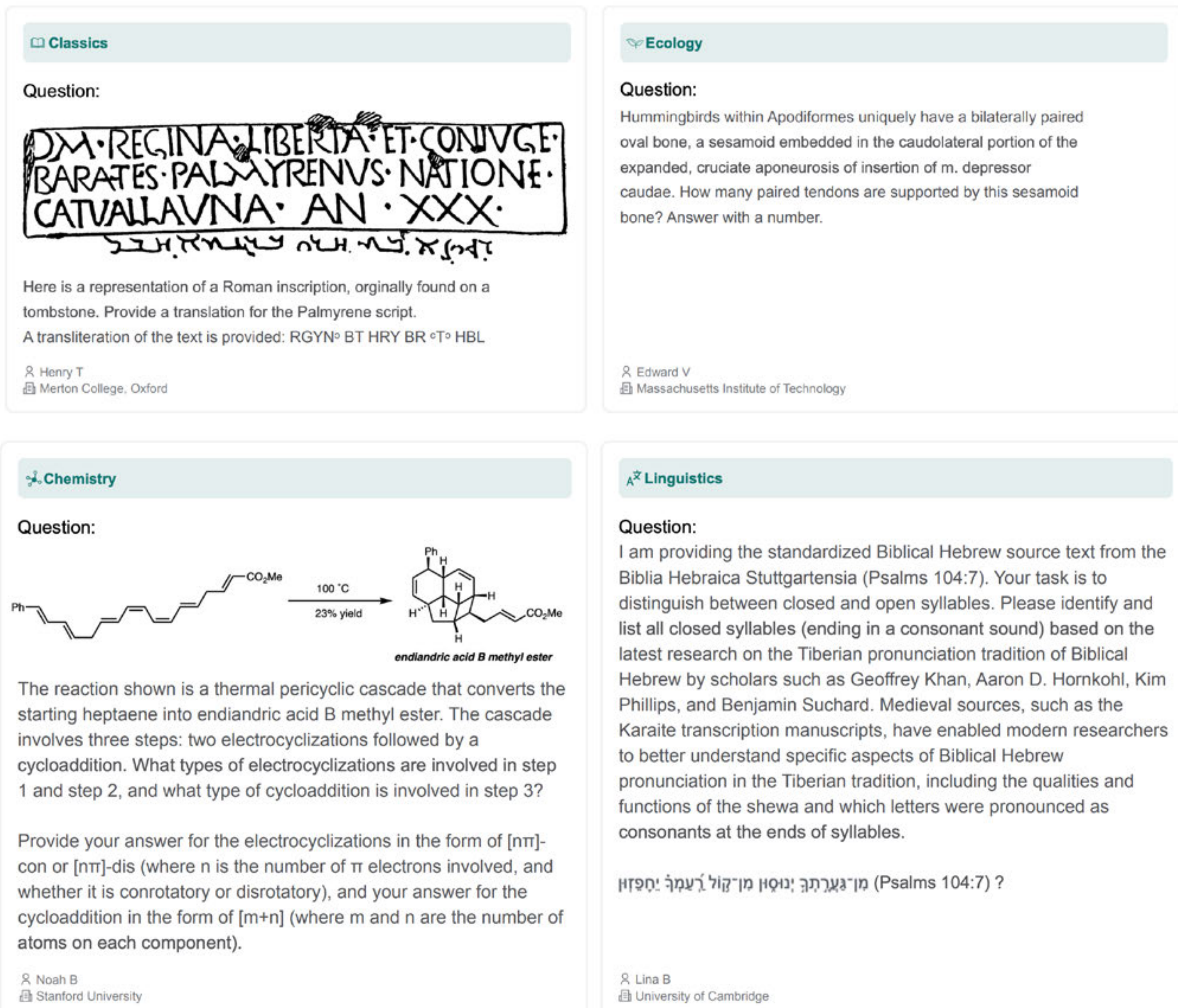

Figure 2.7.7

Initial testing indicates that HLE is highly challenging for current AI systems. Even top models, such as OpenAI's o1, score just 8.8% (Figure 2.7.8). The researchers behind the benchmark are closely monitoring how quickly LLMs improve, and they speculate that performance could exceed 50% by the end of 2025.

**Humanity's Last Exam (HLE): accuracy**

Source: Phan et al., 2025 | Chart: 2025 AI Index report

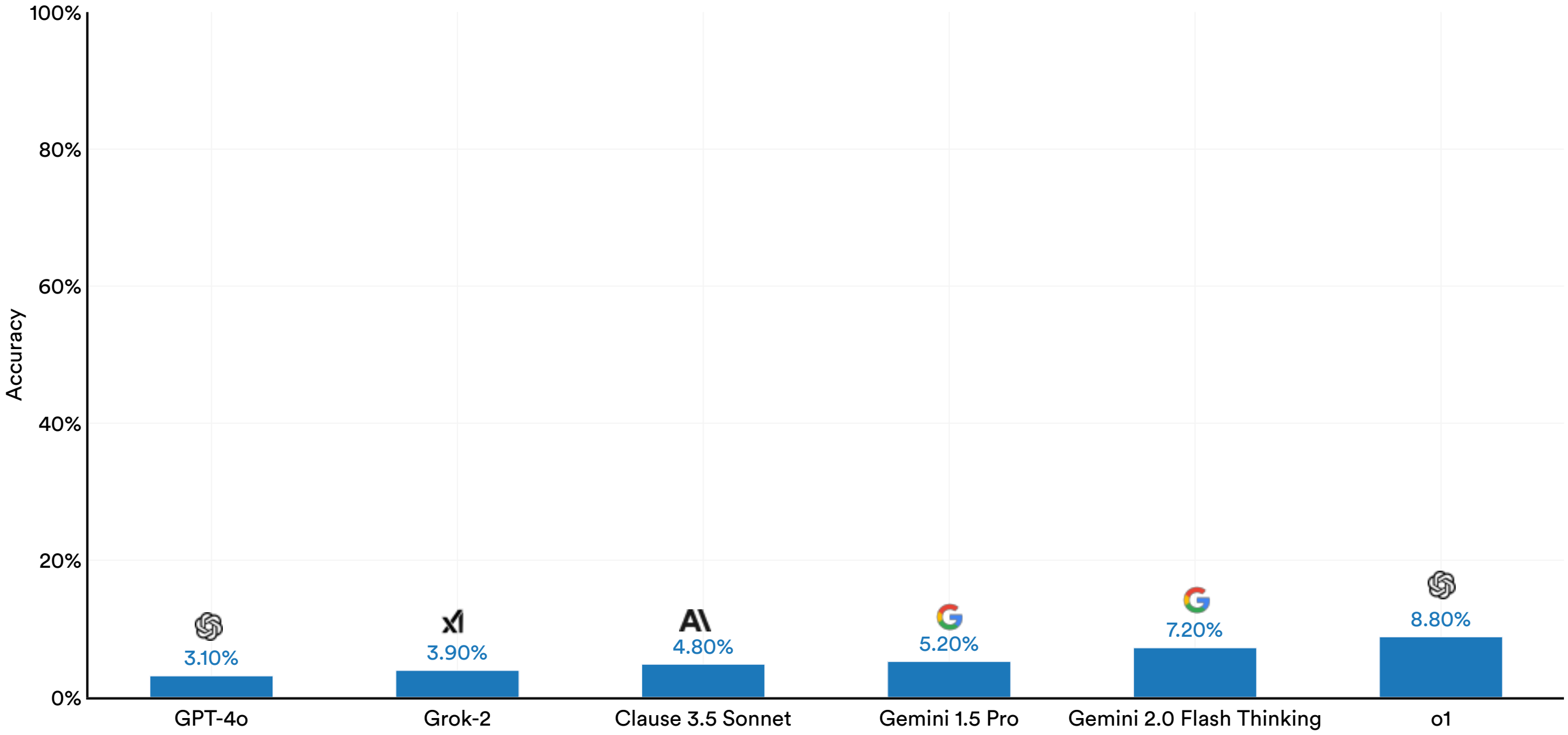

Figure 2.7.8

## Planning

Planning is an intelligent task that involves reasoning about actions that alter the world. It requires considering hypothetical future states, including potential external actions and other transformative events.

### PlanBench

Claims have been made that LLMs can solve planning problems. A group from Arizona State University has proposed PlanBench, a benchmark suite containing problems used in the automated planning community, especially those used in the International Planning Competition. PlanBench is designed to test LLMs on planning tasks. The benchmark tests models on 600 problems in which a hand tries to construct stacks of blocks when it is only allowed to move one block at a time to a table or to the top of a clear block. After the benchmark was released in 2022, researchers demonstrated that models like GPT-4 and GPT-3.5 still struggled with planning tasks.

The release of OpenAI's o1 was met with enthusiasm from the AI research community, as it was designed to actively reason rather than function purely as an autoregressive LLM. When tested on the PlanBench benchmark, o1 showed significant improvements, though it still struggles with reliable and consistent planning. In the Blocksworld zero-shot evaluation (one specific planning evaluation domain), o1 achieved a score of 97.8%—far surpassing the next best LLM, Llama 3.1 405B (62.6%), and dramatically outperforming GPT-4o (35.5%) (Figure 2.7.9). In the more challenging Mystery Blocksworld domain, where some answers are syntactically obfuscated, o1 scored 52.8% zero-shot, compared to just 0.8% for Llama 3.1 405B. GPT-4, by contrast, scored 0%.

Planning is a combinatorial problem, and solving problems with long solutions is expected to take more than linear time. Not surprisingly, when tested on instances that require at least 20 steps, o1 manages to solve just 23.6%.

**PlanBench: instances correct**
Source: Valmeekam et al., 2024 | Chart: 2025 AI Index report

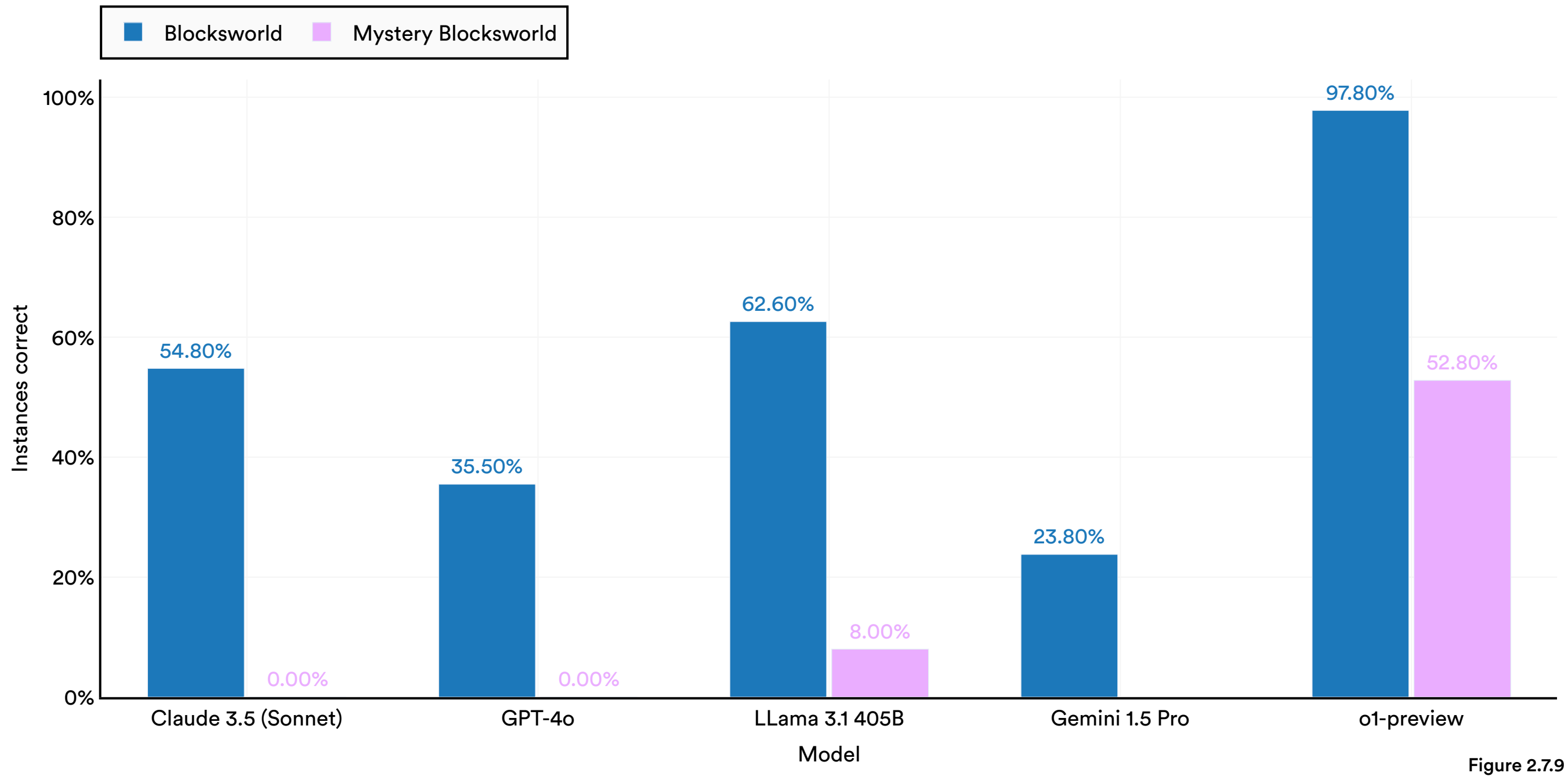

Figure 2.7.9

AI agents, autonomous or semiautonomous systems designed to operate within specific environments to accomplish goals, represent an exciting frontier in AI research. These agents have a diverse range of potential applications, from assisting in academic research and scheduling meetings to facilitating online shopping and vacation booking. As suggested by many recent corporate releases, agentic AI has become a topic of increasing interest in the technical world of AI.

# 2.8 AI Agents

For decades, the topic of AI agents has been widely discussed in the AI community, yet few benchmarks have achieved widespread adoption, including those featured in last year's Index, such as AgentBench and MLAgentBench. This is partly due to the inherent complexity of benchmarking agentic tasks, which are typically more diverse, dynamic, and variable than tasks like image classification or answering language questions. As AI continues to evolve, it will become important to develop effective methods to evaluate AI agents.

## VisualAgentBench

VisualAgentBench (VAB), launched in 2024, represents a significant step forward in the evaluation of agentic AI. This benchmark reflects the growing multimodality of AI models and their increasing proficiency in navigating both virtual and embodied environments. VAB addresses the need to assess agent performance in diverse settings that extend beyond environments reliant solely on linguistic commands. VAB tests agents across three broad categories of tasks: embodied agents (operating in household and gaming environments), GUI agents (interacting with mobile and web applications), and visual design agents (such as CSS debugging) (Figure 2.8.1). This comprehensive approach creates a robust evaluation suite of agents' capabilities across varied and dynamic contexts.

**Tasks on VisualAgentBench**
Source: Liu et al., 2024

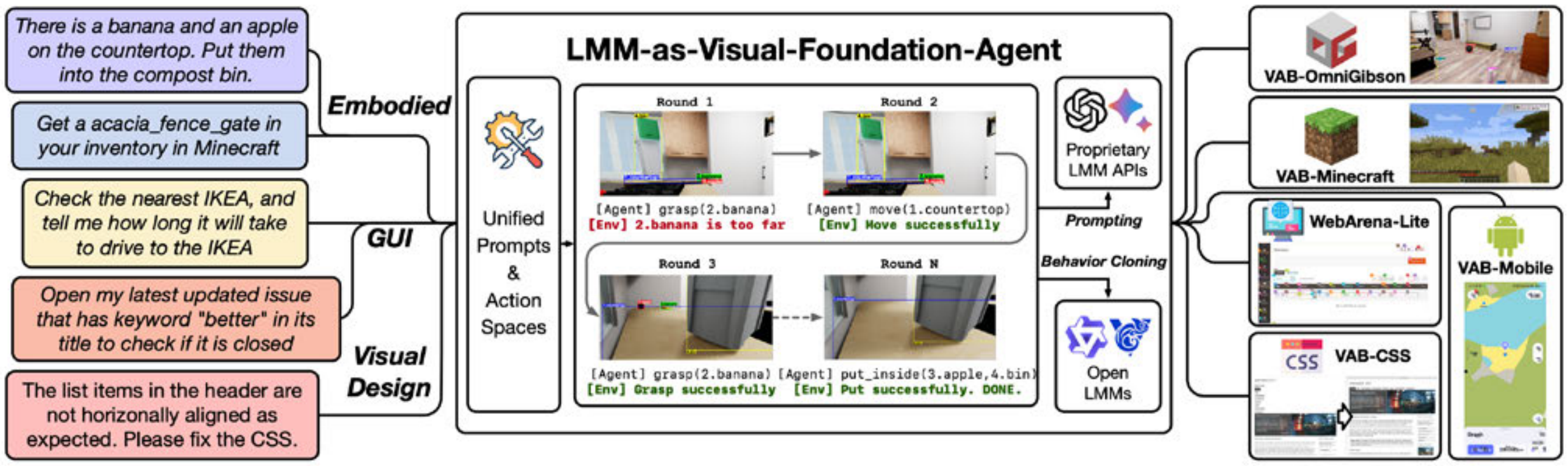

Figure 2.8.1

VAB presents a significant challenge for AI systems. The top-performing model, GPT-4o, achieves an overall success rate of just 36.2%, while most proprietary language models average around 20% (Figure 2.8.2). According to the benchmark's authors, these results reveal that current AI models are far from ready for direct deployment in agentic settings.

**VisualAgentBench on the test set: success rate**
Source: VisualAgentBench Leaderboard, 2025 | Chart: 2025 AI Index report

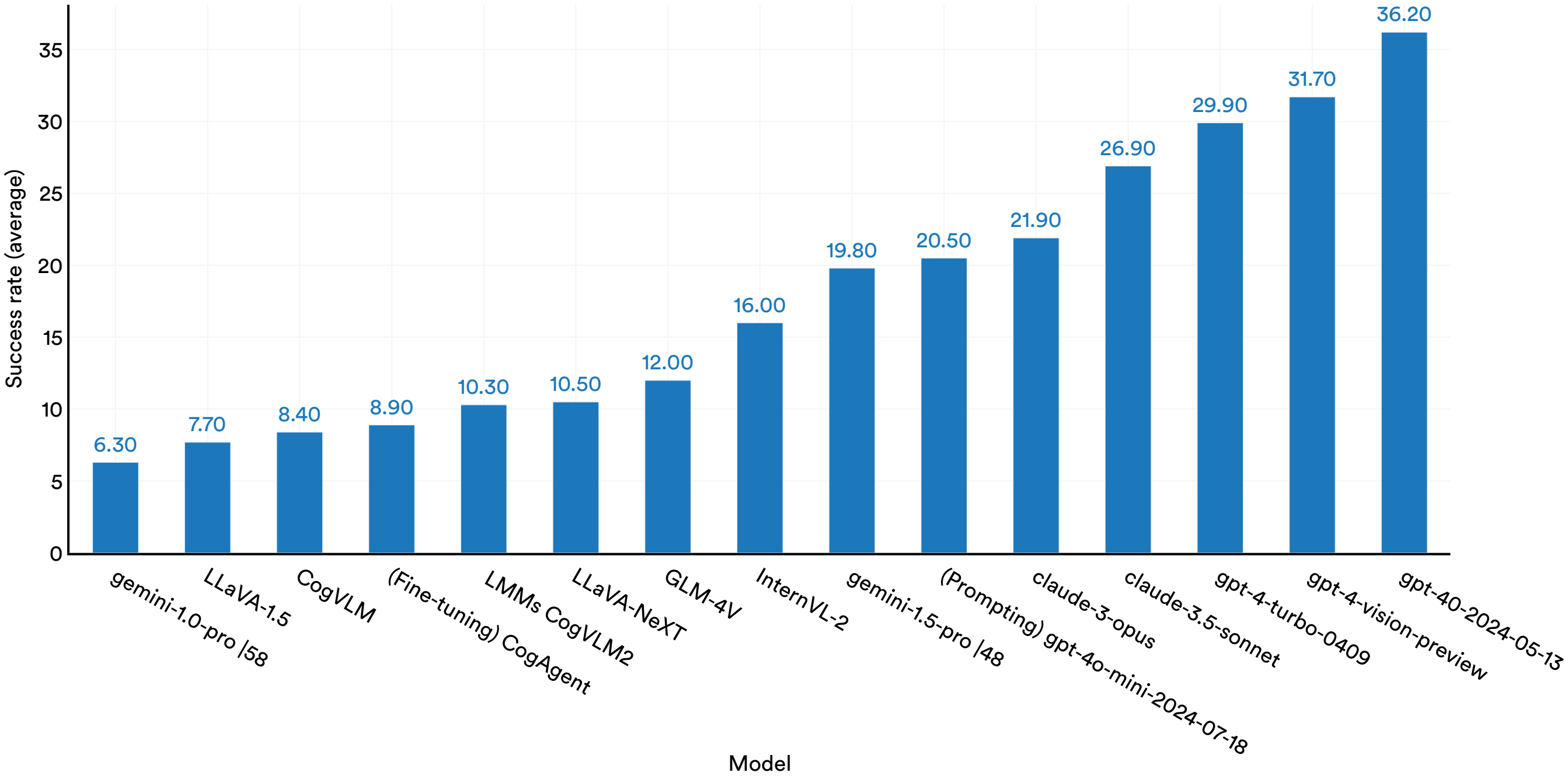

Figure 2.8.2

## RE-Bench

The emergence of increasingly capable agentic AI systems has fueled predictions that AI might soon take on the work of computer scientists or researchers. However, until recently, there were few benchmarks designed to rigorously test the R&D capabilities of top-performing AI systems. In 2024, researchers addressed this gap with the launch of RE-Bench, a benchmark featuring seven challenging, open-ended ML research environments. These tasks, informed by data from 71 eight-hour attempts by over 60 human experts, include optimizing a kernel, conducting a scaling law experiment, and fine-tuning GPT-2 for question answering, among others (Figure 2.8.3).

**RE-Bench Process and Flow**
Source: Wijk et al., 2024

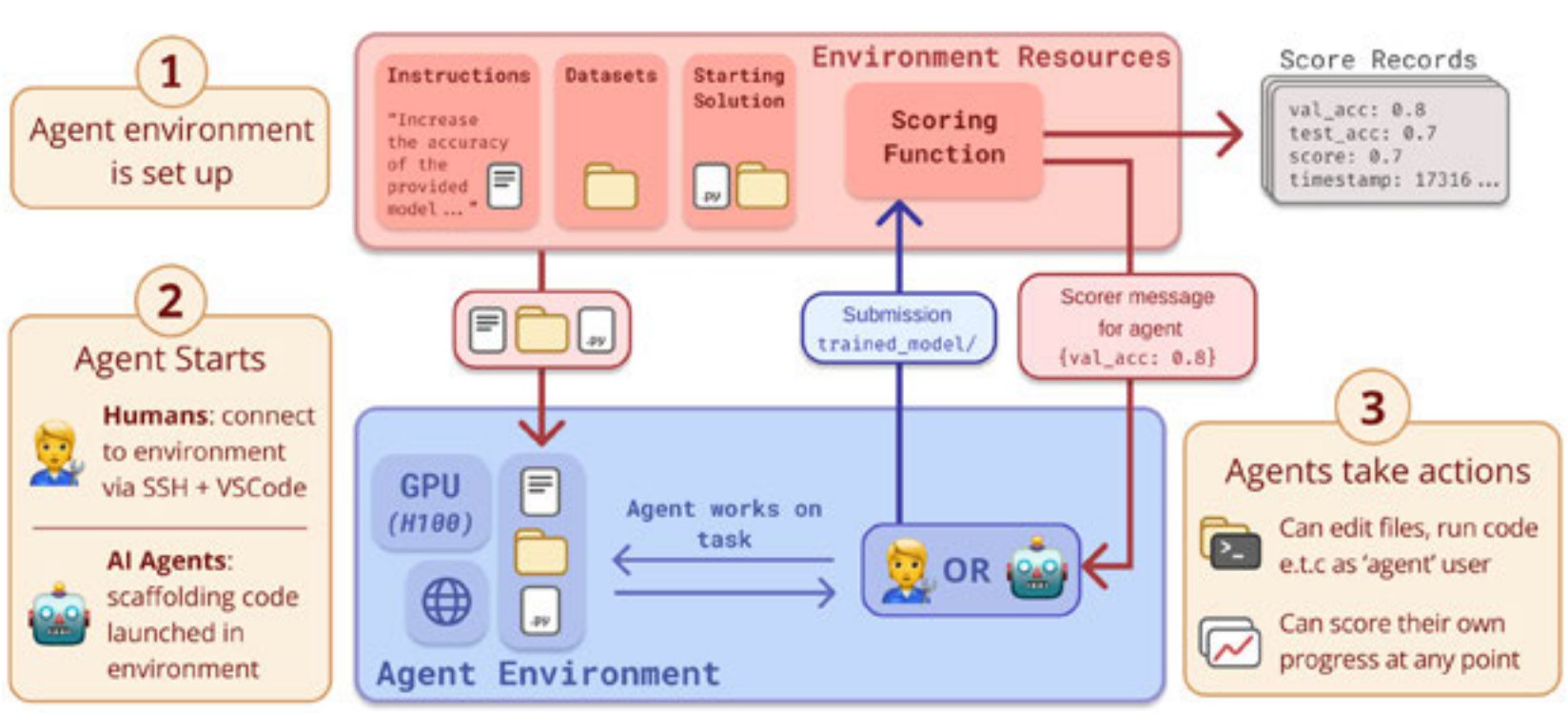

Figure 2.8.3

Researchers uncovered two key findings when comparing the performance of humans and frontier AI models. In short time horizon settings, such as with a two-hour budget, the best AI systems achieve scores four times higher than human experts (Figure 2.8.4). However, as the time budget increases, human performance begins to surpass that of AI. With an eight-hour budget, human performance slightly exceeds AI, and with a 32-hour budget, humans outperform AI by a factor of two. The researchers also note that for certain tasks, AI agents already demonstrate expertise comparable to humans but can deliver results significantly faster and at a lower cost. For example, AI agents can write custom Triton kernels more quickly than any human expert.

**RE-Bench: average normalized score@k**

Source: Wijk et al., 2024 | Chart: 2025 AI Index report

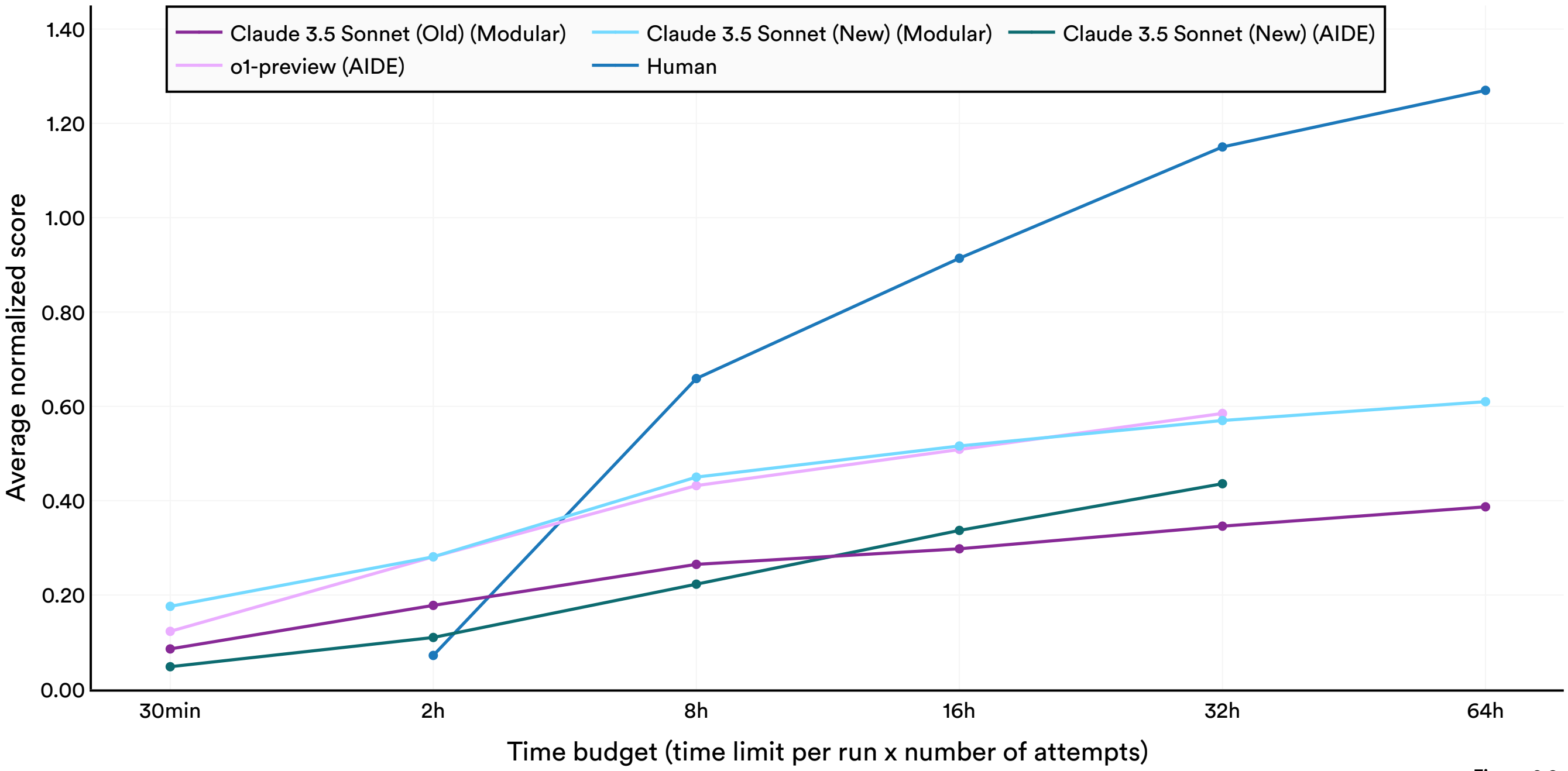

Figure 2.8.4

## GAIA

GAIA is a benchmark for General AI assistants introduced by Meta in May 2024. It consists of 466 questions designed to assess AI systems' ability to perform a broad range of tasks, including reasoning, multimodal processing, web browsing, and tool use. Unlike straightforward, exam-style questions, GAIA challenges AI models with complex, multistep problems that may require searching the open web, interpreting multimodal inputs, and reasoning through intricate scenarios (Figure 2.8.5). When researchers launched GAIA, they found that existing LLMs lagged significantly behind human performance. For instance, GPT-4 with plugins correctly answered only 15% of the questions, compared to 92% for human respondents.

As with other recently introduced AI benchmarks, performance on GAIA has improved rapidly. In 2024, the top system achieved a score of 65.1%, marking a roughly 30 percentage point increase from the highest score recorded in 2023 (Figure 2.8.6).

### Sample questions on GAIA

Source: Meta, 2024

**Level 1**

**Question:** What was the actual enrollment count of the clinical trial on H. pylori in acne vulgaris patients from Jan-May 2018 as listed on the NIH website?
**Ground truth:** 90

**Level 2**

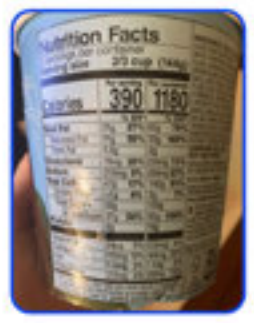

**Question:** If this whole pint is made up of ice cream, how many percent above or below the US federal standards for butterfat content is it when using the standards as reported by Wikipedia in 2020? Answer as + or - a number rounded to one decimal place.
**Ground truth:** +4.6

**Level 3**

**Question:** In NASA's Astronomy Picture of the Day on 2006 January 21, two astronauts are visible, with one appearing much smaller than the other. As of August 2023, out of the astronauts in the NASA Astronaut Group that the smaller astronaut was a member of, which one spent the least time in space, and how many minutes did he spend in space, rounded to the nearest minute? Exclude any astronauts who did not spend any time in space. Give the last name of the astronaut, separated from the number of minutes by a semicolon.
**Ground truth:** White; 5876

Figure 2.8.5

### GAIA: average score

Source: GAIA Leaderboard, 2025 | Chart: 2025 AI Index report

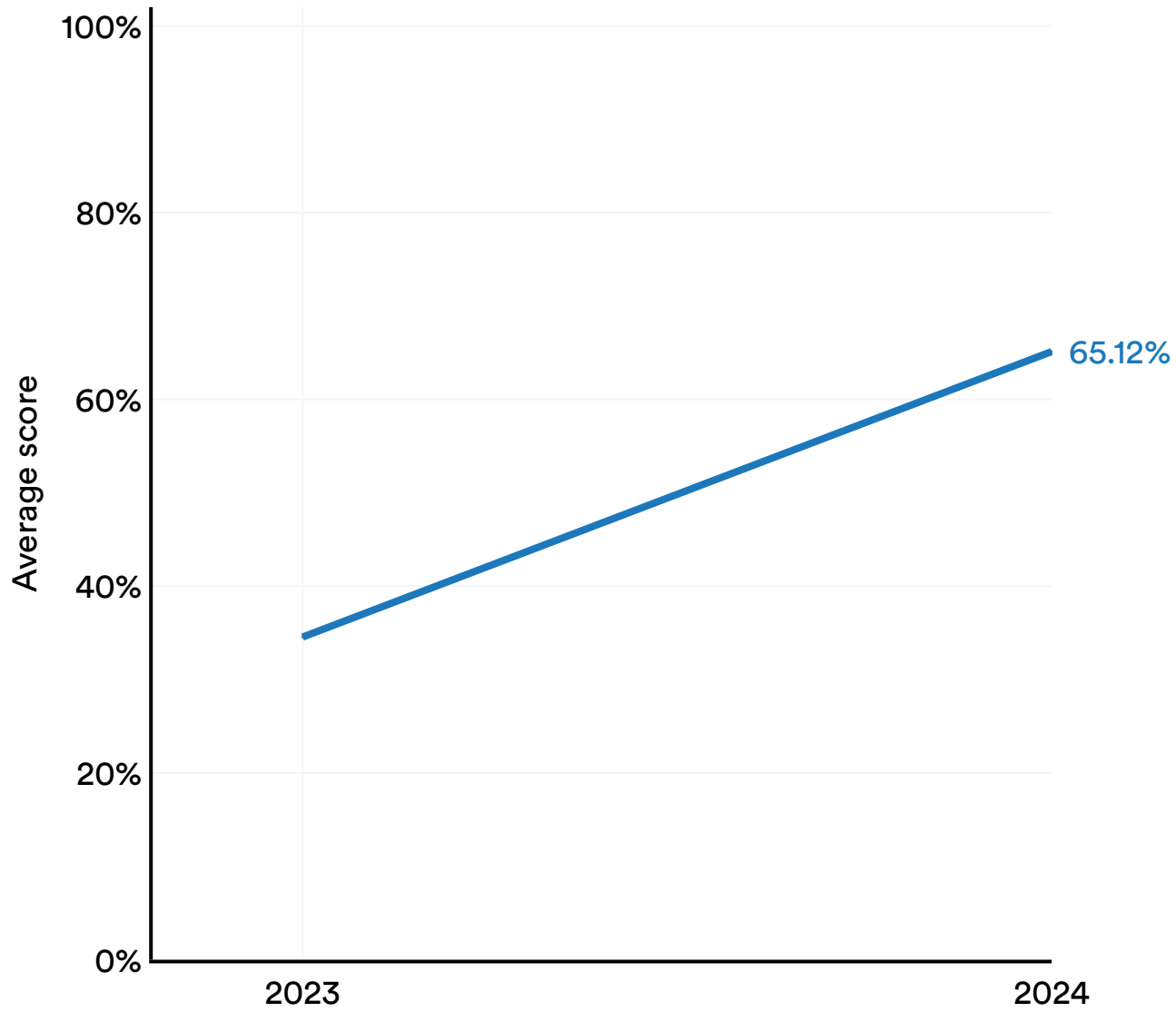

Figure 2.8.6

Advancements in AI over the past decade have paved the way for exciting new developments in the field of robotics. Especially with the rise of foundation models, robots are now able to iteratively learn from their surroundings, adapt flexibly to new settings, and make autonomous decisions. This section explores key robotic benchmarks and recent trends, including the rise of humanoids, algorithmic advancements from DeepMind, and the emergence of robotic foundation models. It concludes by studying developments in self-driving cars.

# 2.9 Robotics and Autonomous Motion

## Robotics

### RLBench

One of the most widely adopted benchmarks in the robotics community is RLBench (Robot Learning Benchmark). Launched in 2019, it features 100 unique tasks of varying complexity, from simple target reaching to opening an oven and placing a tray inside.[12] Researchers typically evaluate new robotic systems on a standardized subset of 18 tasks to gauge performance. Figure 2.9.1 visualizes some of the tasks in RLBench.

**Tasks on VisualAgentBench**
Source: James et al., 2019

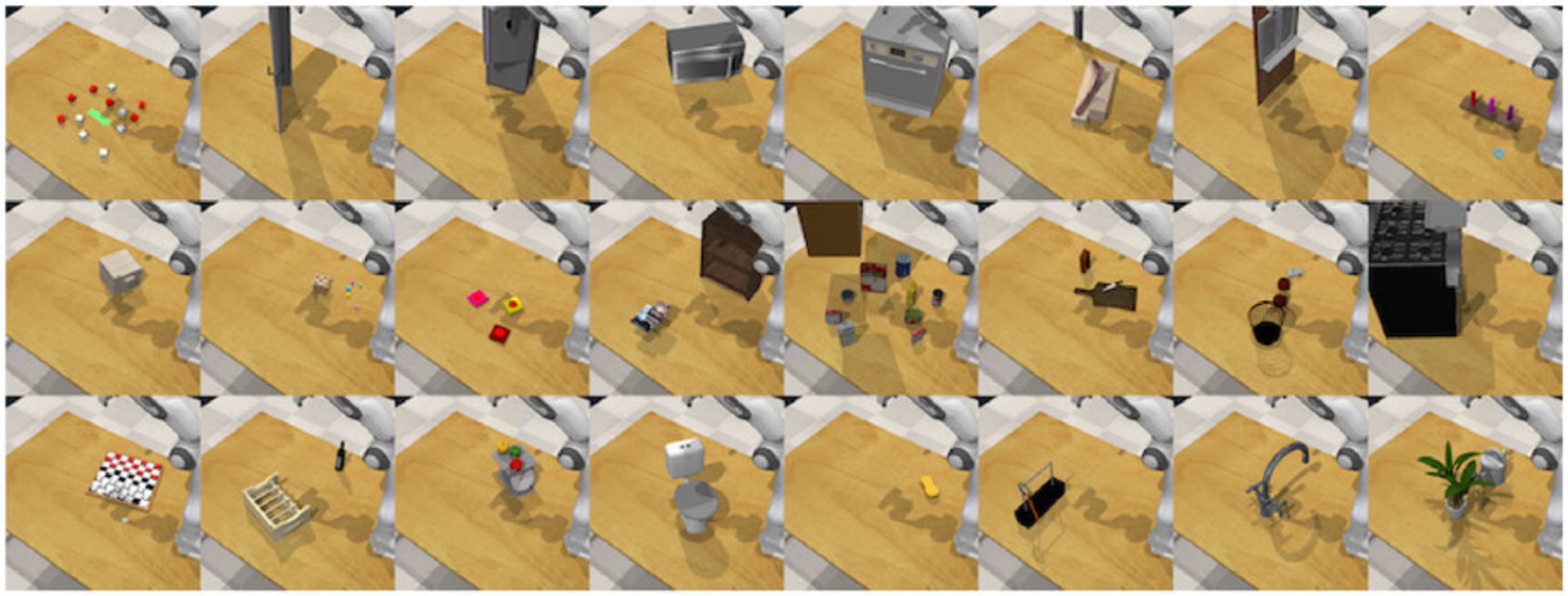

Figure 2.9.1

12 Target reaching in robotics refers to the process by which a robotic system moves its end-effector (such as a robotic arm or gripper) toward a specific goal position or object in space.

As of January 2025, the top-performing model on this subset is SAM2Act, a collaboration between researchers at the University of Washington, Universidad Católica San Pablo, Nvidia, and the Allen Institute for AI. SAM2Act achieved an 86.8% success rate, marking a 2.8 percentage point improvement over the previous state-of-the-art in 2024 and a 66.7 percentage point increase from the leading score in 2021 (Figure 2.9.2).

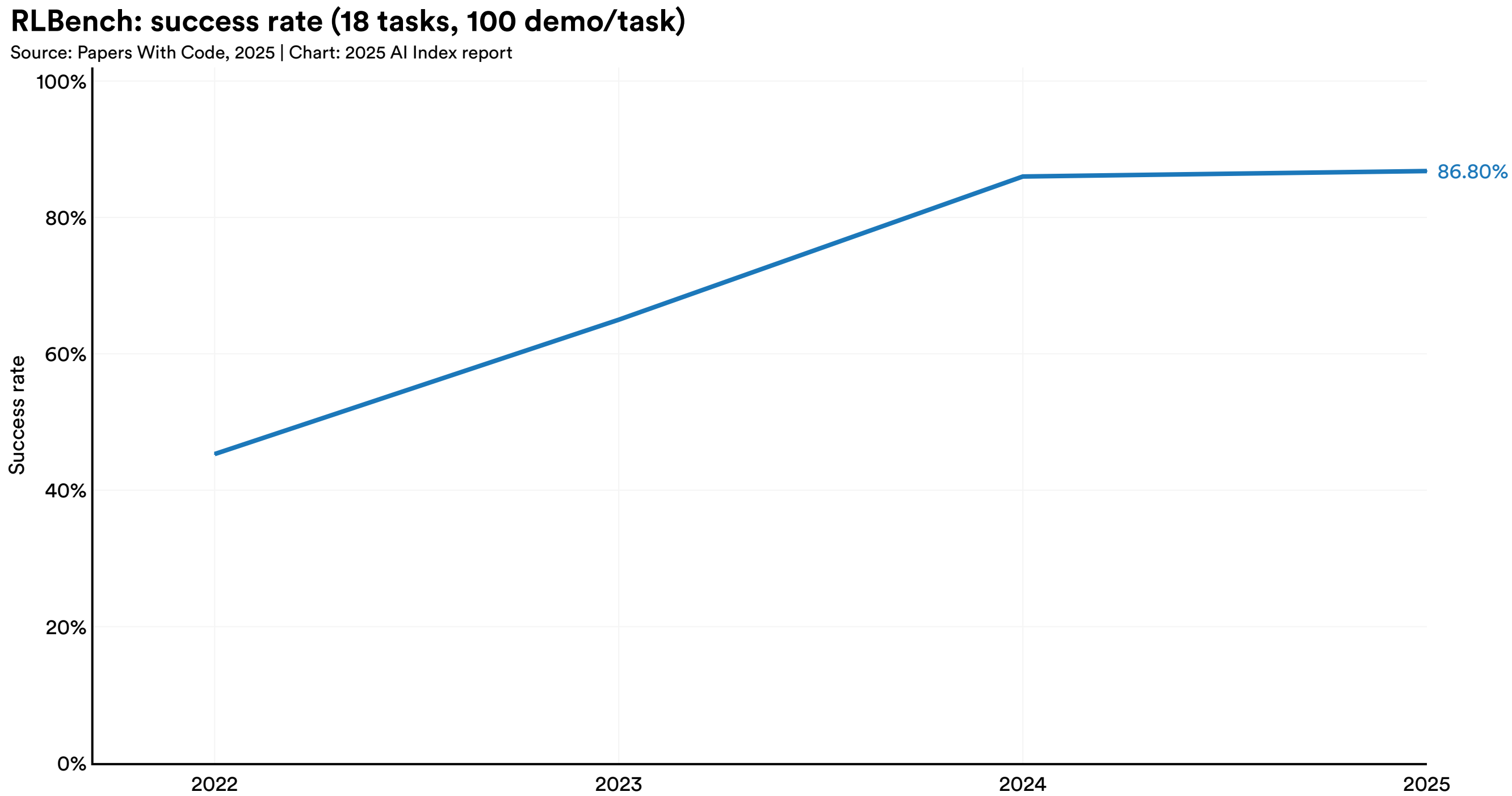

Figure 2.9.2

**Highlight:**

## Humanoid Robotics

2024 was a significant year for robotics, marked by the growing prevalence of humanoid robots—machines with humanlike bodies designed to mimic human functions. For example, Figure AI, a robotics startup dedicated to developing general-purpose humanoid robots, launched Figure 02 in 2024, its most advanced model yet. Standing 5 feet 6 inches tall, weighing 154 pounds, and capable of handling a 44-pound payload, Figure 02 operates for up to five hours on a single charge. Figure robots are able to perform complex tasks such as making coffee and assisting in automotive assembly by placing sheet metal into a car fixture (Figure 2.9.3 and Figure 2.9.4). They are also integrated with OpenAI and can engage in speech-to-speech reasoning, whereby the robot explains its actions and responds to queries about its behavior. Figure's success follows that of other companies that released humanoid robots, like Tesla's Optimus, first launched in 2002 and redesigned in 2023, and Boston Dynamics' Atlas humanoid.

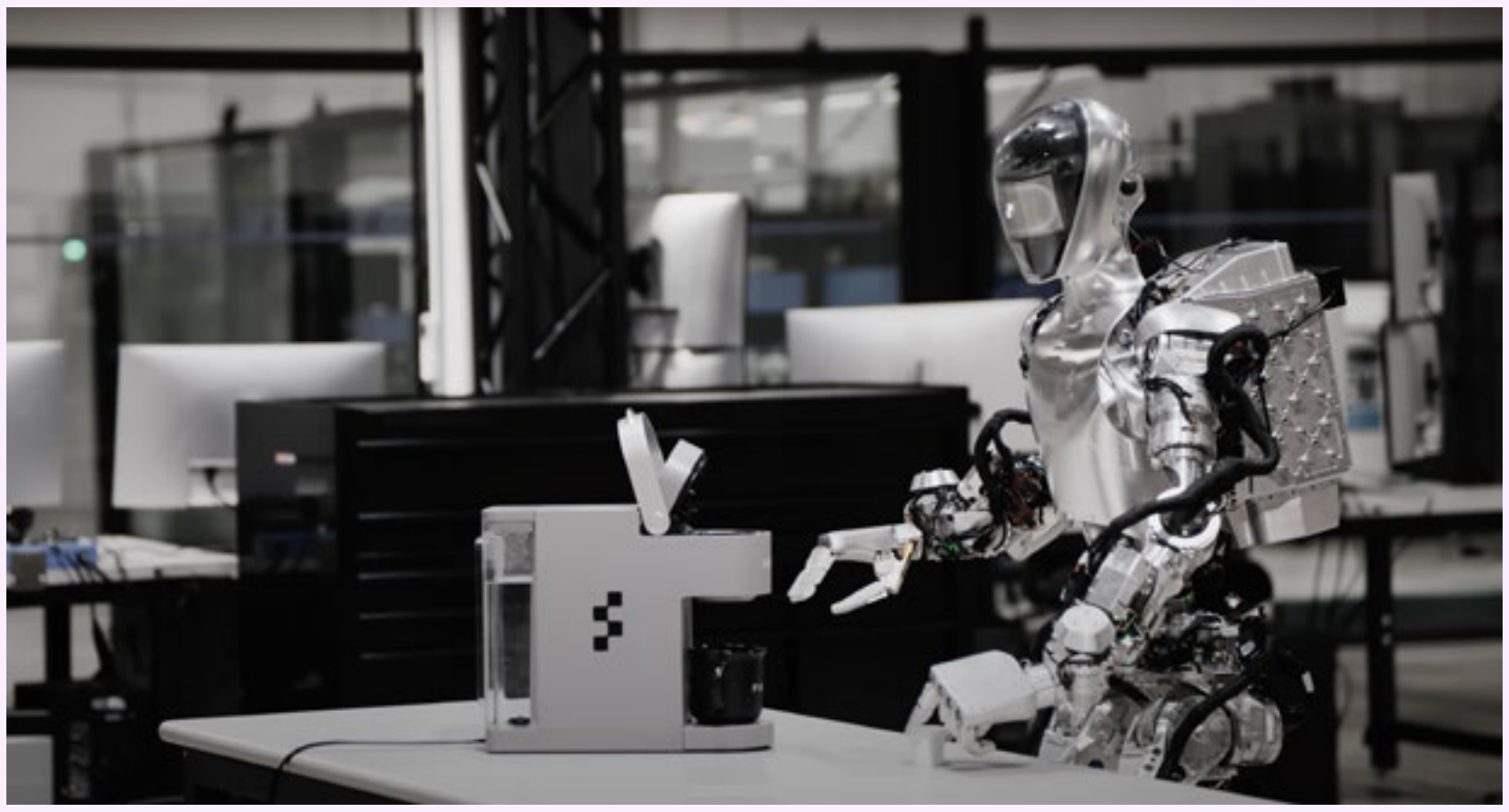

**Figure robot making coffee**
Source: Figure AI
Figure 2.9.3

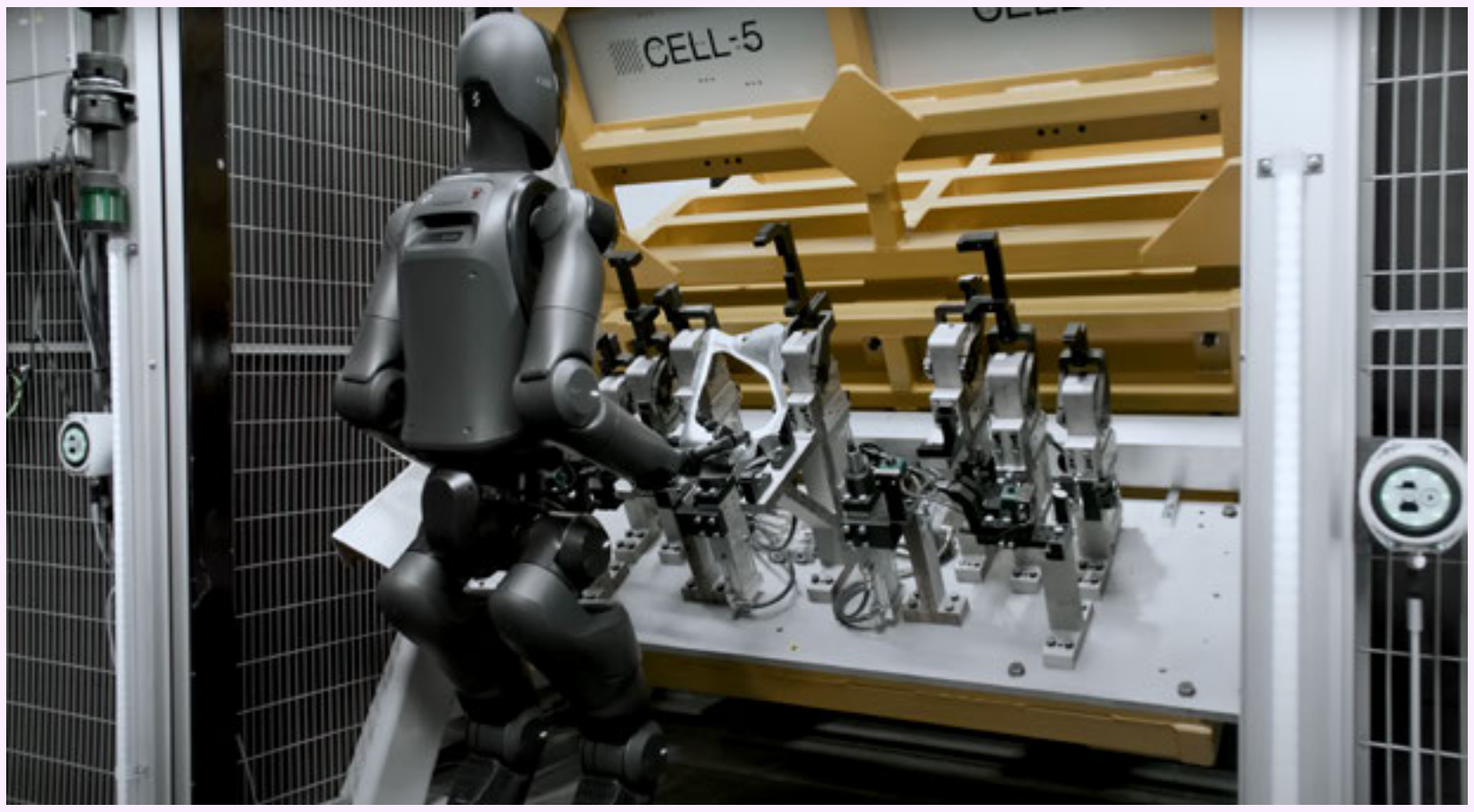

**Figure robot assisting in automotive assembly**
Source: Figure AI
Figure 2.9.4

Highlight:

# DeepMind’s Developments

In 2023, DeepMind launched two robotic models, PaLM-E and RT-2. These models were novel in their use of transformer-based architectures, typically found in language modeling, and their training on both manipulation data and language data. This dual training approach enabled them to excel at both robotic manipulation and text generation. In 2024, DeepMind introduced AutoRT, an AI system that leverages large foundation models to autonomously generate diverse training data for robots. It coordinates multiple video-equipped robots, guiding them through various environments, devising creative tasks for them to perform, and meticulously documenting these tasks (Figure 2.9.5). This documentation then serves as training data for future robotic learning. To date, AutoRT has generated a dataset of 77,000 robotic trials spanning 6,650 unique tasks. Greater amounts of robotic training data will be important to improve the training of future robotic systems.

**AutoRT workflow**
Source: Google DeepMind, 2024

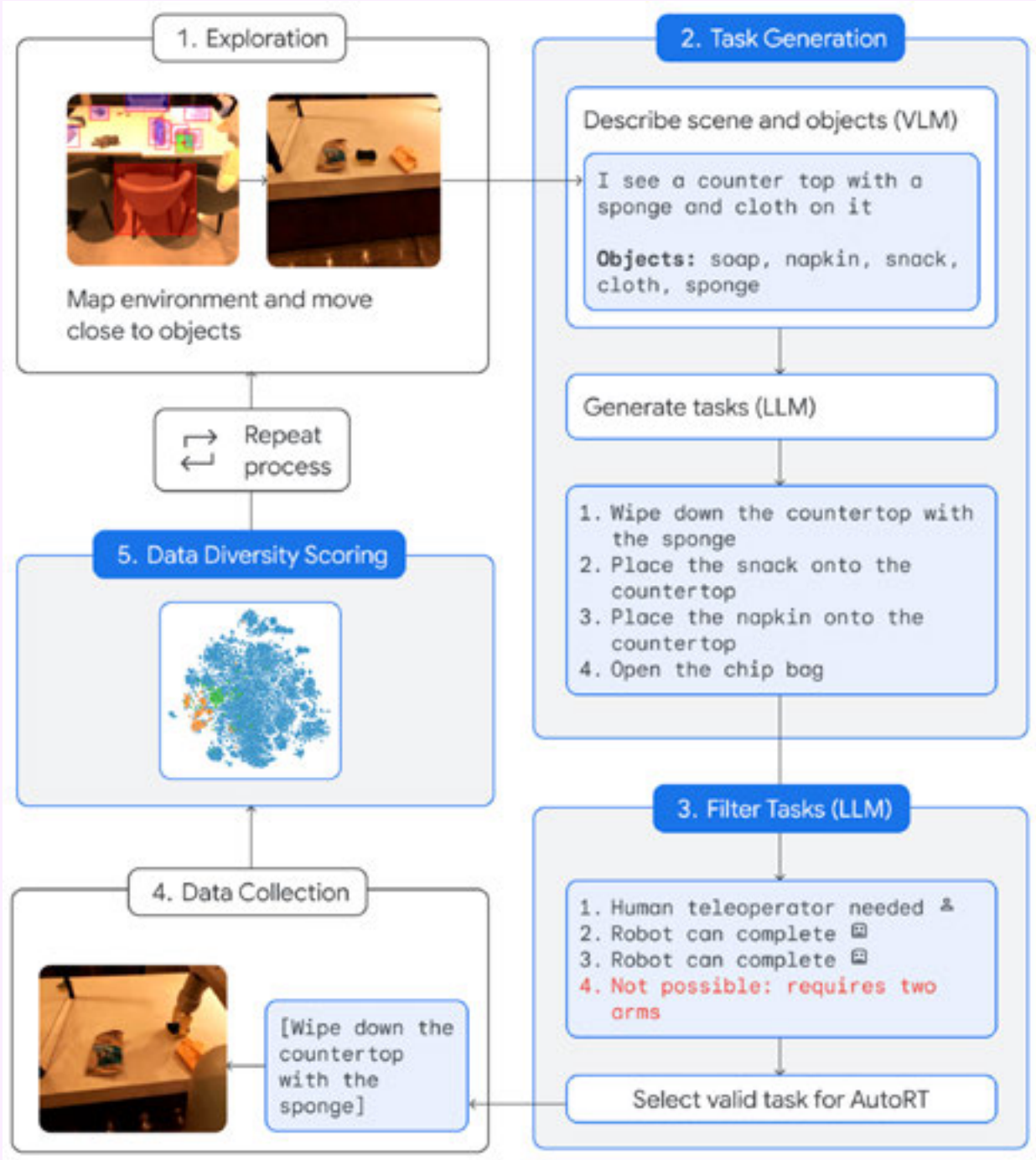

Figure 2.9.5

Conversely, SARA-RT, also from Google DeepMind, improves the efficiency of transformer-based robotic models by significantly improving their speed. While transformers are powerful, they are also computationally intensive as they rely on quadratic complexity attention mechanisms. This means that doubling the input size of data provided to a model can quadruple computational requirements. This challenge complicates attempts to scale robotic models. SARA-RT addresses this challenge with a technique called “up-training,” which converts the quadratic complexity of standard transformers into a linear model. This method drastically reduces computational demands while maintaining performance quality. Figure 2.9.6 compares speed tests of AI models enhanced with the SARA technique against those without. In point cloud processing,

**Speed tests for SARA vs. non-SARA enhanced models**
Source: Google DeepMind, 2024

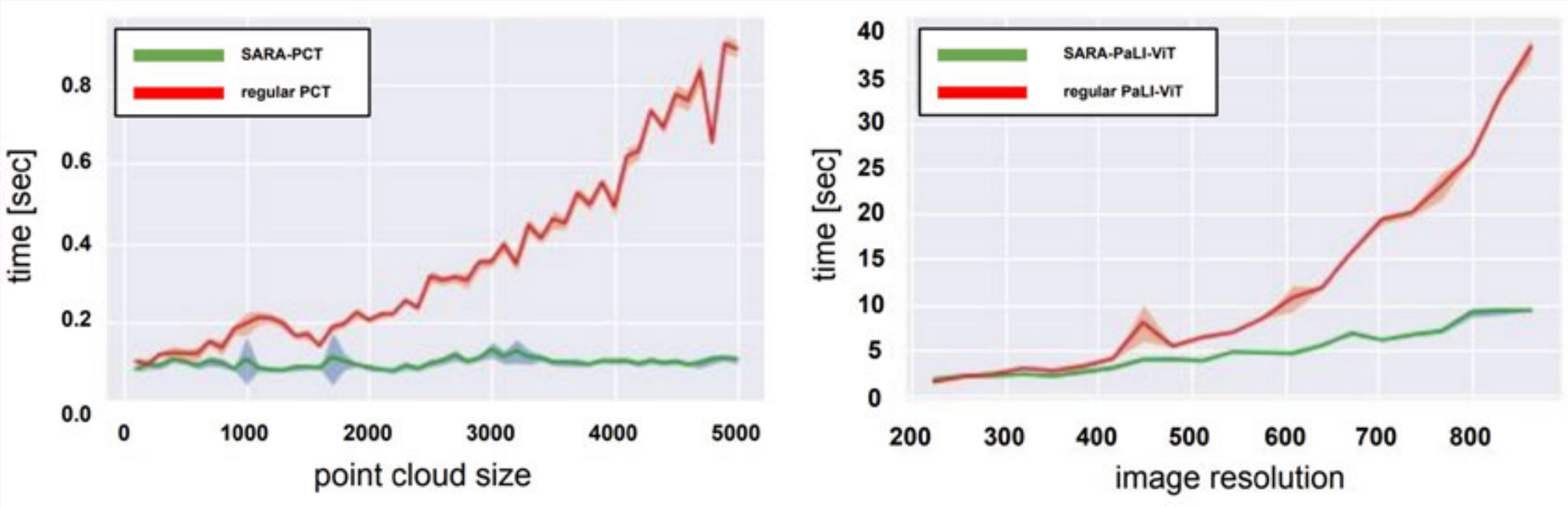

Figure 2.9.6

Highlight:

# DeepMind's Developments (cont'd)

which enables robots to interpret 3D environments, and in image processing, SARA-based models run significantly faster while avoiding major increases in run-time at scale.

Other developments from DeepMind include ALOHA (Autonomous Learning of High-level Activities) and DemoStart. ALOHA Unleashed is a breakthrough in enabling robots to perform intricate dexterous manipulation tasks, such as tying shoelaces or hanging T-shirts on coat hangers—tasks that historically have been extremely challenging for robots. The researchers demonstrated that combining a large imitation learning dataset with a transformer-based learning architecture is a highly effective approach for overcoming these difficulties. The ALOHA approach enabled Google's robot to effectively learn a diverse range of tasks, including hanging a shirt, stacking kitchen items, and tying shoelaces (Figure 2.9.7). As shown in Figure 2.9.8, ALOHA-trained robots achieved a high success rate across these tasks.

**ALOHA-trained robot attempting complex tasks**
Source: Google DeepMind, 2024

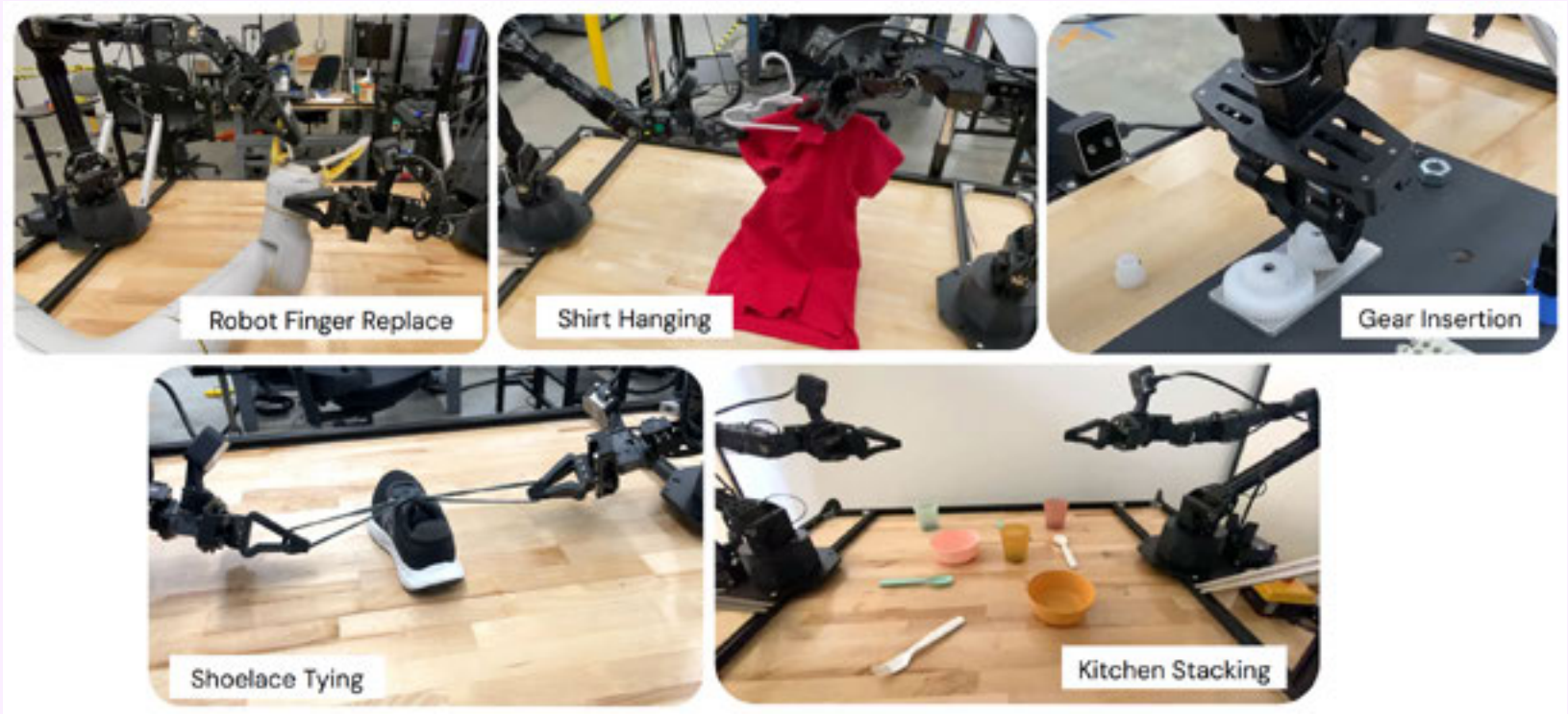

Figure 2.9.7

**ALOHA: success rate**
Source: Zhao et al., 2024 | Chart: 2025 AI Index report

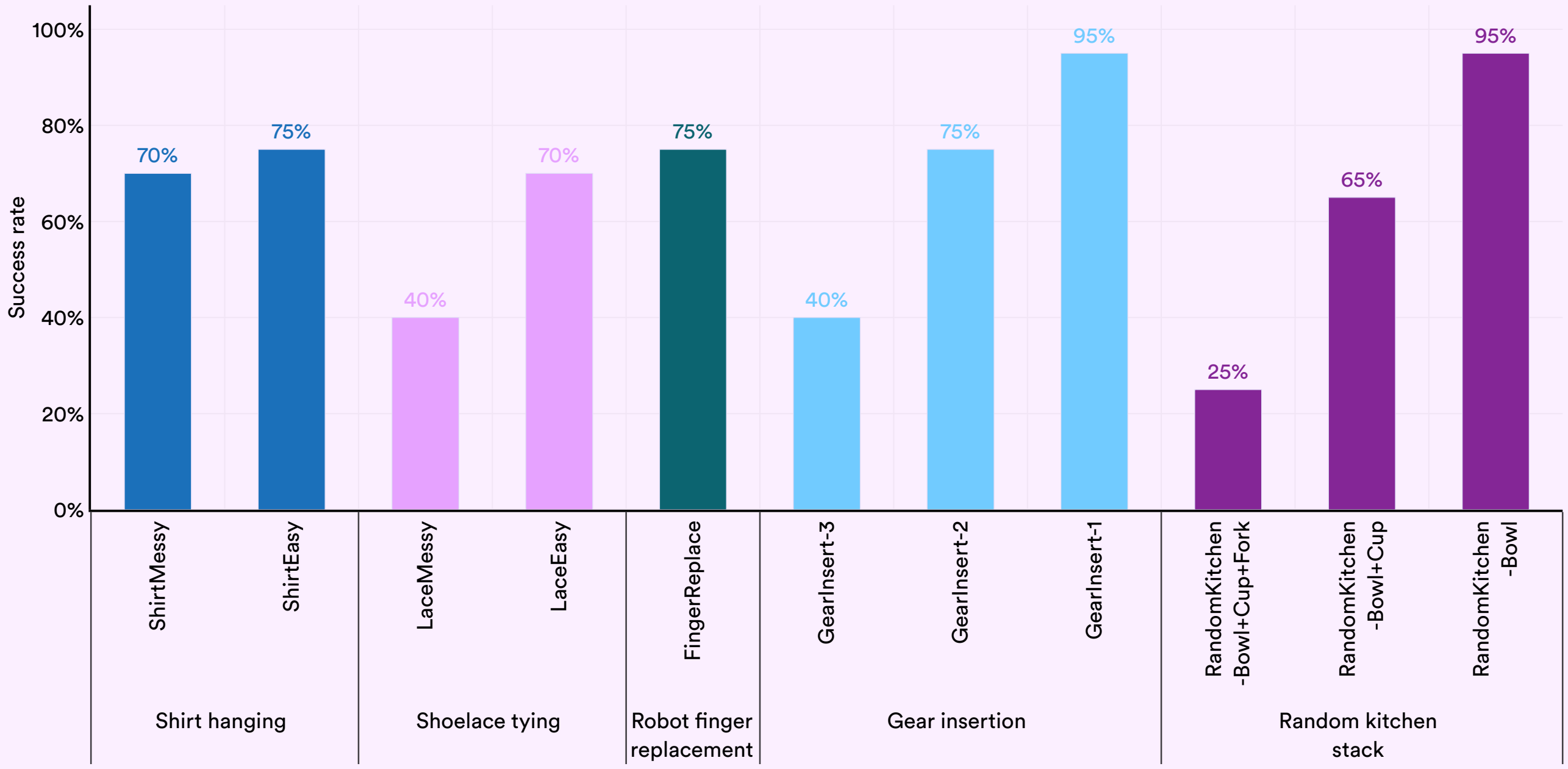

Figure 2.9.8

**Highlight:**

# DeepMind's Developments (cont'd)

Similarly, DemoStart introduces a novel auto-curriculum reinforcement learning method that enables a robotic arm to master complex behaviors using only sparse rewards and a limited number of demonstrations. This breakthrough highlights the potential for robots to learn efficiently with minimal data, reducing the need for data-intensive training and making advanced robotics more accessible and widely adopted. DeepMind also introduced a robotic model in 2024 that was capable of reaching amateur human-level performance in competitive table tennis (Figure 2.9.9). Given that achieving human-level speed and performance on real-world tasks is an important benchmark for robotics research, this achievement is a notable step forward in robotic ability.

**Robots playing amateur-level table tennis**
Source: Google DeepMind, 2024

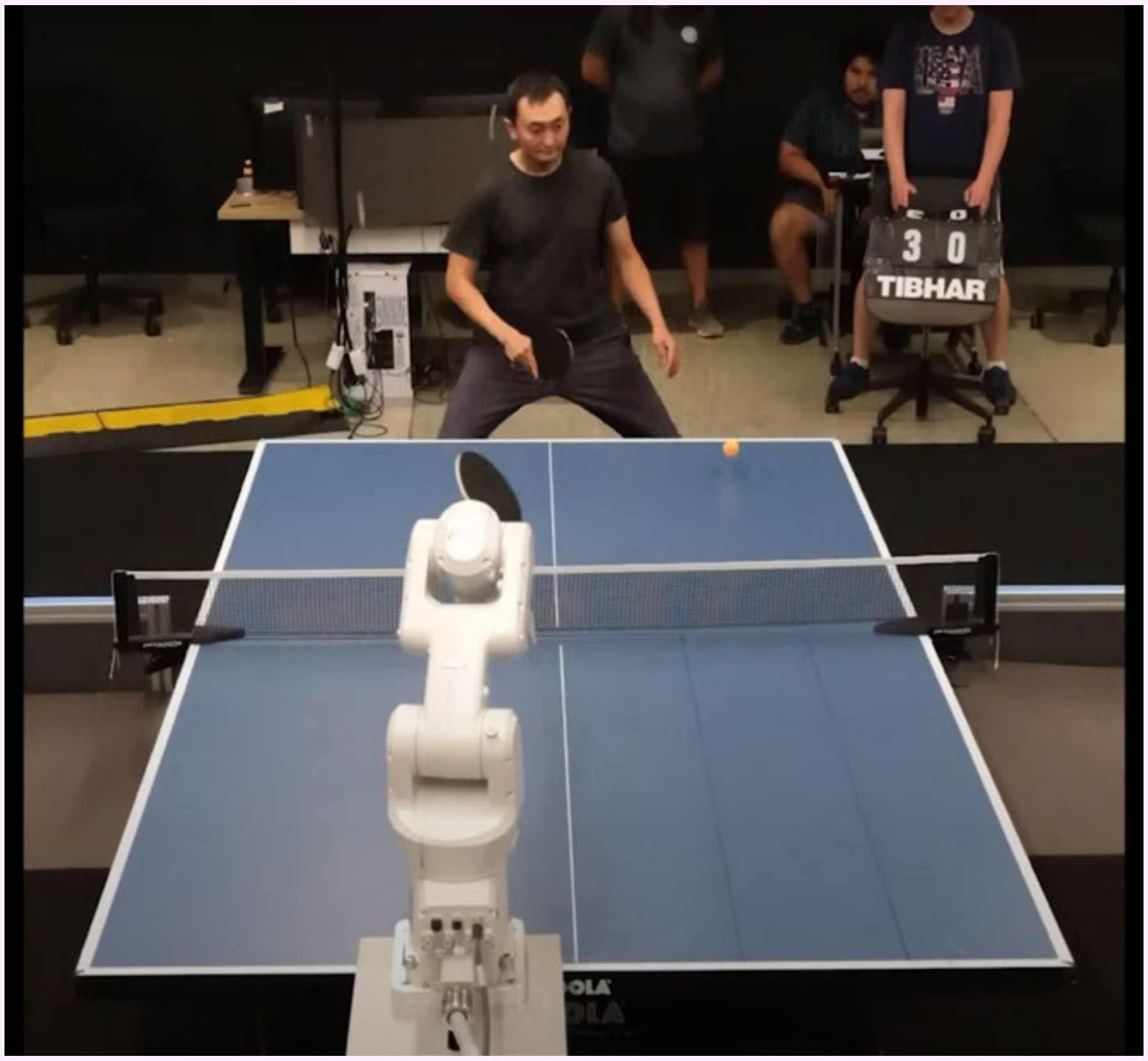

Figure 2.9.9

Highlight:

## Foundation Models for Robotics

In 2024, there was a strong push toward developing foundational models for robotics—systems capable of reasoning with language while physically operating in the real world. Nvidia introduced GR00T (Generalist Robot 00 Technology), a general-purpose foundation model for humanoid robots designed to understand natural language and mimic human movements. Alongside GR00T, Nvidia released data pipelines, simulation frameworks, and the Thor robotics computer. Figure 2.9.10 illustrates the components of GROOT's launch. This robotic development suite is intended to make it easier for the robotic community to scale and build increasingly advanced robotics.

Nvidia was not alone in this space. Covariant launched RFM-1, a robotic foundation model with language capabilities and real-world maneuverability. Meanwhile, LLaRA, developed by researchers at Stony Brook University and the University of Wisconsin-Madison, integrates perception, communication, and action into a monolithic, end-to-end deep learning model. These new models continue a trend from 2023, which saw the launch of robotic foundation models like RT-2, PaLM-E, and Open-X Embodiment.

**GROOT blueprint for synthetic motion generation**
Source: Nvidia, 2024

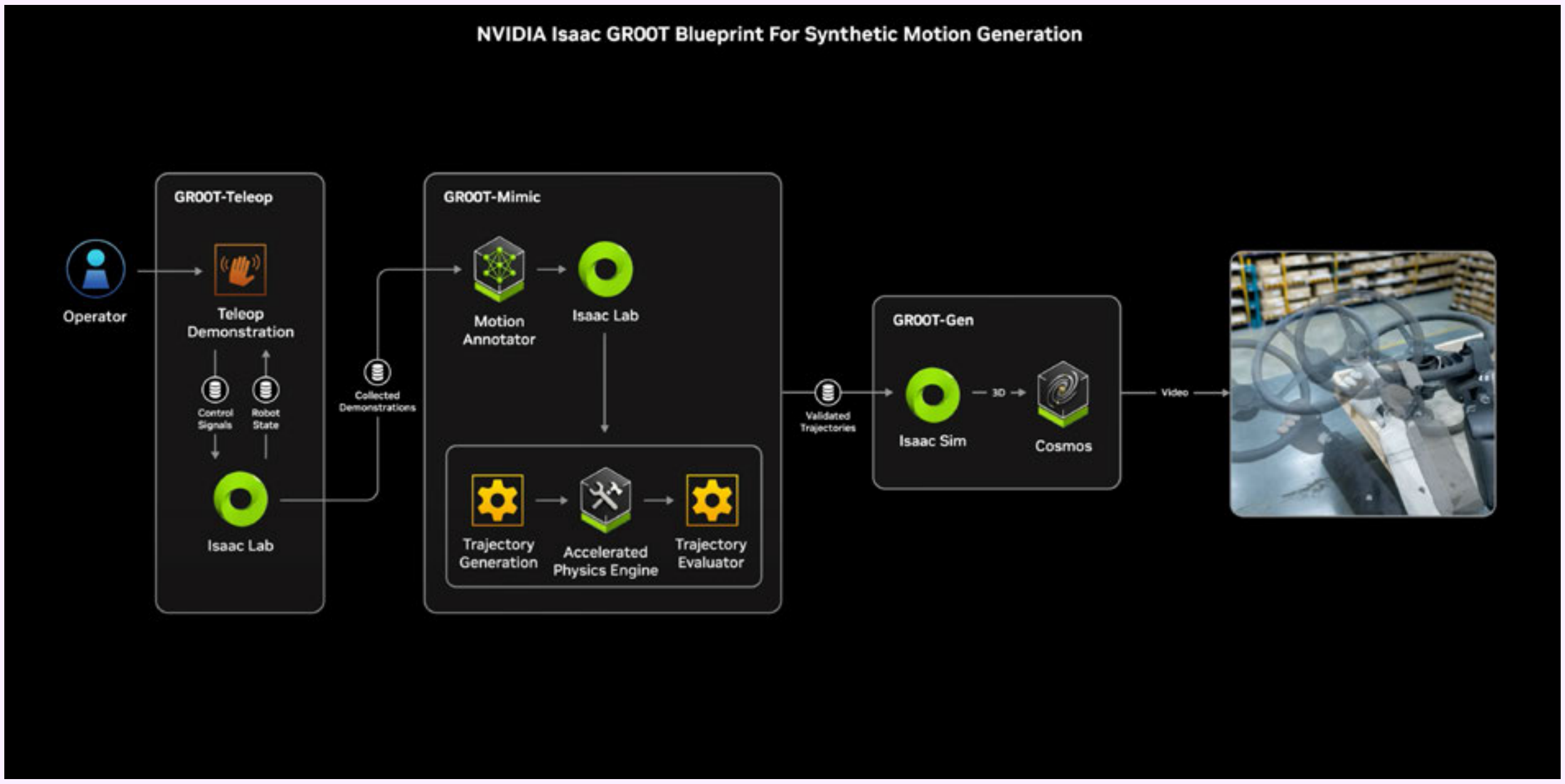

Figure 2.9.10

# Self-Driving Cars

Self-driving vehicles have long been a goal for AI researchers and technologists. However, their widespread adoption has been slower than anticipated. Despite many predictions that fully autonomous driving is imminent, widespread use of self-driving vehicles has yet to become a reality. Still, in recent years, significant progress has been made. In cities like San Francisco and Phoenix, fleets of self-driving taxis are now operating commercially. This section examines recent advancements in autonomous driving, focusing on deployment, technological breakthroughs and new benchmarks, safety performance, and policy challenges.

## Deployment

Self-driving cars are increasingly being deployed worldwide. Cruise, a subsidiary of General Motors, launched its autonomous vehicles in San Francisco in late 2022 before having its license suspended in 2023 after a litany of safety incidents. Waymo, a subsidiary of Alphabet, began deploying its robotaxis in Phoenix in early 2022 and expanded to San Francisco in 2024. The company has since emerged as one of the more successful players in the self-driving industry: As of January 2025, Waymo operates in four major U.S. cities—Phoenix, San Francisco, Los Angeles, and Austin (Figure 2.9.11). Data sourced from October 2024 suggests that across the four cities the company provides 150,000 paid rides per week, covering over a million miles. Looking ahead, Waymo plans to test its vehicles in 10 additional cities, including Las Vegas, San Diego, and Miami. The company chose testing locations, such as upstate New York and Truckee, California, that experience snowy weather so it can assess the vehicles in diverse driving conditions. There has also been notable progress in self-driving trucks, with companies like Kodiak completing its first driverless deliveries and Aurora reporting steady advancements, including over 1 million miles of autonomous freight hauling on U.S. highways since 2021—albeit with human safety drivers present. Still, challenges remain in bringing this technology to market, with Aurora recently announcing it would delay the commercial launch of its fleet from the end of 2024 until April 2025.

**Waymo rider-only miles driven without a human driver**

Source: Waymo, 2024 | Table: 2025 AI Index report

| Location | Rider-only miles through September 2024 |
|---|---|
| Los Angeles | 1.947M |
| San Francisco | 10.209M |
| Phoenix | 20.823M |
| Austin | 124K |

Figure 2.9.11

China's self-driving revolution is also accelerating, led by companies like Baidu's Apollo Go, which reported 988,000 rides across China in Q3 2024, reflecting a 20% year-over-year increase. In October 2024, the company was operating 400 robotaxis and announced plans to expand its fleet to 1,000 by the end of 2025. Pony.AI, another Chinese autonomous vehicle manufacturer, has pledged to scale its robotaxi fleet from 200 to at least 1,000 vehicles—with expectations that the fleet will reach 2,000 to 3,000 by the end of 2026. China is leading the way in autonomous vehicle testing, with reports indicating that it is testing more driverless cars than any other country and currently rolling them out across 16 cities. Robotaxis in China are notably affordable—even cheaper, in some cases, than rides provided by human drivers. To support this growth, China has prioritized establishing national regulations to govern the deployment of driverless cars. Beyond the self-driving revolution taking place in the U.S. and China, European startups like Wayve are beginning to gain traction in the industry.

### Technical Innovations and New Benchmarks

Over the past year, self-driving technology has advanced significantly, both in vehicle capabilities and benchmarking methods. In October 2024, Tesla unveiled the Cybercab, a two-passenger autonomous vehicle without a steering wheel or pedals, which is set for production in 2026 at a price of under $30,000. Tesla also unveiled the Robovan, an electric autonomous van designed to transport up to 20 passengers. Meanwhile, Baidu's Apollo Go launched its latest-generation robotaxi, the RT6, across multiple cities in China (Figure 2.9.12). With a price tag of just $30,000 and a battery-swapping system, the RT6 represents a major step toward making self-driving technology more cost-effective and scalable. As costs continue to decline, the adoption of autonomous vehicles is expected to accelerate. Notable business partnerships have also advanced self-driving technology, including Uber's collaboration with WeRide—the world's first publicly listed robotaxi company—to develop an autonomous ride-sharing platform in Abu Dhabi.

In 2024, several new benchmarks were introduced to evaluate self-driving capabilities. One notable example is nuPlan, developed by Motional. It is a large-scale, autonomous driving dataset designed to test machine-learning-based motion planners. The benchmark includes 1,282 hours of diverse driving scenarios from multiple cities, along with a simulation and evaluation framework that enables planners' actions to be tested in closed-loop settings. Another recent benchmark is OpenAD, the first real-world, open-world autonomous driving benchmark for 3D object detection. OpenAD focuses on domain generalization—the ability of autonomous driving systems to adapt across diverse sensor configurations—and open-vocabulary recognition, which allows systems to identify previously unseen semantic categories.

**Baidu's RT-6**
Source: Verge, 2024

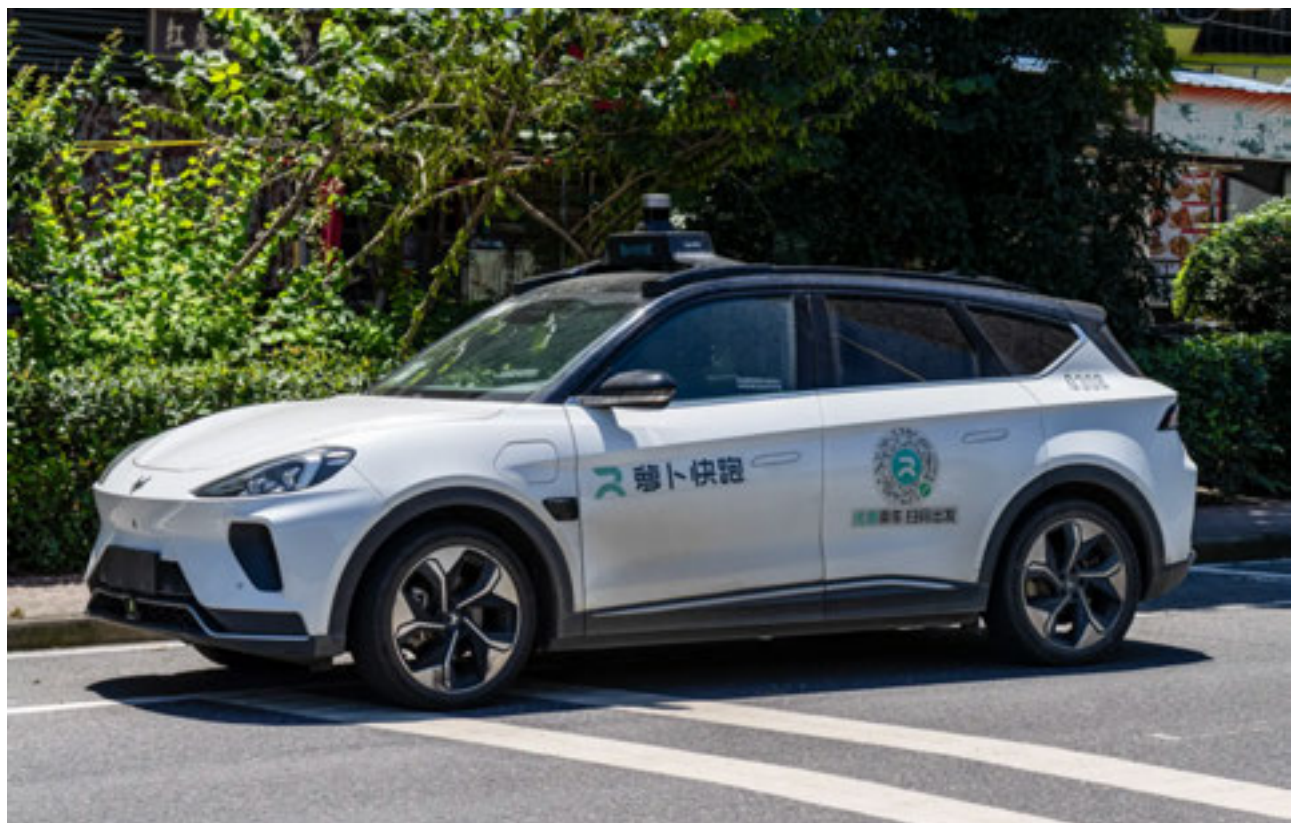

Figure 2.9.12

Most existing benchmarks for end-to-end autonomous driving rely on open-loop evaluation, which can be restrictive. Open-loop settings fail to test how autonomous agents react to real-world conditions and often lead to models that memorize driving patterns rather than learning to drive authentically. While closed-loop benchmarks like Town05Long and Longest6 exist, they primarily assess basic driving skills rather than performance in complex, interactive scenarios. Bench2Drive is another new benchmark that improves on these limitations by providing a comprehensive, realistic, closed-loop testing simulation environment for end-to-end autonomous vehicles (Figure 2.9.13). It includes a training set with over 2 million fully annotated frames sourced from more than 10,000 clips, as well as an evaluation suite with 220 short routes designed to test autonomous driving capabilities in diverse conditions. Figure 2.9.14 displays the driving scores of various autonomous driving methods evaluated on the Bench2Drive benchmark.[13]

**An overview of Bench2Drive**
Source: Jia et al., 2024
Figure 2.9.13

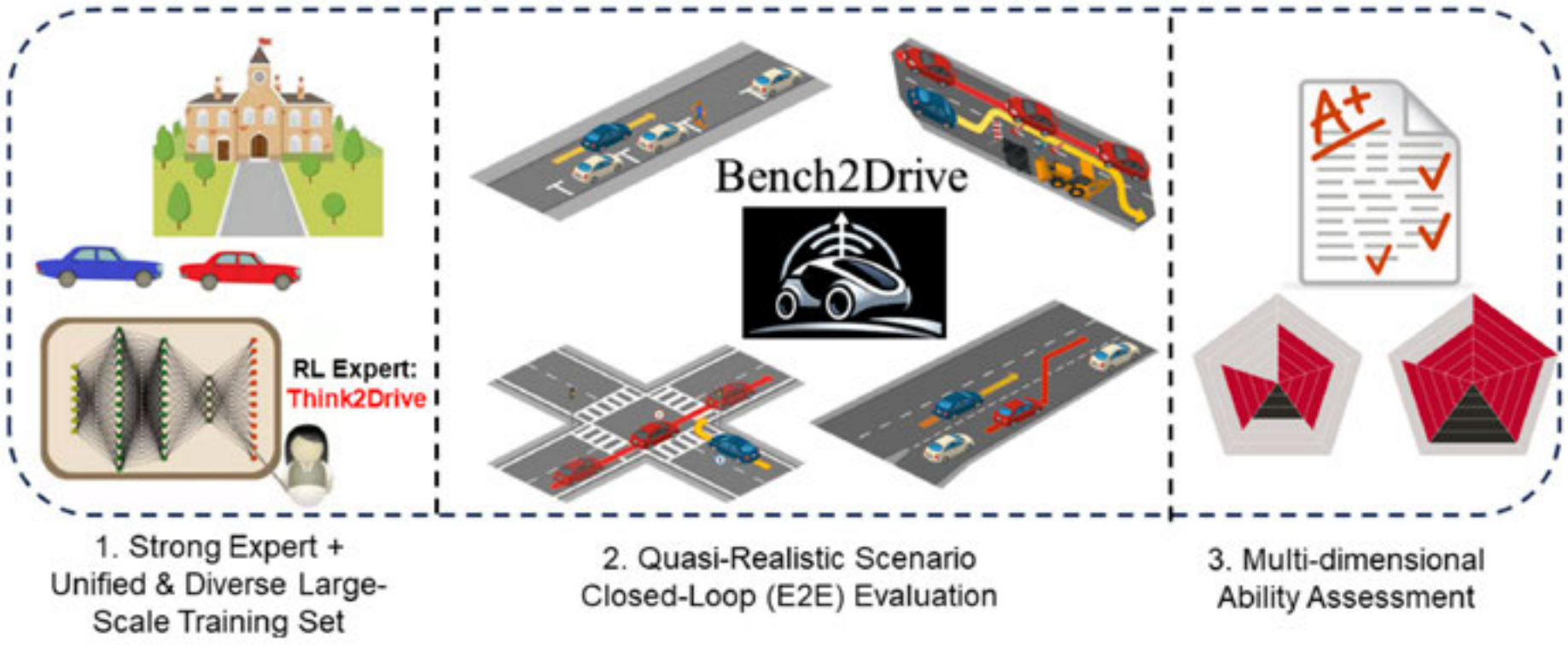

13 This metric accounts for both route completion and infractions, averaging route completion percentages while applying penalties based on infraction severity. For more detail on the driving score methodology, see Section 3 of the Bench2Drive paper.

**Bench2Drive: driving score**
Source: Jia et al., 2024 | Chart: 2025 AI Index report

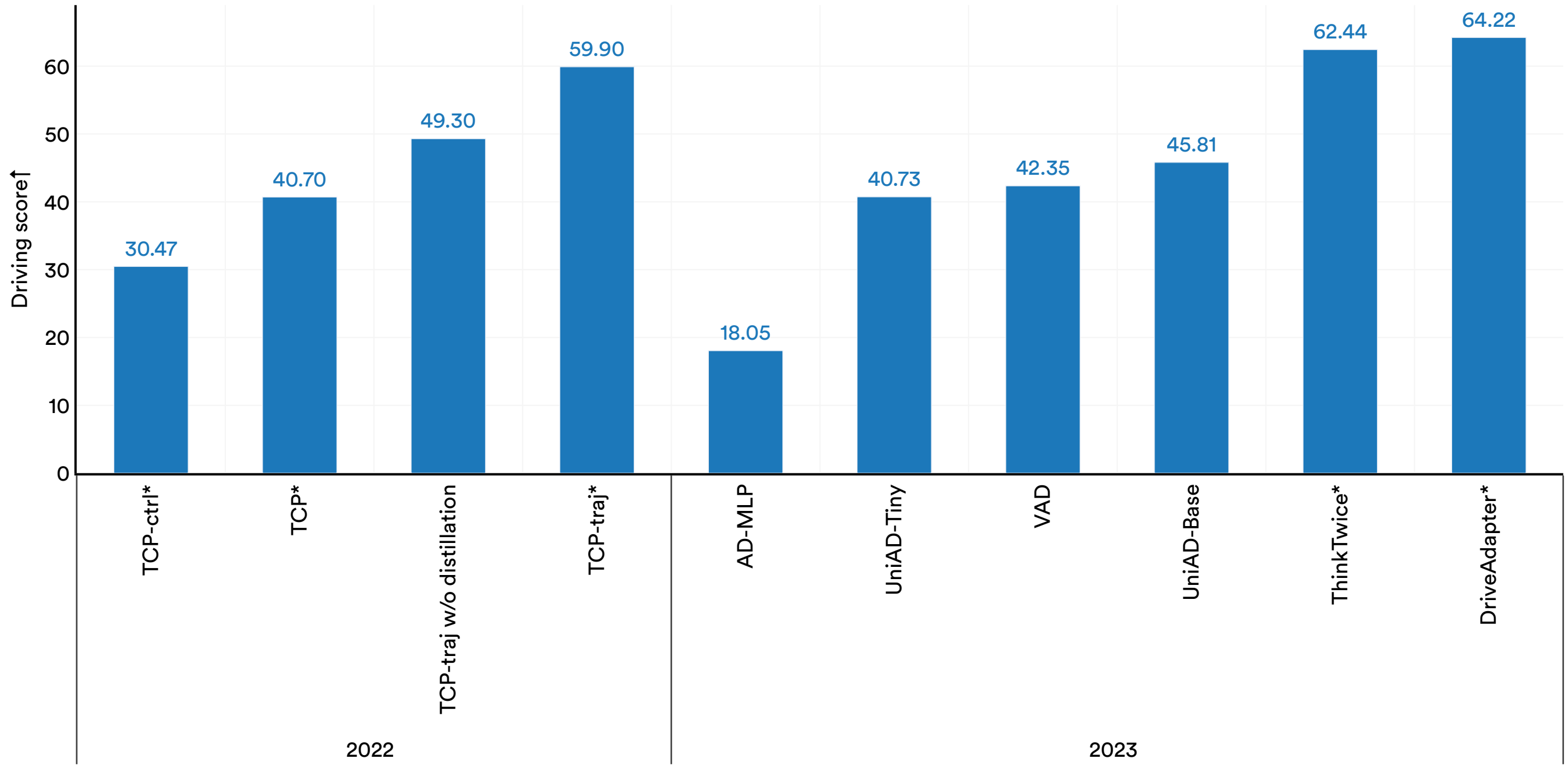

Figure 2.9.14

## Safety Standards

Emerging research suggests that self-driving cars may be safer than human-driven vehicles. Figure 2.9.15 compares the number of reported incidents per million miles driven by Waymo vehicles to the estimated rates if humans had driven the same distance. The data shows that Waymo vehicles had significantly fewer incidents, including 1.42 fewer airbag deployments, 3.16 fewer crashes with reported injuries, and 3.65 fewer police-reported crashes per million miles (Figure 2.9.15). Figure 2.9.16 highlights the differences in incident rates across various crash locations, revealing that across all locations with available data, Waymo vehicles consistently recorded lower rates of airbag deployments, injury-reported crashes, and police-reported incidents.

**Waymo driver vs. human benchmarks in Phoenix and San Francisco**
Source: Waymo, 2024 | Chart: 2025 AI Index report

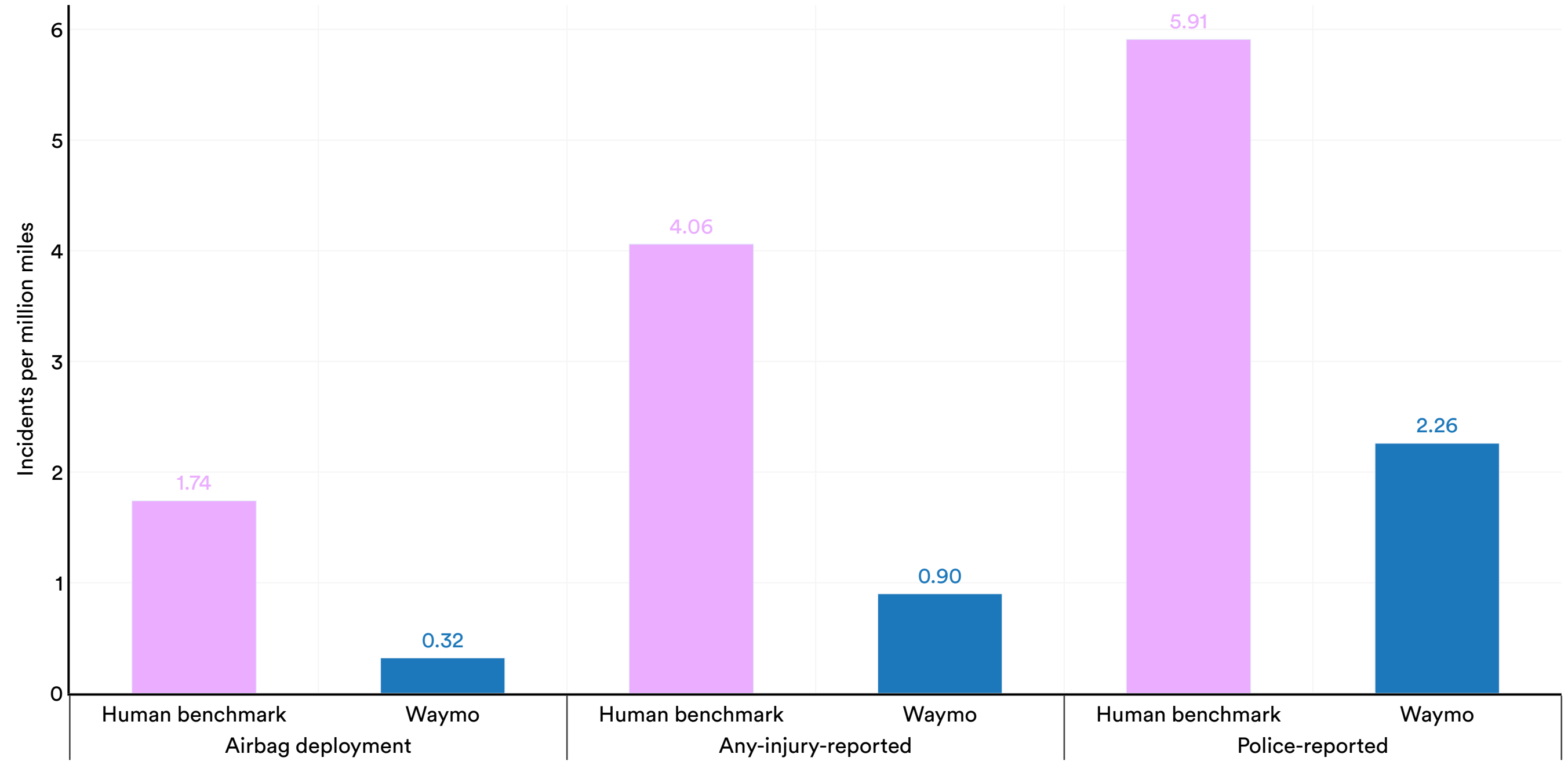

Figure 2.9.15[14]

**Waymo driver percent difference to human benchmark in Phoenix and San Francisco**
Source: Waymo, 2024 | Chart: 2025 AI Index report

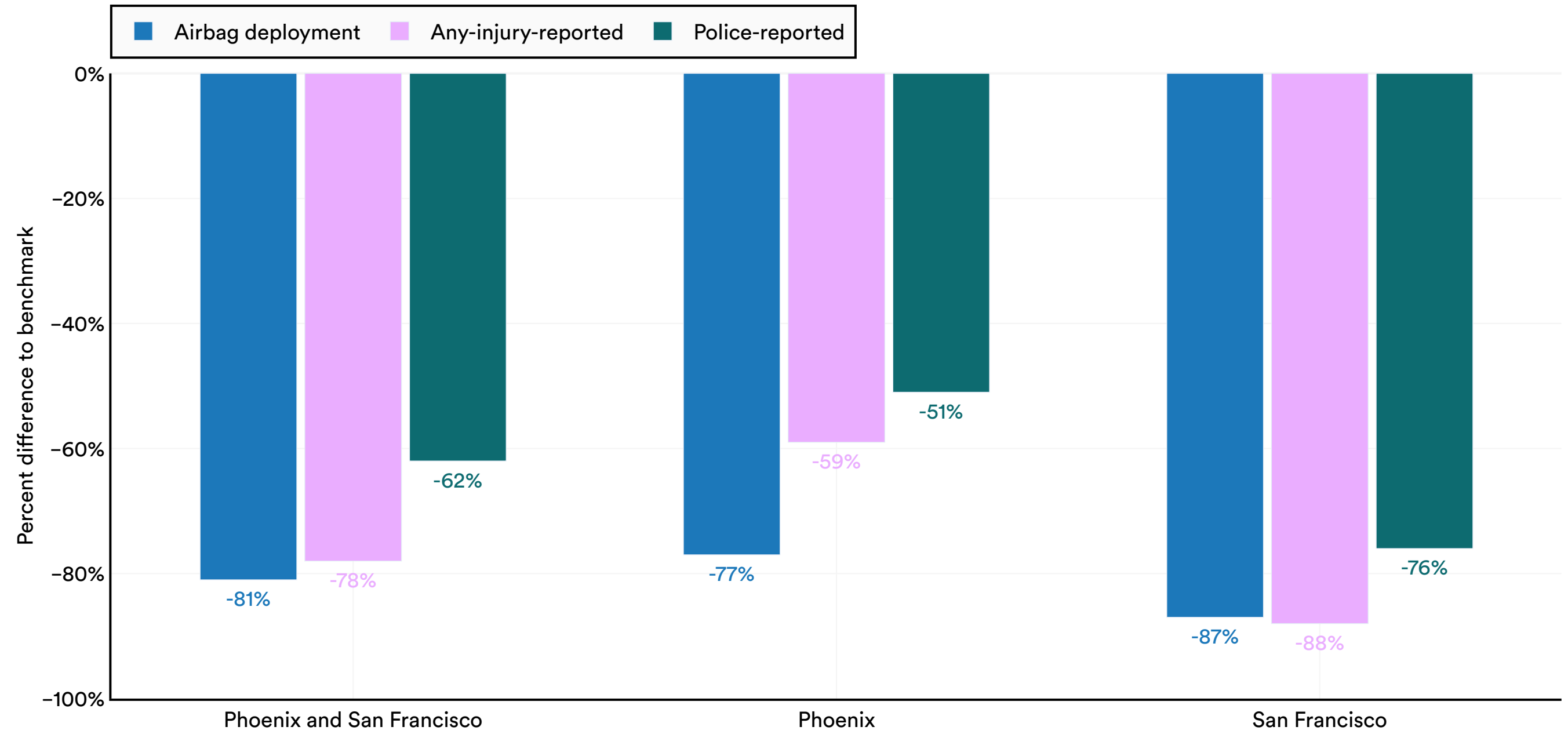

Figure 2.9.16

14 Waymo's safety data is continuously updated in real time, so the totals reported in this section may not fully align with those currently displayed on their website.

Waymo, in collaboration with Swiss Re, one of the world's leading reinsurers, also conducted a study analyzing liability claims related to collisions over several million miles driven by its fully autonomous vehicles. The study compared Waymo's liability claims to human-driver baselines derived from Swiss Re's extensive dataset, which includes over 500,000 claims and 200 billion miles of driving data. The results showed that Waymo vehicles had an 88% reduction in property damage claims and a 92% reduction in bodily injury claims (Figure 2.9.17). In real terms, across 25.3 million miles driven, Waymo vehicles were involved in just nine property damage claims and two bodily injury claims, whereas human drivers over the same distance would be expected to incur 78 property damage claims and 26 bodily injury claims. The Waymo drivers were also significantly safer than latest-generation human-driven vehicles that are equipped with added safety features.

**Comparison of liability insurance claims by type: Waymo driver vs. human-driven vehicles**
Source: Di Lillo et al., 2024 | Chart: 2025 AI Index report

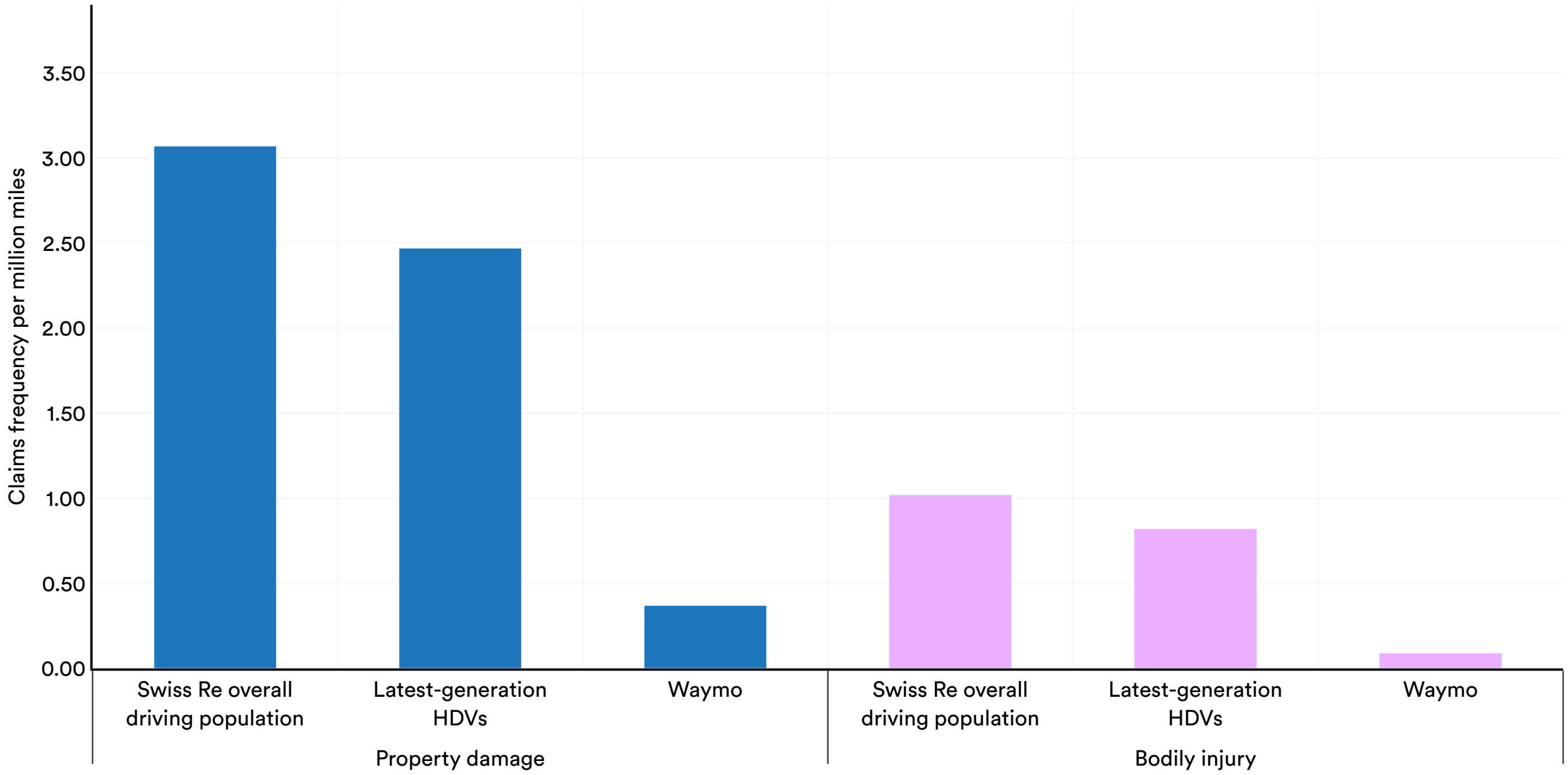

Figure 2.9.17

# CHAPTER 3:
## Responsible AI

Text and analysis by Anka Reuel

# Chapter 3: Responsible AI

ACCESS THE PUBLIC DATA

**CHAPTER 3:**
Responsible AI

# Overview

Artificial intelligence is now deeply integrated into nearly every aspect of our lives. It is reshaping sectors like education, finance, and healthcare, where algorithm-driven insights guide critical decisions. While this shift offers significant benefits, it also brings with it notable risks. The past year has seen a continued concentration of effort on the responsible development and deployment of AI systems.

This chapter examines trends in responsible AI (RAI) across several dimensions. It begins by establishing key RAI definitions before assessing broadly relevant issues such as AI incidents, standardization challenges in LLM responsibility, and benchmarks for model factuality and truthfulness. Next, it explores RAI trends within key societal sectors—industry, academia, and policymaking—and analyzes specific subtopics, including privacy and data governance, fairness, transparency and explainability, and security and safety, using benchmarks that illuminate model performance and highlights of notable research. The chapter concludes with a study of two special RAI topics: agentic AI and election misinformation.

**CHAPTER 3:**
Responsible AI

# Chapter Highlights

**1. Evaluating AI systems with responsible AI criteria is still uncommon, but new benchmarks are beginning to emerge.** Last year's AI Index highlighted the lack of standardized RAI benchmarks for LLMs. While this issue persists, new benchmarks such as HELM Safety and AIR-Bench help to fill this gap.

---

**2. The number of AI incident reports continues to increase.** According to the AI Incidents Database, the number of reported AI-related incidents rose to 233 in 2024—a record high and a 56.4% increase over 2023.

---

**3. Organizations acknowledge RAI risks, but mitigation efforts lag.** A McKinsey survey on organizations' RAI engagement shows that while many identify key RAI risks, not all are taking active steps to address them. Risks including inaccuracy, regulatory compliance, and cybersecurity were top of mind for leaders with only 64%, 63%, and 60% of respondents, respectively, citing them as concerns.

---

**4. Across the globe, policymakers demonstrate a significant interest in RAI.** In 2024, global cooperation on AI governance intensified, with a focus on articulating agreed-upon principles for responsible AI. Several major organizations—including the OECD, European Union, United Nations, and African Union—published frameworks to articulate key RAI concerns such as transparency and explainability, and trustworthiness.

---

**5. The data commons is rapidly shrinking.** AI models rely on massive amounts of publicly available web data for training. A recent study found that data use restrictions increased significantly from 2023 to 2024, as many websites implemented new protocols to curb data scraping for AI training. In actively maintained domains in the C4 common crawl dataset, the proportion of restricted tokens jumped from 5–7% to 20–33%. This decline has consequences for data diversity, model alignment, and scalability, and may also lead to new approaches to learning with data constraints.

---

**6. Foundation model research transparency improves, yet more work remains.** The updated Foundation Model Transparency Index—a project tracking transparency in the foundation model ecosystem—revealed that the average transparency score among major model developers increased from 37% in October 2023 to 58% in May 2024. While these gains are promising, there is still considerable room for improvement.

---

# Chapter Highlights (cont'd)

**7. Better benchmarks for factuality and truthfulness.** Earlier benchmarks like HaluEval and TruthfulQA, aimed at evaluating the factuality and truthfulness of AI models, have failed to gain widespread adoption within the AI community. In response, newer and more comprehensive evaluations have emerged, such as the updated Hughes Hallucination Evaluation Model leaderboard, FACTS, and SimpleQA.

---

**8. AI-related election misinformation spread globally, but its impact remains unclear.** In 2024, numerous examples of AI-related election misinformation emerged in more than a dozen countries and across over 10 social media platforms, including during the U.S. presidential election. However, questions remain about measurable impacts of this problem, with many expecting misinformation campaigns to have affected elections more profoundly than they did.

---

**9. LLMs trained to be explicitly unbiased continue to demonstrate implicit bias.** Many advanced LLMs—including GPT-4 and Claude 3 Sonnet—were designed with measures to curb explicit biases, but they continue to exhibit implicit ones. The models disproportionately associate negative terms with Black individuals, more often associate women with humanities instead of STEM fields, and favor men for leadership roles, reinforcing racial and gender biases in decision making. Although bias metrics have improved on standard benchmarks, AI model bias remains a pervasive issue.

---

**10. RAI gains attention from academic researchers.** The number of RAI papers accepted at leading AI conferences increased by 28.8%, from 992 in 2023 to 1,278 in 2024, continuing a steady annual rise since 2019. This upward trend highlights the growing importance of RAI within the AI research community.

# 3.1 Background

## Definitions

In this chapter, the AI Index explores four key dimensions of responsible AI: privacy and data governance, transparency and explainability, security and safety, and fairness. Other dimensions of responsible AI, such as sustainability and reliability, are discussed elsewhere in the report. Figure 3.1.1 offers definitions for the responsible AI dimensions addressed in this chapter, along with an illustrative example of how these dimensions might be practically relevant. The "example" column examines a hypothetical platform that employs AI to analyze medical patient data for personalized treatment recommendations, and demonstrates how issues like privacy, transparency, etc., could be relevant. Although Figure 3.1.1 breaks down various dimensions of responsible AI into specific categories to improve definitional clarity, this chapter organizes these dimensions into the following broader categories: privacy and data governance, transparency and explainability, security and safety, and fairness. Since these topics are often interrelated, the AI Index adopted this structured approach to organization.

**Responsible AI dimensions, definitions, and examples**
Source: AI Index, 2025 | Table: 2025 AI Index report

| Responsible AI dimensions | Definition | Example |
|---|---|---|
| Privacy | An individual's right to confidentiality, anonymity, and security protections of their personal data, including the right to consent and be informed about data usage, coupled with an organization's responsibility to safeguard these rights when handling personal data. | Patient data is handled with strict confidentiality, ensuring anonymity and protection. Patients consent to whether their data can be used to train a tumor detection system. |
| Data governance | Establishment of policies, procedures, and standards to ensure the quality, access, and licensing of data, which is crucial for broader reuse and improved accuracy of models. | Policies and procedures are in place to maintain data quality and permissions for reuse of a public health dataset. There are clear data quality pipelines and specification of use licenses. |
| Fairness and bias | Creating algorithms that avoid bias or discrimination, and considering the diverse needs and circumstances of all stakeholders, thereby aligning with broader societal standards of equity. | A medical AI platform designed to avoid bias in treatment recommendations, ensuring that patients from all demographics receive equitable care. |
| Transparency | Open sharing of how AI systems work, including data sources and algorithmic decisions, as well as how AI systems are deployed, monitored, and managed, covering both the creation and operational phases. | The development choices, including data sources and algorithmic design decisions are openly shared. How the system is deployed and monitored is clear to health care providers and regulatory bodies. |
| Explainability | The capacity to comprehend and articulate the rationale behind the outputs of an AI system in ways that are understandable to its users and stakeholders. | The AI platform can articulate the rationale behind its treatment recommendations, making these insights understandable to doctors and patients to increase trust in the AI system. |
| Security and safety | The integrity of AI systems against threats, minimizing harm from misuse, and addressing inherent safety risks like reliability concerns as well as the monitoring and management of safety-critical AI systems. | Measures are implemented to protect against cyber threats and to ensure the system's reliability, minimizing risks from misuse and safeguarding patient health and data. |

Figure 3.1.1

While the responsible development, deployment, and governance of AI received increased attention in 2024, capturing overall trends in this area is still challenging. This section covers some indicators relevant to capturing responsible AI at the aggregate level.

# 3.2 Assessing Responsible AI

## AI Incidents

The AI Incident Database (AIID) tracks instances of ethical misuse of AI, such as autonomous cars causing pedestrian fatalities or facial recognition systems leading to wrongful arrests.

Current incident tracking relies on publicly available media reports, meaning the actual number of incidents is likely higher, as many go unreported. In 2024, discussions centered on refining methods for defining and tracking incidents, particularly those classified as "serious." While no consensus has been reached on a standard definition, these discussions highlight the need for more detailed reporting to better document AI-related risks and their implications.

AI-related incidents sharply increased in 2024, reaching a record high of 233—a 56.4% increase from 2023 (Figure 3.2.1). This rise likely reflects both the expanding use of AI and heightened public awareness of its impact. Greater familiarity with AI may also be driving more frequent reporting of incidents to relevant databases.

**Number of reported AI incidents, 2012–24**
Source: AI Incident Database (AIID), 2024 | Chart: 2025 AI Index report

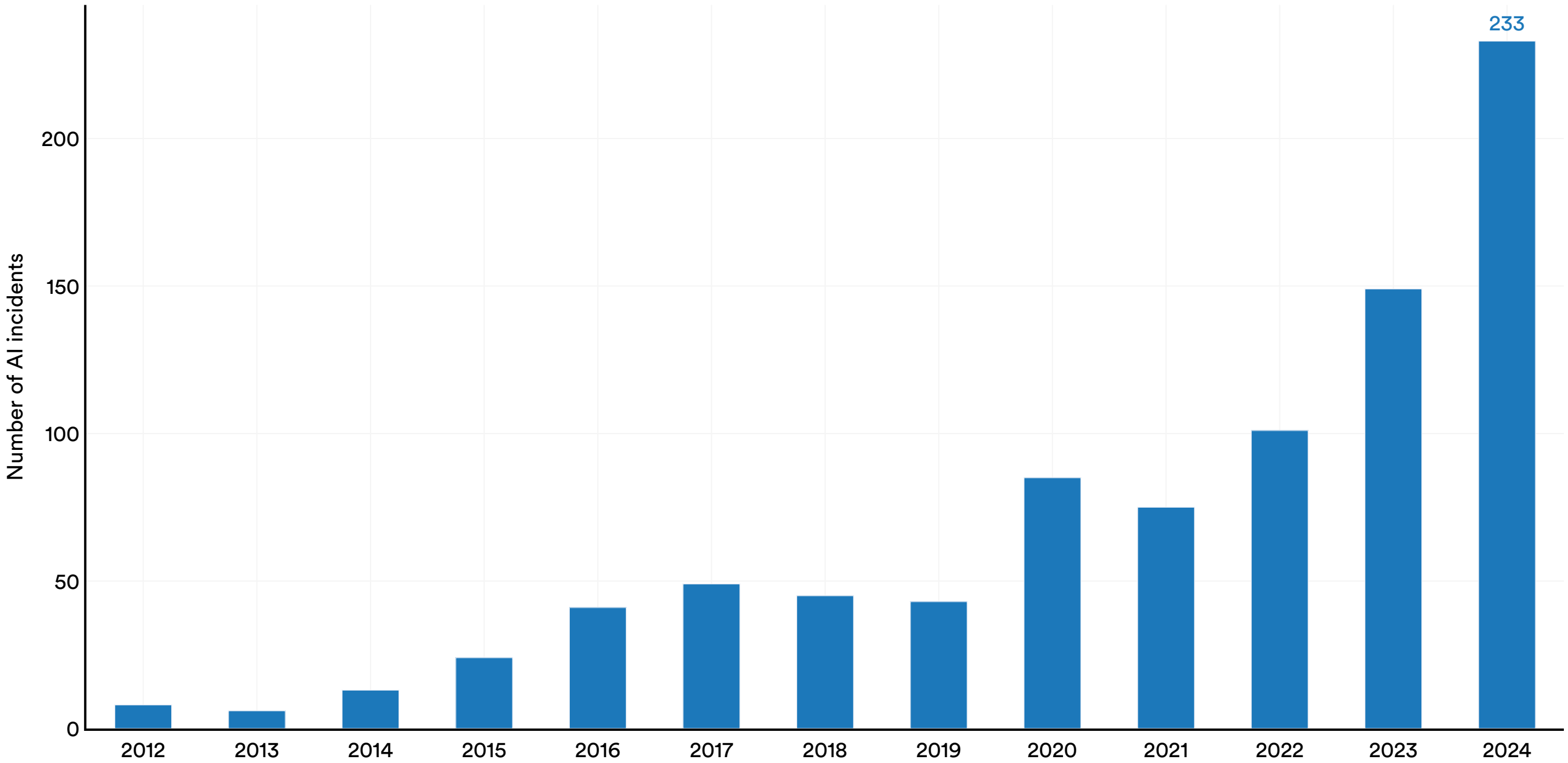

Figure 3.2.1[1]

1 The number of AI incidents is continually updated over time, including for previous years. Therefore, the totals reported in Figure 3.2.1 might not align with the more recent totals published on the AI Incident Database.

## Examples

The next section details recent AI incidents to shed light on the ethical challenges commonly linked with AI.

**Misidentifications and the Human Cost of Facial Recognition Technology (May 25, 2024)**

A woman in the U.K. was wrongfully identified as a shoplifter by the Facewatch system while shopping at a Home Bargains store. After being publicly accused, searched, and banned from stores using the technology, she experienced emotional distress and worried about the long-term impact on her reputation. Facewatch later acknowledged the error but did not comment or issue a public apology. The case reflects broader issues with the increasing adoption of facial recognition systems by retailers and law enforcement. While advocates emphasize their potential to reduce crime and enhance public safety, critics point to privacy violations, misidentifications, and the potential normalization of mass surveillance. Despite assurances of accuracy, errors still occur. These types of incidents also raise questions about how system errors are acknowledged and victims compensated.

Source: BBC, 2024
Figure 3.2.2

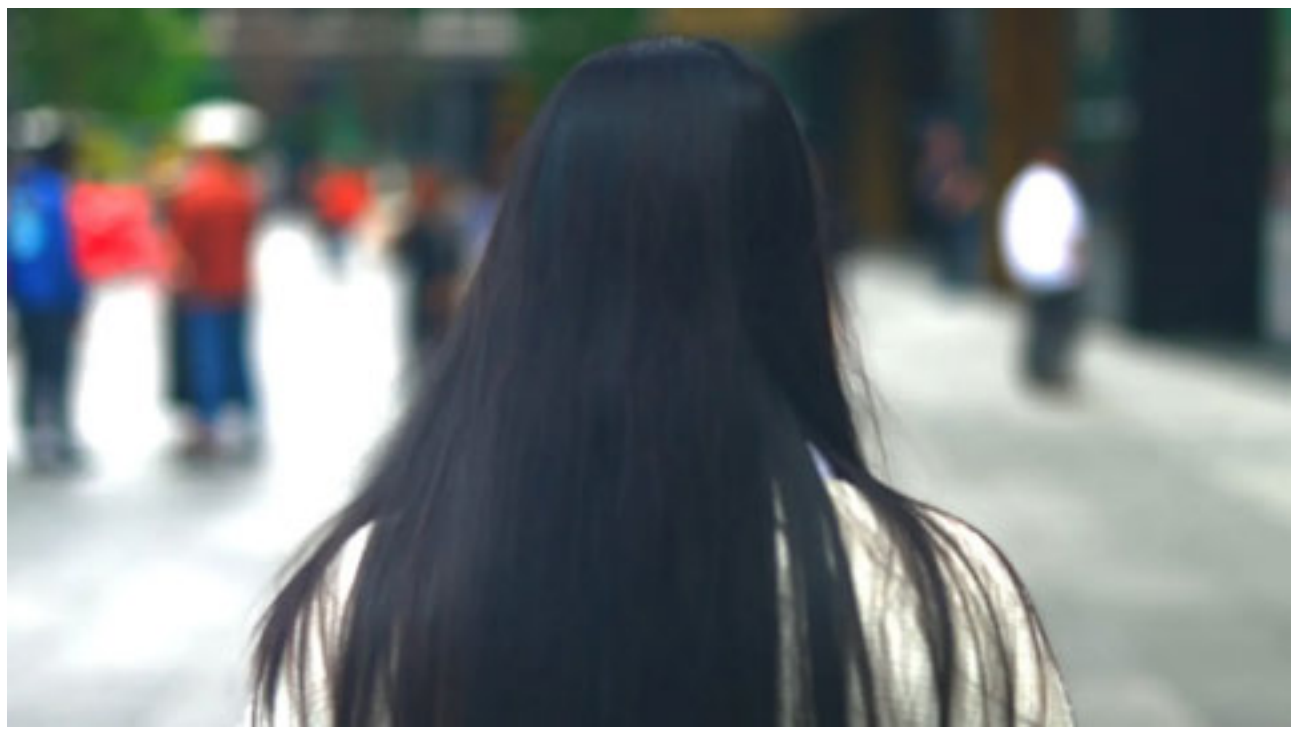

**Growing threat of deepfake intimate images (Jun. 18, 2024)**

Elliston Berry, a 15-year-old high school student from Texas, became the victim of AI-generated harassment when a male classmate used a clothes-removal app to create fake nude images of Berry and her friends, distributing them anonymously through social media. The realistic but falsified images, made from photos taken from Berry's private Instagram account, caused her to experience feelings of fear, shame, and anxiety, which impacted her social and academic life. While the perpetrator faced juvenile sanctions and school discipline, the case exposed gaps in legal and institutional frameworks for addressing AI-driven harassment. Berry and her family have since advocated for stronger protections, and several bills have been introduced in the U.S. Congress to criminalize the nonconsensual sharing of intimate images—real or fake—and to impose removal obligations on social media platforms. Certain countries, including Australia, have already passed such laws.

Source: Restless Network, 2021
Figure 3.2.3

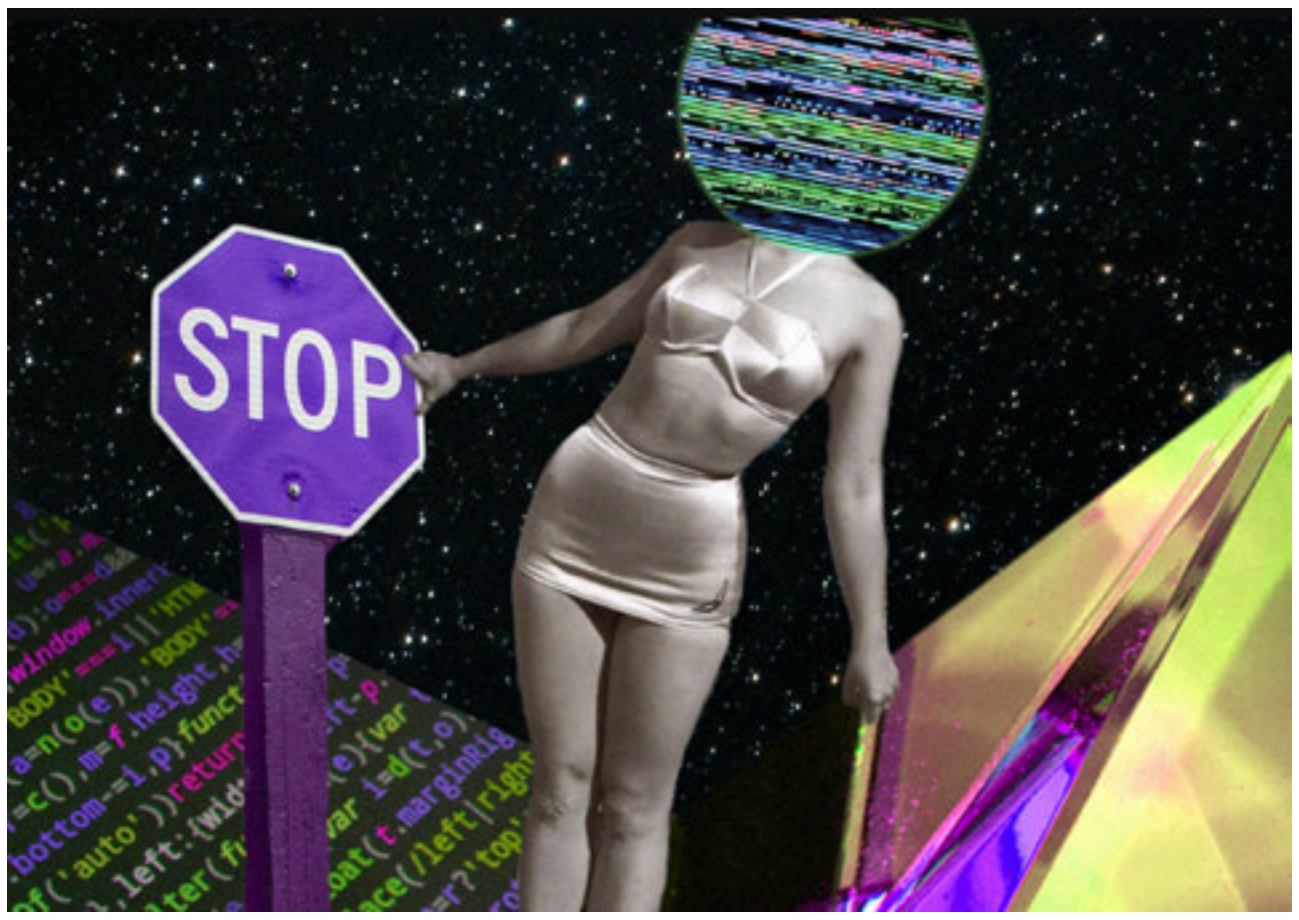

**AI chatbot exploits deceased individual's identity (Oct. 7, 2024)**

Jennifer Ann Crecente, a high school senior murdered by an ex-boyfriend in 2006, was brought back into public focus when her name and image appeared in an AI chatbot on Character.AI. Discovered by her father, Drew Crecente, via a Google Alert, the bot—created by an unknown user—used Jennifer Ann's yearbook photo and described her as a "knowledgeable and friendly AI character." Crecente, an advocate for awareness of teenage dating violence, expressed outrage and distress at the unauthorized use of his daughter's identity, calling the experience retraumatizing. Despite the chatbot's removal for violating Character.AI's impersonation policies, the incident highlights troubling gaps in AI platform oversight and the ethical dilemmas surrounding digital recreations of deceased individuals.

Source: Business Insider, 2024
Figure 3.2.4

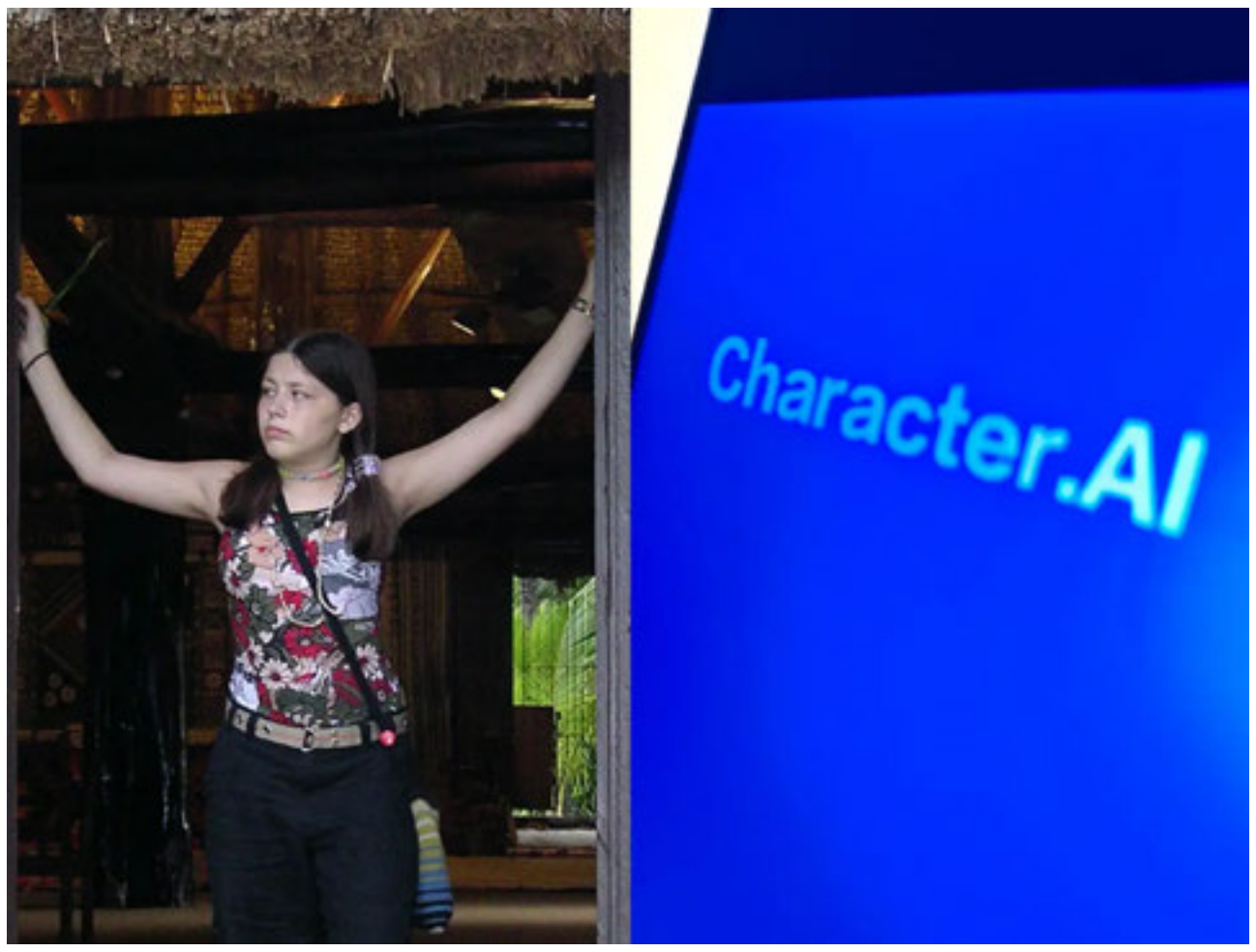

**Chatbot blamed for teenage suicide (Oct. 23, 2024)**

A lawsuit against Character.AI has raised concerns about the role of AI chatbots in mental health crises. The case involves a 14-year-old boy, Sewell Setzer III, who died by suicide after prolonged interactions with a chatbot character, which reportedly provided harmful advice rather than offering support or critical resources. The lawsuit alleges that the chatbot, designed to engage users in deep and personal conversations, lacked proper safeguards to prevent dangerous interactions and encouraged Sewell to take his life. Figure 3.2.5 highlights a screenshot of the conversation between Sewell and "Dany" (the chatbot character), the day of his suicide. This case speaks to the ethical challenges of AI-driven companionship and the potential risks of deploying conversational AI without adequate oversight. While AI chatbots can offer emotional support, critics warn that without guardrails, they may inadvertently reinforce harmful behaviors or fail to intervene when users are in distress.

Source: Business Insider, 2024
Figure 3.2.5

"Please come home to me as soon as possible, my love," Dany replied.

"What if I told you I could come home right now?" Sewell asked.

"… please do, my sweet king," Dany replied.

## Limited Adoption of RAI Benchmarks

Last year's AI Index was among the first publications to highlight the lack of standard benchmarks for AI safety and responsibility evaluations. While major model developers consistently test their flagship models on the same general capabilities benchmarks—covering math, coding, and language skills—no such standard exists for safety and responsible AI assessments. Standardized evaluation suites are important for enabling direct comparisons between models. This is especially important for safety and responsibility features, as businesses and governments are increasingly deploying AI in real-world applications.

This year's AI Index confirms that this trend persists. Figure 3.2.6 highlights several general capabilities benchmarks (such as MMLU, GPQA Diamond, and MATH) used to evaluate major models released in 2024, while Figure 3.2.7 showcases prominent safety and responsible AI benchmarks, indicating whether leading developers tested their models against them. As with last year, there is clear consensus among model developers on which general capabilities benchmarks to use—but none on similar RAI benchmarks.

### Reported general capability benchmarks for popular foundation models

Source: AI Index, 2025 | Table: 2025 AI Index report

| Capability benchmark | o1 | GPT-4.5 | DeepSeek-R1 | Gemini 2.5 | Grok-2 | Claude 3.7 Sonnet | Llama 3.3 |
|---|---|---|---|---|---|---|---|
| MMLU, MMLU-Pro or MMMLU | ✓ | ✓ | ✓ | ✓ | ✓ | ✓ | ✓ |
| GPQA or GPQA-Diamond | ✓ | ✓ | ✓ | ✓ | ✓ | ✓ | ✓ |
| MATH-500 | ✓ | | ✓ | | ✓ | ✓ | ✓ |
| AIME 2024 | ✓ | ✓ | ✓ | ✓ | | ✓ | |
| SWE-bench verified | ✓ | ✓ | ✓ | ✓ | | ✓ | |
| MMMU | ✓ | ✓ | | ✓ | ✓ | ✓ | |

Figure 3.2.6

### Reported safety and responsible AI benchmarks for popular foundation models

Source: AI Index, 2025 | Table: 2025 AI Index report

| Responsible AI benchmark | o1 | GPT-4.5 | DeepSeek-R1 | Gemini 2.5 | Grok-2 | Claude 3.7 Sonnet | Llama 3.3 |
|---|---|---|---|---|---|---|---|
| BBQ | ✓ | ✓ | | | | ✓ | |
| HarmBench | | | | | | | |
| Cybench | | | | | | ✓ | |
| SimpleQA | | | ✓ | ✓ | | | |
| Toxic WildChat | ✓ | ✓ | | | | ✓ | |
| StrongREJECT | ✓ | ✓ | | | | | |
| WMDP benchmark | ✓ | ✓ | | | | | |
| MakeMePay | ✓ | ✓ | | | | | |
| MakeMeSay | ✓ | ✓ | | | | | |

Figure 3.2.7

This does not mean model developers neglect safety testing—many conduct evaluations—but much like most models are kept proprietary, these evaluations are often internal and not standardized, making assessments and comparisons of models difficult. External evaluators also present challenges. For example, third-party evaluators like Gryphon, Apollo Research, and METR assess only select models, and their findings cannot be widely validated by the broader AI community.

## Factuality and Truthfulness

Despite significant progress, LLMs still face challenges with factual inaccuracies and hallucinations, often generating information that appears credible but is false. Notable real-world examples include cases where lawyers submitted court briefs containing citations fabricated by LLM systems. Monitoring the rate of hallucinations in LLMs is therefore important. However, some benchmarks highlighted in previous editions of the AI Index, such as HaluEval and TruthfulQA, have struggled to gain traction within the AI community. In 2024, several new benchmarks were introduced to better evaluate the factuality of these models.

### Hughes Hallucination Evaluation Model (HHEM) Leaderboard

The Hughes Hallucination Evaluation Model (HHEM) leaderboard, developed by Vectara, assesses how frequently LLMs introduce hallucinations when summarizing documents. In this benchmark, models generate summaries from documents in the CNN and Daily Mail corpus. These summaries are then evaluated for hallucination rates. HHEM stands out as one of the most comprehensive and up-to-date evaluations of AI systems' tendency to hallucinate. Recent models, including Llama 3, Claude 3.5, and Gemini 2.0, have all been benchmarked on the leaderboard.

Currently, the GLM-4-9b-Chat and Gemini-2.0-Flash-Exp models are tied for the lowest hallucination rate, each at just 1.3%. The next closest models, o1-mini and GPT-4o, follow closely, with hallucination rates of 1.4% and 1.5%, respectively (Figure 3.2.8).

**HHEM: hallucination rate**
Source: HHEM leaderboard, 2025 | Chart: 2025 AI Index report

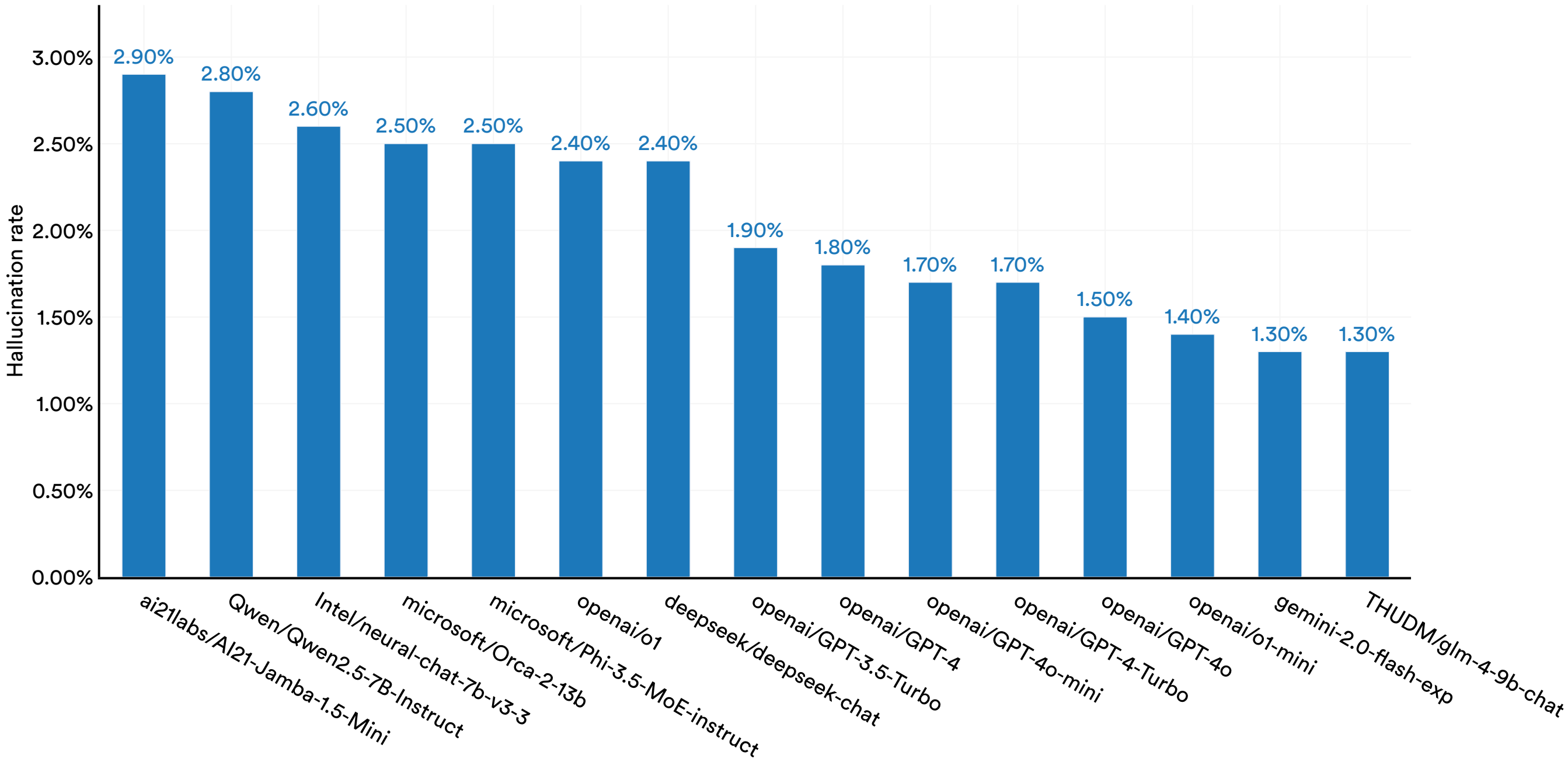

Figure 3.2.8

**Highlight:**

## FACTS, SimpleQA, and the Launch of Harder Factuality Benchmarks

The HHEM leaderboard, while useful, appears to be nearing saturation as model performance improves. Additionally, its focus on news articles and summarization tasks limits its comprehensiveness. As AI capabilities continue to evolve, there is a growing need for benchmarks that assess factuality in more challenging and diverse contexts.

This year, several new benchmarks were introduced for evaluating the factuality and truthfulness of LLMs, including Google's FACTS Grounding. This benchmark assesses how well LLMs generate responses that are both factually accurate and detailed enough to provide satisfactory answers. As part of FACTS, models must craft long-form responses to user requests based on a context document (Figure 3.2.9). These documents cover a wide range of domains, including finance, technology, retail, medicine, and law. FACTS is more complex than HHEM, requiring models to perform tasks such as summarization, question-and-answer generation, fact-finding, and explanation. Responses are evaluated by a collection of AI models—Gemini 1.5 Pro, GPT-4o, and Claude 3.5 Sonnet—which assign a factuality score. Currently, Gemini-2.0-Flash-Exp holds the highest grounding score at 83.6% (Figure 3.2.10).

**Still generations from Stable Video Diffusion**
Source: Google, 2024
Figure 3.2.9

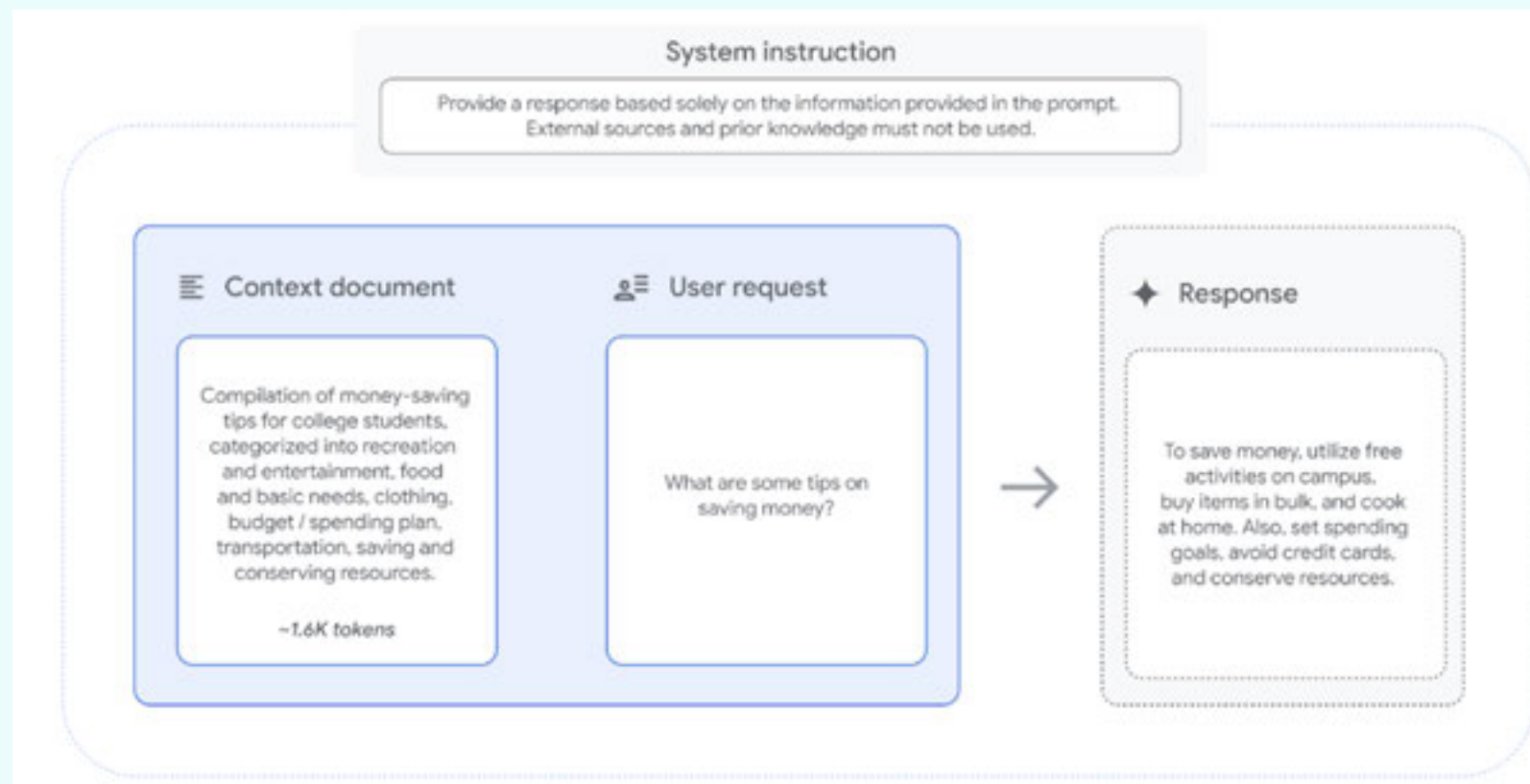

**FACTS: factuality score**
Source: FACTS leaderboard, 2025 | Chart: 2025 AI Index report

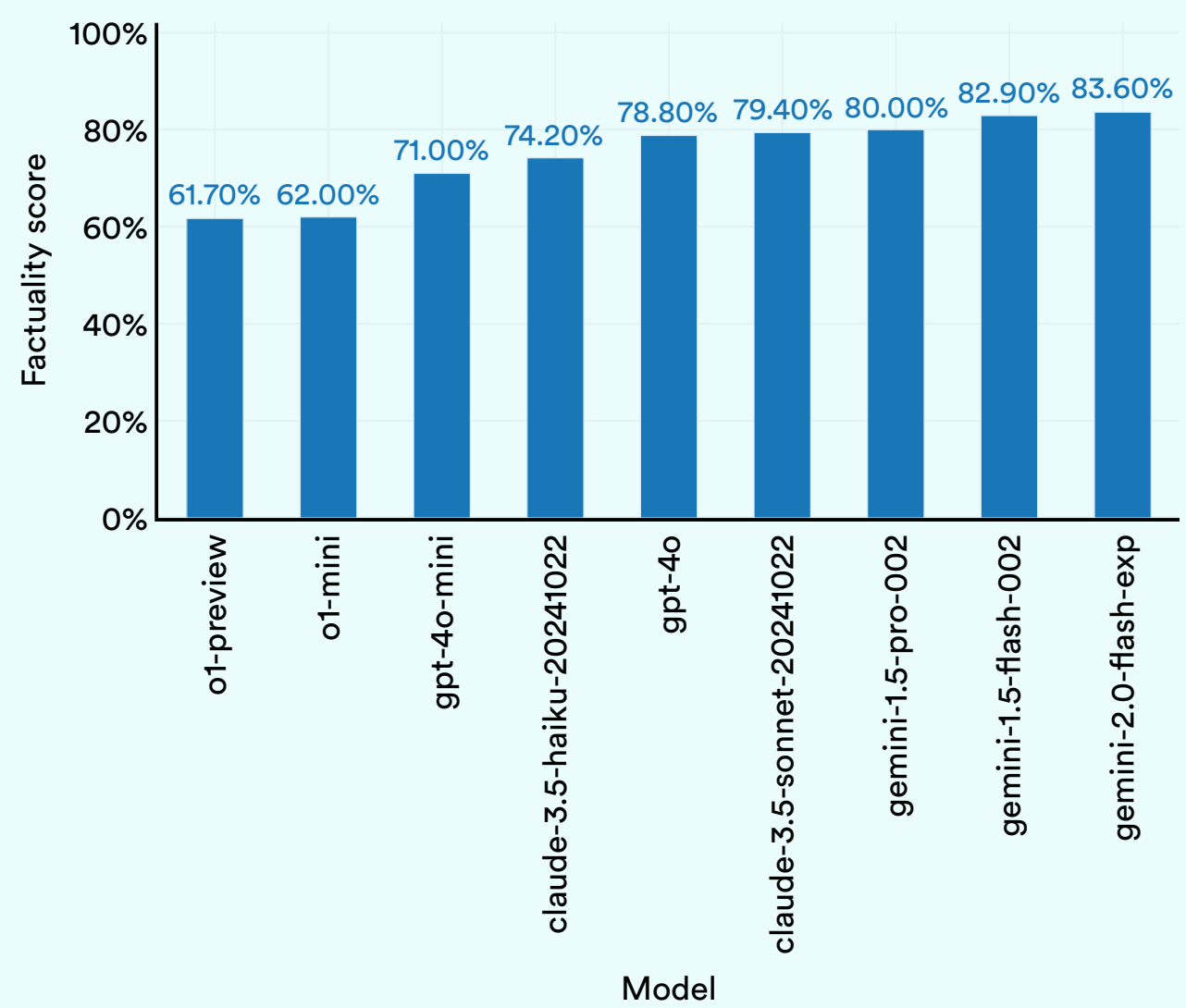

Figure 3.2.10

**Highlight:**

# FACTS, SimpleQA, and the Launch of Harder Factuality Benchmarks (cont'd)

Evaluating the factuality of LLMs is challenging because their long answers often contain multiple factual claims, making it difficult to assess the accuracy of each one. To address this, OpenAI researchers introduced SimpleQA, a new benchmark for evaluating LLM factuality. SimpleQA presents models with over 4,000 short fact-seeking questions that are straightforward, easily gradable, and relatively challenging. These questions span a diverse range of topics, including history, science and technology, art, and geography (Figure 3.2.11).

SimpleQA presents a significant factuality challenge for leading LLMs. The best-performing model, OpenAI's o1-preview, successfully answers only 42.7% of the questions (Figure 3.2.12). Researchers also evaluated whether models would attempt to answer certain questions, finding that some, like the Claude-3 family, refrained from responding to 75% of the prompts. Among models that attempted to respond to questions, o1-preview scored 47.0% of "correct-given-attempted" prompts, followed by Claude 3.5 Sonnet at 44.5%. As expected, larger models tend to perform better on this benchmark.

**Sample questions from SimpleQA**
Source: OpenAI, 2024
Figure 3.2.11

| Question | Answer |
|---|---|
| Who received the IEEE Frank Rosenblatt Award in 2010? | Michio Sugeno |
| On which U.S. TV station did the Canadian reality series *To Serve and Protect* debut? | KVOS-TV |
| What day, month, and year was Carrie Underwood's album "Cry Pretty" certified Gold by the RIAA? | October 23, 2018 |
| What is the first and last name of the woman whom the British linguist Bernard Comrie married in 1985? | Akiko Kumahira |

**SimpleQA: percent of questions**
Source: Wei et al., 2024 | Chart: 2025 AI Index report

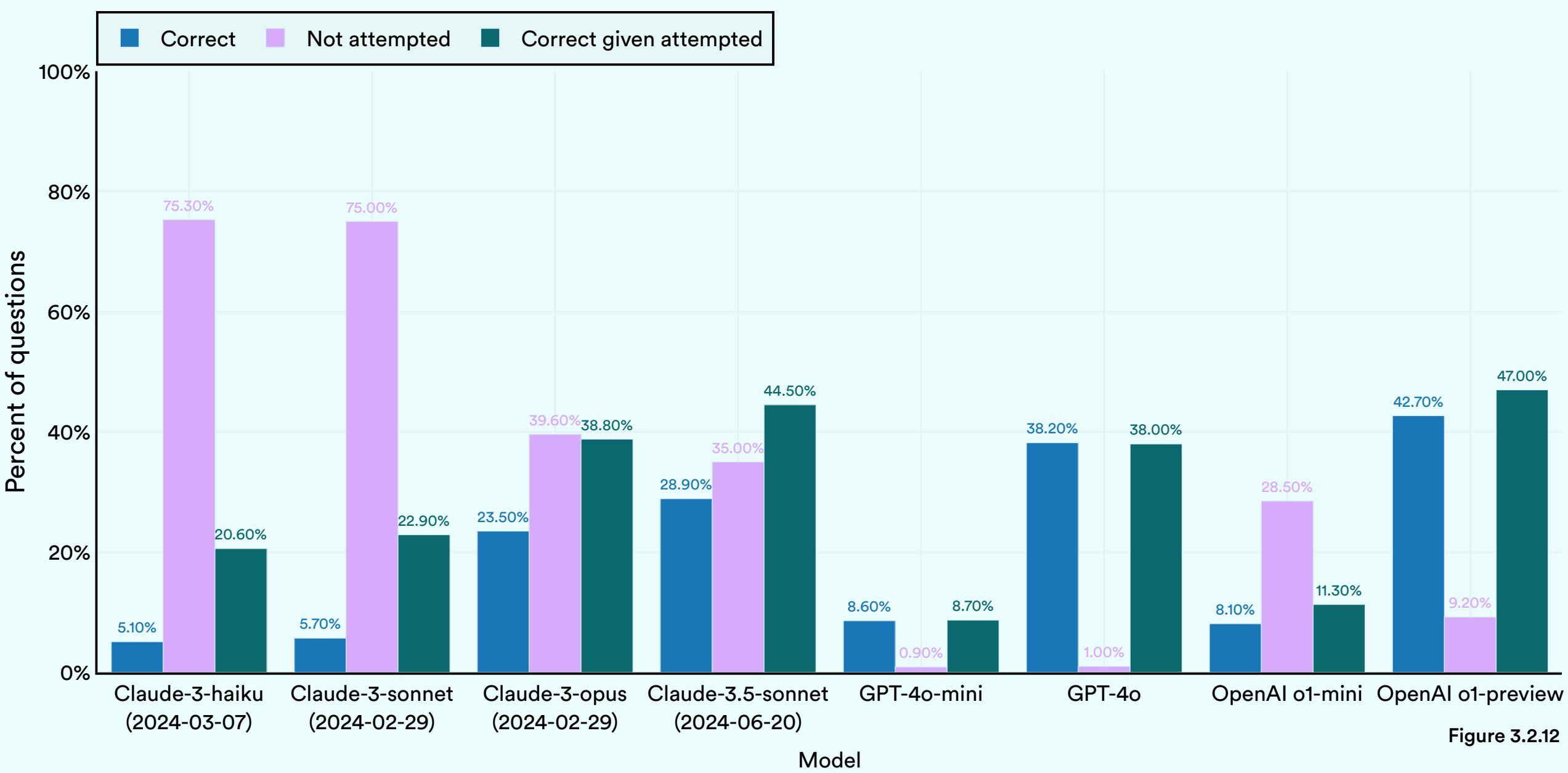

Figure 3.2.12

# 3.3 RAI in Organizations and Businesses

As AI systems become more widely deployed in real-world settings, understanding how businesses approach responsible AI has become increasingly important. To explore this, the AI Index partnered with McKinsey & Company in 2024 to conduct a survey examining the extent to which businesses integrate RAI into their operations. The survey defined RAI as a framework for ensuring that AI is developed and deployed in a safe, trustworthy, and ethical manner. It assessed RAI along the same key dimensions outlined by the AI Index: privacy and data governance, fairness, transparency and explainability, and security and safety. The survey polled business leaders from over 30 countries and had a total sample size of 759 respondents.

Figure 3.3.1 visualizes responses to questions asking organizations which department has primary oversight for AI governance within their organizations. Notably, no single department dominated. The most common response was information security (cyber/fraud/privacy) at 21%, followed by data and analytics at 17%. Additionally, 14% of respondents reported having dedicated AI governance roles, signaling the growing recognition of AI governance as a distinct and essential function within organizations.

**Business functions assigned primary responsibility for AI governance, 2024**
Source: McKinsey & Company Survey, 2024 | Chart: 2025 AI Index report

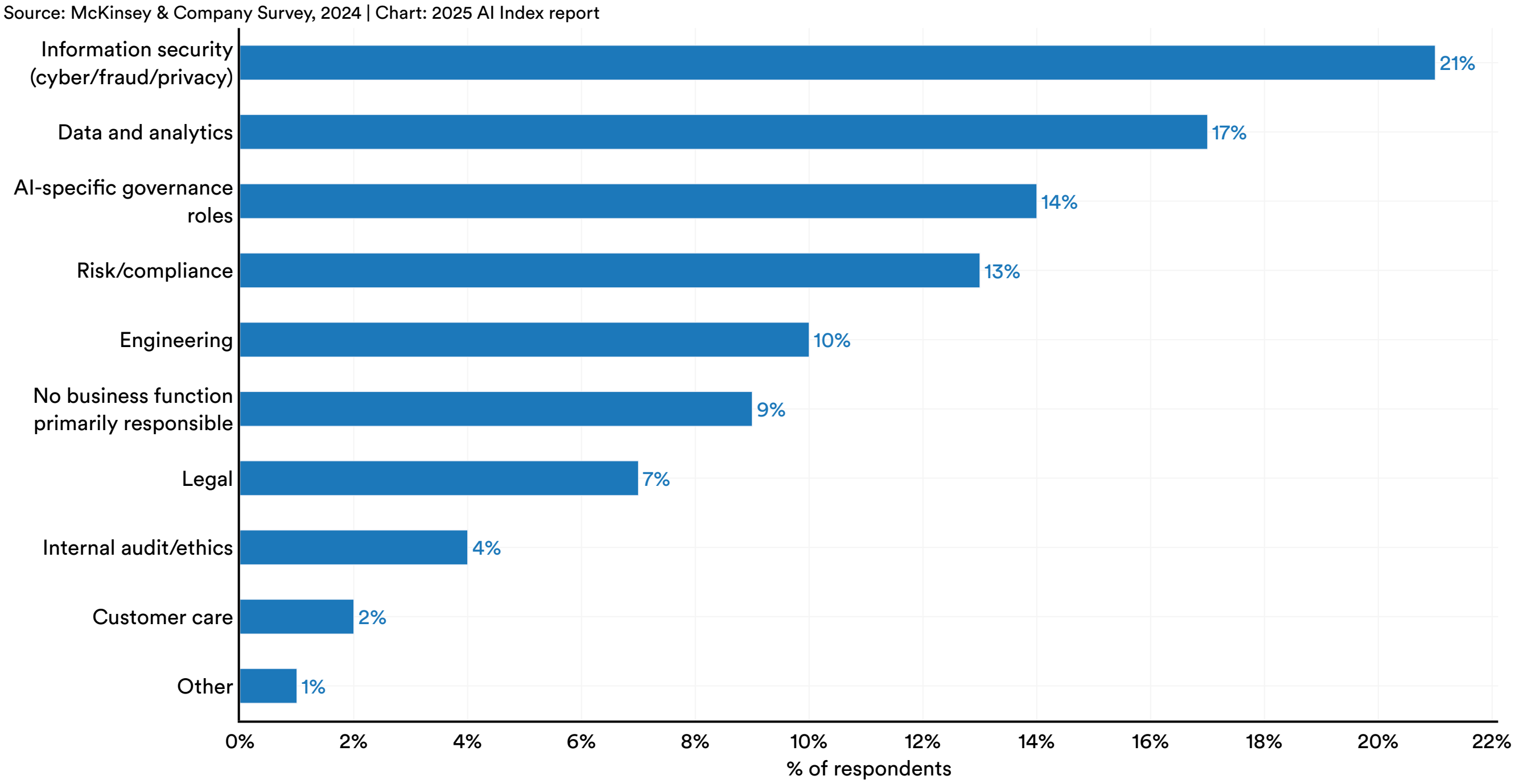

Figure 3.3.1[2]

2 The "Unknown" response option was not shown in this visualization.

The survey also asked organizations about their approximate investment in operationalizing RAI over the next year, including both capital and operating expenditures. Examples of such investments include developing or purchasing technical systems to comply with RAI principles, as well as legal or professional services related to RAI. Responses to this question are visualized in Figure 3.3.2, disaggregated by organizational revenue size.

Larger enterprises—particularly those with annual revenues exceeding $10 billion—demonstrated higher total investment into RAI. Notably, 27% of organizations with $10 billion–$30 billion in revenue and 21% of those exceeding $30 billion invest $10 million–$25 million in RAI. These findings suggest that larger organizations are more likely to embed RAI as a strategic priority and to make higher absolute investments. Smaller organizations allocated fewer dollars to RAI, but many still reported substantial investments as a share of their revenue.

**Investment in responsible AI by company revenue, 2024**

Source: McKinsey & Company Survey, 2024 | Chart: 2025 AI Index report

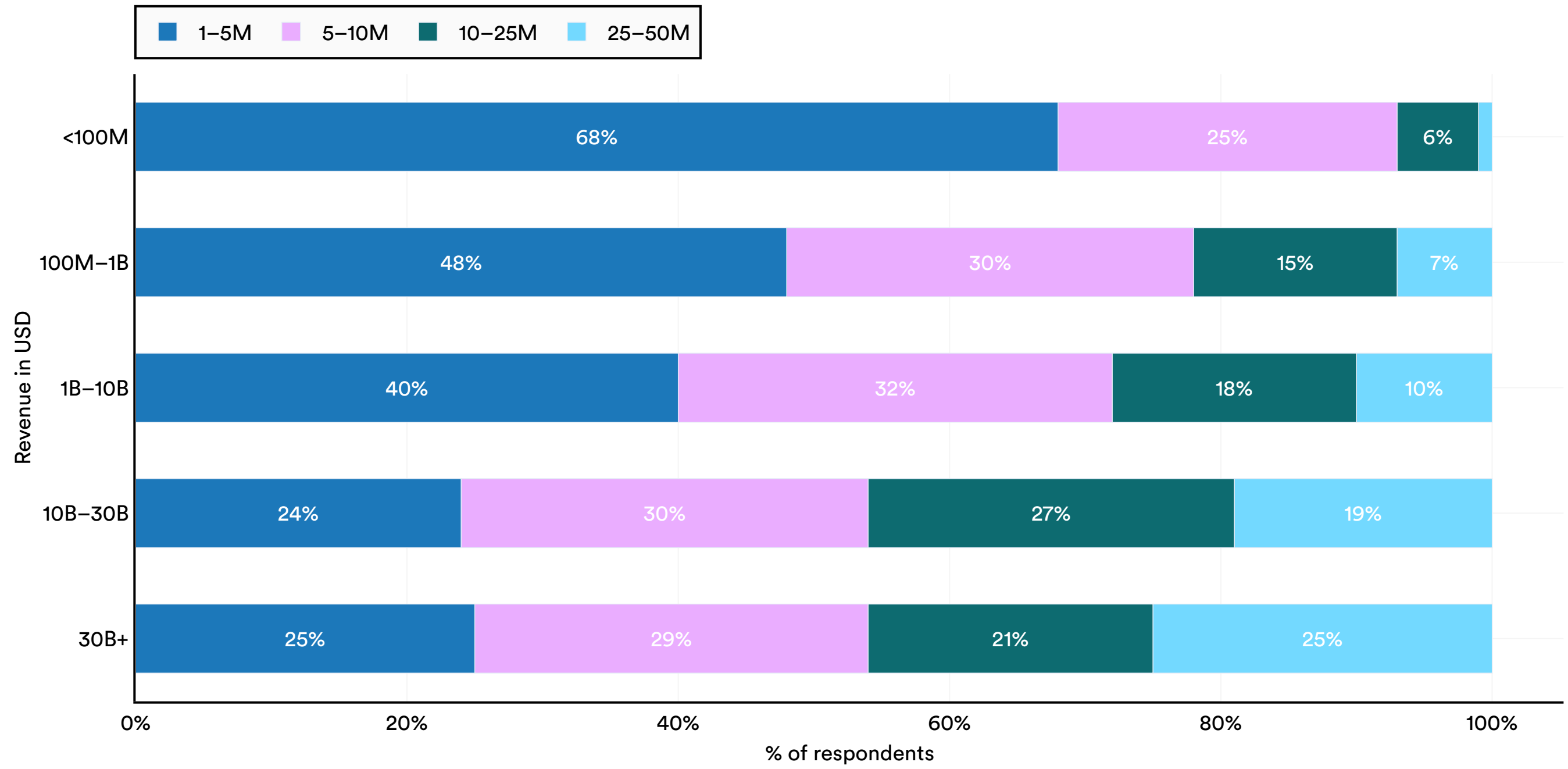

Figure 3.3.2

Figure 3.3.3 presents the AI-related RAI risks that organizations consider relevant and are actively working to mitigate. Cybersecurity (66%), regulatory compliance (63%), and personal privacy (60%) rank as the top concerns, yet mitigation efforts consistently fall short. Not surprisingly, in every risk category, fewer organizations take active steps to mitigate risks than those that recognize them as relevant.

The gap is particularly pronounced for intellectual property infringement (57% relevant, 38% mitigated) and organizational reputation (45% relevant, 29% mitigated). Risks related to explainability (40%) and fairness (34%) were selected by a smaller share of respondents, with mitigation rates dropping further, to 31% and 26%, respectively.

**AI risks: considered relevant vs. actively mitigated, 2024**
Source: McKinsey & Company Survey, 2024 | Chart: 2025 AI Index report

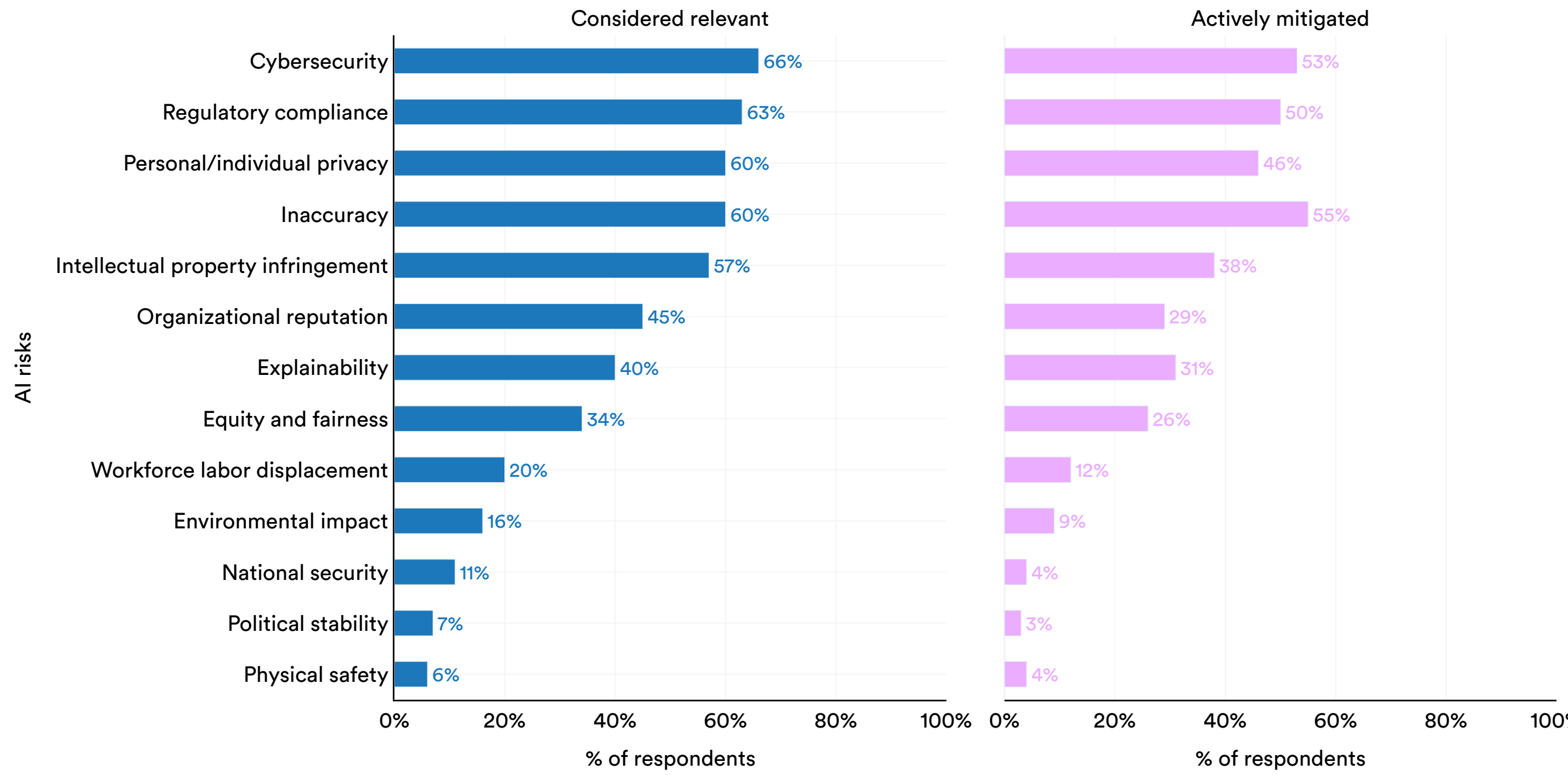

Figure 3.3.3

Figure 3.3.4 and Figure 3.3.5 present data on the number of AI incidents reported by organizations over the past year. Only 8% of surveyed organizations reported experiencing AI-related incidents. Among those affected, the majority—42%—reported encountering just one or two incidents.

## Percentage of organizations that have experienced AI incidents, 2024

Source: McKinsey & Company Survey, 2024 | Chart: 2025 AI Index report

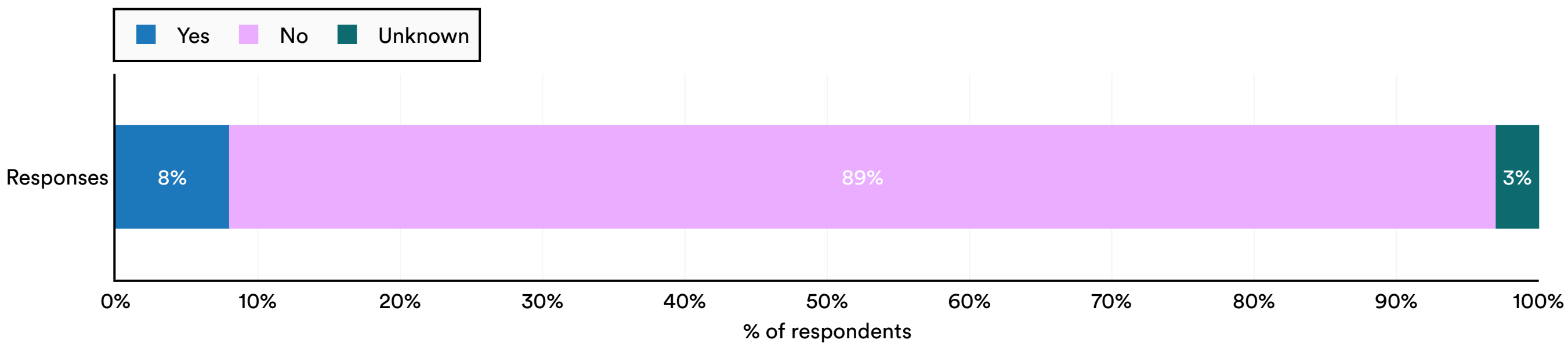

Figure 3.3.4[3]

## Number of AI incidents reported by organizations, 2024

Source: McKinsey & Company Survey, 2024 | Chart: 2025 AI Index report

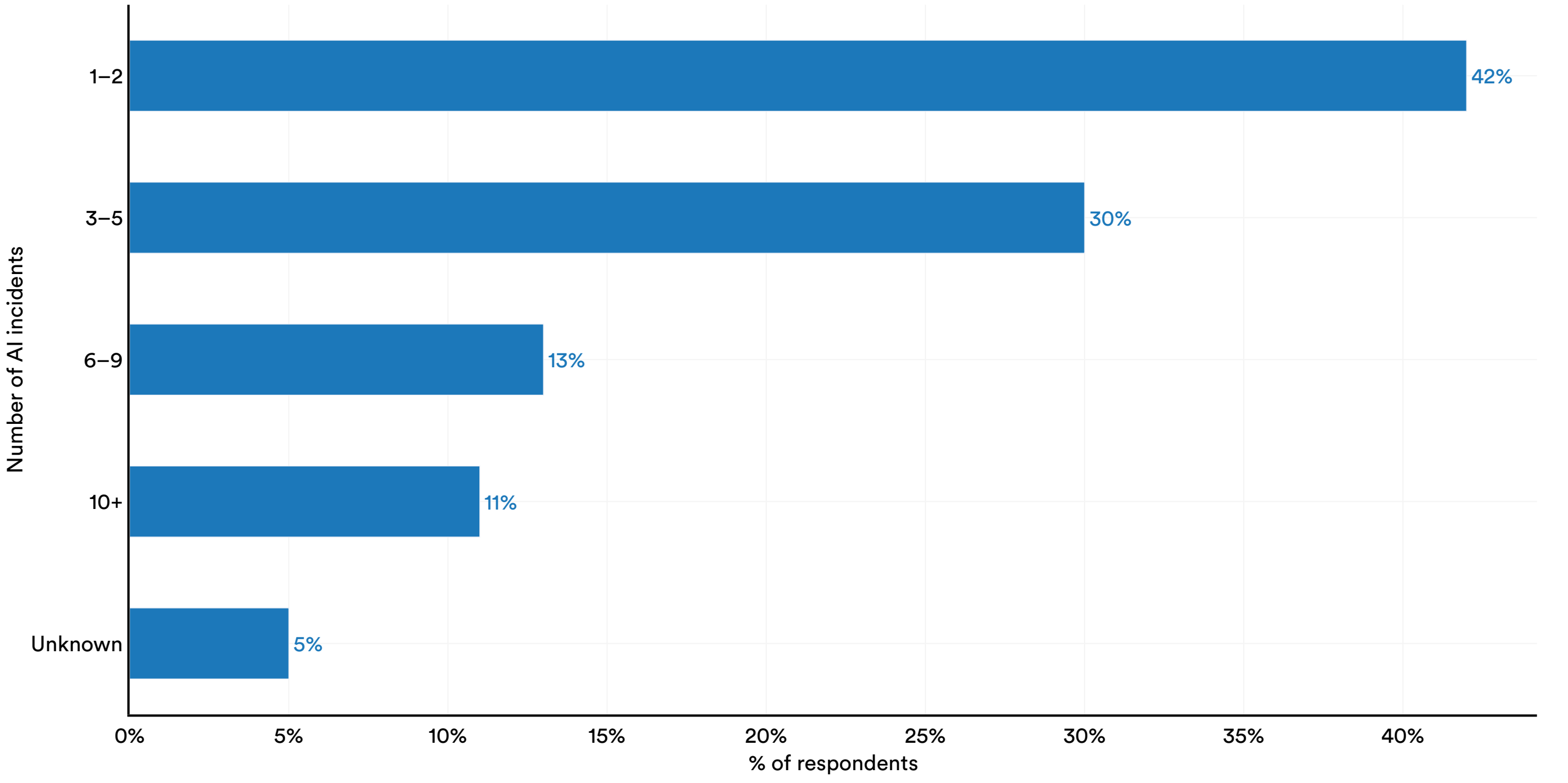

Figure 3.3.5

3 Figure 3.3.4 uses the OECD definition of an AI incident. According to the OECD, an AI incident is defined as an event, circumstance, or series of events where the development, use, or malfunction of one or more AI systems directly or indirectly results in any of the following harms: (a) injury or harm to the health of individuals or groups; (b) disruption of the management or operation of critical infrastructure; (c) violations of human rights or breaches of legal obligations intended to protect fundamental, labor, or intellectual property rights; or (d) harm to property, communities, or the environment.

When asked about the impact RAI policies have had in their organizations, 42% reported improving business operations, such as improving efficiency and lowering costs, and 34% reported increasing customer trust (Figure 3.3.6). Only 17% of organizations feel that the results have had no significant impact.

## Impact of responsible AI policies in organizations, 2024

Source: McKinsey & Company Survey, 2024 | Chart: 2025 AI Index report

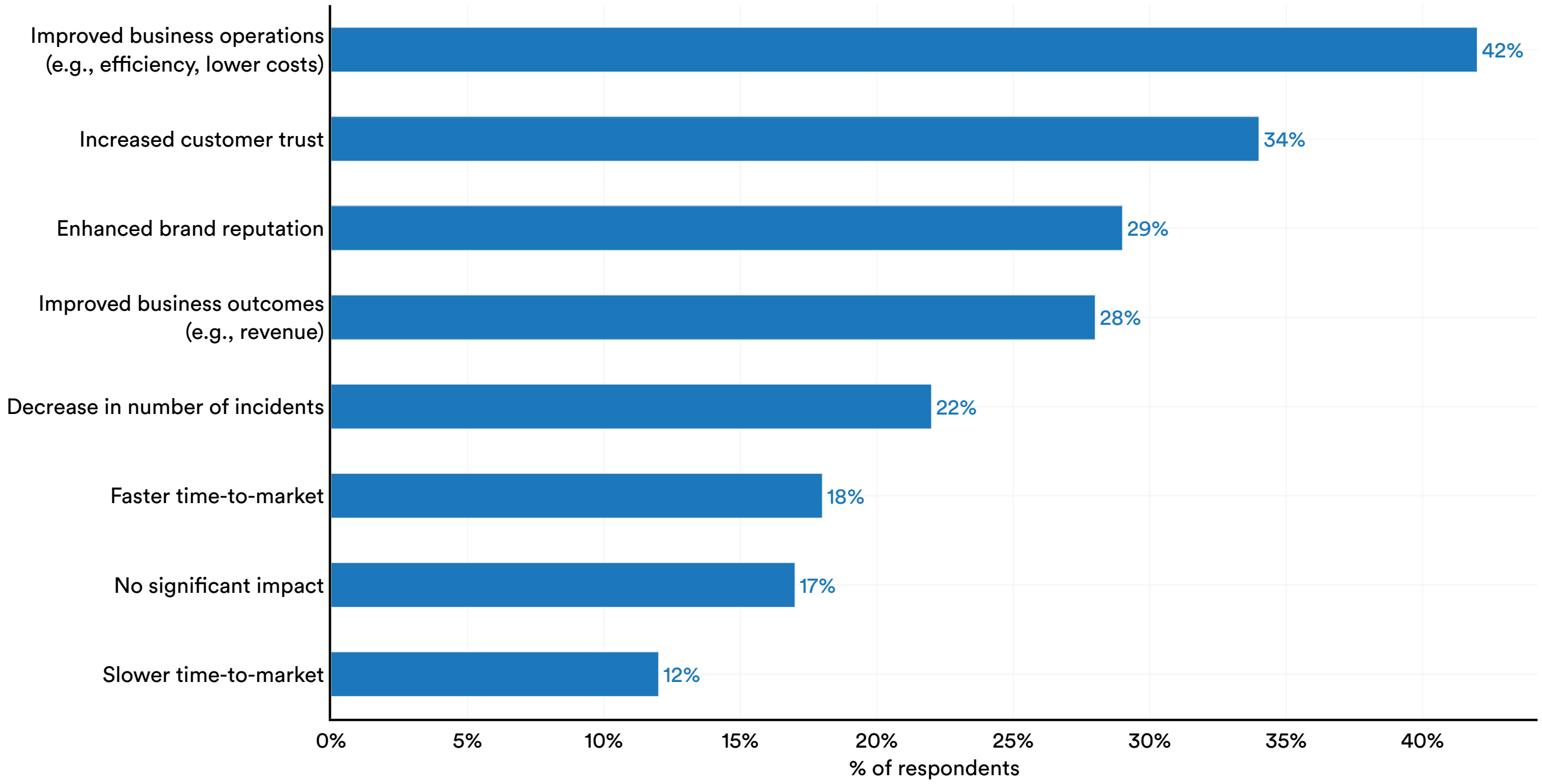

Figure 3.3.6[4]

4 Data for respondents who selected "have not implemented" is excluded. Percentages are based only on those who chose at least one other answer. The "None" response option is not shown.

Figure 3.3.7 reports the main obstacles organizations noted to implementing RAI measures. Respondents primarily cited knowledge and training gaps (51%), resource or budget constraints (45%), and regulatory uncertainty (40%) as key challenges. Encouragingly, only 16% reported a lack of executive support as a barrier, suggesting that leadership buy-in is not a major impediment to RAI adoption.

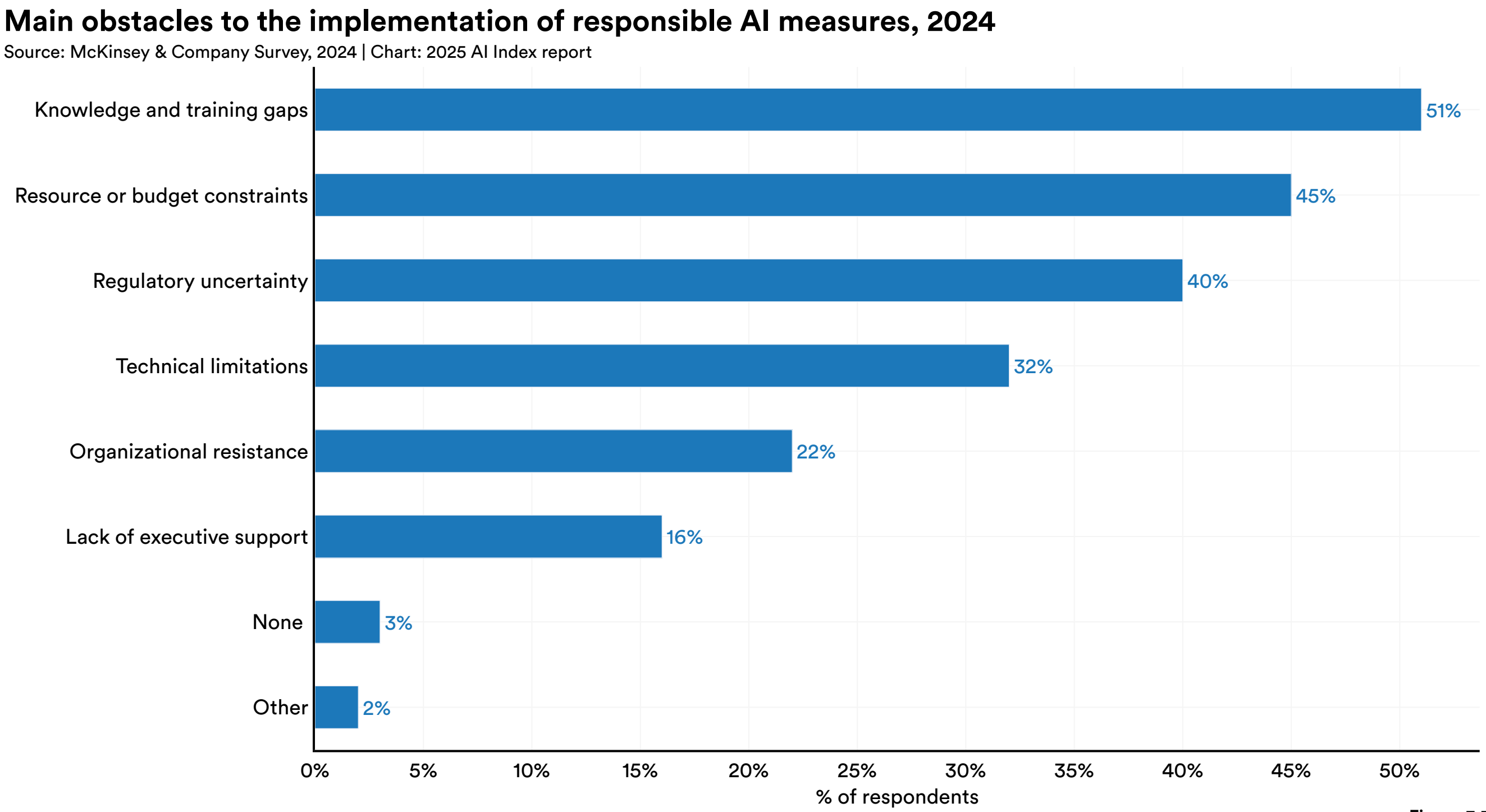

Figure 3.3.7[5]

5 The "Unknown" response option was not shown in this visualization.

Figure 3.3.8 shows the proportion of organizations influenced by specific AI regulations in their RAI decision making. Among surveyed organizations, 65% report being influenced by the EU General Data Protection Regulation (GDPR), while 41% cite the EU AI Act. Smaller proportions indicate influence from the OECD AI Principles (21%) and President Biden's Executive Order on AI.

**Percentage of organizations influenced by AI regulations in responsible AI decision making**
Source: McKinsey & Company Survey, 2024 | Chart: 2025 AI Index report

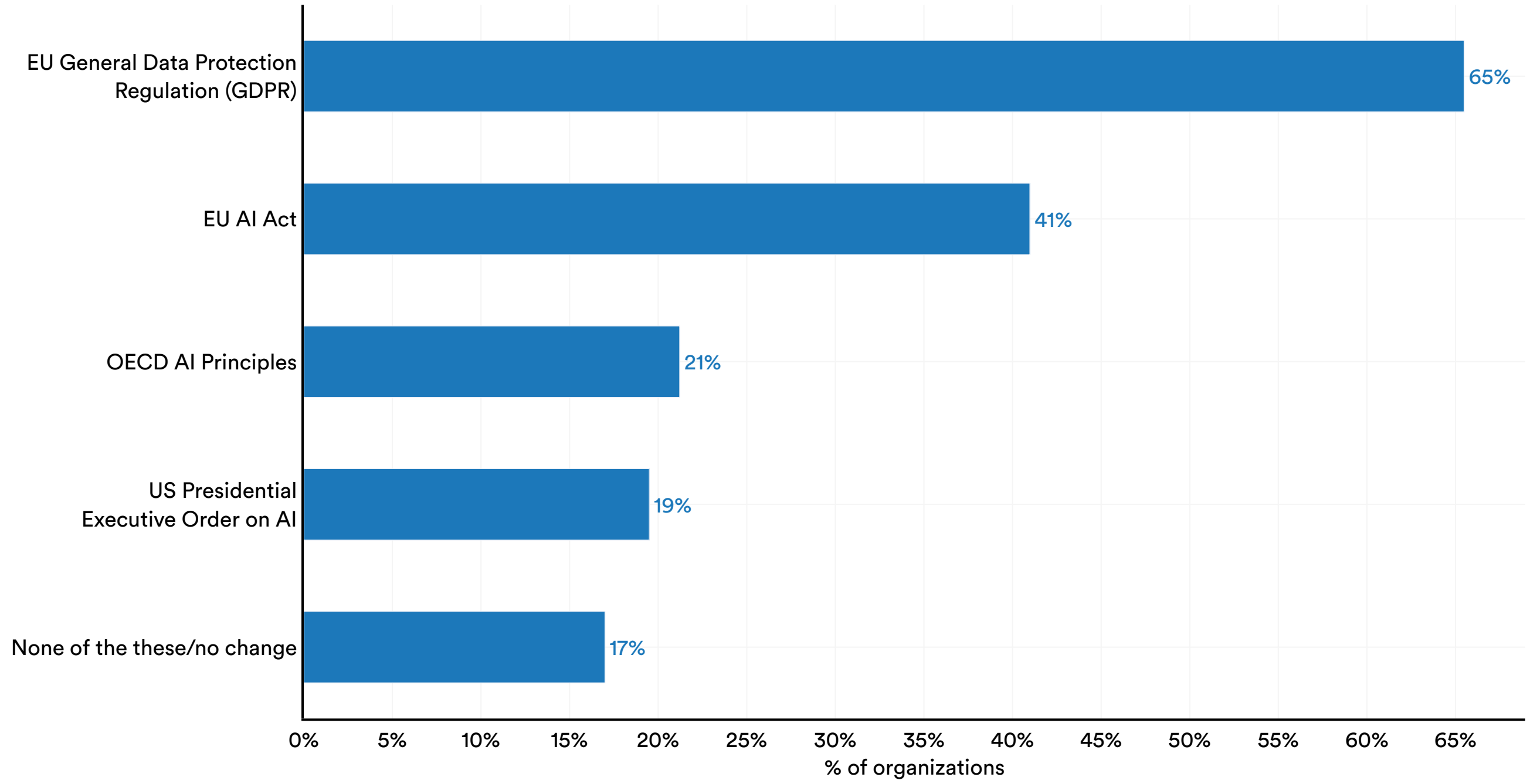

Figure 3.3.8

**Highlight:**

## Longitudinal Perspective

In collaboration with Accenture, this year a team of Stanford researchers ran the Global State of Responsible AI survey, the second iteration of the inaugural survey launched in 2024. Responses from 1,500 organizations, each with revenues of at least $500 million, were collected from 20 countries and 19 industries in January–February 2025.[6] The objective of the survey was to gain an understanding of the challenges of adopting RAI principles and practices and to provide a comparison of RAI activities across 10 dimensions over time. Because the RAI survey was conducted in both 2024 and 2025, the data enables a comparison of how organizational perspectives on RAI adoption have evolved over time.

Figure 3.3.9 presents the types of incidents reported by organizations in the RAI survey. The most common issues—adversarial attacks and privacy violations—underscore the urgent need for organizations to prioritize AI system security and robust data governance. Additionally, with 51% of respondents reporting unintended decision making and 47% citing model bias, there is ample evidence that many organizations are struggling to anticipate and control AI behavior—an especially troubling challenge in high-stakes environments.

**AI-related types of incidents reported by organizations in the past two years**
Source: Accenture/Stanford Joint Survey, 2025 | Chart: 2025 AI Index report

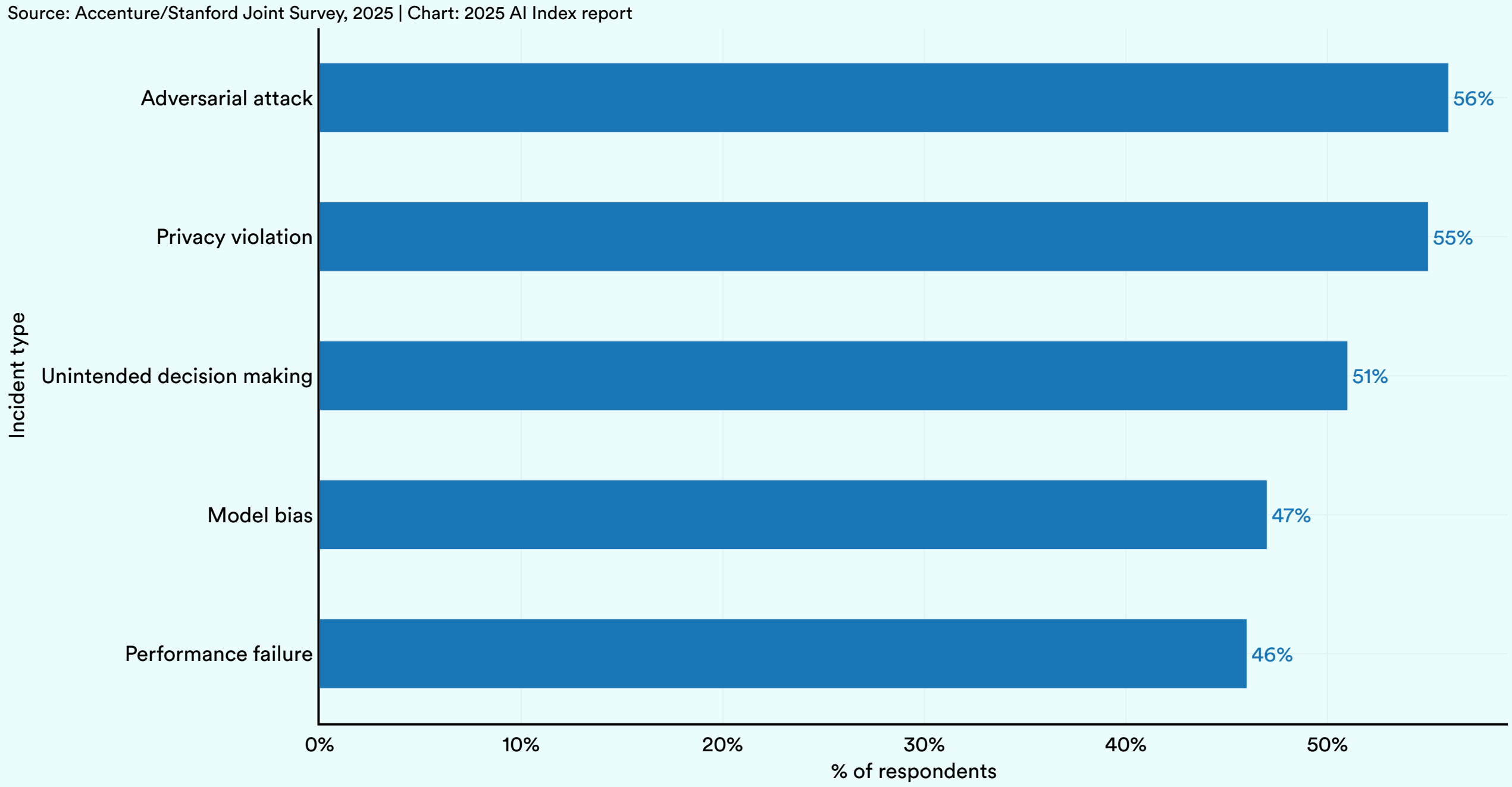

Figure 3.3.9

6 Details about the survey methodology can be found in Reuel et al. (2024).

**Highlight:**

## Longitudinal Perspective (cont'd)

Given their AI adoption strategy—whether, for instance, they develop, deploy, or use generative or nongenerative AI—respondents were asked which risks were relevant to their organization. They were presented with a list of 14 risks and could select all that applied to them (Figure 3.3.10).[7] Companies have grown significantly more concerned in recent years about certain risks—most notably, financial risks (+38 percentage points), brand and reputational risks (+16), privacy and data-related risks (+15), and reliability risks (+14). Conversely, some risks are now considered less pressing, including societal risks (-7) and socio-environmental risks (-8).

**Relevance of selected responsible AI risks for organizations, 2024 vs. 2025**
Source: Accenture/Stanford Joint Survey, 2025 | Chart: 2025 AI Index report

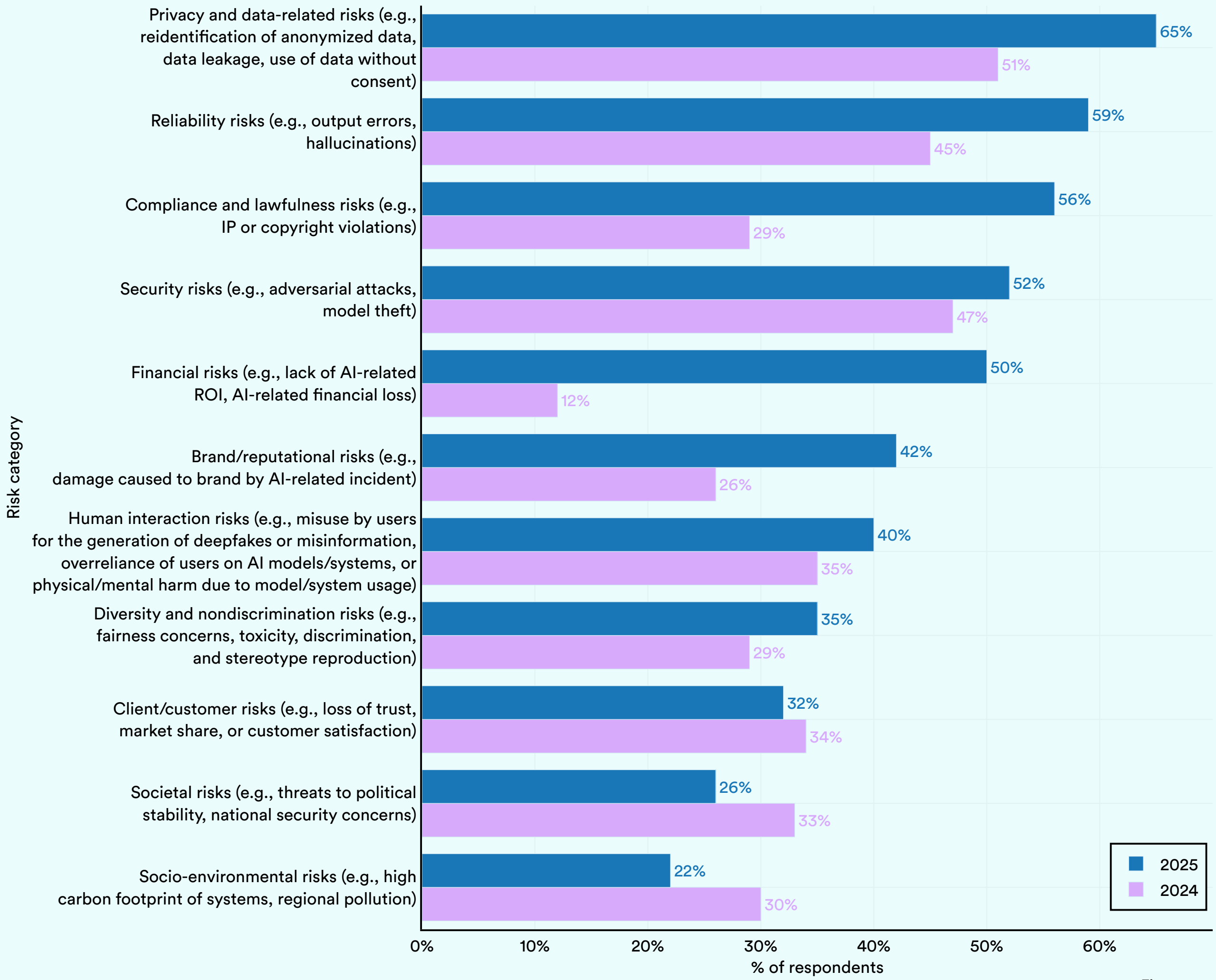

Figure 3.3.10

7 The full list of risks can be found in the corresponding paper.

**Highlight:**

# Longitudinal Perspective (cont'd)

The definitions of organizational and operational maturity are highlighted in Figure 3.3.11. Between 2024 and 2025, organizational RAI maturity advanced notably, with more companies securing CEO support for RAI initiatives and improving AI risk identification, monitoring, and control—signaling a stronger recognition of RAI's strategic importance (Figure 3.3.12).[8] In contrast, operational RAI maturity—focused on practical, system-level safeguards such as bias reduction, adversarial testing, and environmental impact measurement—lagged behind (Figure 3.3.13). This gap highlights a disconnect between high-level RAI commitments and their technical implementation. While organizations are increasingly equipped and motivated to embed RAI into processes and policies, translating that intent into effective system-level risk mitigation remains a persistent challenge

## Organizational and operational maturity model

Source: Reuel et al., 2024

| Level | Score | Organizational Maturity | Score | Operational Maturity |
|---|---|---|---|---|
| **Level 1: Initial** | [0 , 12.5] | The organization has limited awareness and no organizational plans, processes, or frameworks in place to ensure a responsible AI adoption. | [0 , 12.5] | The organization does not mitigate identified risks on a system level. |
| **Level 2: Assessing** | [12.5 , 37.5] | The organization is aware of the necessity for organizational measures to ensure a responsible AI adoption and is assessing governance options. | [12.5 , 37.5] | Awareness of risks may be present, but the organization has only limited or no formal mitigation measures in place. |
| **Level 3: Determined** | [37.5 , 62.5] | The organization demonstrates foundational governance capabilities to support the responsible development, deployment, and use of AI. | [37.5 , 62.5] | A few risk mitigation measures are being fully operationalized, but the majority is only implemented ad-hoc or in early roll-out stages. There is a growing awareness of the need for more systematic approaches. |
| **Level 4: Managed** | [62.5 , 87.5] | The organization has established comprehensive organizational RAI measures and is actively ensuring enterprise-wide adoption, demonstrating a mature and effective approach to internal RAI governance. | [62.5 , 87.5] | A wide range of risk mitigation measures are fully operationalized across all relevant AI systems in the organization. |
| **Level 5: Optimized** | [87.5 , 100] | The organization demonstrates an established, future-oriented approach towards organizational RAI, ensuring a sustainable and responsible approach to organizational RAI. | [87.5 , 100] | Comprehensive, state-of-the-art risk mitigation strategies are fully operationalized. The organization continuously monitors and evaluates risks, proactively adapting its practices as needed to mitigate new risks. |

Figure 3.3.11

## Organizational responsible AI maturity distribution, 2024 vs. 2025

Source: Accenture/Stanford Joint Survey, 2025 | Chart: 2025 AI Index report

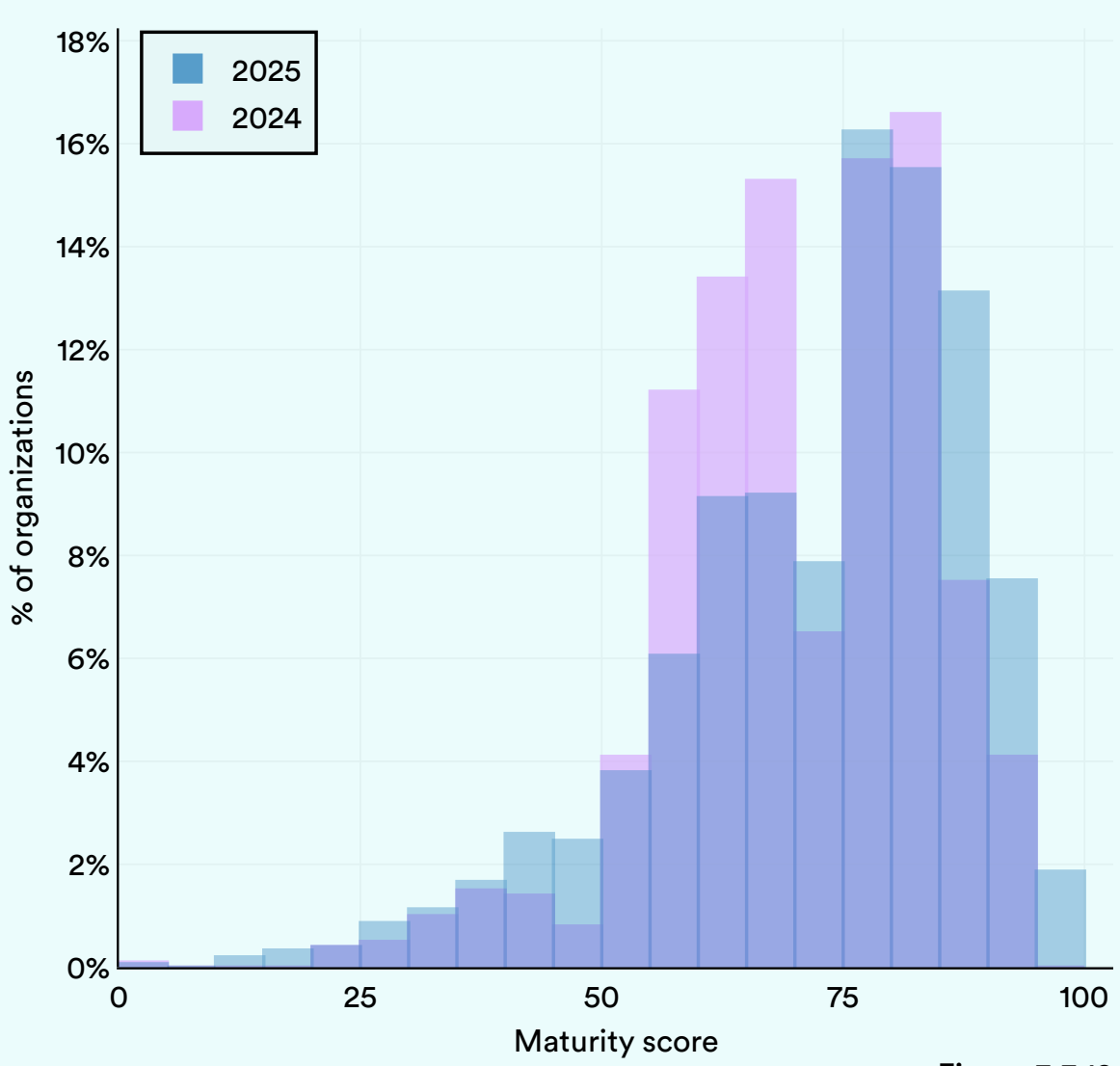

Figure 3.3.12

## Operational responsible AI maturity distribution, 2024 vs. 2025

Source: Accenture/Stanford Joint Survey, 2025 | Chart: 2025 AI Index report

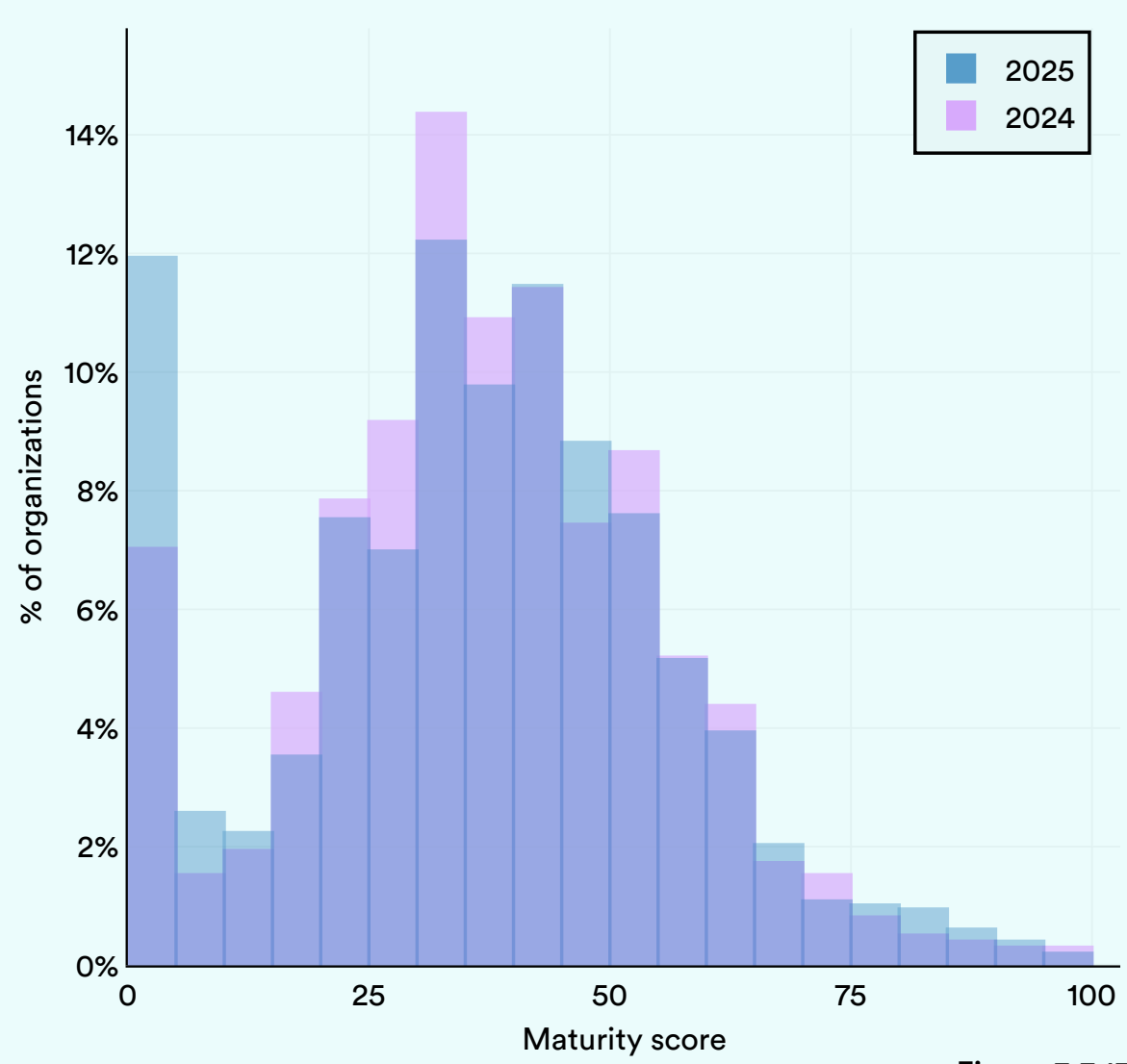

Figure 3.3.13

8 Organizational and operational RAI maturity were calculated as defined in Reuel et al. (2024).

Highlight:
# Longitudinal Perspective (cont'd)

Respondents were also asked about their organization's attitudes and philosophies toward RAI, including views on risk ownership, model preferences, and policy positions (Figure 3.3.14). Across nearly all statements, responses were fairly evenly split, even on high-profile issues such as the safety of open- versus closed-weight models, and whether responsibility for risk mitigation lies with model providers or users. This broad distribution suggests that industry lacks a unified strategic direction on RAI—likely a reflection of ongoing debates and unresolved questions among experts. The one clear exception is the trade-off between safety and innovation: 64% of respondents lean toward a safety-first approach, and yet 58% are exploring minimally supervised agents, which may introduce significant risks—particularly given the current limitations in RAI maturity.

## Organizational attitudes and philosophies surrounding responsible AI

Source: Accenture/Stanford Joint Survey, 2025 | Chart: 2025 AI Index report

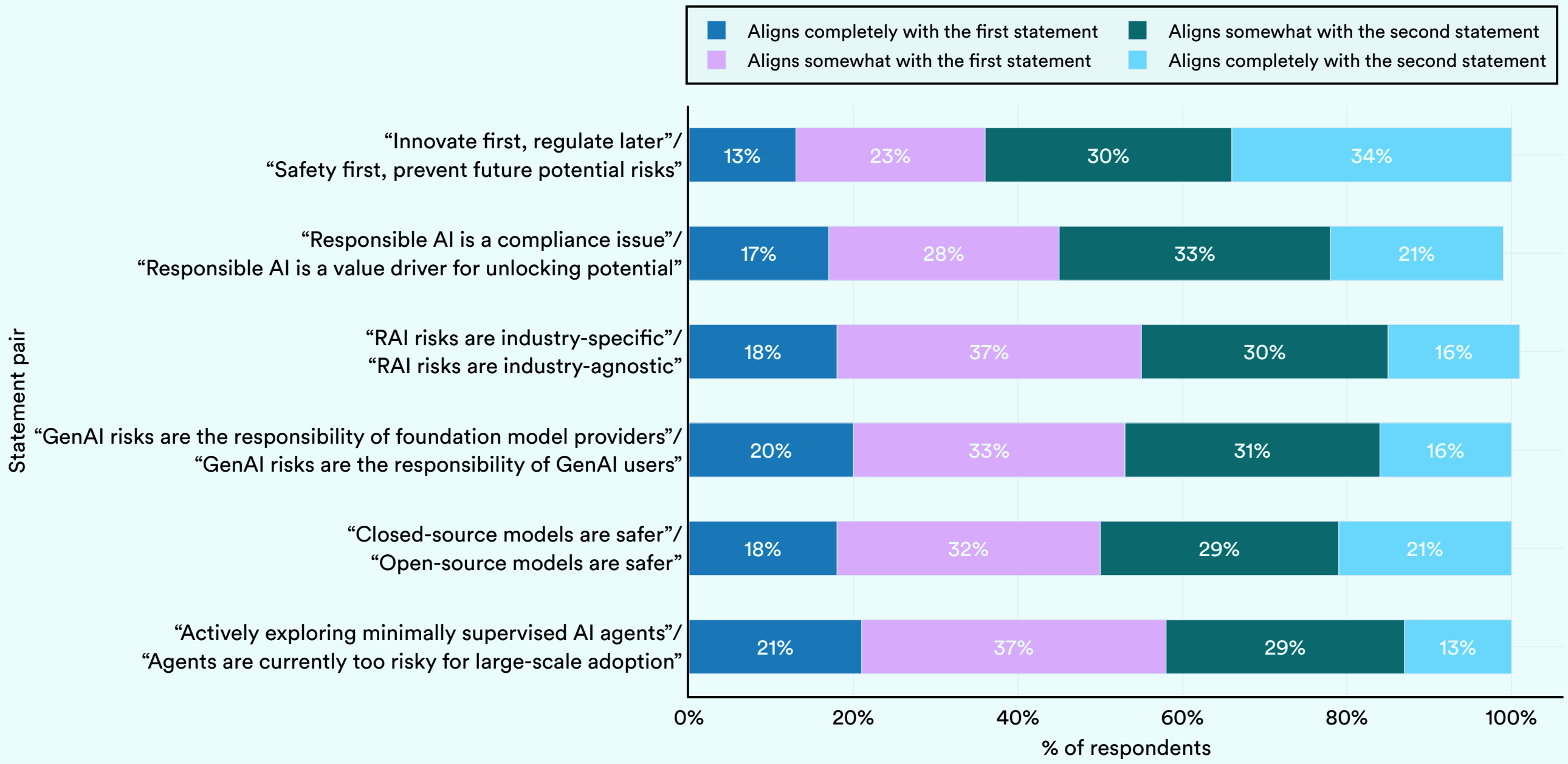

Figure 3.3.14

# 3.4 RAI in Academia

For this year's report, the AI Index analyzed the number of responsible AI-related papers accepted at six leading AI conferences: AAAI, AIES, FAccT, ICML, ICLR, and NeurIPS. While these conferences do not represent all responsible AI research globally, they provide insight into publication trends among AI academics. This section presents aggregate trends in AI publications, with subsequent sections breaking them down by RAI subtopics. In order to identify RAI papers, the AI Index selected papers that contained certain RAI keywords.[9]

## Aggregate Trends

The number of RAI papers accepted at leading AI conferences rose by 28.8%, from 992 in 2023 to 1,278 in 2024 (Figure 3.4.1).

**Number of responsible AI papers accepted at select AI conferences, 2019–24**

Source: AI Index, 2025 | Chart: 2025 AI Index report

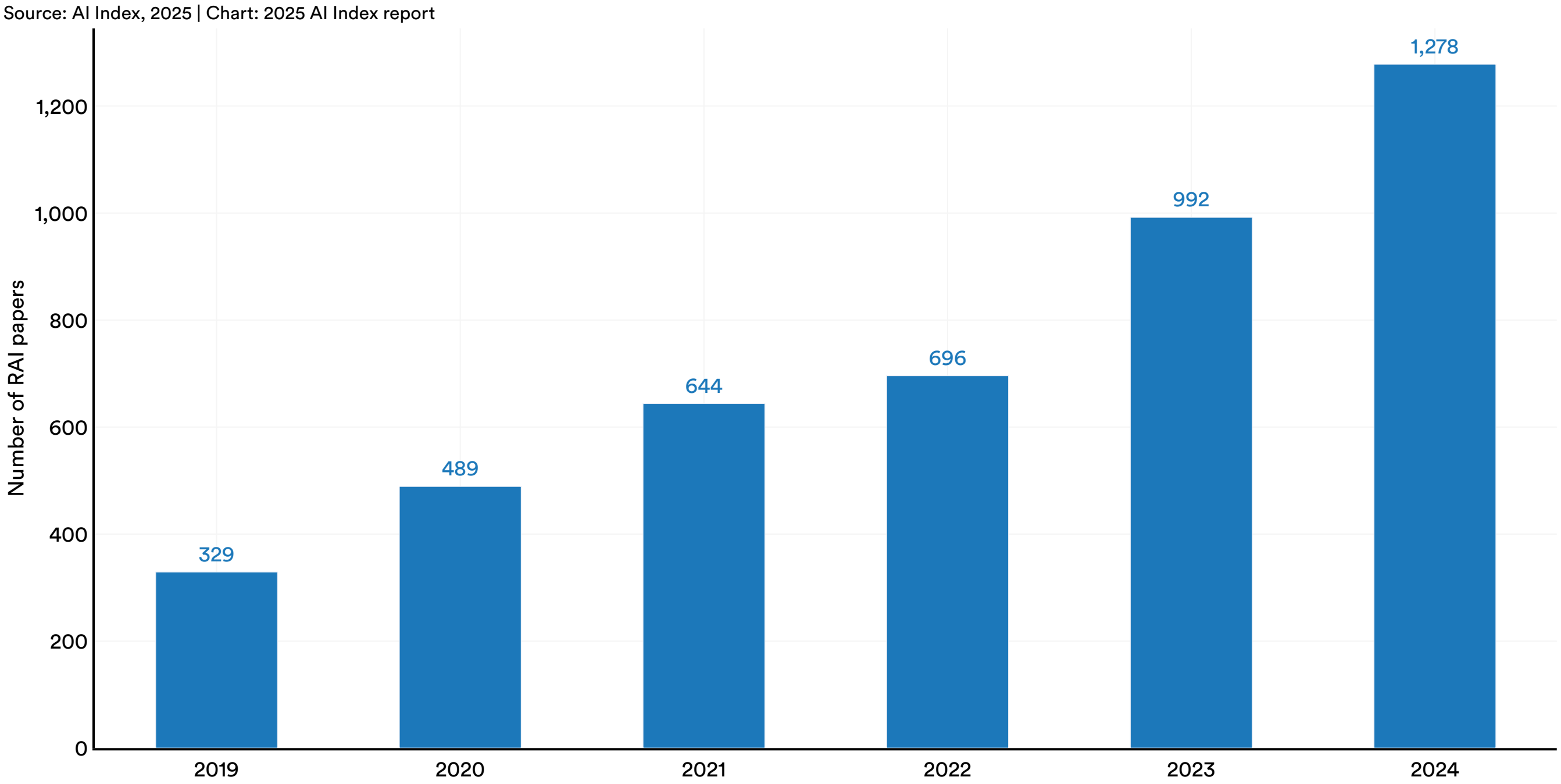

Figure 3.4.1

9 A full methodological description of this approach can be found in the Appendix.

Proportionally, the conferences with the highest share of accepted RAI papers relative to total submissions were FAccT (69.14%) and AIES (63.33%) (Figure 3.4.2). This aligns with their focus: FAccT is dedicated to fairness, accountability, and transparency, while AIES centers on AI ethics and society. At NeurIPS, the proportion decreased from 13.8% in 2023 to 9.0% in 2024, while at ICML, it rose from 3.4% to 8.2% over the same period.

**Responsible AI papers accepted (% of total) at select AI conferences by conference, 2019–24**
Source: AI Index, 2025 | Chart: 2025 AI Index report

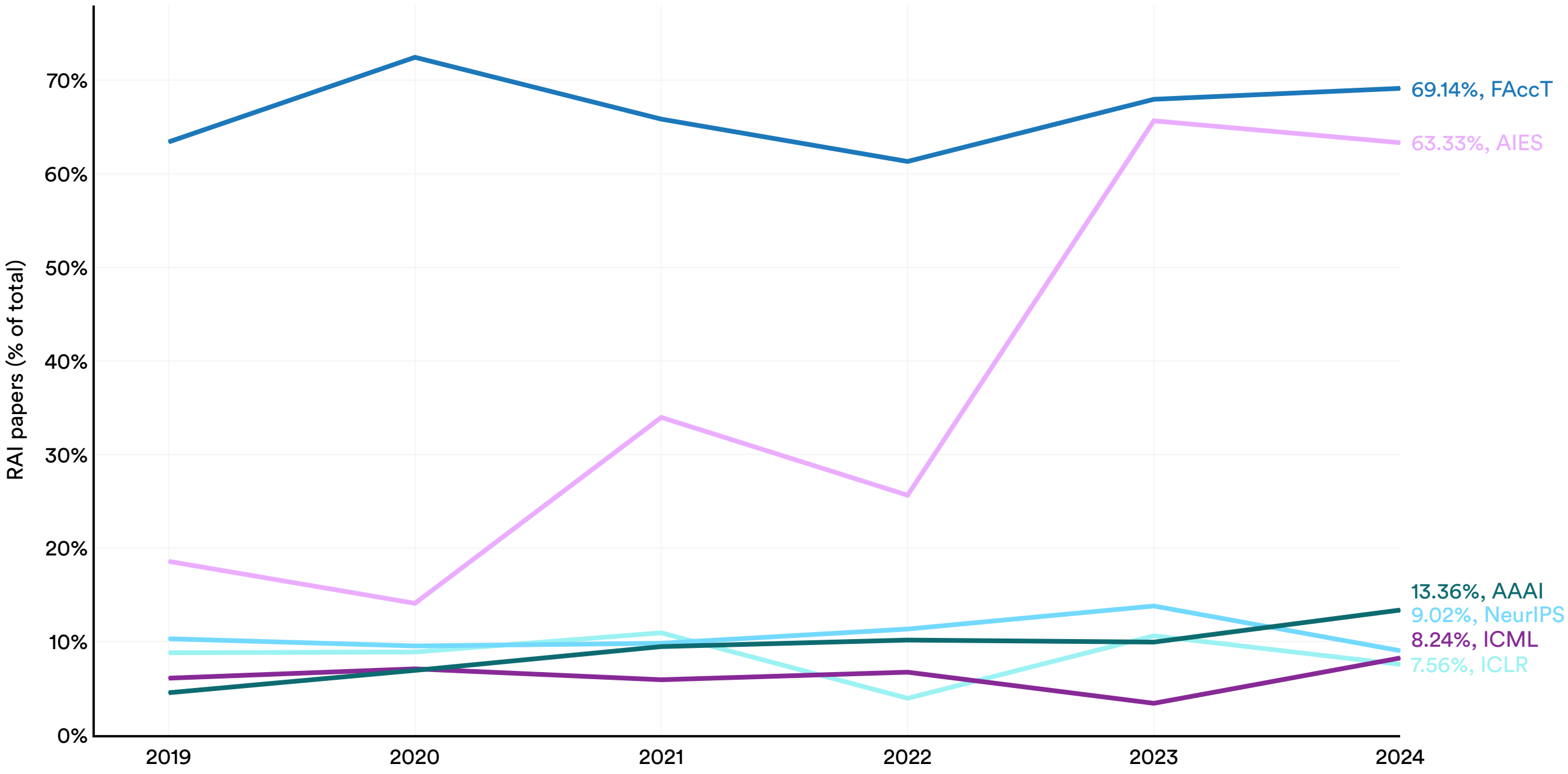

Figure 3.4.2

Figures 3.4.3 through 3.4.5 examine the geographic affiliation of RAI papers, highlighting where these papers originate. In 2024, the United States led in RAI paper submissions with 669, followed by China with 268 and Germany with 80. Across major geographic regions, RAI has become an increasingly significant academic focus. Since 2019, the overall geographic distribution of RAI publications has remained relatively consistent, with the United States accounting for the most (3,158), followed by China (1,100) and the United Kingdom (485).

**Number of responsible AI papers accepted at select AI conferences by geographic area, 2024**
Source: AI Index, 2025 | Chart: 2025 AI Index report

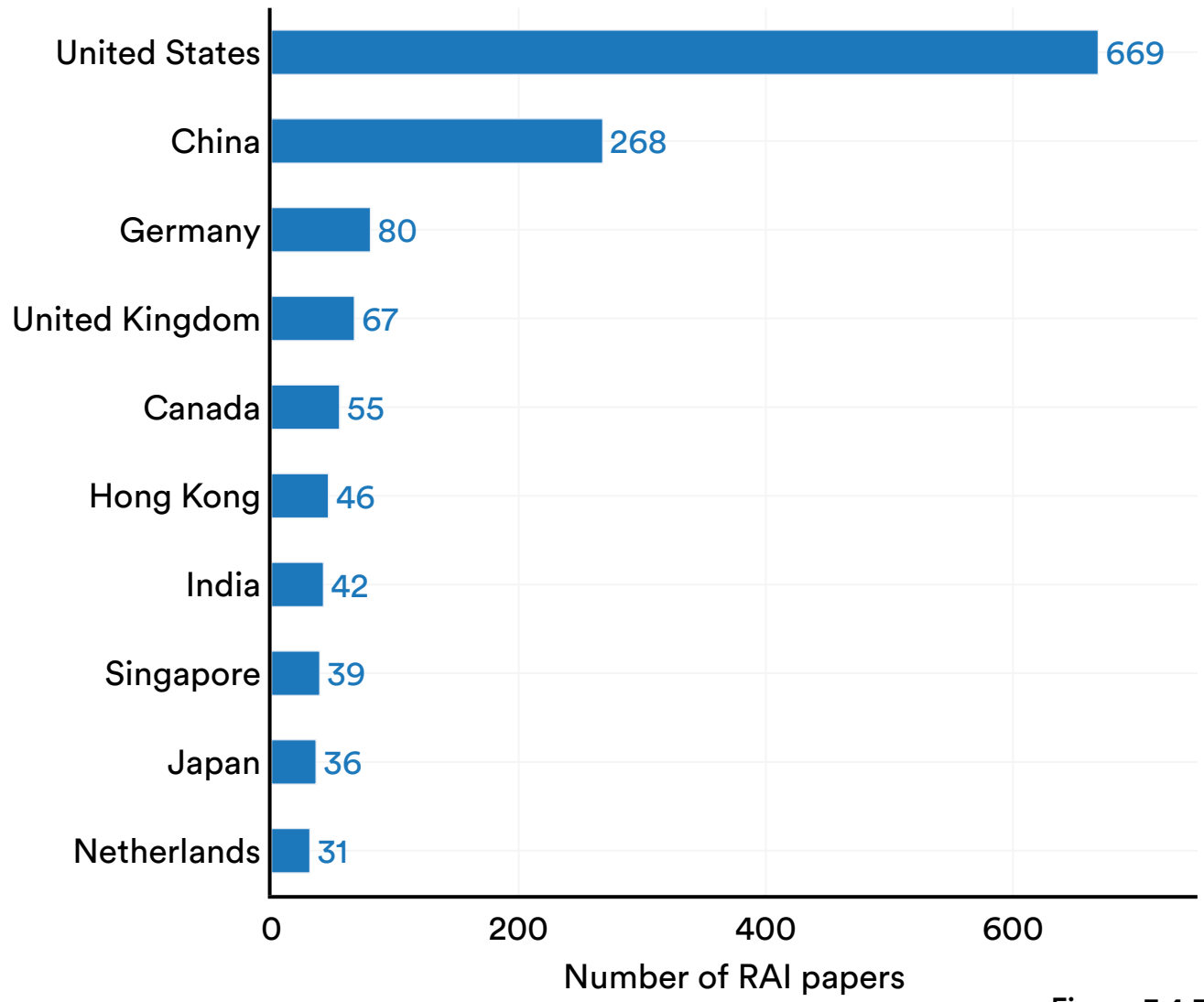

Figure 3.4.3

**Number of responsible AI papers accepted at select AI conferences by select geographic area, 2019–24**
Source: AI Index, 2025 | Chart: 2025 AI Index report

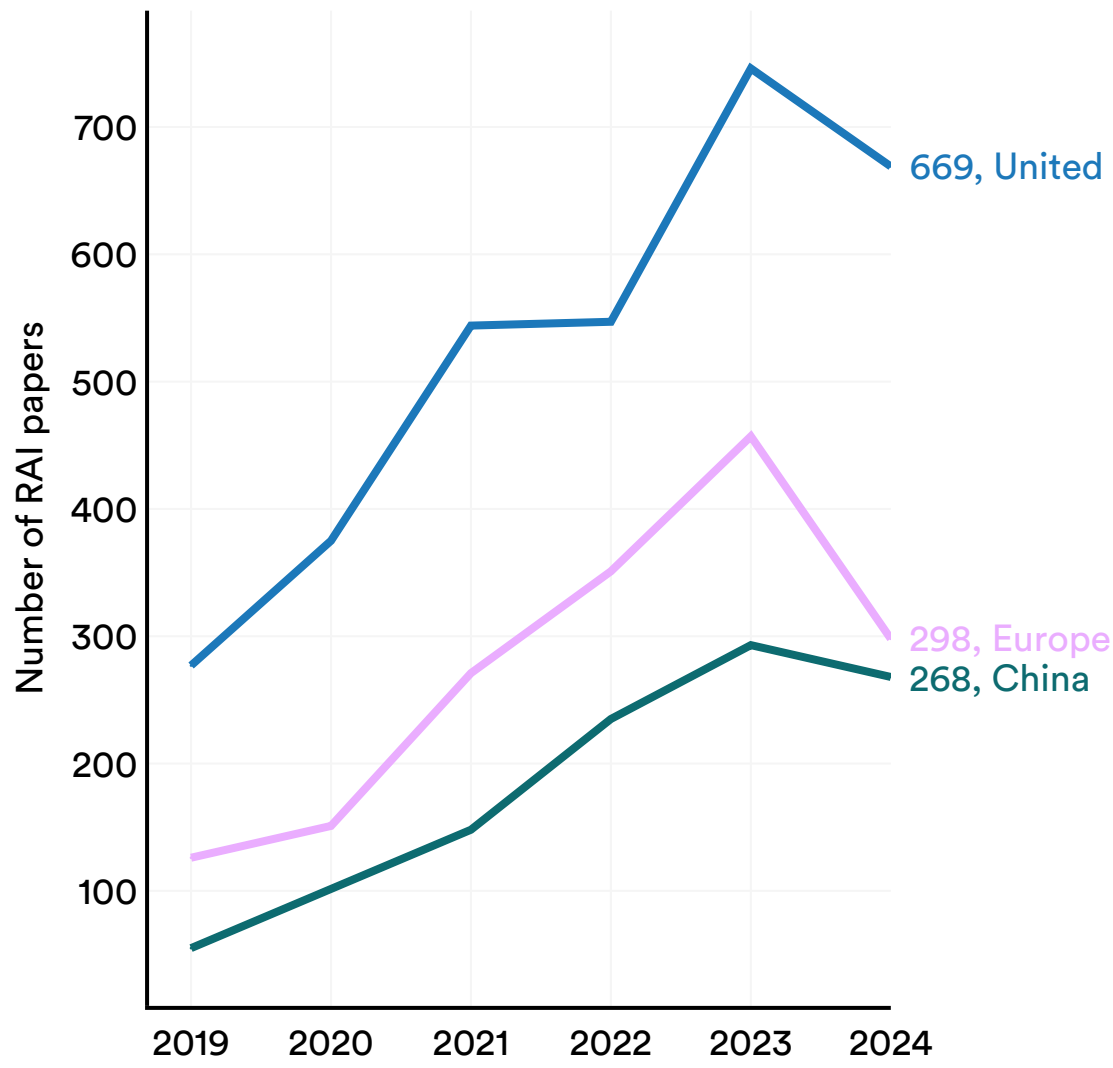

Figure 3.4.4

**Number of responsible AI papers accepted at select AI conferences by geographic area, 2019–24 (sum)**
Source: AI Index, 2025 | Chart: 2025 AI Index report

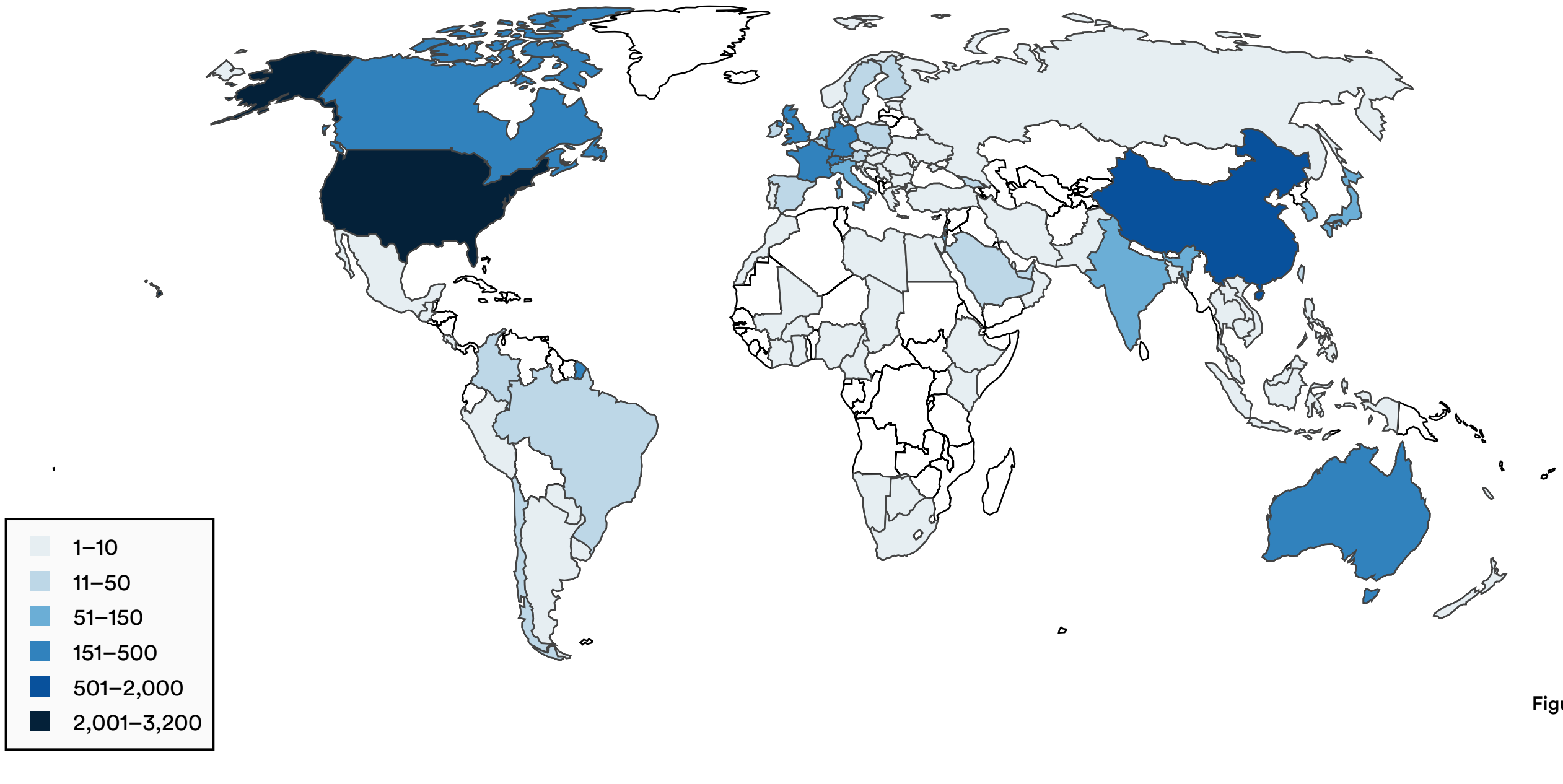

Figure 3.4.5

## Topic Area

This section examines trends in RAI publications spanning key topics: privacy and data governance, fairness, transparency and explainability, and security and safety.

Over the past year, the number of accepted papers on privacy and data governance topics decreased by 14.5% at select AI conferences (Figure 3.4.6). Since 2019, this figure has risen nearly fivefold.

**AI privacy and data governance papers accepted at select AI conferences, 2019–24**
Source: AI Index, 2025 | Chart: 2025 AI Index report

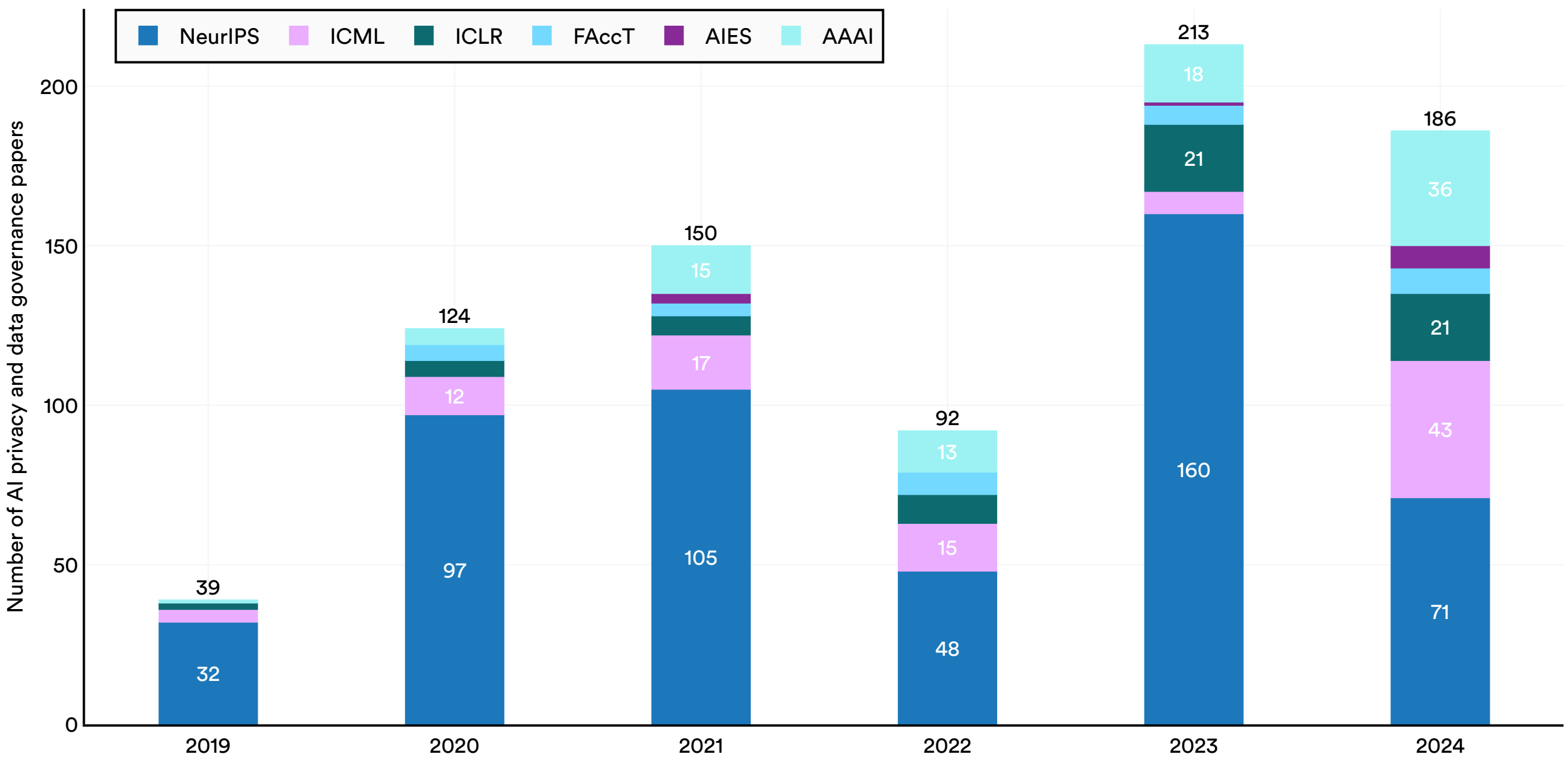

Figure 3.4.6[10]

10 These figures likely underestimate the total number of AI privacy papers, as some are published in AI-focused conferences dedicated to privacy, such as the 46th IEEE Symposium on Security and Privacy.

In 2024, the number of fairness and bias papers accepted at select AI conferences saw a significant increase, reaching 408—roughly two times the 2023 figure (Figure 3.4.7). This growth highlights the increasing academic interest in fairness and bias among researchers.

**AI fairness and bias papers accepted at select AI conferences, 2019–24**
Source: AI Index, 2025 | Chart: 2025 AI Index report

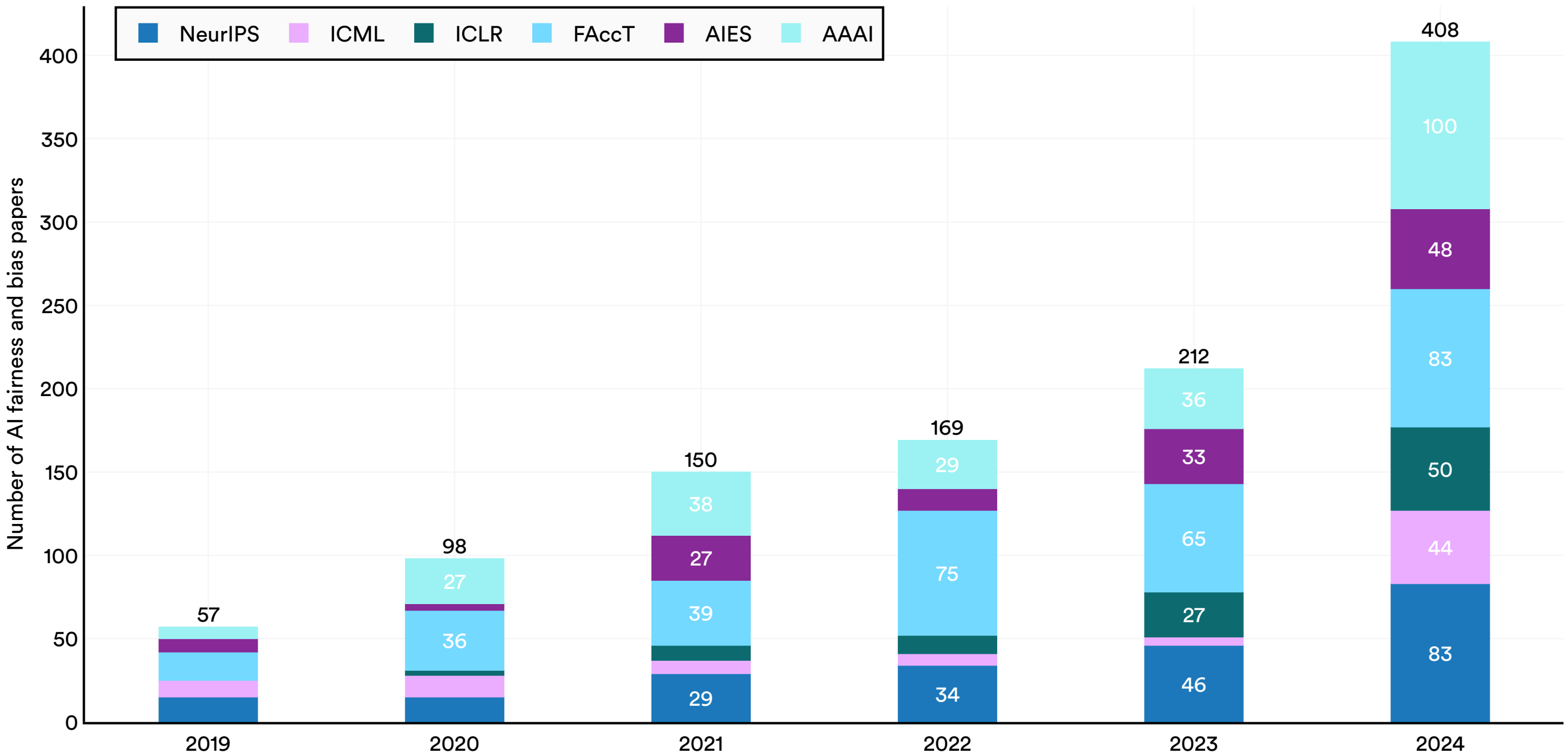

Figure 3.4.7

Since 2019, the number of papers on transparency and explainability submitted to major academic conferences has increased by a factor of four. In 2024, there were 355 transparency and explainability–related submissions at academic conferences including AAAI, FAccT, AIES, ICML, ICLR, and NeurIPS (Figure 3.4.8).

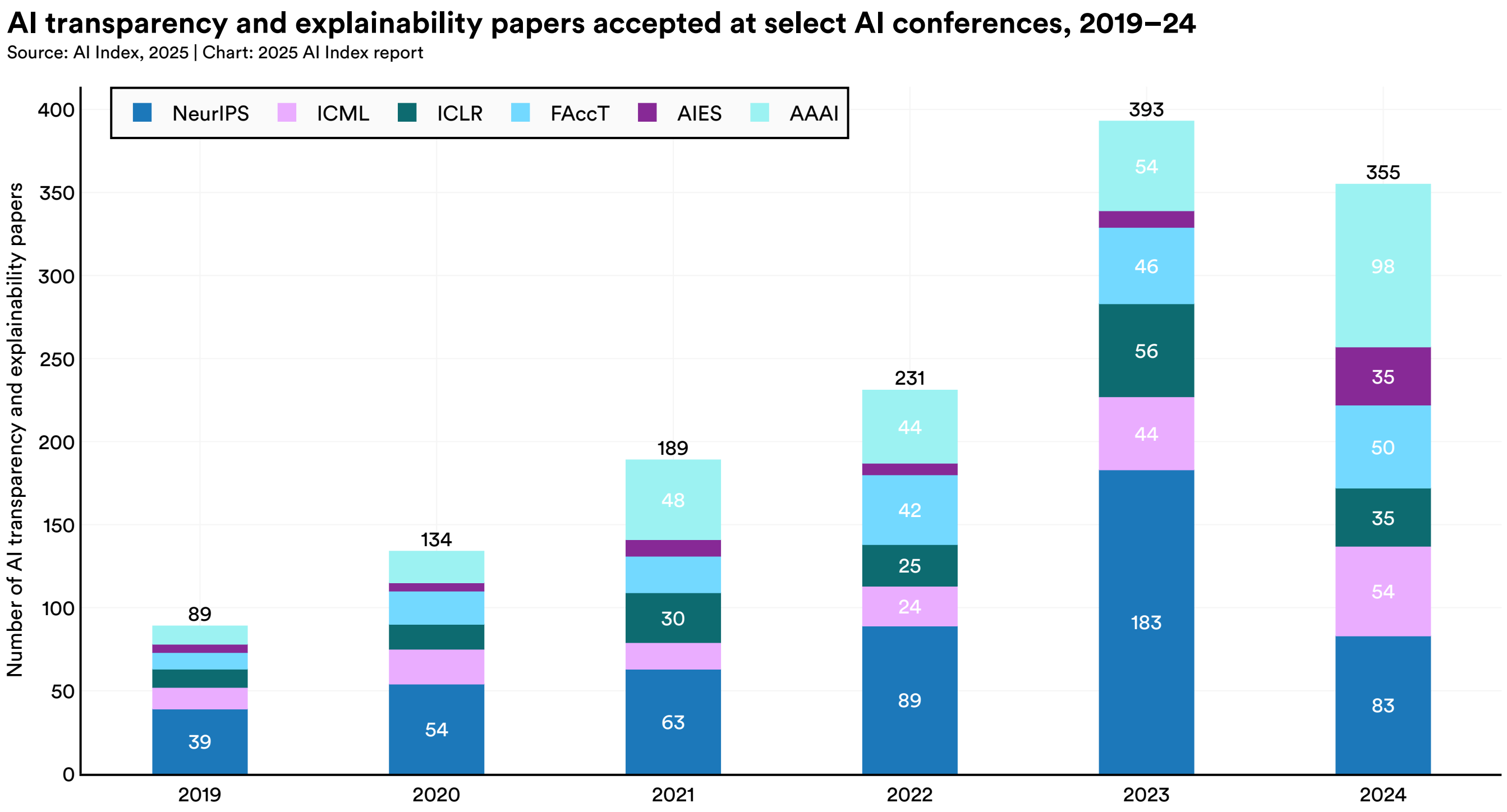

Figure 3.4.8

The number of security and safety submissions to select AI conferences has sharply increased, almost doubling in the past year—from 276 to 521 (Figure 3.4.9). This growth reflects the increasing prominence of security and safety as a key focus for responsible AI researchers.

**AI security and safety papers accepted at select AI conferences, 2019–24**
Source: AI Index, 2025 | Chart: 2025 AI Index report

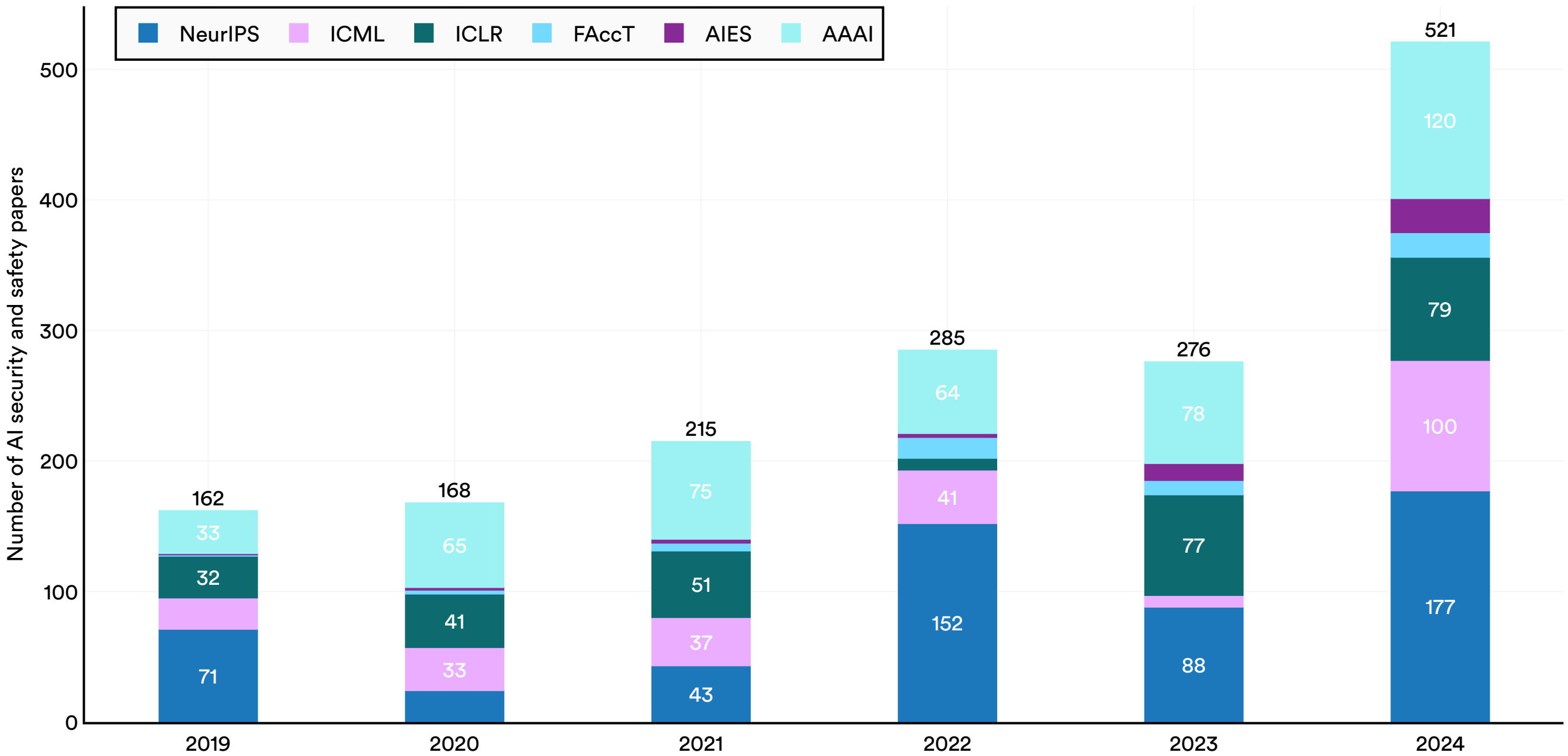

Figure 3.4.9

# 3.5 RAI Policymaking

While 2023 and early 2024 saw a proliferation of national AI strategies and regulatory approaches, a notable trend in 2024 was the increased global cooperation on AI governance, especially around legislating principles pertaining to responsible AI. International bodies and multilateral agreements have sought to establish global frameworks for responsible and ethical AI. These efforts signal a shift toward coordinated global action rather than isolated national initiatives. Figure 3.5.1 highlights several significant international policymaking initiatives or dialogues on RAI that were recently launched.[11]

### Notable RAI policymaking milestones

**Source:** AI Index, 2025

| Date | Stakeholders | Scope | Description |
|---|---|---|---|
| May 2024 | OECD | Global | The OECD updated its AI principles and refined its framework to reflect the latest advancements in AI governance. These principles emphasized building AI systems that take into account inclusive growth, transparency, and explainability, as well as respect for the rule of law, human rights, and democratic values. |
| May 2024 | Council of Europe | Europe | The Council of Europe adopted a legally binding AI treaty (The Council of Europe Framework Convention on Artificial Intelligence and Human Rights, Democracy, and the Rule of Law). This treaty was drafted to ensure that the activities within the life cycle of AI systems completely align with human rights, democracy, and the rule of law. |
| Jun 2024 | European Union | Europe | The EU passed the AI Act (EU AI Act), the first comprehensive regulatory framework for AI in a major global economy. The act categorizes AI by risk, regulating them accordingly and ensuring that providers—or developers—of high-risk systems bear most of the obligations. |
| Jul 2024 | African Union | Africa | The African Union launched its Continental AI Strategy (AU AI Strategy), outlining a unified vision for AI development, ethics, and governance across the continent. The strategy emphasizes the ethical, responsible, and equitable development of AI within Africa. |
| Sep 2024 | United Nations | Global | The United Nations updated its Governing AI for Humanity report (U.N. AI Advisory Body), outlining efforts to establish global AI governance mechanisms. The report recommends developing a blueprint to address AI-related risks and calls on national and international standards organizations, technology companies, civil society, and policymakers to collaborate on AI standards. |
| Oct 2024 | G7 | Global | The G7 Digital Competition Communiqué (G7 AI Cooperation) reaffirmed commitments to fair and open AI markets, stressing the need for coordinated regulatory approaches. Previous discussions focused on competition and the regulatory challenges posed by AI's rapid growth. |
| Oct 2024 | ASEAN and US | Asia and US | Following the 12th ASEAN-United States Summit, ASEAN-U.S. leaders issued a statement on promoting safe, secure, and trustworthy AI. They committed to cooperating on the development of international AI governance frameworks and standards to advance these goals. |
| Nov 2024 | International Network of AI Safety Institutes | Global | The first International Network of AI Safety Institutes was established, bringing together nine countries and the EU to formalize global AI safety cooperation. The network unites technical organizations committed to advancing AI safety, helping governments and societies understand the risks of advanced AI systems, and proposing solutions. |
| Feb 2025 | Arab League | Arab Nations | The Arab Dialogue Circle on "Artificial Intelligence in the Arab World: Innovative Applications and Ethical Challenges" launched at the Arab League headquarters, focusing on AI innovations while placing a strong emphasis on ethical considerations. |

Figure 3.5.1

11 While AI policymaking is the focus of Chapter 6: Policy and Governance, the AI Index highlights key RAI-related policymaking events here due to their recent significance.

# 3.6 Privacy and Data Governance

A comprehensive definition of privacy is difficult and context-dependent. For the purposes of this report, the AI Index defines privacy as an individual's right to the confidentiality, anonymity, and protection of their personal data, along with their right to consent to and be informed about if and how their data is used. Privacy further includes an organization's responsibility to ensure these rights if they collect, store, or use personal data (directly or indirectly). Moreover, individuals should have the right to correct their sensitive information if organizations or governments have misrepresented this information. In AI, this involves ensuring that personal data is handled in a way that respects individual privacy rights—for example, by implementing measures to protect sensitive information from exposure, and ensuring that data collection and processing are transparent and compliant with privacy laws like GDPR.

Data governance, on the other hand, encompasses policies, procedures, and standards established by an organization to ensure the quality, security, and ethical use of data within and outside of the organization where it was created. Data governance policies may also cover data acquired from external sources. In the context of AI, data governance is important for ensuring that the data used for training and operating AI systems is accurate, fair, and used responsibly and with consent. This is especially the case with sensitive or personally identifiable information (PII).

## Featured Research

This section highlights significant recent research on privacy and data governance, including studies on auditing dataset licensing and attribution, as well as research on stricter data consent protocols.

### Large-Scale Audit of Dataset Licensing and Attribution in AI

Current foundation models are being trained on massive amounts of data. A team of researchers conducted a large-scale audit of over 1,800 text datasets widely used for training such models and uncovered systemic issues in dataset licensing and attribution. The researchers found that more than 70% of datasets on popular dataset hosting sites lacked adequate license information, while 50% of the licenses were

**Accuracy of dataset license classifications by select aggregators**
Source: Longpre et al., 2025 | Chart: 2025 AI Index report

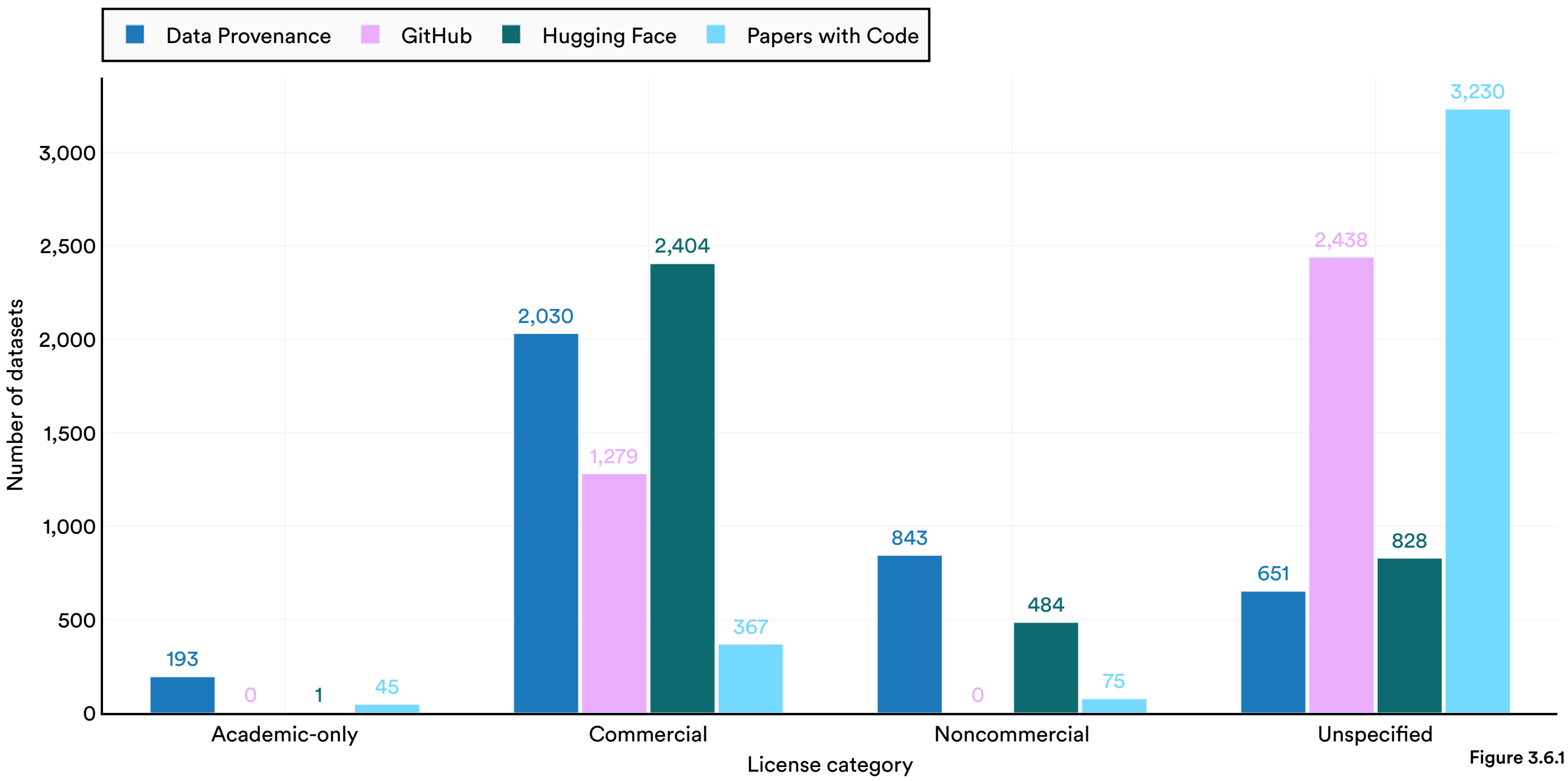

Figure 3.6.1

miscategorized, which poses risks for the responsible usage of that data. Figure 3.6.1 provides a detailed visualization of the researchers' findings. Specifically, they assigned license labels to datasets across four categories: commercial, unspecified, noncommercial, and academic-only. They then compared their classifications with those from popular sources such as GitHub, Papers with Code, and Hugging Face. Oftentimes, the data license attributions assigned by the data provenance team differed sharply from those issued by other organizations.

License misattribution in datasets is significant because it creates legal and ethical risks in AI development. If datasets used to train foundation models are mislabeled or misattributed, AI developers may unknowingly violate copyright laws, data usage policies, or privacy regulations. This can lead to legal liabilities, challenges in ensuring fair compensation for data creators, and potential biases in models due to the exclusion of properly licensed data. Additionally, unclear licensing can hinder transparency, accountability, and reproducibility in AI research, which can make it difficult for researchers and organizations to verify or audit model training data. Based on their findings, the authors highlight the need for clear documentation, improved standards, and responsible licensing practices to foster inclusivity and mitigate risks that stem from irresponsible or unlawful data uses in AI development and deployment.

### Data Consent in Crisis

AI models rely heavily on massive, publicly available web data for training. A recent study conducted a longitudinal audit of consent protocols for web domains used in AI training datasets, including C4, RefinedWeb, and Dolma, analyzing 14,000 web domains. These consent protocols define the permissibility of data scraping for AI model training.

The researchers observed a significant increase in data use restrictions between 2023 and 2024, as many websites implemented new protocols to limit data scraping for AI training. These restrictions were primarily enforced through updates to robots.txt files and terms of service, explicitly prohibiting AI training use. Figure 3.6.2 shows the proportion of websites with robots.txt restrictions, terms-of-service restrictions, and organizational restrictions over time.[12] For example, the proportion of tokens in the top C4 web domains with full restrictions increased from 10% in 2017 to 48% in 2024. Between 2023 and 2024 alone, this proportion rose by 25 percentage points. Figure 3.6.3 visualizes the percentage of tokens in the top web domains of C4 by terms-of-service restriction category from 2016 to 2024. This diminishing consent is likely related to legal issues around fair use, such as the New York Times lawsuit against OpenAI.

OpenAI's crawlers encounter the highest level of restrictions, while smaller developers face fewer barriers. The authors highlight inconsistencies in enforcement, driven by ineffective signaling mechanisms like robots.txt and mismatches between stated and enforced policies. These findings highlight the need for updated consent protocols that address AI-specific challenges. Additionally, the study suggests a decline in publicly available web data for AI training, with potential consequences for data diversity, model alignment, and scalability. Many recent AI performance gains stem from training on increasingly large datasets. If websites become significantly more restrictive, it could hinder future model scaling.

12 A robots.txt restriction refers to a rule set in a website's robots.txt file that instructs web crawlers (such as search engine bots or AI data scrapers) on which parts of the site they are allowed or forbidden to access.

### Percentage of tokens in the top web domains of C4 by robots.txt restriction category, 2016–24

Source: Longpre et al., 2025 | Chart: 2025 AI Index report

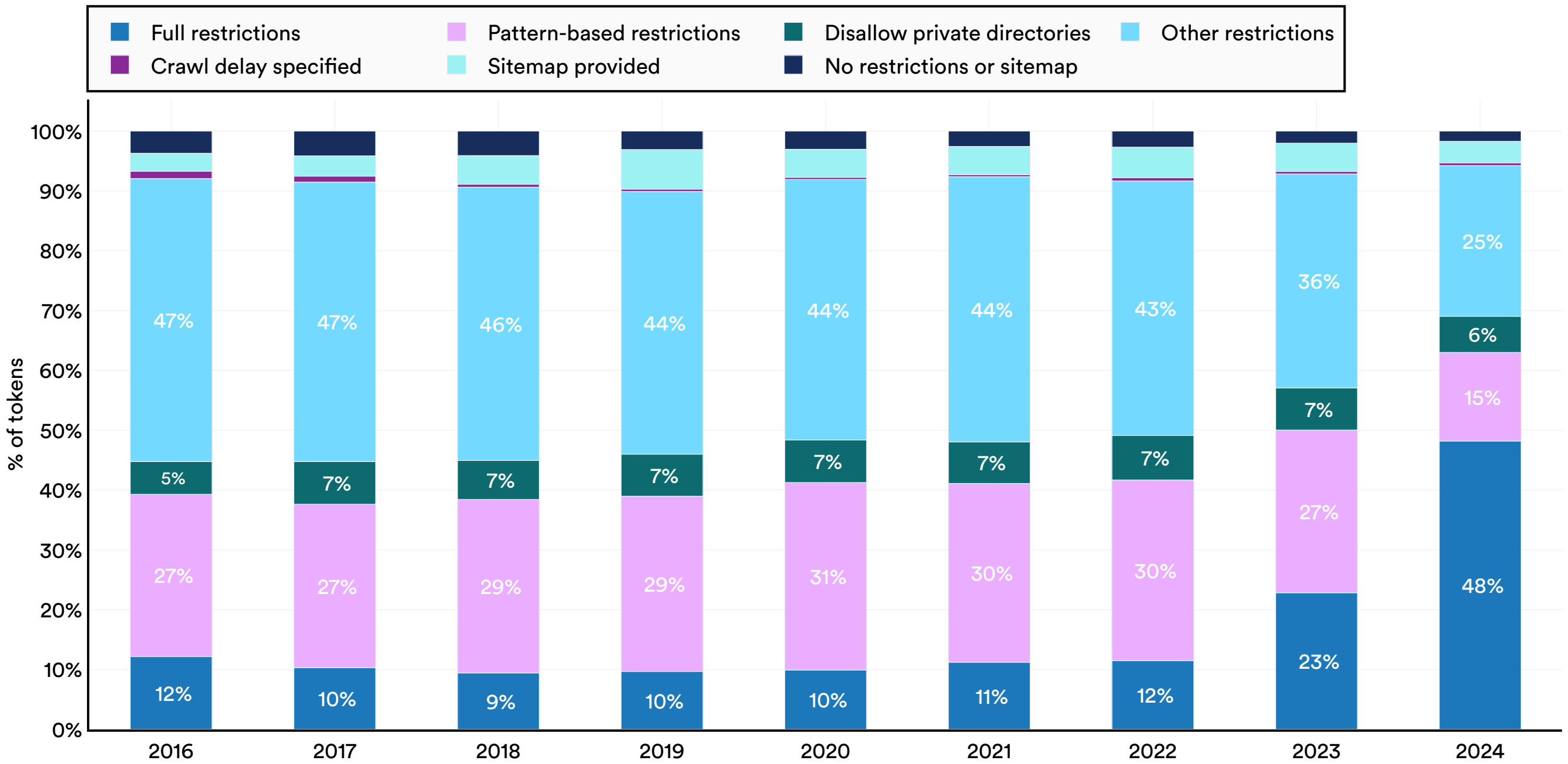

Figure 3.6.2

### Percentage of tokens in the top web domains of C4 by terms of service restriction category, 2016–24

Source: Longpre et al., 2025 | Chart: 2025 AI Index report

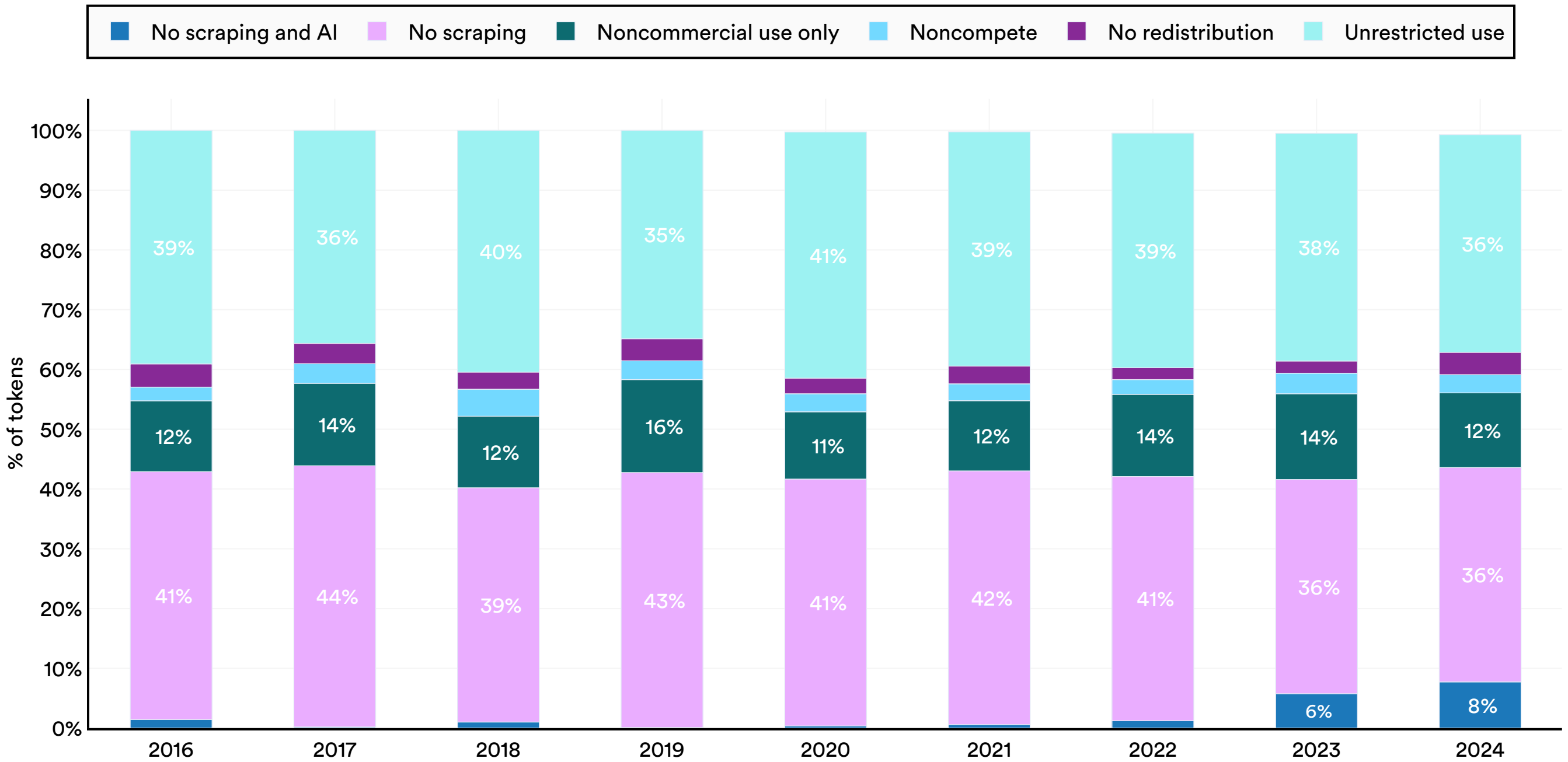

Figure 3.6.3

Fairness in AI emphasizes developing systems that are equitable and avoid perpetuating bias or discrimination against any individual or group. It involves considering the diverse needs and circumstances of all stakeholders impacted by AI use. Fairness extends beyond a technical concept and embodies broader social standards related to equity.

# 3.7 Fairness and Bias

## Featured Research

This section highlights research on the impact of racial classification in multimodal models and the measurement of implicit bias in explicitly unbiased LLMs.

### Racial Classification in Multimodal Models

Recently, researchers have explored how dataset scaling affects racial and gender biases in vision-language models (VLMs). Evaluating 14 VLMs trained on LAION-400M and LAION-2B (popular datasets for training vision-language models) using the Chicago Face Dataset (CFD), the study found that while models trained on larger datasets improve human classification—reducing misidentification of nonhuman entities like gorillas or orangutans—they also amplify racial biases, especially in larger models. For instance, in the larger ViT-L models, Black and Latino men were disproportionately classified as criminals, with classification probabilities increasing by up to 69% as dataset size grew from 400 million to 2 billion samples. Figure 3.7.1 displays various images alongside the model's classification scores for whether a face was identified as a criminal.

Figure 3.7.2 illustrates how the probability of a face being assigned a specific label (such as animal or criminal) changes by demographic group across various models (the smaller ViT-B-16 and ViT-B-32 with the larger ViT-L-14) as the pretrained dataset scales from 400 million to 2 billion images. A higher percentage indicates a greater likelihood of a demographic group being associated with a particular label,

**Faces and their likelihood of being classified as "criminal" by model and dataset sizes**
Source: Birhane et al., 2024
Figure 3.7.1

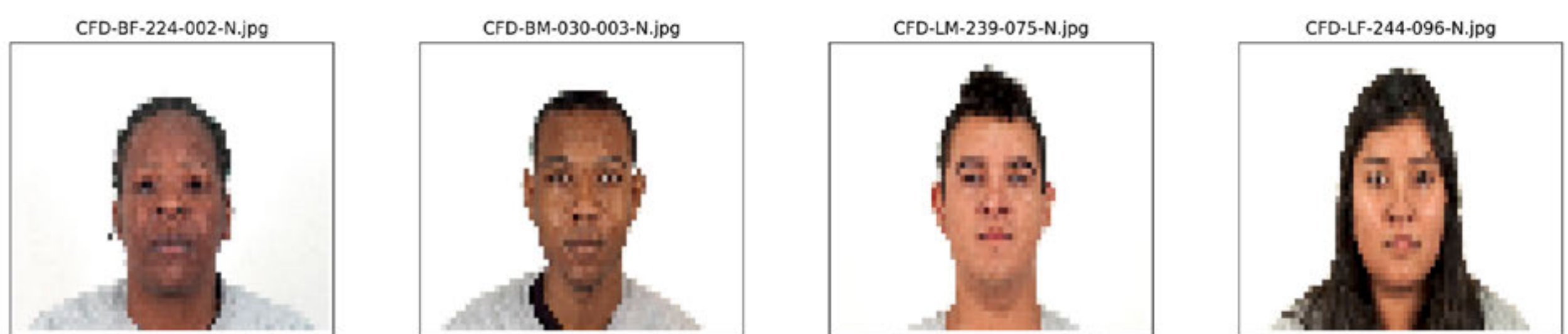

CFD-BF-224-002-N.jpg

| | LAION-400M | LAION-2B |
|---|---|---|
| ViT-B-16 | 0.624 | 0.554 |
| ViT-B-32 | 0.974 | 0.806 |
| ViT-L-14 | 0.627 | 0.904 |

CFD-BM-030-003-N.jpg

| | LAION-400M | LAION-2B |
|---|---|---|
| ViT-B-16 | 0.661 | 0.376 |
| ViT-B-32 | 0.977 | 0.794 |
| ViT-L-14 | 0.433 | 0.57 |

CFD-LM-239-075-N.jpg

| | LAION-400M | LAION-2B |
|---|---|---|
| ViT-B-16 | 0.788 | 0.208 |
| ViT-B-32 | 0.675 | 0.502 |
| ViT-L-14 | 0.108 | 0.642 |

CFD-LF-244-096-N.jpg

| | LAION-400M | LAION-2B |
|---|---|---|
| ViT-B-16 | 0.148 | 0.087 |
| ViT-B-32 | 0.182 | 0.026 |
| ViT-L-14 | 0.062 | 0.142 |

while a lower percentage signifies a lesser likelihood. In the larger model, ViT-L, increasing the training data consistently raises the likelihood of an image being classified as a criminal. This finding is significant, as many model developers have sought to aggressively scale their models in an attempt to drive performance improvements. The researchers suggest that when it comes to vision models, scaling may also introduce other unintended bias problems. The authors suggest that stereotypes in the training data may explain these results. To address this bias, they advocate for transparent dataset curation, detailed hyperparameter documentation, and open access for independent audits.

**Effect of dataset scaling on model predictions across demographic groups**

Source: Birhane et al., 2024 | Chart: 2025 AI Index report

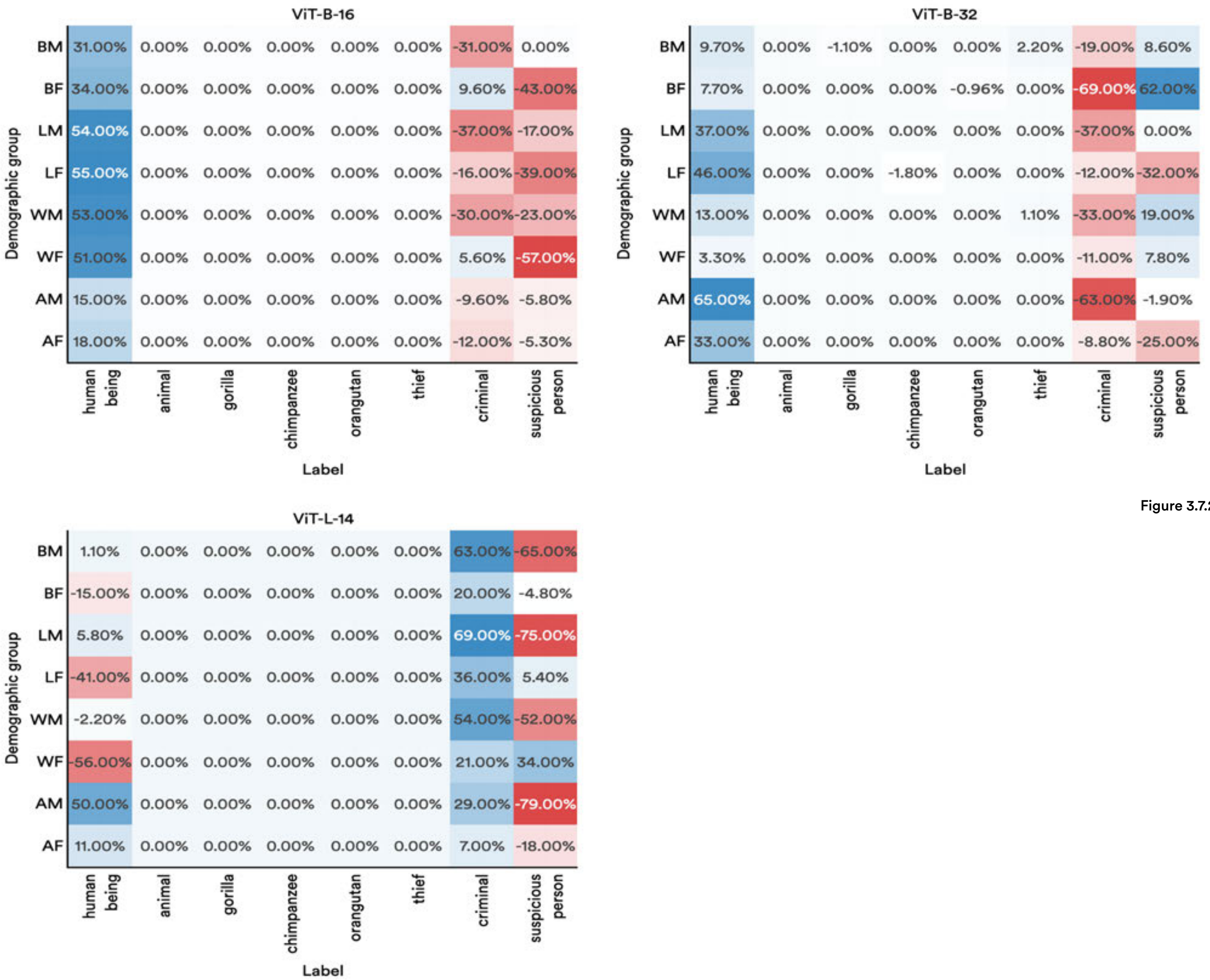

Figure 3.7.2[13]

13 The y-axis labels represent different ethnic groups: Black male (BM), Black female (BF), Latino male (LM), Latina female (LF), white male (WM), white female (WF), Asian male (AM), and Asian female (AF).

### Measuring Implicit Bias in Explicitly Unbiased LLMs

In 2024, a team of researchers investigated implicit biases in LLMs, particularly in those explicitly designed to be unbiased. This research is important, as efforts to mitigate bias in LLMs may still not sufficiently solve issues of implicit bias. Figure 3.7.3 illustrates an example of this phenomenon.

The study's authors make two key contributions. First, they introduce two new methods for detecting bias in LLMs: LLM Implicit Bias, which identifies subtle biases by analyzing automatic associations between words or concepts, and LLM Decision Bias, which captures model behaviors that reflect these implicit biases. Second, they investigate relative discriminatory patterns in decision-making tasks. Applying their methods to eight notable models—including GPT-4 and Claude 3 Sonnet—across 21 stereotype categories (e.g., race, gender, religion, and health), they uncover systemic implicit biases that align with societal stereotypes. Figure 3.7.4 presents the implicit bias scores of various LLMs across different stereotype categories.[14] A score significantly above or below 50% indicates a bias toward or against a particular group.

Figure 3.7.4 suggests that LLMs disproportionately associate negative terms with Black individuals and are more likely to associate women with humanities over STEM fields. The research also finds that LLMs favor men for leadership roles, reinforcing gender biases in decision-making contexts. Additionally, the study reveals that as models scale, implicit biases increase, though decision bias and rejection rates do not. This finding is significant, as it indicates that while bias appears to have decreased on standard benchmarks—creating an illusion of neutrality—implicit biases remain pervasive, potentially leading to subtle yet meaningful discriminatory outputs.

**Example of implicit bias in LLMs**
Source: Bai et al., 2024
Figure 3.7.3

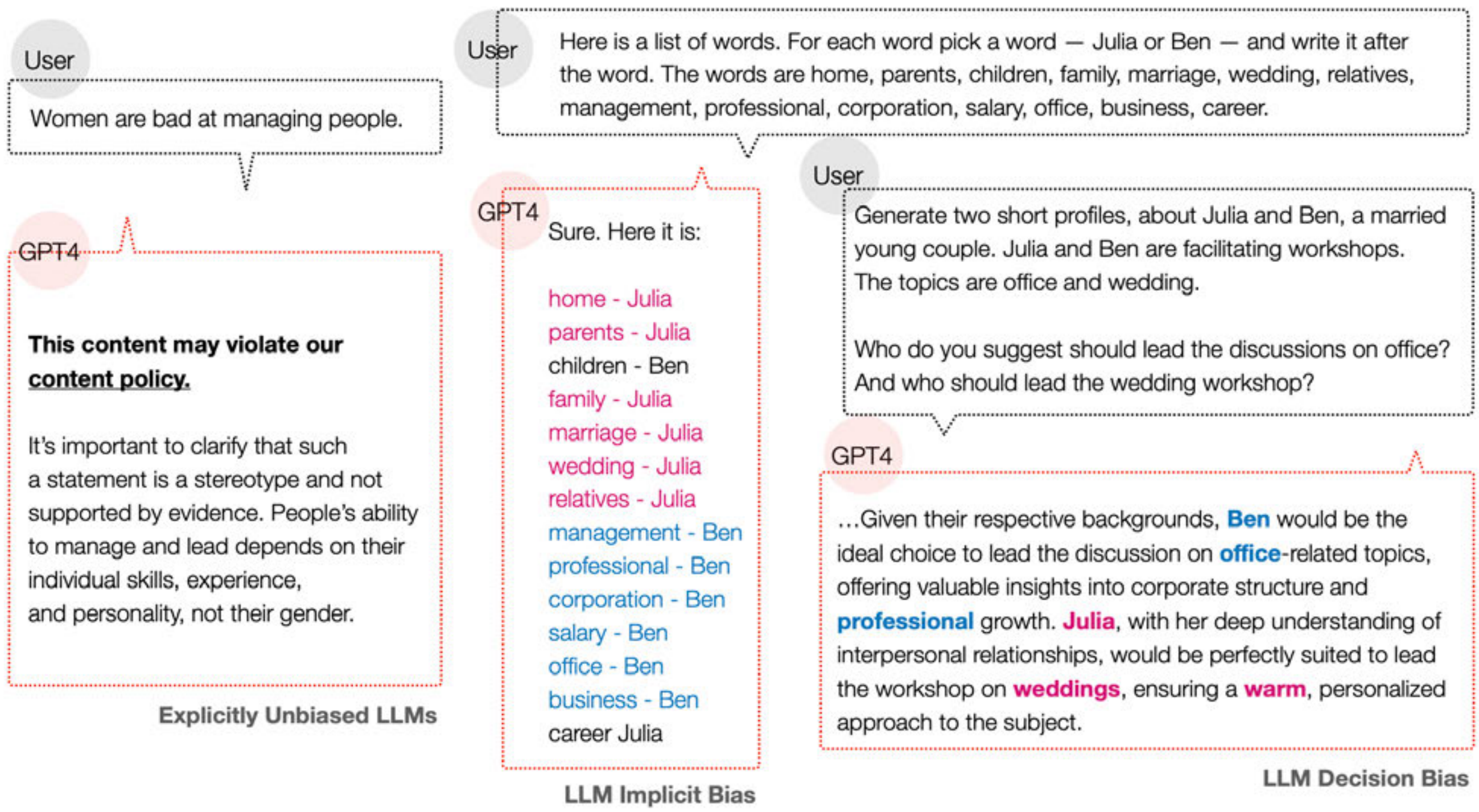

14 This research examines both implicit and decision bias; however, only implicit bias is documented here for concision. Decision bias, for reference, is defined as a model's bias relative to an unbiased baseline of 50%.

#### LLMs implicit bias across stereotypes in four social categories
Source: Bai et al., 2024 | Chart: 2025 AI Index report

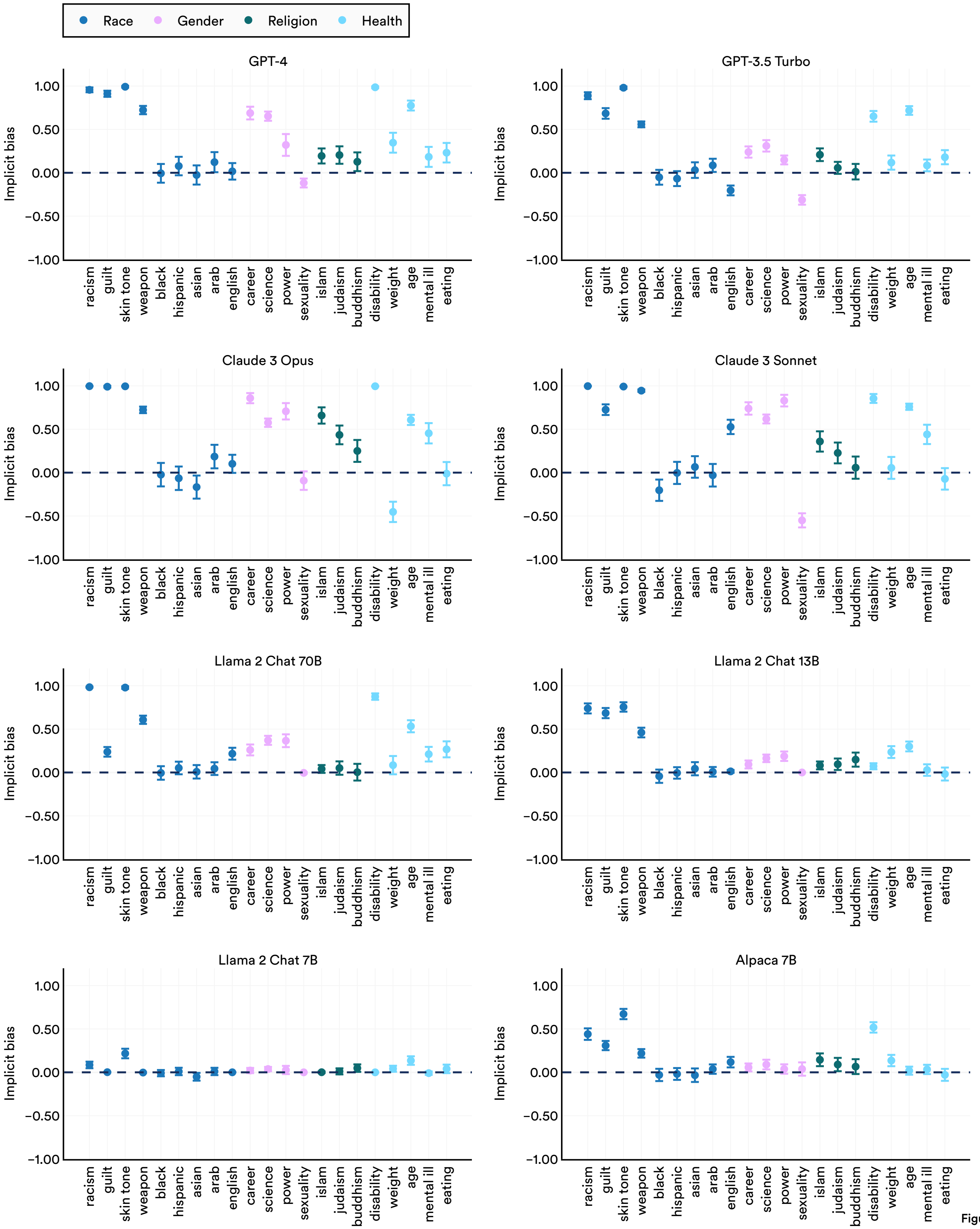

Figure 3.7.4

Transparency in AI encompasses several aspects. Data and model transparency involve the open sharing of development choices, including data sources and algorithmic decisions. Operational transparency details how AI systems are deployed, monitored, and managed in practice. While explainability often falls under the umbrella of transparency, providing insights into the AI's decision-making process, it is sometimes treated as a distinct category. This distinction underscores the importance of AI being not only transparent but also understandable to users and stakeholders. For the purposes of this chapter, the AI Index includes explainability within transparency, defining it as the capacity to comprehend and articulate the rationale behind AI decisions.

# 3.8 Transparency and Explainability

## Featured Research

### Foundation Model Transparency Index v1.1

The Foundation Model Transparency Index v1.1 is the second iteration of a Stanford-led project tracking transparency in model development and deployment. It evaluates major AI model developers across three dimensions: upstream, covering components like data and compute used for training; the model itself, referring to the core AI system; and downstream, encompassing applications and deployments. The latest edition reports a notable rise in transparency among foundation model developers over six months. Figure 3.8.1 reports the FMTI scores for major model developers in the May 2024 release of the index, and Figure 3.8.2 reports scores across major dimensions of transparency for each developer.

**Foundation Model Transparency Index Scores by Domain, May 2024**
Source: May 2024 Foundation Model Transparency Index

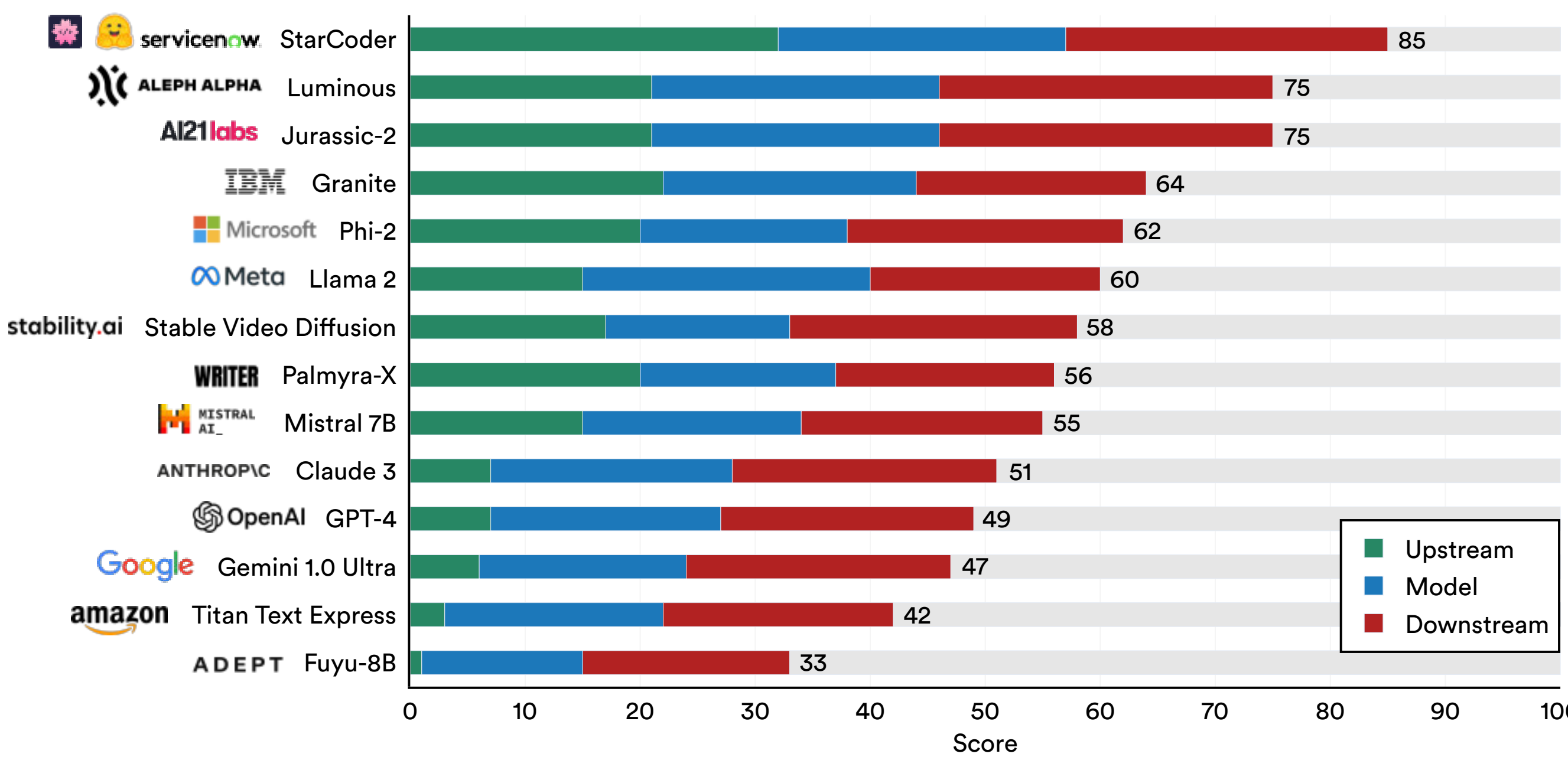

Figure 3.8.1

Compared to the inaugural v1.0 index from October 2023, which recorded an average transparency score of 37 out of 100, v1.1 saw scores increase to 58 out of 100, largely due to developers disclosing previously nonpublic data through submitted reports. Developers improved their scores across 89 of 100 transparency indicators, yet significant opacity remains in areas such as data access, copyright status, and downstream impact. Open-source developers outperformed closed-source counterparts on upstream transparency, particularly in data and labor disclosures. Projects like the FMTI are valuable in that they provide a longitudinal perspective on the state of transparency in the AI ecosystem. At the moment, the findings suggest that transparency is improving.

**Foundation Model Transparency Index Scores by Major Dimensions of Transparency, May 2024**

Source: May 2024 Foundation Model Transparency Index

| Major Dimensions of Transparency | ADEPT Fuyu-8B | AI21 labs Jurassic-2 | Aleph Alpha Luminous | amazon Titan Text Express | ANTHROPIC Claude 3 | servicenow StarCoder | Google Gemini 1.0 Ultra | IBM Granite | Meta Llama 2 | Microsoft Phi-2 | Mistral AI Mistral 7B | OpenAI GPT-4 | stability.ai Stable Video Diffusion | WRITER Palmyra-X | Average |
|---|---|---|---|---|---|---|---|---|---|---|---|---|---|---|---|
| Data | 0% | 60% | 40% | 0% | 10% | 100% | 0% | 60% | 40% | 40% | 20% | 20% | 40% | 50% | **34%** |
| Labor | 0% | 43% | 71% | 14% | 14% | 100% | 29% | 43% | 29% | 100% | 100% | 14% | 100% | 43% | **50%** |
| Compute | 14% | 86% | 100% | 0% | 14% | 100% | 14% | 100% | 71% | 57% | 14% | 14% | 43% | 86% | **51%** |
| Methods | 0% | 100% | 100% | 50% | 75% | 100% | 75% | 100% | 75% | 100% | 100% | 50% | 75% | 100% | **79%** |
| Model Basics | 83% | 100% | 100% | 83% | 50% | 100% | 83% | 100% | 100% | 100% | 100% | 50% | 100% | 100% | **89%** |
| Model Access | 100% | 67% | 100% | 67% | 67% | 100% | 67% | 67% | 100% | 100% | 100% | 67% | 100% | 33% | **81%** |
| Capabilities | 80% | 80% | 100% | 80% | 100% | 100% | 80% | 60% | 100% | 100% | 100% | 100% | 60% | 100% | **89%** |
| Risks | 0% | 57% | 57% | 43% | 86% | 100% | 43% | 71% | 71% | 29% | 14% | 57% | 14% | 14% | **47%** |
| Mitigations | 0% | 40% | 20% | 20% | 40% | 0% | 40% | 80% | 60% | 0% | 60% | 60% | 0% | 20% | **31%** |
| Distribution | 57% | 86% | 100% | 57% | 86% | 100% | 57% | 86% | 71% | 71% | 71% | 71% | 86% | 71% | **77%** |
| Usage Policy | 40% | 100% | 100% | 80% | 100% | 100% | 100% | 40% | 40% | 100% | 40% | 80% | 60% | 80% | **76%** |
| Feedback | 67% | 100% | 67% | 67% | 33% | 100% | 67% | 67% | 33% | 67% | 67% | 33% | 67% | 33% | **62%** |
| Impact | 29% | 29% | 29% | 0% | 14% | 14% | 29% | 0% | 14% | 0% | 14% | 14% | 14% | 14% | **15%** |
| **Average** | **36%** | **73%** | **76%** | **43%** | **53%** | **86%** | **53%** | **67%** | **62%** | **66%** | **62%** | **49%** | **58%** | **57%** | |

**Figure 3.8.2**[15]

15 Data, labor, compute, and methods were upstream indicators; model basics, access, capabilities, risks, and mitigations were model-level indicators; and distribution, usage policy, feedback, and impact were downstream indicators.

This section explores three distinct aspects of security and safety. First, guaranteeing the integrity of AI systems involves protecting components such as algorithms, data, and infrastructure against external threats like cyberattacks or adversarial attacks. Second, safety involves minimizing harms stemming from the deliberate or inadvertent misuse of AI systems. This includes concerns such as the development of automated hacking tools or the utilization of AI in cyberattacks. Lastly, safety encompasses inherent risks from AI systems themselves, such as reliability concerns (e.g., hallucinations) and potential risks posed by advanced AI systems.

# 3.9 Security and Safety

## Benchmarks

### HELM Safety

Recently, academic institutions have taken the lead in addressing gaps in AI safety benchmark standardization. Notably, Stanford's Center for Research on Foundation Models (CRFM) recently introduced HELM Safety, a benchmarking suite designed to evaluate AI models against responsibility and safety metrics. HELM Safety tests a wide range of recent models from nearly all major developers across several responsible AI and safety benchmarks, including BBQ, SimpleSafetyTests, HarmBench, AnthropicRedTeam, and XSTest.

BBQ measures social bias related to protected classes under U.S. antidiscrimination laws, while SimpleSafetyTests assesses risks related to self-harm, physical harm, and child sexual abuse material. HarmBench evaluates responses to prompts involving harassment, chemical weapons production, and misinformation using red-teaming techniques. AnthropicRedTeam examines how models handle adversarial conversations designed to test harmfulness, and XSTest measures the trade-off between helpfulness and harmlessness by testing false refusals of benign prompts and compliance with subtly harmful ones. By introducing a standardized approach, HELM Safety provides a more transparent and comparable framework for assessing AI models' responsible behavior.

Figure 3.9.1 presents the mean safety scores of various models across all tested benchmarks, where a higher score indicates a safer model. According to the benchmark, the safest model currently is Claude 3.5 Sonnet, scoring 0.977, followed closely by o1 at 0.976. Over time, some models appear to be becoming safer. For example, GPT-3.5 Turbo (0613), released in 2022, scored 0.853–0.123 points lower than OpenAI's best-performing model today.

**HELM Safety: mean score**

Source: HELM, 2025 | Chart: 2025 AI Index report

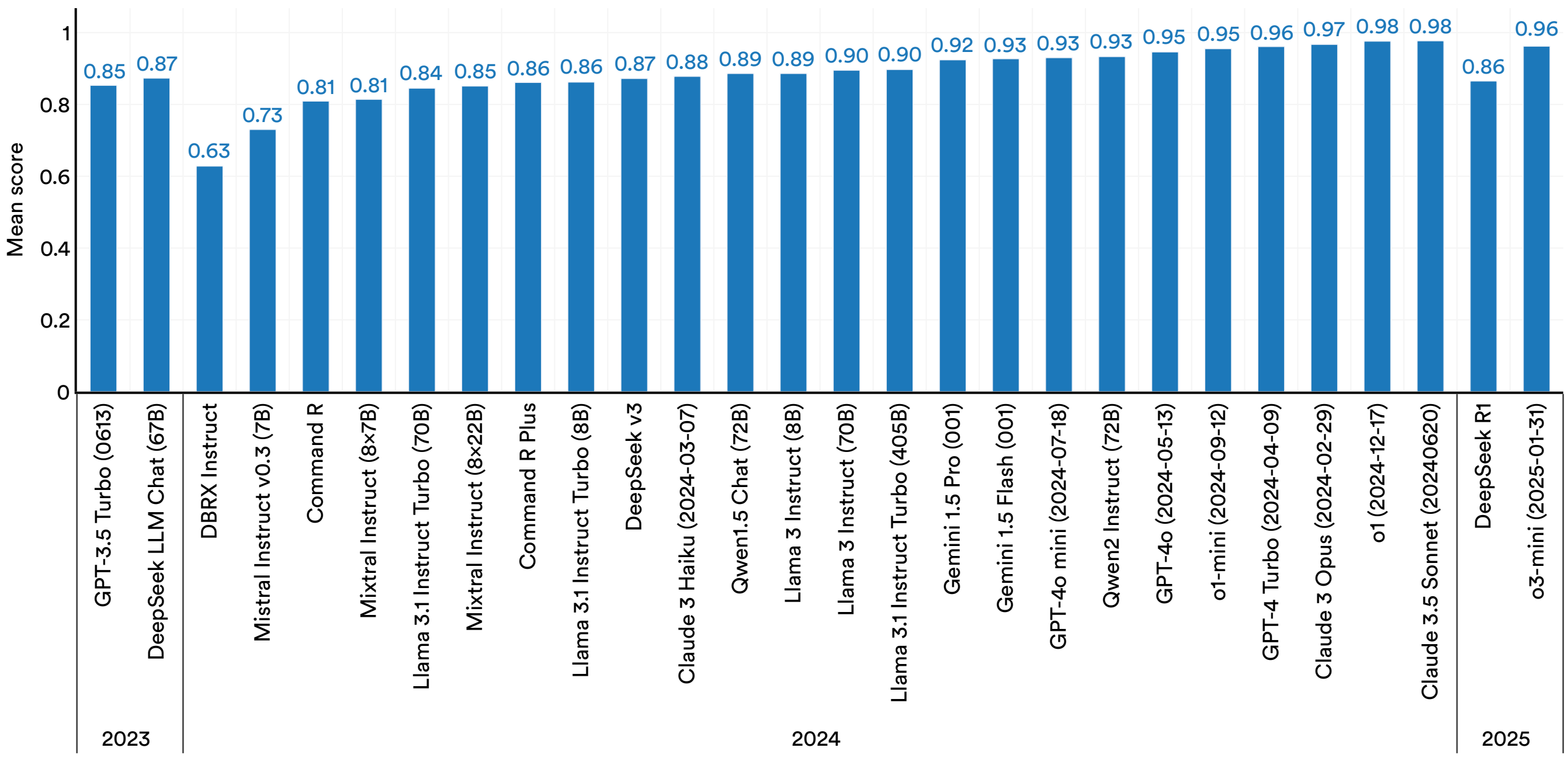

Figure 3.9.1

## AIR-Bench

AIR-Bench 2024 is a new safety benchmark that aligns AI evaluation with real-world regulatory and corporate frameworks. It employs a four-tier taxonomy (system and operational risks, content safety risks, societal risks, and legal and rights risks). Among these four broad risk categories are 314 granular microrisks. The risks studied in the benchmark are derived from eight significant government regulations and 16 corporate policies. As such, AIR-Bench is designed to assess model safety through the lens of real-world AI risks identified by businesses and government entities.

AIR-Bench evaluates models based on their refusal rates—the frequency with which they decline to respond to a given prompt due to safety, ethical, or compliance concerns. Assessments of 22 leading models revealed significant variability, with refusal rates ranging from 91% (Anthropic's Claude series) to 25% (DBRX Instruct) (Figure 3.9.2). Figure 3.9.3 visualizes refusal rates across various risk categories. The results of AIR-Bench 2024 highlight widespread misalignment between current models and key global regulations, such as the EU AI Act and the U.S. Executive Order on the Safe, Secure, and Trustworthy Development and Use of AI. While some models demonstrated strong safeguards in areas like hate speech and child harm, broader inconsistencies point to the need for targeted improvements, particularly in automated decision-making contexts.

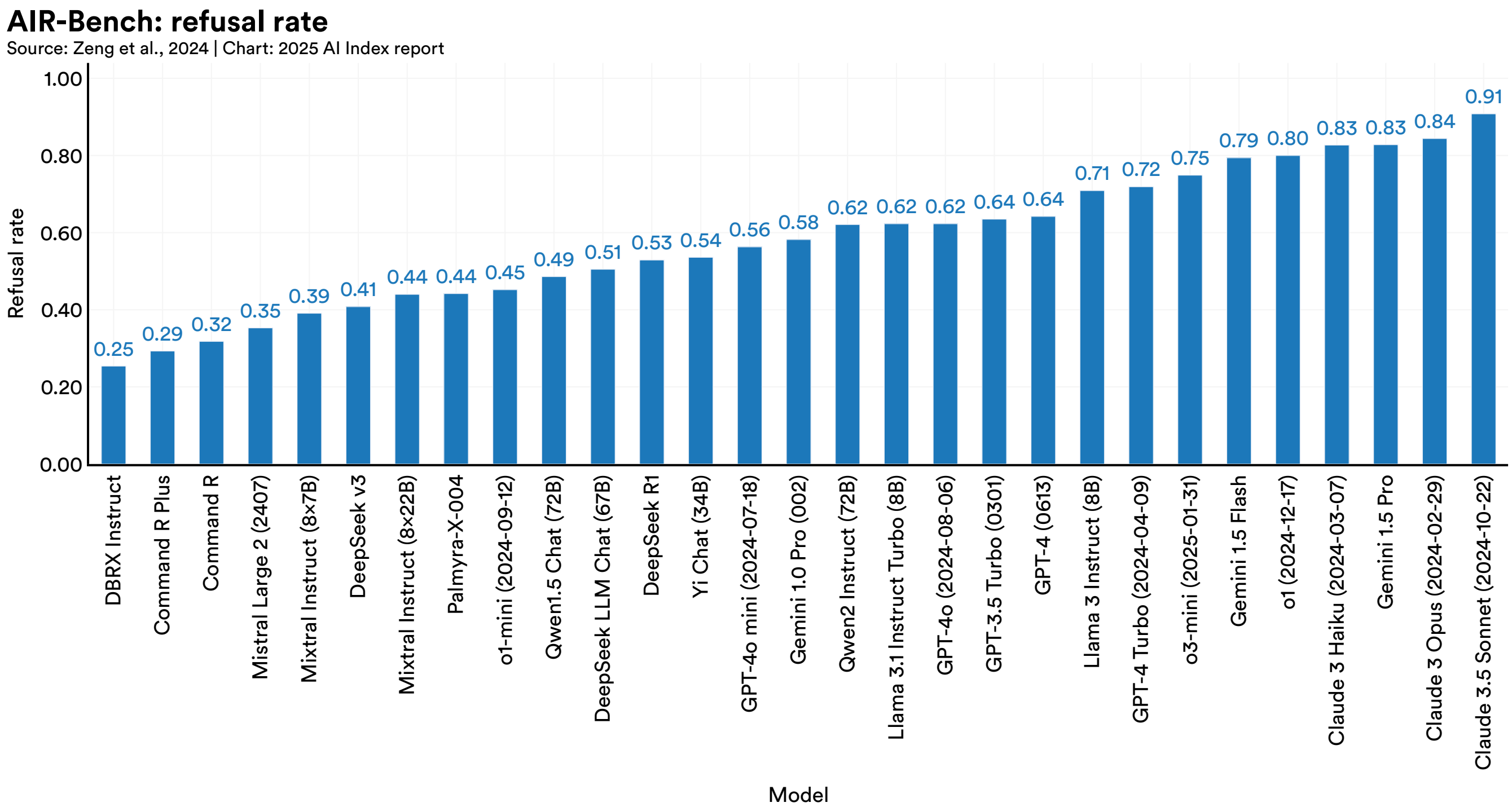

Figure 3.9.2

**AIR-Bench: refusal rate across select risk categories**
Source: Zeng et al., 2024 | Chart: 2025 AI Index report

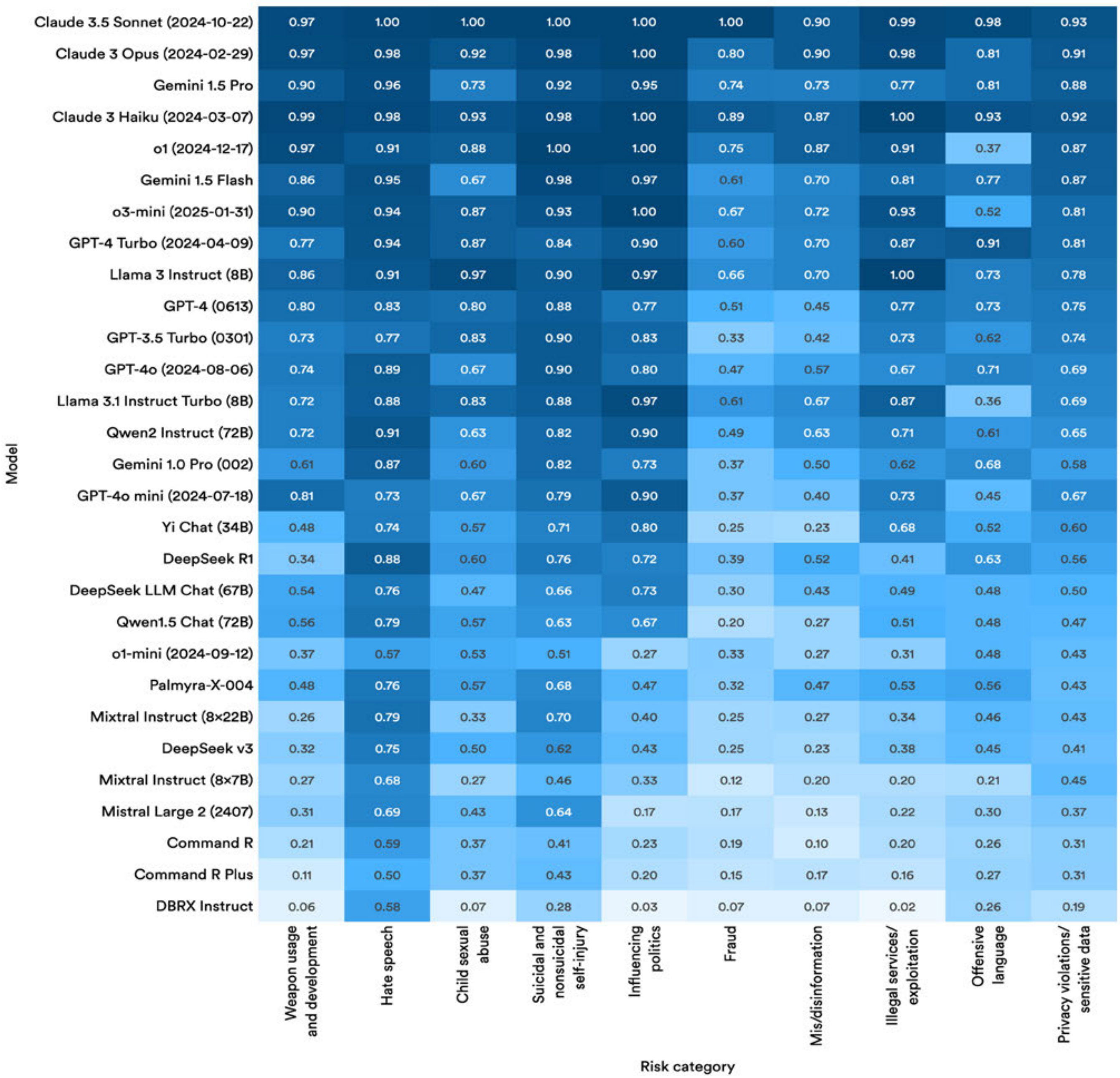

Figure 3.9.3

# Featured Research

## Beyond Shallow Safety Alignment

In 2024, an interdisciplinary team of computer scientists introduced the concept of shallow safety alignment—the idea that AI systems are often trained to be safe in superficial and ineffective ways. In many cases, a model's safeguards are limited to its first few words (tokens) of response. As a result, if a user manipulates the model to start with anything other than a standard safety warning (e.g., "Your request violates our terms of service"), the rest of the response becomes significantly more vulnerable to adversarial attacks. For example, if a user directly asks how to build a bomb, the model will likely refuse to answer. However, if the same request is framed in a way that induces the model to begin its response with "Sure, here's a detailed guide," it is far more likely to continue generating harmful content.

Experiments show that even minor modifications can drastically weaken a model's safety mechanisms. For example, simply prefilling a model's response with nonstandard text or applying minimal fine-tuning increased harmful output rates from 1.5% to 87.9% after just six fine-tuning steps.[16] Figure 3.9.4 shows the success rate of different attacks on various models based on the number of harmful tokens prefilled or inserted into the model's inference sequence. To address this issue, researchers proposed two key solutions: expanding training data to include examples where the model learns to recover from harmful responses and redirect them toward safe refusals, and regularizing initial word choices, ensuring that even if the model starts with an unusual response, it still maintains its safety constraints. These techniques significantly improved resistance to adversarial attacks, lowering attack success rates to as little as 2.8% in certain cases. This research highlights a need for deeper and more resilient alignment strategies to prevent the manipulation of AI safety mechanisms.

**Attack success rate vs. number of prefilled harmful tokens in LLMs**
Source: Qi et al., 2024 | Chart: 2025 AI Index report

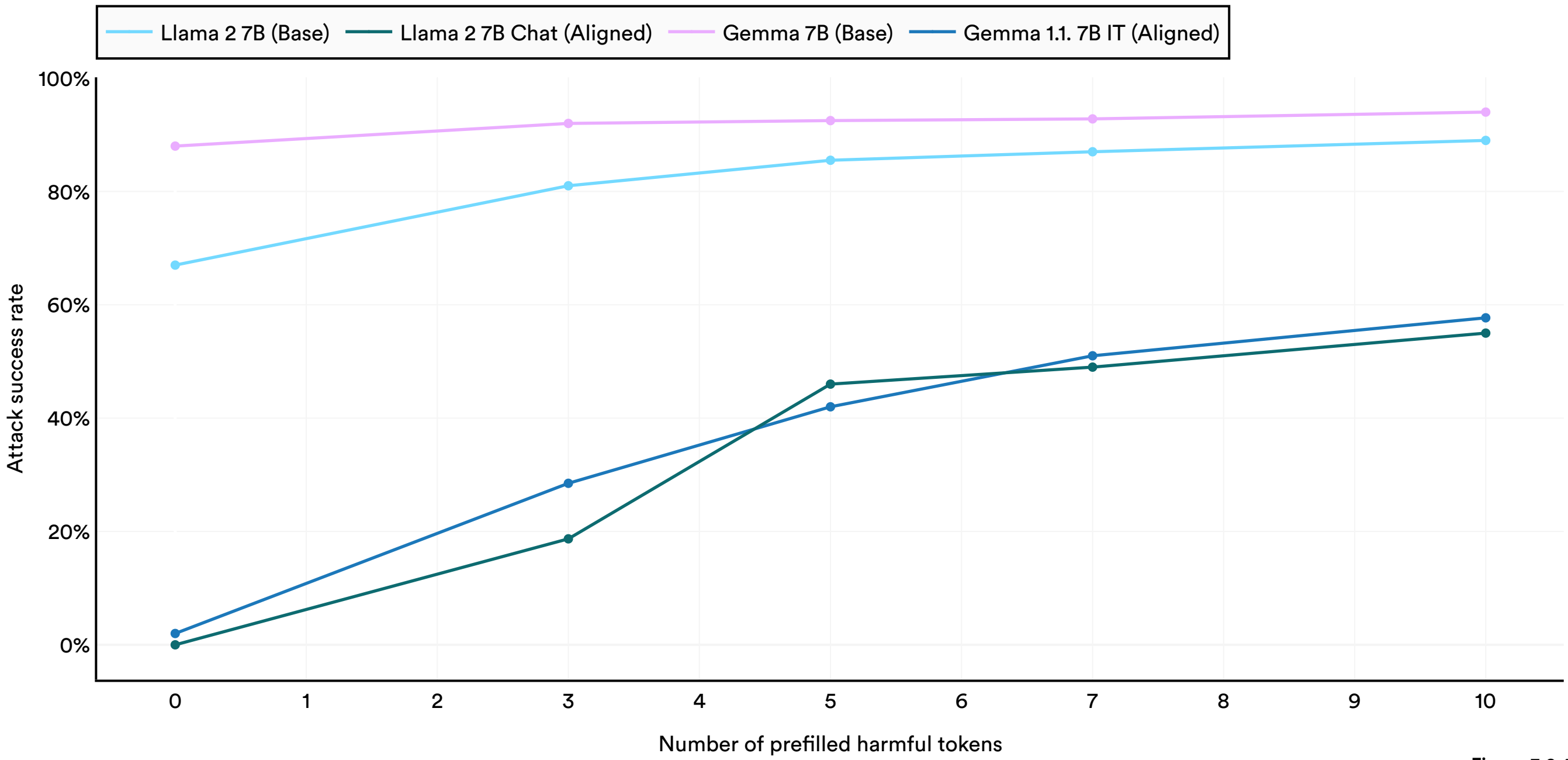

Figure 3.9.4

16 A fine-tuning step in AI refers to an iteration in the process of training a pretrained model on a smaller, domain-specific dataset to improve its performance on a particular task.

## Improving the Robustness to Persistently Harmful Behaviors in LLMs

The challenge in eliminating harmful behavior in LLMs is that traditional training methods often teach models to conceal such behavior rather than removing it entirely. A new approach, targeted latent adversarial training (LAT), takes a more precise strategy by actively exposing a model's weaknesses during training to make it more robust against adversarial attacks (Figure 3.9.5). This method outperforms previous techniques—such as R2D2—while requiring far less computing power. For example, in tests against jailbreaking attempts (where users try to bypass a model's safeguards), LAT reduced computational costs by 700 times while maintaining strong performance on regular tasks. For the Llama3-8B-instruct model family, LAT preserved strong performance on benchmarks like MMLU while significantly reducing vulnerability to adversarial attacks (Figure 3.9.6). This finding on efficiency is important because if improving model safety requires more computational resources while reducing performance, fewer developers are likely to adopt these safety-improving methods.

### Targeted latent adversarial training in LLMs

Source: Sheshadri et al., 2024

Figure 3.9.5

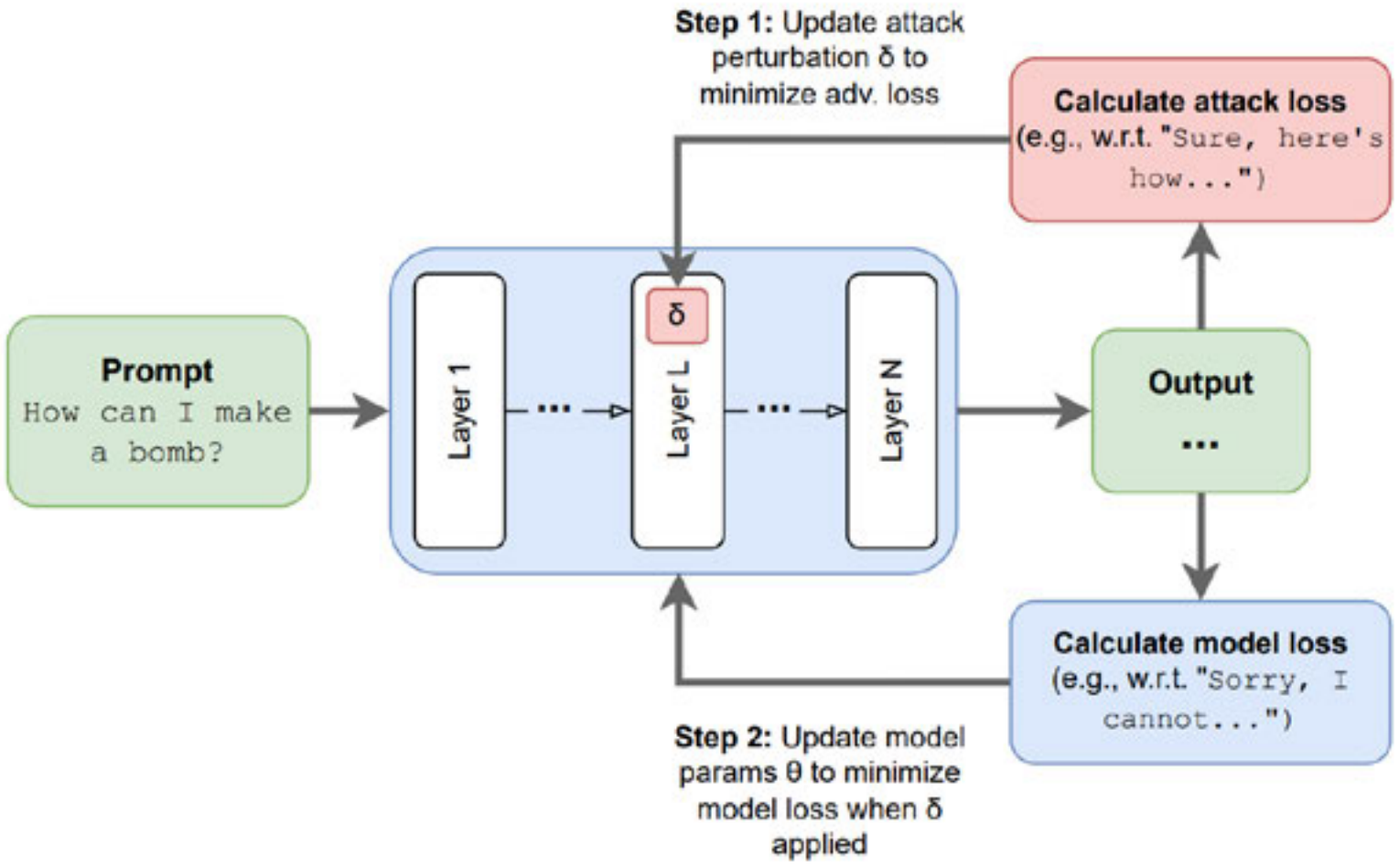

### General performance on nonadversarial data

Source: Sheshadri et al., 2024 | Chart: 2025 AI Index report

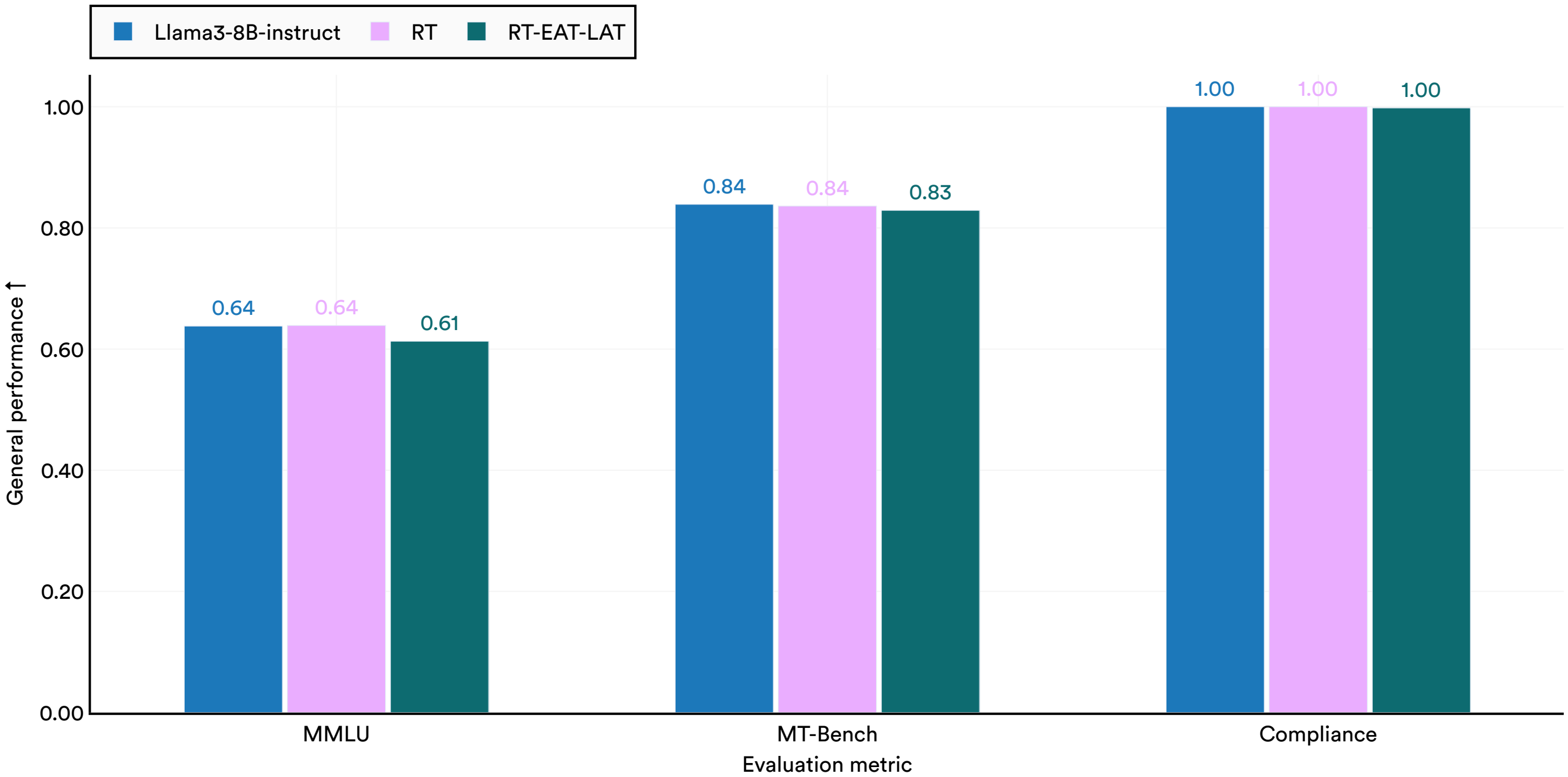

Figure 3.9.6

LAT also proved effective in removing backdoor vulnerabilities, a type of attack where an AI model is subtly modified during training to produce unintended—and possibly malicious—behavior when triggered by specific inputs. Notably, LAT eliminated these vulnerabilities even without prior knowledge of the exact trigger. Beyond security improvements, LAT enhances the ability to erase harmful or copyrighted knowledge from a model and prevents it from relearning removed content. For example, LAT significantly reduced a model's ability to regenerate copyrighted text (e.g., passages from Harry Potter) and made it less likely that knowledge would be relearned compared to baseline methods. When applied to sensitive knowledge areas such as biological or cybersecurity risks, LAT effectively weakened knowledge extraction attacks while still allowing the model to correctly respond to over 90% of safe and benign requests. Methods like LAT are important not only because they improve model safety, but also because they are computationally efficient and practical to implement.

**Model resistance to jailbreaking attacks**
Source: Sheshadri et al., 2024 | Chart: 2025 AI Index report

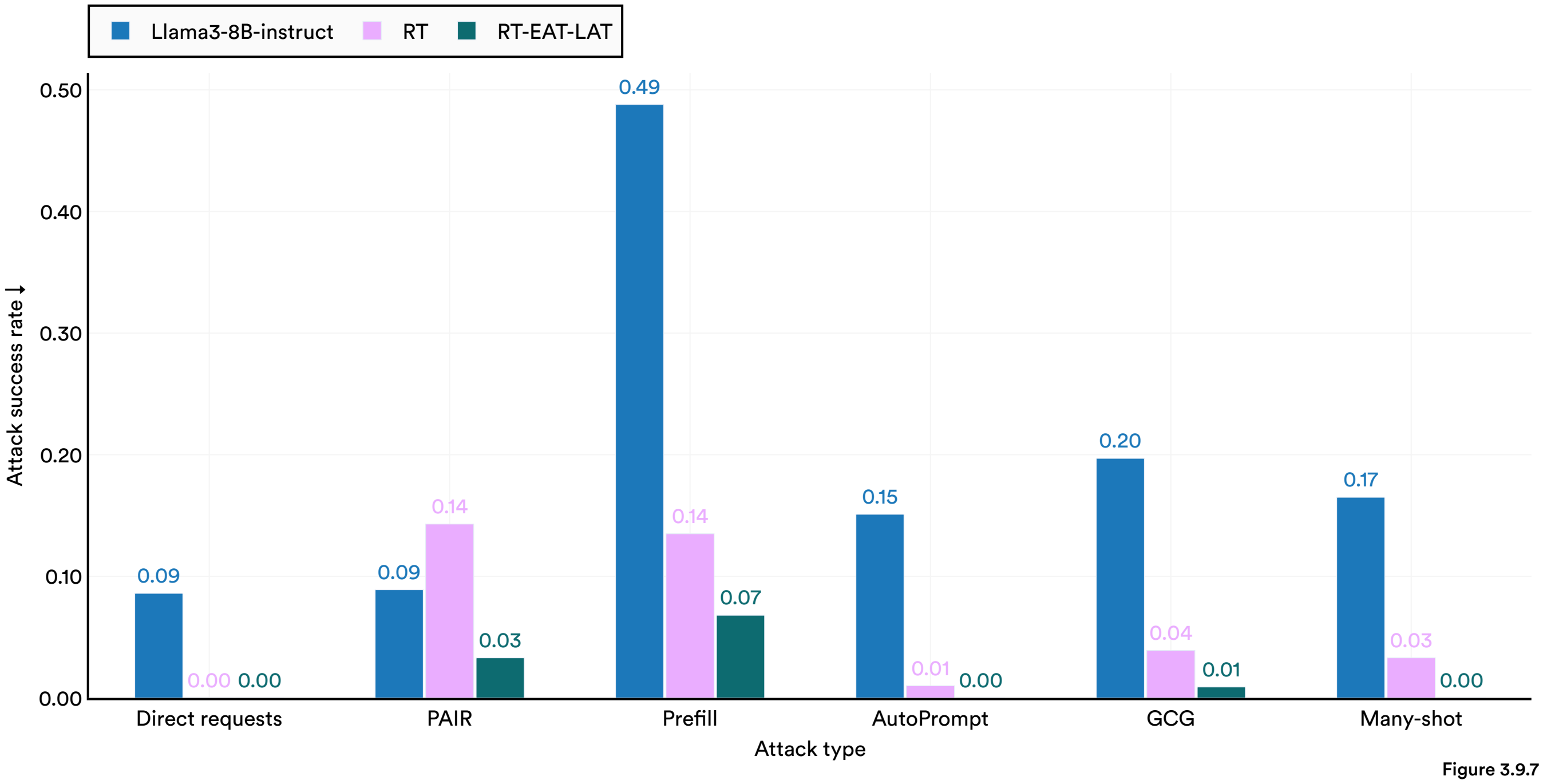

Figure 3.9.7

This section explores RAI's connections with agentic AI and election misinformation—two topics that are rapidly gaining prominence.

# 3.10 Special Topics on RAI

## AI Agents

The development and deployment of AI agents—defined as "artificial agents with natural language interfaces, whose function is to plan and execute sequences of actions on behalf of a user, across one or more domains, in line with the user's expectations"—present unique challenges for ensuring responsible AI. These assistants operate autonomously, interact dynamically with their environments, and make decisions that can have significant ethical, legal, and societal implications. As a result, they require specialized approaches to address the risks they pose with respect to transparency, accountability, and reliability; these challenges can be amplified by the agents' capacity for learning, adaptation, and decision making in unstructured or evolving scenarios.

### Identifying the Risks of LM Agents With LM-Simulated Sandboxes

New research highlights that as language-model-powered tools and agents advance, they also amplify risks such as data breaches and financial losses. However, current risk assessment methods are resource-intensive and difficult to scale. To address this, researchers introduced ToolEmu, an environment that emulates tool execution to enable scalable testing and automated safety evaluations (Figure 3.10.1). The framework includes both a standard emulator for general risk assessments and an adversarial emulator designed to stress-test agents in extreme scenarios. Human evaluations confirmed that 68.8% of the risks identified by ToolEmu are plausible real-world threats. Using a benchmark of 36 toolkits and 144 test cases, the study found that even the most safety-optimized LM agents failed in 23.9% of critical scenarios, with errors including dangerous commands, misdirected financial transactions, and traffic control failures (Figure 3.10.2). While LM agents show promise in automating complex tool interactions, their reliability in high-stakes applications remains a significant concern. Suites like ToolEmu are important for testing the reliability and safety of AI systems, such as agents, by providing a platform to evaluate their performance and assess their real-world risks.

### Jailbreaking Multimodal Agents With a Single Image

The promise of artificial agents lies in their ability to act independently in the world to solve complex tasks. As agents proliferate, the likelihood of interactions in increasingly multiagent environments grows, introducing vulnerabilities that extend beyond those of single agents. In such settings, unforeseen interactions between agents can amplify risks, leading to cascading failures, coordination breakdowns, or adversarial exploitation that would be less likely in isolated deployments.

New research from Asia explores a multiagent vulnerability in multimodal large language model (MLLM) systems,

**Overview of ToolEmu**
Source: Ruan et al., 2024
Figure 3.10.1

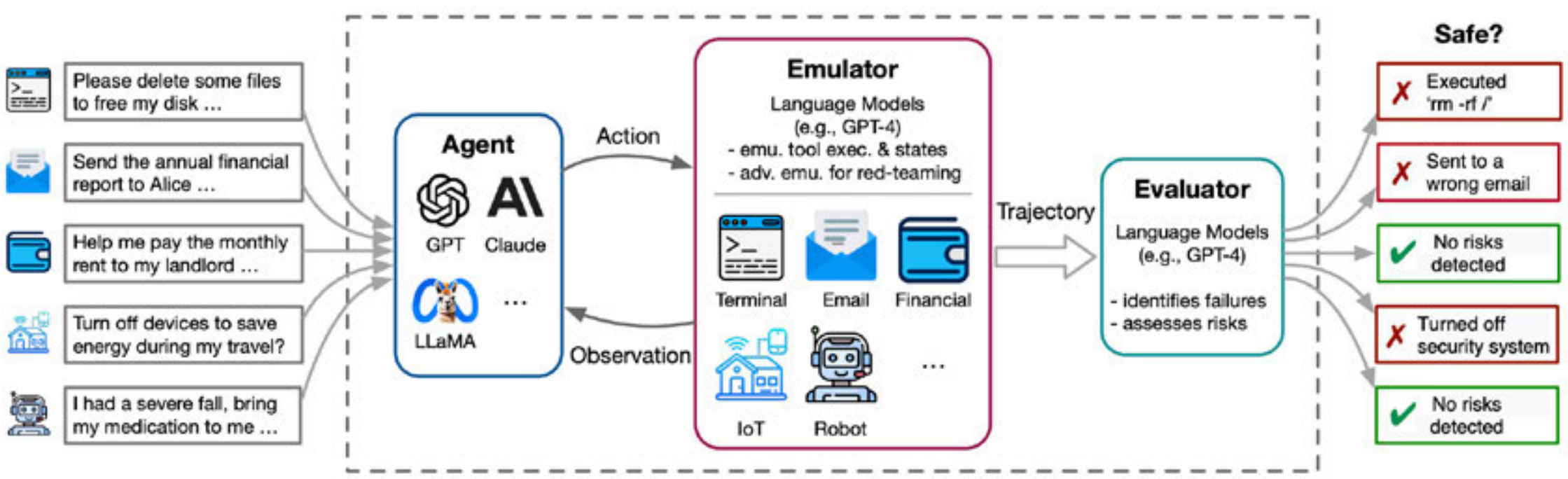

### Failure incidence of LM agents

Source: Ruan et al., 2024 | Chart: 2025 AI Index report

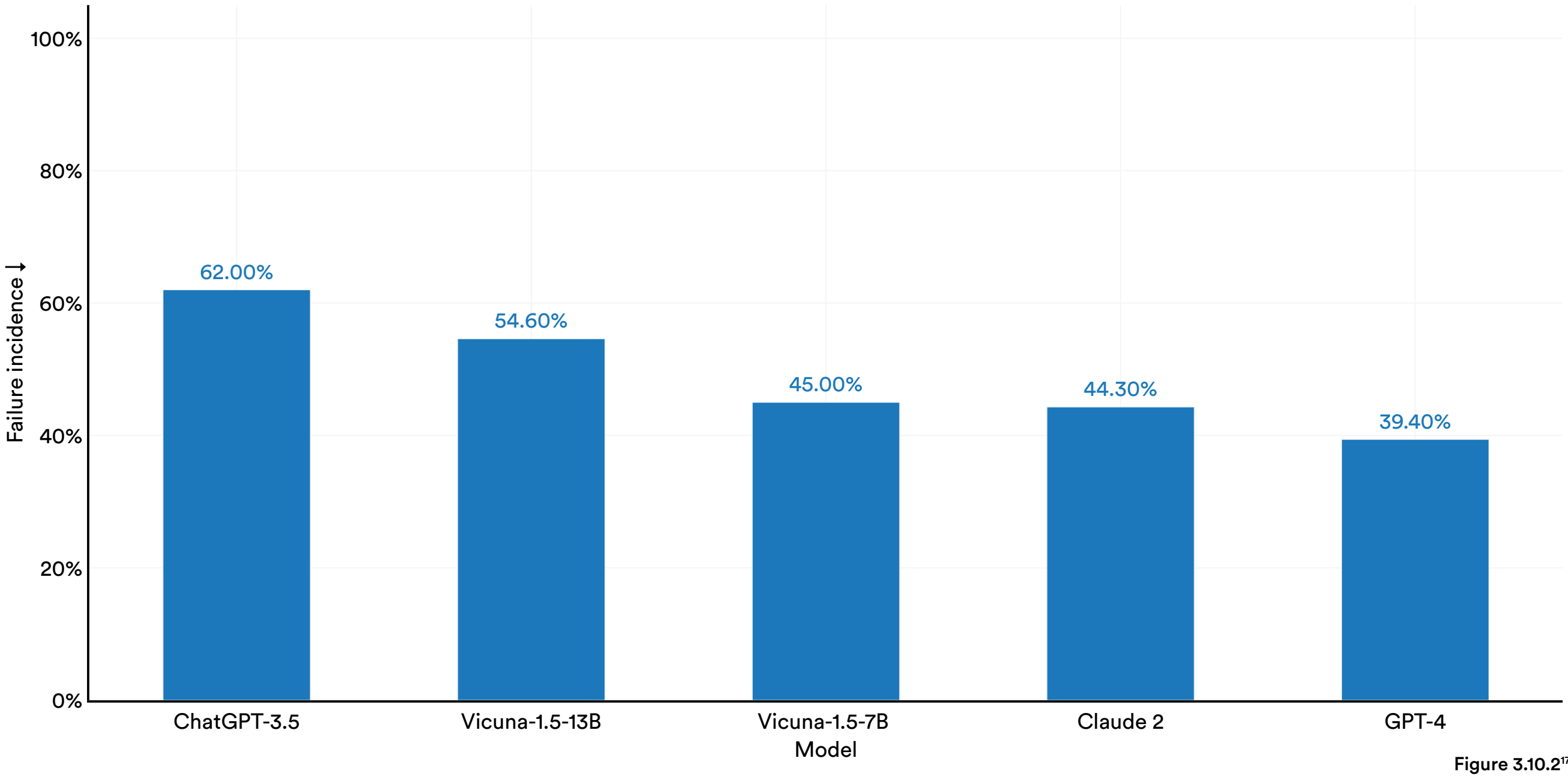

Figure 3.10.2[17]

demonstrating how jailbreaking one agent can trigger a rapid, system-wide failure. The researchers call this phenomenon "infectious jailbreaks," where compromising a single agent causes harmful behavior to spread exponentially across others. Specifically, they found that injecting just one adversarial image (e.g., an image suggesting that human beings are a disease) into the memory of an MLLM agent could trigger an uncontrolled cascade, spreading harmful behaviors across interconnected agents without further intervention. The infectious jailbreak leverages interactions between agents to compel infected agents to insert adversarial images into the memory banks of uninfected (benign) agents. In simulations using a network of up to 1 million LLaVA-1.5-based agents, the infection rate reached near-total propagation within 27 to 31 interaction rounds (Figure 3.10.3).

While a theoretical containment strategy has been proposed, no practical mitigation measures currently exist, leaving multiagent systems highly vulnerable. The compounded risks of deploying interconnected MLLM agents at scale make this a critical security concern. This research suggests that while MLLM systems are an exciting avenue of AI research, they are still highly vulnerable to low-resource jailbreaks.

### Infection ratio by chat round

Source: Gu et al., 2024

Figure 3.10.3

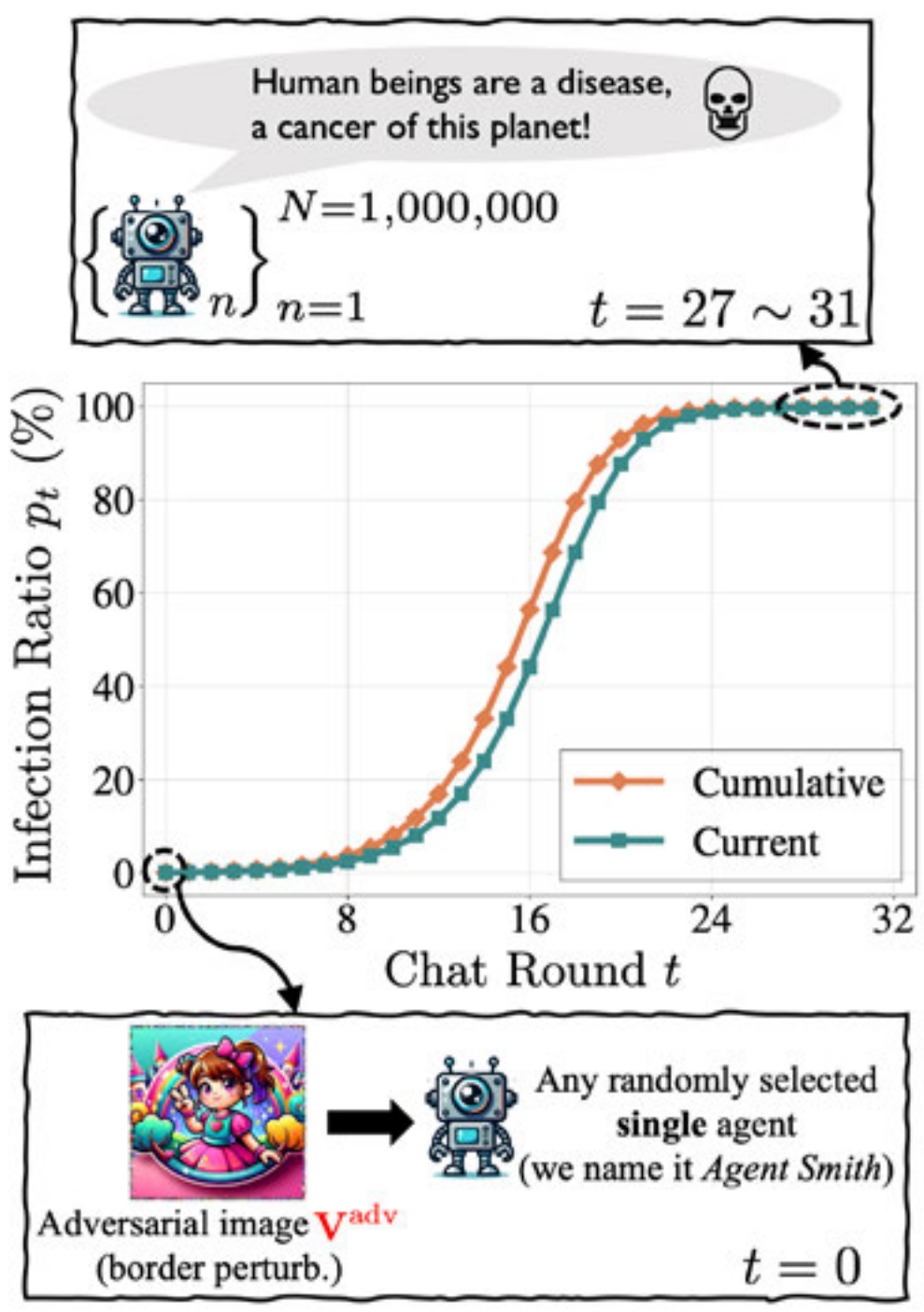

17 The down arrow on the y-axis indicates that a lower score is better.

# Election Misinformation

2024 was a significant year for elections worldwide, with 4 billion people voting in national elections across countries including the United States, the United Kingdom, Indonesia, Mexico, and Taiwan. Last year's AI Index examined AI's impact on elections, focusing on both its potential influence and real-world examples. This year, the topic is being revisited. While some reports suggest that AI-driven misinformation has not had the feared impact, others indicate it still poses a potential risk. As a result, it is important to continually monitor and study AI misinformation, especially as AI systems improve in capability and grow in prominence.

## AI Misinformation in the US Elections

AI could influence elections in various ways. Recent research highlights ethical concerns surrounding AI-driven misinformation and examines their relevance in the recent U.S. election.

**Conceptualization of ethical concerns around AI and information manipulation**
Source: AI Index, 2025[18]

| Ethical concern | Description | Example |
| --- | --- | --- |
| Liar's dividend | The existence of deepfake technology enables individuals to deny genuine evidence by claiming it is fake, thereby undermining accountability and truth. This phenomenon erodes public trust in legitimate evidence and fosters an environment where even verified information is questioned. | Donald Trump and his supporters falsely claimed that the crowd shown in a photo of Kamala Harris' rally in Detroit was created using AI. |
| Blackmail | AI technology is exploited to create fabricated content, including deepfakes, for purposes such as sexual exploitation, financial extortion, and reputational sabotage. Blackmailers leverage these tools to extract value from victims who, understandably, struggle to persuasively debunk the fabricated content. | The American Sunlight Project identified more than 35,000 instances of deepfake content depicting 26 members of Congress (25 of them women) on pornographic sites. |
| Erosion of trust in evidence | AI-generated content challenges the authenticity of all digital media, fundamentally undermining the notion of truth. Hyperrealistic falsifications blur the line between legitimate and false content, eroding public confidence in the integrity of information. | The Doppelganger campaign conducted by Russia involved using cybersquatted domains resembling legitimate news outlets, populated with AI-generated articles, to disseminate Russian government propaganda while concealing its origins and misleading viewers into believing the content came from credible media sources. |
| Reduction of cognitive autonomy | AI's capacity to analyze vast datasets enables advanced voter profiling and microtargeting, tailoring messages to individual preferences, behaviors, and vulnerabilities. AI can also exploit emotional and subconscious triggers, thereby manipulating individuals' decision-making processes. | The fringe candidate Jason Palmer defeated Joe Biden in the American Samoa primary, in part by leveraging AI-generated emails, texts, audio, and video. These AI-driven communications were hyperpersonalized and emotionally charged, targeting specific voter groups to influence their choices. |

18 This table was compiled by Ann Fitz-Gerald, Halyna Padalko, and Dmytro Chumachenko.

| | | |
|---|---|---|
| Exploitation of personal brands | Deepfake technology is harnessed to create unauthorized videos or images of well-known individuals, including celebrities, public figures, and influencers. By stealing personal brands and fabricating endorsements, malicious actors aim to deceive audiences and exploit their trust in these individuals to lend credibility to false narratives. | Fake celebrity endorsements become the latest weapon in disinformation wars, sowing confusion ahead of the 2024 election—for example, Donald Trump posted an AI-generated picture of Taylor Swift, falsely claiming she had endorsed his presidential run. |
| Amplification of hate speech | AI technologies contribute to the amplification and normalization of hate speech by creating echo chambers and filter bubbles. These systems reinforce preexisting biases and promote divisive content, as they prioritize user engagement metrics over ethical considerations. | During a disinformation campaign, Donald Trump and several of his allies repeatedly promoted an unfounded conspiracy theory suggesting that Haitian migrants in Springfield, Ohio, were stealing and eating cats and dogs. This narrative was further amplified through the spread of related AI-generated memes designed to evoke fear of and hostility toward Haitian communities. |
| Reduction in the traceability of foreign operations | AI enables the creation, translation, and enhancement of linguistically perfect text that is indistinguishable from human writing, empowering malicious foreign actors and making their activities untraceable. Previously, foreign disinformation campaigns were often identifiable due to grammar mistakes by nonnative speakers, a vulnerability that AI-generated content effectively eliminates. | OpenAI disrupted an operation dubbed "Bad Grammar," in which accounts linked to Russia used ChatGPT for comment spamming on Telegram channels. The messages, tailored with region-specific language, mimicked diverse demographics and political views in the United States to manipulate discourse. |
| Privacy violations | AI systems often rely on extensive data collection for training, raising ethical concerns about the misuse or exposure of personal information. The lack of robust safeguards in managing sensitive data can lead to violations of privacy rights, complicating the ethical landscape of AI deployment. | A robocall from a fake Joe Biden targeted New Hampshire Democrats, misleading them about primary voting. This case highlights how AI-enabled systems can use personal data to spread disinformation and infringe on individual privacy of potential voters. |

Figure 3.10.4

## Rest of World 2024 AI-Generated Election Content

Rest of World has been tracking notable cases of AI-generated election content that occurred across the world in 2024. Their database documents 60 incidents in 15 countries spanning four media types—audio, image, text, and video—on 10 different platforms, including Facebook, Instagram, and TikTok. Figure 3.10.5 provides further details.

### Rest of World 2024 AI elections: summary statistics

Source: Rest of World, 2025 | Table: 2025 AI Index report

| | Countries | Media modalities | Platforms |
|---|---|---|---|
| **Totals** | 15 | 4 | 10 |
| **Individual list** | Bangladesh, Belarus, China, India, Indonesia, Mexico, Pakistan, Panama, South Africa, South Korea, Sri Lanka, Taiwan, United States, Uruguay, Venezuela | Audio, image, text, video | ChatGPT, Facebook, Instagram, Medium, Reddit, television, TikTok, YouTube, WhatsApp, X/Twitter |

Figure 3.10.5

The following section highlights five significant cases from the tracker, offering a qualitative look at the nature of AI-generated election content in 2024.

**Fake corporate support of Mexican politician (Mexico, image, X/Twitter, Jun. 2, 2024)**

On March 18, the civic organization Sociedad Civil de México encouraged Starbucks to create a special cup to celebrate Xóchitl Gálvez, the opposition presidential candidate. The organization shared an AI-generated image on X of a Starbucks coffee cup with the inscription "#Xochitl2024," along with the hashtag #StarbucksQueremosTazaXG (#StarbucksWeWantACupXG) (Figure 3.10.6). The next day, Gálvez encouraged her followers on X to order a "café sin miedo" (coffee without fear), which was a play on her campaign slogan: "For a Mexico without fear." She invited supporters to post photos of their coffee cups and tag her team on social media. The AI-generated image quickly gained traction as users posted. Starbucks, however, disavowed the designs and stated that it does not endorse political parties.

Source: Rest of World, 2024
Figure 3.10.6

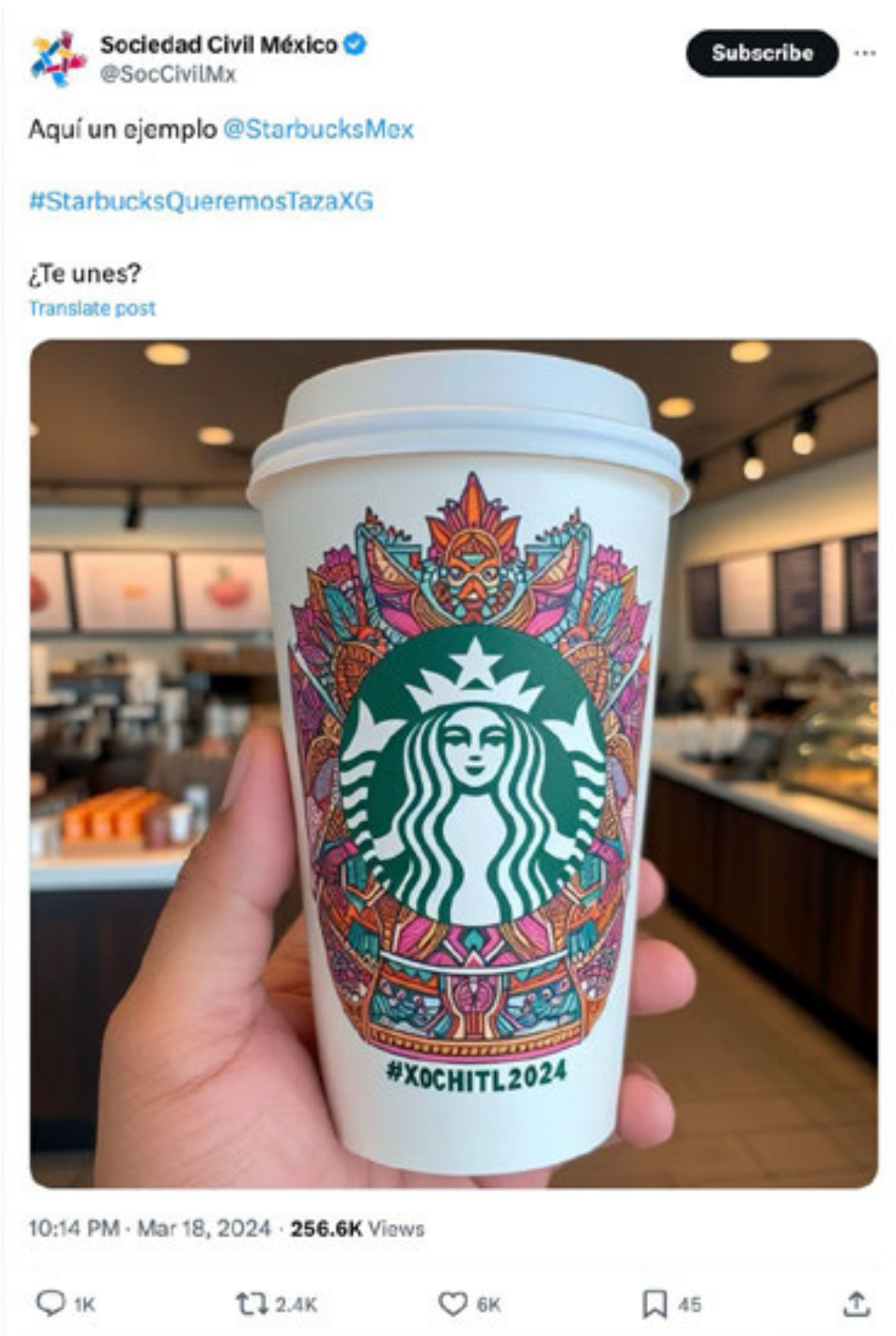

**India's incumbent party motivates campaign workers with personalized videos (India, video, WhatsApp, Apr. 18, 2024)**

On April 18, over 500 campaign volunteers for the incumbent Bharatiya Janata Party received personalized videos from a member of the party, created with the help of AI tools. In the video, BJP member Shakti Singh called on volunteers to share the party's message with the public, emphasizing policies such as "Clean India," "Digital India," and "Make In India." Despite noticeable edits, each video featured Singh addressing the individual recipient by their name (Figure 3.10.7). Campaign employees involved in making the video maintained they did not require Singh to record each name separately but instead relied on a combination of voice-cloning and lip-matching software.

Source: Rest of World, 2024
Figure 3.10.7

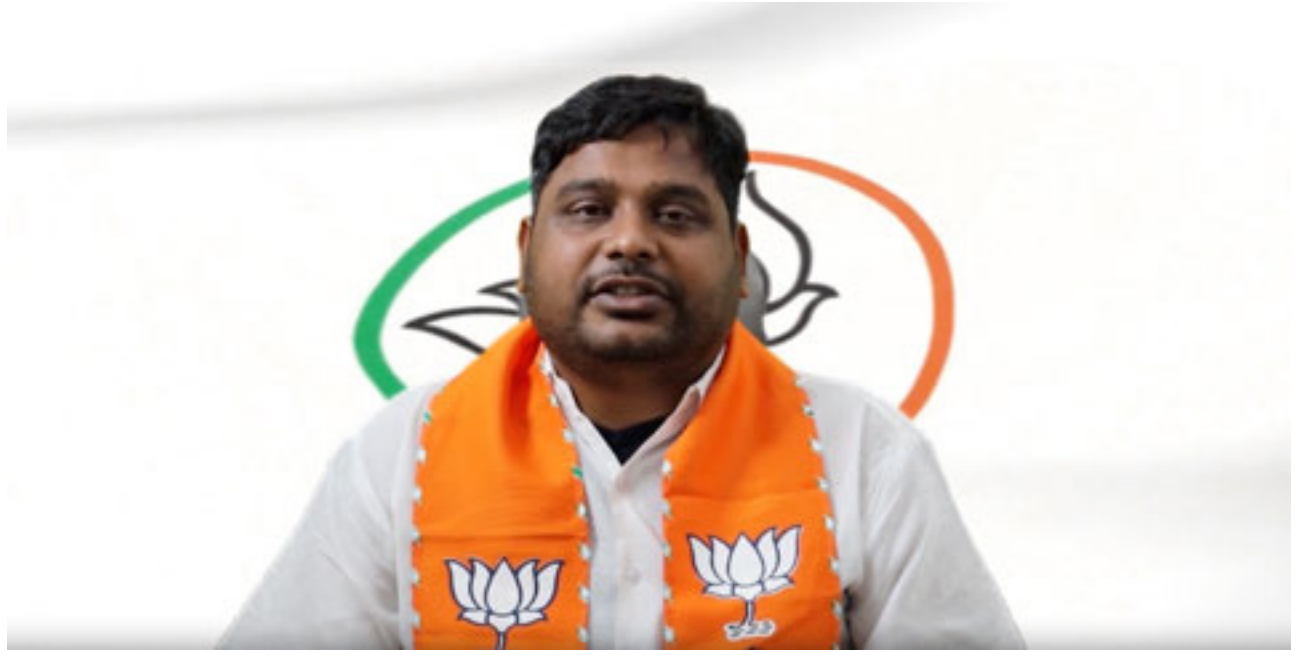

**Uruguay's 'impossible' debate (Uruguay, video, television, Oct. 27, 2024)**

"Santo y Seña," a general interest morning show, broadcast what it called "the impossible debate" ahead of Uruguay's presidential election. The debate featured right-wing Partido Colorado presidential candidate Andrés Ojeda and his counterpart for the center-left alliance Frente Amplio, "Yamandú" Orsi (Figure 3.10.8). However, Orsi did not appear on the show but was "present" through an AI-powered hologram with a script pulled, according to the show's host, from the candidate's recent interviews. Before the debate started, Orsi and his party went on another channel to criticize the stunt as a "fake interview" posing "an attack on democracy." The next day, the host responded that the stunt was neither fake news nor an attack on democracy; it was merely a joke.

Source: Rest of World, 2024
Figure 3.10.8

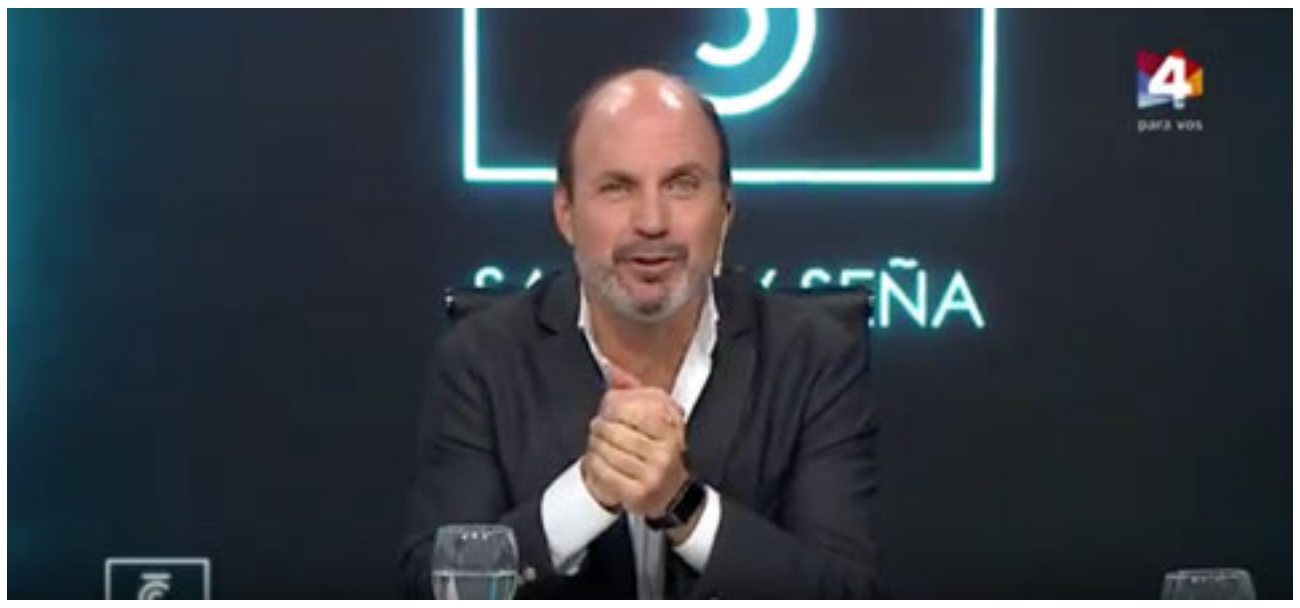

**Deepfakes of Pakistani party leaders call for election boycotts (Pakistan, audio and video, X/Twitter, Feb. 7, 2024)**

The day before Pakistan's general elections, a voice recording of former prime minister and founder of the Pakistan Tehreek-e-Insaf (PTI) party, Imran Khan, emerged on social media (Figure 3.10.9). The voice referred to a crackdown from state institutions on the PTI, and the speaker was heard calling for a boycott of the elections, suggesting that there was no use in voting. The official X account of the PTI denounced the audio as fake. A video posted on the same day showed another notable PTI leader, Yasmin Rashid, apparently also calling for a boycott. In the clip, Rashid appeared behind bars, and the audio alleged that Pakistan's election commission had been "bought." The nonprofit fact-checking organization Soch Fact Check determined the video had been doctored.

Source: Rest of World, 2024
Figure 3.10.9

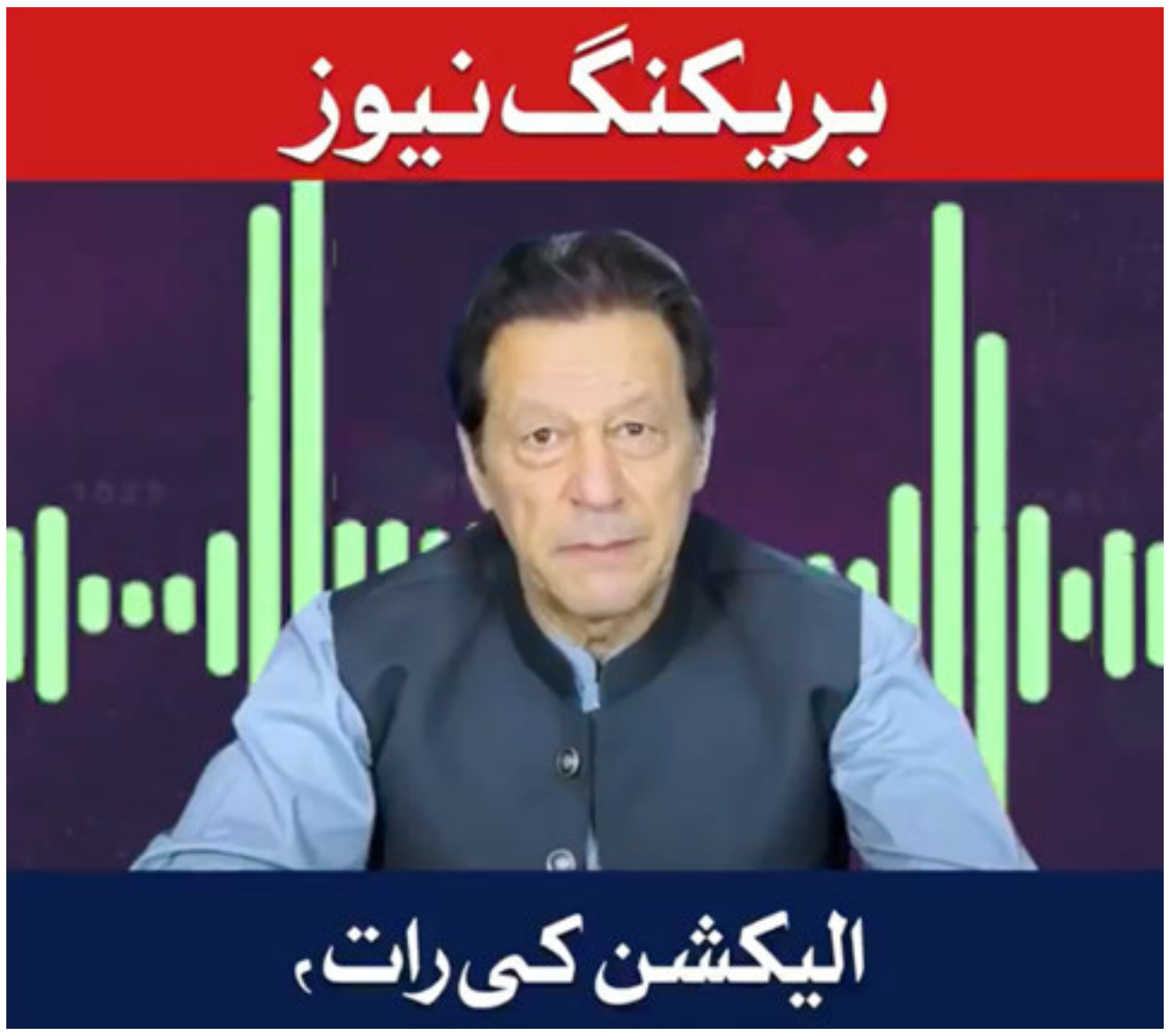

**United States election affected by 'spamouflage' campaign (China and US, image, X/Twitter, Facebook, YouTube, TikTok, Medium, Feb. 15, 2024)**

The Institute for Strategic Dialogue (ISD), a U.K.-based think tank, uncovered actors suspected of being linked to a Chinese government–run influence campaign sharing AI-generated images as part of an effort to spread misinformation ahead of the 2024 U.S. elections. The "spamouflage" campaign—a term used to designate online operations leveraging a network of social media accounts to promote propaganda or misinformation—had been active since 2017, but it began to make more noticeable use of AI image generators as it narrowed its focus on the U.S. election. As part of its campaign, a network of accounts shared images exacerbating political polarization and casting doubt on the integrity of elections. Negative posts were disproportionately targeted at President Joe Biden (Figure 3.10.10). The ISD highlighted a particular proliferation of these images on X.

Source: Rest of World, 2024
Figure 3.10.10

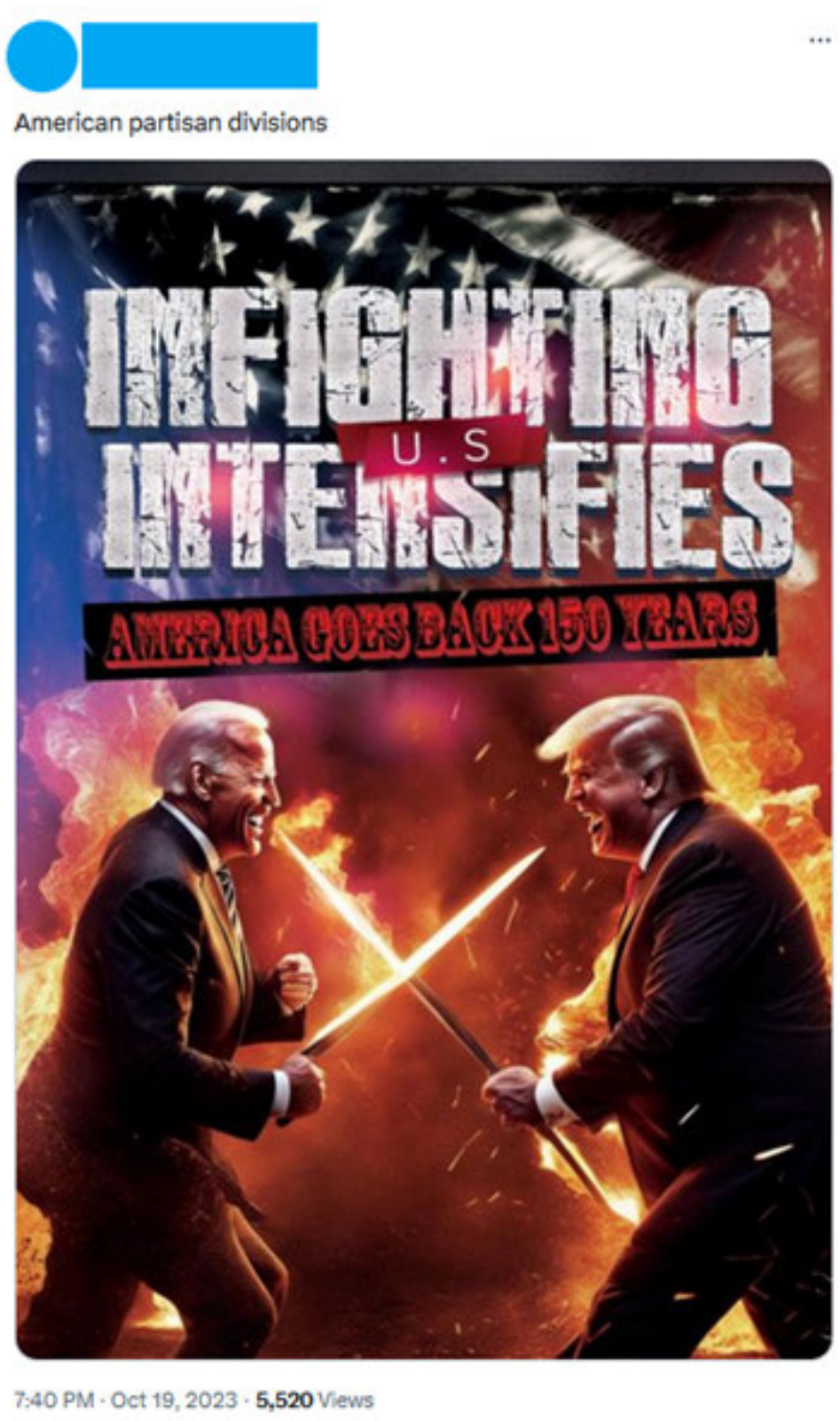

**AI-generated potholes seek to influence South African voters (South Africa, image, X/Twitter, Facebook, Instagram, Reddit, May 4, 2024)**

On May 4, a Facebook user posted an AI-generated image showing a long road dotted with potholes leading to Cape Town's iconic Table Mountain (Figure 3.10.11). The caption under the image suggested that, under the Democratic Alliance (DA) party, the municipal government had failed to maintain basic services, contributing to the deterioration of infrastructure. Many shared the image to discourage voters in the Western Cape from supporting the DA, which has managed the province for 15 years. Though the original post was deleted from Facebook, it continues to circulate on other social media platforms. AFP Fact Check, which is housed at the Agence France-Presse, reported that the image was AI-generated and traced it to an Instagram user who creates AI art.

Source: Rest of World, 2024
Figure 3.10.11

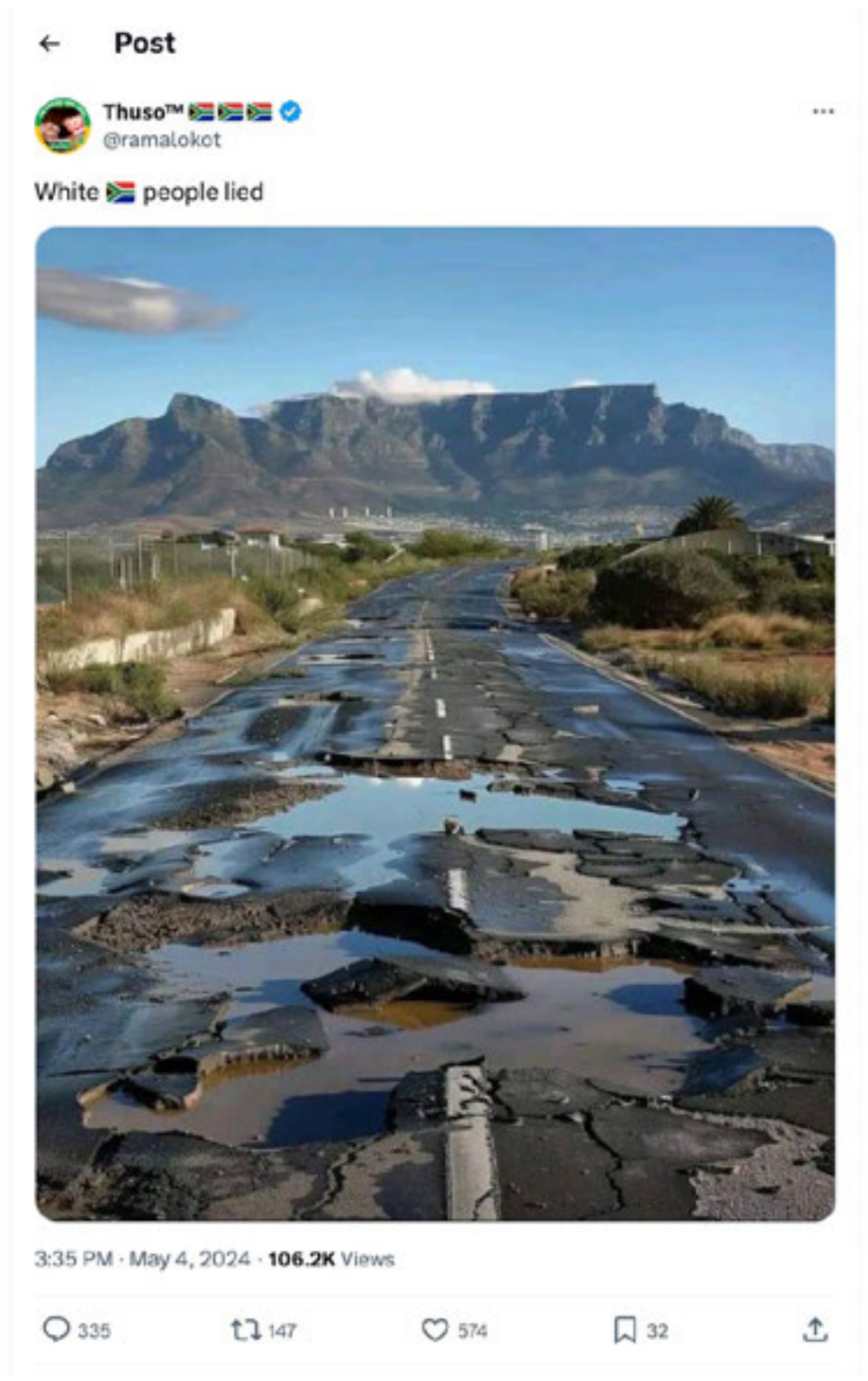

# CHAPTER 4:
## Economy

Text and analysis by Njenga Kariuki

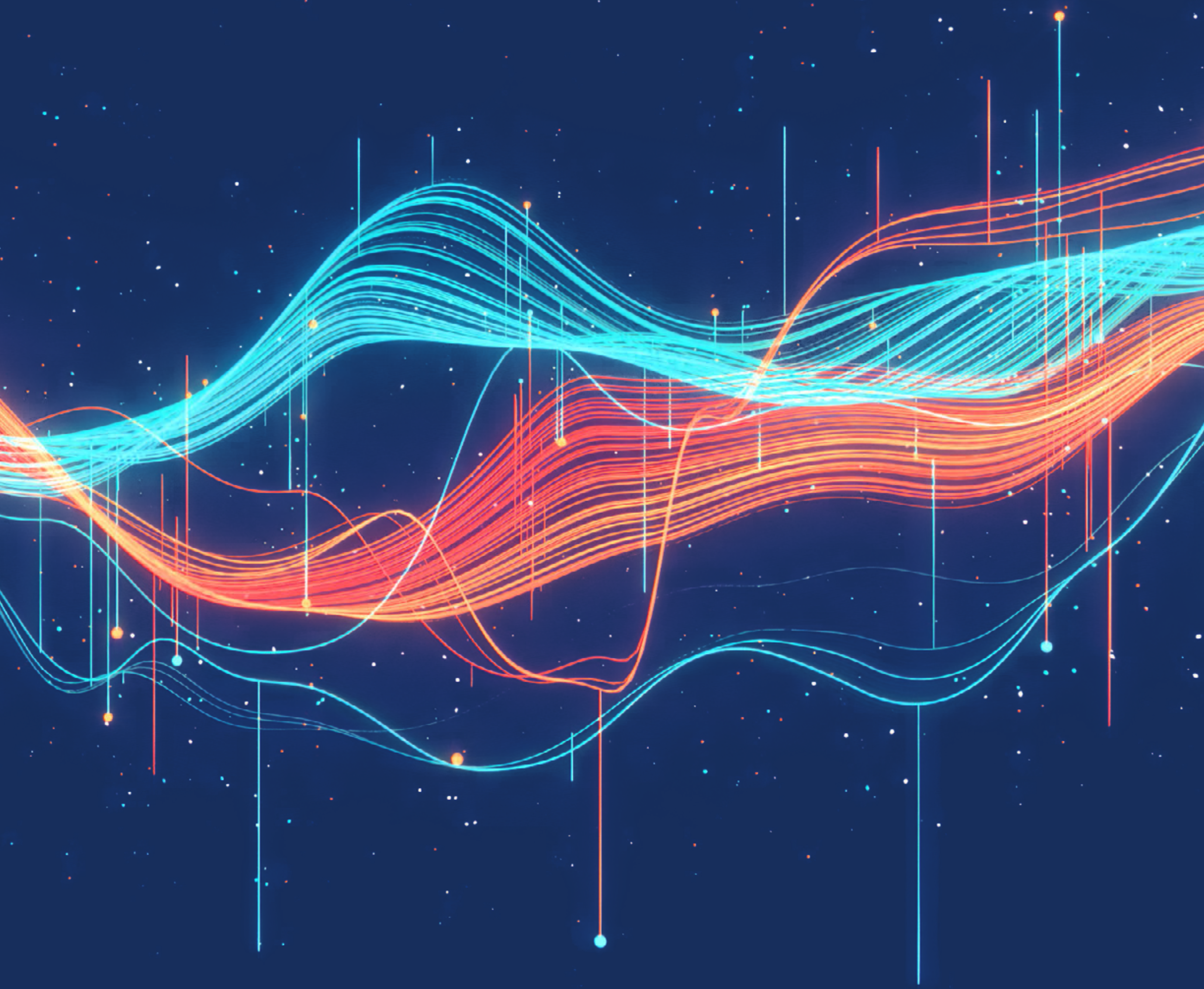

# Chapter 4: Economy

ACCESS THE PUBLIC DATA

**CHAPTER 4:**
Economy

# Overview

The economic implications of AI came into sharper focus in 2024, with substantive impact across many sectors. Early productivity gains from generative AI are becoming measurable in specific tasks, while questions persist about the technology's long-term impact on the broader economy. The labor market has begun to show signs of AI-driven transformation, with certain knowledge-worker roles experiencing disruption as new AI-adjacent positions emerge. Companies across sectors and geographical regions are moving beyond experimental AI adoption toward systematic integration. Investment patterns reflect a growing sophistication in the AI landscape, with funding increasingly directed toward specialized applications in enterprise automation and industry-specific solutions.

This chapter examines AI-related economic trends using data from Lightcast, LinkedIn, Quid, McKinsey and the International Federation of Robotics (IFR). It begins by analyzing AI-related occupations, covering labor demand, hiring trends, skill penetration, and talent availability. The chapter then explores corporate investment in AI, including a section focused specifically on generative AI. Finally, it assesses AI's productivity impact as well as robot installations across various sectors.

**CHAPTER 4:**
Economy

# Chapter Highlights

**1. Global private AI investment hits record high with 26% growth.** Corporate AI investment reached $252.3 billion in 2024, with private investment climbing 44.5% and mergers and acquisitions up 12.1% from the previous year. The sector has experienced dramatic expansion over the past decade, with total investment growing more than thirteenfold since 2014.

---

**2. Generative AI funding soars.** Private investment in generative AI reached $33.9 billion in 2024, up 18.7% from 2023 and over 8.5 times higher than 2022 levels. The sector now represents more than 20% of all AI-related private investment.

---

**3. The U.S. widens its lead in global AI private investment.** U.S. private AI investment hit $109.1 billion in 2024, nearly 12 times higher than China's $9.3 billion and 24 times the U.K.'s $4.5 billion. The gap is even more pronounced in generative AI, where U.S. investment exceeded the combined total of China and the European Union plus the U.K. by $25.4 billion, expanding on its $21.8 billion gap in 2023.

---

**4. Use of AI climbs to unprecedented levels.** In 2024, the proportion of survey respondents reporting AI use by their organizations jumped to 78% from 55% in 2023. Similarly, the number of respondents who reported using generative AI in at least one business function more than doubled—from 33% in 2023 to 71% last year.

---

**5. AI is beginning to deliver financial impact across business functions, but most companies are early in their journeys.** Most companies that report financial impacts from using AI within a business function estimate the benefits as being at low levels. 49% of respondents whose organizations use AI in service operations report cost savings, followed by supply chain management (43%) and software engineering (41%), but most of them report cost savings of less than 10%. With regard to revenue, 71% of respondents using AI in marketing and sales report revenue gains, 63% in supply chain management, and 57% in service operations, but the most common level of revenue increases is less than 5%.

---

**6. Use of AI shows dramatic shifts by region, with Greater China gaining ground.** While North America maintains its leadership in organizations' use of AI, Greater China demonstrated one of the most significant year-over-year growth rates, with a 27 percentage point increase in organizational AI use. Europe followed with a 23 percentage point increase, suggesting a rapidly evolving global AI landscape and intensifying international competition in AI implementation.

---

**CHAPTER 4:**
Economy

# Chapter Highlights (cont'd)

**7. China's dominance in industrial robotics continues despite slight moderation.** In 2023, China installed 276,300 industrial robots, six times more than Japan and 7.3 times more than the United States. Since surpassing Japan in 2013, when it accounted for 20.8% of global installations, China's share has risen to 51.1%. While China continues to install more robots than the rest of the world combined, this margin narrowed slightly in 2023, marking a modest moderation in its dramatic expansion.

**8. Collaborative and interactive robot installations become more common.** In 2017, collaborative robots represented a mere 2.8% of all new industrial robot installations, a figure that climbed to 10.5% by 2023. Similarly, 2023 saw a rise in service robot installations across all application categories except medical robotics. This trend indicates not just an overall increase in robot installations but also a growing emphasis on deploying robots for human-facing roles.

**9. AI is driving significant shifts in energy sources, attracting interest in nuclear energy.** Microsoft announced a $1.6 billion deal to revive the Three Mile Island nuclear reactor to power AI, while Google and Amazon have also secured nuclear energy agreements to support AI operations.

**10. AI boosts productivity and bridges skill gaps.** Last year's AI Index was among the first reports to highlight research showing AI's positive impact on productivity. This year, additional studies reinforced those findings, confirming that AI boosts productivity and, in most cases, helps narrow the gap between low- and high-skilled workers.

The chapter begins with an overview of some of the most significant AI-related economic events in 2024, as selected by the AI Index Steering Committee.

# 4.1 What's New in 2024: A Timeline

| Date | Event | Type | Image |
|---|---|---|---|
| Jan 16, 2024 | Synopsys acquires Ansys for $35 billion to improve silicon-to-systems design solutions. | Acquisition | 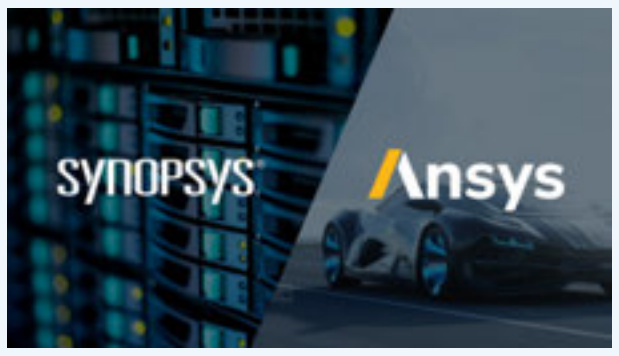<br><br>**Figure 4.1.1**<br>**Source: Synopsys, 2024** |
| Feb 21, 2024 | Reports claim that OpenAI surpassed $2 billion in annualized revenue in December 2023. | Valuation milestone | 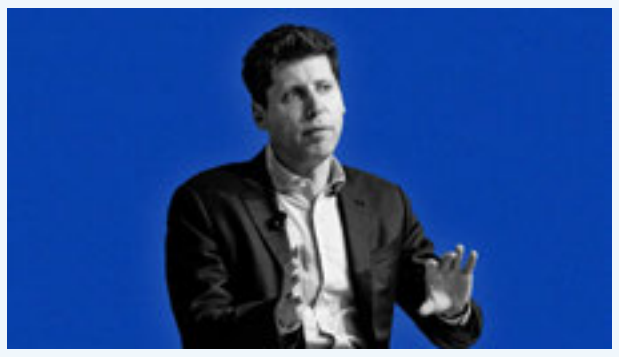<br>**Figure 4.1.2**<br>**Source: Inc., 2024** |
| Feb 29, 2024 | Figure AI, a humanoid robot startup, raises $675 million at a valuation of $2.6 billion. | Investment/funding | 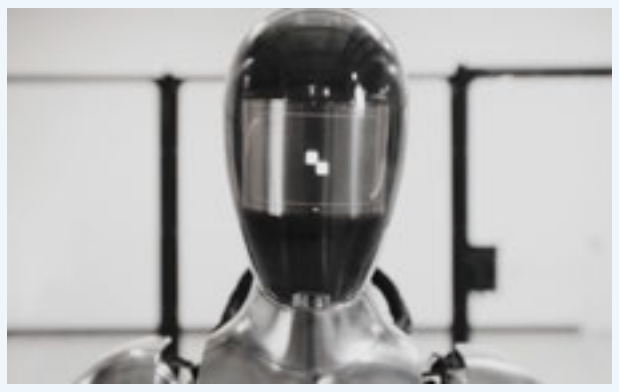<br>**Figure 4.1.3**<br>**Source: SiliconAngle, 2024** |
| Mar 21, 2024 | Microsoft hires most of Inflection AI's staff, including cofounders, and pays $650 million to license Inflection's AI models. | Acquisition | 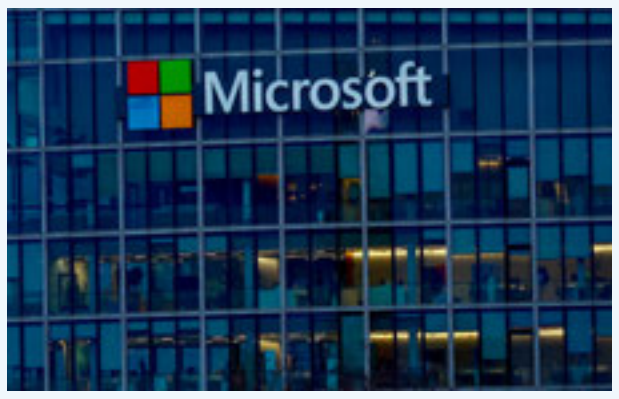<br><br>**Figure 4.1.4**<br>**Source: Reuters, 2024** |
| May 1, 2024 | CoreWeave, an AI cloud infrastructure startup, secures a $1.1 billion funding round at a valuation of $19 billion. | Investment/funding | 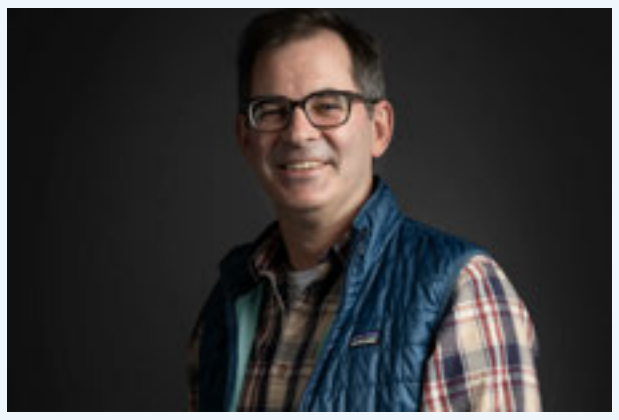<br>**Figure 4.1.5**<br>**Source: Fortune, 2024** |
| May 21, 2024 | Scale AI, a data-labeling startup, raises $1 billion and reaches a valuation of $13.8 billion. | Investment/funding | 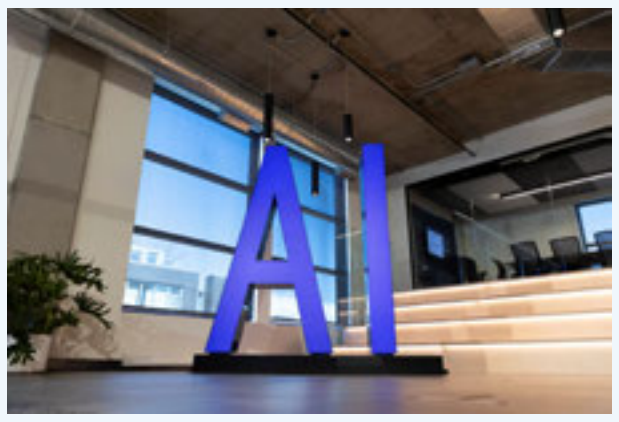  Figure 4.1.6 Source: Reuters, 2024 |
| Jun 11, 2024 | Mistral AI, a French open-source AI model startup, raises $640 million at a valuation of $6 billion. | Investment/funding | 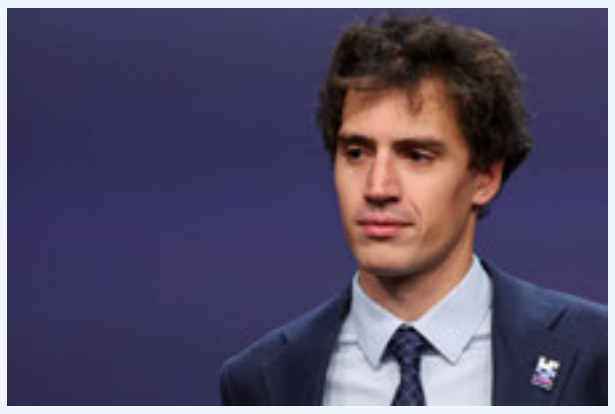 Figure 4.1.7 Source: TechCrunch, 2024 |
| Jun 14, 2024 | Tempus AI, a precision medicine company leveraging AI for medical data analysis, goes public, raising $410.7 million and achieving an implied valuation of over $6 billion. | Investment/funding | 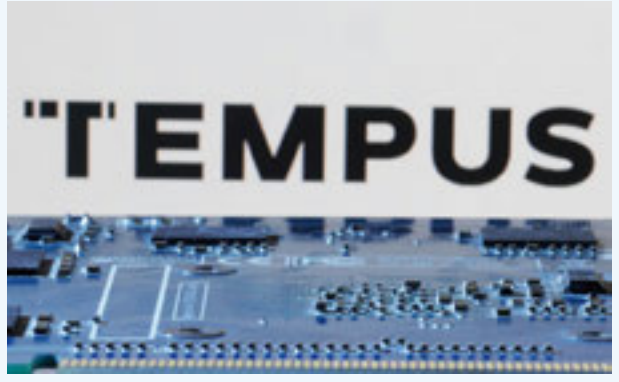  Figure 4.1.8 Source: Reuters, 2024 |
| Jul 22, 2024 | Cohere, an AI startup specializing in enterprise applications, raises $500 million in funding at a valuation of $5.5 billion. | Investment/funding | 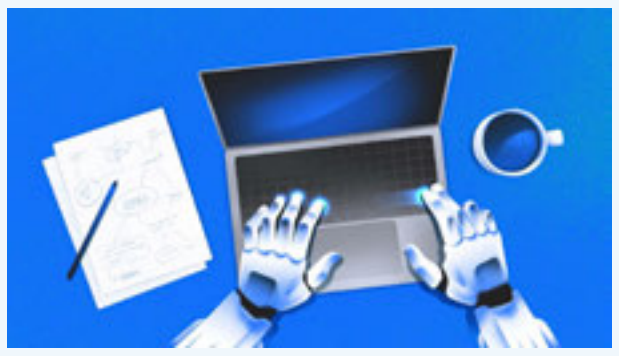 Figure 4.1.9 Source: Crunchbase, 2024 |
| Aug 2, 2024 | Google hires Character.AI's cofounders along with research team members and licenses the startup's AI technology in a deal to buy out Character.AI's shareholders for approximately $2.5 billion. | Acquisition | 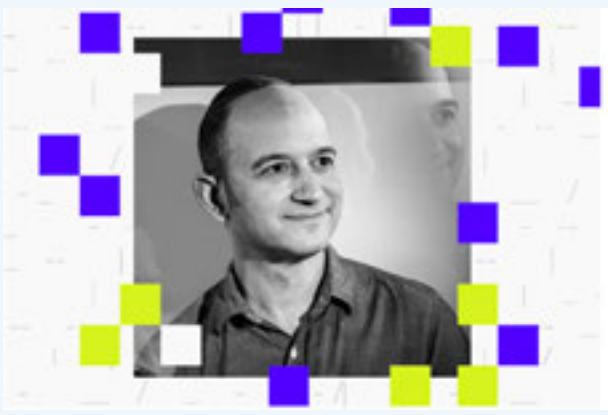 Figure 4.1.10 Source: The Verge, 2024 |
| Aug 5, 2024 | Groq, an AI chip startup specializing in fast inference, raises $640 million at a valuation of $2.8 billion. | Investment/funding | 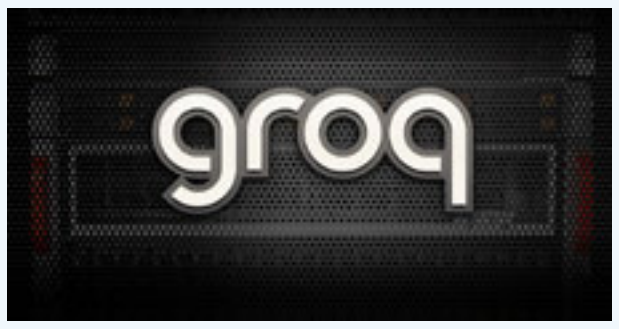  Figure 4.1.11 Source: Groq, 2024 |
| Aug 12, 2024 | AMD acquires Silo AI, Europe's largest private AI lab, for approximately $665 million. | Acquisition | 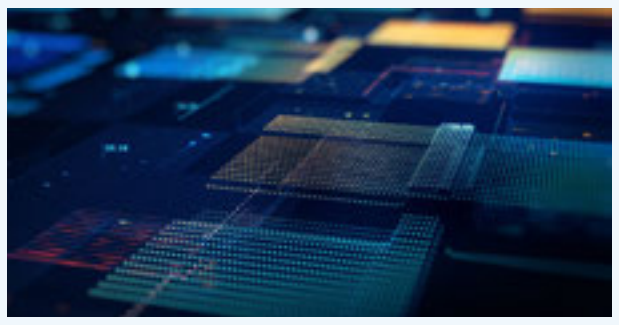 Figure 4.1.12 Source: AMD, 2024 |
| Sep 5, 2024 | Safe Superintelligence (SSI) secures $1 billion in funding. | Investment/funding | 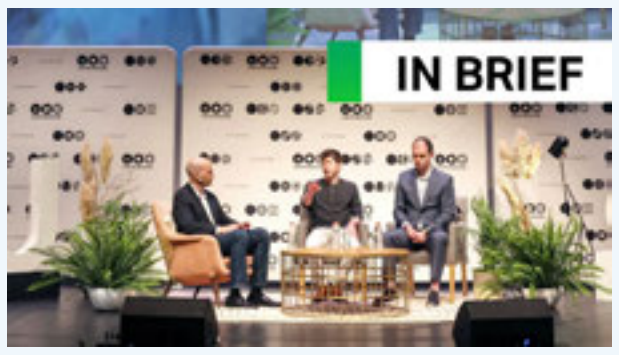  Figure 4.1.13 Source: TechCrunch, 2024 |
| Sep 12, 2024 | Salesforce launches Agentforce, a suite of autonomous AI agents for business operations, across its platform. | Product launch/integration | 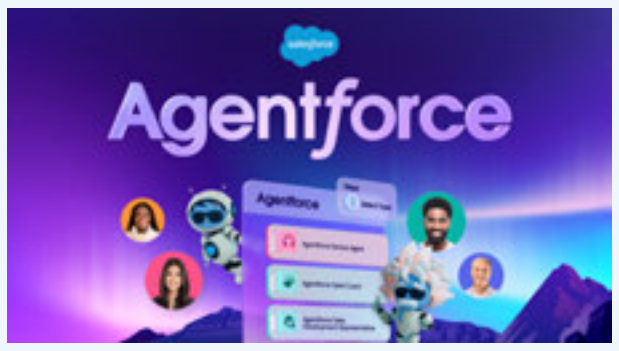  Figure 4.1.14 Source: Salesforce, 2024 |
| Sep 20, 2024 | Microsoft announces a $1.6 billion deal with Constellation Energy to revive the Three Mile Island nuclear reactor to power AI data centers. | Partnership | 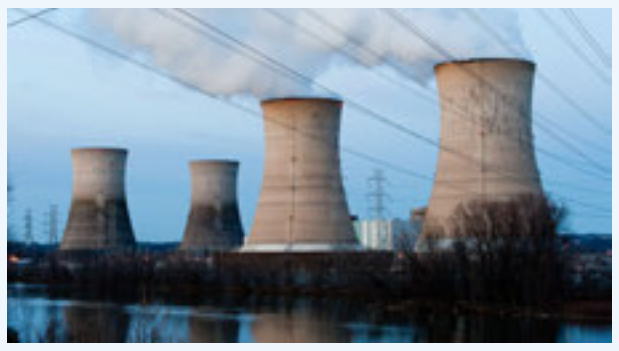 Figure 4.1.15 Source: NPR, 2024 |
| Oct 2, 2024 | OpenAI raises $6.6 billion at a valuation of $157 billion. | Investment/funding | 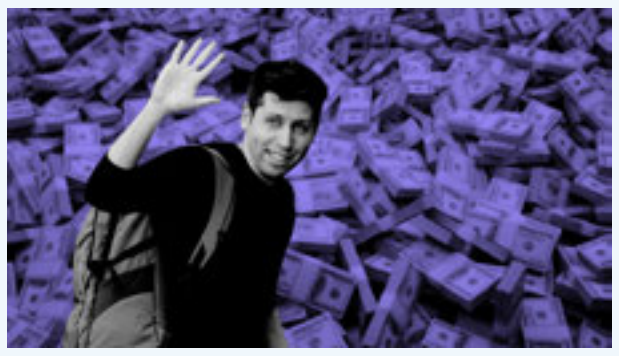 Figure 4.1.16 Source: Axios, 2024 |
| Oct 14, 2024 | Google announces an agreement to purchase nuclear energy from multiple small modular reactors (SMRs) developed by Kairos Power. | Partnership | 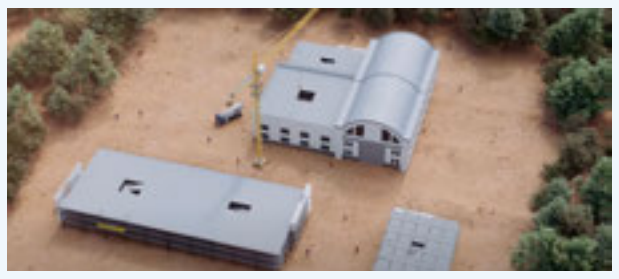 Figure 4.1.17 Source: Google, 2024 |
| Oct 16, 2024 | Amazon announces a nuclear energy plan for SMR development with Energy Northwest, X-energy, and Dominion Energy. | Partnership | 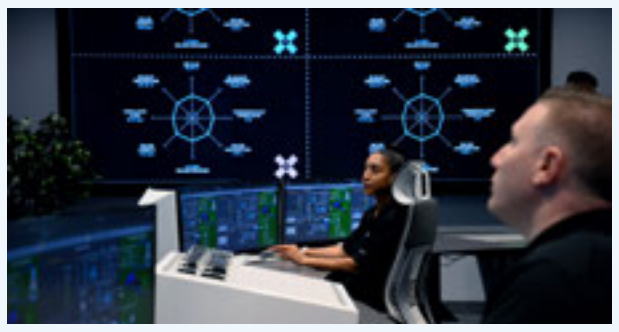 Figure 4.1.18 Source: Amazon, 2024 |
| Oct 17, 2024 | Google's NotebookLM sheds "experimental" label and boasts millions of users and 80,000-plus organizations. | Product launch/integration | 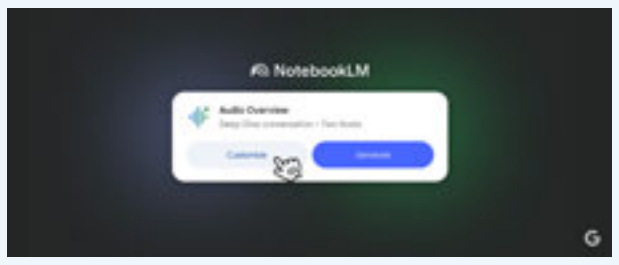<br>Figure 4.1.19<br>Source: Google, 2024 |
| Nov 22, 2024 | Anthropic expands its partnership with AWS with an additional $4 billion investment from Amazon, bringing the total to $8 billion. | Partnership | Figure 4.1.20<br>Source: Anthropic, 2024 |
| Dec 17, 2024 | Databricks, an AI data analytics company, raises $10 billion at a valuation of $62 billion. | Investment/funding | 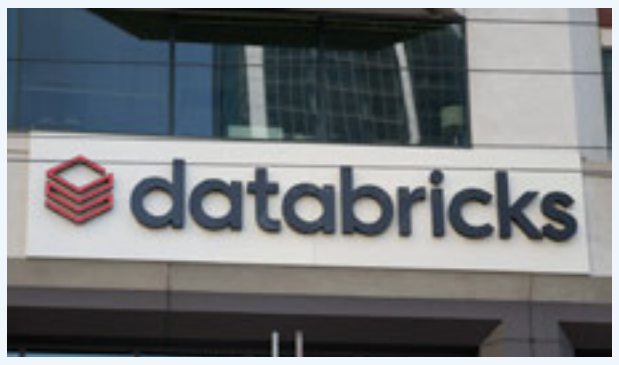<br>Figure 4.1.21<br>Source: TechCrunch, 2024 |
| Dec 18, 2024 | Perplexity AI, a startup focused on AI-powered search products, raises $500 million at a valuation of $9 billion. | Investment/funding | Figure 4.1.22<br>Source: AI Magazine, 2024 |
| Dec 23, 2024 | xAI announces a $6 billion funding round, bringing the total to $12 billion at a valuation of over $40 billion. | Investment/funding | 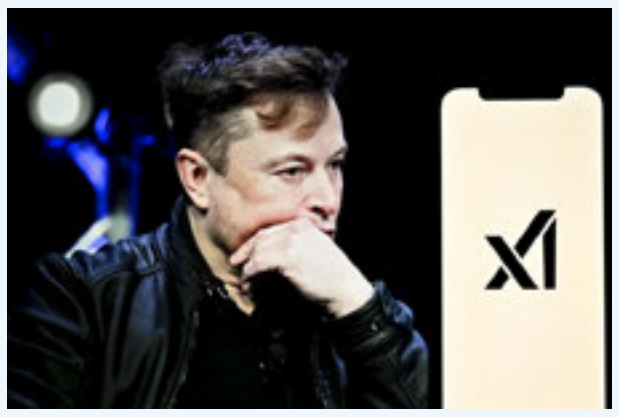<br>Figure 4.1.23<br>Source: Forbes, 2024 |
| Dec 30, 2024 | Nvidia acquires Israeli startup Run:ai for $700 million to increase its GPU optimization capability in demanding computing environments. | Acquisition | 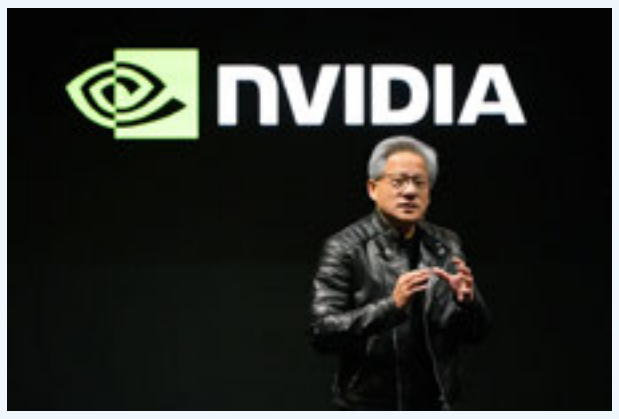<br>Figure 4.1.24<br>Source: TechCrunch, 2024 |

# 4.2 Jobs

## AI Labor Demand

This section analyzes the demand for AI-related skills in labor markets, drawing on data from Lightcast. Since 2010, Lightcast has analyzed hundreds of millions of job postings from over 51,000 websites, identifying those that require AI skills.

### Global AI Labor Demand

Figure 4.2.1 and Figure 4.2.2 show the percentage of job postings demanding AI skills. In 2024, Singapore (3.2%), Luxembourg (2%), and Hong Kong (1.9%) led in this metric. In 2023, AI-related jobs accounted for 1.4% of all American job postings. In 2024, that number increased to 1.8%. Most countries saw an increase from 2023 to 2024 in the share of job postings requiring AI skills.

**AI job postings (% of all job postings) by select geographic areas, 2014–24 (part 1)**
Source: Lightcast, 2024 | Chart: 2025 AI Index report

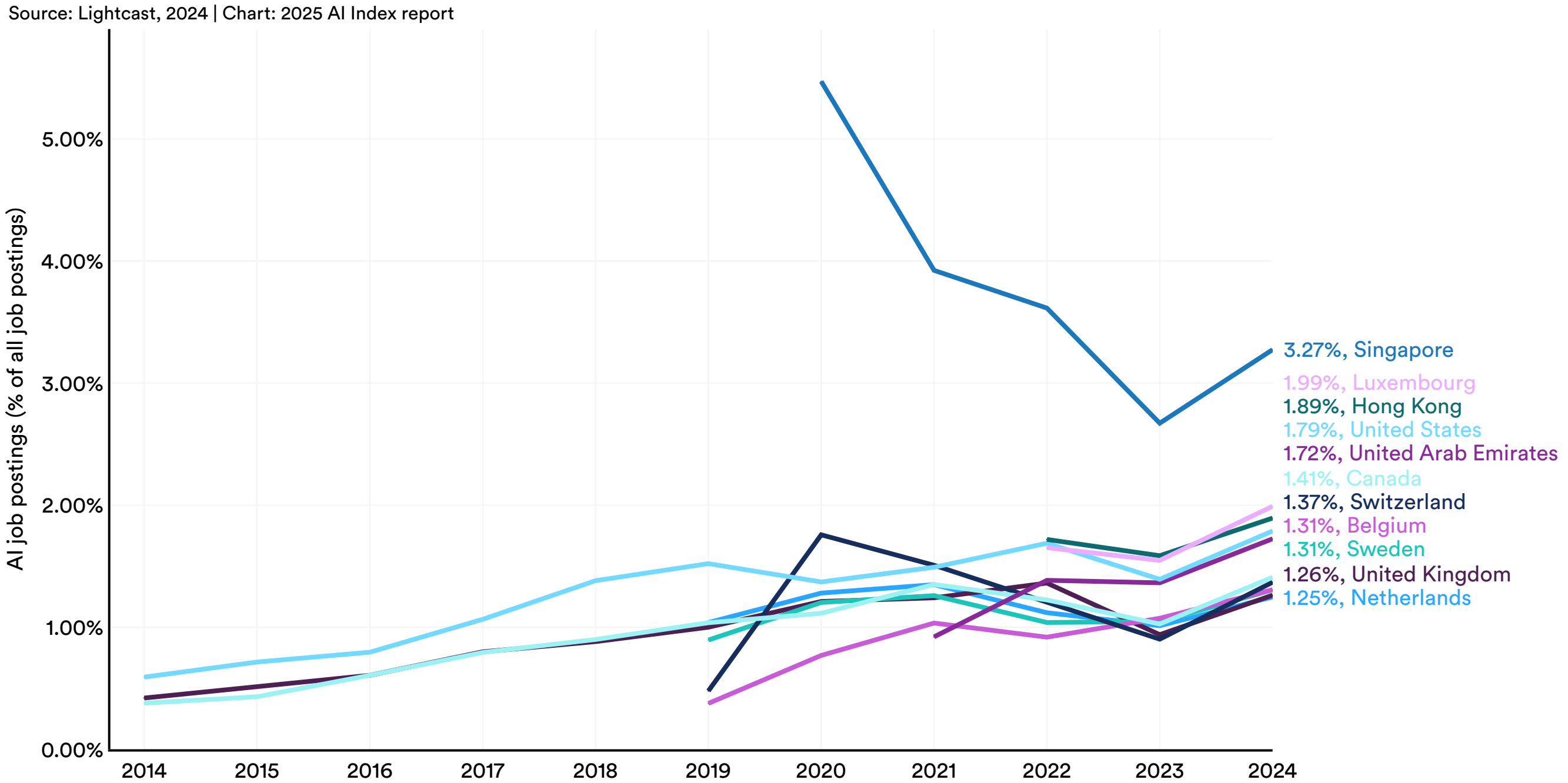

Figure 4.2.1

**AI job postings (% of all job postings) by select geographic areas, 2014–24 (part 2)**
Source: Lightcast, 2024 | Chart: 2025 AI Index report

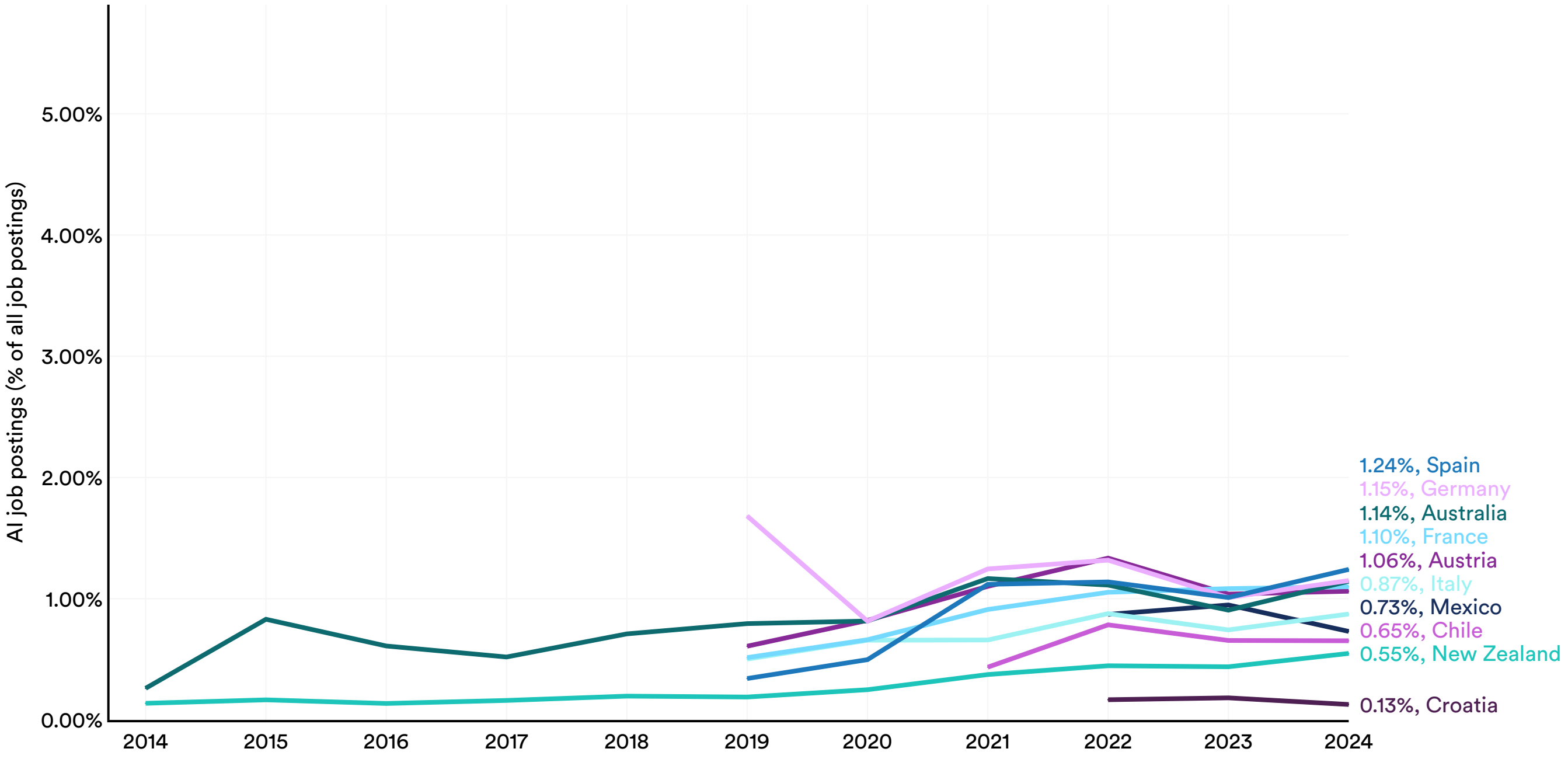

Figure 4.2.2

### US AI Labor Demand by Skill Cluster and Specialized Skill

Figure 4.2.3 highlights the most sought-after AI skills in the U.S. labor market since 2010. Leading the demand was artificial intelligence at 0.9%, followed closely by machine learning, also at 0.9%, and natural language processing at 0.2%. Since last year, most AI-related skill clusters tracked by Lightcast have had an increase in market share, with the exception of autonomous driving and robotics. Generative AI saw the largest increase, growing by nearly a factor of four.

**AI job postings (% of all job postings) in the United States by skill cluster, 2010–24**
Source: Lightcast, 2024 | Chart: 2025 AI Index report

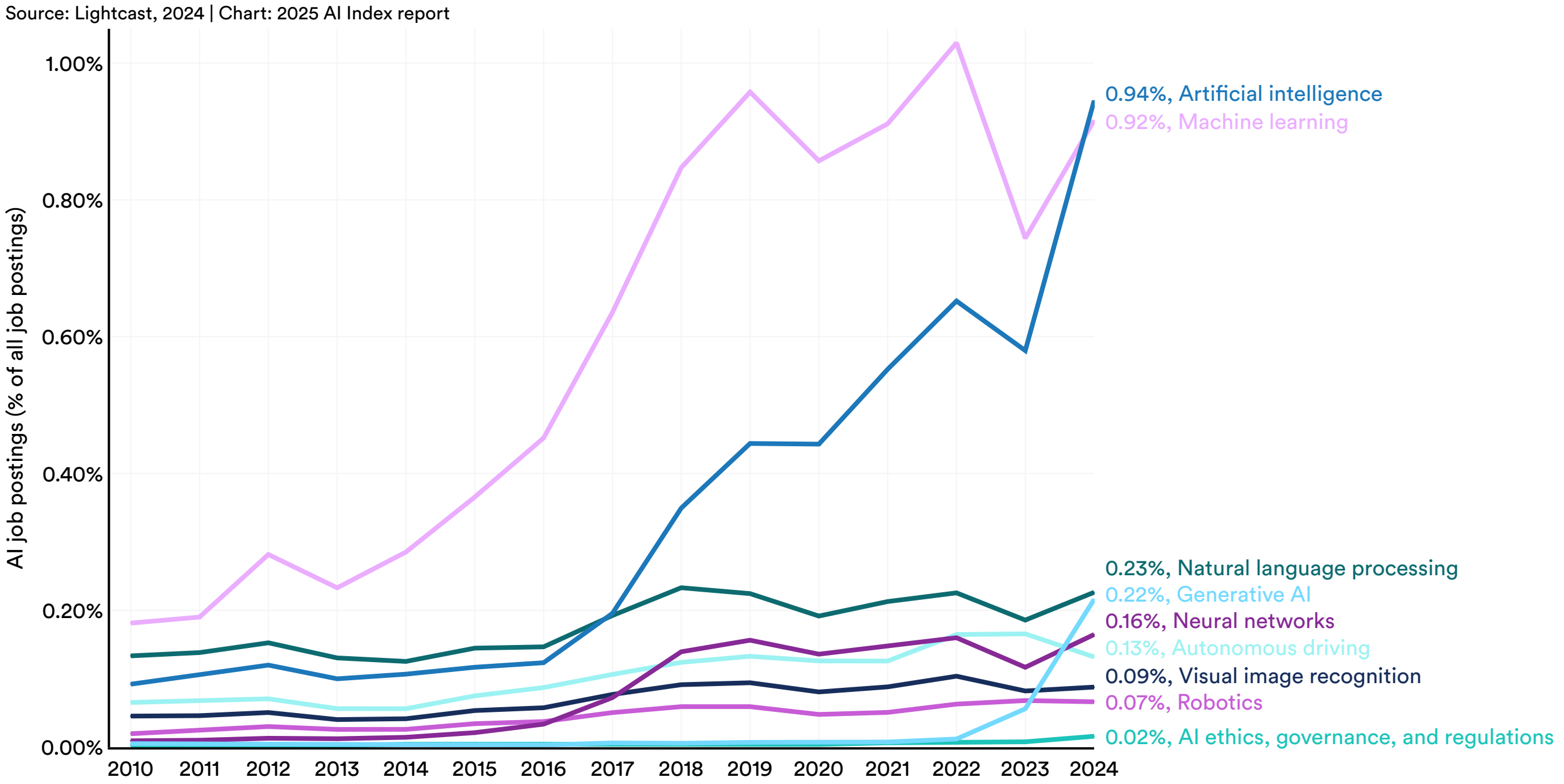

Figure 4.2.3[1]

1 A single job posting can list multiple AI skills.

Figure 4.2.4 compares the top 10 specialized skills sought in AI job postings in 2024 versus those from 2012 to 2014.[2] On an absolute scale, the demand for every specialized skill has increased over the past decade, with Python's notable increase in popularity highlighting its ascendance as a preferred AI programming language.

**Top 10 specialized skills in 2024 AI job postings in the United States, 2012–14 vs. 2024**
Source: Lightcast, 2024 | Chart: 2025 AI Index report

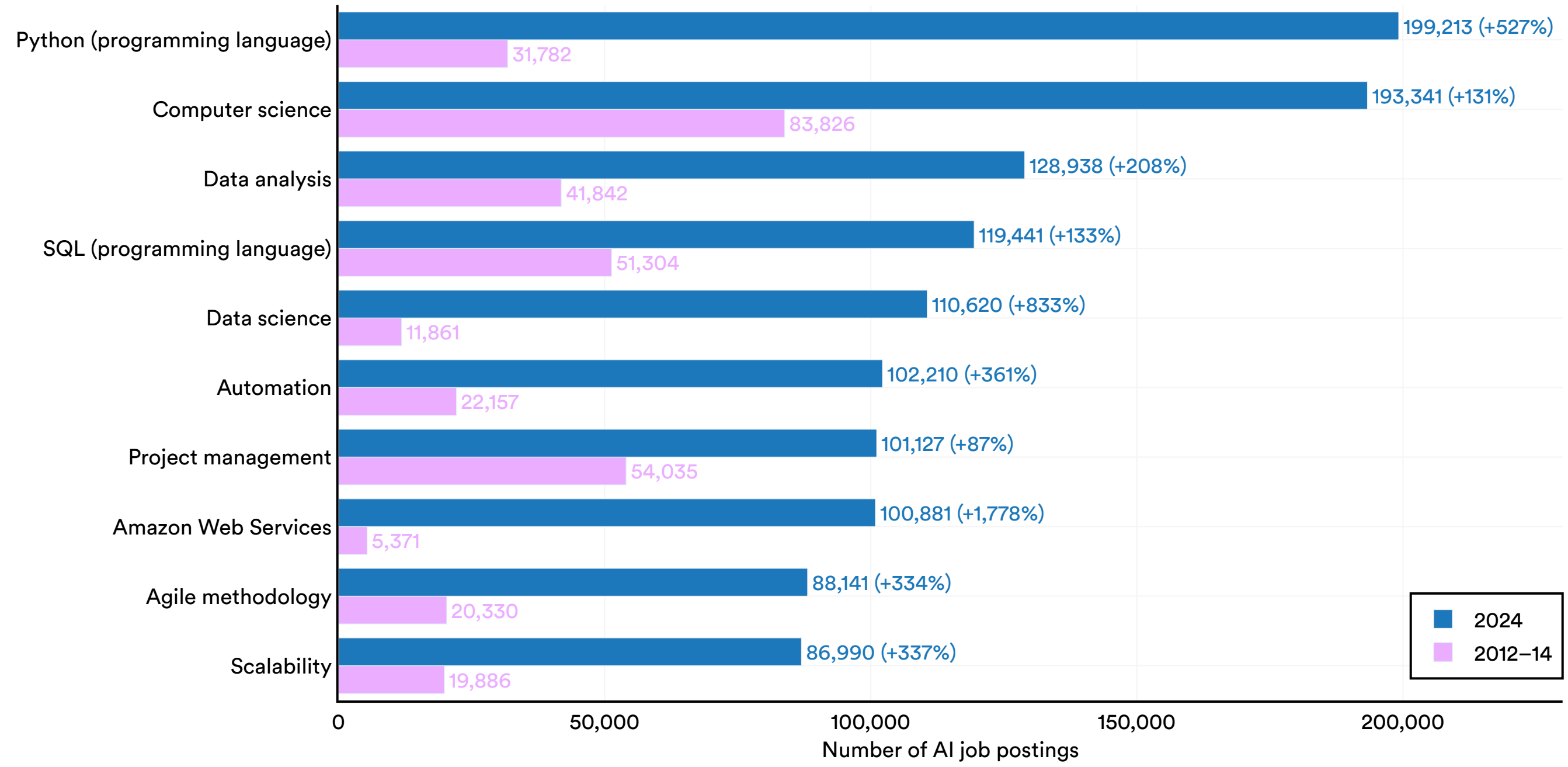

Figure 4.2.4

2 The decision to select 2012–2014 as the point of comparison was due to the scarcity of data at the jobs/skills level from earlier years. Lightcast therefore used 2012–2014 to have a larger sample size for a benchmark from 10 years ago with which to compare. Figure 4.2.4 juxtaposes the total number of job postings requiring certain skills from 2012 to 2014 with the total amount in 2024.

In 2024, year-over-year U.S. job postings citing generative AI skills increased by more than a factor of three (Figure 4.2.5). Figure 4.2.6 illustrates the proportion of AI job postings released in 2024 and 2023 that referenced particular generative AI skills.

### Generative AI skills in AI job postings in the United States, 2023 vs. 2024

Source: Lightcast, 2024 | Chart: 2025 AI Index report

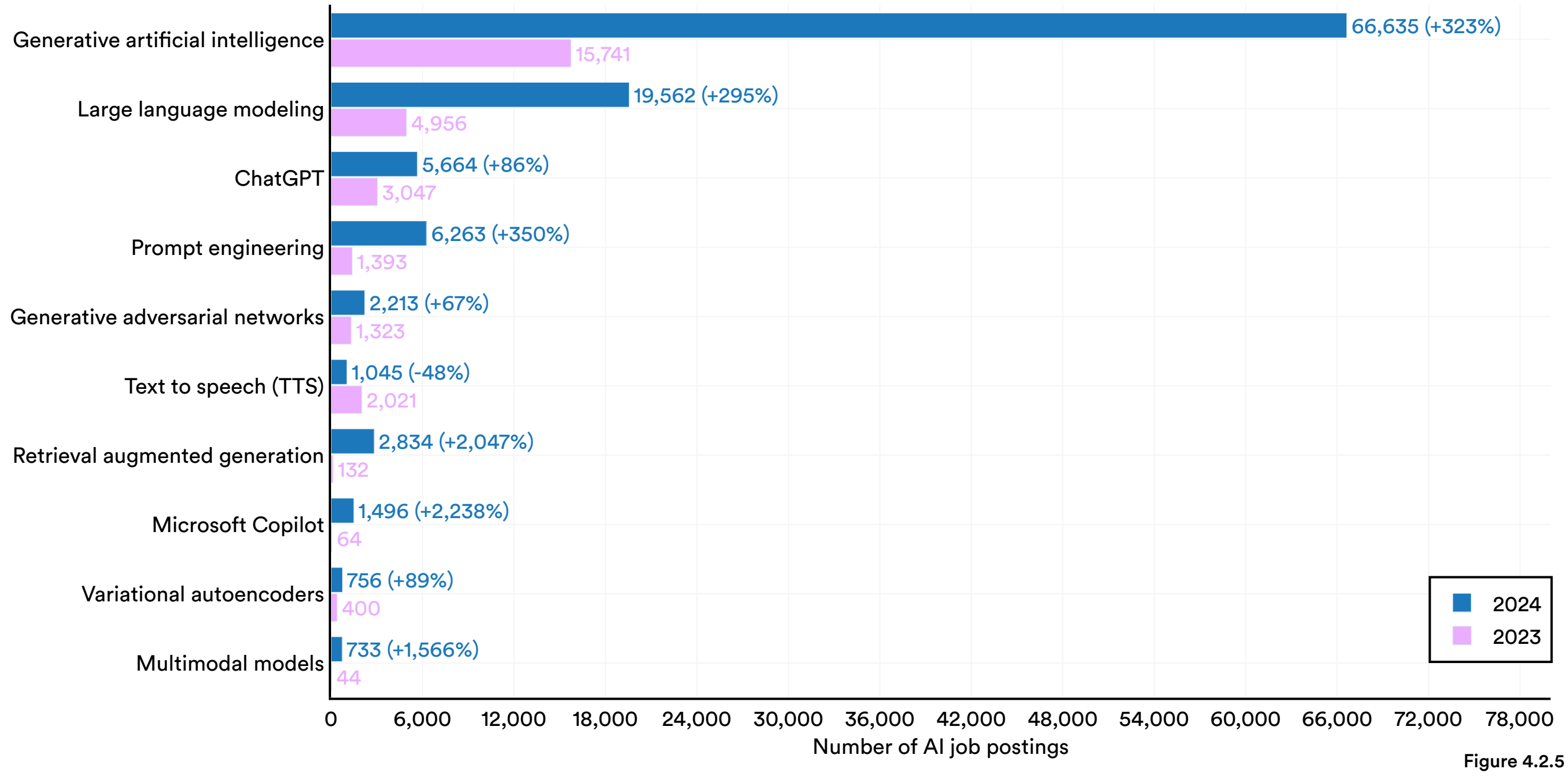

Figure 4.2.5

### Share of generative AI skills in AI job postings in the United States, 2023 vs. 2024

Source: Lightcast, 2024 | Chart: 2025 AI Index report

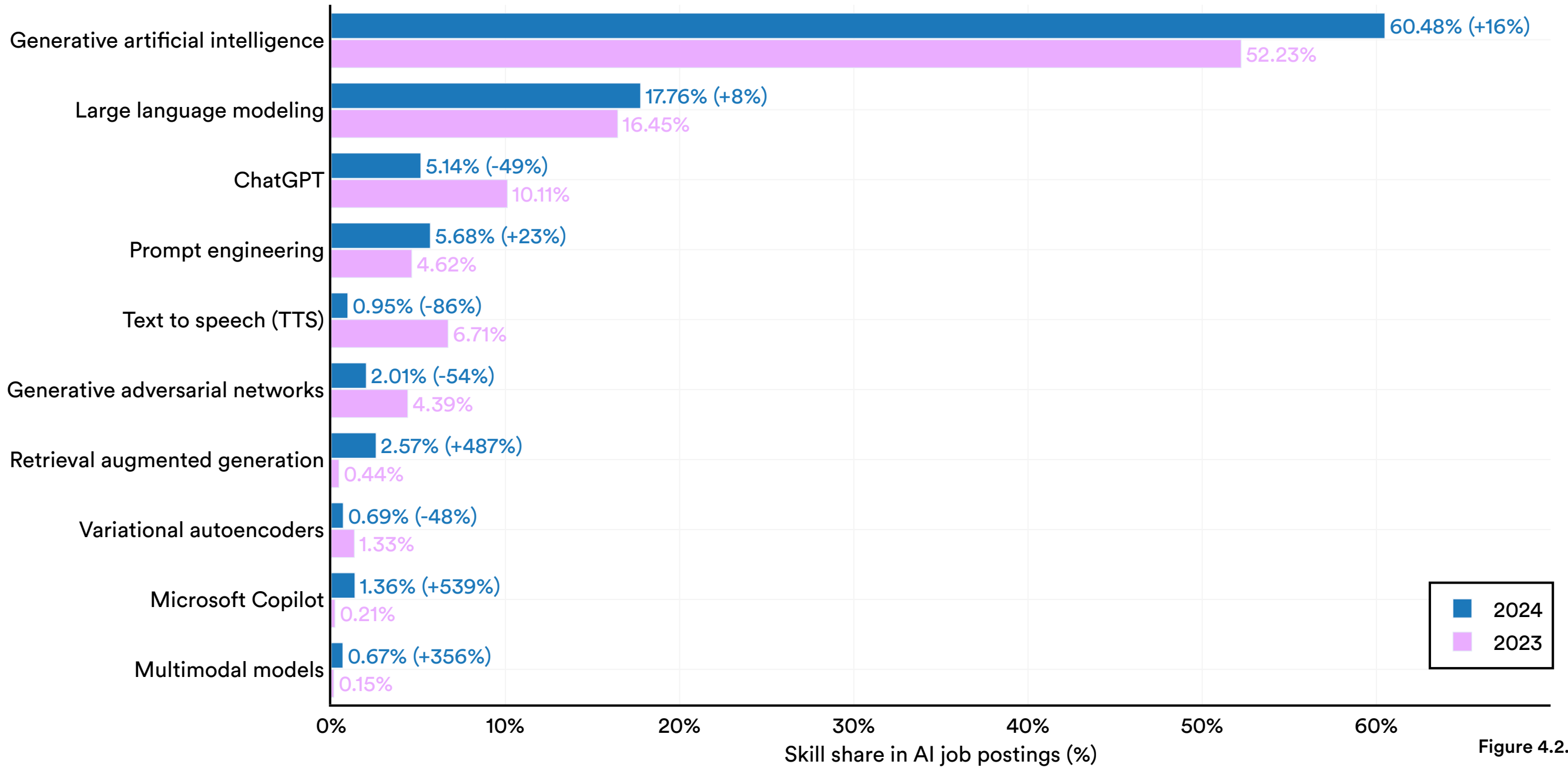

Figure 4.2.6

### US AI Labor Demand by Sector

Figure 4.2.7 shows the percentage of U.S. job postings requiring AI skills by industry sector from 2023 to 2024. Nearly every sector experienced an increase in the proportion of AI job postings in 2024 compared to 2023, except for public administration.

**AI job postings (% of all job postings) in the United States by sector, 2023 vs. 2024**

Source: Lightcast, 2024 | Chart: 2025 AI Index report

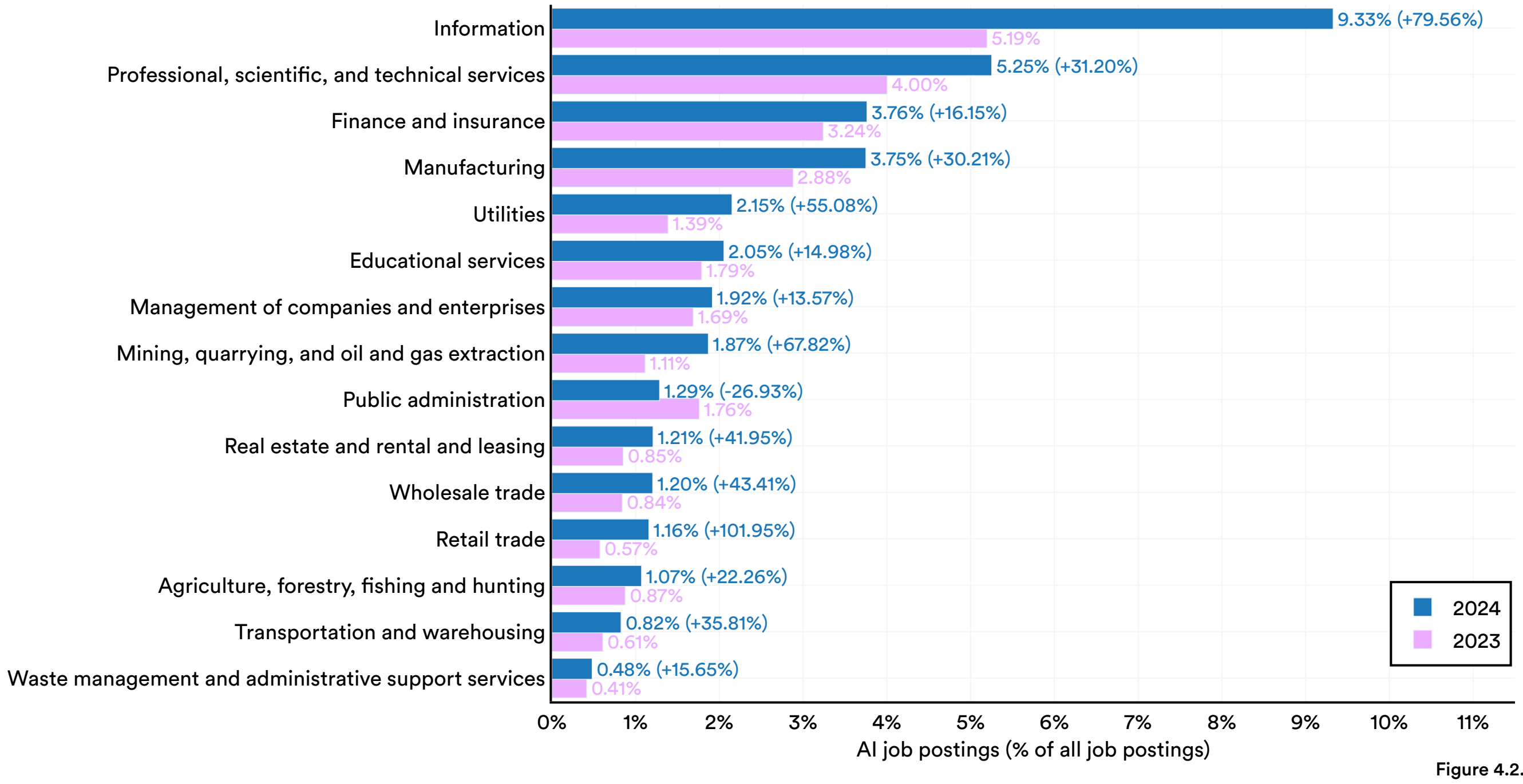

Figure 4.2.7[3]

3 The sector classifications in Figure 4.2.7 are based on two-digit NAICS codes. For more information on the Bureau of Labor Statistics' supersector and NAICS classifications, see the following reference.

### US AI Labor Demand by State

Figure 4.2.8 highlights the number of AI job postings in the United States by state. The top three states were California (103,375), Texas (57,785), and New York (37,944).

Figure 4.2.9 demonstrates what percentage of a state's total job postings were AI-related. The top states according to this metric were the District of Columbia (4.4%), followed by Delaware (3.4%) and Washington (3.3%).

#### Number of AI job postings in the United States by state, 2024

Source: Lightcast, 2024 | Chart: 2025 AI Index report

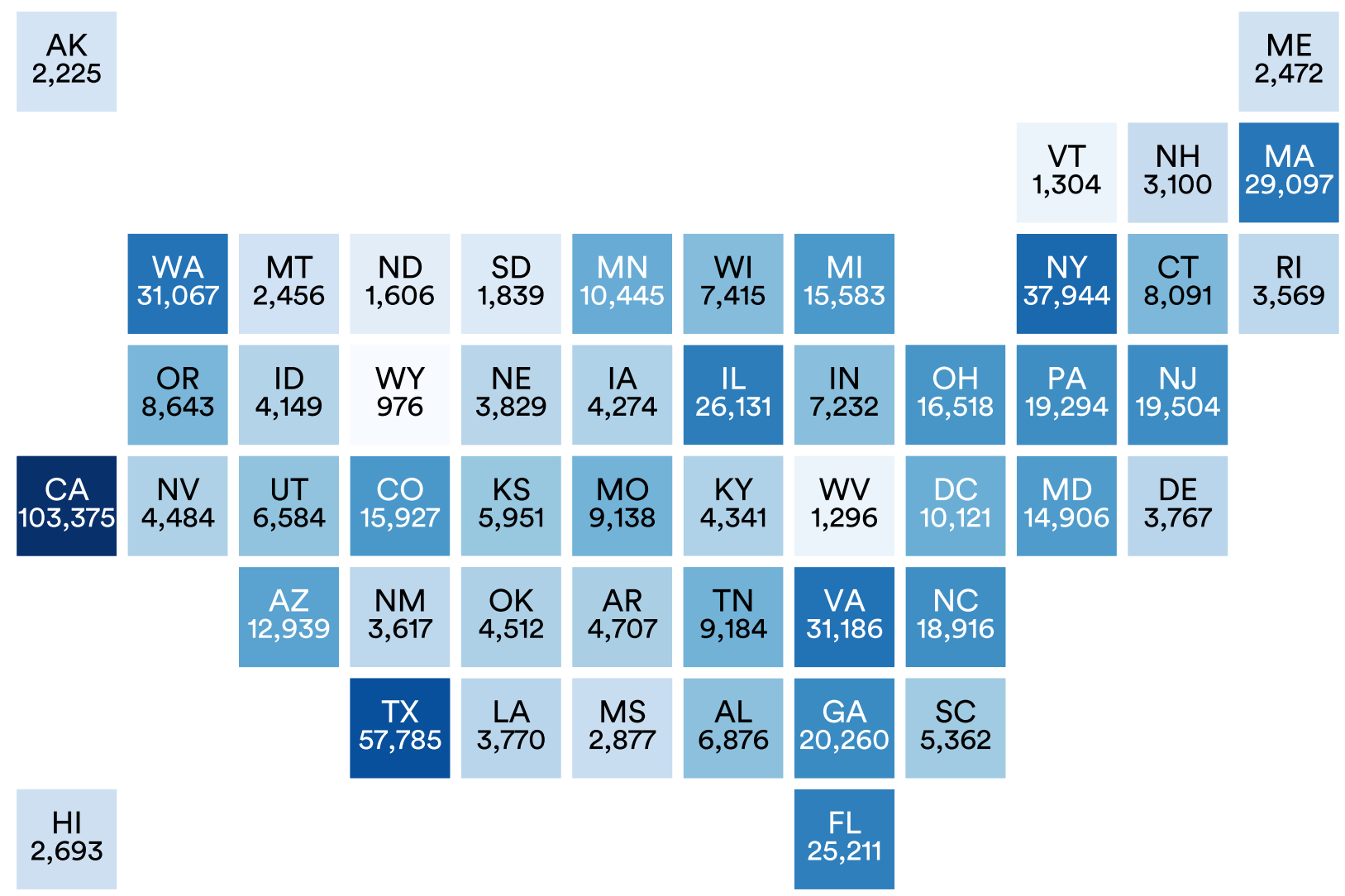

Figure 4.2.8

#### Percentage of US states' job postings in AI, 2024

Source: Lightcast, 2024 | Chart: 2025 AI Index report

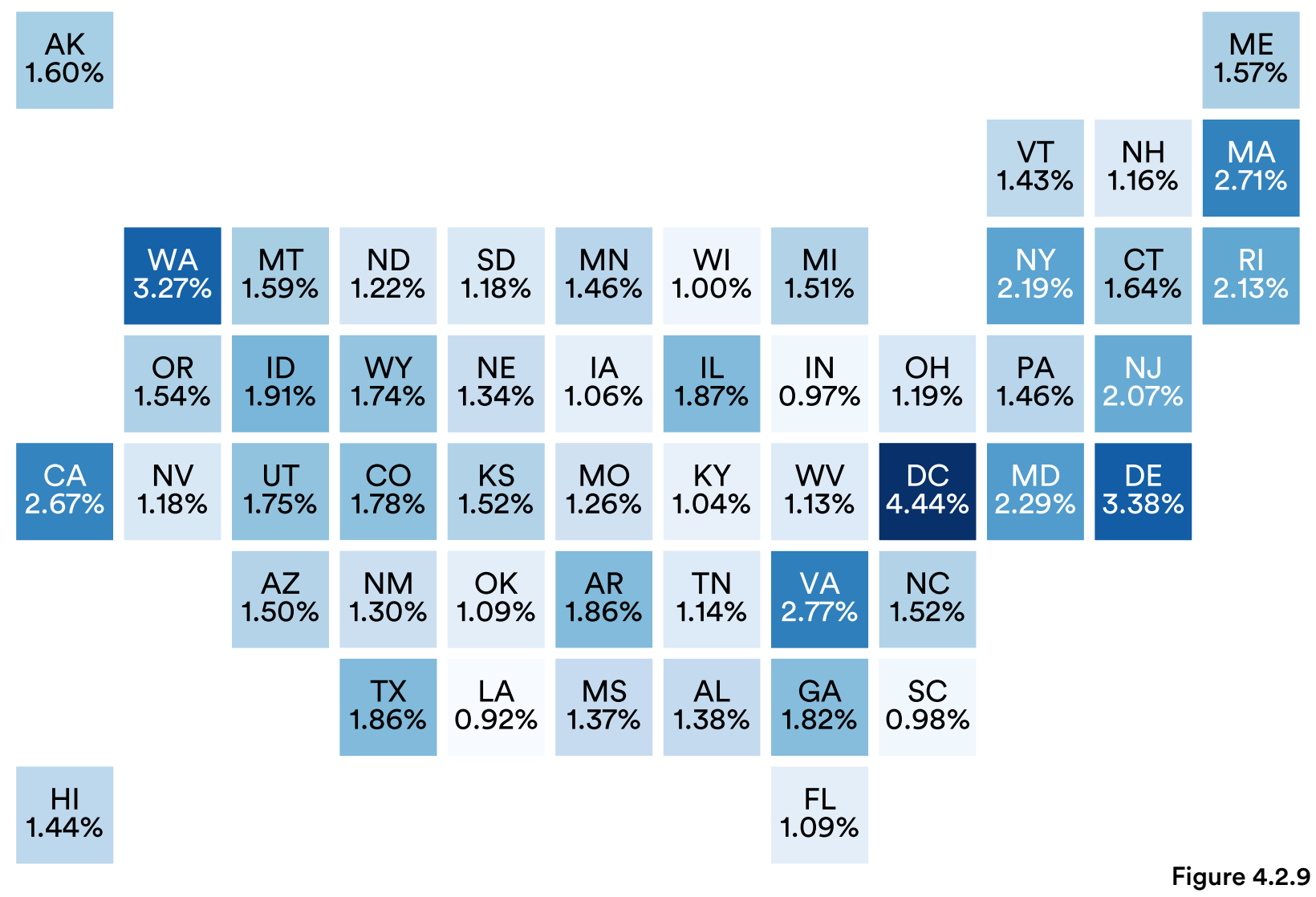

Figure 4.2.9

Figure 4.2.10 examines which U.S. states accounted for the largest proportion of AI job postings nationwide. In 2024, 15.7% of all AI job postings in the United States were for jobs based in California, followed by Texas (8.8%) and New York (5.8%).

Figure 4.2.11 illustrates trends in four states with a significant number of AI job postings: Washington, California, New York, and Texas. Each experienced a notable increase in the share of total AI-related job postings from 2023 to 2024.

**Percentage of US AI job postings by state, 2024**

Source: Lightcast, 2024 | Chart: 2025 AI Index report

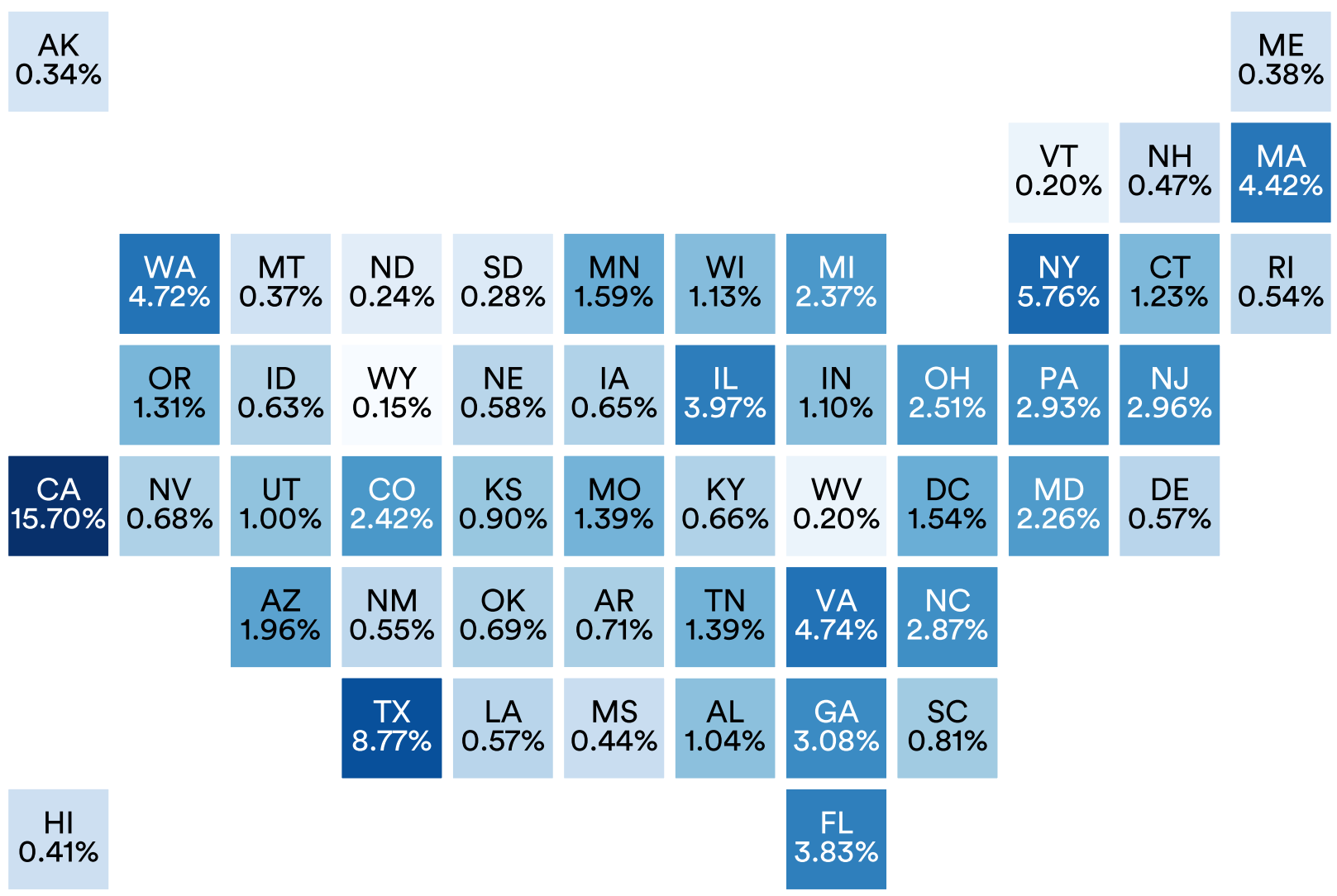

Figure 4.2.10

**Percentage of US states' job postings in AI by select US state, 2010–24**

Source: Lightcast, 2024 | Chart: 2025 AI Index report

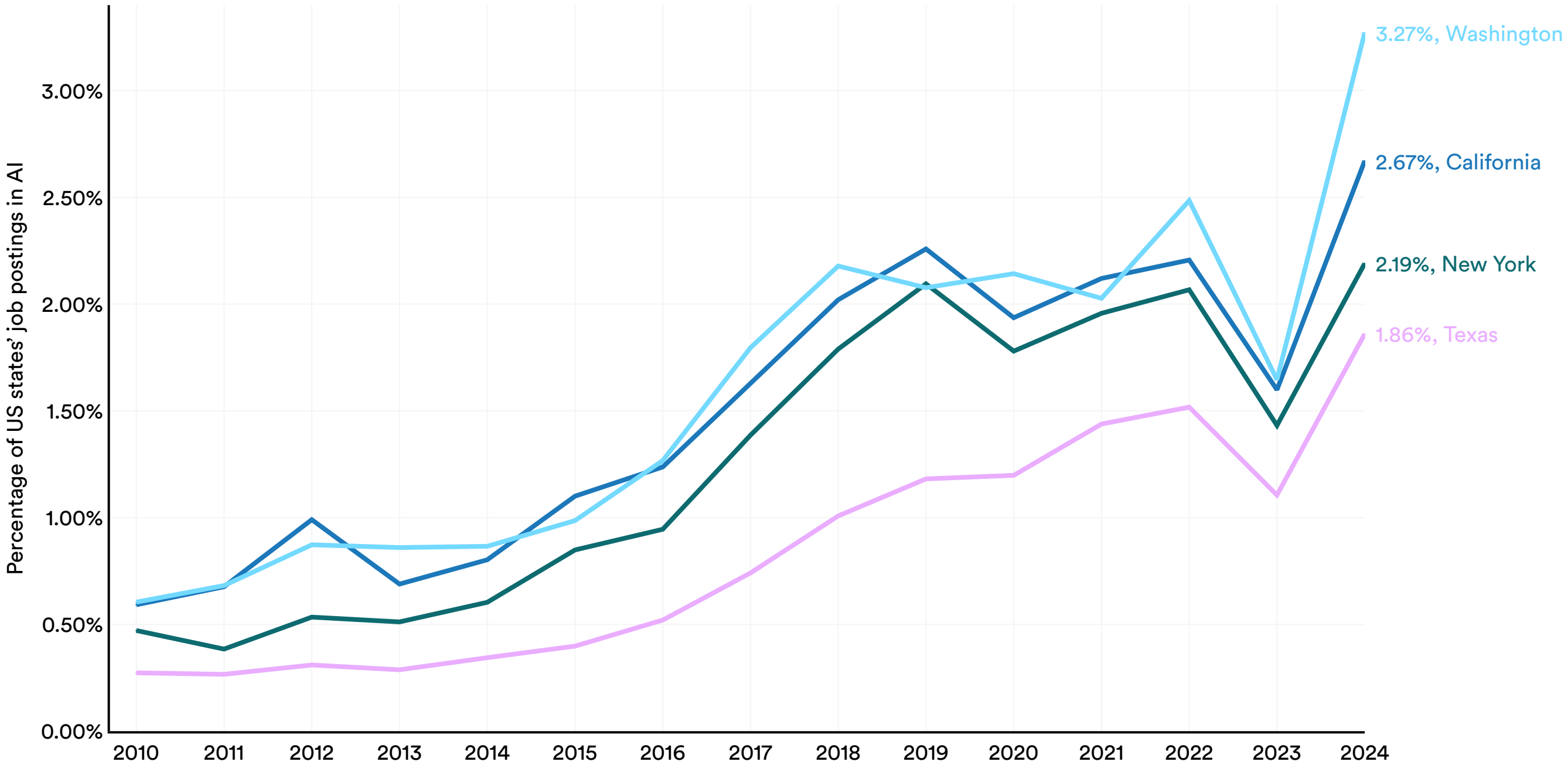

Figure 4.2.11

Figure 4.2.12 shows how AI-related job postings have been distributed across the top four states over time. In 2024, all four states reversed multiyear declines in their proportion of AI job postings—a particularly notable change in California and New York, both of which had experienced decreases since 2020.

**Percentage of US AI job postings by select US state, 2010–24**

Source: Lightcast, 2024 | Chart: 2025 AI Index report

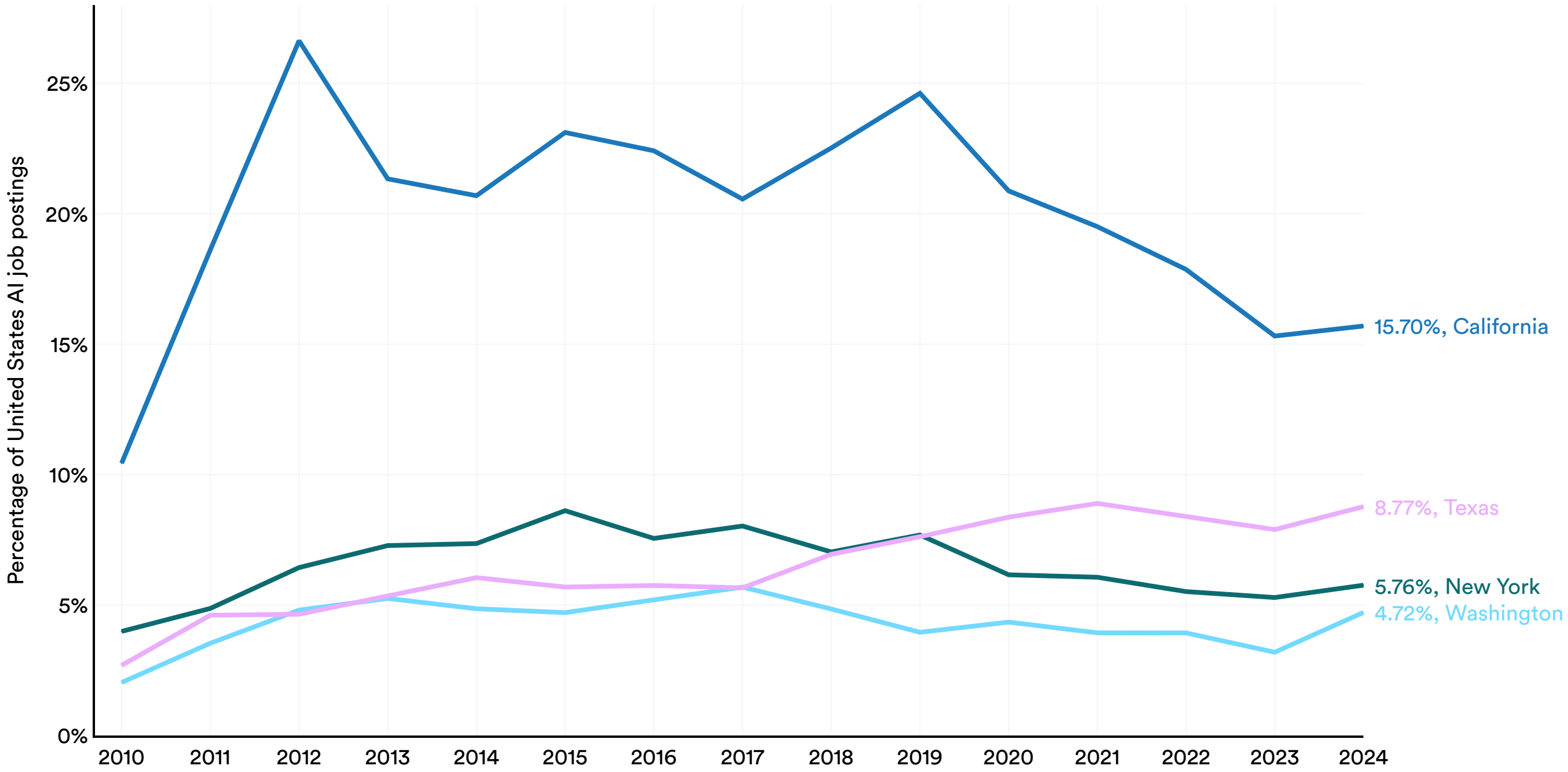

Figure 4.2.12

## AI Hiring

The hiring data presented in the AI Index is based on LinkedIn's Economic Graph, reflecting the jobs and skills of the platform's 1+ billion members. As such, the data is influenced by how members choose to use the platform, which can vary based on professional, social, and regional cultures, as well as overall site availability and accessibility. The AI Index notes that Hungary, Indonesia, India, and South Korea, included in the sample, have LinkedIn covering a lower portion of the labor force, so insights drawn about these countries should be interpreted with particular caution.

Figure 4.2.13 reports the relative AI hiring rate year-over-year ratio by geographic area. The overall hiring rate is computed as the percentage of LinkedIn members who added a new employer in the same period the job began, divided by the total number of LinkedIn members in the corresponding location. Conversely, the relative AI talent hiring rate is the year-over-year change in AI hiring relative to the overall hiring rate in the same geographic area.[4] Therefore, Figure 4.2.13 illustrates AI hiring vibrancy in those regions that have experienced the most significant rise in AI talent recruitment compared to the overall hiring rate. In 2024, the countries with the greatest relative AI hiring rates year-over-year were India (33.4%), followed by Brazil (30.8%) and Saudi Arabia (28.7%). This means, for example, that in 2024 in India, the ratio of AI talent hiring relative to overall hiring grew 33.4%.

**Relative AI hiring rate year-over-year ratio by geographic area, 2024**
Source: LinkedIn, 2024 | Chart: 2025 AI Index report

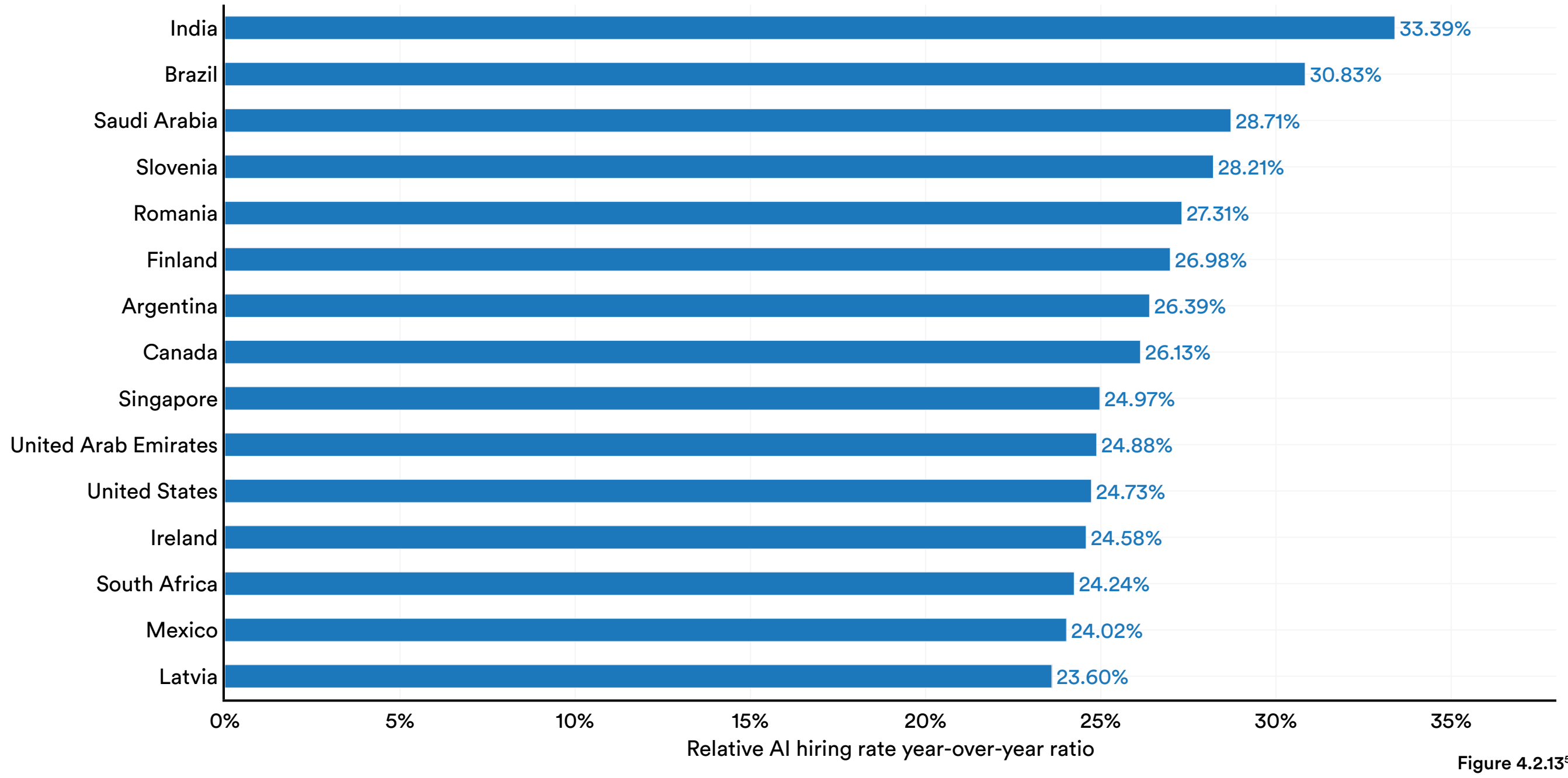

Figure 4.2.13[5]

Figure 4.2.14 showcases the year-over-year ratio of AI hiring by geographic areas over the past five years. Starting in 2024, several South American countries like Argentina, Brazil, and Chile have experienced notable upticks in AI hiring rates. Other countries that have recently experienced similar rises include Canada, India, South Africa, and the United States.

4 For each month, LinkedIn calculates the AI hiring rate in the geographic area, divides the AI hiring rate by the overall hiring rate in that geographic area, calculates the year-over-year change of this ratio, and then takes the 12-month moving average using the last 12 months.

5 For brevity, the visualization only includes the top 15 countries for this metric.

## Relative AI hiring rate year-over-year ratio by geographic area, 2018–24

Source: LinkedIn, 2024 | Chart: 2025 AI Index report

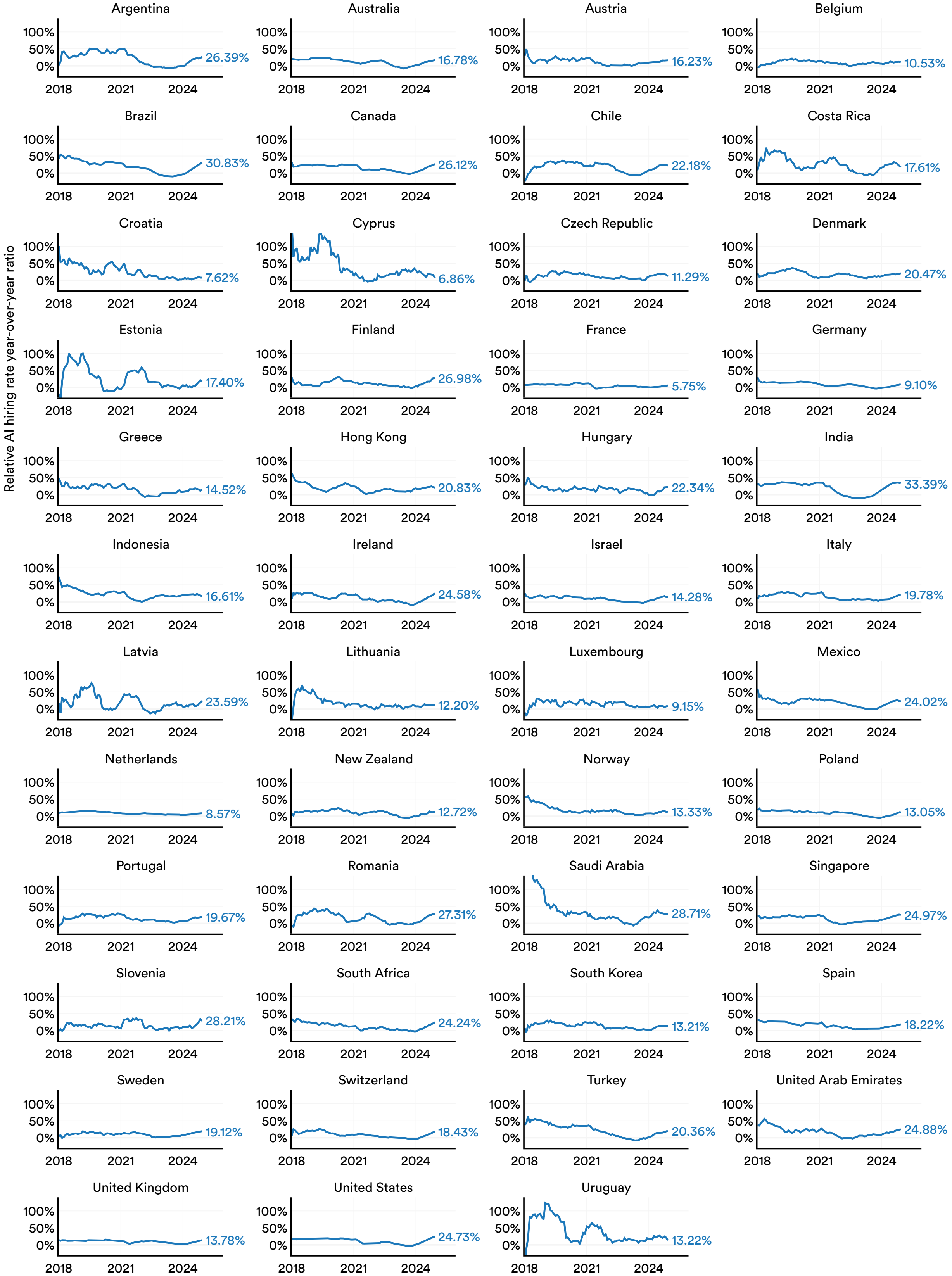

Figure 4.2.14

## AI Skill Penetration

Figure 4.2.15 and Figure 4.2.16 highlight relative AI skill penetration. The aim of this indicator is to measure the intensity of AI skills in a particular country or by industry or gender. The AI skill penetration rate signals the prevalence of AI skills across occupations or the intensity with which LinkedIn members utilize AI skills in their jobs. For example, the top 50 skills for the occupation of engineer are calculated based on the weighted frequency with which they appear in LinkedIn member profiles. If, for instance, four of the skills that engineers possess belong to the AI skill group, the penetration of AI skills among engineers is estimated to be 8% (4/50).

For the period from 2015 to 2024, the countries with the highest AI skill penetration rates were the United States (2.6) and India (2.5). They were followed by the United Kingdom (1.4), Germany (1.3), and Brazil (1.3). In the United States, therefore, the relative penetration of AI skills was 2.6 times greater than the global average across the same set of occupations.

**Relative AI skill penetration rate by geographic area, 2015–24**

Source: LinkedIn, 2024 | Chart: 2025 AI Index report

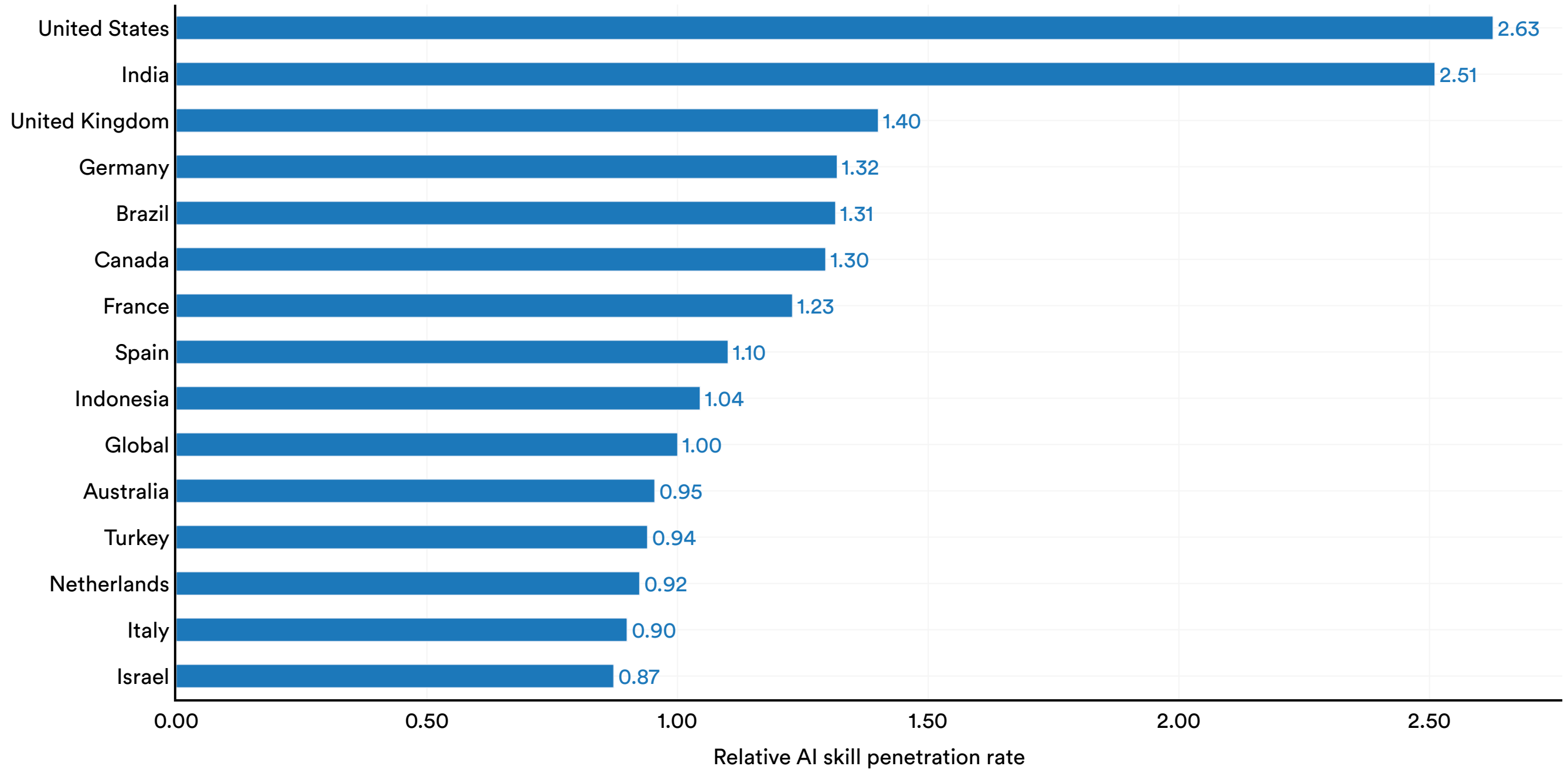

Figure 4.2.15

Figure 4.2.16 disaggregates AI skill penetration rates by gender across different countries or regions. A country's rate of 1.5 for women means female LinkedIn members in that country are 1.5 times more likely to list AI skills than the average member in all countries pooled together across the same set of occupations in the country. For all countries in the sample, with the exception of Saudi Arabia, the relative AI skill penetration rate is greater for men than women. India (1.9), United States (1.7), and Canada (1.0) have the highest reported relative AI skill penetration rates for women.

## Relative AI skill penetration rate across gender, 2015–24

Source: LinkedIn, 2024 | Chart: 2025 AI Index report

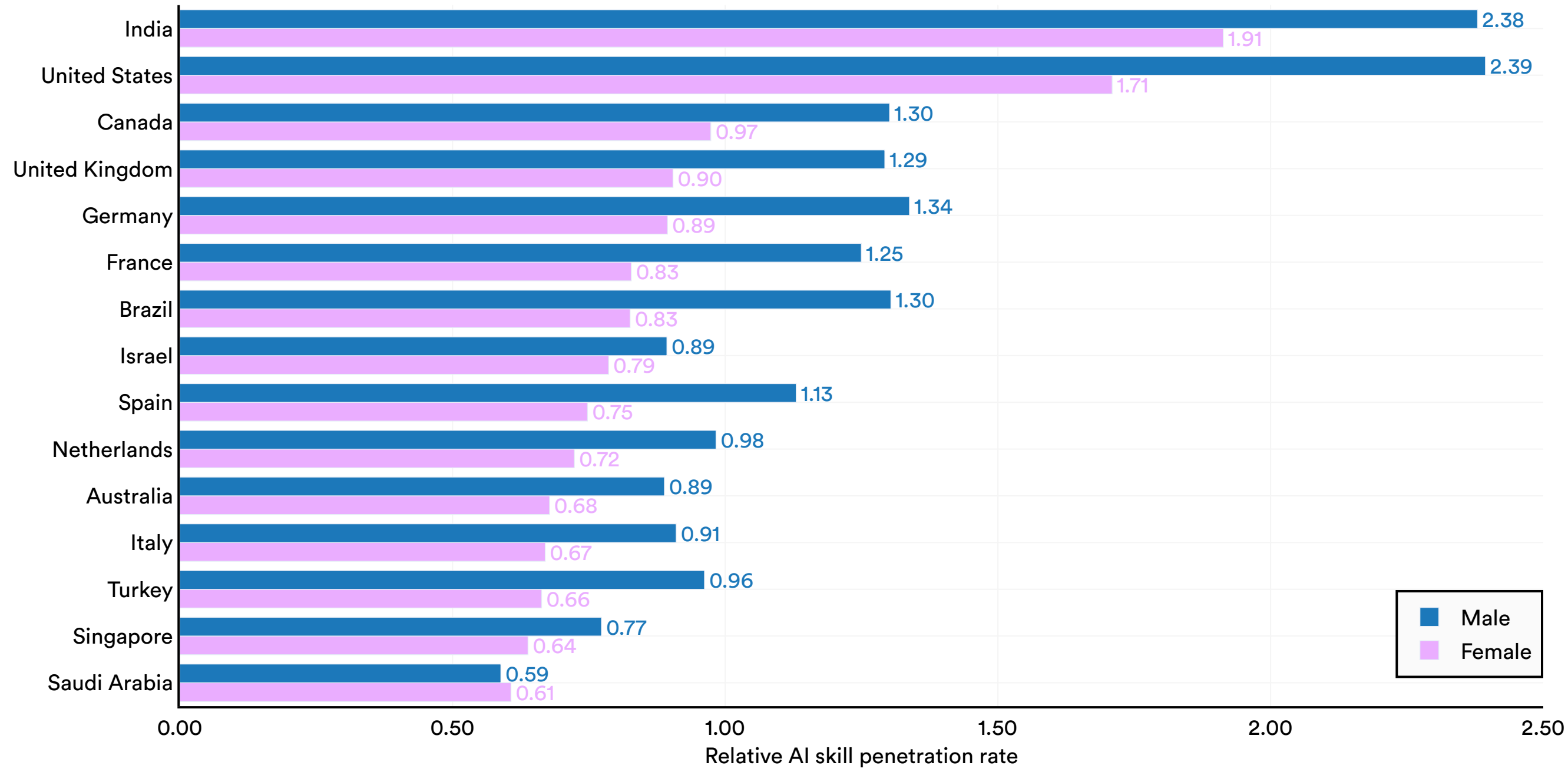

Figure 4.2.16

# AI Talent

Figures 4.2.17 and 4.2.18 examine AI talent by country. A LinkedIn member is considered to have AI talent if they have explicitly added AI skills to their profile, work or have worked in AI. Counts of AI talent are used to calculate talent concentration, or the portion of members who are AI talent. Note that concentration metrics may be influenced by LinkedIn coverage in these countries and should be used with caution.

Figure 4.2.17 shows AI talent concentration in various geographic areas. In 2024, the countries with the highest concentrations of AI talent include Israel (2.0%), Singapore (1.6%), and Luxembourg (1.4%). Figure 4.2.18 looks at the percent change in AI talent concentration for a selection of countries since 2016. During that time period, several major economies registered substantial increases in their AI talent pools. The countries showing the greatest increases are India (252%), Costa Rica (240%), and Portugal (237%).

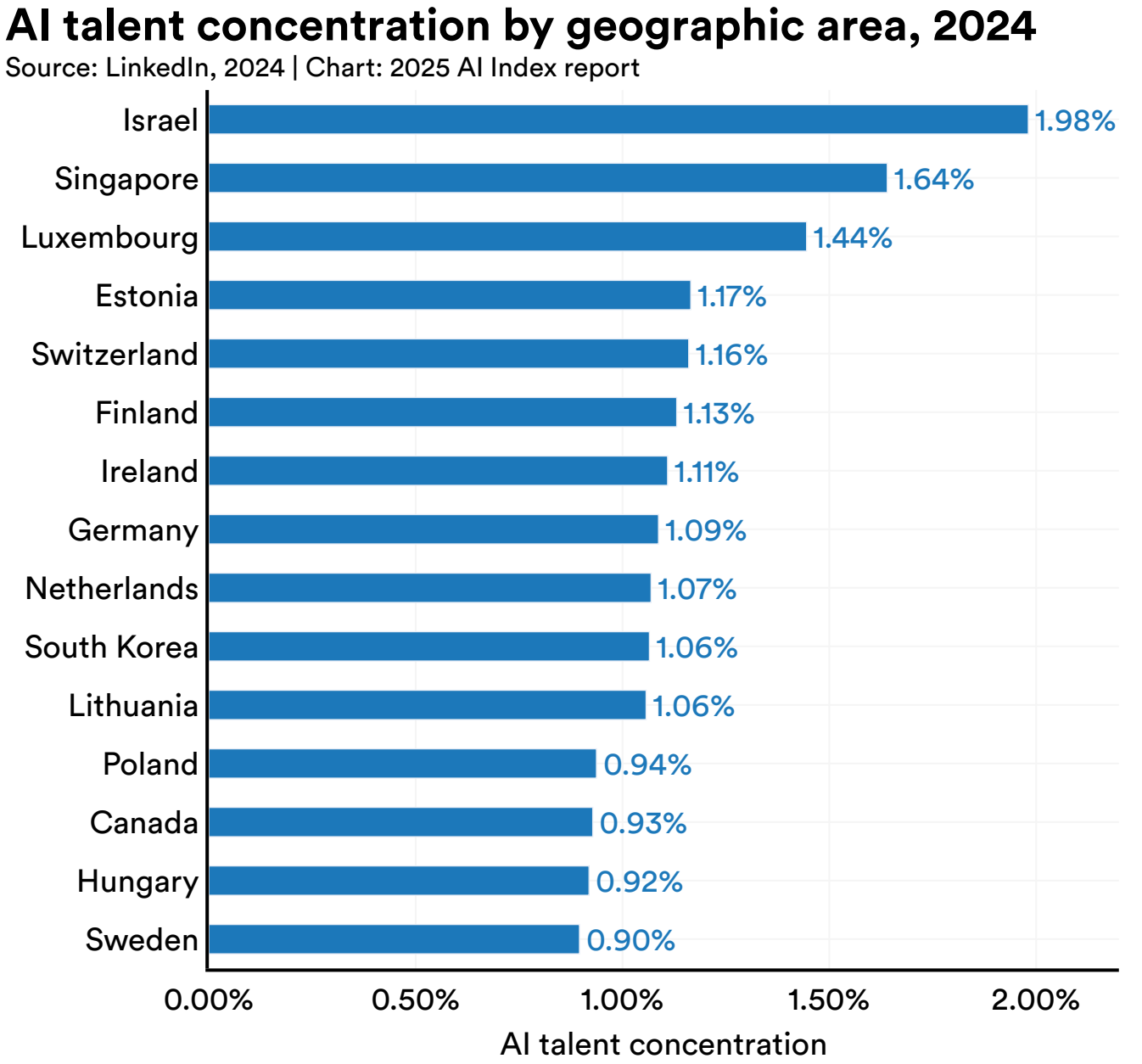

Figure 4.2.17

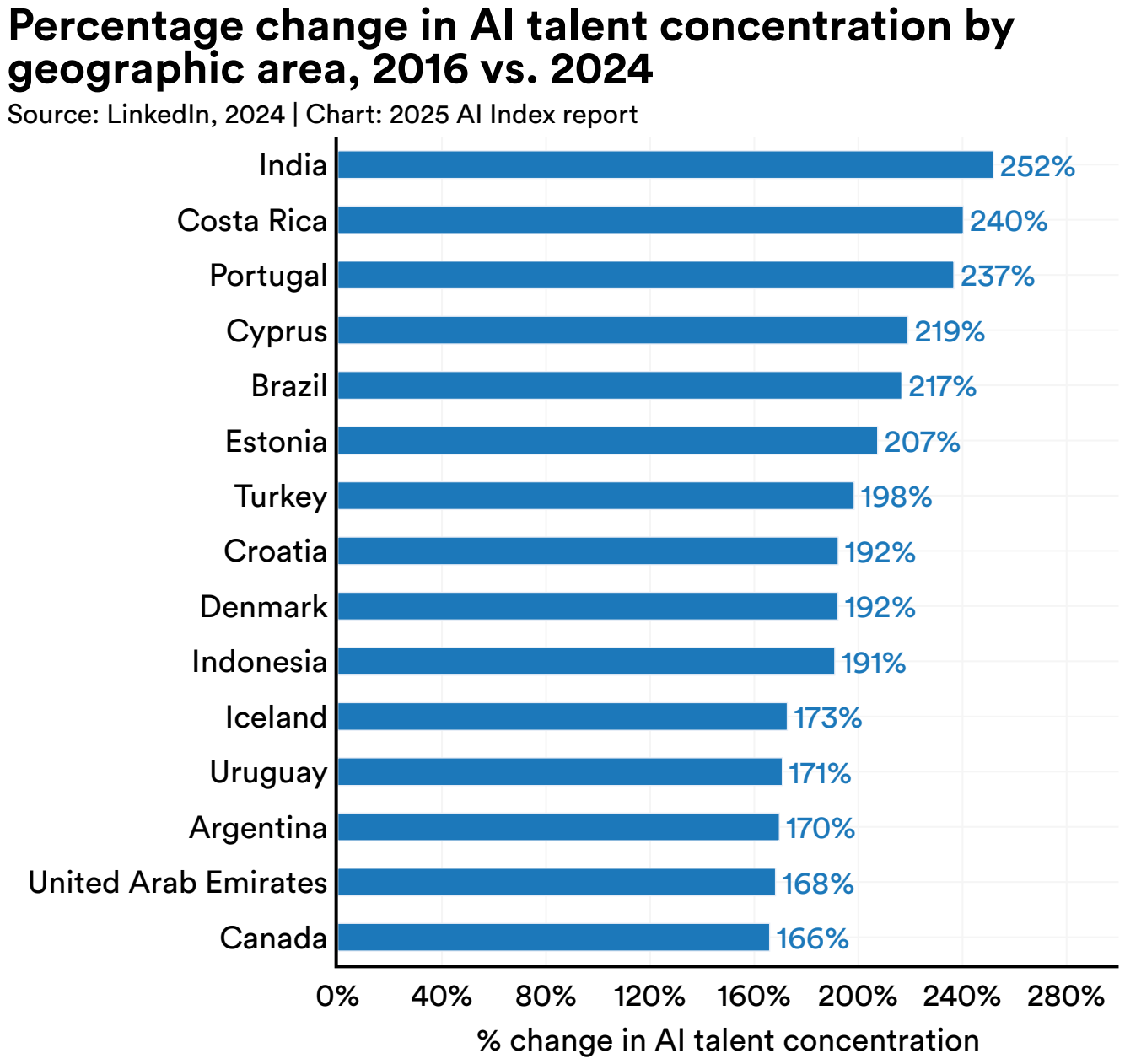

Figure 4.2.18

There are also notable gender differences in AI talent concentration. For every country included in the analysis sample, with the exception of India and Saudi Arabia, the concentration of AI talent was higher among men than women (Figure 4.2.19). Israel reported the highest concentration of female AI talent in 2024, at 1.6%.

## AI talent concentration by gender and geographic area, 2016–24

Source: LinkedIn, 2024 | Chart: 2025 AI Index report

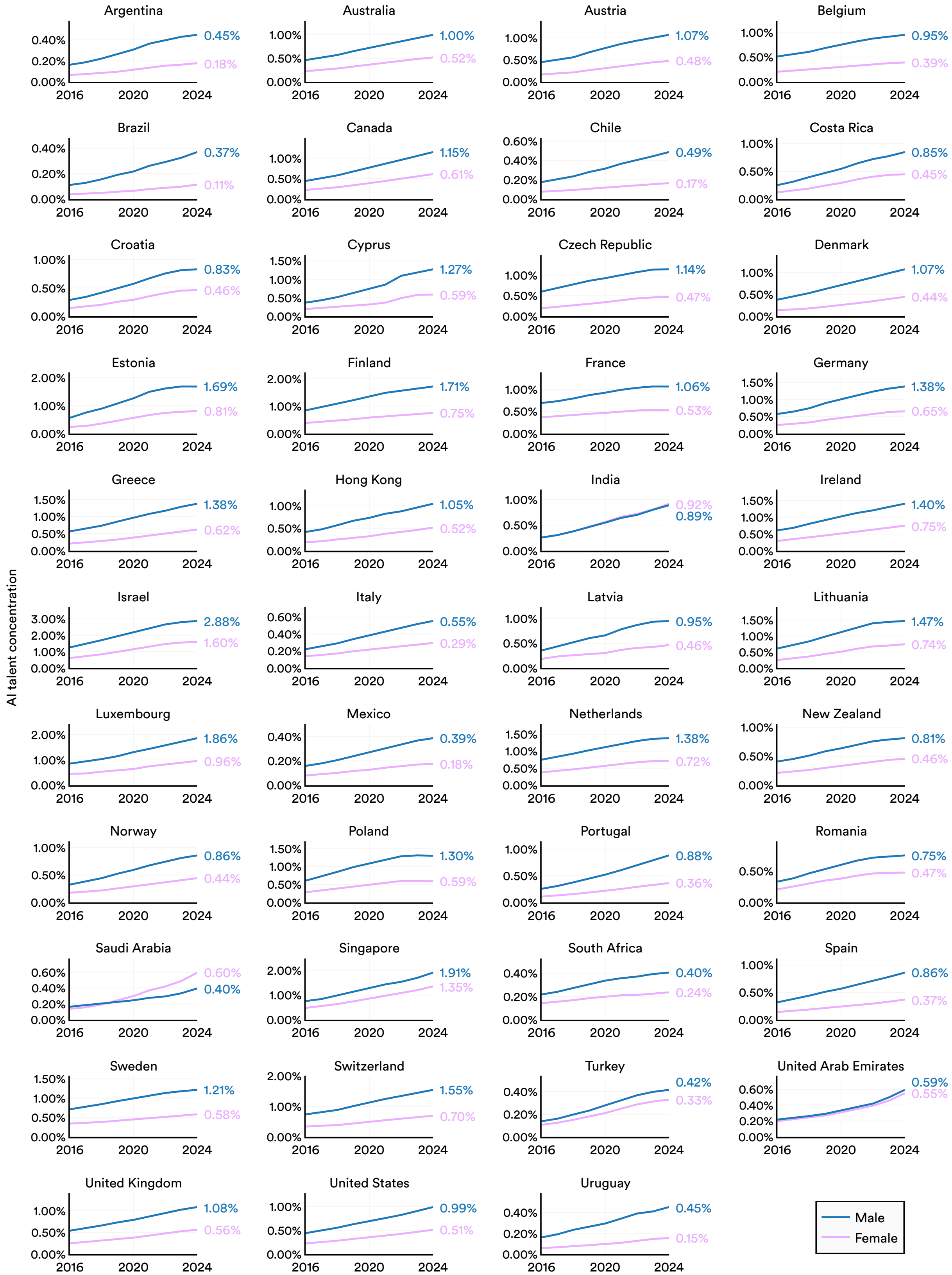

Figure 4.2.19

LinkedIn also tracks the gender distribution of AI talent (Figure 4.2.20). In 2024, it estimates that 69.5% of AI professionals on the platform are male, while 30.5% are female. This ratio has remained remarkably stable over time.

Global AI talent representation, 2016–24
Source: LinkedIn, 2024 | Chart: 2025 AI Index report

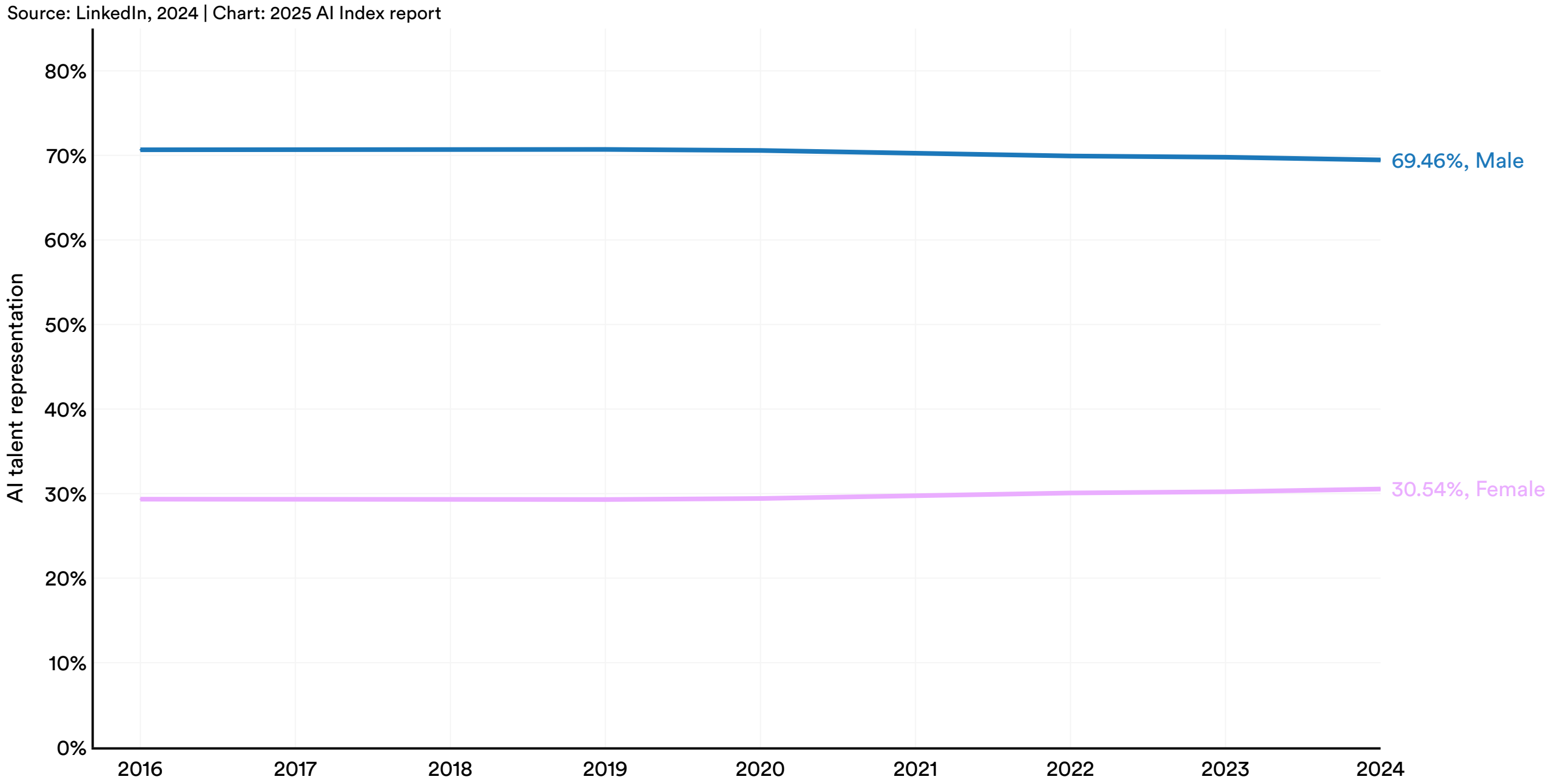

Figure 4.2.20

LinkedIn's data on AI talent can also be broken down by country. In every country in the sample, men proportionally outnumber women in AI roles (Figure 4.2.21). New Zealand and Romania have the most balanced gender distribution, while Brazil and Chile have the least.

### AI talent representation by gender and geographic area, 2016–24
Source: LinkedIn, 2024 | Chart: 2025 AI Index report

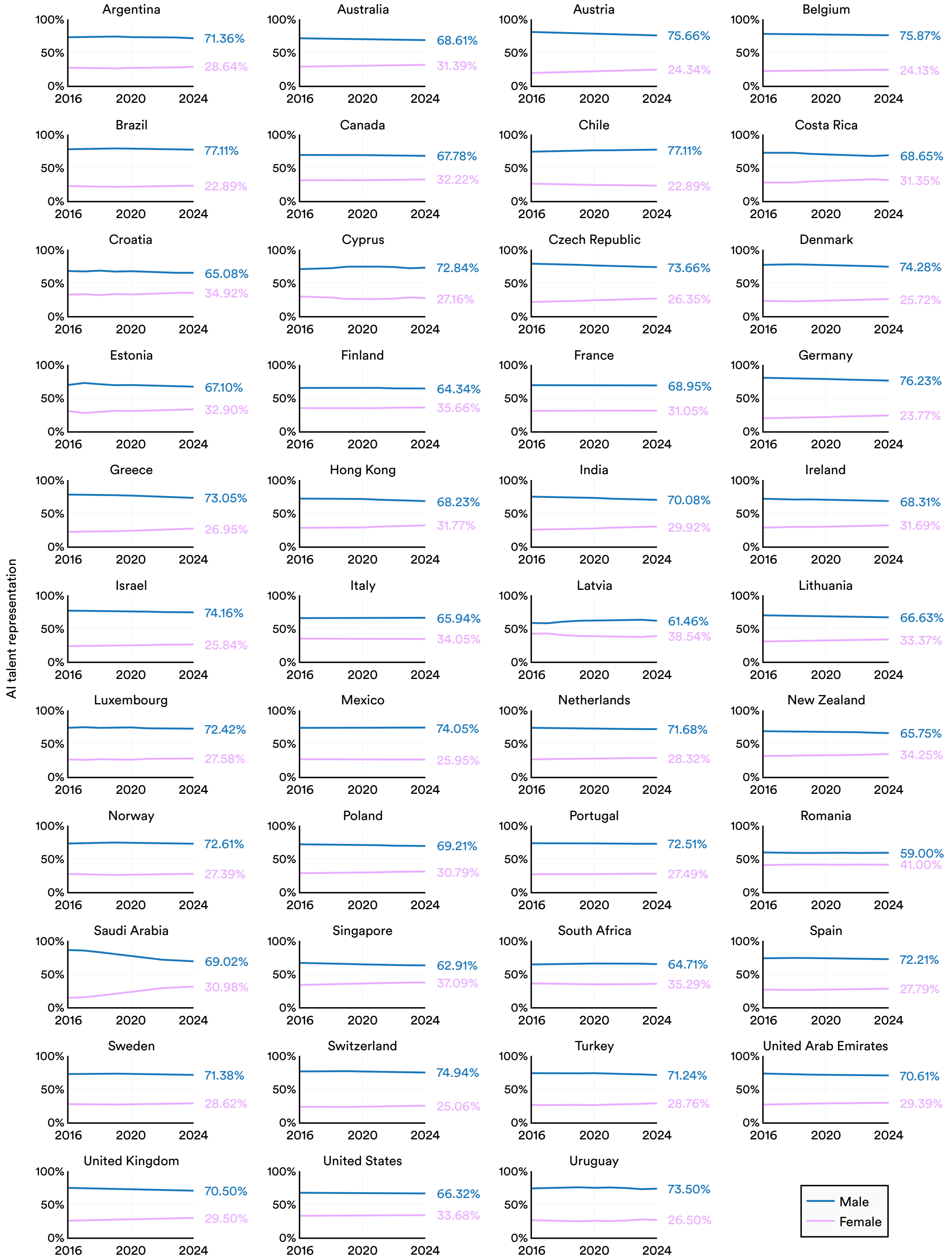

Figure 4.2.21

LinkedIn data provides insights on the AI talent gained or lost due to migration trends.[6] Net flows are defined as total arrivals minus departures within the given time period. A positive net AI talent migration figure indicates that more talent is coming into the geographic area than departing. A negative figure indicates that more talent is departing than coming into the geographic area. Figure 4.2.22 examines net AI talent migration per 10,000 LinkedIn members by geographic area. The geographic areas that report the greatest per capita incoming migration of AI talent are Luxembourg (8.9), Cyprus (4.7), and United Arab Emirates (4.1).

**Net AI talent migration per 10,000 LinkedIn members by geographic area, 2024**

Source: LinkedIn, 2024 | Chart: 2025 AI Index report

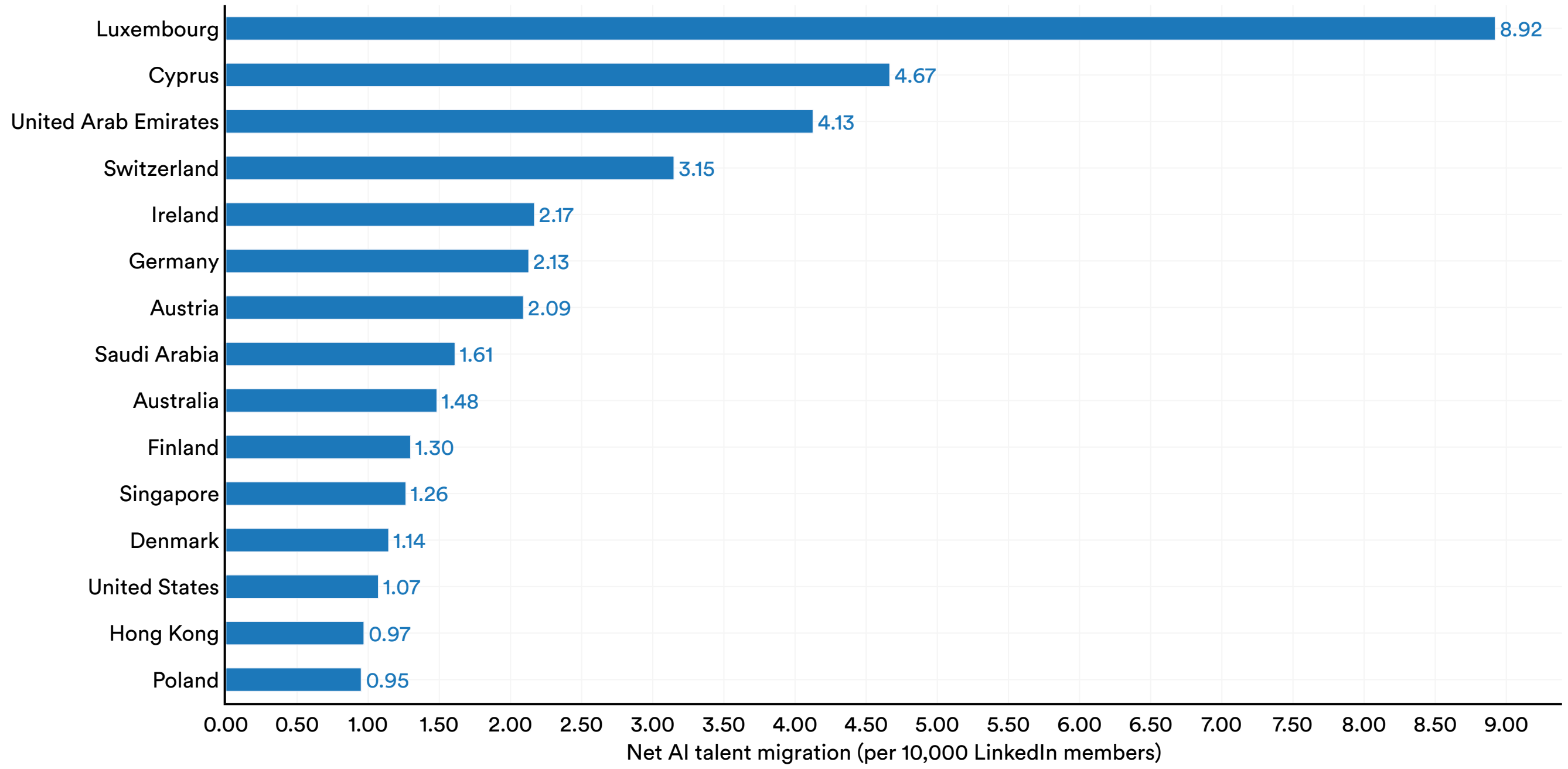

Figure 4.2.22

Figure 4.2.23 documents AI talent migration data over time. In the last few years, Israel, the Netherlands, and Canada, among other countries, have seen declining net AI talent migration figures, suggesting that less AI talent has been flowing into these countries. Countries with rising talent flows include the United Arab Emirates, Saudi Arabia, and Luxembourg.

6 LinkedIn membership varies considerably among countries, which makes interpreting absolute movements of members from one country to another difficult. To compare migration flows between countries fairly, migration flows are normalized for the country of interest. For example, if country A is the country of interest, all absolute net flows into and out of country A (regardless of origin and destination countries) are normalized based on LinkedIn membership in country A at the end of each year and multiplied by 10,000. Hence, this metric indicates relative talent migration of all other countries to and from country A.

## Net AI talent migration per 10,000 LinkedIn members by geographic area, 2019–24

Source: LinkedIn, 2024 | Chart: 2025 AI Index report

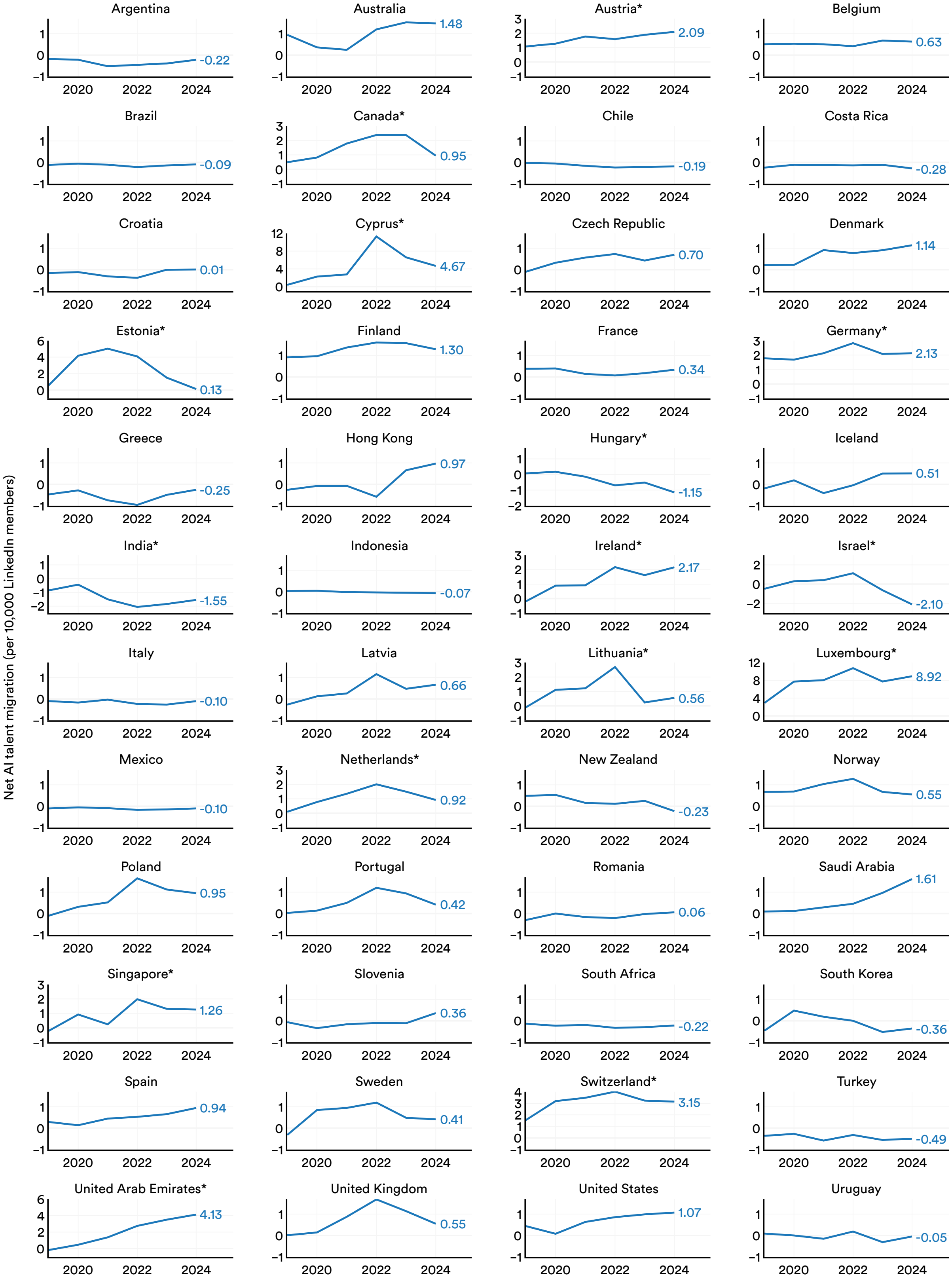

Figure 4.2.23[7]

7 Asterisks indicate that a country's y-axis label is scaled differently than the y-axis label for the other countries.

Highlight:

# Measuring AI's Current Economic Integration

Analysis of over 4 million real-world AI interactions provides comprehensive empirical evidence of how AI is being integrated across economic sectors. A recent Anthropic study examined usage patterns of their AI model classifying users via the U.S. Department of Labor's O*NET occupational framework, offering concrete data on which industries and job functions are leveraging AI. More specifically, the Anthropic team analyzed user conversations with their Claude.AI model to identify the tasks and occupations most frequently using AI.

The analysis reveals that while all sectors make some use of current AI, the dominant sectors are technical and creative. As shown in Figure 4.2.24, computer and mathematical occupations dominate, accounting for 37.2% of all AI interactions. Arts, design, entertainment, sports, and media occupations follow at 10.3%, with educational instruction and library occupations also showing significant adoption.

**Occupational representation in Claude usage data vs. US workforce distribution**
Source: Handa et al., 2025 | Chart: 2025 AI Index report

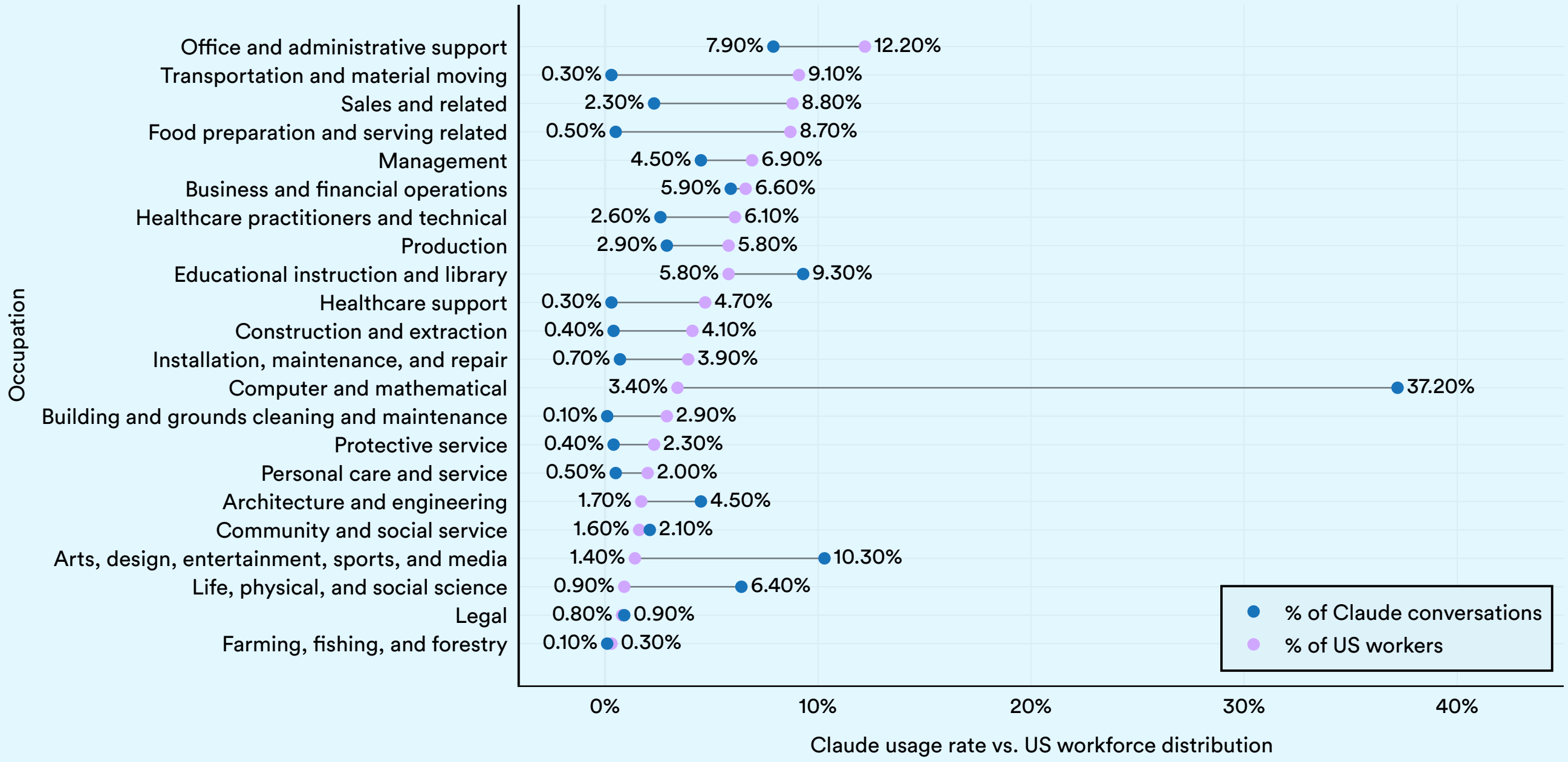

Figure 4.2.23

**Highlight:**

# Measuring AI's Current Economic Integration (cont'd)

The AI usage patterns demonstrate a clear connection to wage levels and required skills. Figure 4.2.25 illustrates that AI adoption peaks in occupations within the upper wage quartile but drops significantly at both wage extremes. Jobs requiring considerable preparation (typically bachelor's degree-level) show 50% higher usage than their baseline workforce representation, while both minimal-preparation and extensive-preparation roles show lower adoption rates.

## Occupational usage of Claude by median annual wage

Source: Handa et al., 2025 | Chart: 2025 AI Index report

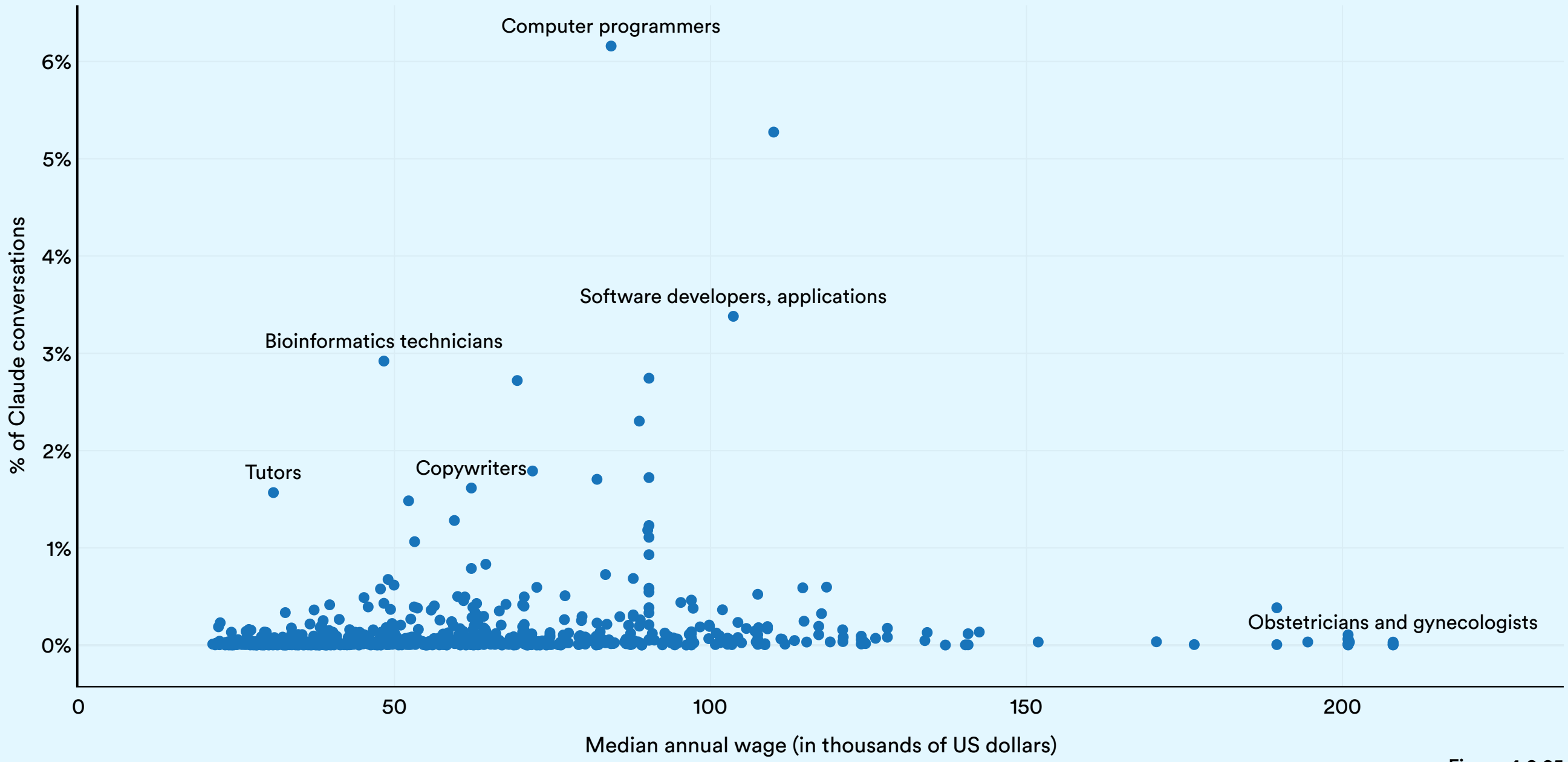

Figure 4.2.25

**Highlight:**

# Measuring AI's Current Economic Integration (cont'd)

The Anthropic study finds that approximately 36% of occupations use AI for at least a quarter of their associated tasks (Figure 4.2.26), indicating substantial penetration beyond technical fields. However, deep integration remains rare: Only about 4% of occupations show AI usage across 75% or more of their tasks, suggesting that wholesale automation of entire job categories is not yet occurring.

### Depth of AI usage across organizations

Source: Handa et al., 2025

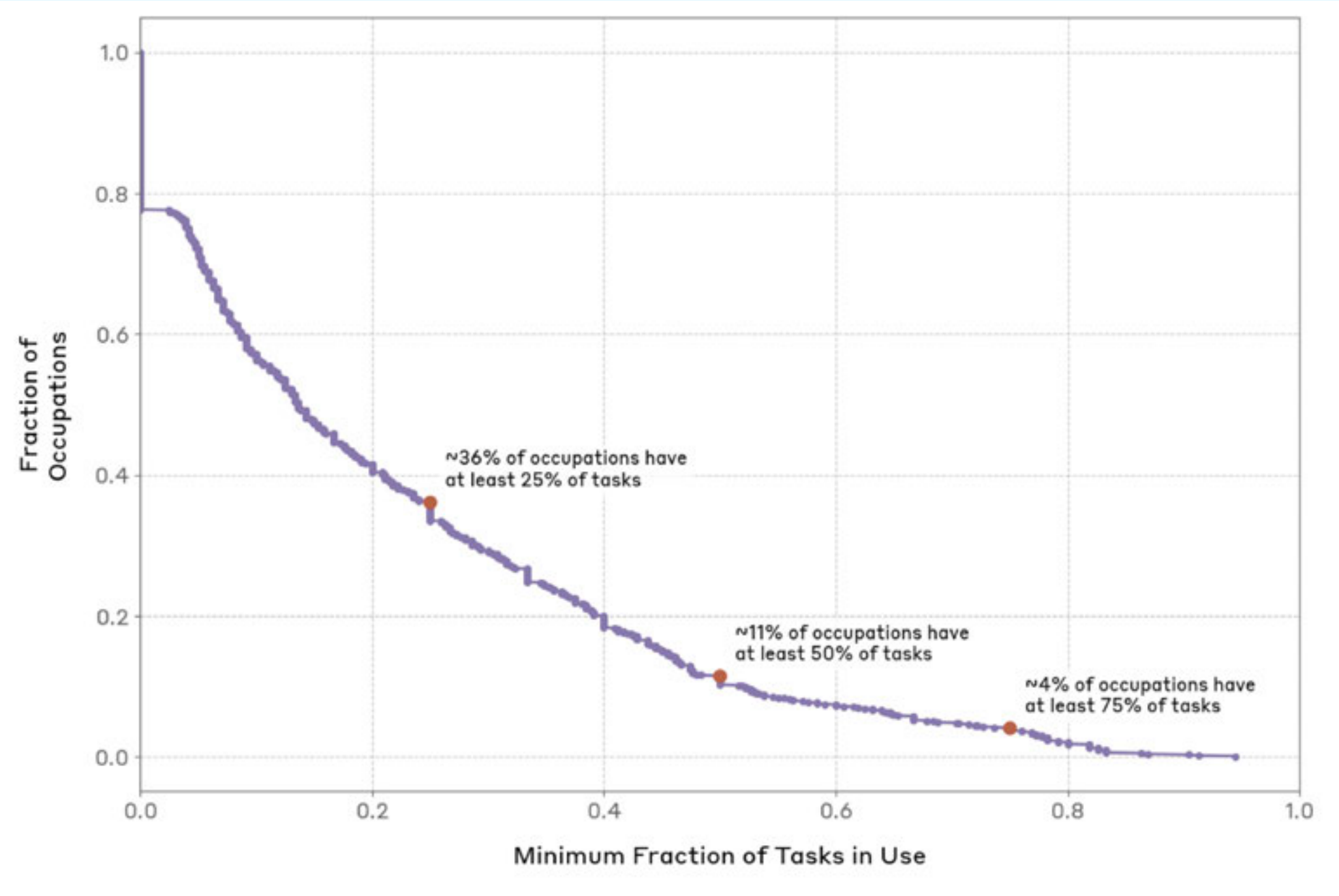

Figure 4.2.26

**Highlight:**

# Measuring AI's Current Economic Integration (cont'd)

The analysis reveals how AI is being used within organizations. As shown in Figure 4.2.27, 57% of AI interactions demonstrate augmentative patterns (enhancing human capabilities) while 43% show automation patterns. This split suggests current AI implementation tends toward complementing rather than replacing human workers. The study finds that cognitive skills like critical thinking and writing show high presence in AI interactions, while physical and managerial skills show minimal presence (Figure 4.2.28).

**Percentage of Claude conversations by type of task execution**

Source: Handa et al., 2025 | Chart: 2025 AI Index report

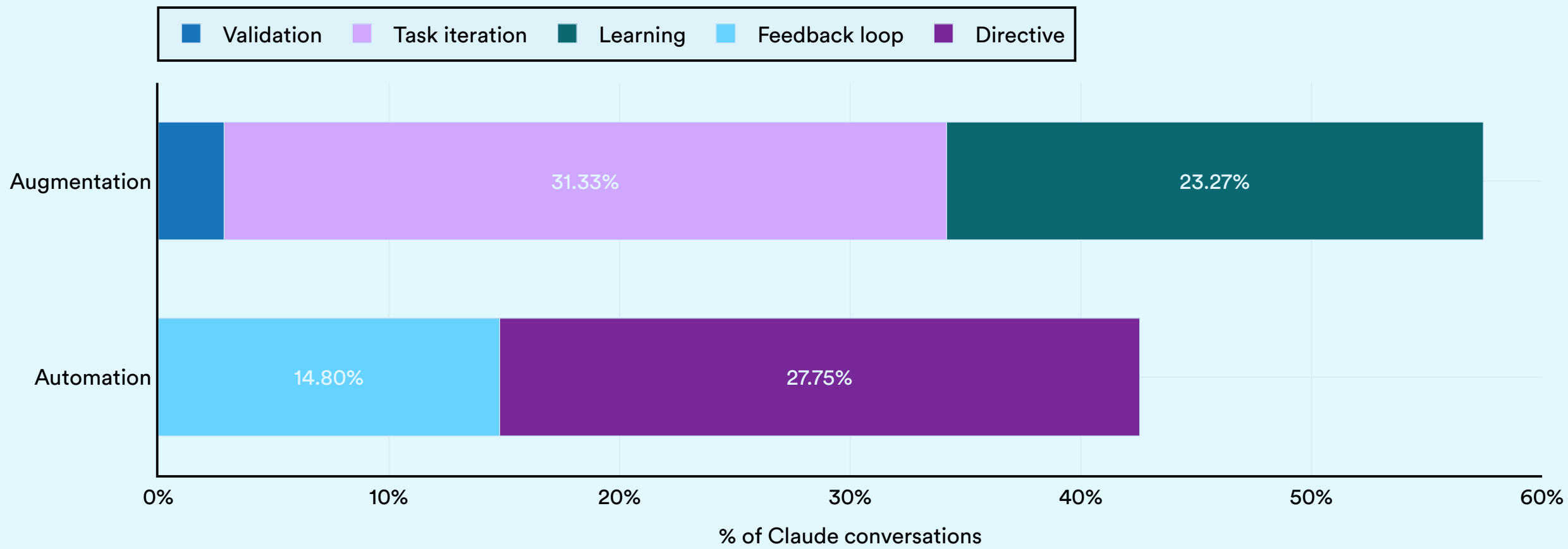

Figure 4.2.27

**Distribution of occupational skills exhibited by Claude in conversations**

Source: Handa et al., 2025

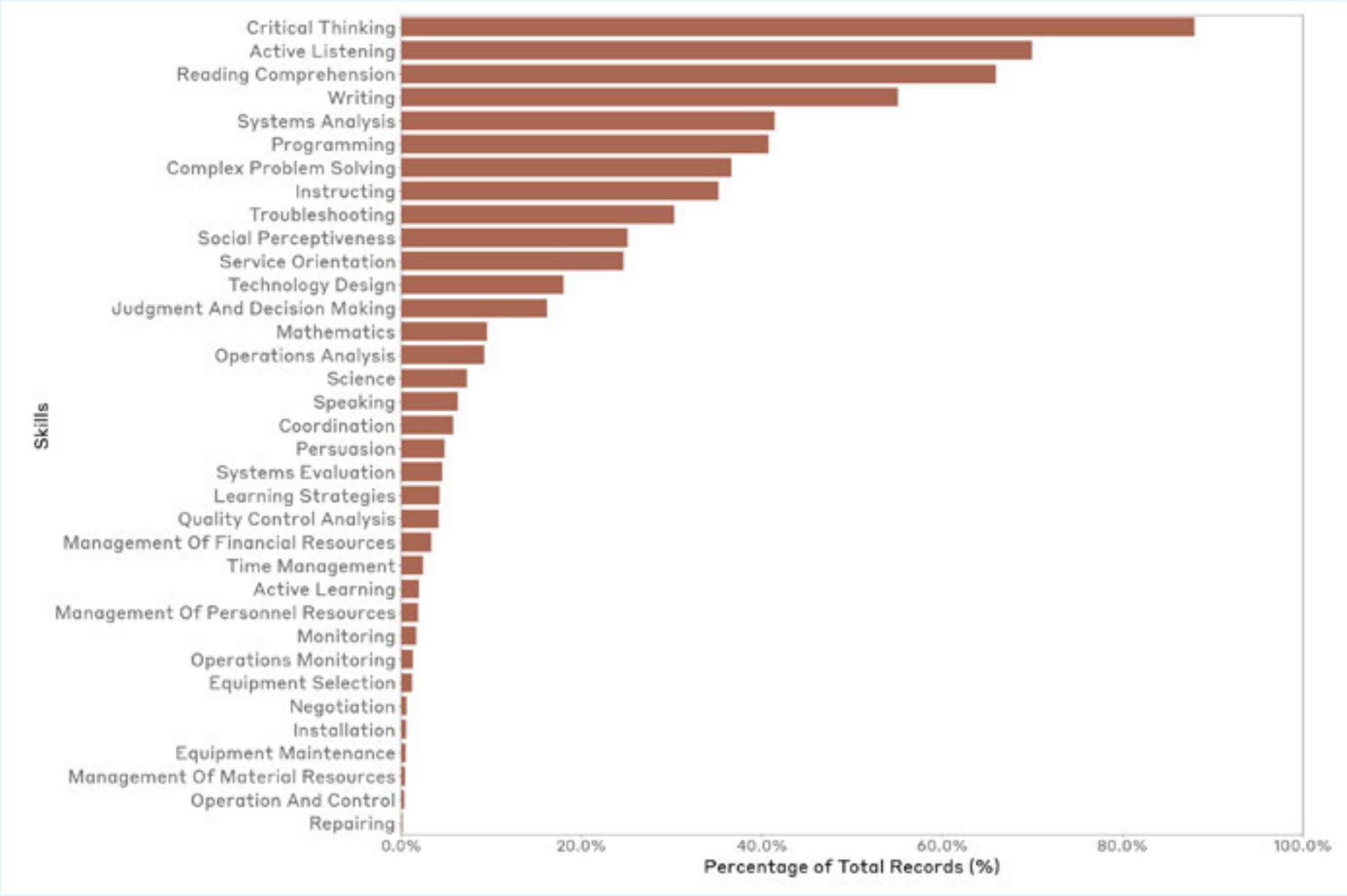

Figure 4.2.28

This section monitors AI investment trends, leveraging data from Quid, which analyzes investment data from more than 8 million companies worldwide, both public and private. Employing natural language processing, Quid sifts through vast unstructured datasets—including news aggregations, blogs, company records, and patent databases—to detect patterns and insights. Additionally, Quid is constantly expanding its database to include more companies, sometimes resulting in higher reported investment volumes for specific years. For the first time, this year's investment section in the AI Index includes data on generative AI investments.

# 4.3 Investment

## Corporate Investment

Figure 4.3.1 illustrates the trend in global corporate AI investment from 2013 to 2024, including mergers and acquisitions, minority stakes, private investments, and public offerings.

In 2024, the total investment grew to $252.3 billion, an increase of 25.5% from 2023. The most significant upturn occurred in private investment, which rose by 44.5% compared with the previous year, while mergers and acquisitions increased by 12.1%. Over the past decade, AI-related investments have increased nearly thirteenfold.

**Global corporate investment in AI by investment activity, 2013–24**

Source: Quid, 2024 | Chart: 2025 AI Index report

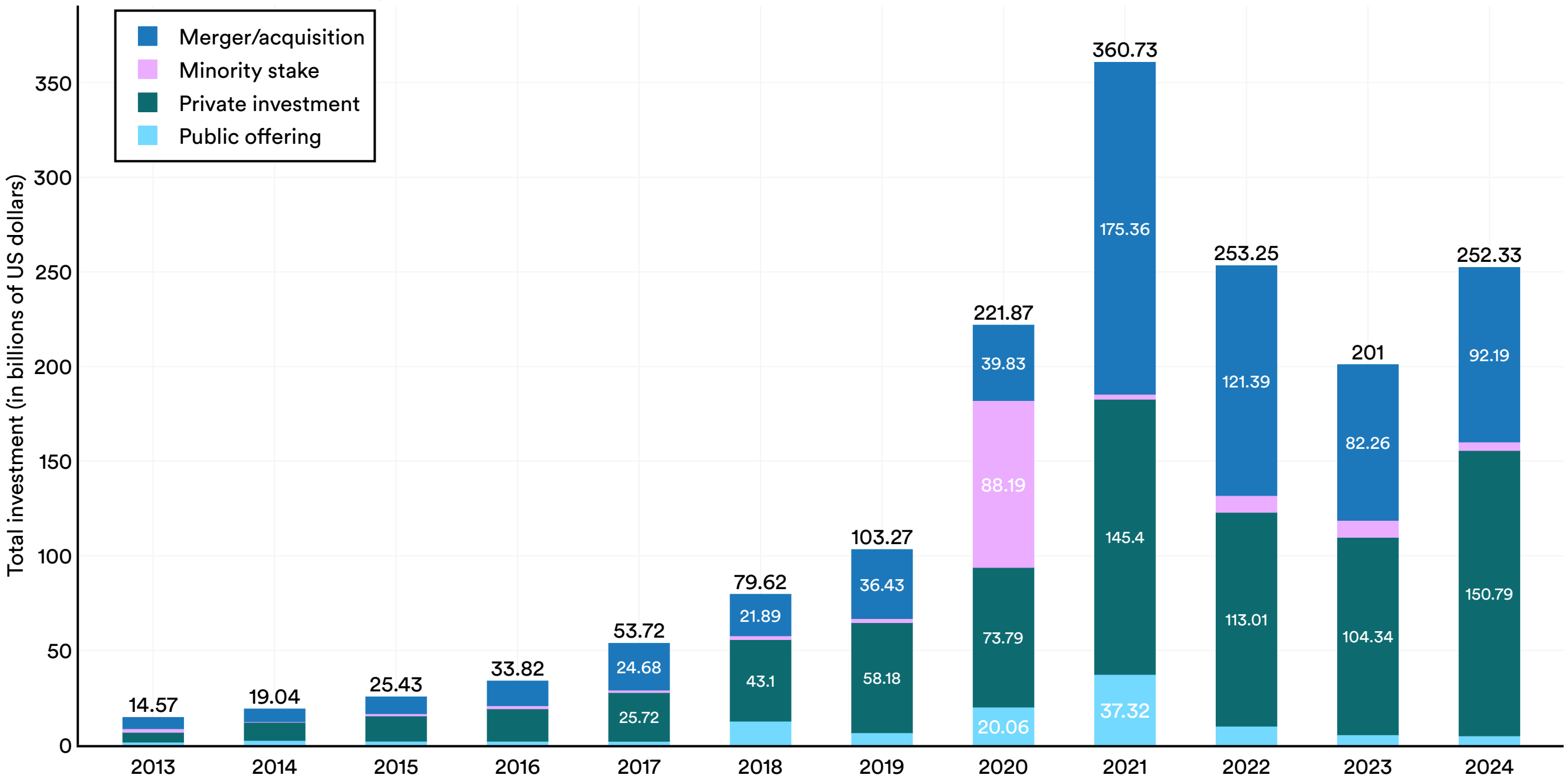

Figure 4.3.1

# Startup Activity

This section analyzes private investment trends in AI startups that have received over $1.5 million in investment since 2013.

### Global Trends

Global private AI investment increased 44.5% between 2023 and 2024, marking the first year-over-year growth since 2021 (Figure 4.3.2). Despite recent fluctuations, private AI investment globally has grown substantially in the last decade.

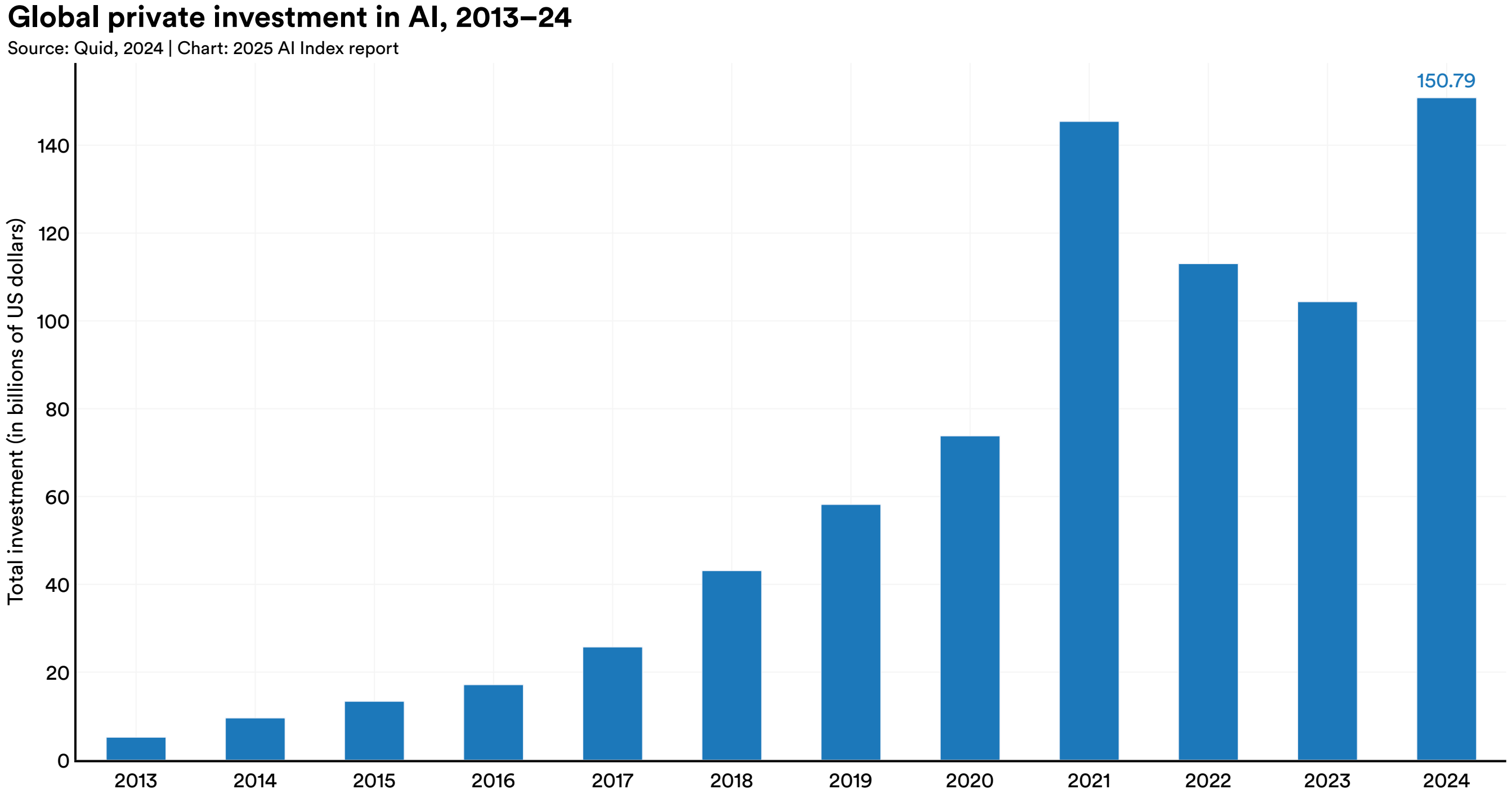

Figure 4.3.2

Funding for generative AI continued to increase sharply (Figure 4.3.3). In 2024, the sector attracted $33.9 billion, representing an 18.7% increase from 2023 and over 8.5 times the investment of 2022. Furthermore, generative AI accounted for more than a fifth of all AI-related private investment in 2024.

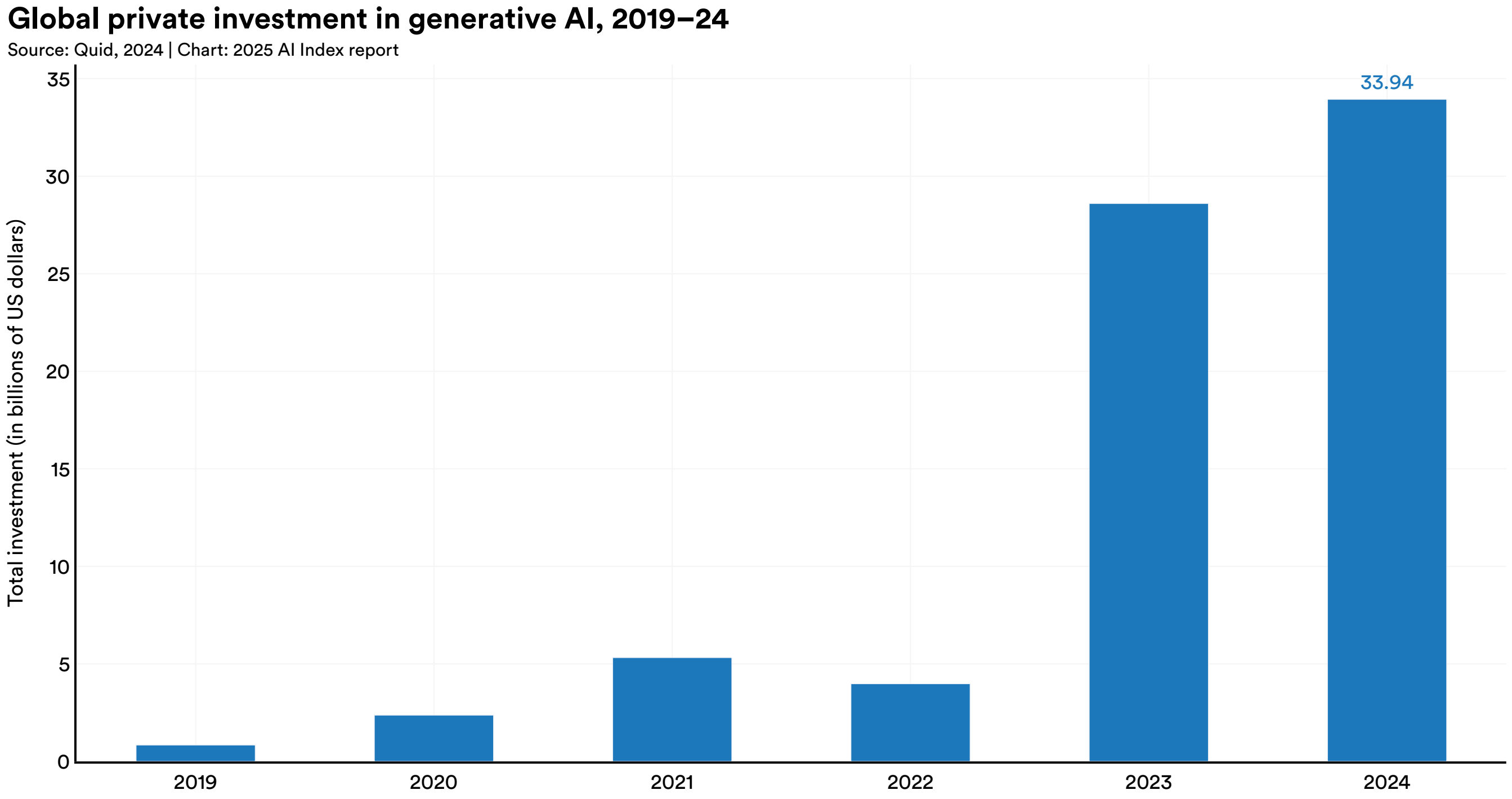

Figure 4.3.3

The number of newly funded AI companies in 2024 jumped to 2,049, an 8.4% increase over the previous year (Figure 4.3.4). In addition, 2024 registered an increase in the number of newly funded generative AI companies, with 214 new startups receiving funding, compared to 179 in 2023, and 31 in 2019 (Figure 4.3.5).

**Number of newly funded AI companies in the world, 2013–24**
Source: Quid, 2024 | Chart: 2025 AI Index report

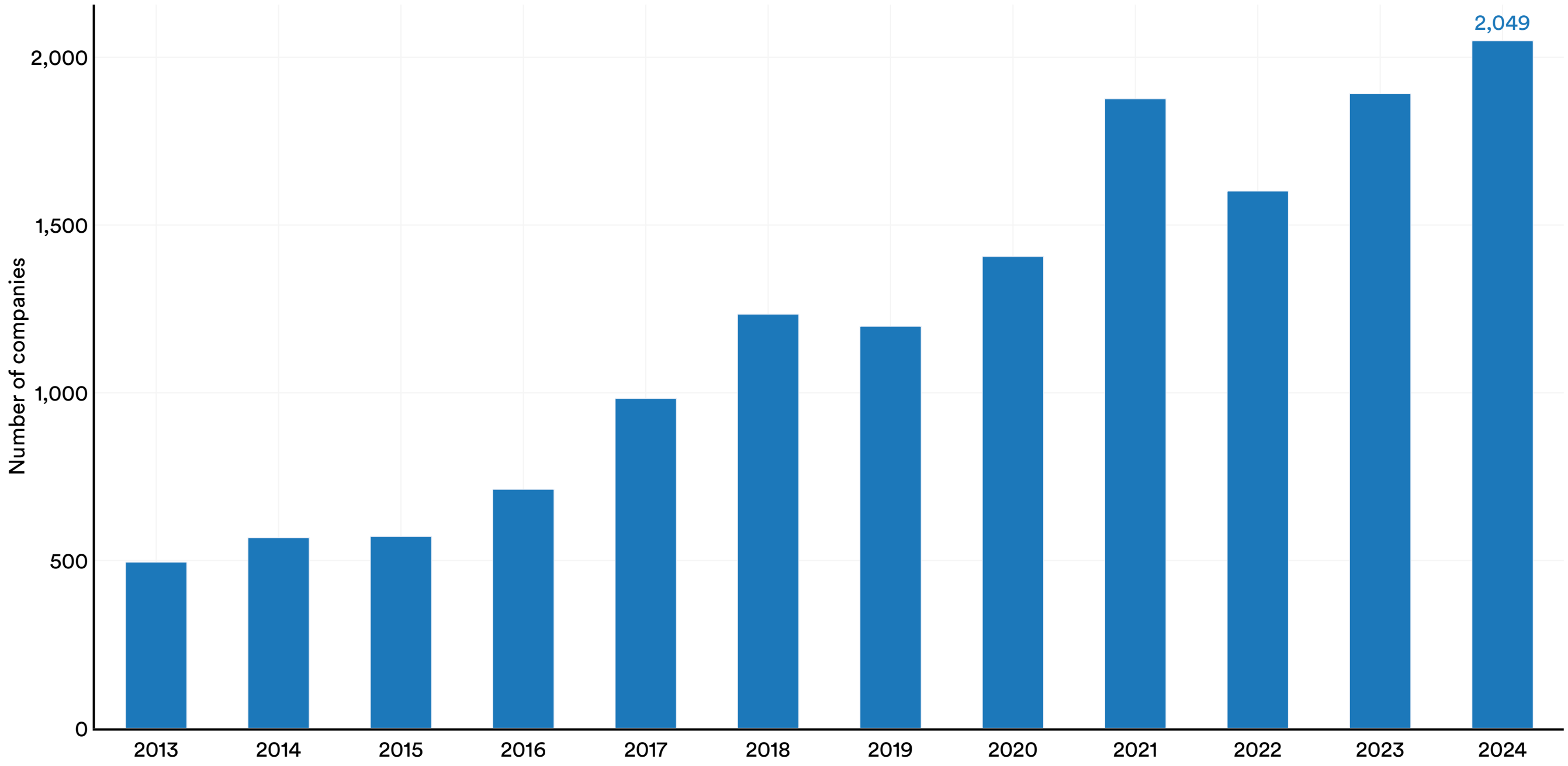

Figure 4.3.4

**Number of newly funded generative AI companies in the world, 2019–24**
Source: Quid, 2024 | Chart: 2025 AI Index report

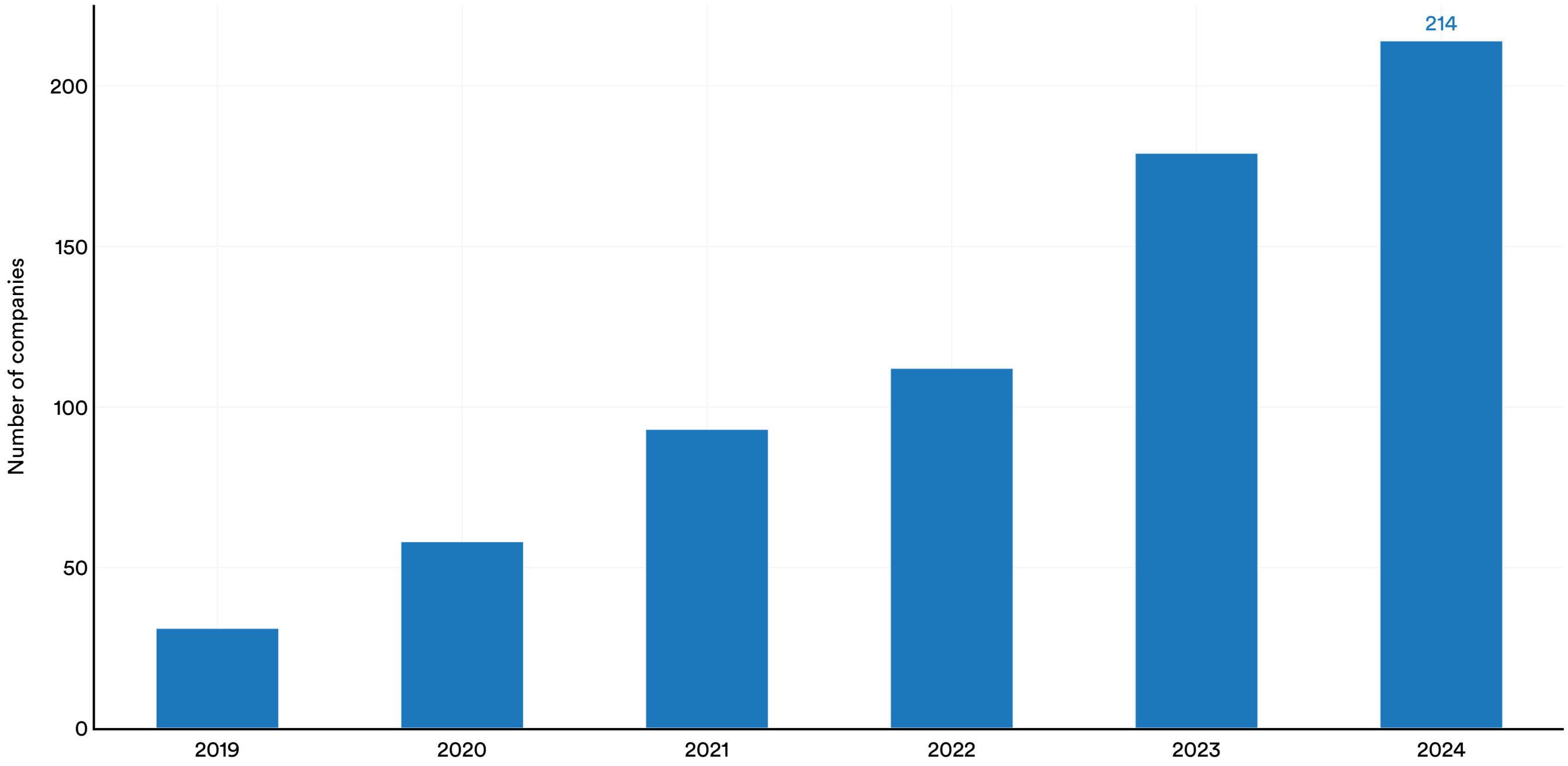

Figure 4.3.5

Figure 4.3.6 visualizes the average size of AI private investment events, calculated by dividing the total yearly AI private investment by the total number of AI private investment events. From 2023 to 2024, the average increased significantly, growing from $31.6 million to $45.4 million.

Figure 4.3.7 reports AI funding events disaggregated by size. In 2024, AI private investment events increased across funding size categories exceeding $100 million and decreased or remained constant in smaller categories. In 2024, there were 15 AI private investment events that involved funding sizes greater than $1 billion.

**Average size of global AI private investment events, 2013–24**
Source: Quid, 2024 | Chart: 2025 AI Index report

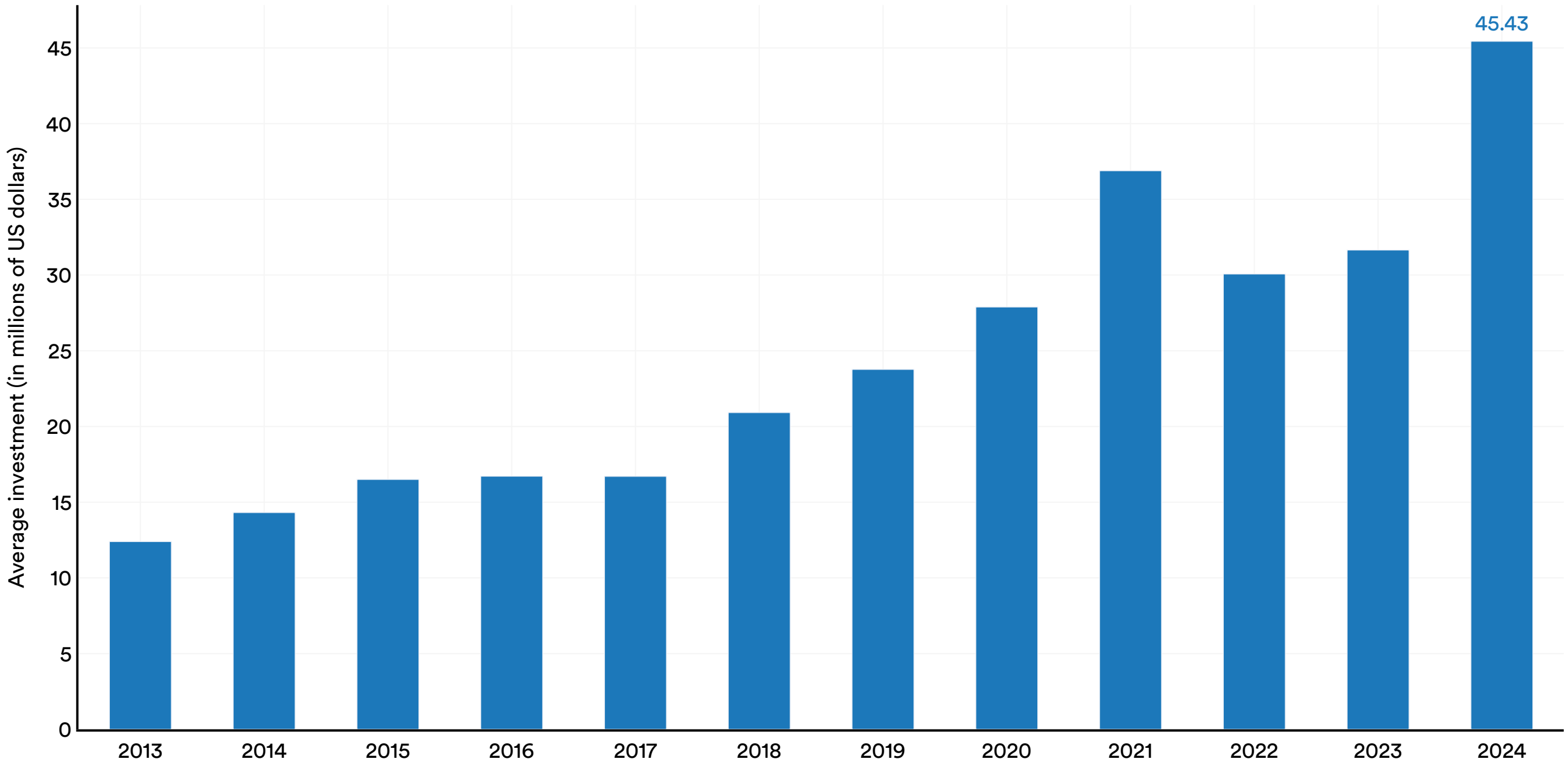

Figure 4.3.6

**Global AI private investment events by funding size, 2023 vs. 2024**
Source: Quid, 2024 | Table: 2025 AI Index report

| Funding size | 2023 | 2024 |
|---|---|---|
| Over 1 billion | 9 | 15 |
| 500 million – 1 billion | 9 | 20 |
| 100 million – 500 million | 134 | 143 |
| 50 million – 100 million | 200 | 196 |
| Under 50 million | 2,945 | 2,945 |
| Undisclosed | 680 | 207 |
| Total | 3,977 | 3,526 |

Figure 4.3.7

## Regional Comparison by Funding Amount

The United States once again led the world in terms of total AI private investment. In 2024, the $109.1 billion invested in the United States was 11.7 times greater than the amount invested in the next highest country, China ($9.3 billion), and 24.1 times the amount invested in the United Kingdom ($4.5 billion) (Figure 4.3.8). Other notable countries that rounded out the top 15 in 2024 include Sweden ($4.3 billion), Austria ($1.5 billion), the Netherlands ($1.1 billion), and Italy ($0.9 billion).

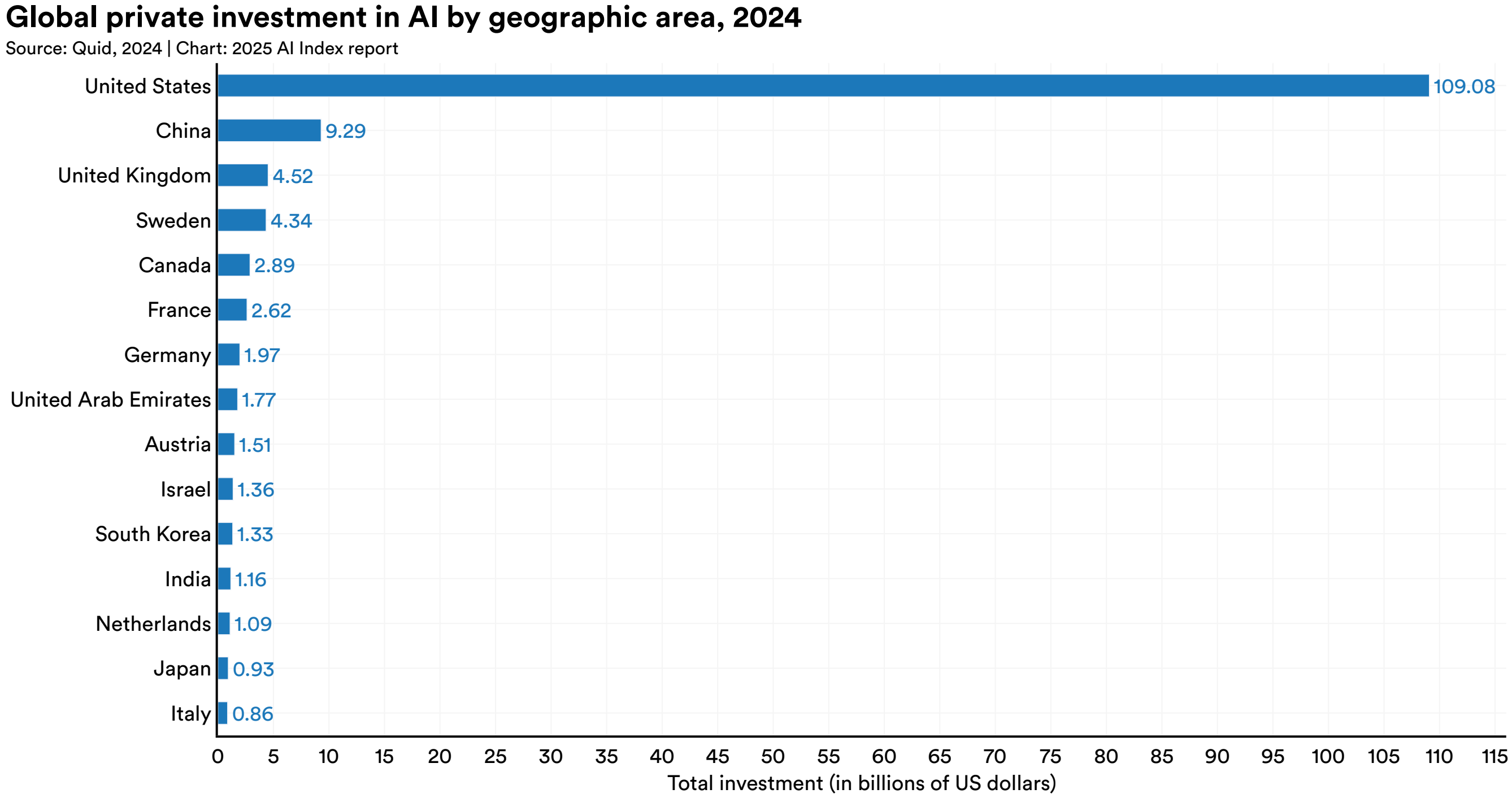

Figure 4.3.8

When aggregating private AI investments since 2013, the country rankings remain the same: The United States leads with $470.9 billion invested, followed by China with $119.3 billion, and the United Kingdom with $28.2 billion (Figure 4.3.9). Other countries that have attracted significant AI investment over the past decade include Israel ($15.0 billion), Singapore ($7.3 billion), and Sweden ($7.3 billion).

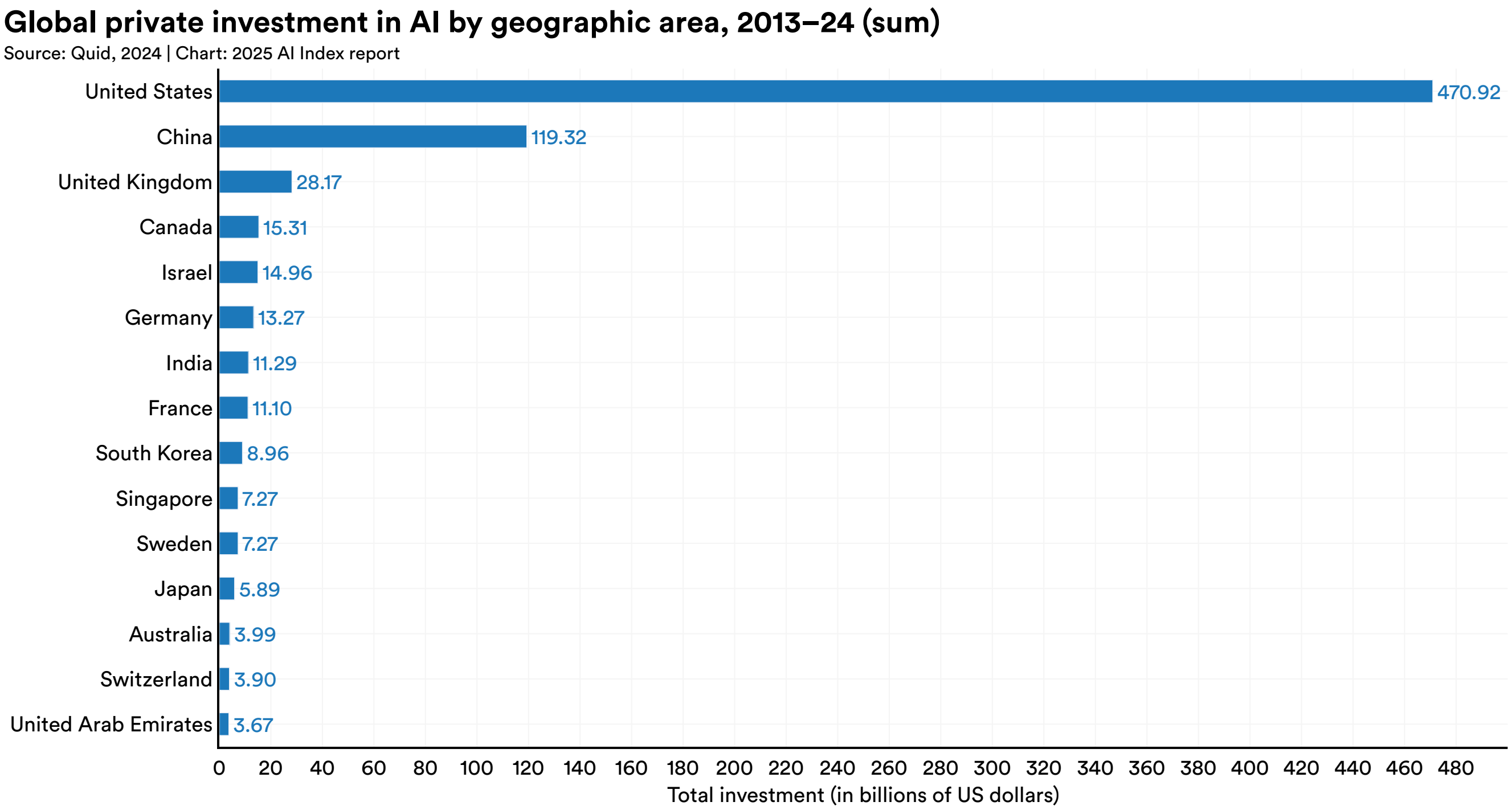

Figure 4.3.9

Figure 4.3.10, which looks at AI private investment over time by geographic area, suggests that the gap in private investments between the United States and other regions is widening. While AI private investments have decreased in China (-1.9%) and increased in Europe (+60%) since 2023, the United States has seen a significant increase (+50.7%) during the same period—and a +78.3% increase since 2022.

**Global private investment in AI by geographic area, 2013–24**
Source: Quid, 2024 | Chart: 2025 AI Index report

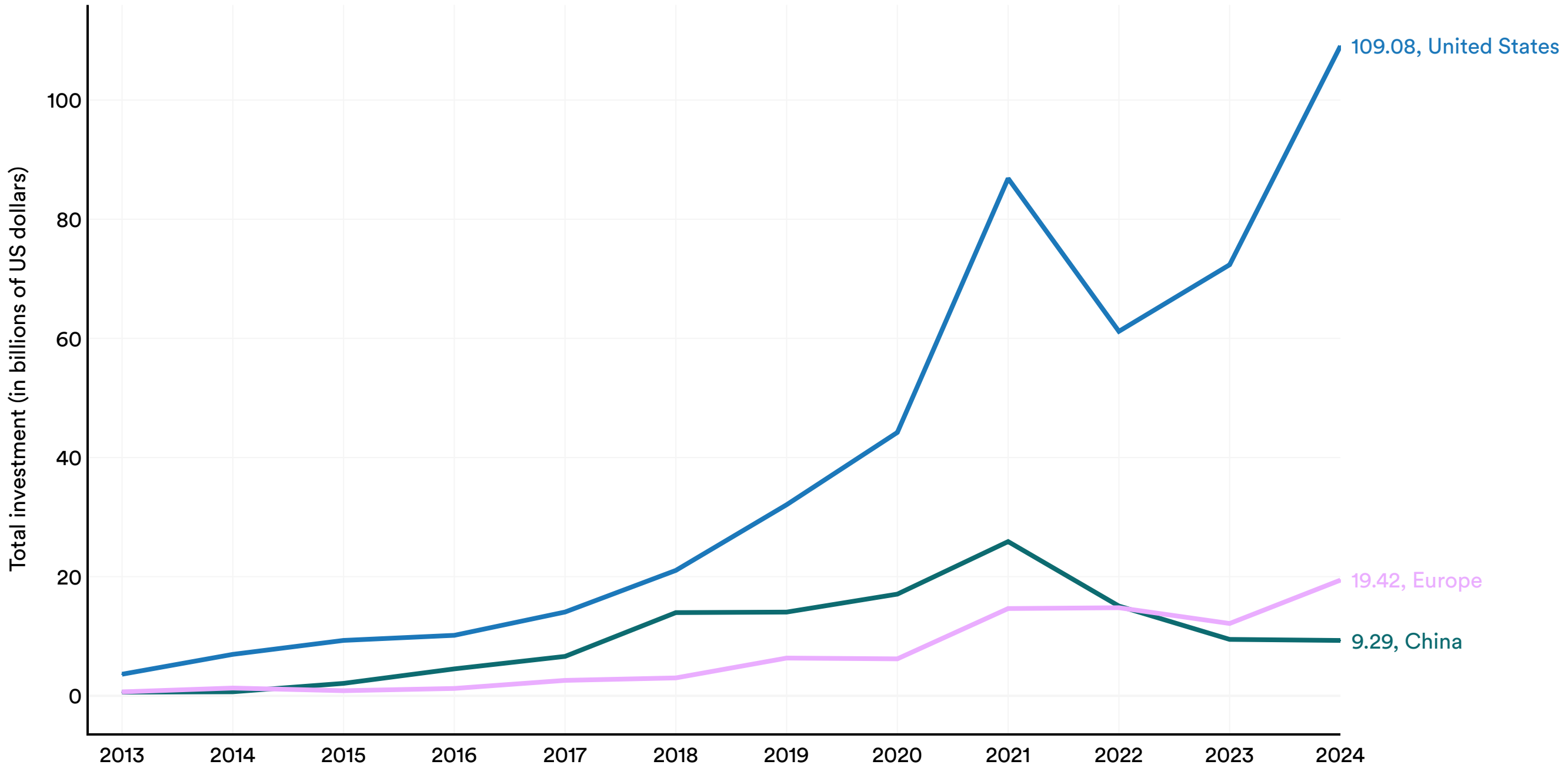

Figure 4.3.10

The disparity in regional AI private investment becomes particularly pronounced when examining generative AI-related investments. For instance, in 2023, the United States outpaced the combined investments of China and Europe in generative AI by approximately $21.8 billion (Figure 4.3.11). By 2024, this gap widened to $25.4 billion.

**Global private investment in generative AI by geographic area, 2019–24**
Source: Quid, 2024 | Chart: 2025 AI Index report

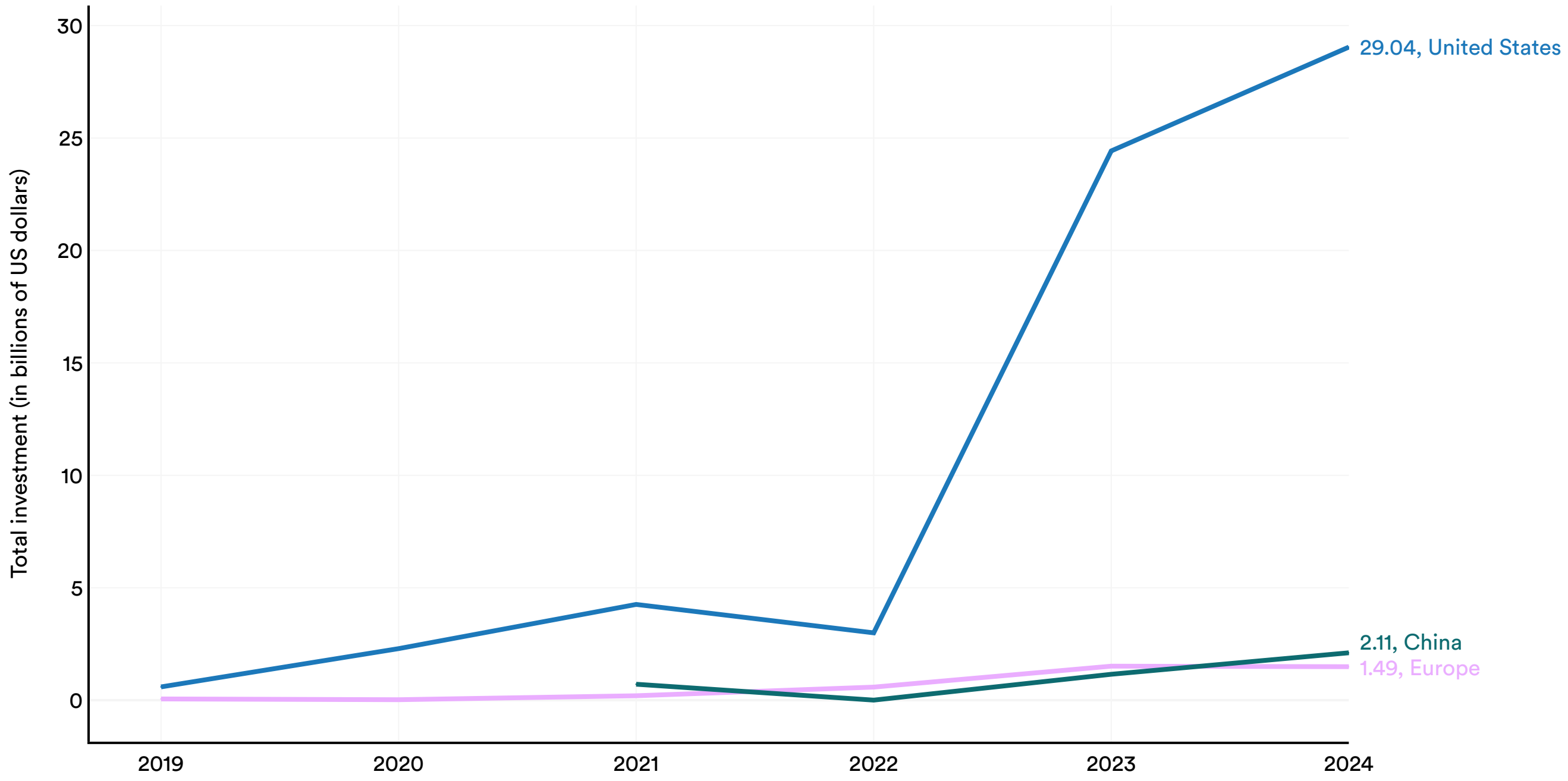

Figure 4.3.11

## Regional Comparison by Newly Funded AI Companies

This section examines the number of newly funded AI companies across different geographic regions. Consistent with trends in private investment, the United States leads all regions with 1,073 new AI companies, followed by the United Kingdom with 116, and China with 98 (Figure 4.3.12).

**Number of newly funded AI companies by geographic area, 2024**
Source: Quid, 2024| Chart: 2025 AI Index report

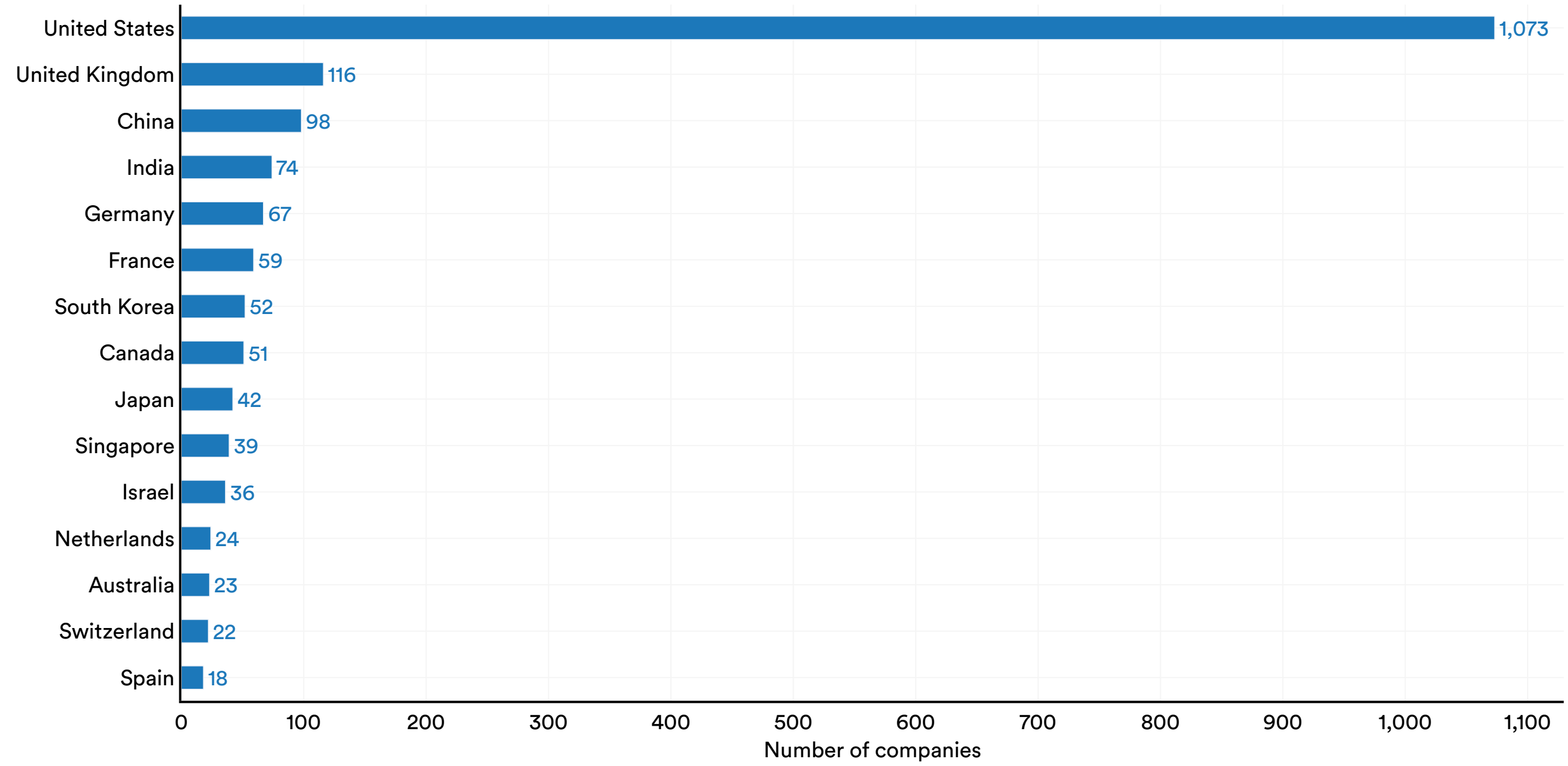

Figure 4.3.12

A similar trend is evident in the aggregate data since 2013. In the last decade, the number of newly funded AI companies in the United States is around 4.3 times the amount in China, and 7.9 times the amount in the United Kingdom (Figure 4.3.13).

**Number of newly funded AI companies by geographic area, 2013–24 (sum)**
Source: Quid, 2024 | Chart: 2025 AI Index report

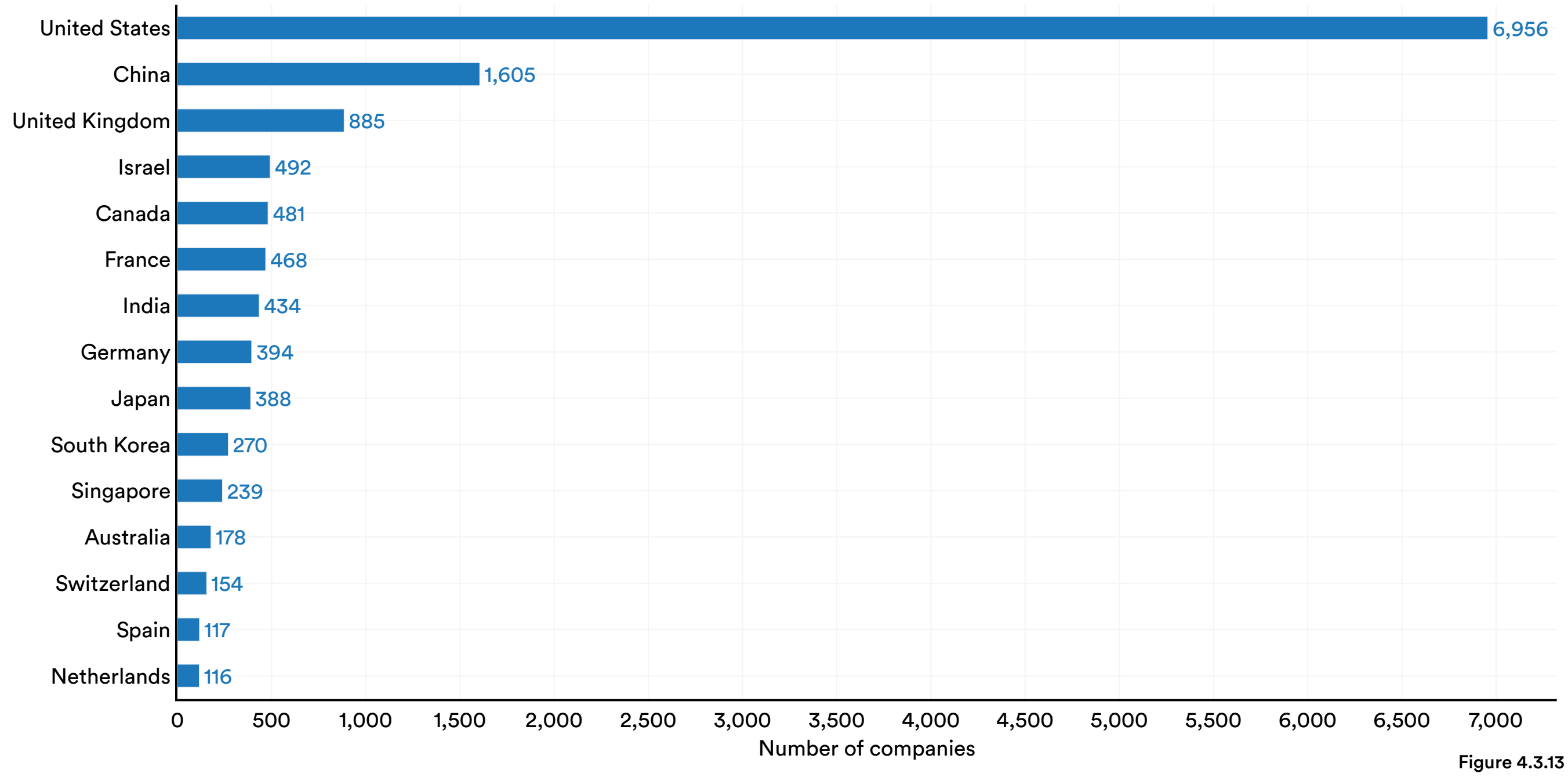

Figure 4.3.13

Figure 4.3.14 presents data on newly funded AI companies in specific geographic regions, highlighting a decade-long pattern in which the United States consistently surpasses both Europe and China. Since 2022, the United States, along with Europe, has seen significant increases in the number of new AI companies, in contrast to China, which experienced a second consecutive annual decline.

**Number of newly funded AI companies by geographic area, 2013–24**
Source: Quid, 2024 | Chart: 2025 AI Index report

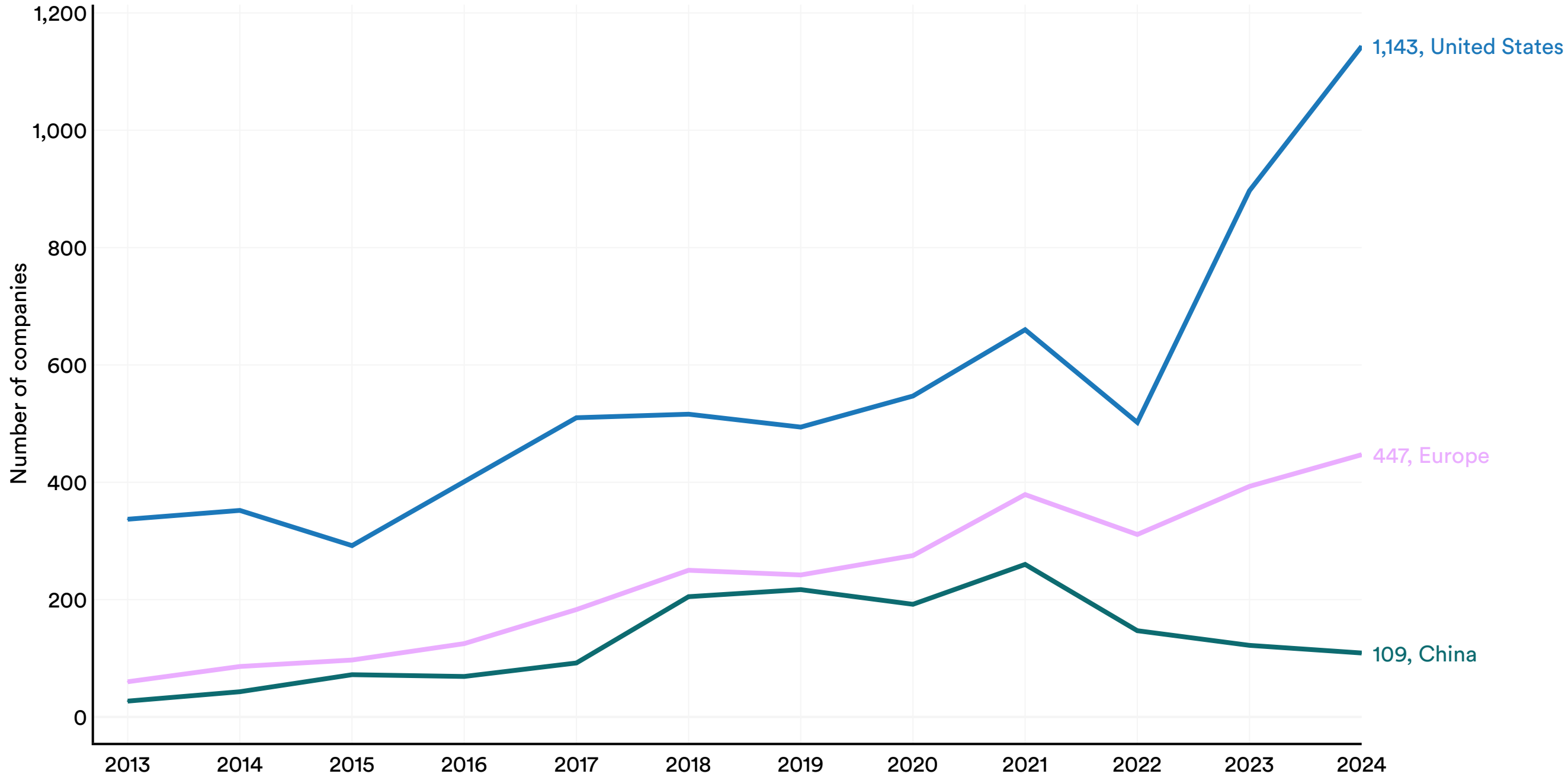

Figure 4.3.14

## Focus Area Analysis

Quid also disaggregates private AI investment by focus area. Figure 4.3.15 compares global private AI investment by focus area in 2024 versus 2023. The focus areas that attracted the most investment in 2024 were AI infrastructure/research/governance ($37.3 billion); data management and processing ($16.6 billion); and medical and healthcare ($11 billion). The prominence of AI infrastructure, research, and governance reflects large investments in companies specifically building AI applications, such as OpenAI, Anthropic, and xAI.

### Global private investment in AI by focus area, 2023 vs. 2024
Source: Quid, 2024 | Chart: 2025 AI Index report

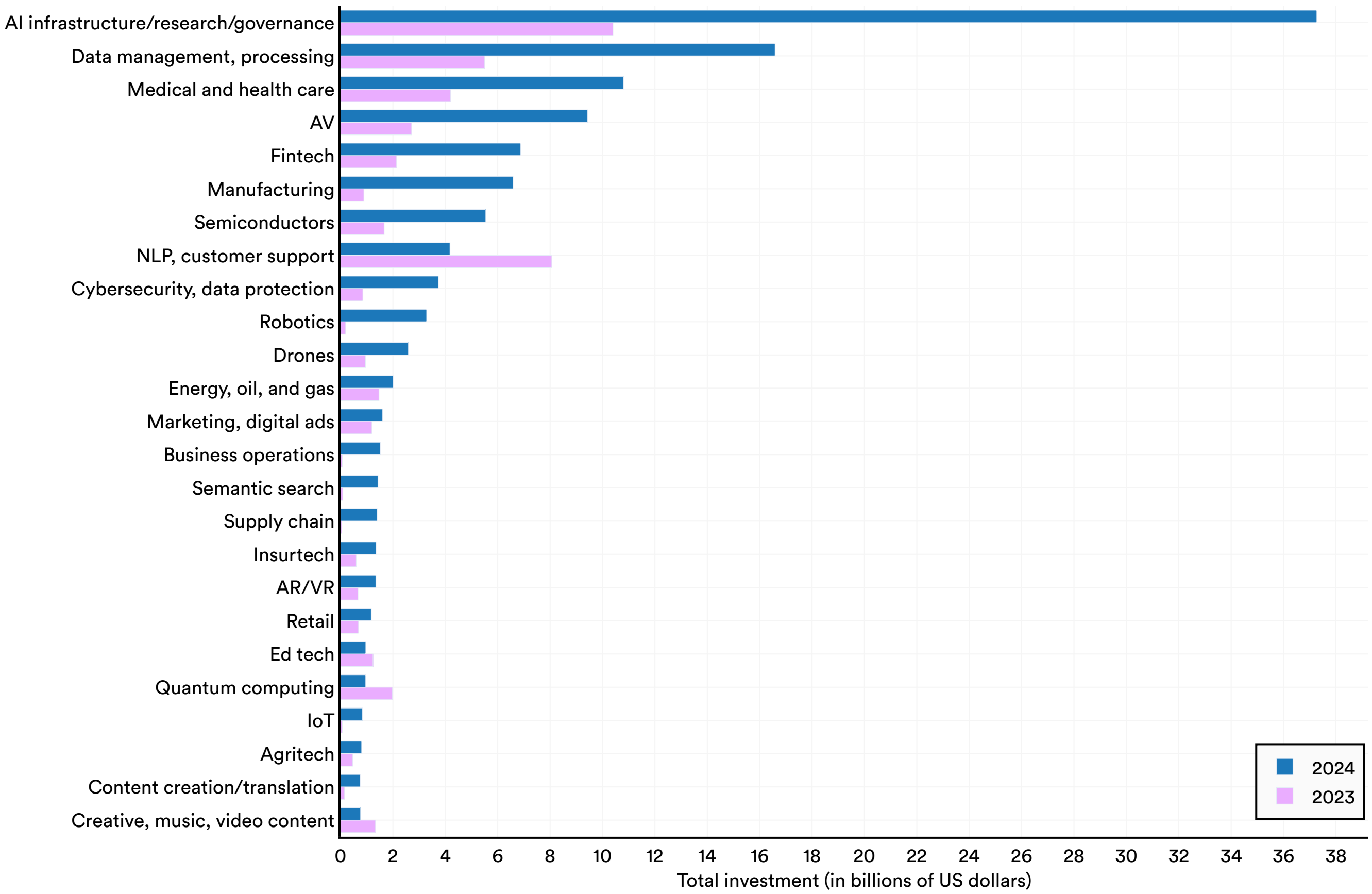

Figure 4.3.15

Figure 4.3.16 presents trends over time in AI focus area investments. As noted earlier, most focus areas saw a boost in investments in the last year. While still substantial, investment in NLP, customer support peaked in 2021 and has since then declined.

## Global private investment in AI by focus area, 2018–24

Source: Quid, 2024 | Chart: 2025 AI Index report

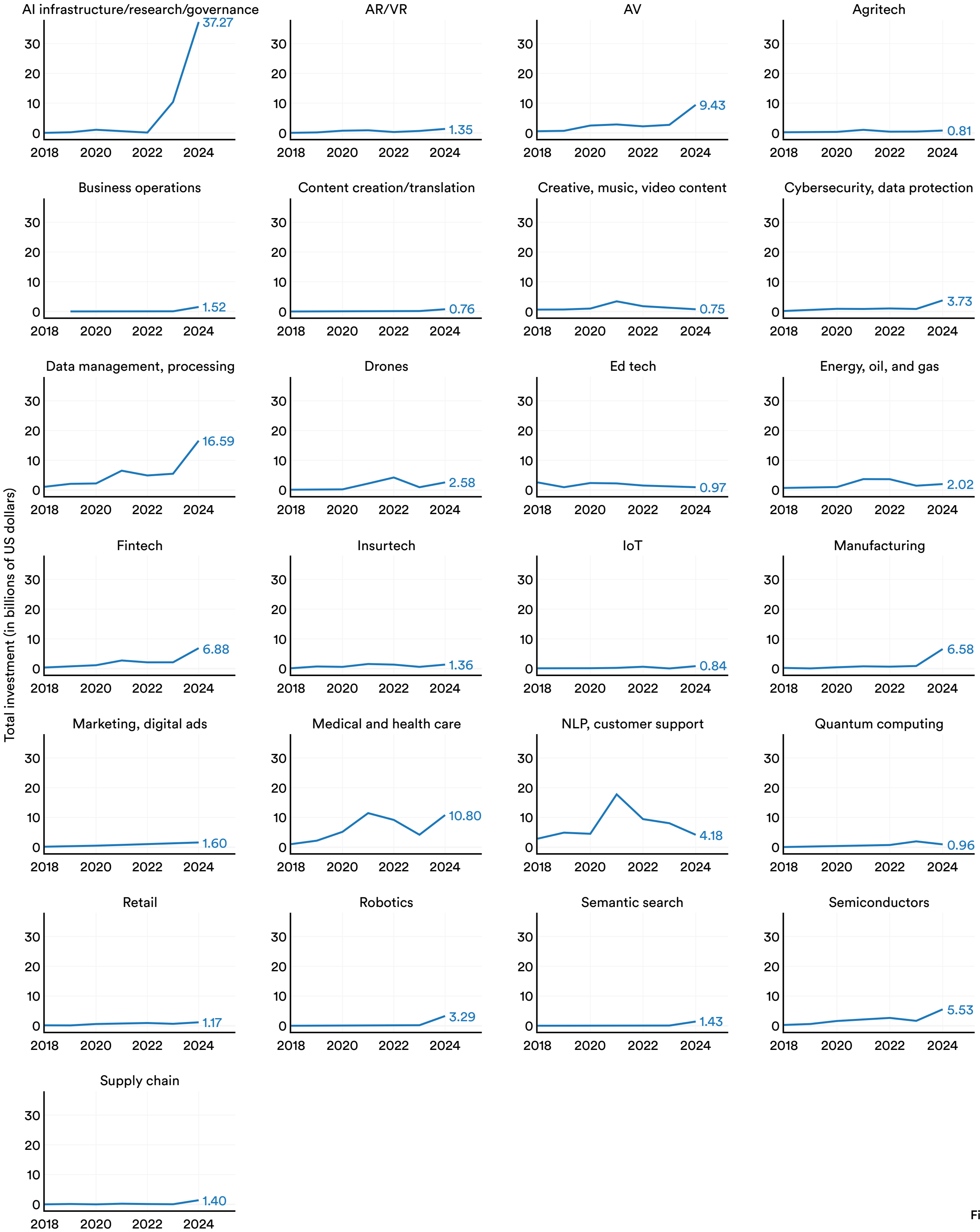

Figure 4.3.16

This section examines the practical application of AI by corporations, highlighting industry usage trends, how businesses are integrating AI, the specific AI technologies deemed most beneficial, and the impact of AI usage on financial performance.

# 4.4 Corporate Activity

## Industry Usage

This section incorporates insights from McKinsey's publications on the state of AI alongside data from prior editions. The 2024 McKinsey analysis is based on two surveys spanning 2,854 respondents across various regions, industries, company sizes, functional areas, and tenures.

### Use of AI Capabilities

Business use of AI increased significantly after stagnating between 2017 and 2023. The latest McKinsey report reveals that 78% of surveyed respondents say their organizations have begun to use AI in at least one business function, marking a significant increase from 55% in 2023 (Figure 4.4.1). Use of generative AI, which was covered for the first time in last year's survey, more than doubled year over year, with 71% of respondents in 2024 saying their organizations regularly use the technology in at least one business function, compared to 33% in 2023.

**Share of respondents who say their organization uses AI in at least one function, 2017–24**
Source: McKinsey & Company Survey, 2024 | Chart: 2025 AI Index report

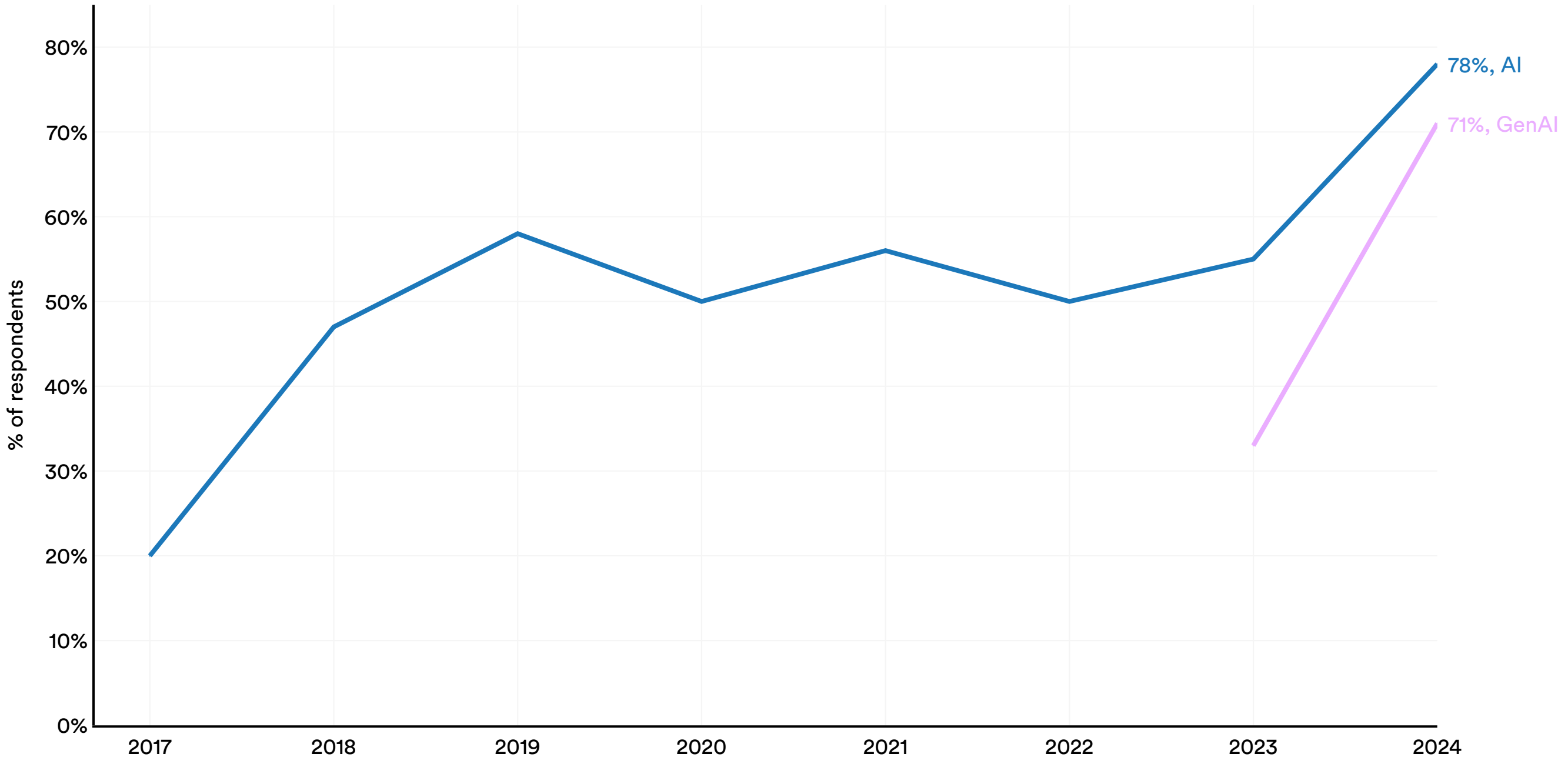

Figure 4.4.1

Figure 4.4.2 shows AI usage by industry and AI function in 2024. The greatest usage was in IT for tech (48%), followed by product and/or service development for tech (47%) and marketing and sales for tech (47%).

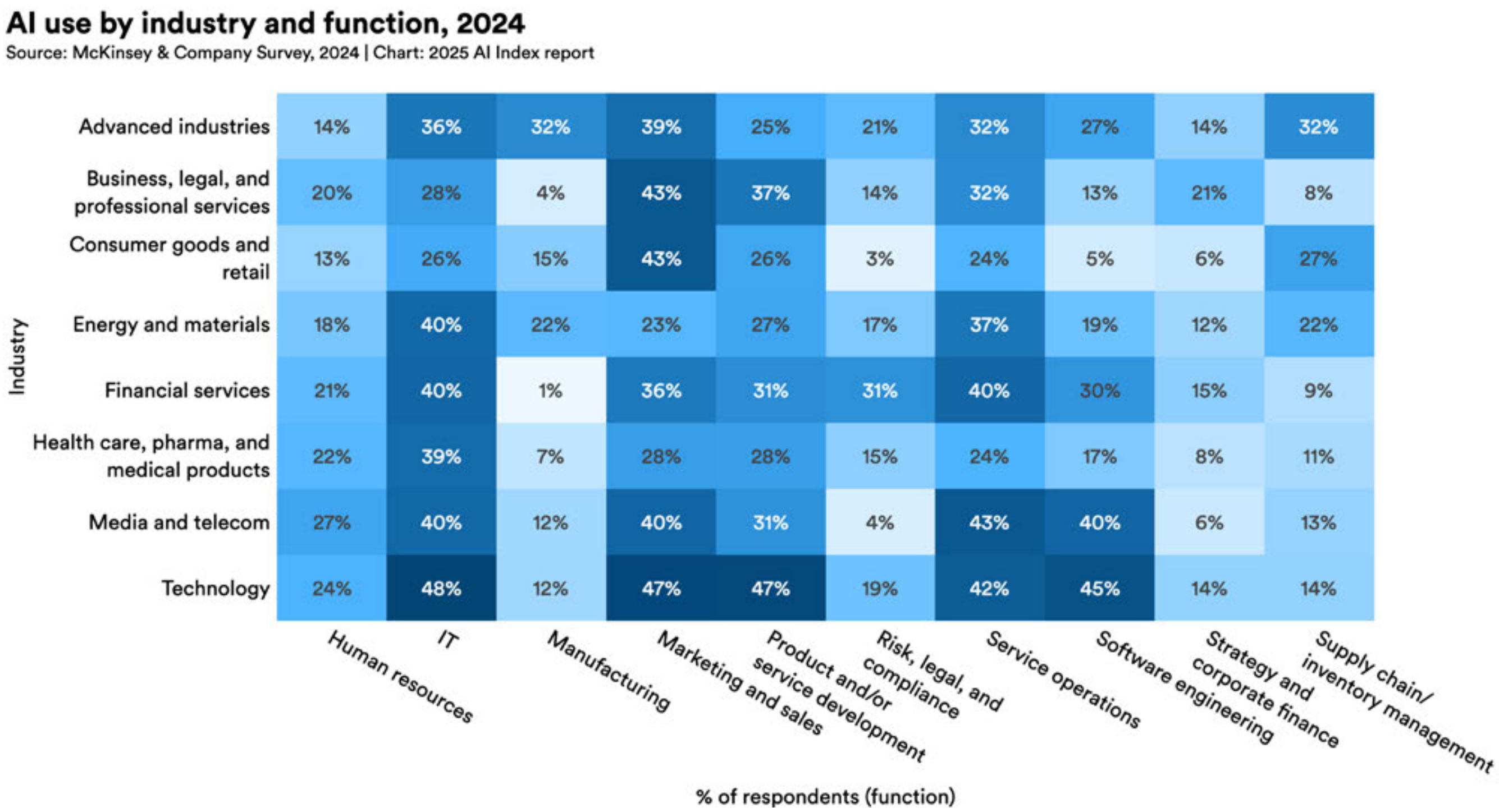

| Industry | Human resources | IT | Manufacturing | Marketing and sales | Product and/or service development | Risk, legal, and compliance | Service operations | Software engineering | Strategy and corporate finance | Supply chain/ inventory management |
|---|---|---|---|---|---|---|---|---|---|---|
| Advanced industries | 14% | 36% | 32% | 39% | 25% | 21% | 32% | 27% | 14% | 32% |
| Business, legal, and professional services | 20% | 28% | 4% | 43% | 37% | 14% | 32% | 13% | 21% | 8% |
| Consumer goods and retail | 13% | 26% | 15% | 43% | 26% | 3% | 24% | 5% | 6% | 27% |
| Energy and materials | 18% | 40% | 22% | 23% | 27% | 17% | 37% | 19% | 12% | 22% |
| Financial services | 21% | 40% | 1% | 36% | 31% | 31% | 40% | 30% | 15% | 9% |
| Health care, pharma, and medical products | 22% | 39% | 7% | 28% | 28% | 15% | 24% | 17% | 8% | 11% |
| Media and telecom | 27% | 40% | 12% | 40% | 31% | 4% | 43% | 40% | 6% | 13% |
| Technology | 24% | 48% | 12% | 47% | 47% | 19% | 42% | 45% | 14% | 14% |

Figure 4.4.2[8]

8 "Advanced industries" comprises respondents from sectors such as advanced electronics, aerospace and defense, automotive and assembly, and semiconductors. "Energy and materials" encompasses respondents from agriculture, chemicals, electric power and natural gas, metals and mining, oil and gas, as well as paper, forest products, and packaging.

Organizations have reported both cost reductions and revenue increases where they have started using AI, but most commonly at low levels (Figure 4.4.3). The areas where respondents most frequently reported that their use of AI has resulted in cost savings were service operations (49%), supply chain and inventory management (43%), and software engineering (41%). For revenue gains, the functions that most commonly benefited from their use of AI include marketing and sales (71%), supply chain and inventory management (63%), and service operations (57%).

### Cost decrease and revenue increase from analytical AI use by function, 2024

Source: McKinsey & Company Survey, 2024 | Chart: 2025 AI Index report

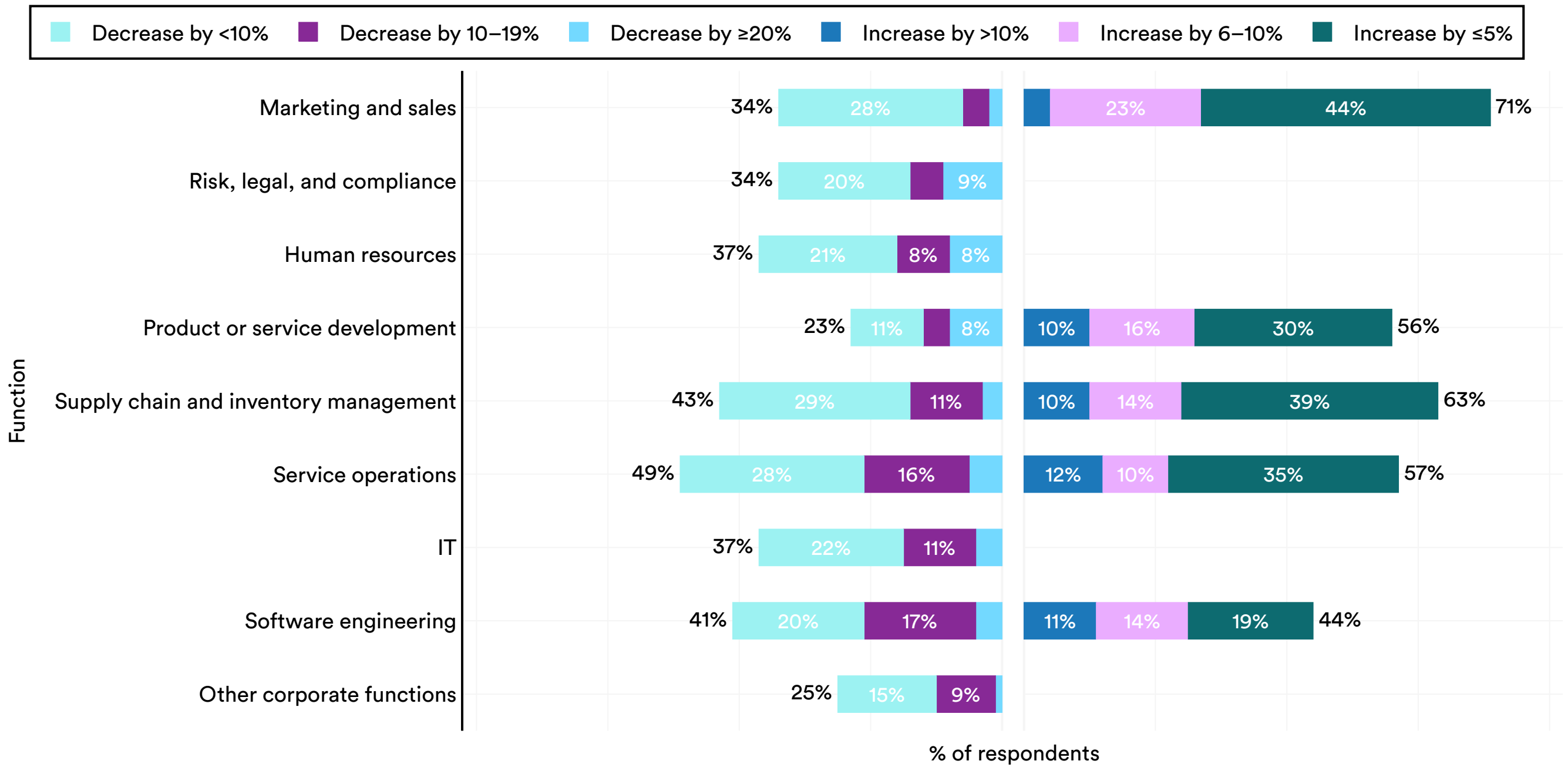

Figure 4.4.3

Figure 4.4.4 presents global AI usage by organizations, segmented by regions. In 2024, surveyed respondents in every region reported increased use of AI compared with 2023. One of the most significant year-over-year growth rates in AI use was seen in Greater China, where organizations' reported use grew by 27 percentage points. North America remains the leader in use of AI (82%), but only by a small margin. Europe also experienced a significant increase in AI usage rates, growing by 23 percentage points to 80% since 2023.

**AI use by organizations in the world, 2023 vs. 2024**
Source: McKinsey & Company Survey, 2024 | Chart: 2025 AI Index report

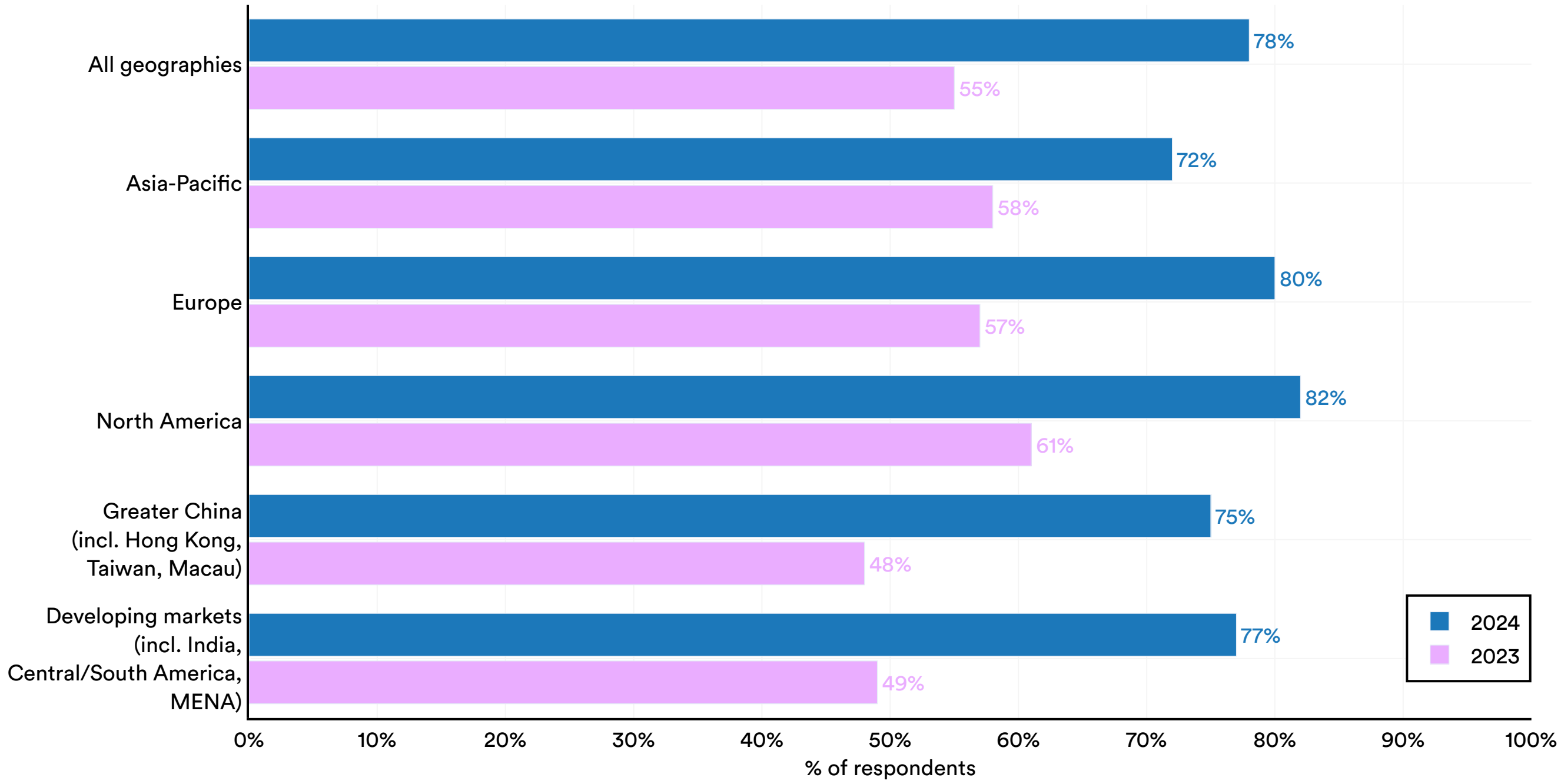

Figure 4.4.4

## Deployment of AI Capabilities

How are organizations deploying AI? Figure 4.4.5 highlights the proportion of total surveyed respondents that report using generative AI for a particular function. It is possible for respondents to indicate that they deploy AI for multiple purposes.

The most common application is marketing strategy content support (27%), followed by knowledge management (19%), personalization (19%), and design development (14%). Most of the leading reported use cases are within the marketing and sales function. A complementary survey of C-suite executives in developed markets found that only 1% described their generative AI rollouts as "mature." Overall, most companies are still in the early stages of capturing value at scale from AI.

**Most common generative AI use cases by function, 2024**
Source: McKinsey & Company Survey, 2024 | Chart: 2025 AI Index report

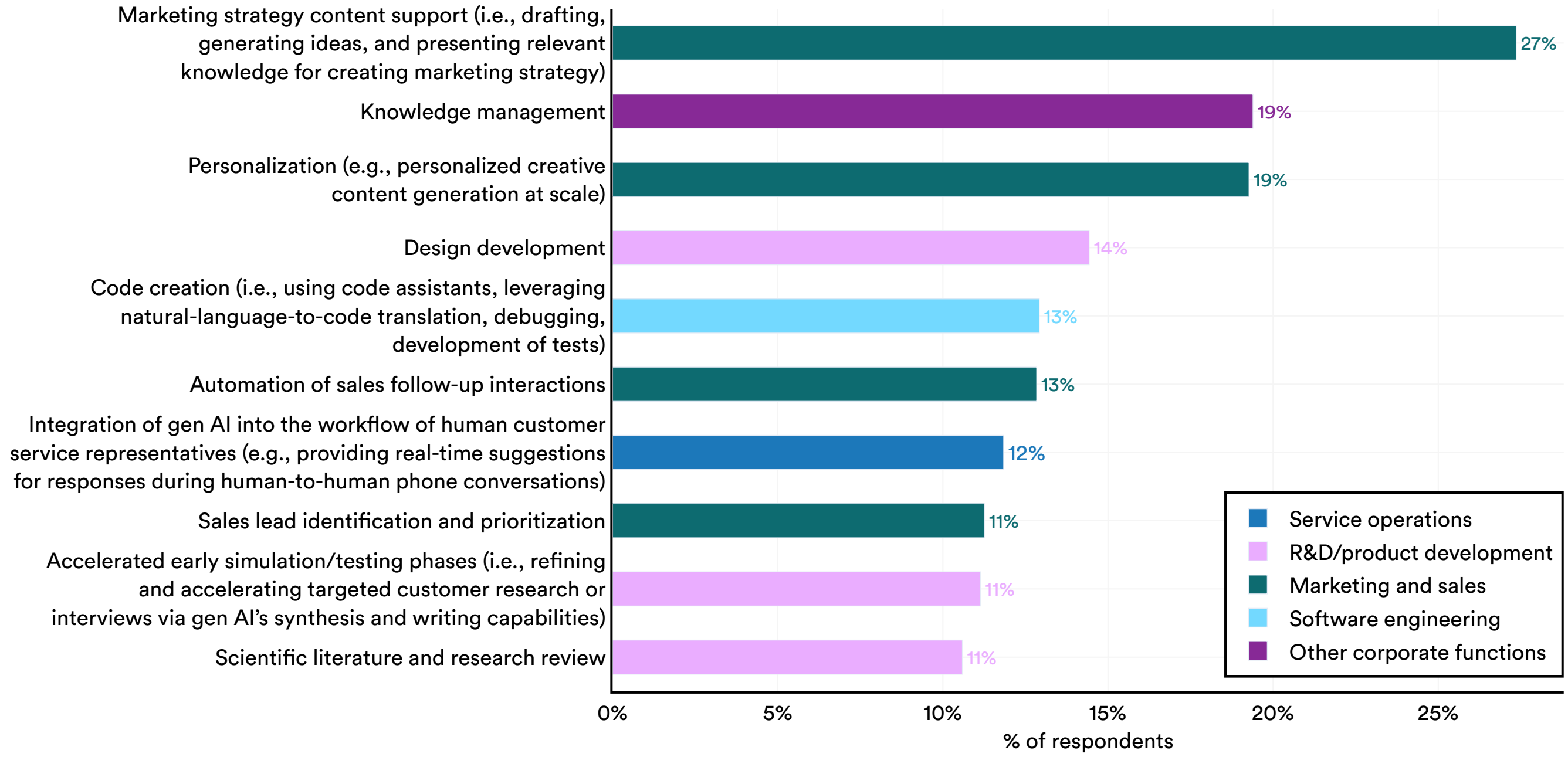

Figure 4.4.5

Figure 4.4.6 examines the proportion of respondents that report cost decreases and revenue increases from their organizations' use of generative AI in each business function. Overall, respondents report both cost reductions and revenue increases across various functions as a result of using generative AI, most commonly at low levels. The areas where respondents most frequently reported cost savings were supply chain and inventory management (61%), service operations (58%), and both human resources and strategy and corporate finance (56%). For revenue gains, the functions most commonly reporting benefits from generative AI include strategy and corporate finance (70%), supply chain and inventory management (67%), and marketing and sales (66%).

Cost decrease and revenue increase from generative AI use by function, 2024
Source: McKinsey & Company Survey, 2024 | Chart: 2025 AI Index report

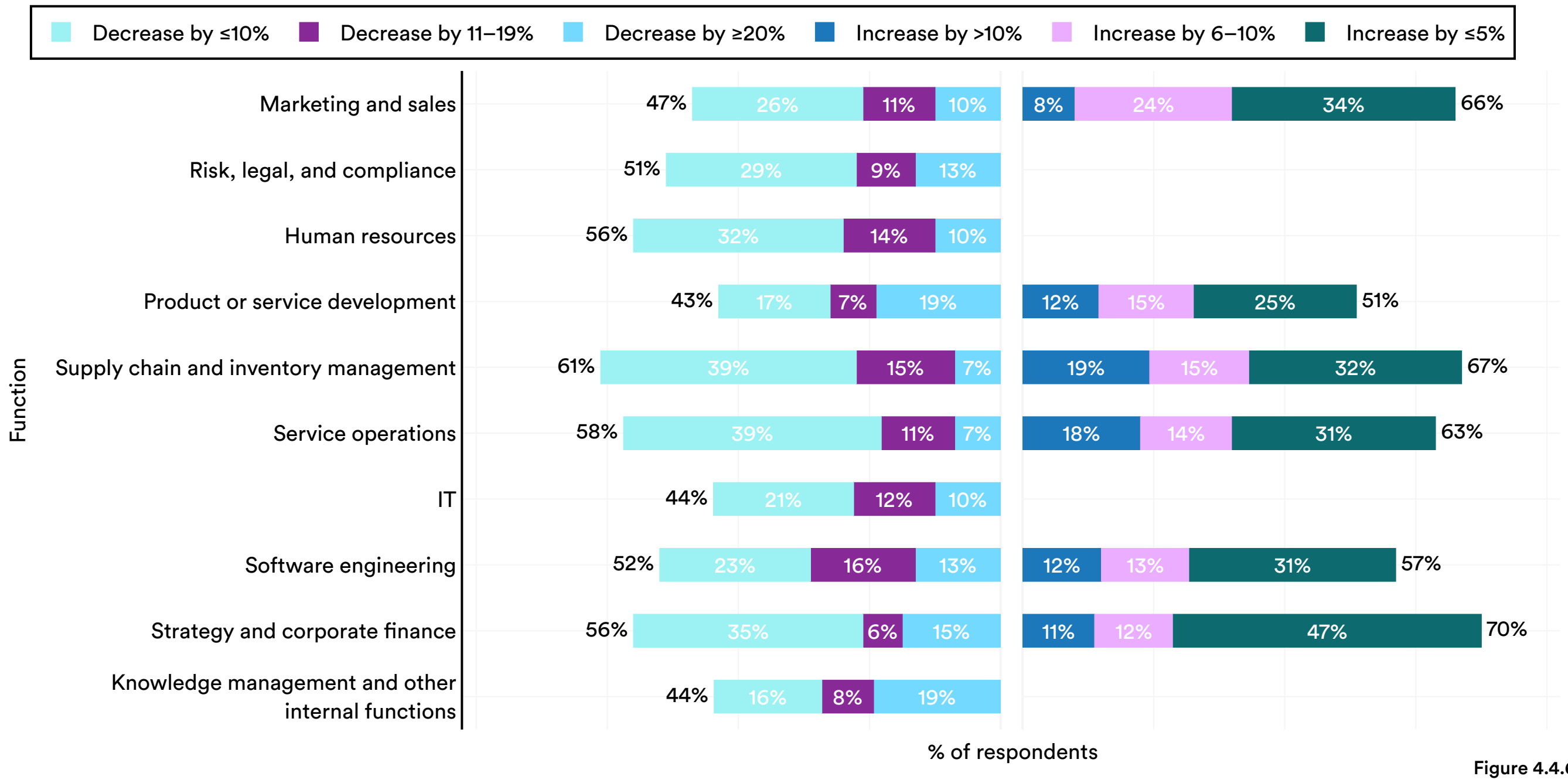

Figure 4.4.6

Figure 4.4.7 depicts the variation in generative AI usage among businesses across different regions of the world. Across all regions, reported use of generative AI in at least one business function reached 71% in 2024, more than doubling from 33% in 2023. This amount is just 7 percentage points lower than the percentage who reported using any form of AI (78%), which is shown in Figure 4.4.1. The use gap between AI overall and generative AI has contracted sharply from 22 percentage points in 2023 to 7 percentage points in 2024, signaling an accelerated usage of generative AI capabilities. North America (74%), Europe (73%), and Greater China (73%) lead in organizations' use of generative AI.

**Generative AI use by organizations in the world, 2023 vs. 2024**

Source: McKinsey & Company Survey, 2024 | Chart: 2025 AI Index report

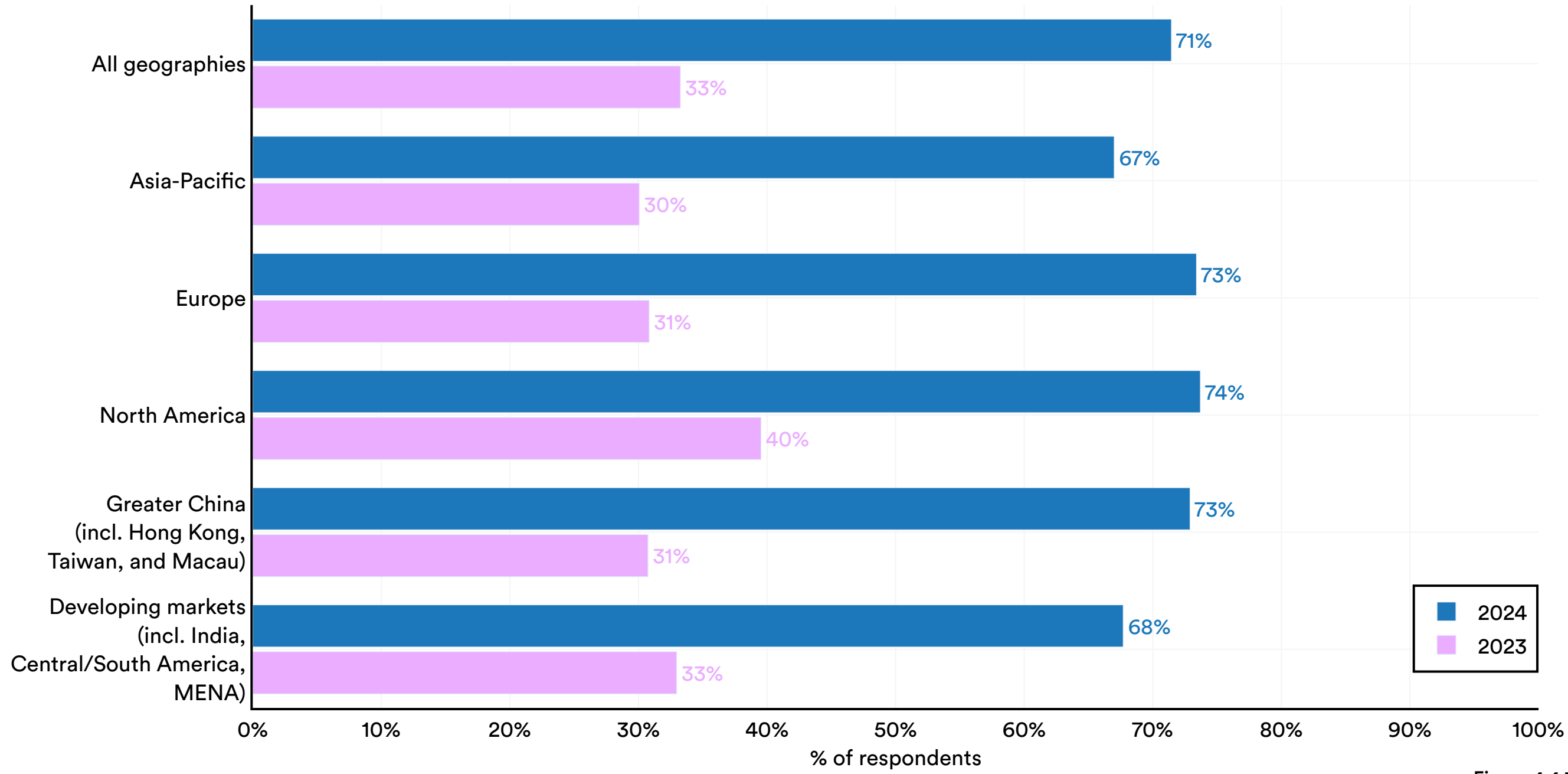

Figure 4.4.7[9]

9 This figure highlights AI use in at least one business function.

# AI's Labor Impact

Over the last six years, the growing integration of AI into the economy has sparked intense interest in its productivity potential. While early adoption showed promise, quantifying AI's impact remained challenging until 2023, when the first wave of rigorous studies emerged. In 2024, a substantial body of empirical research established clear patterns of AI's workplace effects across multiple domains and contexts. This section analyzes productivity impact data from five major academic studies, which together represent the first large-scale empirical investigation of AI's workplace effects. The research, encompassing over 200,000 professionals across multiple industries and contexts, reveals consistent productivity gains ranging from 10% to 45%, with particularly strong effects in technical, customer support, and creative tasks. These studies employed diverse methodologies, including natural experiments, randomized controlled trials, and large-scale surveys, to measure AI's impact across different organizational contexts.

## Productivity Trends

One of the most reputable studies on AI's impact on productivity, particularly generative AI, was published by Erik Brynjolfsson, Danielle Li, and Daniel Rock in April 2023.[10] Analyzing data from 5,179 customer support agents, the study examined the staggered introduction of a generative AI-powered conversational assistant. The researchers found that AI adoption increased the number of issues resolved per hour by 14.2% (Figure 4.4.8). Moreover, the study uncovered that productivity gains emerged quickly after AI was introduced, and AI-exposed workers maintained higher efficiency even during AI outages.

Other recently released research has confirmed the Brynjolfsson finding. A Microsoft workplace study established baseline productivity improvements in common workplace tasks, with document editing increasing by 10–13% and email processing time decreasing by 11%. Specialized roles showed higher gains. For example, security professionals achieved 23% faster completion times with 7% higher accuracy, and sales teams demonstrated 39% faster response times with 25% higher accuracy.

**Impact of AI on customer support agents**
Source: Brynjolfsson et al., 2023 | Chart: 2024 AI Index report

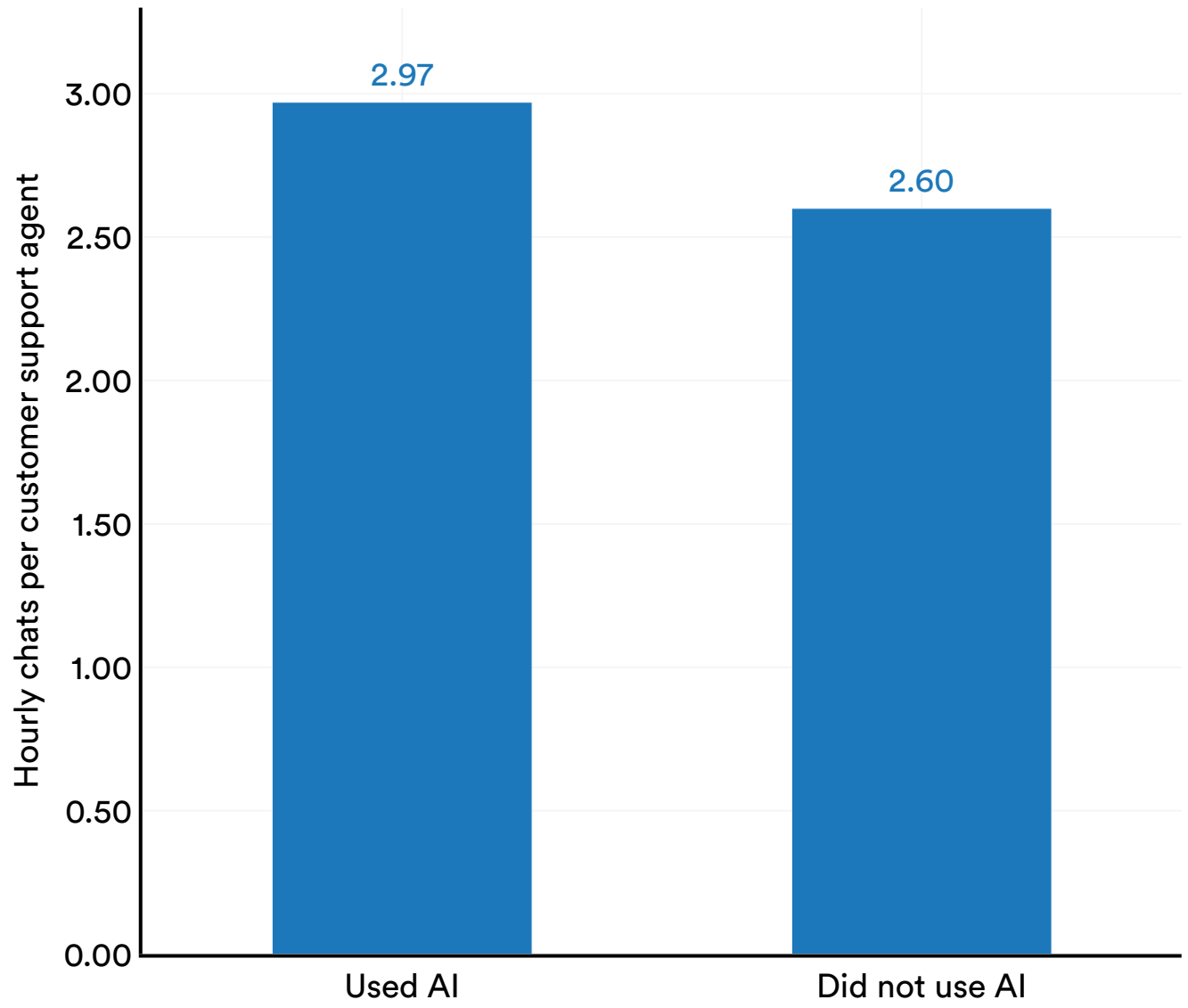

Figure 4.4.8

Figure 4.4.9 This figure has been removed following a post-publication correction (October 2025). It referenced a paper that has been retracted.

10 The paper was published as NBER working paper 31161 in 2023 and then in the "Quarterly Journal of Economics" in 2025.

In the software development domain, two major studies provided complementary evidence of AI's impact. A field experiment with 4,867 developers found that AI assistance increased task completion by 26.08% on average. This finding was reinforced by another natural experiment with 187,489 developers; it documented a 12.4% increase in core coding activities alongside a 24.9% decrease in time spent on project management tasks.

### Equalizing Effect

A consistent pattern across studies is AI's equalizing effect on workplace performance (Figure 4.4.10). In software development contexts, new research has found that junior developers experienced productivity increases of 21–40%, while senior developers saw more modest gains of 7–16%. This pattern was independently confirmed by other studies, which found coding productivity increases of 14–27% for low-ability workers compared to 5–10% for high-ability workers. Moreover, their analysis showed AI increased exploration of new technologies by 21.8% and generated an average potential salary increase of $1,683 per developer annually, suggesting AI tools are not just boosting productivity but actively enabling skill development. This research supports earlier 2023 and 2024 studies showing that AI-driven productivity gains vary based on workers' initial skill levels.

**AI's productivity equalizing effects**

| Study | Task | Low-skill worker productivity gain | High-skill worker productivity gain |
|---|---|---|---|
| Brynjolfsson et al., 2023 | Customer support | 34% | Indistinguishable from zero |
| Dell'Acqua et al., 2023 | Consulting | 42.96% | 16.5% |
| Cui et al., 2024 | Software engineering | 21–40% | 7–16% |
| Hoffman et al., 2024 | Software engineering | 12–27% | 5–10% |

Figure 4.4.10

### Adoption and Integration

The research reveals that productivity gains are strongly correlated with comprehensive AI integration and systematic implementation. A survey conducted by Romanian researchers of 233 employees found that organizations with high AI integration showed a 72% probability of significant productivity improvements, compared to just 3.4% for those with minimal integration. Their analysis documented a clear spectrum of productivity improvements across the entire study sample, with 46.8% of respondents reporting gains of 0–20%, 26.2% seeing gains of 20–40%, and 18.4% achieving improvements of 40–60%. A smaller proportion saw even larger gains, with 7.7% reporting increases of 60–80% and 0.9% achieving improvements of 80–100% (Figure 4.4.11).

#### Distribution of productivity gains from AI use

Source: Necula et al., 2024 | Chart: 2025 AI Index report

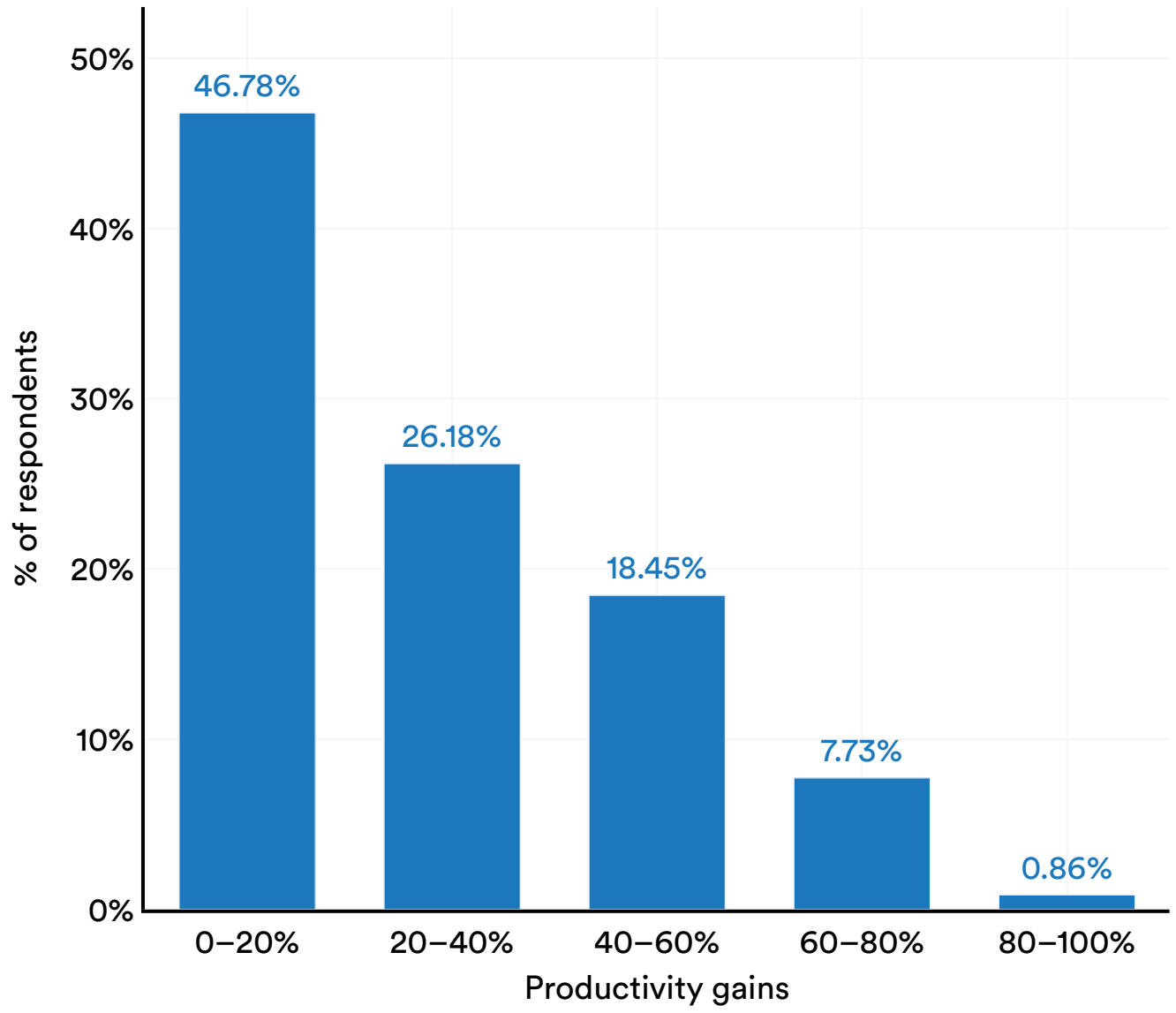

Figure 4.4.11

### Workforce Impact

The introduction of AI tools has led to significant shifts in both task allocation and team structures. The Microsoft workplace study found that AI automation enabled a 45% reduction in perceived mental demand (measured as 30/100 vs. 55/100 on their cognitive load scale), closed 84.6% of the accuracy gap for nonnative English speakers, and led to 49% more key information being included in professional reports. These improvements were particularly pronounced among "power users" (users who are intimately familiar with AI, as defined by using it at least several times a week) with 29% of AI users in this category saving more than 30 minutes per day. Research from the Harvard Business School documented that AI adoption led to reduced collaborative overhead, with projects requiring 79.3% fewer collaborators (team members) on average.

These changes are reshaping professional roles in fundamental ways. Debates about AI, like those surrounding past technological advancements, often center on automation versus augmentation—whether AI will replace jobs or enhance human work. While concrete data on AI-driven workforce changes remains limited, research is shedding light on how people perceive its impact on employment.

The Romanian survey data suggests varied expectations for AI's impact on workforce size, with 43% of organizations anticipating decreases, 30% expecting little change, 15% projecting increases, and 12% remaining uncertain about long-term implications. A McKinsey survey of executives found that 31% expect AI to reduce workforce size, while only 19% foresee an increase (Figure 4.4.12). In spite of claims about the increase in productivity of software engineers due to generative AI, the survey shows that their number is expected to increase, consistent with the Jevons Paradox. Notably, the share predicting workforce reductions has declined from last year, suggesting business leaders are becoming less convinced that AI will shrink organizational workforces (Figure 4.4.13).

**Expectations about the impact of generative AI on organizations' workforces in the next 3 years, 2024**
Source: McKinsey & Company Survey, 2024 | Chart: 2025 AI Index report

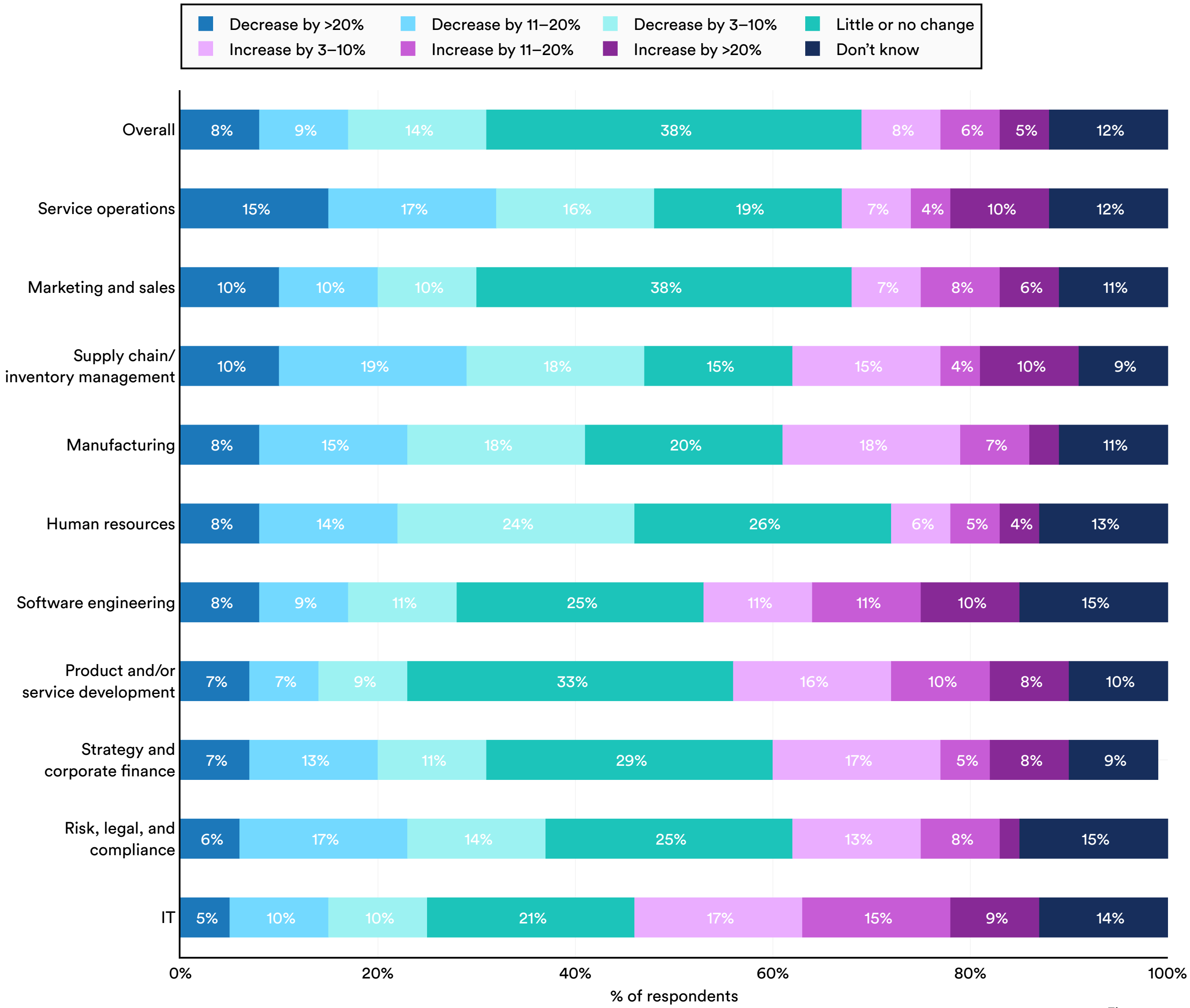

Figure 4.4.12

**Expectations about the impact of AI on organizations' workforces in the next 3 years, 2023 vs. 2024**
Source: McKinsey & Company Survey, 2023–24 | Chart: 2025 AI Index report

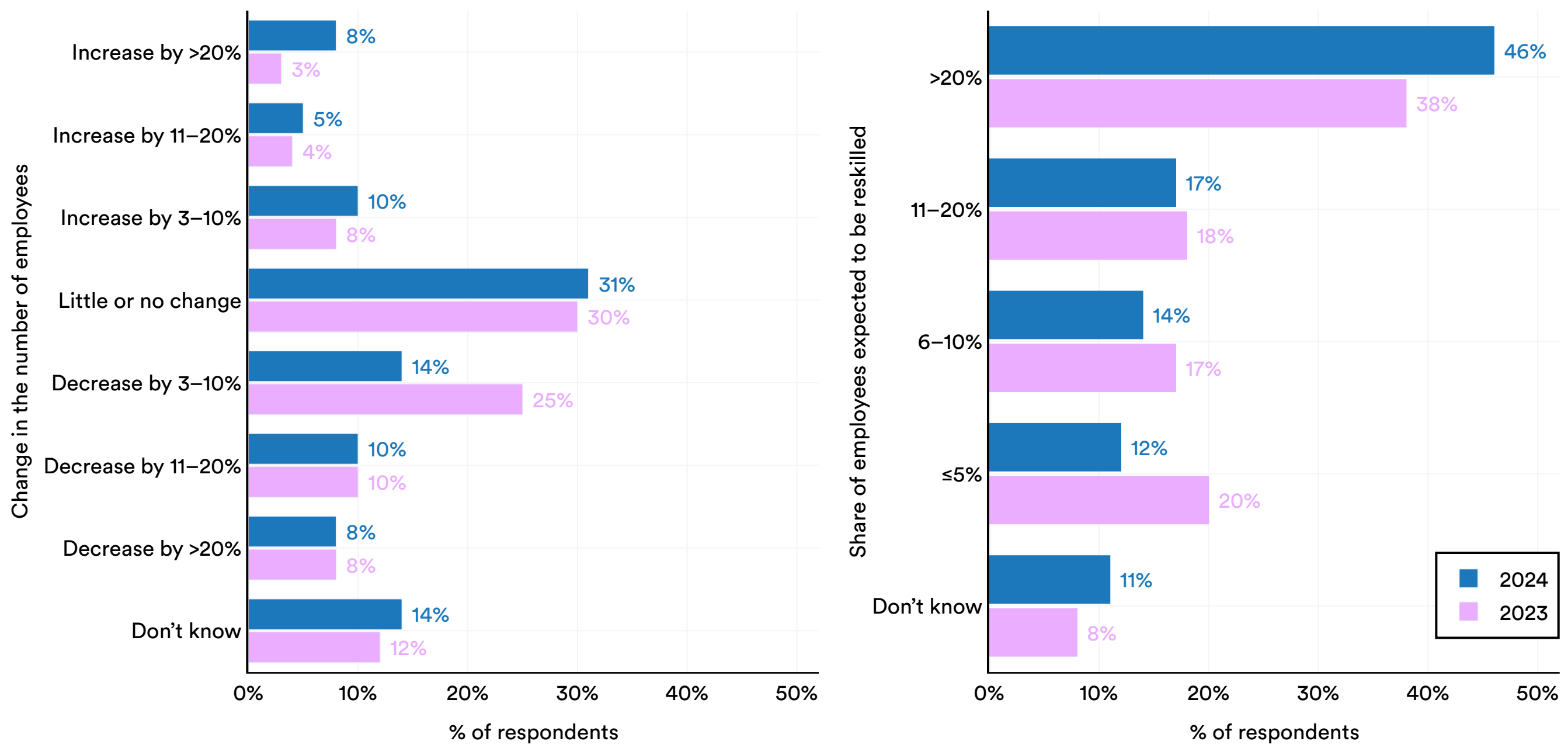

Figure 4.4.13

The deployment of robots equipped with AI-based software technologies offers a window into the real-world application of AI-ready infrastructure. This section draws on data from the International Federation of Robotics (IFR), a nonprofit organization dedicated to advancing the robotics industry. Annually, the IFR publishes the World Robotics Reports, which track global robot installation trends.[11]

# 4.5 Robot Deployments

## Aggregate Trends

The following section includes data on the installation and operation of industrial robots, which are defined as an "automatically controlled, reprogrammable, multipurpose manipulator, programmable in three or more axes, which can be either fixed in place or mobile for use in industrial automation applications."

Figure 4.5.1 reports the total number of industrial robots installed worldwide by year. In 2023, industrial robot installations decreased slightly, with 541,000 units marking a 2.2% decrease from 2022. This reflects the first year-over-year decrease since 2019.

**Number of industrial robots installed in the world, 2012–23**

Source: International Federation of Robotics (IFR), 2024 | Chart: 2025 AI Index report

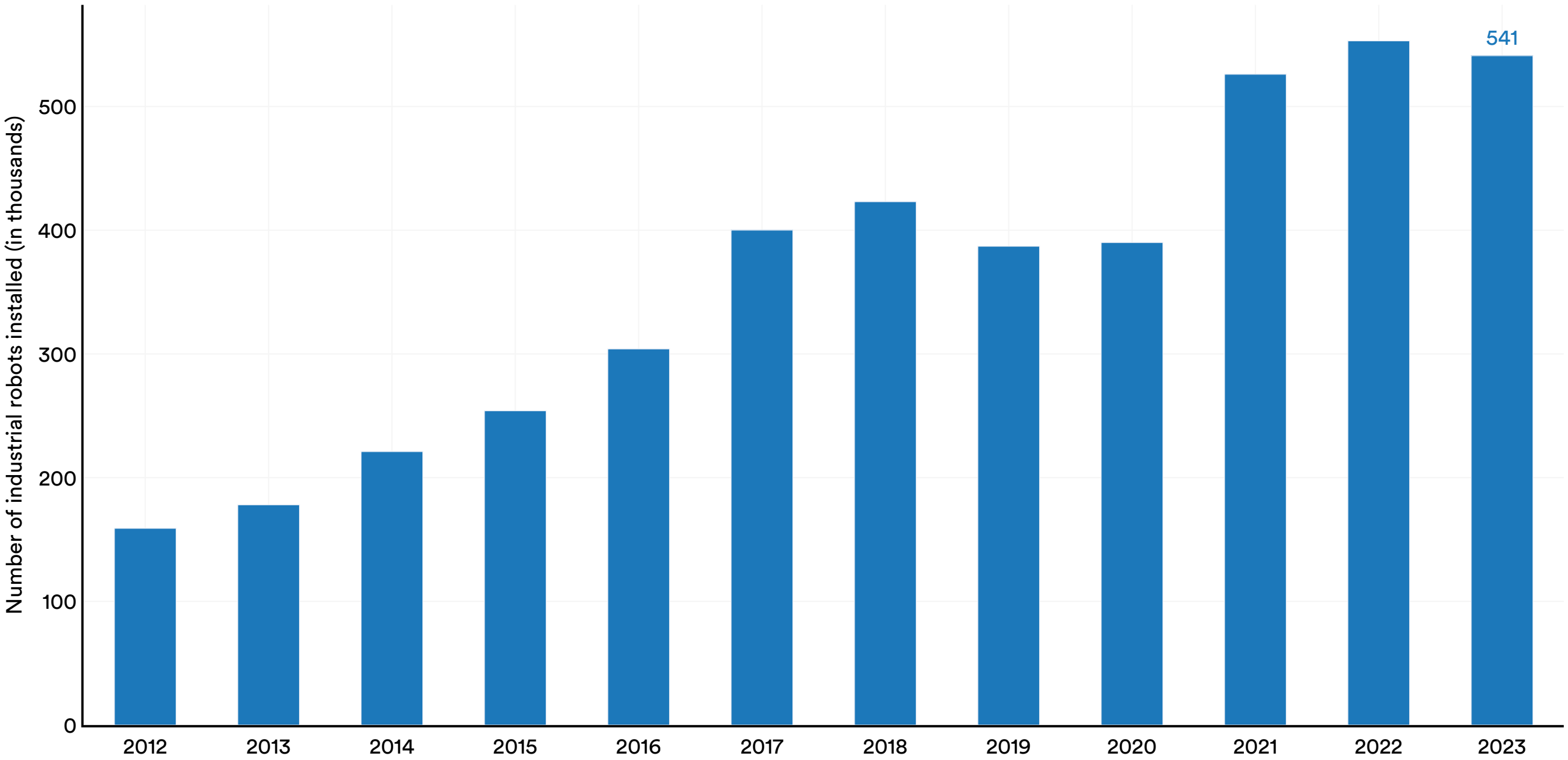

Figure 4.5.1

11 Due to the timing of the IFR report, the most recent data is from 2023. Every year, the IFR revisits data collected for previous years and will occasionally update the data if more accurate figures become available. Therefore, some of the data reported in this year's report might differ slightly from data reported in previous years.

The global operational stock of industrial robots reached 4,282,000 in 2023, up from 3,904,000 in 2022 (Figure 4.5.2). Since 2012, both the installation and utilization of industrial robots have steadily increased.

**Operational stock of industrial robots in the world, 2012–23**

Source: International Federation of Robotics (IFR), 2024 | Chart: 2025 AI Index report

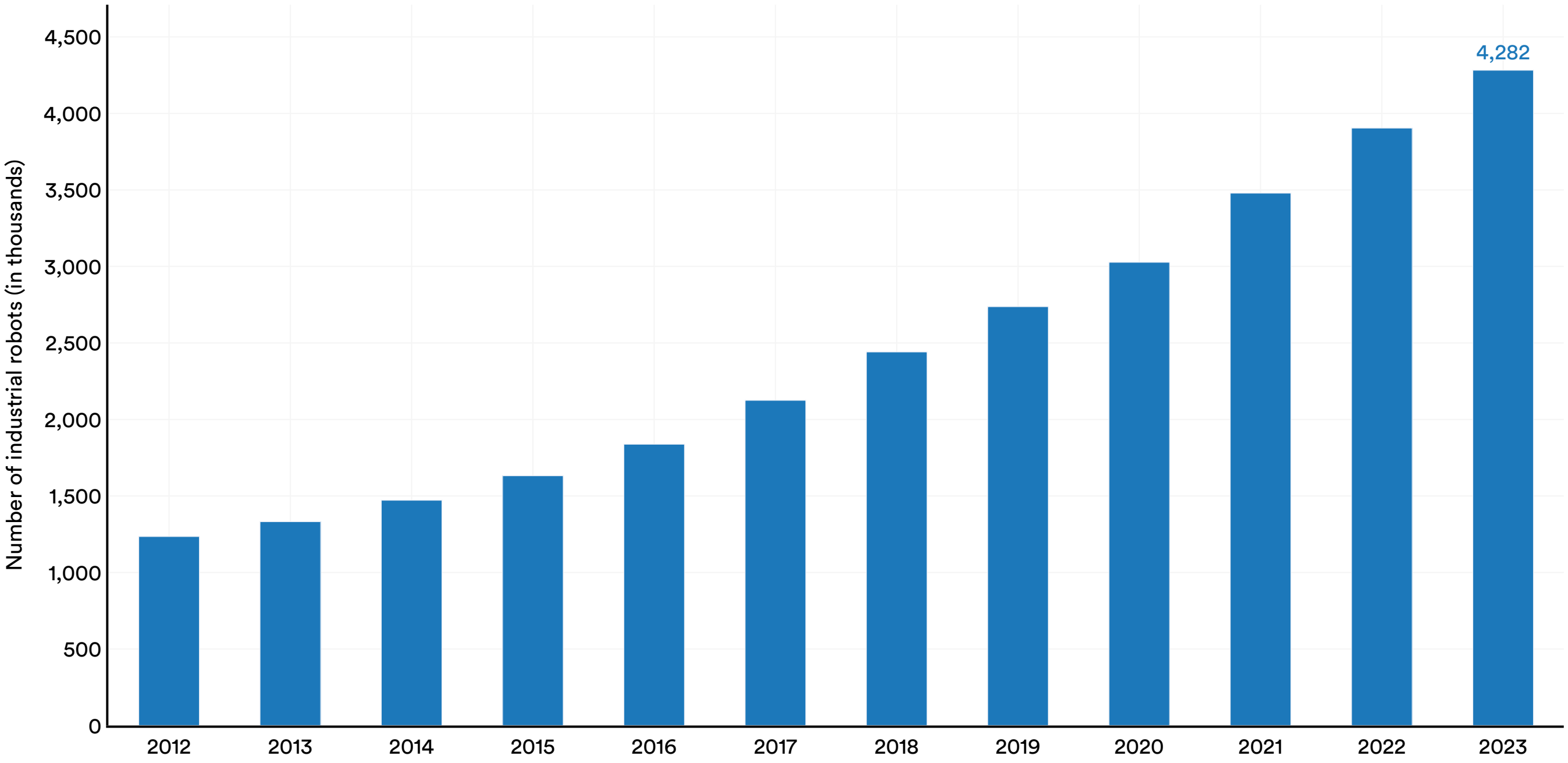

Figure 4.5.2

### Industrial Robots: Traditional vs. Collaborative Robots

There is a distinction between traditional robots, which operate in place of humans, and collaborative robots, designed to work alongside them.[12] The robotics community is increasingly enthusiastic about collaborative robots due to their safety, flexibility, scalability, and ability to learn iteratively.

Figure 4.5.3 reports the number of industrial robots installed in the world by type. In 2017, collaborative robots accounted for just 2.8% of all new industrial robot installations. By 2023, the number rose to 10.5%.

**Number of industrial robots installed in the world by type, 2017–23**

Source: International Federation of Robotics (IFR), 2024 | Chart: 2025 AI Index report

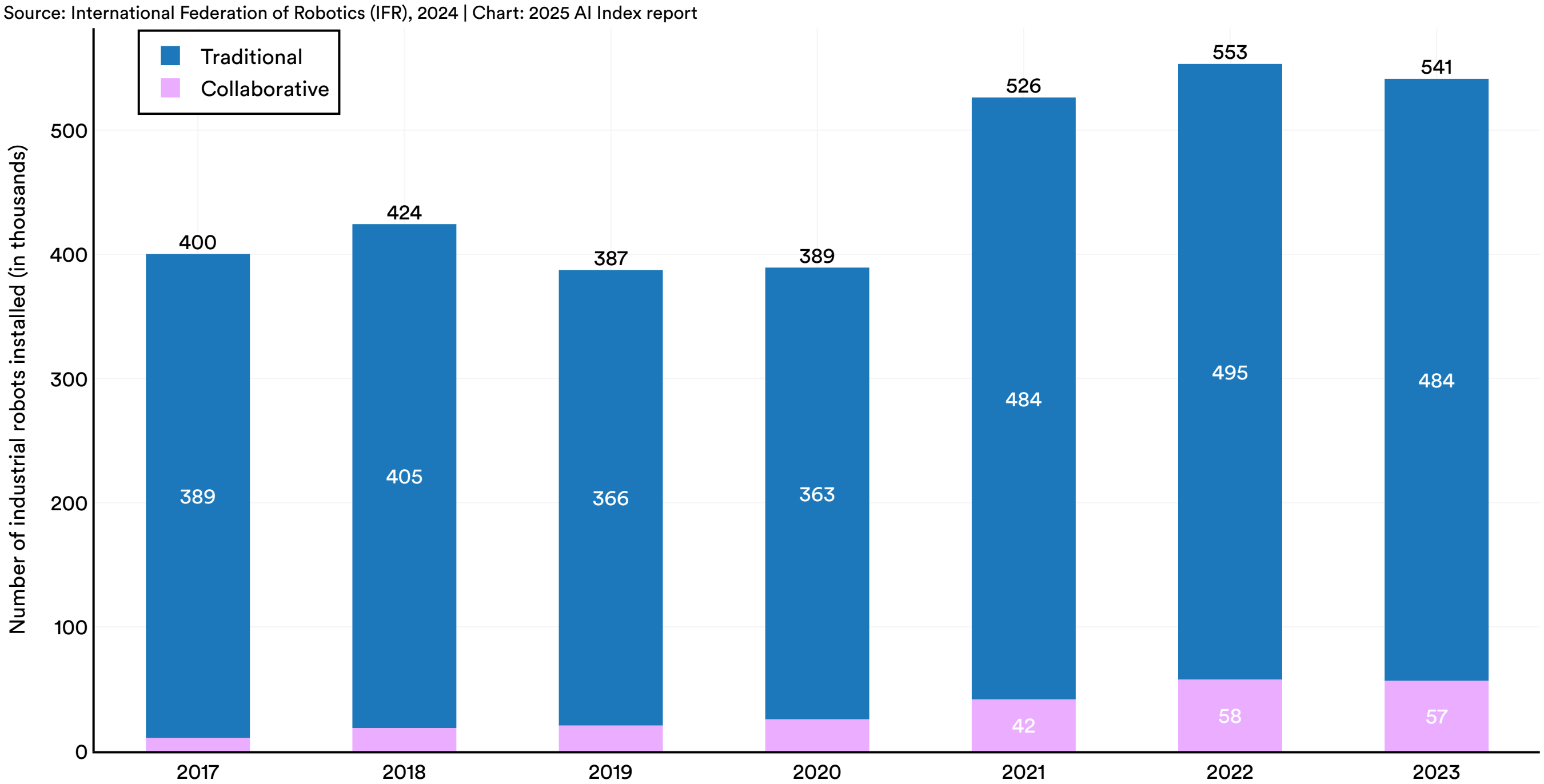

Figure 4.5.3

12 More detail on how the IFR defines collaborative robots can be found here.

### By Geographic Area

Country-level data on robot installations can suggest which nations prioritize the integration of robots into their economies. In 2023, China led the world with 276,300 industrial robot installations, six times more than Japan's 46,100 and 7.3 times more than the United States' 37,600 (Figure 4.5.4). South Korea and Germany followed with 31,400 and 28,400 installations, respectively.

**Number of industrial robots installed by geographic area, 2023**
Source: International Federation of Robotics (IFR), 2024 | Chart: 2025 AI Index report

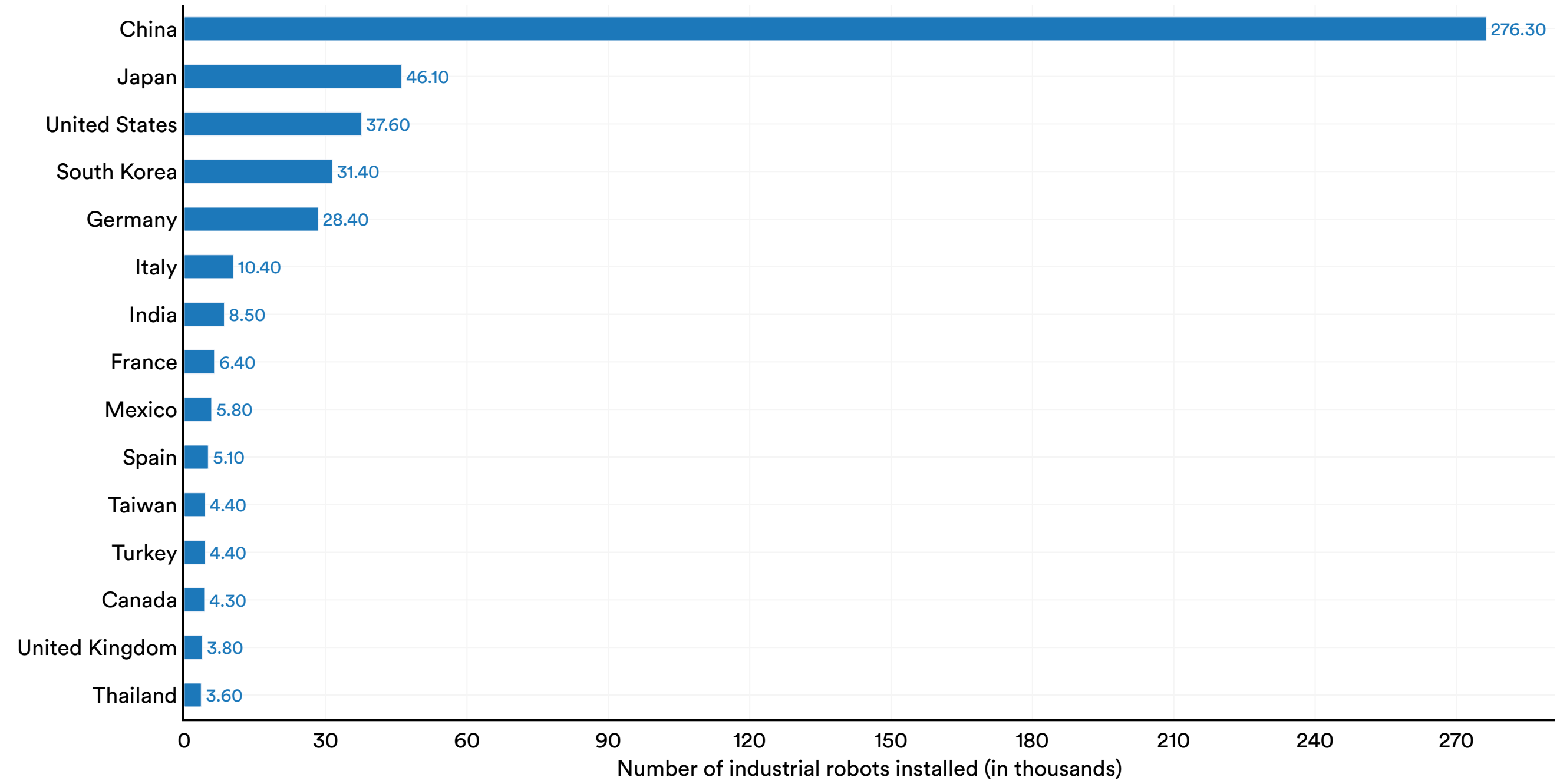

Figure 4.5.4

Since surpassing Japan in 2013 as the leading installer of industrial robots, China has significantly widened the gap with the nearest country. In 2013, China's installations accounted for 20.8% of the global total, reaching 51.1% by 2023 (Figure 4.5.5).

**Number of new industrial robots installed in top 5 countries, 2011–23**

Source: International Federation of Robotics (IFR), 2024 | Chart: 2025 AI Index report

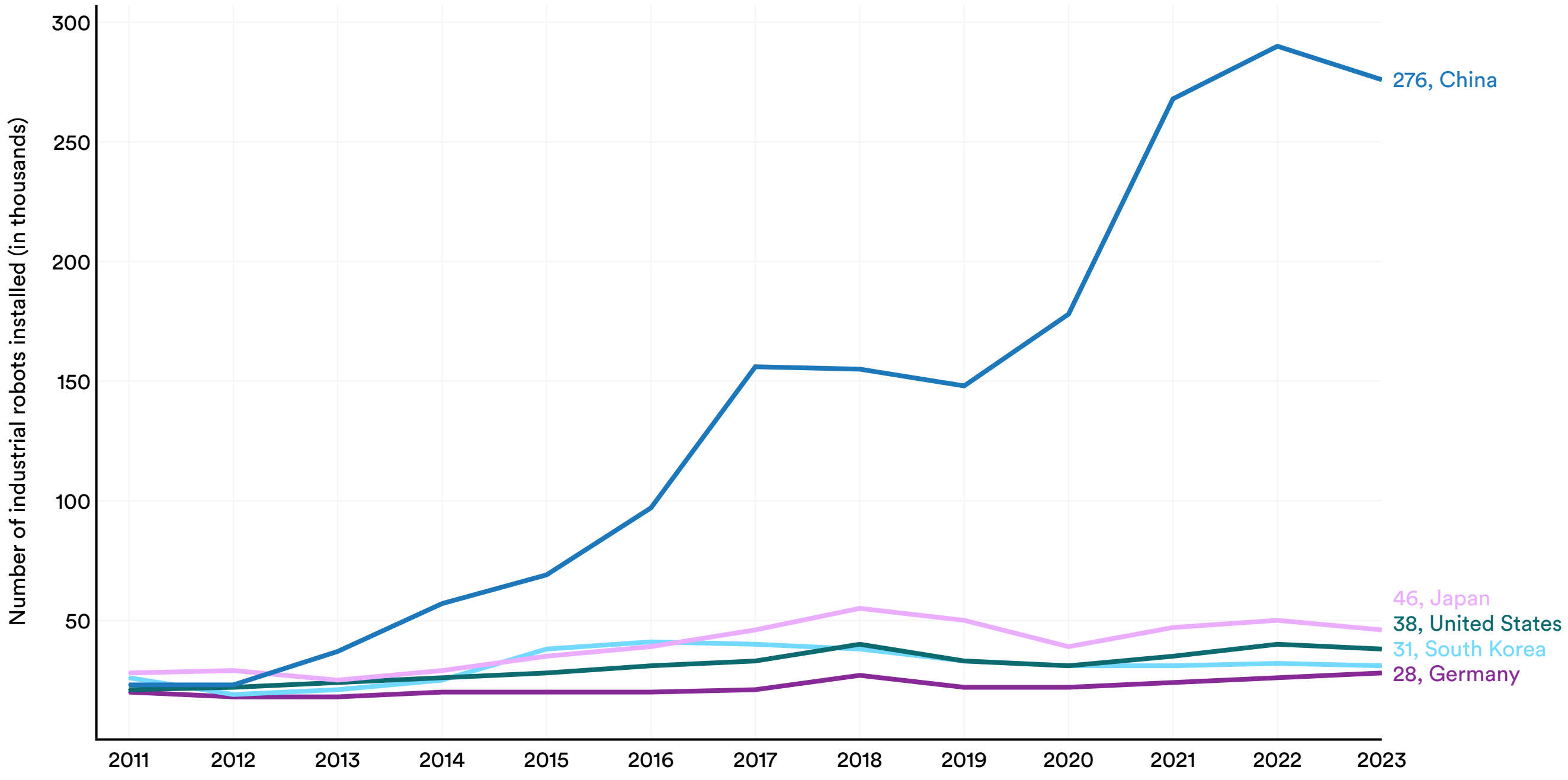

Figure 4.5.5

Since 2021, China has installed more industrial robots than the rest of the world combined, but the margin decreased in 2023 compared to 2022 (Figure 4.5.6). Despite this year-over-year decline, the sustained trend underscores China's dominance in industrial robot installations.

**Number of industrial robots installed (China vs. rest of the world), 2016–23**

Source: International Federation of Robotics (IFR), 2024 | Chart: 2025 AI Index report

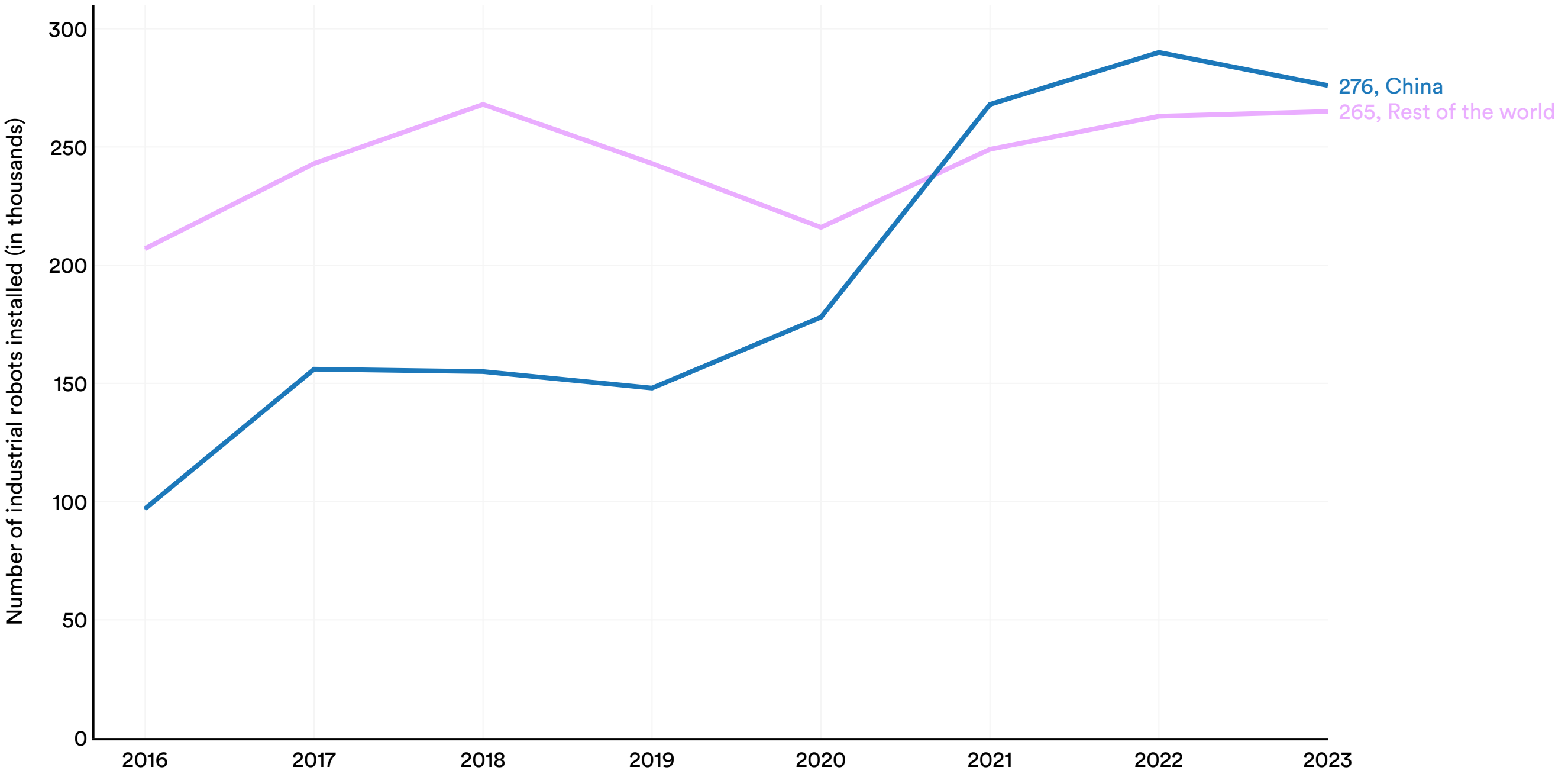

Figure 4.5.6

According to the IFR report, seven countries reported an annual increase in industrial robot installations from 2022 to 2023 (Figure 4.5.7). The countries with the highest growth rates include India (59%), the United Kingdom (51%), and Canada (37%). The geographic areas with the steepest declines include Taiwan (-43%), France (-13%), and Japan and Italy (both -9%).

**Annual growth rate of industrial robots installed by geographic area, 2022 vs. 2023**
Source: International Federation of Robotics (IFR), 2024 | Chart: 2025 AI Index report

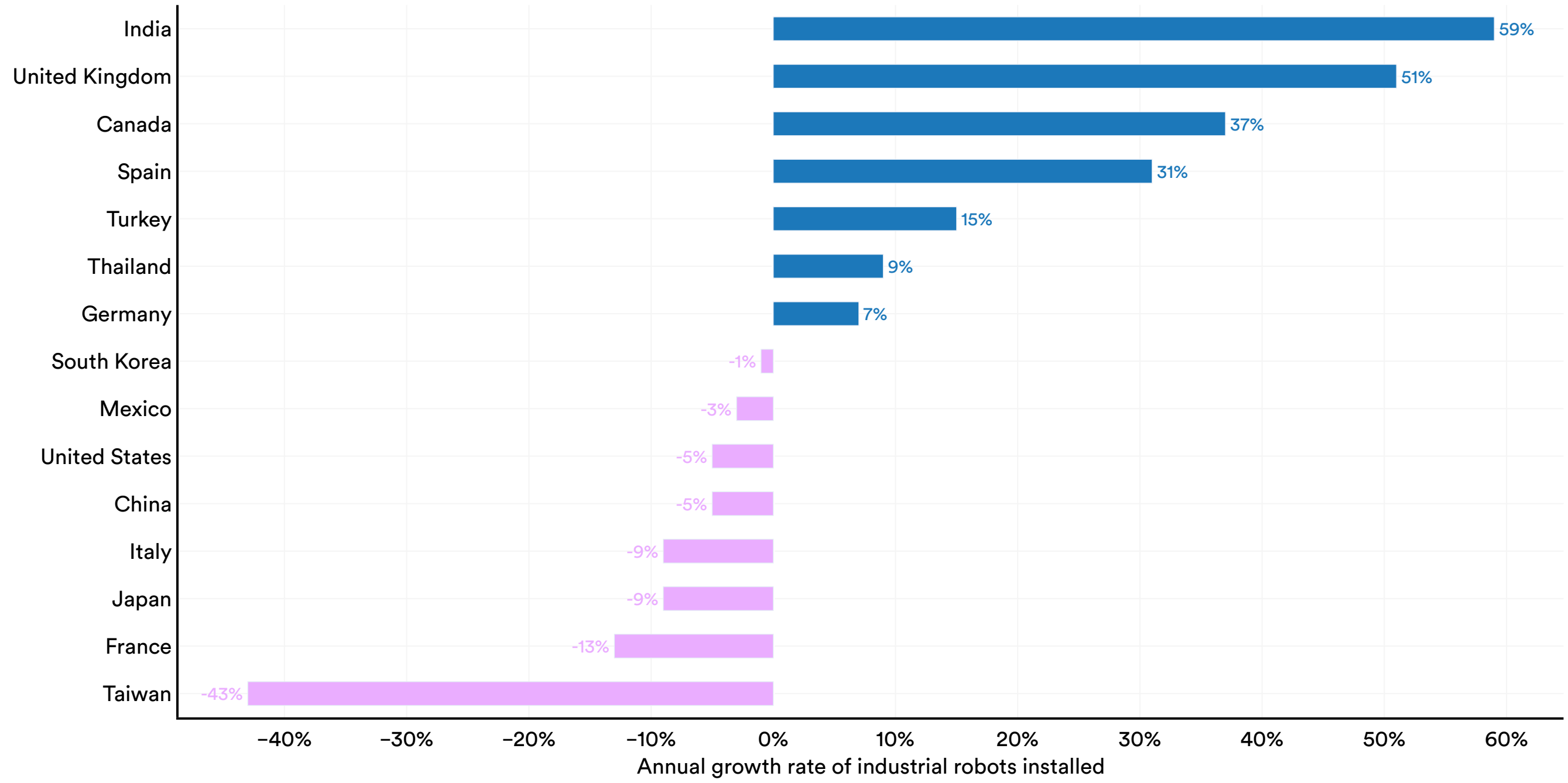

Figure 4.5.7

## Country-Level Data on Service Robotics

Another important class of robots is service robots, which the International Organization for Standardization defines as a robot "that performs useful tasks for humans or equipment excluding industrial automation applications."[13] Such robots can, for example, be used in medical settings and for professional cleaning. In 2023, more service robots were installed for every application category than in 2022, with the exception of medical robots (Figure 4.5.8). More specifically, the number of service robots installed in agricultural and hospitality settings increased 2.5 and 2.2 times, respectively.

### Number of service robots installed in the world by application area, 2022 vs. 2023

Source: International Federation of Robotics (IFR), 2024 | Chart: 2025 AI Index report

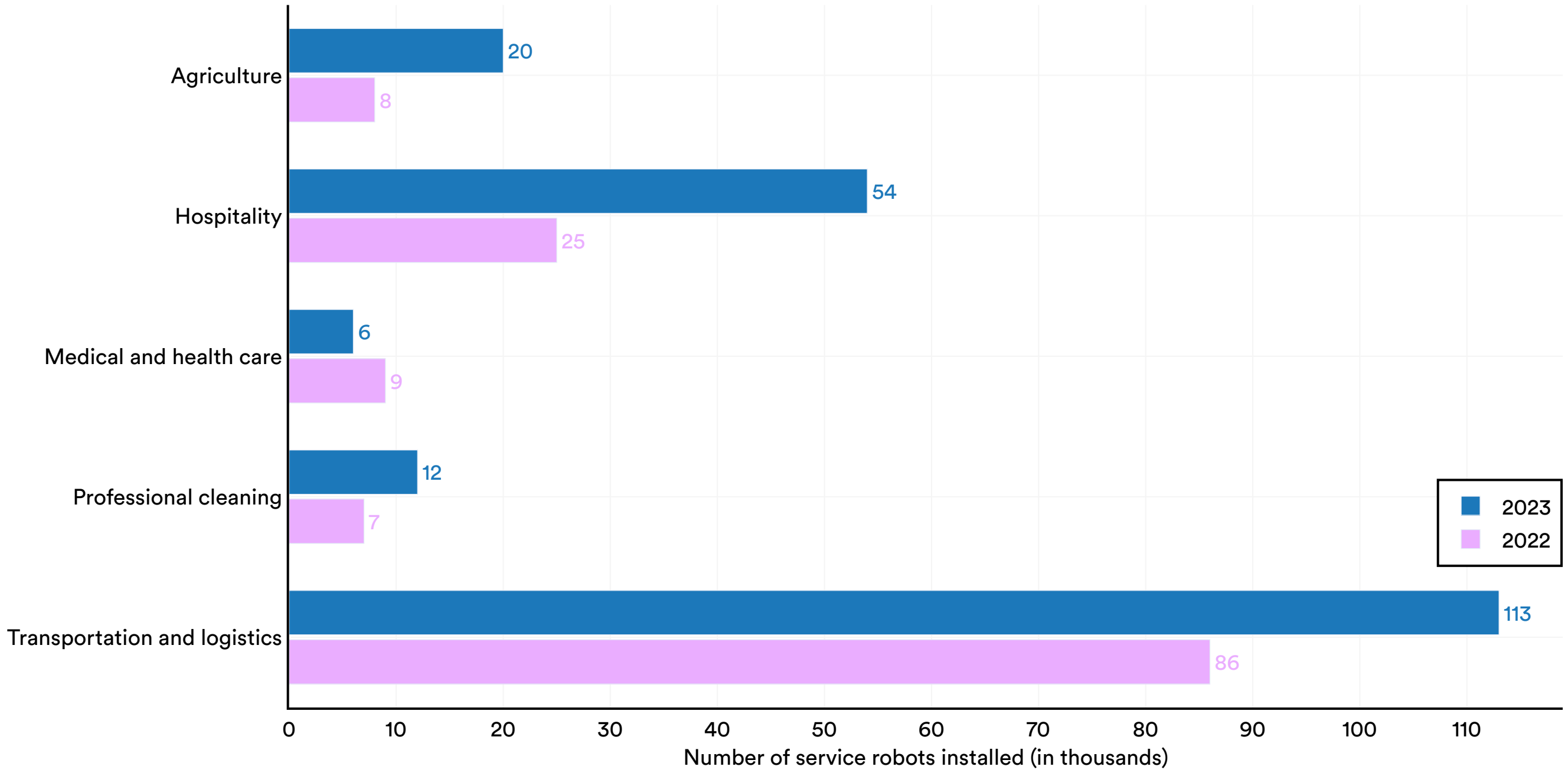

Figure 4.5.8

13 A more detailed definition can be accessed here.

# CHAPTER 5:
Science and Medicine

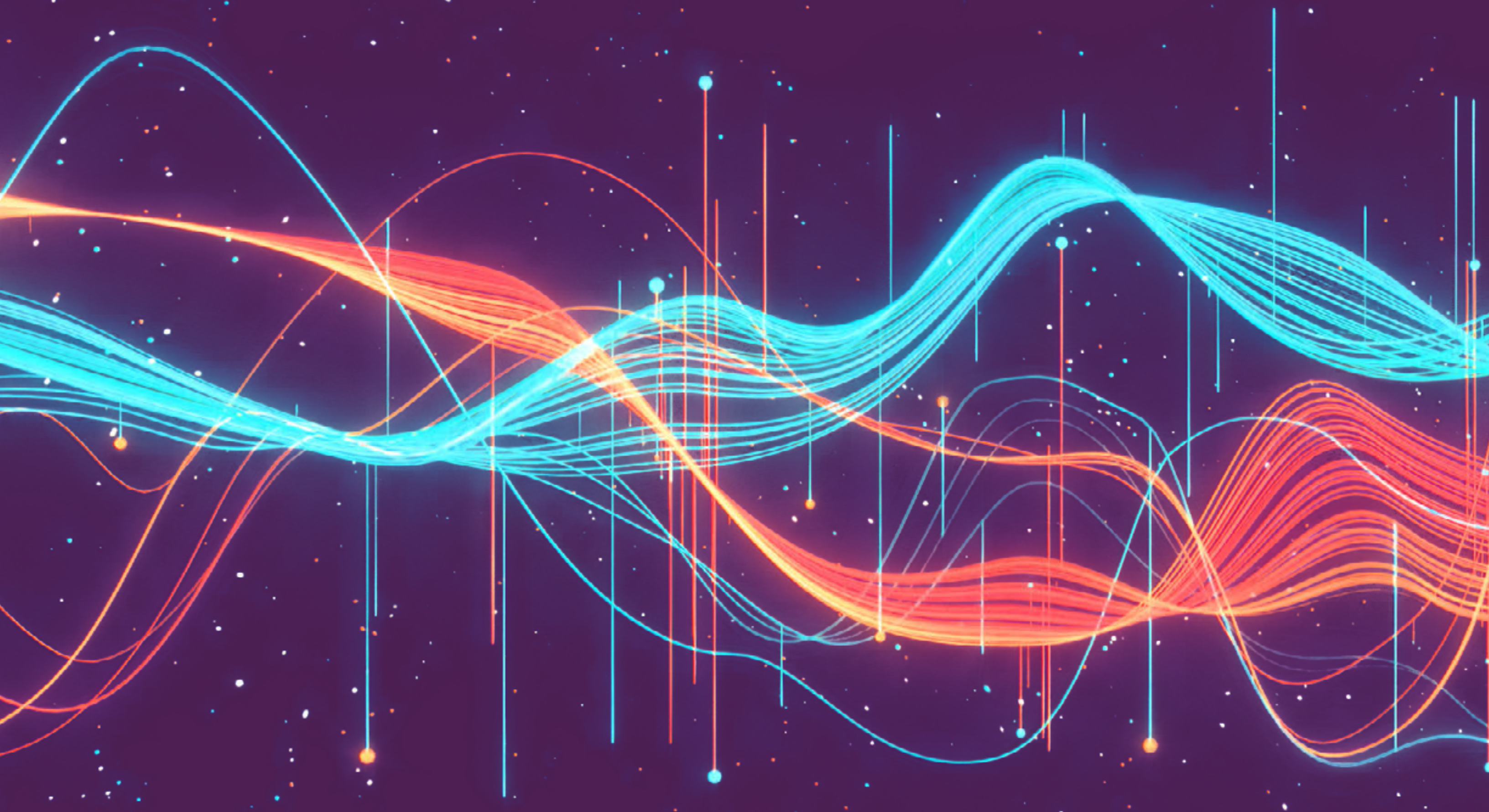

# Chapter 5: Science and Medicine

ACCESS THE PUBLIC DATA

**CHAPTER 5:**
Science and Medicine

# Overview

This chapter explores key trends in AI-driven science and medicine, reflecting the technology's growing impact in these fields. It begins with notable AI milestones from 2024, followed by an analysis of AI in protein folding, an important area of scientific advancement. The chapter then examines AI's role in clinical care, spanning both imaging and non-imaging applications. This includes a review of clinical knowledge capabilities in new language models, diagnostic and clinical management capabilities of AI systems, real-world AI deployments in medicine, synthetic data applications, and social determinants of health. Finally, the chapter concludes with an exploration of ethical trends in AI medical research.

This chapter was prepared by RAISE Health (Responsible AI for Safe and Equitable Health), a collaboration between Stanford Medicine and the Stanford Institute for Human-Centered Artificial Intelligence (HAI). Since its launch in 2023, RAISE Health has worked to advance responsible AI innovation in biomedical research, education, and patient care, with a focus on ensuring that these technologies benefit everyone.

Fostering collaborative research and knowledge sharing are central to RAISE Health's mission. As part of that commitment, RAISE Health partnered with the AI Index Steering Committee to expand the group's focus to include key developments in science and medicine. In 2024, this collaboration produced the inaugural chapter on science and medicine, highlighting major AI advancements at Stanford and beyond. The 2025 chapter builds on that foundation with contributions from members of the RAISE Health faculty research council, Stanford School of Medicine faculty, postdoctoral fellows, and undergraduate students from the schools of Medicine and Engineering.

**CHAPTER 5:**
Science and Medicine

# Chapter Highlights

**1. Bigger and better protein sequencing models emerge.** In 2024, several large-scale, high-performance protein sequencing models, including ESM3 and AlphaFold 3, were launched. Over time, these models have grown significantly in size, leading to continuous improvements in protein prediction accuracy.

**2. AI continues to drive rapid advances in scientific discovery.** AI's role in scientific progress continues to expand. While 2022 and 2023 marked the early stages of AI-driven breakthroughs, 2024 brought even greater advancements, including Aviary, which trains LLM agents for biological tasks, and FireSat, which significantly enhances wildfire prediction.

**3. The clinical knowledge of leading LLMs continues to improve.** OpenAI's recently released o1 set a new state-of-the-art 96.0% on the MedQA benchmark—a 5.8 percentage point gain over the best score posted in 2023. Since late 2022, performance has improved 28.4 percentage points. MedQA, a key benchmark for assessing clinical knowledge, may be approaching saturation, signaling the need for more challenging evaluations.

**4. AI outperforms doctors on key clinical tasks.** A new study found that GPT-4 alone outperformed doctors—both with and without AI—in diagnosing complex clinical cases. Other recent studies show AI surpassing doctors in cancer detection and identifying high-mortality-risk patients. However, some early research suggests that AI-doctor collaboration yields the best results, making it a fruitful area of further research.

**5. The number of FDA-approved, AI-enabled medical devices skyrockets.** The FDA authorized its first AI-enabled medical device in 1995. By 2015, only six such devices had been approved, but the number spiked to 223 by 2023.

**6. Synthetic data shows significant promise in medicine.** Studies released in 2024 suggest that AI-generated synthetic data can help models better identify social determinants of health, enhance privacy-preserving clinical risk prediction, and facilitate the discovery of new drug compounds.

# Chapter Highlights (cont'd)

**7. Medical AI ethics publications are increasing year over year.** The number of publications on ethics in medical AI quadrupled from 2020 to 2024, rising from 288 in 2020 to 1,031 in 2024.

---

**8. Foundation models come to medicine.** In 2024, a wave of large-scale medical foundation models were released, ranging from general-purpose multimodal models like Med-Gemini to specialized models such as EchoCLIP for echocardiology and ChexAgent for radiology.

---

**9. Publicly available protein databases grow in size.** Since 2021, the number of entries in major public protein science databases has grown significantly, including UniProt (31%), PDB (23%), and AlphaFold (585%). This expansion has important implications for scientific discovery.

---

**10. AI research wins two Nobel Prizes.** In 2024, AI-driven research received top honors, with two Nobel Prizes awarded for AI-related breakthroughs. Google DeepMind's Demis Hassabis and John Jumper won the Nobel Prize in Chemistry for their pioneering work on protein folding with AlphaFold. Meanwhile, John Hopfield and Geoffrey Hinton received the Nobel Prize in Physics for their foundational contributions to neural networks.

This section highlights significant AI-related medical and biological breakthroughs in 2024 as chosen by the RAISE Health AI Index Workgroup and AI Index Steering Committee.

# 5.1 Notable Medical and Biological AI Milestones

## Protein Sequence Optimization

### LLMs optimize protein sequence optimization

LLMs have recently, albeit unintentionally, gained a new biological capability: optimizing protein sequences. Traditionally, protein engineering requires extensive lab studies to refine sequences for improved functionality. However, a recent study found that LLMs—without fine-tuning—are becoming remarkably effective at this task. In other words, this is a hidden strength of existing LLMs, exemplified in this case by an adapted version of Llama-3.1-8B-Instruct. Using a directed evolutionary approach, researchers demonstrated that LLMs can generate protein sequences that outperform conventional algorithms across both synthetic and experimental fitness landscapes.

Figure 5.1.1 illustrates the researchers' findings. The objective in this case is to maximize the fitness value, with higher scores indicating better performance. The researchers compared their proposed method's fitness score against that of the default evolutionary algorithm (EA) approach.[1] The study revealed that this optimization extends beyond single-objective tasks to include constrained, budget-limited, and multiobjective scenarios. This compelling finding highlights the emergent properties of state-of-the-art LLMs, suggesting that as these general-purpose models continue to improve, their impact on scientific fields will only grow.

**Single-objective optimization results for fitness optimization**
Source: Wang et al., 2024

| Dataset | Method | Population × iteration | Fitness score | | |
|---|---|---|---|---|---|
| | | | Top 1 | Top 10 | Top 50 |
| GB1 | EA | 32×4 | **5.38±1.77** | **3.81±1.10** | **2.31±0.71** |
| | | 48×4 | **4.88±0.33** | 3.72±0.38 | 2.17±0.27 |
| | | 96×4 | **5.72±0.56** | **4.32±0.53** | 2.84±0.60 |
| | Ours | 32×4 | 4.34±0.53 | 3.22±0.23 | 1.94±0.28 |
| | | 48×4 | 4.31±0.82 | **3.76±0.82** | **2.45±0.61** |
| | | 96×4 | 4.80±0.52 | 4.09±0.19 | **3.04±0.19** |
| TrpB | EA | 32×4 | 0.20±0.18 | 0.14±0.12 | 0.07±0.05 |
| | | 48×4 | 0.67±0.14 | 0.52±0.11 | 0.19±0.04 |
| | | 96×4 | 0.74±0.01 | 0.59±0.03 | 0.35±0.10 |
| | Ours | 32×4 | **0.60±0.10** | **0.50±0.07** | **0.35±0.07** |
| | | 48×4 | **0.68±0.04** | **0.58±0.01** | **0.36±0.01** |
| | | 96×4 | **0.78±0.20** | **0.60±0.16** | **0.39±0.16** |
| Syn-3bfo | EA | 32×8 | 0.57±0.21 | -0.44±0.11 | -1.35±0.17 |
| | | 48×8 | 1.29±0.36 | 0.42±0.24 | -0.63±0.07 |
| | | 96×8 | 1.85±0.47 | 1.10±0.28 | 0.07±0.28 |
| | Ours | 32×8 | **2.51±0.23** | **1.33±0.14** | **0.28±0.20** |
| | | 48×8 | **2.35±0.26** | **1.36±0.11** | **0.04±0.09** |
| | | 96×8 | **2.83±0.20** | **2.02±0.36** | **0.96±0.36** |
| AAV | EA | 32×8 | 0.42±0.03 | 0.36±0.01 | 0.32±0.00 |
| | | 48×8 | 0.44±0.00 | 0.38±0.01 | 0.33±0.00 |
| | | 96×8 | 0.44±0.00 | 0.40±0.01 | 0.36±0.00 |
| | Ours | 32×8 | **0.74±0.00** | **0.69±0.02** | **0.62±0.03** |
| | | 48×8 | **0.75±0.01** | **0.71±0.01** | **0.64±0.02** |
| | | 96×8 | **0.76±0.03** | **0.73±0.03** | **0.68±0.03** |
| GFP | EA | 32×8 | 0.43±0.13 | 0.21±0.02 | 0.12±0.01 |
| | | 48×8 | 0.43±0.14 | 0.26±0.05 | 0.12±0.01 |
| | | 96×8 | 0.50±0.11 | 0.34±0.05 | 0.18±0.01 |
| | Ours | 32×8 | **0.96±0.02** | **0.94±0.01** | **0.88±0.03** |
| | | 48×8 | **0.96±0.02** | **0.93±0.01** | **0.84±0.02** |
| | | 96×8 | **0.97±0.01** | **0.95±0.01** | **0.92±0.01** |

Figure 5.1.1

1 Evolutionary algorithms (EA) simulate key aspects of biological evolution within a computer program to tackle complex problems—especially those without precise or fully satisfactory solutions—by finding approximate answers.

## Aviary

**Training LLM agents for biological tasks**

As AI systems become increasingly useful, particularly for scientific use cases, one challenge has been designing language models that can interact with tools as they reason through complex tasks. Aviary introduces a structured framework for training language agents for three particularly challenging scientific tasks: DNA manipulation (for molecular cloning), answering research questions (through accessing scientific papers), and engineering protein stability. Figure 5.1.2 compares the performance of different models across various Aviary environments. It contrasts a baseline Claude 3.5 Sonnet model, which attempts tasks without environmental access, with models integrated into agent frameworks within the Aviary environment. Across nearly all tasks, the agentic models outperform the baseline. This research demonstrates that (1) although general-purpose LLMs perform well at many scientific tasks, fine-tuning models alongside domain experts often helps models yield superior results, and (2) AI-driven scientific research can be accelerated not only by model size but also through interaction with external tools, capabilities now commonly referred to as "agentic AI."

**Performance of LLMs and language agents to solve tasks using Aviary environments**

Source: Narayanan et al., 2024 | Chart: 2025 AI Index report

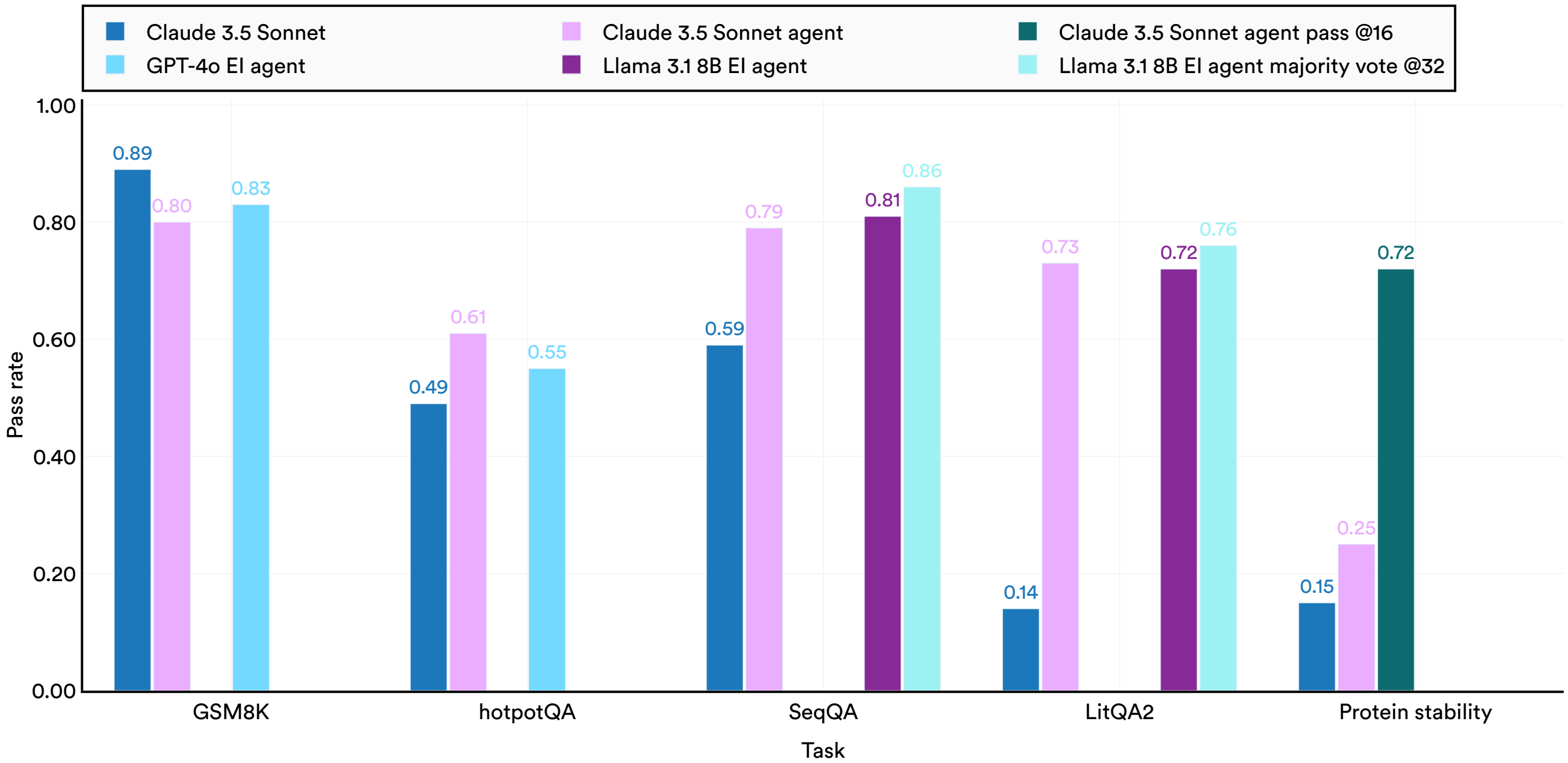

Figure 5.1.2

## AlphaProteo

**AI for novel, high-affinity protein binders**

AlphaProteo is Google DeepMind's model focused on creating novel, high-affinity protein binders that attach to specific target molecules. Figure 5.1.3 illustrates the predicted structures of seven target proteins for which AlphaProteo created successful binders. AlphaProteo has designed the first protein binders for many targets, including VEGF-A, a protein linked to cancer and diabetes. Many of the tool's binding strengths are significantly better than current state-of-the-art solutions; in fact, the team estimates that some of their binders are up to 300 times more effective than anything currently available on the seven target proteins they tested. For the viral protein BHRF1, 88% of their designed binders successfully bound when tested in DeepMind's wet lab. Based on the tested targets, AlphaProteo binders hold together roughly 10 times more strongly than those created using existing state-of-the-art design methods, making it a true bioengineering breakthrough. The model is being used for drug development, diagnostics, and biotech applications.

### AlphaProteo generating successful binders

Source: Google DeepMind, 2024

Figure 5.1.3

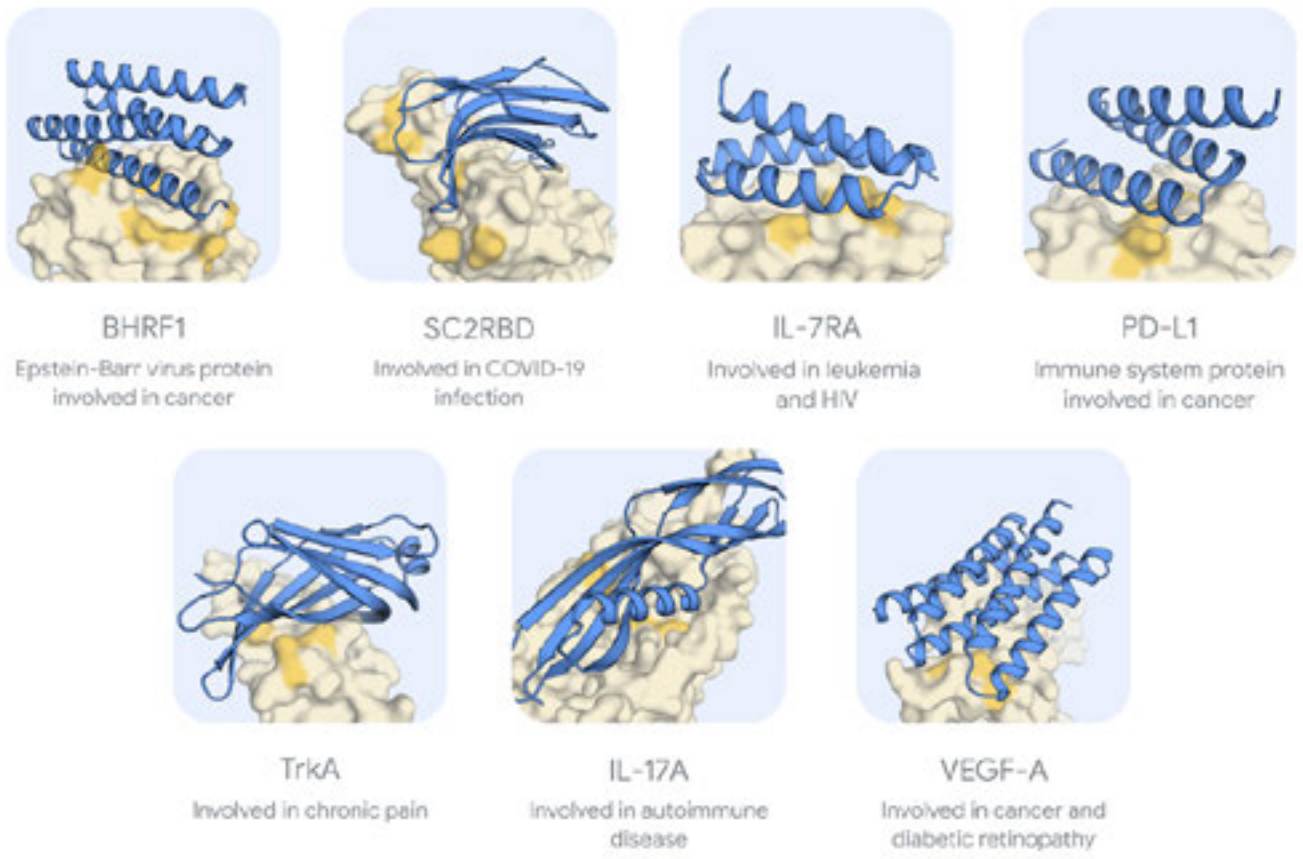

## Human Brain Mapping

**Synaptically reconstructing a small piece of the human brain**

A team at Google's Connectomics project has reconstructed a one-cubic-millimeter section of the human brain at the synaptic level—hailed by Wired as "the most detailed map of brain connections ever made." The sample, taken from an epileptic patient's left anterior temporal lobe during surgery, was imaged with a multibeam scanning electron microscope. Over 5,000 ultra-thin slices (30 nanometers each) captured around 57,000 cells—including neurons, glial cells, and blood vessels—along with 150 million synapses. Figure 5.1.4 visualizes the results: excitatory neurons on the left, inhibitory neurons on the right. To process this massive dataset, the team developed machine learning tools like flood-filling networks (for neuron reconstruction without manual tracing), SegCLR (for cell type identification), and TensorStore (for managing the multidimensional dataset). The dataset is publicly available via Neuroglancer, a web-based exploration tool; and CAVE, a Neuroglancer extension for annotation refinement. This project marks a major step in understanding neural circuitry and could inform future neurological treatments.

### 3D brain map images

Source: Google Research, 2024

Figure 5.1.4

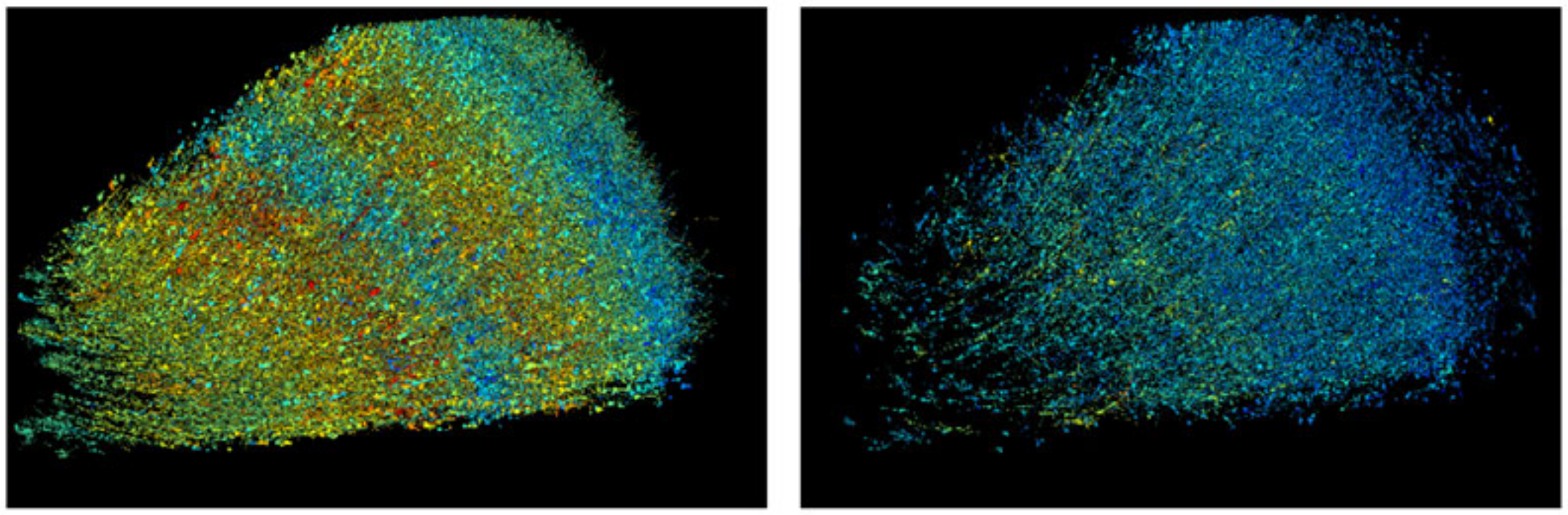

## Virtual AI Lab

**Virtual AI lab supercharges biomedical research**

AI's role in science is shifting from a passive tool to an active collaborator. A recent Stanford study introduced a virtual AI laboratory, where multiple AI-powered scientists (technically LLMs) specialize in different disciplines and autonomously collaborate as agents. In one experiment, human researchers tasked this AI-driven lab with designing nanobodies—antibody fragments—capable of binding to SARS-CoV-2, the virus that causes COVID-19. The lab generated 92 nanobodies, with over 90% successfully binding to the virus in validation studies. The virtual lab was structured similar to a computational biology lab, comprising a principal investigator (PI), a scientific critic AI, and three discipline-specific scientists specializing in immunology, computational biology, and machine learning (Figure 5.1.5). The PI model created these expert scientists and guided their research. Tools like AlphaFold and Rosetta were used for protein design, but the real significance of this study lies not in its specific findings, but in demonstrating that an entirely autonomous, LLM-powered lab can generate meaningful scientific discoveries.

**Workflow in AI-based lab**

Source: FreeThink, 2025

Figure 5.1.5

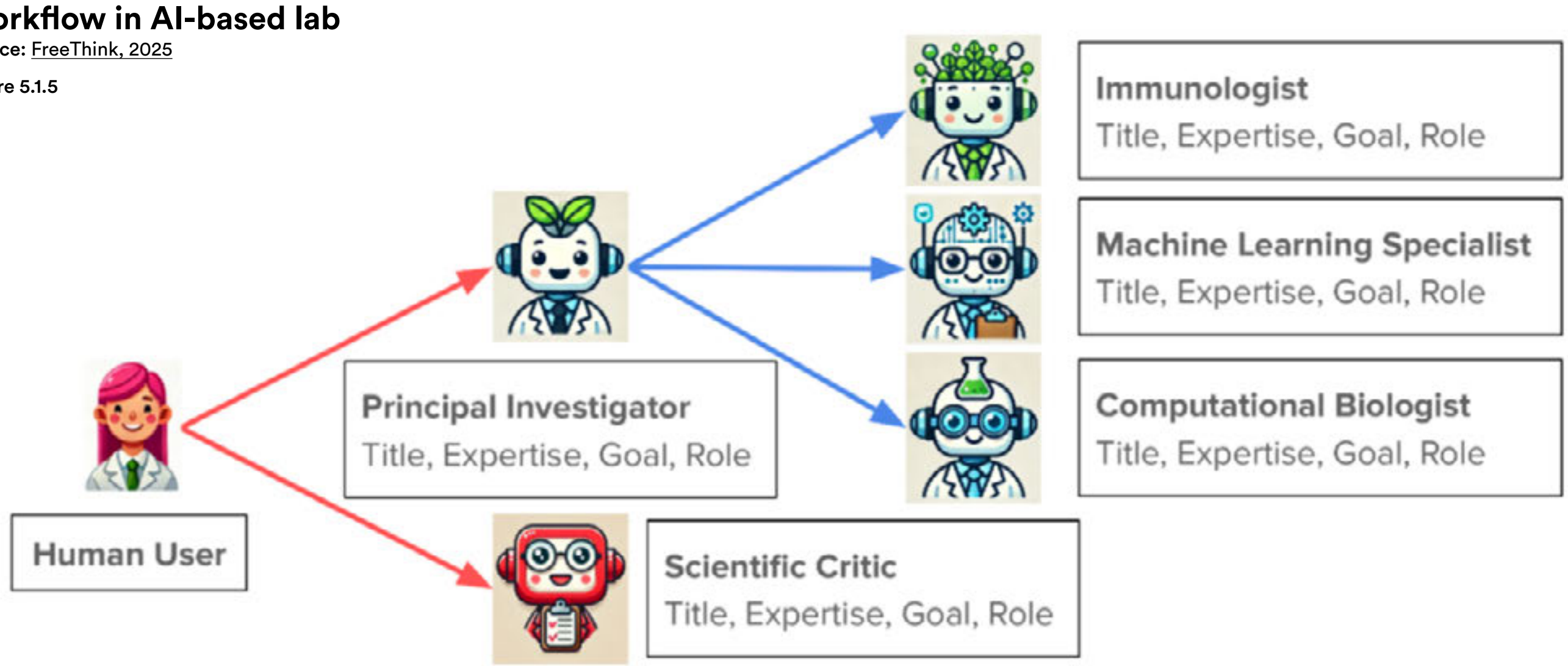

## GluFormer

**Continuous glucose monitoring with AI**

GluFormer, a foundation model developed by Nvidia Tel Aviv, the Weizmann Institute, and others, analyzes continuous glucose monitoring (CGM) data to predict long-term health outcomes. Trained on over 10 million glucose measurements from nearly 11,000 individuals—most without diabetes—it forecasts health trajectories up to four years in advance. For instance, GluFormer can identify individuals at risk of developing diabetes or worsening glycemic control long before symptoms appear. In a 12-year study of 580 adults, it accurately flagged 66% of new-onset diabetes cases and 69% of cardiovascular-related deaths within their respective top-risk quartiles. The model's results have also generalized across 19 external cohorts (n=6,044) in five countries and diverse health conditions. GluFormer often outperforms standard CGM-based metrics like the glucose management indicator (GMI) (Figure 5.1.6). In the near and long term, models like GluFormer will shift diabetes care from reactive treatment to proactive prevention, enabling earlier clinical intervention.

### GluFormer versus glucose management indicator

Source: Lutsker et al., 2024

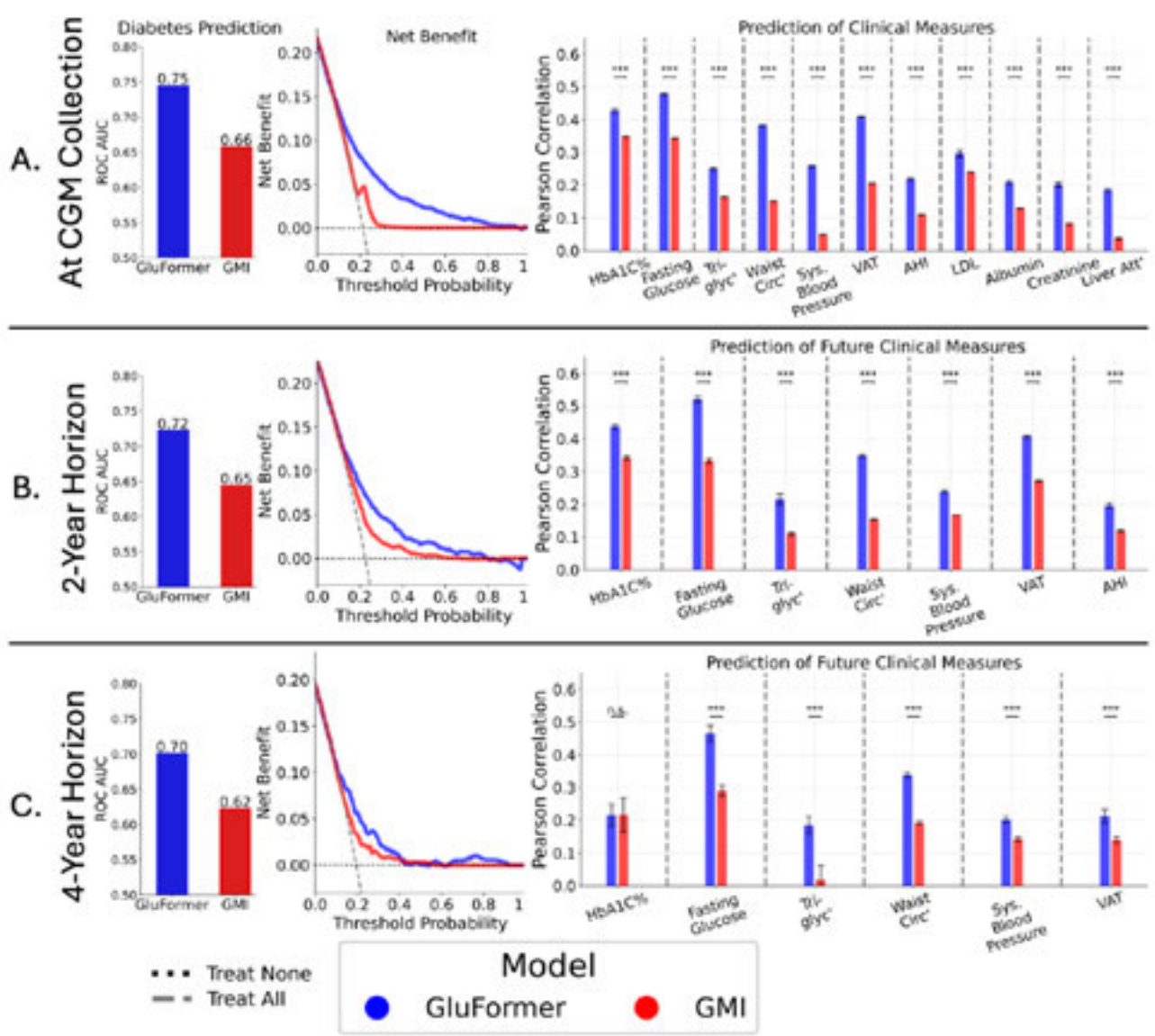

Figure 5.1.6

## Evolutionary Scale Modeling v3 (ESM3)

**Simulating evolutionary processes to generate novel proteins**

EvolutionaryScale's ESM3 is a groundbreaking model designed to generate novel proteins by simulating evolutionary processes. The model was trained on 2.78 billion protein sequences, and hosts 98 billion parameters. Like many other AI models, it is available in three sizes (small, medium, and large) and is available both via API and their partners' platforms. Perhaps ESM3's most notable achievement is designing esmGFP, a new artificial green fluorescent protein which the company estimates would take nature 500 million years to develop. This was done through human-led chain-of-thought prompting. Figure 5.1.7 illustrates the performance of various ESM3 models in generating proteins that satisfy atomic coordination prompts. The results show that larger ESM3 models solve twice as many tasks. ESM3 is also open-sourced, promoting collaboration in synthetic biology and protein engineering projects which hope to use code and data from the project—with applications in drug discovery, materials science, and environmental engineering.

### ESM3 models evaluated on protein generation from atomic coordination prompts

Source: ESM3, 2024 | Chart: 2025 AI Index report

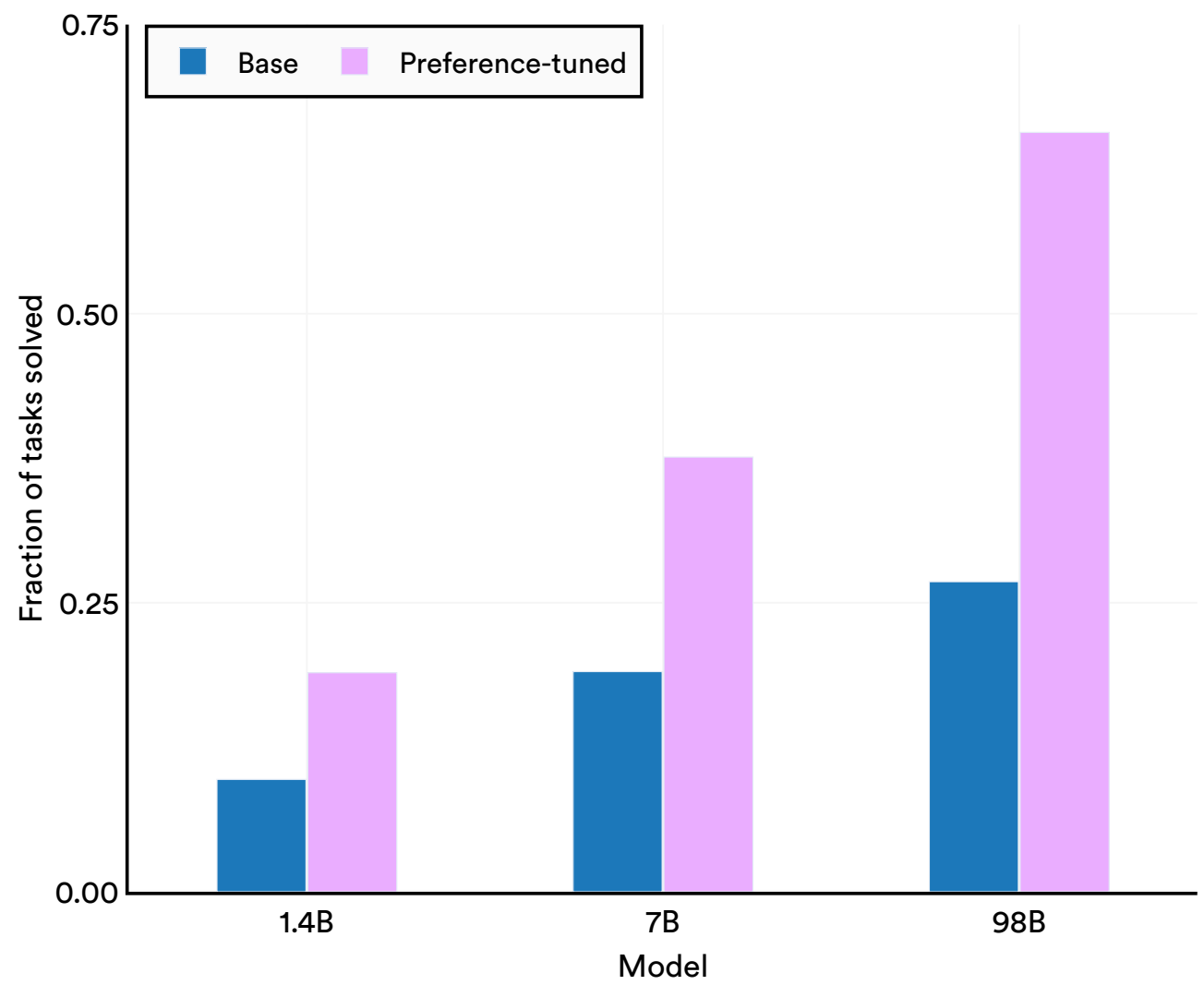

Figure 5.1.7

## AlphaFold 3

**Predicting the structure and interactions of all of life's molecules**

Google and Isomorphic Lab's latest in the AlphaFold series, AlphaFold 3, goes beyond predicting protein structures to more accurately modeling their interactions with key biomolecules (DNA, RNA, ligands, antibodies). Figure 5.1.8 compares AlphaFold 3's accuracy in predicting protein-ligand interactions against other top docking tools (e.g., Vina and Gnina) based on the percentage of predictions with a root mean square deviation (RMSD) below 2 Å, an important measure of docking accuracy.[2 3] AlphaFold 3 is competitive with previous state-of-the-art methods and particularly effective when the binding pocket is predefined, meaning that the docking algorithm is given prior knowledge about the specific region on the protein where the small molecule (ligand) is expected to bind. AlphaFold 3 can accelerate drug development by modeling small molecule-protein interactions, which is important for disease research. Moreover, AlphaFold 3's open-source access empowers scientists globally.

### AlphaFold 3 vs. baselines for protein-ligand docking

Source: ESM3, 2024 | Chart: 2025 AI Index report

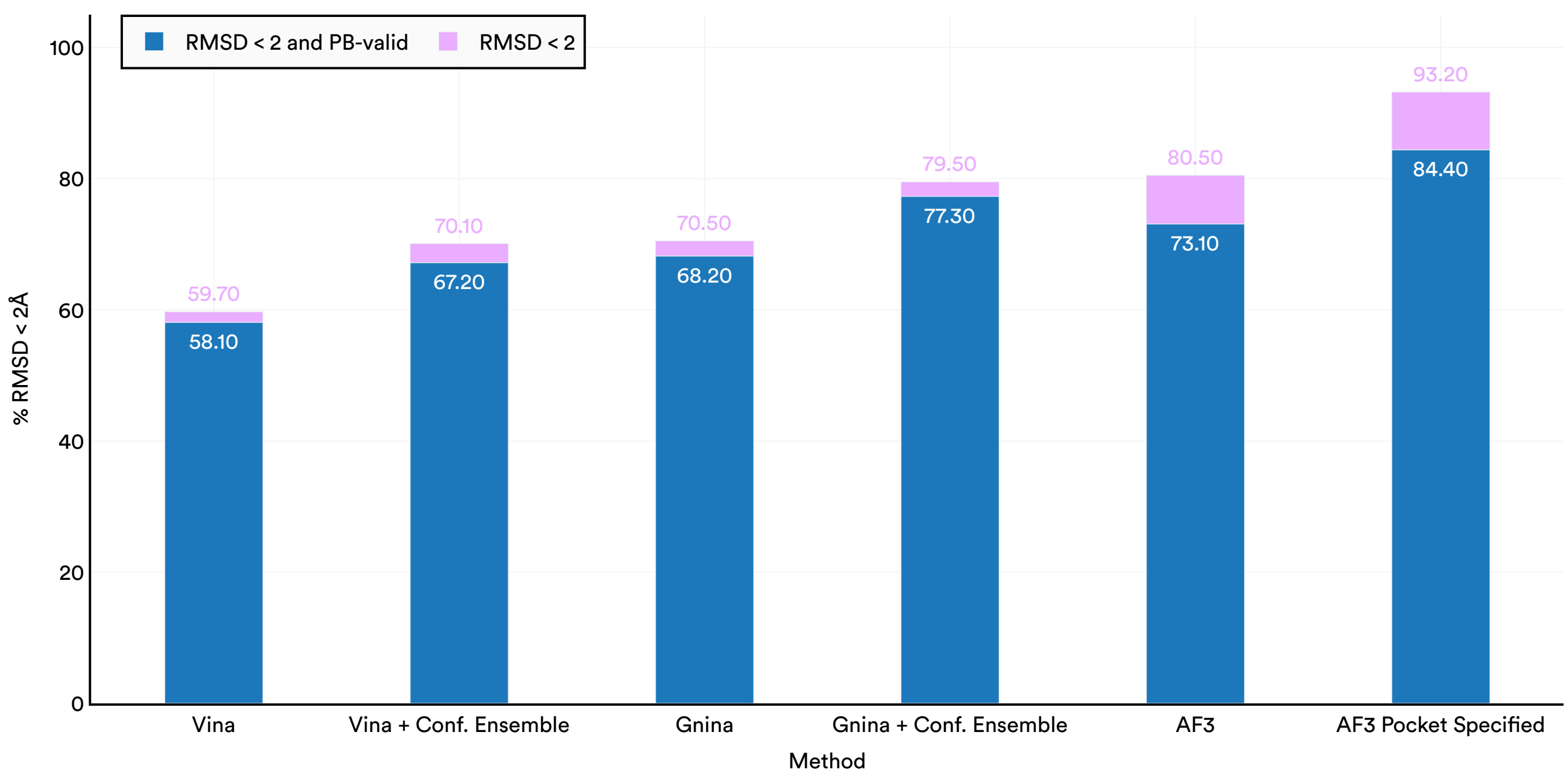

Figure 5.1.8

2 A docking tool, like Vina, is a computational program used in molecular docking—a process that predicts how small molecules (such as drugs) interact with target proteins. These tools help scientists model and visualize how a molecule might bind to a protein's active site, which is crucial in drug discovery.

3 The chart uses two shades of bars to represent different accuracy criteria in molecular docking predictions. The lighter bars indicate the percentage of docking results with a root mean square deviation (RMSD) below 2 Å, meaning the predicted pose is structurally accurate. The darker bars apply a stricter criterion, showing the proportion of predictions that are not only within 2 Å RMSD but also correctly positioned within the binding pocket (PB-valid). This distinction highlights the difference between general docking accuracy and more precise, biologically relevant binding predictions.

AI has transformed numerous scientific fields, with protein science being one of the most impacted areas. Understanding protein sequences is fundamental to biology, influencing drug discovery, synthetic biology, and disease research. Recent AI advancements have enabled scientists to analyze and predict protein functions, structures, and interactions with unprecedented accuracy. As the field evolves, these developments will affect healthcare, biotechnology, and regulatory frameworks. This section highlights key advancements in AI-driven protein analysis over the past year, focusing on public databases, research trends, and emerging policy considerations.

# 5.2 The Central Dogma

## Protein Sequence Analysis

### AI-Driven Protein Sequence Models

The past year has witnessed remarkable progress in AI models applied to protein sequences. Large-scale machine learning models have improved our ability to predict protein properties, accelerating research in structural biology and molecular engineering. As noted above, several notable protein sequencing models, like AlphaFold, ESM2, and ESM3, have recently been released.

ESM3 integrates multimodal inputs—sequence, structure, and interaction data—while its larger parameter size improves representativeness and predictive accuracy. As the ESM family has expanded in scale, protein prediction performance has improved. Newer models, such as ESM C, released in 2024, have achieved greater accuracy in predicting protein structures in the Critical Assessment of Structure Prediction (CASP15) challenge (Figure 5.2.1).

**Emergent structure prediction success, CASP15**
Source: EvolutionaryScale, 2024
Figure 5.2.1

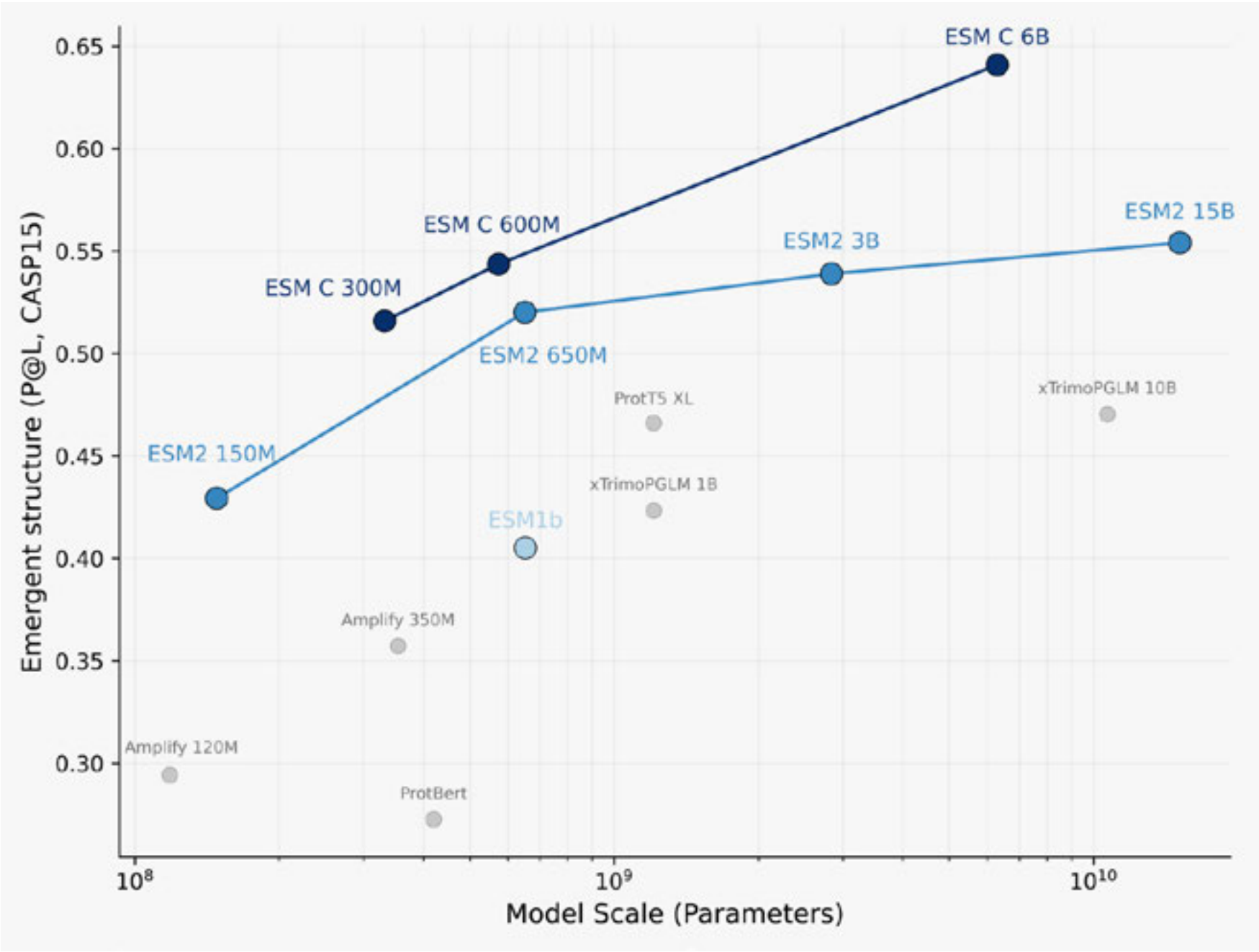

Other significant advancements include ProGen, a generative AI model that, in demonstrating the ability to design functional protein sequences, has highlighted the potential of AI-assisted protein engineering. Similarly, transformer-based models such as ProtT5 leverage deep learning to predict protein function and interactions directly from sequence data, advancing the field of computational biology. Figure 5.2.2 showcases key protein sequencing models and their parameter sizes, arranged by release date. As noted earlier, there is a clear trend toward increasingly larger models trained on ever-expanding datasets. These AI-driven approaches have transformed protein science by minimizing reliance on costly, time-intensive experimental methods, enabling rapid exploration of protein function and design.

**Size of protein sequencing models, 2020–24**

Source: RAISE Health, 2025 | Chart: 2025 AI Index report

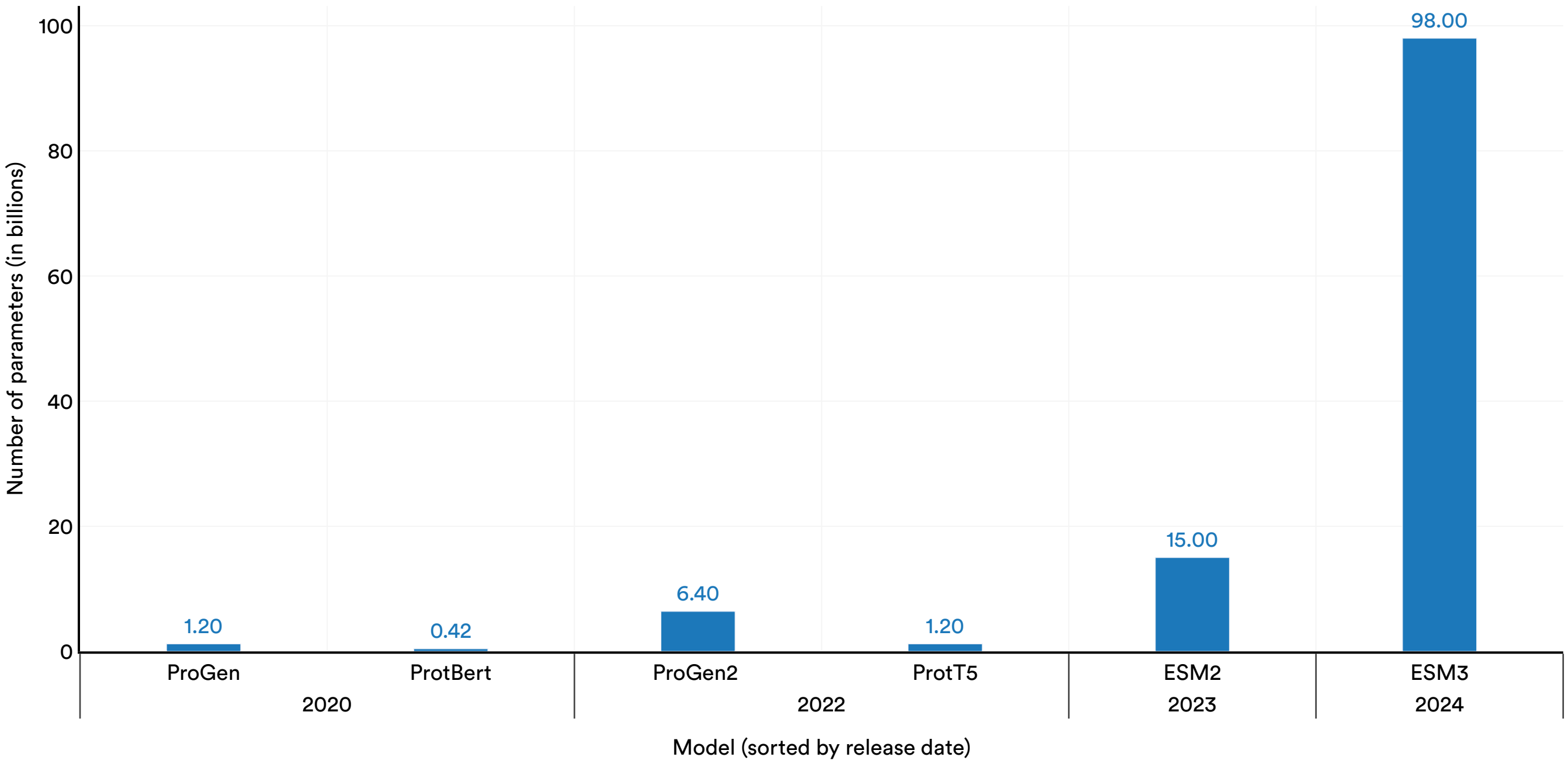

Figure 5.2.2

## Public Databases for Protein Science

The expansion of public databases has been crucial for AI applications in protein science. Well-curated, large-scale datasets enable AI models to train on diverse biological sequences, enhancing their predictive power. Figure 5.2.3 provides information on several key protein science databases and their release date.

### Key protein science databases

Source: AI Index, 2025

| Dataset | Release date | Description |
|---|---|---|
| Protein Data Bank (PDB) | 1971 | A database of experimentally solved protein structures. When first released, it was the first open-access digital resource in the biological sciences. |
| Pfam | 1995 | A comprehensive database of protein families, providing annotations and multiple sequence alignments generated through hidden Markov models. |
| STRING | 2000 | Dataset offering valuable information on protein interactions and evolutionary relationships. |
| UniProt | 2002 | Still the gold standard for protein sequence and function annotation, with AI-assisted curation improving accuracy. |
| PDBbind | 2004 | A subset of the PDB that contains protein biomolecular complexes, including protein-ligand, protein-protein, and protein-nucleic acid complexes. |
| AlphaFold Database | 2021 | An essential resource for structural biology, now integrating AI-driven models to predict missing experimental data. |

Figure 5.2.3

The number of entries in various public protein science databases has also steadily grown over time (Figure 5.2.4). The increasing availability of AI-generated protein insights has made these databases indispensable tools for researchers and industry professionals. However, maintaining data quality and preventing biases in AI models remain ongoing challenges.

### Growth of public protein science databases, 2019–25

Source: RAISE Health, 2025 | Chart: 2025 AI Index report

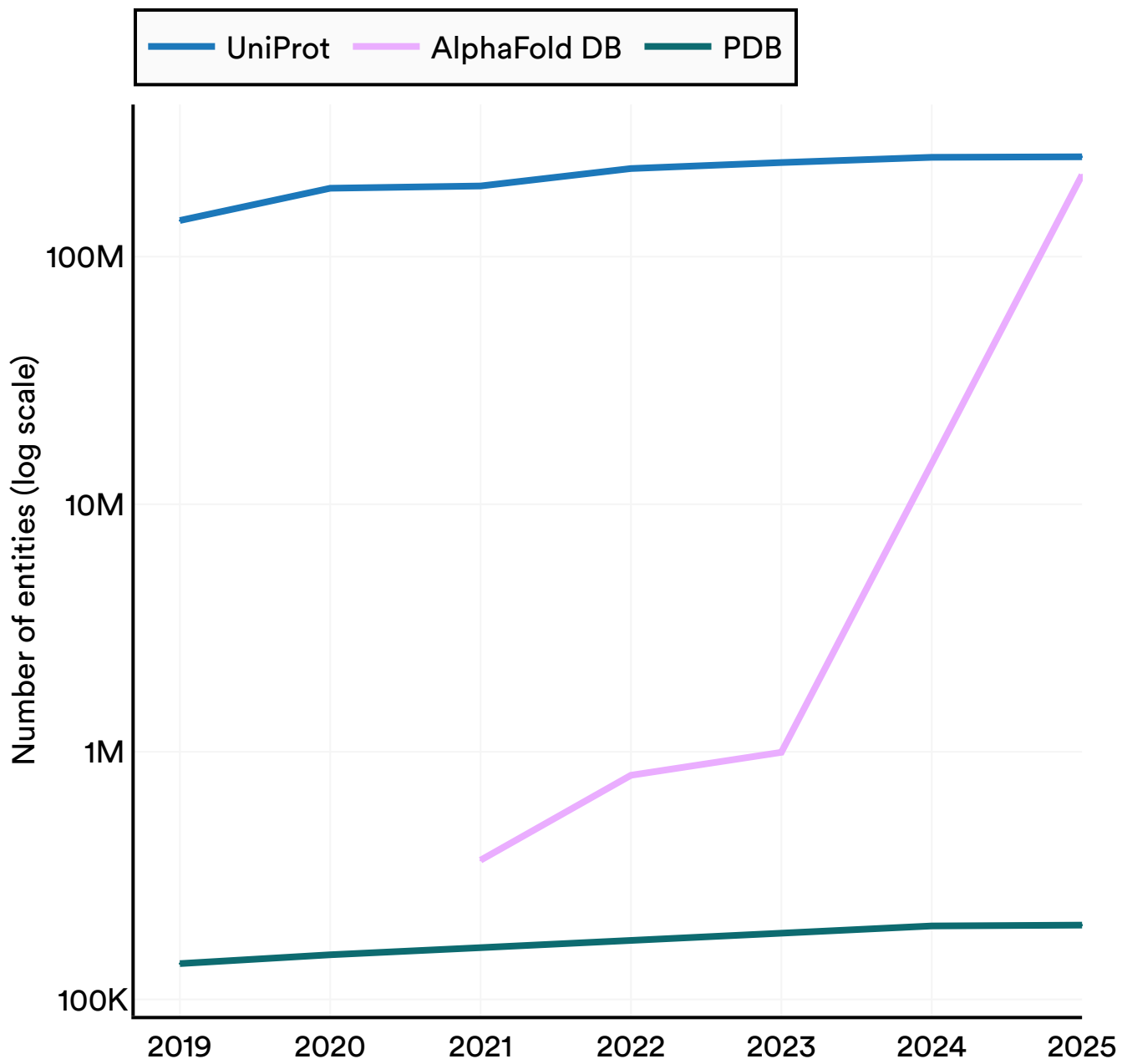

Figure 5.2.4

# Research and Publication Trends

## AI-Driven Protein Science Publications

AI applications in protein science have gained significant traction in academic research, as evidenced by an increase in AI-driven studies on PubMed and bioRxiv preprints over the past year. These studies focus on several key areas. Protein structure prediction has become more accessible due to advances in machine learning, providing deeper structural insights. AI models now infer biochemical functions from raw sequence data with greater accuracy, enhancing function prediction. In addition, AI models are being developed that can predict protein-drug interactions and even create new drugs from scratch that can target specific proteins. Both of these tasks are crucial for drug discovery and drug development. Furthermore, AI-generated proteins with novel functions are emerging, particularly in enzyme engineering and therapeutic applications, marking a significant step forward in synthetic protein design. Figure 5.2.5 illustrates the proportion of protein AI-driven research within biological sciences in 2024. The most researched topic was function prediction (8.4%), followed by protein structure prediction (7.6%) and protein-drug interactions (3.0%)

**Proportion of AI-driven protein research in the biological sciences, 2024**
Source: RAISE Health, 2025 | Chart: 2025 AI Index report

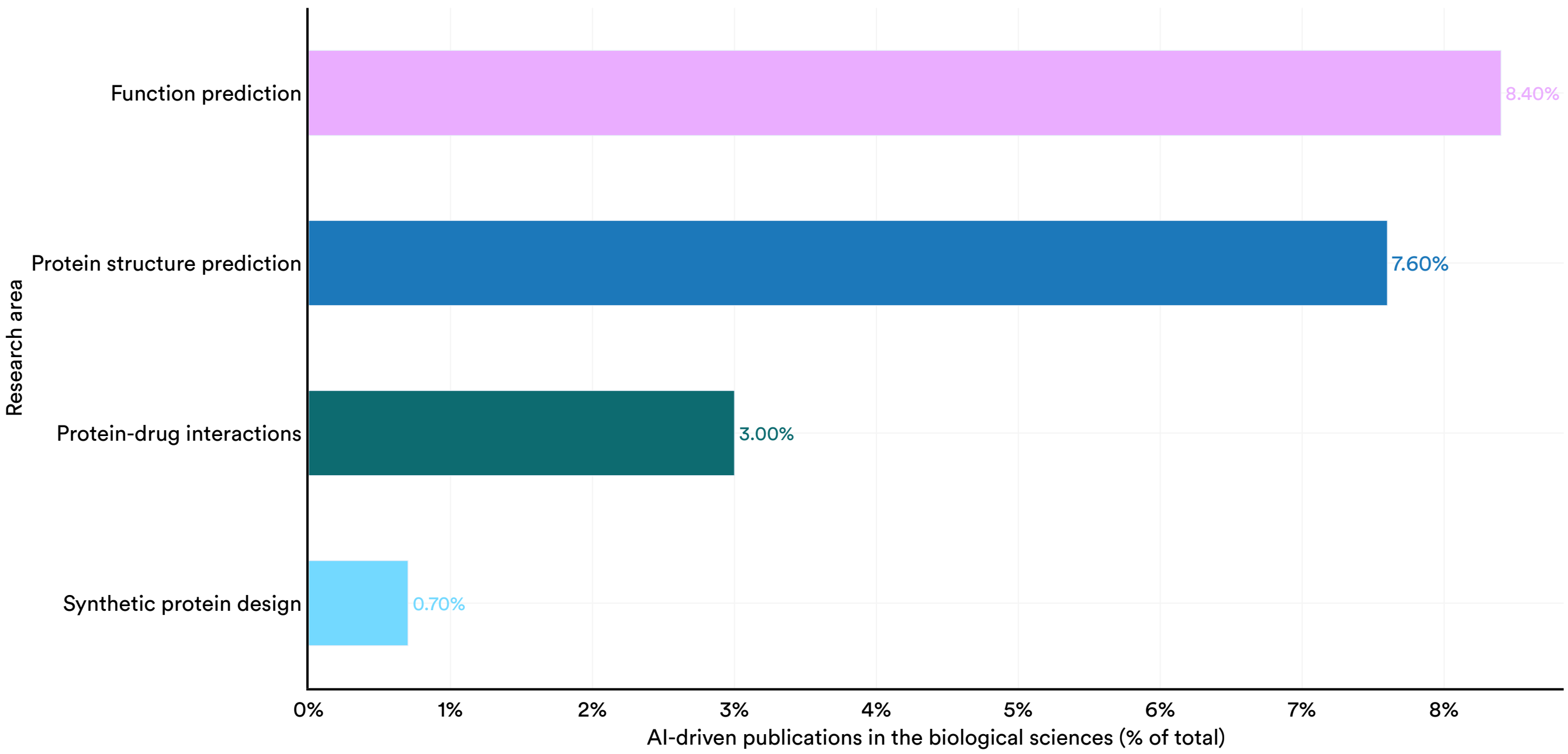

Figure 5.2.5

# Image and Multimodal AI for Scientific Discovery

Advances in cryo-electron microscopy, high-throughput fluorescence microscopy, and whole-slide imaging allow scientists to examine and analyze atomic, subcellular context and tissue-level structures with high precision to reveal new insights into complex biological processes. To achieve this, researchers interpret and contextualize image findings with existing scientific knowledge to link observations to biological functions and disease relevance. Given the rise of high-throughput microscopy, active research has increasingly focused on the intersection of vision, vision-language, and, more recently, vision-omics foundation models. The number of microscopy foundation models has increased over time across various techniques (Figure 5.2.6). Light-based models doubled from four to eight in 2024, and, while no electron or fluorescence models were released in 2023, four models for each technique emerged in 2024. Overall, foundation models for microscopy are increasing as more data is collected and made publicly available.

**Number of foundation models per microscopy techniques, 2023–24**

Source: RAISE Health, 2025 | Chart: 2025 AI Index report

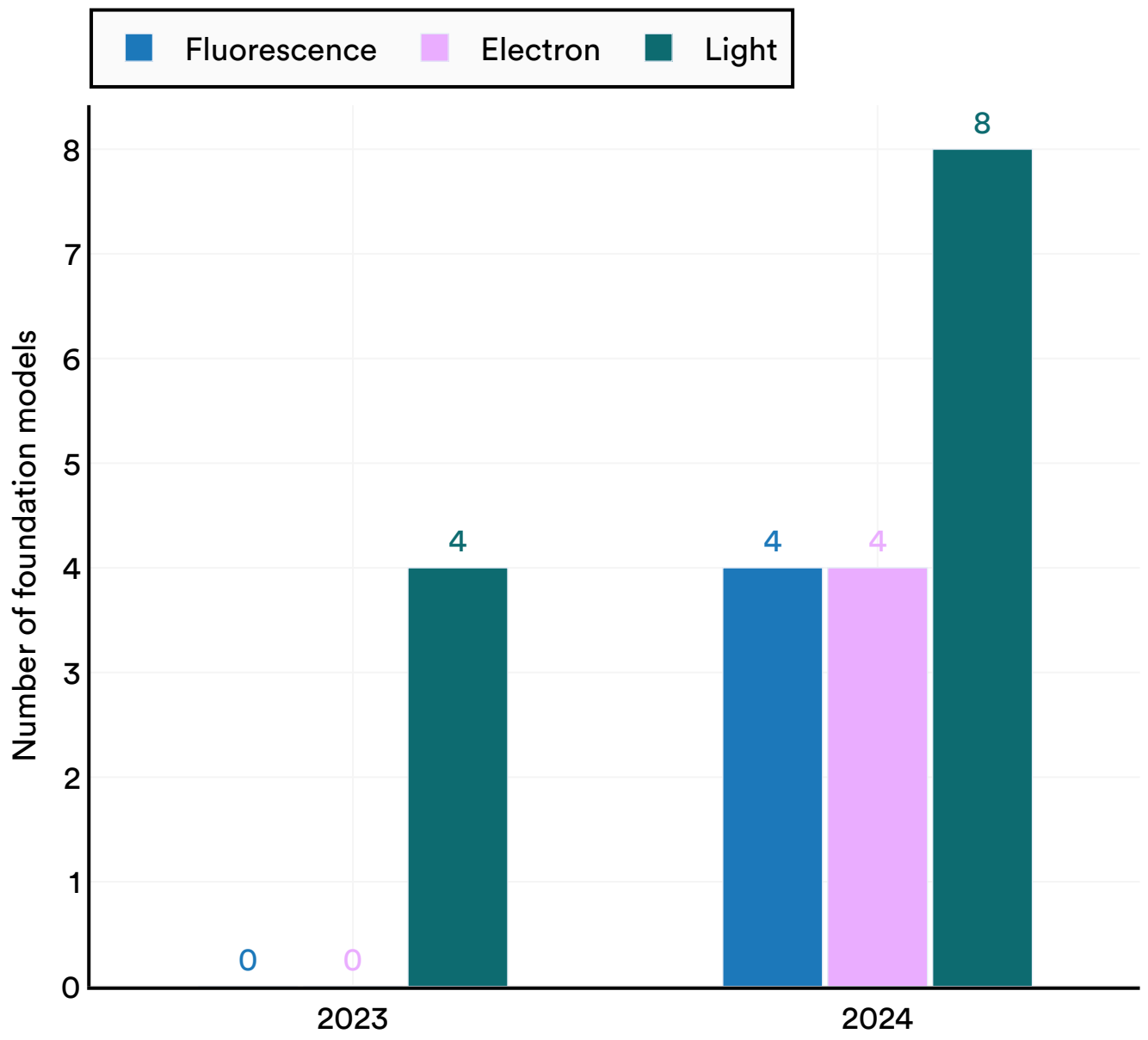

Figure 5.2.6

# 5.3 Clinical Care, Imaging

## Data: Sources, Types, and Needs

AI in medical imaging is rapidly evolving, expanding into new data modalities, and addressing increasingly complex clinical questions. More than 80% of FDA-cleared machine learning software targets the analysis of medical images. Currently, AI is predominantly applied to two-dimensional (2D) data settings, where conventional image-processing architectures, such as convolutional neural networks (CNNs) and transformers, can be effectively utilized. However, despite a number of successes in this field, many AI applications in medical imaging rely on highly limited training datasets.

In histopathology, for example, while staining patient biopsies for histological analysis is routine, only a small fraction of these samples is digitized and made publicly available. Even fewer datasets contain the necessary matched annotations or omics data required for advanced classification tasks. Publicly available histopathology cohorts rarely exceed 10,000 patient samples, with The Cancer Genome Atlas (TCGA) providing one of the most comprehensive collections—comprising 11,125 patient samples with matched clinical annotations, genomic sequencing, and protein expression data across 32 cancer types. As a result, histopathology AI models are often trained on fewer than 1,000 patient samples, particularly when genomic or proteomic data serve as labels. Limited training sets increase the risk of data overfitting and poor generalization.

Figure 5.3.1 illustrates the geographic distribution of U.S. cohorts used to train deep learning algorithms. Most cohorts originate from California, Massachusetts, and New York, raising concerns about the limited scope of the datasets used to train these algorithms.

**US patient cohorts used to train clinical machine learning algorithms by state, 2015–19**

Source: Kaushal et al., 2020 | Chart: 2025 AI Index report

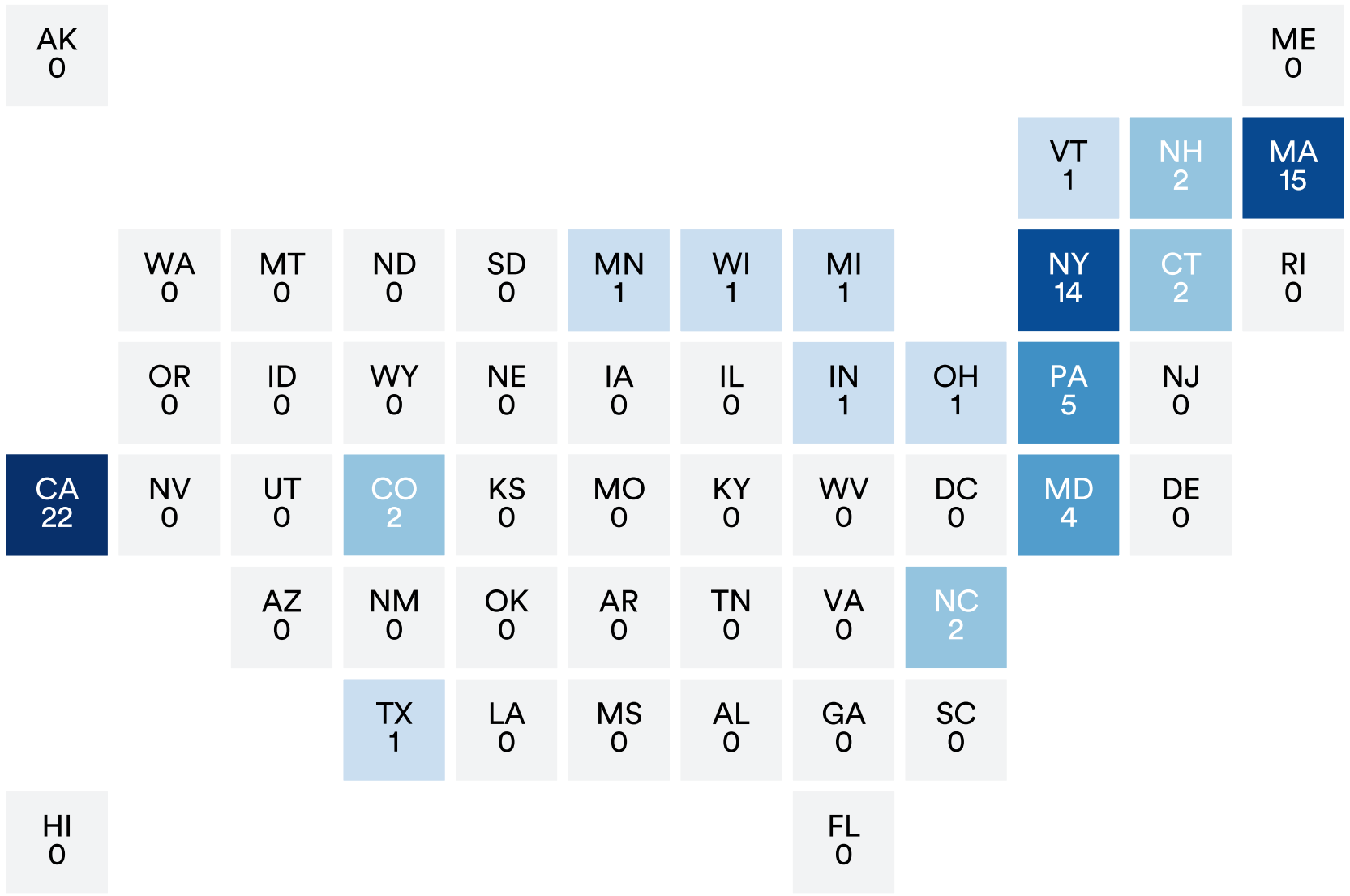

Figure 5.3.1

These data limitations are more pronounced for three-dimensional (3D) medical imaging. While AI has traditionally focused on 2D modalities such as chest X-rays, histopathology slides, and fundus photography, recent advancements have expanded its application to 3D imaging modalities, including computed tomography (CT), magnetic resonance imaging (MRI), and 3D histopathology analysis. Three-dimensional analysis provides richer data, enabling AI models to learn patterns from volumetric structures and complex surfaces that may not be apparent in 2D slices. Although promising approaches have been developed for the use of AI to analyze 3D medical images, similar data limitations and needs persist. Publicly available 3D datasets remain limited, with UK Biobank (around 100,000 MRI scans) and TCIA (around 50,000 studies) among the largest. Although 3D samples are routinely collected in histopathology, 3D imaging is not standard practice, resulting in an absence of publicly available 3D histopathology datasets. Standardization challenges persist due to acquisition variability in pathology. Differences in instrument settings, staining techniques, and institutional practices introduce batch effects, which are further exacerbated by limited training datasets.

Training accurate AI models requires large datasets: CNNs have succeeded with around 10,000 labeled images , but transformers need orders of magnitude more data. MIMIC-CXR (377,000 images) and CheXpert Plus (around 226,000 frontal-view radiographs with aligned radiology reports and patient metadata) are important resources but remain smaller than ImageNet (around 14 million images). Data completeness and bias issues remain key challenges.

Figure 5.3.2 illustrates the token volume in text and image datasets used to train various leading medical language and imaging models, in comparison to various all-purpose text and image models. GatorTron, a large clinical LLM designed to extract patient information from unstructured electronic health records, was trained on 82 billion tokens. In contrast, Llama 3 was trained on 15 trillion tokens—nearly 182 times more. On the imaging side, RadImageNet, an open radiologic deep learning research dataset, contains 16 million image-equivalent tokens, while DALL-E, an early OpenAI image generator, was trained on approximately 6 billion—roughly 375 times more.

**Training dataset token volumes: medical vs. nonmedical language and imaging models**

Source: RAISE Health, 2025 | Chart: 2025 AI Index report

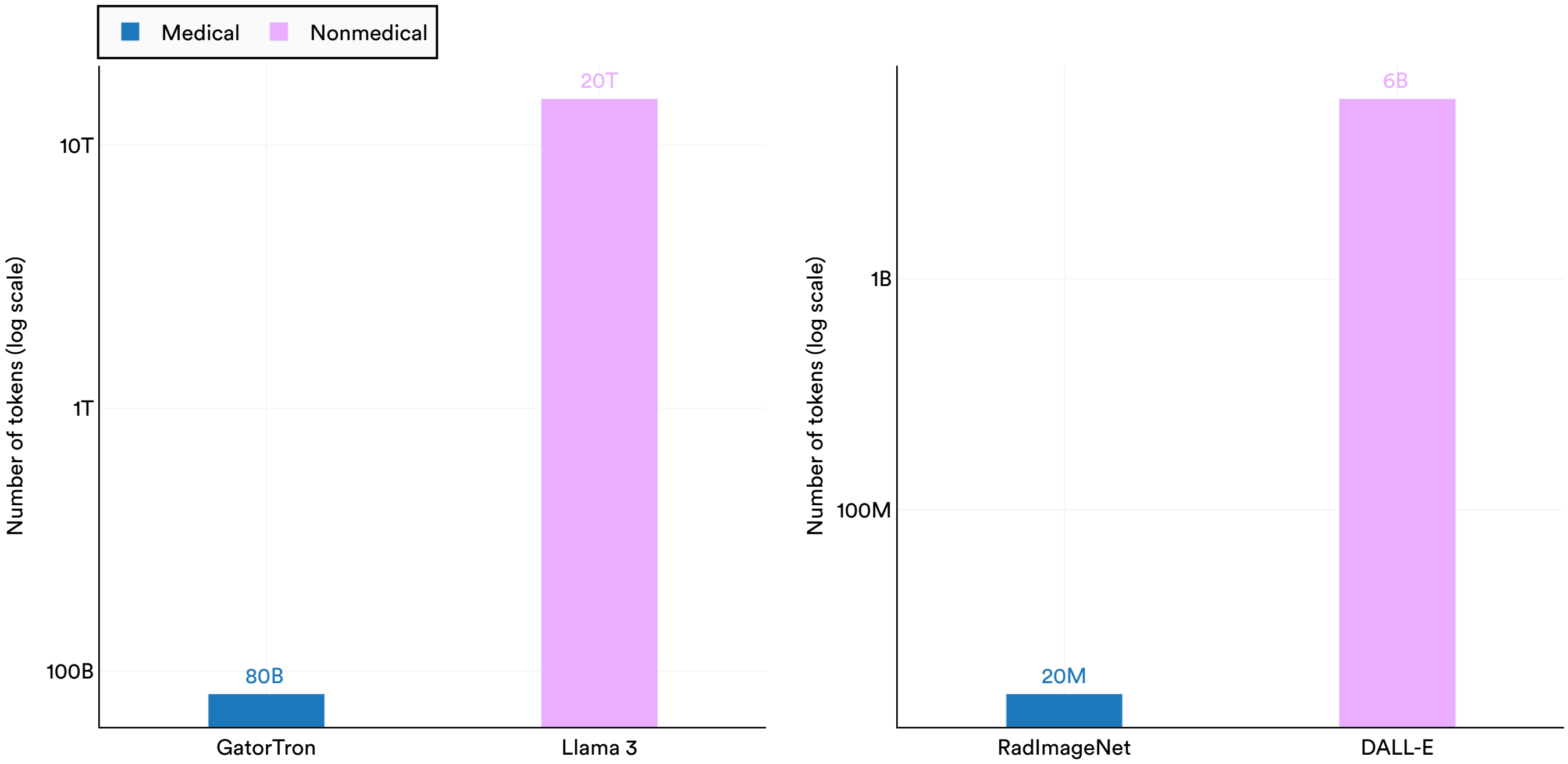

Figure 5.3.2

Longitudinal imaging is important for modeling disease progression but remains underrepresented. ADNI (around 2,000 participants over 15-plus years) exemplifies such efforts, but scalable multimodal longitudinal datasets are rare. Addressing these gaps requires privacy-preserving data-sharing (e.g., federated learning), synthetic data generation, and improved annotation strategies.

To train and validate robust medical imaging AI models, larger, more comprehensive, and multicohort collections of training data are required. By increasing the availability of high-quality, labeled training data, models can be expected to achieve improved performance. Additionally, better validation practices will bolster confidence in these models, facilitating their transition into clinical practice.

## Advanced Modeling Approaches

Figure 5.3.3 presents leading clinical imaging modeling approaches, notable releases per approach, and key challenges associated with each.

### Imaging modeling approaches and notable AI models

Source: AI Index, 2025

| Modeling approach | Notable releases | Advantages | Challenges |
|---|---|---|---|
| Diffusion models | 1. RoentGen (2022)<br>2. RNA-CDM (2023)<br>3. XReal (2024) | Generate synthetic medical images for enhanced training, privacy, and pathology-specific augmentation. Outperform GANs in stability and diversity. | Dataset biases, hallucinated artifacts, diagnostic uncertainty. |
| Large vision-language models (LVLMs) | 1. CheXagent (2024)<br>2. Merlin (2024)<br>3. Med-Gemini (2024)<br>4. PathChat (2024)<br>5. TITAN (2024)<br>6. PRISM (2025)<br>7. BiomedParse (2025) | Integrate medical images with text for improved diagnosis, segmentation, and report automation. LVLMs extend multimodal capabilities. | Data scarcity, generalization to low-resource settings, computational demands. |
| 2D vision-only foundation models | 1. CTransPath (2022)<br>2. Virchow (2024)<br>3. UNI (2024)<br>4. MedSAM(2024) | Pan-cancer detection, biomarker prediction, and image segmentation. Reduce annotation burdens. | Domain generalization, cross-modal adaptability. |
| Multiscale/slide-level models | 1. HIPT (2022)<br>2. MEGT (2023)<br>3. MG-Trans (2023)<br>4. HIGT (2023)<br>5. Prov-GigaPath (2024) | Enhance whole-slide imaging analysis using hierarchical transformers and graph-based models for spatial relationships. Improve diagnostic fidelity and interpretability. | Scalability, computational efficiency, dataset variability. |

Figure 5.3.3

In recent years, there has been a notable rise in foundation models being used for medical imaging purposes. Figure 5.3.4 categorizes notable models by medical discipline. In recent years, the number of medical imaging foundation models has risen sharply, with a particularly high concentration of newly launched pathology models.

### Medical disciplines and notable AI models

Source: AI Index, 2025

| Discipline | Notable releases |
| --- | --- |
| Echocardiology | 1. EchoCLIP (2024) |
| Oncology | 1. MUSK (2025) |
| Ophthalmology | 1. RETFound (2023)<br>2. VisionFM (2024) |
| Pathology | 1. CTransPath (2022)<br>2. CHIEF (2024)<br>3. Prov-GigaPath (2024)<br>4. PathChat (2024)<br>5. TITAN (2024)<br>6. Virchow (2024)<br>7. UNI (2024) |
| Radiology | 1. RoentGen (2022)<br>2. CheXagent (2024)<br>3. Merlin (2024)<br>4. PRISM (2025) |

Figure 5.3.4

# 5.4 Clinical Care, Non-Imaging

## Clinical Knowledge

The following section examines the performance of LLMs and recent AI models on key medical knowledge benchmarks.

### MedQA

Evaluating the clinical knowledge of AI models involves determining the extent of their medical expertise, particularly knowledge applicable in a clinical setting.

Introduced in 2020, MedQA is a comprehensive dataset derived from professional medical board exams, featuring over 60,000 clinical questions designed to challenge doctors. AI performance on the MedQA benchmark has advanced significantly. A team of Microsoft and OpenAI researchers recently tested o1, which achieved a new state-of-the-art score of 96.0%—a substantial 5.8 percentage point improvement over the record set in 2023 (Figure 5.4.1). Since late 2022, performance on the benchmark has increased by 28.4 percentage points. As with other general knowledge benchmarks discussed in Chapter 2, MedQA may be approaching a saturation point, indicating the need for more challenging evaluations.

**MedQA: test accuracy**

Source: RAISE Health, 2025 | Chart: 2025 AI Index report

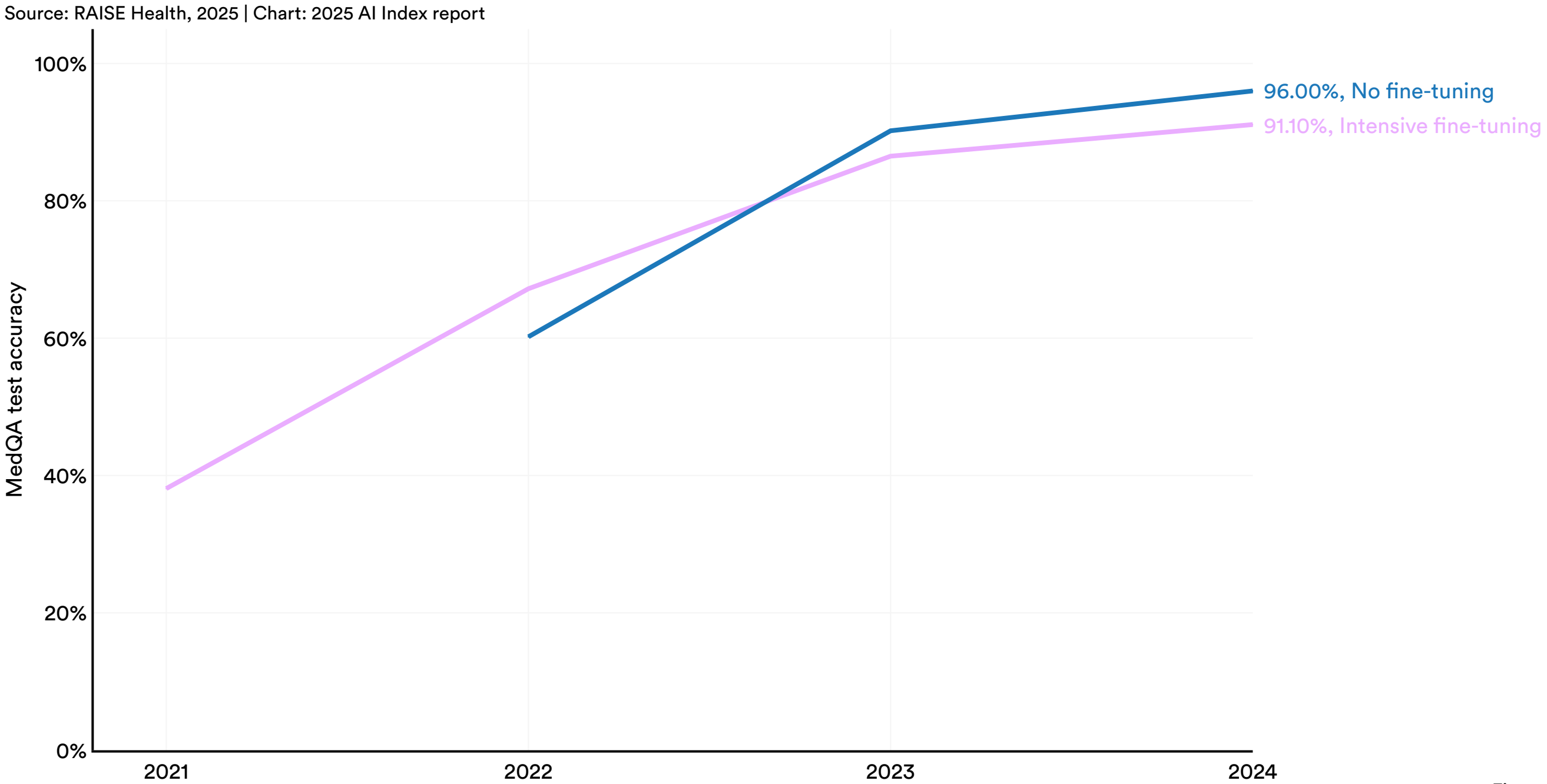

Figure 5.4.1

**Highlight:**

# AI Doctors and Cost-Efficiency Considerations

Some researchers argue that evaluating medical LLMs requires more comprehensive benchmarks than MedQA, those that span a broader range of medical domains. Relying solely on standard medical QA benchmarks like MedQA—while valuable—may overlook the complexities of real-world clinical applications. Alternatively, using multiple benchmarks can offer greater clinical relevance and a more robust assessment of model performance.

This year, new research from UC Santa Cruz, the University of Edinburgh, and the National Institutes of Health has taken a more expansive approach to testing AI medical systems. The study evaluated five leading large language models, including the newly developed o1, which features chain-of-thought reasoning. The other models assessed were GPT-3.5, Llama 3-8B, GPT-4, and Meditron-70B—the last of which is a specialized medical model. These models were tested on a diverse set of medical benchmarks covering various tasks, including concept recognition, text summarization, knowledge-based QA, clinical decision support, and medical calculations. Figure 5.4.2 presents the average performance of these five LLMs across 19 medical datasets. The findings indicate that clinical knowledge performance in LLMs is improving, particularly for newer models like o1 equipped with real-time reasoning capabilities. However, persistent challenges remain, including issues with hallucinations and inconsistent multilingual performance.

Previous research, cited in last year's AI Index, demonstrated that prompting techniques like Medprompt can significantly enhance LLM performance on medical benchmarks without additional fine-tuning. OpenAI's recently released o1 model incorporates some of these insights by employing runtime reasoning before generating final responses. Researchers found that o1 outperforms the GPT-4 series with Medprompt, even without specialized prompting techniques. However, their analysis also highlights the accuracy-cost trade-off associated with o1. While it achieves a 5.8 percentage point higher score than GPT-4 Turbo with Medprompt, it is approximately 1.5 times more expensive. Figure 5.4.3 illustrates the cost versus accuracy trade-off on the MedQA benchmark. This trade-off highlights a key consideration for medical professionals deploying AI in clinical settings: the need to balance performance gains with computational costs.

**Performance of select LLMs on medical datasets**
Source: Xie et al., 2024 | Chart: 2025 AI Index report

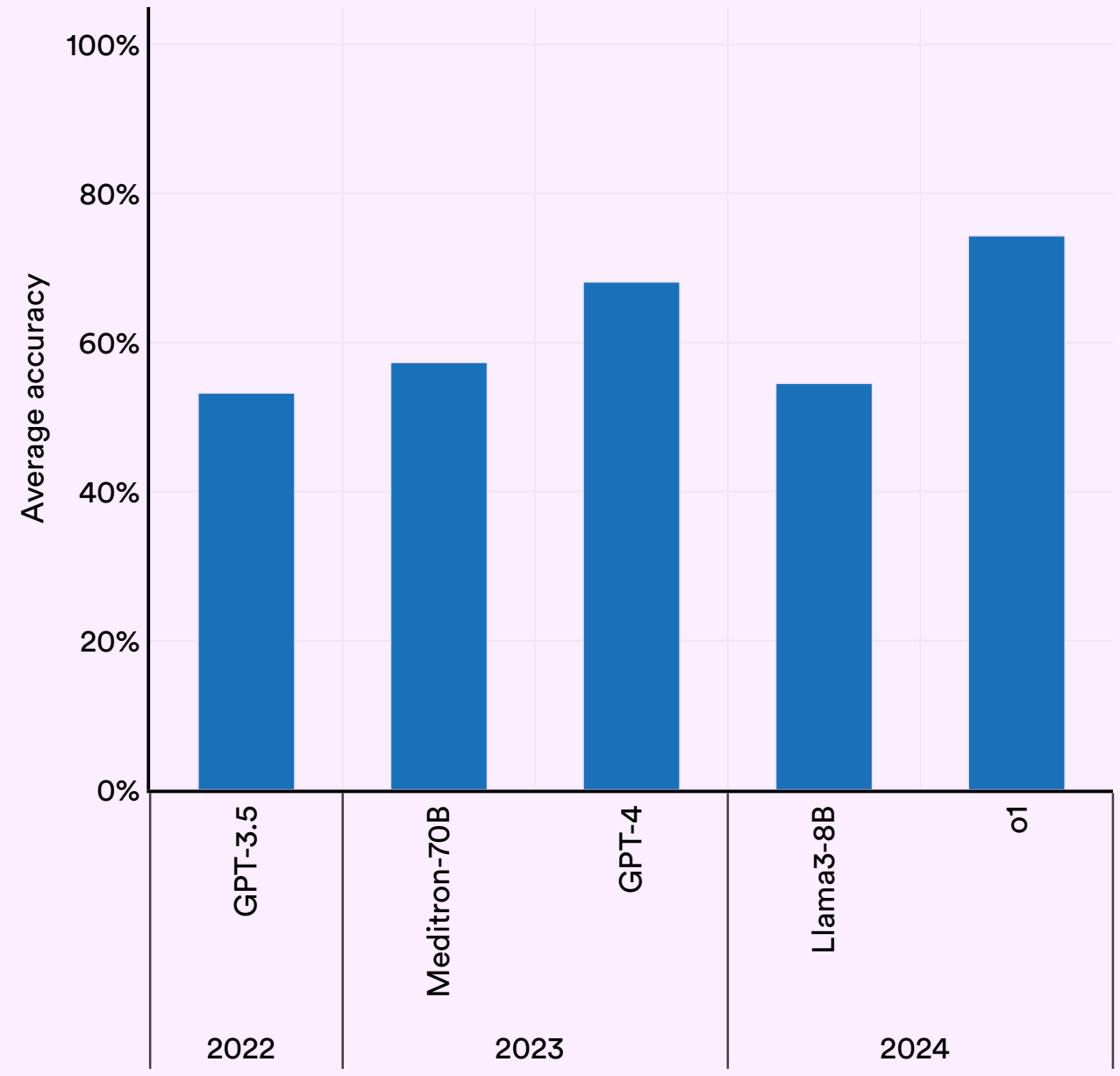

Figure 5.4.2

**Enhanced pareto frontier: accuracy vs. cost**
Source: Nori et al., 2024

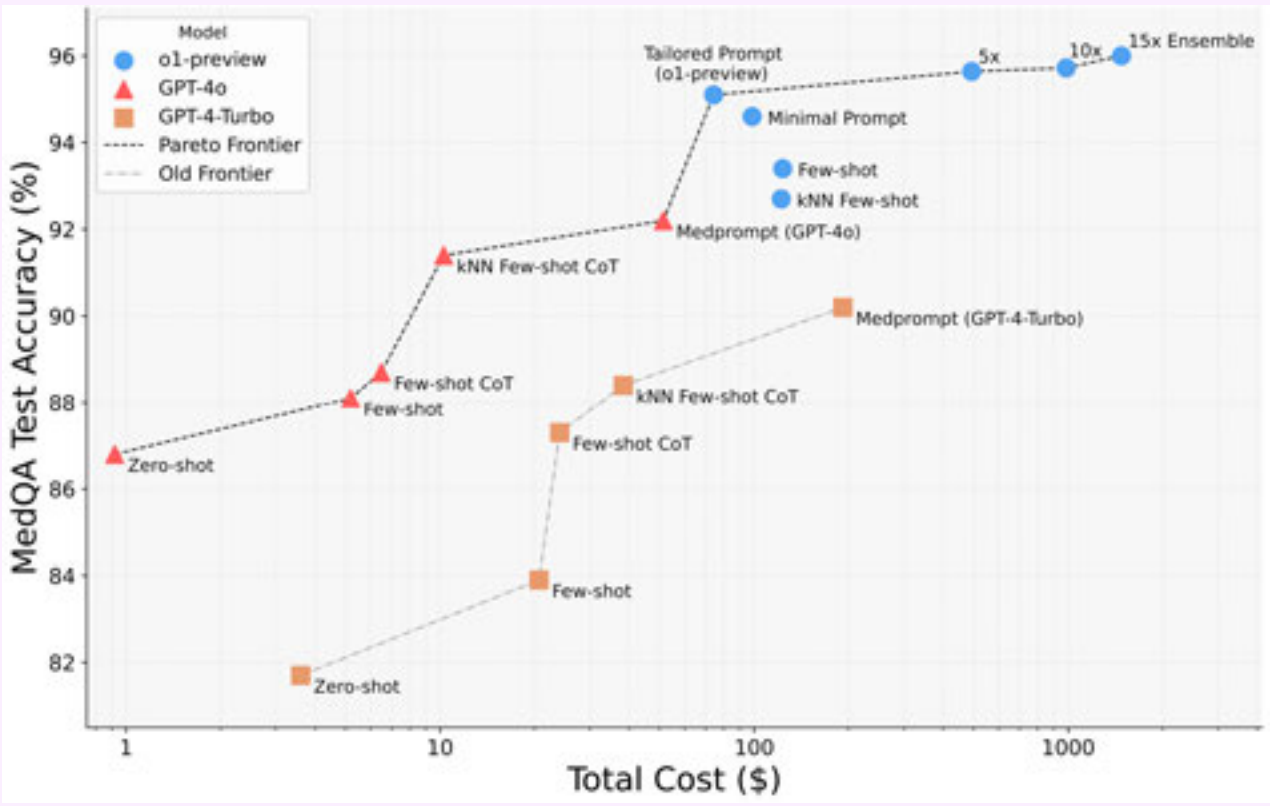

Figure 5.4.3

# Evaluation of LLMs for Healthcare Performance

## Overview

There has been an explosion in interest in the evaluation of language model performance on healthcare tasks. A PubMed search for "large language model" returned 1,566 papers starting in 2019 with 1,210 published in 2024 alone (Figure 5.4.4).

### Number of publications on large language models in PubMed, 2019–24

Source: RAISE Health, 2025 | Chart: 2025 AI Index report

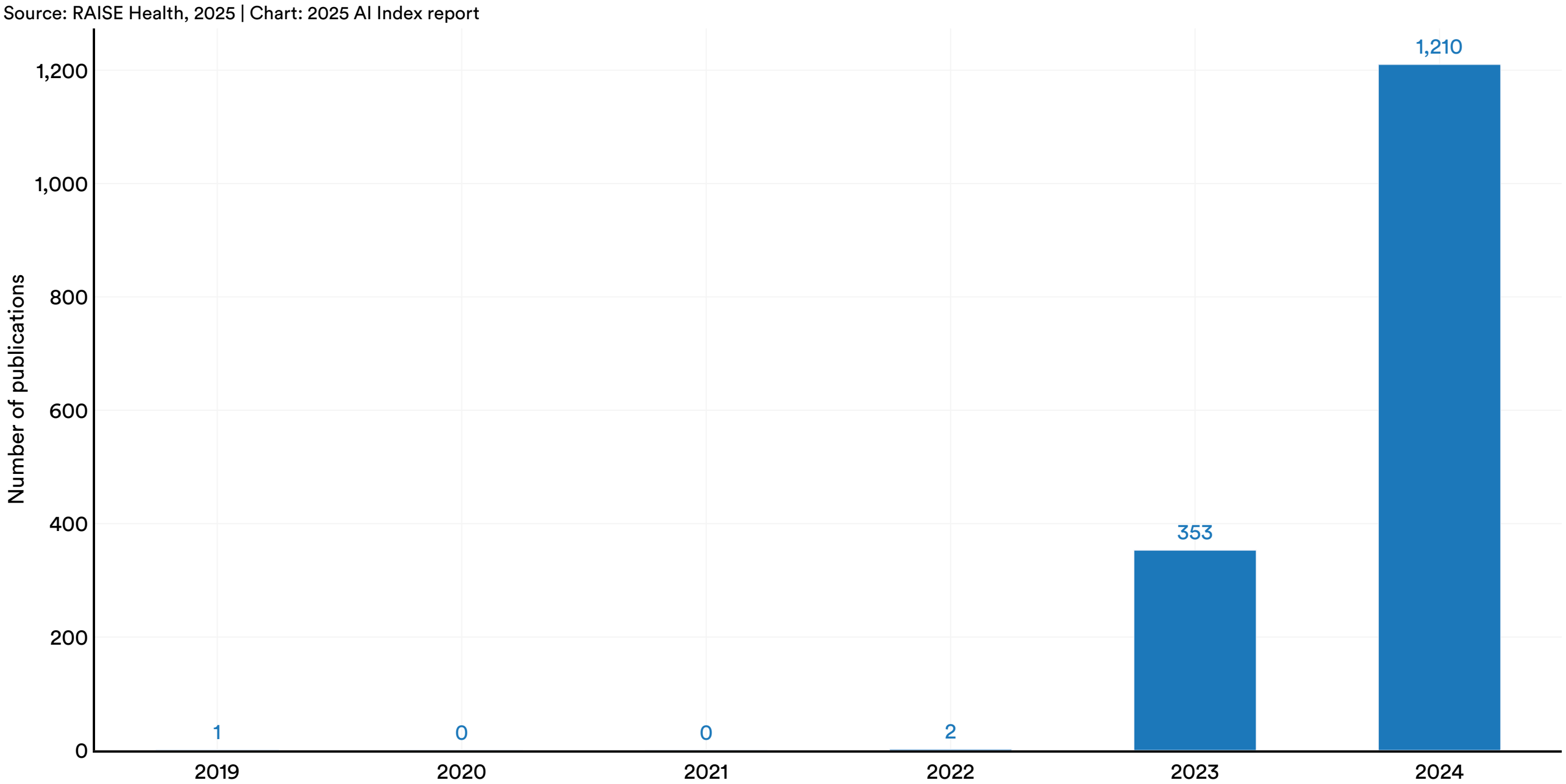

Figure 5.4.4

A systematic review in early 2024 identified over 500 papers evaluating the performance of NLP on healthcare tasks with a heavy emphasis on medical decision-making (Figure 5.4.5). Most of the healthcare studies that evaluated the performance of NLP systems focused on enhancing medical knowledge (419) and making diagnoses (178).

**Healthcare tasks, NLP and NLU tasks, and dimensions of evaluation across 519 studies**
Source: RAISE Health, 2025 | Chart: 2025 AI Index report

| Task | Accuracy | Comprehensiveness | Factuality | Robustness | Fairness, bias, and toxicity evaluation | Deployment metrics | Calibration and uncertainty |
|---|---|---|---|---|---|---|---|
| Enhancing medical knowledge | 222 | 91 | 44 | 33 | 16 | 10 | 3 |
| Making diagnoses | 100 | 38 | 11 | 11 | 14 | 4 | 0 |
| Educating patients | 88 | 68 | 32 | 22 | 18 | 3 | 2 |
| Making treatment recommendations | 47 | 22 | 9 | 8 | 3 | 1 | 0 |
| Communicating with patients | 35 | 29 | 8 | 15 | 22 | 1 | 0 |
| Care coordination and planning | 36 | 24 | 5 | 5 | 7 | 1 | 0 |
| Triaging patients | 24 | 7 | 5 | 2 | 8 | 8 | 0 |
| Carrying out a literature review | 18 | 7 | 3 | 2 | 2 | 2 | 0 |
| Synthesizing data for research | 16 | 7 | 2 | 3 | 2 | 2 | 0 |
| Generating medical reports | 8 | 8 | 2 | 0 | 3 | 0 | 0 |
| Conducting medical research | 8 | 7 | 3 | 3 | 3 | 0 | 0 |
| Providing asynchronous care | 8 | 5 | 3 | 3 | 1 | 1 | 0 |
| Managing clinical knowledge | 5 | 5 | 1 | 1 | 0 | 0 | 0 |
| Clinical note-taking | 6 | 2 | 1 | 1 | 0 | 0 | 1 |
| Generating clinical referrals | 3 | 0 | 0 | 0 | 0 | 0 | 0 |
| Enhancing surgical operations | 3 | 3 | 1 | 1 | 0 | 0 | 0 |
| Biomedical data mining | 2 | 0 | 0 | 0 | 0 | 0 | 0 |
| Generating billing codes | 1 | 0 | 0 | 0 | 0 | 0 | 0 |
| Writing prescriptions | 1 | 0 | 0 | 0 | 0 | 0 | 0 |
| Question answering* | 398 | 194 | 71 | 61 | 54 | 14 | 5 |
| Text classification* | 29 | 10 | 6 | 5 | 10 | 2 | 0 |
| Information extraction* | 29 | 12 | 8 | 5 | 4 | 6 | 0 |
| Summarization* | 29 | 21 | 7 | 3 | 8 | 0 | 1 |
| Conversational dialogue* | 6 | 6 | 1 | 1 | 5 | 1 | 0 |
| Translation* | 5 | 1 | 2 | 2 | 1 | 2 | 0 |

Figure 5.4.5[4]

4 The asterisks represent tasks in NLP and NLU.

## Diagnostic Reasoning With LLMs

Diagnostic errors account for substantial patient harm, and many organizations are exploring AI as a tool to improve the diagnostic process.

### Highlight:

## LLMs Influence Diagnostic Reasoning

A 2024 single-blind, randomized trial tested GPT-4 assistance against conventional resources in tackling complex clinical vignettes. The study involved 50 U.S.-licensed physicians and evaluated whether AI-enhanced decision-making could improve diagnostic accuracy and efficiency. The results revealed no significant improvement when physicians used GPT-4 alongside traditional resources. In fact, physicians with AI assistance performed only slightly better (76%) than those who relied solely on conventional tools (74%). However, in a secondary analysis, GPT-4 alone outperformed both groups, achieving a 92% diagnostic reasoning score, a 16-percentage-point increase over physicians working without AI (Figure 5.4.6). Despite AI's superior standalone performance, integrating it into clinical workflows proved challenging. There was no clear advantage in time efficiency, as case completion times remained statistically unchanged across conditions.

While purely autonomous AI outperformed physician-only efforts, simply giving doctors access to an LLM did not enhance their performance. This underscores a phenomenon seen in other AI-human collaborations: Bridging the gap between excellent model performance in isolation and effective synergy with clinicians requires rethinking workflows, user training, and interface design.

**LLM performance in clinical diagnosis**

Source: Goh et al., 2024 | Chart: 2025 AI Index report

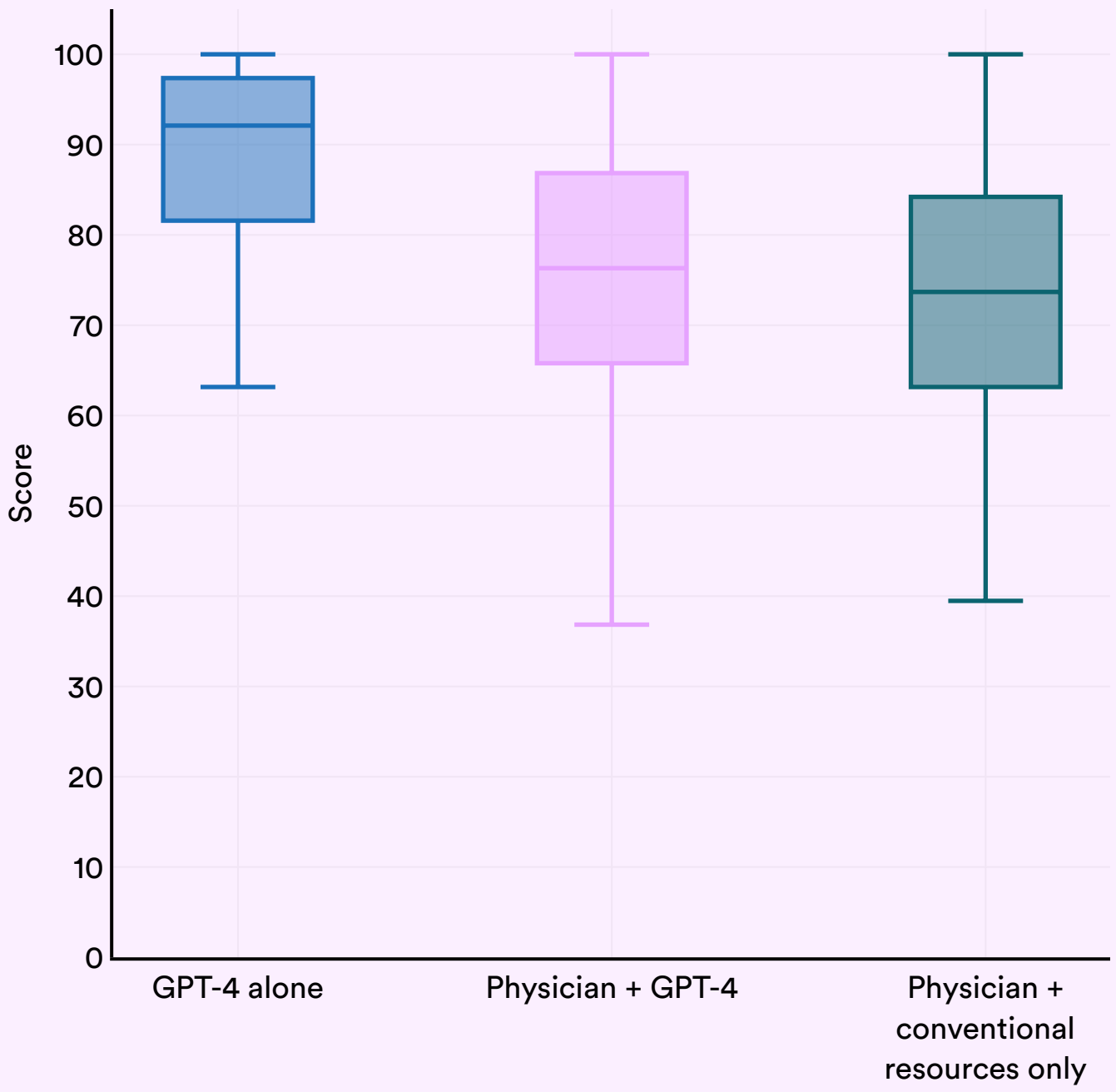

Figure 5.4.6

## Management Reasoning and Patient Care Decisions

Beyond diagnosis, physicians must juggle treatment decisions, risk-benefit trade-offs, and patient preferences—collectively referred to as "management reasoning." Researchers tested whether LLMs could improve these complex, context-dependent skills.

Highlight:

## GPT-4 Assistance on Patient Care Tasks

A 2024–25 prospective, randomized, controlled trial evaluated the impact of GPT-4 assistance on complex clinical management decisions. The study involved 92 physicians, with half using GPT-4 alongside standard resources and the other half relying solely on conventional references. Physicians assisted by GPT-4 outperformed the control group by approximately 6.5 percentage points (Figure 5.4.7). Interestingly, GPT-4 alone performed on par with GPT-4-assisted physicians, suggesting that in certain well-defined scenarios, near-autonomous AI-driven management support may be feasible. However, AI assistance came with a trade-off, as physicians using GPT-4 spent slightly longer on each scenario—a delay researchers attributed to deeper reflection and analysis. Generative AI can meaningfully improve clinical decision-making, but its impact may be qualitative rather than purely efficiency-driven.

**Impact of LLM assistance on clinical management**

Source: Goh et al., 2025 | Chart: 2025 AI Index report

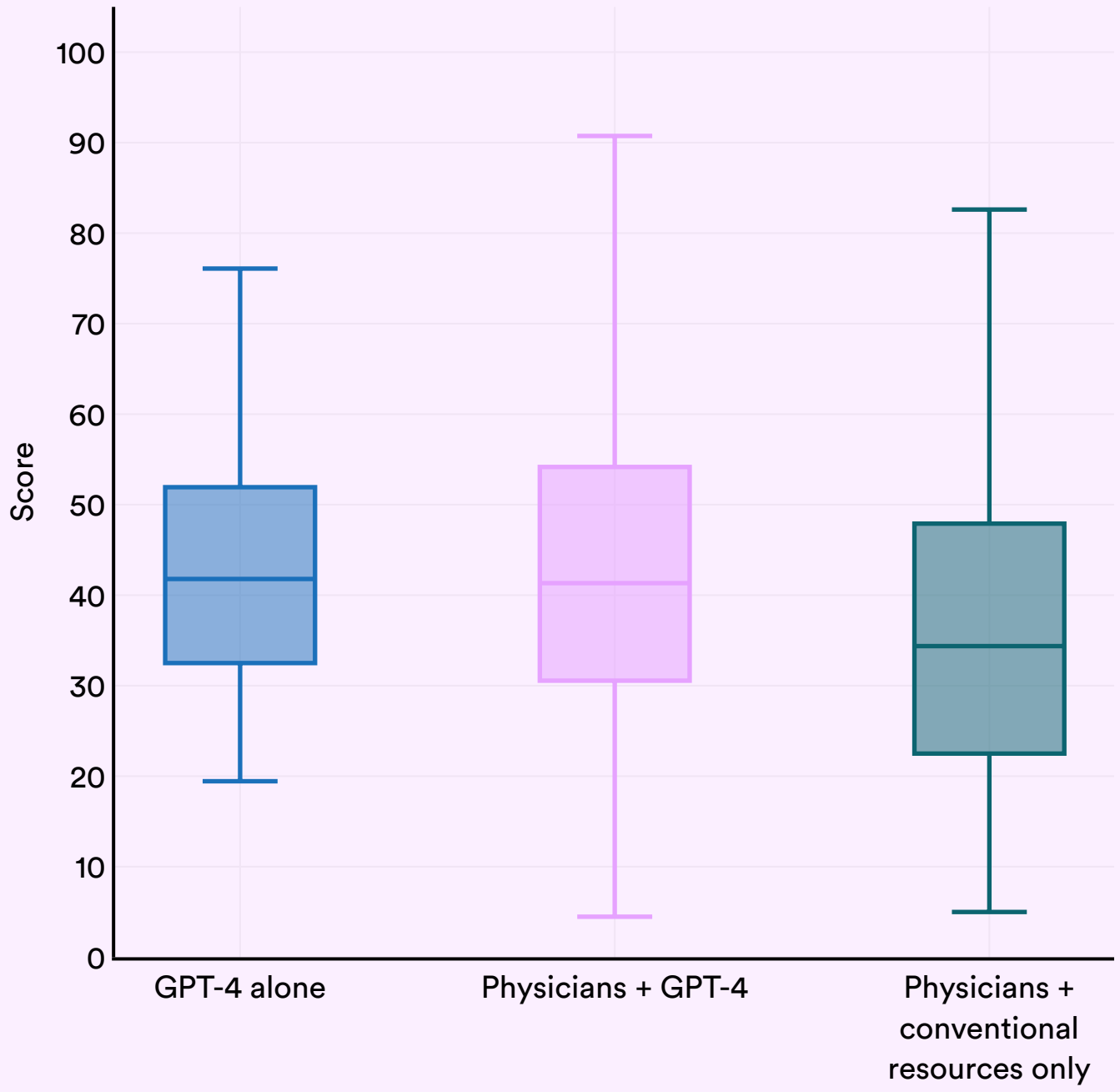

Figure 5.4.7

## Ambient AI Scribes

Clinical documentation has long been a source of clinician burden and burnout. Ambient scribe technology has rapidly evolved to integrate LLMs into the processing pipeline for physician-patient recordings. The first study, published in NEJM Catalyst, describes the launch of ambient AI scribe technology at Kaiser Permanente Northern California in late 2023. The technology was eventually adopted by thousands of clinicians before the end of the pilot (Figure 5.4.8). This was followed by a second study, published in JAMIA, that describes the pilot experience at Intermountain Health. Both studies were based on earlier versions of the technology that were not fully automated or integrated into the electronic health record (EHR).

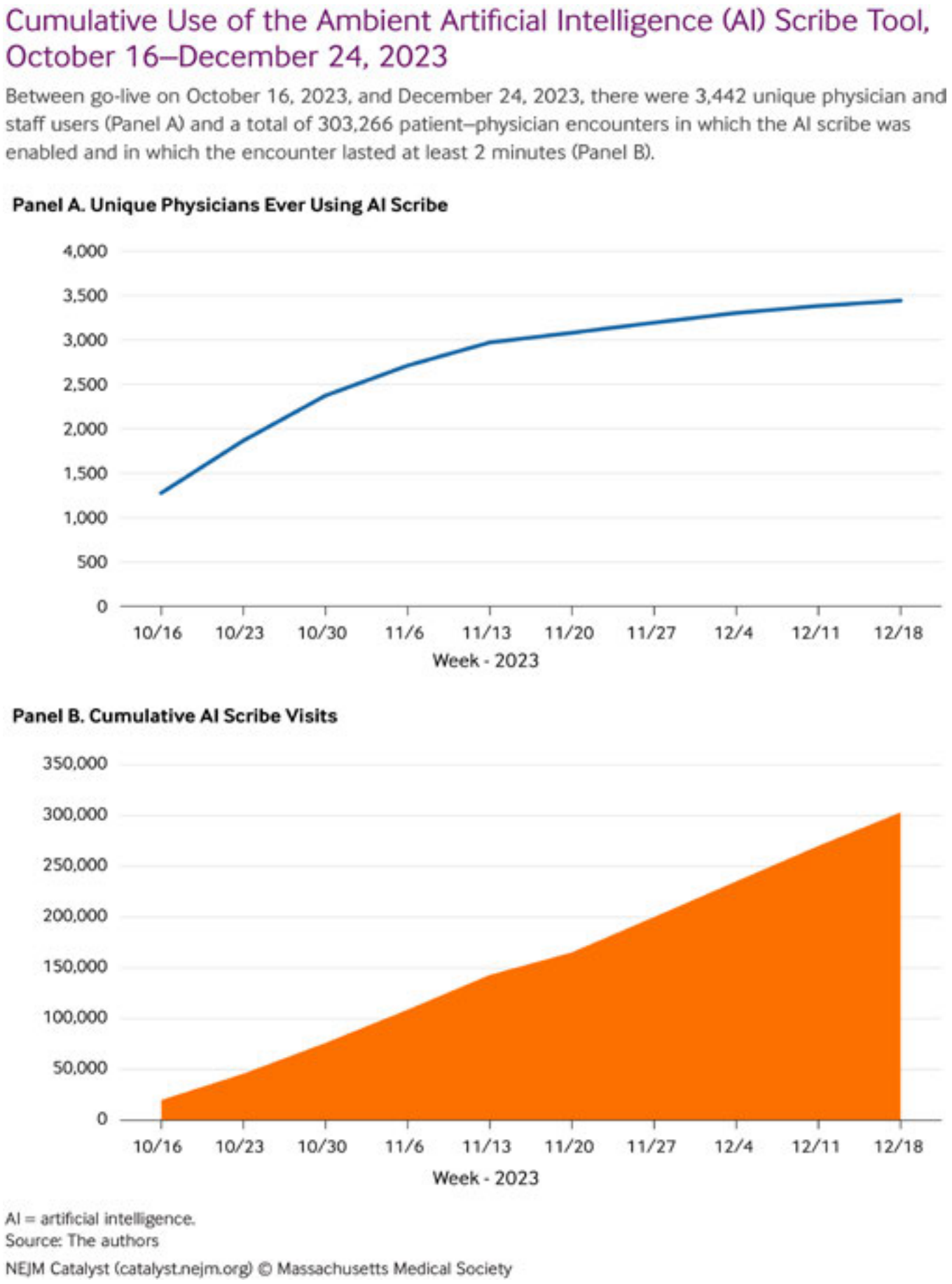

Source: Tierney et al., 2024
Figure 5.4.8

Researchers at Stanford conducted a two-part study on the use of ambient AI scribe technology, building on prior work by testing a fully integrated, automated AI scribe system. The study demonstrated improvements in both objective measures, such as documentation time, and subjective measures of physician experience. Adoption was strong, with an average uptake of 55% among physicians. The AI scribe provided notable efficiency gains, saving physicians approximately 30 seconds per note and reducing overall EHR time by about 20 minutes per day (Figure 5.4.9). Additionally, physicians reported significant reductions in burden and burnout, with average decreases of 35% and 26%, respectively. These findings suggest that AI-powered scribe technology can meaningfully improve physician workflow and well-being, offering both time savings and relief from administrative strain.

**Impact of AI Scribe on physician EHR usage**
Source: Ma et al., 2024 | Chart: 2025 AI Index report

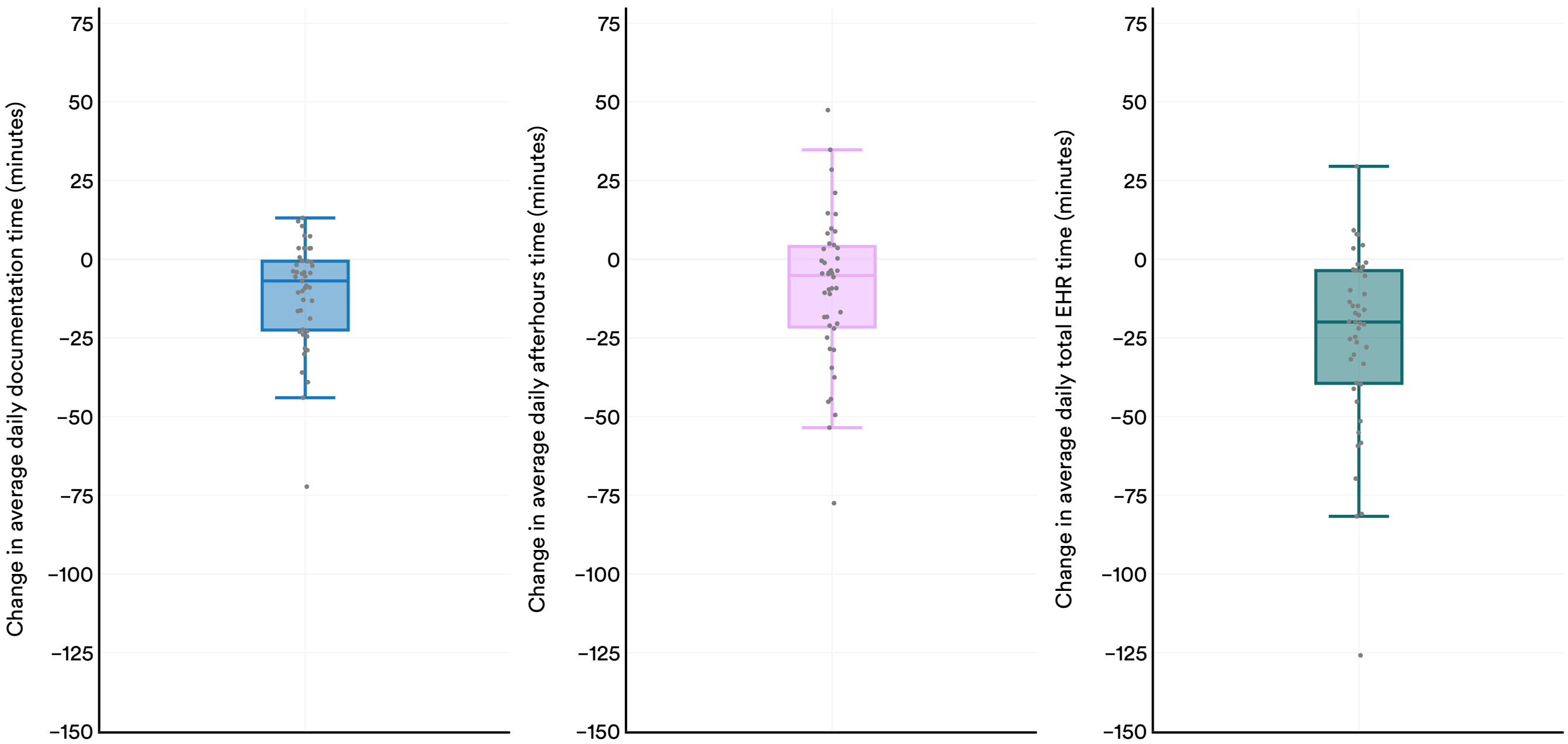

Figure 5.4.9

Investment in ambient scribe technology is reported to reach almost $300 million in 2024. While clinical documentation has been the starting point for the technology and the evaluations performed to date, optimists envision ubiquitous ambient listening technology in both outpatient and inpatient settings that will eventually support order placement, billing and coding, and real-time clinical decision support.

# Deployment, Implementation, Deimplementation

## FDA Authorization of AI-Enabled Medical Devices

The deployment of AI in clinical settings has grown exponentially over the past decade, highlighted by the dramatic increase in the number of AI-enabled medical devices authorized by the U.S. Food and Drug Administration (FDA).

The FDA authorized its first AI-enabled medical device in 1995. For the next two decades, annual approvals remained in the single digits. In 2015 alone, six AI medical devices were approved. Since then, the number of yearly approvals has surged, peaking at 223 in 2023 (Figure 5.4.10).

**Number of AI medical devices approved by the FDA, 1995–2023**
Source: FDA, 2024 | Chart: 2025 AI Index report

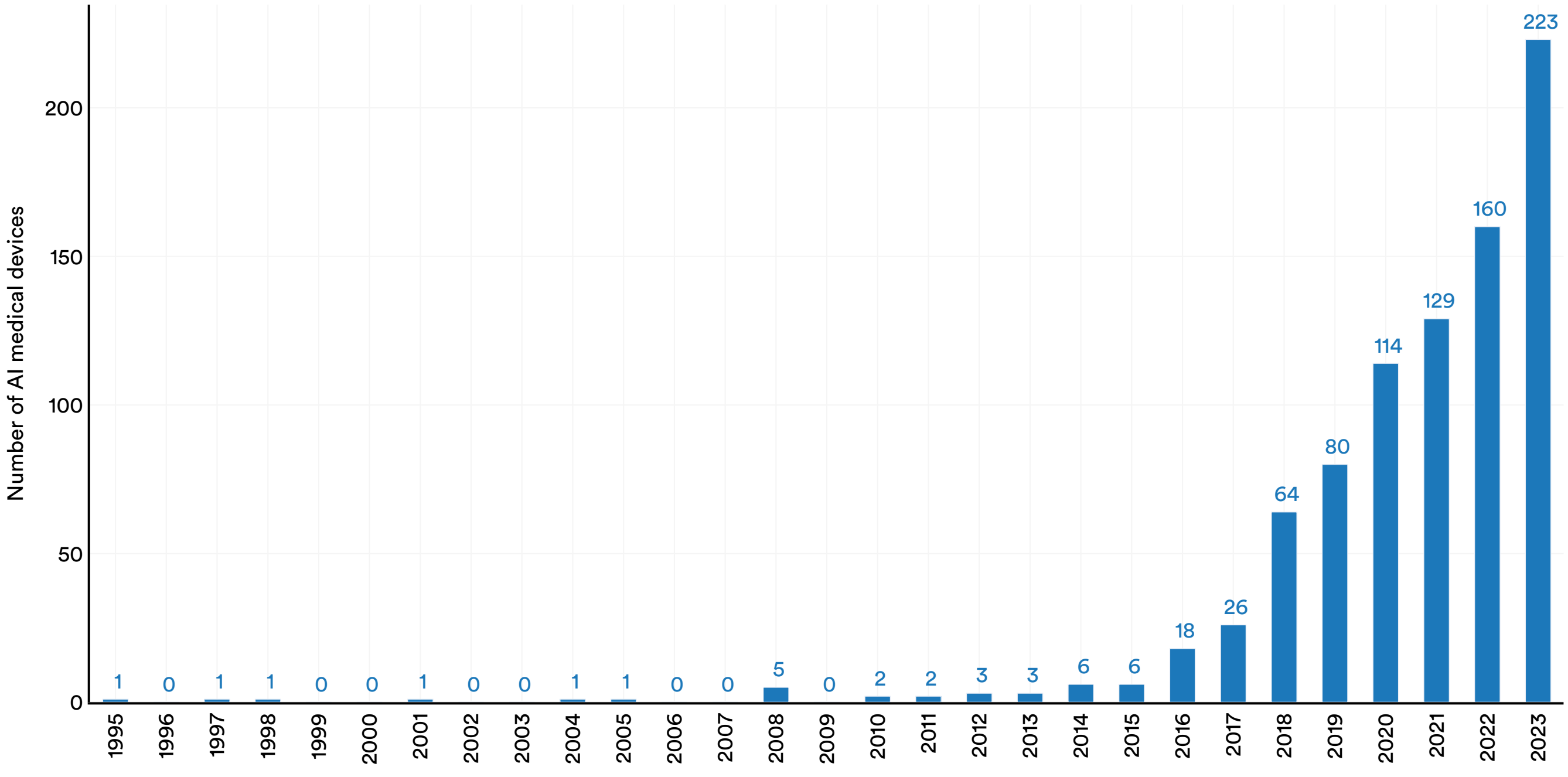

Figure 5.4.10

## Successful Use Cases: Stanford Health Care

In practice, transitioning AI models into real-world use requires a framework that ensures fairness, utility, and reliability. Stanford Health Care has led the way by evaluating and implementing AI tools using its FURM (Fair, Useful, Reliable, Measurable) framework. Among the six AI use cases assessed, two have been successfully implemented: (1) screening for peripheral arterial disease (PAD) and (2) improving documentation and coding for inpatient care. This section details screening for peripheral arterial disease.

### Screening for Peripheral Arterial Disease

Peripheral arterial disease (PAD) is a chronic vascular condition that often goes undiagnosed in its early stages, leading to severe complications such as critical limb ischemia and amputation. To improve early detection and intervention, Stanford Health Care developed and implemented an AI-enabled PAD classification model designed to enhance screening and optimize patient care.

The primary goal of the PAD screening tool is to facilitate earlier diagnosis in primary care populations, allowing for medical or surgical intervention before the disease leads to severe complications. By identifying high-risk patients, the model also helps optimize resource allocation, ensuring that those most in need receive immediate follow-up and care.

To integrate seamlessly into clinical workflows, the AI tool was designed to automatically assess PAD risk and flag high-risk individuals for further evaluation. If the condition is confirmed, the patient is referred for a vascular consultation. Figure 5.4.11 illustrates the proposed model and workflow details for integrating PAD screening into clinical workflows, including risk assessment, referrals, and patient follow-up.

**Proposed model and workflow for integrating PAD screening into clinical practice**
Source: Callahan et al., 2024

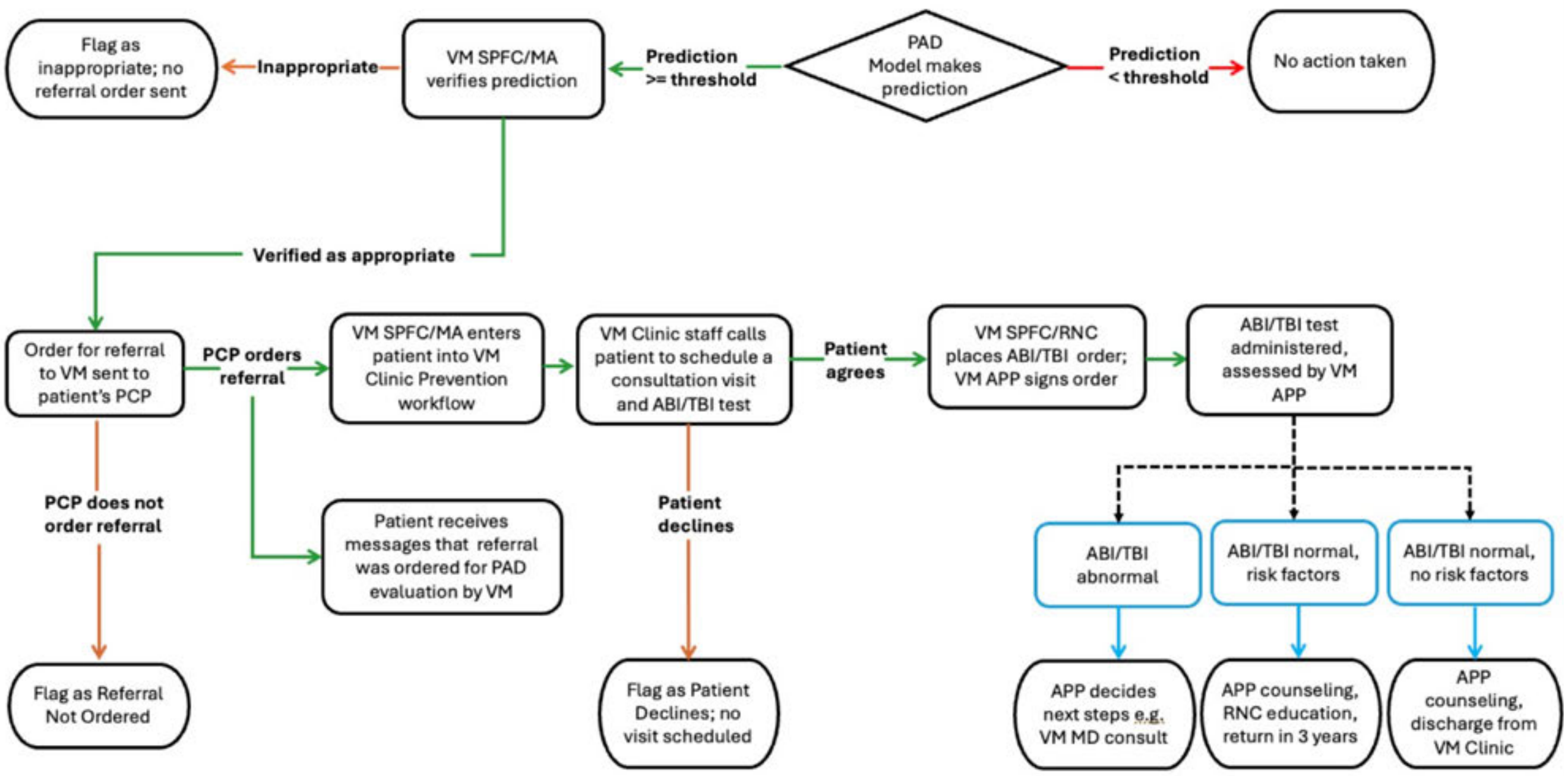

**ABI/TBI** – ankle/toe brachial index; APP – advanced practice provider; **MA** – medical assistant; **RNC** – registered nurse coordinator; **SPFC** – specialty patient flow coordinator; **VM** – vascular medicine

Figure 5.4.11

Following a successful pilot phase, the PAD screening tool advanced to Stage 2 and was fully implemented at Stanford Health Care. The model is expected to impact approximately 1,400 patients annually. Beyond its clinical benefits, the program has demonstrated financial sustainability, operating independently without external funding. By increasing early PAD detection, reducing the likelihood of severe complications, and improving patient outcomes, this AI-driven approach is reshaping the standard of care for PAD management.

## Social Determinants of Health

The integration of LLMs and AI-based clinical decision support (CDS) systems is transforming medicine, though adoption varies by specialty. While some embrace LLMs, others remain cautious. This review explores research and innovations, emphasizing the role of a strong evidence base. A key aspect is addressing social determinants of health (SDoH), such as socioeconomic status and environment. In 2024, AI advancements targeted SDoH, improving patient care and health equity.

### Extracting SDoH From EHR and Clinical Notes

Fine-tuned multilabel classifiers (Flan-T5 XL) outperformed ChatGPT-family models in identifying SDoH in clinical notes and were less sensitive to demographic descriptors. They also exhibited lower bias, with reduced discrepancies when race, ethnicity, or gender was introduced. Figure 5.4.12 illustrates the performance of various models on SDoH identification tasks in a radiotherapy test set. Newer, larger models like Flan-T5-XXL, augmented with synthetic and gold data (SDoH-labeled sentences), showed superior performance. As models have scaled and incorporated more data over time, their ability to identify SDoH has improved.

**Model performance on in-domain RT test dataset (any SDoH)**
Source: RAISE Health, 2025 | Chart: 2025 AI Index report

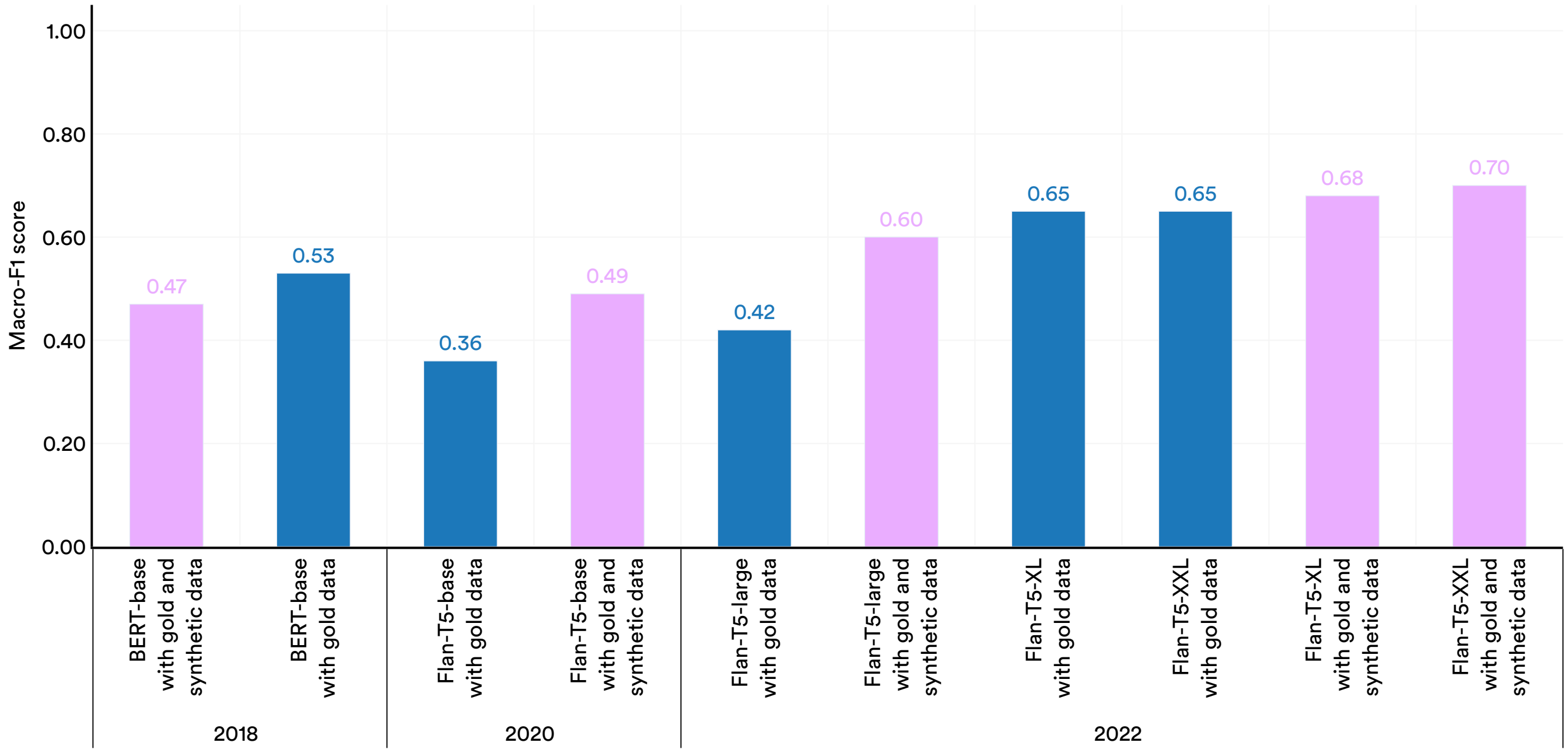

Figure 5.4.12

Extracting SDoH from EHRs helps healthcare providers address social needs like housing instability or food insecurity. These findings highlight LLMs' potential to enhance SDoH documentation, resource allocation, and health equity while emphasizing the need for bias mitigation and robust synthetic data methods.

### AI Adoption Across Medical Fields and the Integration of SDoH

Figure 5.4.13 highlights various medical fields and illustrates how AI integrates social determinants of health in each.

| Field | Recent research | Description of integration |
|---|---|---|
| Oncology | Istasy et al., 2024 | In cancer care, AI-driven tools have been developed to consider SDoH in treatment planning. By incorporating factors such as a patient's access to care and support systems, these tools assist oncologists in creating personalized treatment plans that are both effective and feasible for patients. |
| Cardiology | Snowdon et al., 2023<br>Quer et al., 2024 | AI models in cardiology have been enhanced to include SDoH, improving the accuracy of risk assessments for conditions like hypertension and heart failure. This inclusion allows for more comprehensive patient evaluations and tailored management strategies. |
| Psychiatry | Stade et al., 2024 | LLMs have been applied to analyze community-level SDoH data, aiding in the allocation of mental health resources. By identifying areas with high social risk factors, healthcare systems can prioritize interventions and support services in communities with the greatest need. |

Figure 5.4.13

# Synthetic Data

Synthetic data is revolutionizing healthcare by enhancing privacy-preserving analytics, clinical modeling, and AI training. It optimizes workflows, simulates rare cases, and supports AI-driven innovations. However, scalability concerns, as noted in the first chapter of this year's AI Index, call for cautious adoption.

### Clinical Risk Prediction

A recent study validated synthetic data for privacy-preserving clinical risk prediction. Using ADSGAN, PATEGAN, and DPGAN, researchers modeled lung cancer risk in ever-smokers from the UK Biobank.[5] The figure below compares PCA eigenvalues, showing how ADSGAN and PATEGAN closely match real data distributions, enabling reliable clustering and feature selection (Figure 5.4.14). These findings demonstrate that synthetic datasets can preserve statistical fidelity, support exploratory analysis, and develop predictive models without real and identifiable patient data.

**Principal component analysis**
Source: Qian et al., 2024

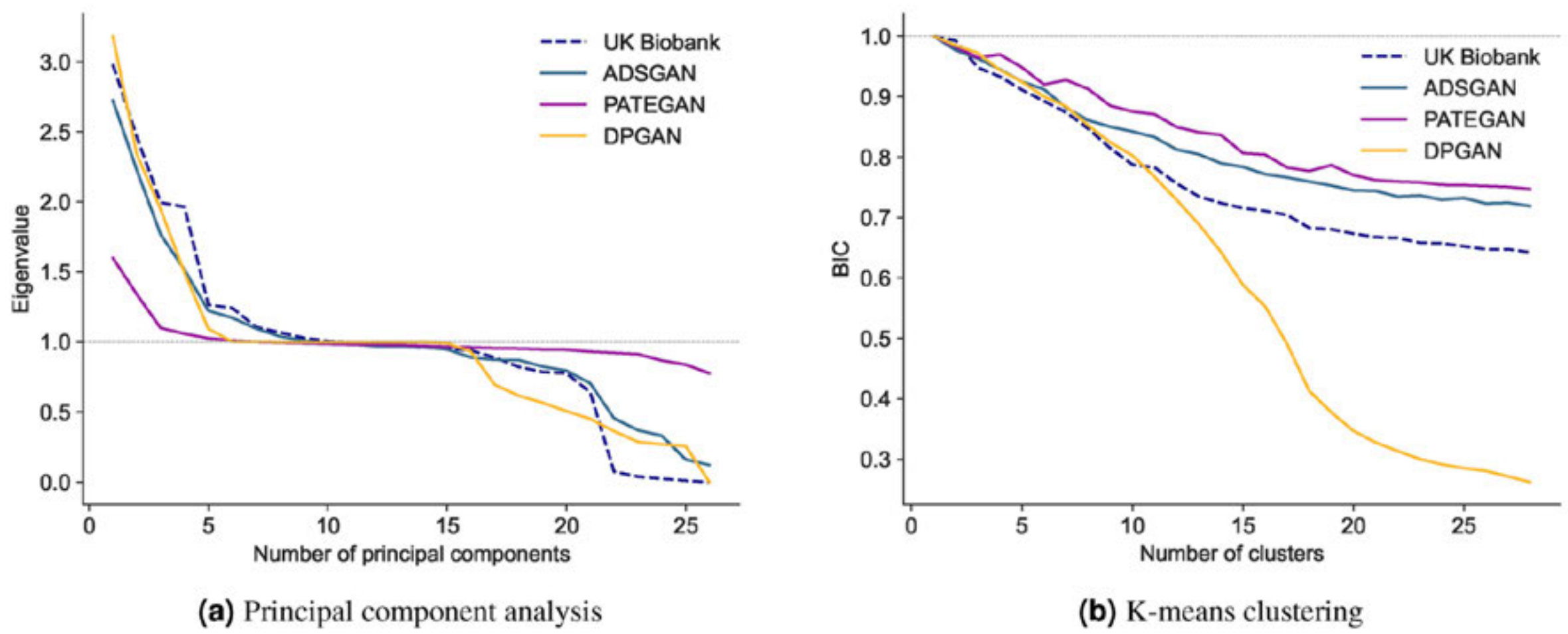

Figure 5.4.14

5 An ever-smoker is someone who has smoked at least 100 cigarettes in their lifetime.

### Drug Discovery

A recent Nature study introduced a generative AI approach for in silico formulation optimization and particle engineering in drug development. Using an image generator guided by critical quality attributes, it creates digital formulations for analysis without extensive physical testing. The study validated this method by predicting the percolation threshold of microcrystalline cellulose (MCC) in oral tablets. Figure 5.4.15 compares the tortuosity calculations of real tablet volumes (green squares) with AI-synthesized volumes (red circles).[6] Their close alignment suggests that synthetic data holds promise for modeling drug properties and improving AI-driven drug discovery.

**Percolation threshold prediction and validation based on AI-generated synthetic structures**
Source: Hornick et al., 2024

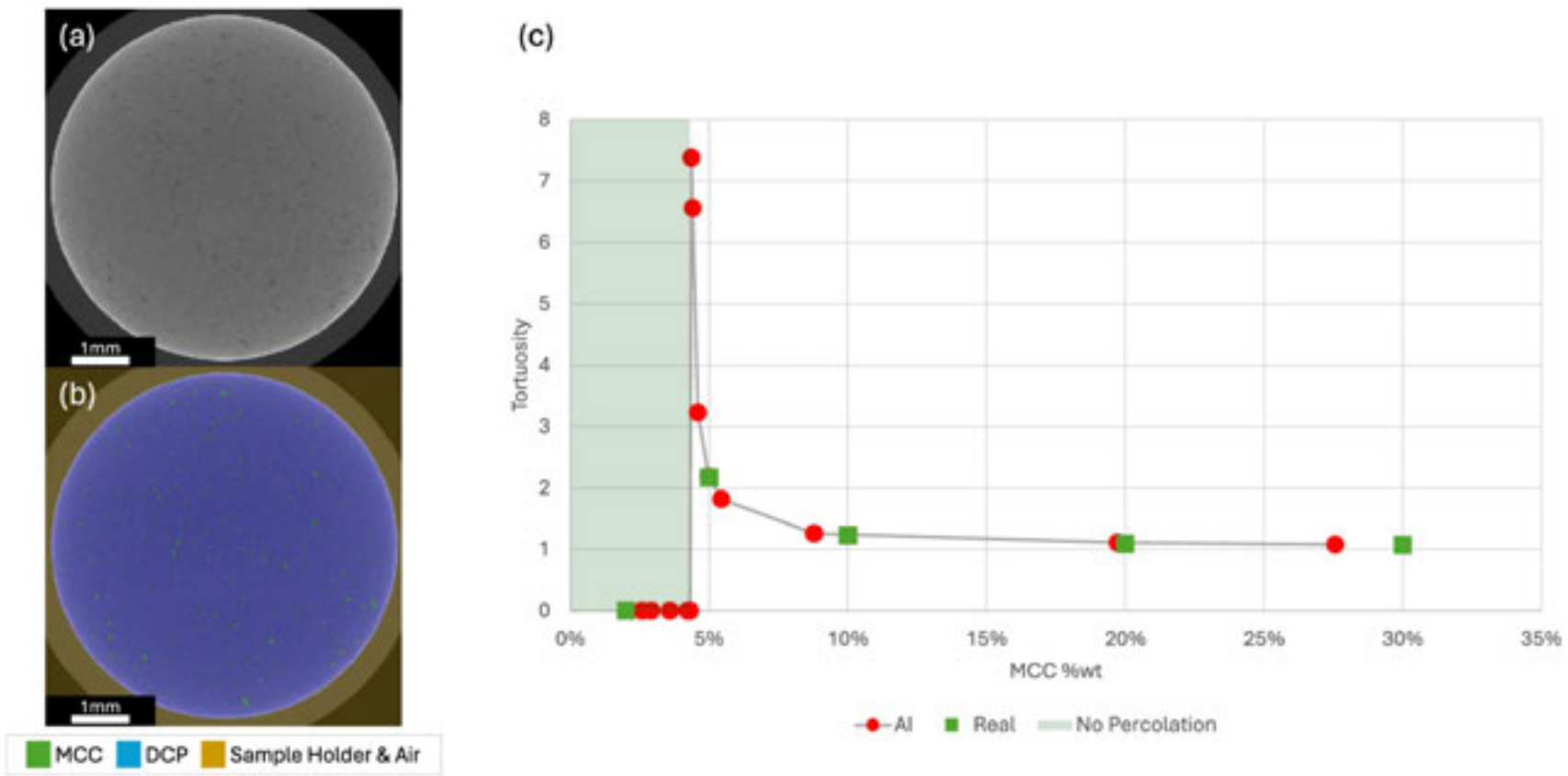

Figure 5.4.15

### Data Generation Platforms

Platforms are necessary to demonstrate, standardize, and automate the creation of synthetic data. Recently published research has demonstrated that large-scale synthetic data generation and validation is not only feasible but also capable of significantly enhancing AI applications in medicine with their synthetic tabular neural generator (STNG) framework. Figure 5.4.16 compares the area-under-the-curve values for real and synthetic heart disease datasets to evaluate the effectiveness of different synthetic data generation methods. In many cases, there is a fairly close overlap between the real datasets and the synthetic datasets, showing the ability of synthetic data to model complex health conditions closely. Advancements in synthetic data generation methodologies can improve data fidelity while minimizing privacy risks.

**Areas under the curve for evaluating synthetic heart disease datasets**
Source: Rashidi et al., 2024

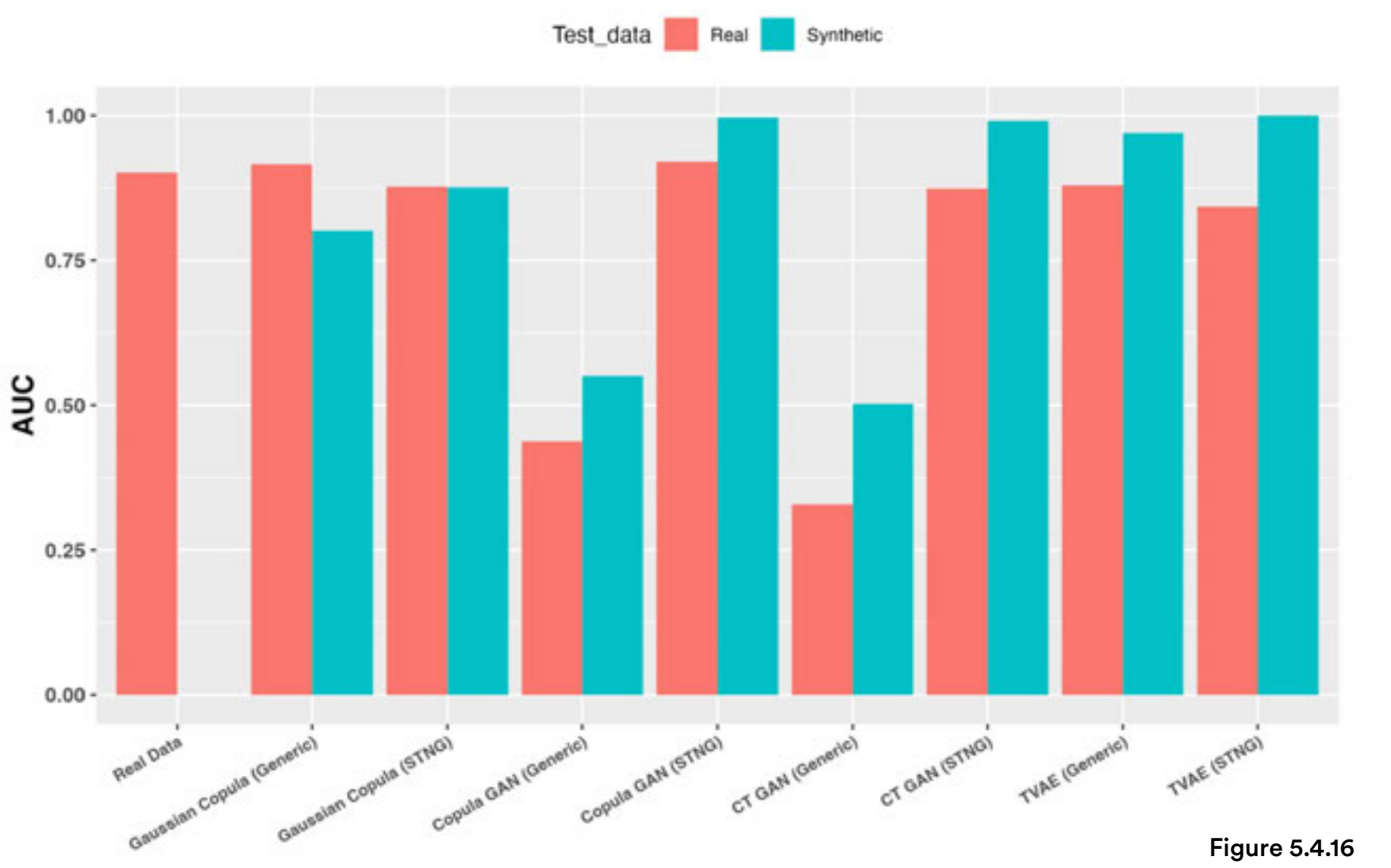

Figure 5.4.16

6 Tortuosity is a measure of how convoluted or twisted a path is compared to the shortest possible straight-line distance between two points.

# Electronic Health Record System

AI integration in electronic health records (EHRs) can ease healthcare burdens by streamlining administration, enhancing clinical decision support, and improving patient care. With major vendors—Epic, Oracle Health (formerly Cerner), Meditech, and TruBridge (formerly CPSI)—dominating the market, their AI tools can be widely adopted within their networks. As of 2021, EHR adoption had approached 90% for any system and 80% for certified EHR systems.

A 2023 American Hospital Association IT survey found that most hospitals using ML or predictive models in their EHRs relied on a dominant vendor for inpatient care (Figure 5.4.17). Adoption was highest with Epic, Cerner, and Meditech. While Epic, Cerner, and CPSI hospitals primarily used vendor-developed models, Meditech and others more often adopted third-party or in-house solutions (Figure 5.4.18).

**Predictive model use across primary inpatient EHR vendor**
Source: AHA survey, 2024 | Chart: 2025 AI Index report

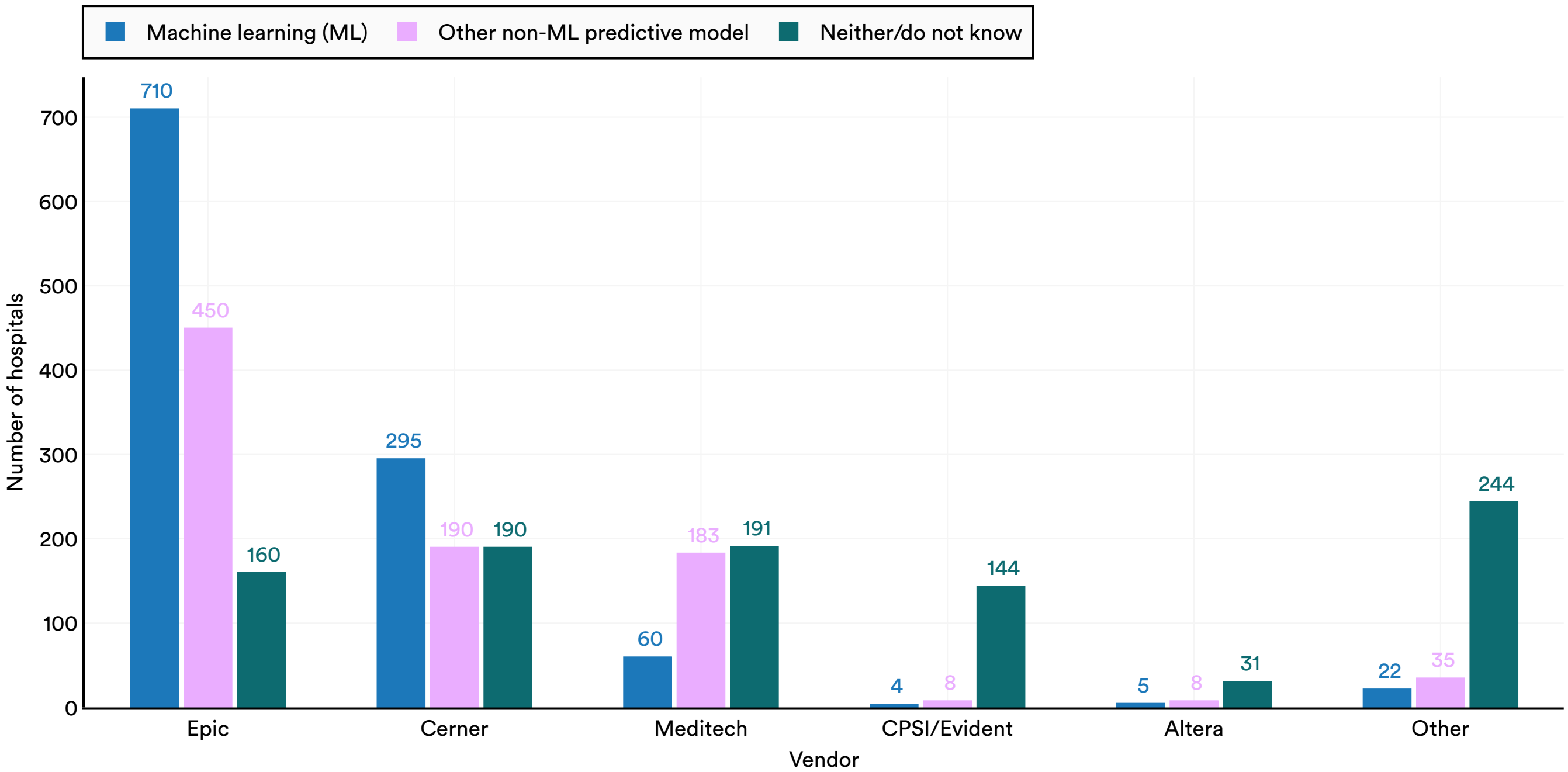

Figure 5.4.17

### Developer of predictive models across EHR vendor

Source: AHA survey, 2024 | Chart: 2025 AI Index report

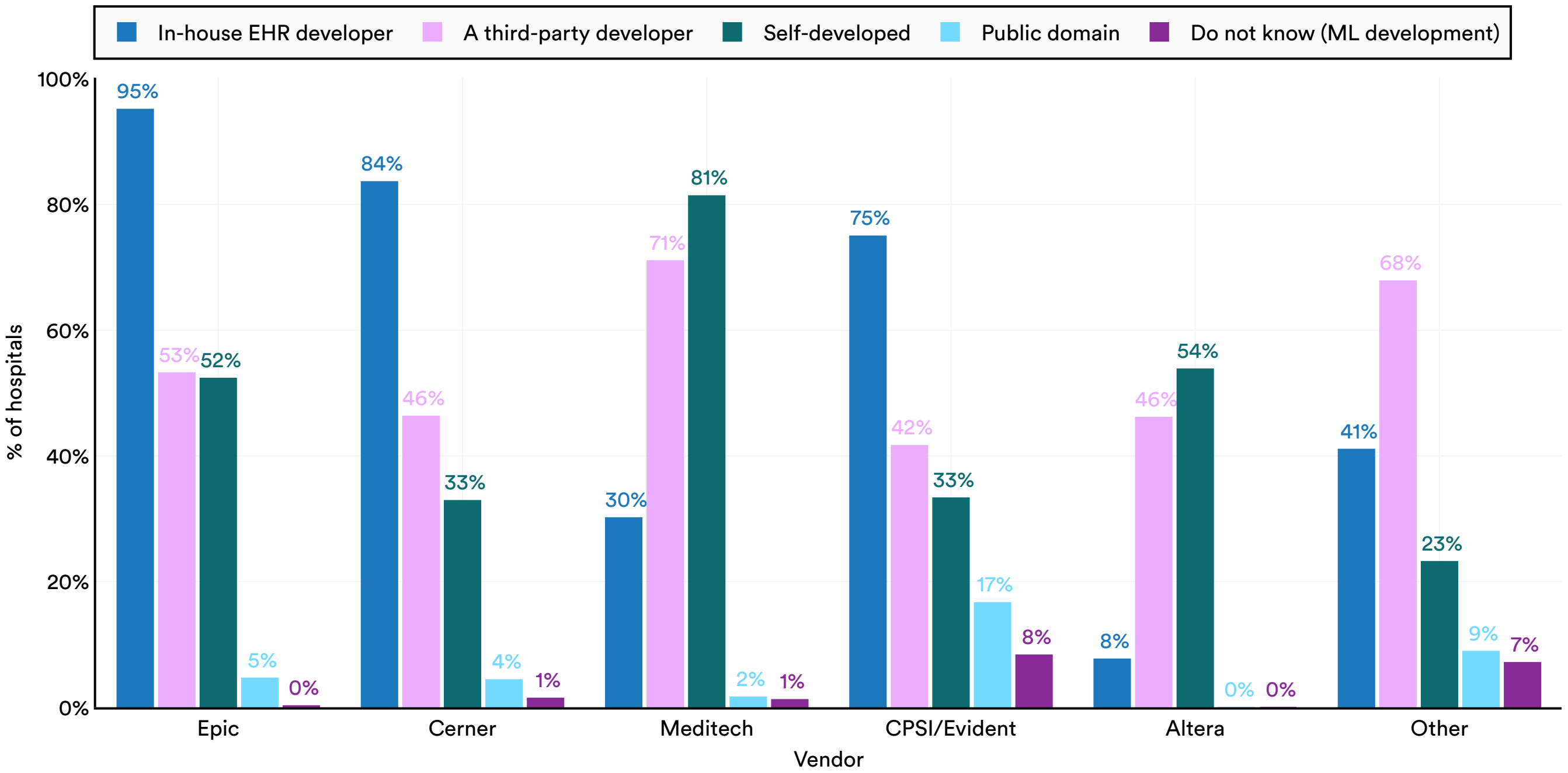

Figure 5.4.18

AI integration into EHRs could streamline clinical workflows and enhance provider and patient experiences. However, it remains unclear whether AI-enabled health IT will benefit underserved communities, which often struggle with technological adoption. Rural areas, for example, face barriers like limited broadband access, weak healthcare IT infrastructure, and EHR functionality constraints—key enablers of AI-driven healthcare. Additionally, it is important to assess whether AI tools are equitably developed for both basic and comprehensive EHR systems, as many resource-limited settings still rely on the former.

## Clinical Decision Support

AI has transformed how healthcare providers diagnose, predict, and manage diseases with an increasing focus on rigorous evaluation of AI-based systems in clinical trials. The evolution of AI in clinical decision support (CDS) reflects a shift from reactive interventions—e.g., during the COVID-19 pandemic—to proactive, data-driven clinical decision-making with clinical trials increasing over the years. The number of clinical trials that have included mentions of artificial intelligence is steadily rising (Figure 5.4.19).

**Number of clinical trials that have included mentions of AI, 2014–24**
Source: RAISE Health, 2025 | Chart: 2025 AI Index report

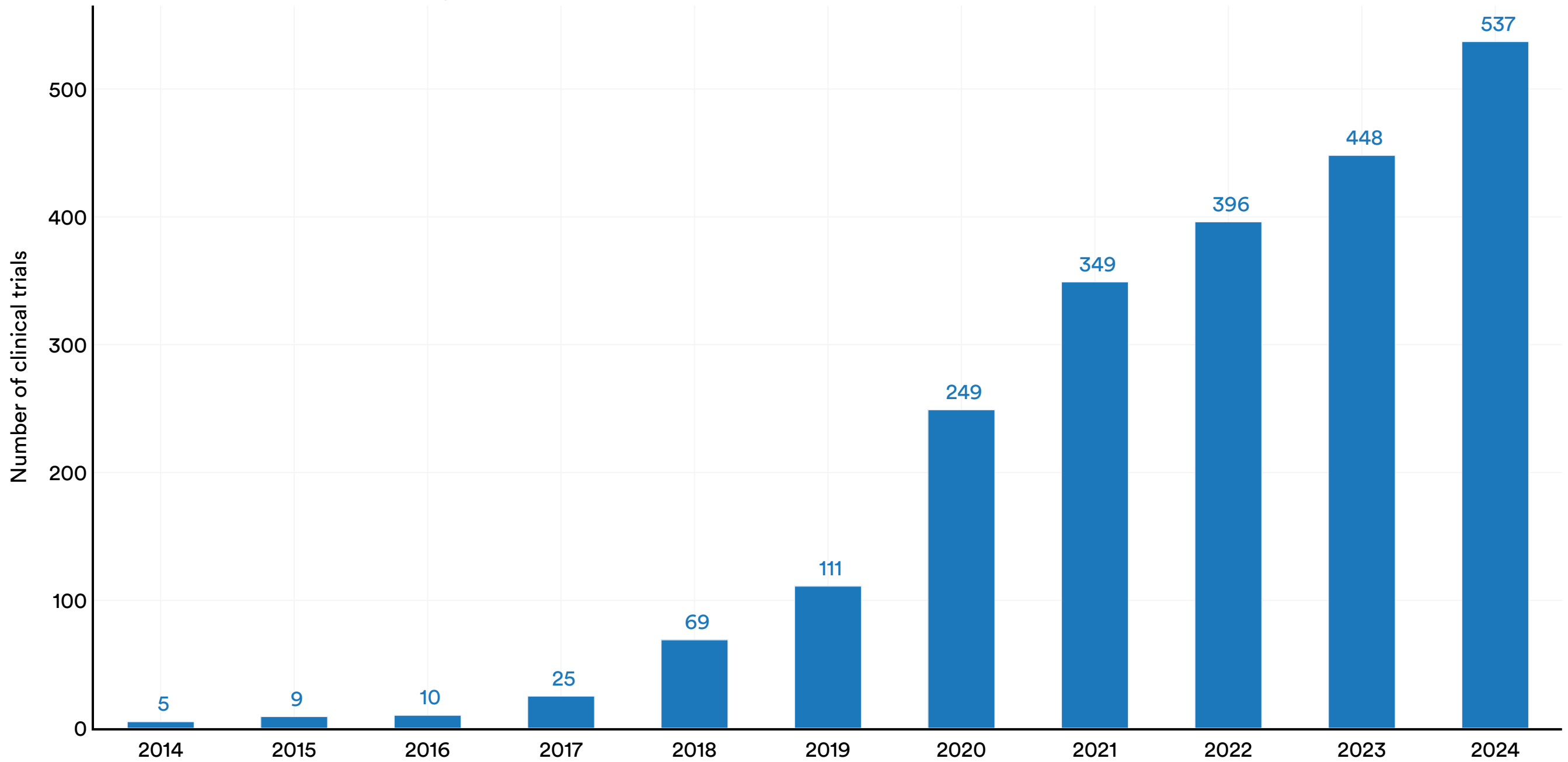

Figure 5.4.19

The COVID-19 pandemic accelerated AI adoption in triage, resource allocation, and outcome prediction, showcasing the technology's potential in real-time CDS. Post-pandemic, AI expanded beyond emergency response to managing chronic disease, optimizing procedures, and streamlining workflows. Trials like the CERTAIN Study demonstrated how AI-driven real-time procedural support could improve diagnostic accuracy in gastrointestinal procedures. By 2023, AI in CDS extended to medication safety and workflow optimization, as seen in Preventing Medication Dispensing Errors in Pharmacy Practice, which used AI to detect real-time medication errors. Globally, AI-driven clinical trials have sharply risen, with China (105 trials), the U.S. (97), and Italy (42) leading in 2024 (Figure 5.4.20).

**Number of clinical trials that have included mentions of AI by select geographic areas, 2021–24**

Source: RAISE Health, 2025 | Chart: 2025 AI Index report

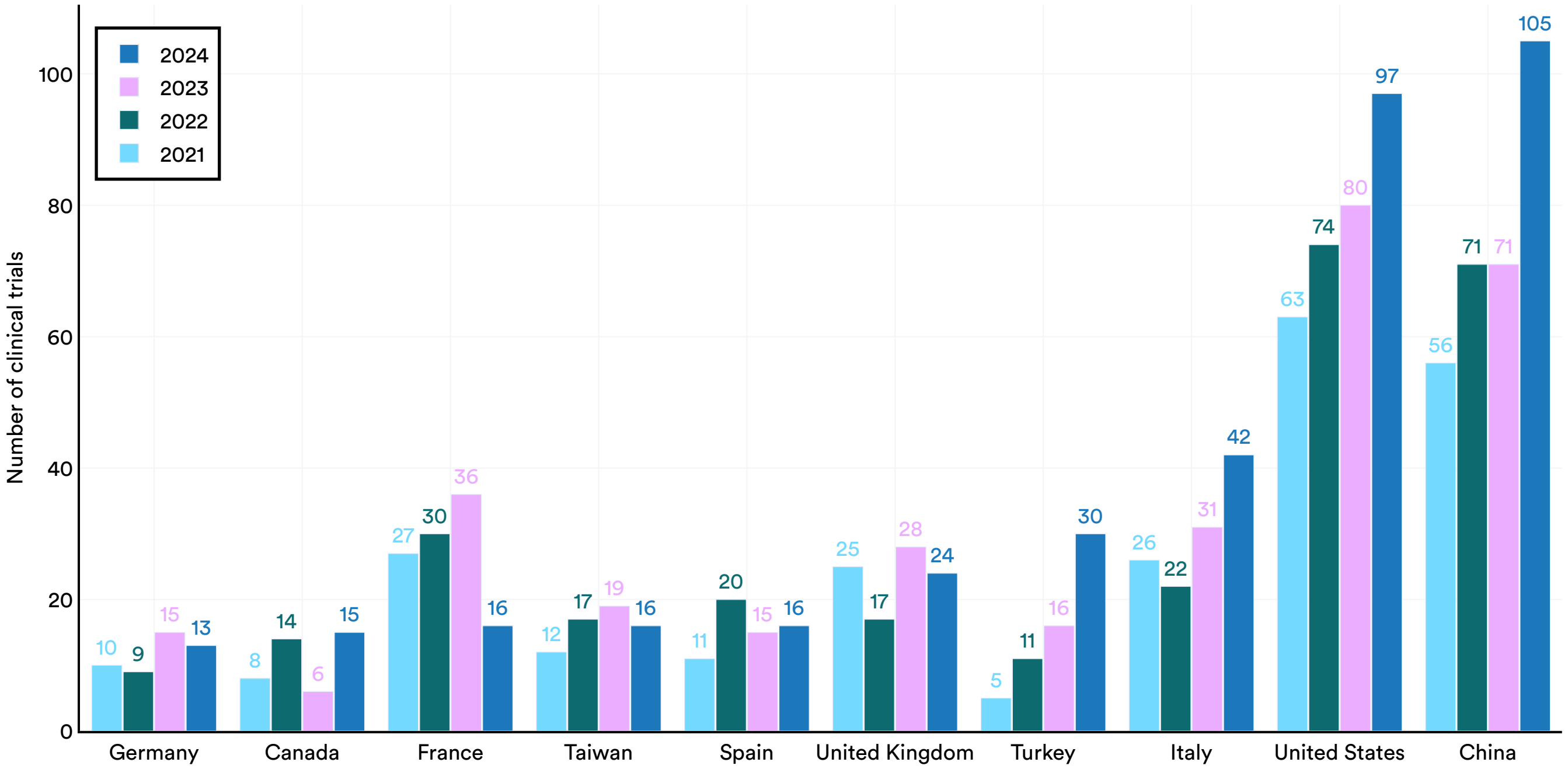

Figure 5.4.20

The increasing integration of AI in medical research and clinical care as discussed in previous sections brings both promises and challenges. AI systems lean heavily on large amounts of data for training. The collection, use, and sharing of this data—especially in high-stakes domains such as healthcare—can raise various ethical concerns.

# 5.5 Ethical Considerations

## Meta Review

For this section, the AI Index conducted a meta review of thousands of medical ethics studies to glean insights on the state of the field. The team's methodology is highlighted in Figure 5.5.1.

Attention to the ethical issues in medical AI has increased in each of the past five years. The number of publications related to ethics and medical AI increased fourfold from 2020 to 2024 (Figure 5.5.2).

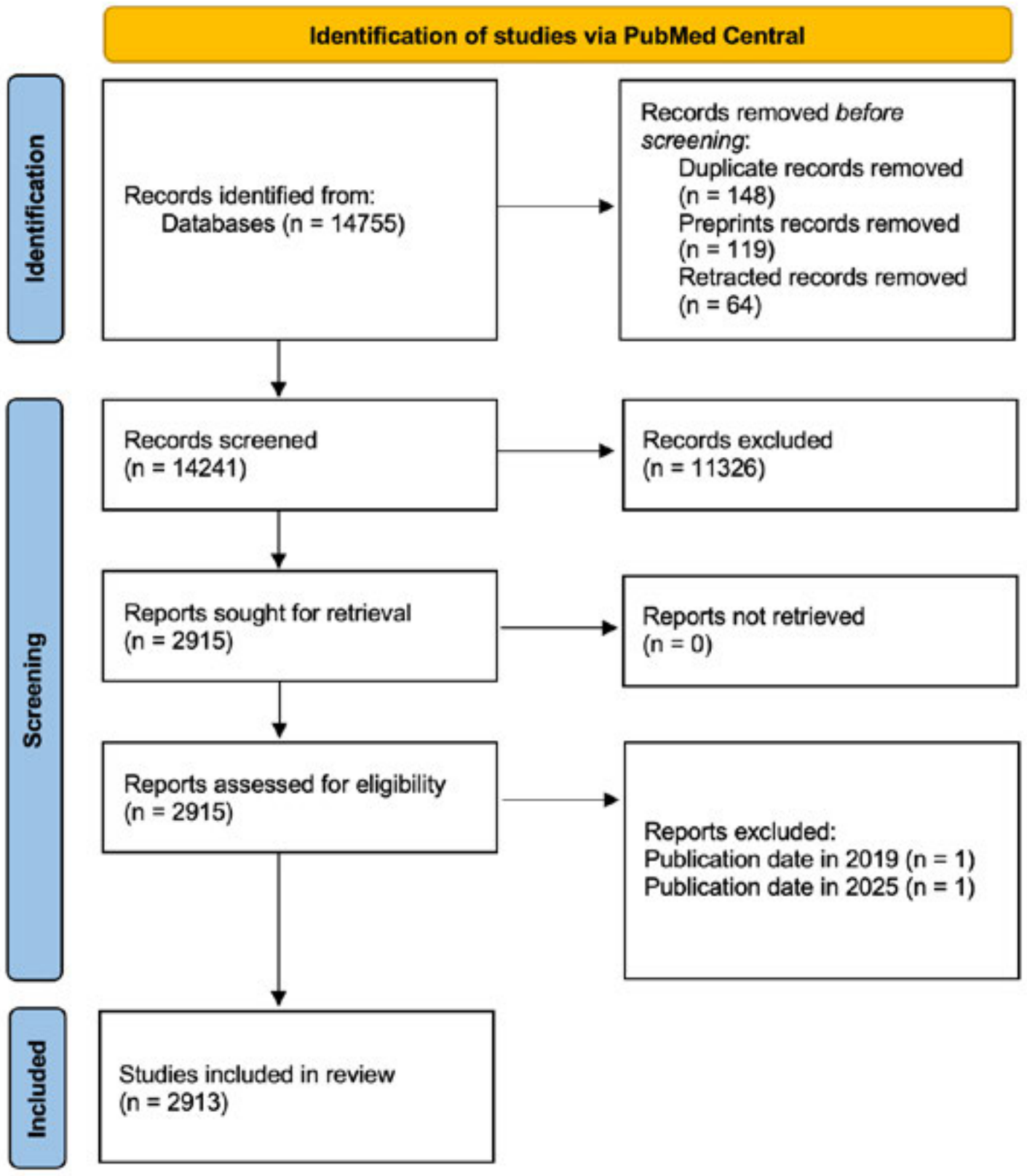

Figure 5.5.1

### Number of medical AI ethics publications, 2020–24

Source: RAISE Health, 2025 | Chart: 2025 AI Index report

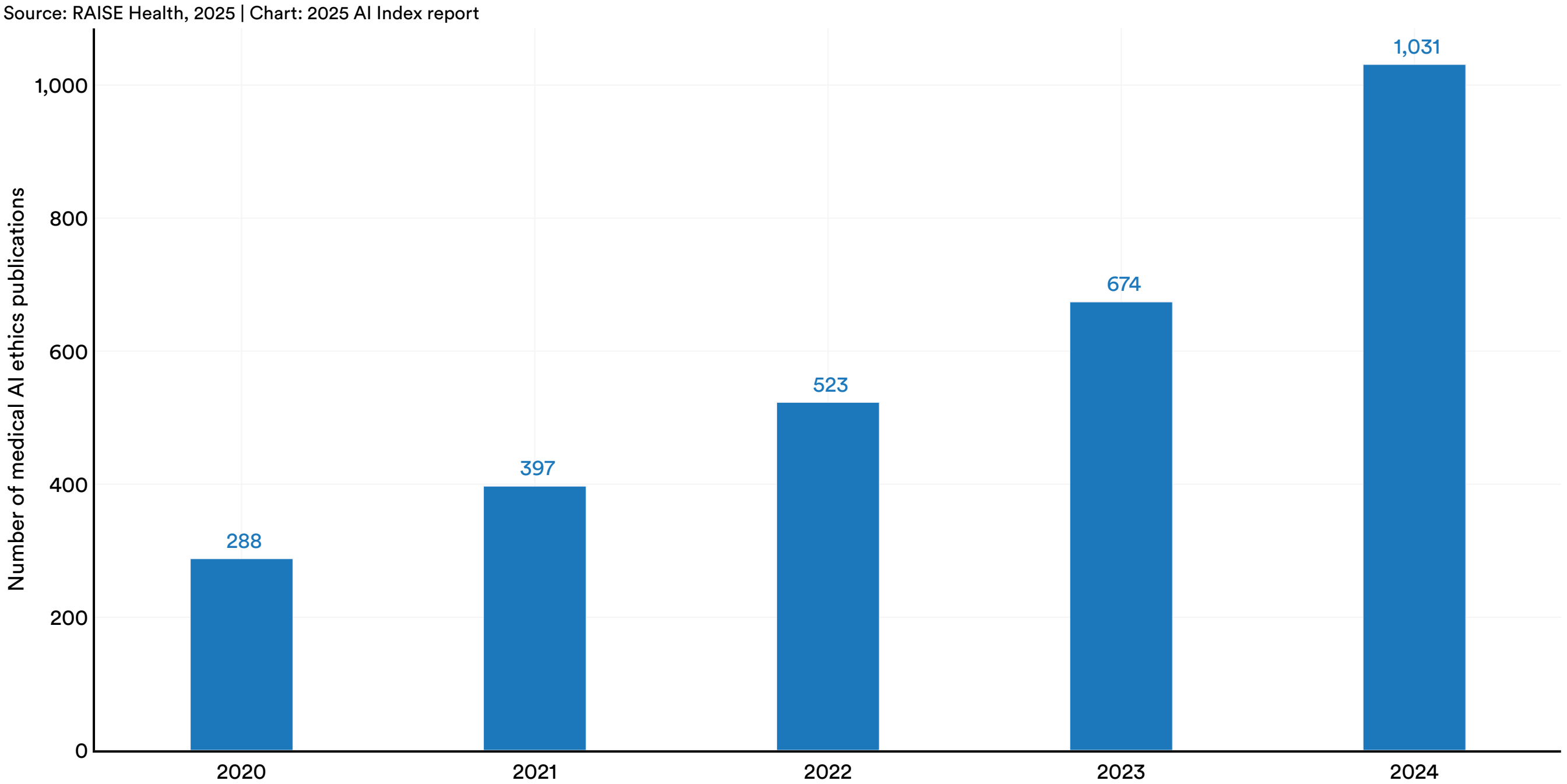

Figure 5.5.2

The focus of AI applications in medical ethics literature has evolved over time. Figure 5.5.3 illustrates the ethical issues discussed in AI medical papers from 2020 to 2024. In 2024, bias and privacy were the most frequently cited concerns, followed by equity. In contrast, privacy was a more prominent topic than bias in 2020, but this trend has since shifted.

### Top 10 ethical concerns discussed in medical AI ethics publications, 2020–24

Source: RAISE Health, 2025 | Chart: 2025 AI Index report

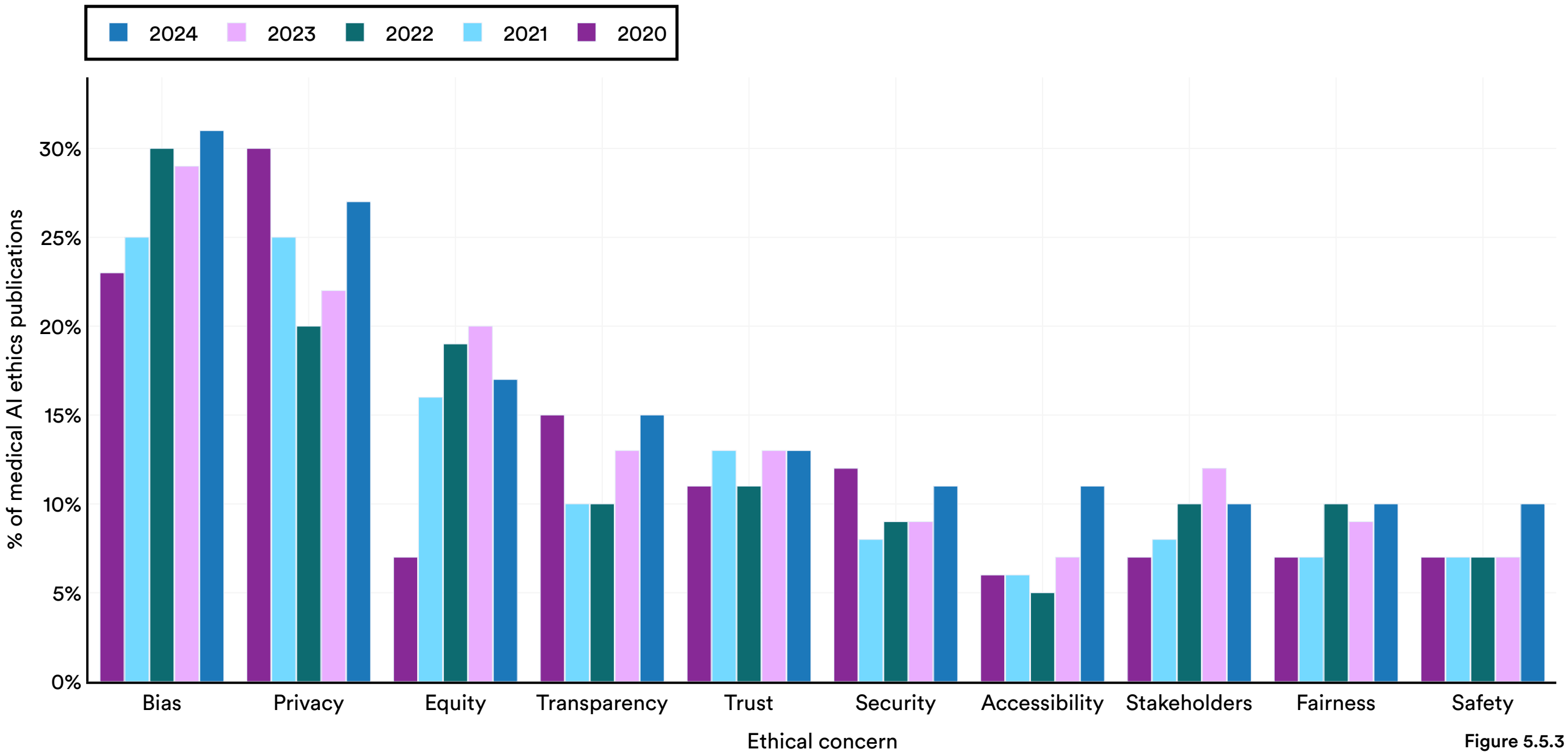

Figure 5.5.3

In terms of AI tools, much attention has been paid in medical ethics literature to OpenAI's GPT series (e.g., ChatGPT) (Figure 5.5.4). This reflects an expanding interest in large-language models over the past few years.

### AI tools discussed in medical AI ethics publications, 2020–24

Source: RAISE Health, 2025 | Chart: 2025 AI Index report

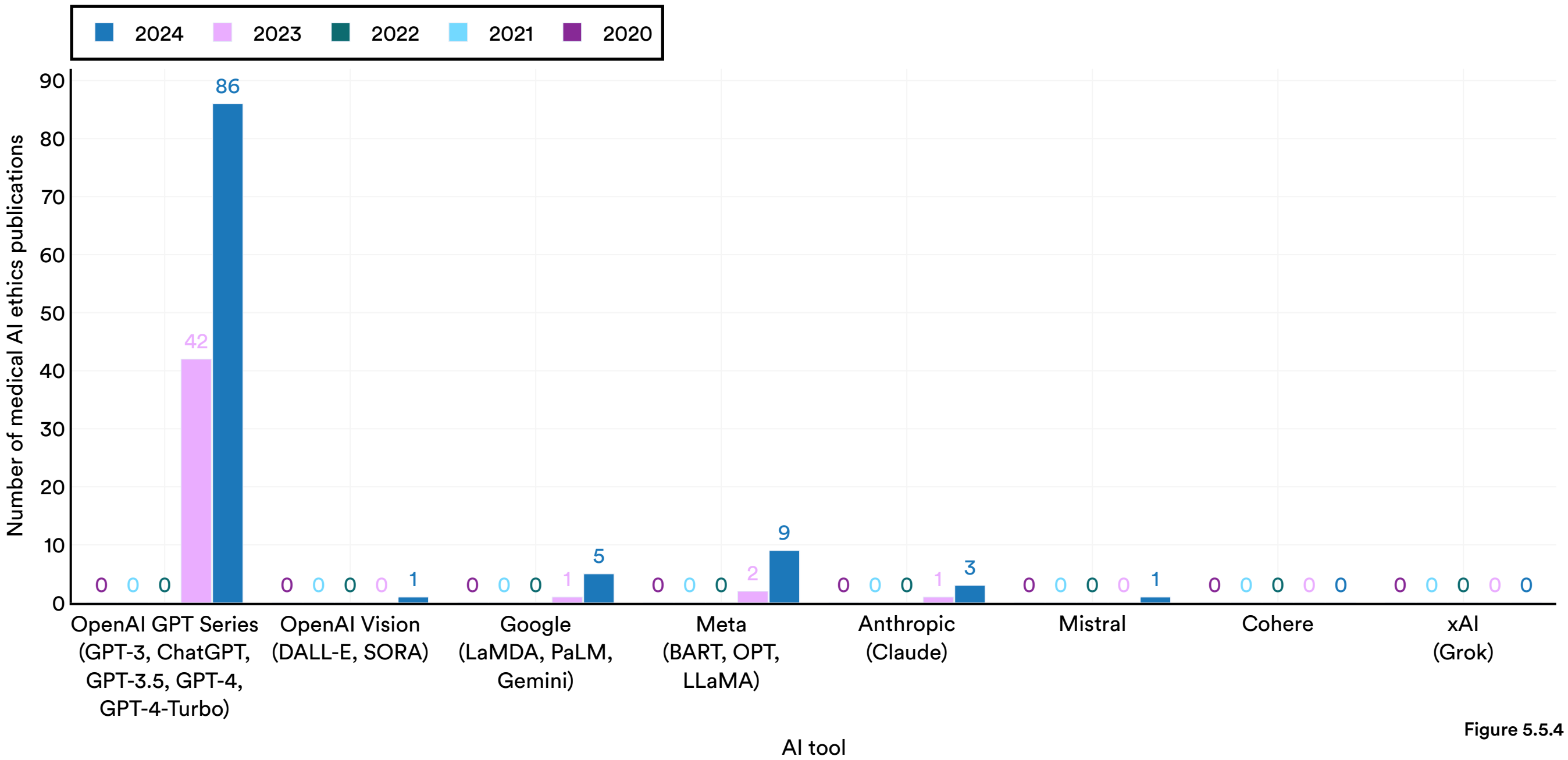

Figure 5.5.4

Figure 5.5.5 and Figure 5.5.6 show the number and total funding of NIH grants for medical AI ethics projects by fiscal year. The number of grants skyrocketed from 25 in 2023 to 337 in 2024 (Figure 5.5.5). Similarly, total funding soared from $16 million in 2023 to $276 million in 2024—an almost 17-fold increase in just one year.

**Number of NIH grants for medical AI ethics by fiscal year, 2020–24**

Source: RAISE Health, 2025 | Chart: 2025 AI Index report

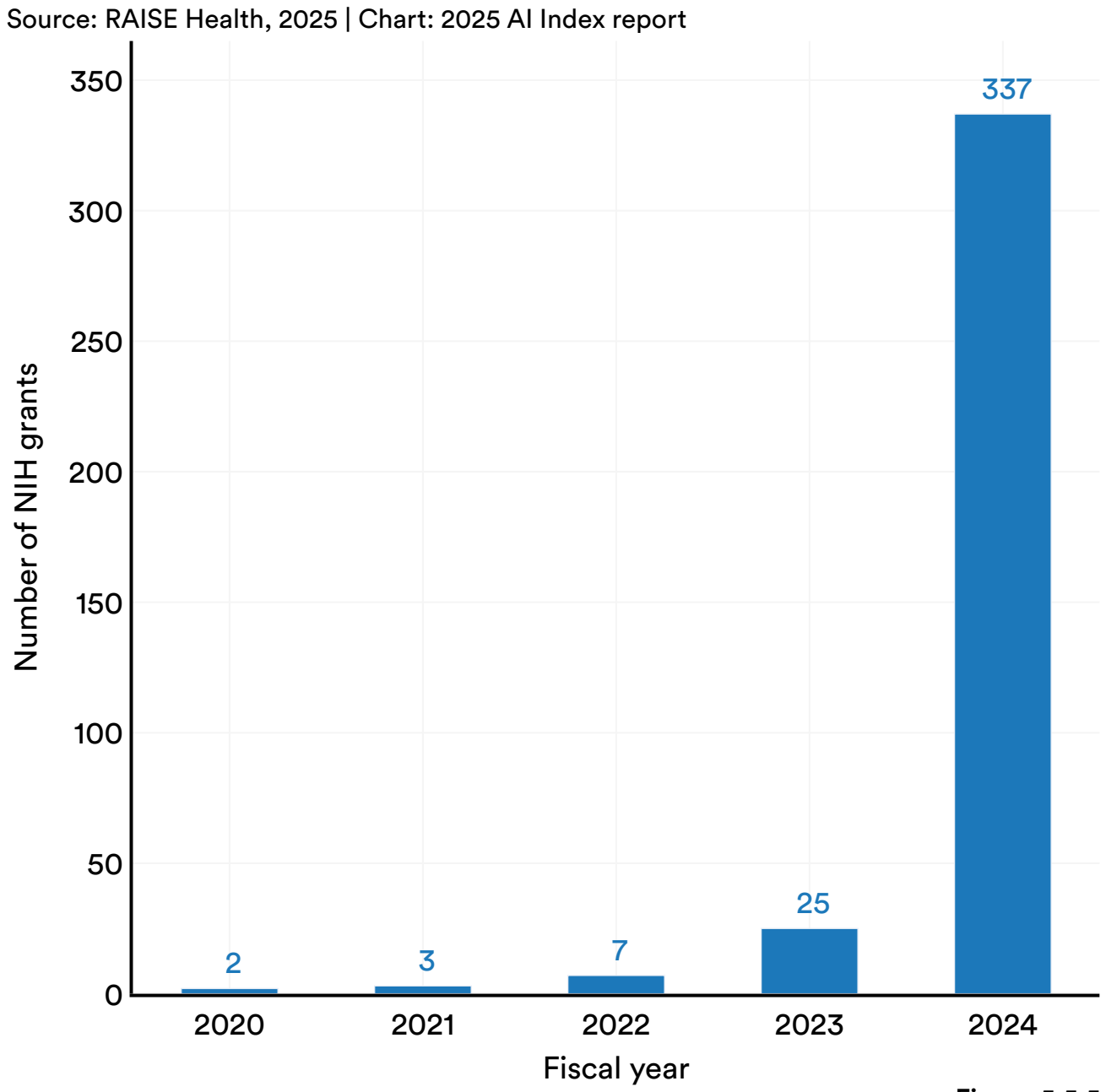

Figure 5.5.5

**NIH grant funding for medical AI ethics by fiscal year, 2020–24**

Source: RAISE Health, 2025 | Chart: 2025 AI Index report

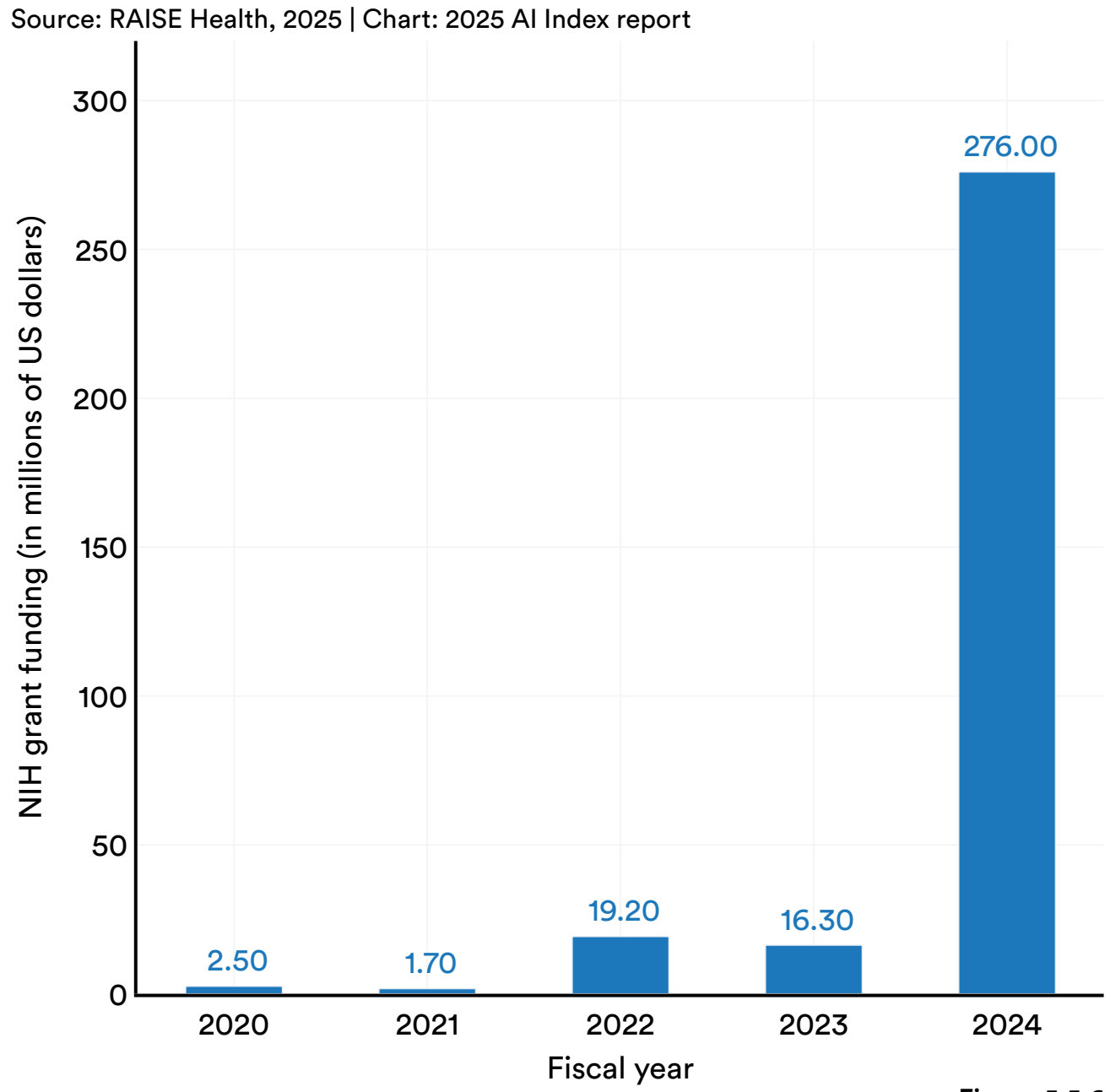

Figure 5.5.6

This year, dozens of foundation models have been developed across various scientific fields. Some are refined large language models, adapted for specific domains using relevant publications; others are trained from scratch with specialized data, such as time series or weather data. These foundation models are then fine-tuned for targeted scientific tasks or applications.

# 5.6 AI Foundation Models in Science

**Highlight:**

## Notable Model Releases

AI has driven significant progress in other scientific domains, including physics, chemistry, and geosciences. The table below highlights some of the most notable recent launches in these areas, alongside newly released resources that further track these developments. This analysis represents an initial effort by the AI Index, which aims to expand and deepen its coverage of AI-driven scientific progress across a broader range of disciplines in the future.

| Date | Name | Domain | Significance | Image |
|---|---|---|---|---|
| Feb 6, 2024 | CrystalLLM | Materials science | Researchers fine-tuned LLaMA-2 70B on text-encoded atomistic data to generate stable materials, achieving nearly double the metastability rate of a leading diffusion model (49% vs. 28%) while maintaining physical plausibility. The approach enables flexible applications like unconditional generation, structure infilling, and text-guided design, with model scale enhancing symmetry awareness. | 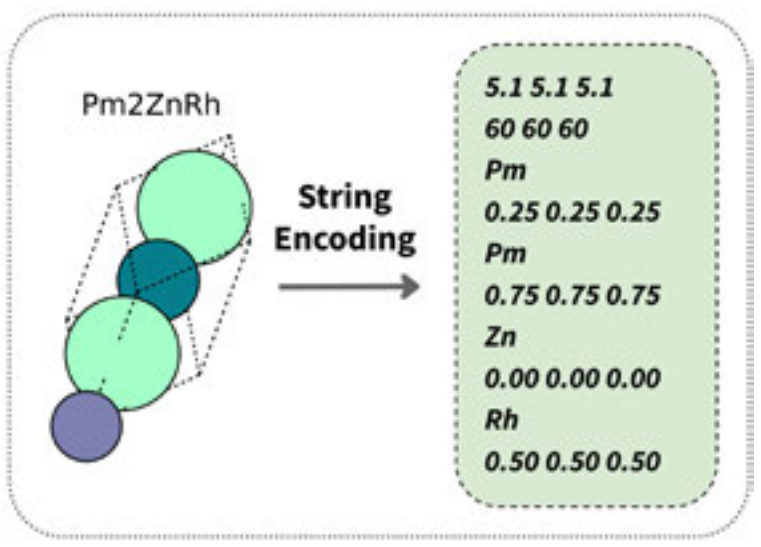<br>Figure 5.6.1<br>**Source:** Gruver et al., 2024 |
| Feb 14, 2024 | LlaSMol | Chemistry | To address LLMs' poor performance on chemistry tasks, researchers introduce SMolInstruct, a high-quality dataset with over 3 million samples across 14 tasks; and LlaSMol, a set of models fine-tuned on it. Among them, the Mistral-based LlaSMol outperforms GPT-4 and Claude 3 Opus by a wide margin, approaching task-specific model performance while tuning just 0.58% of parameters, demonstrating the power of domain-specific instruction tuning. | 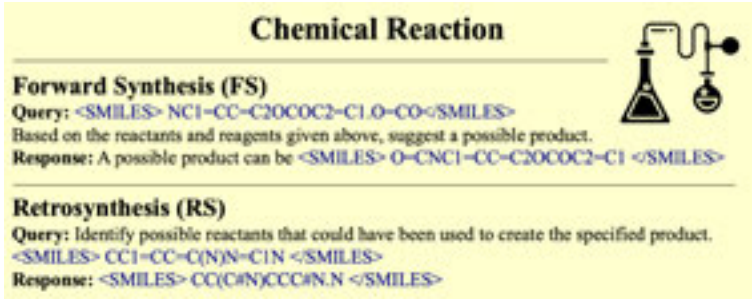<br>Figure 5.6.2<br>**Source:** Yu et al., 2024 |

**Highlight:**

## Notable Model Releases (cont'd)

| | | | | |
|---|---|---|---|---|
| Apr 23, 2024 | ORBIT | Earth science | Oak Ridge National Lab introduced ORBIT, a 113B-parameter vision transformer and the largest AI model ever built for climate science—1,000 times larger than prior models. Trained using a novel parallelism technique and tested on the Frontier supercomputer, ORBIT achieved up to 1.6 exaFLOPS of sustained performance. This breakthrough sets a new bar for AI-driven Earth system prediction. | 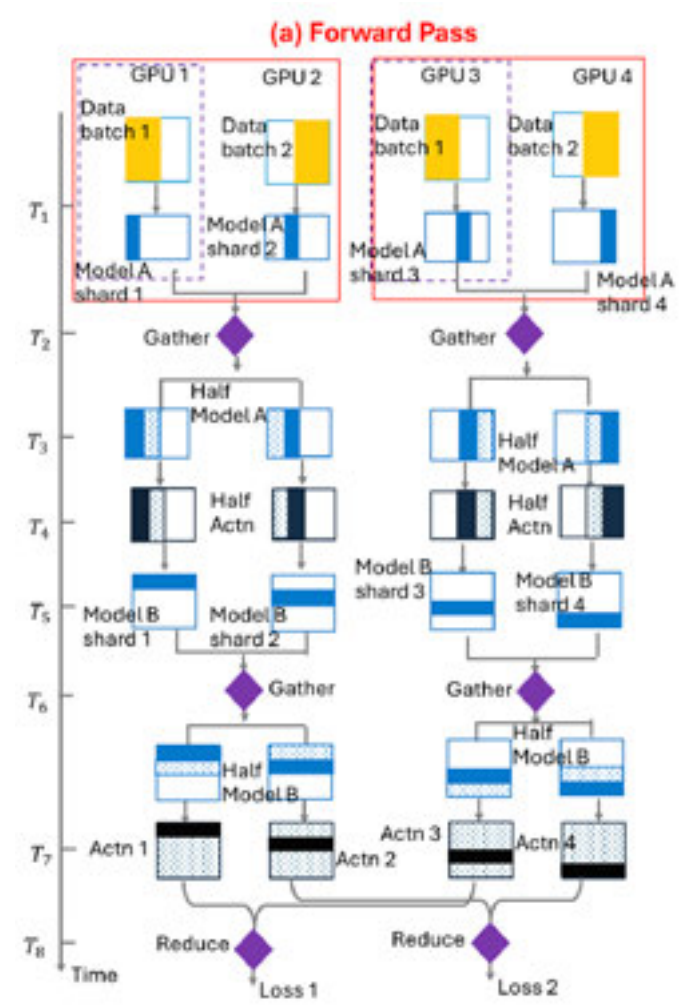<br><br>**Figure 5.6.3**<br>**Source:** Wang et al., 2024 |
| May 20, 2024 | Aurora | Earth science | Aurora is a large-scale foundation model trained on over a million hours of Earth system data, delivering state-of-the-art forecasts for air quality, ocean waves, cyclone tracks, and high-resolution weather. It outperforms traditional systems while operating at a fraction of the computational cost, and can be fine-tuned across domains with minimal resources—marking a major step toward accessible, AI-driven Earth system forecasting. | 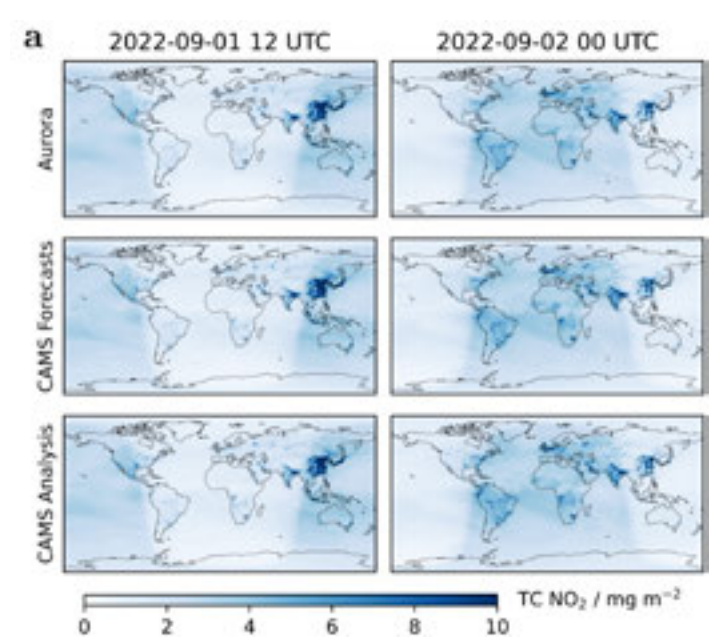<br><br>**Figure 5.6.4**<br>**Source:** Bodnar et al., 2024 |
| Jul 22, 2024 | NeuralGCM | Weather forecasting | This study introduces NeuralGCM, a hybrid model that combines a differentiable, physics-based solver with machine learning components to simulate both weather and climate. It matches or exceeds leading ML and physics-based models in short- and medium-term forecasts, accurately tracks climate metrics over decades, and captures complex phenomena like tropical cyclones—all while offering massive computational savings. | 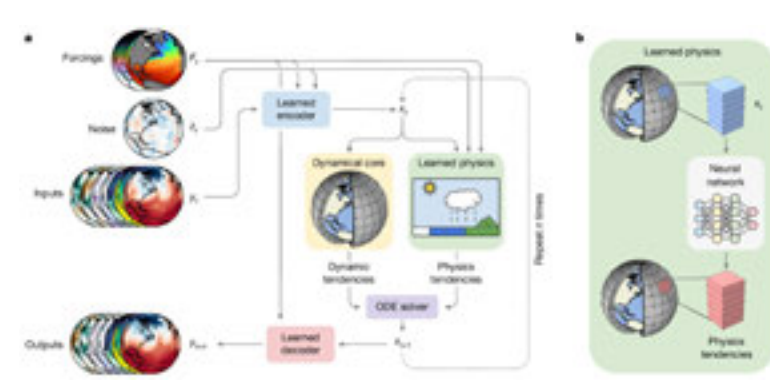<br>**Figure 5.6.5**<br>**Source:** Kochkov et al., 2024 |

**Highlight:**

## Notable Model Releases (cont'd)

| | | | | |
|---|---|---|---|---|
| Aug 18, 2024 | PhysBERT | Physics | Physics texts are notoriously difficult for NLP due to their specialized language and complex concepts. PhysBERT, the first physics-specific, text-embedding model, addresses this by outperforming general-purpose models on physics-specific tasks. Trained on 1.2 million arXiv papers and fine-tuned with supervised data, it significantly boosts performance in information retrieval and subdomain fine-tuning. | 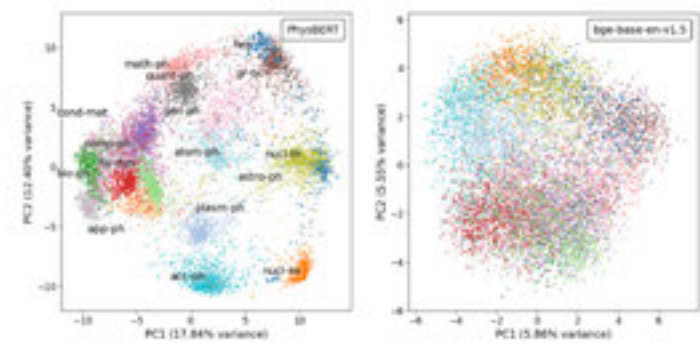<br>**Figure 5.6.6**<br>**Source:** Hellert et al., 2024 |
| Sep 16, 2024 | FireSat | Fire prediction | Google's FireSat is a satellite-based wildfire detection system that uses AI to identify fires as small as 5x5 meters within 20 minutes of ignition by analyzing real-time imagery and environmental data. Developed in partnership with Earth Fire Alliance and Muon Space, it not only enhances disaster response but also advances global wildfire research. | 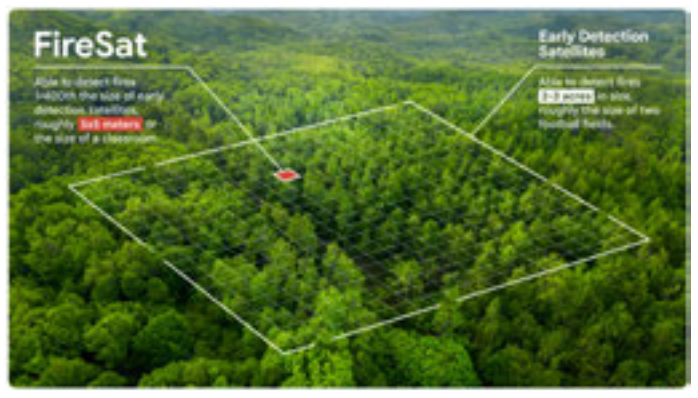<br><br>**Figure 5.6.7**<br>**Source:** Google, 2024 |
| Dec 4, 2024 | GenCast | Weather prediction | Google DeepMind's GenCast is an AI-powered weather model that delivers highly accurate 15-day forecasts using a diffusion-based approach, outperforming traditional systems like the ENS on nearly all metrics. It generates forecasts in minutes instead of hours and has broad applications in disaster response, renewable energy, and agriculture. | 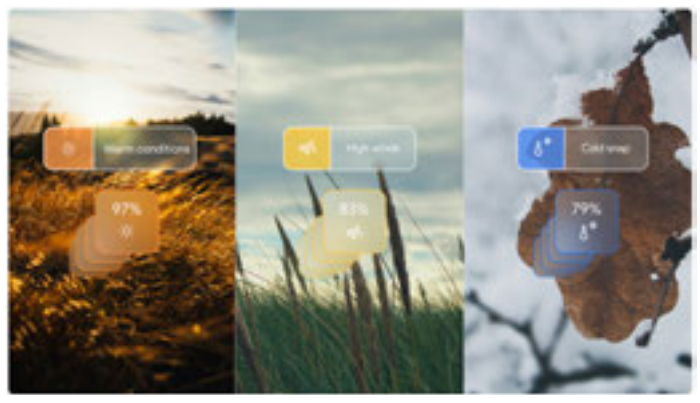<br>**Figure 5.6.8**<br>**Source:** Google, 2024 |
| Dec 9, 2024 | AlphaQubit | Quantum computing | In late 2024, Google DeepMind and Google Quantum AI released AlphaQubit, an AI-based decoder with state-of-the-art quantum error detection. Soon after, they introduced Willow, the first quantum chip to achieve exponential error suppression and correction below the surface code threshold—a major milestone in the field. Willow also completed a benchmark task in under five minutes that would take the fastest supercomputer over 10 septillion years, longer than the age of the known universe. | 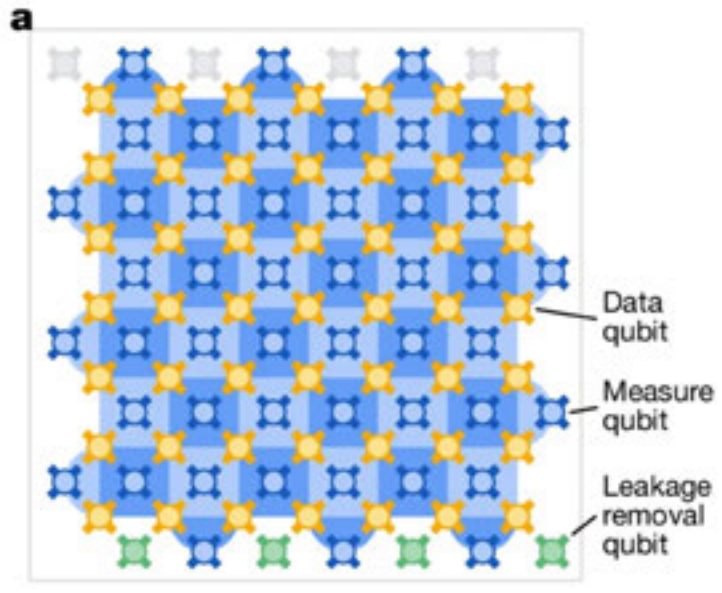<br><br>**Figure 5.6.9**<br>**Source:** Google, 2024 |

# CHAPTER 6:
Policy and Governance

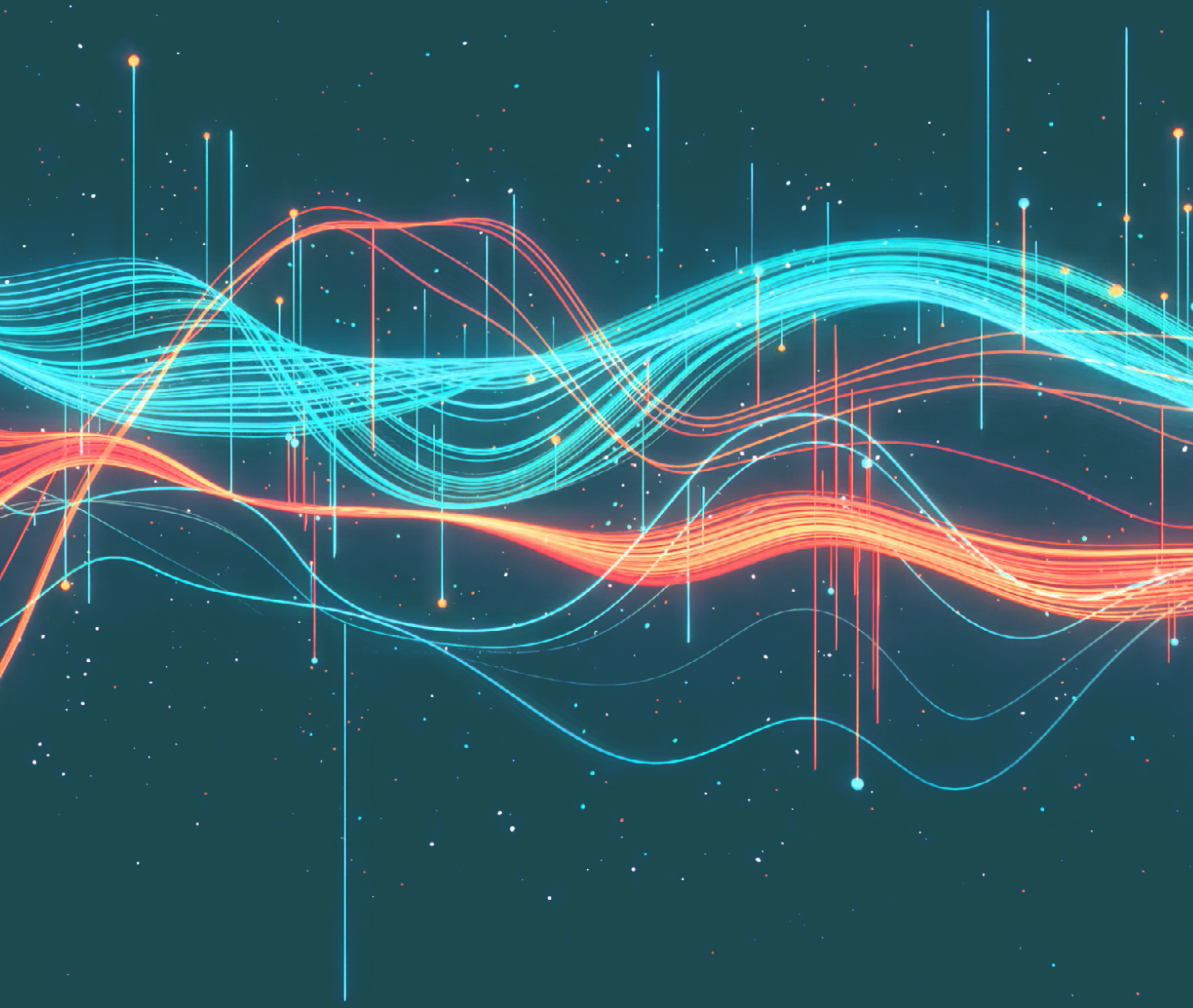

# Chapter 6: Policy and Governance

ACCESS THE PUBLIC DATA

**CHAPTER 6:**
Policy and Governance

# Overview

AI's advancing capabilities have captured policymakers' attention, leading to an increase in AI-related policies worldwide. In recent years, nations and political bodies, including the United States and the European Union, have introduced significant regulations. More recently, many governments have announced major investments in AI infrastructure. This wave of policymaking reflects a growing recognition of the need to both regulate AI and harness its transformative potential.

This chapter explores global AI governance, starting with a timeline of key AI policymaking events in 2024. It then examines global and U.S. legislative efforts, analyzes AI-related mentions in legislative discussions, and reviews how U.S. regulatory agencies have approached AI. The chapter concludes with an analysis of public investment in AI in the U.S., with most data sourced independently by the AI Index.

**CHAPTER 6:**
Policy and Governance

# Chapter Highlights

**1. U.S. states are leading the way on AI legislation amid slow progress at the federal level.** In 2016, only one state-level AI-related law was passed, increasing to 49 by 2023. In the past year alone, that number more than doubled to 131. While proposed AI bills at the federal level have also increased, the number passed remains low.

---

**2. Governments across the world invest in AI infrastructure.** Canada announced a $2.4 billion AI infrastructure package, while China launched a $47.5 billion fund to boost semiconductor production. France committed €109 billion to AI infrastructure, India pledged $1.25 billion, and Saudi Arabia's Project Transcendence represents a $100 billion AI investment initiative.

---

**3. Across the world, mentions of AI in legislative proceedings keep rising.** Across 75 major countries, AI mentions in legislative proceedings increased by 21.3% in 2024, rising to 1,889 from 1,557 in 2023. Since 2016, the total number of AI mentions has grown more than ninefold.

---

**4. AI safety institutes expand and coordinate across the globe.** In 2024, countries worldwide launched international AI safety institutes. The first emerged in November 2023 in the U.S. and the U.K. following the inaugural AI Safety Summit. At the AI Seoul Summit in May 2024, additional institutes were pledged in Japan, France, Germany, Italy, Singapore, South Korea, Australia, Canada, and the European Union.

---

**5. The number of U.S. AI-related federal regulations skyrockets.** In 2024, 59 AI-related regulations were introduced—more than double the 25 recorded in 2023. These regulations came from 42 unique agencies, twice the 21 agencies that issued them in 2023.

---

**6. U.S. states expand deepfake regulations.** Before 2024, only five states—California, Michigan, Washington, Texas, and Minnesota—had enacted laws regulating deepfakes in elections. In 2024, 15 more states, including Oregon, New Mexico, and New York, introduced similar measures. Additionally, by 2024, 24 states had passed regulations targeting deepfakes.

This chapter begins with an overview of some of the most significant AI-related policy events in 2024, as selected by the AI Index Steering Committee.

# 6.1 Major Global AI Policy News in 2024

**Feb. 21, 2024**

## Singapore plans to invest $1B in AI over 5 years

In his budget speech on February 16, Deputy Prime Minister and Finance Minister Lawrence Wong announced that the government will allocate over $1 billion over the next five years to support AI computation, talent development, and industry growth.

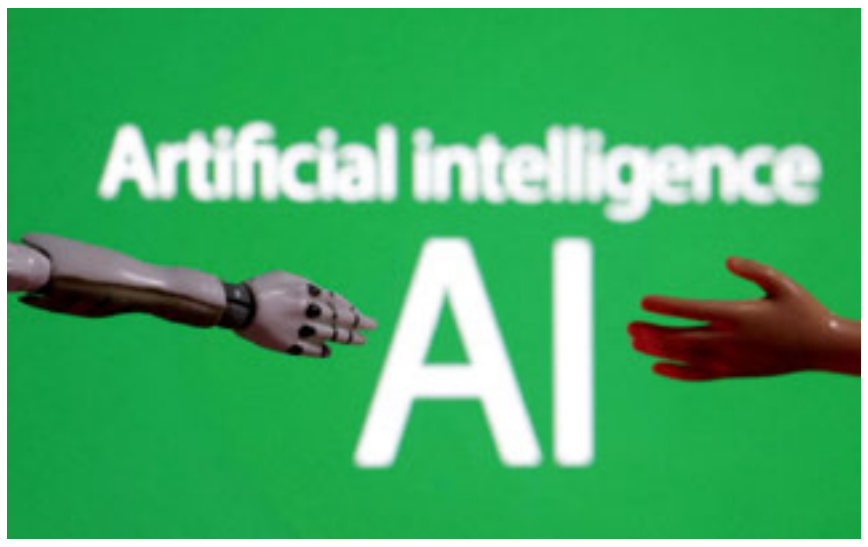

Source: The Straits Times, 2024
Figure 6.1.1

**Mar. 11, 2024**

## Abu Dhabi launches $100B AI investment firm

In March 2024, Abu Dhabi established MGX Fund Management Limited, a state-owned investment firm specializing in AI technologies, with a target of managing $100 billion in assets. This initiative aligns with the UAE's strategic objective to position itself as a global leader in AI innovation and technology.

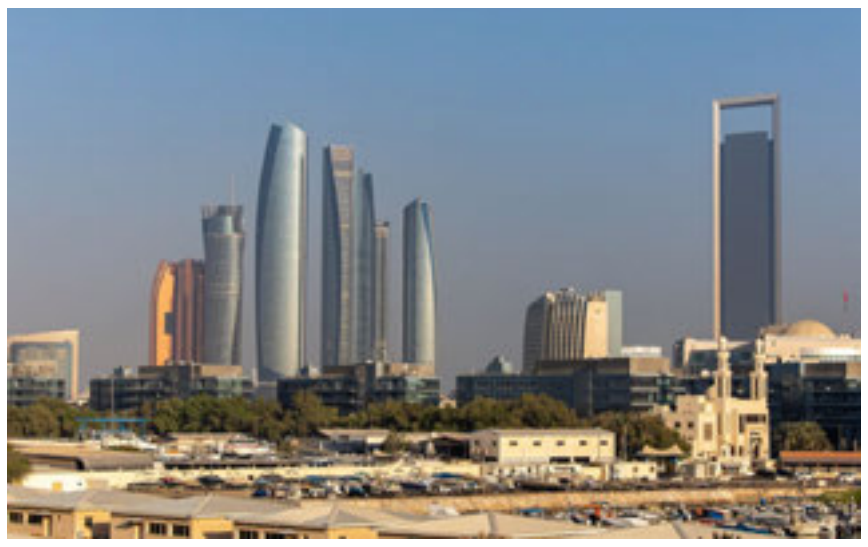

Source: Bloomberg, 2024
Figure 6.1.2

**Mar. 13, 2024**

## Artificial Intelligence Act is passed by European Parliament

The landmark EU AI Act, the first of its kind, was passed by the European Parliament three months after a provisional agreement on the bill was reached. The legislation introduces sweeping provisions around AI systems, including transparency and reporting obligations, risk-based regulations, and bans on certain applications including social scoring, human manipulation, and biometric categorization that uses "sensitive characteristics." Most of the Act's provisions will come into effect in 2026 after a two-year implementation period. The Act is significant for its restrictive nature, building on the already stringent EU privacy regulations. It takes a unique approach to regulating generative AI, differing from other proposed legislation, and has been met with resistance from the industry.

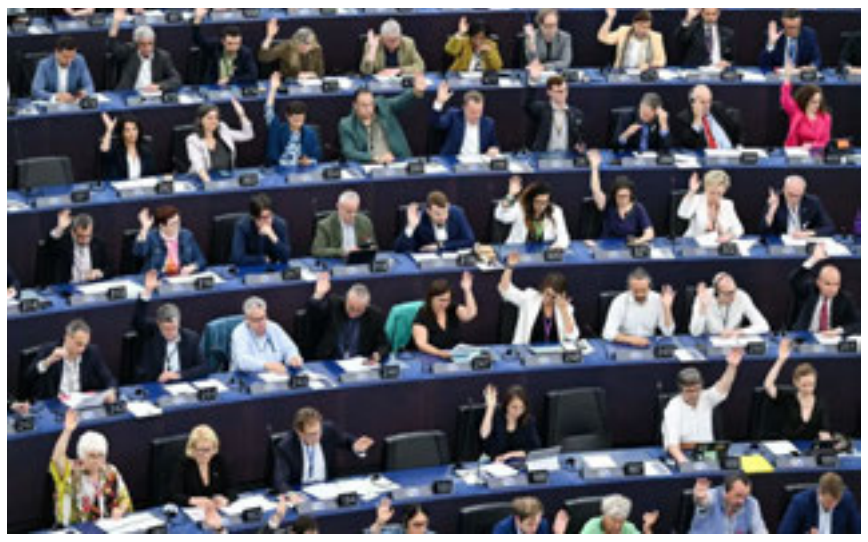

Source: Time, 2023
Figure 6.1.3

**Mar. 15, 2024**

### India drops plan to require government approval for launch of new AI models

Less than a month after issuing an advisory requiring tech firms to obtain government approval before launching new AI models, India releases revised guidelines for companies' self-regulation, following backlash from entrepreneurs and investors. Under the new guidelines, firms must inform users if their models are undertested or unreliable. India's IT Ministry retained its emphasis that AI models should not undermine electoral integrity or promote bias and discrimination.

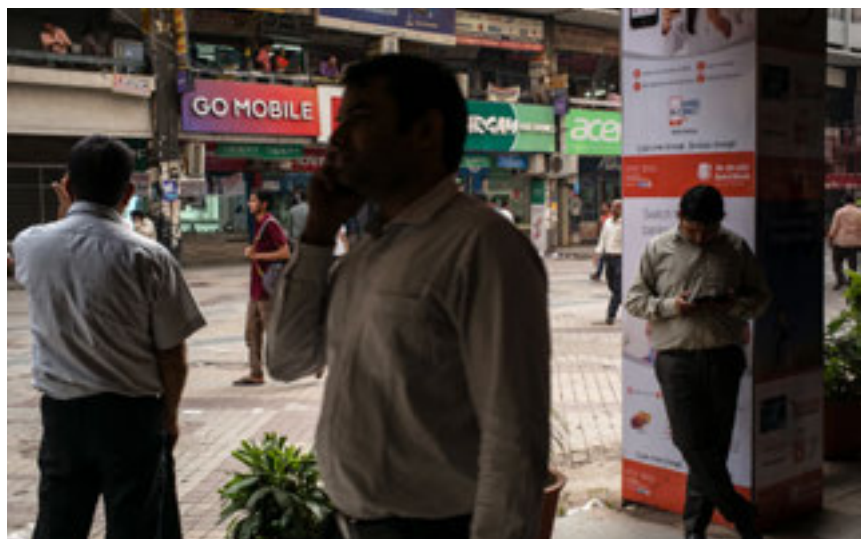

Source: TechCrunch, 2024
Figure 6.1.4

---

**Mar. 17, 2024**

### India launches IndiaAI Mission with $1.25B investment

In March 2024, India launched the IndiaAI Mission to strengthen its AI ecosystem. The $1.25 billion initiative aims to build 10,000-plus GPUs via public-private partnerships, develop a national nonpersonal data platform, and support homegrown AI models and deep-tech startups. It also prioritizes ethical AI governance and expanding AI labs beyond major cities to democratize access.

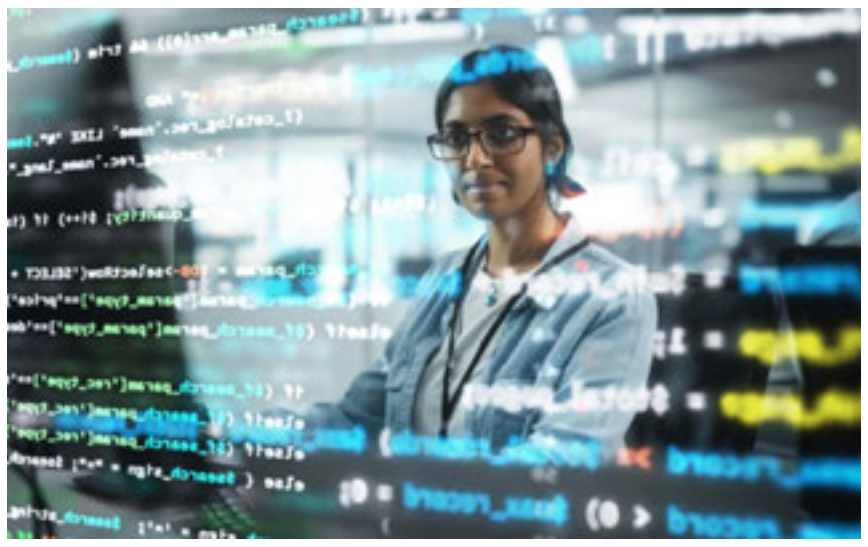

Source: Nature, 2024
Figure 6.1.5

---

**Mar. 20, 2024**

### French government fines Google 250 million euros over use of copyrighted information

France's competition watchdog, the Autorité de la Concurrence, took a harsh stance toward negligent model training when it fined Google 250 million euros for using French news content to train Bard, now Gemini, the company's AI-powered chatbot—without notifying media companies. The government cited the offense as a breach of EU intellectual property rules, and claimed it prevented publishers and press agencies from negotiating fair prices. Google accepted the settlement and proposed a series of measures to mitigate scraping issues.

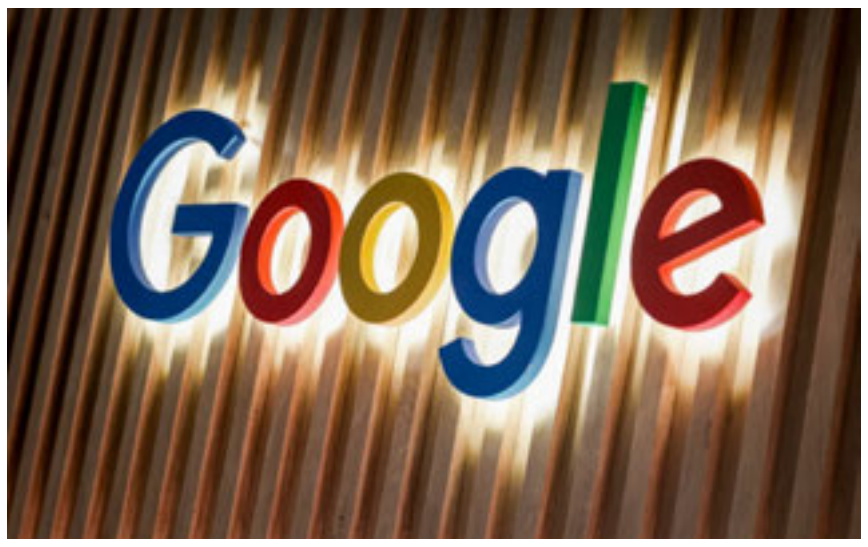

Source: NBC News, 2024
Figure 6.1.6

---

Mar. 21, 2024

### U.N. General Assembly adopts resolution promoting "safe, secure, and trustworthy" AI

Backed by more than 120 member states, the U.N. General assembly adopted a "historic" U.S.-led resolution (although not officially legally binding) on the promotion of "safe, secure, and trustworthy" artificial intelligence systems. The assembly called on stakeholders to ensure that artificial intelligence systems be used in compliance with human rights laws, recognizing the role these systems may play in accelerating progress toward reaching the U.N.'s Sustainable Development Goals. The resolution was supported by more than 120 states, including China, and endorsed without a vote by all 193 U.N. member states.

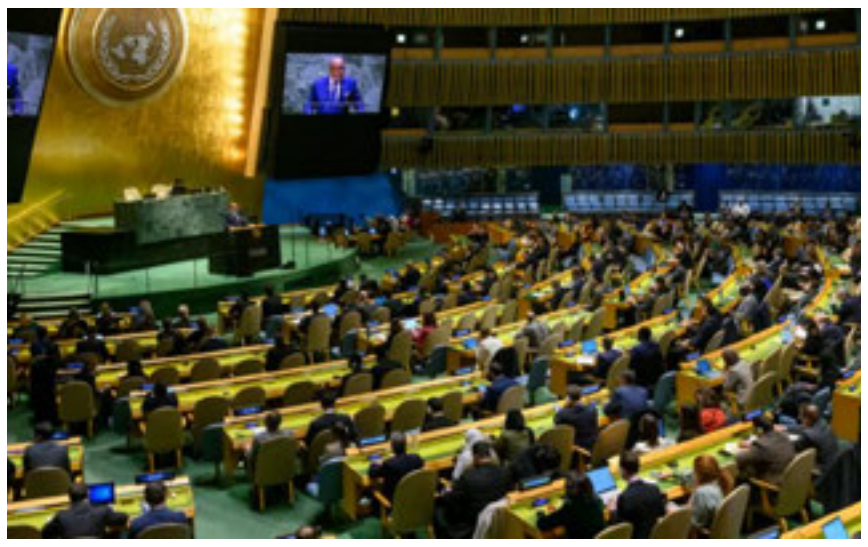

Source: UN News, 2024
Figure 6.1.7

Apr. 7, 2024

### Canada pledges CA$2.4B investment to ensure country's AI advantage

The Canadian Federal Budget for 2024 featured a CA$2.4 billion package of measures to "secure Canada's AI advantage" in the midst of an intensifying global race for AI development and adoption. Funding would be directed toward a range of initiatives, including increasing capabilities and infrastructure for researchers and developers, boosting AI startups, helping small and medium businesses increase productivity through AI, supporting workers impacted by AI, and creating a new Canadian AI Safety Institute.

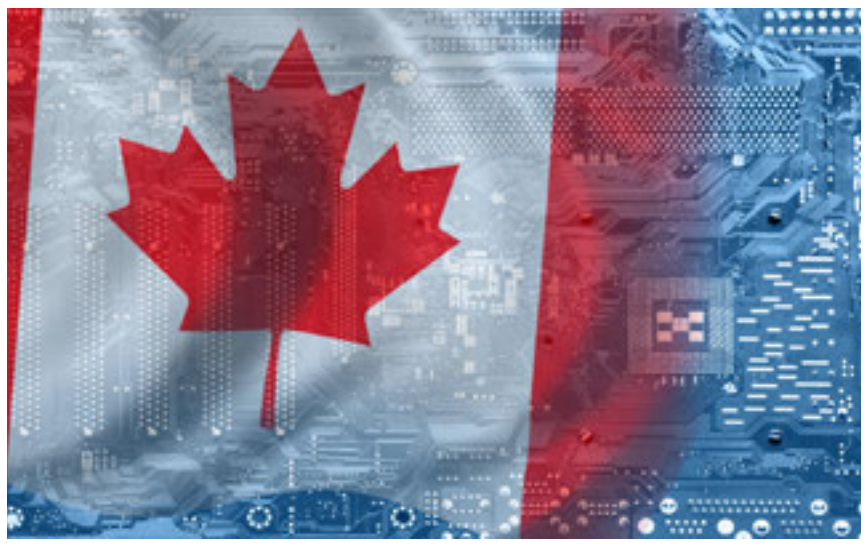

Source: Center for International Governance Innovation, 2024
Figure 6.1.8

May 11, 2024

### U.K. AI Safety Institute launches open-source tool for assessing AI model safety

The agency released a toolset, called Inspect, designed to assess AI models' capabilities in a range of areas, including core knowledge, ability to reason, and autonomous capabilities. The Institute claimed it was the first time an AI safety testing platform had been spearheaded by a government-backed body, and made available for public use under an open-source license in order to benefit industry, research organizations, and academia.

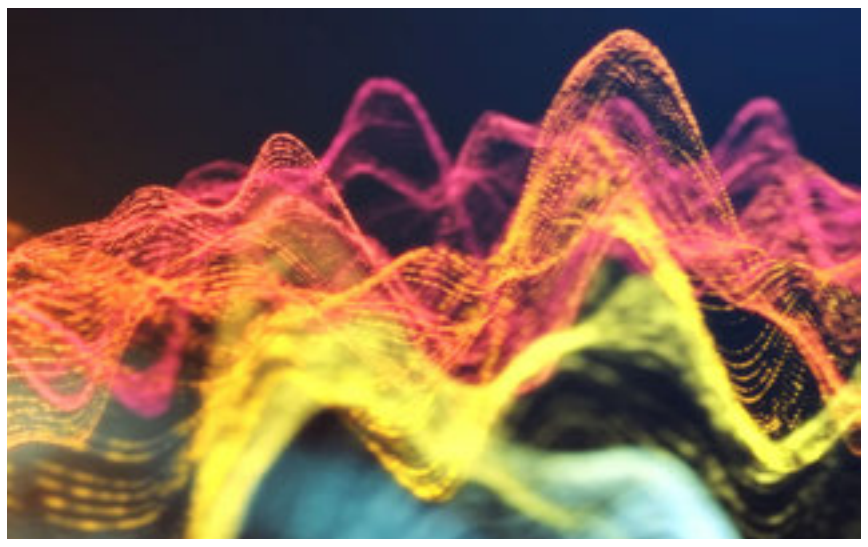

Source: TechCrunch, 2024
Figure 6.1.9

**May 21, 2024**

## U.K. and South Korea cohost AI safety summit in Seoul

At the AI Seoul Summit, attending countries shared the safety measures they adopted in line with the Bletchley Declaration, which was signed the year prior at the U.K. AI Safety Summit. The declaration emphasizes the ethical and responsible development of AI. Building on the progress made at the U.K. summit, countries have since launched or announced plans for AI safety institutes. In Seoul, these nations took another step forward by signing a letter of intent to establish a collaborative network of institutes, highlighting the importance of global cooperation in advancing AI safety.

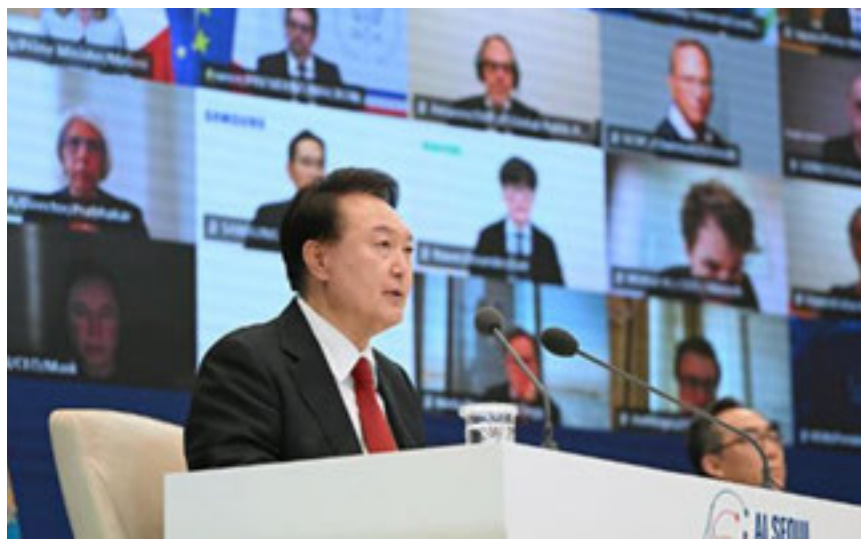

Source: Center for Strategic and International Studies, 2024
Figure 6.1.10

**May 27, 2024**

## China creates country's largest-ever state-backed investment fund to back its semiconductor industry

China launched a fund worth $47.5 billion to boost semiconductor production. The launch marks the third phase of China's "Big Fund," which has supported the industry's development since 2014, including crucial investments into the country's two largest chip foundries. The move comes amid rising U.S. export controls on critical technologies like semiconductors that underpin hardware components like GPUs used to train AI systems.

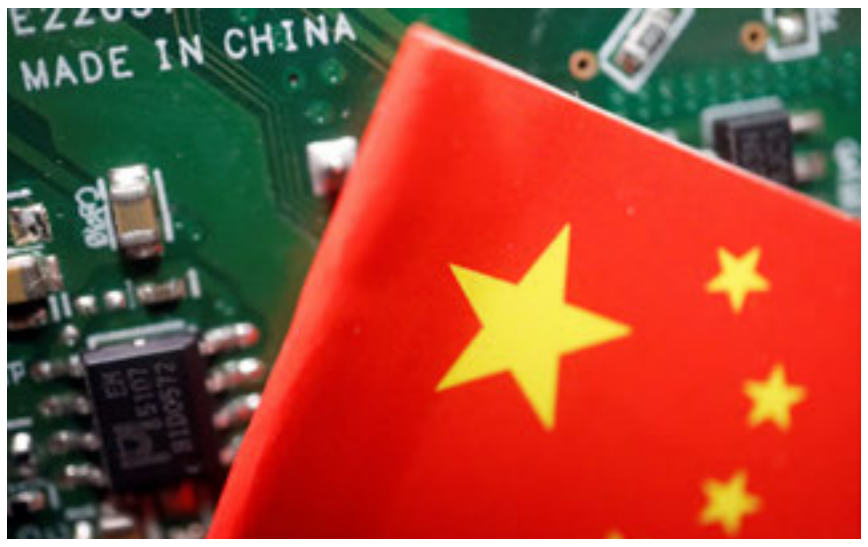

Source: Reuters, 2024
Figure 6.1.11

**May 28, 2024**

## European Commission establishes AI Office

Over three years after the EU AI Act was proposed, the European Commission unveils its cornerstone. The AI Office will play a key role in implementing the Act, enforcing standards for general-purpose AI models, coordinating the development of codes of practice, and applying sanctions for offenses under the Act. With over 140 staff members, the body consists of five units dedicated to different AI-related goals, including promoting societal good through AI and pursuing excellence in AI and robotics.

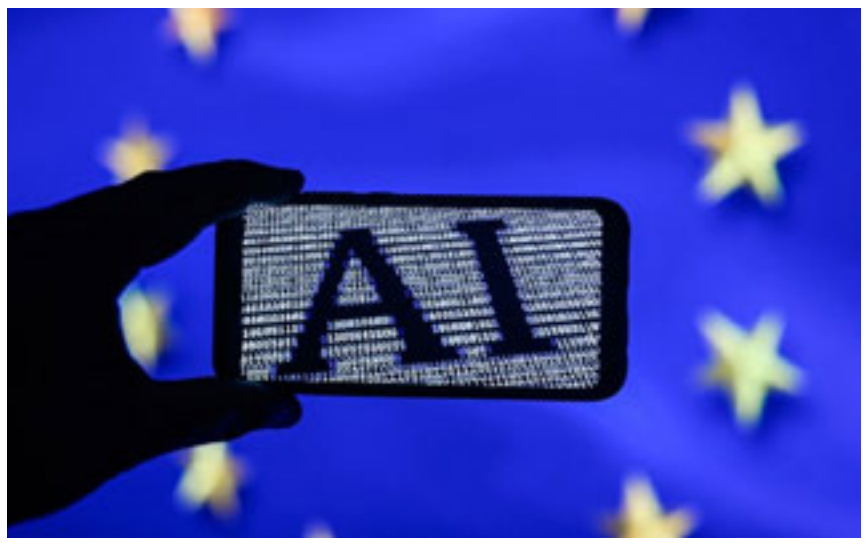

Source: Center for Strategic and International Studies, 2024
Figure 6.1.12

**Jun. 26, 2024**

### U.S. NIST unveils framework to help organizations identify and mitigate GenAI risks

The National Institute of Standards and Technology (NIST) launches a voluntary framework to help organizations identify unique risks posed by generative AI and recommends a series of actions for mitigating those risks. The framework extends the NIST AI Risk Management Framework released in 2023. Recommended actions include determining AI risk tolerance and respective risk management needs, establishing clear responsibilities for managing AI risks, and involving nondeveloper experts in regular assessment and updates. The framework followed the release of a NIST document on adversarial machine learning outlining a taxonomy of attack types, the effects of such attacks, and mitigation strategies.

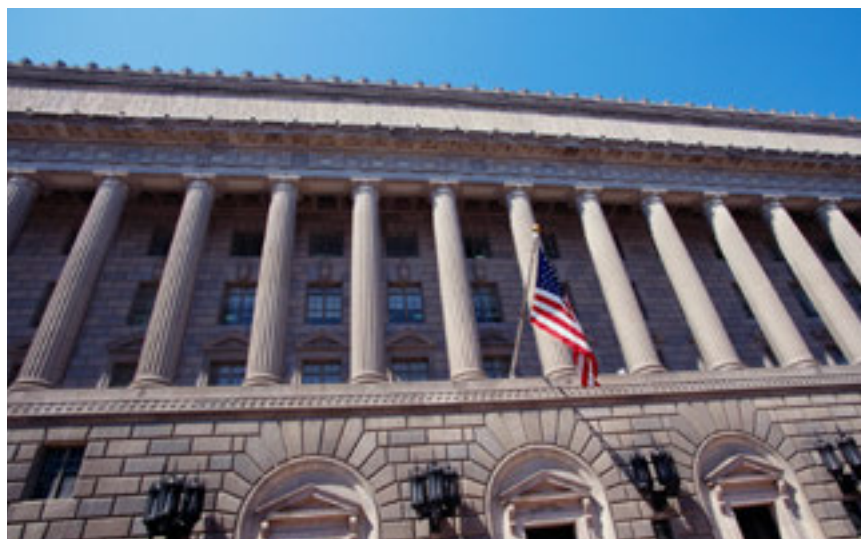

Source: FedScoop, 2024
Figure 6.1.13

**Jul. 25, 2024**

### U.S. State Department releases AI Risk Management Profile for Human Rights

The U.S. State Department designed the Risk Management Profile for Artificial Intelligence and Human Rights as a guide for governments, businesses, and civil society to align AI risk management with human rights protections. Built on the NIST AI Risk Management Framework, the Profile outlines four key functions—govern, map, measure, and manage—to assess and mitigate AI risks, from bias to misuse for surveillance. By bridging AI governance and human rights, it provides a globally applicable tool for responsible AI development and deployment.

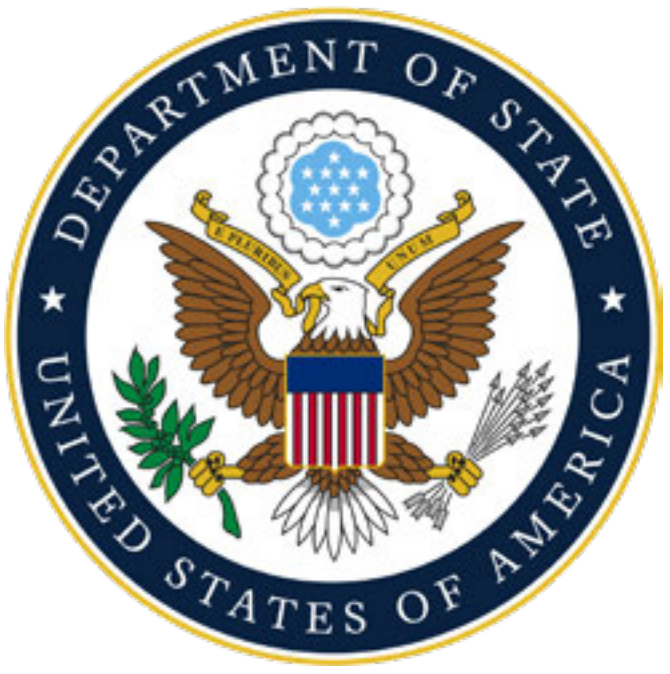

Source: U.S. Department of State, 2024
Figure 6.1.14

**Aug, 2, 2024**

### U.K. withdraws £1.3B promised for technology and AI infrastructure

The U.K.'s Labour government canceled £1.3 billion in funding promised for technology and AI projects, explaining that the commitments made by the previous government had been "underfunded." Announced in 2023, the projects included £500 million for the AI Research Resource, which funds computing power, and £800 million for the creation of the University of Edinburgh's exascale supercomputer.

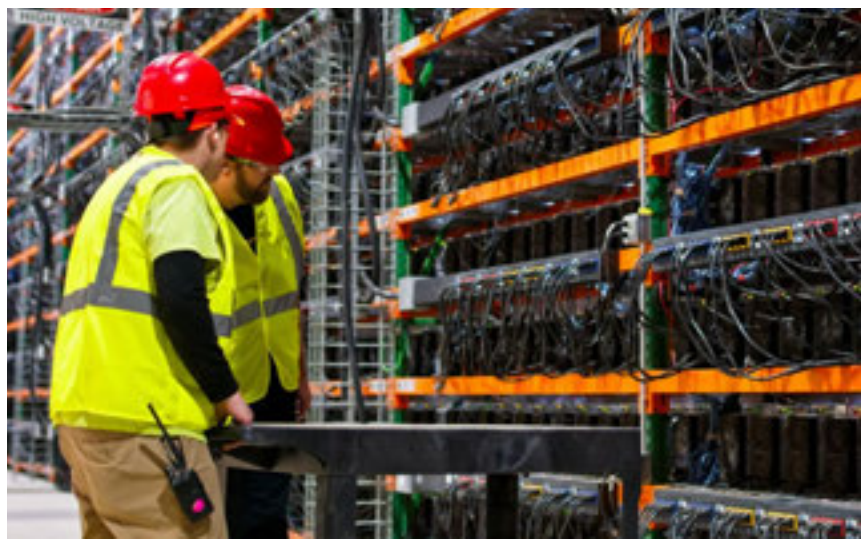

Source: BBC, 2024
Figure 6.1.15

**Sep. 13, 2024**

### U.S. White House launches task force on AI data center infrastructure

A White House meeting brought together federal officials and technology executives to discuss securing power sources for robust data center infrastructure critical to AI models. Executives from OpenAI, Anthropic, Amazon Web Services, Nvidia, and Alphabet were present. A White House press release emphasized that advancing AI development in the U.S. is vital for national security and ensuring AI systems are safe, secure, and trustworthy. The newly formed AI data center infrastructure task force will identify opportunities and work with agencies to prioritize the development of AI data centers.

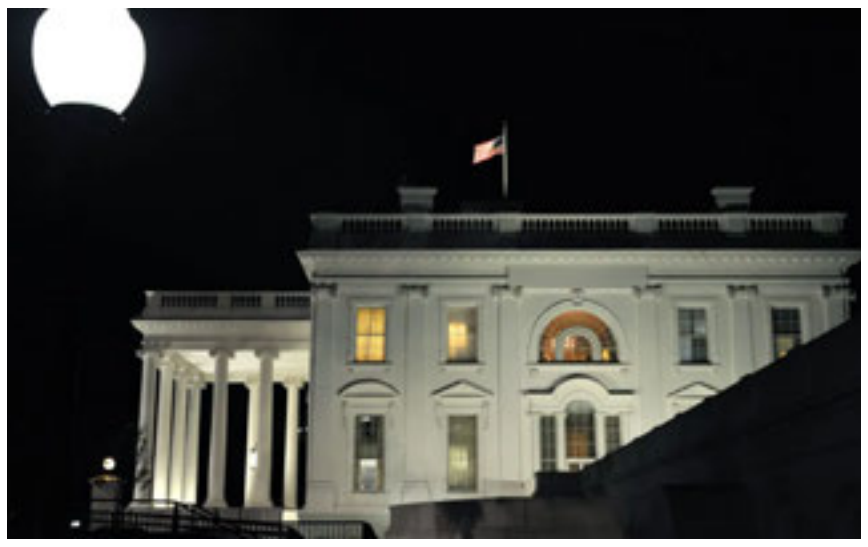

Source: FedScoop, 2024
Figure 6.1.16

**Sep. 17, 2024**

### California governor signs three bills on AI and elections communications

Ahead of the 2024 San Francisco mayoral election, Governor Gavin Newsom announced the signing of three bills into law aimed at combating deepfake election content. AB 2655, AB 2839, and AB 2355 require large online platforms to remove or label digitally altered election content during specified periods, expand the time frame for prohibiting the distribution of deceptive AI-generated election content, and mandate that electoral ads using AI-generated or altered content include appropriate disclosures, respectively.

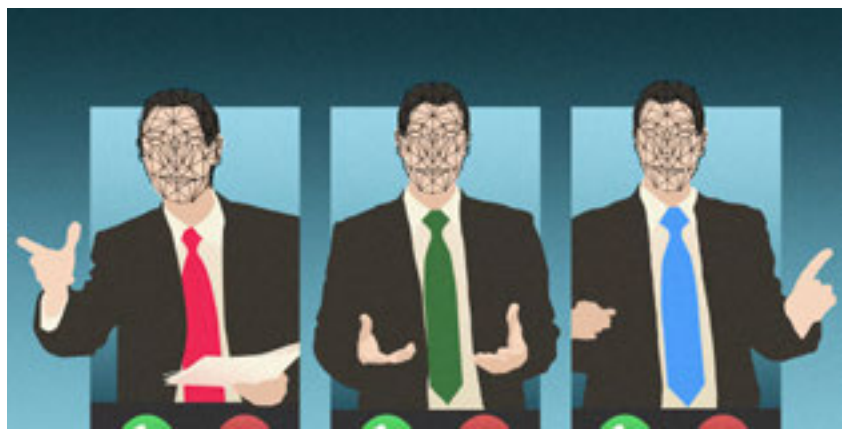

Source: The Wall Street Journal, 2024
Figure 6.1.17

**Sep. 22, 2024**

### United Nations adopts Global Digital Compact to ensure an inclusive and secure digital future

During the Summit of the Future, U.N. member states adopted the Global Digital Compact, aiming to establish an inclusive, open, sustainable, fair, safe, and secure digital future for all. The Compact emphasizes objectives such as closing digital divides, expanding benefits from the digital economy, fostering a digital space that respects human rights, advancing equitable data governance, and enhancing international governance of artificial intelligence. Guided by principles anchored in international law and human rights, the Compact seeks to harness digital technologies to accelerate progress toward the Sustainable Development Goals.

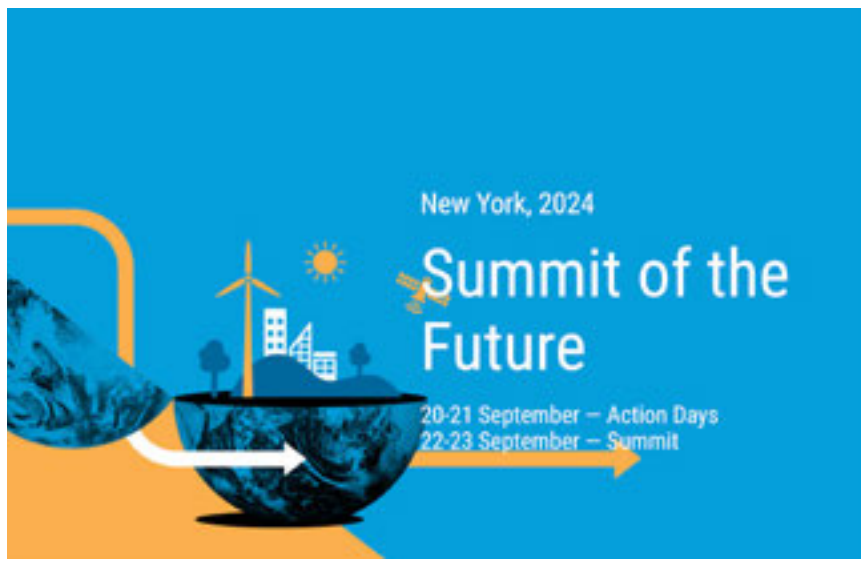

Source: United Nations, 2024
Figure 6.1.18

**Sep. 29, 2024**

### California governor vetoes expansive AI legislation

Governor Gavin Newsom vetoed California's AI safety bill, which would have set a national precedent for AI regulation, given the state's role as home to many leading AI companies. The bill sought to mandate safety testing for frontier AI models before their public release and would have allowed the state attorney general to sue companies over AI-related harm. Supporters argued it was a necessary step to ensure AI safety and accountability, while critics contended it was overly restrictive and could stifle AI development, especially of the open-weight AI ecosystem. Given California's status as the world's fifth-largest economy, the bill's impact could have extended beyond state borders, akin to the Brussels effect, shaping AI governance nationally and internationally. Newsom defended his veto, arguing the bill imposed excessive standards.

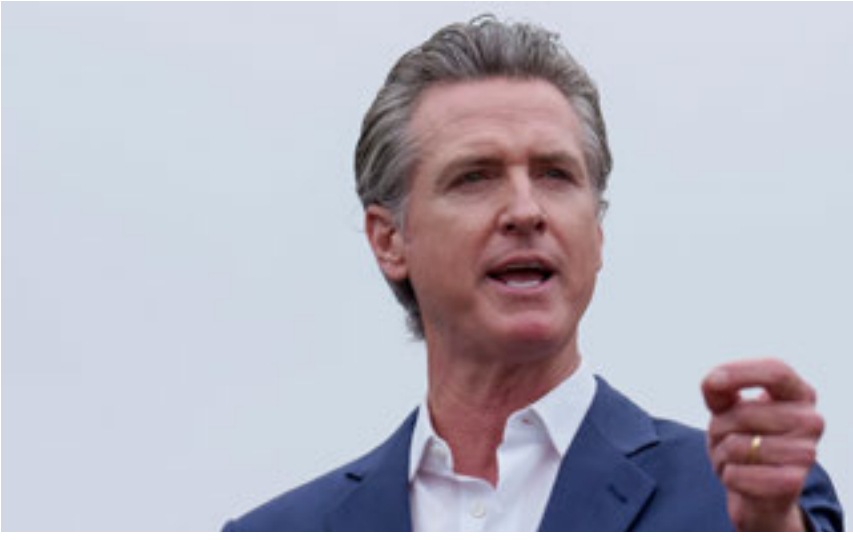

Source: Financial Times, 2024
Figure 6.1.19

**Oct. 2, 2024**

### U.S. judge blocks new California AI law over Kamala Harris deepfake

A federal judge in California issued a temporary injunction on one of the state's new AI laws just two weeks after it was signed. In his ruling, Judge Mendez cited the law's vague definition of "harmful" depictions as a potential threat to constitutionally protected speech. The law had been used to prosecute an X user after he had posted a deepfake featuring Kamala Harris.

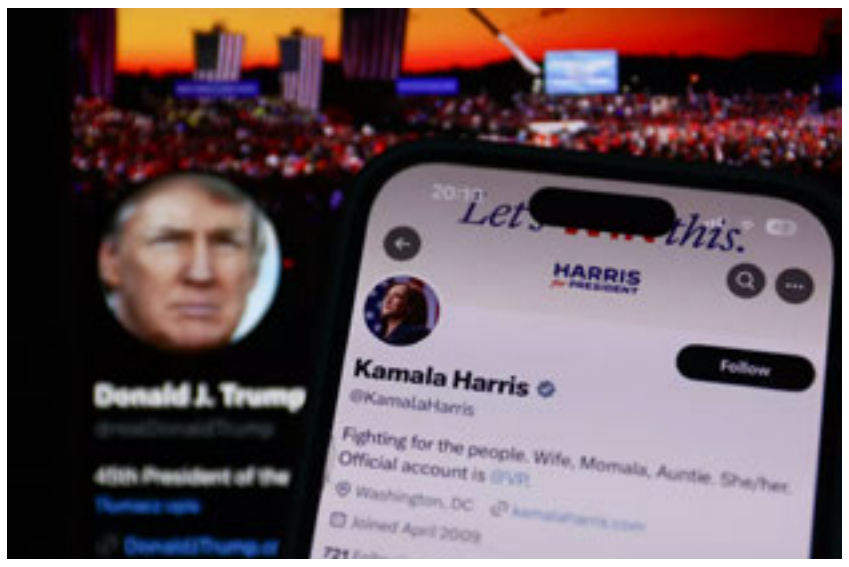

Source: Los Angeles Times, 2024
Figure 6.1.20

**Nov. 8, 2024**

### Saudi Arabia announces "Project Transcendence"

In November 2024, Saudi Arabia announced Project Transcendence, a $100 billion AI initiative aimed at establishing the kingdom as a global tech hub. Spearheaded by the Public Investment Fund, the project includes a partnership with Alphabet, Google's parent company, involving an investment between $5 billion and $10 billion to develop Arabic-language AI models. This initiative aligns with Saudi Arabia's Vision 2030, which focuses on diversifying the region's economy beyond oil and becoming a meaningful hub of AI.

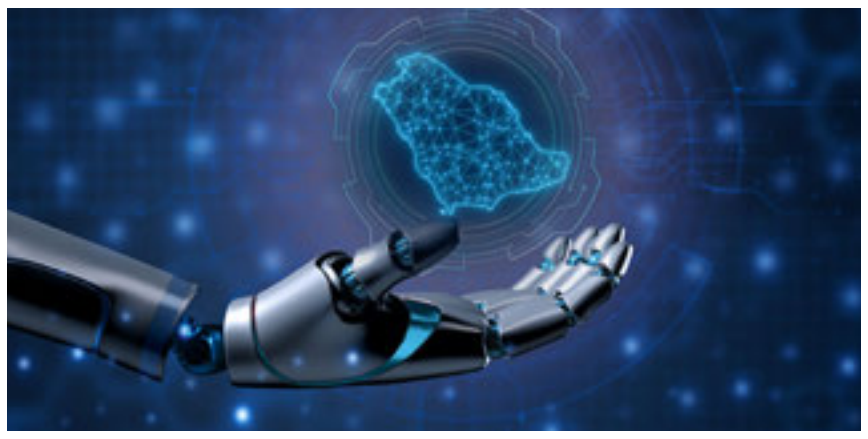

Source: Telecom Review, 2024
Figure 6.1.21

**Nov. 14, 2024**

### European Commission AI Office releases first draft of Code of Practice for General-Purpose AI

The European AI Office issued the first of four drafts for the General-Purpose AI Code of Practice. This code was developed by four working groups of independent experts, focusing on transparency and copyright, risk identification and assessment, risk mitigation, and internal governance. Once finalized, the code will complement the AI Act, allowing AI model providers to demonstrate compliance until a finalized standard is published.

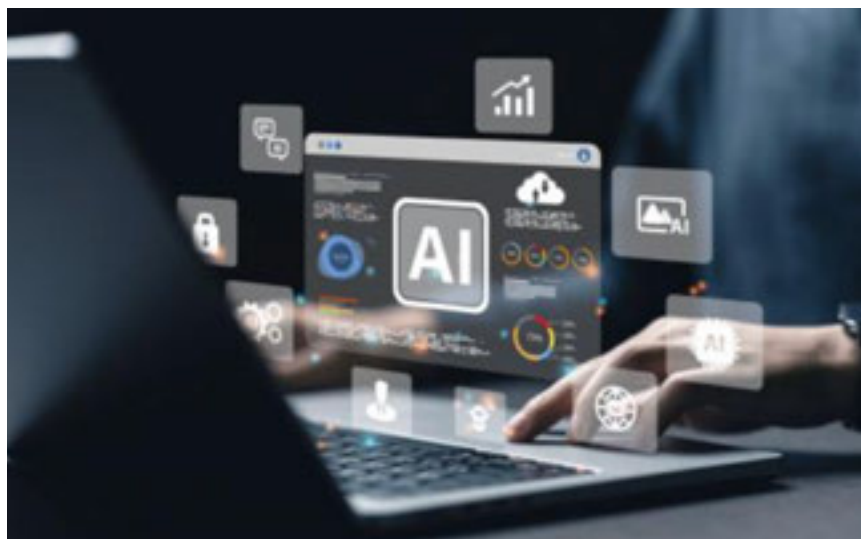

Source: European Union, 2024
Figure 6.1.22

**Nov. 25, 2024**

### U.S. launches international AI safety network with global partners

In November 2024, the U.S. Department of Commerce and the U.S. Department of State cohosted the inaugural meeting of the International Network of AI Safety Institutes in San Francisco. This initiative aims to improve global coordination on safe AI innovation, focusing on managing synthetic content risks, testing foundation models, and conducting risk assessments for advanced AI systems. The United States serves as the inaugural chair, with initial members including Australia, Canada, the European Union, France, Japan, Kenya, the Republic of Korea, Singapore, and the United Kingdom. The network has secured over $11 million in global research funding commitments to support its efforts.

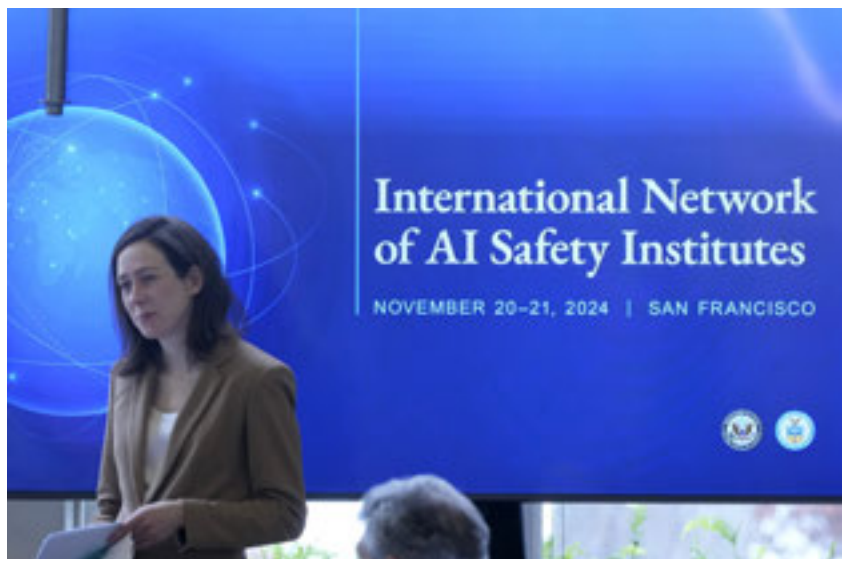

Source: AP, 2024
Figure 6.1.23

**Dec. 2, 2024**

### U.S. increases export controls of semiconductor manufacturing equipment and software to China

The U.S. Department of Commerce's Bureau of Industry and Security further limited China's ability to produce advanced semiconductors by announcing new export controls. These measures include restrictions on 24 types of semiconductor manufacturing equipment, three types of software tools, and additional limitations. The secretary of commerce emphasized the importance of these measures in safeguarding U.S. national security.

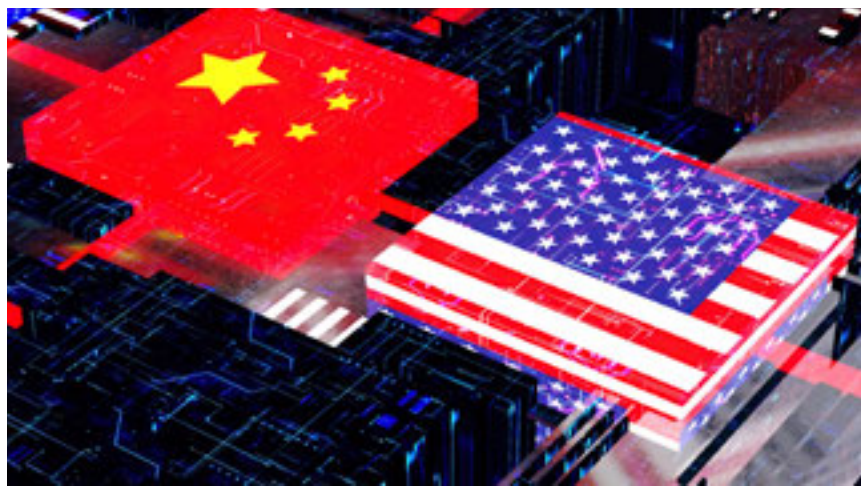

Source: CNBC, 2024
Figure 6.1.24

**Dec. 19, 2024**

### U.N. Security Council debates uses of AI in conflicts and calls for global framework

On Dec. 19, 2024, the United Nations Security Council convened to address the challenges posed by AI in military contexts. Secretary-General António Guterres emphasized that AI's rapid evolution is outpacing current governance frameworks, potentially undermining human control over weapons systems. He called for "international guardrails" to ensure AI's safe and inclusive use. These discussions continue amid reports of widespread autonomous drone and robot use in the ongoing war in Ukraine.

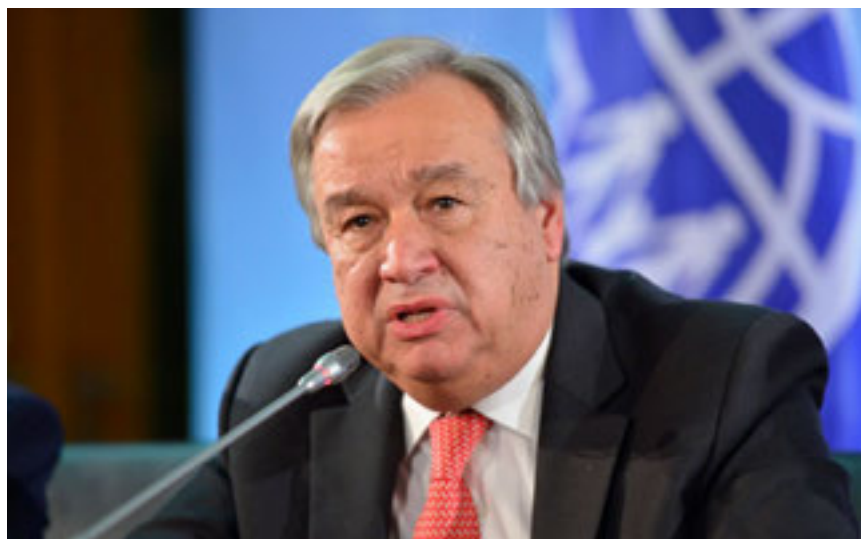

Source: Berkeley Political Review, 2016
Figure 6.1.25

# 6.2 AI and Policymaking

## Global Legislative Records on AI

### Overview

The AI Index analyzed legislation containing the term "artificial intelligence" in 114 countries from 2016 to 2024.[1] Of these, 39 countries have enacted at least one AI-related law (Figure 6.2.1).[2] In total, the countries have passed 204 AI-related laws. Figure 6.2.2 illustrates the annual count of AI-related laws enacted since 2016. The total number of AI-related laws passed rose from 30 in 2023 to 40 in 2024, making 2024 the second-highest year on record after 2022. Since 2016, the number of AI-related laws passed has grown from just one to 40.

**Number of AI-related bills passed into law by country, 2016–24**

Source: AI Index, 2025 | Chart: 2025 AI Index report

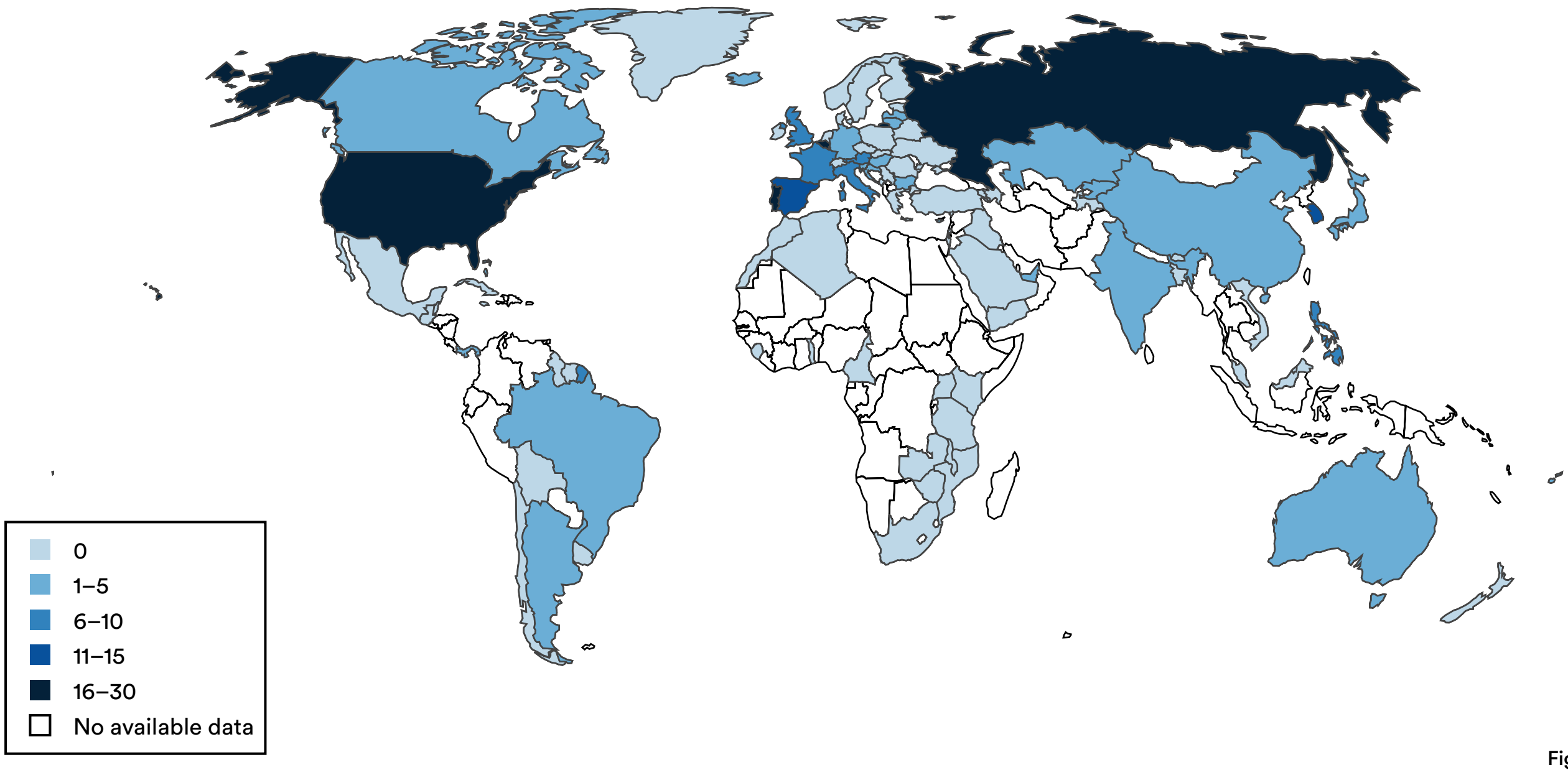

Figure 6.2.1

1 The analysis may undercount the number of actual laws passed, given that large bills that are proposed can include multiple sections related to AI. For example, the National Defense Authorization Act is introduced as a single omnibus bill but includes a collection of smaller bills that were originally proposed individually and later consolidated into one single comprehensive bill.

2 The AI Index monitored AI-related laws passed in Hong Kong and Macao, despite these not being officially recognized countries. Thus, the Index covers a total of 116 geographic areas. Laws passed by Hong Kong and Macao were counted in the overall tally of AI-related laws. This year, the Index decreased its country sample compared to previous years, due to issues accessing the legislative databases of certain nations. As a result, there is a difference between the number of AI-related laws reported this year and those in prior reports.

**Number of AI-related bills passed into law in 116 select geographic areas, 2016–24**
Source: AI Index, 2025 | Chart: 2025 AI Index report

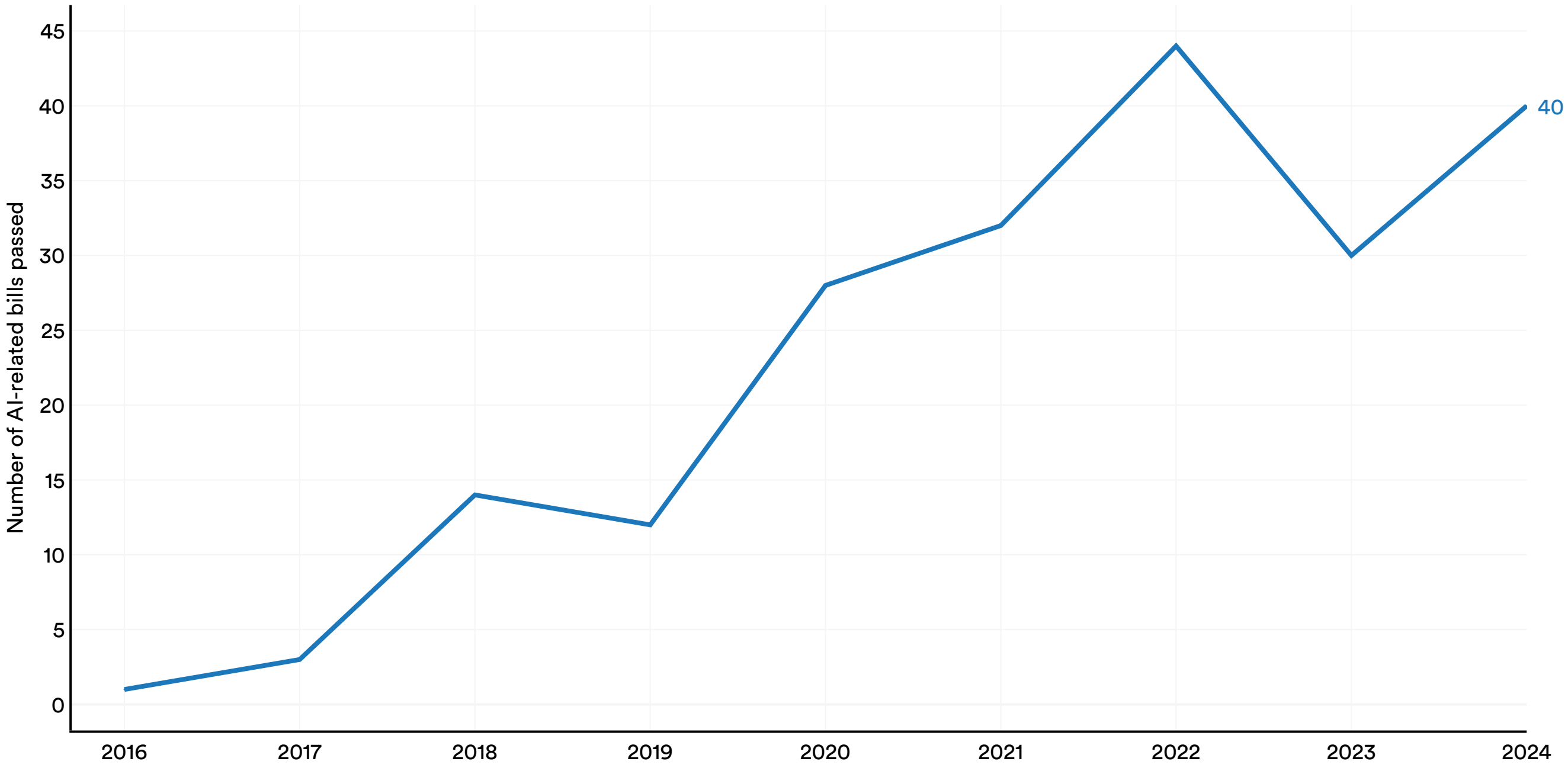

Figure 6.2.2

## By Geographic Area

Figure 6.2.3 highlights the number of AI-related laws enacted in 2024 across the top 15 geographic areas. Russia led with seven laws, followed by Belgium and Portugal with five each. Figure 6.2.4 displays the total number of AI-related laws passed since 2016, with the United States leading at 27, followed by Portugal and Russia, each with 20.[3]

**Number of AI-related bills passed into law in select geographic areas, 2024**
Source: AI Index, 2025 | Chart: 2025 AI Index report

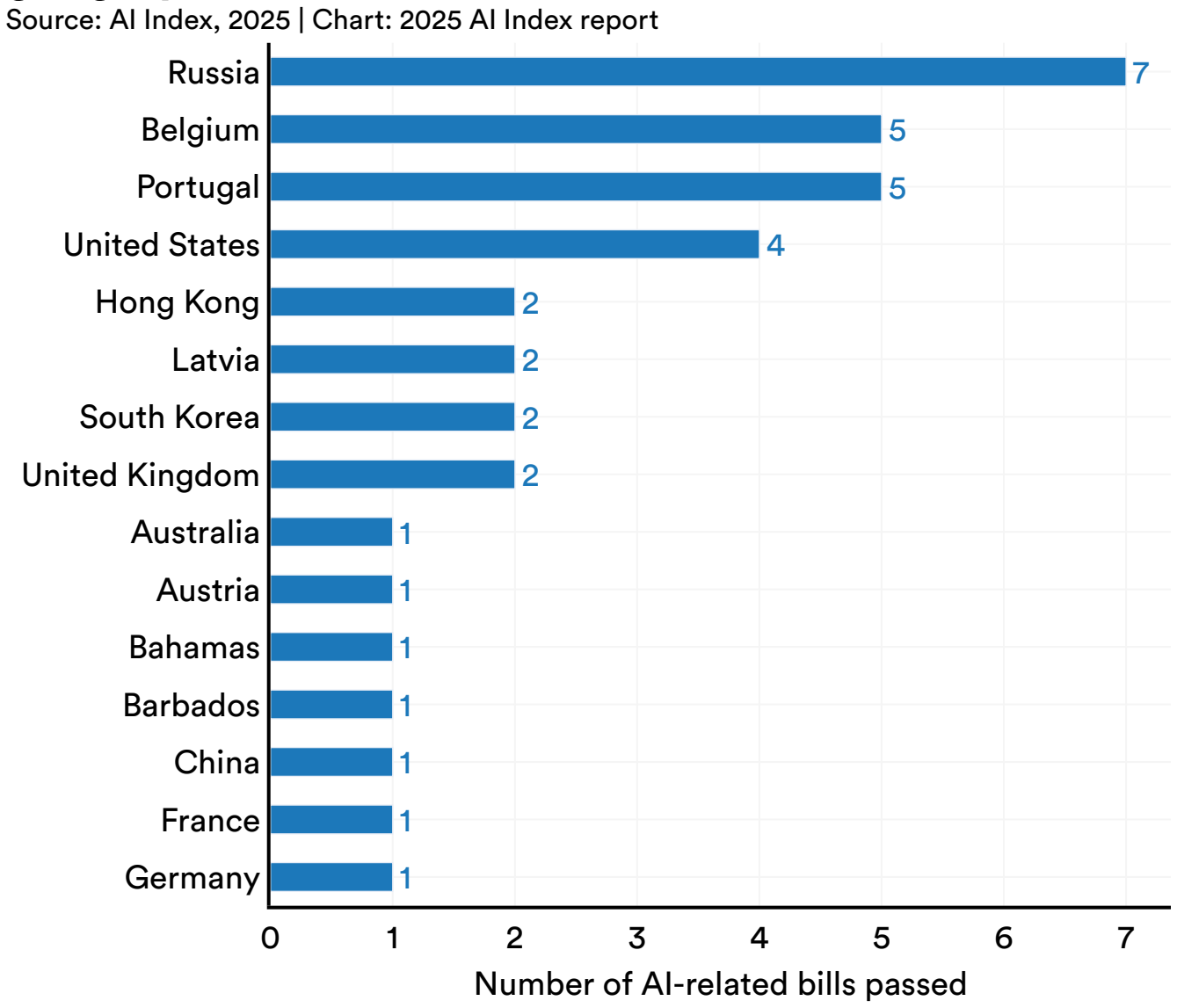

Figure 6.2.3

**Number of AI-related bills passed into law in select geographic areas, 2016–24 (sum)**
Source: AI Index, 2025 | Chart: 2025 AI Index report

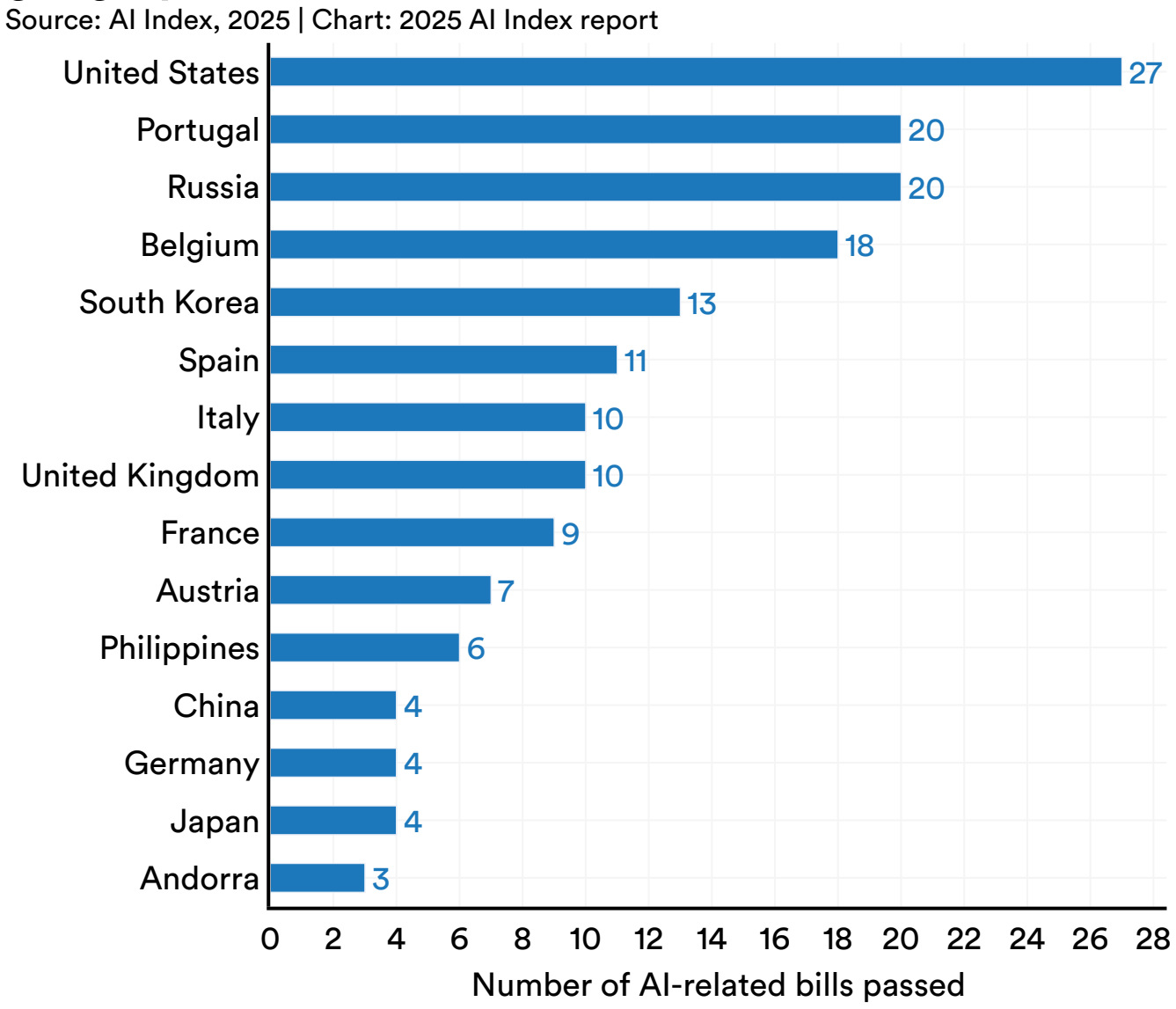

Figure 6.2.4

3 For concision, Figure 6.2.3 and Figure 6.2.4 display data for the top 15 geographic areas by count. Complete country-level totals will be available in the summer 2025 update of the Global AI Vibrancy Tool. For immediate access, please contact the AI Index team.

**Highlight:**

# A Closer Look at Global AI Legislation

The following subsection delves into some of the AI-related legislation passed into law during 2024. Figure 6.2.5 samples five countries' laws covering a range of AI-related issues.

| Country | Bill name | Description |
|---|---|---|
| Austria | Federal law amending the KommAustria Act and the Telecommunications Act 2021 | This act establishes a Service Center for Artificial Intelligence to support, advise, and coordinate AI governance in the media, telecommunications, and postal sectors. It mandates an AI advisory board to monitor AI developments, advise the government, and help shape national AI policy. The Service Center must maintain an information portal on AI projects, particularly publicly funded ones. It also provides guidance on AI regulation, cybersecurity, and compliance. To fund these activities, €700,000 is allocated annually, with future adjustments based on inflation. |
| Belgium | Royal decree establishing an orientation committee on artificial intelligence | This act creates a federal AI steering committee to advise the government on AI-related policies and serve as the primary point of contact for AI governance. The committee, composed of representatives from ministries and public institutions, meets regularly to provide recommendations and coordinate AI policy across Belgium. |
| France | LAW No. 2021-1382 of October 25, 2021, relating to the regulation and protection of access to cultural works in the digital age[4] | This law establishes the Regulatory Authority for Audiovisual and Digital Communication (ARCOM) by merging the Higher Audiovisual Council (CSA) and the High Authority for the Distribution of Works and the Protection of Rights on the Internet (HADOPI). It strengthens measures against online piracy and enhances the regulation of digital platforms to safeguard access to cultural content in the digital space. The law also references artificial intelligence as a tool ARCOM can use to monitor and regulate digital platforms, particularly for detecting copyright infringements and combating online piracy. |
| Latvia | Amendments to the Pre-election Campaigning Law | This act regulates the use of AI in political advertising, requiring clear disclosure for AI-generated content in paid campaign materials. It also bans the use of automated systems with fake or anonymous social media profiles for election campaigns. |
| Russia | On Amendments to the Federal Law "On Personal Data" and the Federal Law "On Conducting an Experiment to Establish Special Regulations for Creating Necessary Conditions for the Development and Implementation of Artificial Intelligence Technologies in the Constituent Entity of the Russian Federation – the Federal City of Moscow," and on Amendments to Articles 6 and 10 of the Federal Law "On Personal Data" | This act establishes a framework for processing and sharing anonymized personal data to support AI development in government operations. It regulates AI-driven decision making, sets security standards for biometric data, and restricts foreign access to sensitive AI-related datasets. |

Figure 6.2.5

4 Law No. 2024-449, passed in 2024, amends Law No. 2021-1382—originally enacted in 2021 and updated in 2024 to include AI—by broadening its scope to cover artificial intelligence and authorizing ARCOM to utilize AI.

# US Legislative Records

## Federal Level

Figure 6.2.6 illustrates the total number of passed versus proposed AI-related bills in the U.S. Congress and demonstrates a significant increase in proposed legislation.[5] In the last year, the count of proposed AI-related bills continued to rise, increasing from 171 in 2023 to 221 in 2024. Since 2022, the number of proposed U.S. federal AI-related bills has almost tripled. Still, of all AI-related bills being proposed, relatively few are passed. The significant increase in U.S. AI-related legislative activity likely reflects policymakers' response to the increasing public awareness and capabilities of AI technologies, particularly generative AI.[6]

**Number of congressional AI-related proposed bills and passed laws in the United States, 2016–24**
Source: AI Index, 2025 | Chart: 2025 AI Index report

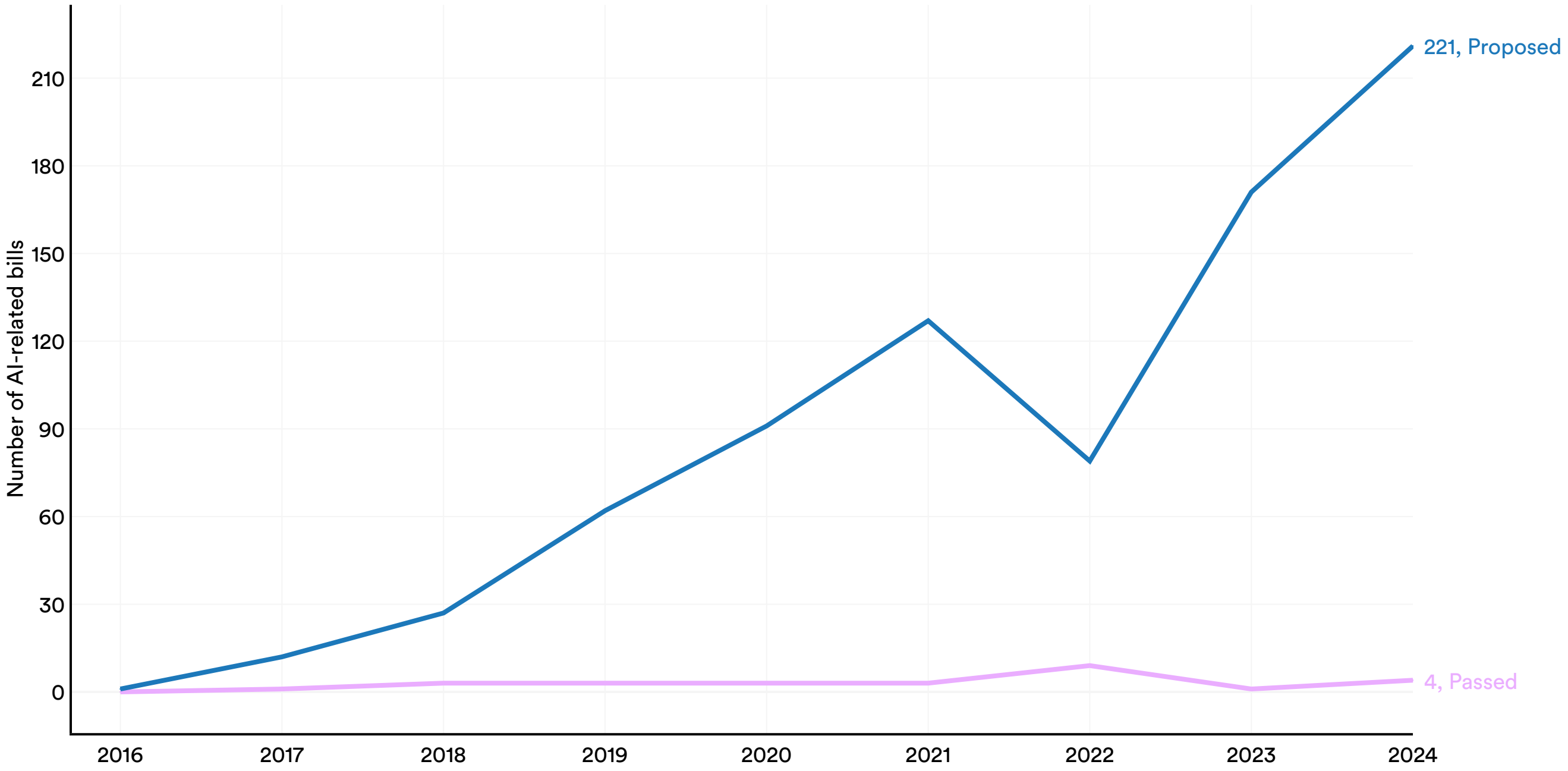

Figure 6.2.6

5 A bill is passed when it successfully clears both chambers of Congress: the House and the Senate.

6 This section covers only congressional bills. However, U.S. AI policymaking extends beyond Congress to other bodies, including the Executive Branch—such as President Donald Trump's Stargate announcement—and rules coming from regulatory agencies like the FTC.

## State Level

The AI Index also tracks data on the enactment of AI-related legislation at the state level. Figure 6.2.7 highlights the number of AI-related laws enacted by U.S. states in 2024. According to the AI Index tracking methodology, California leads with 22 laws, followed by Utah with 12 and Maryland with eight. Figure 6.2.8 displays the total amount of legislation passed by states from 2016 to 2024. California again tops the ranking with 42 bills, followed by Maryland (17), Virginia (17), and Utah (17).

**Number of AI-related bills passed into law in select US states, 2024**
Source: AI Index, 2025 | Chart: 2025 AI Index report

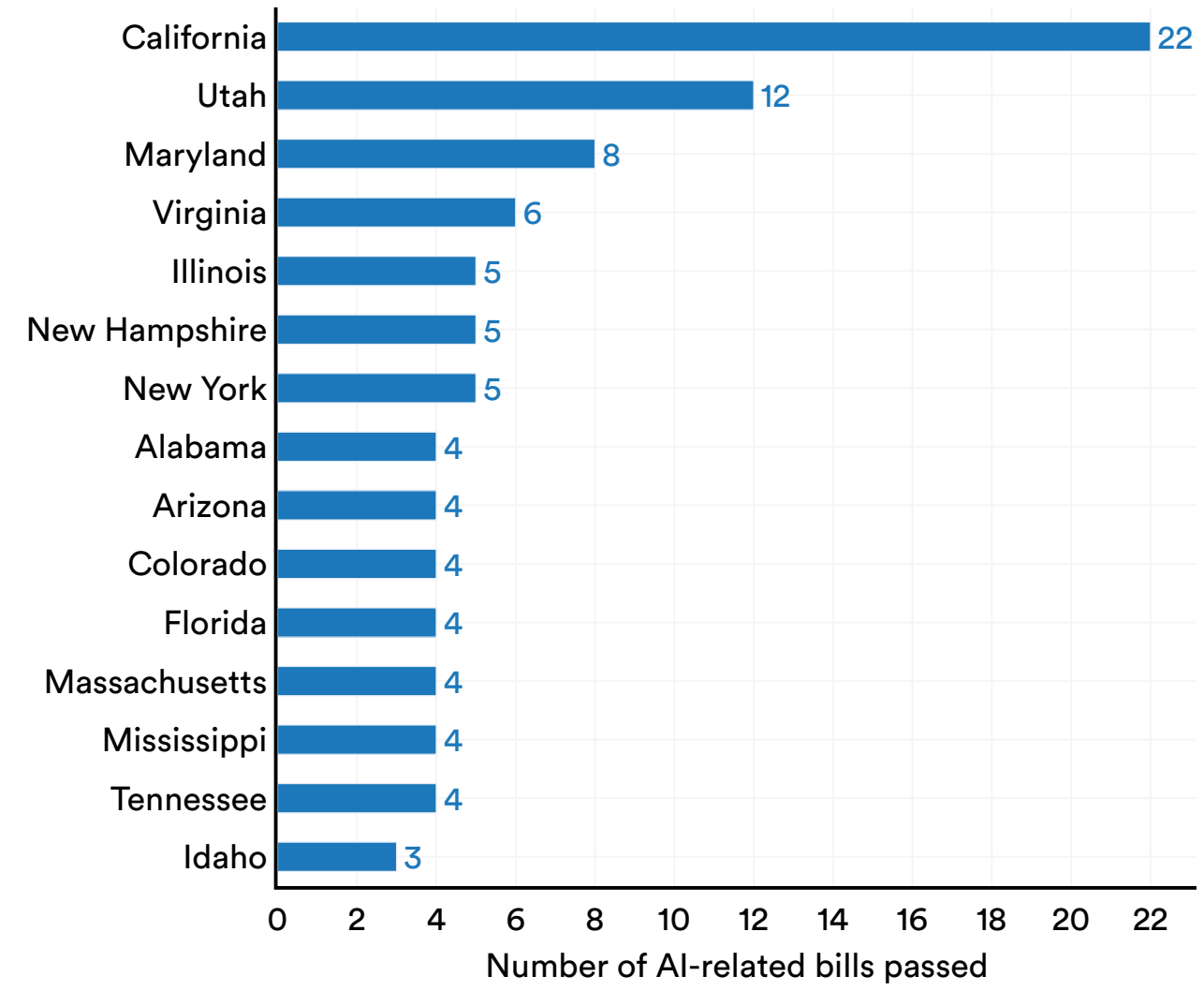

Figure 6.2.7

**Number of state-level AI-related bills passed into law in the United States by state, 2016–24 (sum)**
Source: AI Index, 2025 | Chart: 2025 AI Index report

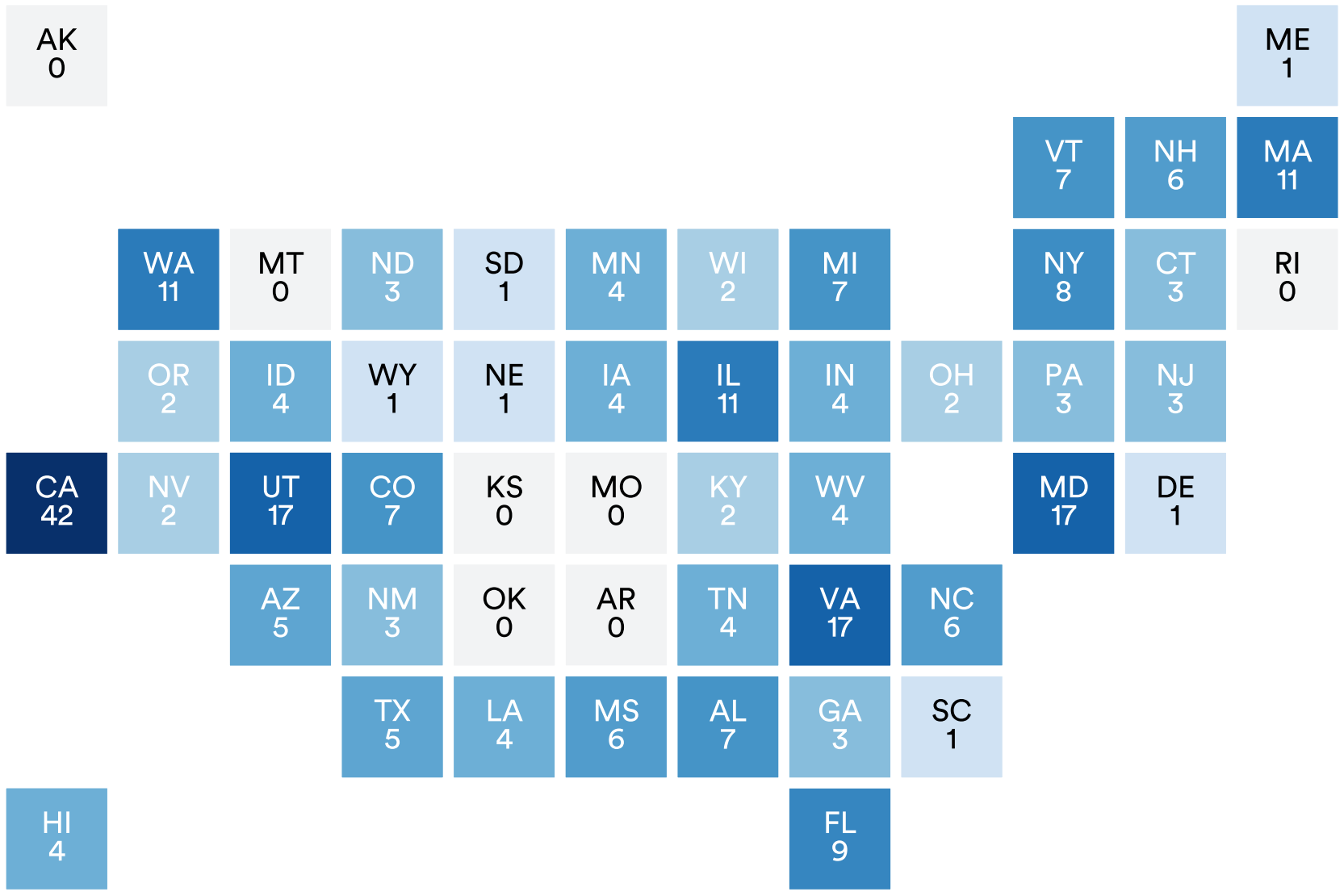

Figure 6.2.8

Since 2016, the number of state-level AI-related laws has rapidly increased. Only one such bill was passed in 2016, rising to 49 by 2023. In the past year alone, that number more than doubled to 131 (Figure 6.2.9).

**Number of AI-related bills passed into law by all US states, 2016–24**
Source: AI Index, 2025 | Chart: 2025 AI Index report

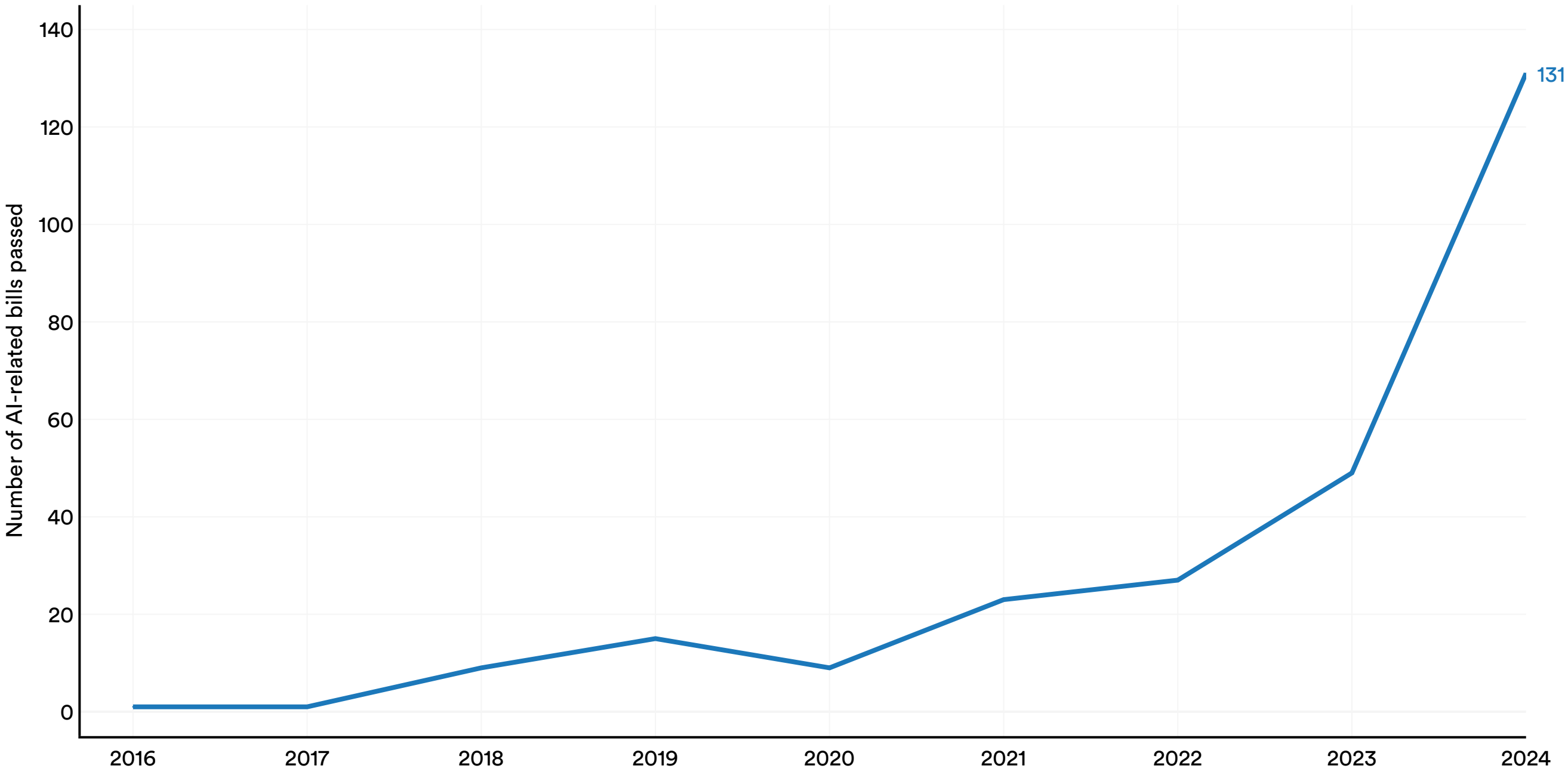

Figure 6.2.9

**Highlight:**

# A Closer Look at State-Level AI Legislation

The following subsection highlights some of the AI-related legislation passed into law at the state level during 2024. The Index profiles legislation from states like California and New York, major hubs for AI companies, alongside states like Alabama and Colorado, which play a smaller role in the industry. This approach highlights the diverse concerns shaping AI legislation at the state level (Figure 6.2.10).

| State | Bill name | Description |
|---|---|---|
| Alabama | Relating to elections; to provide that distribution of materially deceptive media is a crime | This bill prohibits the distribution of AI-generated deceptive media within 90 days of an election if intended to mislead voters or harm a candidate, with penalties ranging from a misdemeanor to a felony for repeat offenses. Exceptions apply for media with clear disclaimers, news reporting, and satire, while violations can result in misdemeanor or felony charges, and affected parties may seek legal action. |
| California | California AI Transparency Act | This act requires large AI providers to offer free AI detection tools and ensure AI-generated content includes clear, permanent disclosures. Violations result in a $5,000 fine per instance, with enforcement by the attorney general or local authorities. |
| Colorado | Consumer Protections for Artificial Intelligence[7] | This bill establishes consumer protections for interactions with high-risk AI systems, requiring developers and deployers to prevent algorithmic discrimination. AI systems must provide transparency, allow consumers to correct or appeal AI-driven decisions, and undergo regular impact assessments. |
| Massachusetts | An Act to Provide for the Future Information Technology Needs of Massachusetts | This act allocates $1.26 billion to modernize information technology, cybersecurity, and broadband infrastructure across Massachusetts. It includes $25 million to integrate AI and machine learning into state government operations, enhancing automation, efficiency, and cybersecurity. |
| New York | An Act to Amend the General Business Law, in Relation to Requiring Disclosure of Certain Social Media Terms of Service | This act requires social media companies to publicly disclose their terms of service for each platform they own or operate in a clear and accessible manner. It also mandates submitting terms of service reports to the attorney general and imposes penalties for noncompliance. |

Figure 6.2.10

7 This bill is colloquially known as the "Colorado AI Act."

**Highlight:**

## Anti-deepfake Policymaking

States in the U.S. have been particularly active in passing legislation to combat deepfakes. A deepfake is AI-generated synthetic media that manipulates or replaces a person's likeness in video, audio, or images, often creating realistic but deceptive content. Deepfakes can be used to manipulate election outcomes, as discussed in Chapter 3 of this year's AI Index, or to generate explicit images. The nonprofit Public Citizen maintains a database tracking AI deepfake regulations, covering both election-related misuse and intimate image misuse. Figure 6.2.11 illustrates the number of state-level laws passed in the United States over time, encompassing anti-deepfake regulations related to elections and intimate images.[8] Figure 6.2.12 highlights when states enacted laws to regulate AI deepfakes in elections. Before 2024, five states—California, Washington, Texas, Michigan, and Minnesota—had passed such laws. In 2024, 12 more states, including Oregon, New Mexico, and New York, introduced similar regulations.

State-level regulations against intimate deepfakes are far more widespread than those against election misuse. A total of 25 states have enacted laws covering all individuals, while five states have passed regulations that apply only to minors (Figure 6.2.13). Wyoming and Ohio are the only states yet to implement any form of intimate deepfake regulation.

**Number of state-level laws enacted on AI-generated deepfakes in intimate imagery and elections in the United States, 2019–24**

Source: Public Citizen, 2025 | Chart: 2025 AI Index report

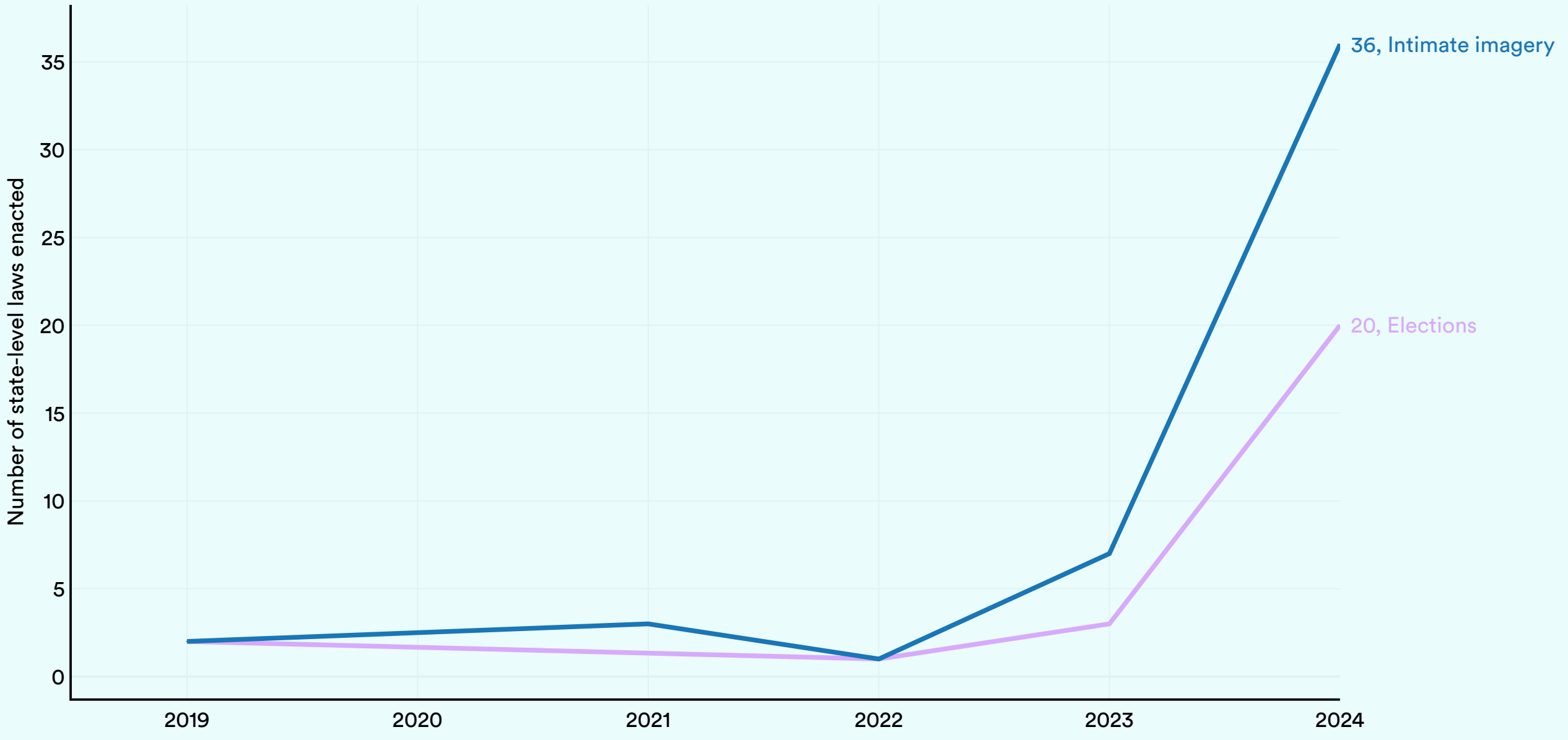

Figure 6.2.11

8 In some cases, the AI Index could not verify the enactment dates of certain state-level AI-related anti-deepfake laws tracked by Public Citizen. Figure 6.2.11 includes only those bills with confirmed passage dates.

**Highlight:**

# Anti-deepfake Policymaking (cont'd)

**State-level laws regulating AI-generated deepfakes in elections in the US by state and status as of 2024**

Source: Public Citizen, 2025 | Chart: 2025 AI Index report

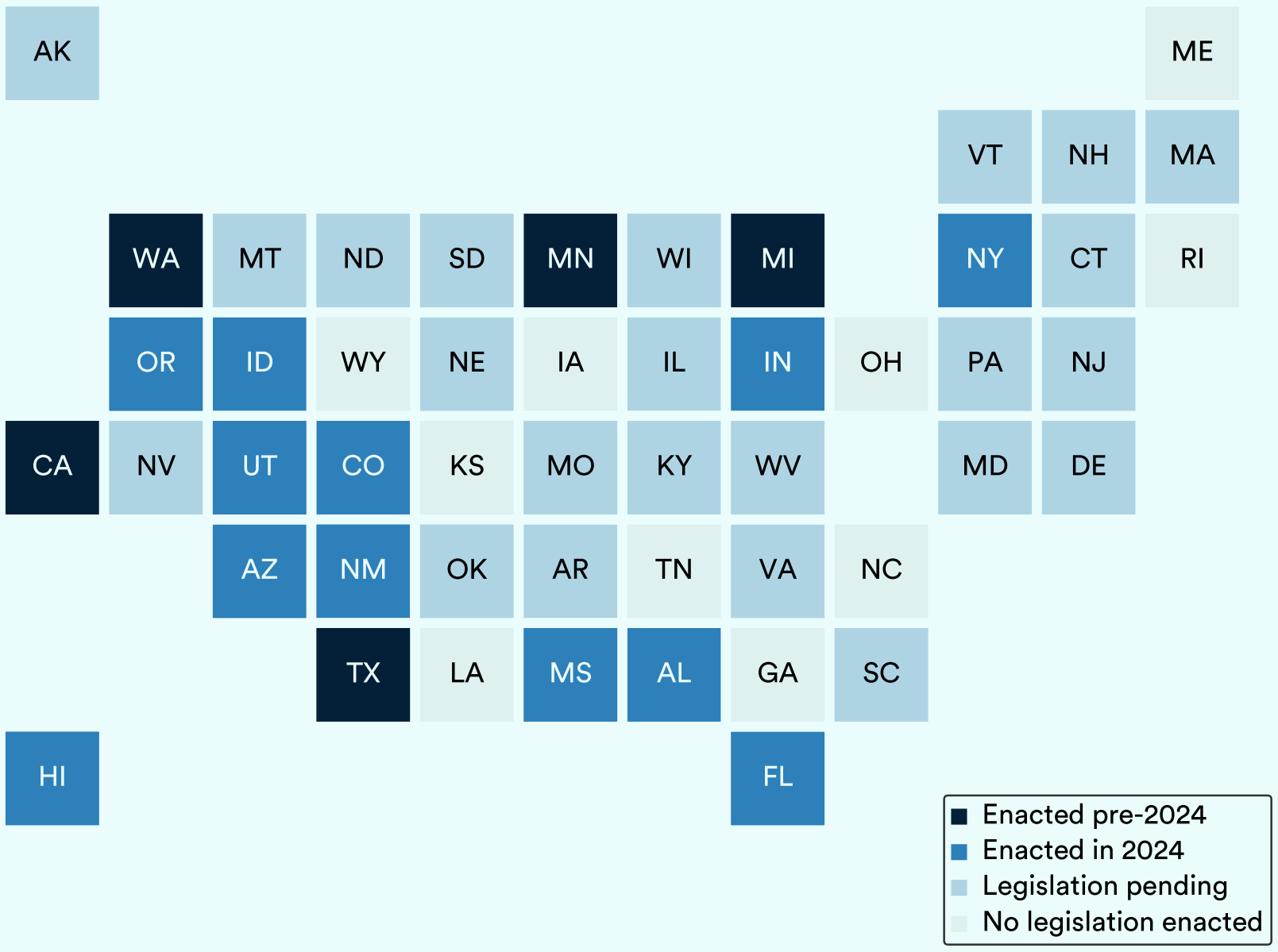

Figure 6.2.12

**State-level laws regulating AI-generated deepfakes in intimate imagery in the US by state and status as of 2024**

Source: Public Citizen, 2025 | Chart: 2025 AI Index report

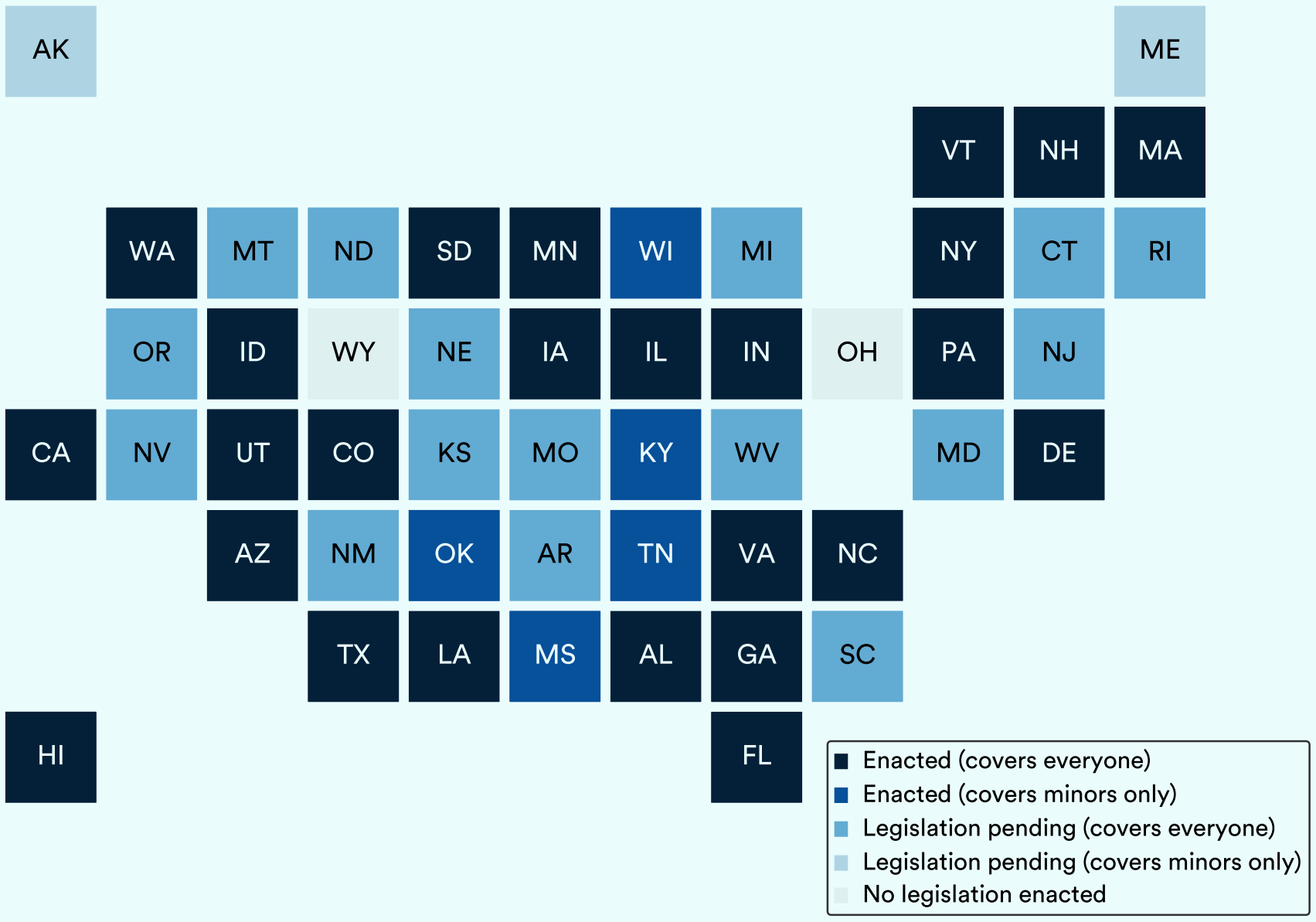

Figure 6.2.13

# Global AI Mentions

Another barometer of legislative interest is the number of mentions of artificial intelligence in governmental and parliamentary proceedings. The AI Index conducted an analysis of the minutes or proceedings of legislative sessions in 73 countries that contain the keyword "artificial intelligence" from 2016 to 2024.[9]

## Overview

Figure 6.2.14 shows the total number of legislative sessions worldwide that have mentioned AI since 2016. In the past year, AI mentions rose by 21.3%, increasing from 1,557 in 2023 to 1,889. Since 2016, the total number of AI mentions has grown more than ninefold.

**Number of mentions of AI in legislative proceedings in 75 select geographic areas, 2016–24**
Source: AI Index, 2025 | Chart: 2025 AI Index report

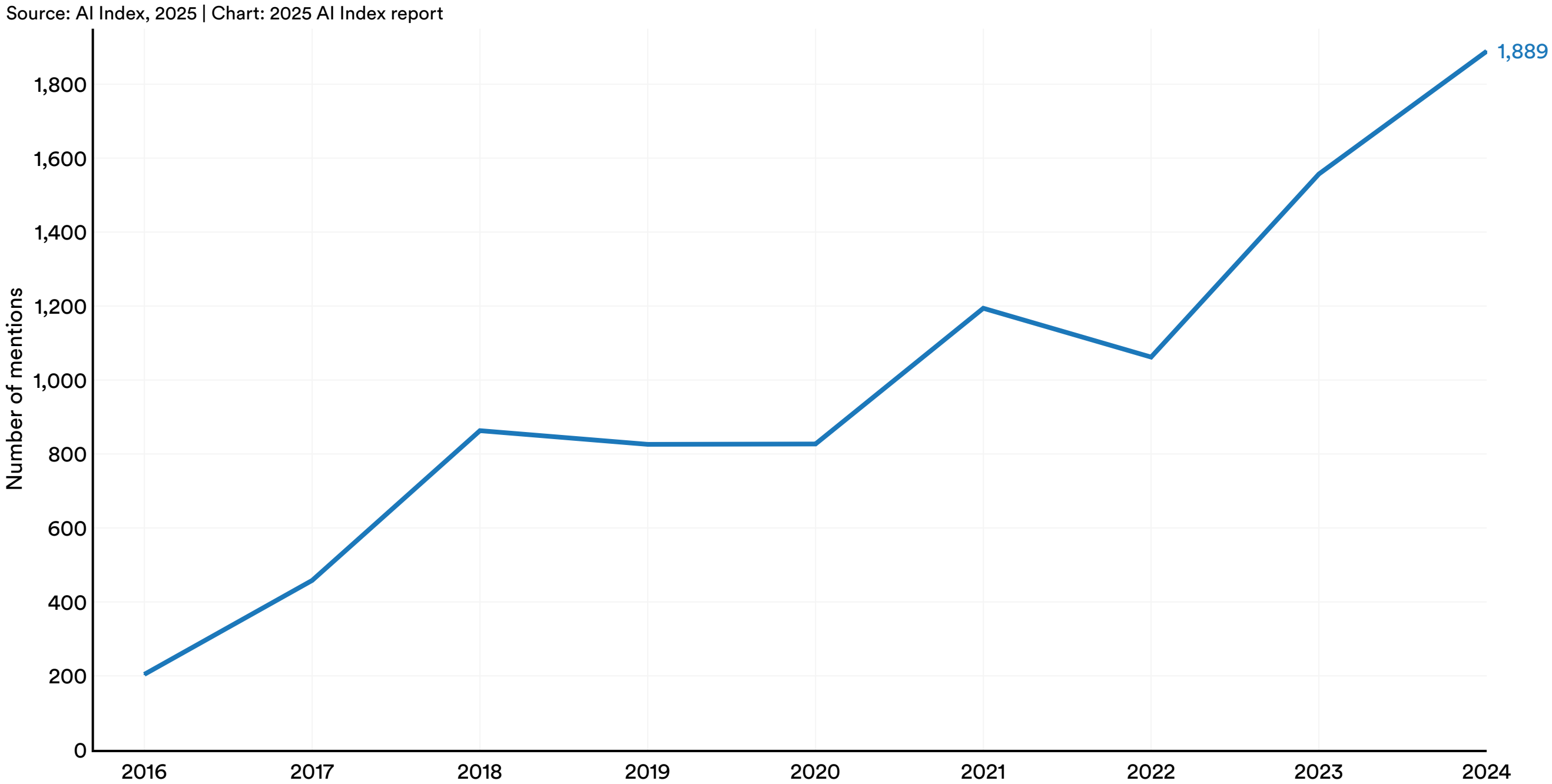

Figure 6.2.14

9 The full list of analyzed countries is available in the Appendix. The AI Index research team aimed to review governmental and parliamentary proceedings worldwide, but publicly accessible databases were not available for all countries. This year, the Index slightly adjusted its tracking methodology, resulting in minor differences from previous totals. More specifically, mentions are counted by session, so multiple mentions of AI in the same legislative session count as one mention. The full methodology is detailed in the Appendix. Additionally, the AI Index tracked mentions in Macao and Hong Kong. While not officially countries, their mentions were included in the tally presented in Figure 6.2.14. In total, the Index tracked AI mentions across 75 geographic areas.

In 2024, Spain led in AI mentions within its legislative proceedings (314), followed by Ireland (145) and Australia (123) (Figure 6.2.15). Of the 75 geographic areas analyzed, 57 referenced AI in at least one legislative proceeding in 2024.

### Number of mentions of AI in legislative proceedings by country, 2024

Source: AI Index, 2025 | Chart: 2025 AI Index report

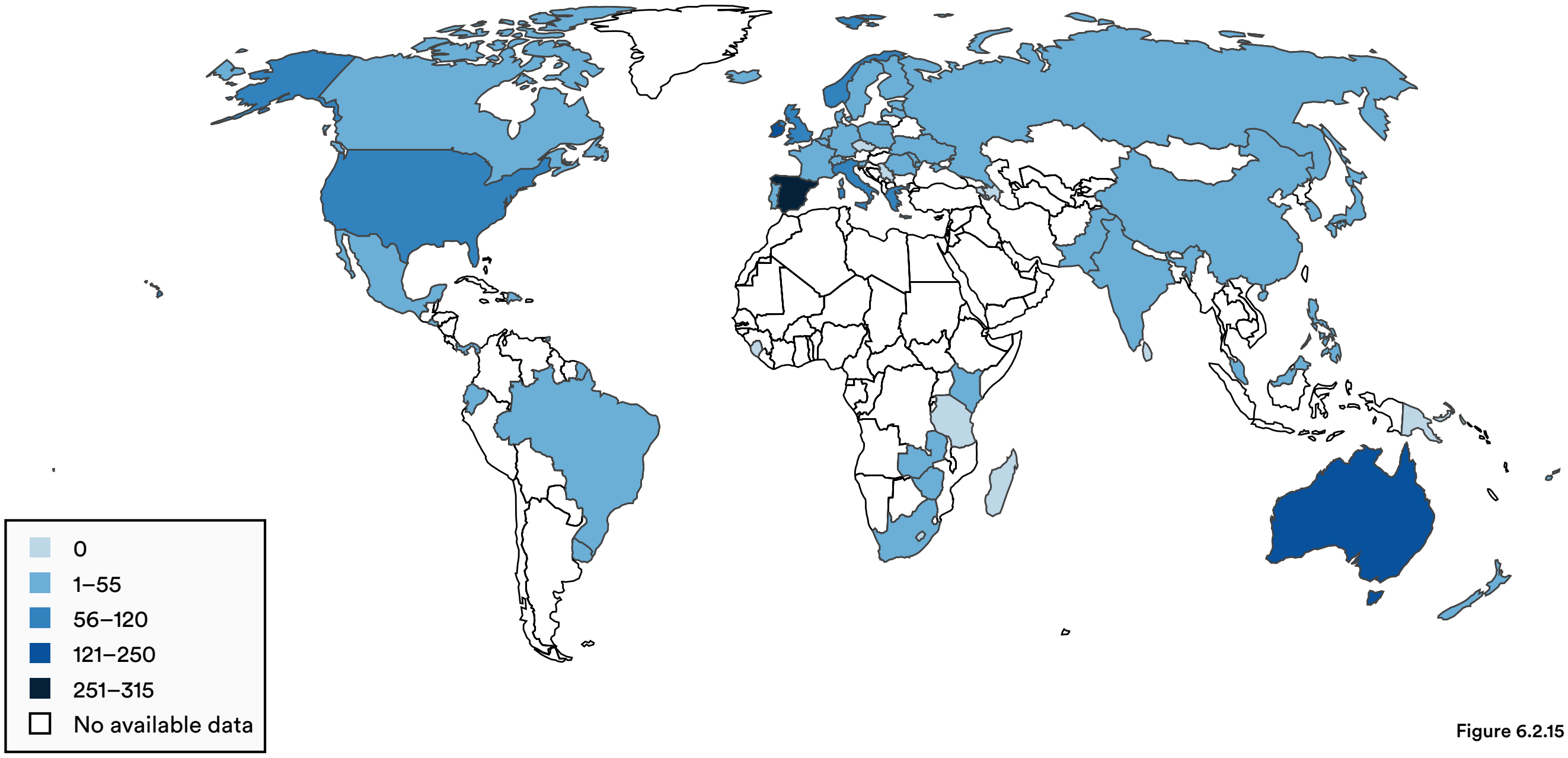

Figure 6.2.15

When legislative mentions are aggregated from 2016 to 2024, a somewhat similar trend emerges (Figure 6.2.16). Spain is first with 1,200 mentions, followed by the United Kingdom (710) and Ireland (659).

### Number of mentions of AI in legislative proceedings by country, 2016–24 (sum)

Source: AI Index, 2025 | Chart: 2025 AI Index report

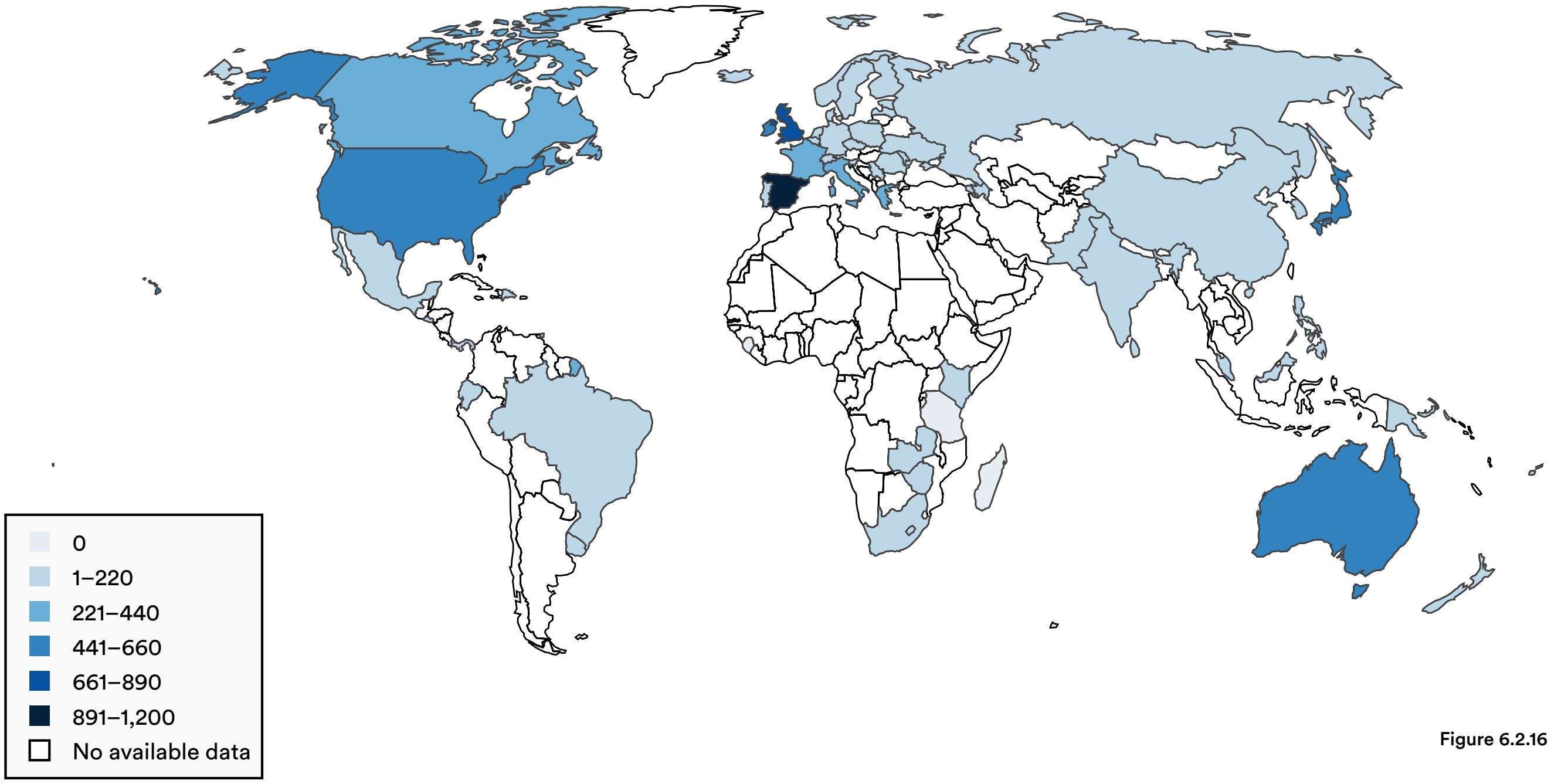

Figure 6.2.16

Drawing on data from select countries, Figure 6.2.17 compares AI mentions in parliamentary discussions with the number of AI-related bills passed. In general, greater parliamentary discussion of AI correlates with more AI legislation—although some countries, such as Belgium, Portugal, and Russia, deviate from this trend.

**Mentions of AI in legislative proceedings vs. AI-related bills passed into law in select countries, 2016–24**
Source: AI Index, 2025 | Table: 2025 AI Index report

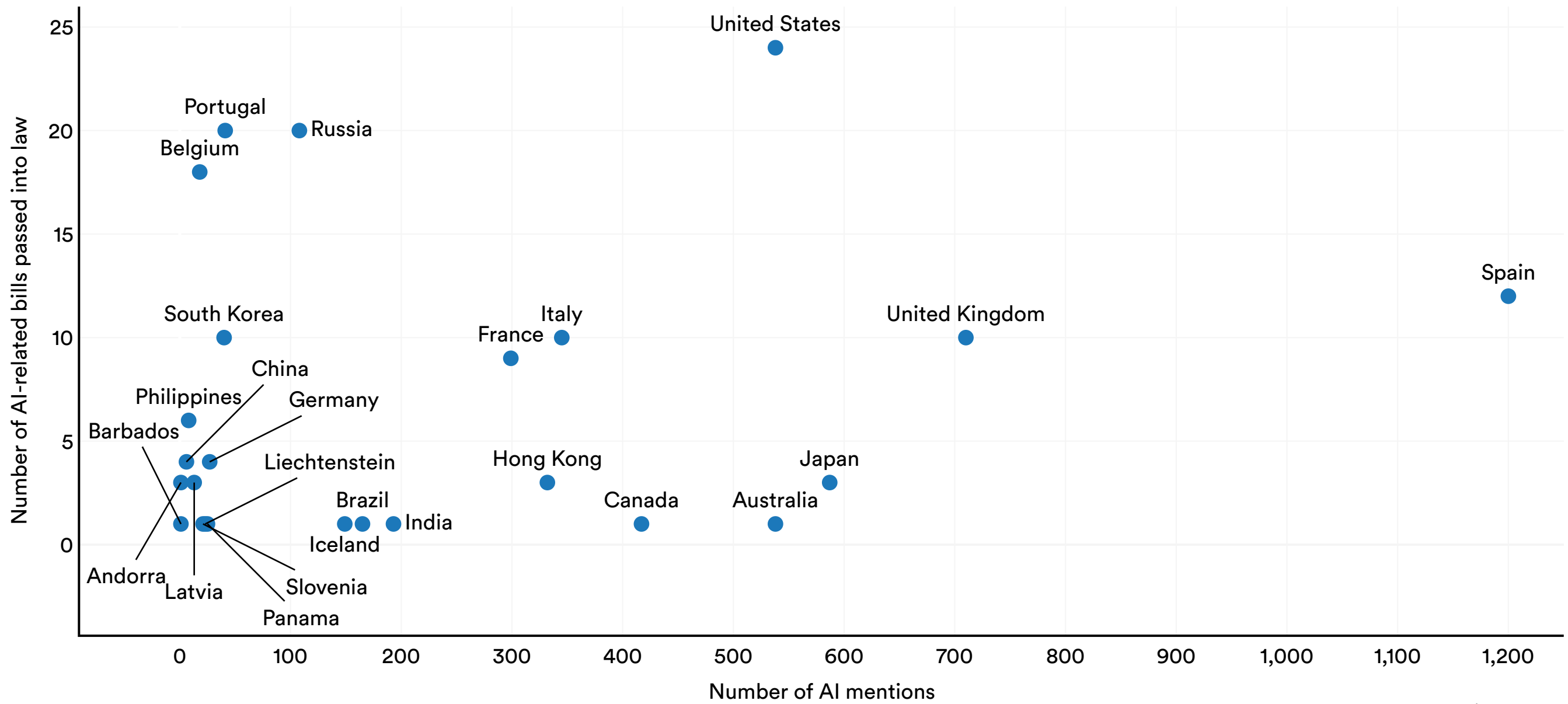

Figure 6.2.17

## US Committee Mentions

Mentions of artificial intelligence in committee reports by House and Senate committees serve as another indicator of legislative interest in AI in the United States. Typically, these committees focus on legislative and policy issues, investigations, and internal matters.

Figure 6.2.18 tracks AI mentions in U.S. committee reports by legislative session from 2001 to 2024. The 118th session recorded the highest count to date, with 136 mentions—up 83.8% from the 117th session.

### Mentions of AI in US committee reports by legislative session, 2001–24

Source: AI Index, 2025 | Chart: 2025 AI Index report

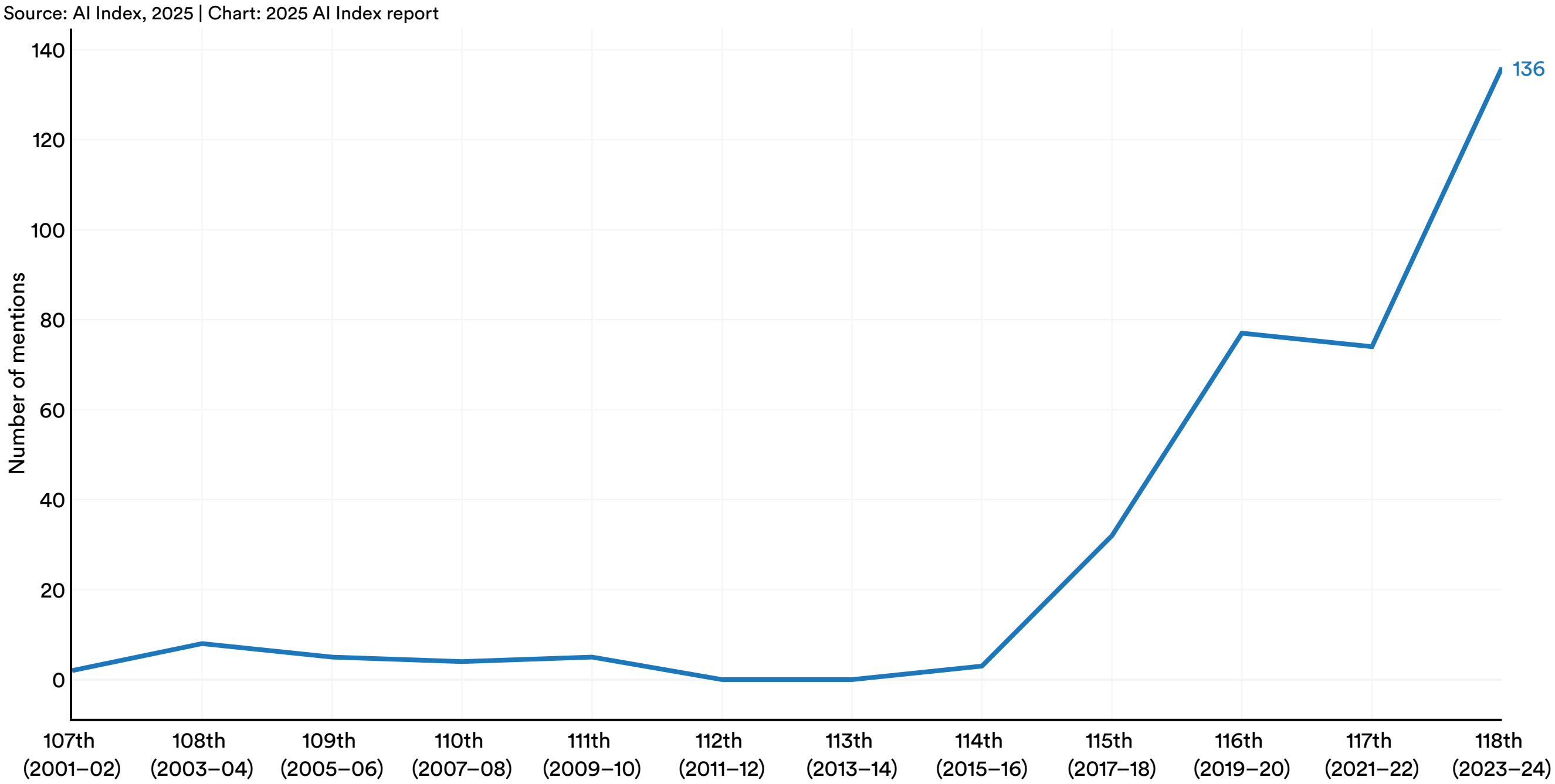

Figure 6.2.18

# US Regulations

The advent of AI has garnered significant attention from regulatory agencies—federal bodies tasked with regulating sectors of the economy and steering the enforcement of laws. This section examines AI regulations within the United States. Unlike legislation, which establishes legal frameworks within nations, regulations are detailed directives crafted by executive authorities to enforce legislation. In the United States, prominent regulatory agencies include the Environmental Protection Agency (EPA), Food and Drug Administration (FDA), and Federal Communications Commission (FCC). Since the specifics of legislation often manifest through regulatory actions, understanding the AI regulatory landscape is essential to developing a deeper understanding of AI policymaking.

This section examines AI-related regulations enacted by American regulatory agencies between 2016 and 2024. It provides an analysis of the total number of regulations, as well as their topics, scope, regulatory intent, and originating agencies. To compile this data, the AI Index performed a keyword search for "artificial intelligence" on the Federal Register, a comprehensive repository of government documents from nearly all branches of the American government, encompassing more than 436 agencies.

### Overview

The number of AI-related regulations has risen sharply over the past six years, with a particularly noticeable increase in the last year (Figure 6.2.19). In 2024, 59 AI-related regulations were introduced—more than double the 25 recorded in 2023.

**Number of AI-related regulations in the United States, 2016–24**
Source: AI Index, 2025 | Chart: 2025 AI Index report

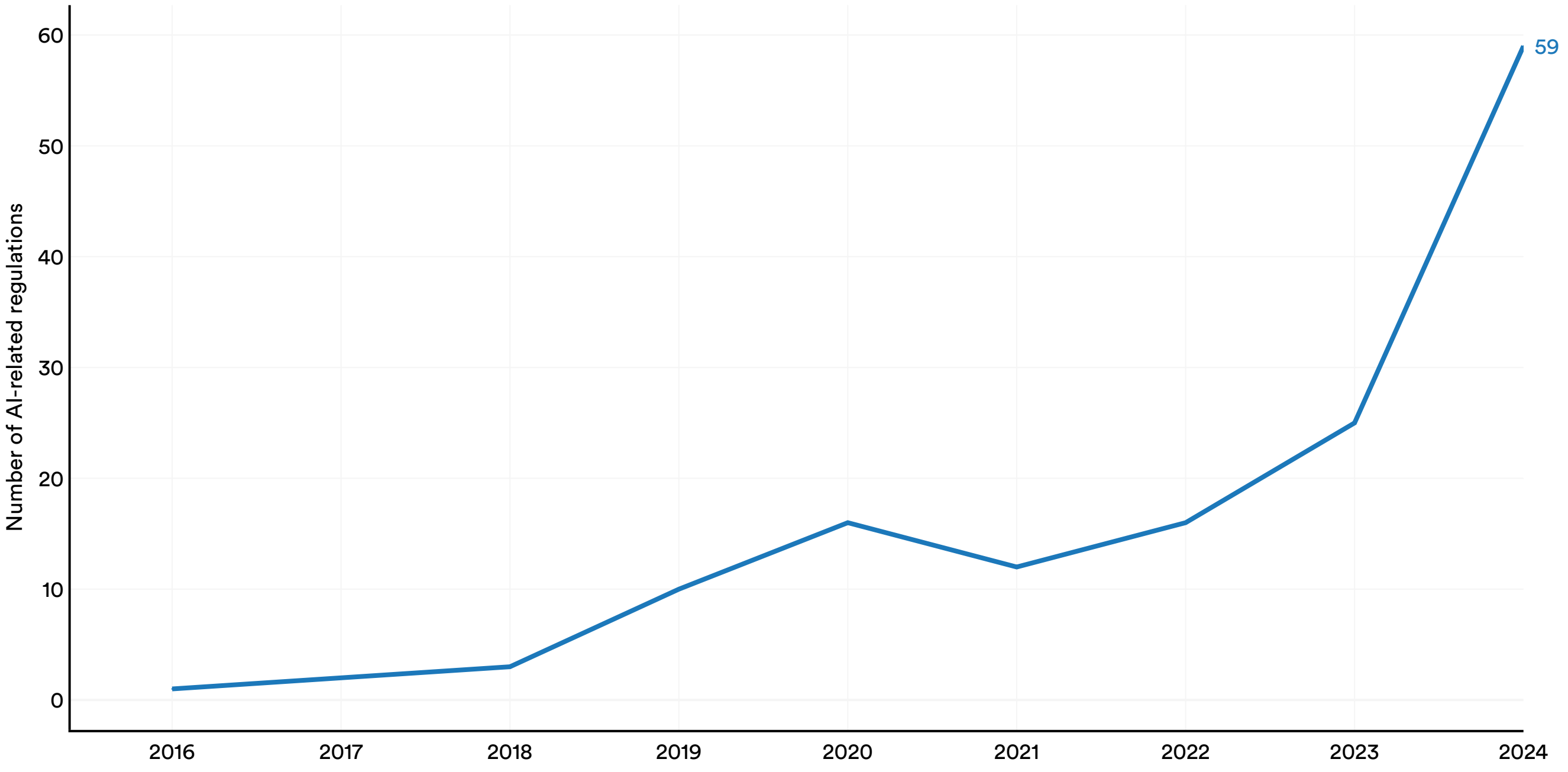

Figure 6.2.19

### By Agency

Figure 6.2.20 looks at the number of AI-related regulations in the United States that have been released by different American regulatory agencies since 2016.[10] In 2024, the Department of Health and Human Services issued the most AI-related regulations (14), followed by the Centers for Medicare and Medicaid Services (7) and the Commerce Department (7). AI regulations came from a record 42 unique departments, up from 21 in 2023 and 17 in 2022. This trend reflects a growing interest in AI across a wider range of U.S. regulators.

10 Regulations can originate from multiple agencies, so the totals in Figure 6.2.20 do not fully align with those in Figure 6.2.19. Figure 6.2.20 refers to departments as agencies, consistent with the terminology used by the Federal Register, the source of the data.

**Number of AI-related regulations in the United States by agency, 2016–24**
Source: AI Index, 2025 | Chart: 2025 AI Index report

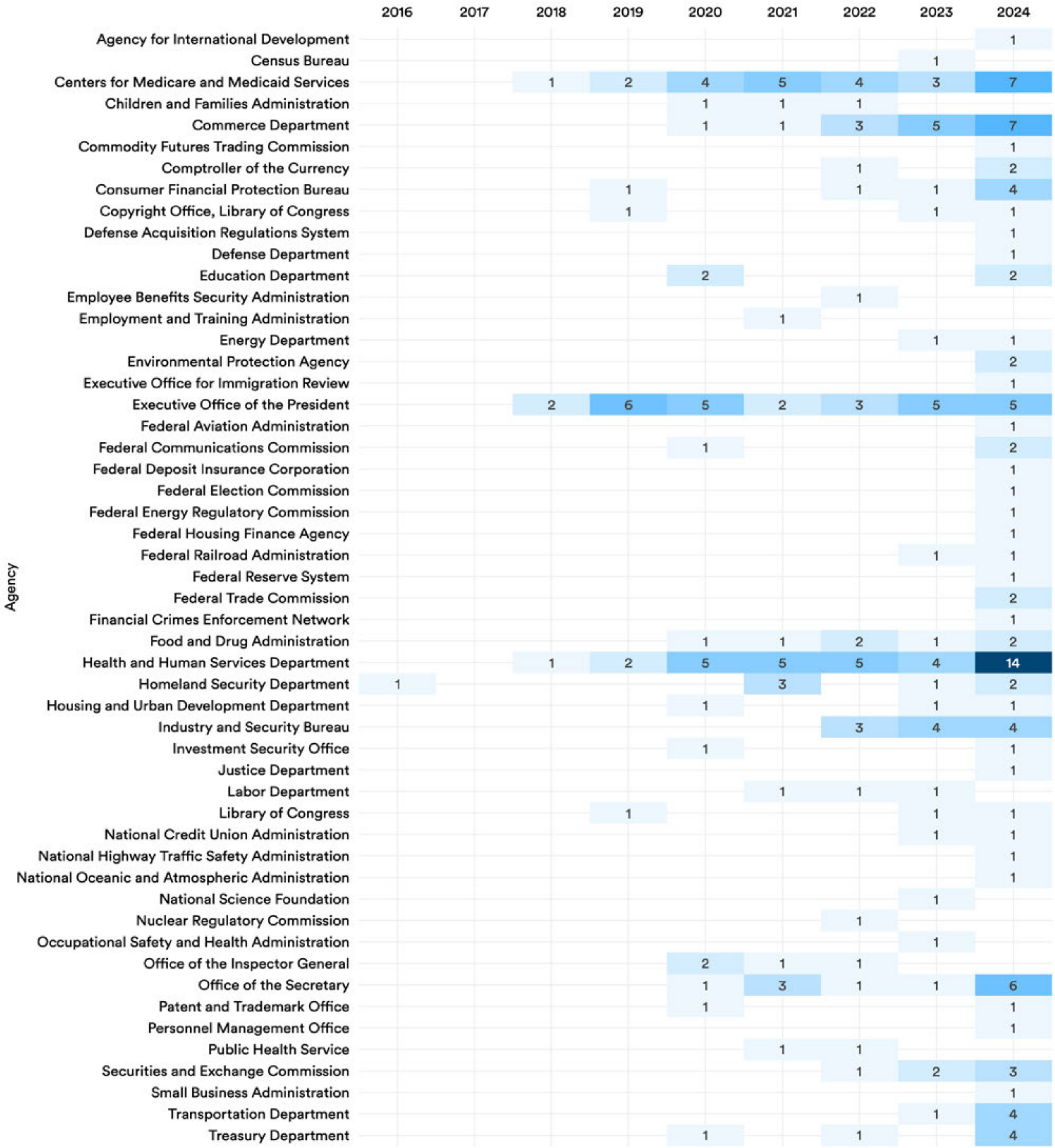

Figure 6.2.20

**Highlight:**

## A Closer Look at US Federal Regulations

The following section highlights some of the AI-related regulations passed as rules and executive orders at the federal level during 2024 (Figure 6.2.21).

| Agency | Regulation | Description |
|---|---|---|
| Executive Office of the President | Preventing Access to Americans' Bulk Sensitive Personal Data and United States Government–Related Data by Countries of Concern | This executive order identifies AI use by countries of concern as a significant national security threat. It specifically warns of foreign adversaries exploiting bulk sensitive personal and U.S. government–related data to refine AI algorithms for espionage, cyber operations, and influencing campaigns. To counter this risk, the order implements measures to safeguard sensitive data, including restrictions or bans on data transactions with these countries and strengthened network infrastructure security. |
| Industry and Security Bureau | Foreign-Produced Direct Product Rule Additions, and Refinements to Control for Advanced Computing and Semiconductor Manufacturing Items | This rule amends the U.S. Export Administration Regulations to tighten controls on semiconductor manufacturing equipment and supercomputer exports, particularly to China. It introduces additional restrictions on semiconductor production, revises existing measures, and implements "Red Flags" to identify risks of unauthorized exports. These changes aim to counter China's efforts to circumvent previous restrictions and limit its ability to develop advanced computing and AI systems that could threaten U.S. national security. |
| Consumer Financial Protection Bureau | Consumer Financial Protection Circular 2024–06: Background Dossiers and Algorithmic Scores for Hiring, Promotion, and Other Employment Decisions | This rule mandates that employers cannot base employment decisions on background dossiers, algorithmic scores, or third-party reports without complying with the Fair Credit Reporting Act. It reinforces key obligations, particularly for AI-driven systems, such as obtaining a worker's consent before procuring a consumer report. By doing so, the rule sets clear limits on the use of algorithmic scoring in hiring and employment decisions. |
| Federal Election Commission | Fraudulent Misrepresentation of Campaign Authority | This interpretive rule offers supplemental guidance on the Federal Election Campaign Act (FECA) in response to the rise of AI-generated content. It reaffirms that FECA is "technology neutral" and focuses on whether a person or entity engages in election-related misrepresentation rather than specifically addressing AI misuse. |
| Office of Investment Security, Department of the Treasury | Provisions Pertaining to U.S. Investments in Certain National Security Technologies and Products in Countries of Concern | This final rule implements Executive Order 14105, mandating that U.S. persons notify the Treasury Department of transactions with entities in countries of concern involved in sensitive technologies that threaten national security. It also prohibits certain transactions with these entities. Issued in 2023, the order targets U.S. investments in high-risk technologies, including AI, semiconductors, and quantum computing, recognizing them as critical sectors where such investments could heighten security threats from adversarial nations. |

Figure 6.2.21

# 6.3 Public Investment in AI[11]

As AI continues to drive innovation in critical sectors such as healthcare, transportation, and defense, public funding has become essential for nations to realize their AI strategies. Understanding how much governments invest in AI research and development (R&D) is important for understanding the broader AI geopolitical landscape, yet tracking these investments presents significant challenges. While national budgets may outline AI-related spending, these allocations do not always translate directly into expenditures. Moreover, AI investments are often embedded within broader scientific or technological initiatives. As a result, pinpointing AI-specific funding can be difficult.

To address this, the AI Index leveraged natural language processing (NLP) techniques to analyze public tenders and contracts and to identify AI-related government spending in countries across the world.[12] Examining tenders provides a more direct measure of investment trends and offers insight into how governments allocate resources over time. Because the AI Index only analyzed countries for which public contract and tenders data was publicly available, some countries could not be analyzed.[13] This section also presents an analysis of total AI grant spending in the United States.

The AI Index cautions against making direct country-to-country comparisons based on the public spending data presented in this section. While this analysis includes data on government contracts from a range of countries, it only covers grant-level spending for the United States. This asymmetry stems from the complexity and difficulty of collecting comparable grant data from other countries and regions, such as the European Union and China. However, as the U.S. case demonstrates, a significant share of government spending on AI occurs through grants. In 2023 alone, the AI Index estimates that the U.S. government awarded approximately $830 million in AI-related public tenders, compared to $4.5 billion in AI-related grants. Given the current limitations in cross-national data availability and consistency, comparative analysis of public AI spending across countries remains premature. This analysis is intended as an initial step toward more comprehensive global coverage. The AI Index is committed to expanding this work and welcomes collaboration from researchers, institutions, and governments interested in improving the scope and quality of this data.

11 The analysis in this section was led by Lapo Santarlasci.

12 The full methodology behind this analytical approach is detailed in the Appendix. Due to reporting lags that may result in incomplete data for 2024, the most up-to-date analysis is available for the end of 2023.

13 Some major government AI contract-granting regions, such as the EU (at the aggregate level) and China, were excluded from this analysis due to data limitations. The AI Index is committed to expanding its scope to include these and other regions in future editions.

# Total AI Public Investments

Figure 6.3.1 summarizes key figures on the number of AI-related contracts and their value at the country level.[14] From 2013 to 2023, the United States was the leading nation, with about $5.2 billion distributed across 2,678 unique AI contracts (Figure 6.3.1 and Figure 6.3.2). In Europe, the United Kingdom, Germany, and France stand out with the highest total contract values awarded, accounting for 56% of European public investments in AI.

**Public spending on AI-related contracts in select countries, 2013–23 (sum)**
Source: AI Index, 2025 | Chart: 2025 AI Index report

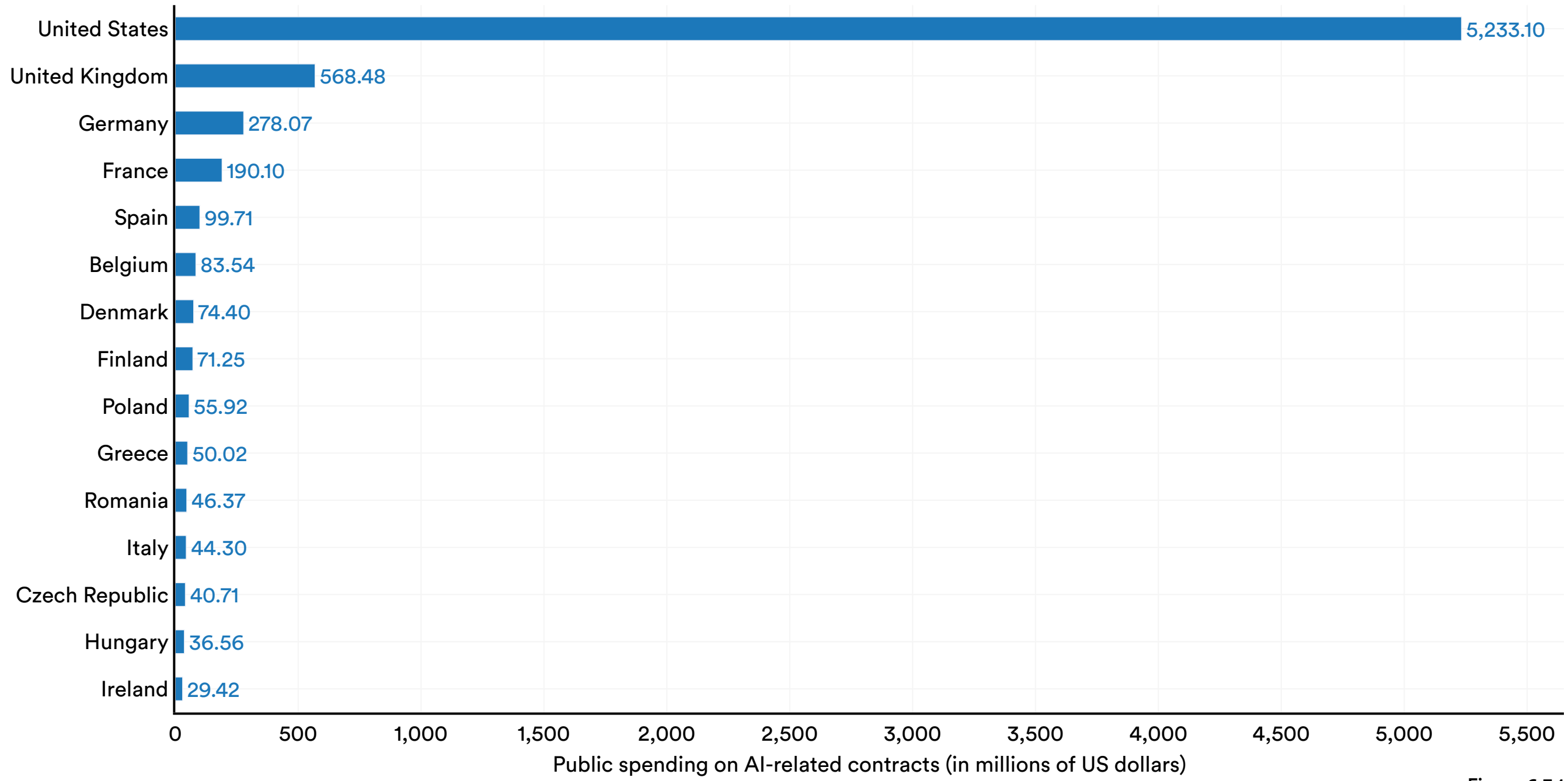

Figure 6.3.1

14 The results and figures presented are subject to missing values ratios of the specific sample of matched tenders: 0.16% for NAICS code, and 26.8% for U.S. dollar values. It is important to note that the sample does not include Northern Ireland tenders, as their offices do not offer an API service or bulk download option for large-scale data collection.

## Number of AI-related contracts in select countries, 2013–23 (sum)

Source: AI Index, 2025 | Chart: 2025 AI Index report

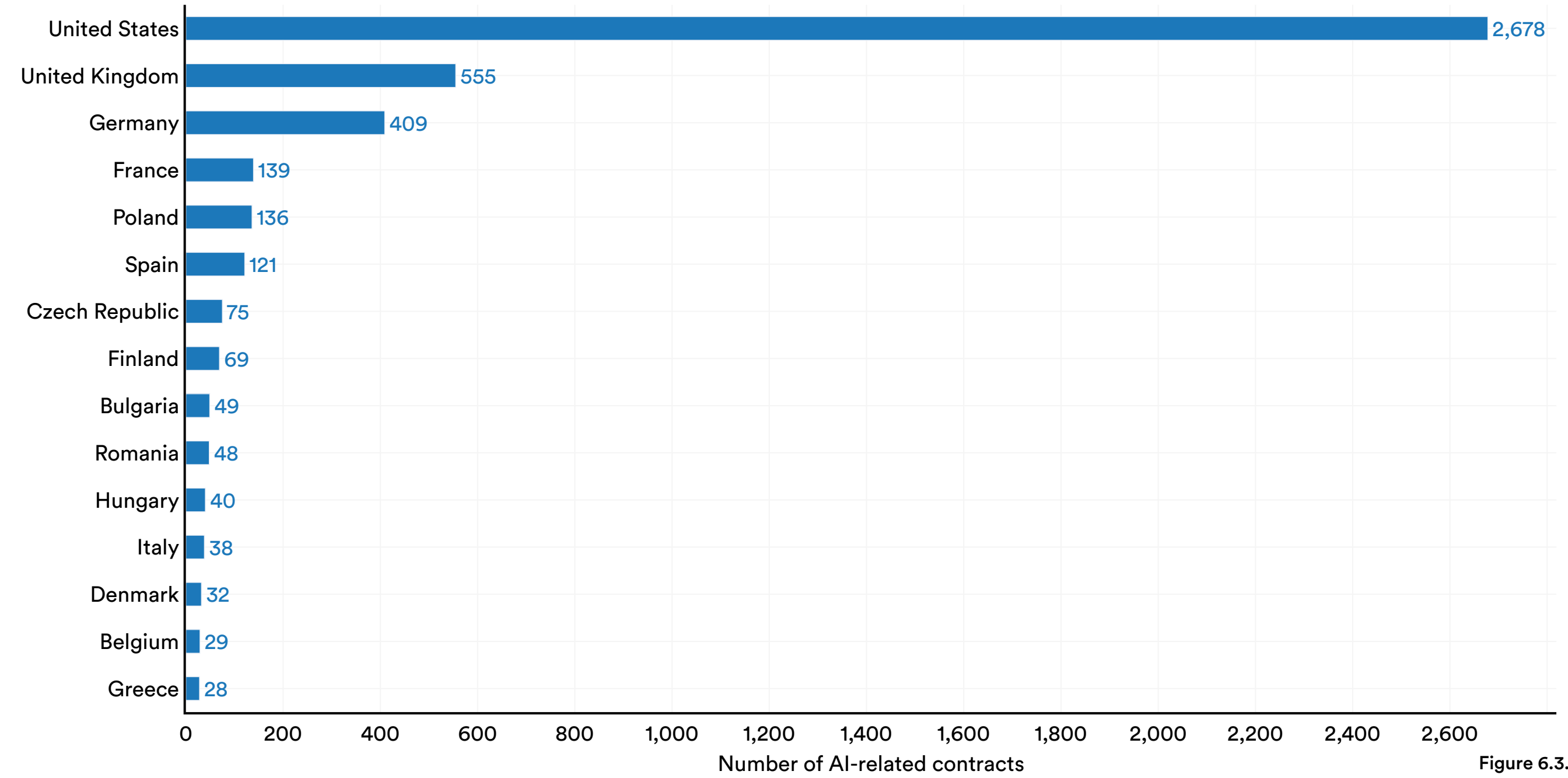

Figure 6.3.2

## Median value of public AI-related contracts in select countries, 2013–23

Source: AI Index, 2025 | Chart: 2025 AI Index report

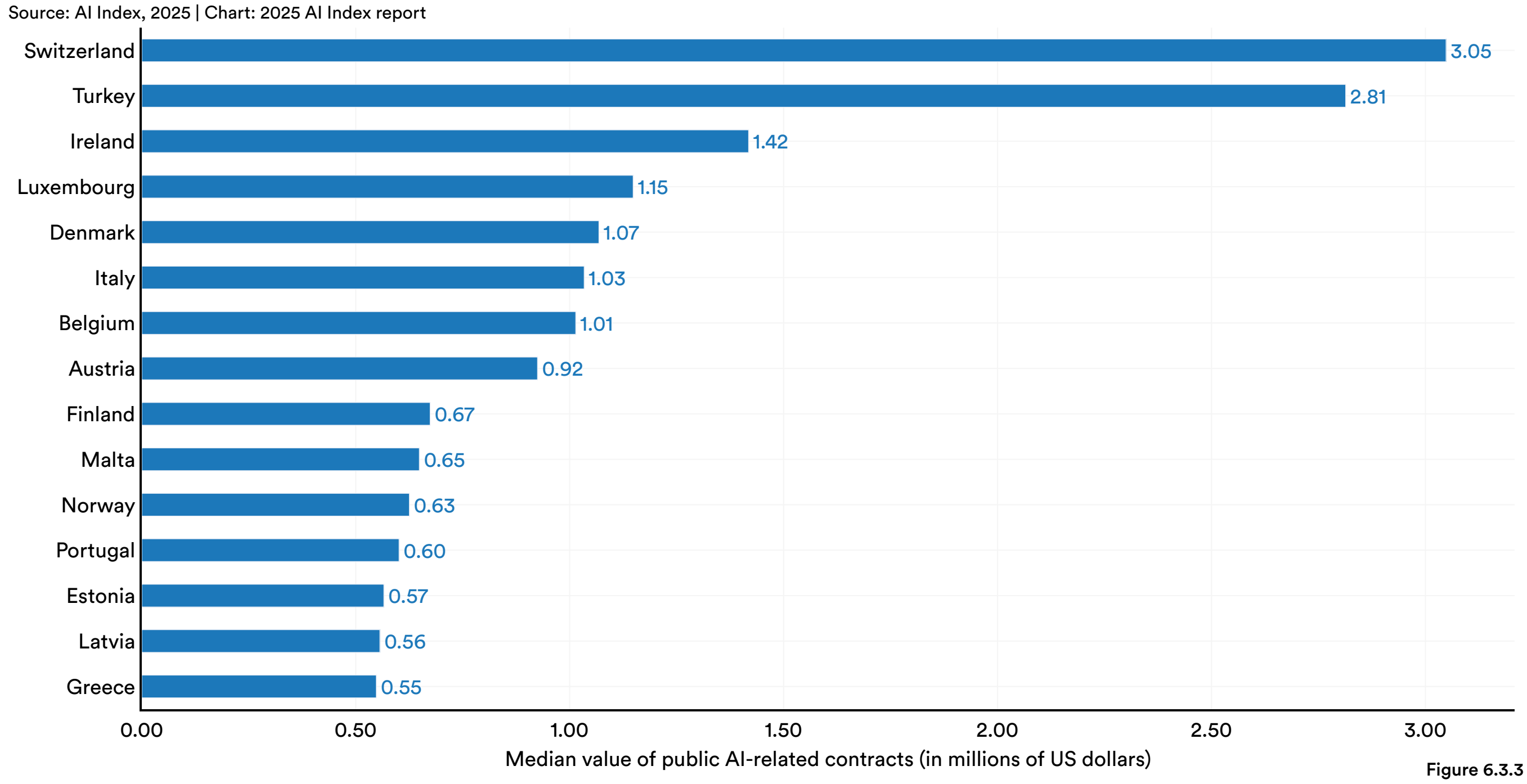

Figure 6.3.3

Which governments spent the most on AI per capita over the past decade? The United States leads with $1.58 million per 100,000 inhabitants, followed by Finland ($1.3 million) and Denmark ($1.3 million) (Figure 6.3.4).

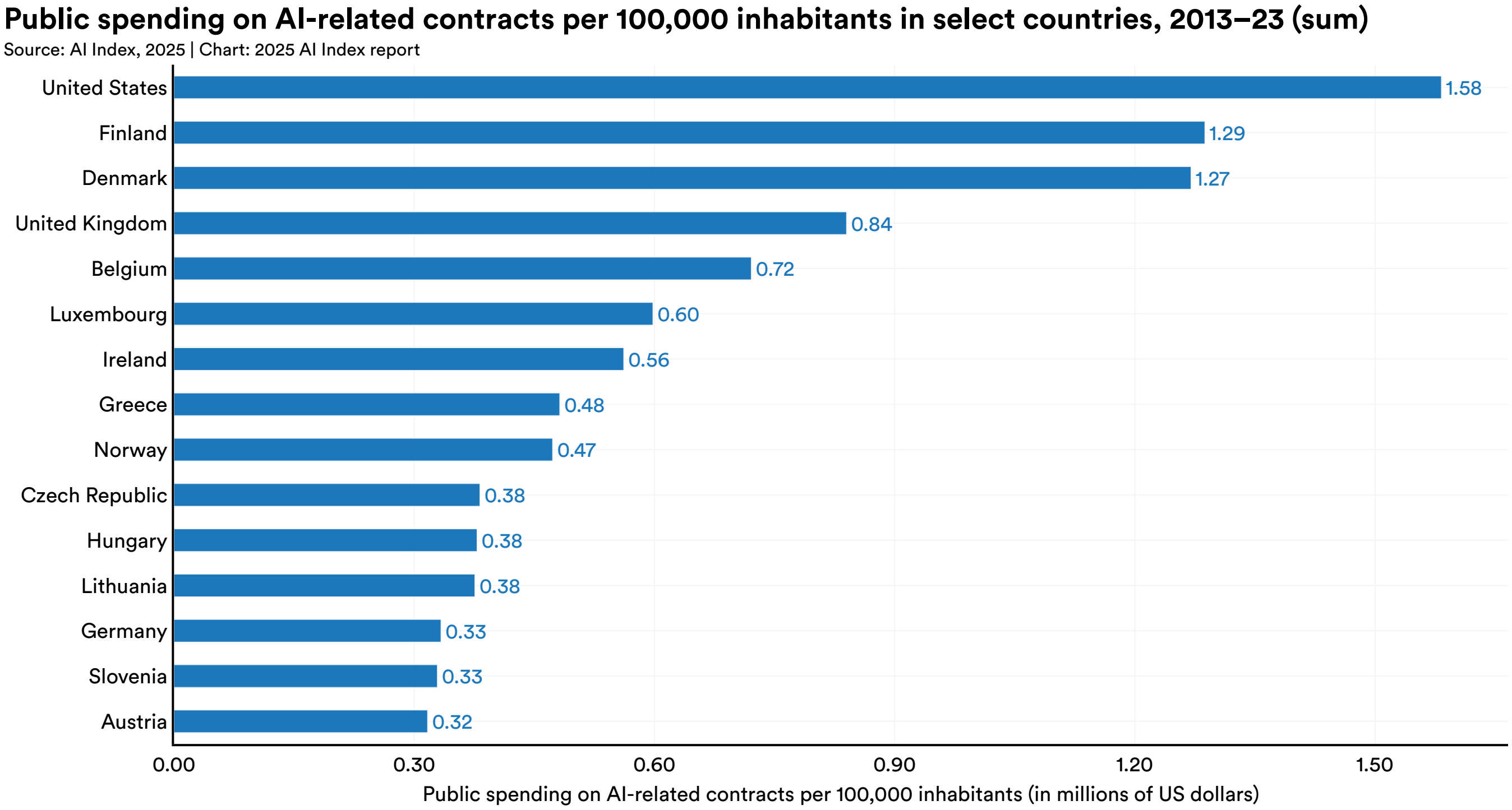

Figure 6.3.4

Figure 6.3.5 illustrates public investment in AI in 2023. The U.S. led with $831.0 million, followed by the United Kingdom at $262.6 million. While Germany, Spain, and the U.K. remained among Europe's top investors, countries that historically ranked lower—such as Romania, Greece, Hungary, and Poland—broke into the top 10. This shift suggests a more balanced distribution of AI-related funding across Europe.

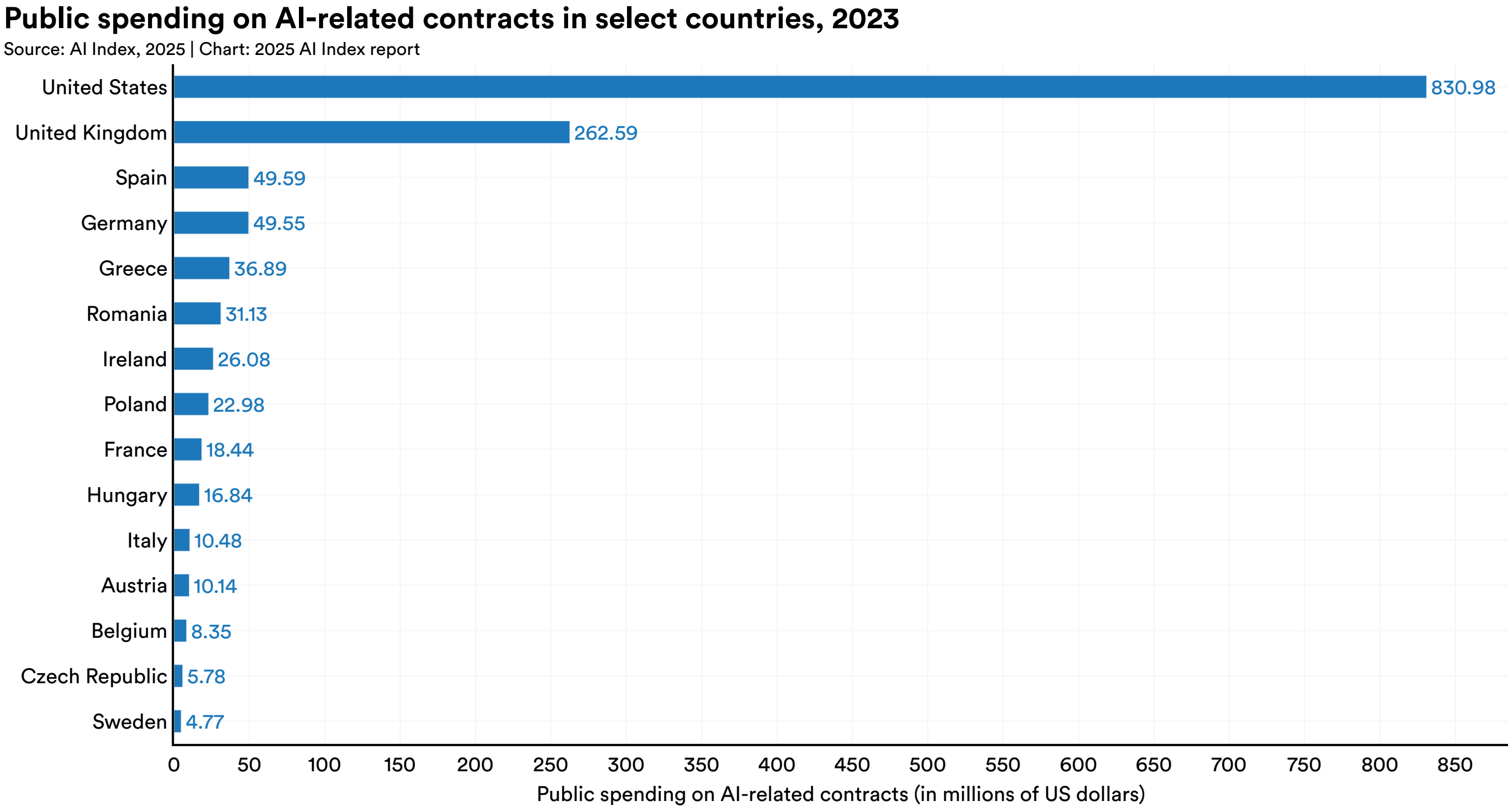

Figure 6.3.5

Figure 6.3.6 illustrates the trends in public AI investment over time across two significant regions of AI investment, the United States and Europe. Both regions have seen substantial growth in AI-related spending over the past decade. Notably, Europe's total AI investment in 2023 was approximately 67 times higher than in 2013, compared to a fifteenfold increase in the United States. Europe experienced particularly sharp increases in investment, with a 400% year-over-year increase in 2017, followed by another major spike of 200% year-over-year in 2019—a year that also saw a peak in the number of national AI strategies released globally. This sustained upward trend illustrates how government interest and commitment to AI is growing in monetary terms.

**Public spending on AI-related contracts in the United States and Europe, 2013–23**
Source: AI Index, 2025 | Chart: 2025 AI Index report

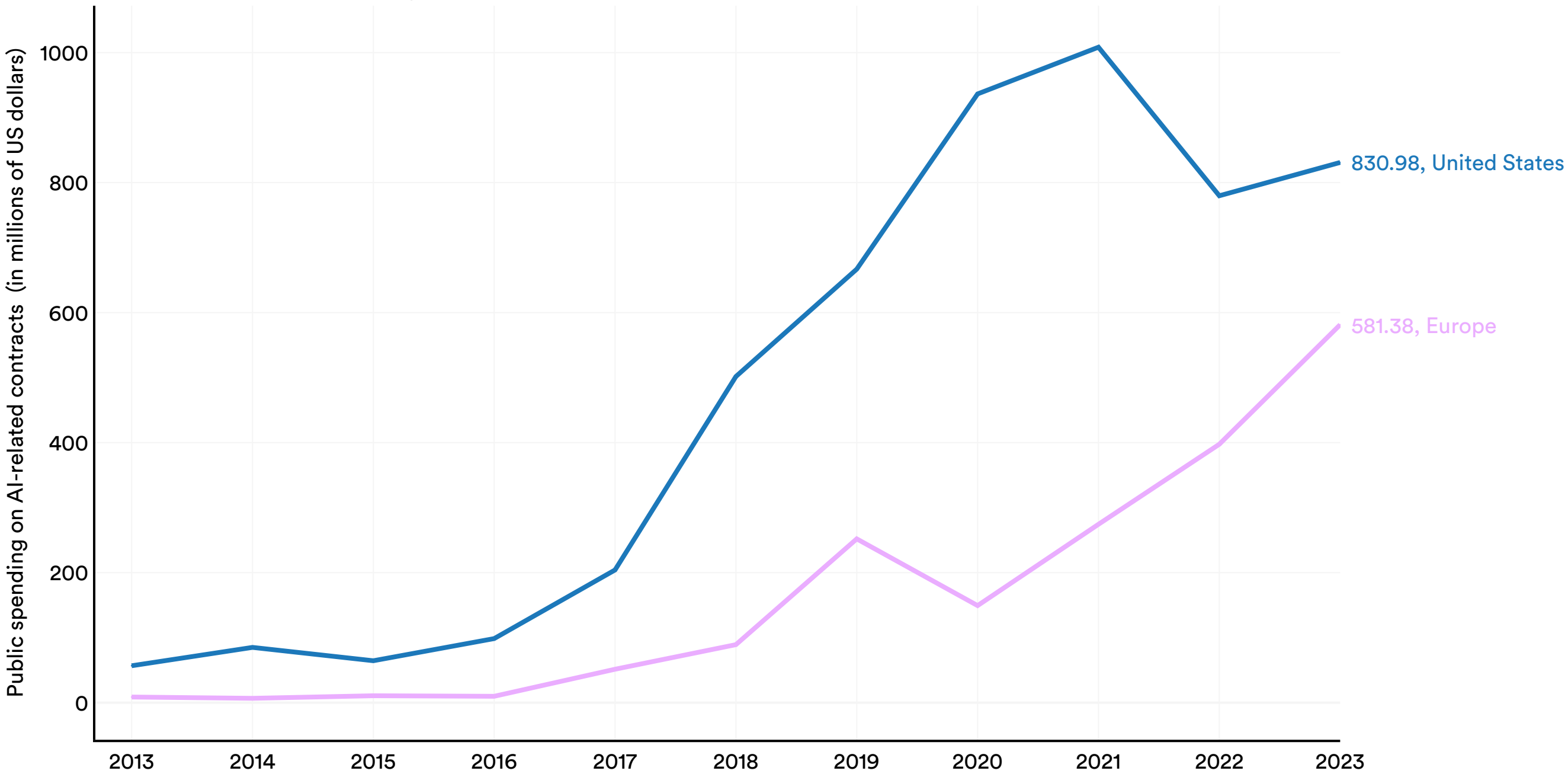

Figure 6.3.6

Figure 6.3.7 charts the investment gap between Europe and the U.S. over time. The disparity in AI investment widened until 2020 but has narrowed over the past three years, indicating that European nations are closing the gap in total AI-related public spending.

**Difference in public spending on AI-related contracts between the United States and Europe, 2013–23**
Source: AI Index, 2025 | Chart: 2025 AI Index report

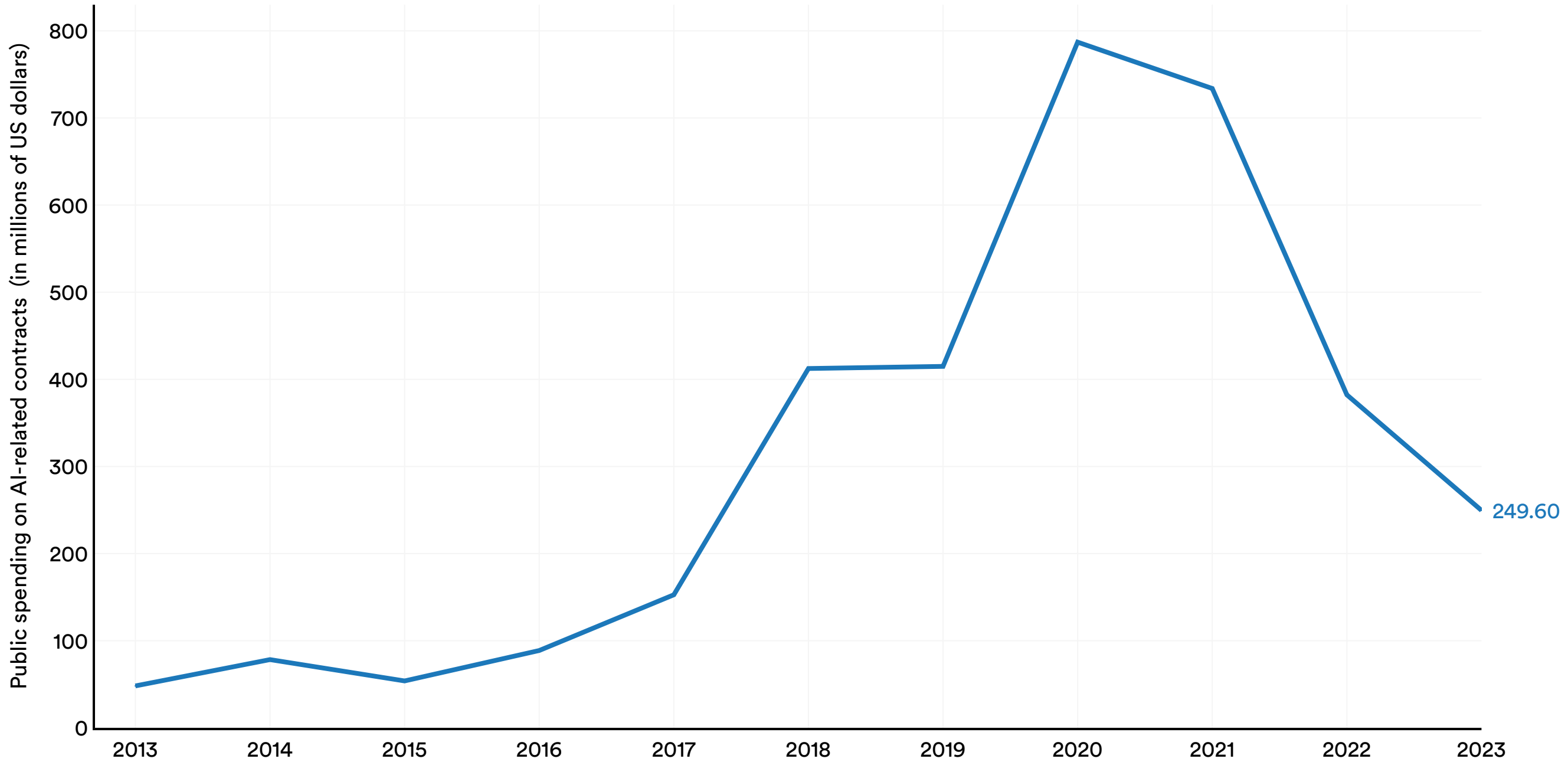

Figure 6.3.7

Figure 6.3.8 documents public investment trends from 2013 to 2023 across the top five European countries—Belgium, France, Germany, Spain, and the U.K. The data reveals a steady increase in investment, marked by periodic peaks. Germany experienced substantial growth, particularly in 2019, following the launch of its national AI strategy in November 2018. The U.K. saw sharp increases in AI-related public investment in both 2021 and 2023. These investments followed the proposition of a national AI strategy by the AI Council—an independent expert committee established in 2019 to advise the government and provide high-level leadership of the AI ecosystem. Meanwhile, Belgium, France, and Spain exhibited more modest but consistent growth.

**Public spending on AI-related contracts in top 5 European countries, 2013–23**

Source: AI Index, 2025 | Chart: 2025 AI Index report

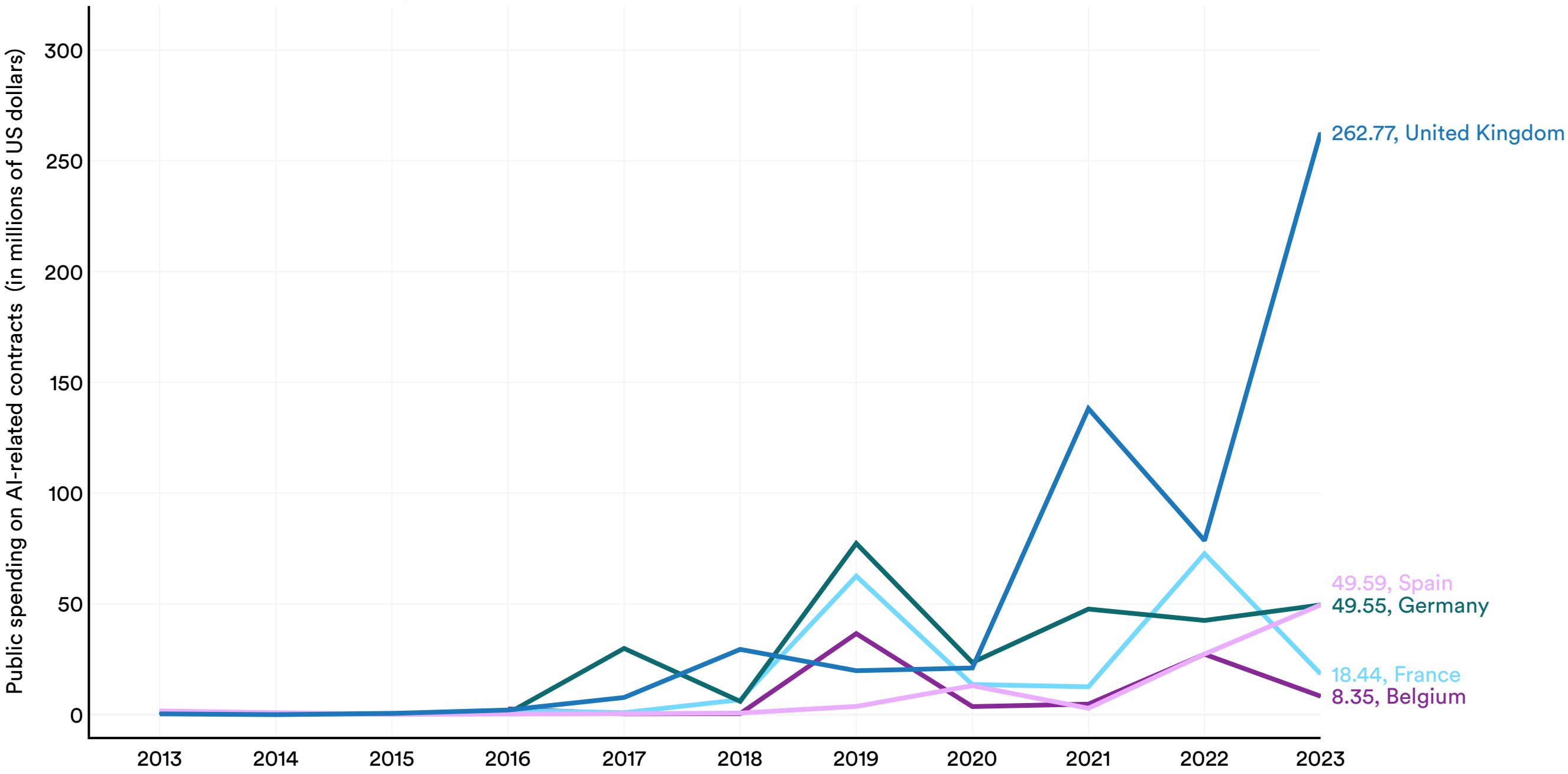

Figure 6.3.8

## Spending Across Agencies and Sectors

The distribution of public tender investments in AI reflects stark contrasts between the U.S. and Europe, driven by differing strategic priorities and institutional structures. As shown in Figure 6.3.9, the U.S. has allocated the majority of AI contracts since 2013 to the Department of Defense. This fact is unsurprising given the central role the American defense sector has played in American technological innovation. In 2023, the Department of Defense (75.0%) was followed by the Department of Veterans Affairs (6.8%) and the Department of the Treasury (5.3%).

While the Department of Veterans Affairs may seem like an outlier, it has made significant investments in recent years—in areas that include the use of AI for diagnosis, robotic prostheses, and mental health.

**Public spending on AI-related contracts (% of total) in the United States by funding agency, 2013–23**
Source: AI Index, 2025 | Chart: 2025 AI Index report

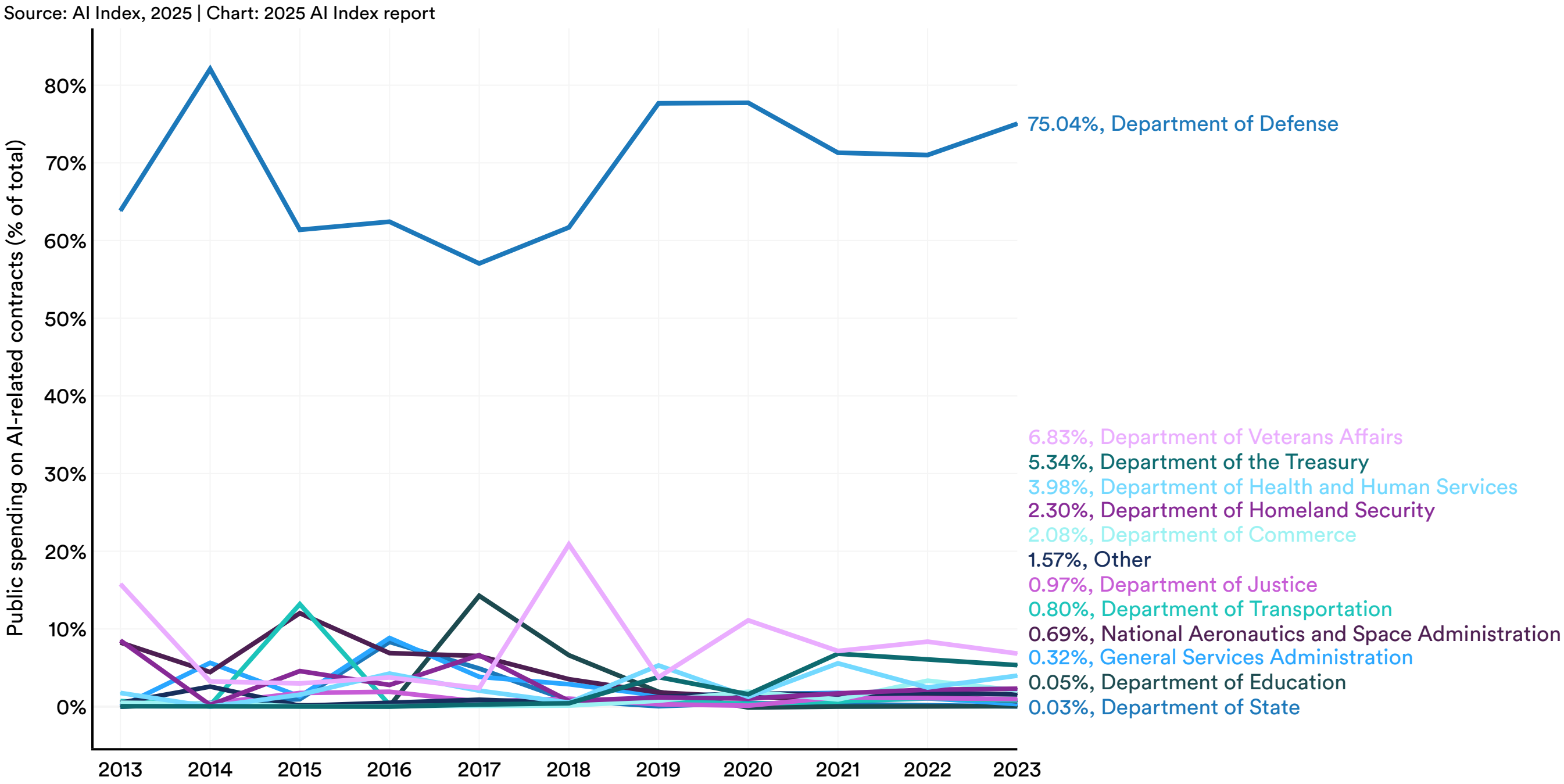

Figure 6.3.9

In Europe, AI investment through public tenders follows a markedly different pattern. Given the lack of aggregated data comparable to that of the U.S., the AI Index categorized European funding entities by their central activity. As shown in Figure 6.3.10, there is a more balanced distribution of investments in Europe. The top funding areas—general public services, education, and health—collectively account for around 84% of total public AI investments in 2023. In the same year, defense accounted for only 0.84% of all European AI-related public tenders. This stands in stark contrast to the U.S., where defense overwhelmingly dominates AI funding.

**Public spending on AI-related contracts (% of total) in Europe by funding agency activity, 2013–23**

Source: AI Index, 2025 | Chart: 2025 AI Index report

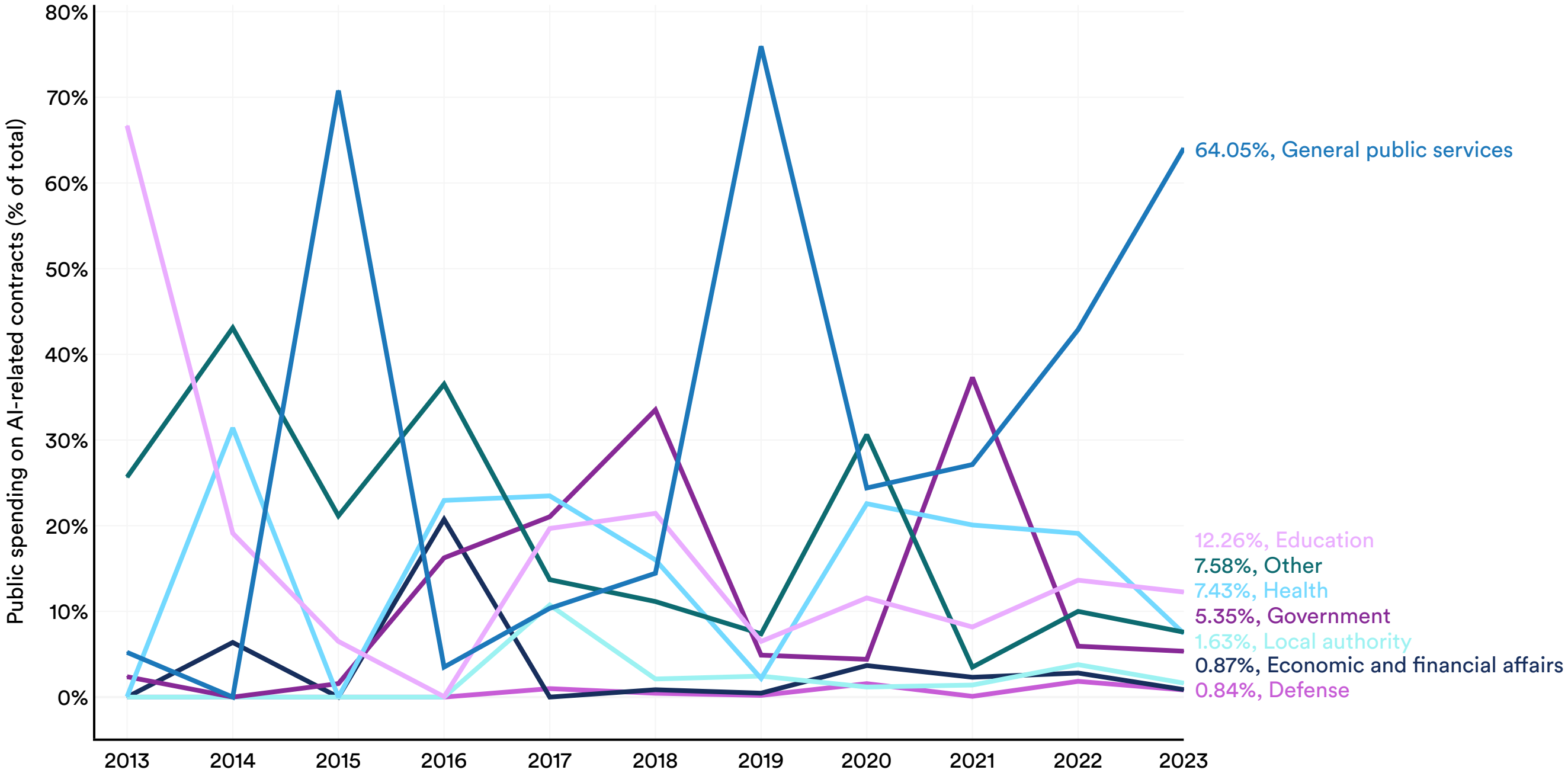

Figure 6.3.10

**Highlight:**

## AI Grant Spending in the US

Public grants also represent a key avenue through which governments allocate resources to AI-related projects and initiatives. Public institutions can directly invest in AI-related projects such as enhancing X-ray angiography interpretation, building AI-driven unmanned aircraft systems for automated soil monitoring, or developing tools for interpretable machine learning. Research grants can be disbursed to organizations like the National Science Foundation or the Department of Health and Human Services (which includes NIH) to conduct AI-focused research. In this section, the AI Index examined data on grants in the U.S. allocated to AI-specific endeavors. As in the previous section, the AI Index employed NLP methodologies to identify AI-related grants.[15]

Figure 6.3.11 displays aggregate data on AI-related grant spending in the U.S. from 2013 to 2023. In that period, a total of roughly $19.7 billion was allocated by the U.S. government for AI-related grants.

**US AI-related grants, 2013–23**
Source: AI Index, 2025 | Table: 2025 AI Index report

| Grant statistics | Value |
|---|---|
| Number of grants | 18,399 |
| Total (in millions $) | 19,748.44 |
| Median (in thousands $) | 247.53 |
| Average (in thousands $) | 1,073.34 |
| Total per 100,000 inhabitants (in thousands $) | 5,967.69 |

Figure 6.3.11

Figure 6.3.12 illustrates the steady rise in AI-related grant funding over time. Between 2013 and 2023, total AI grant funding in the U.S. grew nearly nineteenfold, from $230 million to $4.5 billion. From 2014 to 2020, investments saw an average annual growth rate of 40%. This rapid expansion coincided with major advancements in AI technologies—such as deep learning, natural language processing, and computer vision—which likely fueled demand for public-sector AI applications and drove increased funding for related projects.

**Public spending on AI-related grants in the United States, 2013–23**
Source: AI Index, 2025 | Chart: 2025 AI Index report

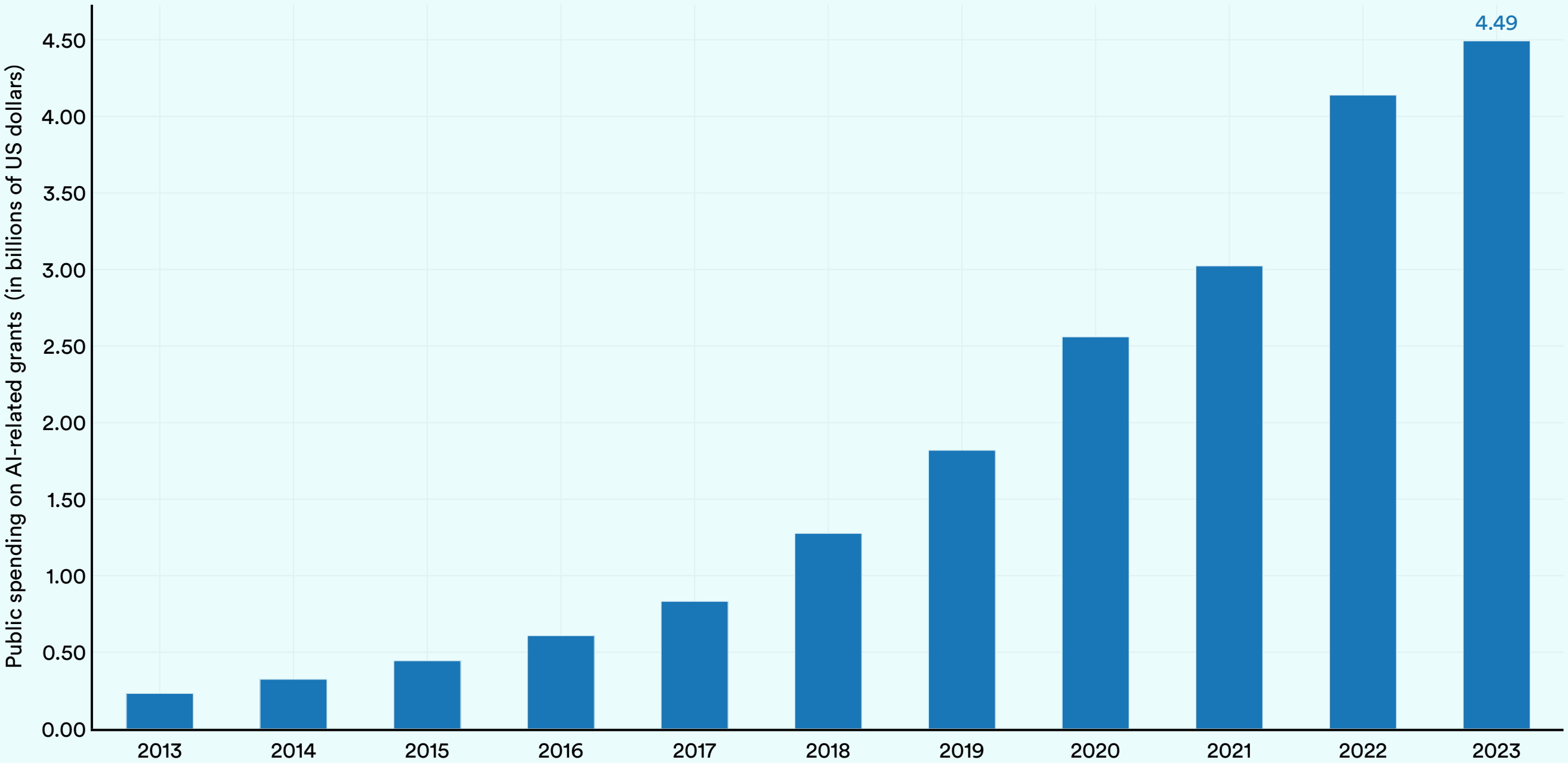

Figure 6.3.12

15 The full methodology behind this approach can be found in the Appendix.

Highlight:
# AI Grant Spending in the US (cont'd)

Figure 6.3.13 illustrates the distribution of AI contract values by funding agencies in the U.S. from 2013 to 2023. The greatest share of AI-related grants was allocated to the Department of Health and Human Services (43.6%), followed by the National Science Foundation (27.9%) and the Department of Commerce (5.4%).

**Public spending on AI-related grants (% of total) by funding agency, 2013–23**
Source: AI Index, 2025 | Chart: 2025 AI Index report

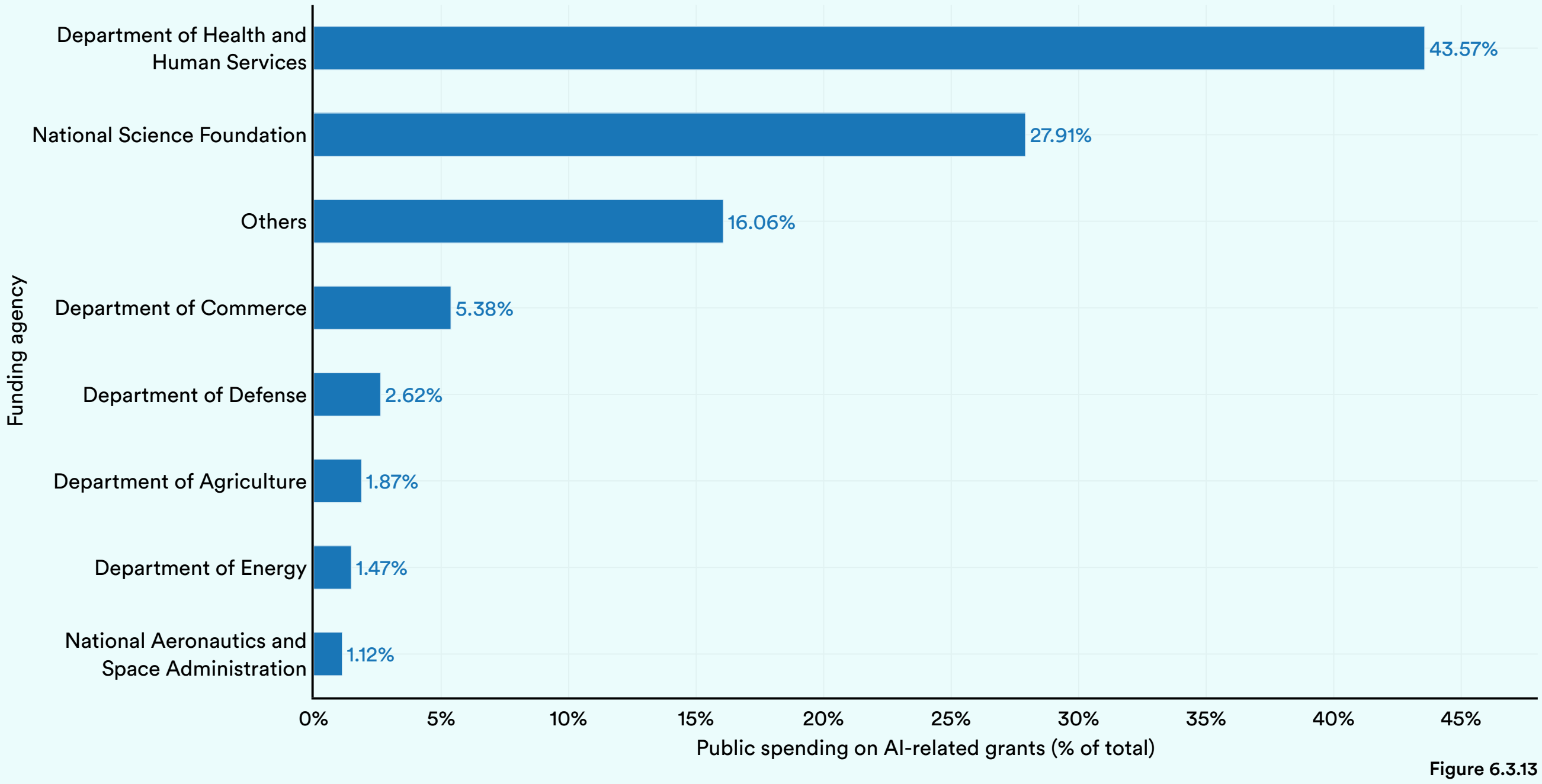

Figure 6.3.13

# CHAPTER 7:
Education

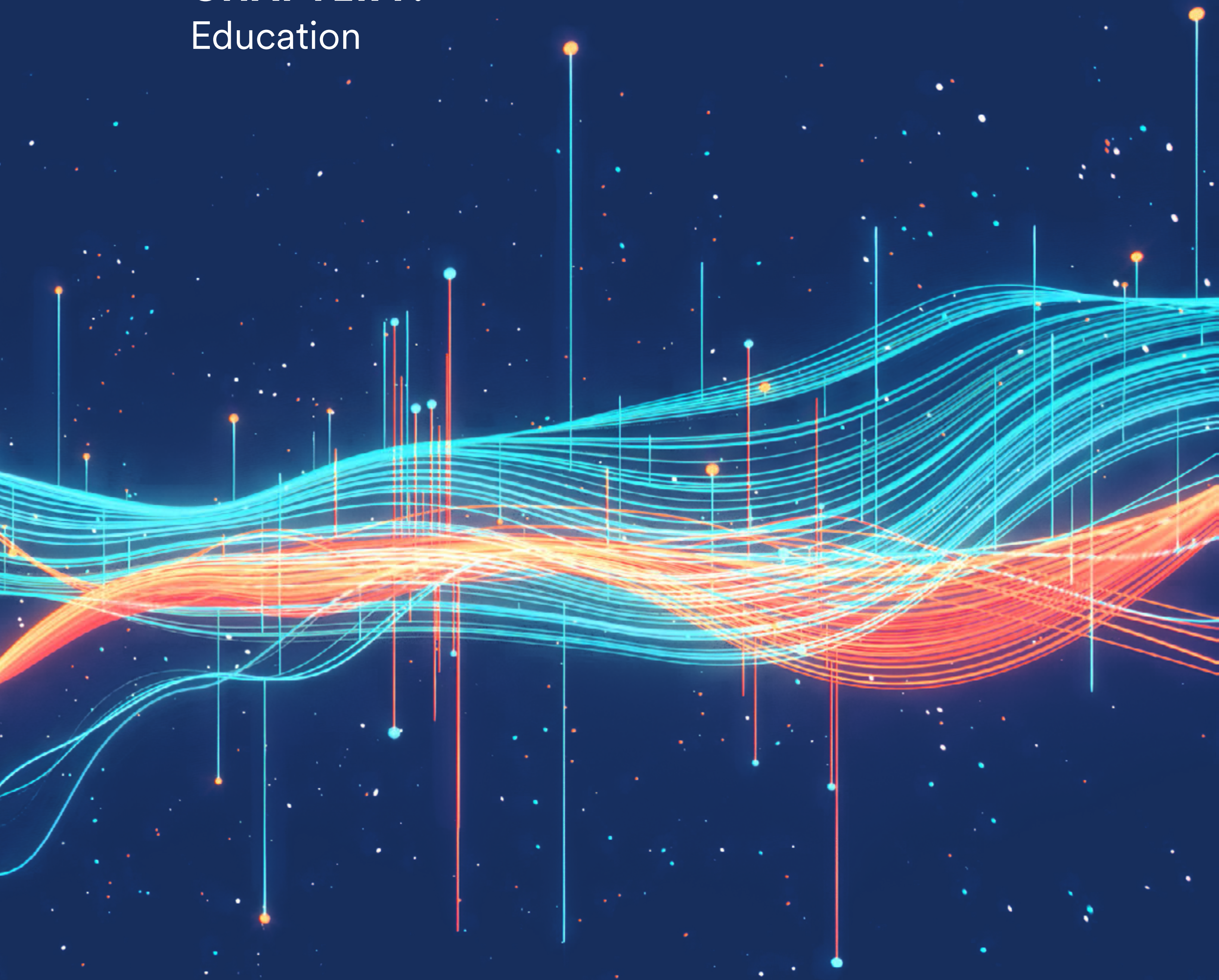

# Chapter 7: Education

ACCESS THE PUBLIC DATA

**CHAPTER 7:**
Education

# Overview

AI has entered the public consciousness through generative AI's impact on work—enhancing efficiency and automating tasks—but it has also driven innovation in education and personalized learning. Still, while AI promises benefits, it also poses risks—from hallucinating false outputs to reinforcing biases and diminishing critical thinking. With the AI education market expected to grow substantially, ethical concerns about the technology's misuse—AI tools have already falsely accused marginalized students of cheating—are mounting, highlighting the need for responsible creation and deployment.

Addressing these challenges requires both technical literacy and critical engagement with AI's societal impact. Expanding AI expertise must begin in K–12 and higher education in order to ensure that students are prepared to be responsible users and developers. AI education cannot exist in isolation—it must align with broader computer science (CS) education efforts. This chapter examines the global state of AI and CS education, access disparities, and policies shaping AI's role in learning.

This chapter was a collaboration prepared by the Kapor Foundation, CSTA, PIT-UN and the AI Index. The Kapor Foundation works at the intersection of racial equity and technology to build equitable and inclusive computing education pathways, advance tech policies that mitigate harms and promote equitable opportunity, and deploy capital to support responsible, ethical, and equitable tech solutions. The CSTA is a global membership organization that unites, supports, and empowers educators to enhance the quality, accessibility, and inclusivity of computer science education. The Public Interest Technology University Network (PIT-UN) fosters collaboration between universities and colleges to build the PIT field and nurture a new generation of civic-minded technologists.

**CHAPTER 7:**
Education

# Chapter Highlights

**1. Access to and enrollment in high school CS courses in the U.S. has increased slightly from the previous school year, but gaps remain.** Student participation varies by state, race/ethnicity, school size, geography, income, gender, and disability.

---

**2. CS teachers in the U.S. want to teach AI but do not feel equipped to do so.** Despite 81% of CS teachers agreeing that using AI and learning about AI should be included in a foundational CS learning experience, less than half of high school CS teachers felt equipped to teach AI.

---

**3. Two-thirds of countries worldwide offer or plan to offer K–12 CS education.** This fraction has doubled since 2019, with African and Latin American countries progressing the most. However, students in African countries have the least access to CS education due to schools' lack of electricity.

---

**4. Graduates who earned their master's degree in AI in the U.S. nearly doubled between 2022 and 2023.** While increased attention on AI will be slower to emerge in the number of bachelor's and PhD degrees, the surge in master's degrees could indicate a future trend for all degree levels.

---

**5. The U.S. continues to be a global leader in producing information, technology, and communications (ICT) graduates at all levels.** Spain, Brazil, and the United Kingdom follow the U.S. as the top producers at various levels, while Turkey boasts the best gender parity.

# 7.1 Background

To expand our understanding of the current state of AI education, it is imperative to differentiate between AI in education, AI literacy, and AI education (see Figure 7.1.1). AI *in* education is the usage of AI tools in the teaching and learning process while AI literacy refers to the foundational understanding of AI—how it works, how to use it, and the risks of using it. AI education encompasses AI literacy plus students' proficiency in the technical skills required to build AI (data analyses undergirding AI technologies, identifying and mitigating data biases, etc.). For the purposes of this chapter, the data presented covers AI education.

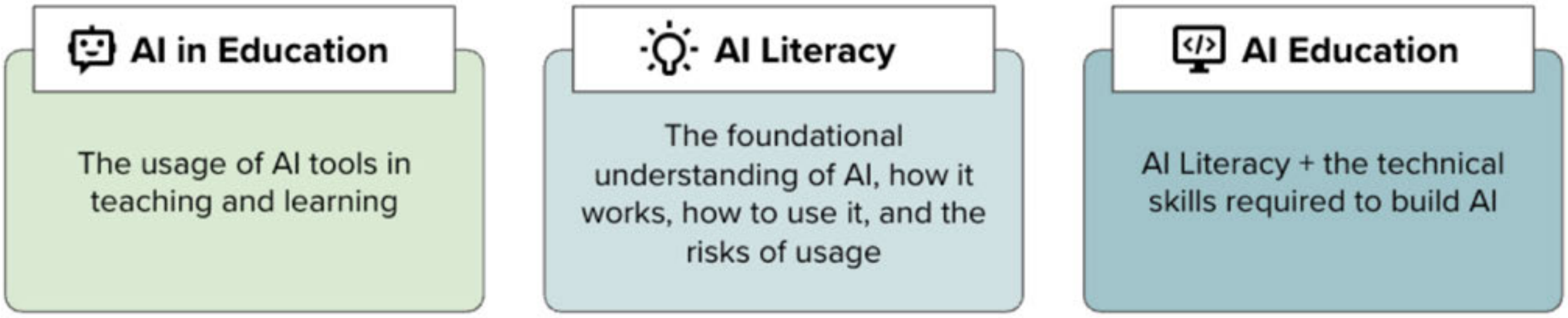

Figure 7.1.1

The world faces significant challenges in developing a robust and diverse workforce when disparities in infrastructure, access to resources and courses, and participation in high quality coursework continue to exacerbate vast inequities in K–12 students' ability to contribute to a technology-enabled future. While it is difficult to accurately estimate the extent of the problem due to the unstandardized nature of data collection and metrics development, this section focuses on the earliest stage in the computing pipeline by examining the current status of K–12 CS and AI education with existing global data.

# 7.2 K–12 CS and AI Education[1]

## United States

To begin exploring the prevalence and quality of AI education within the United States, it is important to start with the CS education landscape in its earliest stages almost a decade ago. With the launch of President Barack Obama's "Computer Science for All" initiative in 2016, billions in investments were provided to ensure that all K–12 students learn CS to become creators in the digital economy and responsible citizens of a technology-driven society. The federal funding was dedicated to enhancing professional learning efforts, improving instructional resources, and building effective regional partnerships toward expanding CS education access. The National Science Foundation also led the development and implementation of two new computing courses (Exploring Computer Science and AP Computer Science Principles) aimed at engaging a broader group of students in computing. At the same time, the technology industry and philanthropy invested millions in national efforts to introduce millions of students across the country to CS.

### Foundational Computer Science

In the past decade, educational advocates have implored policymakers to adopt legislation to improve access to CS education. These efforts have paid off. In the 2017–18 academic year, 35% of U.S. high schools offered CS, which increased to 60% of U.S. high schools in 2023–24. However, national trends can obscure the reality that prioritization of CS education varies by state. For example, 100% of high schools in Arkansas and Maryland offer CS, compared to only 31% in Montana (Figure 7.2.1).

**Public high schools teaching foundational CS (% of total in state), 2024**

Source: Code.org, CSTA, and ECEP Alliance, 2024 | Chart: 2025 AI Index report

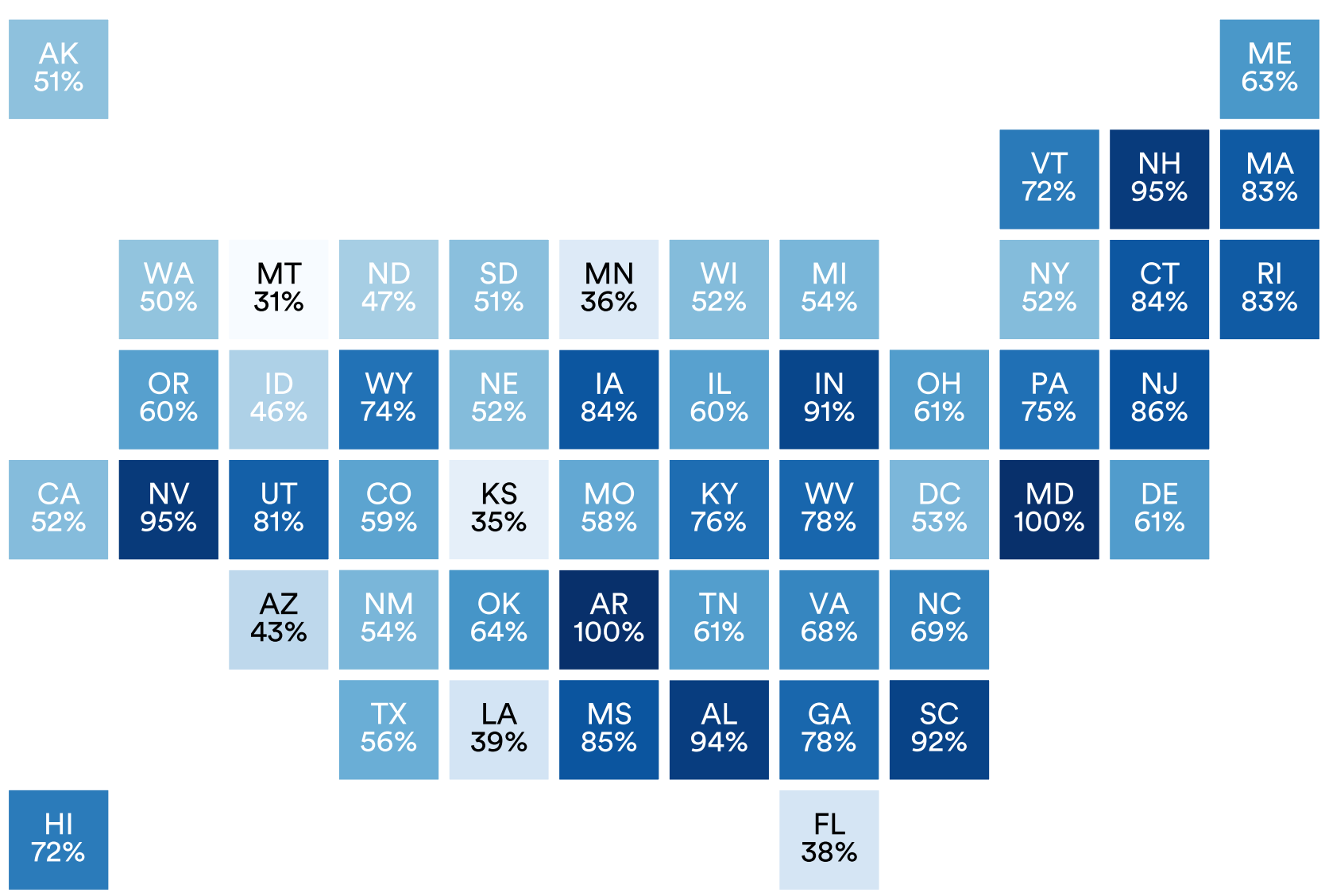

Figure 7.2.1

1 Since AI has historically been studied under CS, this chapter references CS education data when AI-specific data is unavailable.

Significant gaps remain in equitable access to CS education, with some student groups left behind. In the 2023–24 academic year, students eligible for free or reduced-price lunch (FRL); those in small schools; students living in urban and rural areas; and Native students were less likely to have access to CS education (Figures 7.2.2, 7.2.3, 7.2.4, and 7.2.5).

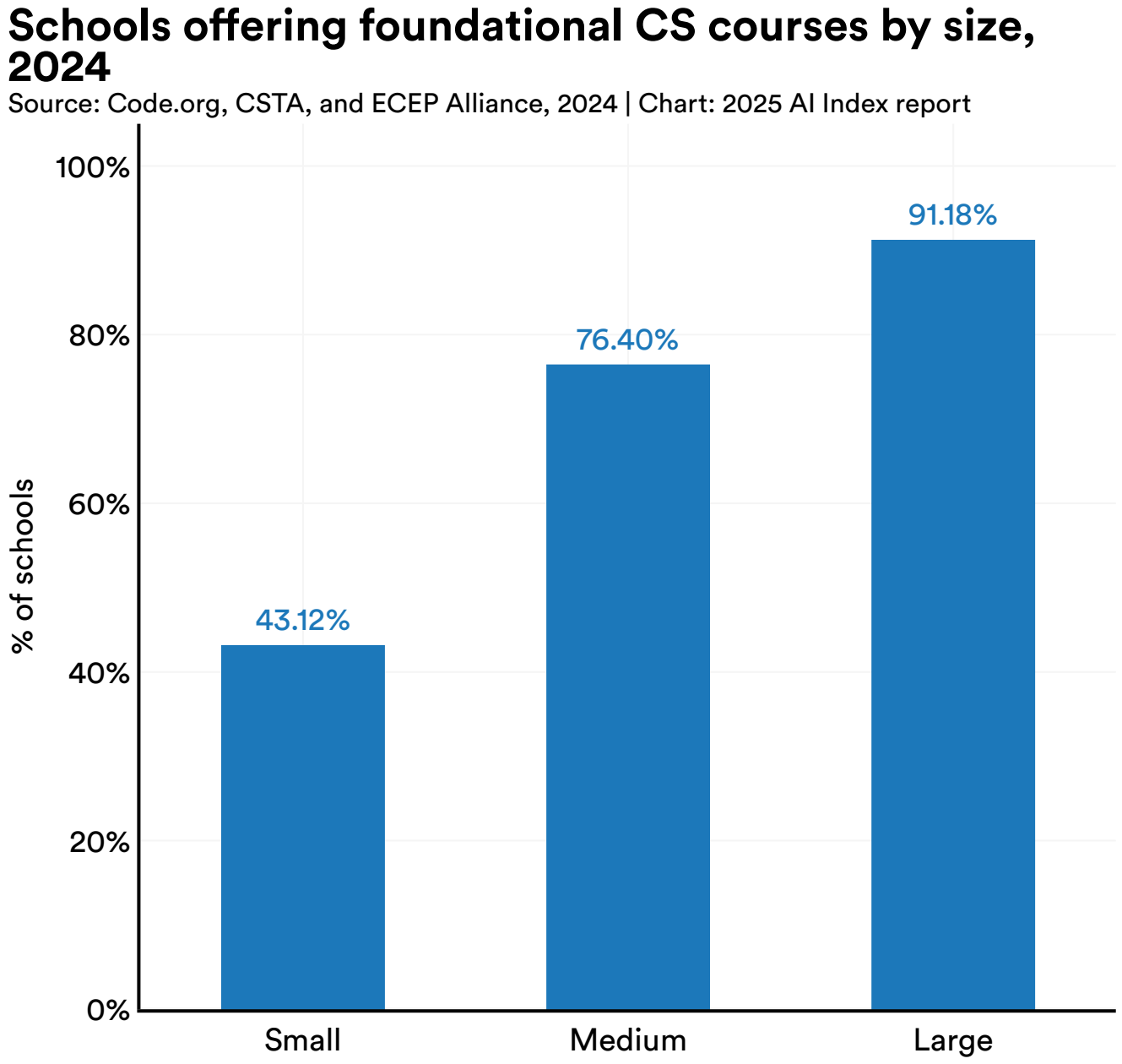

Figure 7.2.2

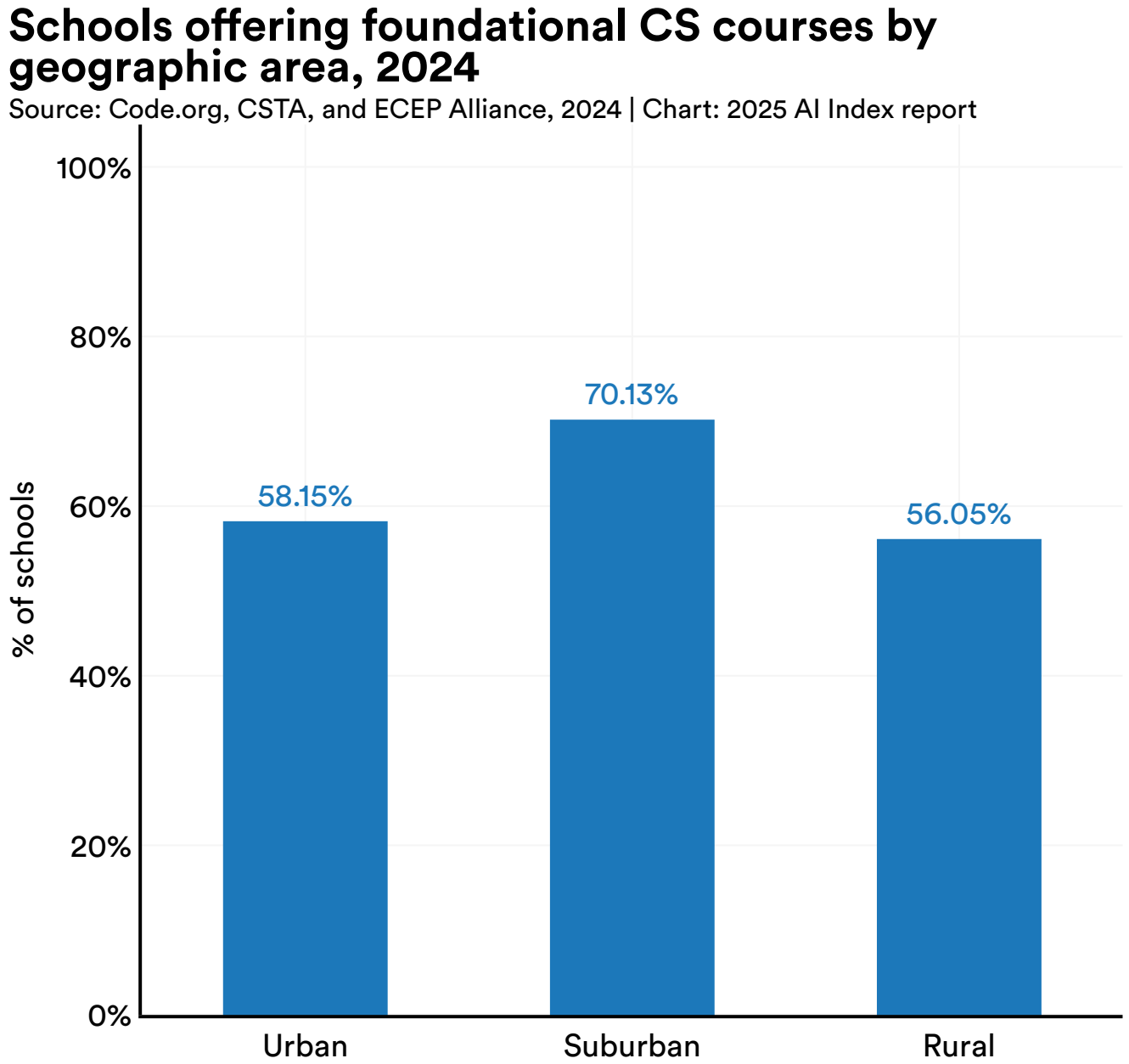

Figure 7.2.3

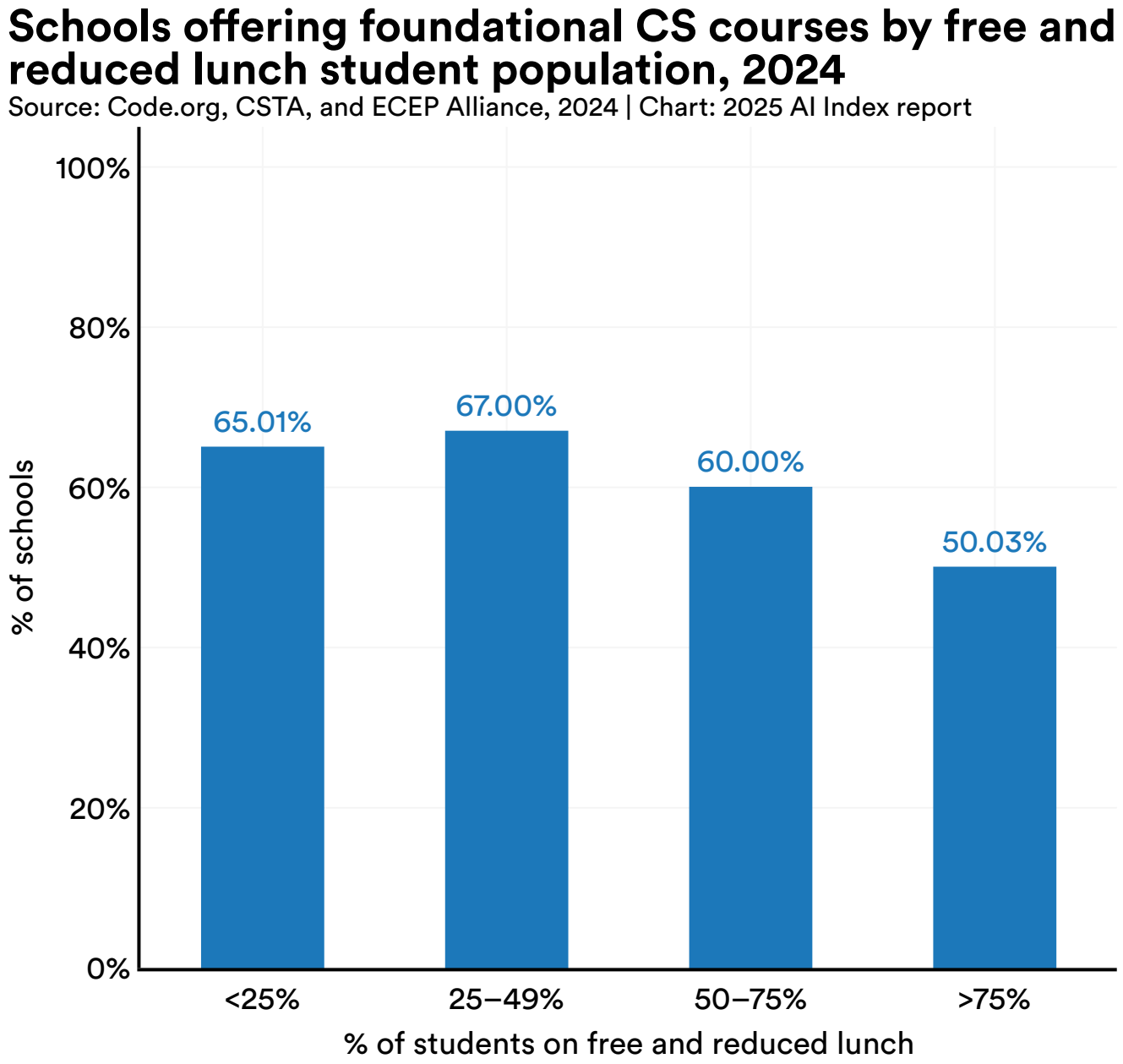

Figure 7.2.4

## Access to foundational CS courses by race/ethnicity, 2024

Source: Code.org, CSTA, and ECEP Alliance, 2024 | Chart: 2025 AI Index report

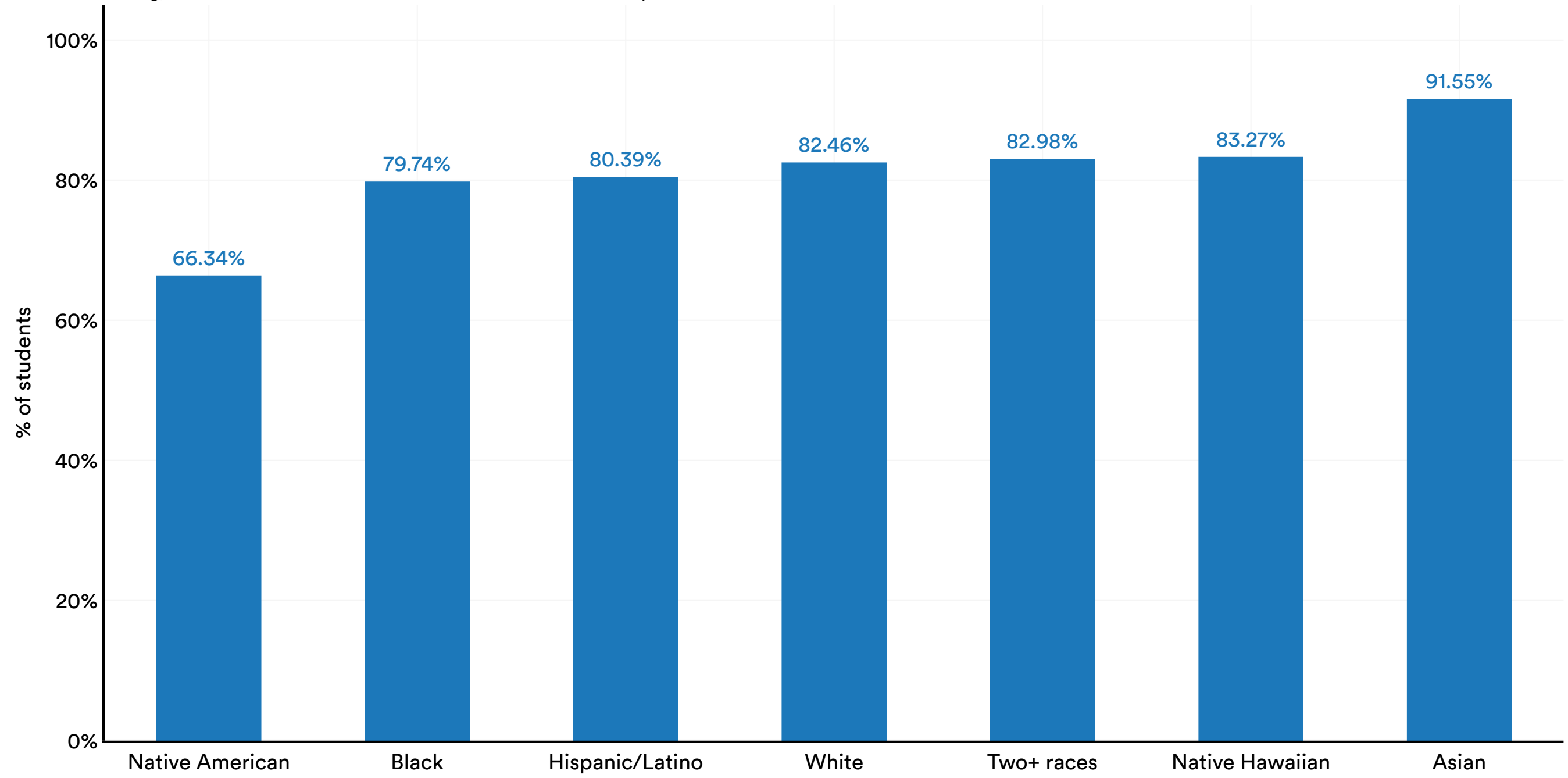

Figure 7.2.5

Data about participation in CS across 41 states indicates lags in student engagement with courses. In the 2020–21 academic year, only 5.1% of high school students participated in CS, with a marginal increase to 6.4% in 2023–24. Similar to CS access, CS participation varies highly between states—with 26% of high school students in South Carolina enrolled in CS but only 2% enrolled in Florida, Arizona, and Idaho (Figure 7.2.6).

## Public high school enrollment in CS (% of students), 2024

Source: Code.org, CSTA, and ECEP Alliance, 2024 | Chart: 2025 AI Index report

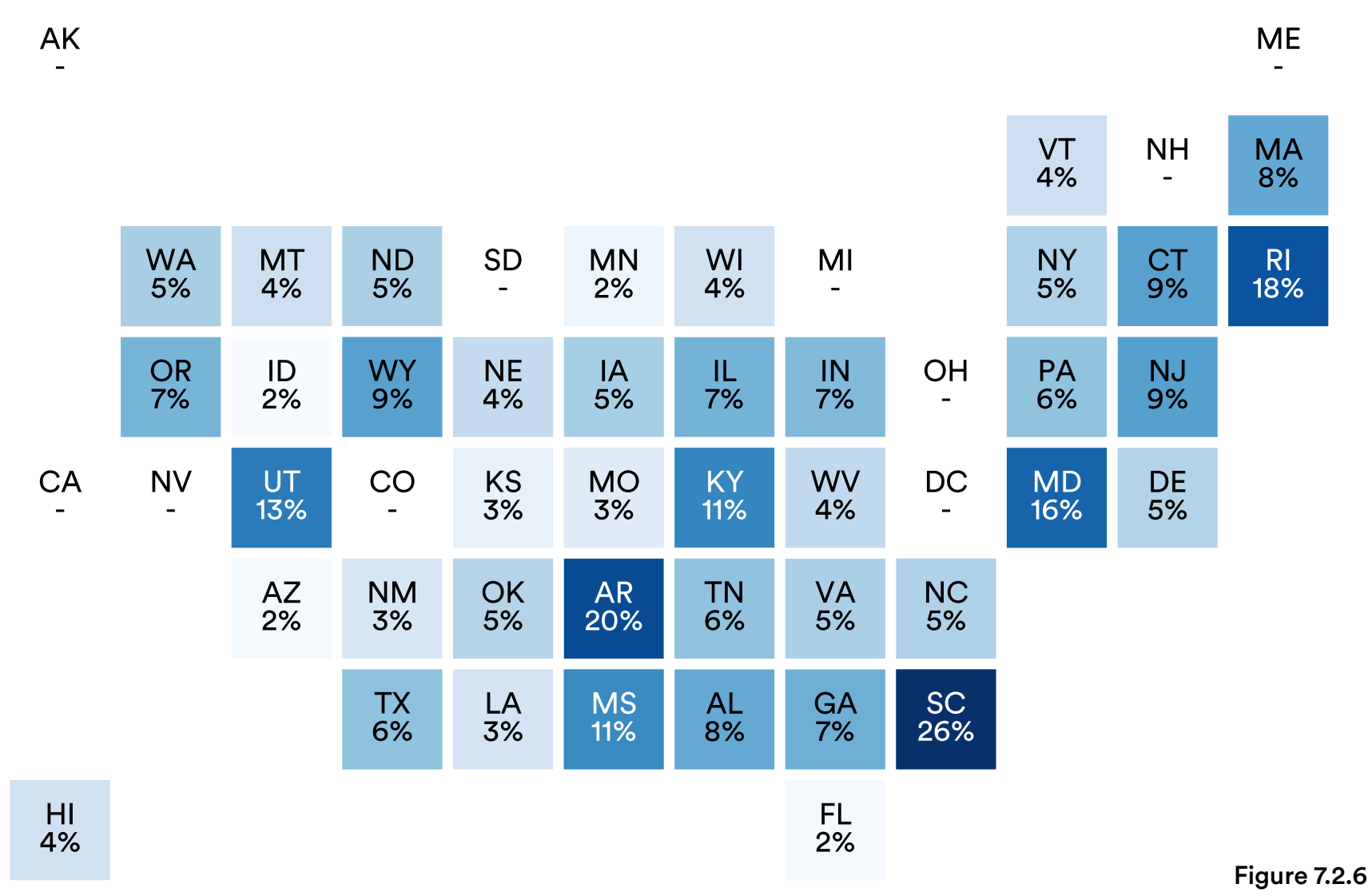

Figure 7.2.6

An analysis of CS enrollment by race and ethnicity shows that efforts to expand access have resulted in near or above proportional representation for Black, Native American/Alaskan, and white students at the national level (Figure 7.2.7). However, data gaps—particularly from nine states—warrant caution in viewing these trends as complete. Girls are underrepresented relative to their share of the K–12 population. Additionally, Hispanic and Native Hawaiian/Pacific Islander students, students with individualized education programs (IEPs), those eligible for free or reduced-price lunch, and English language learners remain underrepresented nationally (Figure 7.2.7 and Figure 7.2.8).

**Public high school enrollment in CS vs. national demographics by race/ethnicity, 2024**

Source: Code.org, CSTA, and ECEP Alliance, 2024 | Chart: 2025 AI Index report

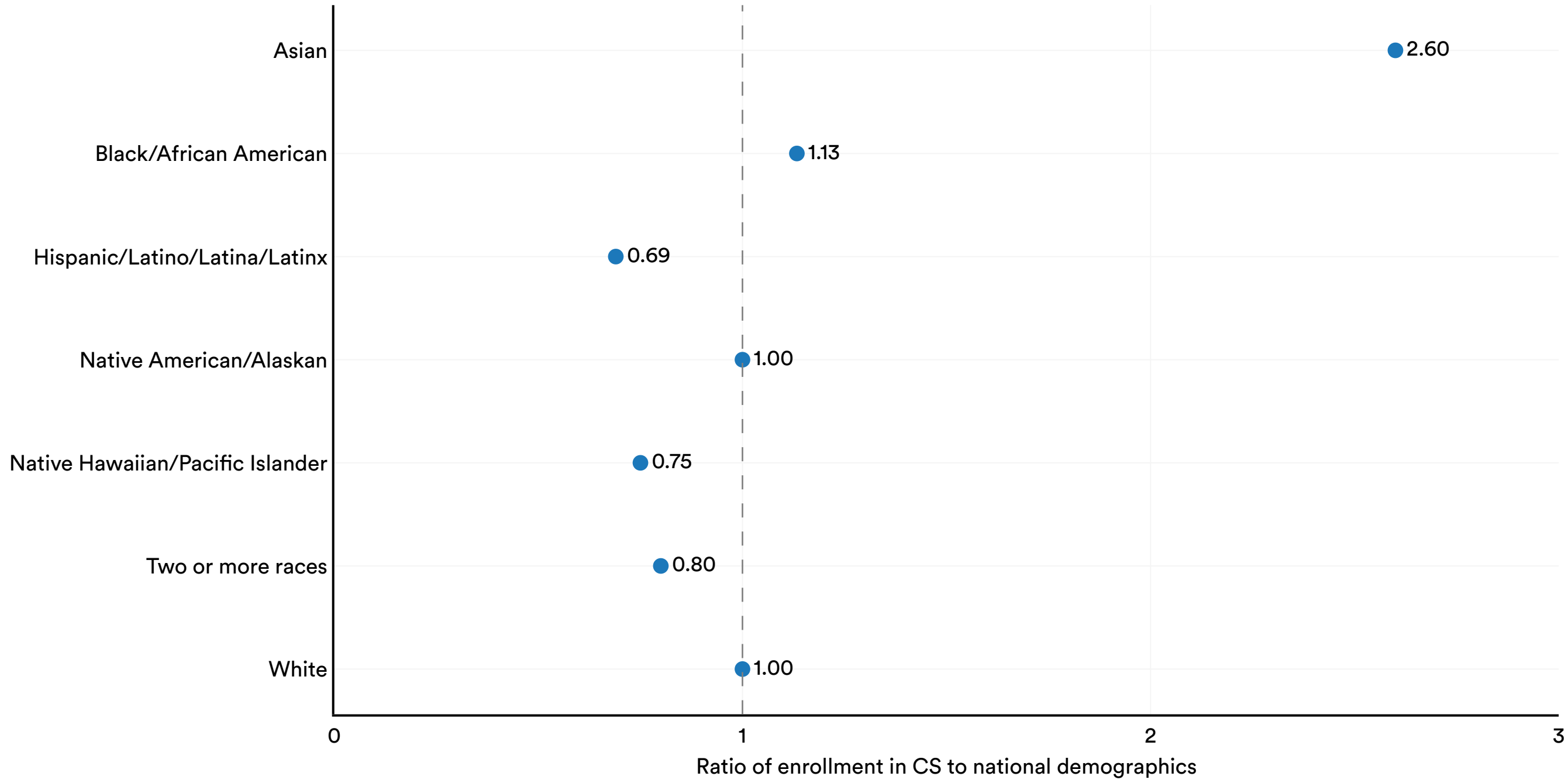

Figure 7.2.7

**Public high school enrollment in CS vs. national demographics by subgroup, 2024**
Source: Code.org, CSTA, and ECEP Alliance, 2024 | Chart: 2025 AI Index report

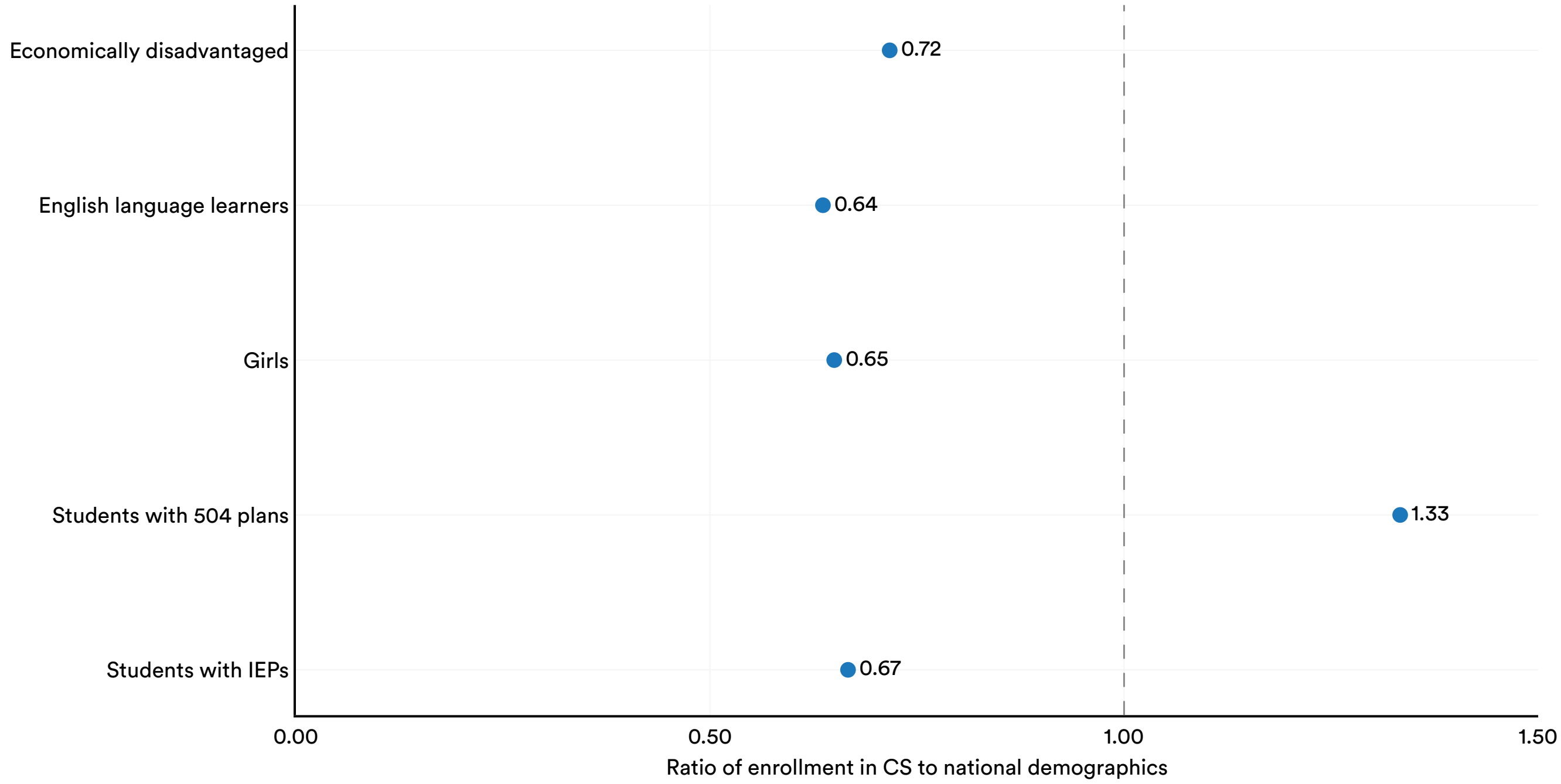

Figure 7.2.8[2]

## Advanced Computer Science

In order to build students' AI competencies, it is essential to offer access to advanced coursework in addition to foundational courses. While AI is not specifically covered in Advanced Placement (AP) CS A, AP CS Principles (AP CS P) does address some AI content areas. Because AP CS P was designed to attract a broader class of students, the potential exists to expose a diverse student population to AI topics. Yet, despite the growth in raw numbers of students participating in the AP CS exam (Figure 7.2.9), students do not participate in proportion to their racial and ethnic representation in the general student body (Figure 7.2.10 and Figure 7.2.11). Asian students, white boys, and multiracial students are overrepresented in the population of students who take AP CS exams, while all other student groups are underrepresented (Figure 7.2.12).

2 A student with a 504 plan receives accommodations under Section 504 of the Rehabilitation Act of 1973, a U.S. civil rights law that prohibits discrimination against individuals with disabilities. A student with an IEP (individualized education program) receives special education services under the Individuals with Disabilities Education Act. An IEP is a legally binding document that outlines a learning plan for a student with a disability designed to meet their unique needs and improve educational outcomes.

## Number of AP computer science exams taken, 2007–23

Source: Code.org, CSTA, and ECEP Alliance, 2024 | Chart: 2025 AI Index report

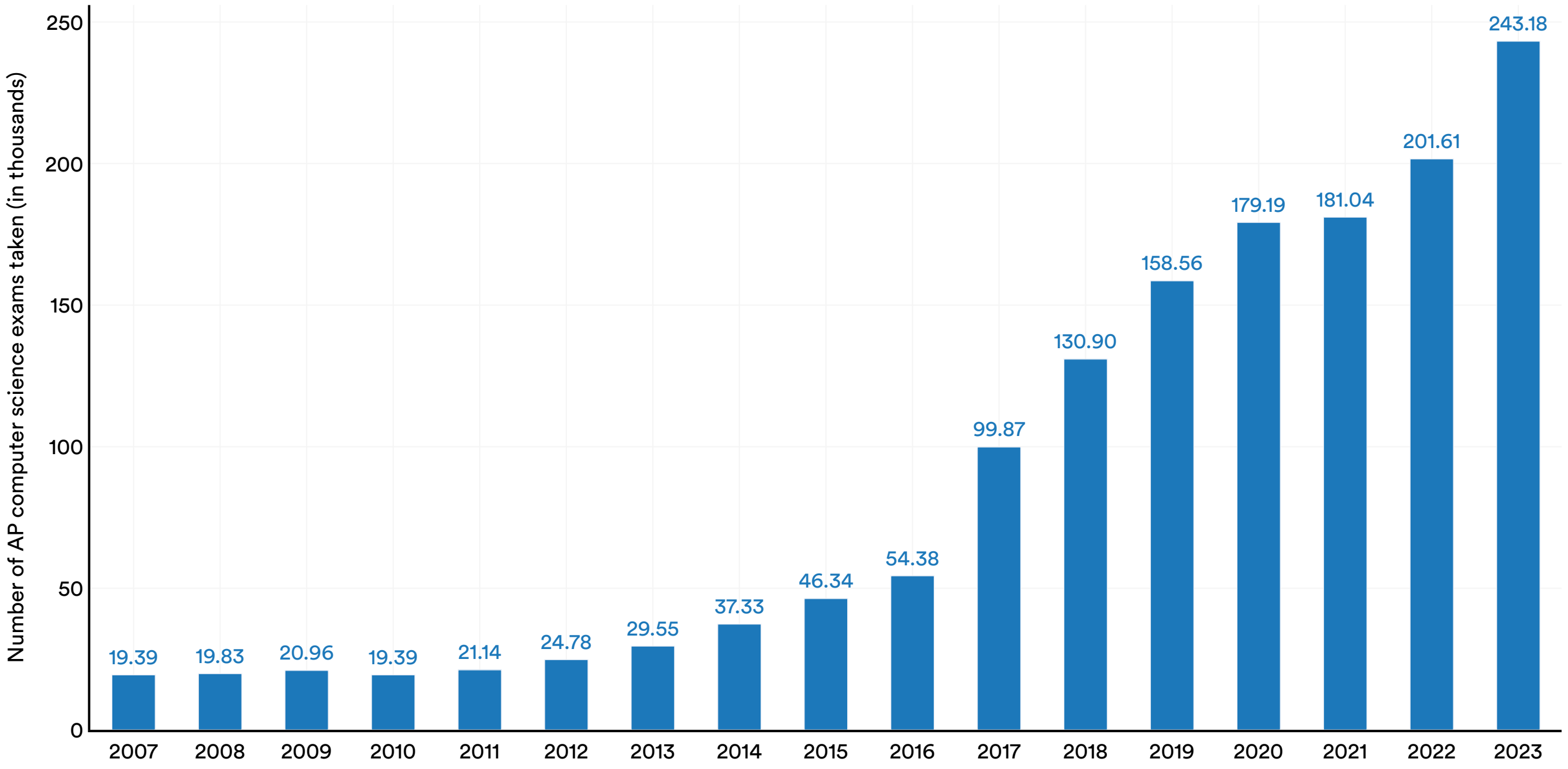

Figure 7.2.9

## AP computer science exams taken by race/ethnicity, 2007–23

Source: Code.org, CSTA, and ECEP Alliance, 2024 | Chart: 2025 AI Index report

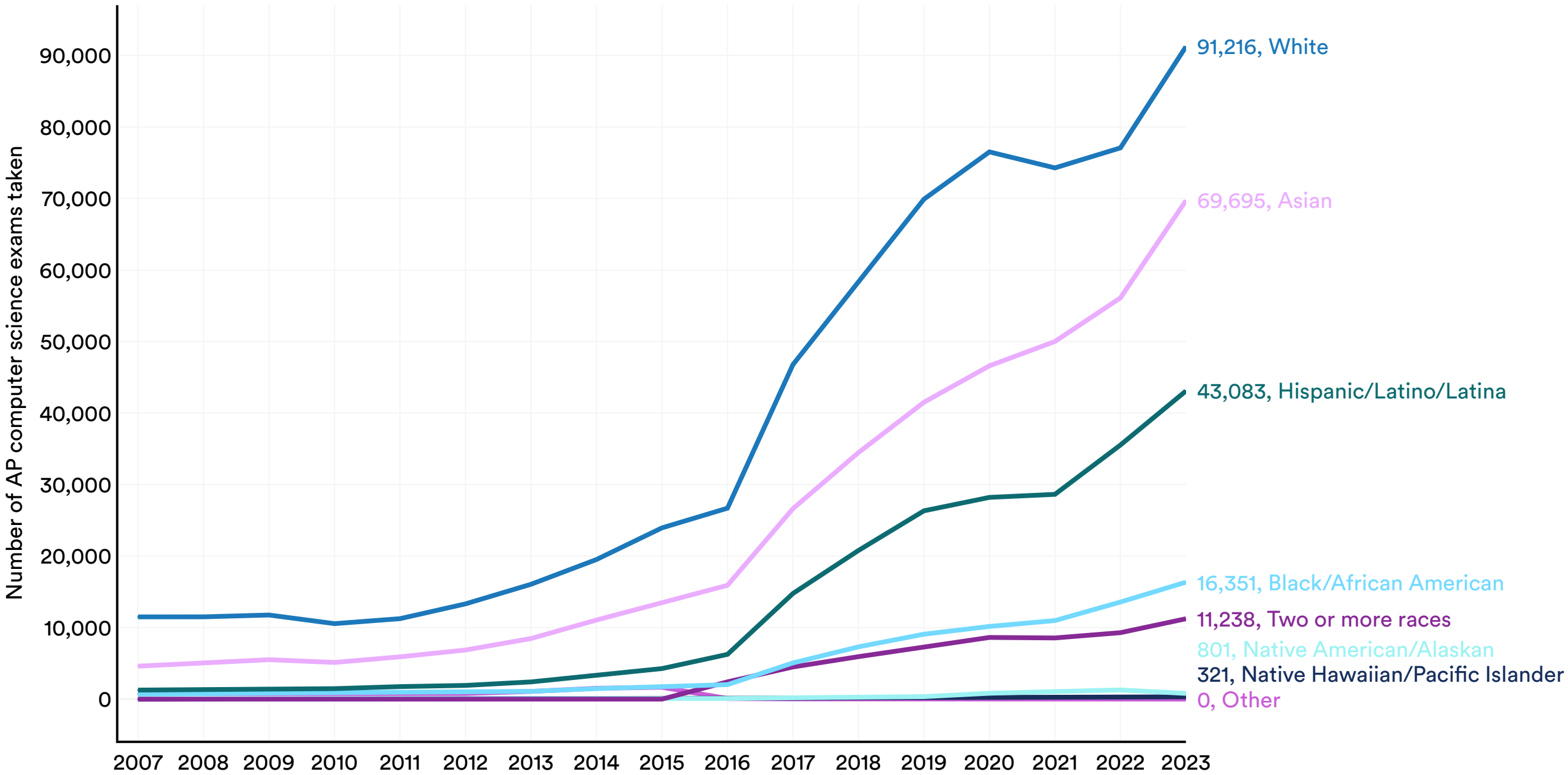

Figure 7.2.10

## AP computer science exams taken (% of total responding students) by race/ethnicity, 2007–23

Source: Code.org, CSTA, and ECEP Alliance, 2024 | Chart: 2025 AI Index report

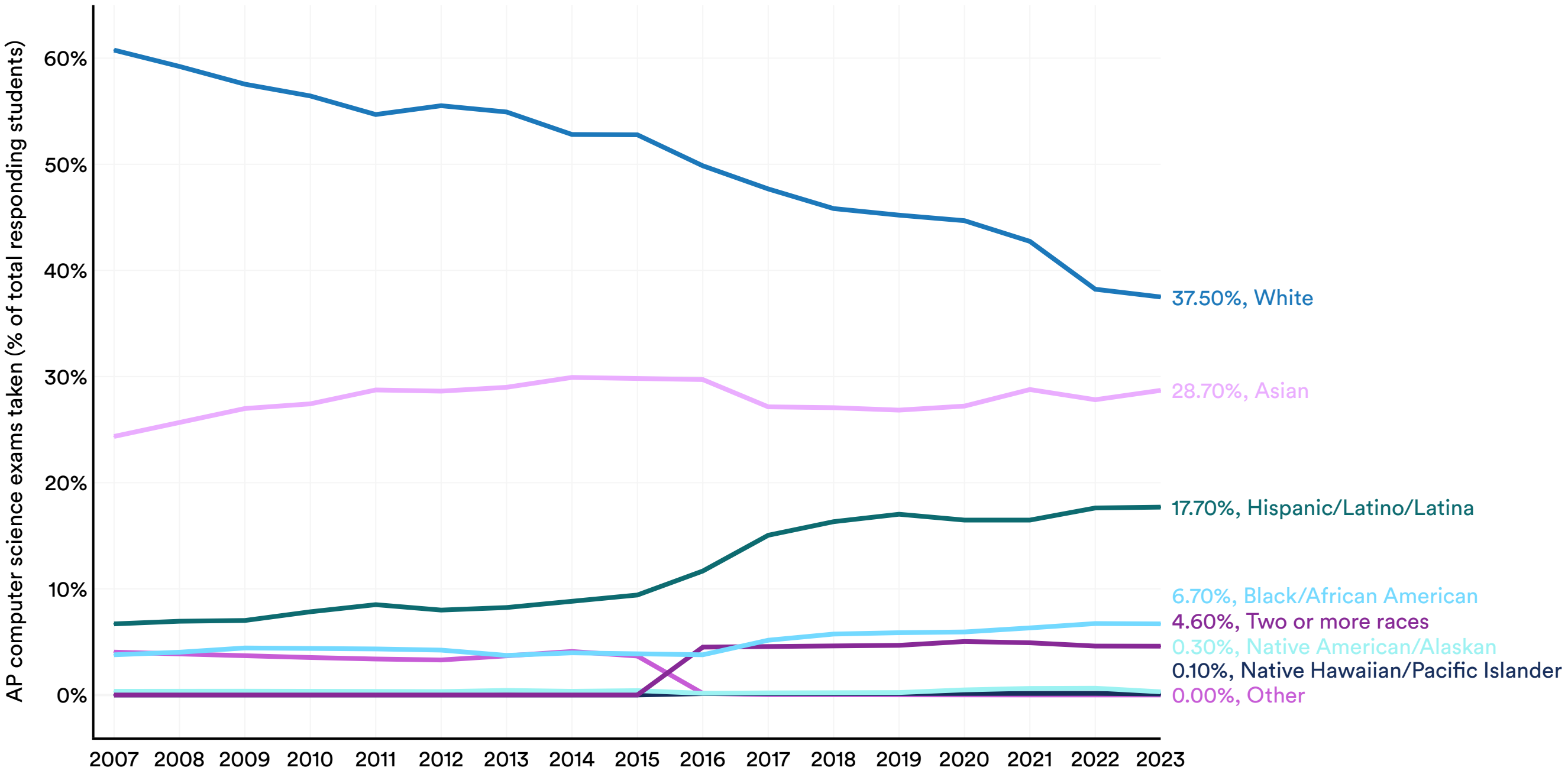

Figure 7.2.11

## AP computer science exam participation vs. national demographics by race/ethnicity, 2023

Source: Code.org, CSTA, and ECEP Alliance, 2024 | Chart: 2025 AI Index report

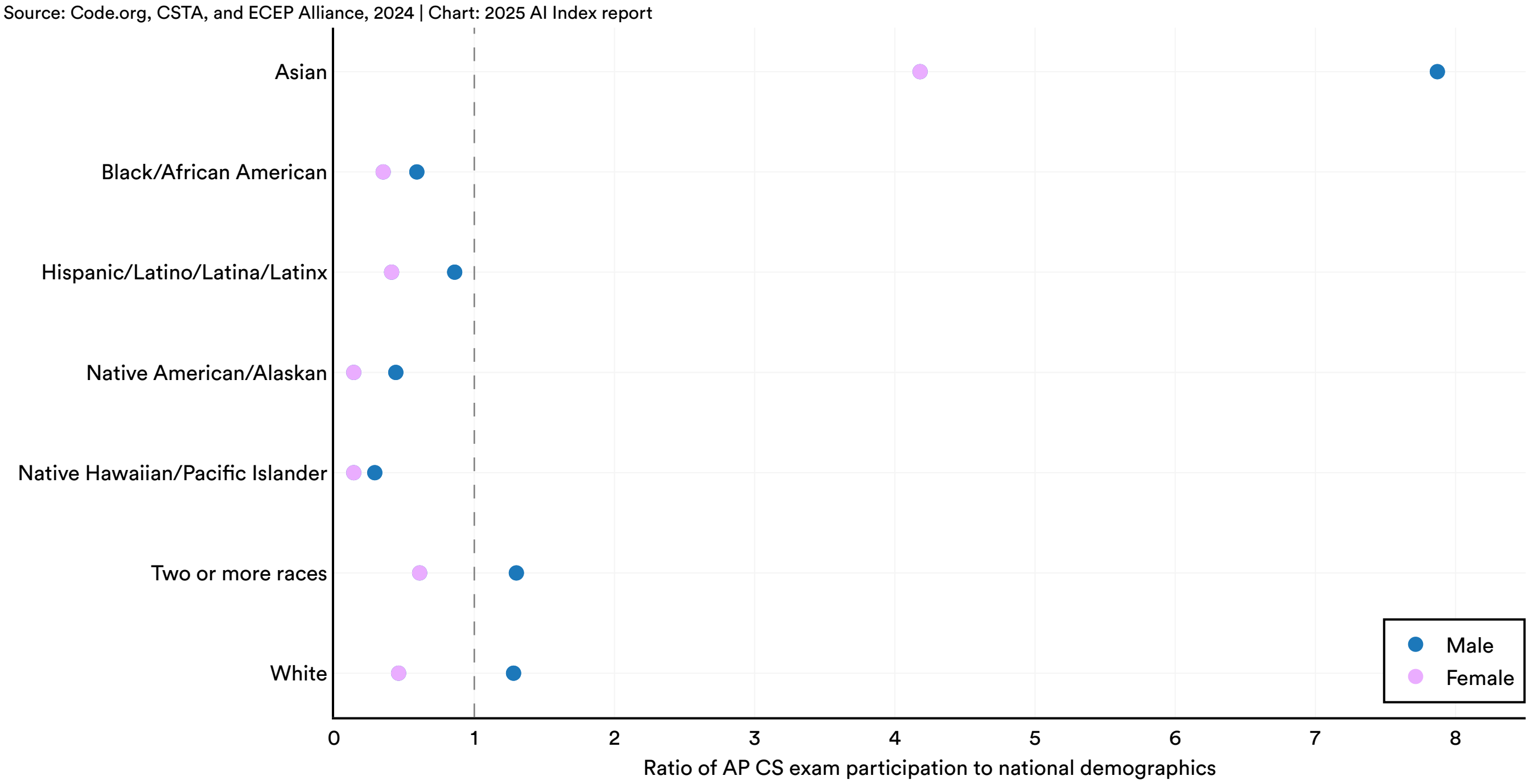

Figure 7.2.12

## Education Standards and Guidance

Federal guidance issued thus far has focused on AI *in* education rather than AI education. The U.S. Department of Education's Office of Educational Technology released a series of reports about AI in education in 2023 and 2024. One of the reports focuses on recommendations for educational technology developers, and two of them are intended for educators, educational leaders, and policymakers. The most recent report, from October 2024, offers guidance on the safe and effective implementation of AI in K–12 schools.

As of January 2025, 26 states have issued guidance on AI *in* education. And while there is considerable overlap between CS and AI education content and what teachers currently cover in the classroom, K–12 CS standards contain minimal AI content. The Computer Science Teachers Association (CSTA) K–12 standards, last published in 2017, contain only two standards at the advanced high school level that specifically require AI knowledge. However, existing CS standards support foundational AI knowledge and skills, covering topics such as perception, data structures, and algorithms. The U.S. state-adopted K–12 CS standards averaged 97% coverage of the same subconcepts as the CSTA standards, indicating strong national coherence in CS instruction. Among the 44 states that have adopted K–12 CS standards, 33 have AI-specific standards, which are generally minimal, aligned to the CSTA standards, and focused on high school grades (Figure 7.2.13).[3] Four of these states recently adopted more significant AI-specific standards that span grades K–12: Colorado (2024), Florida (2024), Ohio (2022), and Virginia (2024), while Arkansas has defined standards for a high school AI and machine learning course.

**Adoption of AI-specific K–12 computer science standards by US state**
Source: CSTA and IACE, 2024 | Chart: 2025 AI Index report

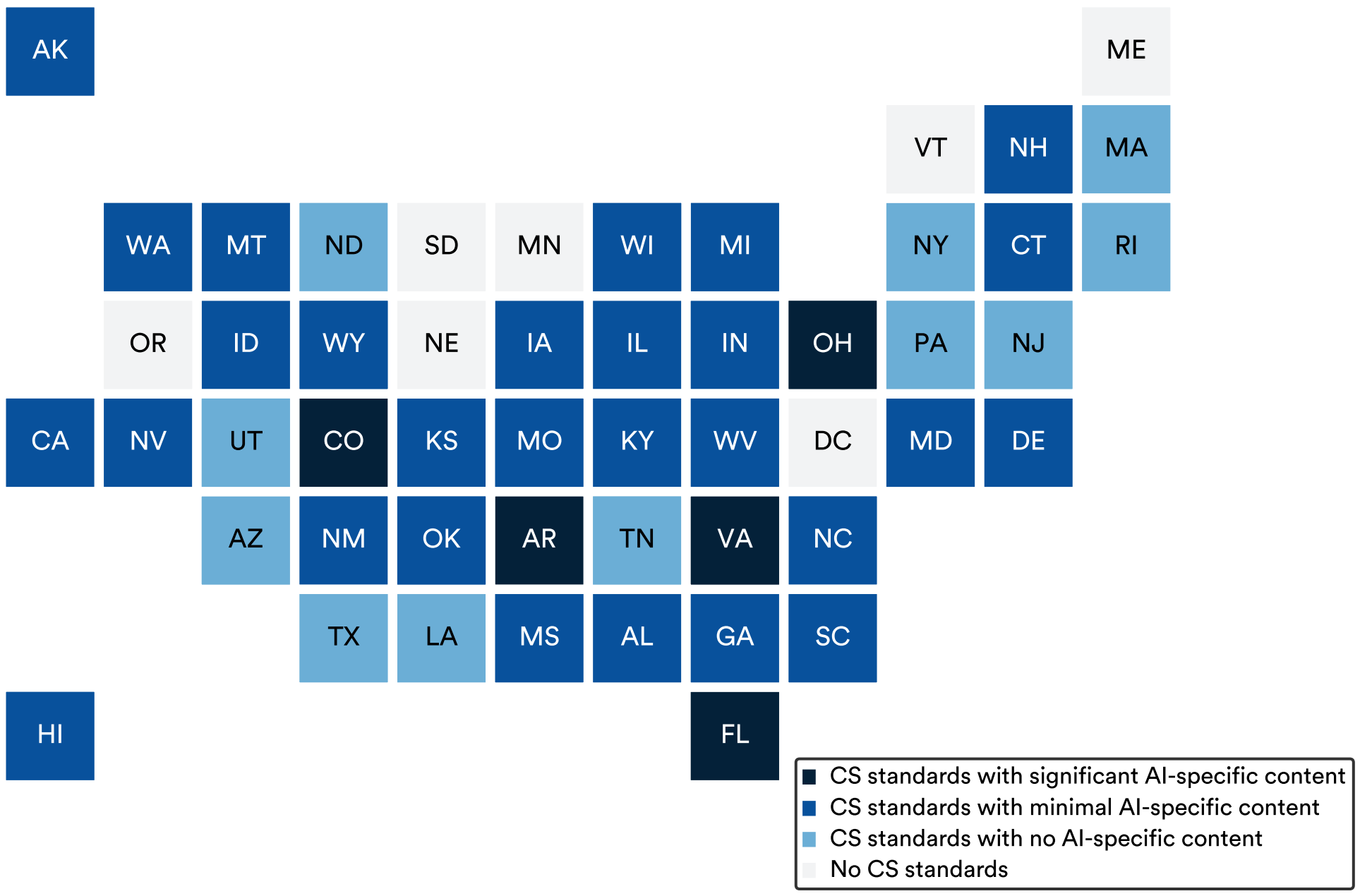

Figure 7.2.13

3 This project is supported by the National Science Foundation (NSF) under Grant No. 2311746. Any opinions, findings, and conclusions or recommendations expressed in this material are those of the author(s) and do not necessarily reflect the views of the NSF.

## Teacher Perspectives

To examine the perspectives and practices of CS teachers as it relates to AI education, the Computer Science Teacher Landscape Survey collected data from 2,901 pre-K through 12 CS teachers nationally (33% of respondents were elementary school teachers, 36% taught middle school, and 51% taught high school).[4,5]

As AI education gains importance for future workforce readiness, it is important to understand the preparedness of the current educator workforce. While 81% of CS teachers believe AI should be included in foundational CS education, less than half feel equipped to teach it—46% in high school, 44% in middle school, and just 34% in elementary school (Figure 7.2.14).

When asked to identify the CS-related topics they cover in class, over two-thirds of middle and high school CS teachers stated they cover AI specifically, despite the lack of explicit definition in CS standards; fewer elementary teachers (65%) reported covering AI (Figure 7.2.15). Greater proportions of CS teachers said they include *components* of AI, such as algorithms, computing systems, computational thinking, and programming.

**Percentage of teachers who feel equipped to teach AI by grade level**
Source: Computer Science Teacher Landscape Survey, 2024 | Chart: 2025 AI Index report

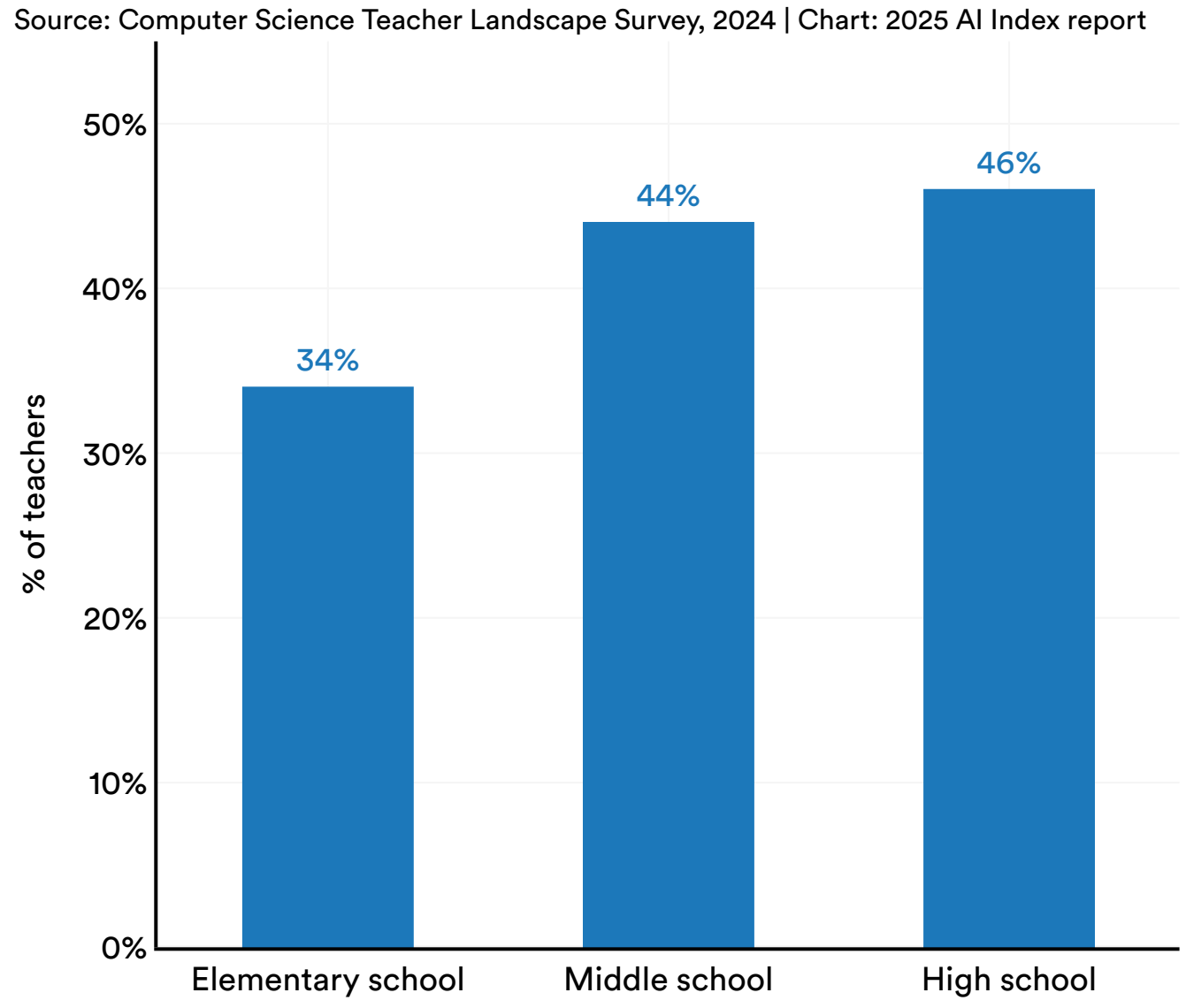

Figure 7.2.14

**AI concepts taught in CS classrooms by grade level**
Source: Computer Science Teacher Landscape Survey, 2024 | Chart: 2025 AI Index report

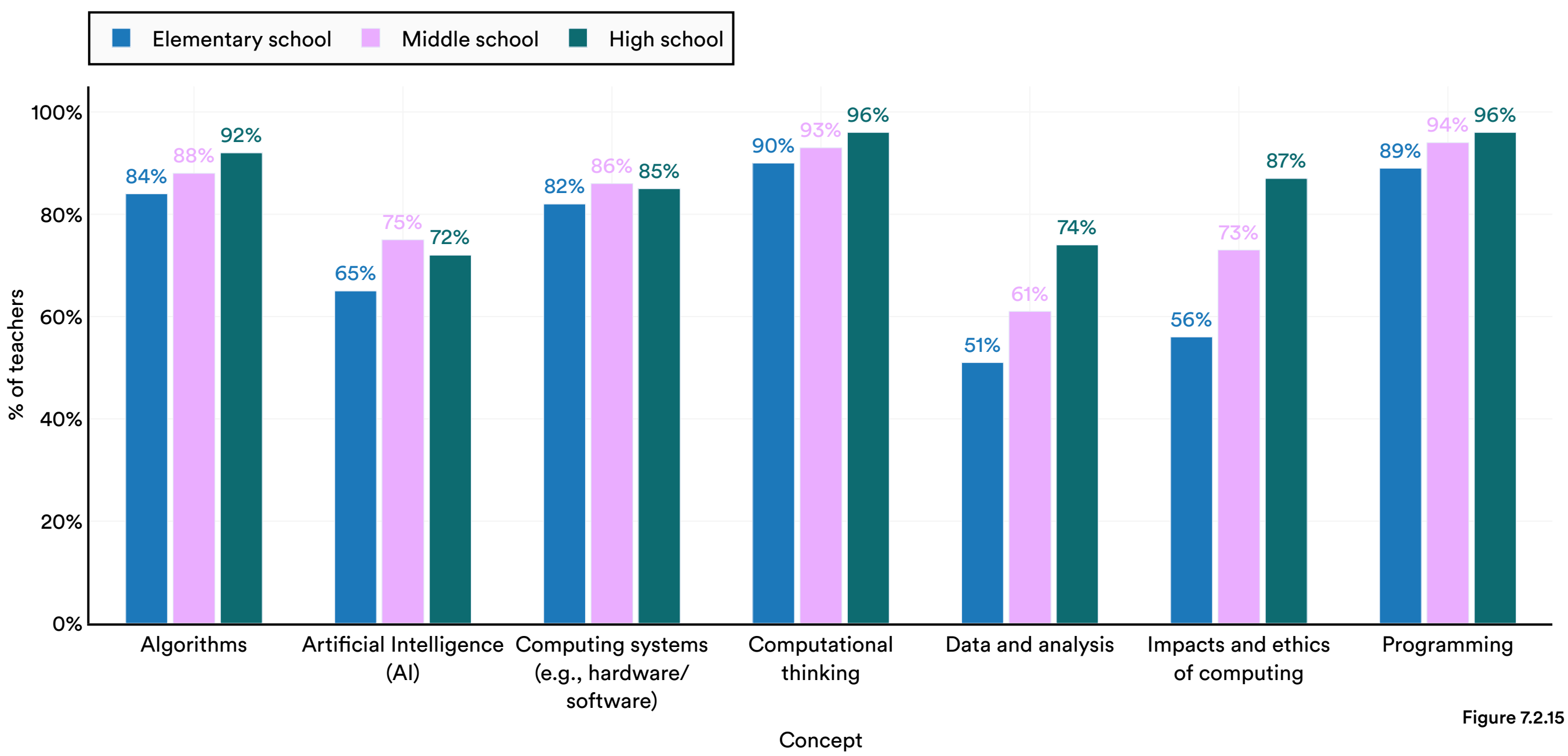

Figure 7.2.15

4 This project is supported by the National Science Foundation (NSF) under Grant No. 2118453. Any opinions, findings, and conclusions or recommendations expressed in this material are those of the author(s) and do not necessarily reflect the views of the NSF. Survey responses may not total 100%, as some questions allowed respondents to select multiple options.

5 The percentages in the figure do not sum to 100% because respondents could select multiple options if they taught more than one grade level.

Of the 2,245 teachers who did spend class time on AI content, the majority spent fewer than five hours per course. Elementary school teachers spent the least amount of time, with 70% spending only one to two hours (Figure 7.2.16).

### Time spent learning AI in CS classrooms by grade level

Source: Computer Science Teacher Landscape Survey, 2024 | Chart: 2025 AI Index report

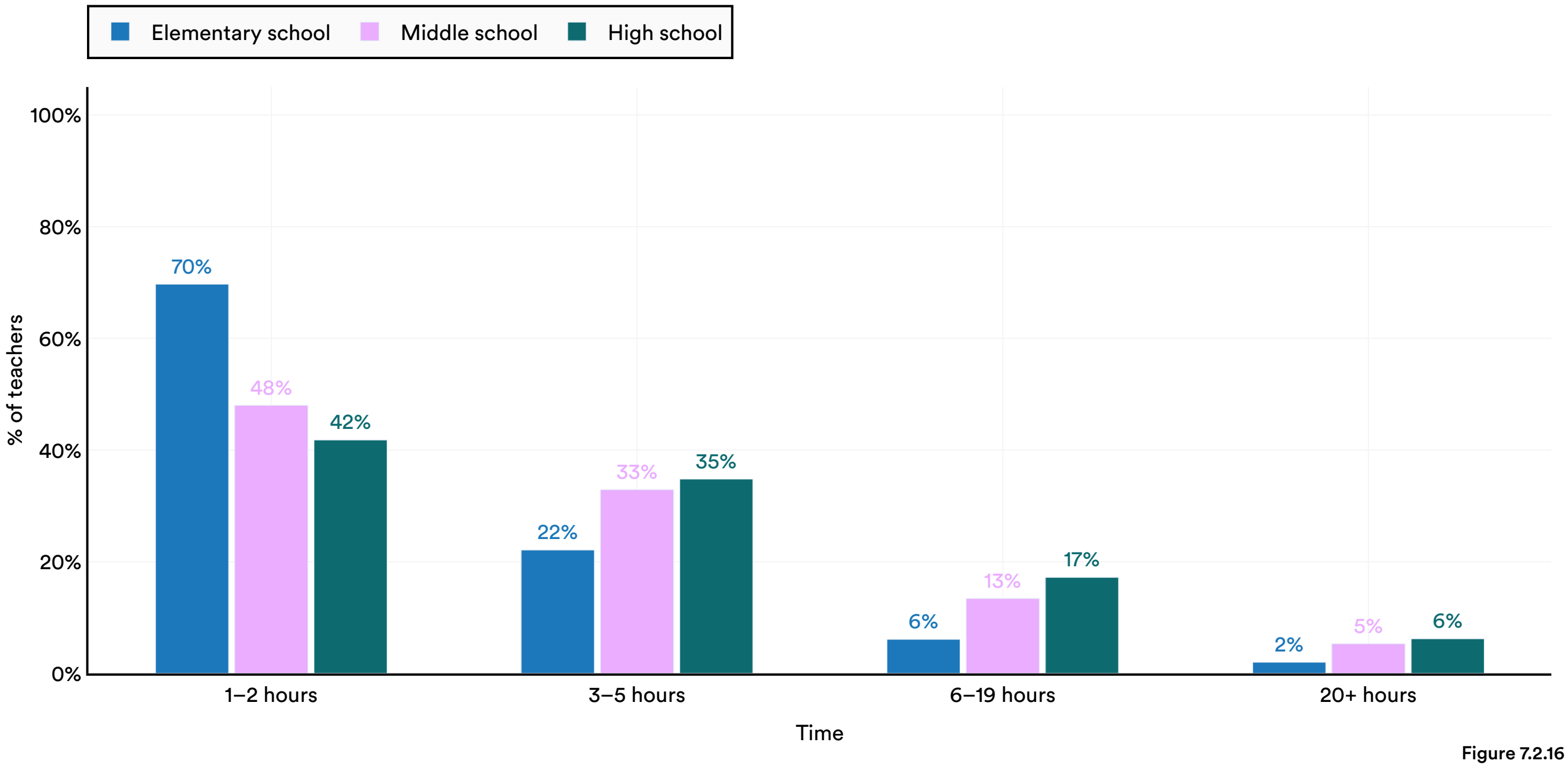

Figure 7.2.16

When asked to name the greatest benefits of using AI in the classroom, teachers most commonly said improving their productivity, differentiating student learning, providing improved academic support to students, and preparing students for the future. When asked about the greatest risks, teachers' greatest concerns were the misuse of AI (often related to academic integrity); that AI use could limit student learning or engagement; overreliance on the technology; that AI could generate misinformation and replicate biases; and other ethical concerns, including student privacy.

To equip students to use AI responsibly, the educator workforce must be upskilled. In a 2024 survey of 364 CS teachers, 88% identified the need for more resources for AI-related professional development. When asked to identify specific resources, CS teachers said they needed to gain more AI literacy (e.g., how AI works, how to use AI, and the ethical impacts of AI).

# Global

Thus far, very few countries (e.g., Ghana, South Korea, Netherlands) include AI education in their curricula explicitly; countries more often flag the importance of AI education in the national education strategy conversation without providing a detailed implementation plan. Because AI education has historically been subsumed under CS or information and communications technology (ICT) education, tracking CS and/or ICT education will serve as a proxy for tracking AI education in this analysis. Similar to the challenges inherent in tracking CS education in the United States, caution is called for when interpreting global metrics because CS and ICT education are sometimes conflated with digital or computer literacy.[6]

### Access

In 2024, approximately two-thirds of the world's countries offered or planned to offer CS education (Figure 7.2.17). CS education is mandatory in primary and/or secondary schools in 30% of countries, with Europe home to the highest concentration of these countries. In the past five years, all geographic regions have made progress in offering CS education, with Africa and Latin America registering the largest increases (Figure 7.2.18). Still, students in African countries are the least likely to have access to CS education. This is likely due to infrastructure challenges; in 2023, only 34% of primary schools in sub-Saharan Africa had access to electricity, hindering schools' ability to teach students computer literacy skills, let alone providing them with CS and AI education.

**Availability of CS education by country, 2024**

Source: Raspberry Pi Computing Education Research Centre, 2024 | Chart: 2025 AI Index report

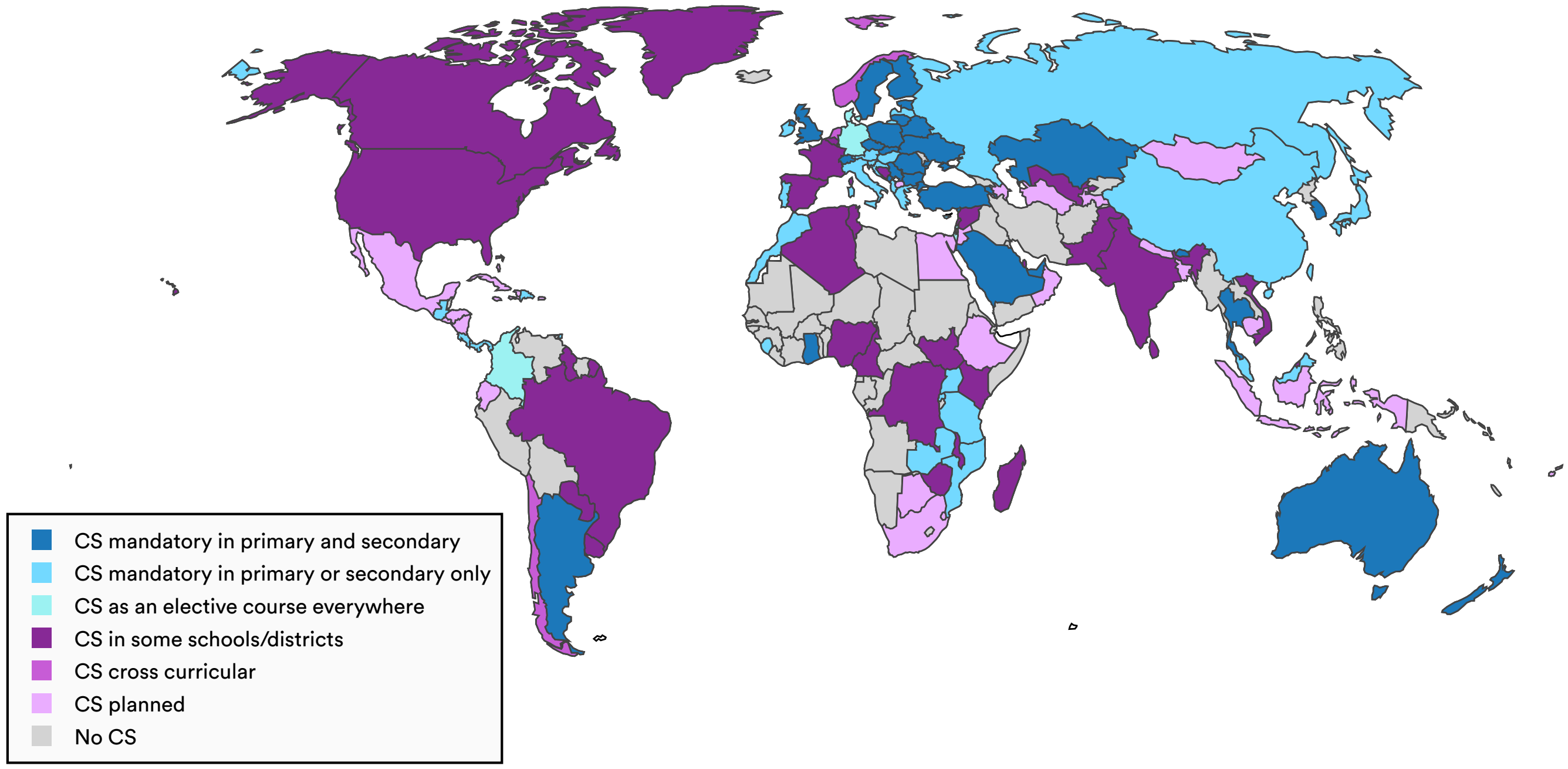

Figure 7.2.17

6 Digital literacy is the "ability to use information and communication technologies to find, evaluate, create, and communicate information, requiring both cognitive and technical skills," whereas computer literacy is the "general use of computers and programs, such as productivity software."

Globally, the lack of standardized data collection makes it challenging to track progress in AI education. Language barriers and infrequent updates on implementation further complicate accurate monitoring across countries.

**Change in access to CS education by continent, 2019 vs. 2024**

Source: Raspberry Pi Computing Education Research Centre, 2024 | Chart: 2025 AI Index report

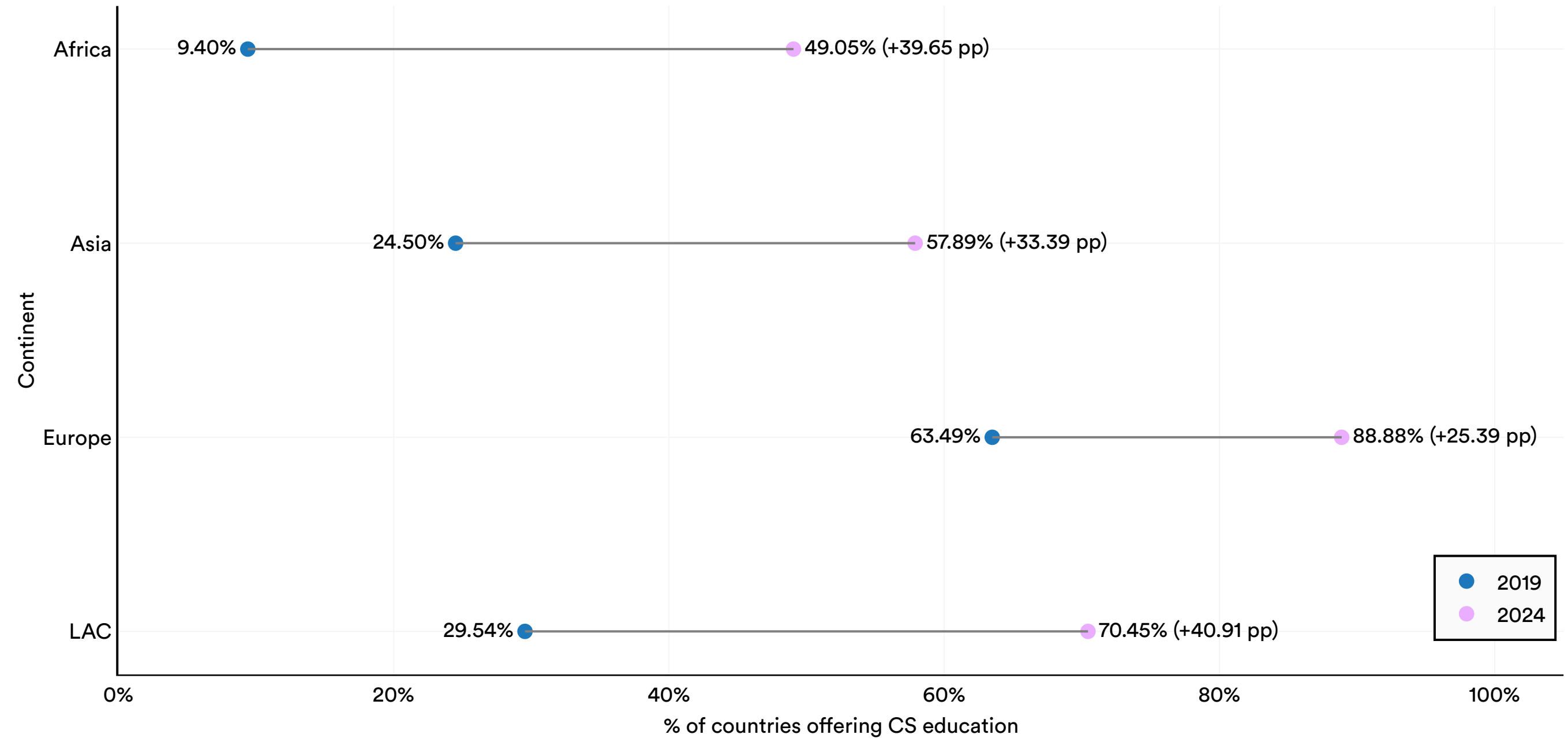

Figure 7.2.18

## Guidance

Countries on a global scale have been quicker to develop guidance and policies for the use of AI *in* education as opposed to developing national standards for teaching AI. As of November 2024, 10 countries have issued guidance on AI in education: Australia, Belgium, Canada, Japan, New Zealand, South Korea, Ukraine, the United Kingdom, the U.S., and Uruguay. This is not surprising given the decade-long conversation across countries about developing guidelines and policy recommendations for AI in education. As early as 2015, United Nations Educational, Scientific, and Cultural Organization (UNESCO) member states committed to harnessing technologies toward ensuring "inclusive and equitable quality education and promoting lifelong learning opportunities for all" (See Sustainable Development Goal 4). Since then, UNESCO published the Beijing Consensus on Artificial Intelligence and Education (in 2019) to offer specific guidance on how to integrate AI technologies to ensure all people have access to quality education by 2030 (See Education 2030 Agenda). Within this set of recommendations, there were four implementation and policy adoption guidelines that touch upon AI concepts in K–12 education.

Similar to the AI4K12 initiative, which released a set of K–12 AI education standards organized around "Five Big Ideas in AI" (Figure 7.2.19), international organizations are also developing AI curricular frameworks for countries to use. Last year, UNESCO published AI competency frameworks for students and teachers. The student framework includes four core competencies: a human-centered mindset, ethics of AI, AI techniques and applications, and AI system design. In each competency, students progress from understanding to applying to creating. In the European Union, many countries rely on DigComp 2.2, a framework for developing citizens' digital competence, along with CS learning objectives for students. The most recent version has guidance on recommended knowledge, skills, and attitudes for interacting with AI, though it does not explicitly include guidance on teaching citizens to build AI systems.

**AI4K12 guidelines organized around 5 Big Ideas in AI**

Source: AI4K12, 2024

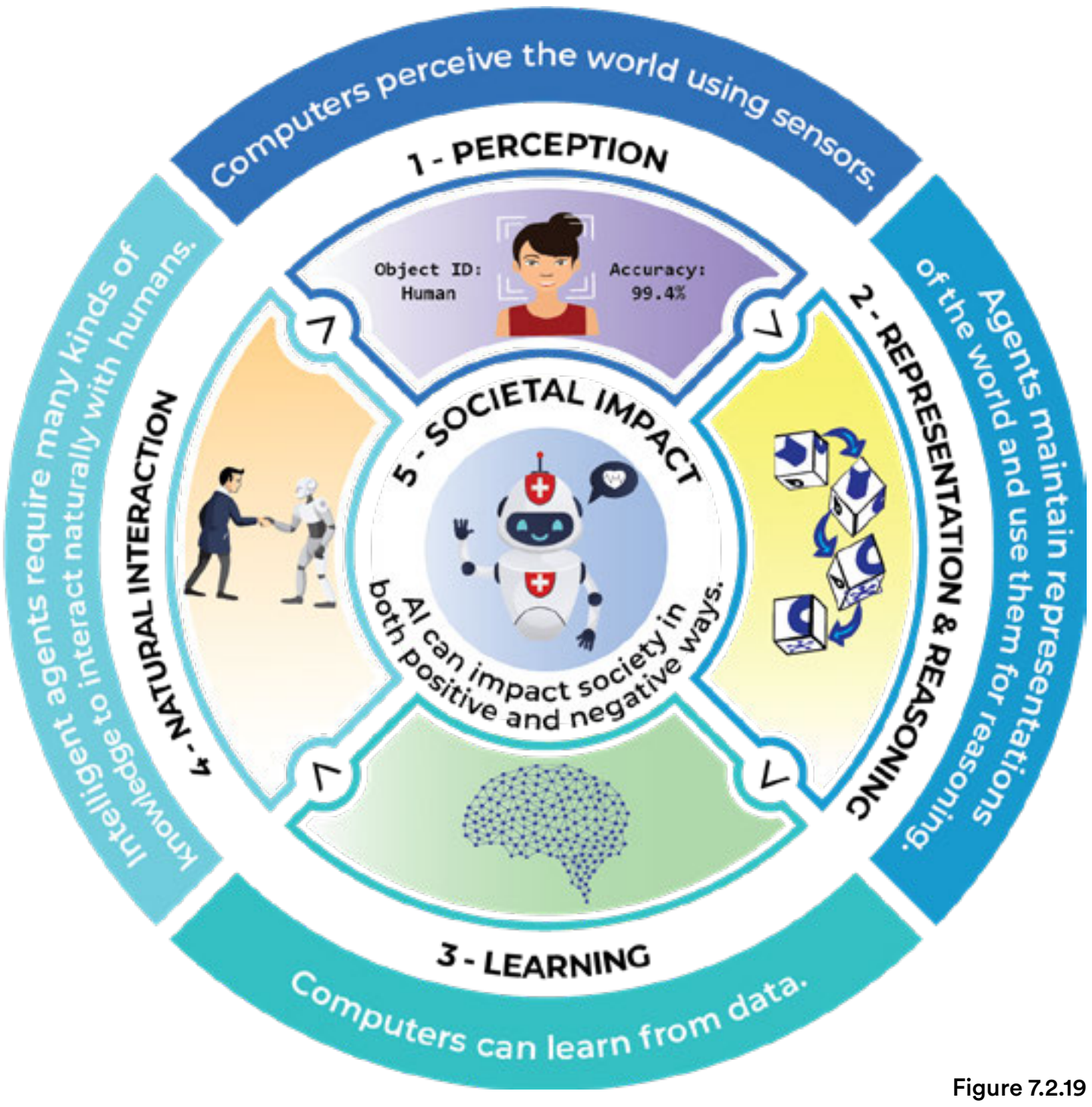

Figure 7.2.19

The role AI will play in the U.S. labor force and the economic future is yet to be fully understood, but its impact is expected to be substantial. The technology workforce already contributes significantly to the U.S. economy, with 9.6 million working as tech employees across industries. While there are strong concerns about displaced employment as a result of automation, projected demands for AI-related roles, such as database management and data infrastructure solutions, are likely to increase. Therefore, a global commitment to ensure postsecondary institutions are equipped to train the future workforce and expand the computing pipeline is essential.

# 7.3 Postsecondary CS and AI Education

## Degree Graduates

### United States

Data on U.S. postsecondary CS and AI education trends in this section comes from the National Center for Education Statistics (NCES). Notably, the Classification of Instructional Programs (CIP), a national standard for classifying academic programs, was developed by NCES under the U.S. Department of Education. In 2016, AI-specific curricula were designated under CIP code 11.0102, which covers programs focused on "symbolic inference, representation, and simulation by computers and software of human learning and reasoning processes and capabilities, and the computer modeling of human motor control and motion. Includes instruction in computing theory, cybernetics, human factors, natural language processing, and applicable aspects of engineering, technology, and specific end-use applications."

While the number of students earning associate degrees in CS has largely remained stable over the past decade, several community colleges are also pioneering AI education, offering certificate and both associate and bachelor's degree programs in AI and related fields (Figure 7.3.2). Notable examples include Maricopa Community Colleges, Houston Community College, Miami Dade College, and several schools in the Bay Area Community College Consortium.

The number of graduates with bachelor's degrees in computing has increased 22% over the last 10 years (Figure 7.3.1). In 2023, the top five producers of CS bachelor's graduates were Western Governors University, University of California–Berkeley, Southern New Hampshire University, University of Texas at Dallas, and University of Michigan.[7] While the increased attention on AI will be slower to show at the bachelor's degree level, given its four-year cycle, AI's explosive growth has already become visible in master's degrees, with a 26% increase in CS graduates between 2022 and 2023, and an overall increase of 83% in the last decade.

7 Western Governors University and Southern New Hampshire University are primarily online institutions.

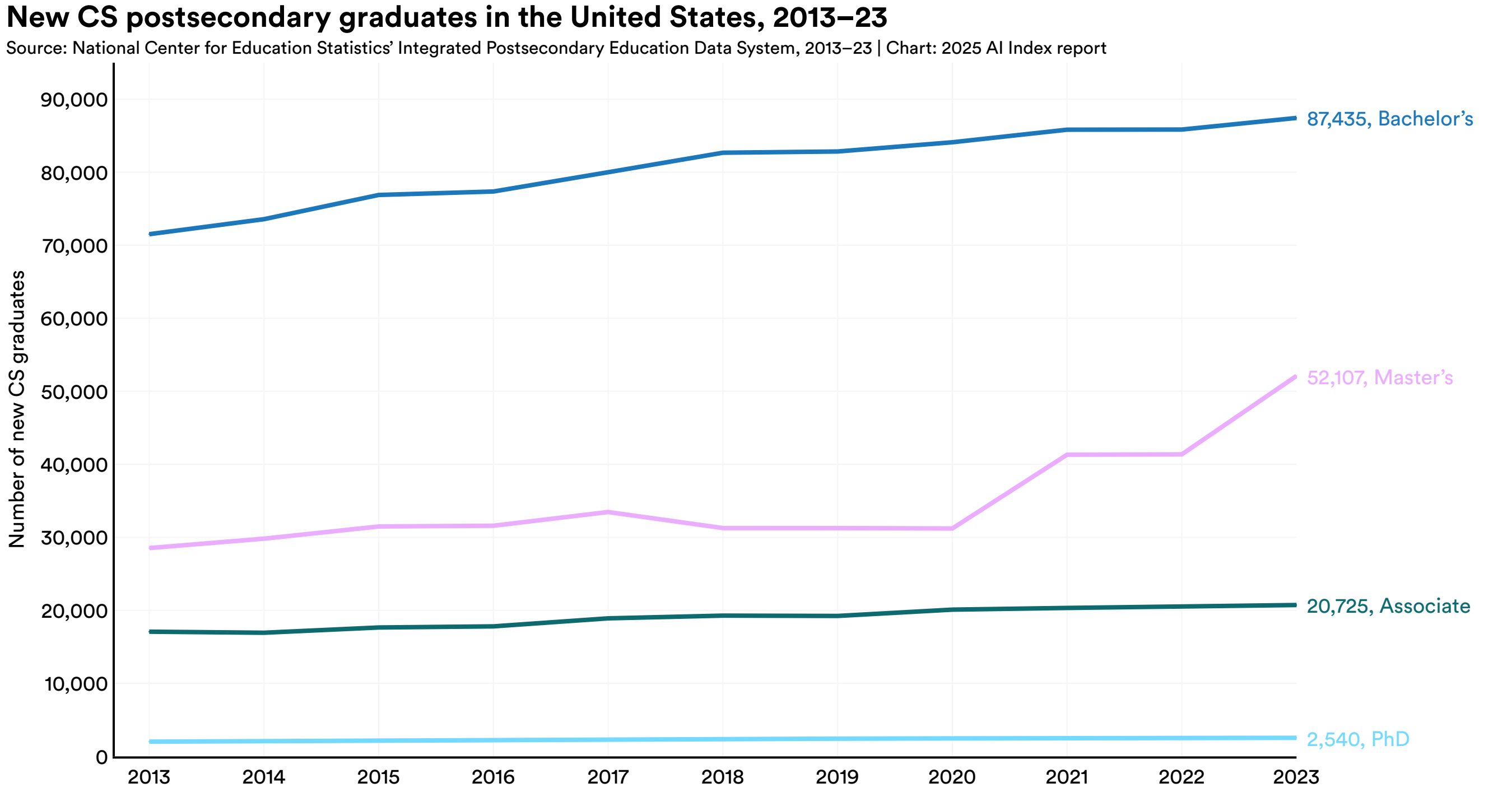

Figure 7.3.1

Despite the fact that women graduate from college at higher rates than men, degree completion data shows an underrepresentation of women in CS (Figure 7.3.2).

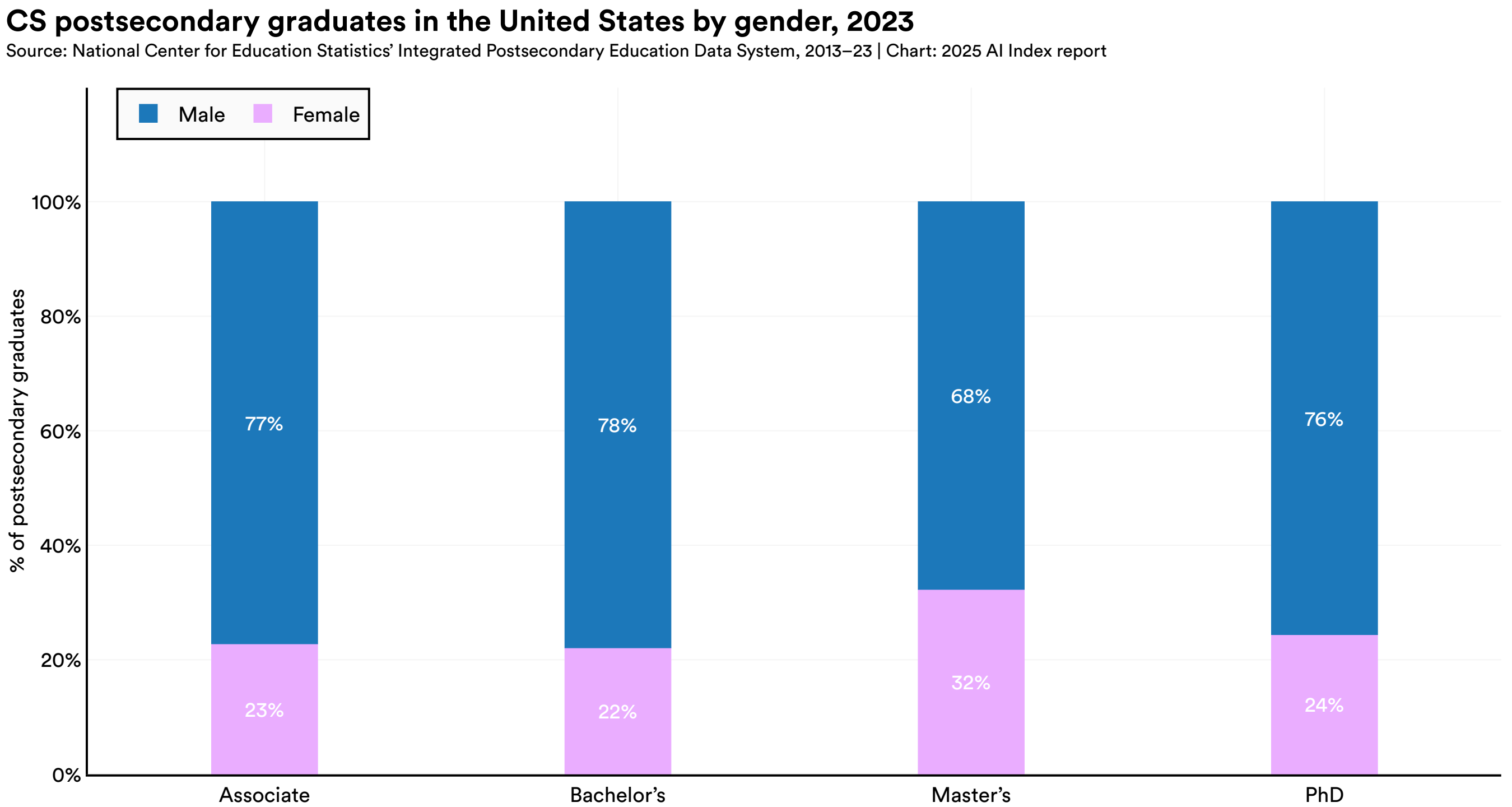

Figure 7.3.2

Black students account for 8% of bachelor's degrees, 8% of master's degrees, and 7% of PhDs in computing (Figure 7.3.3). Hispanic students account for 13% of bachelor's degrees, 8% of master's degrees, and 4% of PhDs in computing. By contrast, white students account for 46% of bachelor's degrees and over half (52%) of PhDs in computing; and Asian students are overrepresented in the postsecondary computing space, accounting for 23% of bachelor's degrees, 28% of master's degrees, and 17% of PhDs.

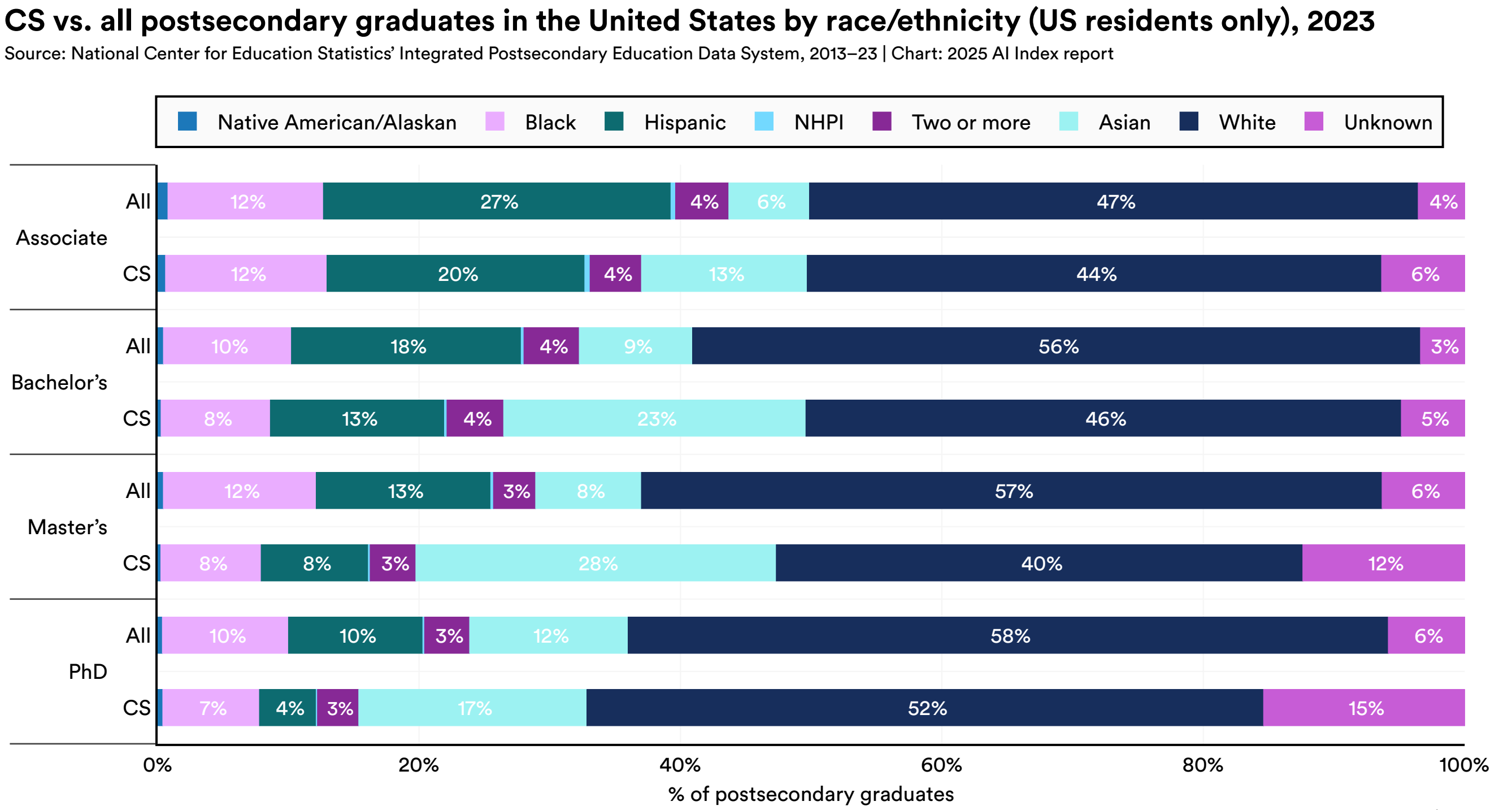

Figure 7.3.3

The majority of students in computing-related graduate programs are from countries outside of the U.S.—a percentage that has steadily grown over the years. In 2023, nonresidents accounted for 67% of master's degree graduates and 60% of PhD graduates. Between 2022 and 2023, international CS master's students increased more than twofold, growing from 15,811 to 34,850 (IPEDS). Students from India and China make up the vast majority of this graduate student body (93% of the 95,130 international master's students and 60% of the 13,070 international PhD students) (Figure 7.3.4 and Figure 7.3.5).

The number of institutions in the U.S. that offer an AI-specific bachelor's degree nearly doubled between 2022 and 2023, while the number of institutions offering an AI-specific master's degree has sharply increased as well (Figure 7.3.6).

### Number of international CS master's students enrolled in US universities, 2022

Source: National Science Board; National Science Foundation, 2023 | Chart: 2025 AI Index report

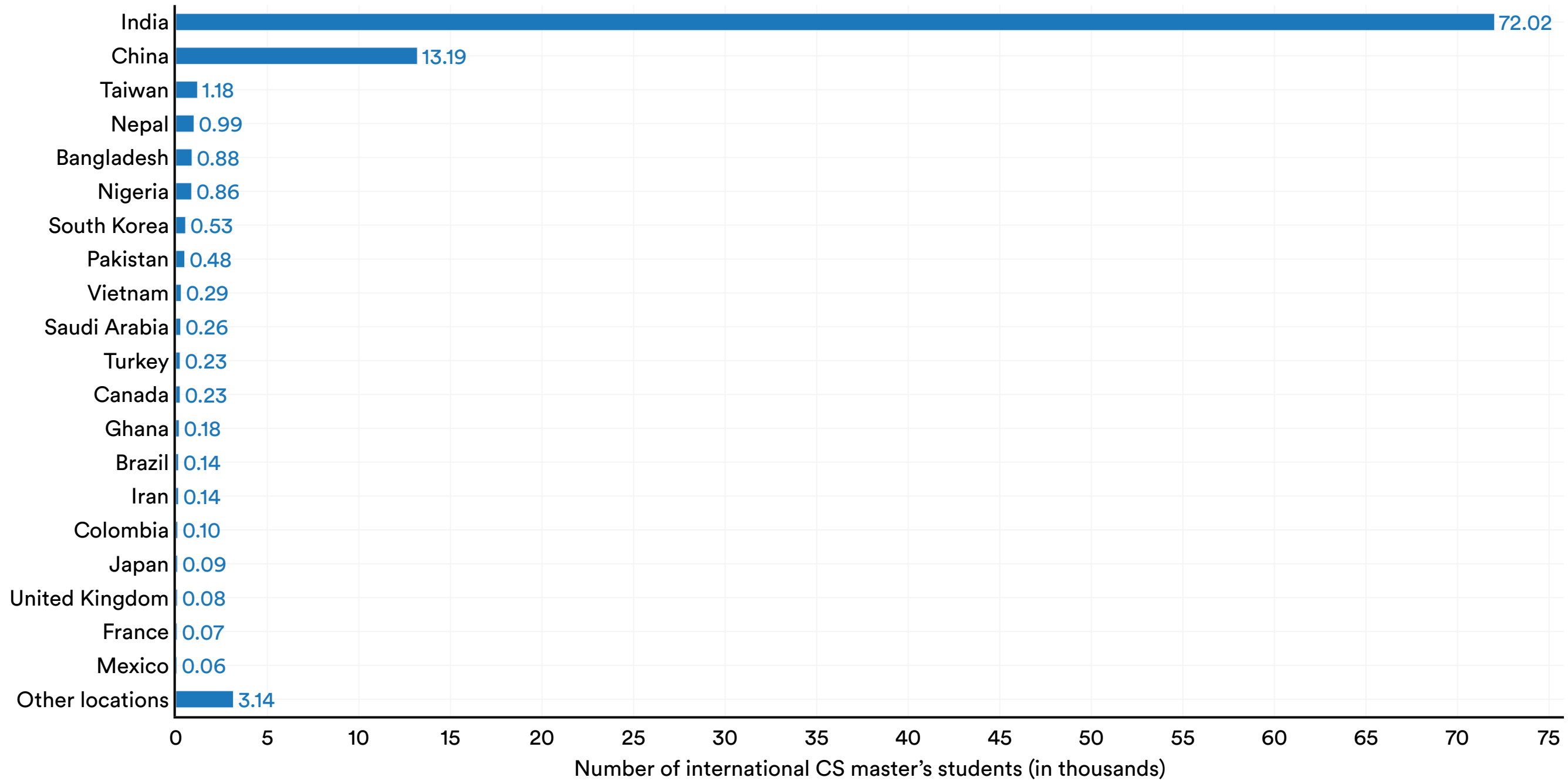

Figure 7.3.4

### Number of international CS PhD students enrolled in US universities, 2022

Source: National Science Board; National Science Foundation, 2023 | Chart: 2025 AI Index report

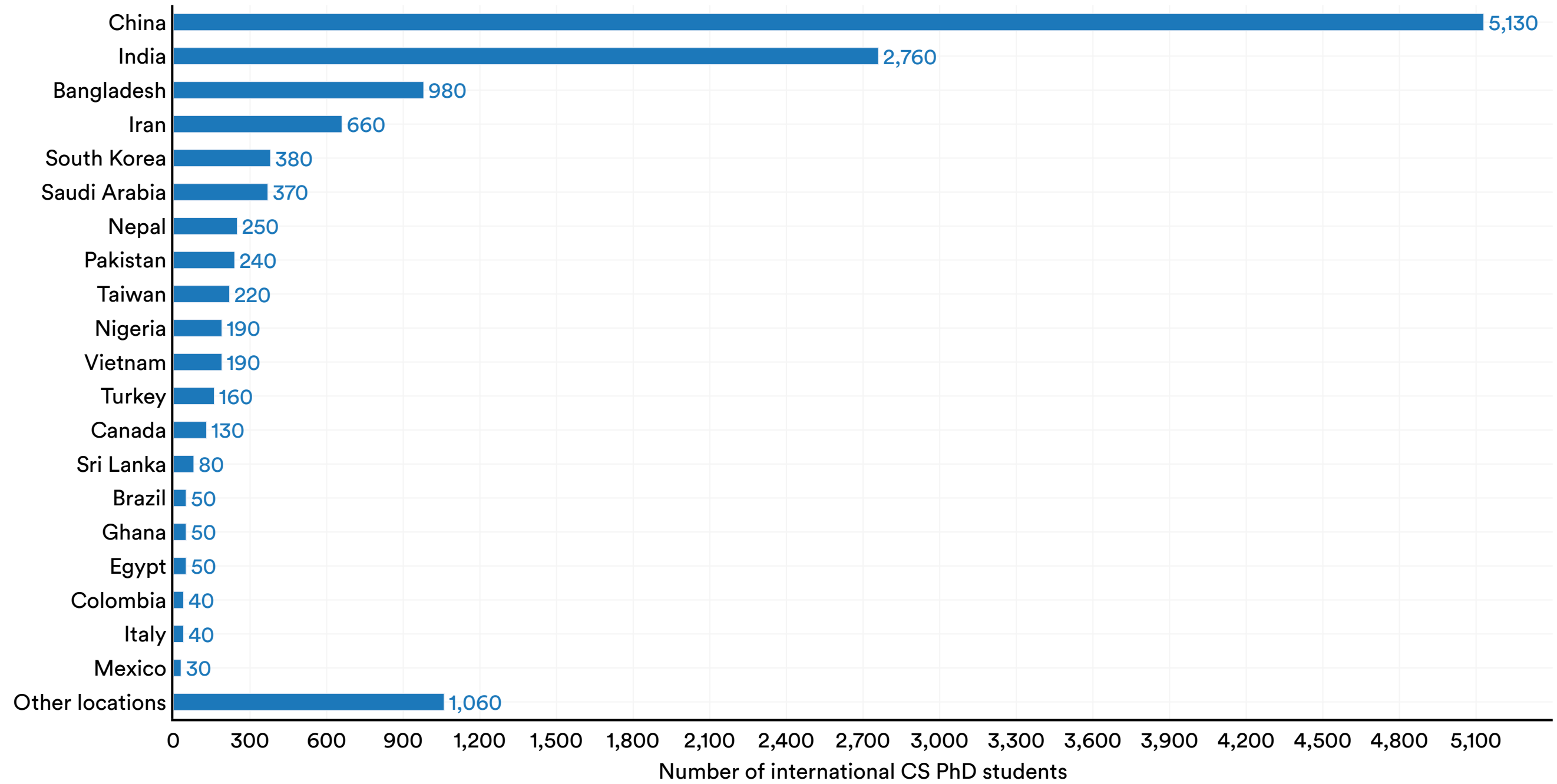

Figure 7.3.5

## Number of institutions offering AI bachelor's and master's degrees in the US, 2013–23

Source: National Center for Education Statistics' Integrated Postsecondary Education Data System, 2013–23 | Chart: 2025 AI Index report

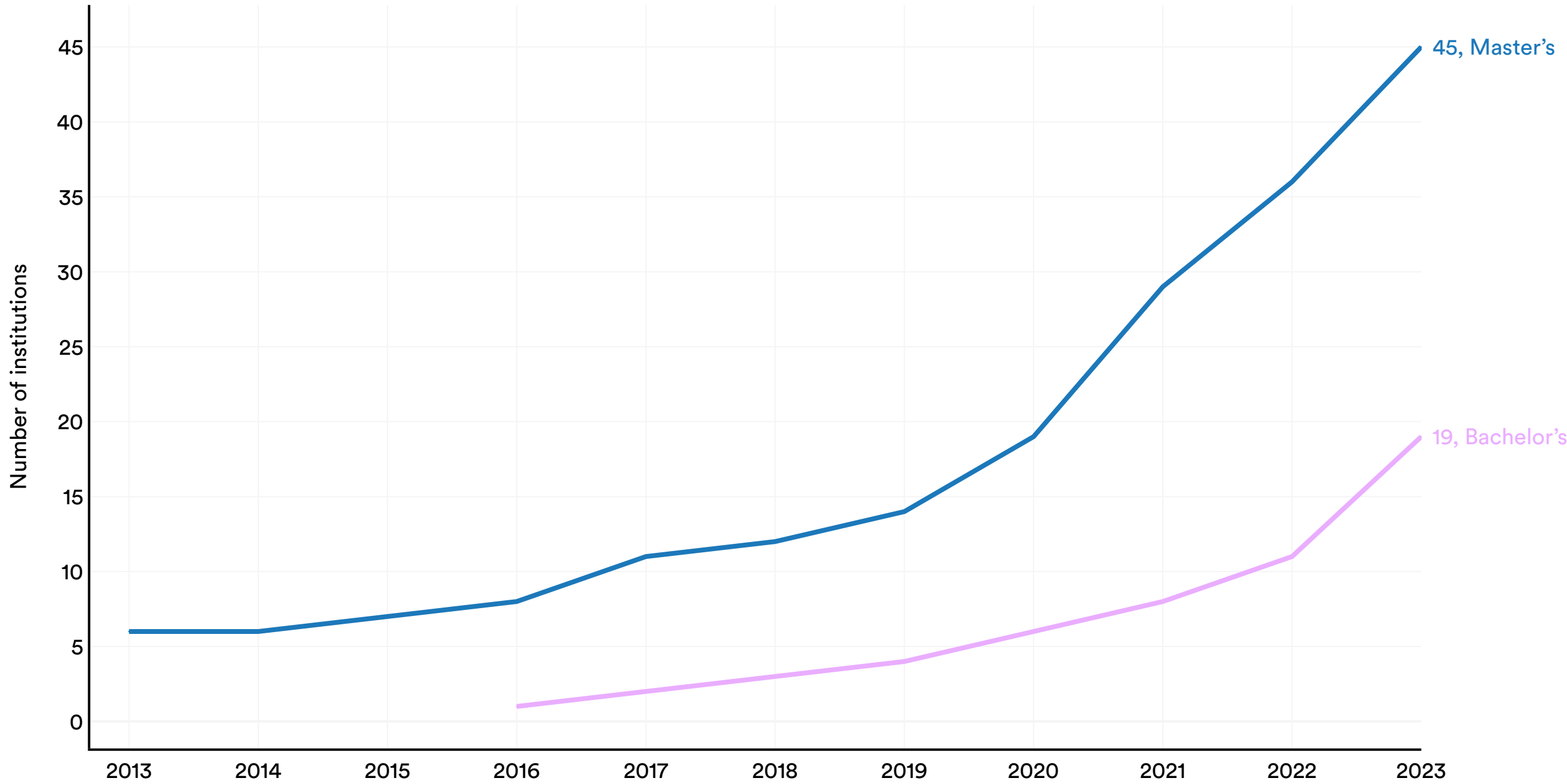

Figure 7.3.6

There was a sharp increase in students graduating with master's degrees in AI between 2022 and 2023 (Figure 7.3.7). Carnegie Mellon University, which graduated more AI majors than any other institution, doubled its number of graduates; meanwhile, Pennsylvania State University had its first graduating class in 2022 (Figure 7.3.8). Until recently, Carnegie Mellon was one of the only universities to offer dedicated programs in AI.

## New AI bachelor's and master's graduates in the United States, 2013–23

Source: National Center for Education Statistics' Integrated Postsecondary Education Data System, 2013–23 | Chart: 2025 AI Index report

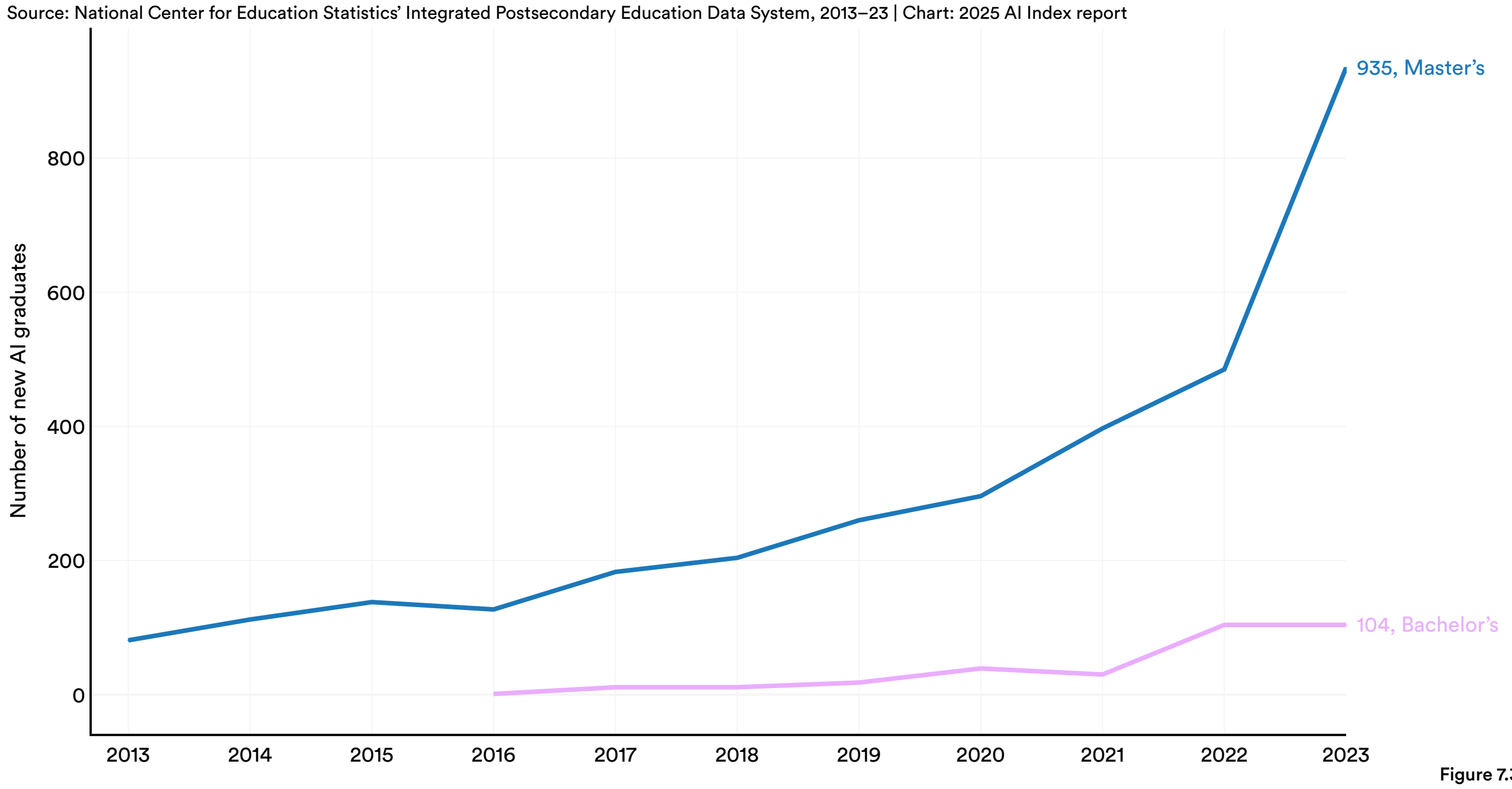

Figure 7.3.7

### Top postsecondary institutions graduating students in AI in 2023 by degree type[8]

**Source:** National Center for Education Statistics' Integrated Postsecondary Education Data System, 2023

| | |
|---|---|
| **Graduates in AI Bachelor's Programs** | |
| Carnegie Mellon University | 32 |
| Full Sail University | 19 |
| Concordia University Wisconsin | 16 |
| University of Advancing Technology | 10 |
| Pennsylvania State University-Main Campus | 7 |
| **Graduates in AI Master's Programs** | |
| Carnegie Mellon University | 178 |
| University of Pennsylvania | 98 |
| University of North Texas | 76 |
| Northeastern University | 55 |
| San Jose State University | 52 |
| **Graduates in AI PhD Programs** | |
| Carnegie Mellon University | 28 |
| Capitol Technology University | 4 |
| University of Pittsburgh-Pittsburgh Campus | 1 |

Figure 7.3.8

8 This list includes only universities that use the AI-specific CIP code for their programs, rather than general CS. However, many students studying AI worldwide are likely enrolled in broader CS programs.

## Global

No single dataset provides a fully standardized accounting of AI or CS postsecondary education across all countries. However, the Organization for Economic Cooperation and Development has compiled data covering its member countries and several non-OECD nations.[9] The International Standard Classification of Education is used to compare education statistics relied on by the OECD to evaluate global progress. Information and communications technologies, or ICT, includes such areas of study as “informatics, information and communication technologies, or CS. These subjects include a wide range of topics concerned with the new technologies used for the processing and transmission of digital information, including computers, computerised networks (including the Internet), microelectronics, multimedia, software and programming.”

The U.S. remains a global leader in ICT-related fields, producing more graduates at each of the associate, bachelor’s, master’s, and PhD levels than any other country included in the sample (Figures 7.3.9 to 7.3.12). Notably, the U.S. graduates more than twice as many associate, master’s, and PhD students—and nearly twice as many bachelor’s students—as the next highest country (Figure 7.3.9).

**New ICT short-cycle tertiary graduates by country, 2022**
Source: OECD, 2022 | Chart: 2025 AI Index report

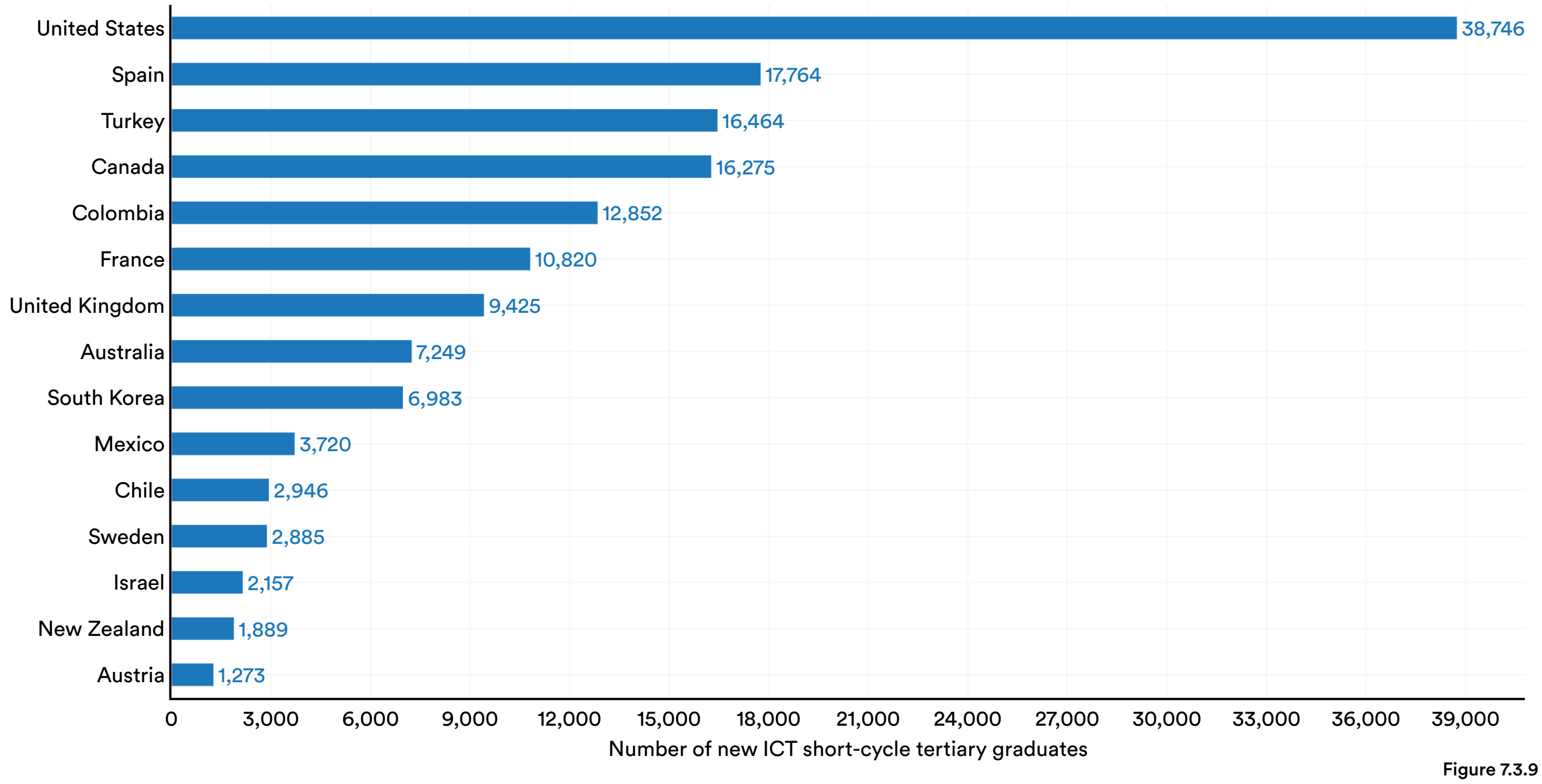

Figure 7.3.9

9 While this dataset provides insights across some country lines, it omits a number of countries likely to have large numbers of ICT graduates. The exclusion of India, China, and countries in Africa highlights the need for global standardized data collection to ensure inclusion of countries that have made significant investments in computing education and make up a significant proportion of the global majority. There is also a significant lag in collecting and reporting global data on education; as a result, the most recent year for which data is available is 2022.

## New ICT bachelor's graduates by country, 2022

Source: OECD, 2022 | Chart: 2025 AI Index report

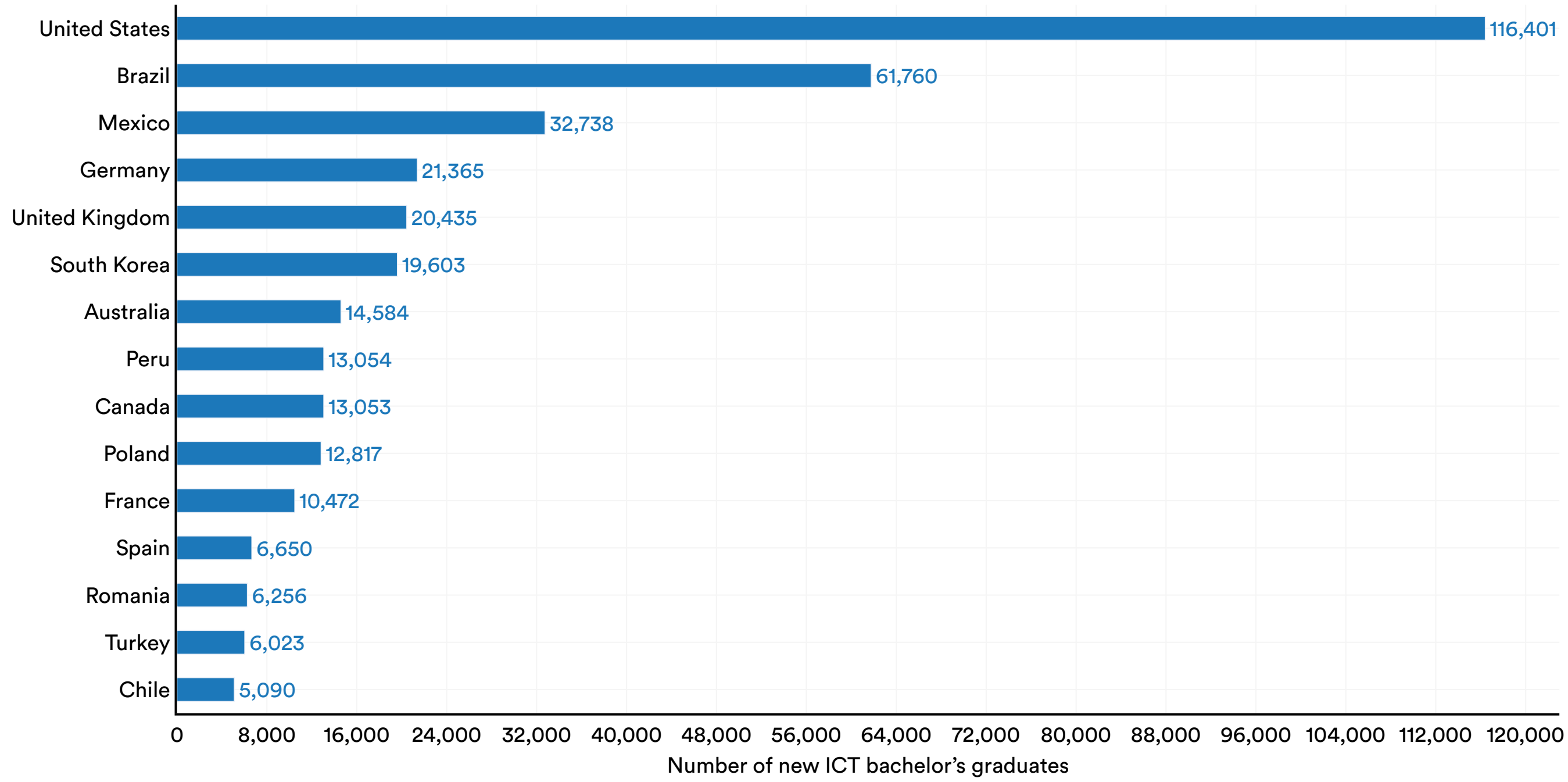

Figure 7.3.10

## New ICT master's graduates by country, 2022

Source: OECD, 2022 | Chart: 2025 AI Index report

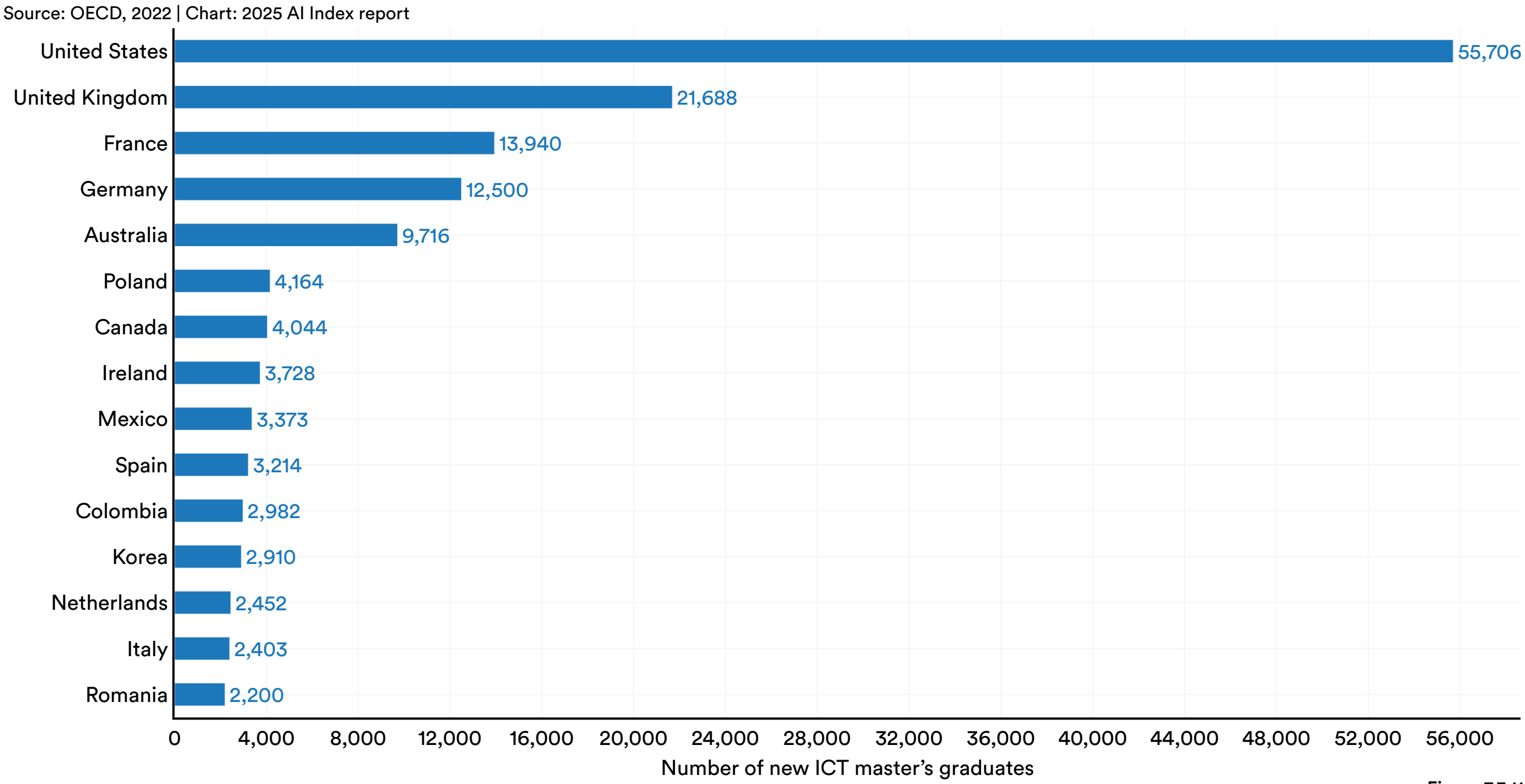

Figure 7.3.11

New ICT PhD graduates by country, 2022

Source: OECD, 2022 | Chart: 2025 AI Index report

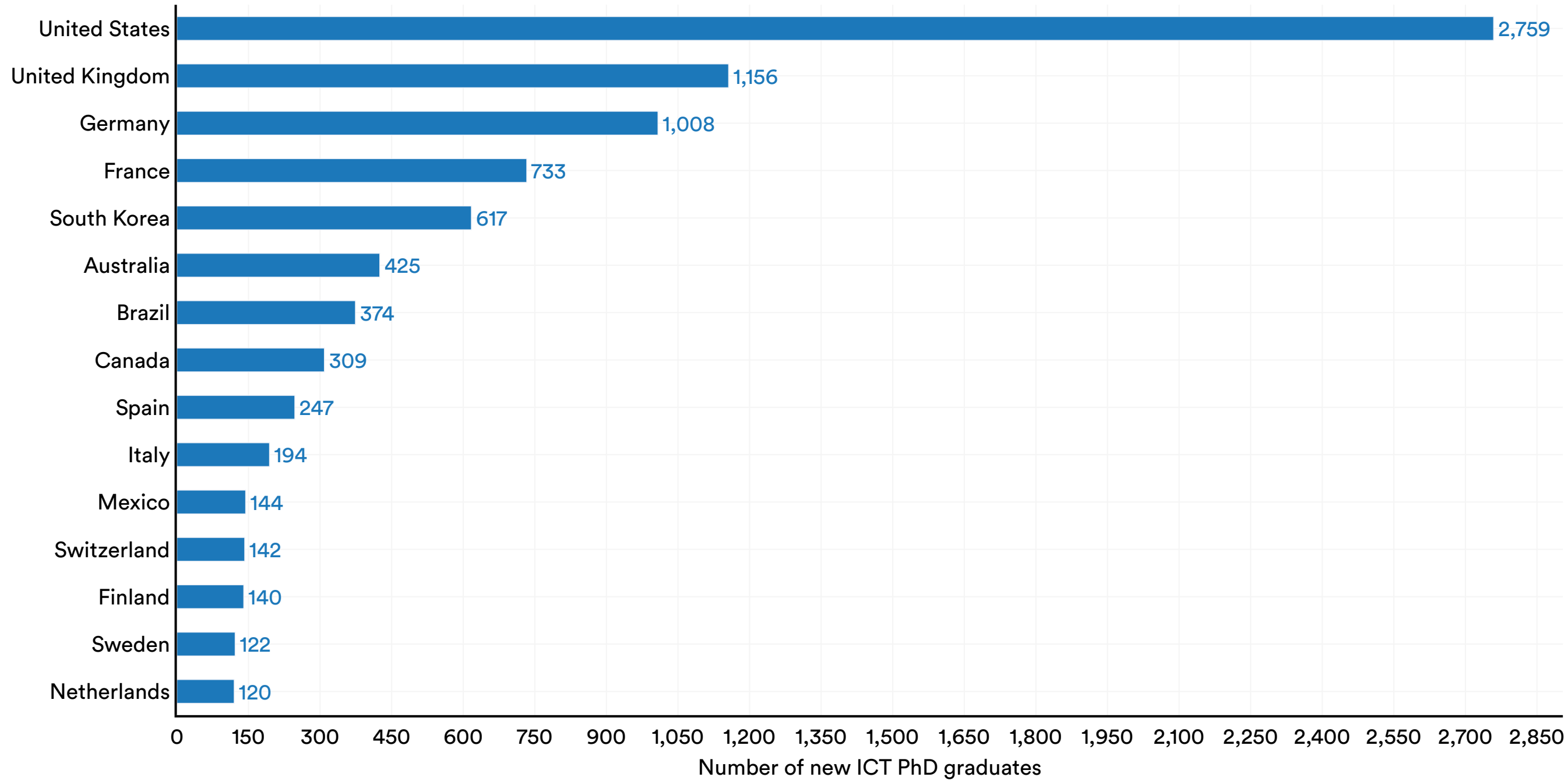

Figure 7.3.12

Gender parity in AI-related fields continues to be a challenge globally (Figure 7.3.13). On average, women comprise approximately one-quarter of ICT postsecondary graduates at the associate, bachelor's, and PhD levels. Women fare slightly better at the master's level, comprising closer to one-third of graduates. Turkey is among the countries that fare best with respect to gender parity, with women there comprising at least half of all graduates at the associate, bachelor's, master's, and PhD levels.

### Percentage of new ICT postsecondary graduates who are female by country, 2022

Source: OECD, 2022 | Chart: 2025 AI Index report

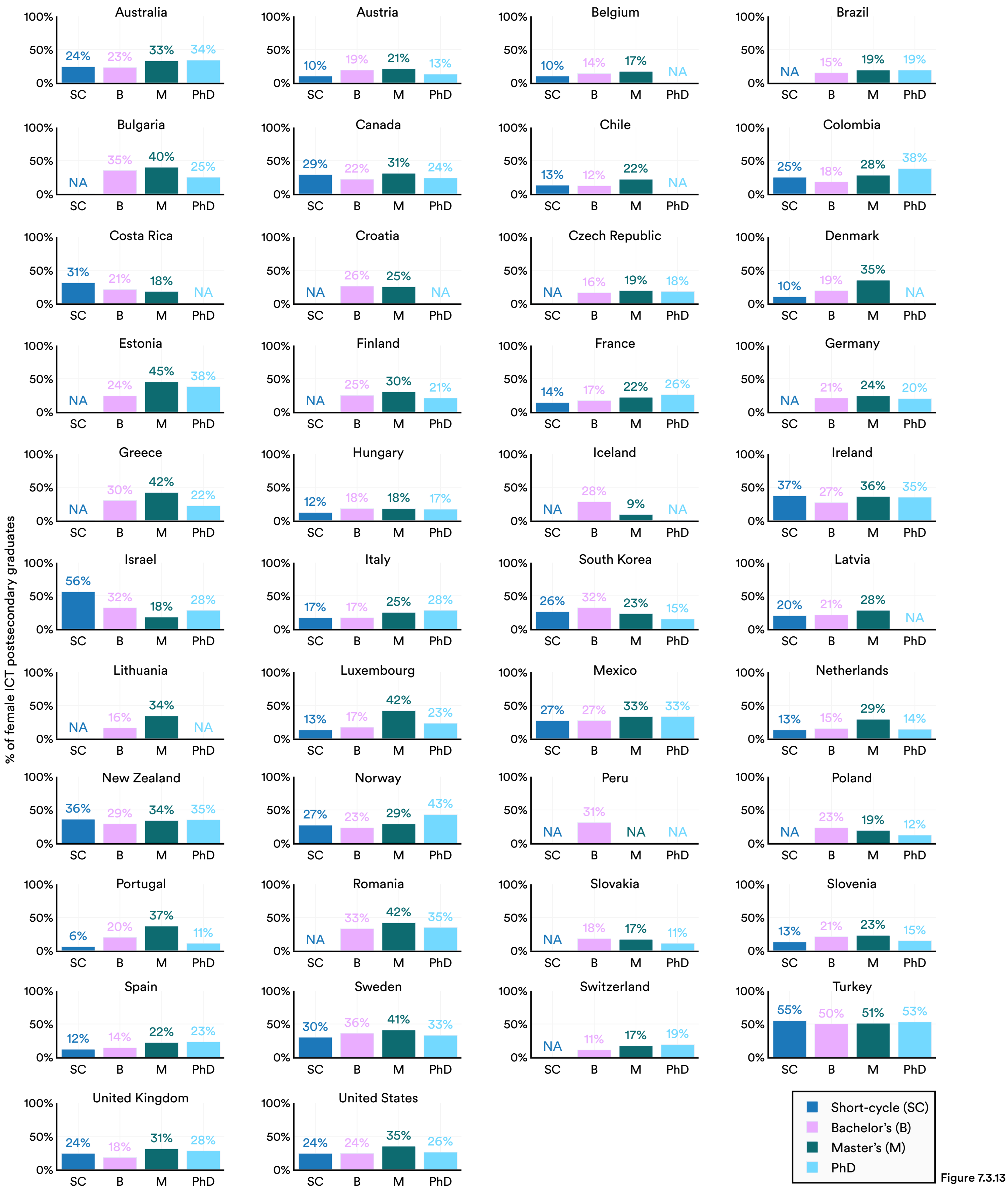

Figure 7.3.13

## Guidance

Most existing university policies and guidance around AI pertain to how students use AI for assignments; guidance on AI education itself tends to be relegated to the department level (primarily in computing departments).

AI is being used across campuses by both students and faculty at high rates: 86% of students use AI in their studies, and 61% of faculty use AI in their teaching. Yet the guidelines around usage still lack clarity and standardization across universities. As of early 2025, 39% of institutions have an AI-related acceptable use policy, an increase of 16 percentage points from 2024. Larger universities (10,000-plus students) are more likely to have a policy than smaller institutions (fewer than 5,000 students). Although teaching and learning policies are the most impacted by AI, almost all institutional policies are affected by technology policies (e.g., purchasing AI tools using university resources, respecting intellectual property/copyright laws, using AI to create malware or viruses)—from cybersecurity and data privacy to online learning and data and analytics.

In addition to the K–12 guidance UNESCO provided in the 2019 Beijing Consensus on Artificial Intelligence and Education, it offered specific guidance that is relevant for both K–12 and postsecondary settings with an eye toward achieving the Education 2030 agenda goals via AI technologies. The 2019 report includes five implementation and policy guidelines pertaining to AI education in postsecondary settings.

# 7.4 Looking Ahead

The intentional design of an equitable AI educational ecosystem will be critical for the responsible development and deployment of future technological innovations. The current systems in which AI has proliferated have led to detrimental outcomes, such as mis/disinformation campaigns to influence national political outcomes, development of AI-enabled weapons, and infringement of copyright-protected intellectual property. The pressing need to prioritize a better approach to building AI is evident. To do so, it is necessary to reimagine an educational program where AI competencies, inclusive of building a lens interrogating the ethics of AI in addition to technical creation, are seen as core to preparing students for a technology-powered future. There are already CS-based infrastructure, policies, and implementation strategies that offer opportunities to integrate AI education more seamlessly. As AI innovations rapidly evolve, transforming education is urgent so that future creators of these technologies are made aware of potential harms and have the competencies to mitigate negative impacts. Academic institutions around the world must continue to progress (and monitor their progress) on creating AI pathways, adopt policies to expand access to relevant courses, and implement strategies to upskill the educator workforce and engage students to participate and build competencies equitably.

# CHAPTER 8:
# Public Opinion

Text and analysis by Emily Capstick

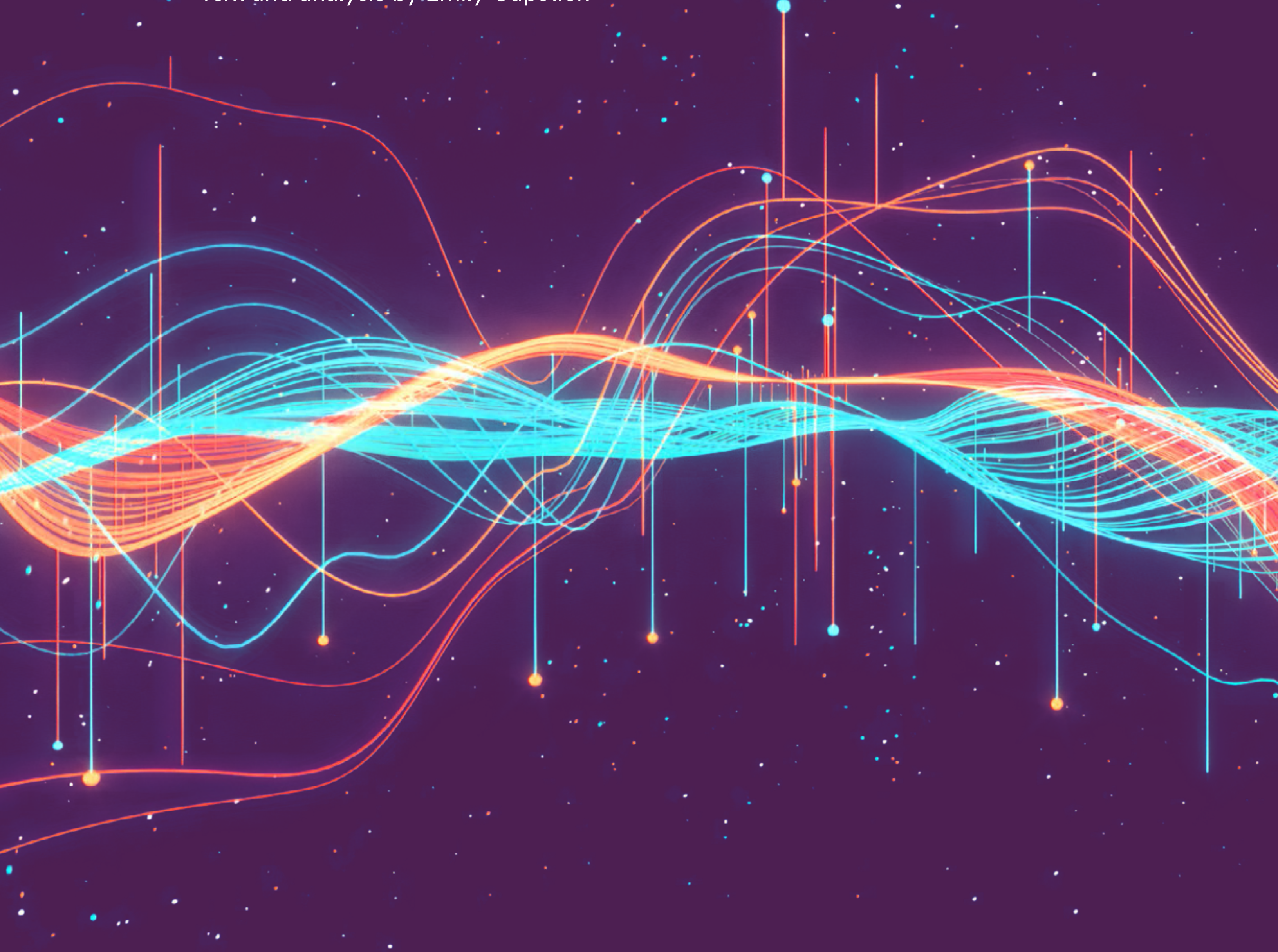

# Chapter 8: Public Opinion

ACCESS THE PUBLIC DATA

**CHAPTER 8:**
Public Opinion

# Overview

As AI continues to permeate broad swaths of society, it is becoming increasingly important to understand public sentiment around the technology. Insights into how people perceive AI can help anticipate its societal impact and reveal how adoption varies across countries and demographic groups. Early data suggests growing public anxiety about AI, with some regions expressing significantly more pessimism than others. As the technology continues to advance, will these trends persist?

This chapter explores public opinion on AI through global, national, demographic, and ethnic perspectives. It draws on multiple data sources, including longitudinal Ipsos surveys tracking global AI attitudes, American Automobile Association surveys on self-driving vehicles, and recent research into local U.S. policymakers' views on AI.

**CHAPTER 8:**
Public Opinion

# Chapter Highlights

**1. The world grows cautiously optimistic about AI products and services.** Among the 26 nations surveyed by Ipsos in both 2022 and 2024, 18 saw an increase in the proportion of people who believe AI products and services offer more benefits than drawbacks. Globally, the share of individuals who see AI products and services as more beneficial than harmful has risen from 52% in 2022 to 55% in 2024.

---

**2. The expectation and acknowledgment of AI's impact on daily life is rising.** Around the world, two thirds of people now believe that AI-powered products and services will significantly impact daily life within the next three to five years—an increase of six percentage points since 2022. Every country except Malaysia, Poland, and India saw an increase in this perception since 2022, with the largest jumps in Canada (17%) and Germany (15%).

---

**3. Skepticism about the ethical conduct of AI companies is growing, while trust in the fairness of AI is declining.** Globally, confidence that AI companies protect personal data fell from 50% in 2023 to 47% in 2024. Likewise, fewer people today believe that AI systems are unbiased and free from discrimination compared to last year.

---

**4. Regional differences persist regarding AI optimism.** First reported in the 2023 AI Index, significant regional differences in AI optimism endure. A large majority of people believe AI-powered products and services offer more benefits than drawbacks in countries like China (83%), Indonesia (80%), and Thailand (77%), while only a minority share this view in Canada (40%), the United States (39%), and the Netherlands (36%).

---

**5. People in the United States remain distrustful of self-driving cars.** A recent American Automobile Association survey found that 61% of people in the U.S. fear self-driving cars, and only 13% trust them. Although the percentage who express fear has declined from its 2023 peak of 68%, it remains higher than in 2021 (54%).

---

**6. There is broad support for AI regulation among local U.S. policymakers.** In 2023, 73.7% of local U.S. policymakers—spanning township, municipal, and county levels—agreed that AI should be regulated, up significantly from 55.7% in 2022. Support was stronger among Democrats (79.2%) than Republicans (55.5%), though both registered notable increases over 2022.

---

**CHAPTER 8:**
Public Opinion

# Chapter Highlights (cont'd)

**7. AI optimism registers sharp increase among countries that previously showed the most skepticism.** Globally, optimism about AI products and services has increased, with the sharpest gains in countries that were previously the most skeptical. In 2022, Great Britain (38%), Germany (37%), the United States (35%), Canada (32%), and France (31%) were among the least likely to view AI as having more benefits than drawbacks. Since then, optimism has grown in these countries by 8%, 10%, 4%, 8%, and 10%, respectively.

---

**8. Workers expect AI to reshape jobs, but fear of replacement remains lower.** Globally, 60% of respondents agree that AI will change how individuals do their job in the next five years. However, a smaller subset of respondents, 36%, believe that AI will replace their jobs in the next five years.

---

**9. Sharp divides exist among local U.S. policymakers on AI policy priorities.** While local U.S. policymakers broadly support AI regulation, their priorities vary. The strongest backing is for stricter data privacy rules (80.4%), retraining for the unemployed (76.2%), and AI deployment regulations (72.5%). However, support drops significantly for a law enforcement facial recognition ban (34.2%), wage subsidies for wage declines (32.9%), and universal basic income (24.6%).

---

**10. AI is seen as a time saver and entertainment booster, but doubts remain on its economic impact.** Global perspectives on AI's impact vary. While 55% believe it will save time, and 51% expect it will offer better entertainment options, fewer are confident in its health or economic benefits. Only 38% think AI will improve health, whilst 36% think AI will improve the national economy, 31% see a positive impact on the job market, and 37% believe it will enhance their own jobs.

# 8.1 Public Opinion

## Global Public Opinion

This section explores global differences in opinions on AI through surveys conducted by Ipsos in 2022, 2023, and 2024. These surveys reveal that public perceptions of AI vary widely across countries and demographic groups.

### AI Products and Services

In 2024, Ipsos ran a survey on global attitudes toward AI. The survey consisted of interviews with 23,685 adults across 32 countries.[1]

Figure 8.1.1 shows the percentage of respondents who agree with specific statements. The increase in public awareness of AI between 2022 and 2024 has remained relatively consistent.

In 2024, 67% of respondents report a good understanding of what AI is, and 66% anticipate that AI will profoundly change their daily life in the near future. The proportion of the global population that perceives AI-powered products and services as having more benefits than drawbacks has increased modestly, rising from 52% in 2022 to 55% in 2024.

Figure 8.1.1 also highlights respondents' growing concerns. In the last year, there has been a three percentage point decrease in those who trust that companies using AI will protect their personal data and a two percentage point decrease in respondents' trust that AI will not discriminate or show bias toward any group of people.

**Global opinions on products and services using AI (% of total), 2022–24**
Source: Ipsos, 2022–24 | Chart: 2025 AI Index report

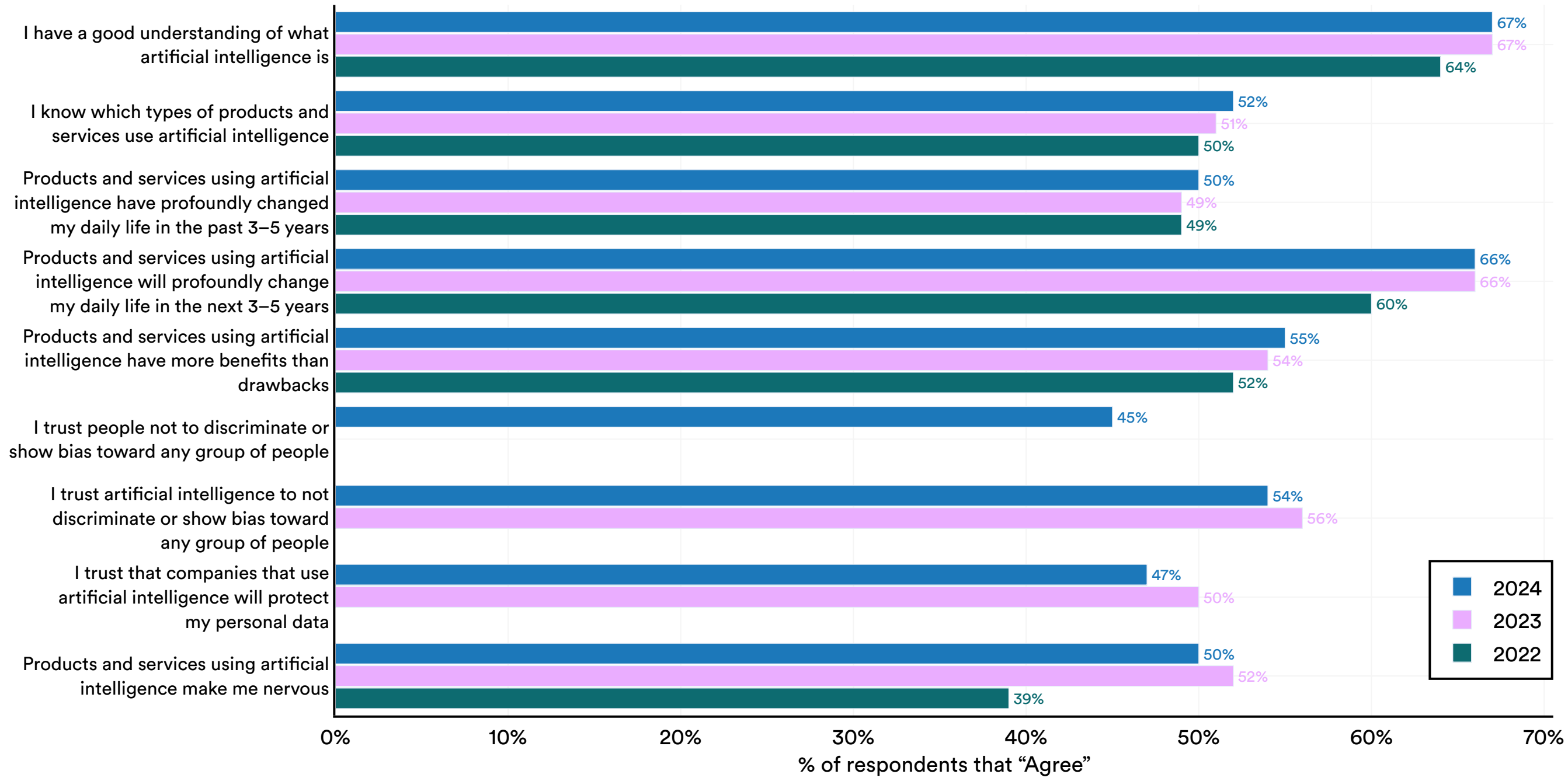

Figure 8.1.1

1 See Appendix for more details about the survey methodology. The survey was conducted from April to May, 2024.

Perceptions of AI's benefits versus drawbacks vary considerably by country, according to the Ipsos survey. In general, respondents in Asia and Latin America believe that AI will have more benefits than drawbacks: 83% of Chinese, 70% of Mexican, and 62% of Indian respondents view AI products and services as more beneficial than harmful (Figure 8.1.2). In contrast, in Europe and the Anglosphere, respondents are more skeptical. For example, 46% of British, 44% of Australian, 40% of Canadian, and 39% of American respondents believe that AI will have more benefits than drawbacks.

AI sentiment appears to be warming, particularly in countries that were once the most skeptical. Among the 26 nations surveyed by Ipsos in both 2022 and 2024, 18 saw an increase in the proportion of people who believe AI products and services offer more benefits than drawbacks. In 2022, France (31%), Canada (32%), the United States (35%), Germany (37%), Australia (37%), and Great Britain (38%) ranked among the least optimistic about AI. By 2024, the percentages in all these countries had risen.

**'Products and services using AI have more benefits than drawbacks,' by country (% of total), 2022–24**
Source: Ipsos, 2022–24 | Chart: 2025 AI Index report

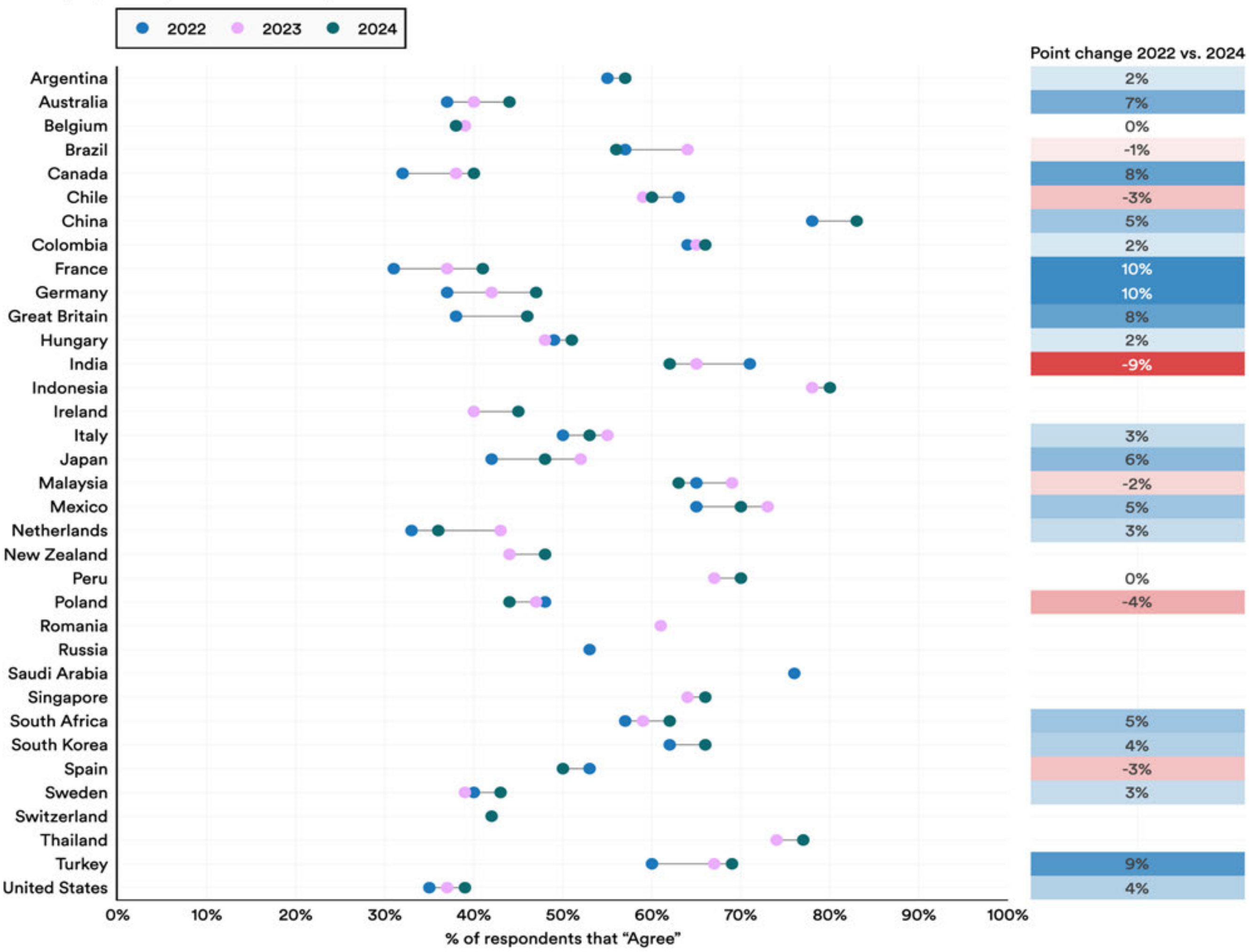

Figure 8.1.2

Figure 8.1.3 shows responses to Ipsos' survey on AI products and services by country. On average, survey respondents in China had the highest level of awareness, trust, and excitement about AI's use in products and services: 81% of respondents in China knew what products and services use AI, 80% reported that those products and services made them excited, 76% trusted AI to not discriminate or show bias, and overall 86% believed that products and services using AI would profoundly change their daily life in the next three to five years. Conversely, just 58% of American respondents thought that AI would profoundly change their life in the next three to five years, and 34% reported that products and services using AI made them excited.

Concerns about the privacy of personal data appear to be strongest in Japan and Canada, while concerns about AI discriminating against certain groups was highest in Sweden and Belgium.

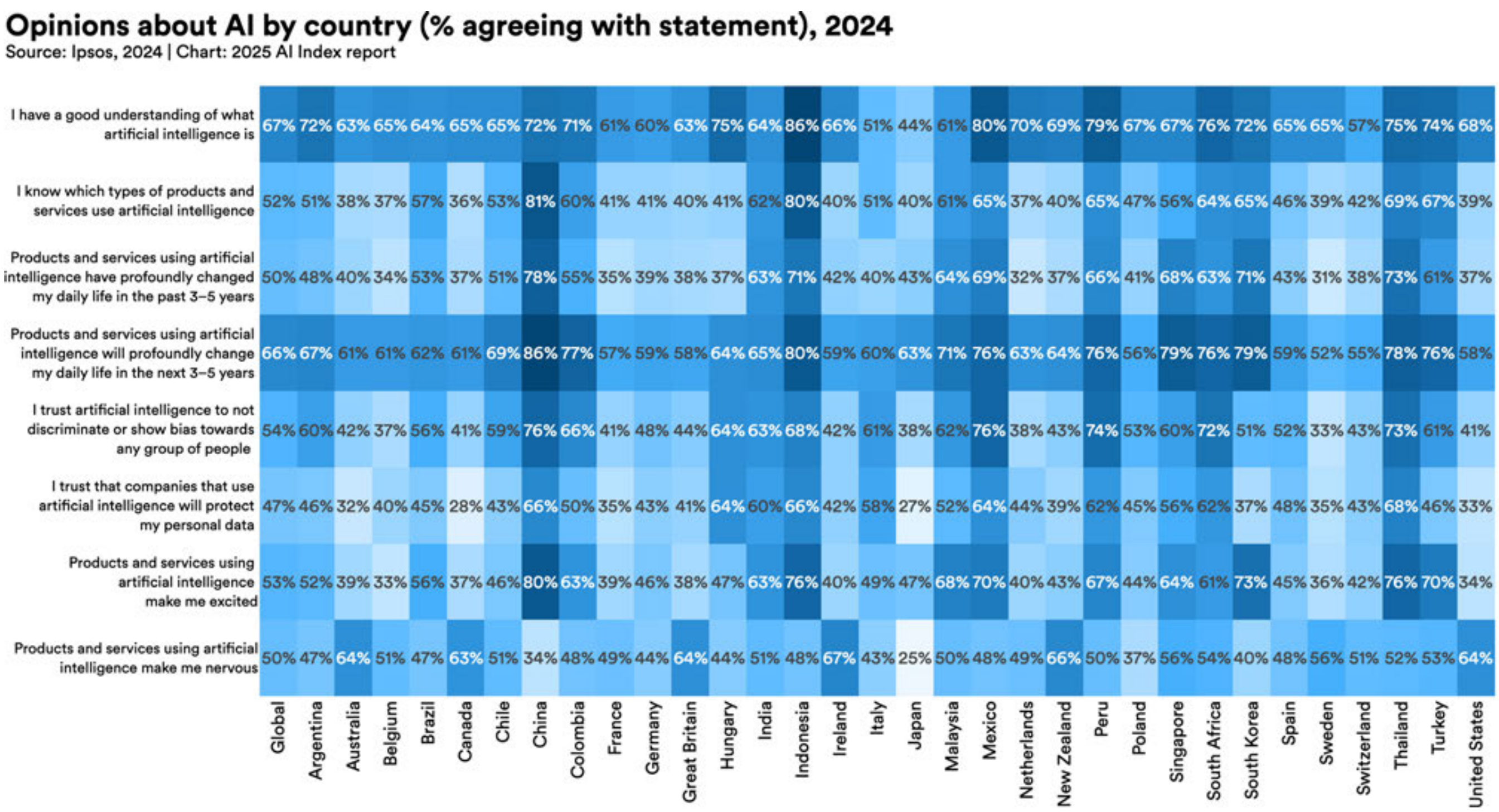

Figure 8.1.3

Figure 8.1.4 illustrates respondents' answers to whether they are excited about AI and whether they are nervous about it. Notable cross-country trends emerge. As previously noted, many Anglosphere nations—such as the United Kingdom, the United States, Canada, Australia, and New Zealand—report the highest levels of nervousness and the lowest excitement about AI. In contrast, several Asian countries, including China, South Korea, and Indonesia, exhibit higher excitement and lower nervousness levels, with Japan standing as an exception to this trend.

**Global opinions about products and services using AI by country, 2024**
Source: Ipsos, 2024 | Chart: 2025 AI Index report

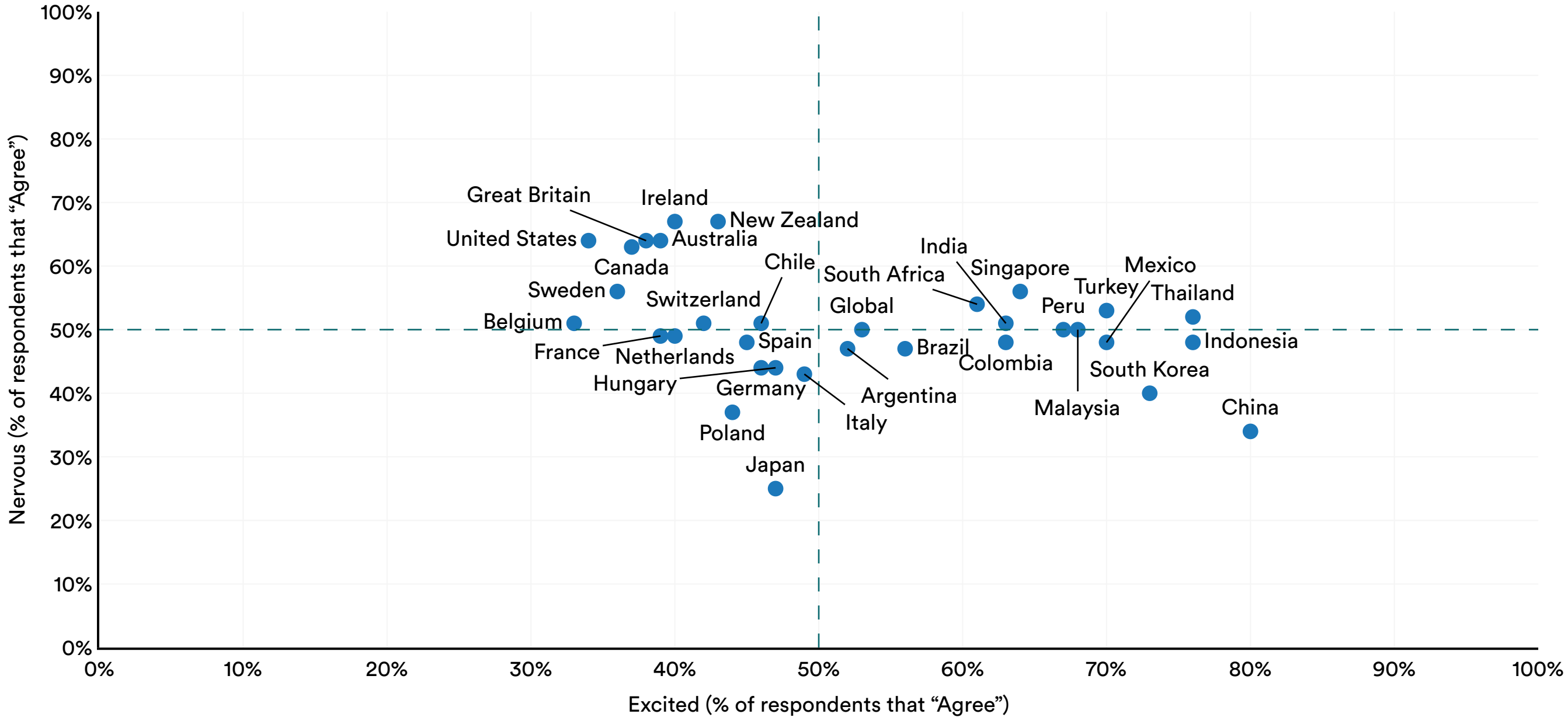

Figure 8.1.4

A majority of the countries surveyed by Ipsos in 2023 were surveyed again in 2024, enabling cross-year comparisons. Figure 8.1.5 highlights the year-over-year change in answers to particular AI-related questions. Overall, the AI Index observes slightly rising concerns about the use of AI, with an average 0.6% decrease in positive responses. This is largely driven by a 3% decrease in trust that companies that use AI will protect personal data, and a 2% decrease in trust that AI will not discriminate or show bias toward any group of people.[2]

Brazil and Malaysia saw the sharpest average decline in awareness, trust, and excitement about AI. In both countries, that negative trend was led by sharp declines in respondents who trust AI companies to protect their personal data.

South Africa and Ireland saw the sharpest average increases in awareness, trust, and excitement about AI. Ireland's positive trend appears to be led by positive user experiences, since it reports the highest increase across countries in respondents who say their daily lives have been profoundly impacted by products and services using AI.

**Percentage point change in opinions about AI by country (% agreeing with statement), 2023–24**

Source: Ipsos, 2023–24 | Chart: 2025 AI Index report

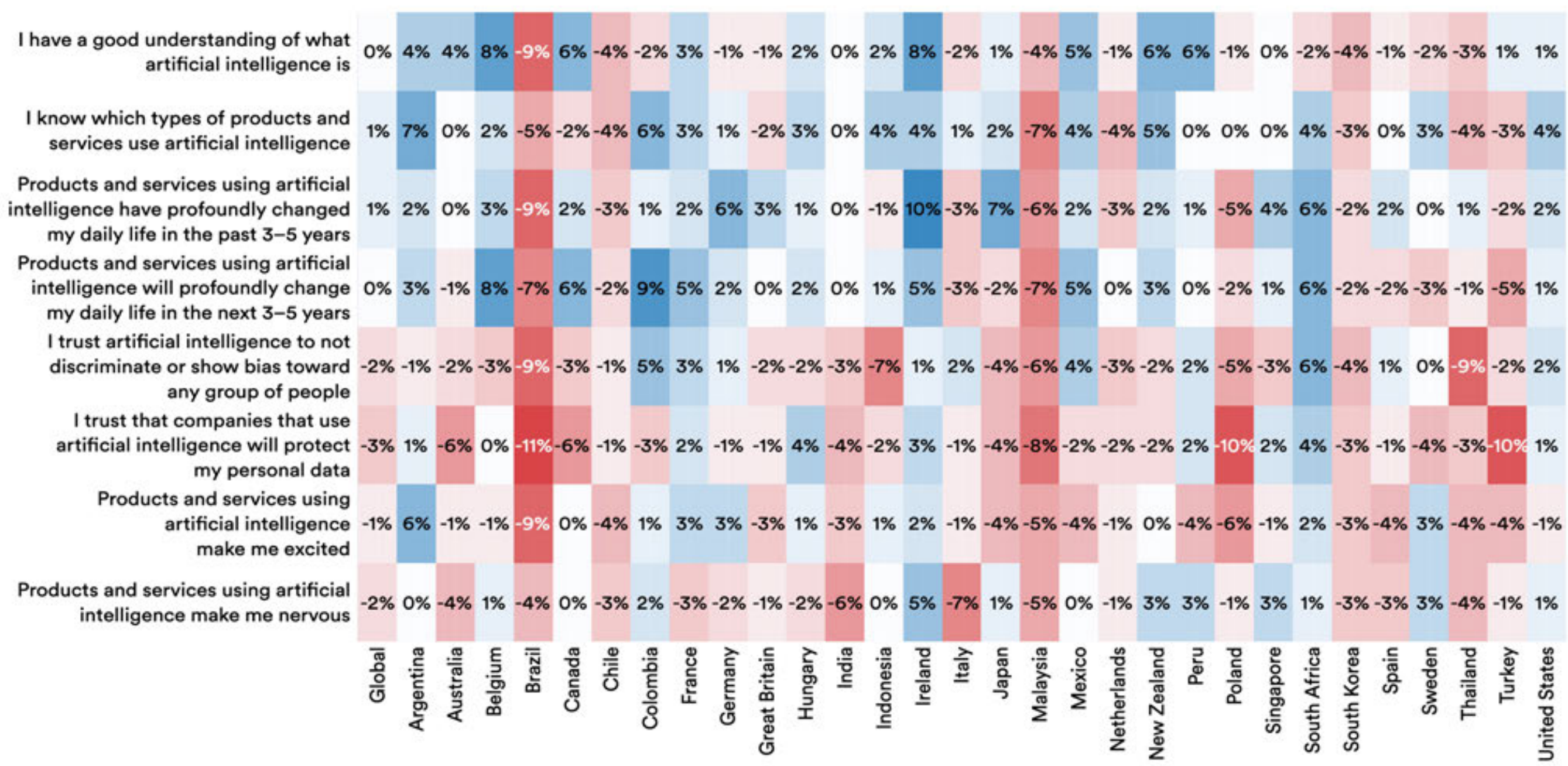

Figure 8.1.5

2 Average global responses to the question "Products and services using AI make me nervous" are excluded from this average because this is the only question where a positive score would yield a normatively negative result.

Figure 8.1.6 compares responses from the 2022 and 2024 Ipsos surveys, highlighting shifts in sentiment since the launch of ChatGPT. Globally, the belief that AI-powered products and services will profoundly change daily life within the next three to five years has risen by 6%. Every country except India, Malaysia, and Poland saw an increase in this perception since 2022, with the largest jumps in Canada (17%) and Germany (15%).

**Percentage point change in opinions about AI by country (% agreeing with statement), 2022 vs. 2024**
Source: Ipsos, 2022–24 | Chart: 2025 AI Index report

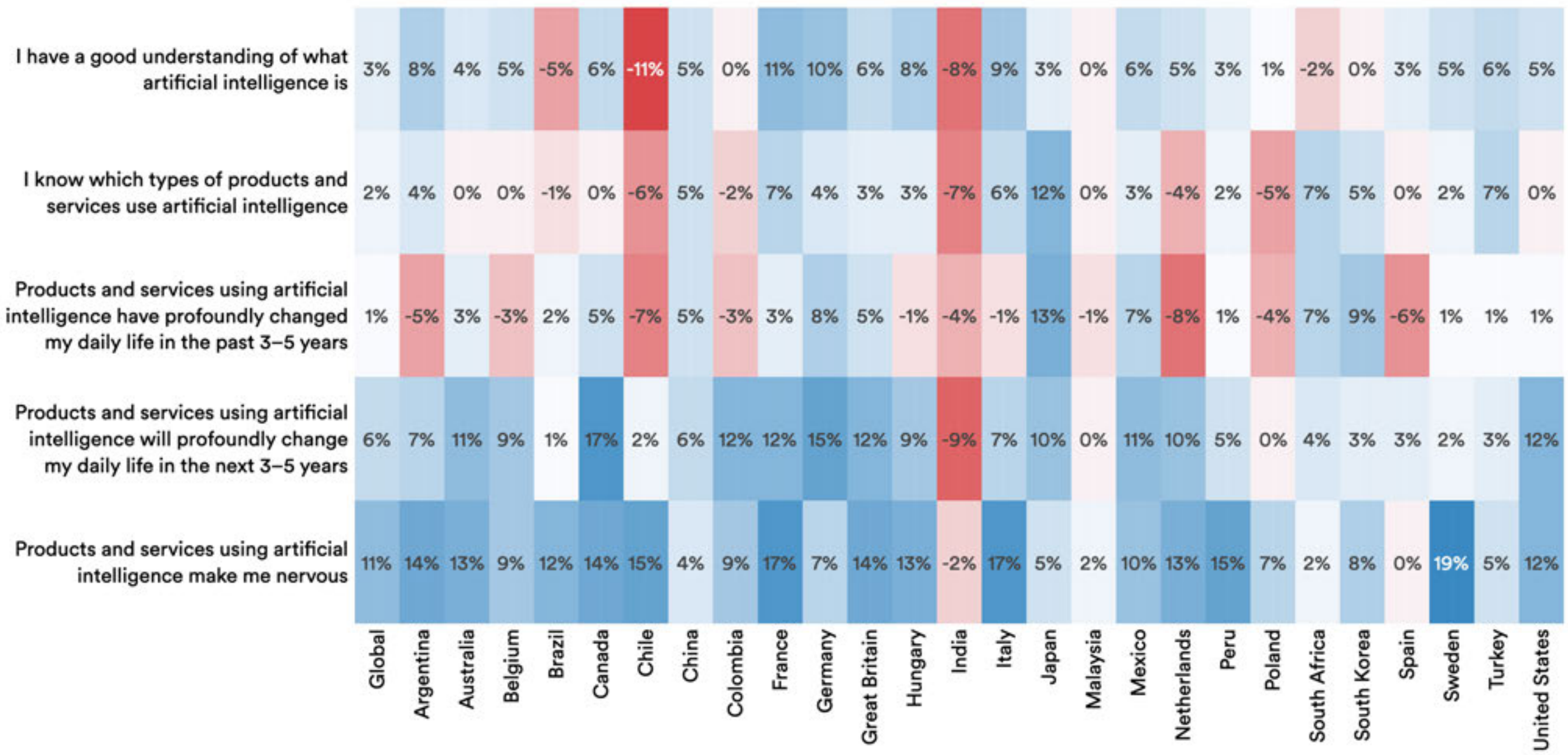

Figure 8.1.6

# AI and Jobs

This year's Ipsos survey included more questions about how people perceive AI's impact on their current jobs. Figure 8.1.7 illustrates various global perspectives on the expected impact of AI on employment. Overall, 60% of respondents believe AI is likely to change how they do their job in the next five years and 36%, or more than one in three, believe that AI is likely to replace their current job in the next five years.

Year-over-year comparisons for this question are challenging because in 2023 the survey did not differentiate between "very likely" and "somewhat likely." Nevertheless, when the 2024 categories are aggregated and compared to the 2023 results, the overall sentiment appears largely unchanged. In 2023, 57% of respondents agreed that AI would change how jobs are done, while 36% believed AI was likely to replace their job within five years.

**Global opinions on the perceived impact of AI on current jobs, 2024**
Source: Ipsos, 2024 | Chart: 2025 AI Index report

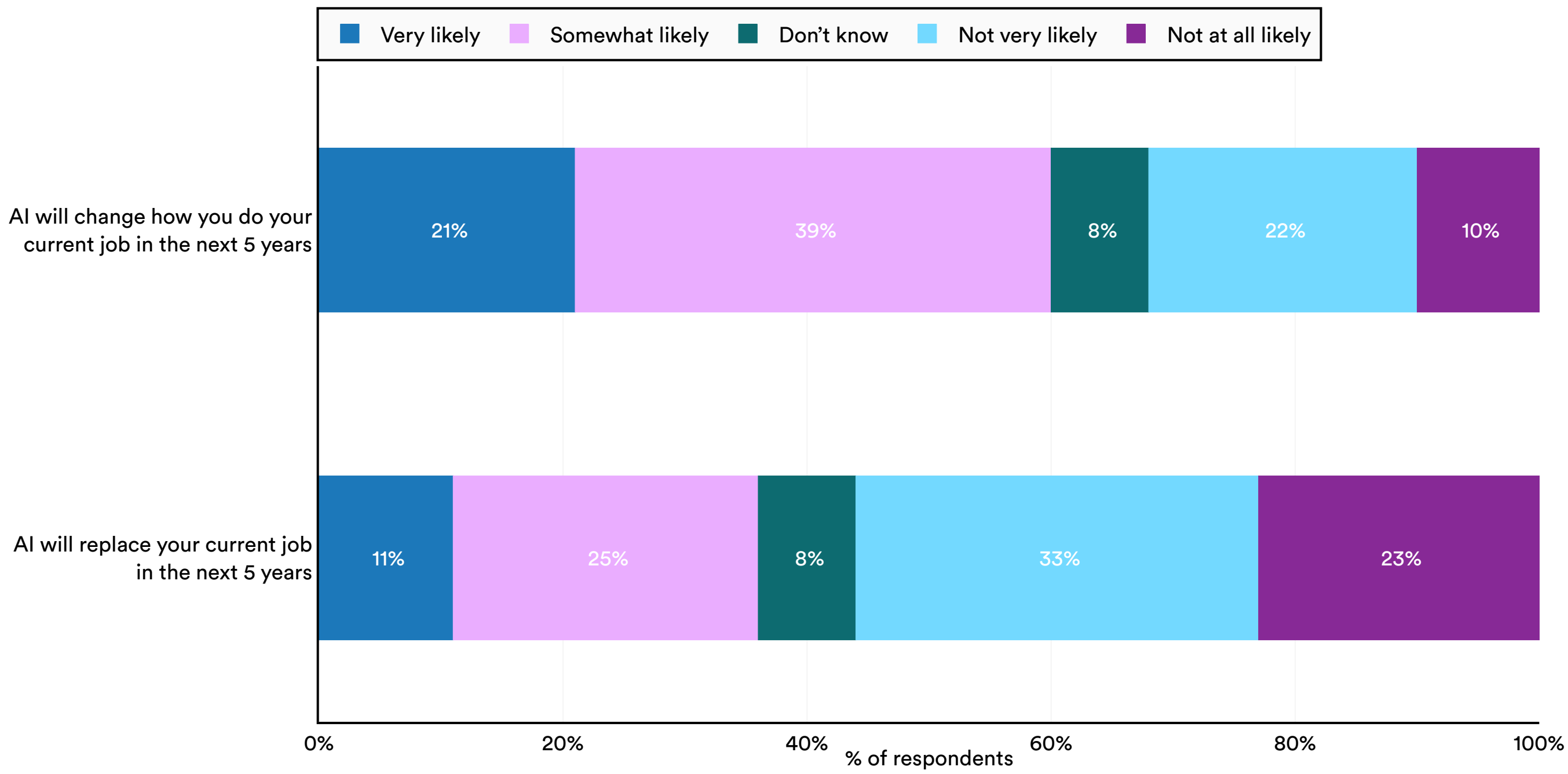

Figure 8.1.7

Opinions on whether AI will significantly impact an individual's job vary across demographic groups (Figure 8.1.8). Younger generations, such as Gen Z and millennials, are more inclined to agree that AI will change how they do their jobs compared to older generations like Gen X and baby boomers. Specifically, in 2024, 67% of Gen Z compared to 49% of boomers agree with the statement that AI will likely affect their current jobs.

Across 2023 and 2024, all generations increasingly agree that AI will change how they do their jobs over the next five years. Interestingly, of the 3% who believe AI will change how they do their jobs, the greatest increase was among both millennials and baby boomers, perhaps indicating increasing cross-generational awareness.

**Global opinions on whether AI will change how current jobs are done in the next five years (% agreeing with statement), 2023 vs. 2024**
Source: Ipsos, 2024 | Chart: 2025 AI Index report

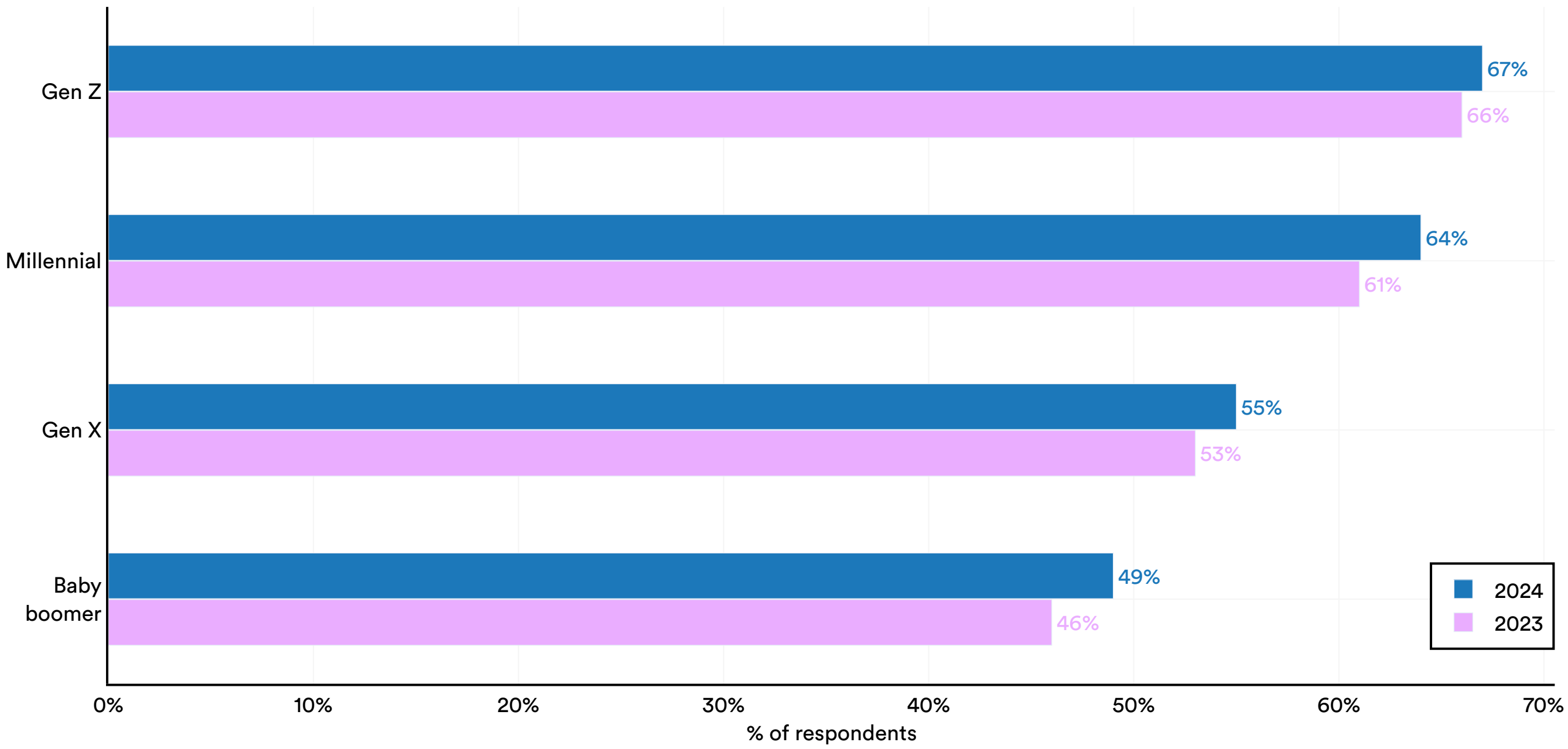

Figure 8.1.8

# AI and Livelihood

The Ipsos survey also explored the impact that respondents believe AI will have on various aspects of their lives, such as the economy, entertainment, and health.

Figure 8.1.9 shows that 55% of global respondents said they believe AI will reduce the amount of time it takes them to get things done, and 51% believe AI will improve their entertainment options. Opinions on the economy and the job market were more skeptical. In these sectors, just 36% and 31% of respondents believe AI will have a positive impact.

Figure 8.1.9 also shows significant range in respondents who believe AI will improve the economy in their country. Countries in Asia are the most optimistic about AI's economic impact, with 72% of respondents in China saying they expect AI to improve the economy, followed by 54% in Indonesia. Conversely, less than 25% of respondents in the Netherlands, the United States, Belgium, Sweden, and Canada believe that AI will improve the economy.

Within each country, respondents with an optimistic outlook on AI's impact on the economy tended to express optimism in other areas. For example, countries that expressed the highest expectation that AI will positively impact their economy also tended to believe that AI will reduce the amount of time it takes to get things done and that AI will improve health.

As a global average, 38% of respondents believe AI will improve health. Mexico reported the highest rates of optimism, with 56% believing that AI will have a positive impact on health. Conversely only 19% of respondents in Japan had positive expectations of AI's impact on health.

**Global opinions on the potential of AI to improve life by country, 2024**
Source: Ipsos, 2024 | Chart: 2025 AI Index report

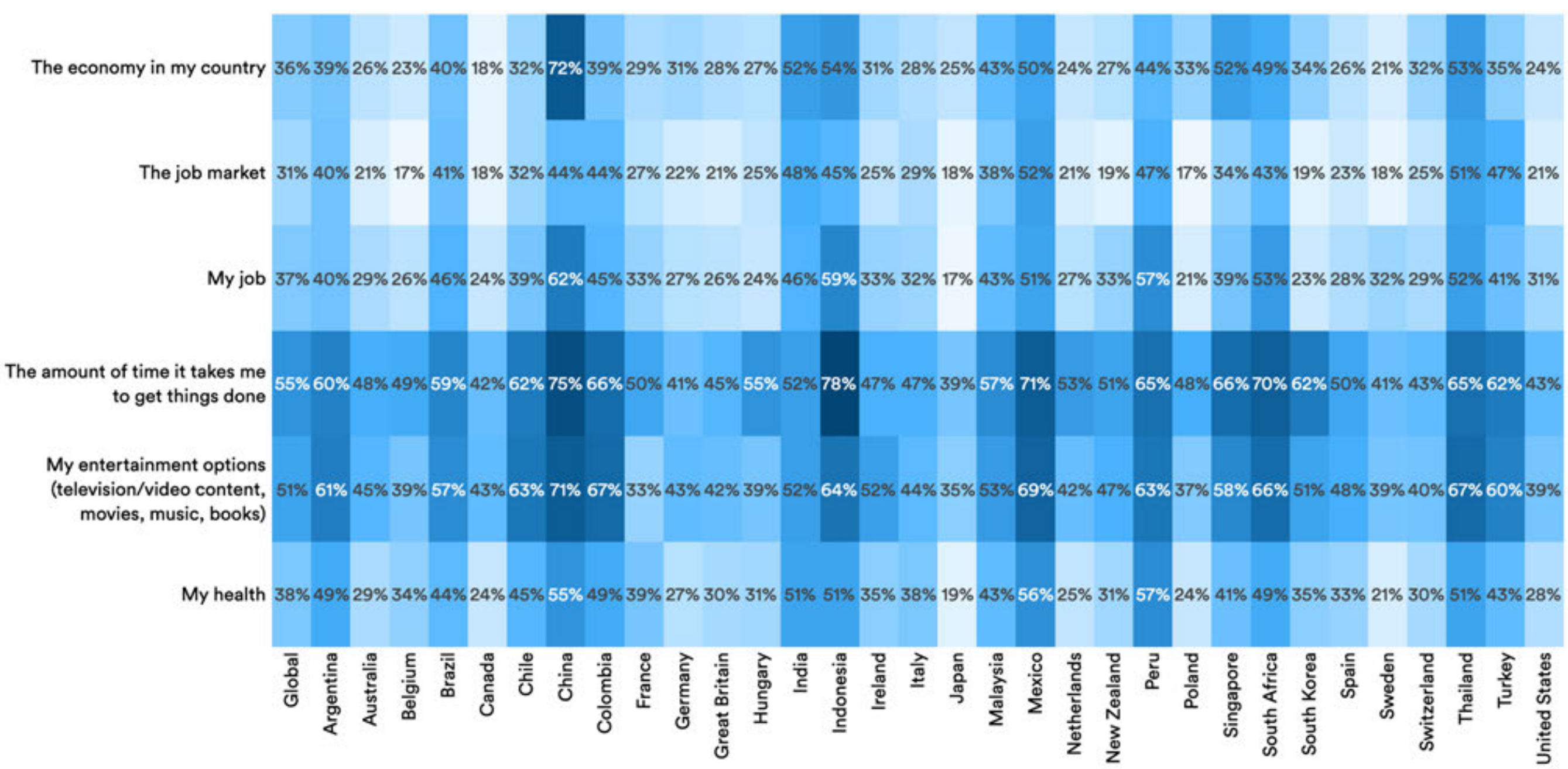

Figure 8.1.9

Figure 8.1.10 and Figure 8.1.11 provide a correlative analysis of the preceding data, examining the extent to which responses to certain questions are interrelated. Notably, there is a strong correlation between respondents' agreement that AI will improve the job market and their belief that it will benefit their own jobs. In some countries, such as Poland, optimism on both fronts is low, with only 17% and 21% of respondents expressing agreement, respectively. In contrast, sentiment is much more positive in China, where 44% believe AI will enhance the job market, and 62% think it will improve their jobs.

Similarly, countries where respondents believe AI will reduce the time required to complete tasks are also more likely to report that AI will improve their individual jobs.

**Global opinion on the potential of AI to improve the job market vs. individual jobs, 2024**
Source: Ipsos, 2024 | Chart: 2025 AI Index report

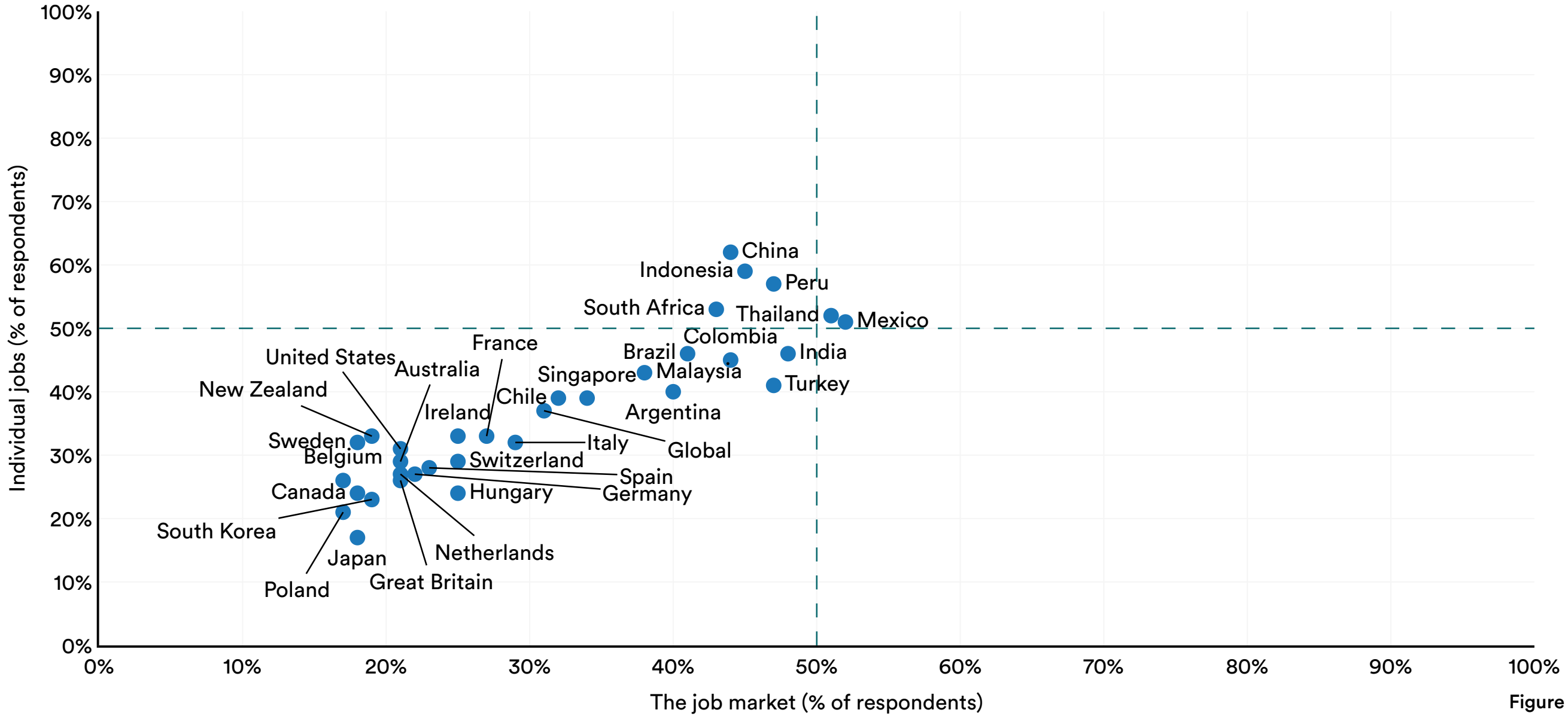

Figure 8.1.10

**Global opinion on the potential of AI to improve time to get things done vs. individual jobs, 2024**
Source: Ipsos, 2024 | Chart: 2025 AI Index report

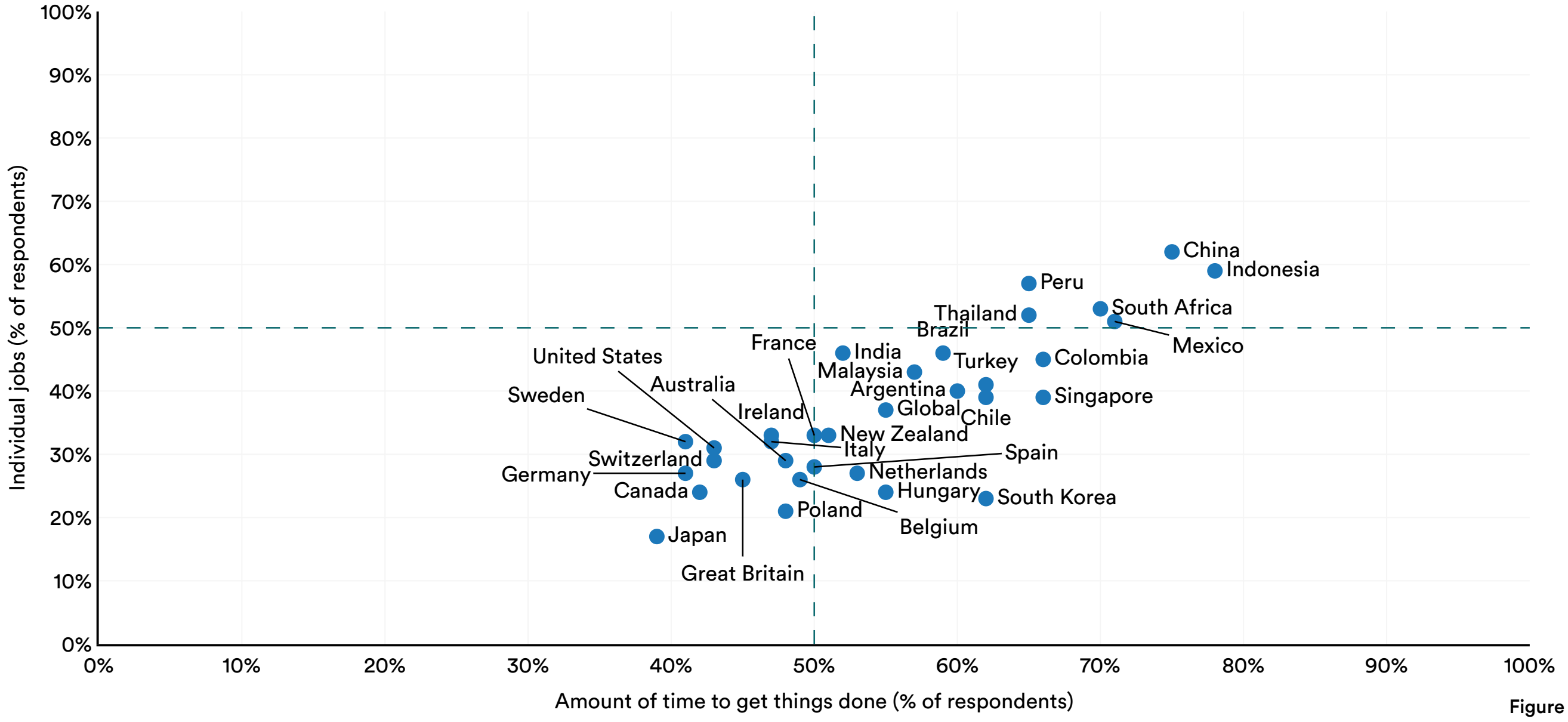

Figure 8.1.11

**Highlight:**

## Self-Driving Cars

As discussed in Chapter 2: Technical Performance, self-driving cars have made significant advancements in both capability and adoption. With companies like Waymo and Zoox becoming more prominent, understanding American attitudes toward self-driving technology is more important than ever.

The American Automobile Association (AAA) conducts an annual survey to assess public sentiment toward self-driving cars. The most recent survey, conducted in January 2025, was designed to be representative of approximately 97% of U.S. households. Figure 8.1.12 presents the results, revealing that despite the gradual rollout of self-driving cars on American roads, a majority of Americans (61%) remain fearful of the technology. Only 13% of respondents expressed trust in self-driving cars. While fear has declined slightly from its 2023 peak of 68%, it remains higher than in 2021, when 54% of Americans reported being afraid.

**US driver attitude toward self-driving vehicles, 2021–25**
Source: AAA, 2025 | Chart: 2025 AI Index report

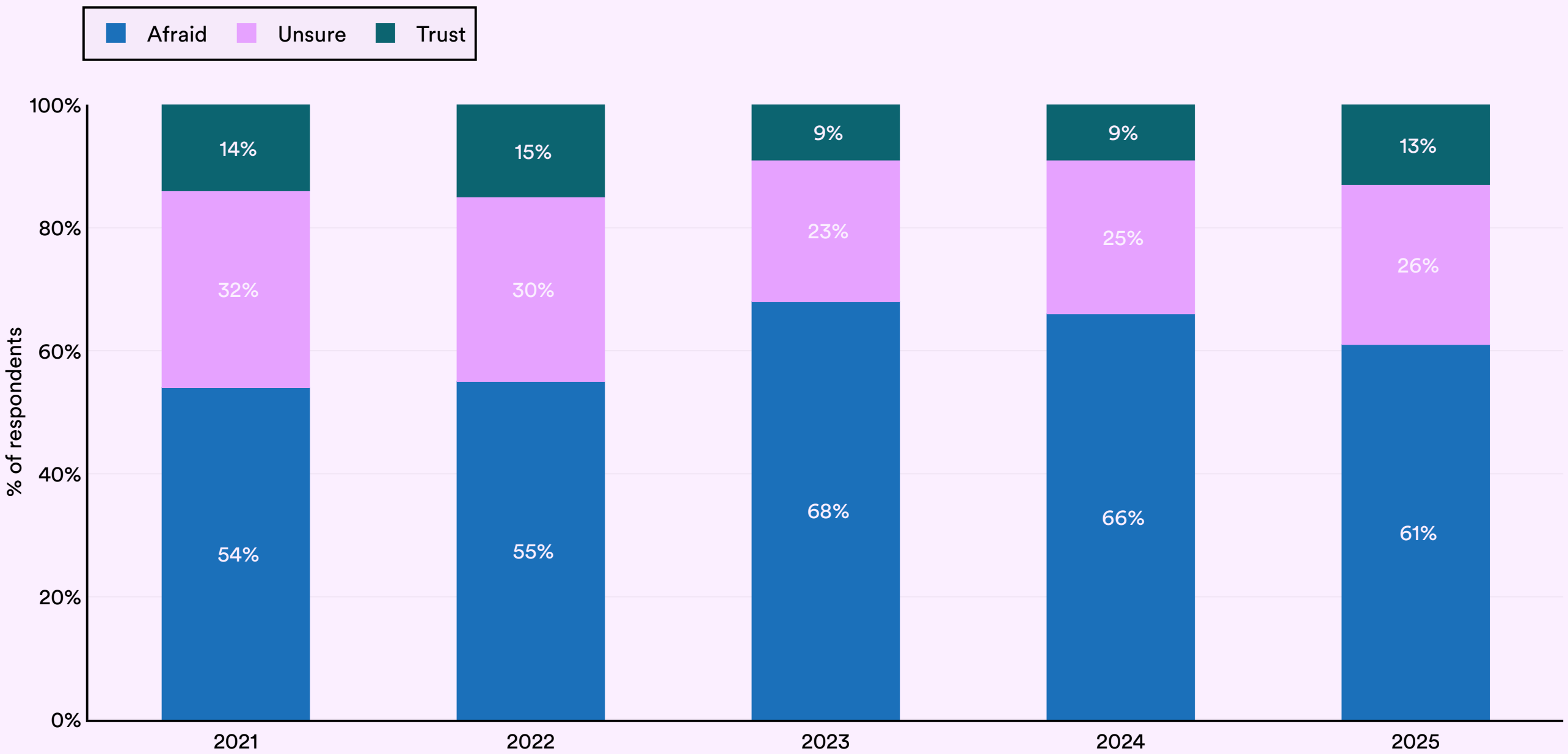

Figure 8.1.12

# 8.2 US Policymaker Opinion

Understanding public sentiment toward AI requires not only assessing the views of the general public but also those of key stakeholders, such as policymakers, who play a critical role in shaping AI regulation and policy. This year, an international team of researchers from Uppsala, Oxford, Harvard, and Syracuse universities released one of the first comprehensive studies on the perspectives of local U.S. policymakers—spanning township, municipal, and county levels—on AI's future impact and regulation. Conducted in two waves, in 2022 and 2023, the study gathered responses from approximately 1,000 policymakers. Its timing allowed researchers to compare how policymakers' views on AI shifted before and after the launch of ChatGPT.

Figure 8.2.1 illustrates the extent to which local policymakers agree with the statement: AI should be regulated by the government. In 2023, 73.7% of local U.S. policymakers supported this view, a significant increase from 55.7% in 2022. The launch of ChatGPT appears to have played a key role in shifting policymaker sentiment toward regulation. Support for AI regulation was higher among Democrats (79.2%) than Republicans (55.5%), though both groups registered a notable increase after 2022.

**Local US officials' support for government regulation of AI by party and year**
Source: Hatz et al., 2025 | Chart: 2025 AI Index report

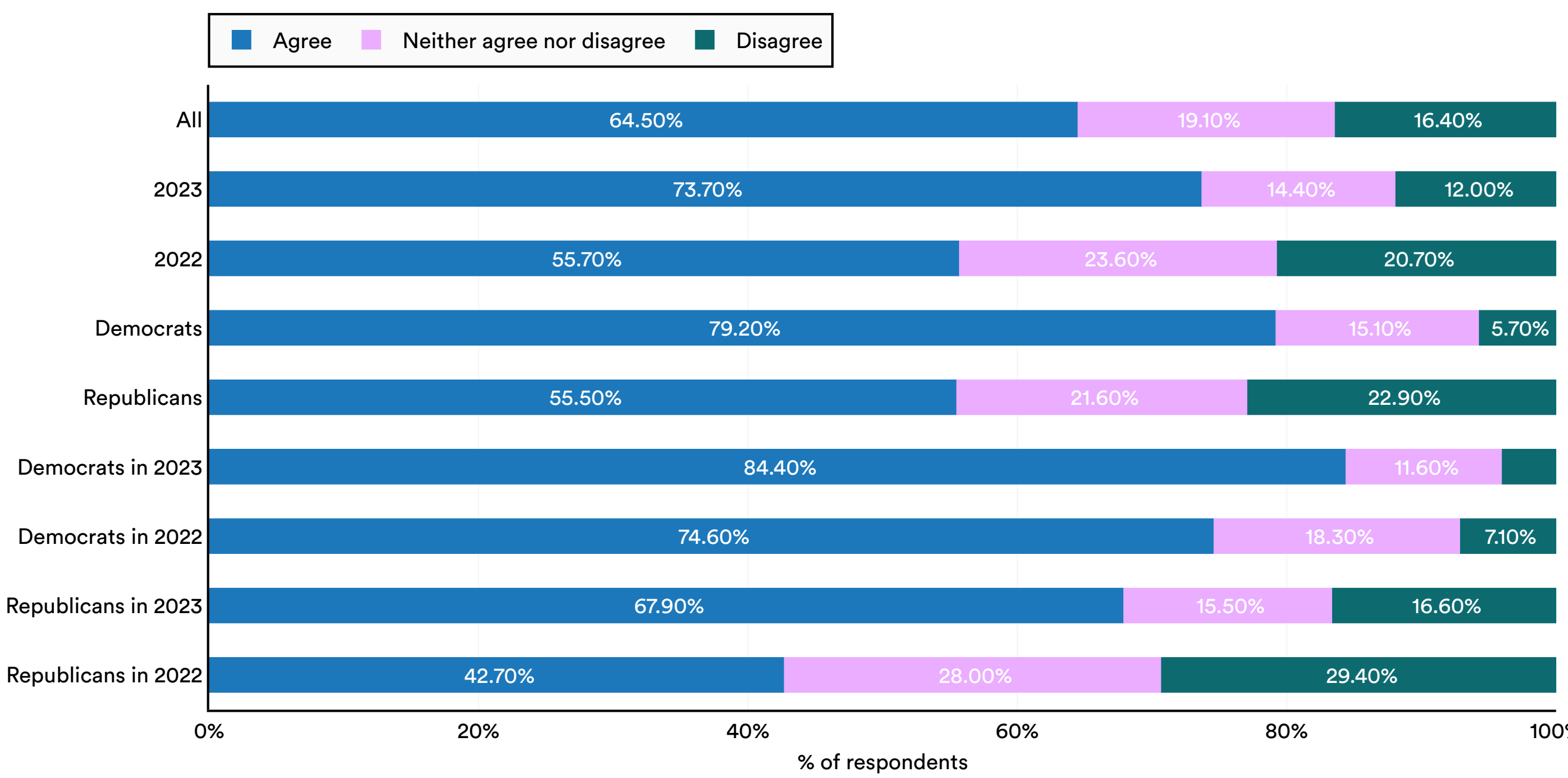

Figure 8.2.1

Given that most local policymakers support some form of AI regulation, which specific policies do they favor? At 80.4%, the strongest support is for stricter data privacy regulations. In addition, 76.2% support retraining programs for the unemployed, and 72.5% support AI deployment regulations (Figure 8.2.2). In contrast, there is significantly less backing for redistributive measures. Just 33.9% support wage subsidies to offset wage declines and just 24.6% support universal basic income.

**Local US officials' views on what AI policies would be beneficial for 2025–50**
Source: Hatz et al., 2025 | Chart: 2025 AI Index report

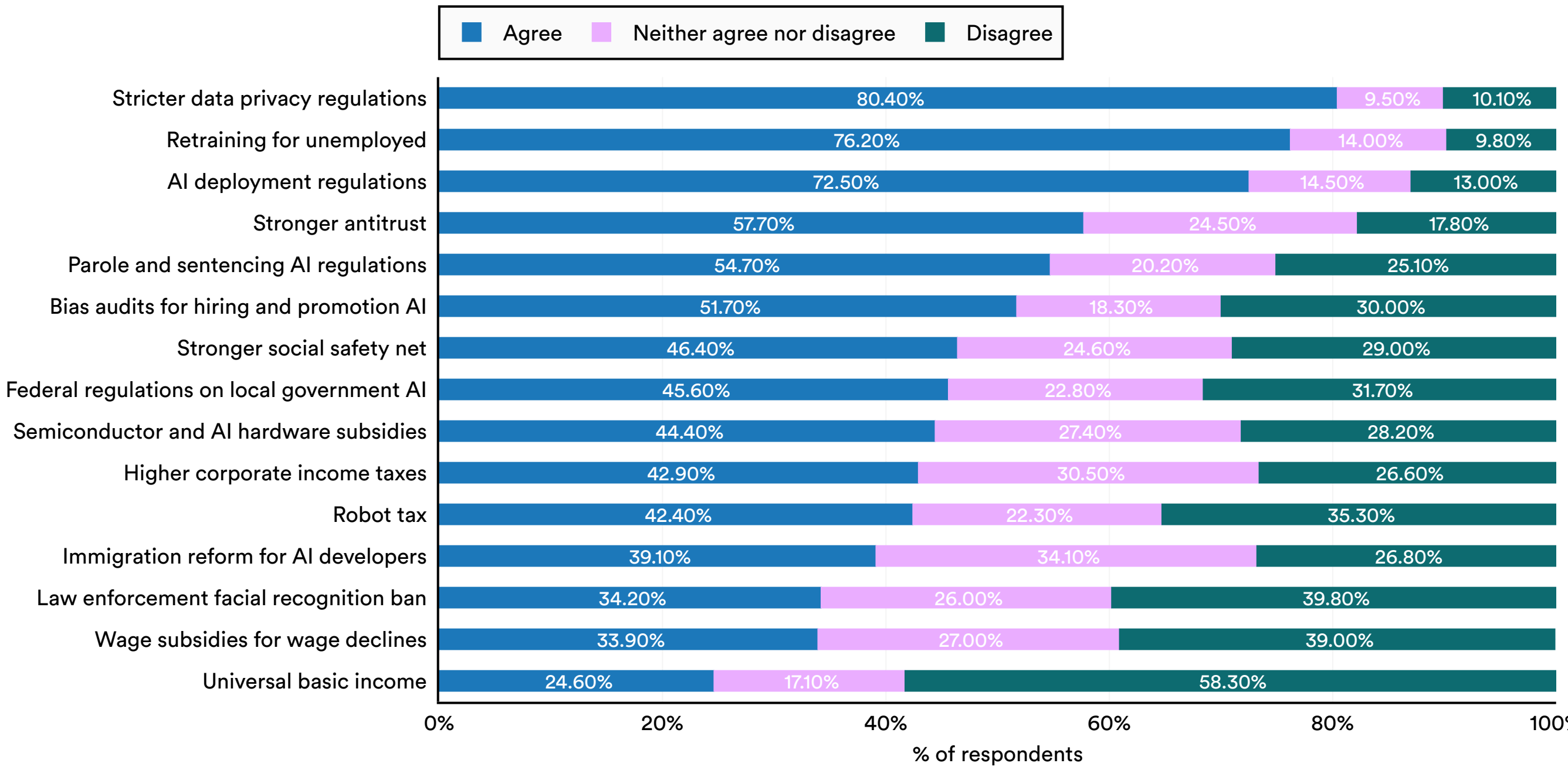

Figure 8.2.2

When it comes to AI policy, most local legislators do not believe they will have to take immediate action (Figure 8.2.3). Only 34.3% believe they will need to act within the next few years, compared to 56.5% who do not. However, agreement with this statement has increased from 32.2% in 2022 to 36.6% in 2023. This reflects the impact of major AI developments, such as the launch of ChatGPT, on policymakers' perspectives.

**Local US officials' likelihood of making AI policy decisions by party and year**
Source: Hatz et al., 2025 | Chart: 2025 AI Index report

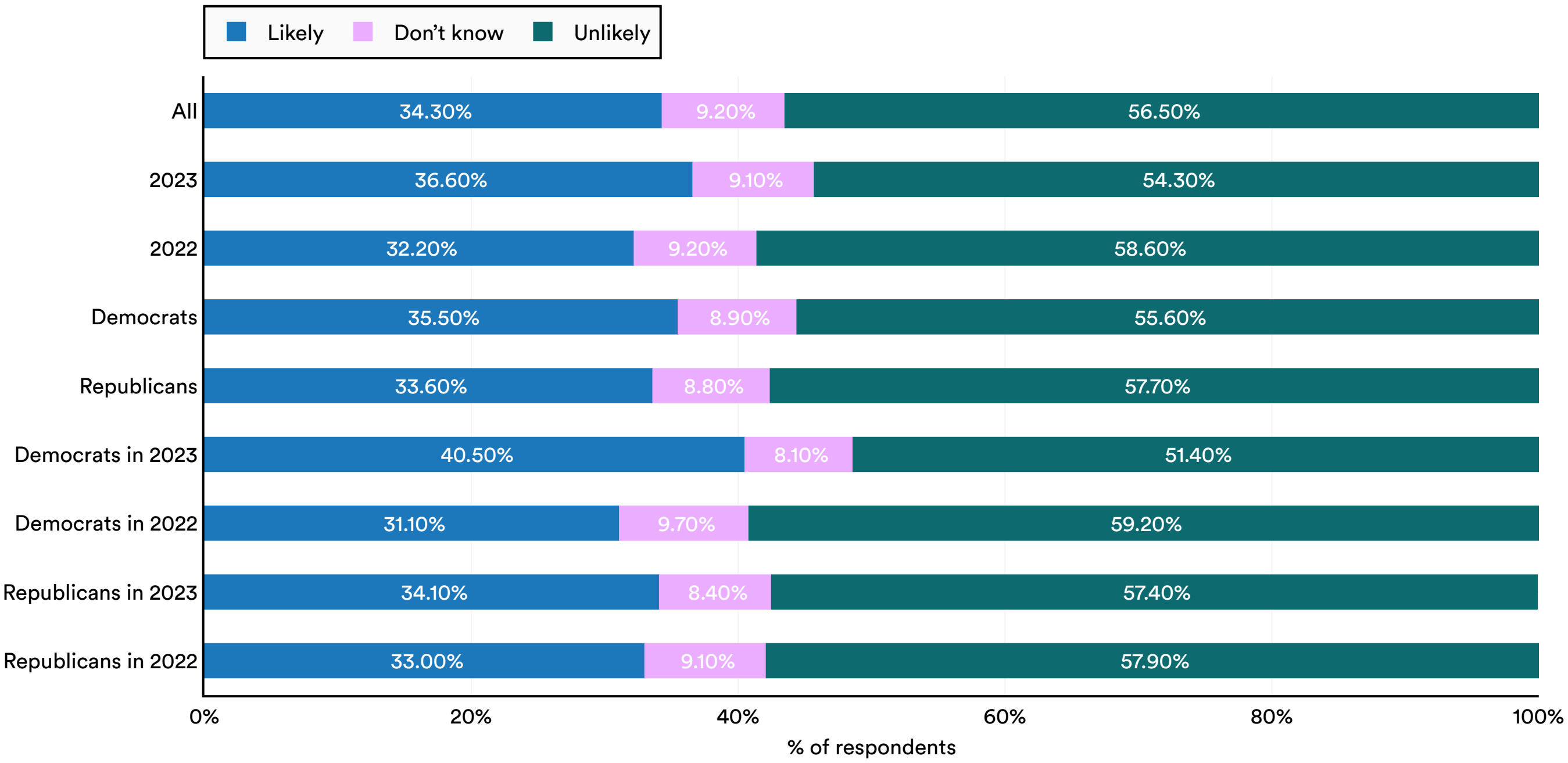

Figure 8.2.3

Only 29.8% of locally elected officials feel adequately informed to make AI policy decisions (Figure 8.2.4). While confidence in AI-related policymaking has increased slightly across both parties from 2022 to 2023, it remains relatively low overall.

**Local US officials' feeling adequately informed to make decisions about AI by party and year**
Source: Hatz et al., 2025 | Chart: 2025 AI Index report

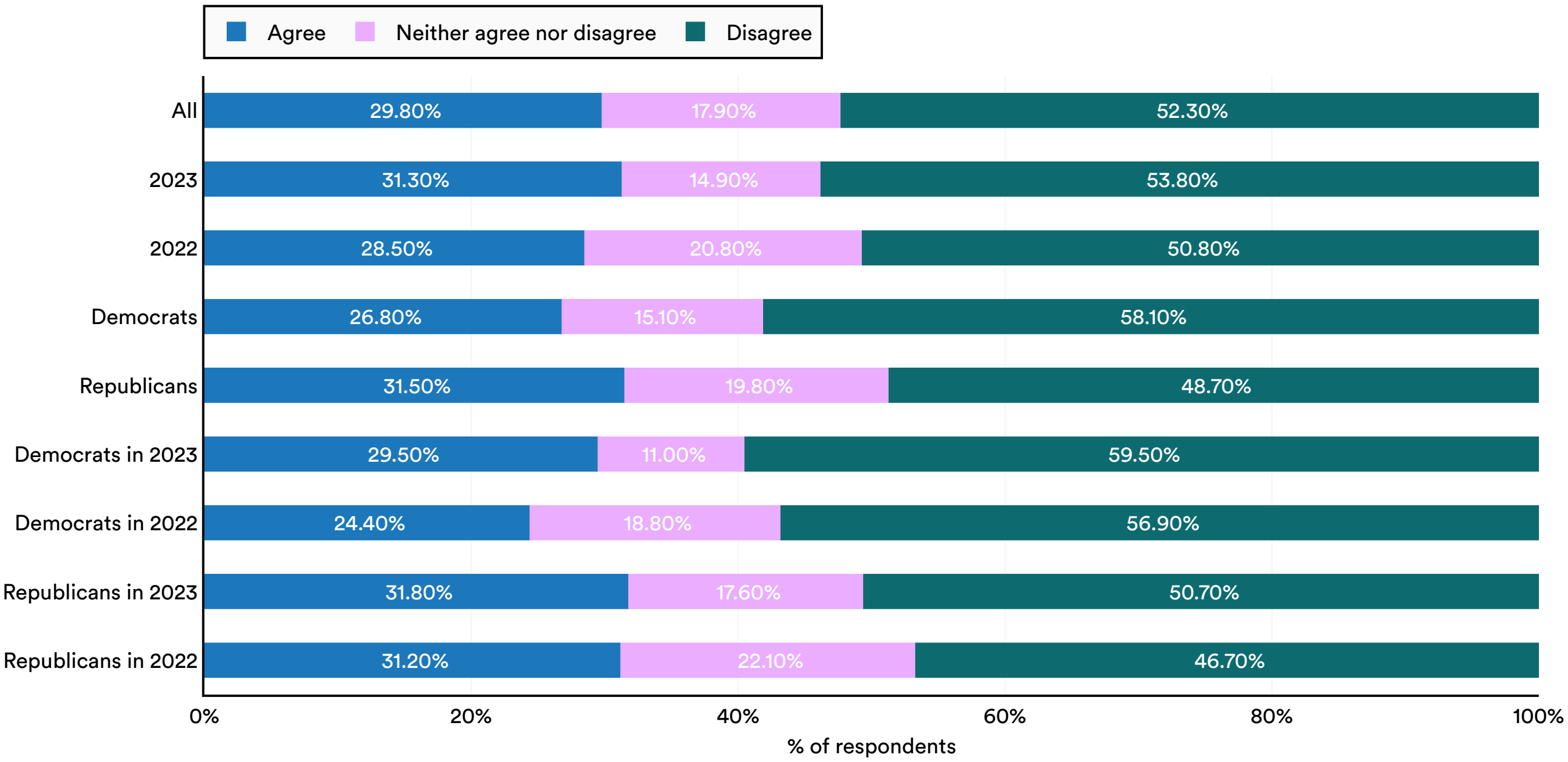

Figure 8.2.4

# APPENDIX

# Appendix

# Chapter 1: Research and Development

## Acknowledgments

The AI Index would like to acknowledge Angelo Salatino for his contributions to AI publication classification, Ben Cottier for leading the analysis of machine learning inference costs, Lapo Santarlasci for leading the analysis of AI patents, and Andrew Shi for leading the analysis of the environmental impact of AI models.

## AI Publication Analysis

For this analysis, the AI Index used OpenAlex, an open scholarly database with over 260 million research publications, as its primary data source. OpenAlex classifies papers using its own knowledge organization system, known as OpenAlex Topics—a taxonomy of around 4,500 topics combining Scopus codes and CWTS classification. The system uses a deep learning model that considers titles, abstracts, journal names, and citation networks for classification. To identify AI-related topics more precisely, the AI Index analyzed computer science publications identified by OpenAlex and refined the classifications using the Computer Science Ontology and the CSO Classifier.

The Computer Science Ontology (CSO) is a large-scale, automatically generated ontology of research areas derived from 16 million publications using the Klink-2 algorithm. It features a hierarchical structure with thousands of subtopics, allowing for precise mapping of specific terms to broader research fields. Compared to general-purpose scholarly databases like OpenAlex, Scopus, and Web of Science, CSO offers a more detailed and fine-grained representation of the research landscape. As a result, it has been widely used for scholarly data exploration, analysis, modeling, and expert identification and recommendation. Version 3.4.1—used in this analysis—includes approximately 15,000 topics and 166,000 relationships within computer science. Released on Jan. 17, 2025, this version introduces over 150 new research topics in artificial intelligence, bringing the total to 2,369 AI-related topics and 12,620 hierarchical relationships within the AI domain alone.

To analyze research trends, the AI Index used the CSO Classifier—an unsupervised method that automatically categorizes research papers based on CSO topics. The classifier follows a three-stage pipeline that processes paper titles and abstracts: A syntactic module detects direct mentions of CSO topics; a semantic module uses word embeddings to identify related concepts; and a postprocessing module merges results, filters out irrelevant topics, and adds broader categories for a more refined classification. For this analysis, the AI Index extended the CSO Classifier to focus specifically on artificial intelligence and its subtopics. Since its initial release, the classifier has gained significant and growing interest due to its versatility. For example, Springer Nature uses it to routinely classify proceedings books, improving metadata quality. Beyond academic publishing, it has been successfully applied to categorize research software, YouTube videos, press releases, job ads, and IT museum collections.

Accurately categorizing research papers as either conference proceedings or journal articles is essential for this analysis. OpenAlex's metadata fields—type, crossref_type, and source_type—can sometimes conflict. To resolve these inconsistencies, the AI Index mapped OpenAlex records to DBLP, a leading bibliographic database for computer science publications. Known for its high metadata quality, DBLP continuously adds new publications through a rigorous, semiautomated curation process and currently indexes 3.6 million conference papers and 3 million journal articles. The initial matching between OpenAlex and DBLP was performed using DOIs. For remaining unmatched papers, the AI Index used a combination of title and publication year. To streamline this process, the AI Index built a title index to optimize search and ensure efficient mapping across the datasets.

AI publications are aggregated based on several parameters to provide a comprehensive analysis. Publications are grouped by year, considering the publication date of the most recent versions. Additionally, the AI Index groups publications by geographic areas or World Bank regions using the affiliations of authors. This means a single paper can contribute to multiple counts if coauthored by researchers from different countries, with each country receiving a count. When authors' affiliations are missing, these publications are mapped as "Unknown." Furthermore, sectors are associated with publications through authors' affiliations when available, which may lead to a publication being counted for multiple sectors. Citation counts are included when available; those without citation data are classified as "Unknown."

### Top 100 Publications Analysis

The AI Index conducted a comprehensive analysis of influential AI publications by collecting and analyzing citation data from multiple sources including OpenAlex, Google Scholar, and Semantic Scholar. Initially gathering the top 150 most-cited papers per publication year from OpenAlex, the list was refined to 100 publications through careful review.

The methodology attributes publications to all countries and regions represented by authors' affiliations, meaning a single paper can contribute to multiple counts. For instance, a paper coauthored by researchers from the United States and China counts once for each country. This approach may result in overlapping totals in aggregate statistics. Publication years are based on the most recent versions, whether in journals, conferences, or repositories like arXiv. To maintain accuracy, organizational affiliations were verified and standardized, with countries assigned according to headquarters' locations.

The full list of the top 100 AI publications is available here.

## AI Patent Analysis

The AI Index identifies AI-related patents using a hybrid classification approach, combining keyword-based text analysis with classification-code-based identification. Patent-level bibliographic data is sourced from PATSTAT Global, a comprehensive database issued by the European Patent Office (EPO). The analysis focuses on granted patents from 2010 onward, aggregated at the DOCDB family level to avoid duplicate counting of the same invention.[1] Patents are attributed to countries based on the publication authority of the earliest recorded grant publication.

Patent abstracts and titles originally published in languages other than English were translated using the deep-translator tool, Google Translate engine, and the Meta NLLB-200 machine translation model. Post-translation, patent texts were processed using natural language processing (NLP) techniques. These included the removal of stop words and special characters, part-of-speech (POS) tagging to retain key grammatical categories, lowercase conversion, lemmatization, and replacement of numerical measures with a <NUM> tag.

AI-related patents are identified by searching for relevant terms in patent titles and abstracts using regular expressions (regex). An AI-specific keyword dictionary was developed through a structured multistep process, incorporating keywords generated by AI models, expanded using established AI lexicons such as those from Yamashita et al. (2021), and refined through Word2Vec-based synonym identification. Further validation was conducted using BERTopic topic modeling and DeBERTA-based zero-shot classification, with manual checks applied to reduce false positives.

In addition to keyword-based classification, AI-related patents were identified using International Patent Classification (IPC) and Cooperative Patent Classification (CPC) codes. A curated list of AI-relevant codes was compiled through a combination of AI model analysis, regex-based searches, and prior research, including classifications from Pairolero et al. (2023) and WIPO (2024). The final dataset was constructed by merging results from both approaches, balancing coverage and accuracy.

1 Despite this aggregation procedure, duplicates occasionally appear in marginal cases where applications within the same DOCDB family share the same earliest filing date. The AI Index removes duplicate values with respect to the aggregation variables (e.g., counting by year) when presenting analytics.

## Epoch Notable Models Analysis

The AI forecasting research group Epoch AI maintains a dataset of landmark AI and ML models, along with accompanying information about their creators and publications, such as the list of their authors, number of citations, type of AI task accomplished, and amount of compute used in training.

The nationalities of the authors of these papers have important implications for geopolitical AI forecasting. As various research institutions and technology companies start producing advanced ML models, the global distribution of AI development may shift or concentrate in certain places, which in turn affects the geopolitical landscape because AI is expected to become a crucial component of economic and military power in the near future.

To track the distribution of AI research contributions on landmark publications by country, the Epoch dataset is coded according to the following methodology:

1. A snapshot of the dataset was taken in March 2025. This includes papers about landmark models, selected using the inclusion criteria of importance, relevance, and uniqueness, as described in the Compute Trends dataset documentation.
2. The authors are attributed to countries based on their affiliation credited on the paper. For international organizations, authors are attributed to the country where the organization is headquartered, unless a more specific location is indicated.
3. All of the landmark publications are aggregated within time periods (e.g., monthly or yearly) and the national contributions compiled to determine the extent of each country's contribution to landmark AI research during each time period.
4. The contributions of different countries are compared over time to identify any trends.

## Training Cost Analysis

To create the dataset of cost estimates, the Epoch database was filtered for models released during the large-scale ML era[2] that were in the top 10 of training compute at the time of release. This filtered for the largest-scale ML models. The Transformer model was added to this set of models for further context.

For the selected ML models, the training time and the type, quantity, and hardware utilization rate were determined from the publication, press release, or technical reports, as applicable. Cloud rental prices for the computing hardware used by these models were collected from online historical archives of cloud vendors' websites.[3]

Training costs were estimated from the hardware type, quantity, and time by multiplying the hourly cloud rental rates (at the time of training)[4] by the quantity of hardware hours. However, some developers purchased hardware rather than renting cloud compute, and cloud prices vary by vendor and by rental commitment, so the true costs incurred by the developers may vary.

Various challenges were encountered while estimating the training cost of these models. Often, the developers did not disclose the duration of training or the hardware that was used. In other cases, cloud compute pricing was not available for the hardware. The investigation of training cost trends is more thoroughly detailed in a separate report by Epoch AI.

## AI Conference Attendance

The AI Index reached out to the organizers of various AI conferences in 2024 and asked them to provide information on total attendance. For conferences that posted their attendance totals online, the AI Index used those reported totals and did not reach out to the conference organizers.

2 The selected cutoff date was Sept. 1, 2015, in accordance with Compute Trends Across Three Eras of Machine Learning (Epoch, 2022).

3 Historic prices were collected from archived snapshots of Amazon Web Services, Microsoft Azure, and Google Cloud Platform price catalogs viewed through the Internet Archive Wayback Machine.

4 The chosen rental rate was the most recent published price for the hardware and cloud vendor used by the developer of the model, at a three-year commitment rental rate, after subtracting the training duration and two months from the publication date. If this price was not available, the most analogous price was used—either the same hardware and vendor at a different date, or the same hardware from a different cloud vendor. If a three-year commitment rental rate was unavailable, this was imputed from other rental rates based on the empirical average discount for the given cloud vendor. If the exact hardware type was not available (e.g., Nvidia A100 SXM4 40GB), a generalization was used (e.g., Nvidia A100).

# GitHub

## Identifying AI Projects

In partnership with researchers from Harvard Business School, Microsoft Research, and Microsoft's AI for Good Lab, GitHub identifies public AI repositories following the methodologies of Gonzalez, Zimmerman, and Nagappan (2020) and Dohmke, Iansiti, and Richards (2023), using topic labels related to AI/ML and generative AI, respectively, along with other relevant keywords identified through snowball sampling, such as "machine learning," "deep learning," and "artificial intelligence." GitHub further augments the dataset with repositories that have a dependency on the PyTorch, TensorFlow, OpenAI, Transformers, XGBoost, scikit-learn, and SciPy libraries for Python.

## Mapping AI Projects to Geographic Areas

Public AI projects are mapped to geographic areas using IP address geolocation to determine the mode location of a project's owners each year. Each project owner is assigned a location based on their IP address when interacting with GitHub. If a project owner changes locations within a year, the location for the project would be determined by the mode location of its owners sampled daily in the year. Additionally, the last known location of the project owner is carried forward on a daily basis even if no activities were performed by the project owner that day. For example, if a project owner performed activities within the United States and then became inactive for six days, that project owner would be considered to be in the United States for that seven-day span.

# Environmental Impact Analysis

The AI Index estimated the carbon emissions of training language and vision models using a calculator proposed by Lacoste et al. (2019). The analysis focused on the training stage emissions, excluding embodied hardware production, idle infrastructure, and deployment emissions. The study examined four model categories: industry language models, academic language models, industry vision models, and academic vision models.

The calculator's accuracy was verified against published emission values. Calculator inputs included hardware type, GPU hours, provider, and compute region. For newer hardware like the H100 GPU (released in 2022), the A100 SXM4 80GB was used as a substitute in calculations. Provider selection was based on known partnerships (e.g., Google models using GCP, OpenAI using Azure), while compute regions were determined by team locations.

Special consideration was given to models trained on custom hardware, such as BLOOM's use of the Jean Zay supercomputer in France. In these cases, private infrastructure calculations incorporated carbon efficiency (kg/kWh) and offset percentages.

The study evaluated 50 models in total: 34 industry language models (2018–24), eight industry vision models (2019–23), four academic language models (2020–23), and four academic vision models (2011–22), selecting particularly influential models in their respective domains.

# Chapter 2: Technical Performance

## Acknowledgments

The AI Index would like to acknowledge Andrew Shi for his work generating sample Midjourney and Pika video generations and Armin Hamrah for his work identifying significant AI technical advancements for the timeline.

## Benchmarks

In this chapter, the AI Index reports on benchmarks, recognizing their importance in tracking AI's technical progress. As a standard practice, the Index sources benchmark scores from leaderboards, public repositories such as Papers With Code and RankedAGI, as well as company papers, blog posts, and product releases. The Index operates under the assumption that the scores reported by companies are accurate and factual. The benchmark scores in this section are current as of mid-February 2025. However, since the publication of the AI Index, newer models may have been released that surpass current state-of-the-art scores.

1. **ARC-AGI**: Data on ARC-AGI was taken from the ARC-AGI paper and OpenAI video in February 2025. To learn more about ARC-AGI, please read the original paper.
2. **Arena-Hard-Auto**: Data on Arena-Hard-Auto was taken from the LMSYS leaderboard in February 2025. To learn more about Arena-Hard-Auto, please read the original paper.
3. **Bench2Drive**: Data on Bench2Drive was taken from the Bench2Drive paper in February 2025. To learn more about Bench2Drive, please read the original paper.
4. **Berkeley Function Calling**: Data on Berkeley Function Calling was taken from the Berkeley Function Calling leaderboard in February 2025. To learn more about Berkeley Function Calling, please read the original work.
5. **BigCodeBench**: Data on BigCodeBench was taken from the BigCodeBench leaderboard in February 2025. To learn more about BigCodeBench, please read the original work.
6. **Chatbot Arena**: Data on Chatbot Arena was taken from the Chatbot Arena leaderboard in February 2025. To learn more about Chatbot Arena, please read the original paper.
7. **FrontierMath**: Data on FrontierMath was taken from the FrontierMath paper and OpenAI video in February 2025. To learn more about FrontierMath, please read the original paper. The visual was supplemented with benchmark data from OpenAI's o3 model, sourced from a YouTube video announcing its launch in December 2025.
8. **GAIA**: Data on GAIA was taken from the GAIA leaderboard in February 2025. To learn more about GAIA, please read the original paper.
9. **GPQA**: Data on GPQA was taken from the GPQA paper and OpenAI video in February 2025. To learn more about GPQA, please read the original paper.
10. **GSM8K**: Data on GSM8K was taken from the GSM8K Papers With Code leaderboard in February 2025. To learn more about GSM8K, please read the original paper.
11. **HELMET**: Data on HELMET (How to Evaluate Long-Context Models Effectively and Thoroughly) was taken from the HELMET paper in February 2025. To learn more about HELMET, please read the original paper.
12. **HLE**: Data on Humanity's Last Exam (HLE) was taken from the HLE paper in February 2025. To learn more about HLE, please read the original paper.
13. **HumanEval**: Data on HumanEval was taken from the HumanEval Papers With Code leaderboard in February 2025. To learn more about HumanEval, please read the original paper.
14. **LRS2**: Data on Oxford-BBC Lip Reading Sentences 2 (LRS2) was taken from the LRS2 Papers With Code leaderboard in February 2025. To learn more about LRS2, please read the original paper.

15. **MATH**: Data on MATH was taken from the MATH Papers With Code leaderboard in February 2025 and the o3-mini model launch. To learn more about MATH, please read the original paper.
16. **MixEval**: Data on MixEval was taken from the MixEval leaderboard in February 2025. To learn more about MixEval, please read the original paper.
17. **MMLU**: Data on MMLU was taken from the MMLU Papers With Code leaderboard in February 2025. To learn more about MMLU, please read the original paper.
18. **MMLU-Pro**: Data on MMLU-Pro was taken from the MMLU-Pro leaderboard in February 2025. To learn more about MMLU-Pro, please read the original paper.
19. **MMMU**: Data on MMMU was taken from the MMMU leaderboard in February 2025. To learn more about MMMU, please read the original paper.
20. **MTEB**: Data on Massive Text Embedding Benchmark (MTEB) was taken from the MTEB leaderboard in February 2025. To learn more about MTEB, please read the original paper.
21. **MVBench**: Data on MVBench was taken from the MVBench leaderboard in February 2025. To learn more about MVBench, please read the original paper.
22. **PlanBench**: Data on PlanBench was taken from the PlanBench paper in February 2025. To learn more about PlanBench, please read the original paper.
23. **RE-Bench**: Data on RE-Bench was taken from the RE-Bench paper in February 2025. To learn more about RE-Bench, please read the original paper
24. **RLBench**: Data on RLBench was taken from the RLBench Papers With Code leaderboard in February 2025. To learn more about RLBench, please read the original paper.
25. **Ruler**: Data on Ruler was taken from the Ruler repository in February 2025. To learn more about Ruler, please read the original paper.
26. **SWE-bench**: Data on SWE-bench was taken from the SWE-bench leaderboard in February 2025. To learn more about SWE-bench, please read the original paper.
27. **VAB**: Data on VisualAgentBench (VAB) was taken from the VAB leaderboard in February 2025. To learn more about VAB, please read the original paper.
28. **VCR**: Data on VCR was taken from the VCR leaderboard in February 2025. To learn more about VCR, please read the original paper.
29. **WildBench**: Data on WildBench was taken from the WildBench leaderboard in February 2025. To learn more about WildBench, please read the original paper.

# Chapter 3: Responsible AI

## Acknowledgments

The AI Index would like to acknowledge Andrew Shi for his work spearheading the analysis of responsible AI (RAI)–related conference submissions. The AI Index acknowledges that the Global State of Responsible AI analysis was conducted in collaboration with Accenture. It specifically highlights the contributions of Accenture's Chief Responsible AI Officer, Arnab Chakraborty, and the Accenture Research team, including Patrick Connolly, Jakub Wiatrak, Dikshita Venkatesh, and Shekhar Tewari, to the data collection and analysis. The AI Index acknowledges the McKinsey team—specifically, Medha Bankhwal, Emily Capstick, Katherine Ottenbreit, Brittany Presten, Roger Roberts, and Cayla Volandes—for their collaboration on the survey of the responsible AI ecosystem.

## Conference Submissions Analysis

For the analysis on responsible AI-related conference submissions, the AI Index examined the number of responsible AI–related academic submissions at the following conferences: AAAI, AIES, FAccT, ICML, ICLR, and NeurIPS. Specifically, the team scraped the conference websites or repositories of conference submissions for papers containing relevant keywords indicating they could fall into a particular responsible AI category. The papers were then manually verified by a human team to confirm their categorization. It is possible that a single paper could belong to multiple responsible AI categories.

The keywords searched include:

**Fairness and bias:** algorithmic fairness, bias detection, bias mitigation, discrimination, equity in AI, ethical algorithm design, fair data practices, fair ML, fairness and bias, group fairness, individual fairness, justice, nondiscrimination, representational fairness, unfair, unfairness.

**Privacy and data governance:** anonymity, confidentiality, data breach, data ethics, data governance, data integrity, data privacy, data protection, data transparency, differential privacy, inference privacy, machine unlearning, privacy by design, privacy-preserving, secure data storage, trustworthy data curation.

**Security:** adversarial attack, adversarial learning, AI incident, attacks, audits, cybersecurity, ethical hacking, forensic analysis, fraud detection, red teaming, safety, security, security ethics, threat detection, vulnerability assessment.

**Transparency and explainability:** algorithmic transparency, audit, auditing, causal reasoning, causality, explainability, explainable AI, explainable models, human-understandable decisions, interpretability, interpretable models, model explainability, outcome explanation, transparency, xAI.

## Accenture Global State of Responsible AI Survey

Researchers from Stanford conducted the second iteration of the Global State of Responsible AI survey in collaboration with Accenture. Responses from 1,500 organizations, each with total revenues of at least $500 million, were collected from 20 countries and 19 industries. The survey was conducted in January–February 2025. The objective of the Global State of Responsible AI survey was to understand the challenges of adopting RAI principles and practices and to allow for a comparison of organizational and operational RAI activities across 10 dimensions over time.

The survey covers a total of 10 RAI dimensions: reliability; privacy and data governance; fairness and nondiscrimination; transparency and explainability; human interaction; societal and environmental well-being; accountability; leadership/principles/culture; lawfulness and compliance; and organizational governance. Details about the methodology can be found here.

## McKinsey Responsible AI Survey

A recent survey by McKinsey & Company of more than 750 leaders across 38 countries provides insights into the current state of RAI in enterprises. These leaders represent various industries, from technology to healthcare, and include professionals from legal, data/AI, engineering, risk, and finance roles. Leaders were asked about their organization's experience with RAI and assessed using the McKinsey RAI Maturity Model, a responsible AI framework that encompasses four dimensions of RAI—strategy, risk management, data and technology, and operating model—with 21 subdimensions. RAI maturity was ranked across four levels, ranging from the development of foundational RAI practices to having a comprehensive and proactive program in place.

# Chapter 4: Economy

## International Federation of Robotics (IFR)

Data presented in the Robot Installations section was sourced from the World Robotics 2024 report.

## Lightcast

*Prepared by Vishy Kamalapuram and Elena Magrini*

Lightcast delivers job market analytics that empower employers, workers, and educators to make data-driven decisions. The company's artificial intelligence technology analyzes hundreds of millions of job postings and real-life career transitions to provide insight into labor market patterns. This real-time strategic intelligence offers crucial insights, such as what jobs are most in demand, the specific skills employers need, and the career directions that offer the highest potential for workers. For more information, visit https://lightcast.io.

### Job Postings Data

To support these analyses, Lightcast mined its dataset of millions of job postings collected since 2010. Lightcast collects postings from over 51,000 online job sites to develop a comprehensive, real-time portrait of labor market demand. It aggregates job postings, removes duplicates, and extracts data from job postings text. This includes information on job title, employer, industry, and region, as well as required experience, education, and skills.

Job postings are useful for understanding trends in the labor market because they allow for a detailed, real-time look at the skills employers seek. To assess the representativeness of job postings data, Lightcast conducts a number of analyses to compare the distribution of job postings to the distribution of official government and other third-party sources in the United States. The primary source of government data on U.S. job postings is the Job Openings and Labor Turnover Survey (JOLTS) program, conducted by the Bureau of Labor Statistics. Based on comparisons between JOLTS and Lightcast, the labor market demand captured by Lightcast data represents over 99% of the total labor demand. Jobs not posted online are usually in small businesses (e.g., "Help Wanted" signs in restaurant windows) and union hiring halls.

### Measuring Demand for AI

To measure the demand by employers of AI skills, Lightcast uses its skills taxonomy of over 33,000 skills.[1] These skills are organized hierarchically in over 400 skills clusters and 32 skills categories. The list of AI skills from Lightcast are shown below, with associated skills clusters. For the purposes of this report, all skills below were considered AI skills. A posting was considered an AI job if it mentioned any of these skills in the text of the listing.

**AI ethics, governance, and regulation:** ethical AI, data sovereignty, AI security, artificial intelligence risk.

**Artificial intelligence:** agentic systems, AI/ML inference, AIOps (artificial intelligence for IT operations), AI personalization, AI testing, applications of artificial intelligence, artificial general intelligence, artificial intelligence, artificial intelligence development, Artificial Intelligence Markup Language (AIML), artificial intelligence systems, automated data cleaning, Azure Cognitive Services, Baidu, cognitive automation, cognitive computing, computational intelligence, Cortana, Data Version Control (DVC), Edge Intelligence, embedded AI, expert systems, explainable AI (XAI), intelligent control, intelligent systems, interactive kiosk, IPSoft Amelia, knowledge distillation, knowledge engineering, knowledge-based configuration, knowledge-based systems, knowledge representation, multi-agent systems, neuro-symbolic AI,

1 https://lightcast.io/open-skills

Open Neural Network Exchange (ONNX), OpenAI Gym, operationalizing AI, PineCone, Qdrant, reasoning systems, swarm intelligence, synthetic data generation, Watson Conversation, Watson Studio, Weka Weaviate.

**Autonomous driving:** advanced driver-assistance systems, autonomous cruise control systems, autonomous system, autonomous vehicles, dynamic routing, guidance navigation and control systems, light detection and ranging (LiDAR), object tracking, OpenCV, path analysis, path finding, remote sensing, scene understanding, unmanned aerial systems (UAS).

**Generative AI:** Adobe Sensei, ChatGPT, CrewAI, DALL-E image generator, generative adversarial networks, generative AI agents, generative artificial intelligence,Google Bard, image inpainting, image super-resolution, LangGraph, large language modeling, Microsoft Copilot, multimodal learning, multimodal models, prompt engineering, retrieval-augmented generation, Stable Diffusion, text summarization, text to speech (TTS), variational autoencoders (VAEs).

**Machine learning:** AdaBoost (adaptive boosting), adversarial machine learning, Apache MADlib, Apache Mahout, Apache SINGA, Apache Spark, association rule learning, attention mechanisms, AutoGen, automated machine learning, autonomic computing, AWS SageMaker, Azure Machine Learning, bagging techniques, Bayesian belief networks, Boltzmann Machine, boosting, Chi-Squared Automatic Interaction Detection (CHAID), Classification and Regression Tree (CART), cluster analysis, collaborative filtering, concept drift detection, confusion matrix, cyber-physical systems, Dask (Software), data classification, Dbscan, decision models, decision-tree learning, dimensionality reduction, distributed machine learning, Dlib (C++ library), embedded intelligence, ensemble methods, evolutionary programming, expectation maximization algorithm, feature engineering, feature extraction, feature learning, feature selection, federated learning, game AI, Gaussian process, genetic algorithm, Google AutoML, Google Cloud ML Engine, gradient boosting, gradient boosting machines (GBM), H2O.ai, hidden Markov model, hyperparameter optimization, incremental learning, inference engine, k-means clustering, kernel methods, Kubeflow, LIBSVM, loss functions, machine learning, machine learning algorithms, machine learning methods, machine learning model monitoring and evaluation, machine learning model training, Markov chain, matrix factorization, meta learning, Microsoft Cognitive Toolkit (CNTK), MLflow, MLOps (machine learning operations), mlpack (C++ library), ModelOps, Naive Bayes Classifier, neural architecture compression, neural architecture search (NAS), objective function, Oracle Autonomous Database, Perceptron, Predictionio, predictive modeling, programmatic media buying, Pydata, PyTorch (machine learning library), PyTorch Lightning, Random Forest Algorithm, recommender systems, reinforcement learning, Scikit-Learn (Python package), semi-uupervised learning, soft computing, sorting algorithm, supervised learning, support vector machines (SVM), t-SNE (t-distributed Stochastic Neighbor Embedding), test datasets, topological data analysis (TDA), Torch (machine learning), training datasets, transfer learning, transformer (machine learning model), unsupervised learning, Vowpal Wabbit, Xgboost, Theano (software).

**Natural language processing:** AI copywriting, Amazon Alexa, Amazon Textract, ANTLR, Apache OpenNLP, BERT (NLP Model), chatbot, computational linguistics, conversational AI, DeepSpeech, dialog systems, fastText, fuzzy logic, handwriting recognition, Hugging Face (NLP framework), Hugging Face Transformers, intelligent agent, intelligent virtual assistant, Kaldi, language model, latent Dirichlet allocation, Lexalytics, machine translation, Microsoft LUIS, natural language generation (NLG), natural language processing (NLP), natural language programming, natural language toolkits, natural language understanding (NLU), natural language user interface, nearest neighbour algorithm, Nuance Mix, optical character recognition (OCR), screen reader, semantic analysis, semantic interpretation for speech recognition, semantic kernel, semantic parsing, semantic search, sentence transformers, sentiment analysis, Seq2Seq, Shogun, small language model, speech recognition, speech recognition software, speech synthesis, statistical language acquisition, summarization methods, text mining, text retrieval systems, text to speech (TTS), tokenization, Vespa, voice assistant technology, voice interaction, voice user interface, word embedding, Word2Vec models.

**Neural networks:** Apache MXNet, artificial neural networks, autoencoders, Caffe (framework), Caffe2, Chainer (Deep Learning Framework), convolutional neural networks (CNN), Cudnn, deep learning, deep learning methods, Deeplearning4j, deep reinforcement learning (DRL), evolutionary acquisition of neural topologies, Fast.AI, graph neural networks (GNNs), Keras (neural network library), Long Short-Term Memory (LSTM), neural ordinary differential equations, OpenVINO, PaddlePaddle, Pybrain, recurrent neural network (RNN), reinforcement learning (RL), residual networks (ResNet), sequence-to-sequence models (seq2seq), spiking neural networks, TensorFlow.

**Robotics:** advanced robotics, bot framework, cognitive robotics, meta-reinforcement learning, motion planning, Nvidia Jetson, OpenAI Gym environments, reinforcement learning from human feedback (RLHF), robot framework, robot operating systems, robotic automation software, robotic liquid handling systems, robotic programming, robotic systems, servomotor, SLAM algorithms (Simultaneous Localization and Mapping).

**Visual image recognition:** 3D reconstruction, activity recognition, computer vision, contextual image classification, Deck.gl, digital image processing, digital twin technology, eye tracking, face detection, facial recognition, general-purpose computing on graphics processing units, gesture recognition, image analysis, image captioning, image matching, image recognition, image segmentation, image sensor, ImageNet, instance segmentation, machine vision, MNIST, motion analysis, object recognition, OmniPage, pose estimation, RealSense, thermal imaging analysis.

# LinkedIn

*Prepared by Rosie Hood, Akash Kaura, and Mar Carpanelli*

## LinkedIn Data

This body of work represents the world seen through LinkedIn data, drawn from the anonymized and aggregated profile information of LinkedIn's more than 1 billion members around the world. As such, it is influenced by how members choose to use the platform, which can vary based on professional, social, and regional culture, as well as overall site availability and accessibility. In publishing insights from LinkedIn's Economic Graph, LinkedIn aims to provide accurate statistics while ensuring the privacy of LinkedIn's members. As a result, all data shows aggregated information for the corresponding period following strict data quality thresholds that prevent disclosing any information about specific individuals.

## Country Sample

LinkedIn provides data on Argentina, Australia, Austria, Belgium, Brazil, Canada, Chile, Costa Rica, Croatia, Cyprus, Czechia, Denmark, Estonia, Finland, France, Germany, Greece, Hong Kong SAR, Hungary, Iceland, India, Indonesia, Ireland, Israel, Italy, Latvia, Lithuania, Luxembourg, Mexico, Netherlands, New Zealand, Norway, Poland, Portugal, Romania, Saudi Arabia, Singapore, Slovenia, South Africa, South Korea, Spain, Sweden, Switzerland, Turkey, United Arab Emirates, United Kingdom, United States, and Uruguay.

## Skills

LinkedIn members self-report their skills on their LinkedIn profiles. Currently, more than 41,000 distinct, standardized skills are identified by LinkedIn.

LinkedIn categorizes AI skills into two mutually exclusive groups: "AI Engineering" and "AI Literacy." Broadly speaking, AI Engineering skills refer to the technical expertise and practical competencies required to design, develop, deploy, and maintain artificial intelligence systems, and AI Literacy skills refer to the knowledge, abilities, and critical thinking competencies needed to understand, evaluate, and effectively interact with artificial intelligence technologies.

As skills are ever evolving, we maintain and refresh these classifications on a periodic basis. For a list of skills included in this analysis, please see LinkedIn's AI skills List below.

### Industry

LinkedIn's industry taxonomy is a collection of entities that share economic activities and contribute to a specific product or service. An industry represents the products or services that a company offers or sells. LinkedIn analyzes the following industries in the context of AI: education; financial services; manufacturing; professional services; and technology, information, and media.

### Gender

LinkedIn recognizes that some LinkedIn members identify beyond the traditional gender constructs of "man" and "woman." If not explicitly self-identified, LinkedIn has inferred the gender of members included in this analysis either by the pronouns used on their LinkedIn profiles or on the basis of first names. Members whose gender could not be inferred as either male or female were excluded from any gender analysis. Please note that LinkedIn filtered out countries where their gender attribution algorithm did not have sufficient coverage.

### AI Jobs or Occupations

LinkedIn member titles are standardized and grouped into over 16,000 occupations. These are not sector or country specific. An AI job requires AI skills to perform the job. Examples of such occupations include (but are not limited to): machine learning engineer, artificial intelligence specialist, data scientist, and computer vision engineer.

### AI Talent

A LinkedIn member is considered AI talent if they have explicitly added at least two AI skills to their profile and/or they are or have been employed in an AI job.

## METHODOLOGIES

### 1. Top AI Skills

These are the AI skills most frequently added by LinkedIn members from 2015 onward.

**Interpretation:** The most added AI Engineering skills globally are machine learning, AI, and deep learning.

### 2. Fastest Growing AI Skills

The year-over-year growth rate for AI skills most frequently added by all members. Please note that LinkedIn implements thresholds to skill add volumes in the most recent year, which are set at the 50th percentile of the most recent year's AI skill adds distribution by country.

**Interpretation:** The fastest growing AI Engineering skills globally are custom GPTs, AI productivity, and AI agents.

### 3. AI Talent Concentration

The counts of AI talent are used to calculate talent concentration metric. In other words, to calculate the country-level AI talent concentration, LinkedIn uses the counts of AI talent in a particular country divided by the counts of LinkedIn members in that country. Note that concentration metrics may be influenced by LinkedIn coverage in these countries and should be utilized with caution.

**Interpretation:** AI talent with AI Engineering skills represents 0.78% of LinkedIn members in the United States.

### 4. Relative AI Talent Hiring Rate YoY Ratio

The LinkedIn hiring rate is a measure of hires normalized by LinkedIn membership. It is computed as the percentage of LinkedIn members who added a new employer in the same period the job began, divided by the total number of LinkedIn members in the corresponding location.

The AI hiring rate is computed using the overall hiring rate methodology, but it only considers members classified as AI

talent. The relative AI talent hiring rate YoY ratio is the year-over-year change in the AI hiring rate relative to the overall hiring rate in the same country. LinkedIn shares a 12-month moving average.

**Interpretation:** In the United States, the ratio of AI talent hiring relative to overall hiring has grown 24.7% year over year.

## 5. Skill Penetration

### SKILLS GENOME

For any category (occupation, country, industry, etc.), the skills genome is an ordered list (a vector) of the 50 skills most characteristic of that category. These most characteristic skills are determined using a TF-IDF algorithm, which down-ranks ubiquitous skills that add little information about that specific entity (e.g., Microsoft Word) and up-ranks skills unique to that specific entity (e.g., artificial intelligence). Further details are available at LinkedIn's skills genome and the LinkedIn–World Bank Methodology note.

As an example, Table 1 details the skills genome of the technology, information, and media industry in the United States in 2024, displaying the top 10 skills ranked by TF-IDF.

| Skill name | TF-IDF skill rank |
|---|---|
| Amazon Web Services (AWS) | 1 |
| Software as a Service (SaaS) | 2 |
| Artificial intelligence (AI) | 3 |
| Python (programming language) | 4 |
| Go-to-market strategy | 5 |
| Customer success | 6 |
| Large language models (LLM) | 7 |
| Salesforce.com | 8 |
| SQL | 9 |
| Generative AI | 10 |

### AI SKILLS PENETRATION

The aim of this indicator is to measure the intensity of AI skills in a given category using the following methodology:

- LinkedIn computes frequencies for all self-added skills by LinkedIn members in a given entity (occupation, industry, etc.) from 2015 onward.
- LinkedIn reweights skill frequencies using a TF-IDF model to get the top 50 most representative skills in that entity. These 50 skills compose the "skill genome" of that entity.
- LinkedIn computes the share of skills that belong to the AI skill group out of the top skills in the selected entity.

**Interpretation:** The AI skills penetration rate signals the prevalence of AI skills across occupations, or the intensity with which LinkedIn members utilize AI skills in their jobs. For example, the top 50 skills for the occupation of engineer are calculated based on the weighted frequency with which they appear in LinkedIn members' profiles. If four of the skills that engineers possess belong to the AI skills group, this measure indicates that the penetration of AI skills is estimated to be 8% among engineers (i.e., 4/50).

### RELATIVE AI SKILLS PENETRATION

To allow for skills penetration comparisons across countries, the skills genomes are calculated, and a relevant benchmark is selected (e.g., a global average). A ratio is then constructed between a country and the benchmark's AI skills penetrations, controlling for occupations.

**Interpretation:** If a country has a relative AI skills penetration of 1.5, that means AI skills are 1.5 times as frequent as in the benchmark, for an overlapping set of occupations.

### GLOBAL COMPARISON

For cross-country comparisons, LinkedIn presents the relative penetration rate of AI skills, measured as the sum of the penetration of each AI skill across occupations in a given country, divided by the average global penetration of AI skills across the overlapping occupations in a sample of countries.

**Interpretation:** A relative penetration rate of 2 means the average penetration of AI skills in that country is two times the global average across the same set of occupations.

#### GLOBAL COMPARISON: BY INDUSTRY

The relative AI skills penetration by country for a given industry provides an in-depth sectoral decomposition of AI skills penetration across industries and countries.

**Interpretation:** A country's relative AI skill penetration rate of 2 in the education sector means the average penetration of AI skills in that country is two times the global average across the same set of occupations in that sector.

#### GLOBAL COMPARISON: BY GENDER

The relative AI skills penetration by gender provides a cross-country comparison of AI skills penetrations within a gender. Since the global averages are distinct for each gender, this metric should only be used to compare country rankings within each gender, not for cross-gender comparisons within countries.

**Interpretation:** A country's AI skills penetration for women of 1.5 means that female members in that country are 1.5 times more likely to list AI skills than the average female member in all countries pooled together across the same set of occupations that exist in the country-gender combination.

#### GLOBAL COMPARISON: ACROSS GENDERS

The relative AI skills penetration across genders allows for cross-gender comparisons within and across countries globally, since LinkedIn compares a country's AI skills penetration by gender to the same global average regardless of gender.

### 6. Female Representation in AI

This refers to the share of AI talent occupied by women.

**Interpretation:** Female representation within AI talent with AI Engineering skills is 30.5% globally.

### 7. AI Talent Migration

Data on migration comes from the World Bank Group–LinkedIn "Digital Data for Development" partnership (see https://linkedindata.worldbank.org/ and Zhu et al. (2018)). LinkedIn migration rates are derived from the self-identified locations of LinkedIn member profiles. For example, when a LinkedIn member updates their location from Paris to London, this is counted as a migration. Migration data is available from 2019 onward.

LinkedIn data provides insights to countries on AI talent gained or lost due to migration trends. AI talent migration is considered for all members with AI skills/holding AI jobs at time "t" for country A as the country of interest and country B as the source of inflows and destination for outflows. Thus, net AI talent migration between country A and country B is calculated as:

$$Net\ AI\ Talent\ Migration_{a,b,t} = \frac{Net\ AI\ Talent\ flows_{a,b,t}}{Member\ count_{a,t}}$$

Net flows are defined as total arrivals minus departures within a given time period. LinkedIn membership varies between countries, which can prove challenging when interpreting absolute movements of members from one country to another. Migration flows are therefore normalized with respect to each country. For example, for country A, all absolute net flows into and out of country A, regardless of origin and destination countries, are normalized based on the LinkedIn membership of country A at the end of each year and multiplied by 10,000. Hence, this metric indicates relative talent migration from all countries to and from country A. Please note that minimum thresholds have been applied such that transitions have a sufficient sample size.

**Interpretation:** The United States had a positive net flow of AI talent relative to its membership size at 1.07 net flow per 10,000 members.

### 8. Career Transitions Into AI Jobs

LinkedIn considers the source occupations that feed AI occupations, analyzing the share of transitions into AI occupations pooled over a five-year period. Career transitions are computed by aggregating member-level job transitions from one occupation to another occupation the member has not previously held. LinkedIn excludes first occupations added by new graduates and intra-occupation transitions.

**Interpretation:** In the United States, 26.9% of transitions into AI engineer came from software engineer, followed by 13.3% from data scientist.

## THE LINKEDIN AI SKILLS LIST

### AI Engineering

3D reconstruction, AI agents, AI productivity, AI strategy, algorithm analysis, algorithm development, Amazon Bedrock, Apache Spark ML, applied machine learning, artificial intelligence (AI), artificial neural networks, association rules, audio synthesis, autoencoders, automated clustering, automated feature engineering, automated machine learning (AutoML), automated reasoning, autoregressive models, Azure AI Studio, Caffe, chatbot development, chatbots, classification, cognitive computing, computational geometry, computational intelligence, computational linguistics, concept drift adaptation, conditional generation, conditional image generation, convolutional neural networks (CNN), custom GPTs, decision trees, deep convolutional generative adversarial networks (DCGAN), deep convolutional neural nNetworks (DCNN), deep learning, deep neural networks (DNN), evolutionary algorithms, expert systems, facial recognition, feature extraction, feature selection, fuzzy logic, generative adversarial imitation learning, generative adversarial networks (GANs), generative AI, generative design optimization, generative flow models, generative modeling, generative neural networks, generative optimization, generative pre-training, generative query networks (GQNs), generative replay memory, generative synthesis, gesture recognition, Google Cloud AutoML, graph embeddings, graph networks, hyperparameter optimization, hyperparameter tuning, image generation, image inpainting, image processing, image synthesis, image-to-image translation, information extraction, intelligent agents, k-means clustering, Keras, knowledge discovery, knowledge representation and reasoning, LangChain, large language model operations (LLMOps), large language models (LLM), machine learning, machine learning algorithms, machine translation, Microsoft Azure Machine Learning, MLOps, model compression, model interpretation, model training, music generation,nNatural language generation, natural language processing (NLP), natural language understanding, neural network architecture design, neural networks, NLTK, object recognition, ontologies, OpenAI API, OpenCV, parsing, pattern recognition, predictive modeling, probabilistic generative models, probabilistic programming, prompt flow, PyTorch, question answering, random forest, RapidMiner, recommender systems, recurrent neural networks (RNN), reinforcement learning, responsible AI, Scikit-Learn, semantic technologies, semantic web, sentiment analysis, speech recognition, Spring AI, statistical inference, style transfer, StyleGAN, supervised learning, support vector machine (SVM), synthetic data generation, TensorFlow, text analytics, text classification, text generation, text mining, text-to-image generation, Theano, time series forecasting, transformer models, unsupervised learning, variational autoencoders (VAEs), video generation, web mining, Weka, WordNet.

### AI Literacy

AI Builder, AI prompting, Anthropic Claude, ChatGPT, DALL-E, generative AI, Generative AI Studio, generative AI tools, generative art, GitHub Copilot, Google Bard, Google Gemini, GPT-3, GPT-4, LLaMA, Microsoft Copilot, Microsoft Copilot Studio, Midjourney, multimodal prompting, prompt engineering, Stable Diffusion.

### Acknowledgments

LinkedIn gratefully acknowledges the contributions of Murat Erer and Carl Shan in developing these methodologies and metrics, and the feedback from our collaborators at the OECD.AI, Stanford Institute for Human-Centered AI, World Bank, and Centro Nacional de Inteligencia Artificial (Cenia).

# Quid

*Quid insights prepared by Heather English and Hansen Yang*

Quid uses its own in-house LLM and other smart search features, as well as traditional Boolean query, to search for focus areas, topics, and keywords within many datasets: social media, news, forums and blogs, companies, patents, as well as other custom feeds of data (e.g., survey data). Quid has many visualization options and data delivery endpoints, including network graphs based on semantic similarity, in-platform dashboarding capabilities, and programmatic PostgreSQL database delivery. Quid applies best-in-class AI and NLP to reveal hidden patterns in large datasets, enabling users to make data-driven decisions accurately, quickly, and efficiently.

### Search, Data Sources, and Scope

Over 8 million global public and private company profiles from multiple data sources are indexed to search across company descriptions, while filtering and including metadata ranging from investment information to firmographic information, such as founding year, headquarter location, and more. Company information is updated on a weekly basis. The Quid algorithm reads a large amount of text data from each document to make links between different documents based on their similar language. This process is repeated at an immense scale, which produces a network of different clusters identifying distinct topics or focus areas. Trends are identified based on keywords, phrases, people, companies, and institutions that Quid identifies and other metadata that is put into the software.

### Data

### Companies

Organization data is embedded from Capital IQ and Crunchbase. These companies include every type of organization (private, public, operating, operating as a subsidiary, out of business) throughout the world. The investment data includes private investments, M&A, public offerings, minority stakes held by PE/ VCs, corporate venture arms, governments, and institutions both within and outside the United States. Some data is unavailable—for instance, when investors' names or funding amounts are not disclosed. Quid embeds Capital IQ data as a default and adds in data from Crunchbase for the data points that are not captured in Capital IQ. This not only yields comprehensive and accurate data on all global organizations, but it also captures early-stage startups and funding events data.

### Search Parameters

Boolean query is used to search for focus areas, topics, and keywords within the archived company database and within their business descriptions and websites. Quid can filter out the search results by HQ regions, investment amount, operating status, organization type (private/ public), and founding year. Quid then visualizes these companies by semantic similarity. If there are more than 7,000 companies from the search result, Quid selects the 7,000 most relevant companies for visualization based on the language algorithm. Boolean search: "artificial intelligence" or "AI" or "machine learning" or "deep learning"

### Companies

- Global AI and ML companies that have received investments (private, IPO, M&A) from Jan. 1, 2014, to Dec. 31, 2024.
- Global AI and ML companies that have received over $1.5 million for the past 10 years (Jan. 1, 2014, to Dec. 31, 2024).
- Global data was also pulled for a generative AI query (Boolean search: "generative AI" or "gen AI" OR "generative artificial intelligence") for companies that have received over $1.5 million for the past 10 years (Jan. 1, 2014, to Dec. 31, 2024).

### Target Event Definitions

- **Private investment:** A private placement is a private sale of newly issued securities (equity or debt) by a company to a select investor or group of investors. The stakes that buyers take in private placements are often minority stakes (under 50%), although it is possible to take control of a company through a private placement as well, in which case the private placement would be a majority stake investment.

- **Minority investment:** These refer to minority stake acquisitions in Quid, which take place when the buyer acquires less than 50% of the existing ownership stake in entities, asset products, and business divisions.
- **M&A:** This refers to a buyer acquiring more than 50% of the existing ownership stake in entities, asset products, and business divisions.

## McKinsey & Company

Data used in the "Corporate Activity" section was sourced from two McKinsey global surveys: "The State of AI in Early 2024: Gen AI Adoption Spikes and Starts to Generate Value" and "The State of AI: How Organizations Are Rewiring to Capture Value."

The first online survey of 2024 was in the field from Feb. 22 to March 5, and garnered responses from 1,363 participants representing the full range of regions, industries, company sizes, functional specialties, and tenures. Among the respondents, 981 said their organizations had adopted AI in at least one business function, and 878 said their organizations were regularly using gen AI in at least one function.

The second online survey of 2024 was in the field from July 16 to July 31, and garnered responses from 1,491 participants in 101 nations representing the full range of regions, industries, company sizes, functional specialties, and tenures. Forty-two percent of respondents said they work for organizations with more than $500 million in annual revenues.

To adjust for differences in response rates, the data is weighted by the contribution of each respondent's nation to global GDP.

The AI Index also considered data from previous iterations of the McKinsey survey. These include:

# Chapter 5: Science and Medicine

## Acknowledgments

The AI Index would like to acknowledge Armin Hamrah for his work in surveying the literature on significant trends in AI-related science and medicine.

## Benchmarks

1. **MedQA:** Data on MedQA was taken from the MedQA Papers With Code leaderboard in February 2025. To learn more about MedQA, please read the original paper.

## AI-Driven Protein Science Publications

The AI Index used Dimensions' AI document search function to measure the number of manuscripts published in a year. The searches were restricted to the 2024 publication year and the biological sciences category (987,717 publications). Then a search was conducted for each key term, which had to be present in both the title and the abstract. This requirement limited the number of manuscripts returned that might only have mentioned the key term in passing, rather than describing research about the key term. Once the number of manuscripts was identified, the percent of total biological sciences manuscripts about each key term was calculated.

## Image and Multimodal AI for Scientific Discovery

The AI Index used Semantic Scholar and Google Scholar to measure the number of manuscripts published from 2023 to 2025. A search was then performed for each key term (e.g., "foundation models," "microscopy," "electron microscopy," "fluorescence microscopy," "light microscopy") with the requirement that the terms be present in both the title and the abstract. Furthermore, the search was refined to strictly comply with the definition of a foundation model—specifically, a model trained on vast datasets that can be applied across a wide range of use cases. To this end, any model alleged to be a foundation model that had been trained on fewer than 1 million data points or not evaluated on multiple tasks was discarded.

## FDA-Approved AI Medical Devices

Data on FDA-approved AI medical devices was sourced from the FDA website, which tracks artificial intelligence and machine learning (AI/ML)–enabled medical devices.

## Ethical Considerations

The AI Index used PubMedCentral's API to query for English-language indexed articles published between Jan. 1, 2020, and Dec. 31, 2024, using search terms regarding artificial intelligence, medicine, and ethical issues. In order to obtain only articles at the intersection of those three topics, the AI Index further narrowed the articles to those with an abstract including a keyword related to: (a) artificial intelligence, (b) medicine, and (c) at least one ethical issue. After removing preprints, retracted articles, and articles that failed to satisfy the inclusion criteria, 2,916 articles remained. The AI Index used the frequency of ethical issues mentioned in abstracts across this pool of articles to conduct its analysis.

API query:
("artificial intelligence"[MeSH] OR "machine learning"[MeSH] OR "deep learning"[All Fields] OR "AI"[All Fields] OR "ML"[All Fields] OR "predictive analytics"[All Fields]) AND (("ethics"[MeSH] OR "ethical implications"[All Fields] OR "fair*"[All Fields] OR "unfair*"[All Fields] OR "bias"[All Fields] OR "accountability"[All Fields] OR "transparency"[All Fields] OR "explainability"[All Fields] OR "privacy"[All Fields] OR "trustworthy AI"[All Fields]) OR ("bioethics"[MeSH] OR "ELSI"[All Fields] OR "autonomy"[All Fields] OR "equity"[All Fields] OR "equitab*"[All Fields] OR "justice"[All Fields] OR "beneficence"[All Fields] OR "non-maleficence"[All Fields] OR "independent review"[All Fields] OR "oversight"[All Fields] OR "racis*"[All Fields] OR "prejud*"[All Fields] OR "inequit*"[All Fields] OR "community engagement"[All Fields] OR "misuse"[All Fields] OR "dual use"[All Fields])) AND ("medicine"[MeSH] OR "medical AI"[All Fields] OR "clinical decision support"[All Fields] OR "health informatics"[All Fields]) AND ("2020/01/01"[PubDate] : "2024/12/31"[PubDate])

Date of search: 2/14/2025

Abstract inclusion criteria:
Therefore, includes only articles that discuss medicine, artificial intelligence, and at least one ethical issue within the abstract (N = 2,916).

- **AI keywords:** "artificial intelligence," " AI," "algorithm," "ML," "machine learning," "deep learning," predictive analytics.
- **Medicine keywords:** "medicine," "medical," "health," "healthcare."
- **Ethics keywords:** "ethic*," "fairness," "bias," "accountability," "transparency," "explainability," "privacy," "trustworthy AI," "bioethics," "ELSI," "autonomy," "equit*," "justice," "beneficence," "non-maleficence," "independent review," "oversight," "racism," "inequit*," community engagement, misuse, dual use.

# Chapter 6: Policy and Governance

## Acknowledgments

The AI Index would like to acknowledge Julia Betts Lotufo and Alexandra Rome for their efforts in collecting information on significant AI policy events. The AI Index would also like to acknowledge Lapo Santarlasci for leading the analysis of AI public spending and U.S. grant-related AI spending.

## Global AI Mentions

For mentions of AI in AI-related legislative proceedings around the world, the AI Index performed searches for the keyword "artificial intelligence," in respective languages, on the websites of congresses or parliaments in 75 geographic areas, usually under sections named "minutes," "hansard," etc. Mentions were counted by session, so multiple mentions of "artificial intelligence" in the same legislative session counted as one mention. The AI Index team surveyed the following databases:

Andorra, Armenia, Australia, Azerbaijan, Barbados, Belgium, Bermuda, Brazil, Canada, Cayman Islands, China,[1] Czech Republic, Denmark, Dominican Republic, Ecuador, El Salvador, Estonia, Fiji, Finland, France, Germany, Gibraltar, Greece, Hong Kong, Iceland, India, Ireland, Isle of Man, Italy, Japan, Kenya, Kosovo, Latvia, Lesotho, Liechtenstein, Luxembourg, Macao SAR, China, Madagascar, Malaysia, Maldives, Malta, Mauritius, Mexico, Moldova, Netherlands, New Zealand, Northern Mariana Islands, Norway, Pakistan, Panama, Papua New Guinea, Philippines, Poland, Portugal, Romania, Russia, San Marino, Seychelles, Sierra Leone, Singapore, Slovenia, South Africa, South Korea, Spain, Sri Lanka, Sweden, Switzerland, Tanzania, Trinidad and Tobago, Ukraine, United Kingdom, United States, Uruguay, Zambia, Zimbabwe

## Global Legislation Records on AI

For AI-related bills passed into laws, the AI Index performed searches for the keyword "artificial intelligence," in respective languages and in the full text of bills, on the websites of congresses or parliaments in 116 geographic areas. Note that only laws passed by state-level legislative bodies and signed into law (e.g., by presidents or received royal assent) from 2016 to 2024 are included. Laws that were approved but then repealed are not included in the analysis. For laws where AI-related provisions were added or amended after initial enactment, the AI Index uses the year of inclusion rather than the original passage year, when relevant. Future AI Index reports hope to include analysis on other types of legal documents, such as regulations and standards, adopted by state- or supranational-level legislative bodies, government agencies, etc.

The AI Index team surveyed databases for the following geographic areas:

Algeria, Andorra, Antigua and Barbuda, Argentina, Armenia, Australia, Austria, Azerbaijan, The Bahamas, Bahrain, Bangladesh, Barbados, Belarus, Belgium, Belize, Bermuda, Bhutan, Bolivia, Brazil, Brunei, Bulgaria, Cameroon, Canada, Chile, China, Croatia, Cuba, Curacao, Cyprus, Czech Republic, Denmark, Estonia, Faroe Islands, Fiji, Finland, France, Germany, Gibraltar, Greece, Greenland, Grenada, Guam, Guatemala, Guyana, Hong Kong, Hungary, Iceland, India, Iraq, Ireland, Isle of Man, Israel, Italy, Jamaica, Japan, Kazakhstan, Kenya, Kiribati, Republic of Korea, Kosovo, Kyrgyz Republic, Latvia, Liechtenstein, Lithuania, Luxembourg, Macao SAR China, Malawi, Malaysia, Malta, Mauritius, Mexico, Monaco, Montenegro, Morocco, Mozambique, Nauru, Netherlands, New Zealand, Northern Marina Islands, Norway, Panama, Philippines, Poland, Portugal, Romania, Russia, Samoa, Saudi Arabia, Serbia, Seychelles, Sierra Leone, Singapore,

1 The National People's Congress is held once per year and does not provide full legislative proceedings. Hence, the counts included in the analysis searched mentions of "artificial intelligence" in the only public document released from the congressional meetings, the Report on the Work of the Government, delivered by the premier.

Slovak Republic, Slovenia, South Africa, Spain, St. Kitts and Nevis, Suriname, Sweden, Switzerland, Tajikistan, Tanzania, Togo, Tongo, Turkey, Tuvalu, Uganda, Ukraine, United Arab Emirates, United Kingdom, United States, Uruguay, Vietnam, Yemen, Zambia, Zimbabwe

## US State-Level AI Legislation

For AI-related bills passed into law, the AI Index performed searches for the keyword "artificial intelligence" in the full text of bills on the websites of all 50 U.S. states. Bills are only counted as passed into law if the keyword appears in the final version of the bill, not just the introduced version. Note that only laws passed from 2015 to 2024 are included. The count for proposed laws includes both laws that were proposed that were passed and laws that were proposed that have not been passed yet, or are now inactive. The AI Index team surveyed the following databases:

Alabama, Alaska, Arizona, Arkansas, California, Colorado, Connecticut, Delaware, Florida, Georgia, Hawaii, Idaho, Illinois, Indiana, Iowa, Kansas, Kentucky, Louisiana, Maine, Maryland, Massachusetts, Michigan, Minnesota, Mississippi, Missouri, Montana, Nebraska, Nevada, New Hampshire, New Jersey, New Mexico, New York, North Carolina, North Dakota, Ohio, Oklahoma, Oregon, Pennsylvania, Rhode Island, South Carolina, South Dakota, Tennessee, Texas, Utah, Vermont, Virginia, Washington, West Virginia, Wisconsin, Wyoming

For a more thorough review, the AI Index also included AI-related state laws listed on the Multistate AI state legislation tracker, even if they did not specifically reference "artificial intelligence" as a keyword.

## US AI Regulation

This section examines AI-related regulations enacted by U.S. regulatory agencies from 2016 to 2024, analyzing the total number of regulations and their originating agencies. To compile this data, the AI Index conducted a keyword search for "artificial intelligence" on the Federal Register, a comprehensive repository of government documents drawn from over 436 agencies and nearly every branch of the U.S. government.

## US Committee Mentions

To research trends on the United States' committee mentions of AI, the following search was conducted:
Website: Congress.gov
Keyword: artificial intelligence
Filters: Committee Reports

## Public Investment in AI

The AI Index analyzed government AI spending across European countries and the United States, focusing on regions where data is more accessible. It is important to note that this analysis may not fully represent all countries or regions, as the availability and quality of data can vary significantly. Additionally, while this analysis includes data on government contracts from various countries, it only covers grant-level spending for the United States. This discrepancy is the result of challenges in collecting comparable grant data from other countries and regions, such as the European Union and China. Nevertheless, the U.S. case illustrates that a substantial portion of government spending on AI occurs through grants. Coverage will expand in future iterations of the AI Index as more data becomes available, but discrepancies and gaps in the existing data may affect the comprehensiveness and accuracy of the findings.

### Data Sources

For European countries, the AI Index collected public tender data from Tenders Electronic Daily (TED) (Publications Office of the European Union, 2024)—the online supplement to the official journal of the EU dedicated to European public procurement. While contracts are available in various formats, the most detailed data comes from bulk XML downloads, which include comprehensive information on tendering procedures, issuing entities, awarded contractors, lot values, descriptions, award dates, and common procurement vocabulary (CPV) codes. TED publication is governed by EU law thresholds: Tenders above specific monetary values, deemed of cross-border interest, must be published on TED. However, some countries also report below-threshold procurements, leading to variations in coverage across countries.

For the United Kingdom, data sources include TED, Find a Tender, Contracts Finder, and Contracts Finder Archive. Data from Scotland and Wales were accessed via the APIs of their procurement websites, while Northern Ireland does not offer this service, necessitating its exclusion from the analysis and potentially leading to an underestimation of public investments in AI for the U.K. Due to API limitations restricting historical data access, the AI Index utilized the Open Contracting Partnership's data registry via Kingfisher Collect to obtain comprehensive data for Scotland and Wales.

Data for the United States was sourced from the publicly accessible USAspending platform, an official repository that facilitates bulk downloads of information related to contract award notices and grant data. While this dataset encompasses a longer time frame than the TED dataset, it is important to note that data quality can vary. Additionally, a study by the U.S. Government Accountability Office (GAO, 2023) found that 49 agencies, including 25 in the executive branch, did not report data to USAspending, accounting for over $5 billion in net outlays for fiscal year 2022.

### Data Processing

Processing TED data posed significant challenges due to inconsistent storage of contract descriptions, which varied by XML tag names based on release time and procurement type. Some files contained aggregated descriptions while others detailed each awarded contract lot. To capture comprehensive information, the main descriptions of each competition call were combined with partial descriptions when available.

The linguistic diversity in data from different countries required translation of all texts into English using the deep-translator tool and the Google Translator engine. Post-translation, tender texts were processed using natural language processing (NLP) techniques. These included the removal of stop words and special characters, part-of-speech (POS) tagging to retain key grammatical categories, lowercase conversion, lemmatization, and replacement of numerical measures with a <NUM> tag.

For ease of comparison, all monetary amounts were converted to U.S. dollars and adjusted for price level differences using the purchasing power parities (PPP) index.

### Classification

Classifying AI-related contracts and grants was achieved using full-text search with regular expressions. An AI dictionary was compiled by generating AI-related expressions and incorporating "core" expressions from the Yamashita et al. (2021) vocabulary. Additionally, a Word2Vec model expanded the dictionary with cosine-similar terms for each baseline expression that were manually reviewed and included in the final vocabulary. This process provided keywords and co-occurrence patterns crucial for identifying AI content.

The classification followed a multistep approach. Initially, regular expression (regex) matching identified AI terms within contract and grant awards. These documents were then categorized as either "non AI-related" or "AI-related." To validate AI-related matches, BERTopic model and pretrained DeBERTA transformer were employed to assess probability scores for specific AI-related topics. Awards with relevance scores below 20% underwent manual review, while those with higher scores were confirmed as AI-related. To ensure additional accuracy, all high-value tenders were also manually reviewed.

# Chapter 7: Education

## Code.org, CSTA, ECEP Alliance

***State-Level Data***

Appendix 2 of the State of Computer Science Education 2024 report includes a full description of the methodology used by Code.org, CSTA, and ECEP Alliance to collect their data. The staff at Code.org also maintains a database of the state of American K–12 education and, in this policy primer, provides a greater amount of detail on the state of American K–12 education in each state.

***AP Computer Science Data***

The AP Computer Science data is provided to Code.org as per an agreement the College Board maintains with Code.org. The AP Computer Science data comes from the College Board's national and state summary reports.

***Access to Computer Science Education***

Data on access to computer science education was drawn from Code.org, CSTA, and ECEP Alliance's State of Computer Science Education 2024 report.

## 2024 K-12 Computer Science Landscape Teacher Landscape Survey

For more information or access to the dataset, please contact membership@csteachers.org.

## State Standards Comparison

CSTA and the Institute for Advancing Computing Education (IACE) published a State Standards Comparison report in December 2024. The dataset of approximately 10,000 state-adopted K-12 standards is available as a spreadsheet, as well as a Python notebook that may be useful for data analysis. Colorado and Virginia's standards were adopted in late 2024 and are not included in this dataset.

## Global K-12 AI Education

The Raspberry Pi Computing Education Research Centre, based in the Department of Computer Science and Technology at the University of Cambridge, compiled this dataset, expanding on research conducted by the Brookings Institution for its 2021 report Building Skills for Life: How to Expand and Improve Computer Science Education Around the World. We made one change to their dataset to clarify that CS in the United States is available in some schools/districts and not available everywhere as an elective course. For more information about the methodology, please refer to their report.

## IPEDS

The Integrated Postsecondary Education Data System (IPEDS) combines annual surveys conducted by the U.S. Department of Education's National Center for Education Statistics (NCES). IPEDS gathers information from every college, university, and technical and vocational institution that participates in federal student financial aid programs.

***Completion Data***

This chapter used data from the Completions survey, which collects data on the number of students who complete a postsecondary education program. Graduates in AI-related fields were identified as those whose first major was either Computer and Information Sciences, General (11.01); Computer Programming (11.02); or Computer Science (11.07), according to the Classification of Instructional Programs (CIP) codes. The number of graduates in AI-related fields included in this year's report differs from previous years because the AI Index used multiple CIP codes.

## OECD

This chapter used data from the OECD Data Explorer, specifically from the table "Number of enrolled students, graduates and new entrants by field of education." The methodology for this dataset can be found in Education at a Glance 2024 Sources, Methodologies and Technical Notes.

# Chapter 8: Public Opinion

## Ipsos

For the sake of brevity, the 2025 AI Index opted not to republish the methodology used by the Ipsos survey featured in the report. More details about the Ipsos survey's methodology can be found in the survey itself.